\def\isarxiv{1}

\def\paperTitle{Evaluating Frontier LLMs on PhD-Level Mathematical Reasoning: A Benchmark on a Textbook in Theoretical Computer Science about Randomized Algorithms}

\def\paperAuthor{
Yang Cao\thanks{Wyoming Seminary.}\and 
Yubin Chen\thanks{San Jose State University.}\and 
Xuyang Guo\thanks{Guilin University of Electronic Technology.}\and 
Zhao Song\thanks{University of California, Berkeley. \texttt{magic.linuxkde@gmail.com}.}\and 
Song Yue\thanks{Northeastern University.}\and 
Jiahao Zhang\and 
Jiale Zhao\thanks{Guangdong University of Technology.} 
}

\ifdefined\isarxiv
\documentclass[11pt]{article}
\usepackage[numbers]{natbib}
\else
\documentclass{article}
\usepackage{cpal_2025}
\fi

\ifdefined\isarxiv

\usepackage{amsmath}
\usepackage{amsthm}
\usepackage{amssymb}
\usepackage{algorithm}
\usepackage{subfig}
\usepackage{algpseudocode}
\usepackage{graphicx}
\usepackage{grffile}
\usepackage{wrapfig,epsfig}
\usepackage{url}
\usepackage{xcolor}
\usepackage{epstopdf}

\usepackage{tcolorbox}
\usepackage{listings} 

\usepackage{bbm}
\usepackage{dsfont}

\else 

\usepackage{url}

\fi

\ifdefined\isarxiv

\else
\usepackage{amsmath}
\usepackage{amsthm}
\usepackage{amssymb}
\usepackage{algorithm}
\usepackage{subfig}
\usepackage{algpseudocode}
\usepackage{graphicx}
\usepackage{grffile}
\usepackage{wrapfig,epsfig}
\usepackage{url}
\usepackage{xcolor}
\usepackage{epstopdf}

\usepackage{bbm}
\usepackage{dsfont}

\usepackage{tcolorbox}
\usepackage{listings} 

\usepackage{hyperref}
\fi
 
\allowdisplaybreaks

\ifdefined\isarxiv

\usepackage{tikz}
\usepackage{hyperref}  
\hypersetup{colorlinks=true,citecolor=blue,linkcolor=blue} 
\usetikzlibrary{arrows}
\usepackage[margin=1in]{geometry}
\fi
 
\graphicspath{{./figs/}}

\theoremstyle{plain}
\newtheorem{theorem}{Theorem}[section]
\newtheorem{lemma}[theorem]{Lemma}
\newtheorem{definition}[theorem]{Definition}

\newtheorem{corollary}[theorem]{Corollary}

\newtheorem{remark}[theorem]{Remark}

\newtheorem{problem}[theorem]{Problem}

\usepackage{multirow}

\newcommand{\wh}{\widehat}
\newcommand{\wt}{\widetilde}

\newcommand{\R}{\mathbb{R}}

\renewcommand{\tilde}{\wt}
\renewcommand{\hat}{\wh}

\DeclareMathOperator*{\E}{{\mathbb{E}}}

\DeclareMathOperator{\poly}{poly}

\DeclareMathOperator{\rank}{rank}

\begin{document}

\ifdefined\isarxiv

\date{}
\title{\paperTitle}
\author{\paperAuthor}

\else

\title{\paperTitle}
\author{\paperAuthor}

\fi

\ifdefined\isarxiv
\begin{titlepage}
  \maketitle
  \begin{abstract}
    The rapid advancement of large language models (LLMs) has led to significant breakthroughs in automated mathematical reasoning and scientific discovery. Georgiev, G{\'o}mez-Serrano, Tao, and Wagner~\cite{geo+25} demonstrate that AI systems can explore new constructions and improve existing bounds, illustrating the growing potential of LLMs to accelerate mathematical discovery. Similarly, Bubeck et al.~\cite{bce+25} show that GPT-5 can meaningfully contribute to scientific workflows, from proposing hypotheses to generating proofs and analyses. Despite these advances, a rigorous evaluation of these models on canonical, graduate-level mathematical theory remains necessary to understand their baseline reasoning capabilities. In this paper, we present a comprehensive benchmark of four frontier models: GPT-5-Thinking, Gemini-3-Pro, Claude-Sonnet-4.5-Thinking, and Grok-4 against the classic curriculum of Randomized Algorithms by Motwani and Raghavan~\cite{mr95}.

We tasked each model with generating formal LaTeX proofs for a series of lemmas and exercises spanning the textbook. We find that while the top-tier models (Gemini, and Claude) achieve a high accuracy rate (approx. 66\%), demonstrating a robust grasp of probabilistic method and formal logic, other models lag significantly in consistency (approx. 40\%). We provide a qualitative analysis of the generated proofs, highlighting differences in conciseness, hallucination rates, and logical structure. Our results suggest that while frontier models have reached a threshold of proficiency suitable for graduate-level pedagogical assistance and formalization, significant variance exists in their reliability for rigorous mathematical derivation.
The code and the full set of LLM-generated responses are open-sourced and publicly available at \url{https://github.com/magiclinux/math_benchmark_probability}.

  \end{abstract}
  \thispagestyle{empty}
\end{titlepage}

{\hypersetup{linkcolor=black}
}
\newpage

\else

\maketitle

\begin{abstract}
The rapid advancement of large language models (LLMs) has led to significant breakthroughs in automated mathematical reasoning and scientific discovery. Georgiev, G{\'o}mez-Serrano, Tao, and Wagner~\cite{geo+25} demonstrate that AI systems can explore new constructions and improve existing bounds, illustrating the growing potential of LLMs to accelerate mathematical discovery. Similarly, Bubeck et al.~\cite{bce+25} show that GPT-5 can meaningfully contribute to scientific workflows, from proposing hypotheses to generating proofs and analyses. Despite these advances, a rigorous evaluation of these models on canonical, graduate-level mathematical theory remains necessary to understand their baseline reasoning capabilities. In this paper, we present a comprehensive benchmark of four frontier models: GPT-5-Thinking, Gemini-3-Pro, Claude-Sonnet-4.5-Thinking, and Grok-4 against the classic curriculum of Randomized Algorithms by Motwani and Raghavan~\cite{mr95}.

We tasked each model with generating formal LaTeX proofs for a series of lemmas and exercises spanning the textbook. We find that while the top-tier models (Gemini, and Claude) achieve a high accuracy rate (approx. 66\%), demonstrating a robust grasp of probabilistic method and formal logic, other models lag significantly in consistency (approx. 40\%). We provide a qualitative analysis of the generated proofs, highlighting differences in conciseness, hallucination rates, and logical structure. Our results suggest that while frontier models have reached a threshold of proficiency suitable for graduate-level pedagogical assistance and formalization, significant variance exists in their reliability for rigorous mathematical derivation.
The code and the full set of LLM-generated responses are open-sourced and publicly available at \url{https://github.com/magiclinux/math_benchmark_probability}.
\end{abstract}

\fi



\section{Introduction}

At the forefront of mathematical research, computational systems are playing an increasingly critical role. The new generation of computational tools can investigate and produce novel constructions in the vast mathematical structure space, therefore reshaping the way we obtain mathematical discoveries~\cite{w21, fbh+22, rbn+24, ceww24}. Beyond merely accelerating symbolic reasoning or numerical simulation, these systems now explore large combinatorial and geometric search spaces, uncover previously unknown extremal examples, and even assist in formulating new conjectures. Their sustained progress has led to the emergence of workflows where empirical discovery, computational verification, and partial formal reasoning interact closely. These workflows broaden what can be achieved without extensive human effort.

At the same time, large language models have steadily advanced in the field of writing, programming, and planning~\cite{jdw+24, lcle25}. Their role in mathematics and science is undergoing a rapid transformation, evolving from a simple error checker to an intelligent assistant capable of handling complex research tasks. Moreover, modern LLMs now contribute to mathematical work in ways that go beyond surface level assistance. They can assemble multi-step arguments, carry out routine symbolic manipulations, and generate computational checks, allowing them to engage with problems that require structured reasoning.

Although these advances demonstrate the enormous potential of large language models in mathematical reasoning, existing achievements often rely on the specific research problem, and there is still a lack of rigorous and comprehensive evaluation of the foundational capabilities of models in systematic, PhD-level mathematical theory. 
Most prior benchmarks either rely on short problems solvable without deep integrative reasoning, or do not assess a model's ability to navigate multi-step derivations around a theory's internal logic. As a result, it remains unclear whether LLMs possess genuine understanding of modern mathematical reasoning, especially their capacity to handle extended arguments, maintain global structural coherence, and apply the key ideas of the theory in a clear way. Thus, a crucial question emerges, 

\begin{center}
    {\it Has LLM truly mastered the mathematical reasoning framework of the modern theoretical research?}
\end{center}


The classic textbook Randomized Algorithms~\cite{mr95} provides an ideal testing environment for this question. The book spans fourteen chapters and covers a wide range of problem types, each requiring multi-step argumentation grounded in formal probability theory. Unlike short, self-contained mathematical questions commonly used in existing benchmarks, the end-of-chapter exercises in Randomized Algorithms demand sustained reasoning across multiple conceptual layers, which makes the textbook ideal for testing whether LLMs can reason like a PhD student, producing not only correct answers but also logically consistent solutions across multiple steps.

Based on this observation, we construct a systematic evaluation benchmark that uses the full set of end-of-chapter exercises as a comprehensive testbed for studying the reasoning quality, formal precision and internal consistency of four state-of-the-art models: GPT-5-Thinking, Grok-4, Gemini-3-Pro, and Claude-Sonnet-4.5. These models represent the highest level of capability currently available and therefore provide a meaningful overview of the frontier of automated mathematical reasoning.

To support this benchmark, we design a complete workflow that first converts exercises into formal theorem statements, after which each model transforms produces fully structured LaTeX proofs. Additionally, the model can self-check the correctness of its own proof and identify potential errors. This provides a solid and repeatable starting point for quantitatively evaluating the levels of models in mathematical reasoning.

Our benchmark emphasizes a structured, multi-stage workflow with human oversight at critical points. By separating problem formalization from reasoning, and introducing human verification both before and after automated steps, we ensure that errors are caught early and do not propagate through the pipeline. Models are required to explain their reasoning in formal language and generate proofs. Overall, our six-stage process, ranging from human-guided problem collection to meta-verification of model generated proofs, provides a rigorous and transparent framework for assessing the capabilities of modern LLMs in advanced mathematical problem solving. The resulting performance distribution across models and chapters is summarized in Table~\ref{tab:model_result}, which reports the counts of correct, failed, and incorrect solutions for each model.

Overall, our primary contributions are as follows:

\begin{itemize}
    \item We design a controlled setting in which models automatically generate full, formal LaTeX proofs, and we develop an evaluation framework assessing correctness, logical structure, and probabilistic reasoning quality.
    \item We conducted a comprehensive study on the mathematical reasoning abilities of GPT-5-Thinking, Gemini-3-Pro, Claude-Sonnet-4.5, and Grok-4.
    \item We believe that state-of-the-art models have reached a reliability level suitable for both teaching assistants and as tools for tackling complex mathematical tasks.
\end{itemize}

{\bf Roadmap.} We provide an overview of related work in Section~\ref{sec:related_work}. Section~\ref{sec:recent_gpt} surveys recent advances in using GPT-based models for solving mathematical problems. Our methodology and evaluation workflow are detailed in Section~\ref{sec:method_workflow}. Finally, Section~\ref{sec:conclusion} offers concluding remarks.

\begin{table}[!ht]
\centering
\resizebox{\textwidth}{!}{
\setlength{\tabcolsep}{3pt}
\begin{tabular}{|c|c|c|c|c|c|c|c|c|c|c|c|c|c|}
\hline
\multirow{2}{*}{\textbf{Chapter}} & \multirow{2}{*}{\textbf{Count}} & \multicolumn{3}{c|}{\textbf{GPT}} & \multicolumn{3}{c|}{\textbf{Grok}} & \multicolumn{3}{c|}{\textbf{Gemini}} & \multicolumn{3}{c|}{\textbf{Claude}} \\
\cline{3-14}
& & \textbf{Correct} & \textbf{Failed} & \textbf{Incorrect}
  & \textbf{Correct} & \textbf{Failed} & \textbf{Incorrect}
  & \textbf{Correct} & \textbf{Failed} & \textbf{Incorrect}
  & \textbf{Correct} & \textbf{Failed} & \textbf{Incorrect} \\
\hline
1 & 15 & 14 & 0 & 1 & 9 & 3 & 3 & 14 & 1 & 0 & 14 & 0 & 1 \\
\hline
2 & 12 & 4 & 0 & 8 & 5 & 6 & 1 & 7 & 0 & 5 & 5 & 0 & 7 \\
\hline
3 & 15 & 11 & 1 & 3 & 7 & 8 & 0 & 12 & 2 & 1 & 13 & 1 & 1 \\
\hline
4 & 22 & 17 & 0 & 5 & 11 & 8 & 3 & 20 & 1 & 1 & 19 & 0 & 3 \\
\hline
5 & 14 & 7 & 3 & 4 & 7 & 6 & 1 & 11 & 3 & 0 & 12 & 0 & 2 \\
\hline
6 & 29 & 7 & 4 & 18 & 4 & 15 & 10 & 18 & 5 & 6 & 14 & 0 & 15 \\
\hline
7 & 26 & 13 & 9 & 4 & 6 & 18 & 2 & 17 & 9 & 0 & 23 & 0 & 3 \\
\hline
8 & 28 & 4 & 0 & 24 & 3 & 18 & 7 & 13 & 2 & 13 & 11 & 0 & 17 \\
\hline
9 & 12 & 5 & 3 & 4 & 2 & 9 & 1 & 5 & 6 & 1 & 8 & 3 & 1 \\
\hline
10 & 18 & 7 & 0 & 11 & 7 & 6 & 5 & 7 & 7 & 4 & 11 & 0 & 7 \\
\hline
11 & 7 & 6 & 0 & 1 & 3 & 4 & 0 & 4 & 3 & 0 & 7 & 0 & 0 \\
\hline
12 & 28 & 2 & 0 & 26 & 6 & 20 & 2 & 12 & 5 & 11 & 6 & 1 & 21 \\
\hline
13 & 13 & 11 & 0 & 2 & 5 & 8 & 0 & 7 & 6 & 0 & 12 & 1 & 0 \\
\hline
14 & 14 & 12 & 0 & 2 & 9 & 4 & 1 & 10 & 3 & 1 & 13 & 0 & 1 \\
\hline
\textbf{Total} & \textbf{253} & \textbf{120} & \textbf{20} & \textbf{113} & \textbf{84} & \textbf{133} & \textbf{36} & \textbf{157} & \textbf{53} & \textbf{42} & \textbf{168} & \textbf{5} & \textbf{89} \\
\hline
\multicolumn{2}{|c|}{\textbf{Accuracy}} & \multicolumn{3}{c|}{47.4\%} & \multicolumn{3}{c|}{33.2\%} & \multicolumn{3}{c|}{62.1\%} & \multicolumn{3}{c|}{66.4\%} \\
\hline
\end{tabular}
}
\caption{Performance comparison across different chapters. \textbf{Count} denotes the number of questions. For each model, the three sub-columns represent (from left to right): the number of verifiably \textbf{Correct} answers, the number of \textbf{Failed} solutions, and the number of verifiably \textbf{Incorrect} answers. {\bf GPT} denotes GPT-5-Thinking. {\bf Grok} denotes Grok-4. {\bf Gemini} denotes Gemini-3-Pro. {\bf Claude} denotes Claude-Sonnet-4.5-Thinking. (claude-sonnet-4-5-20250929-thinking). }
\label{tab:model_result}
\end{table}

\section{Related Work}\label{sec:related_work}

\subsection{Math problem benchmark}
The rise of artificial intelligence in the field of mathematics is driving our research on complex mathematical problems into a new era. In recent years, a large number of achievements~\cite{dvp+21,dlq22} have shown that AI can solve mathematical problems effectively. In addition, the ability to generate mathematical reasoning and proofs has been systematically evaluated through a series of benchmarks. Early mathematical evaluation work mostly focused on the most basic arithmetic operations~\cite{rr15} and elementary algebra problems~\cite{lydb17}. MiniF2F~\cite{zhp22} is one of the most representative math problem benchmarks, covering fields such as algebra, geometry, number theory, and combinatorics, and providing an early, systematic assessment of LLM's fundamental mathematical abilities. The Math~\cite{lzq+24} has promoted the standardization of evaluation for high school and undergraduate difficulty math problems. Meanwhile, benchmarks such as GSM8K~\cite{ckb+21}, NaturalProofs~\cite{wlb+21}, MathVista~\cite{lbx+23}, and OlympiadBench~\cite{hlb+24} further extend the assessment landscape to graphical mathematics, proof verification, natural-language mathematical reasoning, and visually grounded mathematical tasks. However, despite the rapid expansion of these math benchmarks, the systematic evaluation of LLM performance for high-level mathematical theories is still limited. To fill this gap, we introduce LLMathBench, a benchmark targeting higher-level mathematical domains.

\subsection{Video Generation Benchmark}

With the rapid development of modern text-to-video generation models, evaluating their performance has become an increasingly important research topic. Early video generation benchmark, such as FETV~\cite{llr+23} laid the foundation by proposing a fine-grained evaluation protocol that integrates major content, controllable attributes, and prompt complexity, together with metrics like FID~\cite{hru+17}, FVD~\cite{usk+18}, and CLIPScore~\cite{hhf+21} for assessing visual fidelity and text-video alignment. With the emergence of powerful diffusion-based video models~\cite{hsg+22, wgw+23, ytz+24}, benchmark designs have rapidly diversified. StoryBench~\cite{bmv+23} introduces a story-centered evaluation that focuses on action coherence and narrative continuity, while subsequent benchmarks extend the evaluation to composition generalization~\cite{fls+24, shl+24}, temporal consistency~\cite{llz+24, jxth24}, object counting~\cite{ghh+25}, physical consistency~\cite{msl+24, ghs+25}, and text control~\cite{ghs+25_text}. In addition, T2vworldbench~\cite{cgs+25} introduces an evaluation dimension centered on world-knowledge reasoning,  assessing a model’s ability to demonstrate factual grounding, commonsense understanding, and knowledge-driven video generation. Comprehensive frameworks such as EvalCrafter~\cite{lcl+24} and VBench~\cite{hhy+24, hzx+24} offer broad coverage and human evaluation metrics, becoming widely adopted standards.

\subsection{Image Generation Benchmark}

The accelerated advancement of text-to-image generation, especially driven by diffusion-based architectures, has led to the creation of increasingly comprehensive evaluation benchmarks. Early benchmarks were based on subtitles, such as image pairs on classic datasets like MS COCO~\cite{lmb+14}, and mainly examined object presence and basic scene fidelity using pretrained perception models~\cite{rdn+22, hlk+23}. With diffusion-based models achieving strong fidelity and Consistency, DrawBench~\cite{scs+22}, DALL-EVAL~\cite{czb23}, and HE-T2I~\cite{pss+22} have brought in richer prompt sets, covering counting, combinatoricity, visual reasoning, and social bias. More recently, \cite{zwd+25} systematically analyzes the failure modes of diffusion models on counting tasks and proposes a noise regulation strategy that substantially improves counting accuracy. Combination benchmark testing, including T2I-CompBench~\cite{hsx+23}, ConceptMix~\cite{wyh+24}, and GenAI-Bench~\cite{llp+24_genai} focus on assessing object attribute binding, relational structures, and complex scene combinations. For commonsense and physics-based evaluations, Commonsense-T2I~\cite{fhl+24} and PhyBench~\cite{msl+24} evaluate adherence to real-world knowledge and physical plausibility. More recent multimodal reasoning suites, for instance, SeedBench~\cite{lgg+24} and MMVet~\cite{yyl+23} further extend evaluation to cross-modal inference.

\section{Recent Papers about Using GPT to solve math questions}\label{sec:recent_gpt}

A growing body of recent work has begun to systematically examine how modern LLMs can assist in solving research level mathematical questions~\cite{fk25, dmn25, ix25, am25, jr25, sal25, geo+25}. These papers span diverse areas, ranging from combinatorics and probability to geometry, analysis, and number theory, collectively illustrate both the promise and limitations of current models. Across this literature, the results show a consistent pattern: LLMs can generate ideas, counterexamples or attempt many different constructions to maximize or minimize a given quantitative parameter. However, the construction of fully rigorous proofs typically requires human verification. 

Across these papers, the prompting setups follow a common pattern: authors present a concise, self-contained problem statement, occasionally accompanied by relevant background results, and ask the model to propose constructions, outline proof ideas, or supply intermediate steps that may guide the solution. The tasks on which LLMs perform well share a similar character: they involve exploratory reasoning, such as suggesting potential structures, identifying reasonable directions for a proof, or formulating helpful intermediate claims. In contrast, problems requiring long or technically delicate arguments still rely heavily on human verification, with LLM contributions serving mainly as sources of ideas rather than complete proofs.

In the following, we summarize the main findings of these papers, such as the recent OpenAI report and Terence Tao’s analysis of LLM assisted research workflows, emphasizing for each the mathematical questions on which they focus, the prompting methodologies adopted and the mechanisms through which proposed proofs or ideas were validated.

\subsection{Point Convergence of Nesterov's method via LLM Assistance}
In~\cite{jr25}, the authors investigate a foundational optimization question \footnote{First announced by the authors on \url{https://x.com/ErnestRyu/status/1980759528984686715}}, the point convergence of Nesterov's accelerated gradient (NAG) 

method~\cite{nes83}: consider the optimization problem
$
    \min_{x\in \R^n} f(x),
$
where $f: \R^n \to \R$ is $L$-smooth and convex. The NAG iteration reads
\begin{align*}
    x_{k+1} &= y_k - \frac{1}{L} \nabla f(y_k), \\
    y_{k+1} &= x_{k+1} + \frac{t_k-1}{t_{k+1}} (x_{k+1} - x_k)
\end{align*}
with suitable initialization and sequence $\{t_k\}$, and is known to achieve the accelerated rate $f(x_k)-\inf f \le O(1/k^2)$. 

However, whether the iterates themselves converge to a minimizer $x_\infty \in \arg \min f$ had remained open.

In~\cite{jr25}, the authors resolve this open problem, proving
\begin{align*}
    x_k\rightarrow x_\infty,\qquad
    y_k\rightarrow x_\infty,\qquad
    x_\infty\in \arg \min f.
\end{align*}
The proof was discovered with the assistance of ChatGPT through an iterative prompting workflow. In this process, the authors initially provided the model with the known proof in the continuous-time setting in LaTeX form and presented the desired extension as a theorem statement. ChatGPT was then prompted repeatedly to explore potential derivations and propose draft proofs based on prior outputs. The author carefully filtered out invalid arguments, consolidated consistent facts, and verified all proof details.


\subsection{Counterexample for Non-Interactive Correlation Distillation via LLM Suggested}
In~\cite{ix25}, the authors present a concrete counterexample to the Non-Interactive Correlation Distillation (NICD)~\cite{yang07} 

with erasures problem on the hypercube $\{-1,1\}^n$. Let $f:\{-1,1\}^n \to \{-1,1\}$ be a Boolean function, and for an erasure parameter $p\in[0,1]$, let $z$ denote a vector where each bit of $x\in\{-1,1\}^n$ is independently erased with a probability of $p$. The objective is $\E|f(z)|$, which measures the expected absolute value under erasures. A classical question is whether the $n$-bit majority function $M_n$ maximizes this expectation among all unbiased Boolean functions for a given $p$. Formally, the problem asks whether
\begin{align*}
    \max_{f:\{-1,1\}^n \to \{-1,1\},\ \E[f]=0} \E|f(z)| = \E|M_n(z)|.
\end{align*}

The authors ask GPT-5 Pro to explore candidate counterexamples using prompts that describe the problem, the function domain, and the evaluation criterion $\E|f(z)|$. Rather than giving a fully specified function, the prompts request the model to generate candidate Boolean functions that may outperform the majority function.

GPT-5 Pro suggested a 5-bit Boolean function $f_5$ achieving
\begin{align*}
    \E|f_5(z)| > \E|M_5(z)| \quad \mathrm{for~} p=0.40,
\end{align*}

providing a concrete counterexample to the NICD conjecture for this parameter. The authors verified the computation by hand, ensuring that all expectations are correctly evaluated without reliance on computers.

\subsection{Submodular Optimization Conjectures via LLM Assistance}

In~\cite{fk25}, the authors examine whether large language models can aid the discovery of new conjectures in combinatorial optimization, focusing on questions in submodular maximization~\cite{nwf78}. The experimental setting centers on problems of the form $\max_{S \in C} f(S)$, where $f$ may be assumed submodular, i.e., 
\begin{align*}
    f(A)+f(B) \ge f(A\cup B)+f(A\cap B),
\end{align*}
or, in weakened settings, $\gamma$-weakly submodular for some $\gamma \in [0, 1]$. Constraints $C$ include standard cases such as cardinality constraints, or other feasible families considered in submodular maximization.

For each conjecture, the prompt supplied a concise statement of the problem, without offering hints or proof strategies. GPT-5 was tasked with attempting either a proof or a counterexample based solely on those statement. The solution process followed a human and model collaboration paradigm. GPT-5 generated partial derivations and preliminary proof sketches. The author then examined these outputs, removed incorrect steps, reconstructed missing arguments, and provided all mathematically details. Each inequality and case distinction was verified manually and independently of the model-generated reasoning. In instances where GPT-5 appeared to supply a complete argument, the essential steps were recomputed for correctness.

\subsection{Sidon Set Counterexamples via LLM Assistance}

In~\cite{am25}, the authors address Erd\H{o}s's conjecture~\cite{erd75, erd77} 
that every finite Sidon set~\cite{sid32} 
can be extended to a finite perfect difference set. The mathematical problem involves finite Sidon sets $A$, where all differences $a - a'$ of distinct $a,a' \in A$ are distinct, or equivalently, all sums $a + a'$ are distinct up to re-ordering. The main goal is to determine whether given Sidon sets can be extended to finite perfect difference sets modulo $v$, where $v$ is a positive integer or a prime of the form $p^2+p+1$.

The prompts used by the authors to interact with the LLM typically asked for formalizations or verifications of conjectured counterexamples. Other prompts requested full proofs, including checking difference and sum properties, and producing counterexamples in the style of a research article. The authors employed a workflow in which GPT generated nearly all of the Lean proof code, including detailed verification of counterexamples, while human authors checked the correctness of the model's output.

\subsection{Taylor Expansion via LLM Collaboration}

In~\cite{sal25}, the authors study a problem from convex analysis: deriving a Taylor expansion for the gradient of the biconjugation operator~\cite{fb53} 
around a strictly convex function. Specifically, they consider functions $\phi: \R^d \to \R$ and test functions $h$, aiming to establish
\begin{align}\label{eq:sal25_main}
    \nabla (\phi + t h)^{**}(x) = \nabla \phi(x) + t \nabla h(x) + o(t)
\end{align}
for almost every $x \in \R^d$, where ${}^{**}$ denotes the biconjugation operator.

To guide the model, the author issued queries requesting direct proofs of Eq.~\eqref{eq:sal25_main} and explanations of middle steps. Throughout the process, GPT-5-pro proposed key conjectures, derived partial lemmas, and occasionally outlined full proof sketches. Some of its reasoning contained mistakes, for example regarding strong convexity. The authors reviewed the model's output, corrected errors, and completed the proofs rigorously. In the collaboration, the language model provided ideas and intermediate derivations, while the human experts ensured correctness and finalized the verification.

\subsection{Quantitative Malliavin–Stein Analysis via LLM Guidance}

In~\cite{dmn25}, the authors study the ability of LLMs to turn qualitative Malliavin–Stein conclusions~\cite{np09} into quantitative formulations equipped with explicit convergence rates. Concretely, starting from the result of~\cite{bks25}, the model is asked to provide explicit convergence rates for sums of multiple Wiener--It\^o integrals of orders $p$ and $q$:
\begin{align*}
    X = I_p(f), \qquad Y = I_q(g), \qquad Z = X+Y,
\end{align*}
Satisfying $E[Z^2]=1$. The goal is to estimate the total variation distance
\begin{align*}
    d_{\mathrm{TV}}(Z,N(0,1)) \le \sqrt{6\,\kappa_4(Z)}, \qquad \kappa_4(Z)=\E[Z^4]-3.
\end{align*}

Prompts used in this setting ask the model to derive a quantitative version of an existing qualitative theorem, to provide the proof sketches or to produce detailed arguments. For instance, one prompt asks the model to turn the qualitative fourth moment theorem into a total variation bound depending only on the fourth cumulant.

The model generated derivations that are then checked line by line by human verification. In the Poisson extension, the prompts ask the model to adapt the Gaussian argument, to account for mixed moments such as $\E[X^3Y]$, and to provide counterexamples as needed.

\subsection{Mathematical Constructions via LLM Search}

In~\cite{geo+25}, the authors present AlphaEvolve, a large language model-guided evolutionary coding agent capable of proposing, testing, and refining algorithmic solutions to complex mathematical and scientific problems.

The types of questions addressed cover multiple domains, including mathematical analysis, combinatorics, geometry, and number theory. Typical goals of AlphaEvolve are to construct complex mathematical objects that satisfy desirable quantitative properties, such as achieving a certain inequality with a good numerical constant.

Prompts used for AlphaEvolve consisted of brief problem statements without any hints or suggested proof strategies. The system treats the prompts as a search problem and generates candidate constructions by iteratively modifying and evaluating solutions.

Problem solving involves multiple rounds of interaction between AlphaEvolve and human experts. Initially, AlphaEvolve generates programs that either produce candidate solutions or implement search heuristics to improve existing constructions. Humans then review these outputs, verify numerical evaluations and reconstruct missing arguments.

\section{Methodology and Workflow}\label{sec:method_workflow}

Our benchmark pipeline is structured into six distinct stages, designed to maintain a clear separation between problem formalization and reasoning, and to incorporate human review at key points. This workflow ensures reproducibility and mitigates the error propagation often observed in automated mathematical reasoning chains. The code is available at \url{https://github.com/magiclinux/math_benchmark_probability}.

The six stages are: (1) Human-in-the-loop question collection and preprocessing, involving the extraction of textbook problems and consolidation of multi-page content, addition of referenced lemmas, and cleanup of low quality inputs; (2) Automated statement formalization, in which vision-capable models convert the prepared problems into standardized \LaTeX{}; (3) Human verification of problem statements, to correct hallucinations, syntax issues, and semantic deviations; (4) Automated proof generation, using frontier models under a structured workload assignment and adaptive timing strategy; (5) Automated proof verification, through an AI checker that evaluates logical soundness and completeness; and (6) Human meta-verification, consisting of a sampled review of model judgments to assess reliability and identify systematic errors.

\subsection{Stage 1: Question Collection and Preprocessing (Human)}

We systematically extracted problems from the standard textbook \textit{Randomized Algorithms}~\cite{mr95}. To ensure consistency, we organized the inputs into four categories:
\begin{itemize}
    \item \textbf{Direct Screenshots:} For exercises contained within a single page with clear typography.
    \item \textbf{Multi-page Merging:} For problems spanning pages, we concatenated screenshots to preserve context.
    \item \textbf{Dependency Resolution:} We manually appended referenced lemmas or theorems to make problem statements self-contained.
    \item \textbf{Quality Correction:} Images with poor OCR potential were manually retyped and rendered in \LaTeX{} to ensure the vision models received high-fidelity inputs.
\end{itemize}

\subsection{Stage 2: Statement Formalization (AI)}

We used models with image understanding capabilities (Claude Sonnet 4.5) to transcribe image-based problems into formal \LaTeX{} code. We employed a strict system prompt ($T=0.2$) to enforce the use of standard mathematical environments and prohibit conversational filler. We moved the prompt style and Algorithm~\ref{alg:formalize} to the Appendix~\ref{app:prompt_style} and Appendix~\ref{app:more_algorithm} due to the space constraints.

\subsection{Stage 3: Statement Verification (Human)}

A researcher manually reviewed all formalized statements. This stage focused on correcting hallucinated variables, fixing minor syntactic errors in formulas, and ensuring the semantic equivalence between the source image and the \LaTeX{} output. Only verified statements were passed to the reasoning stage.


\subsection{Stage 4: Proof Generation (AI)}
We evaluated multiple frontier models (Claude-Sonnet-4.5, GPT-5-Thinking, Gemini-3-Pro, Grok-4). To manage the computational cost and differing response times of the models, we implemented a Double Timing Strategy alongside a partitioned execution plan.

\paragraph{Partitioned Execution.} 
To ensure consistent coverage, the workload was distributed by textbook chapters. Each researcher was assigned a specific chapter and a single model architecture. This setup ensured a consistent environment for all problems within a specific mathematical topic, reducing unnecessary variation.

\paragraph{Adaptive Double Timing Strategy.}
Fixed timeout thresholds are often inefficient for mathematical reasoning: short proofs waste time waiting for a timeout, while complex proofs may be cut off prematurely. We implemented an adaptive strategy (Algorithm~\ref{alg:generate_proofs}) that begins with an aggressive baseline timeout ($T_{\mathrm{init}} = 300$ seconds).

If a model fails to return a response within this window (indicating a timeout), the attempt is not considered a failure immediately. Instead, it escalates the resource allocation by doubling the timeout threshold ($T_{\mathrm{new}} \leftarrow T_{\mathrm{old}} \times 2$) and retrying the generation. This process repeats until a valid proof is generated or the maximum ceiling ($T_{\mathrm{max}} = 1200$ seconds) is reached. This approach improves overall runtime while allowing complex problems more time to complete.

The proof generation prompt enforces rigorous mathematical style:

\begin{tcolorbox}[colback=gray!5!white, colframe=gray!75!black, title=Proof Generation Prompt]
\begin{lstlisting}[basicstyle=\ttfamily, breaklines=true, columns=fullflexible]
You are a rigorous mathematical proof writer. Given the following formalized theorem or lemma in LaTeX, provide a complete and formal mathematical proof.

FORMALIZED STATEMENT:
{formalized_statement}

Requirements:
1. Output ONLY LaTeX code in a code block
2. Do NOT include any explanations, text, or markdown outside the code block
3. Do NOT use \section{} or \subsection{}
4. Do NOT use itemize, enumerate, or similar list environments
5. Write all equations inline using $...$ or in display mode using \[...\] or $$...$$
6. Begin with \begin{proof} and end with \end{proof}
7. The proof must be rigorous, complete, and written in formal mathematical style
8. Include all necessary steps and justifications
9. Use proper mathematical notation and logical structure

Provide the complete formal proof in LaTeX format.
\end{lstlisting}
\end{tcolorbox}

\paragraph{Variation in Model Proof Styles.}
In addition to differences in runtime behavior, the frontier models we evaluated also exhibit distinct stylistic patterns in the proofs they generate. These differences affect both the readability of the responses and the difficulty of subsequent human verification. GPT-5-Thinking and Grok-4 tend to produce proofs primarily in natural English, focusing on high-level reasoning. Claude-Sonnet-4.5, which excels at symbolic manipulation, often uses equations and detailed step-by-step reasoning. In contrast, Gemini-3-Pro produces outputs that are substantially more formal, often resembling structured textbook proofs.

\subsection{Stage 5: Automated Proof Verification (AI)}

Given the scale of the benchmark, we used Claude-Sonnet-4.5 for automatic proof checking. The model checks each generated proof against the formalized statements and returns a binary correctness score. The full procedure is presented in Algorithm~\ref{alg:verify_proofs}, and the verification prompt evaluates whether the generated proof is correct, complete, and rigorous:


\begin{tcolorbox}[colback=gray!5!white, colframe=gray!75!black, title=Verification Prompt]
\begin{lstlisting}[basicstyle=\ttfamily, breaklines=true, columns=fullflexible]
You are a mathematical proof verification expert. Your task is to verify whether the provided proof is correct and complete.

FORMALIZED STATEMENT:
{formalized_statement}

PROPOSED PROOF:
{proof}

Carefully analyze the proof and determine if it is:
1. Mathematically correct (no logical errors)
2. Complete (all steps are justified)
3. Rigorous (meets formal mathematical standards)

You must respond with ONLY a single digit:
- Output "1" if the proof is correct, complete, and rigorous
- Output "0" if the proof has any errors, gaps, or issues

Do not provide any explanation. Output only "0" or "1".
\end{lstlisting}
\end{tcolorbox}



\subsection{Stage 6: Meta-Verification (Human)}

To assess the accuracy of the automated verifier, we randomly sample 20\% of all (Problem, Proof, Verdict) triplets for human review. Let the divergence between model-based evaluation and human evaluation on this sampled subset be denoted by $\Delta$. If $\Delta \le 20\%$, we accept the model's assessment for the full dataset and compute the final accuracy accordingly. Otherwise, the verification for this batch is deemed invalid, and the evaluation process is rerun.








\section{Conclusion}\label{sec:conclusion}
In this paper, we conducted a systematic evaluation of frontier large language models on a comprehensive suite of graduate-level exercises from Randomized Algorithms. By separating statement formalization, proof generation, and verification into distinct stages, our framework reveals both the strengths and weaknesses of current LLMs in formal mathematical reasoning. The results show clear differences in reliability across models, certain models consistently produce precise and fully formal proofs, whereas others exhibit instability, including disrupted logical flow, irregular proof structure and occasional hallucination step. Overall, our findings indicate that LLMs have achieved meaningful competence in graduate-level proof synthesis, yet consistency and robustness still present significant challenges. These observations point to clear directions for future work.

\ifdefined\isarxiv
\bibliographystyle{alpha}
\bibliography{ref}
\else
\bibliography{ref}
\fi


\clearpage
\appendix
\thispagestyle{empty}
\onecolumn

\begin{center}
    \textbf{\LARGE Appendix }
\end{center}


{\bf Roadmap.}
Section~\ref{app:prompt_style} introduces the formalization prompt style in our workflow. In Section~\ref{app:more_algorithm}, we present the algorithm of the whole procedure. From Section~\ref{app:problems_chapter1} to Section~\ref{app:problems_chapter14}, we show our detailed problem statement and proofs generated by LLMs.

\section{Prompt Style}\label{app:prompt_style}

The formalization prompt that converts image content to the \LaTeX format:

\begin{tcolorbox}[colback=gray!5!white, colframe=gray!75!black, title=Formalization Prompt]
\begin{lstlisting}[basicstyle=\ttfamily, breaklines=true, columns=fullflexible]
You are a mathematical formalization expert. Your task is to convert the theorem or lemma shown in the image into formal LaTeX code.

Requirements:
1. Output ONLY LaTeX code in a code block
2. Do NOT include any explanations, text, or markdown outside the code block
3. Do NOT use \section{} or \subsection{}
4. Do NOT use itemize, enumerate, or similar list environments
5. Write the theorem/lemma using \begin{theorem} or \begin{lemma} environment
6. Use proper mathematical notation and formatting
7. Be precise and rigorous in the formalization

Output the formalized theorem/lemma in LaTeX format.
\end{lstlisting}
\end{tcolorbox}







\section{More Algorithm}\label{app:more_algorithm}

\begin{algorithm}[!ht]
\caption{Formalization Procedure}
\label{alg:formalize}
\begin{algorithmic}[1]
\State {\bf data structure} \textsc{BenchmarkSystem}
\Procedure{Formalize}{}
  \State $\mathcal{S} \gets \emptyset$
  \For{each image $I_j \in \mathcal{I}$}
      \State $\mathrm{response} \gets M_{\mathrm{vision}}.\textsc{Generate}(I_j, P_{\mathrm{form}}, T)$
      \State $S_j \gets \textsc{ExtractLatex}(\mathrm{response})$
      \State $\mathcal{S} \gets \mathcal{S} \cup \{S_j\}$
  \EndFor
\EndProcedure
\State {\bf end data structure}
\end{algorithmic}
\end{algorithm}

\begin{algorithm}[!ht]
\caption{Proof Verification Procedure}
\label{alg:verify_proofs}
\begin{algorithmic}[1]
\State {\bf data structure} \textsc{BenchmarkSystem}
\Procedure{VerifyProofs}{}
  \State $\mathcal{V} \gets \emptyset$
  \For{each $(S_j, M_i) \in \mathrm{keys}(\mathcal{R})$}
      \State $R_{ij} \gets \mathcal{R}[S_j, M_i]$
      \State $\mathrm{prompt} \gets P_{\mathrm{verify}}.\textsc{Format}(S_j, R_{ij})$
      \State $\mathrm{response} \gets M_{\mathrm{verifier}}.\textsc{Generate}(\mathrm{prompt}, T)$
      \State $v_{ij} \gets \textsc{ExtractBinaryResult}(\mathrm{response})$ \Comment{Parse 0 or 1}
      \State $\mathcal{V}[S_j, M_i] \gets v_{ij}$
  \EndFor
  \State $\mathcal{V}_{\mathrm{sample}} \gets \textsc{RandomSample}(\mathcal{V}, k=0.2|\mathcal{V}|)$
  \State $\mathcal{V}_{\mathrm{human}} \gets \textsc{HumanVerify}(\mathcal{V}_{\mathrm{sample}})$
  \State \textsc{ValidateAgreement}$(\mathcal{V}_{\mathrm{sample}}, \mathcal{V}_{\mathrm{human}})$
\EndProcedure
\State {\bf end data structure}
\end{algorithmic}
\end{algorithm}

\begin{algorithm}[!ht]
\caption{Benchmark System Data Structure}
\label{alg:benchmark_system}
\begin{algorithmic}[1]
\State {\bf data structure} \textsc{BenchmarkSystem}
\State {\bf members}
\State \hspace{4mm} $\mathcal{I} = \{I_1, \ldots, I_n\}$ \Comment{Input images of problems}
\State \hspace{4mm} $\mathcal{S} = \{S_1, \ldots, S_n\}$ \Comment{Formalized LaTeX statements}
\State \hspace{4mm} $\mathcal{M} = \{M_1, \ldots, M_k\}$ \Comment{Set of models to evaluate}
\State \hspace{4mm} $\mathcal{R}$ \Comment{Generated proofs, indexed by $(S_j, M_i)$}
\State \hspace{4mm} $\mathcal{F}$ \Comment{Failed proof attempts, indexed by $(S_j, M_i)$}
\State \hspace{4mm} $\mathcal{V}$ \Comment{Verification results $\{0,1\}$, indexed by $(S_j, M_i)$}
\State \hspace{4mm} $T_0$ \Comment{Initial timeout (e.g., 300s)}
\State \hspace{4mm} $K$ \Comment{Maximum timeout multiplier (e.g., 4)}
\State \hspace{4mm} $T$ \Comment{Generation temperature (e.g., 0.2)}
\State \hspace{4mm} $P_{\mathrm{form}}, P_{\mathrm{proof}}, P_{\mathrm{verify}}$ \Comment{Prompt templates}
\State \hspace{4mm} $M_{\mathrm{vision}}, M_{\mathrm{verifier}}$ \Comment{Models for formalization and verification}
\State {\bf end members}
\end{algorithmic}
\end{algorithm}

  \begin{algorithm}[!ht]
  \caption{Proof Generation Procedure with Doubling Timeout}
  \label{alg:generate_proofs}
  \begin{algorithmic}[1]
  \State {\bf data structure} \textsc{BenchmarkSystem}
  \Procedure{GenerateProofs}{}
      \State $\mathcal{R} \gets \emptyset$, $\mathcal{F} \gets \emptyset$
      \For{each model $M_i \in \mathcal{M}$}
          \State $\mathcal{S}_{\mathrm{pending}} \gets \mathcal{S}$ \Comment{All problems for this model}
          \State $t \gets T_0$ \Comment{Start with base timeout}
          \While{$t \leq K \cdot T_0 \And \mathcal{S}_{\mathrm{pending}} \neq \emptyset$}
              \State $\mathcal{S}_{\mathrm{retry}} \gets \emptyset$ \Comment{Problems that timeout in this round}
              \For{each $S_j \in \mathcal{S}_{\mathrm{pending}}$}
                  \State $\mathrm{prompt} \gets P_{\mathrm{proof}}.\textsc{Format}(S_j)$
                  \State $(R_{ij}, \sigma) \gets M_i.\textsc{Generate}(\mathrm{prompt}, T, \mathrm{timeout}=t)$
                  \If{$\sigma = \textsc{Success}$}
                      \State $\mathcal{R}[S_j, M_i] \gets R_{ij}$
                  \ElsIf{$\sigma = \textsc{Timeout}$}
                      \State $\mathcal{S}_{\mathrm{retry}} \gets \mathcal{S}_{\mathrm{retry}} \cup \{S_j\}$
                  \Else
                      \State $\mathcal{F}[S_j, M_i] \gets \sigma$
                  \EndIf
              \EndFor
              \State $\mathcal{S}_{\mathrm{pending}} \gets \mathcal{S}_{\mathrm{retry}}$
              \State $t \gets 2t$ \Comment{Double timeout}
          \EndWhile
          \For{each $S_j \in \mathcal{S}_{\mathrm{pending}}$}
              \State $\mathcal{F}[S_j, M_i] \gets \textsc{MaxTimeoutExceeded}$
          \EndFor
      \EndFor
  \EndProcedure
  \State {\bf end data structure}
  \end{algorithmic}
  \end{algorithm}
\section{Problems in Chapter 1}\label{app:problems_chapter1}


\subsection{Problem 1}\label{sec:problem_01_01}

\subsubsection{Problem Statement}



\begin{theorem}[Von Neumann, 1949]
Let $C$ be a biased coin with unknown probability $\Pr[\mathrm{HEADS}] = p \in (0,1)$. 

(a) There exists a randomized procedure that, using flips of $C$, generates a single unbiased coin-flip (i.e., an event $U$ with $\Pr[U = \mathrm{HEADS}] = \Pr[U = \mathrm{TAILS}] = 1/2$) such that the expected number of flips of $C$ required is at most $1/(p(1-p))$. One such procedure is as follows: Flip $C$ twice consecutively. If the outcomes are $\mathrm{HEADS}$-$\mathrm{TAILS}$, output $\mathrm{HEADS}$; if $\mathrm{TAILS}$-$\mathrm{HEADS}$, output $\mathrm{TAILS}$; otherwise, discard the pair and repeat the procedure.

(b) There exists an extension of the procedure in (a) that, given $n$ flips of $C$, extracts the maximum possible number of independent unbiased coin-flips.
\end{theorem}

\subsubsection{GPT} 

\begin{proof}
\textbf{(a)} We are given a biased coin with $\Pr[\text{HEADS}] = p$ and $\Pr[\text{TAILS}] = 1 - p$, where $p \in (0,1)$ is unknown. We aim to construct a scheme that generates unbiased coin-flips using this biased coin, with the expected number of biased coin flips per unbiased flip being at most $\frac{1}{p(1-p)}$.

The proposed scheme works as follows:
- Flip the biased coin twice.
- If the outcome is HT (first flip heads, second flip tails), output HEADS.
- If the outcome is TH (first flip tails, second flip heads), output TAILS.
- If the outcome is HH or TT, discard the flips and repeat the process.

We now compute the expected number of biased coin flips required to generate one unbiased coin-flip.

Let $X$ denote the number of biased coin flips needed to produce one unbiased coin-flip. The possible outcomes of two flips are:
- HT: This occurs with probability $p(1-p)$ and results in one unbiased flip.
- TH: This occurs with probability $(1-p)p$ and results in one unbiased flip.
- HH: This occurs with probability $p^2$ and requires discarding the outcome and repeating the process.
- TT: This occurs with probability $(1-p)^2$ and requires discarding the outcome and repeating the process.

The expected number of flips is the weighted average of the flips in each case. In cases where HH or TT occurs, we need to perform the process again, and so the expected number of flips must account for the recursion. Let $E$ be the expected number of flips to generate one unbiased flip. Then, we have the following equation for $E$:

\[
E = 2 \cdot \left[ p(1-p) + (1-p)p \right] + (p^2 + (1-p)^2) \cdot (E + 2)
\]

This equation accounts for:
- $2$ flips when the outcome is HT or TH (with probability $p(1-p) + (1-p)p$),
- The recursion when the outcome is HH or TT (with probability $p^2 + (1-p)^2$).

Expanding and simplifying this equation:

\[
E = 2 \cdot (2p(1-p)) + (1 - 2p + 2p^2) \cdot (E + 2)
\]

\[
E = 4p(1-p) + (1 - 2p + 2p^2) \cdot E + 2(1 - 2p + 2p^2)
\]

\[
E = 4p(1-p) + E(1 - 2p + 2p^2) + 2(1 - 2p + 2p^2)
\]

Now, collect terms involving $E$:

\[
E - E(1 - 2p + 2p^2) = 4p(1-p) + 2(1 - 2p + 2p^2)
\]

Factor out $E$:

\[
E \cdot (2p - 2p^2) = 4p(1-p) + 2(1 - 2p + 2p^2)
\]

Simplifying the right-hand side:

\[
E \cdot 2p(1-p) = 4p(1-p) + 2(1 - 2p + 2p^2)
\]

Finally, solving for $E$:

\[
E = \frac{2}{p(1-p) + (1-p)p} = \frac{1}{p(1-p)}
\]

Thus, the expected number of biased coin flips required to generate one unbiased coin-flip is exactly $\frac{1}{p(1-p)}$, as desired.

\textbf{(b)} Given $n$ flips of the biased coin, we seek an extension of the scheme that extracts the largest possible number of independent unbiased coin-flips.

Let $k$ be the number of unbiased coin-flips we can extract from $n$ flips of the biased coin. Each pair of flips (HT or TH) contributes exactly one unbiased coin-flip, while HH and TT result in discards. Hence, the number of successful pairs (HT or TH) is maximized when the number of HH and TT outcomes is minimized. The expected number of successful pairs from $n$ biased coin flips is $\frac{n}{2}$, because the probability of HT or TH on any pair of flips is $2p(1-p)$, and the expected number of successful pairs is given by:

\[
\frac{n}{2} \cdot 2p(1-p) = n p(1-p)
\]

Thus, the largest possible number of independent unbiased coin-flips that can be extracted from $n$ flips of the biased coin is $n p(1-p)$. The expected number of biased coin flips needed to extract one unbiased coin-flip is $\frac{1}{p(1-p)}$, so the total number of unbiased coin-flips that can be extracted from $n$ biased coin flips is maximized at $n p(1-p)$, as required.

\end{proof}

\subsubsection{Grok} 

\begin{proof}
We prove parts (a) and (b) separately.

\noindent {\bf (a)} Consider the given randomized procedure. Let $X_1, X_2$ denote the outcomes of two consecutive flips of $C$, where $\Pr[X_i = \mathrm{HEADS}] = p$ and $\Pr[X_i = \mathrm{TAILS}] = 1-p$ for $i=1,2$, and the $X_i$ are independent. A trial consists of generating such a pair $(X_1, X_2)$. The trial is successful if $X_1 \neq X_2$, in which case the procedure outputs $U = X_1$. If $X_1 = X_2$, the pair is discarded and the procedure repeats with a new independent pair.

The probability of success in a single trial is
\[
\Pr[X_1 \neq X_2] = \Pr[\mathrm{HT}] + \Pr[\mathrm{TH}] = p(1-p) + (1-p)p = 2p(1-p).
\]
Conditional on success, the probability that $U = \mathrm{HEADS}$ is
\[
\Pr[U = \mathrm{HEADS} \mid \text{success}] = \frac{\Pr[\mathrm{HT}]}{\Pr[X_1 \neq X_2]} = \frac{p(1-p)}{2p(1-p)} = \frac{1}{2}.
\]
Similarly, $\Pr[U = \mathrm{TAILS} \mid \text{success}] = 1/2$. Thus, conditional on terminating, $U$ is an unbiased coin-flip, i.e., $\Pr[U = \mathrm{HEADS}] = \Pr[U = \mathrm{TAILS}] = 1/2$.

The number of trials until the first success follows a geometric distribution with success probability $r = 2p(1-p)$, so the expected number of trials is $1/r = 1/(2p(1-p))$. Since each trial requires 2 flips, the expected number of flips is
\[
2 \cdot \frac{1}{2p(1-p)} = \frac{1}{p(1-p)}.
\]

\noindent {\bf (b)} To extend the procedure to $n$ flips of $C$, partition the $n$ flips into $\lfloor n/2 \rfloor$ disjoint pairs (discarding the last flip if $n$ is odd). Apply the procedure from part (a) independently to each pair. For each pair that results in a successful trial (i.e., mixed outcomes), output the corresponding unbiased bit $U$ as defined in part (a); discard unsuccessful pairs.

Let $m = \lfloor n/2 \rfloor$. The number $K$ of successful pairs (and thus the number of output unbiased bits) follows a $\mathrm{Binomial}(m, 2p(1-p))$ distribution. Conditional on $K = k$ and the identities of the $k$ successful pairs, the $k$ output bits $U_1, \dots, U_k$ are independent and each uniformly distributed over $\{\mathrm{HEADS}, \mathrm{TAILS}\}$, by the analysis in part (a) applied independently to each pair.

This extracts the maximum possible number of independent unbiased coin-flips from the $n$ flips without knowledge of $p$. Indeed, each pair can yield at most one unbiased bit via this $p$-independent method: a mixed pair carries exactly one bit of $p$-independent randomness (the order of heads and tails), while a uniform pair (HH or TT) carries randomness that depends on the unknown $p$ and cannot be converted to an unbiased bit without additional information. Thus, no $p$-independent procedure can extract more than one unbiased bit per pair, making $K$ the maximum achievable.
\end{proof}

\subsubsection{Gemini}

\begin{proof}
\textbf{(a)} Let $X_1$ and $X_2$ be the outcomes of two consecutive flips of the biased coin, where we denote HEADS by 1 and TAILS by 0. These are independent Bernoulli trials with probabilities $\Pr[X_i=1] = p$ and $\Pr[X_i=0] = 1-p$. The sample space for the pair $(X_1, X_2)$ is $\Omega = \{00, 01, 10, 11\}$. The probabilities of these elementary events are:
\[
\Pr[00] = (1-p)^2, \quad \Pr[01] = (1-p)p, \quad \Pr[10] = p(1-p), \quad \Pr[11] = p^2.
\]
The von Neumann scheme produces an output only if the outcomes differ, i.e., the event $E = \{X_1 \neq X_2\} = \{01, 10\}$ occurs. The probability of termination in a single round (two flips) is:
\[
\Pr[E] = \Pr[01] + \Pr[10] = p(1-p) + p(1-p) = 2p(1-p).
\]
The scheme maps the event $\{10\}$ (HT) to HEADS and $\{01\}$ (TH) to TAILS. Conditioned on termination (event $E$), the probability of outputting HEADS is:
\[
\Pr[\text{Output HEADS} \mid E] = \frac{\Pr[10]}{\Pr[E]} = \frac{p(1-p)}{2p(1-p)} = \frac{1}{2}.
\]
Similarly, $\Pr[\text{Output TAILS} \mid E] = 1/2$. Thus, the generated output is unbiased.

To analyze the efficiency, let $N$ be the random variable representing the total number of biased coin flips required to produce one output. The scheme operates in rounds of 2 flips. Let $K$ be the number of rounds until the process terminates. Since the rounds are independent, $K$ follows a geometric distribution with success probability $q = 2p(1-p)$. The probability mass function is $\Pr[K=k] = (1-q)^{k-1}q$ for $k \ge 1$.
The expected number of rounds is $E[K] = \frac{1}{q} = \frac{1}{2p(1-p)}$.
Since $N = 2K$, the expected number of flips is:
\[
E[N] = 2 E[K] = 2 \cdot \frac{1}{2p(1-p)} = \frac{1}{p(1-p)}.
\]

\textbf{(b)} Let $\mathbf{X} = (X_1, X_2, \dots, X_n)$ be the sequence of $n$ independent biased coin flips. Let $k = \sum_{i=1}^n X_i$ be the number of HEADS observed in the sequence. The probability of observing a specific sequence $\mathbf{x} \in \{0,1\}^n$ with weight $k$ is:
\[
\Pr[\mathbf{X} = \mathbf{x}] = p^k (1-p)^{n-k}.
\]
Let $S_k = \{ \mathbf{y} \in \{0,1\}^n : \sum y_i = k \}$ be the set of all binary sequences of length $n$ with exactly $k$ ones. The cardinality of this set is $|S_k| = \binom{n}{k}$.
Conditioned on the event that the sequence has weight $k$, denoted by $W_k$, the probability of observing a specific $\mathbf{x} \in S_k$ is:
\[
\Pr[\mathbf{X} = \mathbf{x} \mid W_k] = \frac{\Pr[\mathbf{X} = \mathbf{x}]}{\Pr[W_k]} = \frac{p^k (1-p)^{n-k}}{\binom{n}{k} p^k (1-p)^{n-k}} = \frac{1}{\binom{n}{k}}.
\]
This conditional distribution is uniform over $S_k$ and is independent of the unknown parameter $p$. The scheme in part (a) corresponds to the specific case $n=2$ conditioned on $k=1$, where $|S_1| = 2$, allowing the extraction of exactly 1 bit.

To extract the maximum possible number of independent, unbiased coin flips from the sequence $\mathbf{X}$, we construct a function on the set $S_k$. Since $k$ is a sufficient statistic for $p$, the conditional distribution given $k$ contains all probabilistic information in the sample that is independent of $p$. Therefore, any scheme generating unbiased bits independent of $p$ must rely solely on the relative probability of sequences within $S_k$.
Since the distribution on $S_k$ is uniform, the entropy of the conditional distribution is $\log_2 \binom{n}{k}$. There exists a mapping (an extension of the von Neumann scheme) that converts the uniform choice of an element in $S_k$ into a sequence of unbiased bits. Specifically, one can index the elements of $S_k$ as integers $0, 1, \dots, \binom{n}{k}-1$ and convert this integer index into a binary sequence. While $\binom{n}{k}$ is not necessarily a power of 2, iterative procedures (such as the Elias algorithm or Peres's method) can be used to extract bits such that the expected number of bits approaches the information-theoretic bound $\log_2 \binom{n}{k}$ as $n \to \infty$, or maximizes the yield for fixed $n$ by recursively partitioning $S_k$. Thus, an extension exists that extracts the maximum possible number of unbiased flips allowed by the entropy of the permutation class.
\end{proof}

\subsubsection{Claude}

\begin{proof}
We prove both parts of the theorem.

\textbf{Proof of part (a):}

First, we verify that the scheme produces unbiased coin flips. When we flip the biased coin twice, there are four possible outcomes: HH, HT, TH, and TT. The probabilities of these outcomes are:
\[
\Pr[\text{HH}] = p^2, \quad \Pr[\text{HT}] = p(1-p), \quad \Pr[\text{TH}] = (1-p)p, \quad \Pr[\text{TT}] = (1-p)^2.
\]

The scheme outputs HEADS when we observe HT and outputs TAILS when we observe TH. Crucially, note that $\Pr[\text{HT}] = p(1-p) = (1-p)p = \Pr[\text{TH}]$, so these two outcomes are equally likely.

The probability that the scheme produces an output (either HEADS or TAILS) in a single round is
\[
\Pr[\text{output}] = \Pr[\text{HT}] + \Pr[\text{TH}] = p(1-p) + (1-p)p = 2p(1-p).
\]

Given that the scheme produces an output, the conditional probabilities are:
\[
\Pr[\text{HEADS} \mid \text{output}] = \frac{\Pr[\text{HT}]}{\Pr[\text{output}]} = \frac{p(1-p)}{2p(1-p)} = \frac{1}{2},
\]
and similarly,
\[
\Pr[\text{TAILS} \mid \text{output}] = \frac{\Pr[\text{TH}]}{\Pr[\text{output}]} = \frac{(1-p)p}{2p(1-p)} = \frac{1}{2}.
\]

Therefore, the scheme produces unbiased coin flips.

Now we compute the expected number of flips of the biased coin required to extract one unbiased coin flip. Each round of the scheme uses exactly 2 flips of the biased coin. The probability that a single round produces an output is $2p(1-p)$, so the probability that it does not produce an output is $1 - 2p(1-p) = p^2 + (1-p)^2$.

Let $N$ denote the number of rounds required until we obtain an output. Then $N$ follows a geometric distribution with success probability $2p(1-p)$, so
\[
\mathbb{E}[N] = \frac{1}{2p(1-p)}.
\]

Since each round uses 2 flips, the expected number of flips of the biased coin is
\[
\mathbb{E}[\text{number of flips}] = 2\mathbb{E}[N] = \frac{2}{2p(1-p)} = \frac{1}{p(1-p)}.
\]

This completes the proof of part (a).

\textbf{Proof of part (b):}

Given a sequence of $n$ flips of the biased coin (where $n$ is even for simplicity), we can group these flips into $n/2$ consecutive pairs: (flip 1, flip 2), (flip 3, flip 4), $\ldots$, (flip $n-1$, flip $n$).

Since the flips of the biased coin are independent, these pairs are independent. We apply the scheme from part (a) to each pair independently. For each pair, we either obtain an unbiased coin flip (if the pair is HT or TH) or we obtain no output (if the pair is HH or TT).

By the independence of the pairs and the analysis in part (a), each output generated is an independent unbiased coin flip with $\Pr[\text{HEADS}] = \Pr[\text{TAILS}] = 1/2$. The number of unbiased coin flips obtained is a random variable equal to the number of pairs that result in either HT or TH.

For $n$ flips of the biased coin grouped into $n/2$ pairs, the expected number of unbiased coin flips extracted is
\[
\mathbb{E}[\text{number of outputs}] = \frac{n}{2} \cdot 2p(1-p) = np(1-p).
\]

This scheme maximizes the extraction of unbiased randomness because it processes all available flips and uses the von Neumann scheme, which is optimal in the sense that it extracts unbiased bits whenever consecutive flips differ, an event that occurs with the maximum symmetric probability $2p(1-p)$ for pairs.

This completes the proof of part (b).
\end{proof}

\subsection{Problem 2}\label{sec:problem_01_02}

\subsubsection{Problem Statement}




\begin{theorem}
(a) Given a source of unbiased random bits and a set $S=\{0,\ldots,n-1\}$, a method exists that uses this randomness to generate uniform samples from $S$, and the expected number of random bits required by this sampling algorithm is well defined.

(b) The worst-case number of random bits required by this sampling algorithm is determined both when $n$ is a power of $2$ and when $n$ is not a power of $2$.

(c) The statements in parts (a) and (b) remain valid when the source of randomness consists of uniform random samples from the set $\{0,\ldots,\rho-1\}$ instead of unbiased random bits, and the expected and worst-case numbers of such random samples are determined both when $n$ is a power of $\rho$ and when $n$ is not.
\end{theorem}

\subsubsection{GPT} 

\begin{proof}
Fix $n\ge 1$ and write $S=\{0,1,\dots,n-1\}$. Consider the following bit-driven procedure, which is a standard instance of the Knuth--Yao scheme (equivalently, the fast dice roller): start with $(v,r)=(0,1)$. At each step draw an unbiased bit $B\in\{0,1\}$, set $(v,r)\leftarrow(2v+B,2r)$. Whenever $r\ge n$, if $v<n$ then output $v$ and halt; otherwise replace $(v,r)$ by $(v-n,r-n)$ and continue. The state space is $\{(v,r): r\in\mathbb{N},\,0\le v\le r-1\}$.

Correctness follows from the invariant that, conditional on the current value of $r$, the variable $v$ is uniform on $\{0,1,\dots,r-1\}$ and independent of the future bits. The base case holds at $(v,r)=(0,1)$. The update $(v,r)\mapsto(2v+B,2r)$ preserves uniformity because if $v$ is uniform on $\{0,\dots,r-1\}$ and $B$ is an independent unbiased bit, then $2v+B$ is uniform on $\{0,\dots,2r-1\}$. When $r\ge n$ and $v\ge n$, the replacement $(v,r)\mapsto(v-n,r-n)$ is a translation within the set $\{n,\dots,r-1\}$ to $\{0,\dots,(r-n)-1\}$ and thus preserves uniformity on the new range. Consequently, whenever the algorithm halts at some time with parameter $r\ge n$, the conditional distribution of the output is uniform on $\{0,\dots,n-1\}$, hence the marginal output is uniform on $S$.

For the complexity analysis, let $T$ denote the number of unbiased bits read until halting. It is classical in the Knuth--Yao framework that the optimal expected cost to sample from a discrete distribution $\{p_i\}$ using unbiased bits equals
\[
\mathbb{E}[T]=\sum_{k\ge 0}\Bigl(1-\frac{1}{2^k}\sum_i \big\lfloor 2^k p_i\big\rfloor\Bigr).
\]
For the uniform distribution $p_i=1/n$, this becomes
\[
\mathbb{E}[T]=\sum_{k\ge 0}\Bigl(1-\frac{n\big\lfloor 2^k/n\big\rfloor}{2^k}\Bigr)=\sum_{k\ge 0}\frac{r_k}{2^k},
\]
where $2^k=n\big\lfloor 2^k/n\big\rfloor+r_k$ with $0\le r_k<n$. Let $m=\lceil\log_2 n\rceil$. For $0\le k\le m-1$ we have $\lfloor 2^k/n\rfloor=0$ and hence $r_k=2^k$, so the contribution equals $1$ for each such $k$; thus $\sum_{k=0}^{m-1} r_k/2^k=m$. For $k\ge m$ we have $0\le r_k<n$, so
\[
\sum_{k\ge m}\frac{r_k}{2^k}\le \sum_{k\ge m}\frac{n}{2^k}=\frac{n}{2^m}\cdot\frac{1}{1-1/2}\le 2,
\]
because $n\le 2^m$. Therefore $\mathbb{E}[T]\le m+2=\lceil\log_2 n\rceil+2$, proving the upper bound in part (a). Since the distribution of the output is uniform by the invariant argument and the expected cost is finite, the algorithm halts almost surely.

For the worst-case claims in part (b), if $n=2^k$ then at the $k$-th doubling we have $r=2^k=n$ and $v$ is uniform on $\{0,\dots,n-1\}$, so the algorithm halts deterministically after reading exactly $k=\log_2 n$ bits; this also follows from the rejection view in which one reads exactly $k$ bits to obtain a number $X\in\{0,\dots,2^k-1\}$ and accepts immediately since $2^k=n$. If $n$ is not a power of two, the natural rejection variant that reads $m=\lceil\log_2 n\rceil$ bits per trial accepts with probability $n/2^m\in(1/2,1)$ and otherwise discards and retries independently; any single sample then requires $m$ bits per trial but the number of trials is unbounded, so there is no finite worst-case bound for the total number of bits before acceptance. Equivalently, in the $(v,r)$ formulation the process may, with arbitrarily small but positive probability, avoid the acceptance region $\{(v,r): r\ge n,\,v<n\}$ for an arbitrarily long number of steps, so the supremum of $T$ is infinite, even though $\mathbb{E}[T]$ is finite.

For part (c), suppose now that instead of unbiased bits we have an i.i.d. source of digits uniform in $\{0,1,\dots,p-1\}$ for some integer $p\ge 2$. The same construction applies with the $p$-ary update $(v,r)\leftarrow(pv+B,p r)$, where now $B$ is uniform on $\{0,\dots,p-1\}$, and with the same acceptance and recycling rules when $r\ge n$. The invariant that conditional on $r$ the variable $v$ is uniform on $\{0,\dots,r-1\}$ is preserved by the $p$-ary dilation and by the translation $(v,r)\mapsto(v-n,r-n)$, whence the output is uniform on $S$. The Knuth--Yao expected-cost formula for a $p$-ary source becomes
\[
\mathbb{E}[T_p]=\sum_{k\ge 0}\Bigl(1-\frac{1}{p^k}\sum_i \big\lfloor p^k p_i\big\rfloor\Bigr),
\]
and for the uniform law $p_i=1/n$ we obtain
\[
\mathbb{E}[T_p]=\sum_{k\ge 0}\frac{s_k}{p^k},\qquad p^k=n\big\lfloor p^k/n\big\rfloor+s_k,\quad 0\le s_k<n.
\]
Let $m=\lceil\log_p n\rceil$. For $0\le k\le m-1$ we have $\lfloor p^k/n\rfloor=0$, hence $s_k=p^k$ and the partial sum contributes $m$. For $k\ge m$, since $s_k<n$, we have
\[
\sum_{k\ge m}\frac{s_k}{p^k}\le \sum_{k\ge m}\frac{n}{p^k}=\frac{n}{p^m}\cdot\frac{1}{1-1/p}\le \frac{p}{p-1}.
\]
Therefore $\mathbb{E}[T_p]\le \lceil\log_p n\rceil+\frac{p}{p-1}=\lceil\log_p n\rceil+O(1)$, which proves the claimed expected-value bound. When $n=p^k$ the algorithm halts deterministically at the first time $r=p^k$, after exactly $k=\log_p n$ samples, giving the exact worst-case and exact expected cost. When $n$ is not a power of $p$, the same reasoning as in the binary case shows that although each $p$-ary trial of the rejection scheme uses exactly $\lceil\log_p n\rceil$ samples, the number of trials before acceptance is unbounded, so there is no finite worst-case bound for a single sample; the $(v,r)$ recycling algorithm halts almost surely with the above finite expectation.

Combining the correctness and complexity arguments establishes all three parts.
\end{proof}

\subsubsection{Grok}

(a)Consider the following scheme to generate an unbiased bit using tosses of the biased coin with bias $p$. Repeat the following until an output is produced: toss the coin twice to obtain outcomes $X$ and $Y$, where $X,Y \sim \mathrm{Bern}(p)$ independently. If $X = Y$, repeat the process. If $X \neq Y$, output the value of $X$.

The output is unbiased because
\[
\Pr[X=1 \mid X \neq Y] = \frac{\Pr[X=1,Y=0]}{\Pr[X \neq Y]} = \frac{p(1-p)}{2p(1-p)} = \frac{1}{2},
\]
and similarly $\Pr[X=0 \mid X \neq Y] = 1/2$.

The trials are independent, so repeating the procedure produces independent unbiased bits.

The probability of producing an output in one pair of tosses is $2p(1-p)$. The number of pairs needed is geometric with success probability $2p(1-p)$, so the expected number of pairs is $1/[2p(1-p)]$. Since each pair uses $2$ tosses, the expected number of biased tosses per unbiased bit is $2 \cdot 1/[2p(1-p)] = 1/[p(1-p)]$.

(b)We construct a recursive extension to extract unbiased bits from a fixed sequence of $n$ biased tosses $s = (X_1, \dots, X_n)$ where $X_i \sim \mathrm{Bern}(p)$ independently.

Define the extraction function $E(s)$ recursively. If $|s| < 2$, then $E(s)$ is the empty string. Otherwise, let $m = \lfloor n/2 \rfloor$. Initialize empty strings $U$ and $V$. For $j=1$ to $m$, let $A = X_{2j-1}$ and $B = X_{2j}$. If $A \neq B$, append $A$ to $U$. If $A = B$, append $A$ to $V$. If $n$ is odd, append $X_n$ to $V$. Set $E(s) = U$ concatenated with $E(V)$.

Each bit appended to $U$ is unbiased and independent by the same conditioning argument as in (a), and independent of the recursive call on $V$. Thus, $E(s)$ consists of independent unbiased bits.

This extension extracts the maximum possible number because the asymptotic rate $y(p) = \lim_{n \to \infty} \mathbb{E}[|E(s)|]/n$ satisfies the functional equation
\[
y(p) = 2p(1-p) + [p^2 + (1-p)^2] \, y\left( \frac{p^2}{p^2 + (1-p)^2} \right),
\]
with fixed point $y(p) = H(p) = -p \log_2 p - (1-p) \log_2 (1-p)$, the information-theoretic upper bound on the extractable entropy. Iterating the equation converges to $H(p)$, showing the scheme achieves the maximum asymptotically.

\subsubsection{Gemini} 

\begin{proof}
\textbf{Part (a).} Let $k = \lceil \log_2 n \rceil$. The algorithm generates a candidate integer $x$ by selecting $k$ random bits. This is equivalent to selecting $x$ uniformly at random from the set $M = \{0, 1, \ldots, 2^k - 1\}$. The size of this sample space is $|M| = 2^k$. The algorithm accepts $x$ if $x \in \{0, 1, \ldots, n-1\}$ and rejects otherwise.

Let $E$ be the event that a generated number $x$ is accepted. The probability of success in a single trial is given by
\[
P(E) = \frac{|\{0, \ldots, n-1\}|}{|M|} = \frac{n}{2^k}.
\]
Since the random bits for each attempt are generated independently, the trials constitute a sequence of Bernoulli trials with success probability $p = n/2^k$. Let $T$ be the random variable representing the number of trials required to obtain the first success. The variable $T$ follows a geometric distribution with parameter $p$. The expected number of trials is
\[
\mathbb{E}[T] = \frac{1}{p} = \frac{2^k}{n}.
\]
In each trial, exactly $k = \lceil \log_2 n \rceil$ bits are consumed. Let $B$ be the random variable for the total number of bits generated. Then $B = k \cdot T$. By the linearity of expectation, the expected number of bits is
\[
\mathbb{E}[B] = k \cdot \mathbb{E}[T] = \lceil \log_2 n \rceil \cdot \frac{2^{\lceil \log_2 n \rceil}}{n} = \frac{\lceil \log_2 n \rceil}{n / 2^{\lceil \log_2 n \rceil}}.
\]
This establishes the equality provided in the theorem statement. The upper bound $\lceil \log_2 n \rceil + 2$ follows from the properties of the rejection sampling efficiency relative to the optimal entropy bound for specific values of $n$.

\textbf{Part (b).} We analyze the worst-case behavior based on the value of $n$.
First, consider the case where $n = 2^k$ for some integer $k$. Then $\lceil \log_2 n \rceil = k$. The range of the generated number $x$ is $\{0, \ldots, 2^k - 1\}$, which coincides exactly with the target set $S = \{0, \ldots, n-1\}$. The probability of acceptance is $P(x < n) = n/2^k = 1$. Thus, the algorithm always accepts on the first trial. The number of bits used is exactly $k = \log_2 n$, which is bounded and deterministic.

Second, consider the case where $n$ is not a power of 2. Let $k = \lceil \log_2 n \rceil$. Then $n < 2^k$, implying the probability of acceptance $p = n/2^k$ is strictly less than 1. The probability of rejection in a single trial is $q = 1 - p > 0$. The probability that the algorithm fails to terminate after $m$ trials is $q^m$. Since $q > 0$, $q^m > 0$ for any finite integer $m$. Therefore, there is no finite upper bound on the number of trials required to obtain a success. The worst-case number of bits is unbounded.

\textbf{Part (c).} We generalize the source to uniform samples from $\{0, \ldots, p-1\}$.
Assume $n = p^k$ for some integer $k$. Any integer $y \in \{0, \ldots, n-1\}$ can be uniquely represented in base $p$ as $y = \sum_{i=0}^{k-1} d_i p^i$, where $d_i \in \{0, \ldots, p-1\}$. By generating $k$ independent samples $d_0, \ldots, d_{k-1}$ from the source, we construct $y$ directly. Since the mapping is a bijection, $y$ is uniformly distributed in $S$. This process is deterministic and requires exactly $k = \log_p n$ samples.

Assume $n$ is not a power of $p$. We employ rejection sampling. Let $k = \lceil \log_p n \rceil$. We generate $k$ samples to form a number $x \in \{0, \ldots, p^k - 1\}$. We accept $x$ if $x < n$. The probability of acceptance is $n/p^k$. Following the logic of part (a), the expected number of samples is
\[
\mathbb{E}[\text{samples}] = \frac{k}{n/p^k} = \frac{\lceil \log_p n \rceil}{n / p^{\lceil \log_p n \rceil}}.
\]
This is bounded by $\lceil \log_p n \rceil + O(1)$. Regarding the worst case, since $n < p^k$, the probability of rejection is positive. Similar to part (b), the rejection sampling process may theoretically repeat indefinitely, making the worst-case number of samples unbounded.
\end{proof}

\subsubsection{Claude} 

\begin{proof}
We provide a complete solution to all parts of this problem.

\textbf{Part (a): Algorithm and Expected Number of Bits}

We describe the rejection sampling algorithm for generating uniform samples from $S = \{0, \ldots, n-1\}$ using unbiased random bits.

Let $k = \lceil \log_2 n \rceil$ be the smallest integer such that $2^k \geq n$. The algorithm proceeds as follows:

\textbf{Algorithm:} Generate $k$ independent random bits $b_{k-1}, \ldots, b_1, b_0$ and interpret them as the binary representation of an integer $X = \sum_{i=0}^{k-1} b_i 2^i$. If $X < n$, output $X$. Otherwise, repeat the process.

To verify this produces a uniform distribution, note that in each trial, we generate an integer uniformly from $\{0, \ldots, 2^k - 1\}$. Conditional on $X < n$, the value $X$ is uniformly distributed over $\{0, \ldots, n-1\}$ since each of these values has equal probability $1/2^k$ of being generated, and we condition on the event that $X$ is in this range.

For the expected number of bits, let $T$ denote the number of trials until acceptance. Each trial succeeds with probability $p = n/2^k$, so $T$ follows a geometric distribution with parameter $p$. Thus $\mathbb{E}[T] = 1/p = 2^k/n$.

The total number of bits used is $kT$, so the expected number of bits is:
\[
\mathbb{E}[\text{bits}] = k \cdot \mathbb{E}[T] = k \cdot \frac{2^k}{n} = \frac{k \cdot 2^k}{n}
\]

Since $k = \lceil \log_2 n \rceil$, we have $2^{k-1} < n \leq 2^k$, which gives $2^k < 2n$. Therefore:
\[
\mathbb{E}[\text{bits}] = \frac{k \cdot 2^k}{n} < \frac{2k \cdot n}{n} = 2k = 2\lceil \log_2 n \rceil
\]

More precisely, since $\lceil \log_2 n \rceil \leq \log_2 n + 1$, we have:
\[
\mathbb{E}[\text{bits}] \leq 2(\log_2 n + 1) = 2\log_2 n + 2
\]

\textbf{Part (b): Worst-Case Number of Bits}

The worst-case number of bits is unbounded (infinite) in general, since the rejection sampling algorithm may theoretically require arbitrarily many trials before accepting a sample.

\textbf{Case 1: $n = 2^m$ for some integer $m$.}

When $n$ is a power of 2, we have $k = \lceil \log_2 n \rceil = m = \log_2 n$. In this case, $2^k = n$, so every generated integer $X \in \{0, \ldots, 2^k - 1\}$ satisfies $X < n$. The algorithm never rejects, and the worst-case number of bits is exactly $k = \log_2 n$ bits.

\textbf{Case 2: $n$ is not a power of 2.}

When $n$ is not a power of 2, we have $2^{k-1} < n < 2^k$. Each trial rejects with positive probability $q = (2^k - n)/2^k > 0$. The number of trials $T$ follows a geometric distribution, which has infinite support. Therefore, the worst-case number of bits is infinite.

\textbf{Part (c): Using Base-$p$ Digits Instead of Bits}

Suppose we have uniform random samples from $\{0, \ldots, p-1\}$ (base-$p$ digits).

\textbf{Algorithm:} Let $k = \lceil \log_p n \rceil$ be the smallest integer such that $p^k \geq n$. Generate $k$ independent random digits $d_{k-1}, \ldots, d_1, d_0$ from $\{0, \ldots, p-1\}$ and interpret them as $X = \sum_{i=0}^{k-1} d_i p^i$. If $X < n$, output $X$. Otherwise, repeat.

\textbf{Expected Number of Digits:}

The probability of acceptance in each trial is $n/p^k$. The expected number of trials is $\mathbb{E}[T] = p^k/n$, so the expected number of digits is:
\[
\mathbb{E}[\text{digits}] = k \cdot \frac{p^k}{n}
\]

Since $p^{k-1} < n \leq p^k$, we have $p^k < pn$, giving:
\[
\mathbb{E}[\text{digits}] = \frac{k \cdot p^k}{n} < \frac{pk \cdot n}{n} = pk = p\lceil \log_p n \rceil
\]

\textbf{Worst-Case Number of Digits:}

\textbf{Case 1: $n = p^m$ for some integer $m$.}

When $n$ is a power of $p$, we have $k = m = \log_p n$. Every generated value satisfies $X < n$, so the algorithm never rejects. The worst-case number of digits is exactly $k = \log_p n$ digits.

\textbf{Case 2: $n$ is not a power of $p$.}

When $n$ is not a power of $p$, we have $p^{k-1} < n < p^k$, so rejection occurs with positive probability. The worst-case number of digits is infinite.

This completes the solution to all parts of the problem.
\end{proof}

\subsection{Problem 3}\label{sec:problem_01_03}

\subsubsection{Problem Statement}

\begin{theorem}
 Suppose you are provided with a source of unbiased random bits. Efficient schemes (in terms of expected running time and expected number of random bits used) exist for generating samples from the distribution over the set $\{2,3,\ldots,12\}$ induced by rolling two unbiased dice and taking the sum of their outcomes.
\end{theorem}

\subsubsection{GPT} 

\begin{proof}
Let $(B_i)_{i\ge 1}$ be an i.i.d.\ sequence of unbiased $\{0,1\}$-valued random variables (random bits). We construct a sampler for a single fair die outcome $D\in\{1,\dots,6\}$ using only these bits, and then take two independent copies to obtain $X=D^{(1)}+D^{(2)}$.

Define a procedure that, in one \emph{trial}, reads three fresh bits $b_2,b_1,b_0$, forms the integer $Y=4b_2+2b_1+b_0\in\{0,\dots,7\}$, and accepts if and only if $Y\le 5$, in which case it outputs $D=Y+1$; otherwise it rejects and starts a new trial with three new bits, independent of the past. Since the three bits are unbiased and independent, $Y$ is uniform on $\{0,\dots,7\}$ and hence the per-trial acceptance probability is $p=\mathbb{P}(Y\le 5)=6/8=3/4$. Let $T$ be the (a.s.\ finite) number of trials until the first acceptance. Then $T$ is geometric with parameter $p$, so $\mathbb{P}(T=t)=(1-p)^{t-1}p$ for $t\ge 1$ and $\mathbb{E}[T]=1/p=4/3$.

Correctness of the output distribution for one die is verified as follows. For each $i\in\{1,\dots,6\}$ and each $t\ge 1$,
\[
\mathbb{P}\bigl(\text{output}=i,\ T=t\bigr)=\mathbb{P}(Y_t=i-1)\prod_{s=1}^{t-1}\mathbb{P}(Y_s\in\{6,7\})=\frac{1}{8}\left(\frac{2}{8}\right)^{t-1},
\]
by independence of trials and uniformity of each $Y_s$. Summing over $t$ gives
\[
\mathbb{P}(\text{output}=i)=\sum_{t=1}^\infty \frac{1}{8}\left(\frac{1}{4}\right)^{t-1}=\frac{1}{8}\cdot\frac{1}{1-1/4}=\frac{1}{6},
\]
so the output is uniform on $\{1,\dots,6\}$. The number of random bits used by the single-die sampler equals $3T$, hence $\mathbb{E}[\text{bits for one die}]=3\mathbb{E}[T]=3\cdot\frac{4}{3}=4<\infty$. The expected running time is proportional to $\mathbb{E}[T]$, thus finite and bounded by a constant.

To sample the sum of two unbiased dice, run two independent copies of the above procedure using disjoint blocks of bits to obtain $D^{(1)},D^{(2)}\in\{1,\dots,6\}$, which are independent and each uniform on $\{1,\dots,6\}$. Set $X=D^{(1)}+D^{(2)}\in\{2,\dots,12\}$. For $k\in\{2,\dots,12\}$,
\[
\mathbb{P}(X=k)=\sum_{i=1}^6 \mathbb{P}\bigl(D^{(1)}=i,\ D^{(2)}=k-i\bigr)=\sum_{i=1}^6 \mathbb{P}(D^{(1)}=i)\,\mathbb{P}(D^{(2)}=k-i),
\]
where terms with $k-i\notin\{1,\dots,6\}$ vanish. Since each die is uniform, every admissible pair contributes $1/36$, and the number of admissible pairs equals $\min(k-1,13-k)$. Hence $\mathbb{P}(X=k)=\min(k-1,13-k)/36$, as stated.

The total number of random bits used is the sum of the bits used by the two independent die samplers, so its expectation equals $4+4=8<\infty$, a constant independent of the realized value of $X$. Likewise, the total running time has finite expectation bounded by a constant. Therefore there exists an algorithm, using only unbiased random bits, that samples exactly from the distribution of $X$ with expected running time $O(1)$ and with an expected number of random bits equal to $8$, which is finite and independent of the sample produced. This completes the proof.
\end{proof}

\subsubsection{Grok} 

\begin{proof}
We prove the correctness and efficiency claims for both schemes.

\textbf{Scheme 1: Independent Dice Simulation}

To establish correctness, first verify that a single die roll $d$ generated via Knuth's algorithm satisfies $\Pr(d = k) = \frac{1}{6}$ for each $k \in \{1, 2, \dots, 6\}$. Define $u_s(j)$ as the probability that the eventual output is $j \in \{1, 2, \dots, 6\}$, conditional on starting in state $s \in \{0, 1, \dots, 6\}$.

The relevant equations are as follows. For $s = 4$,
\[
u_4(2) = \frac{1}{2}, \quad u_4(3) = \frac{1}{2}, \quad u_4(j) = 0 \quad \text{for } j \notin \{2, 3\}.
\]
For $s = 5$,
\[
u_5(4) = \frac{1}{2}, \quad u_5(5) = \frac{1}{2}, \quad u_5(j) = 0 \quad \text{for } j \notin \{4, 5\}.
\]
For $s = 3$,
\[
u_3(j) = \frac{1}{2} \cdot \mathbf{1}_{\{j=1\}} + \frac{1}{2} u_1(j), \quad j = 1, \dots, 6.
\]
For $s = 6$,
\[
u_6(j) = \frac{1}{2} \cdot \mathbf{1}_{\{j=6\}} + \frac{1}{2} u_2(j), \quad j = 1, \dots, 6.
\]
For $s = 1$,
\[
u_1(j) = \frac{1}{2} u_3(j) + \frac{1}{2} u_4(j), \quad j = 1, \dots, 6.
\]
For $s = 2$,
\[
u_2(j) = \frac{1}{2} u_5(j) + \frac{1}{2} u_6(j), \quad j = 1, \dots, 6.
\]
For $s = 0$,
\[
u_0(j) = \frac{1}{2} u_1(j) + \frac{1}{2} u_2(j), \quad j = 1, \dots, 6.
\]

Consider the subsystem for states $\{1, 3, 4\}$ and outcomes $j \in \{1, 2, 3\}$ (noting $u_1(j) = u_3(j) = u_4(j) = 0$ for $j > 3$). For each $j \in \{1, 2, 3\}$,
\[
u_3(j) = \frac{1}{2} \mathbf{1}_{\{j=1\}} + \frac{1}{2} u_1(j), \quad u_1(j) = \frac{1}{2} u_3(j) + \frac{1}{2} u_4(j).
\]
For $j=1$, $u_4(1) = 0$, so
\[
u_1(1) = \frac{1}{2} u_3(1), \quad u_3(1) = \frac{1}{2} + \frac{1}{2} u_1(1).
\]
Substituting yields
\[
u_1(1) = \frac{1}{2} \left( \frac{1}{2} + \frac{1}{2} u_1(1) \right) = \frac{1}{4} + \frac{1}{4} u_1(1) \implies \frac{3}{4} u_1(1) = \frac{1}{4} \implies u_1(1) = \frac{1}{3}.
\]
Then $u_3(1) = \frac{1}{2} + \frac{1}{2} \cdot \frac{1}{3} = \frac{2}{3}$. For $j=2$, $u_4(2) = \frac{1}{2}$ and $\mathbf{1}_{\{j=1\}} = 0$, so
\[
u_3(2) = \frac{1}{2} u_1(2), \quad u_1(2) = \frac{1}{2} u_3(2) + \frac{1}{4} = \frac{1}{4} u_1(2) + \frac{1}{4} \implies u_1(2) = \frac{1}{3}.
\]
The case $j=3$ follows identically, yielding $u_1(3) = \frac{1}{3}$. Thus, starting from $s=1$, $\Pr(d \in \{1,2,3\}) = 1$ and $\Pr(d = j \mid s=1) = \frac{1}{3}$ for $j=1,2,3$.

By symmetry, for states $\{2, 5, 6\}$ and $j \in \{4,5,6\}$, we obtain $u_2(j) = \frac{1}{3}$ for $j=4,5,6$ and $u_2(j) = 0$ otherwise.

Finally,
\[
u_0(j) = \frac{1}{2} u_1(j) + \frac{1}{2} u_2(j) = \begin{cases}
\frac{1}{2} \cdot \frac{1}{3} = \frac{1}{6} & j=1,2,3, \\
\frac{1}{2} \cdot \frac{1}{3} = \frac{1}{6} & j=4,5,6.
\end{cases}
\]
Hence, $d \sim \mathrm{Unif}\{1, \dots, 6\}$.

For two independent rolls $d_1, d_2$, the joint distribution is the product measure, so
\[
\Pr(d_1 + d_2 = k) = \sum_{i=\max(1, k-6)}^{\min(6, k-1)} \Pr(d_1 = i) \Pr(d_2 = k-i) = \frac{1}{36} \cdot |\{(i,j) : i+j=k, \, 1 \le i,j \le 6\}| = \frac{n_k}{36},
\]
where $n_k = k-1$ for $k=2,\dots,7$ and $n_k = 13-k$ for $k=8,\dots,12$, as required.

For efficiency, let $E_s$ denote the expected number of coin flips starting from state $s$, with $E_7 = 0$. The equations are
\[
E_4 = 1, \quad E_5 = 1, \quad E_3 = 1 + \frac{1}{2} E_1, \quad E_6 = 1 + \frac{1}{2} E_2,
\]
\[
E_1 = 1 + \frac{1}{2} E_3 + \frac{1}{2} E_4 = \frac{3}{2} + \frac{1}{2} E_3, \quad E_2 = 1 + \frac{1}{2} E_5 + \frac{1}{2} E_6 = \frac{3}{2} + \frac{1}{2} E_6,
\]
\[
E_0 = 1 + \frac{1}{2} E_1 + \frac{1}{2} E_2.
\]
Substituting into $E_1$ gives $E_1 = \frac{3}{2} + \frac{1}{2} (1 + \frac{1}{2} E_1) = 2 + \frac{1}{4} E_1 \implies \frac{3}{4} E_1 = 2 \implies E_1 = \frac{8}{3}$. Then $E_3 = 1 + \frac{1}{2} \cdot \frac{8}{3} = \frac{7}{3}$. By symmetry, $E_2 = \frac{8}{3}$ and $E_6 = \frac{7}{3}$. Thus,
\[
E_0 = 1 + \frac{8}{3} = \frac{11}{3}.
\]
For two dice, the expected bits total $2 \cdot \frac{11}{3} = \frac{22}{3}$. Each flip is $O(1)$ time, so the expected running time is $O(1)$.

\textbf{Scheme 2: Direct Sum Generation}

The Yao algorithm (via Knuth-Yao dynamic programming) constructs a binary decision tree for the sum distribution $\{p_k = n_k / 36 : k=2,\dots,12\}$ by recursively partitioning the unit interval according to the cumulative distribution function $F(x) = \sum_{m=2}^{\lfloor x \rfloor} p_m$, splitting at dyadic rationals (halving intervals) until each subinterval lies entirely within one atomic mass $p_k$. Leaves are labeled with the corresponding $k$, ensuring that the probability of paths to leaves labeled $k$ sums exactly to $p_k$, as the measure of intervals assigned to $k$ totals $p_k$. Thus, $\Pr(d = k) = p_k$ holds by construction.

The tree has 35 transient states (minimized via dynamic programming to balance splits), and the expected number of bits $E_0$ satisfies the system
\[
E_s = 1 + \frac{1}{2} E_{\mathrm{next}_0(s)} + \frac{1}{2} E_{\mathrm{next}_1(s)}, \quad s = 0, \dots, 33, \quad E_{34} = 0,
\]
where transitions follow the constructed tree. Solving this linear system yields $E_0 = \frac{79}{18}$. Each step is $O(1)$ time, so the expected running time is $O(1)$.

\textbf{Comparison}

The naive rejection method (3 bits per die, accepting 6/8 outcomes) requires expected $4$ bits per die (geometric with success probability $3/4$), hence $8$ bits total for the sum. Scheme 1 uses $\frac{22}{3} \approx 7.333$ bits, and Scheme 2 uses $\frac{79}{18} \approx 4.389$ bits. The entropy lower bound is
\[
H = -\sum_{k=2}^{12} p_k \log_2 p_k \approx 3.274,
\]
so Scheme 2 is near-optimal, exceeding $H$ by $\approx 1.115$ bits, while Scheme 1 exceeds it by $\approx 4.059$ bits due to independence overhead.

\end{proof}

\subsubsection{Gemini} 

\begin{proof}
Let $\mathcal{X} = \{2, 3, \ldots, 12\}$ be the support of the random variable $X$. The probability mass function is given by $p_k = \mathbb{P}(X=k)$. We verify that $\sum_{k \in \mathcal{X}} p_k = 1$. Calculating the sum, we have
\[
\sum_{k=2}^{7} \frac{k-1}{36} + \sum_{k=8}^{12} \frac{13-k}{36} = \frac{1+2+3+4+5+6}{36} + \frac{5+4+3+2+1}{36} = \frac{21 + 15}{36} = 1.
\]
We utilize the method of Knuth and Yao to generate $X$ from a sequence of independent unbiased random bits $B_1, B_2, \ldots$ where $B_i \in \{0, 1\}$ with probability $1/2$. For each $k \in \mathcal{X}$, we write the probability $p_k$ as a binary expansion:
\[
p_k = \sum_{j=1}^{\infty} b_{k,j} 2^{-j}, \quad \text{where } b_{k,j} \in \{0, 1\}.
\]
Since $\sum_{k \in \mathcal{X}} p_k = 1$, we have $\sum_{k \in \mathcal{X}} \sum_{j=1}^{\infty} b_{k,j} 2^{-j} = 1$. This equality allows for the construction of a binary tree, known as the DDG (Discrete Distribution Generating) tree, defined as follows:
The root is at depth 0. From any internal node at depth $j-1$, we branch left if $B_j = 0$ and right if $B_j = 1$. At depth $j$, we designate exactly $b_{k,j}$ nodes as terminal leaves labeled with the value $k$, for each $k \in \mathcal{X}$. The remaining nodes at depth $j$ (if any) continue as internal nodes to depth $j+1$. The condition $\sum p_k = 1$ ensures that the probability of the path continuing indefinitely is 0, and the set of leaves forms a complete prefix code (possibly infinite).

The algorithm proceeds by reading bits $B_1, B_2, \ldots$ to traverse the tree starting from the root until a leaf labeled $k$ is reached, at which point it returns $k$. The probability of reaching a specific leaf at depth $j$ is $2^{-j}$. Since there are $b_{k,j}$ leaves labeled $k$ at depth $j$, the total probability of returning $k$ is $\sum_{j=1}^{\infty} b_{k,j} 2^{-j} = p_k$, as required.

Let $N$ be the random variable representing the number of bits consumed (the depth of the reached leaf). The expected number of bits is given by
\[
\mathbb{E}[N] = \sum_{k \in \mathcal{X}} \sum_{j=1}^{\infty} j \cdot b_{k,j} 2^{-j}.
\]
A fundamental result by Knuth and Yao states that for any discrete random variable $X$ with entropy $H(X)$, the expected number of bits $\mathbb{E}[N]$ required by the optimal DDG tree satisfies
\[
H(X) \leq \mathbb{E}[N] < H(X) + 2.
\]
The entropy of $X$ is defined as $H(X) = -\sum_{k \in \mathcal{X}} p_k \log_2 p_k$. Since the support $\mathcal{X}$ has cardinality $|\mathcal{X}| = 11$, the entropy is bounded by the logarithm of the cardinality:
\[
H(X) \leq \log_2(11) < \log_2(16) = 4.
\]
Consequently, the expected number of bits is bounded by
\[
\mathbb{E}[N] < 4 + 2 = 6.
\]
Thus, $\mathbb{E}[N] = O(1)$. Since the algorithm performs a constant amount of work per bit processed (traversing one edge of the tree), the expected running time is proportional to the expected number of bits. Therefore, the expected running time is $O(1)$.
\end{proof}

\subsubsection{Claude} 

\begin{proof}
We establish the existence of an $O(1)$ expected time and $O(1)$ expected bit complexity algorithm for sampling from the distribution of $X$ using the Knuth-Yao method, which is based on the alias method and implicit rejection sampling via the binary expansion of probabilities.

First, observe that the probability mass function is well-defined: 
$$\sum_{k=2}^{12} P(X=k) = \sum_{k=2}^{7} \frac{k-1}{36} + \sum_{k=8}^{12} \frac{13-k}{36} = \frac{1+2+3+4+5+6}{36} + \frac{5+4+3+2+1}{36} = \frac{36}{36} = 1.$$

The Knuth-Yao method constructs a discrete probability tree where each outcome $k \in \{2, 3, \ldots, 12\}$ is assigned a binary codeword such that the length of the codeword for outcome $k$ is $\ell_k = \lceil -\log_2 P(X=k) \rceil$. By the Kraft inequality, since $\sum_{k=2}^{12} 2^{-\ell_k} \leq \sum_{k=2}^{12} P(X=k) = 1$, such a prefix-free code exists.

To execute the algorithm: (1) Initialize an empty binary string $s$. (2) Repeatedly sample independent unbiased random bits and append them to $s$. (3) At each step, check whether the current binary prefix $s$ corresponds to a unique outcome $k$ in the prefix-free code tree. (4) Upon reaching a leaf of the tree (a complete codeword), output the corresponding value $k$ and terminate.

Since the code is prefix-free, the algorithm terminates with probability one. The probability that the algorithm outputs outcome $k$ is exactly $P(X=k)$, which follows from the correspondence between prefix-free binary codewords and probability measures: the probability of generating the specific bit sequence encoding $k$ is $2^{-\ell_k} = 2^{-\lceil -\log_2 P(X=k) \rceil}$, and the distribution over all codewords assigned to $k$ yields $P(X=k)$.

The expected number of bits consumed is
$$\mathbb{E}[\text{bits}] = \sum_{k=2}^{12} P(X=k) \cdot \ell_k = \sum_{k=2}^{12} P(X=k) \cdot \lceil -\log_2 P(X=k) \rceil.$$

Since $P(X=k) \geq \frac{1}{36}$ for all $k$, we have $-\log_2 P(X=k) \leq \log_2 36 \approx 5.17$, so $\ell_k \leq 6$ for all $k$. Therefore,
$$\mathbb{E}[\text{bits}] \leq \sum_{k=2}^{12} P(X=k) \cdot 6 = 6.$$

The expected number of iterations is at most $\mathbb{E}[\text{bits}]$ (since each iteration generates at most one bit before termination), so $\mathbb{E}[\text{time}] = O(1)$.

Thus, the Knuth-Yao scheme produces exact samples from the distribution of $X$ with expected running time and expected bit consumption both bounded by an absolute constant.
\end{proof}

\subsection{Problem 4}\label{sec:problem_01_04}

\subsubsection{Problem Statement}

\begin{theorem}
(a) Given the requirement to generate a random permutation of size $n$ and access to a source of independent and unbiased random bits, a method exists for generating random permutations of size $n$, and the efficiency of such a method is characterized in terms of both time and number of random bits used, together with corresponding lower bounds for this task.

(b) Given the method for generating a random permutation of size $n$ in which $n$ random values $X_1,\ldots,X_n$ are chosen independently from the uniform distribution over the interval $[0,1]$ and the permutation that orders these random variables in ascending order is taken as the output, the claim that this procedure yields a random permutation and the efficiency of this scheme are well defined.

(c) Given the “lazy’’ implementation of the scheme in part (b), where the binary representation of each fraction $X_i$ is viewed as a sequence of unbiased and independent random bits, and at each stage of the sorting algorithm only as many bits of each $X_i$ are chosen as necessary to resolve all comparisons performed up to that point, and where further bits are drawn only when current prefixes do not determine the comparison outcome, tight bounds on the expected number of random bits used by this implementation exist.
\end{theorem}

\subsubsection{GPT} 

\begin{proof}
Fix $n\in\mathbb{N}$.

\textbf{(a)} Let $\mathcal{S}_n$ be the set of all permutations of $\{1,\dots,n\}$, and suppose an algorithm $A$ has access to an infinite stream of independent, unbiased random bits $(B_t)_{t\ge 1}$. Let $\tau$ be the (a.s. finite) stopping time equal to the total number of bits $A$ reads before halting and outputting a permutation $\Pi\in\mathcal{S}_n$. Since $A$ is otherwise deterministic, the output $\Pi$ is a measurable function of the read prefix $(B_1,\dots,B_\tau)$. Assume $A$ is correct and produces the \emph{uniform} distribution on $\mathcal{S}_n$.

For each $\sigma\in\mathcal{S}_n$, let $C_\sigma\subset\{0,1\}^*$ be the set of read bit-strings on which $A$ halts and outputs $\sigma$. Because reading is prefix-by-prefix and halting occurs at the first time the output is determined, the family $\{C_\sigma:\sigma\in\mathcal{S}_n\}$ is prefix-free. Write $L$ for the random codeword length $L=\tau$. The distributional requirement that $\Pi$ is uniform implies that the random code over $\mathcal{S}_n$ induced by the $C_\sigma$ has equiprobable codewords. Shannon's source coding lower bound (or, equivalently, the data processing inequality for entropy together with the prefix condition) yields
\[
\mathbb{E}[L]\;\ge\; H(\Pi)\;=\;\log_2|\mathcal{S}_n|\;=\;\log_2(n!)\,.
\]
Using Stirling's approximation $\log_2(n!)=n\log_2 n - \Theta(n)$ gives $\mathbb{E}[\tau]=\Omega(n\log n)$. If each bit-read and each elementary computation takes $\Theta(1)$ time, then any such generator requires $\Omega(\mathbb{E}[\tau])=\Omega(n\log n)$ time in expectation. This proves the claim.

\textbf{(b)} Let $X_1,\dots,X_n$ be i.i.d. $\mathrm{Unif}[0,1]$, and let $\pi$ be the permutation sorting them in nondecreasing order. Since the $X_i$ have a continuous distribution, with probability $1$ there are no ties, so the order statistics $X_{(1)}<\cdots<X_{(n)}$ are almost surely a strict total order. For any fixed $\sigma\in\mathcal{S}_n$, consider the event $E_\sigma=\{X_{\sigma(1)}<\cdots<X_{\sigma(n)}\}$. Because the joint density of $(X_1,\dots,X_n)$ is the product density $1$ on $[0,1]^n$ and is invariant under coordinate permutations, the sets $\{E_\sigma:\sigma\in\mathcal{S}_n\}$ are congruent simplices of equal Lebesgue measure that partition $[0,1]^n$ up to a null set of ties. Hence $\mathbb{P}(E_\sigma)=1/n!$ for every $\sigma$, implying that $\pi$ is uniform on $\mathcal{S}_n$. The time complexity is that of comparison sorting $n$ keys, which is $O(n\log n)$ comparisons in the comparison model, and $O(1)$ work per comparison, hence $O(n\log n)$ total time.

\textbf{(c)} Represent each $X_i$ by its infinite unbiased binary expansion $X_i=\sum_{k\ge 1} \xi_{i,k}2^{-k}$ with $\xi_{i,k}\in\{0,1\}$ independent and uniform. Consider the lazy procedure that reveals bits of these expansions on demand to resolve comparisons during a comparison-based sort.

Define the \emph{distinguishing prefix length} for $i$ by
\[
D_i \;=\; \min\bigl\{\ell\ge 1:\ \bigl(\xi_{i,1},\dots,\xi_{i,\ell}\bigr)\neq \bigl(\xi_{j,1},\dots,\xi_{j,\ell}\bigr)\text{ for all }j\neq i\bigr\}\,,
\]
with the convention that $D_i=\infty$ if no such $\ell$ exists. Since the $\xi_{i,k}$ are i.i.d. and continuous in the limit, $D_i<\infty$ almost surely for each $i$.

Two fundamental claims will imply the result.

First, a lower bound: any correct lazy algorithm must reveal at least $D_i$ bits of $X_i$ for each $i$. Indeed, if fewer than $D_i$ bits of $X_i$ were revealed, then there would exist some $j\neq i$ whose revealed prefix matches that of $i$, so their relative order would be unresolved; hence no comparison-based sorting procedure could certify a total order. Therefore, if $T$ denotes the total number of revealed bits, then almost surely
\[
T \;\ge\; \sum_{i=1}^n D_i\,,
\]
and thus $\mathbb{E}[T]\ge \sum_{i=1}^n \mathbb{E}[D_i]$.

Second, an upper bound: for any fixed comparison-based sorting algorithm, the lazy procedure never reveals more than $D_i$ bits of $X_i$ plus at most one additional bit per key. More precisely, once $X_i$ has revealed $D_i$ bits, its prefix is distinct from all others and will never need to be extended to resolve any future comparison with any $X_j$. Hence along the entire run, the total number of bits revealed satisfies almost surely
\[
T \;\le\; \sum_{i=1}^n (D_i+1)\,.
\]
Taking expectations gives $\sum_{i=1}^n \mathbb{E}[D_i]\le \mathbb{E}[T]\le \sum_{i=1}^n \mathbb{E}[D_i]+n$.

It remains to estimate $\mathbb{E}[D_i]$. Fix $i$. For $k\ge 0$, the event $\{D_i>k\}$ is that there exists some $j\neq i$ sharing the first $k$ bits with $i$. For a fixed $j$, the probability of sharing the first $k$ bits is $2^{-k}$, and by the union bound
\[
\mathbb{P}(D_i>k) \;\le\; (n-1)2^{-k}\,.
\]
Consequently,
\[
\mathbb{E}[D_i]\;=\;\sum_{k\ge 0}\mathbb{P}(D_i>k)\;\le\;\sum_{k\ge 0}\min\{1,(n-1)2^{-k}\}\;=\;O(\log n)\,.
\]
For a matching lower bound, observe that the random variable $N_k$ counting indices $j\neq i$ that share the first $k$ bits with $i$ is $\mathrm{Bin}(n-1,2^{-k})$. Thus for $k=\lfloor \log_2 n \rfloor - c$ with any fixed constant $c$, we have $\mathbb{E}[N_k]\asymp 2^c$, and by a standard Chernoff bound there exists an absolute constant $p_c\in(0,1)$ such that $\mathbb{P}(N_k\ge 1)\ge p_c$. Hence for all such $k$,
\[
\mathbb{P}(D_i>k)\;\ge\;\mathbb{P}(N_k\ge 1)\;\ge\;p_c\,,
\]
and summing over $k=0,1,\dots,\lfloor \log_2 n \rfloor - c$ yields $\mathbb{E}[D_i]\ge \sum_{k=0}^{\lfloor \log_2 n \rfloor - c} p_c=\Omega(\log n)$. Combining the upper and lower bounds gives $\mathbb{E}[D_i]=\Theta(\log n)$ uniformly in $i$.

Summing over $i$,
\[
\sum_{i=1}^n \mathbb{E}[D_i] \;=\; \Theta(n\log n)\,.
\]
By the sandwich bound on $T$ established above,
\[
\Theta(n\log n)\;\le\;\mathbb{E}[T]\;\le\;\Theta(n\log n)+n\;=\;\Theta(n\log n)\,.
\]
Therefore, the expected number of random bits used by the lazy implementation is $\Theta(n\log n)$, as claimed. Together with part (b), the comparison-based sort performs $O(n\log n)$ comparisons, so the overall time is $O(n\log n)$, and the bit complexity is $\Theta(n\log n)$.

This completes the proof of all parts.
\end{proof}

\subsubsection{Grok} 

\begin{proof}
We proceed in three parts, establishing the method, correctness, efficiency, and tight bounds as stated.

\textbf{Part (a): Method and Lower Bounds}

Consider the Fisher-Yates shuffle algorithm adapted to independent unbiased random bits. Initialize the permutation $\pi = (1, 2, \dots, n)$. For each $k = n, n-1, \dots, 2$, generate a uniform random integer $j \in \{0, 1, \dots, k-1\}$ using rejection sampling on $\lceil \log_2 k \rceil$ bits: interpret the bits as a binary number $r \in \{0, 1, \dots, 2^{\lceil \log_2 k \rceil}-1\}$; if $r < k$, set $j = r$, else resample. Then swap $\pi(k)$ with $\pi(j+1)$.

To prove correctness, we show by induction on $k$ that after the step for $k$, the prefix $\pi(1), \dots, \pi(k-1)$ is fixed and the suffix $\pi(k), \dots, \pi(n)$ is a uniform random permutation of the remaining $n-k+1$ elements. For the base case $k = n+1$, the identity is uniform over $S_n$. Assume true for $k+1$; at step $k$, the suffix from $k$ to $n$ is uniform over the remaining elements, say set $S$ with $|S| = n-k+1$. Choosing $j$ uniform in $\{0,\dots,k-1\}$ places a uniform random element of $S$ at position $k$ (since swapping $\pi(k)$ with $\pi(j+1)$ selects uniformly from the suffix positions), and the new suffix from $k+1$ to $n$ remains uniform over $S$ minus that element. Thus, by induction, at $k=2$, the full $\pi$ is uniform over $S_n$.

For time efficiency, there are $n-1$ steps, each requiring $O(1)$ swaps (constant time) and generating $j$ via rejection sampling. The number of trials per step is geometric with success probability $k / 2^{\lceil \log_2 k \rceil} \geq 1/2$, so expected $O(1)$ trials, each trial using $O(\log k) = O(\log n)$ time to read and interpret bits. Thus, total expected time is $O(n \log n)$.

For bit efficiency, at step $k$, the expected bits per trial is $\lceil \log_2 k \rceil$, and expected trials is $2^{\lceil \log_2 k \rceil}/k \leq 2$, so expected bits per step is at most $2 \lceil \log_2 k \rceil$. Summing, $\sum_{k=2}^n 2 \lceil \log_2 k \rceil = O(n \log n)$. More precisely, without the factor $2$, it approximates $\log_2(n!) \sim n \log_2 n - n \log_2 e + O(\log n)$ by Stirling's formula: $\log_2(n!) = \sum_{k=1}^n \log_2 k = \int_1^n \log_2 x \, dx + O(\log n) = n \log_2 n - n / \ln 2 + O(\log n)$.

For lower bounds, any algorithm outputting a uniform $\pi \in S_n$ must distinguish $n!$ outcomes, each with probability $1/n!$, so by information theory, the entropy $H(\pi) = \log_2(n!)$ bits are required. Since each bit provides at most $1$ bit of entropy, at least $\lceil \log_2(n!) \rceil$ bits are needed in the worst case. For expectation, by Jensen's inequality on the convex function $-\log_2 p$, $E[-\log_2 \Pr(\pi)] \geq -\log_2 E[\Pr(\pi)] = \log_2(n!)$, and since bits consumed cover this entropy, expected bits $\geq \log_2(n!) = \Omega(n \log n)$. For time, outputting $n$ elements requires $\Omega(n)$ operations. For bit operations, accumulating $\Omega(n \log n)$ entropy bits requires $\Omega(n \log n)$ bit reads/processes, even non-comparatively, as each operation handles $O(1)$ bits on average.

\textbf{Part (b): Correctness and Efficiency}

Let $X_1, \dots, X_n \stackrel{\text{iid}}{\sim} \text{Unif}[0,1]$. Define $\pi \in S_n$ such that $X_{\pi(1)} < X_{\pi(2)} < \dots < X_{\pi(n)}$, breaking ties arbitrarily (but $\Pr(X_i = X_j) = 0$ for $i \neq j$ almost surely, so ties have probability zero).

To prove uniformity, fix $\sigma \in S_n$. Then $\Pr(\pi = \sigma) = \Pr(X_{\sigma(1)} < X_{\sigma(2)} < \dots < X_{\sigma(n)})$. Since the $X_i$ are i.i.d. continuous, the joint density is $1$ on $[0,1]^n$, so
\[
\Pr(\pi = \sigma) = \int_{[0,1]^n} \mathbf{1}_{\{x_{\sigma(1)} < x_{\sigma(2)} < \dots < x_{\sigma(n)}\}} \, dx_1 \dots dx_n.
\]
By change of variables $y_k = x_{\sigma(k)}$ for $k=1,\dots,n$, the Jacobian is $1$ (permutation matrix), so the integral is over $0 \leq y_1 < y_2 < \dots < y_n \leq 1$, with volume $\int_{0 \leq y_1 < \dots < y_n \leq 1} dy_1 \dots dy_n = 1/n!$, as this is the probability that $n$ i.i.d. uniforms are in any fixed order. Thus, $\Pr(\pi = \sigma) = 1/n!$ for all $\sigma$, so $\pi$ is uniform on $S_n$.

For efficiency, sorting the pairs $(X_i, i)$ by $X_i$ (stable or with index tiebreak) yields $\pi$ in $O(n \log n)$ time using comparison-based sorts like mergesort or quicksort (expected for quicksort). For bits, exact uniforms require infinite precision, so infinite bits. For approximation, to ensure $\Pr(\text{tie}) < \epsilon$, discretize each $X_i$ on a grid of size $m = \lceil n / \epsilon \rceil$, using $b = \lceil \log_2 m \rceil = O(\log(n/\epsilon))$ bits per $X_i$, totaling $n b = O(n \log n + n \log(1/\epsilon))$ bits. The error in uniformity is $O(\epsilon)$ by coupling arguments, but this exceeds $\log_2(n!)$ by $\Theta(n \log(1/\epsilon))$, hence not bit-optimal.

\textbf{Part (c): Tight Bounds on Expected Bits}

Consider the lazy implementation: represent $X_i = \sum_{k=1}^\infty B_{i,k} 2^{-k}$ with $B_{i,k} \stackrel{\text{iid}}{\sim} \text{Bern}(1/2)$. During a comparison-based sort performing $O(n \log n)$ comparisons, reveal bits of $X_i, X_j$ lazily: compare prefix sums until the first differing bit $d$, where $\Pr(d = k) = 2^{1-k}$ (match first $k-1$ bits with prob $2^{1-k}$), so bits per comparison is geometric with $E[\text{bits per comp}] = \sum_{k=1}^\infty k \cdot 2^{1-k} = 2$. Naively $O(n \log n)$ bits, but this undercounts sharing: bits for $X_i$ are revealed once and reused across its $O(\log n)$ comparisons in a balanced tree like mergesort.

More rigorously, the total bits $T = \sum_{i=1}^n D_i$, where $D_i$ is bits revealed for $X_i$. By linearity, $E[T] = n E[D_1]$. For fixed $i=1$, $D_1$ is the total bits until all relative orders $X_1 \lessgtr X_j$ for $j=2,\dots,n$ are resolved, which determines the rank $R_1 = |\{j: X_j < X_1\}| \sim \text{Unif}\{0,\dots,n-1\}$ marginally (since i.i.d.). The entropy $H(R_1) = \log_2 n + o(1)$ (uniform on $n$ points). The lazy revelation adaptively generates bits of $X_1$ to resolve these $n-1$ comparisons, equivalent to sampling $X_1$ until its quantiles against the other $X_j$ are fixed, but since others are also lazy, it's a joint process.

To bound $E[D_1]$, note that the process builds a binary expansion trie for the $X$'s, and $D_1$ is the depth until $X_1$'s position in the sorted order is fully determined, i.e., it has exactly $r$ elements below it for its rank $r$. By the theory of algorithmic randomness or Knuth's analysis of sorting networks, the expected bits per element is $H(R_1) + O(1) = \log_2 n + O(1)$, as overhead per comparison is amortized $O(1/n)$ but per element $O(1)$. Thus $E[T] = n \log_2 n + O(n)$.

For the precise asymptotic, note that the full permutation entropy is $\log_2(n!) = \sum_{k=1}^n H(R_k | R_1,\dots,R_{k-1})$, where $R_k$ is the rank among remaining, but marginally each $H(R_i) \approx \log_2 n - O(1)$. More tightly, $\log_2(n!) = n \log_2 n - n \log_2 e + \frac{1}{2} \log_2(2\pi n) + o(1)$ by Stirling, and the lazy method reveals exactly the bits needed to encode the inversion table (each inversion resolved at the differing bit), with $O(1)$ expected unused higher bits per $X_i$ (geometric tail), totaling $\log_2(n!) + O(n)$. Lower bound is $\log_2(n!)$ as before, so $E[T] = \log_2(n!) + \Theta(n)$.

For small $n$, e.g., $n=2$: one comparison, $E[D_1 + D_2] = 2 \cdot 2 = 4$, but shared: actually $E[T] = \sum_{k=1}^\infty 2 \cdot \Pr(\text{match first }k-1 \text{ bits}) = 2 \sum_{k=1}^\infty 2^{1-k} k = 4$, wait no: total bits is twice the geometric depth $G \sim \text{Geo}(1/2)$, $E[2G] = 4$, but $\log_2(2!) = 1$, overhead $3 = O(1)$. For $n=3$, detailed computation yields $E[T] \approx 5.17$, vs $\log_2(6) \approx 2.58$, overhead $\approx 2.59 = O(1)$ per element scaled.

Thus, the bounds hold tightly.
\end{proof}

\subsubsection{Gemini} 

\begin{proof}
\textbf{Part (a)}
We propose the Fisher-Yates shuffle (also known as the Knuth shuffle) to generate a random permutation of size $n$. Let $A$ be an array initially containing the sequence $(1, 2, \ldots, n)$. The algorithm proceeds as follows: for each integer $k$ from $n$ down to $2$, we select an integer $j$ uniformly at random from the set $\{1, \ldots, k\}$, and then swap the elements $A[k]$ and $A[j]$. Upon termination, the array $A$ contains a permutation of the initial sequence.

To analyze the correctness, observe that in the step with index $k$, any element currently in the subarray $A[1 \ldots k]$ has an equal probability $1/k$ of being placed in position $k$. By induction, it can be shown that every permutation is generated with probability $1/n!$.

Regarding efficiency, the algorithm performs $O(n)$ swaps and requires generating $n-1$ random integers with varying ranges. The time complexity is $O(n)$ assuming the generation of a random integer and the swap operation take constant time. To generate an integer uniformly from $\{1, \ldots, k\}$ using unbiased random bits, one requires $\Theta(\log k)$ bits in expectation (using rejection sampling or similar techniques). The total expected number of random bits $B$ is therefore proportional to $\sum_{k=2}^n \log_2 k = \log_2(n!) \approx n \log_2 n - n \log_2 e$.

We now prove lower bounds. There are $n!$ distinct permutations of size $n$. Any algorithm that generates a random permutation using random bits can be modeled as a binary decision tree where each leaf corresponds to a permutation. For the output distribution to be uniform, the probability of reaching any specific leaf (or set of leaves corresponding to a specific permutation) must be $1/n!$. By Shannon's Source Coding Theorem, the expected number of bits $E[B]$ required to generate an outcome from a source with entropy $H$ is at least $H$. The entropy of the uniform distribution over $S_n$ is $H(S_n) = \log_2(n!)$. Using Stirling's approximation $\ln(n!) \approx n \ln n - n$, we have:
\[
E[B] \ge \log_2(n!) \approx n \log_2 n - 1.44n.
\]
Thus, the lower bound on the number of random bits is $\Omega(n \log n)$. Since writing the output takes at least $n$ steps, the time complexity lower bound is $\Omega(n)$.

\textbf{Part (b)}
The claim is correct. Let $X_1, \ldots, X_n$ be independent and identically distributed (i.i.d.) random variables with a uniform distribution on $[0, 1]$. The joint probability density function is $f(x_1, \ldots, x_n) = 1$ on the unit hypercube $[0, 1]^n$. Since the distribution is continuous, the probability that any two variables are equal is zero. By symmetry, for any permutation $\sigma \in S_n$, the region defined by $X_{\sigma(1)} < X_{\sigma(2)} < \ldots < X_{\sigma(n)}$ is a simplex with volume $1/n!$. Therefore, the probability that the sorting process yields any specific permutation $\sigma$ is exactly $1/n!$.

The efficiency of this scheme depends on the sorting algorithm used. A comparison-based sorting algorithm requires $\Theta(n \log n)$ comparisons in the average and worst cases. If we assume that generating a real number and comparing two real numbers takes $O(1)$ time, the time complexity is dominated by the sorting, which is $O(n \log n)$. This is less efficient in terms of time complexity than the Fisher-Yates shuffle, which is $O(n)$.

\textbf{Part (c)}
We analyze the expected number of random bits used by the "lazy" implementation. In this scheme, bits of $X_i$ are generated only when necessary to resolve comparisons. This process is equivalent to inserting the $n$ keys into a binary trie (digital search tree) where edges correspond to bits (0 for left, 1 for right). The sequence of bits generated for a specific $X_i$ constitutes the path from the root to a node that uniquely distinguishes $X_i$ from all other $X_j$.

Let $D_i$ be the number of bits of $X_i$ generated to distinguish it from the set $\{X_j\}_{j \neq i}$. The total number of bits generated is $B = \sum_{i=1}^n D_i$. By symmetry, $E[B] = n E[D_1]$. The value $D_1$ is determined by the longest common prefix between $X_1$ and any other $X_j$. Let $L(X, Y)$ denote the length of the longest common prefix of the binary expansions of $X$ and $Y$. Then $D_1 = 1 + \max_{j \neq 1} L(X_1, X_j)$.

For any pair $X_i, X_j$, the probability that their first $k$ bits match is $2^{-k}$. Thus, $P(L(X_i, X_j) \ge k) = 2^{-k}$. The probability that $X_1$ is distinguishable from a specific $X_j$ within $k$ bits is $1 - 2^{-k}$. Since the $X_j$ are independent, the probability that $X_1$ is distinguishable from \emph{all} other $n-1$ variables within $k$ bits is:
\[
P(D_1 \le k) = P\left(\max_{j \neq 1} L(X_1, X_j) < k\right) = \left(1 - 2^{-k}\right)^{n-1}.
\]
The expected value of $D_1$ is given by:
\[
E[D_1] = \sum_{k=0}^\infty P(D_1 > k) = \sum_{k=0}^\infty \left(1 - \left(1 - 2^{-k}\right)^{n-1}\right).
\]
We approximate this sum. Let $m = \lfloor \log_2 n \rfloor$. We split the sum into $k \le m$ and $k > m$.
For $k \le m$, the term $1 - (1 - 2^{-k})^{n-1}$ is bounded by 1. Thus, the first part contributes at most $m+1 \approx \log_2 n$.
For $k > m$, we use the inequality $(1 - x)^{n-1} \ge 1 - (n-1)x$ for $x \in [0, 1]$. Here $x = 2^{-k}$.
\[
1 - \left(1 - 2^{-k}\right)^{n-1} \le 1 - (1 - (n-1)2^{-k}) = (n-1)2^{-k}.
\]
The sum for $k > m$ is bounded by:
\[
\sum_{k=m+1}^\infty (n-1)2^{-k} = (n-1) 2^{-(m+1)} \sum_{j=0}^\infty 2^{-j} = (n-1) 2^{-m-1} \cdot 2 = (n-1) 2^{-m}.
\]
Since $m = \lfloor \log_2 n \rfloor$, we have $2^m > n/2$, so $2^{-m} < 2/n$. Thus, the tail sum is bounded by $(n-1)(2/n) < 2$.
Combining these, $E[D_1] \le \log_2 n + O(1)$.
Consequently, the total expected number of bits is $E[B] = n E[D_1] \le n \log_2 n + O(n)$.

For the lower bound, we recall from Part (a) that generating a random permutation requires at least $\log_2(n!) \approx n \log_2 n - 1.44n$ bits. Since this scheme produces a random permutation, the expected number of bits must satisfy this information-theoretic lower bound.
Therefore, the tight bounds on the expected number of random bits are:
\[
E[B] = \Theta(n \log n).
\]
\end{proof}

\subsubsection{Claude} 

\begin{proof}
We prove each part of the theorem in turn.

\textbf{Part (a):} We establish the lower bounds for generating a uniformly random permutation.

First, we prove the random bit lower bound. There are $n!$ distinct permutations of $\{1, 2, \ldots, n\}$. To generate a uniformly random permutation, any algorithm must be able to produce each of these $n!$ outcomes with equal probability $\frac{1}{n!}$. Using random bits, each outcome corresponds to a subset of possible bit sequences. Since each bit sequence of length $k$ has probability $2^{-k}$, and we need to distinguish between $n!$ equiprobable outcomes, by information-theoretic arguments we require at least $\log_2(n!)$ bits in the worst case.

For the expected number of bits, consider that any randomized algorithm produces a distribution over bit sequences. By the entropy lower bound, the expected number of random bits required is at least the entropy of the uniform distribution over $n!$ outcomes, which is $H = \log_2(n!)$. By Stirling's approximation, $\log_2(n!) = \Theta(n \log n)$. Therefore, any algorithm requires $\Omega(n \log n)$ random bits in expectation.

For the time lower bound, note that any algorithm generating a uniformly random permutation must at minimum specify which permutation is selected. This requires outputting $n$ distinct values, which takes $\Omega(n)$ time. Moreover, in the comparison model, sorting requires $\Omega(n \log n)$ comparisons in the worst case, and generating a random permutation is at least as hard as verifying the relative order of $n$ elements, which also requires $\Omega(n \log n)$ time.

\textbf{Part (b):} We prove that the method produces a uniformly random permutation.

Let $X_1, \ldots, X_n$ be independent random variables, each uniformly distributed on $[0,1]$. Let $\pi$ be the permutation such that $X_{\pi(1)} < X_{\pi(2)} < \cdots < X_{\pi(n)}$. We need to show that $\mathbb{P}(\pi = \sigma) = \frac{1}{n!}$ for every permutation $\sigma$ of $\{1, 2, \ldots, n\}$.

For any permutation $\sigma$, we have
\[
\mathbb{P}(\pi = \sigma) = \mathbb{P}(X_{\sigma(1)} < X_{\sigma(2)} < \cdots < X_{\sigma(n)}).
\]

Since the $X_i$ are continuous random variables with a joint density that is symmetric under permutations, and the event $\{X_{\sigma(1)} < X_{\sigma(2)} < \cdots < X_{\sigma(n)}\}$ has positive probability, we can compute:
\[
\mathbb{P}(X_{\sigma(1)} < X_{\sigma(2)} < \cdots < X_{\sigma(n)}) = \int_{0}^{1} \int_{0}^{x_n} \cdots \int_{0}^{x_2} dx_1 \, dx_2 \cdots dx_n.
\]

By symmetry of the joint distribution, each of the $n!$ possible orderings has the same probability. Since these orderings partition the sample space (with probability 1, no two $X_i$ are equal), we have
\[
\sum_{\sigma \in S_n} \mathbb{P}(\pi = \sigma) = 1,
\]
where $S_n$ denotes the set of all permutations of $\{1, 2, \ldots, n\}$. By symmetry, each permutation has probability $\frac{1}{n!}$.

The time complexity of $O(n \log n)$ follows from using any optimal comparison-based sorting algorithm such as mergesort or heapsort.

\textbf{Part (c):} We analyze the expected number of random bits in the lazy implementation.

In this implementation, each $X_i$ is represented by its binary expansion $X_i = 0.b_{i,1}b_{i,2}b_{i,3}\ldots$, where each bit $b_{i,j}$ is generated independently and uniformly. When comparing $X_i$ and $X_j$, we generate bits sequentially until the first position $k$ where $b_{i,k} \neq b_{j,k}$, which determines the comparison.

For any comparison between $X_i$ and $X_j$, let $K_{ij}$ be the number of bits generated before resolving the comparison. We have $\mathbb{P}(K_{ij} = k) = 2^{-(k-1)} \cdot \frac{1}{2} = 2^{-k}$ for $k \geq 1$, since we need $k-1$ matching bits followed by one differing bit. Therefore,
\[
\mathbb{E}[K_{ij}] = \sum_{k=1}^{\infty} k \cdot 2^{-k} = 2.
\]

Using a comparison-based sorting algorithm like mergesort, the expected number of comparisons is $\Theta(n \log n)$. Let $C$ denote the number of comparisons performed. Then the expected total number of bits is
\[
\mathbb{E}[\text{total bits}] = \mathbb{E}\left[\sum_{\text{comparisons}} K_{ij}\right] = \mathbb{E}[C] \cdot \mathbb{E}[K_{ij}] = \Theta(n \log n) \cdot 2 = \Theta(n \log n),
\]
where we use linearity of expectation and independence of the bit generations across different comparisons.

This establishes the $\Theta(n \log n)$ bound on the expected number of random bits used by the lazy implementation, matching the information-theoretic lower bound up to constant factors.
\end{proof}

\subsection{Problem 5}\label{sec:problem_01_05}

\subsubsection{Problem Statement}

\begin{theorem}
There exists an algorithm that generates a sample $i$ from the set $S = \{0, \dots, n-1\}$ with probability $p_i$ using $O(\log n)$ unbiased random bits per sample, regardless of the values of the $p_i$'s (assuming $\sum_{i=0}^{n-1} p_i = 1$). 
\end{theorem}

\subsubsection{GPT} 

\begin{proof}
We are tasked with showing that there exists an algorithm that, given a source of unbiased random bits, generates samples from the set $S = \{0, 1, \ldots, n-1\}$ such that element $i$ is chosen with probability $p_i$, using $O(\log n)$ random bits per sample, regardless of the values of $p_i$.

Let $p_0, p_1, \ldots, p_{n-1}$ be the probabilities such that $\sum_{i=0}^{n-1} p_i = 1$. We seek to construct a procedure that generates samples from $S$ with the required probabilities using a logarithmic number of random bits.

Step 1: Binary Representation of Probabilities

We begin by defining the cumulative distribution function (CDF) for the probabilities $p_0, p_1, \ldots, p_{n-1}$:
\[
F(i) = \sum_{j=0}^{i} p_j, \quad \text{for } i = 0, 1, \ldots, n-1.
\]
By construction, we have $0 = F(-1) < F(0) < F(1) < \ldots < F(n-1) = 1$. 

Now, note that for each $i \in \{0, 1, \ldots, n-1\}$, the probability of choosing element $i$ is exactly $p_i$. We aim to sample $i$ such that the probability of selecting $i$ corresponds to $p_i$. This can be achieved by generating a random number $r \in [0, 1)$ using a sequence of unbiased random bits, and then selecting the smallest $i$ such that $r \in [F(i-1), F(i))$.

Step 2: Mapping the Interval $[0,1)$ to Elements of $S$

Given that $r$ is generated uniformly in the interval $[0, 1)$, we need to map $r$ to an index $i \in \{0, 1, \ldots, n-1\}$ according to the CDF $F(i)$. This can be done by comparing $r$ with the values $F(i)$.

Define the index $i$ as:
\[
i = \min \left\{ j \in \{0, 1, \ldots, n-1\} : F(j) > r \right\}.
\]
By this procedure, element $i$ is chosen with probability $p_i$, as required.

Step 3: Number of Random Bits Required

To ensure the efficiency of the algorithm, we need to analyze the number of random bits required to generate a sample. The number of possible outcomes is $n$, and we need to represent a real number $r$ from the interval $[0, 1)$ with sufficient precision. 

To determine the number of bits required to uniquely represent the value $r$, observe that the number of distinct intervals $[F(i-1), F(i))$ is exactly $n$. Since each interval corresponds to a unique index, the probability of selecting index $i$ is directly linked to the length of the interval. To distinguish between $n$ intervals, we need at least $\lceil \log_2 n \rceil$ random bits.

Thus, the algorithm requires $O(\log n)$ random bits to generate each sample, as each random bit contributes to refining the precision of the uniformly chosen value $r$.

Step 4: Conclusion

We have constructed an algorithm that, given a source of unbiased random bits, generates samples from $S$ such that the probability of selecting $i$ is $p_i$. The algorithm uses $O(\log n)$ random bits per sample, regardless of the values of $p_i$. Therefore, the theorem is proved.
\end{proof}

\subsubsection{Grok} 

\begin{proof}
We prove the existence of the desired algorithm by induction on $n$. For the base case $n=1$, the set $S$ contains a single element $i=0$, which is output with probability $p_0=1$ using $0$ random bits, satisfying the claim since $O(\log 1)=O(1)$.

Assume that for all $m<n$, there exists an algorithm $\mathcal{A}_m$ that samples an index $j$ from $\{0,\dots,m-1\}$ with probability $q_j$ (where $\sum q_j=1$) using an expected number of at most $c\log m$ unbiased random bits for some constant $c>0$ to be determined. By part (c) of Problem 1.4, there exists an algorithm $\mathcal{B}$ that generates a Bernoulli random variable with success probability $r\in[0,1]$ using an expected number of at most $2$ unbiased random bits, independent of $r$.

For $n\geq 2$, define an algorithm $\mathcal{A}_n$ as follows. Partition $S=\{0,\dots,n-1\}$ into two subsets $L=\{0,\dots,k-1\}$ and $R=\{k,\dots,n-1\}$, where $k=\lfloor n/2\rfloor$, so $|L|=k\leq n/2$ and $|R|=n-k\geq n/2$. Let $q=\sum_{i\in L}p_i$, so $1-q=\sum_{i\in R}p_i$. Run $\mathcal{B}$ with parameter $q$ to generate $B\sim\mathrm{Bernoulli}(q)$, using an expected $2$ bits. If $B=1$, recursively run $\mathcal{A}_{|L|}$ on $L$ with renormalized probabilities $p_i/q$ for $i\in L$; if $B=0$, recursively run $\mathcal{A}_{|R|}$ on $R$ with renormalized probabilities $p_i/(1-q)$ for $i\in R$.

To verify correctness, note that the output index $i$ satisfies
\[
\Pr(i=j)=\begin{cases}
q\cdot(p_j/q)=p_j & \text{if }j\in L,\\
(1-q)\cdot(p_j/(1-q))=p_j & \text{if }j\in R,
\end{cases}
\]
and by the induction hypothesis, the recursive calls sample correctly from the respective conditional distributions. Thus, $\mathcal{A}_n$ samples correctly from $\{p_i\}_{i=0}^{n-1}$.

For the bit complexity, let $X$ denote the number of bits used by $\mathcal{A}_n$. Then,
\[
\mathbb{E}[X]=\mathbb{E}[\text{bits for }B]+\mathbb{E}[\text{bits for subproblem}\mid B=1]\cdot q+\mathbb{E}[\text{bits for subproblem}\mid B=0]\cdot(1-q).
\]
The first term is at most $2$. By the induction hypothesis, the conditional expectation given $B=1$ is at most $c\log(|L|)\leq c\log(n/2)=c\log n-c\log 2$, and similarly given $B=0$ at most $c\log(|R|)\leq c\log(n/2)=c\log n-c\log 2$ (since $|R|\leq n$). Thus,
\[
\mathbb{E}[X]\leq 2+(c\log n-c\log 2)\cdot q+(c\log n-c\log 2)\cdot(1-q)=2+c\log n-c\log 2.
\]
Choosing $c\geq 2/\log 2$ ensures $2-c\log 2\leq 0$, so $\mathbb{E}[X]\leq c\log n$. For finite $n$, the constant $c$ can be increased if necessary to cover base cases, yielding $O(\log n)$ expected bits overall.

By induction, the algorithm exists for all $n$.
\end{proof}

\subsubsection{Gemini} 

\begin{proof}
We demonstrate the existence of such an algorithm by constructing a Discrete Distribution Generating (DDG) tree based on the binary expansions of the probabilities $p_i$.

Let the binary expansion of each probability $p_i$ be denoted by
\[ p_i = \sum_{j=1}^{\infty} b_{i,j} 2^{-j}, \]
where $b_{i,j} \in \{0, 1\}$. Since $\sum_{i=0}^{n-1} p_i = 1$, summing the expansions gives
\[ \sum_{i=0}^{n-1} \sum_{j=1}^{\infty} b_{i,j} 2^{-j} = 1. \]
We construct a binary tree (potentially infinite) where the edges are labeled 0 and 1. A node at depth $j$ (where the root is depth 0) represents a probability mass of $2^{-j}$. We assign a set of leaves to each element $i \in S$ corresponding to the non-zero bits in its binary expansion. Specifically, for every pair $(i, j)$ such that $b_{i,j} = 1$, we designate a unique leaf at depth $j$ labeled with the value $i$. The condition $\sum_{i,j} b_{i,j} 2^{-j} = 1$ ensures that these leaves form a complete prefix-free code, meaning the tree is well-defined and the leaves cover the entire probability space of the random source.

The sampling algorithm proceeds as follows: Initialize a pointer at the root. At each step, consume one random bit from the source. If the bit is 0, move to the left child; if 1, move to the right child. Repeat until a leaf is reached. Output the label $i$ associated with that leaf.

The probability of reaching a specific leaf at depth $j$ is exactly $2^{-j}$. Since the set of leaves labeled $i$ corresponds to the terms in the binary expansion of $p_i$, the total probability of outputting $i$ is $\sum_{j=1}^{\infty} b_{i,j} 2^{-j} = p_i$. Thus, the algorithm samples from the correct distribution.

We now analyze the number of random bits used, denoted by the random variable $L$. The value of $L$ corresponds to the depth of the reached leaf. The expected number of bits is given by
\[ \mathbb{E}[L] = \sum_{i=0}^{n-1} \sum_{j=1}^{\infty} j \cdot b_{i,j} 2^{-j}. \]
We bound the inner sum for a fixed $i$. If $p_i = 0$, the contribution is 0. For $p_i > 0$, let $m_i$ be the index of the first non-zero bit in the expansion of $p_i$. Then $p_i \ge 2^{-m_i}$, which implies $m_i \le \log_2(1/p_i)$. We rewrite the inner sum by shifting the index $k = j - m_i$:
\[ \sum_{j=m_i}^{\infty} j b_{i,j} 2^{-j} = \sum_{k=0}^{\infty} (m_i + k) b_{i, m_i+k} 2^{-(m_i+k)} = m_i p_i + 2^{-m_i} \sum_{k=0}^{\infty} k b_{i, m_i+k} 2^{-k}. \]
Since $b_{i, \cdot} \in \{0, 1\}$, the summation term is bounded by the series $\sum_{k=0}^{\infty} k 2^{-k} = 2$. Thus,
\[ \sum_{j=1}^{\infty} j b_{i,j} 2^{-j} \le m_i p_i + 2^{-m_i}(2). \]
Using $m_i \le \log_2(1/p_i)$ and $2^{-m_i} \le p_i$ (derived from the definition of the most significant bit, noting that strictly $2^{-m_i} \le p_i < 2^{-(m_i-1)}$ implies $2^{-m_i}$ is within a factor of 2 of $p_i$, but for the upper bound we simply use $2^{-m_i} \le p_i$ is false, rather $2^{-m_i} \le p_i$ is true, and $p_i < 2 \cdot 2^{-m_i}$). More precisely, since $p_i \ge 2^{-m_i}$, we have $2^{-m_i} \le p_i$. Also $m_i \le \lceil \log_2(1/p_i) \rceil \le \log_2(1/p_i) + 1$.
Substituting these bounds:
\[ \sum_{j=1}^{\infty} j b_{i,j} 2^{-j} \le (\log_2(1/p_i) + 1) p_i + 2 p_i = p_i \log_2 \frac{1}{p_i} + 3 p_i. \]
Summing over all $i \in S$:
\[ \mathbb{E}[L] \le \sum_{i=0}^{n-1} \left( p_i \log_2 \frac{1}{p_i} + 3 p_i \right) = H(p) + 3 \sum_{i=0}^{n-1} p_i = H(p) + 3, \]
where $H(p) = -\sum p_i \log_2 p_i$ is the entropy of the distribution. The entropy of a distribution over $n$ elements is maximized by the uniform distribution, so $H(p) \le \log_2 n$. Therefore,
\[ \mathbb{E}[L] \le \log_2 n + 3. \]
This demonstrates that the expected number of random bits per sample is $O(\log n)$, regardless of the values of $p_i$.
\end{proof}

\subsubsection{Claude} 

\begin{proof}
We construct an algorithm that samples from the discrete distribution $(p_0, p_1, \ldots, p_{n-1})$ using $O(\log n)$ random bits per sample.

First, we establish the construction. For each $i \in S$, we can represent $p_i$ in binary as $p_i = \sum_{j=1}^{\infty} b_{i,j} 2^{-j}$ where $b_{i,j} \in \{0,1\}$. Since the $p_i$ are probabilities with $\sum_{i=0}^{n-1} p_i = 1$, this binary expansion exists and is well-defined.

We partition the unit interval $[0,1)$ into $n$ disjoint subintervals $I_0, I_1, \ldots, I_{n-1}$ where $I_i = [\sum_{k=0}^{i-1} p_k, \sum_{k=0}^{i} p_k)$ for $i = 0, 1, \ldots, n-1$ (with the convention that the empty sum equals 0). Each interval $I_i$ has length $p_i$.

The algorithm proceeds as follows. Using the result from Problem 1.4(c), we can generate a uniform random real number $U \in [0,1)$ by generating an infinite sequence of unbiased random bits $(b_1, b_2, b_3, \ldots)$ and setting $U = \sum_{j=1}^{\infty} b_j 2^{-j}$. In practice, we generate bits lazily as needed.

To determine which interval $I_i$ contains $U$, we compare $U$ with the cumulative probabilities $C_k = \sum_{j=0}^{k} p_j$ for $k = 0, 1, \ldots, n-1$. We output the unique $i$ such that $C_{i-1} \leq U < C_i$ (where $C_{-1} = 0$).

Now we analyze the number of random bits required. To distinguish between $n$ intervals, we need to determine $U$ to sufficient precision. The key observation is that any two distinct cumulative probabilities $C_i$ and $C_j$ with $|i-j| \geq 1$ differ by at least $\min\{p_k : k \in S\}$. However, in the worst case, this minimum could be exponentially small.

Instead, we use a different approach based on the binary representation of cumulative probabilities. We maintain a set of candidate indices that could contain $U$, initially $\{0, 1, \ldots, n-1\}$. For each bit position $j = 1, 2, 3, \ldots$, we generate the $j$-th random bit $b_j$ of $U$. We compute the first $j$ bits of each cumulative probability $C_k$ and eliminate indices $i$ for which the interval $I_i$ is incompatible with the partial information about $U$.

After generating $\lceil \log_2 n \rceil$ bits, we can distinguish between at least $2^{\lceil \log_2 n \rceil} \geq n$ different values. More precisely, we need at most $\lceil \log_2 n \rceil + c$ bits for some constant $c$ to resolve all ambiguities between the $n$ intervals, since the cumulative probabilities are distinct and their binary representations eventually differ.

To make this rigorous, observe that for any two distinct intervals $I_i$ and $I_j$ with $i < j$, we have $C_i \leq C_{j-1} < C_j$, so the intervals are separated. The binary expansions of $C_i$ and $C_j$ must differ in some bit position. After generating sufficiently many bits, we can determine uniquely which interval contains $U$.

The number of bits required is bounded by $O(\log n)$ because we need only enough precision to distinguish among $n$ intervals. Specifically, we need at most $\lceil \log_2 n \rceil + O(1)$ bits, where the $O(1)$ term accounts for the worst-case separation between adjacent cumulative probabilities.

Therefore, the algorithm uses $O(\log n)$ random bits per sample, and by construction, it outputs $i$ with probability equal to the measure of $I_i$, which is $p_i$. This completes the proof.
\end{proof}

\subsection{Problem 6}\label{sec:problem_01_06}

\subsubsection{Problem Statement}

\begin{theorem}
Consider a sequence of $n$ flips of an unbiased coin. Let $H_i$ denote the absolute value of the excess of the number of heads over the number of tails in the first $i$ flips, for $1 \leq i \leq n$. Define $H = \max_{1 \leq i \leq n} H_i$. Then,
\[
\mathbb{E}[H_i] = \Theta(\sqrt{i})
\]
and
\[
\mathbb{E}[H] = \Theta(\sqrt{n}).
\]
\end{theorem}

\subsubsection{GPT} 

\begin{proof}
We are tasked with proving that for a sequence of $n$ flips of an unbiased coin, where $H_i$ denotes the absolute value of the excess of the number of HEADS over the number of TAILS seen in the first $i$ flips, we have the following asymptotic results:
\[
\mathbb{E}[H_i] = \Theta(\sqrt{i}) \quad \text{and} \quad \mathbb{E}[H] = \Theta(\sqrt{n}),
\]
where $H = \max_i H_i$.

Step 1: Understanding the random walk
Let the sequence of coin flips be denoted by $X_1, X_2, \dots, X_n$, where $X_i = 1$ if the $i$-th flip is a HEAD, and $X_i = -1$ if the $i$-th flip is a TAIL. The excess after the first $i$ flips is given by
\[
S_i = \sum_{j=1}^i X_j.
\]
Thus, $H_i = |S_i|$, where $S_i$ is the running sum of the coin flips up to the $i$-th flip.

Step 2: Distribution of $S_i$
Each $X_i$ is a random variable taking values $1$ and $-1$ with equal probability, and the sum $S_i$ forms a simple symmetric random walk. By the central limit theorem, the random walk $S_i$ behaves asymptotically like a normal distribution with mean 0 and variance $i$, i.e., 
\[
S_i \sim \mathcal{N}(0, i) \quad \text{for large } i.
\]
Therefore, the expected absolute value of $S_i$ (which is $H_i$) is approximately
\[
\mathbb{E}[|S_i|] \sim \sqrt{\frac{2i}{\pi}} \quad \text{for large } i.
\]
This result implies that
\[
\mathbb{E}[H_i] = \Theta(\sqrt{i}).
\]

Step 3: Behavior of $H = \max_i H_i$
Next, we analyze $H = \max_i H_i$. Since $H_i = |S_i|$, we are interested in the maximum of the absolute values of the running sum $S_i$ over all $i$ from $1$ to $n$. It is known from the properties of symmetric random walks that the maximum of the absolute values of a random walk grows asymptotically as $\Theta(\sqrt{n})$. Specifically, we have
\[
\mathbb{E}[H] = \mathbb{E}[\max_{1 \leq i \leq n} |S_i|] \sim \sqrt{\frac{2n}{\pi}}.
\]
Thus, we conclude that
\[
\mathbb{E}[H] = \Theta(\sqrt{n}).
\]

Step 4: Conclusion
We have shown that
\[
\mathbb{E}[H_i] = \Theta(\sqrt{i}) \quad \text{and} \quad \mathbb{E}[H] = \Theta(\sqrt{n}),
\]
which completes the proof.
\end{proof}

\subsubsection{Grok} 

\begin{proof}
Consider the simple symmetric random walk $(S_i)_{1 \leq i \leq n}$ defined by $S_0 = 0$ and $S_i = \sum_{j=1}^i X_j$ for $i \geq 1$, where the $X_j$ are i.i.d.\ with $\mathbb{P}(X_j = 1) = \mathbb{P}(X_j = -1) = 1/2$. Note that $H_i = |S_i|$ for each $i$ and $H = \max_{1 \leq i \leq n} |S_i|$. Moreover, $\mathbb{E}[S_i] = 0$ and $\mathrm{Var}(S_i) = i$, so $\mathbb{E}[S_i^2] = i$. Additionally, $\mathbb{E}[S_i^4] = 3i^2 - 2i$.

To show $\mathbb{E}[H_i] = \Theta(\sqrt{i})$, first establish the upper bound. By the Cauchy--Schwarz inequality,
\[
\mathbb{E}[|S_i|] = \mathbb{E}[1 \cdot |S_i|] \leq \sqrt{\mathbb{E}[1^2] \cdot \mathbb{E}[S_i^2]} = \sqrt{i}.
\]
Thus, $\mathbb{E}[H_i] = O(\sqrt{i})$.

For the lower bound, let $Z = S_i^2$. Then $\mathbb{E}[Z] = i$ and $\mathbb{E}[Z^2] = \mathbb{E}[S_i^4] = 3i^2 - 2i$. By the Paley--Zygmund inequality with $\theta = 1/2$,
\[
\mathbb{P}(Z > i/2) \geq (1 - 1/2)^2 \frac{(\mathbb{E}[Z])^2}{\mathbb{E}[Z^2]} = \frac{1}{4} \cdot \frac{i^2}{3i^2 - 2i} = \frac{1}{4(3 - 2/i)}.
\]
For $i \geq 1$, $3 - 2/i \leq 3$, so $\mathbb{P}(|S_i| > \sqrt{i/2}) \geq 1/12$. Therefore,
\[
\mathbb{E}[|S_i|] \geq \sqrt{i/2} \cdot \mathbb{P}(|S_i| > \sqrt{i/2}) \geq \frac{\sqrt{i}}{12 \sqrt{2}}.
\]
Hence, $\mathbb{E}[H_i] = \Omega(\sqrt{i})$, and so $\mathbb{E}[H_i] = \Theta(\sqrt{i})$.

To show $\mathbb{E}[H] = \Theta(\sqrt{n})$, first establish the upper bound. Consider the process $(Y_k)_{0 \leq k \leq n}$ where $Y_k = S_k^2$. This is a non-negative submartingale since $\mathbb{E}[Y_k \mid \mathcal{F}_{k-1}] = Y_{k-1} + 1 \geq Y_{k-1}$, where $\mathcal{F}_{k-1} = \sigma(X_1, \dots, X_{k-1})$. By Doob's $L^2$ maximal inequality,
\[
\mathbb{E}\left[ \left( \max_{0 \leq k \leq n} Y_k \right)^2 \right] \leq 4 \mathbb{E}[Y_n^2] = 4 \mathbb{E}[S_n^4] = 4(3n^2 - 2n) \leq 12 n^2.
\]
Thus,
\[
\mathbb{E}\left[ \max_{0 \leq k \leq n} S_k^2 \right] \leq \sqrt{\mathbb{E}\left[ \left( \max_{0 \leq k \leq n} Y_k \right)^2 \right]} \leq \sqrt{12} \, n.
\]
By symmetry, $\mathbb{E}\left[ \max_{0 \leq k \leq n} (-S_k)^2 \right] \leq \sqrt{12} \, n$ as well. Therefore,
\[
\mathbb{E}[H^2] = \mathbb{E}\left[ \max_{1 \leq i \leq n} S_i^2 \right] \leq 2 \sqrt{12} \, n = O(n),
\]
and so
\[
\mathbb{E}[H] \leq \sqrt{\mathbb{E}[H^2]} = O(\sqrt{n}).
\]

For the lower bound, note that $H \geq |S_n|$, so $\mathbb{E}[H] \geq \mathbb{E}[|S_n|] = \Theta(\sqrt{n})$. Hence, $\mathbb{E}[H] = \Theta(\sqrt{n})$.
\end{proof}

\subsubsection{Gemini} 

\begin{proof}
Let $X_k$ be a random variable representing the outcome of the $k$-th coin flip, defined such that $X_k = 1$ if the outcome is HEADS and $X_k = -1$ if the outcome is TAILS. Given that the coin is unbiased, the variables $X_1, \dots, X_n$ are independent and identically distributed with $P(X_k = 1) = P(X_k = -1) = 1/2$. Consequently, $\mathbb{E}[X_k] = 0$ and $\text{Var}(X_k) = 1$. Let $S_i$ denote the excess of HEADS over TAILS after $i$ flips. We express this as the partial sum $S_i = \sum_{k=1}^i X_k$. By the theorem statement, $H_i = |S_i|$.

We first prove that $\mathbb{E}[H_i] = \Theta(\sqrt{i})$. To establish the upper bound, we apply Jensen's inequality to the concave function $f(x) = \sqrt{x}$. We observe that $H_i = \sqrt{S_i^2}$, and thus
\[
\mathbb{E}[H_i] = \mathbb{E}\left[\sqrt{S_i^2}\right] \le \sqrt{\mathbb{E}[S_i^2]}.
\]
Since $S_i$ is the sum of independent random variables with mean zero, $\mathbb{E}[S_i^2] = \text{Var}(S_i) = \sum_{k=1}^i \text{Var}(X_k) = i$. Therefore, $\mathbb{E}[H_i] \le \sqrt{i}$, which implies $\mathbb{E}[H_i] = O(\sqrt{i})$.

To establish the lower bound, we analyze the expected absolute distance of the simple symmetric random walk. For an even number of steps $i = 2m$, the expectation is given by the identity
\[
\mathbb{E}[|S_{2m}|] = 2m \binom{2m}{m} 2^{-2m}.
\]
Applying Stirling's approximation, $n! \sim \sqrt{2\pi n} (n/e)^n$, to the binomial coefficient yields
\[
\binom{2m}{m} = \frac{(2m)!}{(m!)^2} \sim \frac{\sqrt{4\pi m} (2m/e)^{2m}}{(\sqrt{2\pi m} (m/e)^m)^2} = \frac{2^{2m}}{\sqrt{\pi m}}.
\]
Substituting this approximation into the expectation formula, we obtain
\[
\mathbb{E}[|S_{2m}|] \sim 2m \cdot \frac{1}{\sqrt{\pi m}} = \sqrt{\frac{4m}{\pi}} = \sqrt{\frac{2i}{\pi}}.
\]
A similar asymptotic analysis holds for odd $i$. Therefore, there exists a constant $c > 0$ such that $\mathbb{E}[H_i] \ge c\sqrt{i}$ for sufficiently large $i$, implying $\mathbb{E}[H_i] = \Omega(\sqrt{i})$. Combining the upper and lower bounds, we conclude $\mathbb{E}[H_i] = \Theta(\sqrt{i})$.

Next, we prove that $\mathbb{E}[H] = \Theta(\sqrt{n})$, where $H = \max_{1 \le i \le n} |S_i|$. For the lower bound, we note that the maximum of a sequence is bounded from below by its last element. Hence, $H \ge |S_n| = H_n$. By the monotonicity of the expectation operator and the result from the first part of the proof,
\[
\mathbb{E}[H] \ge \mathbb{E}[H_n] = \Theta(\sqrt{n}).
\]
Thus, $\mathbb{E}[H] = \Omega(\sqrt{n})$.

For the upper bound, we utilize martingale inequalities. The sequence $S_i$ is a martingale with respect to the natural filtration $\mathcal{F}_i = \sigma(X_1, \dots, X_i)$, as $\mathbb{E}[|S_i|] < \infty$ and $\mathbb{E}[S_{i+1} | \mathcal{F}_i] = S_i$. Since the function $g(x) = x^2$ is convex, the sequence $S_i^2$ is a submartingale. We apply Doob's $L^2$ maximal inequality, which states that for a non-negative submartingale $Z_i$, $\mathbb{E}[\max_{1 \le i \le n} Z_i] \le 4 \mathbb{E}[Z_n]$. Applying this to $S_i^2$, we have
\[
\mathbb{E}\left[\max_{1 \le i \le n} S_i^2\right] \le 4 \mathbb{E}[S_n^2].
\]
Observing that $H^2 = (\max_{1 \le i \le n} |S_i|)^2 = \max_{1 \le i \le n} S_i^2$ and recalling that $\mathbb{E}[S_n^2] = n$, we obtain
\[
\mathbb{E}[H^2] \le 4n.
\]
Finally, by Jensen's inequality for the concave square root function,
\[
\mathbb{E}[H] = \mathbb{E}[\sqrt{H^2}] \le \sqrt{\mathbb{E}[H^2]} \le \sqrt{4n} = 2\sqrt{n}.
\]
This establishes that $\mathbb{E}[H] = O(\sqrt{n})$. Combining the lower and upper bounds, we conclude that $\mathbb{E}[H] = \Theta(\sqrt{n})$.
\end{proof}

\subsubsection{Claude}

\begin{proof}
We prove both parts of the theorem separately.

\textbf{Part 1: $\mathbb{E}[H_i] = \Theta(\sqrt{i})$.}

Let $S_i$ denote the excess of heads over tails after $i$ flips, so $S_i = \sum_{j=1}^{i} X_j$ where $X_j \in \{-1, +1\}$ with equal probability $1/2$ each. Then $H_i = |S_i|$.

First, we establish that $\mathbb{E}[H_i] = \Omega(\sqrt{i})$. By Jensen's inequality applied to the convex function $f(x) = x^2$, we have
\[
(\mathbb{E}[H_i])^2 = (\mathbb{E}[|S_i|])^2 \leq \mathbb{E}[S_i^2].
\]
Since $\mathbb{E}[X_j] = 0$ and $\text{Var}(X_j) = 1$, and the $X_j$ are independent, we have
\[
\mathbb{E}[S_i^2] = \text{Var}(S_i) = i.
\]
Therefore, $\mathbb{E}[H_i] \geq \sqrt{i}/C$ for some constant $C > 0$, giving $\mathbb{E}[H_i] = \Omega(\sqrt{i})$.

Next, we establish the upper bound $\mathbb{E}[H_i] = O(\sqrt{i})$. Using the tail bound for the sum of independent random variables, for any $t > 0$, by Hoeffding's inequality,
\[
\mathbb{P}(|S_i| \geq t) \leq 2\exp\left(-\frac{t^2}{2i}\right).
\]
Using the identity $\mathbb{E}[H_i] = \int_0^\infty \mathbb{P}(H_i \geq t) \, dt$, we compute
\[
\mathbb{E}[H_i] = \int_0^\infty \mathbb{P}(|S_i| \geq t) \, dt \leq \int_0^\infty \min\left(1, 2\exp\left(-\frac{t^2}{2i}\right)\right) dt.
\]
Splitting the integral at $t = \sqrt{i}$ and using the substitution $u = t/\sqrt{i}$, we obtain
\[
\mathbb{E}[H_i] \leq \sqrt{i} + 2\sqrt{i}\int_1^\infty \exp\left(-\frac{u^2}{2}\right) du \leq C\sqrt{i}
\]
for some constant $C > 0$. Thus, $\mathbb{E}[H_i] = O(\sqrt{i})$.

Combining the lower and upper bounds, we conclude $\mathbb{E}[H_i] = \Theta(\sqrt{i})$.

\textbf{Part 2: $\mathbb{E}[H] = \Theta(\sqrt{n})$.}

Since $H = \max_{1 \leq i \leq n} H_i$ and $H_n \leq H$, we immediately have
\[
\mathbb{E}[H] \geq \mathbb{E}[H_n] = \Theta(\sqrt{n}),
\]
establishing the lower bound $\mathbb{E}[H] = \Omega(\sqrt{n})$.

For the upper bound, we use the union bound. For any $t > 0$,
\[
\mathbb{P}(H \geq t) = \mathbb{P}\left(\max_{1 \leq i \leq n} H_i \geq t\right) \leq \sum_{i=1}^{n} \mathbb{P}(H_i \geq t).
\]
Using Hoeffding's inequality, $\mathbb{P}(H_i \geq t) \leq 2\exp(-t^2/(2i))$. Therefore,
\[
\mathbb{P}(H \geq t) \leq \sum_{i=1}^{n} 2\exp\left(-\frac{t^2}{2i}\right) \leq 2n\exp\left(-\frac{t^2}{2n}\right)
\]
for $t \geq \sqrt{n}$, where we used that the sum is dominated by the largest term when $i = n$.

Computing the expectation using $\mathbb{E}[H] = \int_0^\infty \mathbb{P}(H \geq t) \, dt$, we split at $t = c\sqrt{n}$ for appropriately chosen constant $c > 0$:
\[
\mathbb{E}[H] \leq c\sqrt{n} + \int_{c\sqrt{n}}^\infty 2n\exp\left(-\frac{t^2}{2n}\right) dt.
\]
With the substitution $u = t/\sqrt{n}$, the integral becomes
\[
2n\sqrt{n}\int_c^\infty \exp\left(-\frac{u^2}{2}\right) du = O(\sqrt{n}),
\]
since the Gaussian tail integral is exponentially small.

Thus, $\mathbb{E}[H] = O(\sqrt{n})$, and combined with the lower bound, we conclude $\mathbb{E}[H] = \Theta(\sqrt{n})$.
\end{proof}

\subsection{Problem 7}\label{sec:problem_01_07}

\subsubsection{Problem Statement}

\begin{theorem}
There exists a positive constant $c$ such that for all sufficiently large $n$,
\[
\mathbb{E}[L(\pi)] \geq c \sqrt{n},
\]
where $\pi$ is chosen uniformly at random from the set of all permutations of $\{1, 2, \dots, n\}$ and $L(\pi)$ denotes the length of the longest increasing subsequence in $\pi$.
\end{theorem}

\subsubsection{GPT}

\subsubsection{Gemini} 

\begin{proof}
\textbf{Part (a)}
We aim to prove that $\mathbb{E}[L(\pi)] \geq c\sqrt{n}$ for some constant $c$. Let $D(\pi)$ denote the length of the longest decreasing subsequence of the permutation $\pi$. The Erd\H{o}s-Szekeres theorem states that for any sequence of $n$ distinct real numbers, the product of the length of the longest increasing subsequence and the length of the longest decreasing subsequence is at least $n$. Applying this to the permutation $\pi$, we have the inequality $L(\pi) D(\pi) \geq n$. Since the product of two positive numbers is at least $n$, their maximum must be at least $\sqrt{n}$. Therefore, for every $\pi$, the inequality $\max(L(\pi), D(\pi)) \geq \sqrt{n}$ holds.

Consider the reversal mapping $f: S_n \to S_n$ defined by $(f(\pi))_i = n + 1 - \pi_i$ for all $i \in \{1, \ldots, n\}$. This mapping is a bijection on the set of all permutations. A subsequence of indices $i_1 < i_2 < \ldots < i_k$ corresponds to an increasing subsequence in $\pi$ (i.e., $\pi_{i_1} < \pi_{i_2} < \ldots < \pi_{i_k}$) if and only if it corresponds to a decreasing subsequence in $f(\pi)$ (i.e., $n+1-\pi_{i_1} > n+1-\pi_{i_2} > \ldots > n+1-\pi_{i_k}$). Consequently, $L(\pi) = D(f(\pi))$ and $D(\pi) = L(f(\pi))$. Since $\pi$ is chosen uniformly at random from $S_n$, $f(\pi)$ is also uniformly distributed over $S_n$. Therefore, the random variables $L(\pi)$ and $D(\pi)$ are identically distributed.

By the linearity of expectation, $\mathbb{E}[L(\pi)] = \mathbb{E}[D(\pi)]$. We observe that for any non-negative random variables $X$ and $Y$, $\max(X, Y) \leq X + Y$. Taking expectations yields $\mathbb{E}[\max(L(\pi), D(\pi))] \leq \mathbb{E}[L(\pi) + D(\pi)] = 2\mathbb{E}[L(\pi)]$. Combining this with the Erd\H{o}s-Szekeres lower bound, we obtain $\sqrt{n} \leq \mathbb{E}[\max(L(\pi), D(\pi))] \leq 2\mathbb{E}[L(\pi)]$. Dividing by 2, we find $\mathbb{E}[L(\pi)] \geq \frac{1}{2}\sqrt{n}$. Thus, the statement holds with $c = 1/2$.

\textbf{Part (b)}
Yes, the bound is tight. To show this, we establish an upper bound of the form $\mathbb{E}[L(\pi)] \leq C\sqrt{n}$ for some constant $C$. Let $X_k$ be the random variable counting the number of increasing subsequences of length $k$ in $\pi$. There are $\binom{n}{k}$ ways to choose the indices of a subsequence of length $k$. For any fixed set of $k$ indices, the probability that the corresponding values in a random permutation form an increasing sequence is $1/k!$. By the linearity of expectation, we have $\mathbb{E}[X_k] = \binom{n}{k} \frac{1}{k!}$.

Using the standard bound $\binom{n}{k} \leq \frac{n^k}{k!}$, we obtain $\mathbb{E}[X_k] \leq \frac{n^k}{(k!)^2}$. We apply Markov's inequality to bound the probability that an increasing subsequence of length $k$ exists: $P(L(\pi) \geq k) = P(X_k \geq 1) \leq \mathbb{E}[X_k]$. Using the inequality $k! \geq (k/e)^k$, we derive the bound $\mathbb{E}[X_k] \leq \frac{n^k}{(k/e)^{2k}} = \left( \frac{e\sqrt{n}}{k} \right)^{2k}$.

We compute the expectation of $L(\pi)$ using the tail sum formula $\mathbb{E}[L(\pi)] = \sum_{k=1}^n P(L(\pi) \geq k)$. We split this sum at $k_0 = \lfloor e\sqrt{n} \rfloor + 1$. For $k < k_0$, we bound the probability by 1. For $k \geq k_0$, we use the derived upper bound.
\[
\mathbb{E}[L(\pi)] = \sum_{k=1}^{k_0-1} P(L(\pi) \geq k) + \sum_{k=k_0}^n P(L(\pi) \geq k) \leq (k_0 - 1) + \sum_{k=k_0}^n \left( \frac{e\sqrt{n}}{k} \right)^{2k}.
\]
Since $k_0 > e\sqrt{n}$, the ratio $\frac{e\sqrt{n}}{k}$ is strictly less than 1 for all terms in the second summation. The series is dominated by a geometric series which converges to a constant value independent of $n$ (specifically, the terms decay super-geometrically). Thus, the summation contributes a term of order $O(1)$. The first term is bounded by $e\sqrt{n}$. Therefore, $\mathbb{E}[L(\pi)] \leq e\sqrt{n} + O(1)$. Since we have shown $c\sqrt{n} \leq \mathbb{E}[L(\pi)] \leq C\sqrt{n}$, the lower bound is tight up to a constant factor.
\end{proof}

\subsubsection{Claude} 

\begin{proof}
We provide a complete proof for both parts of the theorem.

\textbf{Part (a):} We prove that $\mathbb{E}[L(\pi)] \geq c\sqrt{n}$ for some positive constant $c$.

We use the Erdős-Szekeres theorem as our fundamental tool. The Erdős-Szekeres theorem states that any permutation of length $n$ contains either an increasing subsequence of length at least $\lceil\sqrt{n}\rceil$ or a decreasing subsequence of length at least $\lceil\sqrt{n}\rceil$.

To prove this, suppose for contradiction that a permutation $\pi$ of length $n$ has both its longest increasing subsequence of length strictly less than $\lceil\sqrt{n}\rceil$ and its longest decreasing subsequence of length strictly less than $\lceil\sqrt{n}\rceil$. For each element $\pi(i)$ in the permutation, let $(a_i, d_i)$ be the ordered pair where $a_i$ is the length of the longest increasing subsequence ending at position $i$, and $d_i$ is the length of the longest decreasing subsequence ending at position $i$.

We claim that if $i \neq j$, then $(a_i, d_i) \neq (a_j, d_j)$. To see this, suppose without loss of generality that $i < j$. If $\pi(i) < \pi(j)$, then any increasing subsequence ending at position $i$ can be extended by appending $\pi(j)$, so $a_j > a_i$. If $\pi(i) > \pi(j)$, then any decreasing subsequence ending at position $i$ can be extended by appending $\pi(j)$, so $d_j > d_i$. In either case, $(a_i, d_i) \neq (a_j, d_j)$.

By our assumption, we have $a_i < \lceil\sqrt{n}\rceil$ and $d_i < \lceil\sqrt{n}\rceil$ for all $i$. Therefore, each pair $(a_i, d_i)$ takes values in the set $\{1, 2, \ldots, \lceil\sqrt{n}\rceil - 1\} \times \{1, 2, \ldots, \lceil\sqrt{n}\rceil - 1\}$, which has cardinality at most $(\lceil\sqrt{n}\rceil - 1)^2 < n$ for $n \geq 4$. Since we have $n$ distinct pairs but fewer than $n$ possible values, we reach a contradiction by the pigeonhole principle.

Therefore, every permutation of length $n$ must have either $L(\pi) \geq \lceil\sqrt{n}\rceil$ or a decreasing subsequence of length at least $\lceil\sqrt{n}\rceil$.

Now, for a random permutation $\pi$, let $L(\pi)$ denote the length of the longest increasing subsequence and $D(\pi)$ denote the length of the longest decreasing subsequence. By the Erdős-Szekeres theorem, we have $\max\{L(\pi), D(\pi)\} \geq \lceil\sqrt{n}\rceil$ for all permutations $\pi$.

By symmetry, since the distribution of a uniformly random permutation is the same as the distribution of its reverse, we have $\mathbb{E}[L(\pi)] = \mathbb{E}[D(\pi)]$. 

For any permutation $\pi$, we have $\max\{L(\pi), D(\pi)\} \geq \lceil\sqrt{n}\rceil$, which implies
\[L(\pi) + D(\pi) \geq 2\max\{L(\pi), D(\pi)\} \geq 2\lceil\sqrt{n}\rceil.\]

Taking expectations and using linearity of expectation:
\[\mathbb{E}[L(\pi)] + \mathbb{E}[D(\pi)] \geq 2\lceil\sqrt{n}\rceil.\]

Since $\mathbb{E}[L(\pi)] = \mathbb{E}[D(\pi)]$, we obtain:
\[2\mathbb{E}[L(\pi)] \geq 2\lceil\sqrt{n}\rceil,\]
which gives us $\mathbb{E}[L(\pi)] \geq \lceil\sqrt{n}\rceil \geq \sqrt{n}$.

Therefore, we can take $c = 1$, establishing that $\mathbb{E}[L(\pi)] \geq c\sqrt{n}$ for $c = 1$.

\textbf{Part (b):} The bound in part (a) is essentially tight up to constant factors.

The celebrated Ulam-Hammersley theorem, proven independently by Hammersley (1972) and later refined by Logan-Shepp (1977) and Vershik-Kerov (1977), establishes that
\[\lim_{n \to \infty} \frac{\mathbb{E}[L(\pi)]}{\sqrt{n}} = 2.\]

This shows that $\mathbb{E}[L(\pi)] \sim 2\sqrt{n}$ as $n \to \infty$. Therefore, the bound $\mathbb{E}[L(\pi)] \geq c\sqrt{n}$ is tight in the sense that the expectation grows as $\Theta(\sqrt{n})$, and no better order of growth can be achieved. The optimal constant is $c = 2$ asymptotically.

Thus, the answer to part (b) is: Yes, the bound is tight. The expected length of the longest increasing subsequence in a random permutation is asymptotically $2\sqrt{n}$, which matches the lower bound up to a constant factor.
\end{proof}

\subsection{Problem 8}\label{sec:problem_01_08}

\subsubsection{Problem Statement}

\begin{theorem}
An undirected graph $G$ together with two distinguished vertices $s$ and $t$ is given. An $s$–$t$ cut is a set of edges whose removal from $G$ disconnects $s$ from $t$; an $s$–$t$ cut of minimum cardinality is sought. As the algorithm proceeds, the vertex $s$ may be amalgamated into a new vertex as a result of an edge being contracted; this vertex is called the $s$-vertex (initially, the $s$-vertex is $s$ itself). Similarly, there is a $t$-vertex. As the contraction algorithm runs, an edge between the $s$-vertex and the $t$-vertex is never contracted.

(a) There exist graphs in which the probability that this algorithm finds an $s$–$t$ min-cut is exponentially small.

(b) The number of $s$–$t$ min-cuts in an instance can be exponentially large.
\end{theorem}

\subsubsection{GPT} 

\begin{proof}
(a) We are tasked with showing that there are graphs in which the probability that the contraction-based algorithm finds an $s$-$t$ min-cut is exponentially small. Consider a graph with a large number of distinct $s$-$t$ min-cuts, each of which is relatively small in size.

Let the graph $G = (V, E)$ have $k$ distinct $s$-$t$ min-cuts, each consisting of exactly one edge, such that removing any one of these edges disconnects $s$ from $t$. Suppose that each edge in the graph has an equal probability of being selected during the contraction process.

At each step of the contraction algorithm, edges are contracted randomly, and the process continues until only two vertices remain, which correspond to the $s$-vertex and $t$-vertex. For the algorithm to find an $s$-$t$ min-cut, it must preserve one of the min-cuts throughout the contraction process. However, since the graph contains $k$ distinct min-cuts, the probability that the algorithm finds a specific min-cut is proportional to the number of such cuts.

In the worst case, where each edge of the graph has the same probability of being contracted, the probability of preserving a given $s$-$t$ min-cut throughout the contraction process is at most $1/k$. Since the number of distinct $s$-$t$ min-cuts can be exponential in the size of the graph (specifically, in the number of vertices or edges), the probability that the algorithm finds any one specific $s$-$t$ min-cut is exponentially small in the size of the graph.

Thus, the probability that the algorithm finds an $s$-$t$ min-cut can be exponentially small.

(b) To determine how large the number of $s$-$t$ min-cuts in an instance can be, we observe that each $s$-$t$ min-cut corresponds to a set of edges whose removal disconnects $s$ from $t$. The number of distinct $s$-$t$ min-cuts is bounded by the number of ways to select such sets of edges.

In the worst-case scenario, the graph could contain an exponential number of distinct $s$-$t$ min-cuts. Specifically, in a graph with $n$ vertices, there can be up to $2^{n-2}$ distinct $s$-$t$ min-cuts. This upper bound arises because, in the most general case, each vertex (other than $s$ and $t$) can either be on one side of the cut or the other, leading to $2^{n-2}$ distinct ways to partition the remaining vertices into two disjoint sets. Each partition corresponds to a distinct $s$-$t$ min-cut.

Therefore, the maximum number of $s$-$t$ min-cuts in an instance can be $2^{n-2}$, where $n$ is the number of vertices in the graph.

\end{proof}

\subsubsection{Grok}

\subsubsection{Gemini}

\subsubsection{Claude} 

\begin{proof}
We prove both parts by construction.

\textbf{Proof of Part (a):}

We construct a family of graphs for which the success probability is exponentially small. Let $n \geq 3$ be an integer and define the graph $G_n = (V_n, E_n)$ as follows. The vertex set is $V_n = \{s, t, v_1, v_2, \ldots, v_{n-2}\}$ where $s$ and $t$ are the distinguished vertices. The edge set consists of a path from $s$ to $t$: specifically, we have edges $\{s, v_1\}, \{v_1, v_2\}, \ldots, \{v_{n-3}, v_{n-2}\}, \{v_{n-2}, t\}$, giving us a simple path of length $n-1$ connecting $s$ to $t$.

For this graph, the unique $s$-$t$ min-cut has capacity $1$, achieved by cutting any single edge on the path. The algorithm proceeds by contracting edges uniformly at random from the available edges (all edges except those directly between the current $s$-supervertex and $t$-supervertex).

We analyze the probability that the algorithm preserves the min-cut throughout the contraction process. Initially, there are $n-1$ edges in the graph. At each step, exactly one edge lies between the $s$-supervertex and the $t$-supervertex (the edges on the unique path connecting them). The algorithm must avoid contracting this critical edge at each step.

At the first step, there are $n-1$ edges total, and exactly $1$ of them connects the $s$-component to the $t$-component. However, since we contract edges uniformly at random from those not directly between $s$ and $t$, we must be more careful. Initially, there is no edge directly between $s$ and $t$, so all $n-1$ edges are available. The probability of not contracting an edge that would merge $s$ and $t$ into the same supervertex depends on the structure.

More precisely, consider that after $k$ contractions, we have $n-k$ supervertices remaining. As long as the $s$-supervertex and $t$-supervertex remain separated by a path, there is exactly one edge on the remaining path between them that, if contracted, would violate the constraint or merge the components incorrectly for finding the min-cut.

For the path graph, at each stage, let there be $m$ edges remaining. Exactly $1$ edge is critical (lies on the $s$-$t$ path in a way that its contraction brings $s$ and $t$ closer). The probability of contracting a non-critical edge is $\frac{m-1}{m}$. 

After the first contraction, we have $n-2$ edges. After the second, $n-3$ edges, and so on. The algorithm succeeds if it never merges $s$ and $t$ into the same supervertex prematurely. For a path, the probability of success is at most:
\[
\prod_{i=1}^{n-3} \frac{1}{n-i} = \frac{1}{(n-1)(n-2) \cdots 3} = \frac{2}{(n-1)!}
\]

Since $(n-1)! \geq 2^{n-2}$ for $n \geq 3$, the success probability is at most $\frac{2}{2^{n-2}} = 2^{3-n}$, which is exponentially small in $n = |V_n|$.

\textbf{Proof of Part (b):}

We construct a graph with exponentially many $s$-$t$ min-cuts. Let $k \geq 1$ be an integer and define the graph $H_k$ as follows. The vertex set is $V = \{s, t\} \cup \{u_i, v_i : i = 1, 2, \ldots, k\}$, so $|V| = 2k + 2$. 

For each $i \in \{1, 2, \ldots, k\}$, we add the four edges $\{s, u_i\}, \{s, v_i\}, \{u_i, t\}, \{v_i, t\}$. This creates $k$ parallel "diamond" structures, each consisting of two disjoint paths of length $2$ from $s$ to $t$: one path goes $s \to u_i \to t$ and the other goes $s \to v_i \to t$.

The minimum $s$-$t$ cut has capacity $2k$ (we must separate $s$ from $t$, requiring at least $2$ edges to be cut per diamond structure, for a total of at least $2k$ edges). 

An $s$-$t$ min-cut corresponds to choosing, for each diamond $i$, whether to cut the two edges incident to $u_i$ (namely $\{s, u_i\}$ and $\{u_i, t\}$) or the two edges incident to $v_i$ (namely $\{s, v_i\}$ and $\{v_i, t\}$). Each choice is independent for different values of $i$, giving us $2$ choices per diamond.

Therefore, the total number of distinct $s$-$t$ min-cuts is $2^k$. Since the number of vertices is $n = 2k + 2$, we have $k = \frac{n-2}{2}$, and thus the number of min-cuts is $2^{(n-2)/2} = \Omega(2^{n/2})$, which is exponentially large in $n$.

This completes the proofs of both parts.
\end{proof}

\subsection{Problem 9}\label{sec:problem_01_09}

\subsubsection{Problem Statement}

\begin{theorem}
The FIND algorithm described in Section~1.4 for selecting the $k$th smallest element of a set $S$ of $n$ elements finds the $k$th smallest element in $S$ in expected time $O(n)$.
\end{theorem}

\subsubsection{GPT} 

\begin{proof}
We are given a perfect source of i.i.d.\ unbiased bits and asked to generate samples from the distribution of the sum of two fair six-sided dice:
\[
\Pr\{S=s\}=\frac{f(s)}{36},\qquad s\in\{2,3,\dots,12\},
\]
where $f(2)=1,f(3)=2,f(4)=3,f(5)=4,f(6)=5,f(7)=6,f(8)=5,\dots,f(12)=1$.
We present several schemes ordered by increasing efficiency, analyze expected number of random bits and expected running time, and state the information-theoretic lower bound and the Knuth--Yao optimality guarantee.

\medskip
\noindent\textbf{Notation.} For a randomized algorithm that reads unbiased bits one by one until some stopping rule, we denote by $B$ the (random) total number of bits consumed and by $\mathbb{E}[B]$ its expectation.

\medskip
\noindent\textbf{Information lower bound.}  Let $p_s=\Pr\{S=s\}=f(s)/36$. The entropy of the sum distribution is
\[
H(S) \;=\; -\sum_{s=2}^{12} p_s\log_2 p_s
    \;=\; -\sum_{s=2}^{12} \frac{f(s)}{36}\log_2\!\Big(\frac{f(s)}{36}\Big)
    \approx 3.2744019193\ \text{bits}.
\]
Any algorithm that produces $S$ using unbiased bits must satisfy $\mathbb{E}[B]\ge H(S)$ (Shannon lower bound). Thus no scheme can have expected bit-cost below about $3.2744$ bits.

\medskip
\noindent\textbf{Scheme 1 — \emph{Naive:} simulate two dice separately.}
Generate two independent uniform integers $D_1,D_2\in\{1,\dots,6\}$ and output $S=D_1+D_2$. A simple way to generate a $1\ldots6$ value from unbiased bits is: read $k=3$ bits to obtain an integer $r\in\{0,\dots,7\}$; if $r\le 5$ output $r+1$, otherwise reject and repeat. Each three-bit trial has acceptance probability $6/8=3/4$, so the expected number of trials is $4/3$ and expected bits per die is $3\cdot(4/3)=4$. Thus the expected bit-cost for the sum is
\[
\mathbb{E}[B] = 2\times 4 = 8\ \text{bits}.
\]
Expected running time is $O(1)$ (constant expected number of bit-reads and constant time arithmetic). This scheme is simple but far from information-theoretically optimal (since $8\gg H(S)\approx 3.27$).

\medskip
\noindent\textbf{Scheme 2 — \emph{Direct uniform on 0..35} (fixed-length rejection).}
Observe that there are $36$ equiprobable ordered outcomes of two dice. If we can generate a uniform integer $U\in\{0,\dots,35\}$ we can map it to the sum distribution by setting $(D_1,D_2)=(U\div 6,U\bmod 6)$ (or by an appropriate bijection) and returning $S=D_1+D_2$. A straightforward bit-based method: read $k=6$ bits to obtain $r\in\{0,\dots,63\}$; if $r\le 35$ accept and map to $U=r$, otherwise reject and repeat. Acceptance probability is $36/64=9/16$, so expected trials $16/9$ and expected bits
\[
\mathbb{E}[B] = 6\cdot \frac{16}{9} \approx 10.666\ \text{bits},
\]
worse than Scheme~1. (One can improve this by using variable-length reuse of the leftover range 36--63 rather than discarding it entirely; that leads to the next scheme.)

\medskip
\noindent\textbf{Scheme 3 — \emph{Rejection with remainder reuse} (standard range-reduction).}
Generate $k$ bits to produce $r\in\{0,\dots,2^k-1\}$. If $r<n_0$ for some chosen $n_0$ with $36\mid n_0$ accept and set $U=r\bmod 36$. Otherwise replace $r$ by $r-n_0$ and attempt to reuse it as an index in a smaller range; this is the usual technique of iteratively reducing a leftover range instead of discarding it entirely (see, e.g., algorithms for fast uniform integer generation). For the specific case $36$, one convenient choice is $k=6$, $n_0=36$, and when $36\le r\le 63$ we map $r\mapsto r-36\in\{0,\dots,27\}$ and then draw 5 further bits to expand this into $\{0,\dots,2^t-1\}$ etc. This algorithm has strictly better constant factors than naive fixed-length rejection; its expected number of bits can be analyzed by solving a small recurrence for the expected cost of the residual ranges. Numerically it yields an expected bit-cost that is smaller than the naive 10.666 from Scheme~2 but still typically larger than the optimal Knuth–Yao bound below.

(The precise algebraic expected-cost depends on the chosen reuse strategy; the scheme is standard and practical implementations achieve modest improvements.)

\medskip
\noindent\textbf{Scheme 4.}
Knuth and Yao (1976) show how to convert any discrete distribution with rational probabilities into a binary prefix (decision) tree (a DDG-tree) that uses unbiased bits sequentially and stops at a leaf labeled by the sampled symbol. Applied to our 11-outcome distribution $\{2,\dots,12\}$ with probabilities $p_s=f(s)/36$ (which are rational with denominator $36$), the Knuth--Yao construction yields a randomized algorithm that reads unbiased bits and traverses a (possibly infinite) binary tree; the probability of landing at leaf $s$ equals $p_s$ and the expected number of bits consumed equals the expected depth of the leaf reached.

The fundamental performance guarantee from Knuth--Yao (and standard information-theoretic rounding arguments) is: for a binary source there exists a DDG-tree with
\[
\log_2 36 - \sum_{s} \frac{f(s)}{36}\log_2 f(s) \;=\; H(S)
\qquad\text{and}\qquad
\mathbb{E}[B] \le H(S) + 2.
\]
Specializing to our distribution we obtain the numeric bound
\[
H(S) \approx 3.2744019193 \quad\Longrightarrow\quad
\mathbb{E}[B] \le H(S) + 2 \lesssim 5.2744\ \text{bits}.
\]
Thus Scheme~4 attains expected bit-cost at most about $5.2744$ bits; practical constructions (implementations of the DDG-tree for these 11 rational probabilities) typically achieve expected cost noticeably below that bound (the additive ``$+2$'' is a uniform worst-case slack). In particular Scheme 4 is {\em asymptotically and practically} much more efficient than Schemes 1--3.

\medskip
\noindent\textbf{Remarks on running time and implementation.}
\begin{itemize}
  \item All schemes terminate almost surely and their expected running times are proportional to the expected number of bits read (the per-bit overhead is constant). Hence the expected CPU time is $O(\mathbb{E}[B])$ for a reasonable bit source and constant-time bit-operations.
  \item Scheme~1 is simplest to implement and requires only a tiny constant amount of state; Scheme~2 is also simple. Scheme~3 needs a bit more bookkeeping to reuse remainders. Scheme~4 (Knuth--Yao) requires building the DDG-tree for the given probabilities; the tree is small here (only 11 symbols, rational probabilities with small denominator 36), so the memory and preprocessing costs are trivial in practice, and sampling is extremely fast after the tree is built (just a sequence of bit-reads and pointer moves).
  \item Knuth--Yao is in fact optimal among all binary-decision (bit-by-bit) algorithms in the sense that it achieves the minimum possible expected depth for the distribution; hence no bit-by-bit algorithm can have expected bit-cost strictly smaller than that of the DDG-tree, and the DDG cost is within a small additive constant of the entropy lower bound.
\end{itemize}

\medskip
\noindent\textbf{Conclusion.}  Summarizing:
\begin{itemize}
  \item A very simple scheme (simulate two dice independently by 3-bit rejection per die) uses $\mathbb{E}[B]=8$ bits on average.
  \item Direct fixed-length rejection to generate $\{0,\dots,35\}$ via 6-bit trials uses about $10.666$ bits on average (worse).
  \item Rejection-with-remainder-reuse improves constants but remains above the Knuth--Yao bound.
  \item The Knuth--Yao DDG-tree yields an optimal bit-by-bit scheme with
    \[
    H(S)\approx 3.2744 \le \mathbb{E}[B] \le H(S)+2 \lesssim 5.2744\ \text{bits},
    \]
    and expected running time $O(1)$, so it is the recommended efficient scheme in terms of expected bit-usage and expected time.
\end{itemize}
This completes the construction and efficiency analysis of sampling schemes for the two-dice-sum distribution. \qed

\end{proof}

\subsubsection{Grok} 

\begin{proof}
We prove by strong induction on $n$ that for every $k$ with $1 \le k \le n$, the expected running time $T(n,k)$ of the FIND algorithm satisfies $T(n,k) \le c n$ for some constant $c > 0$ independent of $n$ and $k$.

For the base cases $n = 0$ and $n = 1$, the algorithm terminates in constant time (with $T(0,k) = 0$ vacuously), so $T(0,k) \le c \cdot 0$ and $T(1,1) = O(1) \le c \cdot 1$ hold for sufficiently large $c > 0$.

Now assume $n \ge 2$ and that the claim holds for all $m < n$ and all $1 \le j \le m$. The FIND algorithm selects a pivot uniformly at random from the $n$ elements, which is the $q$th smallest with equal probability $1/n$ for each $q = 1, \dots, n$. The partitioning step requires at most $d n$ time for some constant $d > 0$. If $q \ge k$, the algorithm recurses on the left subarray of $q-1$ elements to find the $k$th smallest there; otherwise, it recurses on the right subarray of $n-q$ elements to find the $(k-q)$th smallest there. In either case, the subproblem size is $S \in \{0, 1, \dots, n-1\}$.

For each fixed $i \in \{0, 1, \dots, n-1\}$, there is at most one $q \in \{1, \dots, n\}$ such that $S = i$ (specifically, $q = i+1$ if recursing left and $k \le i+1$, or $q = n-i$ if recursing right and $k > n-i$, but at most one of these conditions holds). Thus, $\Pr(S = i) \le 1/n$ for each $i$. Let $j_S$ denote the appropriate rank in the subproblem of size $S$. Then
\[
T(n,k) = d n + \E\left[ T(S, j_S) \right],
\]
and by the induction hypothesis,
\[
\E\left[ T(S, j_S) \right] = \sum_{i=0}^{n-1} \Pr(S = i) \, T(i, j_i) \le c \sum_{i=0}^{n-1} \Pr(S = i) \, i \le c \sum_{i=0}^{n-1} \frac{i}{n} = c \cdot \frac{1}{n} \sum_{i=0}^{n-1} i = c \cdot \frac{1}{n} \cdot \frac{(n-1)n}{2} = c \cdot \frac{n-1}{2},
\]
where $j_i$ is the rank corresponding to subproblem size $i$. Therefore,
\[
T(n,k) \le d n + c \cdot \frac{n-1}{2} \le d n + \frac{c n}{2} = \left( d + \frac{c}{2} \right) n.
\]
Choosing $c \ge 4d$ yields $d + c/2 \le c$, so $T(n,k) \le c n$. By induction, $T(n,k) = O(n)$ for all $n \ge 0$ and $1 \le k \le n$.
\end{proof}

\subsubsection{Gemini} 

\begin{proof}
Let $T(n)$ denote the maximum expected number of comparisons performed by the Find algorithm on a set $S$ of size $n$ to select the $k$th smallest element, for any $1 \leq k \leq n$. We aim to show that there exists a constant $C > 0$ such that $T(n) \leq Cn$ for all $n \geq 1$.

The algorithm operates by choosing a pivot element $p$ uniformly at random from $S$. The set $S$ is then partitioned into two subsets: $S_L = \{x \in S \mid x < p\}$ and $S_R = \{x \in S \mid x > p\}$. This partitioning step requires comparing every element in $S \setminus \{p\}$ to $p$, which takes $n-1$ comparisons. Let $c$ be a constant such that the total work for partitioning and overhead is bounded by $cn$.

Let $r$ be the rank of the pivot $p$ in the sorted order of $S$. Since $p$ is chosen uniformly at random, $r$ is uniformly distributed in $\{1, 2, \ldots, n\}$, so $\mathbb{P}(r=i) = \frac{1}{n}$ for each $i$. If $k = r$, the algorithm terminates. If $k < r$, the algorithm recurses on $S_L$, which has size $r-1$. If $k > r$, the algorithm recurses on $S_R$, which has size $n-r$, searching for the $(k-r)$th element.

To establish an upper bound on the expected time, we consider the worst-case scenario for the recursive step. The algorithm recurses on a subproblem of size $r-1$ or $n-r$. Regardless of the specific value of $k$, the size of the subproblem is at most $\max(r-1, n-r)$. Therefore, the expected running time satisfies the recurrence:
\[
T(n) \leq cn + \frac{1}{n} \sum_{i=1}^{n} T(\max(i-1, n-i)).
\]
We analyze the summation term. As $i$ ranges from $1$ to $n$, the term $\max(i-1, n-i)$ takes values in the range $[\lfloor \frac{n}{2} \rfloor, n-1]$. Specifically, for any $j$ such that $\lfloor \frac{n}{2} \rfloor \leq j \leq n-1$, the value $j$ appears at most twice in the sequence of maximums (once when the left side is larger, and once when the right side is larger). Thus, we can bound the sum:
\[
\sum_{i=1}^{n} T(\max(i-1, n-i)) \leq 2 \sum_{j=\lfloor n/2 \rfloor}^{n-1} T(j).
\]
We proceed by strong induction. For the base case, let $n_0$ be a small constant; $T(n) \leq Cn$ holds for sufficiently large $C$. Assume the inductive hypothesis $T(m) \leq Cm$ holds for all $m < n$. Substituting this into the recurrence yields:
\[
T(n) \leq cn + \frac{2}{n} \sum_{j=\lfloor n/2 \rfloor}^{n-1} Cj = cn + \frac{2C}{n} \sum_{j=\lfloor n/2 \rfloor}^{n-1} j.
\]
The arithmetic series sum is bounded as follows:
\[
\sum_{j=\lfloor n/2 \rfloor}^{n-1} j \leq \sum_{j=1}^{n-1} j - \sum_{j=1}^{\lfloor n/2 \rfloor - 1} j = \frac{(n-1)n}{2} - \frac{(\lfloor n/2 \rfloor - 1)\lfloor n/2 \rfloor}{2}.
\]
Using the inequality $\lfloor n/2 \rfloor \geq (n-1)/2$, we observe that the sum of the upper half of the integers up to $n$ is bounded by $\frac{3}{4}n^2$. More formally, $\sum_{j=\lfloor n/2 \rfloor}^{n-1} j < \frac{3}{8}n^2 + O(n)$. For the purpose of the asymptotic bound, it suffices to note that $\frac{2}{n} \sum_{j=\lfloor n/2 \rfloor}^{n-1} j \leq \frac{2}{n} \cdot \frac{3n^2}{8} = \frac{3n}{4}$.
Substituting this back into the inequality for $T(n)$:
\[
T(n) \leq cn + C \left( \frac{3n}{4} \right) = n \left( c + \frac{3}{4}C \right).
\]
To satisfy the inductive step $T(n) \leq Cn$, we require:
\[
c + \frac{3}{4}C \leq C \iff c \leq \frac{1}{4}C \iff C \geq 4c.
\]
By choosing $C = 4c$, the inductive step holds. Thus, $T(n) \leq 4cn$, which implies $T(n) = O(n)$.
\end{proof}

\subsubsection{Claude} 

\begin{proof}
We analyze the expected running time of the randomized selection algorithm (Find algorithm) that selects the $k$th smallest element from a set $S$ of $n$ elements.

The algorithm works as follows: it selects a pivot element uniformly at random from $S$, partitions $S$ into elements less than the pivot, equal to the pivot, and greater than the pivot, and recursively searches in the appropriate partition.

Let $T(n)$ denote the expected running time of the algorithm on an input of size $n$. The partitioning step takes $\Theta(n)$ time since we must compare each element with the pivot. Let $c \cdot n$ denote the time for partitioning, where $c$ is a positive constant.

When we select a pivot uniformly at random and partition around it, suppose the pivot has rank $i$ (meaning it is the $i$th smallest element). Then the partitioning produces a subproblem of size at most $\max(i-1, n-i)$. If $k \leq i-1$, we recurse on the left partition of size $i-1$. If $k > i$, we recurse on the right partition of size $n-i$.

The expected running time satisfies the recurrence:
\[
T(n) \leq cn + \frac{1}{n}\sum_{i=1}^{n} T(\max(i-1, n-i))
\]

Note that $\max(i-1, n-i) = i-1$ when $i > \lceil n/2 \rceil$ and $\max(i-1, n-i) = n-i$ when $i \leq \lceil n/2 \rceil$. By symmetry, we have:
\[
\frac{1}{n}\sum_{i=1}^{n} T(\max(i-1, n-i)) \leq \frac{2}{n}\sum_{i=\lceil n/2 \rceil}^{n} T(i-1) = \frac{2}{n}\sum_{j=\lceil n/2 \rceil - 1}^{n-1} T(j)
\]

where we substituted $j = i-1$. Thus:
\[
T(n) \leq cn + \frac{2}{n}\sum_{j=\lceil n/2 \rceil}^{n-1} T(j)
\]

We prove by induction that $T(n) \leq an$ for some constant $a > 0$ to be determined.

Base case: For small values of $n$, we can choose $a$ large enough so that $T(n) \leq an$ holds.

Inductive step: Assume $T(j) \leq aj$ for all $j < n$. Then:
\[
T(n) \leq cn + \frac{2}{n}\sum_{j=\lceil n/2 \rceil}^{n-1} aj = cn + \frac{2a}{n}\sum_{j=\lceil n/2 \rceil}^{n-1} j
\]

We compute:
\[
\sum_{j=\lceil n/2 \rceil}^{n-1} j = \sum_{j=1}^{n-1} j - \sum_{j=1}^{\lceil n/2 \rceil - 1} j = \frac{(n-1)n}{2} - \frac{(\lceil n/2 \rceil - 1)\lceil n/2 \rceil}{2}
\]

For simplicity, assume $n$ is even (the odd case is similar). Then $\lceil n/2 \rceil = n/2$ and:
\[
\sum_{j=n/2}^{n-1} j = \frac{(n-1)n}{2} - \frac{(n/2-1)(n/2)}{2} = \frac{n^2 - n - n^2/4 + n/2}{2} = \frac{3n^2 - n}{8}
\]

Therefore:
\[
T(n) \leq cn + \frac{2a}{n} \cdot \frac{3n^2 - n}{8} = cn + \frac{a(3n^2 - n)}{4n} = cn + \frac{a(3n - 1)}{4} \leq cn + \frac{3an}{4}
\]

For the induction to close, we need $cn + \frac{3an}{4} \leq an$, which simplifies to $cn \leq \frac{an}{4}$, or equivalently $c \leq \frac{a}{4}$. Choosing $a = 4c$ satisfies this inequality, giving us $T(n) \leq 4cn$.

Therefore, $T(n) = O(n)$, and the Find algorithm runs in expected time $O(n)$.
\end{proof}

\subsection{Problem 10}\label{sec:problem_01_10}

\subsubsection{Problem Statement}

\begin{theorem}
Consider the setting of Example 1, where $n$ sailors are each randomly and independently assigned to one of $n$ cabins, with each assignment equally likely. Let $P_n$ denote the probability that no sailor returns to her own cabin. Then,
\[
\lim_{n \to \infty} P_n = \frac{1}{e}.
\]
\end{theorem}

\subsubsection{GPT} 

\begin{proof}
We are tasked with proving that

\[
\lim_{n \to \infty} P_n = \frac{1}{e},
\]

where $P_n$ denotes the probability that no sailor returns to her own cabin when $n$ sailors are randomly assigned to cabins.

Let the sailors be labeled as $1, 2, \dots, n$ and the cabins as $1, 2, \dots, n$. The total number of possible assignments of sailors to cabins is $n!$. We are interested in the number of derangements of $n$ sailors, which is the number of ways to assign the sailors to cabins such that no sailor returns to her own cabin. Denote the number of derangements of $n$ elements as $D_n$. Thus, the probability $P_n$ is given by

\[
P_n = \frac{D_n}{n!}.
\]

We will now derive an asymptotic formula for $D_n$ and use it to compute the limit of $P_n$ as $n \to \infty$.

The number of derangements $D_n$ can be expressed using the inclusion-exclusion principle. The number of permutations where at least one sailor returns to her own cabin is the total number of permutations, $n!$, minus the number of permutations where one or more sailors return to their own cabins. By inclusion-exclusion, we have

\[
D_n = n! \left( 1 - \frac{1}{1!} + \frac{1}{2!} - \frac{1}{3!} + \cdots + (-1)^n \frac{1}{n!} \right).
\]

Thus, the probability $P_n$ becomes

\[
P_n = \frac{D_n}{n!} = 1 - \frac{1}{1!} + \frac{1}{2!} - \frac{1}{3!} + \cdots + (-1)^n \frac{1}{n!}.
\]

This is the partial sum of the alternating series

\[
\sum_{k=0}^{\infty} \frac{(-1)^k}{k!},
\]

which is known to converge to $\frac{1}{e}$ as $n \to \infty$. Therefore, we conclude that

\[
\lim_{n \to \infty} P_n = \frac{1}{e}.
\]

This completes the proof.
\end{proof}

\subsubsection{Grok} 

\begin{proof}
The number of derangements of $n$ objects, denoted $!n$, can be derived using the principle of inclusion-exclusion. Let $A_i$ be the set of permutations where the $i$-th sailor is in her own cabin, for $i=1,\dots,n$. The number of permutations with no fixed points is
\[
!n = n! - \sum_{i=1}^n |A_i| + \sum_{1 \leq i < j \leq n} |A_i \cap A_j| - \cdots + (-1)^n |A_1 \cap \cdots \cap A_n|.
\]
The size of the intersection of $k$ such sets is $(n-k)!$, and there are $\binom{n}{k}$ such intersections. Thus,
\[
!n = \sum_{k=0}^n (-1)^k \binom{n}{k} (n-k)!.
\]
The probability $P_n$ is then
\[
P_n = \frac{!n}{n!} = \sum_{k=0}^n (-1)^k \frac{\binom{n}{k} (n-k)!}{n!} = \sum_{k=0}^n (-1)^k \frac{1}{k!},
\]
since $\frac{\binom{n}{k} (n-k)!}{n!} = \frac{1}{k!}$.

Consider the Taylor series expansion of $e^{-1}$:
\[
e^{-1} = \sum_{k=0}^\infty \frac{(-1)^k}{k!}.
\]
The partial sum up to $n$ is $P_n$, so the remainder is
\[
R_n = e^{-1} - P_n = \sum_{k=n+1}^\infty \frac{(-1)^k}{k!}.
\]
This is an alternating series with terms decreasing in absolute value to $0$. The absolute value of the remainder satisfies
\[
|R_n| < \frac{1}{(n+1)!}.
\]
As $n \to \infty$, $\frac{1}{(n+1)!} \to 0$, so $|P_n - e^{-1}| \to 0$. Therefore,
\[
\lim_{n \to \infty} P_n = e^{-1} = \frac{1}{e}.
\]
\end{proof}

\subsubsection{Gemini} 

\begin{proof}
Let $S_n$ denote the set of all permutations of the set $\{1, 2, \dots, n\}$, representing the possible assignments of cabins to sailors. The total number of such assignments is the cardinality of the symmetric group, $|S_n| = n!$. Let $A_i$ be the subset of $S_n$ consisting of permutations where the $i$-th sailor returns to her own cabin. Formally, $A_i = \{ \sigma \in S_n : \sigma(i) = i \}$. We seek the probability $P_n$ that a randomly selected permutation is a derangement, meaning it has no fixed points. This is the probability of the event $(\bigcup_{i=1}^n A_i)^c$. Thus,
\[
P_n = \frac{|S_n \setminus \bigcup_{i=1}^n A_i|}{n!} = 1 - \frac{|\bigcup_{i=1}^n A_i|}{n!}.
\]
By the Principle of Inclusion-Exclusion, the cardinality of the union of the sets $A_i$ is given by
\[
\left| \bigcup_{i=1}^n A_i \right| = \sum_{k=1}^n (-1)^{k-1} \sum_{1 \le i_1 < i_2 < \dots < i_k \le n} \left| A_{i_1} \cap A_{i_2} \cap \dots \cap A_{i_k} \right|.
\]
Consider the intersection of $k$ specific sets $A_{i_1}, \dots, A_{i_k}$. An element in this intersection fixes the $k$ indices $i_1, \dots, i_k$. The remaining $n-k$ elements can be permuted in any fashion. Therefore, the size of the intersection is
\[
\left| \bigcap_{j=1}^k A_{i_j} \right| = (n-k)!.
\]
The number of ways to choose $k$ distinct indices from the set $\{1, \dots, n\}$ is given by the binomial coefficient $\binom{n}{k}$. Consequently, the inner summation simplifies to
\[
\sum_{1 \le i_1 < \dots < i_k \le n} (n-k)! = \binom{n}{k}(n-k)! = \frac{n!}{k!(n-k)!}(n-k)! = \frac{n!}{k!}.
\]
Substituting this back into the inclusion-exclusion formula, we obtain
\[
\left| \bigcup_{i=1}^n A_i \right| = \sum_{k=1}^n (-1)^{k-1} \frac{n!}{k!}.
\]
The number of derangements $D_n = |S_n| - |\bigcup_{i=1}^n A_i|$ is therefore
\[
D_n = n! - \sum_{k=1}^n (-1)^{k-1} \frac{n!}{k!} = n! + \sum_{k=1}^n (-1)^k \frac{n!}{k!} = \sum_{k=0}^n (-1)^k \frac{n!}{k!},
\]
where the term for $k=0$ is $(-1)^0 \frac{n!}{0!} = n!$. The probability $P_n$ is then
\[
P_n = \frac{D_n}{n!} = \frac{1}{n!} \sum_{k=0}^n (-1)^k \frac{n!}{k!} = \sum_{k=0}^n \frac{(-1)^k}{k!}.
\]
We now evaluate the limit of $P_n$ as $n \to \infty$. We recognize the expression for $P_n$ as the $n$-th partial sum of the Taylor series expansion for the exponential function $e^x$ evaluated at $x = -1$. The series expansion is given by
\[
e^x = \sum_{k=0}^\infty \frac{x^k}{k!}.
\]
Since this series converges for all $x \in \mathbb{R}$, we substitute $x = -1$ to find
\[
\lim_{n \to \infty} P_n = \lim_{n \to \infty} \sum_{k=0}^n \frac{(-1)^k}{k!} = \sum_{k=0}^\infty \frac{(-1)^k}{k!} = e^{-1}.
\]
Thus, $\lim_{n \to \infty} P_n = \frac{1}{e}$.
\end{proof}

\subsubsection{Claude} 

\begin{proof}
We interpret this problem as follows: each of $n$ sailors independently and uniformly at random chooses one of $n$ cabins. We want to find the probability that no sailor chooses her own cabin (a derangement-like property in this random setting).

Let $A_i$ denote the event that sailor $i$ returns to her own cabin (cabin $i$). We seek $P_n = P\left(\bigcap_{i=1}^n A_i^c\right)$, the probability that none of the sailors returns to her own cabin.

By the inclusion-exclusion principle, we have
$$P\left(\bigcap_{i=1}^n A_i^c\right) = \sum_{k=0}^{n} (-1)^k \sum_{|S|=k} P\left(\bigcap_{i \in S} A_i\right).$$

For any subset $S \subseteq \{1, 2, \ldots, n\}$ with $|S| = k$, the event $\bigcap_{i \in S} A_i$ means that each sailor $i \in S$ chooses cabin $i$. Since each sailor independently chooses uniformly from the $n$ cabins, we have
$$P\left(\bigcap_{i \in S} A_i\right) = \prod_{i \in S} P(A_i) = \prod_{i \in S} \frac{1}{n} = \frac{1}{n^k}.$$

The number of subsets $S$ of size $k$ is $\binom{n}{k}$. Therefore,
$$P_n = \sum_{k=0}^{n} (-1)^k \binom{n}{k} \frac{1}{n^k} = \sum_{k=0}^{n} \frac{(-1)^k}{k!} \cdot \frac{n!}{(n-k)! \cdot n^k}.$$

We can rewrite this as
$$P_n = \sum_{k=0}^{n} \frac{(-1)^k}{k!} \cdot \frac{n(n-1)\cdots(n-k+1)}{n^k} = \sum_{k=0}^{n} \frac{(-1)^k}{k!} \prod_{j=0}^{k-1} \frac{n-j}{n}.$$

For fixed $k$, as $n \to \infty$, we have
$$\prod_{j=0}^{k-1} \frac{n-j}{n} = \prod_{j=0}^{k-1} \left(1 - \frac{j}{n}\right) \to 1.$$

More precisely, for any fixed $k$,
$$\left|\prod_{j=0}^{k-1} \left(1 - \frac{j}{n}\right) - 1\right| \leq \sum_{j=0}^{k-1} \frac{j}{n} = \frac{k(k-1)}{2n} \to 0 \text{ as } n \to \infty.$$

Now, for any fixed $N$, we have
$$\left|P_n - \sum_{k=0}^{N} \frac{(-1)^k}{k!}\right| \leq \left|\sum_{k=0}^{N} \frac{(-1)^k}{k!}\left(\prod_{j=0}^{k-1} \frac{n-j}{n} - 1\right)\right| + \left|\sum_{k=N+1}^{n} \frac{(-1)^k}{k!} \prod_{j=0}^{k-1} \frac{n-j}{n}\right|.$$

The first term tends to $0$ as $n \to \infty$ for fixed $N$ by the above estimate. For the second term, note that $\prod_{j=0}^{k-1} \frac{n-j}{n} \leq 1$, so
$$\left|\sum_{k=N+1}^{n} \frac{(-1)^k}{k!} \prod_{j=0}^{k-1} \frac{n-j}{n}\right| \leq \sum_{k=N+1}^{\infty} \frac{1}{k!}.$$

Since $\sum_{k=0}^{\infty} \frac{1}{k!} = e$ converges, for any $\epsilon > 0$, we can choose $N$ large enough so that $\sum_{k=N+1}^{\infty} \frac{1}{k!} < \epsilon/2$, and then choose $n$ large enough so that the first term is less than $\epsilon/2$.

Therefore,
$$\lim_{n \to \infty} P_n = \sum_{k=0}^{\infty} \frac{(-1)^k}{k!} = e^{-1} = \frac{1}{e}.$$
\end{proof}

\subsection{Problem 11}\label{sec:problem_01_11}

\subsubsection{Problem Statement}

\begin{theorem}
\[ \mathsf{P} \subseteq \mathsf{RP} \subseteq \mathsf{NP} \subseteq \mathsf{PSPACE} \subseteq \mathsf{EXP} \subseteq \mathsf{NEXP}. \]
It is not known whether these inclusions are strict.
\end{theorem}

\subsubsection{GPT} 

\begin{proof}
Recall the standard definitions. Let $\mathsf{P}=\bigcup_{k\ge 1}\mathrm{TIME}(n^k)$, $\mathsf{EXP}=\bigcup_{k\ge 1}\mathrm{TIME}(2^{n^k})$, $\mathsf{NEXP}=\bigcup_{k\ge 1}\mathrm{NTIME}(2^{n^k})$, and $\mathsf{PSPACE}=\bigcup_{k\ge 1}\mathrm{DSPACE}(n^k)$. Let $\mathsf{NP}$ be the class of languages $L$ for which there are a polynomial $p$ and a deterministic polynomial-time machine $V$ such that for all $x$, $x\in L$ if and only if there exists $w\in\{0,1\}^{\le p(|x|)}$ with $V(x,w)=1$. Let $\mathsf{RP}$ be the class of languages $L$ for which there is a probabilistic polynomial-time machine $M$ and a polynomial $q$ such that for all $x$, if $x\in L$ then $\Pr_{r\in\{0,1\}^{q(|x|)}}[M(x;r)=1]\ge 1/2$, and if $x\notin L$ then $\Pr_{r}[M(x;r)=1]=0$.

We first prove $\mathsf{P}\subseteq\mathsf{RP}$. If $L\in\mathsf{P}$ is decided by a deterministic machine $D$ in time $n^k$, define the $\mathsf{RP}$ machine $M$ to ignore its random tape and output $D(x)$. Then for all $x\in L$, $M$ accepts with probability $1$, and for all $x\notin L$, $M$ accepts with probability $0$. Thus $L\in\mathsf{RP}$.

Next we prove $\mathsf{RP}\subseteq\mathsf{NP}$. Let $L\in\mathsf{RP}$ with witness machine $M$ and randomness length $q(n)$. Define $V(x,w)$ to simulate $M(x;w)$ deterministically and accept iff $M$ accepts. If $x\in L$, then at least one $w\in\{0,1\}^{q(|x|)}$ causes $M$ to accept, so $\exists w$ with $V(x,w)=1$. If $x\notin L$, then by one-sided error no $w$ causes acceptance, so no witness exists. The simulation is polynomial time since $M$ runs in polynomial time. Therefore $L\in\mathsf{NP}$.

We prove $\mathsf{NP}\subseteq\mathsf{PSPACE}$. Let $L\in\mathsf{NP}$ with verifier $V$ running in time $n^c$ and witness length bounded by $p(n)$. Decide $L$ deterministically by iterating over all $w\in\{0,1\}^{\le p(|x|)}$, simulating $V(x,w)$; accept iff some branch accepts. This naive search uses at most polynomial space: maintain a counter over $w$ (space $O(p(|x|))$) and the work tape of $V$ (space $O(|x|^c)$), reusing space between iterations. Hence the deterministic decider uses $|x|^{O(1)}$ space and $L\in\mathsf{PSPACE}$. Equivalently, since $\mathsf{NP}\subseteq\mathsf{NPSPACE}$ and by Savitch's theorem $\mathrm{NSPACE}(s)\subseteq\mathrm{DSPACE}(s^2)$ for $s\ge \log n$, taking $s(n)=n^{O(1)}$ yields $\mathsf{NP}\subseteq\mathsf{PSPACE}$.

We prove $\mathsf{PSPACE}\subseteq\mathsf{EXP}$. Let $L\in\mathsf{PSPACE}$ be decided by a deterministic $p$-space machine $D$ with $p(n)=n^k$. On inputs of length $n$, configurations of $D$ are determined by state, head positions (at most polynomially many cells are ever used), and the contents of the used tape cells. The total number of configurations is at most $N(n)=2^{O(p(n))}=2^{n^{O(1)}}$. If $D$ ever repeats a configuration, it would loop; since $D$ halts on all inputs, the computation length is at most $N(n)$. Therefore a deterministic exponential-time simulation can perform a graph search over the at most $N(n)$ configurations to determine acceptance, running in time $2^{n^{O(1)}}$. Hence $L\in\mathsf{EXP}$ and $\mathsf{PSPACE}\subseteq\mathsf{EXP}$.

We prove $\mathsf{EXP}\subseteq\mathsf{NEXP}$. For each $k$, $\mathrm{TIME}(2^{n^k})\subseteq\mathrm{NTIME}(2^{n^k})$ because nondeterministic time bounds subsume deterministic ones by ignoring nondeterminism. Taking the union over all $k$ yields $\mathsf{EXP}\subseteq\mathsf{NEXP}$.

Combining the inclusions, we obtain $\mathsf{P}\subseteq \mathsf{RP}\subseteq \mathsf{NP}\subseteq \mathsf{PSPACE}\subseteq \mathsf{EXP}\subseteq \mathsf{NEXP}$. Concerning strictness, no separation corresponding to any of these inclusions is currently proved; in particular, strictness at any of the first three inclusions would imply $\mathsf{P}\ne\mathsf{NP}$ or $\mathsf{NP}\ne\mathsf{PSPACE}$, and strictness at the last two would separate deterministic from nondeterministic exponential time, all of which remain open problems. This completes the proof of the asserted inclusions and the accompanying statement about the present state of knowledge.
\end{proof}

\subsubsection{Grok} 

\begin{proof}
We prove each inclusion separately.

First, we show $\mathsf{P} \subseteq \mathsf{RP}$. Let $L \in \mathsf{P}$. Then there exists a deterministic Turing machine $M$ deciding $L$ in time $O(n^k)$ for some constant $k$. Construct a randomized Turing machine $M'$ that simulates $M$ on input $x$ of length $n$, ignoring all random bits. Clearly, $M'$ runs in time $O(n^k)$ and accepts $x$ if and only if $M$ does, so $L \in \mathsf{RP}$.

Next, we show $\mathsf{RP} \subseteq \mathsf{NP}$. Let $L \in \mathsf{RP}$. Then there exists a randomized Turing machine $M$ deciding $L$ in time $O(n^k)$ for some constant $k$ such that for $x \in L$, $\Pr[M(x) \text{ accepts}] \geq 1/2$, and for $x \notin L$, $M(x)$ always rejects. Construct a nondeterministic Turing machine $N$ that on input $x$, nondeterministically guesses a sequence $r$ of $O(n^k)$ random bits (which takes $O(n^k)$ time) and simulates $M$ on $x$ using $r$ as the random tape. If the simulation accepts, $N$ accepts. For $x \in L$, there exists at least half of the possible $r$ that cause acceptance, so $N$ has an accepting path. For $x \notin L$, no $r$ causes acceptance, so $N$ rejects. Thus, $N$ decides $L$ in nondeterministic time $O(n^k)$, so $L \in \mathsf{NP}$.

Now, we show $\mathsf{NP} \subseteq \mathsf{PSPACE}$. Let $L \in \mathsf{NP}$. Then there exists a nondeterministic Turing machine $N$ deciding $L$ in time $O(n^k)$ for some constant $k$. By Savitch's theorem, nondeterministic space $O(\log^2 n)$ is contained in deterministic space $O(\log^2 n)$, but here we use the fact that nondeterministic time $O(n^k)$ implies nondeterministic space $O(n^k)$. To decide $L$ in deterministic space $O(n^{2k})$, we can use a recursive procedure to check the existence of an accepting computation path. Specifically, define a configuration of $N$ on input $x$ of length $n$ as the state, tape contents (of length at most $O(n^k)$), and head positions, which can be described in space $O(n^k)$. The number of configurations is at most exponential in $n^k$, but we simulate reachability using Savitch's method: to check if there is a path from initial configuration $c_0$ to an accepting configuration in at most $T = O(n^k)$ steps, we recursively compute whether there exists a midpoint configuration $c_m$ such that $c_0$ reaches $c_m$ in at most $T/2$ steps and $c_m$ reaches an accepting configuration in at most $T/2$ steps. The recursion depth is $O(\log T) = O(k \log n)$, and at each level, we enumerate all possible $c_m$ (space $O(n^k)$ to store one, and recursion stack uses $O(k^2 \log^2 n)$ space), but since we only need to store the current recursion level's $c_0$ and $T/2$, the total space is $O(n^k)$. Enumerating $c_m$ requires only temporary space $O(n^k)$, and the recursion reuses space. Thus, the total space is $O(n^k)$, so $L \in \mathsf{PSPACE}$.

To show $\mathsf{PSPACE} \subseteq \mathsf{EXP}$, let $L \in \mathsf{PSPACE}$. Then there exists a deterministic Turing machine $M$ deciding $L$ using space $O(n^k)$ for some constant $k$. On input $x$ of length $n$, the number of possible configurations is at most $n^{O(n^k)}$ (accounting for states, head positions, and tape contents up to length $O(n^k)$ with alphabet size constant). Starting from the initial configuration, $M$ can be simulated by breadth-first search over configurations, where each step takes time polynomial in the configuration size $O(n^k)$. Since the configuration graph has at most exponentially many nodes (in $n^k$) and out-degree constant, the total simulation time is exponential in $n^k$, i.e., $O(2^{O(n^k)})$. If $M$ accepts or rejects within this time without looping (as space-bounded machines on halting inputs halt), we detect acceptance or rejection accordingly. Thus, $L \in \mathsf{EXP}$.

Finally, we show $\mathsf{EXP} \subseteq \mathsf{NEXP}$. Let $L \in \mathsf{EXP}$. Then there exists a deterministic Turing machine $M$ deciding $L$ in time $O(2^{n^k})$ for some constant $k$. Construct a nondeterministic Turing machine $N$ that on input $x$ of length $n$, nondeterministically guesses the entire computation transcript of $M$ on $x$, which has length $O(2^{n^k})$, and verifies step-by-step that it is a valid accepting computation (if it reaches an accepting state). Verification takes nondeterministic time $O(2^{n^k})$, and for $x \in L$, there exists such a transcript, while for $x \notin L$, no valid accepting transcript exists. Thus, $L \in \mathsf{NEXP}$.

The question of whether these inclusions are strict remains open.
\end{proof}

\subsubsection{Gemini} 

\begin{proof}
We establish the chain of inclusions $P \subseteq RP \subseteq NP \subseteq PSPACE \subseteq EXP \subseteq NEXP$ by proving each containment sequentially.

First, we show $P \subseteq RP$. Let $L \in P$. By definition, there exists a deterministic Turing machine $M$ that decides $L$ in polynomial time. We can view $M$ as a probabilistic Turing machine that does not utilize its random bits. For any input $x \in L$, $M$ accepts with probability $1$, and for any $x \notin L$, $M$ accepts with probability $0$. Since $1 \geq 1/2$, $M$ satisfies the conditions for the class $RP$. Thus, $L \in RP$.

Second, we show $RP \subseteq NP$. Let $L \in RP$. There exists a probabilistic polynomial-time Turing machine $M$ such that if $x \in L$, the probability that $M$ accepts is at least $1/2$, and if $x \notin L$, the probability that $M$ accepts is $0$. We construct a nondeterministic Turing machine $N$ that simulates $M$. On input $x$, $N$ nondeterministically guesses a random string $r$ of length bounded by the running time of $M$, and then deterministically simulates $M$ on input $x$ using $r$ as the source of randomness. If $x \in L$, there exists at least one random string $r$ (in fact, at least half of all such strings) that causes $M$ to accept. Therefore, there exists a computation path of $N$ that accepts. If $x \notin L$, no random string causes $M$ to accept, so no computation path of $N$ accepts. Since the simulation runs in polynomial time, $L \in NP$.

Third, we show $NP \subseteq PSPACE$. Let $L \in NP$. There exists a nondeterministic Turing machine $N$ that decides $L$ in time $p(n)$ for some polynomial $p$, where $n$ is the input length. During any computation path, $N$ can visit at most $p(n)$ tape cells because the head moves at most one cell per step. Therefore, $N$ runs in nondeterministic space $p(n)$, implying $L \in NPSPACE$. By Savitch's Theorem, for any space-constructible function $f(n) \ge \log n$, $NSPACE(f(n)) \subseteq DSPACE(f(n)^2)$. Since $p(n)$ is a polynomial, $p(n)^2$ is also a polynomial. Thus, $NPSPACE \subseteq PSPACE$, and consequently $NP \subseteq PSPACE$.

Fourth, we show $PSPACE \subseteq EXP$. Let $L \in PSPACE$. There exists a deterministic Turing machine $M$ that decides $L$ using space $s(n)$, where $s(n)$ is a polynomial. Let $Q$ be the set of states of $M$ and $\Gamma$ be the tape alphabet. A configuration of $M$ is determined by the current state, the tape contents, and the head position. The number of possible tape contents is $|\Gamma|^{s(n)}$, the number of head positions is $s(n)$, and the number of states is $|Q|$. Thus, the total number of distinct configurations is bounded by $|Q| \cdot s(n) \cdot |\Gamma|^{s(n)}$, which is $2^{O(s(n))}$. Since $M$ halts on all inputs, it cannot repeat a configuration. Therefore, the running time of $M$ is bounded by the total number of configurations. Since $s(n)$ is polynomial, $2^{O(s(n))}$ is exponential. Thus, $M$ runs in deterministic exponential time, so $L \in EXP$.

Finally, we show $EXP \subseteq NEXP$. Let $L \in EXP$. There exists a deterministic Turing machine $M$ that decides $L$ in time $2^{p(n)}$ for some polynomial $p$. A deterministic machine is a special case of a nondeterministic machine where the transition relation allows exactly one next configuration for every current configuration. Therefore, $M$ is also a nondeterministic machine running in time $2^{p(n)}$. Hence, $L \in NEXP$.

Combining these results, we have proven the inclusions $P \subseteq RP \subseteq NP \subseteq PSPACE \subseteq EXP \subseteq NEXP$.
\end{proof}

\subsubsection{Claude} 

\begin{proof}
We prove each inclusion separately by showing that the definitions of the complexity classes immediately yield the stated containments.

\textbf{Inclusion $P \subseteq RP$:} Let $L \in P$. Then there exists a deterministic polynomial-time Turing machine $M$ that decides $L$. We construct a probabilistic polynomial-time Turing machine $M'$ that recognizes $L$ with one-sided error as follows: $M'$ ignores its random bits and simulates $M$ on the input. If $x \in L$, then $M$ accepts, so $M'$ accepts with probability $1$. If $x \notin L$, then $M$ rejects, so $M'$ rejects with probability $1 \geq 1/2$. Thus $L \in RP$, establishing $P \subseteq RP$.

\textbf{Inclusion $RP \subseteq NP$:} Let $L \in RP$. Then there exists a probabilistic polynomial-time Turing machine $M$ such that for all $x$, if $x \in L$ then $\Pr[M \text{ accepts } x] \geq 1/2$, and if $x \notin L$ then $\Pr[M \text{ accepts } x] = 0$. We construct a nondeterministic polynomial-time Turing machine $N$ for $L$ as follows: on input $x$, the machine $N$ nondeterministically guesses a random string $r$ of length polynomial in $|x|$ (the length needed by $M$), and simulates $M(x;r)$ where $r$ represents the random bits. If $x \in L$, then there exists at least one choice of $r$ (in fact, at least half of all choices) such that $M(x;r)$ accepts, so $N$ has an accepting path. If $x \notin L$, then $M(x;r)$ rejects for all $r$, so $N$ has no accepting paths. Therefore $L \in NP$, establishing $RP \subseteq NP$.

\textbf{Inclusion $NP \subseteq PSPACE$:} Let $L \in NP$. Then there exists a nondeterministic polynomial-time Turing machine $M$ and a polynomial $p$ such that $x \in L$ if and only if there exists a certificate $y$ with $|y| \leq p(|x|)$ such that $M(x,y)$ accepts in time at most $p(|x|)$. We construct a deterministic polynomial-space Turing machine $N$ for $L$ as follows: on input $x$, the machine $N$ systematically enumerates all possible certificates $y$ of length at most $p(|x|)$. For each such $y$, it simulates $M(x,y)$ and checks whether $M$ accepts. If any certificate leads to acceptance, $N$ accepts; otherwise $N$ rejects. The space required is polynomial in $|x|$: we need $O(p(|x|))$ space to store the current certificate $y$, and $O(p(|x|))$ space to simulate $M(x,y)$ which runs in polynomial time and hence uses polynomial space. Therefore $L \in PSPACE$, establishing $NP \subseteq PSPACE$.

\textbf{Inclusion $PSPACE \subseteq EXP$:} Let $L \in PSPACE$. Then there exists a deterministic Turing machine $M$ and a polynomial $p$ such that $M$ decides $L$ using at most $p(|x|)$ space on input $x$. The key observation is that any configuration of $M$ on input $x$ can be described using $O(p(|x|))$ bits (to encode the tape contents, head position, and state). Therefore, the total number of distinct configurations is at most $2^{O(p(|x|))}$. Since $M$ is deterministic and decides $L$, it must halt within $2^{O(p(|x|))}$ steps (otherwise it would repeat a configuration and loop forever, contradicting that it decides $L$). Thus we can simulate $M$ in deterministic time $2^{O(p(|x|))}$, which is exponential in $|x|$. Therefore $L \in EXP$, establishing $PSPACE \subseteq EXP$.

\textbf{Inclusion $EXP \subseteq NEXP$:} Let $L \in EXP$. Then there exists a deterministic Turing machine $M$ and a constant $c$ such that $M$ decides $L$ in time $O(2^{n^c})$ on inputs of length $n$. We construct a nondeterministic exponential-time Turing machine $N$ for $L$ as follows: $N$ simply simulates $M$ deterministically without using nondeterminism. The simulation runs in time $O(2^{n^c})$, which is exponential. Therefore $L \in NEXP$, establishing $EXP \subseteq NEXP$.

All five inclusions are thus established. The statement that it is not known whether these inclusions are strict reflects the current state of knowledge in complexity theory: while we believe the inclusions are proper (particularly $P \neq NP$ and $NP \neq EXP$), no proofs of strictness have been found.
\end{proof}

\subsection{Problem 12}\label{sec:problem_01_12}

\subsubsection{Problem Statement}

\begin{theorem}
$\mathrm{RP} \subseteq \mathrm{BPP} \subseteq \mathrm{PP}$.
It is not known whether these inclusions are strict.
\end{theorem}

\subsubsection{GPT} 

\begin{proof}
We will prove the inclusions $RP \subseteq BPP \subseteq PP$ step by step.

\textbf{Step 1:} Prove that $RP \subseteq BPP$.

Let $L \in RP$. By definition, there exists a probabilistic Turing machine $M$ that decides $L$ with the following properties:
\begin{itemize}
    \item If $x \in L$, then $\Pr[M(x) \text{ accepts}] \geq \frac{1}{2}$.
    \item If $x \notin L$, then $\Pr[M(x) \text{ accepts}] \leq \frac{1}{2}$.
\end{itemize}
Now, we show that $L$ is in $BPP$. A language $L$ is in $BPP$ if there exists a probabilistic Turing machine that decides $L$ with the following properties:
\begin{itemize}
    \item If $x \in L$, then $\Pr[M(x) \text{ accepts}] \geq \frac{2}{3}$.
    \item If $x \notin L$, then $\Pr[M(x) \text{ accepts}] \leq \frac{1}{3}$.
\end{itemize}
We can transform the machine $M$ for $L \in RP$ into a new machine $M'$ for $L \in BPP$ by repeating the computation of $M$ multiple times independently and taking a majority vote. Specifically, we run $M$ $k$ times independently, where $k$ is chosen large enough so that the probability of making an error in the majority vote is at most $\frac{1}{3}$.

Let $k$ be large enough such that the probability of $M$ making an error in the majority vote is less than $\frac{1}{3}$. This ensures that the new machine $M'$ behaves as required for $BPP$. Therefore, we have shown that $RP \subseteq BPP$.

\textbf{Step 2:} Prove that $BPP \subseteq PP$.

Let $L \in BPP$. By definition, there exists a probabilistic Turing machine $M$ that decides $L$ with the following properties:
\begin{itemize}
    \item If $x \in L$, then $\Pr[M(x) \text{ accepts}] \geq \frac{2}{3}$.
    \item If $x \notin L$, then $\Pr[M(x) \text{ accepts}] \leq \frac{1}{3}$.
\end{itemize}
To show that $L \in PP$, we use the fact that $PP$ is defined as the class of languages decidable by a probabilistic Turing machine $M$ such that for all $x$, the probability of accepting $x$ satisfies:
\[
\Pr[M(x) \text{ accepts}] \geq \frac{1}{2} \quad \text{if and only if} \quad x \in L.
\]
We can construct a probabilistic machine for $L$ that behaves as follows:
- If $x \in L$, the machine accepts with probability at least $\frac{2}{3}$, and if $x \notin L$, the machine accepts with probability at most $\frac{1}{3}$.
- This can be implemented by having $M$ run $k$ independent trials and taking the majority vote. By the Chernoff bound, the probability of error can be made arbitrarily small, thus ensuring that the machine behaves correctly in the PP sense. 

Thus, we have shown that $BPP \subseteq PP$.

\textbf{Conclusion:} We have shown both inclusions: $RP \subseteq BPP$ and $BPP \subseteq PP$. Therefore, $RP \subseteq BPP \subseteq PP$.

It is not known whether these inclusions are strict.
\end{proof}

\subsubsection{Grok} 

\begin{proof}
We first prove that $\mathrm{RP} \subseteq \mathrm{BPP}$. Let $L \in \mathrm{RP}$. Then there exists a probabilistic polynomial-time Turing machine $M$ such that, for all $x \in L$, $\Pr[M(x) = 1] \geq \frac{1}{2}$, and for all $x \notin L$, $\Pr[M(x) = 1] = 0$. We construct a new machine $M'$ that runs $M$ independently $k = 2$ times and accepts if at least one run accepts (i.e., outputs $1$). The running time of $M'$ is clearly polynomial in $|x|$, as $k$ is constant.

For $x \notin L$, each run of $M$ accepts with probability $0$, so $\Pr[M'(x) = 1] = 0 \leq \frac{1}{3}$. For $x \in L$, let $p = \Pr[M(x) = 1] \geq \frac{1}{2}$. The probability that $M'$ rejects is at most $\left(1 - \frac{1}{2}\right)^2 = \frac{1}{4}$, so $\Pr[M'(x) = 1] \geq 1 - \frac{1}{4} = \frac{3}{4} \geq \frac{2}{3}$. Thus, $M'$ decides $L$ with error probability at most $\frac{1}{3}$ on both sides, so $L \in \mathrm{BPP}$.

Next, we prove that $\mathrm{BPP} \subseteq \mathrm{PP}$. Let $L \in \mathrm{BPP}$. Then there exists a probabilistic polynomial-time Turing machine $M$ such that, for all $x \in L$, $\Pr[M(x) = 1] \geq \frac{2}{3}$, and for all $x \notin L$, $\Pr[M(x) = 1] \leq \frac{1}{3}$. Clearly, $\frac{2}{3} > \frac{1}{2}$ and $\frac{1}{3} < \frac{1}{2}$, so $M$ satisfies the acceptance condition for $\mathrm{PP}$. Thus, $L \in \mathrm{PP}$.

The question of whether these inclusions are strict remains open.
\end{proof}

\subsubsection{Gemini} 

\begin{proof}
We first recall the formal definitions of the complexity classes $RP$, $BPP$, and $PP$ in terms of probabilistic polynomial-time Turing machines (PPTMs). Let $L$ be a language and $M$ be a PPTM. We denote the probability that $M$ accepts input $x$ by $\Pr[M(x) = 1]$.

The class $RP$ consists of languages $L$ for which there exists a PPTM $M$ such that for all $x$:
\[
x \in L \implies \Pr[M(x) = 1] \ge \frac{1}{2}, \quad \text{and} \quad x \notin L \implies \Pr[M(x) = 1] = 0.
\]
The class $BPP$ consists of languages $L$ for which there exists a PPTM $M$ such that for all $x$:
\[
x \in L \implies \Pr[M(x) = 1] \ge \frac{2}{3}, \quad \text{and} \quad x \notin L \implies \Pr[M(x) = 1] \le \frac{1}{3}.
\]
The class $PP$ consists of languages $L$ for which there exists a PPTM $M$ such that for all $x$:
\[
x \in L \implies \Pr[M(x) = 1] > \frac{1}{2}, \quad \text{and} \quad x \notin L \implies \Pr[M(x) = 1] \le \frac{1}{2}.
\]

First, we prove that $RP \subseteq BPP$. Let $L \in RP$. By definition, there exists a PPTM $M$ such that if $x \in L$, $\Pr[M(x)=1] \ge 1/2$, and if $x \notin L$, $\Pr[M(x)=1] = 0$. We construct a new PPTM $M'$ that runs $M$ twice independently on input $x$ and accepts if at least one of the runs accepts. If $x \notin L$, $M$ never accepts, so $M'$ never accepts. Thus, $\Pr[M'(x)=1] = 0 \le 1/3$. If $x \in L$, the probability that $M$ rejects in a single run is at most $1/2$. Since the runs are independent, the probability that $M'$ rejects (i.e., both runs of $M$ reject) is at most $(1/2)^2 = 1/4$. Consequently, $\Pr[M'(x)=1] \ge 1 - 1/4 = 3/4$. Since $3/4 \ge 2/3$, the machine $M'$ satisfies the conditions for $BPP$. Therefore, $L \in BPP$.

Next, we prove that $BPP \subseteq PP$. Let $L \in BPP$. By definition, there exists a PPTM $M$ such that if $x \in L$, $\Pr[M(x)=1] \ge 2/3$, and if $x \notin L$, $\Pr[M(x)=1] \le 1/3$. We observe the relationship between these probability bounds and the threshold $1/2$. If $x \in L$, then $\Pr[M(x)=1] \ge 2/3 > 1/2$. If $x \notin L$, then $\Pr[M(x)=1] \le 1/3 \le 1/2$. Thus, the same machine $M$ satisfies the definition for the class $PP$ without modification. Therefore, $L \in PP$.

Combining these results, we have shown that $RP \subseteq BPP \subseteq PP$.
\end{proof}

\subsubsection{Claude} 

\begin{proof}
We prove the two inclusions separately.

\textbf{Claim 1: $RP \subseteq BPP$.}

Let $L \in RP$. By definition, there exists a probabilistic polynomial-time Turing machine $M$ such that for all inputs $x$:
\begin{itemize}
\item If $x \in L$, then $\Pr[M(x) = 1] \geq \frac{1}{2}$
\item If $x \notin L$, then $\Pr[M(x) = 1] = 0$
\end{itemize}

We construct a BPP machine $M'$ for $L$ as follows: on input $x$, machine $M'$ simulates $M(x)$ and outputs whatever $M$ outputs.

For this machine $M'$, we verify the BPP conditions:
\begin{itemize}
\item If $x \in L$, then $\Pr[M'(x) = 1] \geq \frac{1}{2} > \frac{2}{3} - \frac{1}{6}$, so the acceptance probability is at least $\frac{2}{3}$ can be achieved by standard amplification (running $M$ multiple times independently and accepting if any run accepts).
\item If $x \notin L$, then $\Pr[M'(x) = 1] = 0 < \frac{1}{3}$
\end{itemize}

More precisely, define $M'$ to run $M$ independently $k$ times and accept if any run accepts. Then for $x \in L$, we have $\Pr[M'(x) = 1] = 1 - \Pr[\text{all runs reject}] = 1 - (1 - \Pr[M(x) = 1])^k \geq 1 - (1/2)^k$, which exceeds $\frac{2}{3}$ for $k \geq 2$. For $x \notin L$, we have $\Pr[M'(x) = 1] = 0 < \frac{1}{3}$.

Therefore $L \in BPP$, establishing $RP \subseteq BPP$.

\textbf{Claim 2: $BPP \subseteq PP$.}

Let $L \in BPP$. By definition, there exists a probabilistic polynomial-time Turing machine $M$ such that for all inputs $x$:
\begin{itemize}
\item If $x \in L$, then $\Pr[M(x) = 1] \geq \frac{2}{3}$
\item If $x \notin L$, then $\Pr[M(x) = 1] \leq \frac{1}{3}$
\end{itemize}

By standard amplification techniques, we may assume without loss of generality that the acceptance probability is exponentially close to 1 for yes-instances and exponentially close to 0 for no-instances. Specifically, by running $M$ independently $m = \text{poly}(n)$ times and taking the majority vote, we can construct a machine $M'$ such that:
\begin{itemize}
\item If $x \in L$, then $\Pr[M'(x) = 1] \geq 1 - 2^{-n}$
\item If $x \notin L$, then $\Pr[M'(x) = 1] \leq 2^{-n}$
\end{itemize}

Now we construct a PP machine $M''$ for $L$: on input $x$, machine $M''$ simulates $M'(x)$ and outputs whatever $M'$ outputs.

For this machine $M''$, we verify the PP condition:
\begin{itemize}
\item If $x \in L$, then $\Pr[M''(x) = 1] \geq 1 - 2^{-n} > \frac{1}{2}$
\item If $x \notin L$, then $\Pr[M''(x) = 1] \leq 2^{-n} < \frac{1}{2}$
\end{itemize}

Therefore $L \in PP$, establishing $BPP \subseteq PP$.

Combining both claims, we conclude that $RP \subseteq BPP \subseteq PP$.

The statement that it is not known whether these inclusions are strict reflects the current state of knowledge in computational complexity theory and is not part of what requires proof.
\end{proof}

\subsection{Problem 13}\label{sec:problem_01_13}

\subsubsection{Problem Statement}

\begin{theorem}
$\mathbf{P} = \mathbf{coP}$ and $\mathbf{BPP} = \mathbf{coBPP}$.
\end{theorem}

\subsubsection{GPT} 

\begin{proof}
Recall that $\mathbf{PP}$ is the class of languages $L \subseteq \{0,1\}^\ast$ for which there exists a probabilistic polynomial-time Turing machine $M$ such that for every input $x$, if $x \in L$ then $\Pr[M(x)\text{ accepts}]> \tfrac{1}{2}$ and if $x \notin L$ then $\Pr[M(x)\text{ accepts}]\le \tfrac{1}{2}$. Let $p:\mathbb{N}\to\mathbb{N}$ be a polynomial bounding the number of random bits used by $M$ on inputs of length $n$. For each input $x$, $M$ has exactly $2^{p(|x|)}$ computation paths determined by its random tape. Let $\#\mathrm{acc}_M(x)$ and $\#\mathrm{rej}_M(x)$ denote, respectively, the number of accepting and rejecting paths of $M$ on input $x$, so that $\#\mathrm{acc}_M(x)+\#\mathrm{rej}_M(x)=2^{p(|x|)}$. Define the \emph{gap} function $g_M(x)=\#\mathrm{acc}_M(x)-\#\mathrm{rej}_M(x)$. It is standard that $g_M$ belongs to $\mathrm{GapP}$, the closure of $\#\mathrm{P}$ under subtraction, and that $x \in L$ if and only if $g_M(x)>0$. Indeed, $\Pr[M(x)\text{ accepts}]> \tfrac{1}{2}$ is equivalent to $\#\mathrm{acc}_M(x) > 2^{p(|x|)-1}$, which is equivalent (using $\#\mathrm{rej}_M(x)=2^{p(|x|)}-\#\mathrm{acc}_M(x)$) to $g_M(x)=\#\mathrm{acc}_M(x)-\#\mathrm{rej}_M(x)>0$. Thus one characterization of $\mathbf{PP}$ is
\[
\mathbf{PP}=\{\,L \subseteq \{0,1\}^\ast : \exists g \in \mathrm{GapP}\ \text{such that}\ L=\{x : g(x)>0\}\,\}.
\]
To prove $\mathbf{PP}=\mathrm{co}\text{-}\mathbf{PP}$, let $L \in \mathbf{PP}$ and let $g \in \mathrm{GapP}$ satisfy $L=\{x : g(x)>0\}$. Consider the function $h(x)=1-g(x)$. Since $\mathrm{GapP}$ is closed under subtraction by functions in $\mathrm{FP}$, and the constant function $1$ is in $\mathrm{FP}$, it follows that $h \in \mathrm{GapP}$. For every $x$ one has $h(x)>0$ if and only if $g(x)<1$. Because $g(x)$ is an integer, $g(x)<1$ is equivalent to $g(x)\le 0$. Therefore,
\[
\overline{L}=\{x : g(x)\le 0\}=\{x : h(x)>0\}.
\]
By the above characterization, $\overline{L} \in \mathbf{PP}$. Hence $\mathbf{PP}$ is closed under complement, i.e., $\mathbf{PP}=\mathrm{co}\text{-}\mathbf{PP}$.

We now prove $\mathbf{BPP}=\mathrm{co}\text{-}\mathbf{BPP}$. By definition, $L \in \mathbf{BPP}$ if there exists a probabilistic polynomial-time Turing machine $M$ and a constant $\epsilon<\tfrac{1}{2}$ (w.l.o.g. take $\epsilon=\tfrac{1}{3}$) such that for all inputs $x$, if $x \in L$ then $\Pr[M(x)\text{ accepts}] \ge 1-\epsilon$ and if $x \notin L$ then $\Pr[M(x)\text{ accepts}] \le \epsilon$. Define a probabilistic polynomial-time machine $N$ that, on input $x$, simulates $M(x)$ and outputs the logical negation of $M$'s output (i.e., $N$ accepts iff $M$ rejects). Then for every $x$,
\[
\Pr[N(x)\text{ accepts}] = 1 - \Pr[M(x)\text{ accepts}].
\]
Consequently, if $x \in \overline{L}$ then $x \notin L$ and thus $\Pr[M(x)\text{ accepts}] \le \epsilon$, whence $\Pr[N(x)\text{ accepts}] \ge 1-\epsilon$. If $x \notin \overline{L}$ then $x \in L$ and thus $\Pr[M(x)\text{ accepts}] \ge 1-\epsilon$, whence $\Pr[N(x)\text{ accepts}] \le \epsilon$. Therefore $N$ decides $\overline{L}$ with two-sided error at most $\epsilon$, and so $\overline{L} \in \mathbf{BPP}$. This establishes $\mathrm{co}\text{-}\mathbf{BPP} \subseteq \mathbf{BPP}$. The reverse inclusion follows identically by symmetry, proving $\mathbf{BPP}=\mathrm{co}\text{-}\mathbf{BPP}$.

Combining the two parts yields $\mathbf{PP}=\mathrm{co}\text{-}\mathbf{PP}$ and $\mathbf{BPP}=\mathrm{co}\text{-}\mathbf{BPP}$.
\end{proof}

\subsubsection{Grok} 

\begin{proof}
We first prove that $\mathbf{P} = \mathbf{coP}$.

To show $\mathbf{P} \subseteq \mathbf{coP}$, let $L \in \mathbf{P}$. Then there exists a deterministic Turing machine $M$ running in polynomial time such that for every input $x \in \Sigma^*$, $M$ accepts $x$ if and only if $x \in L$. Define a deterministic Turing machine $M'$ that, on input $x$, simulates $M$ on $x$ and accepts if and only if $M$ rejects $x$. Clearly, $M'$ also runs in polynomial time. Moreover, $M'$ accepts $x$ if and only if $x \notin L$, so the complement $\overline{L} \in \mathbf{P}$. Thus, $L \in \mathbf{coP}$.

To show $\mathbf{coP} \subseteq \mathbf{P}$, let $L \in \mathbf{coP}$. Then $\overline{L} \in \mathbf{P}$, so by the previous direction, $\overline{\overline{L}} = L \in \mathbf{coP}$ implies $L \in \mathbf{P}$ by applying the contrapositive argument symmetrically. Therefore, $\mathbf{P} = \mathbf{coP}$.

We now prove that $\mathbf{BPP} = \mathbf{coBPP}$.

To show $\mathbf{BPP} \subseteq \mathbf{coBPP}$, let $L \in \mathbf{BPP}$. Then there exists a probabilistic Turing machine $M$ running in polynomial time such that, for every $x \in \Sigma^*$,
\[
\Pr[M(x) = 1] \geq \frac{2}{3} \quad \text{if } x \in L, \qquad \Pr[M(x) = 1] \leq \frac{1}{3} \quad \text{if } x \notin L,
\]
where the probabilities are taken over the random bits of $M$. Define a probabilistic Turing machine $M'$ that, on input $x$, simulates $M$ on $x$ using the same random bits and accepts if and only if $M$ rejects $x$. Clearly, $M'$ also runs in polynomial time. For $x \in \overline{L}$ (i.e., $x \notin L$),
\[
\Pr[M'(x) = 1] = \Pr[M(x) = 0] = 1 - \Pr[M(x) = 1] \geq 1 - \frac{1}{3} = \frac{2}{3}.
\]
For $x \notin \overline{L}$ (i.e., $x \in L$),
\[
\Pr[M'(x) = 1] = \Pr[M(x) = 0] = 1 - \Pr[M(x) = 1] \leq 1 - \frac{2}{3} = \frac{1}{3}.
\]
Thus, $\overline{L} \in \mathbf{BPP}$, so $L \in \mathbf{coBPP}$.

To show $\mathbf{coBPP} \subseteq \mathbf{BPP}$, let $L \in \mathbf{coBPP}$. Then $\overline{L} \in \mathbf{BPP}$, so by the previous direction, $\overline{\overline{L}} = L \in \mathbf{coBPP}$ implies $L \in \mathbf{BPP}$ by the symmetric argument. Therefore, $\mathbf{BPP} = \mathbf{coBPP}$.
\end{proof}

\subsubsection{Gemini} 

\begin{proof}
To prove the theorem, we must show that the complexity classes $\mathbf{PP}$ and $\mathbf{BPP}$ are closed under complementation. That is, for any language $L$, if $L \in \mathbf{PP}$ then $\bar{L} \in \mathbf{PP}$, and if $L \in \mathbf{BPP}$ then $\bar{L} \in \mathbf{BPP}$.

\textbf{Part 1: $\mathbf{PP} = \text{co-}\mathbf{PP}$}

Let $L \in \mathbf{PP}$. By definition, there exists a probabilistic polynomial-time Turing machine $M$ with running time bounded by a polynomial $p(n)$ such that for all inputs $x \in \{0,1\}^*$:
\[
x \in L \iff \Pr[M(x) = 1] > \frac{1}{2}.
\]
Conversely, for the complement language $\bar{L}$:
\[
x \in \bar{L} \iff x \notin L \iff \Pr[M(x) = 1] \le \frac{1}{2}.
\]
Let $M_{\text{inv}}$ be a machine that simulates $M$ on input $x$ and returns $1 - M(x)$ (accepts if $M$ rejects, and rejects if $M$ accepts). The acceptance probability of $M_{\text{inv}}$ is:
\[
\Pr[M_{\text{inv}}(x) = 1] = 1 - \Pr[M(x) = 1].
\]
Substituting the conditions for membership in $\bar{L}$:
\begin{align*}
x \in \bar{L} &\implies \Pr[M(x) = 1] \le \frac{1}{2} \implies \Pr[M_{\text{inv}}(x) = 1] \ge \frac{1}{2}, \\
x \notin \bar{L} &\implies \Pr[M(x) = 1] > \frac{1}{2} \implies \Pr[M_{\text{inv}}(x) = 1] < \frac{1}{2}.
\end{align*}
Since $M$ runs in time $p(n)$, the total number of computation paths is $2^{p(n)}$. The probability $\Pr[M(x)=1]$ is a value of the form $k/2^{p(n)}$ for some integer $k$. Therefore, the strict inequality $\Pr[M_{\text{inv}}(x) = 1] < 1/2$ implies:
\[
x \notin \bar{L} \implies \Pr[M_{\text{inv}}(x) = 1] \le \frac{1}{2} - \frac{1}{2^{p(n)}}.
\]
To show $\bar{L} \in \mathbf{PP}$, we must construct a machine $M'$ such that $x \in \bar{L} \implies \Pr[M'(x)=1] > 1/2$ and $x \notin \bar{L} \implies \Pr[M'(x)=1] \le 1/2$. We construct $M'$ as follows:
\begin{enumerate}
    \item Generate a random string $r \in \{0,1\}^{p(n)+1}$.
    \item If $r = 0^{p(n)+1}$, accept immediately.
    \item Otherwise, run $M_{\text{inv}}(x)$.
\end{enumerate}
Let $K = 2^{p(n)+1}$. The probability that $M'$ accepts is:
\[
\Pr[M'(x) = 1] = \frac{1}{K} \cdot 1 + \left(1 - \frac{1}{K}\right) \Pr[M_{\text{inv}}(x) = 1].
\]
We analyze the two cases:
\begin{itemize}
    \item If $x \in \bar{L}$, then $\Pr[M_{\text{inv}}(x) = 1] \ge 1/2$. Thus:
    \[
    \Pr[M'(x) = 1] \ge \frac{1}{K} + \frac{K-1}{K} \cdot \frac{1}{2} = \frac{2 + K - 1}{2K} = \frac{K+1}{2K} = \frac{1}{2} + \frac{1}{2K} > \frac{1}{2}.
    \]
    \item If $x \notin \bar{L}$, then $\Pr[M_{\text{inv}}(x) = 1] \le \frac{1}{2} - \frac{1}{2^{p(n)}} = \frac{1}{2} - \frac{2}{K}$. Thus:
    \[
    \Pr[M'(x) = 1] \le \frac{1}{K} + \frac{K-1}{K} \left(\frac{1}{2} - \frac{2}{K}\right) = \frac{1}{K} + \frac{K-1}{2K} - \frac{2(K-1)}{K^2}.
    \]
    Simplifying the expression:
    \[
    \frac{2K + K(K-1) - 4(K-1)}{2K^2} = \frac{2K + K^2 - K - 4K + 4}{2K^2} = \frac{K^2 - 3K + 4}{2K^2} = \frac{1}{2} - \frac{3K-4}{2K^2}.
    \]
    Since $p(n) \ge 1$, $K \ge 4$, so $3K - 4 > 0$. Therefore, $\Pr[M'(x) = 1] < 1/2$.
\end{itemize}
Thus, $M'$ decides $\bar{L}$ according to the definition of $\mathbf{PP}$. Hence $\bar{L} \in \mathbf{PP}$, and $\mathbf{PP} = \text{co-}\mathbf{PP}$.

\textbf{Part 2: $\mathbf{BPP} = \text{co-}\mathbf{BPP}$}

Let $L \in \mathbf{BPP}$. By definition, there exists a probabilistic polynomial-time Turing machine $M$ such that:
\begin{align*}
x \in L &\implies \Pr[M(x) = 1] \ge \frac{2}{3}, \\
x \notin L &\implies \Pr[M(x) = 1] \le \frac{1}{3}.
\end{align*}
We construct a machine $M'$ for $\bar{L}$ that runs $M(x)$ and outputs $1 - M(x)$. The acceptance probability of $M'$ is $\Pr[M'(x)=1] = 1 - \Pr[M(x)=1]$.
\begin{itemize}
    \item If $x \in \bar{L}$, then $x \notin L$. Therefore $\Pr[M(x)=1] \le 1/3$, which implies:
    \[
    \Pr[M'(x)=1] = 1 - \Pr[M(x)=1] \ge 1 - \frac{1}{3} = \frac{2}{3}.
    \]
    \item If $x \notin \bar{L}$, then $x \in L$. Therefore $\Pr[M(x)=1] \ge 2/3$, which implies:
    \[
    \Pr[M'(x)=1] = 1 - \Pr[M(x)=1] \le 1 - \frac{2}{3} = \frac{1}{3}.
    \]
\end{itemize}
The machine $M'$ satisfies the bounded-error probability requirements for deciding $\bar{L}$. Thus, $\bar{L} \in \mathbf{BPP}$, and $\mathbf{BPP} = \text{co-}\mathbf{BPP}$.
\end{proof}

\subsubsection{Claude} 

\begin{proof}
We prove both statements separately.

\textbf{Part 1: $\mathbf{PP} = \mathrm{co}\text{-}\mathbf{PP}$}

Recall that a language $L$ is in $\mathbf{PP}$ if there exists a polynomial-time probabilistic Turing machine $M$ such that for all inputs $x$:
\begin{itemize}
\item If $x \in L$, then $\Pr[M(x) \text{ accepts}] > \frac{1}{2}$
\item If $x \notin L$, then $\Pr[M(x) \text{ accepts}] \leq \frac{1}{2}$
\end{itemize}

To show $\mathbf{PP} = \mathrm{co}\text{-}\mathbf{PP}$, it suffices to prove that $\mathbf{PP} \subseteq \mathrm{co}\text{-}\mathbf{PP}$ since closure under complement is symmetric.

Let $L \in \mathbf{PP}$, and let $M$ be the corresponding polynomial-time probabilistic Turing machine. We construct a machine $M'$ for $\overline{L}$ as follows: $M'(x)$ simulates $M(x)$ and accepts if and only if $M(x)$ rejects.

For any input $x$, we have $\Pr[M'(x) \text{ accepts}] = \Pr[M(x) \text{ rejects}] = 1 - \Pr[M(x) \text{ accepts}]$.

Now we verify the $\mathbf{PP}$ conditions for $M'$ deciding $\overline{L}$:
\begin{itemize}
\item If $x \in \overline{L}$, then $x \notin L$, so $\Pr[M(x) \text{ accepts}] \leq \frac{1}{2}$. Therefore, $\Pr[M'(x) \text{ accepts}] = 1 - \Pr[M(x) \text{ accepts}] \geq 1 - \frac{1}{2} = \frac{1}{2}$. 
\end{itemize}

However, we need strict inequality. If $\Pr[M(x) \text{ accepts}] < \frac{1}{2}$, then $\Pr[M'(x) \text{ accepts}] > \frac{1}{2}$, which is what we need. If $\Pr[M(x) \text{ accepts}] = \frac{1}{2}$ exactly, we modify our construction.

The key observation is that for a probabilistic Turing machine running in time $p(n)$ with $r(n)$ random bits, the acceptance probability is of the form $\frac{k}{2^{r(n)}}$ for some integer $k$ with $0 \leq k \leq 2^{r(n)}$. Thus, the gap from $\frac{1}{2}$ is at least $\frac{1}{2^{r(n)+1}}$ when the probability is not exactly $\frac{1}{2}$.

We modify $M'$ to first flip an additional independent random bit $b$. If $M(x)$ rejects, $M'$ accepts. If $M(x)$ accepts and $b = 0$, then $M'$ accepts; otherwise $M'$ rejects. This gives:
\[
\Pr[M'(x) \text{ accepts}] = \Pr[M(x) \text{ rejects}] + \frac{1}{2}\Pr[M(x) \text{ accepts}] = 1 - \frac{1}{2}\Pr[M(x) \text{ accepts}]
\]

If $x \in \overline{L}$, then $\Pr[M(x) \text{ accepts}] \leq \frac{1}{2}$, so $\Pr[M'(x) \text{ accepts}] \geq 1 - \frac{1}{4} = \frac{3}{4} > \frac{1}{2}$.

If $x \notin \overline{L}$, then $x \in L$, so $\Pr[M(x) \text{ accepts}] > \frac{1}{2}$, which gives $\Pr[M'(x) \text{ accepts}] < 1 - \frac{1}{4} = \frac{3}{4}$. But we need $\Pr[M'(x) \text{ accepts}] \leq \frac{1}{2}$. This approach fails.

Instead, we use the correct construction: $M'(x)$ simply complements the output of $M(x)$. Since $x \in \overline{L} \iff x \notin L$, we have:
\begin{itemize}
\item If $x \in \overline{L}$, then $\Pr[M(x) \text{ accepts}] \leq \frac{1}{2}$, so $\Pr[M'(x) \text{ accepts}] \geq \frac{1}{2}$.
\item If $x \notin \overline{L}$, then $\Pr[M(x) \text{ accepts}] > \frac{1}{2}$, so $\Pr[M'(x) \text{ accepts}] < \frac{1}{2}$.
\end{itemize}

The issue is when $\Pr[M'(x) \text{ accepts}] = \frac{1}{2}$ exactly. But this occurs precisely when $x \notin L$, which means we still satisfy the $\mathbf{PP}$ definition for $\overline{L}$. Therefore, $\overline{L} \in \mathbf{PP}$, so $L \in \mathrm{co}\text{-}\mathbf{PP}$. This shows $\mathbf{PP} \subseteq \mathrm{co}\text{-}\mathbf{PP}$, and by symmetry, $\mathbf{PP} = \mathrm{co}\text{-}\mathbf{PP}$.

\textbf{Part 2: $\mathbf{BPP} = \mathrm{co}\text{-}\mathbf{BPP}$}

A language $L$ is in $\mathbf{BPP}$ if there exists a polynomial-time probabilistic Turing machine $M$ such that for all inputs $x$:
\begin{itemize}
\item If $x \in L$, then $\Pr[M(x) \text{ accepts}] \geq \frac{2}{3}$
\item If $x \notin L$, then $\Pr[M(x) \text{ accepts}] \leq \frac{1}{3}$
\end{itemize}

Let $L \in \mathbf{BPP}$ with machine $M$. Define $M'(x)$ to complement the output of $M(x)$. Then:
\begin{itemize}
\item If $x \in \overline{L}$, then $x \notin L$, so $\Pr[M(x) \text{ accepts}] \leq \frac{1}{3}$, which gives $\Pr[M'(x) \text{ accepts}] \geq \frac{2}{3}$.
\item If $x \notin \overline{L}$, then $x \in L$, so $\Pr[M(x) \text{ accepts}] \geq \frac{2}{3}$, which gives $\Pr[M'(x) \text{ accepts}] \leq \frac{1}{3}$.
\end{itemize}

Therefore, $M'$ decides $\overline{L}$ with the required error probabilities, so $\overline{L} \in \mathbf{BPP}$. This shows $\mathbf{BPP} \subseteq \mathrm{co}\text{-}\mathbf{BPP}$, and by symmetry, $\mathbf{BPP} = \mathrm{co}\text{-}\mathbf{BPP}$.
\end{proof}

\subsection{Problem 14}\label{sec:problem_01_14}

\subsubsection{Problem Statement}

\begin{theorem}
$\mathsf{NP} \subseteq \mathsf{PP} \subseteq \mathsf{PSPACE}$.
\end{theorem}

\subsubsection{GPT} 

\begin{proof}
Let $\mathsf{PP}$ denote the class of languages $L$ for which there exists a probabilistic polynomial-time Turing machine $M$ and a polynomial $p$ such that on every input $x$ of length $n$, $M$ runs in at most $p(n)$ steps, uses at most $p(n)$ unbiased random bits, and satisfies $\Pr[M(x)\text{ accepts}]>1/2$ if and only if $x\in L$.

First, $\mathsf{NP}\subseteq\mathsf{PP}$. Let $L\in\mathsf{NP}$. Then there exist a polynomial $q$ and a predicate $R(x,y)$ decidable in time $\mathrm{poly}(|x|)$ such that $x\in L$ if and only if there exists $y\in\{0,1\}^{q(|x|)}$ with $R(x,y)=1$. Define a probabilistic polynomial-time machine $M$ that on input $x$ samples $Y$ uniformly from $\{0,1\}^{q(|x|)}$, computes $R(x,Y)$, and accepts if $R(x,Y)=1$; if $R(x,Y)=0$, $M$ flips one additional fair coin and accepts iff it lands heads. Let $A(x)=|\{y\in\{0,1\}^{q(|x|)}:R(x,y)=1\}|$. The acceptance probability of $M$ on $x$ equals
\[
\Pr[M(x)\text{ accepts}]=\frac{A(x)}{2^{q(|x|)}}+\left(1-\frac{A(x)}{2^{q(|x|)}}\right)\cdot\frac{1}{2}
=\frac{1}{2}+\frac{A(x)}{2^{q(|x|)+1}}.
\]
Hence $\Pr[M(x)\text{ accepts}]>1/2$ if and only if $A(x)\ge 1$, which holds if and only if $x\in L$. Therefore $L\in\mathsf{PP}$ and $\mathsf{NP}\subseteq\mathsf{PP}$.

Second, $\mathsf{PP}\subseteq\mathsf{PSPACE}$. Let $L\in\mathsf{PP}$ and let $M$ be a probabilistic polynomial-time machine deciding $L$ with time bound $p(n)$. Without loss of generality, by padding we may assume that on all inputs of length $n$, $M$ uses exactly $m(n)\le p(n)$ random bits and halts in exactly $p(n)$ steps, where any premature halting is replaced by dummy steps that do not change the final decision. Equivalently, there is a nondeterministic polynomial-time machine $N$ whose nondeterministic choices encode the $m(n)$ coin flips of $M$ such that each computation path of $N$ corresponds to a distinct setting of the random tape of $M$, and $N$ accepts a path if and only if $M$ accepts on the corresponding randomness. Let $A(x)$ denote the number of accepting computation paths of $N$ on $x$ and $T(x)=2^{m(|x|)}$ the total number of paths. Then $x\in L$ if and only if $A(x)>T(x)/2$.

We claim that the predicate $A(x)>T(x)/2$ can be decided in polynomial space. Consider the (finite) computation tree $\mathcal{T}_x$ of $N$ on input $x$, whose depth is at most $p(|x|)$ and whose fan-out is at most a constant. Define by structural recursion a function $C(u,t)$ that outputs the number of accepting leaves reachable from configuration $u$ within at most $t$ steps in $\mathcal{T}_x$. If $u$ is an accepting halting configuration, set $C(u,t)=1$ for all $t\ge 0$; if $u$ is a rejecting halting configuration, set $C(u,t)=0$; if $t=0$ and $u$ is nonhalting, set $C(u,0)=0$; otherwise, if from $u$ the machine deterministically moves to $u'$ in one step, set $C(u,t)=C(u',t-1)$; if $u$ has two successor configurations $u_0,u_1$ arising from a nondeterministic (equivalently, random-bit) choice, set $C(u,t)=C(u_0,t-1)+C(u_1,t-1)$. The value $A(x)$ equals $C(u_0,p(|x|))$ where $u_0$ is the start configuration on input $x$.

The above recursion can be evaluated by a depth-first traversal using only polynomial space: each recursive call stores a configuration description and a counter of size $O(p(|x|))$ bits, the recursion depth is at most $p(|x|)$, and intermediate additions involve integers of at most $m(|x|)\le p(|x|)$ bits. Thus $A(x)$ can be computed exactly in space $\mathrm{poly}(|x|)$. Since $T(x)=2^{m(|x|)}$ is computable in $\mathrm{poly}(|x|)$ time and space from $m(|x|)$, we can, still in polynomial space, compare $A(x)$ with $T(x)/2=2^{m(|x|)-1}$ and accept iff $A(x)>2^{m(|x|)-1}$. Therefore $L\in\mathsf{PSPACE}$.

Combining the two inclusions yields $\mathsf{NP}\subseteq\mathsf{PP}\subseteq\mathsf{PSPACE}$, as claimed.
\end{proof}

\subsubsection{Grok} 

\subsubsection{Gemini} 

\begin{proof}
We establish the theorem by proving the two inclusions $NP \subseteq PP$ and $PP \subseteq PSPACE$ sequentially.

First, we prove that $NP \subseteq PP$. Let $L$ be an arbitrary language in $NP$. By the definition of $NP$, there exists a polynomial $p(n)$ and a deterministic polynomial-time verifier $V$ such that for any input $x \in \{0,1\}^*$, $x \in L$ if and only if there exists a witness string $y \in \{0,1\}^{p(|x|)}$ such that $V(x, y) = 1$.

We construct a probabilistic polynomial-time Turing machine $M'$ that decides $L$ according to the definition of $PP$. On input $x$, $M'$ operates as follows:
1. $M'$ flips a fair coin to generate a bit $b \in \{0, 1\}$.
2. If $b = 1$, $M'$ accepts immediately.
3. If $b = 0$, $M'$ generates a string $y$ of length $p(|x|)$ by choosing each bit uniformly at random. Then $M'$ runs $V(x, y)$. If $V(x, y) = 1$, $M'$ accepts; otherwise, $M'$ rejects.

We analyze the acceptance probability of $M'$ on input $x$. Let $N = 2^{p(|x|)}$ be the total number of possible witnesses. Let $k$ be the number of witnesses $y$ for which $V(x, y) = 1$. The probability that $M'$ accepts is:
\[ \Pr[M'(x) \text{ accepts}] = \Pr[b=1] \cdot 1 + \Pr[b=0] \cdot \frac{k}{N} = \frac{1}{2} + \frac{1}{2} \cdot \frac{k}{N} = \frac{1}{2} + \frac{k}{2N}. \]
We consider the two cases for membership in $L$:
Case 1: If $x \in L$, then there exists at least one valid witness, so $k \ge 1$. Therefore,
\[ \Pr[M'(x) \text{ accepts}] \ge \frac{1}{2} + \frac{1}{2N} > \frac{1}{2}. \]
Case 2: If $x \notin L$, then there are no valid witnesses, so $k = 0$. Therefore,
\[ \Pr[M'(x) \text{ accepts}] = \frac{1}{2} + 0 = \frac{1}{2}. \]
By the definition of $PP$, a language is in $PP$ if the acceptance probability is strictly greater than $1/2$ for strings in the language and less than or equal to $1/2$ for strings not in the language. Since $M'$ runs in polynomial time and satisfies these probability conditions, $L \in PP$. Thus, $NP \subseteq PP$.

Next, we prove that $PP \subseteq PSPACE$. Let $L$ be a language in $PP$. By definition, there exists a probabilistic Turing machine $M$ running in time bounded by a polynomial $T(n)$ such that $x \in L \iff \Pr[M(x) \text{ accepts}] > 1/2$. Without loss of generality, we assume that $M$ makes exactly $T(|x|)$ binary random choices on every computation path. Consequently, there are exactly $2^{T(|x|)}$ computation paths.

Let $A(x)$ denote the number of accepting computation paths of $M$ on input $x$. The condition $\Pr[M(x) \text{ accepts}] > 1/2$ is equivalent to $A(x) > 2^{T(|x|)-1}$. We construct a deterministic Turing machine $D$ that decides $L$ using polynomial space. On input $x$, $D$ performs the following:
1. Initialize a counter $C$ to 0.
2. Iterate through every possible random string $r \in \{0,1\}^{T(|x|)}$ in lexicographical order.
3. For each $r$, simulate the deterministic execution of $M$ on input $x$ with the random choices fixed to $r$.
4. If the simulation of $M$ accepts, increment $C$.
5. After checking all $2^{T(|x|)}$ paths, compare $C$ with $2^{T(|x|)-1}$. If $C > 2^{T(|x|)-1}$, accept; otherwise, reject.

We analyze the space complexity of $D$. The machine $D$ needs to store the input $x$, the current random string $r$, the counter $C$, and the space required for the simulation of $M$.
The length of $r$ is $T(|x|)$, which is polynomial.
The maximum value of $C$ is $2^{T(|x|)}$, so $C$ can be stored in $T(|x|) + 1$ bits, which is polynomial.
Since $M$ runs in polynomial time $T(|x|)$, the simulation of $M$ uses at most $T(|x|)$ space.
Therefore, the total space used by $D$ is $O(T(|x|))$, which is polynomial in $|x|$. Thus, $D$ is a PSPACE machine that decides $L$, and we conclude that $PP \subseteq PSPACE$.

Combining these results, we have $NP \subseteq PP \subseteq PSPACE$.
\end{proof}

\subsubsection{Claude} 

\begin{proof}
We prove the two inclusions separately.

\textbf{Part 1: $NP \subseteq PP$.}

Let $L \in NP$. Then there exists a nondeterministic polynomial-time Turing machine $M$ and a polynomial $p$ such that for all $x$, we have $x \in L$ if and only if there exists a certificate $y$ with $|y| \leq p(|x|)$ such that $M$ accepts $(x, y)$ in polynomial time.

We construct a probabilistic polynomial-time Turing machine $M'$ that decides $L$ with the following behavior: On input $x$, the machine $M'$ uniformly at random selects a string $y$ of length $p(|x|)$ and simulates $M$ on $(x, y)$. If $M$ accepts, then $M'$ accepts; otherwise $M'$ rejects.

Now we analyze the acceptance probability. If $x \in L$, then there exists at least one certificate $y$ such that $M$ accepts $(x, y)$. Since there are $2^{p(|x|)}$ possible strings of length $p(|x|)$, and at least one of them is an accepting certificate, we have
\[
\Pr[M' \text{ accepts } x] \geq \frac{1}{2^{p(|x|)}} > 0.
\]

If $x \notin L$, then no certificate $y$ causes $M$ to accept $(x, y)$, so
\[
\Pr[M' \text{ accepts } x] = 0.
\]

However, the standard definition of $PP$ requires that for $x \in L$, we have $\Pr[M' \text{ accepts } x] > \frac{1}{2}$, and for $x \notin L$, we have $\Pr[M' \text{ accepts } x] \leq \frac{1}{2}$.

We modify our construction as follows: Let $M''$ be a probabilistic polynomial-time Turing machine that on input $x$ runs $M'$ for $2^{p(|x|)+1}$ independent trials. The machine $M''$ accepts if more than half of the trials accept.

For $x \in L$, each trial accepts with probability at least $\frac{1}{2^{p(|x|)}}$. The expected number of accepting trials is at least $\frac{2^{p(|x|)+1}}{2^{p(|x|)}} = 2$. By amplification, we can show that $M''$ accepts with probability greater than $\frac{1}{2}$.

More precisely, we use majority voting with $2^{p(|x|)+1} + 1$ trials (an odd number). For $x \in L$, at least one configuration accepts with probability $\frac{1}{2^{p(|x|)}}$, so the probability of accepting in a single trial is at least $\frac{1}{2^{p(|x|)}}$. With sufficiently many independent trials, by the Chernoff bound, the probability that the majority of trials accept can be made greater than $\frac{1}{2}$.

Alternatively, we use the following direct construction: Let $M'''$ be a machine that, on input $x$, uniformly selects a random string $r$ and interprets the first half of the bits of $r$ as a guess for a certificate $y$ and uses the second half as random coins for additional randomization. The machine $M'''$ accepts if either $M$ accepts $(x, y)$, or if a uniformly random bit (from the second half of $r$) is 1. Then for $x \in L$, we have $\Pr[M''' \text{ accepts}] \geq \frac{1}{2^{p(|x|)}} + \frac{1}{2}(1 - \frac{1}{2^{p(|x|)}}) > \frac{1}{2}$, and for $x \notin L$, we have $\Pr[M''' \text{ accepts}] = \frac{1}{2}$.

Thus $L \in PP$, so $NP \subseteq PP$.

\textbf{Part 2: $PP \subseteq PSPACE$.}

Let $L \in PP$. Then there exists a probabilistic polynomial-time Turing machine $M$ and a polynomial $q$ such that $M$ uses at most $q(|x|)$ random bits, and for all $x$, we have $x \in L$ if and only if $\Pr[M \text{ accepts } x] > \frac{1}{2}$.

Since $M$ uses at most $q(|x|)$ random bits, there are at most $2^{q(|x|)}$ possible random strings. For each random string $r$, the computation of $M$ on input $(x, r)$ is deterministic and takes polynomial time and space.

We construct a deterministic polynomial-space Turing machine $M'$ that decides $L$ as follows: On input $x$, the machine $M'$ iterates through all $2^{q(|x|)}$ possible random strings $r$ of length $q(|x|)$. For each $r$, it simulates $M$ on $(x, r)$ deterministically and counts the number of $r$ for which $M$ accepts. Let $c$ be this count.

After examining all random strings, $M'$ accepts $x$ if and only if $c > 2^{q(|x|)-1}$, which is equivalent to $\frac{c}{2^{q(|x|)}} > \frac{1}{2}$.

The space used by $M'$ is polynomial: it needs space to store the counter $c$ (which requires $O(q(|x|))$ bits), the current random string $r$ (which requires $q(|x|)$ bits), and the space to simulate $M$ on $(x, r)$ (which is polynomial in $|x|$ since $M$ runs in polynomial time). The machine $M'$ can reuse the space for each simulation of $M$.

Since $M'$ correctly decides $L$ in polynomial space, we have $L \in PSPACE$. Therefore $PP \subseteq PSPACE$.

Combining both parts, we conclude that $NP \subseteq PP \subseteq PSPACE$.
\end{proof}

\subsection{Problem 15}\label{sec:problem_01_15}

\subsubsection{Problem Statement}

\begin{theorem}
If $\mathrm{NP} \subseteq \mathrm{BPP}$, then $\mathrm{NP} = \mathrm{RP}$.
\end{theorem}

\subsubsection{GPT} 

\begin{proof}
We are given the assumption that $NP \subseteq BPP$. Our goal is to prove that under this assumption, $NP = RP$.

Let $L \in NP$. By definition, there exists a nondeterministic polynomial-time Turing machine $M$ such that for any input $x$, if $x \in L$, there exists a computation path that accepts $x$ within polynomial time, and if $x \notin L$, every computation path rejects $x$ within polynomial time.

Since $NP \subseteq BPP$, there exists a probabilistic polynomial-time Turing machine $P$ that decides $L$ with high probability. Specifically, for $x \in L$, $P$ accepts $x$ with probability at least $2/3$, and for $x \notin L$, $P$ rejects $x$ with probability at least $2/3$.

Now, we aim to show that $L \in RP$. To prove this, we need to show that there exists a randomized algorithm for $L$ such that:
1. If $x \in L$, the algorithm accepts $x$ with probability at least $1/2$.
2. If $x \notin L$, the algorithm rejects $x$ with probability at least $1/2$.

We will construct a randomized algorithm $R$ based on the machine $P$. The algorithm $R$ runs the machine $P$ once on input $x$. If $P$ accepts $x$, then $R$ accepts $x$; otherwise, $R$ rejects $x$. The key observation here is that since $P$ accepts $x \in L$ with probability at least $2/3$ and rejects $x \notin L$ with probability at least $2/3$, $R$ satisfies the conditions for being in $RP$:
- If $x \in L$, the probability that $P$ accepts $x$ is at least $2/3$, so $R$ accepts $x$ with probability at least $2/3$.
- If $x \notin L$, the probability that $P$ rejects $x$ is at least $2/3$, so $R$ rejects $x$ with probability at least $2/3$.

Thus, the algorithm $R$ correctly decides $L$ with probability at least $2/3$ if $x \in L$ and with probability at least $2/3$ if $x \notin L$, which implies that $L \in RP$.

Since $L$ was an arbitrary language in $NP$, we have shown that for every language $L \in NP$, $L \in RP$. Therefore, $NP \subseteq RP$.

Finally, we conclude that since $NP \subseteq RP$ and we were given that $NP \subseteq BPP$, it follows that $NP = RP$.

\end{proof}

\subsubsection{Grok} 

\begin{proof}
Since $\mathrm{RP} \subseteq \mathrm{NP}$ holds unconditionally, it suffices to show that $\mathrm{NP} \subseteq \mathrm{RP}$. Let $L \in \mathrm{NP}$. By assumption, $L \in \mathrm{BPP}$, so there exists a probabilistic polynomial-time Turing machine $A$ such that for all $x \in \{0,1\}^*$,
\[
\Pr[A(x) = 1] \geq \frac{2}{3} \quad \text{if } x \in L, \qquad \Pr[A(x) = 1] \leq \frac{1}{3} \quad \text{if } x \notin L.
\]
Without loss of generality, assume $L = \{ \phi \mid \phi \text{ is a satisfiable 3SAT formula} \}$, since 3SAT is $\mathrm{NP}$-complete. Let $\phi$ be a 3SAT formula on $n$ variables $x_1, \dots, x_n$. We construct a randomized polynomial-time algorithm $B$ for $L$ as follows.

First, amplify $A$ to $\tilde{A}$ by running $A$ independently $c \log n$ times (for a sufficiently large constant $c > 0$) and outputting the majority vote, where $c$ is chosen so that the error probability of $\tilde{A}$ is at most $1/(100n)$ on any input of length $n$. By the Chernoff bound, such a $c$ exists, and each run of $\tilde{A}$ takes polynomial time.

Now, run $\tilde{A}$ on $\phi$. If it rejects, then $B$ rejects. Otherwise, proceed as follows: initialize an empty partial assignment. For $i = 1$ to $n$:
\begin{itemize}
\item Form the formula $\phi[x_i = 0]$ obtained by setting $x_i = 0$ in $\phi$ (along with previous assignments).
\item Run $\tilde{A}$ on $\phi[x_i = 0]$. If it accepts, append $x_i = 0$ to the partial assignment.
\item Otherwise, append $x_i = 1$ to the partial assignment.
\end{itemize}
Let $\sigma$ be the full assignment obtained. Deterministically check if $\sigma$ satisfies $\phi$ in polynomial time. If yes, $B$ accepts; otherwise, $B$ rejects.

The total running time of $B$ is polynomial, as there are $O(n)$ calls to $\tilde{A}$, each taking $O(\poly(n) \log n)$ time.

To show correctness: If $\phi \notin L$ (unsatisfiable), then $\tilde{A}(\phi)$ rejects with probability at least $1 - 1/(100n) > 2/3$, in which case $B$ rejects. If $\tilde{A}(\phi)$ erroneously accepts (probability $\leq 1/(100n)$), the constructed $\sigma$ cannot satisfy $\phi$, so $B$ rejects. Thus, $B$ always rejects when $\phi \notin L$.

If $\phi \in L$ (satisfiable), let $\tau$ be a satisfying assignment for $\phi$. The initial $\tilde{A}(\phi)$ accepts with probability at least $1 - 1/(100n)$. Conditionally on this, consider the execution path. For each $i$, when forming $\phi[x_i = 0]$ (with prior variables fixed to match $\tau$), if $\tau_i = 0$, then $\phi[x_i = 0]$ is satisfiable (extendable by the suffix of $\tau$), so $\tilde{A}$ accepts with probability $\geq 1 - 1/(100n)$, and we set $x_i = 0$ correctly. If $\tau_i = 1$, then $\phi[x_i = 0]$ may be unsatisfiable, but we would set $x_i = 1$ only if $\tilde{A}$ rejects it, which happens with probability $\geq 1 - 1/(100n)$ if unsatisfiable, but actually, since we prefer 0 and fall back to 1, the probability of correctly following the path to $\tau$ is the product over steps where we need to choose correctly.

More precisely, the algorithm may not follow exactly $\tau$, but it will succeed if at every step, when it tries the ``wrong'' value (opposite to some satisfying extension), it correctly detects unsatisfiability and switches. However, to bound the success probability, note that there exists at least one satisfying assignment, and the algorithm branches in a tree of depth $n$, but a better analysis is that the probability of failure is at most the probability that the initial check fails plus the probability that at least one of the $n$ conditional checks errs in a way that derails the construction.

Since at each step, regardless of prior choices (as long as prior partial was satisfiable, which it is if no prior error), there is always at least one choice (0 or 1) that keeps it satisfiable. Specifically: if the current partial $\rho$ is satisfiable, then at least one of $\rho \cup \{x_i=0\}$ or $\rho \cup \{x_i=1\}$ is satisfiable. The algorithm tries 0 first: if it is satisfiable, $\tilde{A}$ accepts w.p. $\geq 1 - \epsilon$ ($\epsilon = 1/(100n)$), and proceeds correctly. If not satisfiable, $\tilde{A}$ rejects w.p. $\geq 1 - \epsilon$, and switches to 1, which must be satisfiable, and sets it (without checking, but since the other wasn't, it sets to 1 correctly). Thus, at each step, conditional on reaching a satisfiable partial, the probability of proceeding to another satisfiable partial is $\geq 1 - \epsilon$ (either accepts the good 0 if good, or rejects the bad 0 and takes the good 1).

Since we start from empty (satisfiable), the probability that all $n$ steps succeed (i.e., we end with a satisfiable full assignment) is $\geq (1 - \epsilon)^n \geq 1 - n\epsilon \geq 1 - n/(100n) = 1 - 1/100$, conditional on the initial accept. Thus, overall, $\Pr[B(\phi) = 1] \geq (1 - \epsilon) (1 - 1/100) \geq 1 - 1/(100n) - 1/100 > 1/2$ for large $n$.

Therefore, $B$ decides $L$ with one-sided error, so $L \in \mathrm{RP}$. Hence, $\mathrm{NP} \subseteq \mathrm{RP}$, and $\mathrm{NP} = \mathrm{RP}$.
\end{proof}

\subsubsection{Gemini} 

\begin{proof}
Since the inclusion $RP \subseteq NP$ holds by definition (a randomized polynomial-time algorithm with one-sided error can be viewed as a non-deterministic polynomial-time algorithm where the random bits serve as the witness), it suffices to prove that $NP \subseteq BPP$ implies $NP \subseteq RP$. Furthermore, since $RP$ is closed under polynomial-time many-one reductions and $SAT$ is $NP$-complete, it is sufficient to show that $SAT \in RP$.

Assume $NP \subseteq BPP$. Then $SAT \in BPP$. By the definition of $BPP$ and standard error amplification techniques, there exists a probabilistic polynomial-time algorithm $M$ and a polynomial $q(n)$ such that for any boolean formula $\phi$ with $n$ variables:
\[
\phi \in SAT \implies \Pr[M(\phi) = 1] \geq 1 - 2^{-(n+1)}
\]
\[
\phi \notin SAT \implies \Pr[M(\phi) = 1] \leq 2^{-(n+1)}
\]
We construct a randomized algorithm $W$ for $SAT$ that demonstrates membership in $RP$. Let $\phi$ be a boolean formula with variables $x_1, \dots, x_n$. The algorithm $W$ attempts to construct a satisfying assignment $a = (a_1, \dots, a_n) \in \{0,1\}^n$ bit by bit, relying on the self-reducibility of $SAT$.

The algorithm $W$ proceeds as follows:
1. Initialize the partial assignment $a$ to the empty string.
2. For $i = 1$ to $n$:
   Let $\phi_{i,0}$ be the formula obtained by substituting $x_1=a_1, \dots, x_{i-1}=a_{i-1}$ and $x_i=0$ into $\phi$.
   Run $M$ on input $\phi_{i,0}$.
   If $M(\phi_{i,0}) = 1$, set $a_i = 0$.
   Otherwise, set $a_i = 1$.
3. After determining $a = (a_1, \dots, a_n)$, deterministically verify if $\phi(a)$ evaluates to true.
4. If $\phi(a)$ is true, output ACCEPT. Otherwise, output REJECT.

We now analyze the error probability of $W$.

Case 1: $\phi \notin SAT$.
If $\phi$ is unsatisfiable, then for any assignment $a \in \{0,1\}^n$, $\phi(a)$ is false. Consequently, the deterministic verification in step 3 will fail, and $W$ will output REJECT with probability 1. Thus, the error probability on NO instances is 0.

Case 2: $\phi \in SAT$.
We define a "correct decision" at step $i$ as choosing a bit $a_i$ such that the partial assignment $(a_1, \dots, a_i)$ can be extended to a satisfying assignment for $\phi$, provided that $(a_1, \dots, a_{i-1})$ could be so extended. Since $\phi \in SAT$, the empty assignment is extendable.
At step $i$, assume $(a_1, \dots, a_{i-1})$ is extendable.
If $\phi_{i,0} \in SAT$, then $a_i=0$ is a valid extension. The algorithm sets $a_i=0$ if $M(\phi_{i,0})=1$. The probability that $M$ fails to accept $\phi_{i,0}$ is at most $2^{-(n+1)}$.
If $\phi_{i,0} \notin SAT$, then $a_i=0$ is not a valid extension. Since the previous partial assignment was extendable, the assignment with $a_i=1$ must be extendable. The algorithm sets $a_i=1$ if $M(\phi_{i,0})=0$. The probability that $M$ incorrectly accepts $\phi_{i,0}$ is at most $2^{-(n+1)}$.

In either subcase, the probability that $W$ makes an incorrect decision at step $i$ (rendering the partial assignment unextendable) is bounded by $2^{-(n+1)}$. By the union bound, the probability that $W$ makes an incorrect decision at any of the $n$ steps is at most:
\[
\sum_{i=1}^n 2^{-(n+1)} = n \cdot 2^{-(n+1)} < \frac{1}{2}
\]
If all decisions are correct, the final assignment $a$ satisfies $\phi$, and $W$ accepts. Therefore:
\[
\Pr[W(\phi) = \text{ACCEPT}] \geq 1 - \frac{1}{2} = \frac{1}{2}
\]
Thus, $W$ is an $RP$ algorithm for $SAT$. This implies $SAT \in RP$, and consequently $NP \subseteq RP$. Combining this with $RP \subseteq NP$, we conclude $NP = RP$.
\end{proof}

\subsubsection{Claude} 

\begin{proof}
Assume $NP \subseteq BPP$. We will show that $NP = RP$.

Since $RP \subseteq NP$ holds unconditionally, it suffices to prove that $NP \subseteq RP$.

Let $L \in NP$ be arbitrary. By our assumption, $L \in BPP$. Thus there exists a probabilistic polynomial-time Turing machine $M$ that decides $L$ with error probability at most $\frac{1}{3}$ on both yes-instances and no-instances.

We construct an $RP$ algorithm for $L$ as follows. On input $x$, we perform the following procedure:

For each input $x$, consider the complement problem. Since $L \in NP$, we have $\overline{L} \in coNP$. By our assumption that $NP \subseteq BPP$, we obtain $coNP \subseteq coBPP = BPP$, where the equality follows from the fact that $BPP$ is closed under complementation. Therefore $\overline{L} \in BPP$.

Now we use amplification. Let $M$ be a $BPP$ machine for $L$ with error probability at most $\frac{1}{3}$. By running $M$ independently $k = O(\log n)$ times and taking the majority vote, we can reduce the error probability to at most $2^{-n}$ for any desired polynomial bound.

We construct an $RP$ machine $M'$ for $L$ as follows: On input $x$ of length $n$, run the amplified version of $M$ on $x$ with error probability at most $2^{-n-1}$. If $M$ accepts, then $M'$ accepts. If $M$ rejects, then $M'$ rejects.

We verify that $M'$ is an $RP$ machine for $L$:

If $x \in L$, then with probability at least $1 - 2^{-n-1} > \frac{1}{2}$, the machine $M$ accepts $x$, and thus $M'$ accepts $x$. Hence $M'$ accepts $x$ with probability at least $\frac{1}{2}$.

If $x \notin L$, then with probability at least $1 - 2^{-n-1}$, the machine $M$ rejects $x$, and thus $M'$ rejects $x$. Since we require an $RP$ machine to have zero error on no-instances, we need to modify our approach.

Instead, we use the following construction: Since $NP \subseteq BPP$ implies $coNP \subseteq BPP$, and $BPP \subseteq \Sigma_2^P \cap \Pi_2^P$ by Lautemann's theorem, we have $NP \subseteq \Sigma_2^P$ and $coNP \subseteq \Pi_2^P$. By Gács-Gavaldà-Szepietowski, if $NP \subseteq BPP$, then the polynomial hierarchy collapses to $\Sigma_2^P = \Pi_2^P$.

Under the assumption $NP \subseteq BPP$, we can apply the following key observation: For any $L \in NP$, there exists a witness-based verification. Combined with the $BPP$ algorithm, we can construct an $RP$ algorithm by derandomization using the polynomial hierarchy collapse.

Specifically, since $NP \subseteq BPP$ implies $PH = BPP$, and using the closure properties of $BPP$ and the witness structure of $NP$, we can construct an $RP$ machine that never errs on no-instances and accepts yes-instances with probability at least $\frac{1}{2}$ by exploiting the $NP$ witness verification combined with the randomized acceptance of $BPP$.

The detailed construction uses the fact that for $x \in L$, there exists a witness $w$ verifiable in polynomial time, and the $BPP$ algorithm can be converted to an $RP$ algorithm by conditioning on the existence of such witnesses. For $x \notin L$, no witness exists, ensuring zero error probability.

Therefore $L \in RP$, which completes the proof that $NP \subseteq RP$, and hence $NP = RP$.
\end{proof}
\section{Problems in Chapter 2}\label{app:problems_chapter2}
\subsection{Problem 1}

\subsubsection{Problem Statement}

\begin{theorem}
For any deterministic evaluation algorithm, there exists an instance of $T_{d,k}$ that forces the algorithm to read the values on all $d^k$ leaves.
\end{theorem}

\subsubsection{GPT} 

\begin{proof}
Fix integers $d\ge 2$ and $k\ge 1$. Let $T_{d^{k}}$ denote the complete rooted $d$-ary tree of height $k$ and hence with exactly $d^{k}$ leaves. We consider an instance obtained by labeling each internal node by a Boolean connective so that the root is an $\wedge$-gate and the gates alternate between $\wedge$ and $\vee$ along each root-to-leaf path. The leaves carry input bits in $\{0,1\}$ whose values are not fixed a priori and may be chosen adversarially subject to the answers already revealed to the algorithm.

Let $\mathcal{A}$ be an arbitrary deterministic evaluation algorithm which adaptively queries leaf values and then outputs the value of the formula computed by the labeled tree. We prove, by induction on $k$, that there exists a choice of leaf values, consistent with all answers provided to $\mathcal{A}$ during its execution, that forces $\mathcal{A}$ to read the value of every leaf.

If $k=1$, the root is an $\wedge$-gate whose $d$ children are leaves. Consider the run of $\mathcal{A}$; because $\mathcal{A}$ is deterministic, there is a well-defined first leaf it queries, then a second, and so on. Answer $1$ to the first $d-1$ queries. After these answers, the root value is still undetermined, since if the last unqueried leaf were $0$ then the root would be $0$, while if it were $1$ then the root would be $1$. Thus $\mathcal{A}$ must query the last leaf as well. Hence all $d=d^{1}$ leaves are read in the worst case, establishing the base.

Assume now that the statement holds for height $k-1$, and consider a tree of height $k\ge 2$ with the root labeled by $\wedge$ and the levels alternating. The $d$ subtrees rooted at the children of the root are complete $d$-ary trees of height $k-1$; denote them by $T^{(1)},\dots,T^{(d)}$. Because $\mathcal{A}$ is deterministic, the sequence in which it first enters these subtrees is a deterministic function of the answers it has received so far. We describe an adversarial strategy that, regardless of this behavior, forces $\mathcal{A}$ to completely evaluate each subtree before moving on, and thereby to read all $d^{k}$ leaves.

Maintain the following invariant while $\mathcal{A}$ executes: whenever $\mathcal{A}$ is exploring some subtree $T^{(j)}$ whose value at its root (call this node $u_{j}$) has not yet been determined by the answers given within $T^{(j)}$, the answers provided inside $T^{(j)}$ are chosen so that, up to the moment $\mathcal{A}$ leaves $T^{(j)}$ or finishes it, the value at $u_{j}$ remains consistent with both possibilities $0$ and $1$. This is possible by the induction hypothesis applied to $T^{(j)}$ with the role of the root connective determined by its depth: since the root is an $\wedge$-gate, each $u_{j}$ is a $\vee$-gate, and the height of $T^{(j)}$ is $k-1$. Concretely, fix any order in which $\mathcal{A}$ queries leaves of $T^{(j)}$ during the current visit to $T^{(j)}$. By the induction hypothesis, there is an assignment to the leaves of $T^{(j)}$ that forces $\mathcal{A}$, restricted to evaluating $T^{(j)}$, to read all of its $d^{k-1}$ leaves before the value at $u_{j}$ is determined. Provide answers inside $T^{(j)}$ according to such an assignment. Thus, as long as other subtrees remain unexamined, $\mathcal{A}$ cannot conclude the value at the root from partial information in $T^{(j)}$.

We now explain how the adversary uses this invariant across distinct subtrees. Whenever $\mathcal{A}$ begins to explore a fresh subtree $T^{(j)}$, instruct answers within $T^{(j)}$ according to the induction hypothesis so that $\mathcal{A}$ must read all $d^{k-1}$ leaves of $T^{(j)}$ before the value at $u_{j}$ is fixed. When $\mathcal{A}$ finally determines $u_{j}$, the adversary sets that value to $1$. Because the root is an $\wedge$-gate, and at least one other subtree remains whose root value is not yet determined, setting the completed $u_{j}$ to $1$ does not allow $\mathcal{A}$ to conclude the global value; the value at the root would still depend on the remaining subtrees. Consequently, $\mathcal{A}$ must proceed to another subtree.

This process repeats until only one subtree, say $T^{(j^{\ast})}$, remains uncompleted. For this final subtree, the adversary again answers according to the induction hypothesis so as to force $\mathcal{A}$ to read all $d^{k-1}$ leaves of $T^{(j^{\ast})}$. At the very last leaf queried in $T^{(j^{\ast})}$, the adversary sets the value at $u_{j^{\ast}}$ to $1$, thereby making every $u_{j}$ equal to $1$, and hence making the root value equal to $1$. Until that last answer is given, the root value is not determined because the final subtree might still evaluate to $0$, which would make the root $0$. Therefore $\mathcal{A}$ cannot stop earlier and must indeed query every leaf of $T^{(j^{\ast})}$.

By construction, in each of the $d$ subtrees, $\mathcal{A}$ was forced to read all $d^{k-1}$ leaves, so the total number of leaves read is $d\cdot d^{k-1}=d^{k}$. This completes the inductive step when the root is an $\wedge$-gate. If instead the root is a $\vee$-gate, the same argument applies with the roles of $0$ and $1$ interchanged and the assignment for the first $d-1$ completed subtrees set to $0$ so as not to allow early termination; the induction hypothesis is used identically inside each subtree because the alternation of connectives is preserved by taking a child subtree.

Therefore, for any deterministic evaluation algorithm $\mathcal{A}$, there exists an instance of $T_{d^{k}}$ (namely, the alternating $\wedge/\vee$ labeling together with an adversarial choice of leaf values) that forces $\mathcal{A}$ to read all $d^{k}$ leaves before the root value is determined. \end{proof}

\subsubsection{Grok} 

\begin{proof}
Let $\sigma$ be the assignment to the leaves of $T_d$ such that $\sigma(\ell) = 1$ for every leaf $\ell$. By the definition of $T_d$, the value of the root under $\sigma$ is $1$. Moreover, for every leaf $\ell$, if $\sigma'$ is the assignment obtained from $\sigma$ by setting $\sigma'(\ell) = 0$ while keeping all other leaf values unchanged, then the value of the root under $\sigma'$ is $0$.

Suppose, toward a contradiction, that there exists a deterministic evaluation algorithm $\mathcal{A}$ that correctly computes the root value of $T_d$ for every assignment, but under $\sigma$, $\mathcal{A}$ reads the values of only a proper subset $S \subsetneq L$ of the leaves $L$ of $T_d$. Let $\ell \in L \setminus S$. Then $\mathcal{A}$ does not read the value at $\ell$ under $\sigma$.

Now consider the execution of $\mathcal{A}$ under $\sigma'$. Since $\sigma$ and $\sigma'$ agree on all leaves in $S$ (where all values are $1$) and $\mathcal{A}$ queries exactly the same sequence of leaves under both assignments (by determinism of $\mathcal{A}$), the values observed by $\mathcal{A}$ under $\sigma$ and $\sigma'$ are identical. Thus, $\mathcal{A}$ performs the same computations and outputs the same root value under both $\sigma$ and $\sigma'$. In particular, since $\mathcal{A}$ correctly outputs $1$ under $\sigma$, it also outputs $1$ under $\sigma'$. However, the correct root value under $\sigma'$ is $0$, contradicting the correctness of $\mathcal{A}$.

Therefore, under $\sigma$, $\mathcal{A}$ must read the values on all $d$ leaves of $T_d$.
\end{proof}

\subsubsection{Gemini}

\begin{proof}
Let $T_{d,k}$ be a complete $d$-ary tree of height $k$ representing an alternating AND-OR function. We define the height of a leaf to be $0$ and the height of the root to be $k$. Without loss of generality, assume the root at level $k$ is an OR gate (the logic is symmetric if the root is an AND gate). We prove a stronger statement by induction on $k$: for any deterministic evaluation algorithm $A$, an adversary can answer queries to the leaves such that $A$ is forced to read all $d^k$ leaves, and the adversary can ensure the final value of the tree is a target value $v \in \{0, 1\}$ chosen by the adversary.

Base Case ($k=0$): The tree consists of a single leaf. To determine the value, the algorithm $A$ must query this leaf. The adversary answers with the target value $v$. The number of leaves read is $1 = d^0$.

Inductive Step: Assume the hypothesis holds for all trees of height $k-1$. Consider a tree of height $k$ with root $r$. Let $T_1, \dots, T_d$ be the subtrees rooted at the children of $r$. Since $r$ is an OR gate, the roots of $T_1, \dots, T_d$ act as AND gates (or are leaves if $k=1$).

Case 1: The adversary chooses the target value $v=0$.
Since $r$ is an OR gate, $r$ evaluates to $0$ if and only if all subtrees $T_1, \dots, T_d$ evaluate to $0$. The adversary adopts the following strategy: whenever $A$ queries a leaf in a subtree $T_i$, the adversary provides an answer consistent with the strategy from the inductive hypothesis that forces $T_i$ to evaluate to $0$. By the inductive hypothesis, $A$ must read all leaves of $T_i$ to determine that its value is $0$. Since the value of $r$ is not determined until all children are confirmed to be $0$, and determining each child requires reading all leaves in its subtree, $A$ is forced to read all leaves in all $d$ subtrees. The total number of queries is $d \times d^{k-1} = d^k$.

Case 2: The adversary chooses the target value $v=1$.
Since $r$ is an OR gate, $r$ evaluates to $1$ if at least one subtree evaluates to $1$. The adversary maintains a count of the number of subtrees that have been fully evaluated by $A$. When $A$ queries a leaf in a subtree $T_i$ that has not yet been fully resolved:
If fewer than $d-1$ subtrees have been fully evaluated previously, the adversary uses the inductive strategy to force $T_i$ to evaluate to $0$.
If $d-1$ subtrees have already been fully evaluated (and thus determined to be $0$), the adversary uses the inductive strategy to force the current subtree $T_i$ to evaluate to $1$.

By the inductive hypothesis, determining the value of any subtree $T_i$ requires reading all its leaves. The algorithm cannot conclude the value of the root is $1$ until it encounters a child with value $1$. The adversary's strategy ensures that the first $d-1$ subtrees resolved by the algorithm evaluate to $0$, providing no certificate for the root. The root's value is only determined after the $d$-th subtree is resolved to $1$. Since resolving each subtree requires reading all its leaves, $A$ must read all leaves of all $d$ subtrees. The total number of queries is $d \times d^{k-1} = d^k$.

The case where the root is an AND gate is symmetric: to force a $1$, the adversary forces all subtrees to $1$; to force a $0$, the adversary forces the first $d-1$ resolved subtrees to $1$ and the last to $0$. In both cases, all $d^k$ leaves are read.

Thus, for any deterministic algorithm, there exists an assignment of leaf values (an instance) that forces the algorithm to read all $d^k$ leaves.
\end{proof}

\subsubsection{Claude} 

\begin{proof}
We prove this by induction on the height $h$ of the tree $T_{d^h}$.

\textbf{Base case:} When $h = 0$, the tree $T_{d^0} = T_1$ consists of a single leaf. Any deterministic evaluation algorithm must read this leaf to determine the tree's value. Thus the claim holds trivially for $h = 0$.

\textbf{Inductive step:} Assume the theorem holds for all trees of height $h - 1$ where $h \geq 1$. We will prove it holds for trees of height $h$.

Consider a tree $T_{d^h}$ of height $h$. The root has $d$ children, each of which is the root of a subtree of height $h - 1$ with $d^{h-1}$ leaves. Let these subtrees be denoted $S_1, S_2, \ldots, S_d$. The total number of leaves in $T_{d^h}$ is $d \cdot d^{h-1} = d^h$.

Let $\mathcal{A}$ be an arbitrary deterministic evaluation algorithm. Since $\mathcal{A}$ is deterministic, its behavior is completely determined by the values it has read so far. We will construct an adversarial instance that forces $\mathcal{A}$ to read all $d^h$ leaves.

The root of $T_{d^h}$ computes either the AND or OR of its children's values. Without loss of generality, suppose the root computes the OR function (the argument for AND is symmetric).

We employ the following adversarial strategy: Initially, we have not assigned values to any leaves. As $\mathcal{A}$ queries leaves, we assign values according to the following rule: whenever $\mathcal{A}$ completely evaluates a subtree $S_i$ (for any $i \in \{1, 2, \ldots, d\}$), we assign values to its leaves such that $S_i$ evaluates to $0$.

By the inductive hypothesis, for each subtree $S_i$ of height $h - 1$ with $d^{h-1}$ leaves, there exists an instance that forces any deterministic algorithm to read all $d^{h-1}$ leaves of $S_i$. We use this adversarial instance for each subtree that $\mathcal{A}$ attempts to evaluate.

Now, suppose for contradiction that $\mathcal{A}$ terminates without reading all $d^h$ leaves. Then there exists at least one subtree $S_j$ such that $\mathcal{A}$ has not read all leaves in $S_j$. 

Consider two cases:

\textbf{Case 1:} $\mathcal{A}$ has not queried any leaf in $S_j$. In this case, $\mathcal{A}$ cannot determine whether $S_j$ evaluates to $0$ or $1$. If we set the leaves of $S_j$ such that $S_j$ evaluates to $1$, then the root (computing OR) would evaluate to $1$. If we set the leaves such that $S_j$ evaluates to $0$, then the root's value depends on the other subtrees. Since $\mathcal{A}$ has not read any leaf from $S_j$, it cannot distinguish between these two scenarios, contradicting the correctness of $\mathcal{A}$.

\textbf{Case 2:} $\mathcal{A}$ has queried some but not all leaves in $S_j$. By the inductive hypothesis applied to $S_j$, there exists an assignment of values to the leaves of $S_j$ consistent with the queries made by $\mathcal{A}$ such that $\mathcal{A}$ cannot correctly determine the value of $S_j$ without reading all its leaves. Thus $\mathcal{A}$ cannot correctly determine the value of the root.

In both cases, we reach a contradiction with the assumption that $\mathcal{A}$ correctly evaluates the tree.

The only way for $\mathcal{A}$ to avoid contradiction is to read all leaves in all subtrees $S_1, S_2, \ldots, S_d$. For each subtree $S_i$, by the inductive hypothesis, $\mathcal{A}$ must read all $d^{h-1}$ leaves. Since there are $d$ such subtrees, $\mathcal{A}$ must read $d \cdot d^{h-1} = d^h$ leaves in total.

Therefore, by induction, the theorem holds for all heights $h \geq 0$.
\end{proof}

\subsection{Problem 2}

\subsubsection{Problem Statement}

\begin{problem}

Given any instance of $T_{2,k}$, the expected number of steps for the above randomized algorithm is at most $3^k$.Generalize the randomized algorithm and analysis to trees $T_{d,k}$ for $d > 2$.
\end{problem}

\subsubsection{GPT} 

\begin{proof}
We are tasked with generalizing the analysis of the randomized algorithm from Section 2.1 for trees $T_{d,k}$, where $d > 2$. We will proceed by using the method of expected value to derive an upper bound on the expected number of steps required by the algorithm.

Let $T_{d,k}$ be a tree of depth $k$ and branching factor $d$. The randomized algorithm works by recursively choosing subtrees at each level, and our goal is to compute the expected number of steps until the algorithm terminates.

For the base case, when $k = 1$, the tree $T_{d,1}$ consists of a root with $d$ children. The expected number of steps is simply the number of children, which is $d$. Hence, for $k = 1$, we have the expected number of steps as $d$.

For $k > 1$, we apply a recursive argument. At each step, the algorithm chooses one of the $d$ subtrees uniformly at random, and proceeds to the next level of the chosen subtree. Let $X_k$ denote the expected number of steps for the tree $T_{d,k}$. We can express $X_k$ as the sum of the current step and the expected number of steps in the chosen subtree, which is a tree $T_{d,k-1}$.

Thus, the recurrence relation for $X_k$ is given by:
\[
X_k = 1 + \frac{1}{d} \sum_{i=1}^{d} X_{k-1} = 1 + X_{k-1}.
\]

This recurrence reflects that at each level, the algorithm takes one step to choose a subtree and then recursively applies the same process to the chosen subtree. Since the recurrence relation is linear, it unravels to:
\[
X_k = 1 + X_{k-1} = 1 + (1 + X_{k-2}) = \dots = k + X_1.
\]
For $X_1 = d$, the expected number of steps for $T_{d,k}$ is:
\[
X_k = k + d.
\]

Since $d$ is a constant greater than 2, the expected number of steps grows linearly with $k$. In this case, the upper bound on the expected number of steps for the randomized algorithm on $T_{d,k}$ is:
\[
X_k \leq d + k.
\]

In particular, for $d = 3$, the expected number of steps is bounded by $3 + k$, which is at most $3^k$ for large $k$, hence establishing the required upper bound.
\end{proof}

\subsubsection{Grok}

\subsubsection{Gemini}

\begin{proof}
We present a proof for the general case of trees $T_{d,k}$ with branching factor $d \ge 2$ and height $2k$ (consisting of $k$ rounds of alternating AND/OR layers), which establishes the result for Theorem 2.1 ($d=2$) and the generalization for Section 2.2 ($d > 2$) simultaneously.

Let $T_{d,k}$ denote a uniform tree of branching factor $d$ and depth $2k$. The leaves are assigned Boolean values. The internal nodes at depth $2j$ are labeled OR and nodes at depth $2j+1$ are labeled AND (or vice versa), for $0 \le j < k$. The randomized algorithm $\mathcal{A}$ evaluates a node by choosing a permutation of its children uniformly at random and evaluating them sequentially. The evaluation of a node terminates as soon as its value is determined (short-circuiting) or when all children have been evaluated.

Let $N(k)$ denote the maximum expected number of leaf evaluations performed by $\mathcal{A}$ on any instance of $T_{d,k}$. We proceed by induction on $k$.

Base Case: For $k=0$, the tree consists of a single leaf. The number of steps is 1.
\[
N(0) = 1.
\]

Inductive Step: Assume that for any instance of $T_{d, k-1}$, the expected number of steps is bounded by $N(k-1)$. Consider a root node $r$ of a tree $T_{d,k}$. Without loss of generality, let $r$ be an OR node. The children of $r$, denoted $v_1, \dots, v_d$, are AND nodes. Each $v_i$ is the root of a subtree of type $T_{d, k-1}$ (shifted by one level, but the cost analysis is identical due to the duality of AND/OR logic regarding worst-case costs).

To maximize the expected cost at the root $r$ (an OR node), the algorithm must be forced to evaluate as many children as possible. The evaluation of an OR node only terminates early if a child evaluates to 1. Therefore, the worst-case assignment for $r$ occurs when $r$ evaluates to 0. This implies that all children $v_1, \dots, v_d$ evaluate to 0. Consequently, the algorithm must evaluate all $d$ children of $r$.

Now consider the evaluation of a child $v_i$, which is an AND node evaluating to 0. The evaluation of $v_i$ terminates as soon as one of its children evaluates to 0. Let $u_{i,1}, \dots, u_{i,d}$ be the children of $v_i$. Since $v_i$ evaluates to 0, at least one child $u_{i,j}$ must evaluate to 0. Let $S_i \subseteq \{1, \dots, d\}$ be the set of indices such that $u_{i,j}$ evaluates to 0, with $|S_i| \ge 1$.

The algorithm evaluates the children of $v_i$ in a random order. The number of children examined, denoted by the random variable $X_i$, is the position of the first child belonging to $S_i$ in the random permutation. The expectation of $X_i$ is given by:
\[
\mathbb{E}[X_i] = \frac{d+1}{|S_i| + 1}.
\]
To maximize this expectation, we must minimize $|S_i|$. Since $|S_i| \ge 1$, the worst case occurs when exactly one child of $v_i$ evaluates to 0 (i.e., $|S_i|=1$). In this case:
\[
\mathbb{E}[X_i] = \frac{d+1}{1+1} = \frac{d+1}{2}.
\]
In this worst-case scenario, for each of the $d$ children of the root, the algorithm performs an expected $\frac{d+1}{2}$ recursive calls to subtrees of parameter $k-1$. By the inductive hypothesis, each recursive call costs at most $N(k-1)$.

The total expected cost $N(k)$ satisfies the recurrence:
\[
N(k) \le d \cdot \mathbb{E}[X_i] \cdot N(k-1) = d \cdot \frac{d+1}{2} \cdot N(k-1).
\]
Let $C_d = \frac{d(d+1)}{2}$. The recurrence becomes $N(k) \le C_d N(k-1)$. Unrolling this recurrence with the base case $N(0)=1$, we obtain:
\[
N(k) \le (C_d)^k = \left( \frac{d(d+1)}{2} \right)^k.
\]

Proof of Theorem 2.1:
For the specific case of $T_{2,k}$, we set $d=2$. Substituting into the general bound:
\[
N(k) \le \left( \frac{2(2+1)}{2} \right)^k = 3^k.
\]
This confirms the statement of Theorem 2.1.

Generalization for 2.2:
For arbitrary $d > 2$, the generalized bound on the expected number of steps is:
\[
\left( \frac{d^2 + d}{2} \right)^k.
\]
\end{proof}

\subsubsection{Claude} 

\begin{proof}
We first establish Theorem 2.1, then generalize to $T_{d,k}$ for $d > 2$.

\textbf{Proof of Theorem 2.1:} Let $T_{2,k}$ denote a complete binary tree of depth $k$, and let $E_i$ denote the expected number of steps to find the target starting from a node at depth $i$ (where the root is at depth $0$ and leaves are at depth $k$). We proceed by backward induction on the depth.

\textit{Base case:} At depth $k$ (the leaves), if we are at the target leaf, we are done in $0$ steps. If we are at a non-target leaf, we must backtrack. Thus $E_k = 0$ when at the target.

\textit{Inductive step:} Consider a node at depth $i < k$. The randomized algorithm selects one of the two children uniformly at random. With probability $\frac{1}{2}$, we select the correct child (the one containing the target), taking $1$ step to move down plus $E_{i+1}$ expected steps from there. With probability $\frac{1}{2}$, we select the wrong child, taking $1$ step to move down, then upon discovering the target is not in that subtree (which requires exploring it completely), we backtrack and must explore the correct subtree. 

Let us denote by $W_i$ the worst-case number of steps to determine that the target is not in a subtree rooted at depth $i$. For the binary tree, $W_i \leq 2^{k-i+1} - 1$ (visiting all nodes in the subtree).

However, we analyze the expected cost more carefully. When at depth $i$, the expected number of steps satisfies:
\[
E_i = 1 + \frac{1}{2}E_{i+1} + \frac{1}{2}(W_{i+1} + 1 + E_{i+1})
\]
where the first term is the initial step down, and in the wrong-choice case we pay $W_{i+1}$ to explore the wrong subtree, then $1$ to backtrack and move to the correct child, then $E_{i+1}$ to continue.

Simplifying: $E_i = 1 + E_{i+1} + \frac{1}{2}W_{i+1}$.

For the randomized algorithm on $T_{2,k}$, we use a tighter analysis. Let $E_i$ denote the expected number of steps starting from depth $i$ when the target is in the subtree. We have $E_k = 0$ and for $i < k$:
\[
E_i = 1 + \frac{1}{2}E_{i+1} + \frac{1}{2}(E_{i+1} + 2(E_{i+1} + 1))
\]
The second term accounts for going wrong: we pay $E_{i+1} + 1$ to explore down and back, then $1$ to go to correct child, then $E_{i+1}$ more.

By careful calculation with $E_k = 0$, we obtain $E_{k-1} = 1 + \frac{1}{2}(0) + \frac{1}{2}(2) = 2$. By induction, $E_i = 3^{k-i} - 1$.

More directly: Let $E_i$ be the expected steps from depth $i$. Then $E_k = 0$ and 
\[
E_i = 1 + \frac{1}{2}E_{i+1} + \frac{1}{2}(E_{i+1} + 1 + E_{i+1} + 1) = 2 + \frac{3}{2}E_{i+1}
\]
Solving this recurrence with $E_k = 0$: $E_i = 2\sum_{j=0}^{k-i-1}(\frac{3}{2})^j = 2 \cdot \frac{(\frac{3}{2})^{k-i} - 1}{\frac{1}{2}} = 4((\frac{3}{2})^{k-i} - 1) \leq 3^{k-i+1}$.

For $i=0$: $E_0 \leq 3^{k+1}$. Actually, more careful analysis gives $E_0 \leq 3^k$ by noting that $E_i = 3^{k-i} - 2$ for $i \geq 1$.

\textbf{Generalization to $T_{d,k}$:} For a complete $d$-ary tree of depth $k$, at each internal node we have $d$ children. The randomized algorithm selects one child uniformly at random with probability $\frac{1}{d}$.

Let $E_i$ denote the expected number of steps from depth $i$. Then $E_k = 0$ and for $i < k$:
\[
E_i = 1 + \frac{1}{d}E_{i+1} + \frac{d-1}{d}(E_{i+1} + 1 + E_{i+1} + 1) = 1 + E_{i+1} + \frac{2(d-1)}{d}(E_{i+1} + 1)
\]
Simplifying: $E_i = 1 + E_{i+1}(1 + \frac{2(d-1)}{d}) + \frac{2(d-1)}{d} = 1 + E_{i+1}\frac{3d-2}{d} + \frac{2(d-1)}{d}$.

Let $\alpha = \frac{3d-2}{d}$ and $\beta = \frac{2(d-1)}{d}$. Then $E_i = \beta + 1 + \alpha E_{i+1}$.

Solving: $E_i = (\beta + 1)\sum_{j=0}^{k-i-1}\alpha^j = \frac{(\beta+1)(\alpha^{k-i}-1)}{\alpha - 1} = \frac{\frac{3d-2+d}{d}(\alpha^{k-i}-1)}{\frac{3d-2-d}{d}} = \frac{(3d-2)(\alpha^{k-i}-1)}{2(d-1)}$.

Since $\alpha = \frac{3d-2}{d}$, we have $E_0 \leq C \cdot \alpha^k$ for an appropriate constant $C$ depending on $d$.

For $d=2$: $\alpha = \frac{4}{2} = 2$, but our refined analysis gives bound $3^k$, which matches $(2+1)^k$. For general $d$, the expected number of steps is $O((\frac{3d-2}{d})^k)$.
\end{proof}

\subsection{Problem 3}

\subsubsection{Problem Statement}

\begin{theorem}[Boppana]
Consider a uniform rooted tree of height $h$ where every leaf is at distance $h$ from the root. The root, as well as any internal node, has three children. Each leaf has a Boolean value associated with it. Each internal node returns the value returned by the majority of its children. The evaluation problem consists of determining the value of the root; at each step, an algorithm can choose one leaf whose value it wishes to read.

(a) Show that for any deterministic algorithm, there is an instance (a set of Boolean values for the leaves) that forces it to read all $n = 3^h$ leaves.

(b) Consider the recursive randomized algorithm that evaluates two sub-trees of the root chosen at random. If the values returned disagree, it proceeds to evaluate the third sub-tree. Show that the expected number of leaves read by this algorithm (on any instance) is at most $h3^{2/3}$.
\end{theorem}

\subsubsection{GPT} 

\begin{proof}
\textbf{Part (a):}

Let the height of the tree be denoted by $h$, and the number of leaves by $n = 3^h$. We aim to show that for any deterministic algorithm, there exists a set of Boolean values for the leaves such that the algorithm must read all $n$ leaves.

Consider the structure of the tree: the root has three children, each of which is the root of a sub-tree of height $h-1$. Each internal node in the tree returns the majority value of its children, and each leaf has a Boolean value.

For any deterministic algorithm, its strategy can be represented as a sequence of choices of which leaves to read at each step. Suppose the algorithm reads some subset of the leaves. The key idea is to show that no matter which subset of leaves the algorithm reads, there always exists a configuration of leaf values that forces the algorithm to read all $n$ leaves.

To construct such a configuration, we will use the following strategy. For each internal node, we assign its children values such that the majority value at each internal node depends on all of its children. Specifically, we set the values of the leaves such that the value of each internal node is determined by the majority of the leaves in its sub-tree, and the algorithm must distinguish between different sub-trees by reading all the leaves in the corresponding sub-tree.

Now, suppose the algorithm reads only a subset of the leaves. Since the majority function is involved, each internal node depends on the majority value of its children. The algorithm must read enough leaves to distinguish the majority value in each sub-tree, and since the values of the leaves are constructed in such a way that no partial read can fully resolve the majority for all sub-trees, the algorithm will be forced to read all $n$ leaves to determine the value of the root. Hence, there exists an instance where the algorithm must read all $n$ leaves.

\textbf{Part (b):}

We now consider the recursive randomized algorithm that evaluates two sub-trees of the root chosen at random. If the values returned by these two sub-trees disagree, the algorithm proceeds to evaluate the third sub-tree. We seek to show that the expected number of leaves read by this algorithm is at most $h^{3/2}$.

Let the height of the tree be $h$, and let $L$ denote the number of leaves read by the algorithm. The algorithm chooses two sub-trees of the root at random and compares their majority values. If the majority values disagree, the algorithm proceeds to evaluate the third sub-tree. Let us denote by $A$ the event that the majority values of the first two sub-trees disagree, and by $B$ the event that they agree.

The key observation here is that if the majority values of the first two sub-trees disagree, then the algorithm will read all the leaves of the third sub-tree. If the majority values agree, then no further leaves are needed from the third sub-tree. Therefore, the number of leaves read at each step depends on the outcome of the random choices made by the algorithm.

We now bound the expected number of leaves read. At each step, the algorithm evaluates two sub-trees, which reduces the problem size by a constant factor. The expected number of leaves read in evaluating each sub-tree is proportional to the height of the sub-tree, and since the tree has height $h$, the total number of leaves read will be proportional to $h^{3/2}$. Hence, the expected number of leaves read by the algorithm is at most $h^{3/2}$.

\end{proof}

\subsubsection{Grok} 

\subsubsection{Gemini}

\begin{proof}
\textbf{Part (a)}
Let $T$ be a uniform rooted tree of height $h$ where every internal node has three children and the total number of leaves is $n = 3^h$. We define the value of a leaf as its Boolean assignment, and the value of an internal node as the majority value of its three children. Let $\mathcal{A}$ be any deterministic algorithm that computes the value of the root by querying leaves.

We prove by induction on $h$ that there exists an assignment of values to the leaves such that $\mathcal{A}$ must query all $3^h$ leaves.

\textit{Base Case ($h=0$):} The tree consists of a single leaf. To determine its value, $\mathcal{A}$ must query the leaf. The number of leaves read is $1 = 3^0$. The claim holds.

\textit{Inductive Step:} Assume the claim holds for trees of height $h-1$. Consider a tree of height $h$ with root $r$ and children $v_1, v_2, v_3$. Since $\mathcal{A}$ is deterministic, it must begin by querying leaves in the subtree rooted at one of the children, say $v_1$ (without loss of generality).
By the inductive hypothesis, there exists an adversarial assignment to the leaves of the subtree at $v_1$ such that $\mathcal{A}$ reads all $3^{h-1}$ leaves in that subtree. The adversary chooses this assignment such that the value of $v_1$ evaluates to $0$.
At this point, the value of the root $r = \text{Majority}(0, v_2, v_3)$ is undetermined. $\mathcal{A}$ must proceed to query leaves in a second subtree, say $v_2$.
By the inductive hypothesis, the adversary can force $\mathcal{A}$ to read all $3^{h-1}$ leaves in the subtree of $v_2$ and set the assignment such that $v_2$ evaluates to $1$.
Now, the root value is $r = \text{Majority}(0, 1, v_3)$, which depends entirely on $v_3$. Thus, $\mathcal{A}$ must query the third subtree.
By the inductive hypothesis, the adversary forces $\mathcal{A}$ to read all $3^{h-1}$ leaves in the subtree of $v_3$. The adversary can set $v_3$ to either $0$ or $1$.
The total number of leaves read is $3^{h-1} + 3^{h-1} + 3^{h-1} = 3 \cdot 3^{h-1} = 3^h$.
Thus, for any deterministic algorithm, there is an instance forcing it to read all $n$ leaves.

\textbf{Part (b)}
Consider the recursive randomized algorithm described. Let $E(h)$ denote the maximum expected number of leaves read by the algorithm on a tree of height $h$ for any instance.

\textit{Base Case ($h=0$):} The tree is a single leaf. The algorithm reads it. $E(0) = 1$.

\textit{Recursive Step:} Consider a root with children having values $c_1, c_2, c_3 \in \{0, 1\}$. The algorithm selects two distinct children uniformly at random to evaluate. Let the random variable $X$ be the number of subtrees evaluated (either 2 or 3). The cost is the sum of the costs of evaluating these subtrees.
There are two cases for the values of the children:

Case 1: All three children have the same value (e.g., $0, 0, 0$).
Regardless of which two children are chosen, their values will agree. The algorithm terminates after evaluating two subtrees.
\[ \text{Cost} = 2 E(h-1) \]

Case 2: Two children have one value and the third has the opposite value (e.g., $1, 1, 0$).
There are $\binom{3}{2} = 3$ possible pairs of children the algorithm can choose: $\{1, 1\}$, $\{1, 0\}$, and $\{1, 0\}$ (treating the children as distinct positions).
\begin{itemize}
    \item With probability $1/3$, the algorithm picks the pair with equal values ($1, 1$). They agree, so the algorithm terminates. Cost: $2 E(h-1)$.
    \item With probability $2/3$, the algorithm picks a pair with differing values ($1, 0$). They disagree, so the algorithm must evaluate the third child. Cost: $3 E(h-1)$.
\end{itemize}
The expected cost for Case 2 is:
\[ \mathbb{E}[\text{Cost}] = \frac{1}{3}(2 E(h-1)) + \frac{2}{3}(3 E(h-1)) = \frac{2}{3}E(h-1) + 2E(h-1) = \frac{8}{3}E(h-1) \]
Since $\frac{8}{3} > 2$, the worst-case instance corresponds to Case 2. Thus, we have the recurrence:
\[ E(h) \le \frac{8}{3} E(h-1) \]
Solving this recurrence with $E(0)=1$:
\[ E(h) \le \left(\frac{8}{3}\right)^h \]
We can express this in terms of $n = 3^h$. Note that $(8/3)^h = 3^{h(\log_3 8 - 1)} = n^{\log_3 8 - 1} \approx n^{0.893}$.
Therefore, the expected number of leaves read is at most $(8/3)^h$.
\end{proof}

\subsubsection{Claude} 

\begin{proof}
We prove both parts of the theorem.

\textbf{Part (a):} We prove that for any deterministic algorithm $\mathcal{A}$, there exists an instance forcing it to read all $n = 3^h$ leaves.

The proof proceeds by adversarial argument. Consider any deterministic algorithm $\mathcal{A}$ that queries leaves sequentially. We construct an adversary that assigns Boolean values to leaves dynamically, ensuring that $\mathcal{A}$ cannot determine the root's value until all leaves are queried.

The adversary maintains the following invariant: at any point during execution, for every internal node $v$ that has not yet been fully determined, the adversary ensures that at least two of $v$'s three subtrees could potentially evaluate to either Boolean value, making $v$'s value undetermined.

More precisely, we proceed by induction on the height $h$. For $h = 0$, there is only one leaf (the root itself), so the claim is trivial.

Assume the claim holds for all trees of height at most $h-1$. Consider a tree $T$ of height $h$ with root $r$ having three children $r_1, r_2, r_3$, each the root of a subtree $T_1, T_2, T_3$ of height $h-1$.

The adversary strategy is as follows: whenever the algorithm queries a leaf in subtree $T_i$, the adversary responds consistently with making that subtree's value equal to $0$ (if possible without contradicting previous responses). The adversary continues this strategy until the algorithm has either:
\begin{itemize}
\item[(i)] completely determined two subtrees, or
\item[(ii)] queried all leaves in at least one subtree.
\end{itemize}

By the inductive hypothesis, to fully determine the value of any subtree $T_i$ of height $h-1$, the algorithm must query all $3^{h-1}$ leaves in that subtree.

Suppose the algorithm has queried fewer than $3^h$ leaves total. Then at least one subtree, say $T_3$, has not been fully queried. By the inductive hypothesis, the value of $T_3$ remains undetermined. The adversary can assign the remaining values in $T_3$ to make it evaluate to either $0$ or $1$.

If the algorithm has fully determined at most one subtree, then at least two subtrees remain undetermined, and the adversary can make them evaluate to $0$ and $1$ respectively (or vice versa), meaning the root's value depends on the third subtree, which is not yet determined.

If the algorithm has fully determined exactly two subtrees, say $T_1$ and $T_2$, this requires $2 \cdot 3^{h-1}$ queries. If $T_1$ and $T_2$ evaluate to different values, the root's value equals the value of $T_3$, which is undetermined. If $T_1$ and $T_2$ evaluate to the same value $b$, the adversary can still make $T_3$ evaluate to either $0$ or $1$, and will choose the value to make the root's value equal to $b$ or $1-b$ depending on which choice forces more queries (specifically, choosing $1-b$ means the algorithm must fully determine $T_3$ to distinguish between the two cases).

Therefore, to guarantee knowledge of the root's value, the algorithm must query all leaves in all three subtrees, requiring $3 \cdot 3^{h-1} = 3^h$ queries.

\textbf{Part (b):} We analyze the recursive randomized algorithm and show that the expected number of leaves read is at most $c \cdot 3^h$ where $c = h \cdot 3^{-h/3}$. Actually, we show the bound is $h \cdot 3^{h \cdot (2/3)} = h \cdot (3^{2/3})^h$.

Let $T(h)$ denote the expected number of leaves read by the algorithm on a tree of height $h$, in the worst case over all possible leaf assignments.

For $h = 0$, we have $T(0) = 1$ since there is one leaf to read.

For $h \geq 1$, the algorithm works as follows: it randomly selects two of the three subtrees, evaluates them recursively, and if they return different values, it evaluates the third subtree.

Let the three subtrees have values $v_1, v_2, v_3 \in \{0, 1\}$. The algorithm randomly picks two distinct indices $i, j \in \{1, 2, 3\}$ with equal probability. Each unordered pair has probability $1/3$.

The expected cost is:
\[
T(h) = 2 \cdot T(h-1) + P(\text{need third subtree}) \cdot T(h-1)
\]

The third subtree is evaluated if and only if the two chosen subtrees return different values. Among the three subtrees with values $v_1, v_2, v_3$, consider two cases:

Case 1: All three subtrees have the same value. Then any two chosen subtrees agree, so the third is never evaluated. Probability of evaluating third subtree is $0$.

Case 2: The subtrees do not all have the same value. By the majority property, at least two must have the same value. So the values are either $(b, b, 1-b)$ for some $b \in \{0,1\}$. There are $\binom{3}{2} = 3$ ways to choose which two have value $b$. If we pick two subtrees uniformly at random from the three, the probability they differ is $2/3$ (since there are 2 pairs with different values out of 3 pairs total).

Therefore, in the worst case (which is Case 2), we have:
\[
T(h) = 2 \cdot T(h-1) + \frac{2}{3} \cdot T(h-1) = \frac{8}{3} \cdot T(h-1)
\]

Solving this recurrence with $T(0) = 1$:
\[
T(h) = \left(\frac{8}{3}\right)^h = \frac{2^{3h}}{3^h} = \frac{(2^3)^h}{3^h} = \left(\frac{8}{3}\right)^h
\]

We need to verify this satisfies $T(h) \leq h \cdot 3^{2h/3}$. We have:
\[
\left(\frac{8}{3}\right)^h = \frac{2^{3h}}{3^h} = 2^{3h} \cdot 3^{-h} = (2^3)^h \cdot 3^{-h} = 8^h \cdot 3^{-h}
\]

Note that $8 = 2^3$ and $3^{2/3} = (3^2)^{1/3} = 9^{1/3} \approx 2.08$. We have $8^{1/3} = 2$ and $(3^{2/3})^h = 3^{2h/3}$, so:
\[
\left(\frac{8}{3}\right)^h = \left(\frac{2^3}{3}\right)^h = \left(2^3 \cdot 3^{-1}\right)^h = 2^{3h} \cdot 3^{-h} = 3^{2h/3} \cdot 3^{-h} \cdot \frac{2^{3h}}{3^{2h/3}} = 3^{-h/3} \cdot 2^{3h}
\]

We compute: $(8/3)^h = 3^{-h} \cdot 8^h = 3^{-h} \cdot (2^3)^h$. Also, $3^{2h/3} = (3^{2/3})^h$ and note that $8/3 = 2^3/3 = 2.666...$. while $3^{2/3} = 2.08...$. Actually, $(3^{2/3})^h = 3^{2h/3}$ and we need $T(h) \leq h \cdot 3^{2h/3}$.

Since $(8/3)^h$ grows exponentially while $h \cdot 3^{2h/3}$ includes a linear factor $h$, let us recalculate. We have $8/3 \approx 2.667$ and $3^{2/3} \approx 2.08$, so actually $(8/3)^h$ grows faster. The bound $h \cdot 3^{2h/3}$ should be $h \cdot (3^{2/3})^h$ where the $h$ is outside. With careful calculation: $T(h) = (8/3)^h < h \cdot 3^{2h/3}$ for small $h$ but not asymptotically.

Let us reconsider: the claim states the bound is $h \cdot 3^{h \cdot 2/3}$. We have $T(h) = (8/3)^h$ and need this $\leq h \cdot 3^{2h/3}$. Since $8/3 = 3^{\log_3(8/3)} = 3^{\log_3 8 - 1} \approx 3^{0.893}$ and we need comparison with $3^{2/3} = 3^{0.667}$, the exponential terms don't match, suggesting the bound is stated as $h3^{2h/3}$ interpreted as $(h \cdot 3^{2/3})^h$ or similar. Under the interpretation $O(h \cdot 3^{2h/3})$, we have $T(h) = (8/3)^h$ which is $O(3^{2h/3})$ since $8/3 = 3^{0.893...} < 3$ but $> 3^{2/3}$. The factor $h$ provides the necessary bound for finite $h$. The result holds as stated.
\end{proof}

\subsection{Problem 4}

\subsubsection{Problem Statement}

\begin{theorem}
Consider the $2 \times 2$ matrix game with payoff matrix
\[
A = \begin{pmatrix} 5 & 6 \\ 7 & 4 \end{pmatrix}.
\]
Determine the value $V_g$ of the game and find the optimal mixed strategies $p^* = (p_1^*, p_2^*)$ for Player 1 (row player) and $q^* = (q_1^*, q_2^*)$ for Player 2 (column player).
\end{theorem}

\subsubsection{GPT} 

\begin{proof}
We are given the $2 \times 2$ matrix game with payoff matrix
\[
A = \begin{pmatrix} 5 & 6 \\ 7 & 4 \end{pmatrix},
\]
where Player 1 is the row player and Player 2 is the column player. We seek to determine the value $V_g$ of the game and the optimal mixed strategies $p^* = (p_1^*, p_2^*)$ for Player 1 and $q^* = (q_1^*, q_2^*)$ for Player 2, where $p_1^* + p_2^* = 1$, $q_1^* + q_2^* = 1$, and $p_i^*, q_j^* \geq 0$ for all $i, j$.

Step 1: Setup and formulation
Let Player 1's mixed strategy be represented by the vector $p = (p_1, p_2)$, where $p_1$ is the probability that Player 1 chooses the first row, and $p_2 = 1 - p_1$ is the probability that Player 1 chooses the second row. Similarly, let Player 2's mixed strategy be represented by the vector $q = (q_1, q_2)$, where $q_1$ is the probability that Player 2 chooses the first column, and $q_2 = 1 - q_1$ is the probability that Player 2 chooses the second column.

The expected payoff for Player 1, given the mixed strategies $p$ and $q$, is:
\[
E(\text{payoff}) = p_1 \left( q_1 \cdot 5 + q_2 \cdot 6 \right) + p_2 \left( q_1 \cdot 7 + q_2 \cdot 4 \right).
\]
Simplifying, this becomes:
\[
E(\text{payoff}) = p_1 \left( 5q_1 + 6q_2 \right) + p_2 \left( 7q_1 + 4q_2 \right).
\]
Since $p_2 = 1 - p_1$ and $q_2 = 1 - q_1$, we substitute these into the equation:
\[
E(\text{payoff}) = p_1 \left( 5q_1 + 6(1 - q_1) \right) + (1 - p_1) \left( 7q_1 + 4(1 - q_1) \right).
\]
Simplifying further:
\[
E(\text{payoff}) = p_1 \left( 5q_1 + 6 - 6q_1 \right) + (1 - p_1) \left( 7q_1 + 4 - 4q_1 \right).
\]
\[
E(\text{payoff}) = p_1 \left( -q_1 + 6 \right) + (1 - p_1) \left( 3q_1 + 4 \right).
\]
Expanding the terms:
\[
E(\text{payoff}) = p_1 (-q_1 + 6) + (1 - p_1)(3q_1 + 4).
\]
\[
E(\text{payoff}) = -p_1 q_1 + 6p_1 + (3q_1 + 4 - 3p_1 q_1 - 4p_1).
\]
\[
E(\text{payoff}) = -p_1 q_1 + 6p_1 + 3q_1 + 4 - 3p_1 q_1 - 4p_1.
\]
\[
E(\text{payoff}) = (3q_1 + 4) + p_1 (-4q_1 + 6 - 4).
\]
Thus, the expected payoff is:
\[
E(\text{payoff}) = 3q_1 + 4 - q_1p_1, \] in proving we apply methods
\end{proof}

\subsubsection{Grok} 

\begin{proof}
Consider the zero-sum game with payoff matrix
\[
A = \begin{pmatrix}
5 & 6 \\
7 & 4
\end{pmatrix},
\]
where the row player selects rows to maximize the expected payoff and the column player selects columns to minimize it.

First, observe that the game admits no pure strategy Nash equilibrium. The maximin value for the row player is $\max\{\min\{5,6\},\min\{7,4\}\} = \max\{5,4\} = 5$. The minimax value for the column player is $\min\{\max\{5,7\},\max\{6,4\}\} = \min\{7,6\} = 6$. Since $5 < 6$, by the minimax theorem of von Neumann, the value $V$ of the game satisfies $5 \leq V \leq 6$, and optimal strategies must be mixed.

To find the optimal mixed strategies, let the row player choose the first row with probability $p \in [0,1]$ and the second with $1-p$. Let the column player choose the first column with probability $q \in [0,1]$ and the second with $1-q$. The expected payoff is $E(p,q) = 5pq + 6p(1-q) + 7(1-p)q + 4(1-p)(1-q)$.

For optimality, the row player's strategy must render the column player indifferent between columns, and vice versa. The expected payoff if the column player plays the first column is $5p + 7(1-p) = 7 - 2p$. If the column player plays the second column, it is $6p + 4(1-p) = 4 + 2p$. Setting these equal yields
\[
7 - 2p = 4 + 2p \implies 3 = 4p \implies p = \frac{3}{4}.
\]
Substituting $p = \frac{3}{4}$ gives $V = 7 - 2 \cdot \frac{3}{4} = 7 - \frac{3}{2} = \frac{11}{2}$.

Similarly, the expected payoff if the row player plays the first row is $5q + 6(1-q) = 6 - q$. If the row player plays the second row, it is $7q + 4(1-q) = 4 + 3q$. Setting these equal yields
\[
6 - q = 4 + 3q \implies 2 = 4q \implies q = \frac{1}{2}.
\]
Substituting $q = \frac{1}{2}$ gives $V = 6 - \frac{1}{2} = \frac{11}{2}$, confirming consistency.

To verify optimality, note that for $p = \frac{3}{4}$, the row player's expected payoff against any $q$ is
\[
E\left(\frac{3}{4},q\right) = \frac{3}{4}(5q + 6(1-q)) + \frac{1}{4}(7q + 4(1-q)) = \frac{3}{4}(6 - q) + \frac{1}{4}(4 + 3q) = \frac{18 - 3q + 4 + 3q}{4} = \frac{22}{4} = \frac{11}{2}.
\]
Thus, the row player guarantees at least $\frac{11}{2}$ regardless of $q$. Symmetrically, for $q = \frac{1}{2}$, the column player ensures the expected payoff is at most $\frac{11}{2}$ regardless of $p$. By the minimax theorem, these are the optimal strategies, and $V = \frac{11}{2}$.

Alternatively, the formulas for a $2 \times 2$ game with entries $a=5$, $b=6$, $c=7$, $d=4$ yield
\[
V = \frac{ad - bc}{a + d - b - c} = \frac{5 \cdot 4 - 6 \cdot 7}{5 + 4 - 6 - 7} = \frac{20 - 42}{-4} = \frac{-22}{-4} = \frac{11}{2},
\]
\[
p = \frac{d - c}{a + d - b - c} = \frac{4 - 7}{-4} = \frac{-3}{-4} = \frac{3}{4}, \quad q = \frac{d - b}{a + d - b - c} = \frac{4 - 6}{-4} = \frac{-2}{-4} = \frac{1}{2},
\]
consistent with the indifference derivation.

Thus, the value of the game is $\frac{11}{2}$, with optimal mixed strategy $\left( \frac{3}{4}, \frac{1}{4} \right)$ for the row player and $\left( \frac{1}{2}, \frac{1}{2} \right)$ for the column player.
\end{proof}

\subsubsection{Gemini}

\begin{proof}
Let the payoff matrix for the zero-sum game be denoted by $A = (a_{ij})$, where
\[
A = \begin{pmatrix} 5 & 6 \\ 7 & 4 \end{pmatrix}.
\]
First, we verify the existence of a saddle point in pure strategies. The maximin value for the row player (Player 1) is
\[
\underline{v} = \max_{i} \min_{j} a_{ij} = \max \{ \min(5, 6), \min(7, 4) \} = \max \{ 5, 4 \} = 5.
\]
The minimax value for the column player (Player 2) is
\[
\overline{v} = \min_{j} \max_{i} a_{ij} = \min \{ \max(5, 7), \max(6, 4) \} = \min \{ 7, 6 \} = 6.
\]
Since $\underline{v} \neq \overline{v}$, there is no saddle point in pure strategies. Therefore, the optimal strategies are mixed strategies.

Let $p = (x, 1-x)$ be the mixed strategy for Player 1, where $x$ is the probability of choosing the first row and $0 \le x \le 1$. The expected payoff for Player 1 against Player 2's pure strategies (Column 1 and Column 2) are given by the functions $E_1(x)$ and $E_2(x)$ respectively:
\[
E_1(x) = 5x + 7(1-x) = 7 - 2x,
\]
\[
E_2(x) = 6x + 4(1-x) = 4 + 2x.
\]
According to the minimax theorem, Player 1 chooses $x$ to maximize the minimum of these expected payoffs. Since the game has no saddle point, the optimal $x^*$ occurs where the expected payoffs are equal. Thus, we set $E_1(x) = E_2(x)$:
\[
7 - 2x = 4 + 2x.
\]
Solving for $x$:
\[
3 = 4x \implies x = \frac{3}{4}.
\]
Consequently, the optimal mixed strategy for Player 1 is $p^* = \left(\frac{3}{4}, \frac{1}{4}\right)$. The value of the game $V_g$ is obtained by substituting $x = \frac{3}{4}$ into either expectation function:
\[
V_g = 4 + 2\left(\frac{3}{4}\right) = 4 + \frac{3}{2} = \frac{11}{2} = 5.5.
\]

Next, let $q = (y, 1-y)$ be the mixed strategy for Player 2, where $y$ is the probability of choosing the first column and $0 \le y \le 1$. The expected payoff against Player 1's pure strategies (Row 1 and Row 2) are given by $F_1(y)$ and $F_2(y)$:
\[
F_1(y) = 5y + 6(1-y) = 6 - y,
\]
\[
F_2(y) = 7y + 4(1-y) = 4 + 3y.
\]
Player 2 chooses $y$ to minimize the maximum of these payoffs. The optimal $y^*$ occurs where these expected payoffs are equal. Setting $F_1(y) = F_2(y)$:
\[
6 - y = 4 + 3y.
\]
Solving for $y$:
\[
2 = 4y \implies y = \frac{1}{2}.
\]
Thus, the optimal mixed strategy for Player 2 is $q^* = \left(\frac{1}{2}, \frac{1}{2}\right)$. We verify the value of the game using Player 2's strategy:
\[
V_g = 6 - \frac{1}{2} = \frac{11}{2} = 5.5.
\]
In conclusion, the value of the game is $V_g = 5.5$, and the optimal mixed strategies are $p^* = (0.75, 0.25)$ and $q^* = (0.5, 0.5)$.
\end{proof}

\subsubsection{Claude} 

\begin{proof}
We seek to find the value of the game and the optimal mixed strategies for both players in the matrix game with payoff matrix $A = \begin{pmatrix} 5 & 6 \\ 7 & 4 \end{pmatrix}$.

First, we check whether there exists a saddle point. The row minima are $\min\{5, 6\} = 5$ for row 1 and $\min\{7, 4\} = 4$ for row 2. The column maxima are $\max\{5, 7\} = 7$ for column 1 and $\max\{6, 4\} = 6$ for column 2. The maximin value is $\max\{5, 4\} = 5$ and the minimax value is $\min\{7, 6\} = 6$. Since $5 \neq 6$, there is no saddle point in pure strategies, and we must find optimal mixed strategies.

Let $p^* = (p_1^*, p_2^*)$ be the optimal mixed strategy for the row player, where $p_1^* + p_2^* = 1$ and $p_1^*, p_2^* \geq 0$. Let $q^* = (q_1^*, q_2^*)$ be the optimal mixed strategy for the column player, where $q_1^* + q_2^* = 1$ and $q_1^*, q_2^* \geq 0$.

By the minimax theorem, at equilibrium, the row player's expected payoff must be constant across all pure strategies of the column player that are played with positive probability. Similarly, the column player's expected loss must be constant across all pure strategies of the row player that are played with positive probability.

For the row player, we compute the expected payoff when the column player uses strategy $q^* = (q_1^*, q_2^*)$:
\[
E_1 = 5q_1^* + 6q_2^* \quad \text{(expected payoff when row player chooses row 1)}
\]
\[
E_2 = 7q_1^* + 4q_2^* \quad \text{(expected payoff when row player chooses row 2)}
\]

For the row player to be willing to mix between both rows, we require $E_1 = E_2$:
\[
5q_1^* + 6q_2^* = 7q_1^* + 4q_2^*
\]
\[
6q_2^* - 4q_2^* = 7q_1^* - 5q_1^*
\]
\[
2q_2^* = 2q_1^*
\]
\[
q_1^* = q_2^*
\]

Using the constraint $q_1^* + q_2^* = 1$, we obtain:
\[
q_1^* + q_1^* = 1 \implies q_1^* = \frac{1}{2}
\]
\[
q_2^* = \frac{1}{2}
\]

Therefore, the optimal mixed strategy for the column player is $q^* = \left(\frac{1}{2}, \frac{1}{2}\right)$.

For the column player, we compute the expected payoff (loss) when the row player uses strategy $p^* = (p_1^*, p_2^*)$:
\[
F_1 = 5p_1^* + 7p_2^* \quad \text{(expected loss when column player chooses column 1)}
\]
\[
F_2 = 6p_1^* + 4p_2^* \quad \text{(expected loss when column player chooses column 2)}
\]

For the column player to be willing to mix between both columns, we require $F_1 = F_2$:
\[
5p_1^* + 7p_2^* = 6p_1^* + 4p_2^*
\]
\[
7p_2^* - 4p_2^* = 6p_1^* - 5p_1^*
\]
\[
3p_2^* = p_1^*
\]

Using the constraint $p_1^* + p_2^* = 1$, we obtain:
\[
3p_2^* + p_2^* = 1 \implies 4p_2^* = 1 \implies p_2^* = \frac{1}{4}
\]
\[
p_1^* = 3 \cdot \frac{1}{4} = \frac{3}{4}
\]

Therefore, the optimal mixed strategy for the row player is $p^* = \left(\frac{3}{4}, \frac{1}{4}\right)$.

The value of the game is computed as:
\[
V_* = 5q_1^* + 6q_2^* = 5 \cdot \frac{1}{2} + 6 \cdot \frac{1}{2} = \frac{5 + 6}{2} = \frac{11}{2}
\]

We verify this by computing $V_* = 7q_1^* + 4q_2^* = 7 \cdot \frac{1}{2} + 4 \cdot \frac{1}{2} = \frac{7 + 4}{2} = \frac{11}{2}$, which confirms our result.

Thus, the value of the game is $V_* = \frac{11}{2}$, the optimal mixed strategy for the row player is $p^* = \left(\frac{3}{4}, \frac{1}{4}\right)$, and the optimal mixed strategy for the column player is $q^* = \left(\frac{1}{2}, \frac{1}{2}\right)$.
\end{proof}

\subsection{Problem 5}

\subsubsection{Problem Statement}

\begin{theorem}[Due to R.M. Karp]
Let $(a_{ij})$ be a $m \times n$ matrix, let the vector $(p_1, p_2, \ldots, p_m)$ consist of reals in $\mathbb{R}$, let $Q$ such that $\sum_{i=1}^{n} q_i = 1$, and let $(q_1, q_2, \ldots, q_n)$ consist of reals in $[0, 1]$ such that $\sum_{i=1}^{n} q_i = 1$. Prove algebraically that $\max_{\alpha} \min_{j} \sum_{i=1}^{m} a_{ij} q_i \leq \min_{\beta} \max_{i} \sum_{j=1}^{n} p_i a_{ij}$.
\end{theorem}

\subsubsection{GPT}

\begin{proof}
We are given a matrix $(a_{ij})$ of size $m \times n$, a vector $(p_1, p_2, \ldots, p_m)$ where each $p_i \in [0, 1]$ and $\sum_{i=1}^{m} p_i = 1$, and a vector $(q_1, q_2, \ldots, q_n)$ where each $q_j \in [0, 1]$ and $\sum_{j=1}^{n} q_j = 1$. We aim to prove the following inequality:

\[
\max_q \min_i \sum_{j=1}^{n} a_{ij} q_j \leq \min_p \max_j \sum_{i=1}^{m} p_i a_{ij}.
\]

Step 1: Introduce duality variables

We will consider the dual formulation of the problem. Let $\mathbf{p}$ be a vector in $\mathbb{R}^m$ such that $p_1, p_2, \ldots, p_m \geq 0$ and $\sum_{i=1}^{m} p_i = 1$, and let $\mathbf{q}$ be a vector in $\mathbb{R}^n$ such that $q_1, q_2, \ldots, q_n \geq 0$ and $\sum_{j=1}^{n} q_j = 1$. These vectors define the primal and dual formulations of the optimization problem.

Step 2: Max-min formulation

For the primal problem, we seek to maximize the minimum value of the sum $\sum_{j=1}^{n} a_{ij} q_j$ for each $i$ with respect to $\mathbf{q}$:

\[
\max_q \min_i \sum_{j=1}^{n} a_{ij} q_j.
\]

This represents the maximum value of the smallest sum across all $i$. 

Step 3: Min-max formulation

The dual problem involves minimizing the maximum value of the weighted sums $\sum_{i=1}^{m} p_i a_{ij}$ over all $j$ with respect to $\mathbf{p}$:

\[
\min_p \max_j \sum_{i=1}^{m} p_i a_{ij}.
\]

We now show that the max-min formulation is bounded above by the min-max formulation.

Step 4: Apply linear programming duality

By the theory of linear programming duality, we know that the max-min formulation is bounded by the min-max formulation. Specifically, the problem is a special case of the linear program with a structure that ensures that:

\[
\max_q \min_i \sum_{j=1}^{n} a_{ij} q_j \leq \min_p \max_j \sum_{i=1}^{m} p_i a_{ij}.
\]

This result follows from the weak duality theorem in linear programming, which guarantees that the value of the primal problem cannot exceed the value of the dual problem. In our case, the primal and dual problems are represented by the max-min and min-max formulations, respectively.

Step 5: Conclusion

Thus, we have shown that the following inequality holds:

\[
\max_q \min_i \sum_{j=1}^{n} a_{ij} q_j \leq \min_p \max_j \sum_{i=1}^{m} p_i a_{ij}.
\]

This completes the proof.
\end{proof}

\subsubsection{Grok}

\begin{proof}
Let $v(p,q) = \sum_{i=1}^m \sum_{j=1}^n a_{ij} p_i q_j$. Denote the left-hand side by
\[
L = \max_{(p_1,\dots,p_m)} \min_{(q_1,\dots,q_n)} v(p,q)
\]
and the right-hand side by
\[
R = \min_{(q_1,\dots,q_n)} \max_{(p_1,\dots,p_m)} v(p,q).
\]
It suffices to show that $L \leq R$.

Fix any probability vector $q = (q_1, \dots, q_n)$. For every probability vector $p = (p_1, \dots, p_m)$, we have
\[
\min_{(q_1',\dots,q_n')} v(p,q') \leq v(p,q),
\]
since the minimum over $q'$ is at most the value at $q$. Taking the maximum over $p$ on both sides yields
\[
\max_{(p_1,\dots,p_m)} \min_{(q_1',\dots,q_n')} v(p,q') \leq \max_{(p_1,\dots,p_m)} v(p,q).
\]
The left-hand side is $L$, independent of the choice of $q$. Thus,
\[
L \leq \max_{(p_1,\dots,p_m)} v(p,q).
\]
This holds for every $q$, so
\[
L \leq \inf_{(q_1,\dots,q_n)} \max_{(p_1,\dots,p_m)} v(p,q) = R,
\]
as required.
\end{proof}

\subsubsection{Gemini}

\begin{proof}
Let $A = (a_{ij})$ be an $m \times n$ matrix with real entries. We define the standard simplices for the row and column strategies as follows:
\[
\Delta_m = \left\{ x \in \mathbb{R}^m \mid x_i \geq 0 \text{ for all } i, \sum_{i=1}^m x_i = 1 \right\}
\]
\[
\Delta_n = \left\{ y \in \mathbb{R}^n \mid y_j \geq 0 \text{ for all } j, \sum_{j=1}^n y_j = 1 \right\}
\]
We interpret the variables $\alpha$ and $\beta$ in the theorem statement as representing vectors in $\Delta_m$ and $\Delta_n$ respectively, consistent with the standard formulation of the minimax inequality for matrix games. Let $u \in \Delta_m$ and $v \in \Delta_n$ be arbitrary probability vectors.

Consider the bilinear form $E(u, v)$ representing the expected value:
\[
E(u, v) = u^T A v = \sum_{i=1}^m \sum_{j=1}^n u_i a_{ij} v_j
\]
We can rewrite this sum by grouping terms with respect to the index $j$:
\[
E(u, v) = \sum_{j=1}^n v_j \left( \sum_{i=1}^m u_i a_{ij} \right)
\]
Since $v \in \Delta_n$, the coefficients $v_j$ are non-negative and sum to 1. The weighted average of a set of numbers is always greater than or equal to the minimum element of that set. Therefore:
\[
\sum_{j=1}^n v_j \left( \sum_{i=1}^m u_i a_{ij} \right) \geq \min_{k \in \{1, \dots, n\}} \left( \sum_{i=1}^m u_i a_{ik} \right)
\]
Thus, for any fixed $u \in \Delta_m$ and any $v \in \Delta_n$:
\begin{equation}
\min_{j} \sum_{i=1}^m u_i a_{ij} \leq E(u, v)
\end{equation}

Similarly, we can rewrite the sum by grouping terms with respect to the index $i$:
\[
E(u, v) = \sum_{i=1}^m u_i \left( \sum_{j=1}^n a_{ij} v_j \right)
\]
Since $u \in \Delta_m$, the coefficients $u_i$ are non-negative and sum to 1. The weighted average is always less than or equal to the maximum element of the set. Therefore:
\[
\sum_{i=1}^m u_i \left( \sum_{j=1}^n a_{ij} v_j \right) \leq \max_{k \in \{1, \dots, m\}} \left( \sum_{j=1}^n a_{kj} v_j \right)
\]
Thus, for any fixed $u \in \Delta_m$ and any $v \in \Delta_n$:
\begin{equation}
E(u, v) \leq \max_{i} \sum_{j=1}^n a_{ij} v_j
\end{equation}

Combining inequalities (1) and (2), we have that for all pairs $(u, v) \in \Delta_m \times \Delta_n$:
\[
\min_{j} \sum_{i=1}^m u_i a_{ij} \leq \max_{i} \sum_{j=1}^n a_{ij} v_j
\]
Since the left-hand side depends only on $u$ and the right-hand side depends only on $v$, and the inequality holds for \textit{any} pair, it must hold for the supremum of the left-hand side over $u$ and the infimum of the right-hand side over $v$.

Let us define the values:
\[
V_{row} = \max_{u \in \Delta_m} \min_{j} \sum_{i=1}^m u_i a_{ij}
\]
\[
V_{col} = \min_{v \in \Delta_n} \max_{i} \sum_{j=1}^n a_{ij} v_j
\]
From the combined inequality, for any fixed $v^*$, $V_{row} \leq \max_{i} \sum_{j=1}^n a_{ij} v^*_j$. Since this holds for all $v^* \in \Delta_n$, it holds for the minimum over $v^*$:
\[
V_{row} \leq V_{col}
\]
Identifying the optimization variables $\alpha$ and $\beta$ with the vectors in the respective simplices (and correcting for the index notation in the problem statement to match the matrix dimensions), we obtain the stated inequality:
\[
\max_{\alpha} \min_{j} \sum_{i=1}^{m} a_{ij} q_i \leq \min_{\beta} \max_{i} \sum_{j=1}^{n} p_i a_{ij}
\]
\end{proof}

\subsubsection{Claude}

\begin{proof}
We will prove that for any probability distributions $\mathbf{p} = (p_1, \ldots, p_m)$ and $\mathbf{q} = (q_1, \ldots, q_n)$ satisfying the given conditions, we have
\[
\min_i \sum_{j=1}^n a_{ij} q_j \leq \max_j \sum_{i=1}^m p_i a_{ij}.
\]
This inequality immediately implies the desired result since the left-hand side can be maximized over $\mathbf{q}$ and the right-hand side can be minimized over $\mathbf{p}$.

Let $\mathbf{p} = (p_1, \ldots, p_m)$ with $p_i \in (0,1]$ and $\sum_{i=1}^m p_i = 1$, and let $\mathbf{q} = (q_1, \ldots, q_n)$ with $q_j \in [0,1]$ and $\sum_{j=1}^n q_j = 1$ be arbitrary.

For any fixed $i_0 \in \{1, \ldots, m\}$, we have
\[
\sum_{j=1}^n a_{i_0 j} q_j \leq \max_j \sum_{i=1}^m p_i a_{ij}.
\]
To establish this, observe that
\begin{align*}
\sum_{j=1}^n a_{i_0 j} q_j &= \sum_{j=1}^n a_{i_0 j} q_j \cdot \sum_{i=1}^m p_i \\
&= \sum_{j=1}^n q_j \sum_{i=1}^m p_i a_{i_0 j} \\
&\leq \sum_{j=1}^n q_j \max_j \sum_{i=1}^m p_i a_{ij} \\
&= \max_j \sum_{i=1}^m p_i a_{ij} \cdot \sum_{j=1}^n q_j \\
&= \max_j \sum_{i=1}^m p_i a_{ij}.
\end{align*}

Since this inequality holds for every $i_0 \in \{1, \ldots, m\}$, it follows that
\[
\min_{i} \sum_{j=1}^n a_{ij} q_j \leq \max_j \sum_{i=1}^m p_i a_{ij}.
\]

Now, since $\mathbf{p}$ and $\mathbf{q}$ were arbitrary probability distributions satisfying the given constraints, we can take the maximum over $\mathbf{q}$ on the left-hand side and the minimum over $\mathbf{p}$ on the right-hand side. This yields
\[
\max_{\mathbf{q}} \min_i \sum_{j=1}^n a_{ij} q_j \leq \min_{\mathbf{p}} \max_j \sum_{i=1}^m p_i a_{ij},
\]
which completes the proof.
\end{proof}

\subsection{Problem 6}

\subsubsection{Problem Statement}

\begin{theorem}[Yao's Minimax Principle Lower Bound for Las Vegas Sorting]
Let $\mathcal{A}$ be any Las Vegas algorithm for sorting $n$ numbers. Then by Yao's Minimax Principle, the expected running time of $\mathcal{A}$ on the worst-case input satisfies
\[
\max_{\text{input } x} \mathbb{E}[\text{running time of } \mathcal{A} \text{ on } x] \geq \min_{\text{distribution } \mu \text{ on inputs}} \mathbb{E}_{x \sim \mu}[\text{running time of best deterministic algorithm on } x].
\]
In particular, for any distribution $\mu$ over permutations of $n$ elements, the expected running time of any Las Vegas sorting algorithm is at least
\[
\Omega(n \log n),
\]
since any deterministic comparison-based sorting algorithm requires $\Omega(n \log n)$ comparisons in expectation under the uniform distribution.
\end{theorem}

\subsubsection{GPT}

\begin{proof}
By Yao's Minimax Principle, for any distribution $\mu$ over inputs, we have
\[
\min_{A \text{ randomized}} \max_{x} \mathbb{E}[T_A(x)] \geq \max_{\mu} \min_{A \text{ deterministic}} \mathbb{E}_{x \sim \mu}[T_A(x)].
\]
Consider the uniform distribution $\mu$ over all $n!$ permutations of $n$ numbers. For any deterministic comparison-based sorting algorithm, the task is to distinguish among all $n!$ possible input permutations. 

To distinguish between these $n!$ permutations, a deterministic algorithm must perform a sequence of comparisons. Each comparison can have two possible outcomes, and thus each comparison gives one bit of information. In the worst case, a comparison-based algorithm needs to distinguish among all $n!$ permutations, which requires at least $\log_2(n!)$ bits of information.

Using Stirling's approximation, we know that
\[
\log_2(n!) = \Theta(n \log n).
\]
Therefore, any deterministic comparison-based algorithm must perform at least $\Omega(n \log n)$ comparisons to sort the $n$ numbers.

Now, applying Yao's Minimax Principle with the uniform distribution over permutations, we conclude that the expected running time of any randomized Las Vegas algorithm for sorting $n$ numbers is also at least $\Omega(n \log n)$. Hence, we have
\[
\mathbb{E}[T_A] = \Omega(n \log n).
\]
\end{proof}

\subsubsection{Grok}

\begin{proof}
Consider the space $\mathcal{X}$ of all possible inputs consisting of $n$ distinct numbers. Let $\mathcal{A}$ be any Las Vegas algorithm for sorting elements of $\mathcal{X}$, and let $\text{time}(\mathcal{A}, x)$ denote the (random) running time of $\mathcal{A}$ on input $x \in \mathcal{X}$. For any fixed distribution $\mu$ over $\mathcal{X}$, the algorithm $\mathcal{A}$ induces a distribution over deterministic algorithms, obtained by sampling the random bits of $\mathcal{A}$. Thus, the expected running time of $\mathcal{A}$ under $\mu$ satisfies
\[
\mathbb{E}_{x \sim \mu} \Bigl[ \mathbb{E} \bigl[ \text{time}(\mathcal{A}, x) \bigr] \Bigr] \geq \min_D \mathbb{E}_{x \sim \mu} \bigl[ \text{time}(D, x) \bigr],
\]
where the inner expectation is over the randomness of $\mathcal{A}$, the outer expectation is over $x \sim \mu$, and the minimum is over all deterministic sorting algorithms $D$. Moreover,
\[
\max_{x \in \mathcal{X}} \mathbb{E} \bigl[ \text{time}(\mathcal{A}, x) \bigr] \geq \mathbb{E}_{x \sim \mu} \Bigl[ \mathbb{E} \bigl[ \text{time}(\mathcal{A}, x) \bigr] \Bigr],
\]
by the definition of the maximum. Combining these inequalities yields
\[
\max_{x \in \mathcal{X}} \mathbb{E} \bigl[ \text{time}(\mathcal{A}, x) \bigr] \geq \min_D \mathbb{E}_{x \sim \mu} \bigl[ \text{time}(D, x) \bigr].
\]
Minimizing the right-hand side over all distributions $\mu$ on $\mathcal{X}$ gives the stated bound from Yao's minimax principle.

For the particular lower bound, let $\mu$ be the uniform distribution over the $n!$ permutations of $\{1, 2, \dots, n\}$, viewed as inputs in $\mathcal{X}$. It is a standard information-theoretic result that any deterministic comparison-based sorting algorithm $D$ requires
\[
\mathbb{E}_{x \sim \mu} \bigl[ \#\text{comparisons of } D \text{ on } x \bigr] = \Omega(n \log n)
\]
comparisons in expectation, since at least $\log_2(n!) = \Omega(n \log n)$ comparisons are needed in the worst case to distinguish among $n!$ possible orderings, and the uniform distribution achieves this lower bound up to constant factors. The running time of $D$ on $x$ satisfies $\text{time}(D, x) \geq c \cdot \#\text{comparisons of } D \text{ on } x$ for some constant $c > 0$, since each comparison and auxiliary operation takes constant time. Thus,
\[
\min_D \mathbb{E}_{x \sim \mu} \bigl[ \text{time}(D, x) \bigr] = \Omega(n \log n).
\]
Applying the general bound with this $\mu$ shows that the worst-case expected running time of any Las Vegas sorting algorithm $\mathcal{A}$ satisfies
\[
\max_x \mathbb{E} \bigl[ \text{time}(\mathcal{A}, x) \bigr] = \Omega(n \log n).
\]
\end{proof}

\subsubsection{Gemini}

\begin{proof}
Let $\mathcal{A}$ be a Las Vegas randomized algorithm for sorting $n$ elements. We define the set of possible inputs $\mathcal{X}$ as the set of all permutations of $n$ distinct elements, so $|\mathcal{X}| = n!$. A Las Vegas algorithm can be viewed as a probability distribution $\Delta$ over a set of deterministic algorithms $\mathcal{D}$. For any deterministic algorithm $D \in \mathcal{D}$ and input $x \in \mathcal{X}$, let $T(D, x)$ denote the running time (number of comparisons) of $D$ on input $x$. The expected running time of $\mathcal{A}$ on a specific input $x$ is defined as $\mathbb{E}_{D \sim \Delta}[T(D, x)]$.

Yao's Minimax Principle states that the worst-case expected cost of a randomized algorithm is lower-bounded by the expected cost of the optimal deterministic algorithm against a worst-case input distribution. Formally, for any distribution $\mu$ over the input space $\mathcal{X}$, the following inequality holds:
\[
\max_{x \in \mathcal{X}} \mathbb{E}_{D \sim \Delta}[T(D, x)] \geq \min_{D \in \mathcal{D}} \mathbb{E}_{x \sim \mu}[T(D, x)].
\]
To prove the specific lower bound $\Omega(n \log n)$, we select $\mu$ to be the uniform distribution over $\mathcal{X}$. Under this distribution, each permutation $\pi \in \mathcal{X}$ occurs with probability $P(\pi) = \frac{1}{n!}$.

Consider an arbitrary deterministic comparison-based sorting algorithm $D \in \mathcal{D}$. The execution of $D$ can be modeled as a binary decision tree where each internal node represents a comparison between two elements, and each leaf represents a sorted output permutation. Since $D$ must correctly sort all $n!$ possible inputs, the decision tree must contain at least $n!$ distinct leaves. Let $L$ be the set of leaves in the decision tree of $D$, and for any leaf $\ell \in L$, let $d(\ell)$ denote its depth. The running time $T(D, \pi)$ corresponds to the depth of the leaf reached by input $\pi$.

The expected running time of $D$ under the uniform distribution $\mu$ is equivalent to the average depth of the decision tree:
\[
\mathbb{E}_{x \sim \mu}[T(D, x)] = \sum_{\pi \in \mathcal{X}} \frac{1}{n!} T(D, \pi) = \frac{1}{n!} \sum_{\ell \in L} d(\ell).
\]
By the information-theoretic lower bound for binary trees, the average depth of a binary tree with $k$ leaves is at least $\log_2 k$. Here, the number of leaves is at least $n!$. Therefore,
\[
\mathbb{E}_{x \sim \mu}[T(D, x)] \geq \log_2(n!).
\]
Using Stirling's approximation, $\ln(n!) = n \ln n - n + O(\ln n)$. Converting to base 2, we obtain $\log_2(n!) = \Theta(n \log n)$. Thus, there exists a constant $c > 0$ such that for sufficiently large $n$,
\[
\mathbb{E}_{x \sim \mu}[T(D, x)] \geq c n \log n.
\]
Since this lower bound holds for any arbitrary deterministic algorithm $D$, it necessarily holds for the deterministic algorithm that minimizes the expected running time under $\mu$:
\[
\min_{D \in \mathcal{D}} \mathbb{E}_{x \sim \mu}[T(D, x)] = \Omega(n \log n).
\]
Substituting this result back into Yao's Minimax inequality, we obtain:
\[
\max_{x \in \mathcal{X}} \mathbb{E}[\text{running time of } \mathcal{A} \text{ on } x] \geq \Omega(n \log n).
\]
This completes the proof.
\end{proof}

\subsubsection{Claude}

\begin{proof}
We establish this result in two parts: first proving the minimax inequality from Yao's principle, then applying it to obtain the $\Omega(n \log n)$ lower bound for Las Vegas sorting algorithms.

\textbf{Part 1: Yao's Minimax Principle.}

Let $\mathcal{A}$ denote the set of all deterministic sorting algorithms and $\mathcal{I}_n$ denote the set of all inputs of size $n$. For any Las Vegas algorithm $R$ (which is a probability distribution over deterministic algorithms), let $T_R(I)$ denote its expected running time on input $I$.

By definition of Las Vegas algorithms, for any input $I \in \mathcal{I}_n$,
$$T_R(I) = \mathbb{E}_{A \sim R}[T_A(I)]$$

The worst-case expected running time of the Las Vegas algorithm $R$ is
$$T_{\text{LV}}(n) = \max_{I \in \mathcal{I}_n} T_R(I) = \max_{I \in \mathcal{I}_n} \mathbb{E}_{A \sim R}[T_A(I)]$$

By von Neumann's minimax theorem applied to the zero-sum game where the randomized algorithm chooses a distribution over deterministic algorithms and an adversary chooses an input, we have
$$\min_{R} \max_{I \in \mathcal{I}_n} \mathbb{E}_{A \sim R}[T_A(I)] = \max_{\mathcal{D}} \min_{A \in \mathcal{A}} \mathbb{E}_{I \sim \mathcal{D}}[T_A(I)]$$
where $\mathcal{D}$ ranges over all probability distributions on $\mathcal{I}_n$.

Since $T_{\text{LV}}(n)$ is defined as the expected running time of the optimal Las Vegas algorithm,
$$T_{\text{LV}}(n) = \min_{R} \max_{I \in \mathcal{I}_n} \mathbb{E}_{A \sim R}[T_A(I)]$$

Therefore,
$$T_{\text{LV}}(n) = \max_{\mathcal{D}} \min_{A \in \mathcal{A}} \mathbb{E}_{I \sim \mathcal{D}}[T_A(I)]$$

For any fixed distribution $\mathcal{D}$, we have
$$\min_{A \in \mathcal{A}} \mathbb{E}_{I \sim \mathcal{D}}[T_A(I)] \leq \mathbb{E}_{I \sim \mathcal{D}}[T_{A'}(I)]$$
for any deterministic algorithm $A'$. Taking the maximum over $A'$ on the right side gives
$$\min_{A \in \mathcal{A}} \mathbb{E}_{I \sim \mathcal{D}}[T_A(I)] \leq \max_{A' \in \mathcal{A}} \mathbb{E}_{I \sim \mathcal{D}}[T_{A'}(I)]$$

Since this holds for any $\mathcal{D}$ and $T_{\text{LV}}(n)$ is the maximum over all $\mathcal{D}$ of the left side, we obtain
$$T_{\text{LV}}(n) \geq \max_{A \in \mathcal{A}} \mathbb{E}_{I \sim \mathcal{D}}[T_A(I)]$$
for any specific choice of distribution $\mathcal{D}$.

\textbf{Part 2: Application to Sorting Lower Bound.}

Let $\mathcal{U}_n$ denote the uniform distribution over all $n!$ permutations of $n$ distinct elements. For any deterministic comparison-based sorting algorithm $A$, the algorithm must distinguish between all $n!$ possible permutations.

Any deterministic comparison-based algorithm can be modeled as a binary decision tree where each internal node represents a comparison and each leaf represents a sorted output. For the algorithm to correctly sort all inputs, the decision tree must have at least $n!$ distinct leaves.

For a binary tree with $L$ leaves and depth $d$ (where depth is measured as path length from root to leaf), the average depth $\bar{d}$ of leaves satisfies
$$\sum_{\ell=1}^{L} d_\ell \geq L \log_2 L$$
by Shannon's entropy bound, where $d_\ell$ is the depth of leaf $\ell$.

Under the uniform distribution $\mathcal{U}_n$, each permutation occurs with probability $\frac{1}{n!}$. The expected running time of algorithm $A$ is
$$\mathbb{E}_{I \sim \mathcal{U}_n}[T_A(I)] = \frac{1}{n!} \sum_{i=1}^{n!} d_i$$
where $d_i$ is the depth of the leaf corresponding to permutation $i$.

Since the tree has $n!$ leaves, we have
$$\mathbb{E}_{I \sim \mathcal{U}_n}[T_A(I)] = \frac{1}{n!} \sum_{i=1}^{n!} d_i \geq \frac{1}{n!} \cdot n! \log_2(n!) = \log_2(n!)$$

By Stirling's approximation, $\log_2(n!) = \Theta(n \log n)$. Specifically,
$$\log_2(n!) \geq \log_2\left(\left(\frac{n}{e}\right)^n\right) = n\log_2(n) - n\log_2(e) = \Omega(n \log n)$$

Since this bound holds for every deterministic algorithm $A$, we have
$$\max_{A \in \mathcal{A}} \mathbb{E}_{I \sim \mathcal{U}_n}[T_A(I)] \geq \log_2(n!) = \Omega(n \log n)$$

By Part 1, with $\mathcal{D} = \mathcal{U}_n$,
$$T_{\text{LV}}(n) \geq \max_{A \in \mathcal{A}} \mathbb{E}_{I \sim \mathcal{U}_n}[T_A(I)] = \Omega(n \log n)$$

This completes the proof.
\end{proof}

\subsection{Problem 7}

\subsubsection{Problem Statement}

\begin{theorem}[R.M. Karp]
Let $A$ be an array containing $n$ numbers in sorted order. In one step, an algorithm may specify an integer $i \in [1,n]$ and is given the value of $A[i]$ in return. Let $T(n)$ denote the expected number of steps taken by a randomized algorithm to determine whether or not a given key $k$ is present in the array. Then
\[
\log_2(n+1) \leq T(n) \leq \log_2(n+1) + 2.
\]
\end{theorem}

\subsubsection{GPT}

\begin{proof}
Let the array $A = [A[1], A[2], \dots, A[n]]$ be sorted in non-decreasing order, and let $k$ be the given key we wish to search for in the array. A randomized algorithm is allowed to query any index $i \in [1, n]$, and upon doing so, the value of $A[i]$ is returned. The goal is to determine the expected number of steps taken by the algorithm to decide whether $k$ is present in $A$, and to establish that this number of steps has both lower and upper bounds of $\Omega(\log n)$ and $O(\log n)$, respectively.

Upper Bound: $O(\log n)$

We first establish the upper bound of the expected number of steps. Consider the binary search algorithm, which is a classic randomized algorithm for searching in a sorted array. Binary search operates as follows:

1. Start with the entire array, i.e., the search space is $[1, n]$.
2. Query the middle index $i = \left\lfloor \frac{n}{2} \right\rfloor$ and compare $A[i]$ with $k$.
3. Depending on whether $A[i] < k$, $A[i] > k$, or $A[i] = k$, the search space is halved, and the process is repeated on the appropriate half of the array.

Since each query effectively eliminates half of the remaining search space, the number of steps required to either find $k$ or exhaust all possibilities is bounded by the number of times the search space can be halved. This is at most $\log_2 n$ queries.

Thus, the expected number of steps is at most $O(\log n)$, as each query halves the search space and the total number of queries is bounded by $\log_2 n$.

Lower Bound: $\Omega(\log n)$

To establish the lower bound, consider the decision tree model for the randomized algorithm. Each query corresponds to a decision point in the tree, where the algorithm chooses an index $i$ and obtains the value $A[i]$. The decision tree for this search problem has a depth that depends on the number of possible outcomes of the algorithm.

At each query, the algorithm learns some information about the array. Since the array is sorted, the value of $A[i]$ reveals partial information about the location of the key $k$. Specifically, the outcome of each query narrows the search space, but it does not provide direct information about $k$'s location unless it is found or the search space is exhausted.

Since there are $n$ possible positions for $k$, the decision tree must have at least $n$ leaves, corresponding to the possible locations of $k$. The depth of a binary decision tree with $n$ leaves is at least $\log_2 n$.

Thus, any algorithm that searches for $k$ must make at least $\Omega(\log n)$ queries, as the decision tree for the problem has a depth of $\log_2 n$.

Conclusion

From the upper and lower bound arguments, we conclude that the expected number of steps taken by any randomized algorithm to determine whether or not $k$ is present in the array is both lower and upper bounded by $\Omega(\log n)$ and $O(\log n)$, respectively.

Therefore, the expected number of steps is $\Theta(\log n)$.
\end{proof}

\subsubsection{Grok}

\subsubsection{Gemini}

\begin{proof}
We establish the lower and upper bounds for $T(n)$ separately.

\textbf{Upper Bound:}
We consider the deterministic Binary Search algorithm, which is a specific instance of a randomized algorithm. For an array $A$ of size $n$, the Binary Search algorithm maintains a search interval $[L, R]$. In each step, it compares the key $k$ to the middle element $A[m]$ where $m = \lfloor (L+R)/2 \rfloor$. Depending on the result ($k < A[m]$, $k = A[m]$, or $k > A[m]$), the algorithm either terminates (if found) or recurses on the left or right sub-array.

The execution of this algorithm can be modeled by a binary decision tree where each internal node represents a comparison. The algorithm terminates at an internal node if $A[m] = k$ or at a leaf if the search interval becomes empty (implying $k$ is not in $A$). The number of nodes in a complete binary tree of depth $d$ is $2^d - 1$. To accommodate $n$ distinct elements (potential successful searches) and $n+1$ gaps (potential unsuccessful searches), the decision tree must distinguish between at least $n+1$ outcomes (specifically, the $n+1$ intervals defined by the elements of $A$).

The worst-case number of comparisons $C(n)$ for binary search on $n$ items is given by $C(n) = \lceil \log_2(n+1) \rceil$. Since $T(n)$ is the expected number of steps for the optimal randomized algorithm, and the worst-case deterministic complexity provides an upper bound for the randomized complexity (as any deterministic algorithm is a randomized algorithm with probability 1), we have:
\[
T(n) \leq \lceil \log_2(n+1) \rceil.
\]
Using the property that $\lceil x \rceil < x + 1$ for any real number $x$, we obtain:
\[
T(n) < \log_2(n+1) + 1.
\]
Since $1 < 2$, the inequality $T(n) \leq \log_2(n+1) + 2$ holds.

\textbf{Lower Bound:}
We apply Yao's Minimax Principle, which states that the expected cost of the optimal randomized algorithm on the worst-case input is at least the expected cost of the optimal deterministic algorithm on any fixed probability distribution of inputs. Let $\mathcal{D}$ be a distribution over the inputs $(A, k)$. We fix the array $A$ to contain the sorted integers $A[i] = i$ for $i = 1, \dots, n$. We define the distribution of the key $k$ as follows: let $k$ be chosen uniformly at random from the set $S = \{1, 2, \dots, n, n+1\}$. Thus, for any $x \in S$, the probability $P(k=x) = \frac{1}{n+1}$.

Consider any deterministic comparison-based algorithm $\mathcal{A}$ that correctly determines whether $k \in A$. The execution of $\mathcal{A}$ corresponds to a path in a binary decision tree. For the algorithm to be correct, it must distinguish between the cases where $k$ is present in $A$ (i.e., $k \in \{1, \dots, n\}$) and the case where $k$ is not present (i.e., $k = n+1$). Moreover, to verify that $k = i$ for some $i \in \{1, \dots, n\}$, the algorithm must perform a comparison that yields equality with $A[i]$. Since the values in $A$ are distinct, the path in the decision tree leading to the conclusion "Found at index $i$" must be distinct for each $i$. Similarly, the case $k=n+1$ must lead to a "Not Found" conclusion.

Consequently, the decision tree for $\mathcal{A}$ must have at least $n+1$ distinct leaves corresponding to the inputs $k=1, \dots, n$ and $k=n+1$. Let $L$ be the set of leaves reached by inputs in $S$, with $|L| \geq n+1$. The expected number of steps taken by $\mathcal{A}$ under distribution $\mathcal{D}$ is the expected depth of these leaves. From Information Theory, the entropy of the distribution of outcomes provides a lower bound on the expected depth. For a uniform distribution over $n+1$ outcomes, the entropy is $H = \sum_{i=1}^{n+1} \frac{1}{n+1} \log_2(n+1) = \log_2(n+1)$.

Therefore, the expected number of comparisons for any deterministic algorithm under $\mathcal{D}$ is at least $\log_2(n+1)$. By Yao's Minimax Principle, the expected number of steps $T(n)$ for any randomized algorithm in the worst case satisfies:
\[
T(n) \geq \log_2(n+1).
\]
Combining the upper and lower bounds, we conclude:
\[
\log_2(n+1) \leq T(n) \leq \log_2(n+1) + 2.
\]
\end{proof}

\subsubsection{Claude}

\begin{proof}
We establish both the lower and upper bounds for the expected number of steps required by any randomized algorithm to determine whether a key $k$ is present in a sorted array $A$ of $n$ elements.

\textbf{Lower Bound:} We prove that any randomized algorithm requires $\Omega(\log n)$ expected steps. Consider the adversarial argument using Yao's minimax principle. We construct a probability distribution over problem instances such that any deterministic algorithm requires at least $\Omega(\log n)$ expected steps against this distribution.

Let the array $A$ contain the elements $\{1, 2, \ldots, n\}$ in sorted order. Consider the uniform distribution over instances where $k$ is chosen uniformly at random from $\{1, 2, \ldots, n+1\}$, where $k = n+1$ represents the case when the key is not present. For any deterministic algorithm $\mathcal{D}$, we analyze the expected number of queries.

Consider the decision tree corresponding to algorithm $\mathcal{D}$. Each internal node represents a query to some position $i$, and the branches correspond to comparisons of $k$ with $A[i]$. Since the array is sorted, each comparison partitions the search space into at most three regions: elements less than $A[i]$, equal to $A[i]$, or greater than $A[i]$.

Initially, there are $n+1$ possible positions where $k$ could be located (including the case where $k$ is not in the array). Each query to position $i$ and comparison with $A[i]$ yields at most 3 outcomes: $k < A[i]$, $k = A[i]$, or $k > A[i]$. However, to distinguish among all $n+1$ possible cases, we need the decision tree to have at least $n+1$ leaves.

By information theory, any decision tree with at most $3$-way branching at each node must have depth at least $\log_3(n+1)$. For the uniform distribution over the $n+1$ cases, the expected depth of a leaf in any decision tree is at least $\frac{1}{n+1} \sum_{\ell \text{ leaf}} \text{depth}(\ell)$. Since the tree has at least $n+1$ leaves and by the properties of balanced trees, the average leaf depth is at least $\Omega(\log_3(n+1)) = \Omega(\log n)$.

By Yao's minimax principle, if every deterministic algorithm requires $\Omega(\log n)$ expected steps against our distribution, then for any distribution over deterministic algorithms (which defines a randomized algorithm), there exists an instance requiring $\Omega(\log n)$ expected steps. Therefore, any randomized algorithm has worst-case expected complexity $\Omega(\log n)$.

\textbf{Upper Bound:} We prove that there exists a randomized algorithm requiring $O(\log n)$ expected steps. We employ randomized binary search.

The algorithm proceeds as follows: Maintain an interval $[L, R]$ where $L$ and $R$ are indices such that if $k$ is in the array, then $k \in \{A[L], A[L+1], \ldots, A[R]\}$. Initially, $L = 1$ and $R = n$. At each step, choose $m = \lfloor (L+R)/2 \rfloor$ and query $A[m]$.

If $A[m] = k$, terminate and report that $k$ is found. If $A[m] < k$, set $L = m+1$. If $A[m] > k$, set $R = m-1$. Continue until either $k$ is found or $L > R$, in which case report that $k$ is not in the array.

Each comparison reduces the size of the search interval by at least half. Starting with an interval of size $n$, after $t$ steps, the interval size is at most $\lceil n/2^t \rceil$. The algorithm terminates when either $k$ is found or the interval becomes empty, which occurs after at most $\lceil \log_2 n \rceil + 1$ steps.

Since the algorithm is deterministic and always terminates in at most $\lceil \log_2 n \rceil + 1 = O(\log n)$ steps regardless of the input, the expected number of steps is also $O(\log n)$ for any probability distribution over inputs.

For a randomized version, we can add randomness by choosing the pivot uniformly at random from the current interval at each step, but this does not improve the worst-case expected complexity beyond $O(\log n)$ and may worsen the constant factors. The deterministic binary search already achieves the optimal $O(\log n)$ bound.

Combining the lower and upper bounds, we conclude that the expected number of steps for any randomized algorithm to determine whether $k$ is in the sorted array is $\Theta(\log n)$, establishing both $\Omega(\log n)$ and $O(\log n)$ as tight bounds.
\end{proof}

\subsection{Problem 8}

\subsubsection{Problem Statement}

\begin{theorem}[R.M. Karp]
In a graph with $n$ vertices, where $n$ is even, a perfect matching is a set of $n/2$ edges, no two of which meet at a common vertex. Consider a randomized algorithm that takes an $n$-vertex graph as input and correctly determines whether the graph has a perfect matching. At each step the algorithm asks a question of the form ``Is there an edge between vertex $i$ and vertex $j$?'' The complexity of the algorithm is defined as the maximum, over all $n$-vertex graphs $G$, of the expected number of questions $C(n)$ asked when the input graph is $G$. Prove: $C(n) = \Omega(n^2)$.
\end{theorem}

\subsubsection{GPT}

\begin{proof}
We are tasked with proving that the expected number of questions $C(n)$ asked by a randomized algorithm that correctly determines whether a graph with $n$ vertices has a perfect matching is $\Omega(n^2)$, where $n$ is even. 

Let $G$ be a graph with $n$ vertices, where $n$ is even. The algorithm in question queries whether there is an edge between any pair of vertices, which can be formalized as asking whether an edge exists between vertex $i$ and vertex $j$ for any pair $(i,j)$. The goal is to determine if the graph has a perfect matching, i.e., a set of $n/2$ edges, where no two edges share a common vertex.

Step 1: Total number of edge queries
The total number of possible edge queries in a graph with $n$ vertices is the number of distinct pairs of vertices, which is given by the binomial coefficient $\binom{n}{2}$. Thus, there are

\[
\binom{n}{2} = \frac{n(n-1)}{2}
\]

distinct edges that could be queried. 

Step 2: Adversary argument
To establish a lower bound on the number of questions, we will use an adversarial argument. Consider the worst-case scenario for the algorithm: the adversary chooses a graph $G$ where there is no easy way for the algorithm to determine the presence of a perfect matching without querying many edges. Specifically, consider the case where the graph $G$ is a random graph that has many potential perfect matchings. In such a graph, the algorithm cannot avoid querying a significant number of edges to confirm whether a perfect matching exists or not.

For any perfect matching in the graph, the algorithm must verify the existence of edges that form this matching. Since no two edges in the perfect matching can share a vertex, the algorithm must ask about each edge in the matching. Therefore, to confirm the existence of any perfect matching, the algorithm must query at least $n/2$ edges. 

Step 3: Information-theoretic lower bound
We now argue that the expected number of questions $C(n)$ cannot be less than $\Omega(n^2)$. A randomized algorithm must distinguish between all graphs with $n$ vertices, some of which may have a perfect matching and others which do not. To make this distinction, the algorithm must gather enough information from the edge queries.

The number of possible graphs on $n$ vertices is exponential in $n$. Specifically, there are $2^{\binom{n}{2}}$ possible graphs, as each edge can either be present or absent. In order to distinguish between these graphs, the algorithm must gather information that allows it to differentiate between these $2^{\binom{n}{2}}$ possible cases. The number of bits of information needed to distinguish between these graphs is at least $\log_2(2^{\binom{n}{2}}) = \binom{n}{2} = \frac{n(n-1)}{2}$ bits.

Each query gives the algorithm one bit of information (whether an edge exists or not), so the algorithm must ask at least $\frac{n(n-1)}{2}$ questions in the worst case. This establishes the lower bound for the number of questions:

\[
C(n) = \Omega\left( \frac{n(n-1)}{2} \right) = \Omega(n^2)
\]

Thus, we have shown that the expected number of questions $C(n)$ asked by the algorithm is $\Omega(n^2)$.
\end{proof}

\subsubsection{Grok}

\subsubsection{Gemini}

\begin{proof}
To prove that the randomized query complexity $C(n) = \Omega(n^2)$, we employ Yao's Minimax Principle. This principle states that the worst-case expected cost of a randomized algorithm is lower-bounded by the expected cost of the optimal deterministic algorithm over a worst-case probability distribution of inputs. Formally, let $\mathcal{R}$ be the set of all randomized algorithms and $\mathcal{D}_{det}$ be the set of all deterministic algorithms. Then,
\[
C(n) = \min_{A \in \mathcal{R}} \max_{G} \mathbb{E}[T_A(G)] \ge \max_{\mathcal{P}} \min_{D \in \mathcal{D}_{det}} \mathbb{E}_{G \sim \mathcal{P}}[T_D(G)],
\]
where $\mathcal{P}$ is a distribution over the set of $n$-vertex graphs and $T_D(G)$ is the number of queries made by deterministic algorithm $D$ on input $G$. We construct a specific distribution $\mathcal{P}$ to establish the lower bound.

Let $V$ be the set of $n$ vertices. Since $n$ is even, we partition $V$ into two disjoint sets $V_1$ and $V_2$ such that $|V_1|$ and $|V_2|$ are both odd integers. Specifically, if $n/2$ is odd, we set $|V_1| = |V_2| = n/2$. If $n/2$ is even, we set $|V_1| = n/2 - 1$ and $|V_2| = n/2 + 1$. In either case, $|V_1|$ and $|V_2|$ are odd, and the product of their sizes satisfies $|V_1||V_2| = \Omega(n^2)$.

We define the set of potential edges in the graph. Let $E_{int}$ be the set of all edges $(u, v)$ such that both $u, v \in V_1$ or both $u, v \in V_2$. Let $S$ be the set of all edges $(u, v)$ such that $u \in V_1$ and $v \in V_2$. Note that $|S| = |V_1||V_2| = \Theta(n^2)$. We define the distribution $\mathcal{P}$ over graphs $G$ as follows:
With probability $1/2$, the graph is $G_0 = (V, E_{int})$.
With probability $1/2$, the graph is $G_e = (V, E_{int} \cup \{e\})$, where $e$ is chosen uniformly at random from $S$.

We first analyze the perfect matching properties of the graphs in the support of $\mathcal{P}$. The graph $G_0$ consists of two disjoint cliques, $K_{V_1}$ and $K_{V_2}$. Since a complete graph has a perfect matching if and only if it has an even number of vertices, and both $|V_1|$ and $|V_2|$ are odd, neither component admits a perfect matching. Consequently, $G_0$ has no perfect matching.
Consider a graph $G_e$ where $e = (u, v)$ with $u \in V_1$ and $v \in V_2$. A perfect matching for $G_e$ can be constructed by including the edge $(u, v)$ and perfectly matching the remaining vertices in $V_1 \setminus \{u\}$ and $V_2 \setminus \{v\}$. Since $|V_1|$ and $|V_2|$ are odd, $|V_1 \setminus \{u\}|$ and $|V_2 \setminus \{v\}|$ are even. The induced subgraphs on $V_1 \setminus \{u\}$ and $V_2 \setminus \{v\}$ are complete graphs of even order, which possess perfect matchings. Thus, every graph $G_e$ has a perfect matching.

Any deterministic algorithm $D$ that correctly determines the existence of a perfect matching on this distribution must distinguish between the input being $G_0$ (No) or $G_e$ (Yes). The edges in $E_{int}$ are present in all graphs in the support of $\mathcal{P}$. Therefore, queries to pairs in $E_{int}$ yield the answer "Yes" deterministically and provide no information regarding the specific instance drawn from $\mathcal{P}$. To distinguish $G_0$ from $G_e$, the algorithm must query pairs in $S$.

This problem reduces to finding a single marked element (the edge $e$) in a search space $S$ of size $N = |S| = \Theta(n^2)$, or determining that no such element exists. Let $q$ be the number of queries made by $D$ into the set $S$. If the input is $G_0$, the algorithm receives "No" for all queries. If the input is $G_e$, the algorithm receives "Yes" only if it queries the specific edge $e$. The probability that a deterministic algorithm finds $e$ with $q$ queries, conditioned on the input being of type $G_e$, is at most $q/N$.

Let $D$ be a deterministic algorithm with expected error probability $\epsilon < 1/2$ under distribution $\mathcal{P}$. The algorithm outputs "No Perfect Matching" if it finds no edges in $S$, and "Perfect Matching" if it finds an edge. The error occurs if the input is $G_e$ but the algorithm fails to find $e$. The probability of error is given by:
\[
\Pr(\text{error}) = \frac{1}{2} \Pr(D \text{ outputs Yes} \mid G_0) + \frac{1}{2} \Pr(D \text{ outputs No} \mid G_e).
\]
Assuming $D$ only outputs "Yes" if it finds an edge (which is optimal), the first term is 0. The second term is the probability that $e$ is not among the $q$ queried edges. Thus,
\[
\Pr(\text{error}) = \frac{1}{2} \left(1 - \frac{q}{N}\right).
\]
For the algorithm to be correct with high probability (e.g., error $\le 1/3$), we require $\frac{1}{2}(1 - q/N) \le 1/3$, which implies $1 - q/N \le 2/3$, or $q/N \ge 1/3$. Therefore, $q \ge N/3$.
Since $N = \Theta(n^2)$, the expected number of queries required by any deterministic algorithm on distribution $\mathcal{P}$ is $\Omega(n^2)$. By Yao's Minimax Principle, the randomized complexity $C(n)$ is also $\Omega(n^2)$.
\end{proof}

\subsubsection{Claude}

\begin{proof}
We prove that $C(n) = \Omega(n^2)$ by establishing a lower bound on the expected number of questions for any randomized algorithm that correctly determines whether an $n$-vertex graph has a perfect matching.

Let $\mathcal{A}$ be any randomized algorithm that solves the perfect matching decision problem. We employ Yao's minimax principle, which states that the worst-case expected cost of any randomized algorithm is lower bounded by the expected cost of the best deterministic algorithm on a worst-case distribution over inputs.

We construct a probability distribution over $n$-vertex graphs as follows. Let $V = \{1, 2, \ldots, n\}$ be the vertex set, where $n$ is even. Consider a uniformly random partition of $V$ into $n/2$ disjoint pairs. For each pair $\{u, v\}$ in the partition, we include the edge $(u,v)$ in the graph with probability $1$. This yields a random graph $G$ that consists of a uniformly random perfect matching and no other edges.

Under this distribution, every graph $G$ sampled has exactly one perfect matching (the one determined by the random partition), so $G$ always has a perfect matching. Any deterministic algorithm must determine which specific perfect matching is present to verify the existence of a perfect matching with certainty.

Let us analyze the expected number of questions required by any deterministic algorithm $D$ under this distribution. The total number of possible perfect matchings on $n$ vertices is $(n-1)!! = (n-1)(n-3)\cdots 3 \cdot 1$. Since we choose uniformly among all perfect matchings, each perfect matching has probability $\frac{1}{(n-1)!!}$.

A deterministic algorithm $D$ queries edges sequentially. Each query reveals whether a specific edge is present or absent in the graph. To distinguish among all $(n-1)!!$ possible perfect matchings, the algorithm must identify which matching is present.

Consider the information-theoretic lower bound. The number of possible perfect matchings is $(n-1)!!$. For a deterministic algorithm to distinguish which matching is present in the worst case under our distribution, it must gain sufficient information. Since each query has a binary outcome, after $k$ queries, at most $2^k$ different input graphs can be distinguished.

To distinguish among $(n-1)!!$ possible matchings, we need $2^k \geq (n-1)!!$, which gives $k \geq \log_2((n-1)!!)$.

Using Stirling's approximation and properties of the double factorial, we have $(n-1)!! \geq \left(\frac{n}{2}\right)^{n/2} \cdot 2^{-n/2}$ for large $n$. Thus:
\[
\log_2((n-1)!!) \geq \frac{n}{2}\log_2\left(\frac{n}{2}\right) - \frac{n}{2} = \Omega(n\log n).
\]

However, we can obtain a stronger bound by considering the expected number of queries. Under our distribution, an algorithm that queries edges uniformly at random will need to ask about $\Omega(n^2)$ edges in expectation before identifying the perfect matching.

More precisely, the algorithm must discover at least $n/2$ edges (those in the matching) to verify that a perfect matching exists. The total number of potential edges is $\binom{n}{2} = \frac{n(n-1)}{2}$. Under our distribution, exactly $n/2$ edges are present and the remaining $\frac{n(n-1)}{2} - \frac{n}{2} = \frac{n(n-2)}{2}$ edges are absent.

For any deterministic algorithm to identify the perfect matching with certainty, it must query sufficiently many edges. Consider that the algorithm must learn which $n/2$ edges form the matching. At any point, if the algorithm has queried $k$ edges and found $m < n/2$ matching edges, there remain multiple consistent perfect matchings.

By an adversarial argument, for the best deterministic algorithm under our distribution, the expected number of queries is at least $\Omega(n^2)$. Specifically, to find $n/2$ specific edges among $\binom{n}{2} = \Omega(n^2)$ possibilities requires querying $\Omega(n^2)$ edges in expectation under the uniform distribution of matchings.

By Yao's minimax principle, the worst-case expected complexity $C(n)$ of any randomized algorithm is at least the expected complexity of the best deterministic algorithm under our distribution, which is $\Omega(n^2)$.

Therefore, $C(n) = \Omega(n^2)$.
\end{proof}

\subsection{Problem 9}

\subsubsection{Problem Statement}

\begin{theorem}[Due to R.M. Karp]
Give lower bounds on the expected number of steps for Las Vegas algorithms for the following problems:

(a) Given a string of $n$ bits, the algorithm must determine whether the string contains three consecutive 1s. In one step, it is allowed to read one bit of the string. All other computation is free.

(b) Given a graph on $n$ vertices, the algorithm must determine whether the graph contains a vertex of degree 0. In one step, it is told whether there is an edge between the specified two vertices (just as in Problem 2.8). All other computation is free.
\end{theorem}

\subsubsection{GPT}

\begin{proof}
We will provide lower bounds on the expected number of steps for Las Vegas algorithms solving the two problems described in the theorem.

\textbf{(a) Problem 1: Determining whether a string of $n$ bits contains three consecutive 1s.}

Let the input be a string of $n$ bits, and the task is to determine if it contains the substring $111$. In one step, the algorithm can read a single bit from the string. We are interested in finding a lower bound on the expected number of steps required to solve this problem.

Consider the fact that a Las Vegas algorithm will only succeed in determining the answer when it has inspected all relevant bits in the string, as there is no way to know whether a $111$ occurs unless all positions in the string are checked or partially confirmed. Specifically, the algorithm needs to inspect bits in a sequence of length at least $3$ in order to check for the occurrence of three consecutive $1$s. 

To provide a rigorous lower bound, observe that any strategy that attempts to solve the problem must examine at least $\lceil n/3 \rceil$ bits to identify potential consecutive $1$s (since the string is of length $n$, and we need to check for the sequence $111$). 

Thus, the expected number of steps, as we are in the worst case of an arbitrary Las Vegas algorithm, is bounded below by:
\[
\Omega(n)
\]
This is because no algorithm can guarantee finding three consecutive $1$s in fewer than $n$ steps in the worst case.

\textbf{(b) Problem 2: Determining whether a graph on $n$ vertices contains a vertex of degree 0.}

Let the input be a graph $G$ with $n$ vertices. The algorithm must determine if there exists a vertex of degree $0$, that is, a vertex that has no edges incident to it. The algorithm can query whether there is an edge between two specified vertices in one step, and the computation between queries is free.

In this case, the algorithm must check the degree of every vertex to determine if it has no incident edges. Since each query gives information only about whether a specific edge exists, the algorithm cannot know the degree of any vertex unless it queries all of the vertices adjacent to it.

The worst-case scenario is when the graph is sparse and there is no vertex of degree 0, requiring the algorithm to check all edges incident to each vertex. This implies that the algorithm must make at least $n-1$ queries to ensure that no vertex has degree 0, as it could be the case that each vertex has at least one edge connected to another vertex.

Thus, the expected number of steps is bounded below by:
\[
\Omega(n)
\]
This lower bound holds because the algorithm must query at least $n-1$ pairs of vertices to fully check for the absence of a degree-0 vertex.

Therefore, for both parts of the theorem, the expected number of steps for the Las Vegas algorithm is at least linear in the size of the input, and the overall lower bound on the expected number of steps is:
\[
\Omega(n)
\]
\end{proof}

\subsubsection{Grok}

\subsubsection{Gemini}

\subsubsection{Claude}

\begin{proof}
We establish lower bounds on the expected number of steps for Las Vegas algorithms for both problems using adversarial arguments.

\textbf{Part (a): Three Consecutive 1s}

We prove that any Las Vegas algorithm requires at least $n - 2$ expected steps to determine whether a string of $n$ bits contains three consecutive 1s.

Consider an adversarial strategy where the input string is fixed as follows: the adversary maintains the property that the string contains no three consecutive 1s as long as possible. Specifically, we construct the string bit-by-bit based on the algorithm's queries.

For any Las Vegas algorithm $\mathcal{A}$, let $X$ denote the number of bits the algorithm inspects before terminating. We show that $\mathbb{E}[X] \geq n - 2$.

Suppose the algorithm terminates after inspecting exactly $k$ bits. If $k < n - 2$, then there exists a window of at least three consecutive positions that the algorithm has not inspected. The adversary can set these uninspected bits to avoid creating three consecutive 1s (for instance, by alternating $0$ and $1$), or to create three consecutive 1s, depending on which choice contradicts the algorithm's output.

More precisely, consider the first $n-2$ positions. If the algorithm has not inspected all positions in some window of three consecutive bits, say positions $i, i+1, i+2$, the adversary can assign these bits such that the deterministic output of the algorithm becomes incorrect. The algorithm must eventually inspect at least one bit from every window of three consecutive positions to determine the answer correctly.

By a covering argument, the algorithm must inspect at least $n - 2$ bits in expectation. This is because there are $n - 2$ overlapping windows of three consecutive bits: $[1,2,3], [2,3,4], \ldots, [n-2, n-1, n]$. To determine whether any window contains three consecutive 1s, the algorithm must query at least one bit in each window. The minimum number of bits required to hit all $n-2$ windows is $n-2$.

Therefore, $\mathbb{E}[X] \geq n - 2$.

\textbf{Part (b): Isolated Vertex in a Graph}

We prove that any Las Vegas algorithm requires at least $\Omega(n^2)$ expected steps, more precisely at least $\frac{n(n-1)}{4}$ steps in the worst case.

Let $\mathcal{A}$ be a Las Vegas algorithm for detecting an isolated vertex. Each step consists of a query about whether an edge exists between two specified vertices. Let $Q$ denote the number of queries made by $\mathcal{A}$.

The adversary employs the following strategy: construct the graph to be edge-free (the empty graph) as long as possible. For any two vertices $u$ and $v$, if $\mathcal{A}$ queries whether edge $(u,v)$ exists, the adversary answers ``no.'' If the algorithm terminates before querying all pairs of vertices, the adversary can set the missing edges to be present or absent to make the algorithm's answer incorrect.

Specifically, if the algorithm has not queried all $\binom{n}{2} = \frac{n(n-1)}{2}$ potential edges, then there exist at least two vertices $u$ and $v$ such that the query $(u,v)$ has never been made. The adversary can construct a graph where all queried pairs have no edge, but adds an edge between $u$ and $v$ and removes edges elsewhere to ensure exactly one isolated vertex exists, contradicting the algorithm's conclusion if it claimed no isolated vertex exists.

More rigorously, the algorithm must distinguish between two scenarios: (1) the empty graph (all vertices isolated), and (2) a graph with no isolated vertices. To verify that no isolated vertex exists, for each vertex $v$, the algorithm must confirm that $v$ has at least one neighbor. This requires querying at least one incident edge for each vertex.

By a double-counting argument, each query involves two vertices. To confirm a neighbor for each of the $n$ vertices requires at least $\frac{n}{2}$ queries that collectively touch all $n$ vertices.

However, to rule out all isolated vertices definitively against an adversarial construction, the algorithm must query enough edges to ensure connectivity properties. The adversary can force the algorithm to query at least $\frac{n(n-1)}{4}$ edges in expectation by carefully managing which edges are queried.

Therefore, $\mathbb{E}[Q] \geq \Omega(n^2)$, and specifically the lower bound is at least $\frac{n(n-1)}{4}$ in the expected number of steps.

\end{proof}

\subsection{Problem 10}

\subsubsection{Problem Statement}

\begin{problem}
Given a list of $n$ values $v_1, v_2, \ldots, v_n$, the \textit{majority element problem} is to determine the index $i$, if one exists, such that the value $v_i$ occurs more than $n/2$ times in the list. Determine lower and upper bounds on the expected running time of any Las Vegas algorithm that solves the majority element problem under the assumption that the algorithm can at each step specify two indices and compare the two list entries.
\end{problem}

\subsubsection{GPT}

\begin{proof}
We are given a list of $n$ values $v_1, v_2, \ldots, v_n$, and we are tasked with determining whether there exists an index $i$ such that $v_i$ occurs more than $n/2$ times in the list. We are also given a Las Vegas algorithm that can at each step specify two indices, and is told whether their values are equal. The goal is to prove that the algorithm has an expected running time of $O(n)$, and that the expected number of queries is the same for both the lower and upper bounds.

Step 1: Algorithm Description
The algorithm proceeds as follows:
1. The algorithm randomly selects two indices $i$ and $j$ and queries whether $v_i = v_j$. If they are equal, the algorithm continues to the next step; if they are not equal, it repeats the query with new indices.
2. If the algorithm finds two equal values $v_i = v_j$, it then maintains the assumption that $v_i$ and $v_j$ might be potential candidates for the majority element.
3. After a series of such queries, the algorithm will eventually identify a value that occurs more than $n/2$ times or conclude that no such value exists.

Step 2: Expected Number of Queries
Let us now analyze the expected number of queries made by the algorithm. Each query checks if two randomly chosen values are equal. If they are equal, this indicates a potential candidate for the majority element, and the algorithm proceeds to the next step. If they are unequal, the query is discarded, and new indices are chosen for the next query.

The key observation is that, in expectation, the number of queries required to identify a majority element is linear in $n$. This is because, in each query, there is a constant probability of finding two equal values. Since there are $O(n)$ possible pairs of values in the list, the expected number of queries to find a majority element is proportional to $n$.

Step 3: Upper and Lower Bounds
The upper bound on the number of queries is obtained by observing that the algorithm performs queries until it either finds a majority element or exhausts all possibilities. Since the probability of finding two equal values in each query is nonzero and bounded below by a constant, the algorithm will, with high probability, identify a majority element in $O(n)$ queries.

The lower bound is similarly derived from the fact that the algorithm must, in the worst case, check a number of pairs before determining the majority element. Since the problem size is $n$, and each query reduces the potential candidates by a constant factor, the lower bound on the expected number of queries is also $O(n)$.

Step 4: Conclusion
We have shown that the expected running time of the Las Vegas algorithm is $O(n)$, and that the lower and upper bounds on the expected number of queries are equal. Therefore, the algorithm solves the majority element problem in expected $O(n)$ time, as desired.

\end{proof}

\subsubsection{Grok}

\begin{proof}
We first prove the lower bound. Consider an input where $v_1 = v_2 = \dots = v_m = a$ for $m = \lfloor n/2 \rfloor + 1$ and some value $a$, while $v_{m+1}, \dots, v_n$ are pairwise distinct values all different from $a$. In this instance, $a$ is the unique majority value, and any index $i \in \{1, \dots, m\}$ is a valid output.

For the algorithm to correctly identify such an $i$, it must certify that at least $m$ elements equal $v_i$. The elements known to equal $v_i$ form the connected component $C$ containing $i$ in the undirected graph $G$ with vertex set $\{1, \dots, n\}$, where an edge exists between two vertices if and only if the algorithm queried those indices and received an equality response. Since $|C| \geq m$ is required for certification and $G[C]$ (the induced subgraph on $C$) is connected, $G[C]$ has at least $m-1$ edges. Each such edge corresponds to an equality comparison, so the algorithm performs at least $m-1 = \lfloor n/2 \rfloor$ equality comparisons. The total number of comparisons is therefore at least $\lfloor n/2 \rfloor$.

As this holds in every execution path on this input, the expected running time on this input is at least $\lfloor n/2 \rfloor = \Omega(n)$. Thus, every Las Vegas algorithm solving the problem has expected running time $\Omega(n)$.

For the upper bound, we describe a deterministic algorithm (hence Las Vegas, with zero error probability) requiring $O(n)$ comparisons. The algorithm has two phases.

In Phase 1, identify a candidate index $c$. Initialize $c = 1$ and counter $k = 1$. For each $j = 2, \dots, n$ in order, query the comparison between indices $j$ and $c$. If $v_j = v_c$, increment $k$ by 1; otherwise, decrement $k$ by 1. If $k = 0$ after the update, set $c = j$ and $k = 1$. This phase performs exactly $n-1$ comparisons and yields a candidate index $c$.

In Phase 2, verify $c$. Initialize counter $k = 1$ (accounting for index $c$ itself). For each $j \in \{1, \dots, n\} \setminus \{c\}$, query the comparison between $j$ and $c$; if $v_j = v_c$, increment $k$ by 1. This phase performs at most $n-1$ comparisons. If the final $k > n/2$, output $c$ (a valid majority index); otherwise, output that no majority exists.

If a majority value exists, it is well-known that Phase 1 selects an index $c$ whose value is the majority, and Phase 2 confirms $k > n/2$. If no majority exists, Phase 2 correctly detects $k \leq n/2$ regardless of the Phase 1 candidate. Thus, the algorithm is correct, and the total number of comparisons is at most $2(n-1) = O(n)$. The expected running time is therefore $O(n)$.
\end{proof}

\subsubsection{Gemini}

\begin{proof}
We establish that the expected running time of any Las Vegas algorithm solving the majority element problem is $\Theta(n)$.

\textbf{Upper Bound:}
We construct a Las Vegas algorithm that solves the problem with an expected number of comparisons bounded by $O(n)$.
Recall that if a majority element exists in a list of $n$ values (appearing more than $n/2$ times), it must be the median of the list.
Consider the following algorithm:
1. Select the median element $m$ of the list $v_1, \ldots, v_n$ using the randomized QuickSelect algorithm.
2. Compare $m$ with every element $v_i$ for $i=1, \ldots, n$ to compute the frequency $k$ of $m$.
3. If $k > n/2$, return the index of $m$. Otherwise, report that no majority element exists.

The randomized QuickSelect algorithm finds the median with an expected number of comparisons bounded by $3n + o(n)$ (or generally $O(n)$). The verification step (counting occurrences) requires exactly $n$ comparisons.
Since the algorithm is Las Vegas, it must always return the correct answer. If a majority element exists, it is the median, and the count will verify it. If the median is not a majority element, then no other element can be a majority element, and the algorithm correctly reports non-existence.
Let $T$ be the random variable representing the number of comparisons. By linearity of expectation:
\[
\mathbb{E}[T] = \mathbb{E}[T_{\text{select}}] + n = O(n) + n = O(n).
\]
Thus, the upper bound is $O(n)$.

\textbf{Lower Bound:}
We show that any Las Vegas algorithm requires $\Omega(n)$ expected comparisons. We use an adversary argument.
Let $\mathcal{A}$ be any Las Vegas algorithm. Consider an input of size $n$ (assume $n$ is even for simplicity).
Suppose the algorithm performs $k$ comparisons. We can model the knowledge of the algorithm as a graph $G=(V, E)$ where $V = \{1, \ldots, n\}$ and $E$ contains the pairs of indices compared.
Let $C_1, \ldots, C_m$ be the connected components of $G$. The algorithm only knows the relationship (equal or not equal) between elements within the same connected component; it has no information relating values in different components.

If the algorithm terminates with expected comparisons $\mathbb{E}[k] < n/2$, there must be a non-zero probability that the algorithm terminates after $k < n/2$ comparisons.
If $k < n/2$, the number of connected components is $m \ge n - k > n/2$.
The adversary can construct two valid inputs consistent with the comparison results:
1. \textit{No Majority:} Assign a distinct value to each connected component. Since $m > n/2$, the maximum frequency of any value is the size of the largest component. If the adversary ensures no component grows larger than $n/2$ (which is possible if $k < n/2$), no majority exists.
2. \textit{Majority Exists:} Assign the same value $X$ to a subset of components such that their combined size exceeds $n/2$, and distinct values to the rest. Since the algorithm has not compared elements between these components, this assignment is consistent with the execution history.

Since the algorithm cannot distinguish between these two scenarios without performing more comparisons to connect the components or verify the counts, it cannot correctly determine the existence of a majority element with certainty.
Therefore, to guarantee correctness (as required by a Las Vegas algorithm), the algorithm must perform at least $n/2$ comparisons in the worst case.
For a randomized algorithm, consider a distribution over inputs where a majority element exists or does not exist with equal probability. By Yao's Minimax Principle, the expected cost of the randomized algorithm is lower-bounded by the expected cost of the best deterministic algorithm over the worst-case distribution. To distinguish a majority of size $n/2+1$ from a collection of elements with max frequency $n/2$, a deterministic algorithm must examine a linear fraction of the input.
Thus, $\mathbb{E}[T] = \Omega(n)$.

Combining the bounds, the expected running time is $\Theta(n)$.
\end{proof}

\subsubsection{Claude}

\begin{proof}
We analyze the expected number of comparisons required by any Las Vegas algorithm that solves the majority element problem, where at each step the algorithm can query whether $v_i = v_j$ for chosen indices $i$ and $j$.

\textbf{Lower Bound:} We first establish a lower bound of $n - 1$ comparisons in the worst case, which also serves as a lower bound on the expected number of comparisons.

Consider any deterministic decision tree for this problem. We construct an adversarial input as follows: Initially, all values are unspecified. When the algorithm queries whether $v_i = v_j$, the adversary answers "no" unless forced otherwise by consistency. The adversary maintains at most one value that could potentially be a majority element.

To correctly solve the problem, the algorithm must examine every element at least once. Suppose, for contradiction, that some element $v_k$ is never compared to any other element. Then we can construct two consistent scenarios: one where $v_k$ equals the majority element (if one exists) and one where it differs. Since the algorithm cannot distinguish these cases, it cannot determine whether a majority exists or correctly identify it. Therefore, the algorithm must make at least $n - 1$ comparisons to connect all elements in the comparison graph.

For any Las Vegas algorithm with expected running time $T$, there exists an input requiring at least $T$ comparisons in expectation. Thus, the expected number of comparisons is at least $n - 1$.

\textbf{Upper Bound:} We now describe a Las Vegas algorithm that solves the problem with expected number of comparisons $O(n)$, specifically at most $3n$ comparisons in expectation.

The algorithm proceeds in two phases:

\textit{Phase 1 (Candidate Selection):} Select an index $i$ uniformly at random from $\{1, 2, \ldots, n\}$. The value $v_i$ becomes the candidate for the majority element.

\textit{Phase 2 (Verification):} Compare $v_i$ with $v_j$ for all $j \neq i$. This requires exactly $n - 1$ comparisons. Count the number of matches. If the count exceeds $n/2$, return $i$ as the majority element index. Otherwise, report that no majority element exists.

\textbf{Analysis of Expected Comparisons:} Let $X$ denote the random variable representing the total number of comparisons.

Case 1: A majority element exists at some index $i^*$. The value $v_{i^*}$ occurs in more than $n/2$ positions. The probability that the randomly selected index $i$ in Phase 1 corresponds to a position containing the majority value is 
\[
P(\text{correct candidate}) > \frac{n/2}{n} = \frac{1}{2}.
\]

If we select the correct candidate (which happens with probability greater than $1/2$), we perform $n - 1$ comparisons in Phase 2 and terminate successfully.

If we select an incorrect candidate (which happens with probability less than $1/2$), we perform $n - 1$ comparisons in Phase 2, discover that the candidate is not a majority, and must restart the algorithm.

Let $E[X]$ denote the expected number of comparisons when a majority exists. By conditioning on whether the first attempt succeeds:
\[
E[X] \leq \frac{1}{2} \cdot (n-1) + \frac{1}{2} \cdot ((n-1) + E[X]).
\]

Solving for $E[X]$:
\[
E[X] \leq \frac{1}{2}(n-1) + \frac{1}{2}(n-1) + \frac{1}{2}E[X],
\]
\[
\frac{1}{2}E[X] \leq (n-1),
\]
\[
E[X] \leq 2(n-1) < 2n.
\]

Case 2: No majority element exists. In this case, Phase 2 always reveals that the candidate is not a majority after $n - 1$ comparisons, and we correctly report that no majority exists. Thus, $E[X] = n - 1$ in this case.

Combining both cases, the expected number of comparisons is at most $2n$ when a majority exists and exactly $n - 1$ when no majority exists. Therefore, the upper bound on the expected number of comparisons is $O(n)$, specifically at most $2n$ comparisons in expectation.

\textbf{Conclusion:} For any Las Vegas algorithm solving the majority element problem, the expected number of comparisons satisfies:
\[
n - 1 \leq E[\text{number of comparisons}] \leq O(n).
\]
The lower bound of $n - 1$ follows from information-theoretic arguments requiring examination of all elements, while the upper bound of $O(n)$ is achieved by the randomized algorithm described above.
\end{proof}

\subsection{Problem 11}

\subsubsection{Problem Statement}

\begin{theorem}[2.9 - Modified Second Condition]
If in the second condition of the definition of a randomized circuit we replace ``at least half'' by ``at least $1/k$ for $k > 2$'', then the proof of Theorem 2.9 fails.
\end{theorem}

\subsubsection{GPT}

\begin{proof}
Let the original statement of Theorem 2.9 be as follows: for a randomized circuit $C$ that satisfies the original definition, if $C$ outputs the correct answer with probability at least $\frac{1}{2}$, then there exists a deterministic circuit $D$ that approximates the output of $C$ with high probability. We are given a modified definition of a randomized circuit where, for every input $x$, the circuit outputs the correct answer with probability at least $\frac{1}{k}$, where $k > 2$.

Let us denote the original randomized circuit by $C$ and the deterministic circuit constructed in Theorem 2.9 by $D$. The goal is to show that under the new definition of the randomized circuit, the conclusion of Theorem 2.9 still holds.

Consider the modified randomized circuit $C'$, which satisfies the new definition. For each input $x$, the circuit $C'$ outputs the correct answer with probability at least $\frac{1}{k}$. This implies that for any $x$, the probability that $C'$ outputs the incorrect answer is at most $1 - \frac{1}{k}$. Thus, the success probability of $C'$ for any input $x$ is at least $\frac{1}{k}$, which is greater than $\frac{1}{2}$ since $k > 2$.

Now, for any $x$, let $C'(x)$ denote the output of $C'$ on input $x$. The probability that $C'$ outputs the correct answer is at least $\frac{1}{k}$, so we have:

\[
\mathbb{P}(C'(x) = \text{correct}) \geq \frac{1}{k}.
\]

By the conclusion of Theorem 2.9, since $C'$ behaves as a randomized circuit with a success probability at least $\frac{1}{2}$, we can apply the argument from Theorem 2.9 to construct a deterministic circuit $D$ that approximates the output of $C'$ with high probability.

We now show that this deterministic circuit $D$ will work for the modified circuit $C'$ as follows. Consider a sequence of $m$ independent executions of the randomized circuit $C'$. By the Law of Large Numbers, the proportion of correct answers among these $m$ executions converges to the true probability, which is at least $\frac{1}{k}$ as $m$ increases. This proportion gives us a good approximation of the correct answer with high probability.

Thus, the deterministic circuit $D$ can be constructed to simulate the behavior of $C'$ with success probability close to $1$ by majority voting or using a similar technique, ensuring that $D$ outputs the correct answer with high probability.

Therefore, the conclusion of Theorem 2.9 holds for the modified definition of the randomized circuit where the success probability is at least $\frac{1}{k}$ for $k > 2$.

\end{proof}

\subsubsection{Grok}

\subsubsection{Gemini}

\begin{proof}
Let $L$ be a language and let $\{C_n\}_{n \in \mathbb{N}}$ be a family of randomized circuits. The standard definition of a randomized circuit recognizing $L$ requires that for any input $x$ of length $n$, the probability over the random inputs $r$ satisfies specific bounds. Typically, the first condition states that if $x \notin L$, $\Pr_r[C_n(x, r) = 1] < 1/2$, and the second condition states that if $x \in L$, $\Pr_r[C_n(x, r) = 1] \geq 1/2$. The proof of Theorem 2.9 (Adleman's Theorem, establishing $\mathbf{BPP} \subseteq \mathbf{P}/\text{poly}$) relies fundamentally on the ability to amplify the success probability of the circuit to be strictly greater than $1 - 2^{-n}$ using a polynomial number of repetitions and a majority vote.

Consider the modified second condition where, for $x \in L$, we require only that $\Pr_r[C_n(x, r) = 1] \geq 1/k$ for some fixed $k > 2$. Observe that since $k > 2$, we have $1/k < 1/2$. Let us assume a worst-case scenario allowed by this definition where the acceptance probability for an input $x \in L$ is exactly $p = 1/k$.

The standard proof constructs a new circuit $C'_n$ that runs $C_n$ on the input $x$ with $m$ independent random strings $r_1, \dots, r_m$ and outputs the majority decision. Let $X_i$ be the indicator random variable such that $X_i = 1$ if $C_n(x, r_i) = 1$ and $X_i = 0$ otherwise. The expected number of accepting runs is $\mathbb{E}[\sum_{i=1}^m X_i] = m \cdot p = m/k$.

For the majority vote to output 1 (correctly accepting $x \in L$), we require $\sum_{i=1}^m X_i > m/2$. However, since $k > 2$, the expected value $m/k$ is strictly less than the threshold $m/2$. By the weak law of large numbers (or Chernoff bounds), as $m$ increases, the probability that the empirical mean $\frac{1}{m}\sum X_i$ exceeds $1/2$ approaches 0, given that the true mean is $1/k < 1/2$. Specifically, the probability that the majority vote circuit $C'_n$ accepts $x$ is bounded by $\Pr[\sum X_i > m/2]$, which decays exponentially with $m$.

Consequently, the error probability for $x \in L$ approaches 1 rather than 0. The standard proof requires the error probability to be reduced to strictly less than $2^{-n}$ to apply the union bound over all $2^n$ possible inputs of length $n$. This ensures the existence of a single random string $r^*$ that is correct for all inputs. Under the modified condition, the amplification step fails to reduce the error, and thus the probabilistic method argument used to demonstrate the existence of the required non-uniform deterministic circuit cannot be applied. Therefore, the proof of Theorem 2.9 fails.
\end{proof}

\subsubsection{Claude}

\begin{proof}
Let us recall that Theorem 2.9 typically concerns the error probability analysis of randomized circuits, where the original definition requires that correct outputs occur with probability at least $1/2$ (that is, at least half the time). We now modify this condition so that correct outputs occur with probability at least $1/k$ for some fixed $k > 2$.

Let $C$ be a randomized circuit computing a function $f$, and suppose that for each input $x$, the circuit produces the correct output $f(x)$ with probability at least $1/k$ over the random choices made by the circuit.

We proceed by establishing the modified error probability bounds through repetition and majority voting, which is the standard technique employed in the original proof of Theorem 2.9.

Consider running the circuit $C$ independently $m$ times on the same input $x$, obtaining outputs $y_1, y_2, \ldots, y_m$. We take the majority value among these outputs as our final answer. For the analysis, let $X_i$ be the indicator random variable where $X_i = 1$ if the $i$-th execution produces the correct output $f(x)$, and $X_i = 0$ otherwise. By hypothesis, we have $\mathbb{E}[X_i] = \Pr[X_i = 1] \geq 1/k$ for all $i$.

Let $S_m = \sum_{i=1}^{m} X_i$ denote the total number of correct outputs among the $m$ executions. Then $\mathbb{E}[S_m] \geq m/k$.

The majority voting succeeds when $S_m > m/2$, that is, when more than half of the executions produce the correct output. We seek to bound the probability that this event fails, namely $\Pr[S_m \leq m/2]$.

Since $k > 2$, we have $1/k < 1/2$, which means the expected number of correct outputs $m/k$ is less than $m/2$. To ensure that the expected number of successes exceeds $m/2$, we need to choose $m$ sufficiently large relative to $k$.

More precisely, to apply concentration inequalities effectively, we require that the majority threshold $m/2$ be strictly less than the expected value $\mathbb{E}[S_m]$. However, since $\mathbb{E}[S_m] \geq m/k$ and $k > 2$, we have $\mathbb{E}[S_m] < m/2$. This indicates that simple majority voting is insufficient when the success probability is only $1/k$ for $k > 2$.

Therefore, we modify the decision rule. Instead of taking the majority, we take the plurality or the most frequently occurring output among the $m$ trials. Alternatively, we increase the number of repetitions to $m = \Omega(k \log(1/\delta))$ where $\delta$ is the desired error probability, and use a threshold rule that declares success when at least $t$ of the $m$ executions agree, where $t$ is chosen appropriately.

Specifically, for a fixed target error probability $\delta$, we set $m = c k \log(1/\delta)$ for a sufficiently large constant $c > 0$. By the Chernoff bound, for independent trials with success probability $p = 1/k$, we have
\[
\Pr[S_m < (1 - \epsilon) m p] \leq \exp\left(-\frac{\epsilon^2 m p}{2}\right)
\]
for any $0 < \epsilon < 1$.

Setting $\epsilon = 1/2$ and $m = c k \log(1/\delta)$, we obtain $mp = c \log(1/\delta)$, and thus
\[
\Pr[S_m < mp/2] \leq \exp\left(-\frac{c \log(1/\delta)}{8}\right) = \delta^{c/8}.
\]

By choosing $c \geq 8$, we ensure that $\delta^{c/8} \leq \delta$, which bounds the error probability by $\delta$.

More generally, using amplification via repetition, the error probability after $m = \Theta(k \log(1/\delta))$ independent executions can be made arbitrarily small. The key modification to the original proof of Theorem 2.9 is that the number of repetitions required scales linearly with $k$, whereas in the original case with $k = 2$, the scaling is by a factor of $2$.

Thus, the conclusion of Theorem 2.9 holds under the modified condition, with the error probability bounds replaced by expressions involving $k$. Specifically, if the original theorem guarantees an error probability of at most $2^{-m}$ after $m$ repetitions with success probability $1/2$, the modified version guarantees an error probability of at most $\exp(-\Theta(m/k))$ after $m$ repetitions with success probability $1/k$.

This establishes that the proof technique and structure of Theorem 2.9 remain valid with the modified condition, provided that the parameters and bounds are adjusted to account for the factor $k$ in the success probability.
\end{proof}

\subsection{Problem 12}

\subsubsection{Problem Statement}

\begin{theorem}
$\mathsf{BPP} \subseteq \mathsf{P/poly}$.
\end{theorem}

\subsubsection{GPT}

\begin{proof}
We begin by recalling the definitions of the classes involved in the statement.

\textbf{Definition 1 (BPP):} The class $BPP$ consists of decision problems that can be solved by a probabilistic Turing machine in polynomial time, with a bounded error probability. Formally, a language $L \in BPP$ if there exists a probabilistic Turing machine $M$ such that for all $x \in \{0,1\}^*$:
\[
\Pr[M(x) = 1 \mid x \in L] \geq \frac{2}{3} \quad \text{and} \quad \Pr[M(x) = 1 \mid x \notin L] \leq \frac{1}{3},
\]
where the probability is taken over the random coin flips of the machine.

\textbf{Definition 2 ($P/\text{poly}$):} The class $P/\text{poly}$ consists of languages that can be decided by a deterministic Turing machine in polynomial time with access to a polynomial-size advice string. Formally, a language $L \in P/\text{poly}$ if there exists a family of deterministic Turing machines $\{M_n\}$, one for each input length $n$, and a polynomial-size advice function $a_n$ such that for all $x \in \{0,1\}^n$:
\[
M_n(x, a_n) \text{ decides } L(x),
\]
where $a_n$ is a string of length polynomial in $n$.

Now, we aim to prove that $BPP = P/\text{poly}$.

\textbf{Step 1: Show that $BPP \subseteq P/\text{poly}$.}

Let $L \in BPP$. By definition, there exists a probabilistic Turing machine $M$ that solves $L$ with error probability bounded by $1/3$ in polynomial time. We now construct a deterministic Turing machine that decides $L$ with access to a polynomial-size advice string.

The idea is to simulate the probabilistic Turing machine $M$ deterministically using a polynomial-size advice string that encodes the random bits required by $M$. Since $M$ operates in polynomial time, the random bits needed by $M$ are polynomial in the size of the input, and we can store these bits in the advice string. Specifically, for each input $x$, let the advice string $a_n$ consist of the sequence of random bits that $M$ would use on input $x$. This string $a_n$ has polynomial size in $n$, the length of the input $x$.

Thus, we define a deterministic Turing machine $M'$ that, given input $x$, reads the advice string $a_n$ and simulates $M$ using the predetermined random bits in $a_n$. The machine $M'$ will decide $L$ correctly because the error probability of $M$ is bounded by $1/3$, and we can make the error probability negligible by using the advice string.

Since $M'$ is deterministic, operates in polynomial time, and uses a polynomial-size advice string, we conclude that $L \in P/\text{poly}$. Therefore, $BPP \subseteq P/\text{poly}$.

\textbf{Step 2: Show that $P/\text{poly} \subseteq BPP$.}

Let $L \in P/\text{poly}$. By definition, there exists a family of deterministic Turing machines $\{M_n\}$, one for each input size $n$, and a polynomial-size advice string $a_n$ such that $M_n(x, a_n)$ decides $L(x)$ for all $x \in \{0,1\}^n$. We now show that $L \in BPP$.

We construct a probabilistic Turing machine $M'$ that decides $L$. The machine $M'$ operates as follows: it uses the advice string $a_n$ deterministically, as it does in the definition of $P/\text{poly}$, and performs the computation deterministically. Since the advice string $a_n$ is polynomial in size, and the machine $M'$ runs in polynomial time, we can simulate the deterministic computation with no error probability.

In particular, the correctness of $M'$ follows directly from the fact that $M_n$ decides $L$ deterministically. Thus, we have constructed a probabilistic machine $M'$ that solves $L$ in polynomial time, and since the computation is deterministic, the error probability is zero.

Therefore, $L \in BPP$, and we conclude that $P/\text{poly} \subseteq BPP$.

\textbf{Conclusion:}

We have shown both $BPP \subseteq P/\text{poly}$ and $P/\text{poly} \subseteq BPP$. Therefore, we conclude that
\[
BPP = P/\text{poly}.
\]
\end{proof}

\subsubsection{Grok}

\begin{proof}
Let $L \in \mathsf{BPP}$. Then there exists a probabilistic Turing machine $M$ running in time $t(n) = n^{O(1)}$ such that for every $x \in \{0,1\}^n$,
\[
\Pr_{r \in \{0,1\}^{m(n)}} \bigl[ M(x,r) = 1 \bigr] \ge \frac{2}{3}
\]
if $x \in L$ and
\[
\Pr_{r \in \{0,1\}^{m(n)}} \bigl[ M(x,r) = 1 \bigr] \le \frac{1}{3}
\]
if $x \notin L$, where $m(n) = O(t(n)) = n^{O(1)}$.

For each $n \in \mathbb{N}$, define $p_n(x) = \Pr_{r \sim \mathsf{Unif}(\{0,1\}^{m(n)})} [M(x,r) = 1]$. We claim that there exists a set $S_n \subseteq \{0,1\}^{m(n)}$ with $|S_n| = O(n)$ such that for every $x \in \{0,1\}^n$,
\[
\left| \frac{1}{|S_n|} \sum_{r \in S_n} \mathbf{1}_{M(x,r)=1} - p_n(x) \right| \le \frac{1}{6}.
\]
To see this, let $\varepsilon = 1/6$ and $k = c n$ for a sufficiently large constant $c > 0$ to be determined. Sample $S_n = \{r_1, \dots, r_k\}$ independently and uniformly at random from $\{0,1\}^{m(n)}$. For a fixed $x$, Hoeffding's inequality implies
\[
\Pr \left[ \left| \frac{1}{k} \sum_{i=1}^k \mathbf{1}_{M(x,r_i)=1} - p_n(x) \right| > \varepsilon \right] \le 2 \exp(-2 k \varepsilon^2).
\]
By the union bound over all $2^n$ possible $x$,
\[
\Pr \left[ \exists x : \left| \frac{1}{k} \sum_{i=1}^k \mathbf{1}_{M(x,r_i)=1} - p_n(x) \right| > \varepsilon \right] \le 2^n \cdot 2 \exp(-2 k \varepsilon^2) = 2^{n+1} \exp(-2 k \varepsilon^2).
\]
Choosing $c$ large enough so that $2^{n+1} \exp(-2 c n \varepsilon^2) < 1$ for all $n \ge 1$ ensures that such an $S_n$ exists (by the probabilistic method).

Now, for each $n$, construct a circuit family $\{C_n\}_{n \in \mathbb{N}}$ where $C_n$ takes inputs $x \in \{0,1\}^n$ and has the following structure: Hardwire the strings $S_n$ into $C_n$. For each $r \in S_n$, include a subcircuit $D_{n,r}$ of size $O(t(n)^2) = n^{O(1)}$ that simulates $M(x,r)$ deterministically (possible since $M$ runs in time $t(n)$ and thus admits a size-$O(t(n)^2)$ circuit for fixed $n$ and $r$). Let $b_i = D_{n,r_i}(x)$ for $i=1,\dots,k$. Then $C_n(x) = 1$ if and only if $\sum_{i=1}^k b_i \ge k/2$.

The total size of $C_n$ is $O(k \cdot t(n)^2) = O(n \cdot n^{O(1)}) = n^{O(1)}$, since the majority computation can be implemented with $O(k^2) = O(n^2)$ gates (or fewer with more efficient methods, but polynomial suffices).

Finally, for $x \in L$, we have $p_n(x) \ge 2/3$, so the average over $S_n$ is at least $2/3 - 1/6 = 1/2$, hence $C_n(x) = 1$. For $x \notin L$, $p_n(x) \le 1/3$, so the average is at most $1/3 + 1/6 = 1/2$, hence $C_n(x) = 0$. Thus $\{C_n\}$ decides $L$, so $L \in \mathsf{P/poly}$.
\end{proof}

\subsubsection{Gemini}

\begin{proof}
Let $L$ be a language in $\mathsf{BPP}$. By the definition of $\mathsf{BPP}$, there exists a probabilistic polynomial-time Turing machine $M$ that decides $L$ with bounded error probability. Using standard error amplification techniques (running the machine multiple times and taking the majority vote), we can construct a probabilistic polynomial-time machine $M'$ such that the error probability is exponentially small. Specifically, let $n$ be the input length. There exists a polynomial $m(n)$ representing the length of the random tape used by $M'$, such that for every $x \in \{0,1\}^n$,
\[ \Pr_{r \in \{0,1\}^{m(n)}} [M'(x, r) \neq \chi_L(x)] < 2^{-n}, \]
where $\chi_L(x) = 1$ if $x \in L$ and $0$ otherwise, and the probability is taken over the uniform distribution of random strings $r$.

We aim to show that there exists a single random string $r^*$ that works correctly for all inputs $x$ of length $n$. For a fixed input $x \in \{0,1\}^n$, let $B_x$ denote the set of "bad" random strings that lead to an incorrect output. That is,
\[ B_x = \{ r \in \{0,1\}^{m(n)} \mid M'(x, r) \neq \chi_L(x) \}. \]
From the error bound established above, the size of this set satisfies $|B_x| < 2^{-n} \cdot 2^{m(n)}$.

Now consider the set of random strings that are bad for \emph{at least one} input $x \in \{0,1\}^n$. Let $B = \bigcup_{x \in \{0,1\}^n} B_x$. By the union bound, the size of $B$ is bounded by the sum of the sizes of the individual sets $B_x$:
\[ |B| \le \sum_{x \in \{0,1\}^n} |B_x| < \sum_{x \in \{0,1\}^n} 2^{-n} \cdot 2^{m(n)} = 2^n \cdot (2^{-n} \cdot 2^{m(n)}) = 2^{m(n)}. \]
Since the total number of possible random strings of length $m(n)$ is $2^{m(n)}$ and $|B| < 2^{m(n)}$, the complement set $\{0,1\}^{m(n)} \setminus B$ is non-empty. Therefore, there exists at least one string $r^* \in \{0,1\}^{m(n)}$ such that $r^* \notin B_x$ for any $x \in \{0,1\}^n$. Consequently, for this fixed string $r^*$, $M'(x, r^*) = \chi_L(x)$ for all $x \in \{0,1\}^n$.

We can now define a sequence of circuits $\{C_n\}_{n \in \mathbb{N}}$ to decide $L$. For each $n$, we select a string $r^*$ satisfying the condition above. Since $M'$ is a polynomial-time Turing machine, the computation $M'(x, r^*)$ on inputs of length $n$ with the fixed advice string $r^*$ can be simulated by a boolean circuit $C_n$ of size polynomial in $n$ and $|r^*|$. Since $|r^*| = m(n)$ is a polynomial in $n$, the size of $C_n$ is polynomial in $n$. The circuit $C_n$ has the string $r^*$ hardwired into its structure. Thus, for all $x \in \{0,1\}^n$, $C_n(x) = M'(x, r^*) = \chi_L(x)$. This demonstrates that $L$ is decided by a family of polynomial-size circuits, implying $L \in \mathsf{P/poly}$.
\end{proof}

\subsubsection{Claude}

\begin{proof}
Let $L \in BPP$. We will show that $L \in P/poly$ by constructing a polynomial-size circuit family that decides $L$.

Since $L \in BPP$, there exists a probabilistic polynomial-time Turing machine $M$ and a polynomial $p(n)$ such that for all inputs $x$ of length $n$:
\begin{itemize}
\item If $x \in L$, then $\Pr_{r \in \{0,1\}^{p(n)}}[M(x,r) = 1] \geq \frac{2}{3}$
\item If $x \notin L$, then $\Pr_{r \in \{0,1\}^{p(n)}}[M(x,r) = 1] \leq \frac{1}{3}$
\end{itemize}
where $r$ denotes the random bits used by $M$.

By amplification, we can modify $M$ to reduce the error probability. Specifically, we can run $M$ independently $k = O(n)$ times with independent random strings and take the majority vote. This gives us a new machine $M'$ using $q(n) = k \cdot p(n) = O(n \cdot p(n))$ random bits such that:
\begin{itemize}
\item If $x \in L$, then $\Pr_{r \in \{0,1\}^{q(n)}}[M'(x,r) = 1] \geq 1 - 2^{-2n-1}$
\item If $x \notin L$, then $\Pr_{r \in \{0,1\}^{q(n)}}[M'(x,r) = 1] \leq 2^{-2n-1}$
\end{itemize}

Now, for each input length $n$, the total number of inputs of length $n$ is $2^n$. The total error probability over all inputs of length $n$ is at most $2^n \cdot 2^{-2n-1} = 2^{-n-1} < 1$.

This means there exists a fixed string $r_n \in \{0,1\}^{q(n)}$ such that $M'(x, r_n)$ gives the correct answer for all inputs $x$ of length $n$. To see this, note that if every string $r$ caused an error on at least one input, then by the union bound, the expected number of inputs on which a random $r$ causes an error would be at least $1$. But we have shown that the expected number of errors is strictly less than $1$, so there must exist some $r_n$ that causes no errors.

We now construct the advice string and circuit family. For input length $n$, the advice string is $\alpha_n = r_n$, which has length $|\alpha_n| = q(n) = O(n \cdot p(n))$, a polynomial in $n$.

The circuit $C_n$ on input $x$ of length $n$ and advice $\alpha_n$ simulates $M'(x, \alpha_n)$. Since $M'$ runs in polynomial time $t(n)$ for some polynomial $t$, the computation of $M'(x, \alpha_n)$ can be encoded as a Boolean circuit of size polynomial in $n$ (and in $|\alpha_n|$, which is itself polynomial in $n$). Specifically, $C_n$ has size bounded by some polynomial $s(n)$.

By construction, for all $x$ of length $n$, we have $C_n(x, \alpha_n) = M'(x, r_n)$, which equals $1$ if and only if $x \in L$ (since $r_n$ is the fixed string that works correctly for all inputs of length $n$).

Therefore, the family $\{C_n\}_{n \in \mathbb{N}}$ with advice strings $\{\alpha_n\}_{n \in \mathbb{N}}$ decides $L$, where each $C_n$ has polynomial size and each $\alpha_n$ has polynomial length. This shows that $L \in P/poly$.

Since $L$ was an arbitrary language in $BPP$, we conclude that $BPP \subseteq P/poly$.
\end{proof}
\section{Problems in Chapter 3}\label{app:problems_chapter3}

\subsection{Problem 1}

\subsubsection{Problem Statement}

\begin{lemma}
Let $Z$ be the random variable representing the number of empty bins when $m$ balls are independently and uniformly distributed into $n$ bins.
The expected number of empty bins is given by $ \E[Z] = n (1 - \frac{1}{n})^m $. 
\end{lemma}

\subsubsection{GPT} 

\begin{proof}
For $i \in \{1,\dots,n\}$ define the indicator random variable $X_i$ by
\begin{align*}
X_i =
\begin{cases}
1 & \text{if bin } i \text{ is empty},\\
0 & \text{otherwise}.
\end{cases}
\end{align*}
Then
\begin{align*}
Z = \sum_{i=1}^n X_i.
\end{align*}
By linearity of expectation,
\begin{align*}
\E[Z] = \sum_{i=1}^n \E[X_i].
\end{align*}
For a fixed bin $i$, since each ball falls into bin $i$ with probability $\frac{1}{n}$ independently, the probability that no ball falls into bin $i$ is
\begin{align*}
\Pr(X_i = 1) = \left(1 - \frac{1}{n}\right)^m.
\end{align*}
Hence
\begin{align*}
\E[X_i] = \Pr(X_i = 1) = \left(1 - \frac{1}{n}\right)^m,
\end{align*}
and therefore
\begin{align*}
\E[Z] = \sum_{i=1}^n \E[X_i] = n \left(1 - \frac{1}{n}\right)^m.
\end{align*}
This establishes the claimed formula.
\end{proof}

\subsubsection{Grok} 

\begin{proof}
Define indicator random variables $I_k$ for $k = 1, \dots, n$, where $I_k = 1$ if bin $k$ is empty and $I_k = 0$ otherwise.
Then $Z = I_1 + \dots + I_n$.

By linearity of expectation,
\[
\E[Z] = \E[I_1 + \dots + I_n] = \E[I_1] + \dots + \E[I_n].
\]

For each $k$, $\E[I_k] = \Pr(I_k = 1) = \Pr(\text{bin } k \text{ is empty})$.

The probability that a specific ball does not fall into bin $k$ is $1 - \frac{1}{n}$.
Since the $m$ balls are placed independently,
\[
\Pr(\text{bin } k \text{ is empty}) = \left(1 - \frac{1}{n}\right)^m.
\]

Thus $\E[I_k] = \left(1 - \frac{1}{n}\right)^m$ for each $k = 1, \dots, n$, and
\[
\E[Z] = n \left(1 - \frac{1}{n}\right)^m.
\]
\end{proof}

\subsubsection{Gemini}

\begin{proof}
Let $X_i$ be an indicator random variable for the event that bin $i$ is empty, for $i = 1, \dots, n$.
The definition of $X_i$ is:
$$
X_i =
\begin{cases}
1 & \text{if bin } i \text{ is empty} \\
0 & \text{otherwise}
\end{cases}
$$
The total number of empty bins, $Z$, is the sum of these indicator variables:
$$
Z = \sum_{i=1}^n X_i
$$
By the linearity of expectation, the expected number of empty bins is the sum of the expectations of the individual indicator variables:
$$
\E[Z] = \E\left[\sum_{i=1}^n X_i\right] = \sum_{i=1}^n \E[X_i]
$$
The expectation of an indicator variable is equal to the probability of the event it indicates. Thus, we have:
$$
\E[X_i] = P(X_i = 1) = P(\text{bin } i \text{ is empty})
$$
For bin $i$ to remain empty, each of the $m$ balls must land in one of the other $n-1$ bins.
The probability that a single ball, when thrown uniformly at random, does *not* land in bin $i$ is $\frac{n-1}{n}$, which can be written as $\left(1 - \frac{1}{n}\right)$.
Since all $m$ ball throws are independent events, the probability that all $m$ balls miss bin $i$ is the product of their individual probabilities:
$$
P(X_i = 1) = \underbrace{\left(1 - \frac{1}{n}\right) \times \left(1 - \frac{1}{n}\right) \times \dots \times \left(1 - \frac{1}{n}\right)}_{m \text{ times}}
$$
This simplifies to:
$$
P(X_i = 1) = \left(1 - \frac{1}{n}\right)^m
$$
Therefore, the expectation of $X_i$ is:
$$
\E[X_i] = \left(1 - \frac{1}{n}\right)^m
$$
Substituting this result back into the sum for $\E[Z]$, we find:
$$
\E[Z] = \sum_{i=1}^n \left(1 - \frac{1}{n}\right)^m
$$
Since the term $\left(1 - \frac{1}{n}\right)^m$ is a constant with respect to the summation index $i$, and there are $n$ terms in the sum, the final expectation is:
$$
\E[Z] = n \left(1 - \frac{1}{n}\right)^m
$$
\end{proof}

\subsubsection{Claude} 

\begin{proof}
For each bin $i \in \{1, 2, \ldots, n\}$, define the indicator random variable $Z_i$ by
\[
Z_i = \begin{cases}
1 & \text{if bin } i \text{ is empty} \\
0 & \text{if bin } i \text{ is not empty}
\end{cases}
\]
Then the total number of empty bins is $Z = \sum_{i=1}^{n} Z_i$.

By linearity of expectation, we have
\[
\mathbb{E}[Z] = \mathbb{E}\left[\sum_{i=1}^{n} Z_i\right] = \sum_{i=1}^{n} \mathbb{E}[Z_i]
\]

For any fixed bin $i$, we compute $\mathbb{E}[Z_i] = \mathbb{P}(Z_i = 1) = \mathbb{P}(\text{bin } i \text{ is empty})$.

Since each of the $m$ balls is independently and uniformly distributed into the $n$ bins, the probability that a single ball does not land in bin $i$ is $1 - \frac{1}{n}$.

Therefore, the probability that all $m$ balls miss bin $i$ is
\[
\mathbb{P}(\text{bin } i \text{ is empty}) = \left(1 - \frac{1}{n}\right)^m
\]

By symmetry, this probability is the same for all bins, so
\[
\mathbb{E}[Z_i] = \left(1 - \frac{1}{n}\right)^m \quad \text{for all } i \in \{1, 2, \ldots, n\}
\]

Substituting into our expression for $\mathbb{E}[Z]$, we obtain
\[
\mathbb{E}[Z] = \sum_{i=1}^{n} \left(1 - \frac{1}{n}\right)^m = n\left(1 - \frac{1}{n}\right)^m
\]
\end{proof}

\subsection{Problem 2}

\subsubsection{Problem Statement}

\begin{lemma}
Suppose $m$ balls are thrown into $n$ bins. Give the best bound you can on $m$ to ensure that the probability of there being a bin containing at least two balls is at least $1/2$.
\end{lemma}

\subsubsection{GPT} 

\begin{proof}
Let the $m$ balls be thrown independently and uniformly at random into $n$ bins. The probability that all $m$ balls land in distinct bins is
\begin{align*}
\Pr[\text{no collision}]
= \prod_{i=0}^{m-1}\left(1-\frac{i}{n}\right).
\end{align*}
and, using the inequality $1-x\le e^{-x}$ valid for all real $x$, we obtain
\begin{align*}
\Pr[\text{no collision}]
\le \prod_{i=0}^{m-1}\exp\!\left(-\frac{i}{n}\right)
=\exp\!\left(-\sum_{i=0}^{m-1}\frac{i}{n}\right)
=\exp\!\left(-\frac{m(m-1)}{2n}\right).
\end{align*}
Therefore the probability of at least one collision satisfies
\begin{align*}
\Pr[\text{some bin has at least two balls}]
=1-\Pr[\text{no collision}]
\ge 1-\exp\!\left(-\frac{m(m-1)}{2n}\right).
\end{align*}
and to guarantee this probability is at least $1/2$ it suffices to require
\begin{align*}
1-\exp\!\left(-\frac{m(m-1)}{2n}\right)\ge\frac{1}{2},
\end{align*}
which is equivalent to
\begin{align*}
\exp\!\left(-\frac{m(m-1)}{2n}\right)\le\frac{1}{2}
\quad\Longleftrightarrow\quad
\frac{m(m-1)}{2n}\ge\ln 2.
\end{align*}
Solving the quadratic inequality $m(m-1)\ge 2n\ln 2$ gives
\begin{align*}
m\ge \frac{1+\sqrt{1+8n\ln 2}}{2}.
\end{align*}
Hence any integer $m$ satisfying
\begin{align*}
m \ge \left\lceil\frac{1+\sqrt{1+8n\ln 2}}{2}\right\rceil
\end{align*}
ensures the probability of a bin containing at least two balls is at least $1/2$. In particular, asymptotically it suffices to take $m\ge \sqrt{2n\ln 2}+O(1)$, so the threshold is $\Theta(\sqrt{n})$.
\end{proof}

\subsubsection{Grok} 

\begin{proof}
Assume $m \leq n$. The total number of ways to distribute $m$ distinguishable balls into $n$ distinguishable bins is $n^m$. The number of ways where all balls are in distinct bins is $P(n,m) = n(n-1)\cdots(n-m+1)$. Thus,
\[
\Pr(\text{all balls in distinct bins}) = \frac{P(n,m)}{n^m} = \prod_{k=1}^{m-1} \left(1 - \frac{k}{n}\right).
\]
The probability that there is at least one bin with at least two balls is
\[
\Pr(\text{collision}) = 1 - \prod_{k=1}^{m-1} \left(1 - \frac{k}{n}\right).
\]
For $x_k = k/n \in [0,1)$, $-\ln(1-x_k) \leq x_k + x_k^2$ (since $-\ln(1-x) = x + x^2/2 + x^3/3 + \cdots \leq x + x^2$ for $x \in [0,1)$). Therefore,
\[
-\ln \prod_{k=1}^{m-1} \left(1 - \frac{k}{n}\right) = \sum_{k=1}^{m-1} -\ln\left(1 - \frac{k}{n}\right) \leq \sum_{k=1}^{m-1} \left(\frac{k}{n} + \left(\frac{k}{n}\right)^2\right) = \frac{m(m-1)}{2n} + \frac{m(m-1)(2m-1)}{6n^2}.
\]
Let $s = m(m-1)/2n$. Then
\[
\frac{m(m-1)(2m-1)}{6n^2} = \frac{2m-1}{3n} \cdot s \leq \frac{2m}{3n} s.
\]
Since $m \leq n$, $2m/(3n) \leq 2/3$, so
\[
-\ln \prod_{k=1}^{m-1} \left(1 - \frac{k}{n}\right) \leq s + \frac{2}{3}s = \frac{5}{3}s.
\]
Hence,
\[
\prod_{k=1}^{m-1} \left(1 - \frac{k}{n}\right) \geq e^{-5s/3}, \quad \Pr(\text{collision}) \leq 1 - e^{-5s/3}.
\]
To ensure $\Pr(\text{collision}) \geq 1/2$, require $1 - e^{-5s/3} \geq 1/2$, so $e^{-5s/3} \leq 1/2$, hence $-5s/3 \leq \ln(1/2) = -\ln 2$, so $5s/3 \geq \ln 2$, thus $s \geq 3\ln 2 / 5$. Therefore,
\[
\frac{m(m-1)}{2n} \geq \frac{3\ln 2}{5} \implies m(m-1) \geq \frac{6n\ln 2}{5} \implies m \geq \sqrt{\frac{6n\ln 2}{5} + \frac{1}{4}} + \frac{1}{2}.
\]
Since $m$ is an integer, the smallest such $m$ is $\lceil \sqrt{6n\ln 2 / 5} + 1/2 \rceil$, but for large $n$ the bound is asymptotically tight up to lower-order terms.

For the lower bound on $m$, use the union bound on pairwise collisions. The probability of a collision in a specific pair of balls is $1/n$, and there are $\binom{m}{2}$ pairs, so by union bound,
\[
\Pr(\text{collision}) \leq \binom{m}{2} \cdot \frac{1}{n} = \frac{m(m-1)}{2n}.
\]
To have $\Pr(\text{collision}) \geq 1/2$, if $m(m-1)/(2n) < 1/2$ then certainly $\Pr < 1/2$. Thus, $m$ must satisfy $m(m-1)/(2n) \geq 1/2$, so
\[
m(m-1) \geq n \implies m \geq \frac{1 + \sqrt{1 + 4n}}{2}.
\]
Combining, the smallest $m$ ensuring $\Pr(\text{collision}) \geq 1/2$ satisfies
\[
\lceil (1 + \sqrt{1 + 4n})/2 \rceil \leq m \leq \lceil \sqrt{6n\ln 2 / 5} + 1/2 \rceil,
\]
and for large $n$, $m = \Theta(\sqrt{n})$.
\end{proof}

\subsubsection{Gemini}

\begin{proof}
Let $E$ be the event that at least one bin contains at least two balls. We analyze the complementary event, $E^c$, which is the event that all $m$ balls land in distinct bins. We assume $m \le n$, otherwise $P(E) = 1$ by the pigeonhole principle, and the condition is trivially satisfied.

The total number of possible assignments of $m$ balls to $n$ bins is $n^m$, as each ball can be placed in any of the $n$ bins independently.

The number of assignments where all $m$ balls are in distinct bins is the number of injective functions from a set of size $m$ to a set of size $n$, which is given by the falling factorial:
$$ |E^c| = P(n, m) = n(n-1)(n-2)\dots(n-m+1) = \frac{n!}{(n-m)!} $$
The probability of the complementary event is therefore:
$$ P(E^c) = \frac{|E^c|}{n^m} = \frac{n!}{ (n-m)! n^m } = \frac{n}{n} \cdot \frac{n-1}{n} \cdot \frac{n-2}{n} \cdot \dots \cdot \frac{n-m+1}{n} $$
This can be written as a product:
$$ P(E^c) = \prod_{i=0}^{m-1} \left( \frac{n-i}{n} \right) = \prod_{i=0}^{m-1} \left( 1 - \frac{i}{n} \right) $$
We use the inequality $1 - x \le e^{-x}$ for all $x \in \mathbb{R}$. Applying this to each term in the product:
$$ P(E^c) \le \prod_{i=0}^{m-1} \exp\left(-\frac{i}{n}\right) = \exp\left( -\sum_{i=0}^{m-1} \frac{i}{n} \right) $$
The sum in the exponent is the sum of the first $m-1$ integers:
$$ \sum_{i=0}^{m-1} i = \frac{(m-1)m}{2} $$
Substituting this back, we get an upper bound on $P(E^c)$:
$$ P(E^c) \le \exp\left( -\frac{m(m-1)}{2n} \right) $$
We want to find $m$ such that the probability of a collision is at least $1/2$, i.e., $P(E) \ge 1/2$.
$$ P(E) = 1 - P(E^c) \ge \frac{1}{2} \implies P(E^c) \le \frac{1}{2} $$
To guarantee $P(E^c) \le 1/2$, it is sufficient to find $m$ such that our upper bound is less than or equal to $1/2$:
$$ \exp\left( -\frac{m(m-1)}{2n} \right) \le \frac{1}{2} $$
Taking the natural logarithm of both sides:
$$ -\frac{m(m-1)}{2n} \le \ln\left(\frac{1}{2}\right) = -\ln(2) $$
Multiplying by $-2n$ (and reversing the inequality):
$$ m(m-1) \ge 2n \ln(2) $$
We seek the smallest integer $m$ that satisfies this quadratic inequality, $m^2 - m - 2n \ln(2) \ge 0$. The positive root of the corresponding equation $m^2 - m - 2n \ln(2) = 0$ is:
$$ m_0 = \frac{1 + \sqrt{1 - 4(1)(-2n \ln 2)}}{2} = \frac{1 + \sqrt{1 + 8n \ln 2}}{2} $$
Therefore, the condition $P(E) \ge 1/2$ is ensured if $m$ is an integer such that:
$$ m \ge \frac{1 + \sqrt{1 + 8n \ln 2}}{2} $$
For large $n$, we have $\sqrt{1 + 8n \ln 2} \approx \sqrt{8n \ln 2} = 2\sqrt{2n \ln 2}$, so the bound is $m \approx \frac{1}{2} + \sqrt{2n \ln 2}$.
A more direct approximation from $m^2 \approx m(m-1) \ge 2n \ln(2)$ gives the asymptotic bound $m \gtrsim \sqrt{2n \ln 2}$.
The best bound derived from this method is $m = \left\lceil \frac{1 + \sqrt{1 + 8n \ln 2}}{2} \right\rceil$.
\end{proof}

\subsubsection{Claude} 

\begin{proof}
We seek the minimum value of $m$ such that the probability of at least one collision (a bin containing at least two balls) is at least $1/2$.

Let $A$ denote the event that all balls land in distinct bins. Then the probability of at least one collision is $1 - \Pr[A]$.

When throwing $m$ balls independently and uniformly at random into $n$ bins, we have
\[
\Pr[A] = \frac{n(n-1)(n-2)\cdots(n-m+1)}{n^m} = \frac{n!}{(n-m)! \cdot n^m}
\]
provided $m \leq n$. This can be rewritten as
\[
\Pr[A] = \prod_{i=0}^{m-1} \left(1 - \frac{i}{n}\right).
\]

Using the inequality $1 - x \leq e^{-x}$ for all $x \geq 0$, we obtain
\[
\Pr[A] \leq \prod_{i=0}^{m-1} e^{-i/n} = \exp\left(-\frac{1}{n}\sum_{i=0}^{m-1} i\right) = \exp\left(-\frac{m(m-1)}{2n}\right).
\]

For the probability of a collision to be at least $1/2$, we require
\[
1 - \Pr[A] \geq \frac{1}{2},
\]
which is equivalent to $\Pr[A] \leq 1/2$.

Using our upper bound, a sufficient condition is
\[
\exp\left(-\frac{m(m-1)}{2n}\right) \leq \frac{1}{2}.
\]

Taking the natural logarithm of both sides yields
\[
-\frac{m(m-1)}{2n} \leq -\ln 2,
\]
which simplifies to
\[
m(m-1) \geq 2n \ln 2.
\]

For large $m$, we approximate $m(m-1) \approx m^2$, giving
\[
m^2 \geq 2n \ln 2,
\]
hence
\[
m \geq \sqrt{2n \ln 2} = \sqrt{n \ln 4}.
\]

Since $\ln 4 \approx 1.386$, we have $m \geq \sqrt{1.386n} \approx 1.177\sqrt{n}$.

To obtain the exact threshold, we solve $m(m-1) = 2n \ln 2$ using the quadratic formula:
\[
m = \frac{1 + \sqrt{1 + 8n \ln 2}}{2}.
\]

For integer values, we take
\[
m = \left\lceil \frac{1 + \sqrt{1 + 8n \ln 2}}{2} \right\rceil.
\]

This establishes that $m = \Theta(\sqrt{n})$ balls suffice to ensure a collision probability of at least $1/2$, with the best bound being
\[
m = \left\lceil \sqrt{2n \ln 2} + \frac{1}{2} \right\rceil = \left\lceil 1.177\sqrt{n} + 0.5 \right\rceil.
\]
\end{proof}

\subsection{Problem 3}

\subsubsection{Problem Statement}

\subsubsection{GPT} 

\subsubsection{Grok} 

\subsubsection{Gemini}

\subsubsection{Claude}

\subsection{Problem 4}

\subsubsection{Problem Statement}

\begin{lemma}
Consider the following experiment, which proceeds in a sequence of rounds. For the first round, we have $n$ balls, which are thrown independently and uniformly at random into $n$ bins. After round $i$, for $i \geq 1$, we discard every ball that fell into a bin by itself in round $i$. The remaining balls are retained for round $i+1$, in which they are thrown independently and uniformly at random into the $n$ bins. There is a constant $c$ such that with probability $1 - o(1)$, the number of rounds is at most $c \log \log n$.
\end{lemma}

\subsubsection{GPT}

\begin{proof}
Let $m_t$ denote the number of balls present at the beginning of round $t$ (so $m_0=n$). In a round in which $m$ balls are thrown independently and uniformly into $n$ bins we call a ball a \emph{survivor} if it lands in a bin that contains at least one other ball (equivalently we discard balls that land alone in their bin). Let $S(m)$ denote the number of survivors produced from $m$ balls thrown into $n$ bins. For a fixed ball the probability that no other of the remaining $m-1$ balls lands in its bin equals $(1-1/n)^{m-1}$, hence the probability that this fixed ball is a survivor is
\[
1-(1-1/n)^{m-1}.
\]
Thus
\begin{align*}
\E[S(m)] \;=\; m\Bigl(1-(1-1/n)^{\,m-1}\Bigr).
\end{align*}

We will analyze two regimes and then stitch them together.

\medskip

\noindent\textbf{(A) The sublinear regime: $m\le n$.}
Using the bound $1-e^{-x}\le x$ for $x\ge 0$ together with $(1-1/n)^{m-1}\ge e^{-(m-1)/n}$ (or directly the inequality $1-(1-1/n)^{m-1}\le (m-1)/n$), we obtain for $1\le m\le n$
\begin{align*}
\E[S(m)] \le m\cdot\frac{m-1}{n}\le \frac{m^2}{n}.
\end{align*}
Moreover the indicators ``ball $j$ is a survivor'' are negatively associated (standard property of occupancy indicators), so standard Chernoff-type bounds apply: for every fixed $m\le n$ and every $\delta\in(0,1)$,
\begin{align*}
\Pr\bigl(S(m) \ge (1+\delta)\E[S(m)]\bigr) \le \exp\!\bigl(-\Omega(\delta^2 \E[S(m)])\bigr).
\end{align*}
Consequently, whenever $\E[S(m)] \ge C\log n$ for a suitable large constant $C$, the random variable $S(m)$ satisfies with probability at least $1-n^{-10}$ (say)
\begin{align*}
S(m) \le 2\E[S(m)] \le 2\frac{m^2}{n}.
\end{align*}
If $\E[S(m)]\le C\log n$ then Markov's inequality already gives $S(m)\le (C'\log n)$ with very high probability for a slightly larger constant $C'$.

Thus in every round in which $m_t\le n$ we have, with probability $1-n^{-10}$,
\begin{align*}
m_{t+1} \le \max\Bigl\{2\frac{m_t^2}{n},\; C'\log n\Bigr\}. 
\end{align*}

\medskip

\noindent\textbf{(B) The large regime: $m>n$.}
If $m>n$ then the survival probability for a given ball is $1-(1-1/n)^{m-1}\le 1$, and in fact for $m\ge \alpha n$ with a constant $\alpha>0$ we have
\[
1-(1-1/n)^{m-1} \le 1-e^{-\alpha+o(1)} <1,
\]
so there is a constant factor shrinkage of the number of balls in each such round. Concretely, there exists a constant $\beta\in(0,1)$ (for example $\beta= e^{-1/2}$) and a constant number of rounds $t_0=O(1)$ such that after $t_0$ rounds we will have $m_{t_0}\le n$ with probability $1-o(1)$. The failure probability can be made $o(1)$ by choosing $t_0$ large enough and applying concentration in each of these constant many rounds (Chernoff/negative-association again). From now on we condition on the high-probability event that the process enters the sublinear regime $m\le n$ within $O(1)$ rounds.

\medskip

\noindent\textbf{(C) Doubly-exponential decay and iteration.}
We now study the recurrence
\begin{align*}
m_{t+1} \le \max\Bigl\{2\frac{m_t^2}{n},\; C'\log n\Bigr\}
\end{align*}
which holds with probability at least $1-n^{-10}$ for each round while $m_t\le n$. Let $T$ be the first time $t$ with $m_t\le C'\log n$. As long as $m_t> C'\log n$ we have $m_{t+1}\le 2m_t^2/n$. Define scaled variables $a_t := \frac{m_t}{n}$. The recurrence becomes for $a_t\in(0,1]$,
\begin{align*}
a_{t+1} \le 2 a_t^2 .
\end{align*}
Taking base-$2$ logarithms of the reciprocal, set $b_t := -\log_2 a_t$ (so $a_t = 2^{-b_t}$). The recurrence implies
\begin{align*}
2^{-b_{t+1}} \le 2\cdot 2^{-2b_t}\quad\Longrightarrow\quad b_{t+1} \ge 2 b_t - 1.
\end{align*}
From $b_{t+1} \ge 2 b_t -1$ and $b_0 = -\log_2(m_{t_0}/n)\ge -\log_2 1=0$ it follows by simple induction that for some absolute constants $c_1,c_2>0$,
\begin{align*}
b_t \ge c_1 2^{t} - c_2,
\end{align*}
hence
\begin{align*}
a_t \le 2^{-c_1 2^{t} + c_2} = \exp\!\bigl(-(\ln 2)\,c_1 2^{t} + O(1)\bigr).
\end{align*}
Equivalently,
\begin{align*}
m_t = n a_t \le n\exp\!\bigl(-\Theta(2^{t})\bigr).
\end{align*}
Therefore the number of balls falls from $n$ down to at most $C'\log n$ in
\[
t = O(\log\log n)
\]
rounds (the constant hidden by $O(\cdot)$ depends only on the constants in the recurrence). More precisely, there is a constant $c$ such that after $c\log\log n$ rounds we have $m_t\le C'\log n$.

Finally, once $m_t\le C'\log n$ the process finishes in an additional $O(\log\log n)$ rounds with overwhelming probability: indeed when $m=O(\log n)$ we have $\E[S(m)]\le O((\log n)^2/n)=o(1)$, so after one round with probability $1-o(1)$ there are $o(1)$ survivors and hence the process terminates within a constant number of further rounds with probability $1-o(1)$. Combining the phases and taking a union bound over the (polylogarithmic number of) rounds we obtain that with probability $1-o(1)$ the total number of rounds until termination is at most $C\log\log n$ for some absolute constant $C$.

This completes the proof.
\end{proof}

\subsubsection{Grok} 

Let $X_i$ denote the number of balls retained after round $i$, so $X_0 = n$. A ball is retained after round $i$ if and only if it shares its bin with at least one other ball from the $X_{i-1}$ balls thrown in round $i$. The number of bins with exactly $k$ balls in round $i$ is $\binom{X_{i-1}}{k} \frac{n^{X_{i-1}-k} (n-1)^k}{n^{X_{i-1}}} \leq \frac{X_{i-1}^k}{k! n^{k-1}}$. The expected number of singleton bins is
\[
\mathbb{E}[X_i \mid X_{i-1}] = X_{i-1} \cdot \frac{n-1}{n} \left(1 - \frac{1}{n}\right)^{X_{i-1}-1} \leq X_{i-1} e^{-(X_{i-1}-1)/n}.
\]
Define events $\mathcal{A}_i = \{X_i \leq X_{i-1} e^{-X_{i-1}/(2n)}\}$ and $\mathcal{B}_i = \{X_{i-1} \leq n^{1/2}\}$. For $X_{i-1} > n^{1/2}$, by the Chernoff bound and Markov's inequality,
\[
\Pr(\mathcal{A}_i^c \mid X_{i-1}) \leq \exp\left(-\frac{X_{i-1}^2}{8n}\right) \leq \exp\left(-\frac{n}{8}\right).
\]
Thus, $\Pr(\mathcal{A}_i^c \cap \mathcal{B}_i^c) \leq \exp(-n/8)$. The number of rounds until $X_{i-1} \leq n^{1/2}$ is at most the smallest $t$ such that $n \cdot \exp(-n^{-1} \cdot 2^{-t-1}) \leq n^{1/2}$, so $2^{-t-1} \leq 1/2$, hence $t \leq 1 + \log_2 \log_2 n$. The probability that any $\mathcal{A}_i^c$ occurs before reaching $\mathcal{B}_i$ is at most $(1 + \log_2 \log_2 n) \exp(-n/8) = o(1)$.

Once $X_{i-1} \leq n^{1/2}$, the expected number of retained balls satisfies $\mathbb{E}[X_i \mid X_{i-1}] \leq X_{i-1} (1 - 1/e + o(1))$. By Markov's inequality, $\Pr(X_i \geq 1 \mid X_{i-1}) \leq X_{i-1} (1 - 1/e + o(1))$. Iterating, the probability that the process lasts $k$ more rounds is at most $(1 - 1/e + o(1))^k n^{1/2}$. Choosing $k = \lceil 3e \log \log n \rceil$ yields probability at most $(\log n)^{-3} = o(1)$.

The total number of rounds is at most $(1 + \log_2 \log_2 n) + 3e \log \log n \leq c \log \log n$ for $c = 1 + 3e + 2/\log 2$, with probability $1 - o(1)$.

\subsubsection{Gemini}

\begin{proof}
Let $K_i$ be the random variable for the number of balls at the beginning of round $i$, with $K_1 = n$. Let $k_i = E[K_i]$.
The number of balls retained for round $i+1$, $K_{i+1}$, is the number of balls from round $i$ that were *not* singletons.
The expectation of $K_{i+1}$ conditioned on $K_i = k$ is given by
$$ E[K_{i+1} | K_i = k] = k \cdot P(\text{a given ball is not singleton}) = k \left( 1 - \left(1 - \frac{1}{n}\right)^{k - 1} \right) $$
Let $f(k) = k \left( 1 - \left(1 - \frac{1}{n}\right)^{k - 1} \right)$.
The number of balls $K_i$ can be shown to be strongly concentrated around its expectation $k_i$ (which we define by the recurrence $k_{i+1} = f(k_i)$, $k_1 = n$) with high probability. We analyze the number of rounds by tracking $k_i$.

We divide the process into two phases.

\textbf{Phase 1: $k_i \ge n / \log n$}
In this phase, $k_i \ge n / \log n$. We have
$$ k_{i+1} = f(k_i) \le k_i \left( 1 - \left(1 - \frac{1}{n}\right)^{n - 1} \right) $$
As $n \to \infty$, $\left(1 - \frac{1}{n}\right)^{n - 1} = \left(1 - \frac{1}{n}\right)^n \left(1 - \frac{1}{n}\right)^{-1} \to e^{-1} \cdot 1 = e^{-1}$.
Thus, $k_{i+1} \le k_i (1 - e^{-1} + o(1))$.
Let $\alpha = 1 - e^{-1} < 1$. For large $n$, $k_{i+1} \le \alpha' k_i$ for some constant $\alpha' \in (\alpha, 1)$.
The number of balls decreases by a constant factor in each round.
The number of rounds $T_1$ required to transition from $k_1 = n$ to $k_{T_1} < n / \log n$ satisfies
$$ n (\alpha')^{T_1} \le n / \log n $$
$$ (\alpha')^{T_1} \le 1 / \log n $$
$$ T_1 \log(\alpha') \le -\log \log n $$
$$ T_1 \ge \frac{-\log \log n}{\log(\alpha')} = \frac{\log \log n}{-\log(\alpha')} $$
Since $\alpha'$ is a constant, $T_1 = O(\log \log n)$.

\textbf{Phase 2: $k_i < n / \log n$}
Let $T_1$ be the first round such that $k_{T_1} < n / \log n$.
For $i \ge T_1$, we have $k = k_i \ll n$.
We use the binomial expansion for $(1 - 1/n)^{k-1}$:
$$ \left(1 - \frac{1}{n}\right)^{k-1} = 1 - \frac{k-1}{n} + \binom{k-1}{2} \frac{1}{n^2} - O\left(\frac{k^3}{n^3}\right) $$
Substitute this into $f(k)$:
$$ k_{i+1} = f(k_i) = k_i \left( 1 - \left[ 1 - \frac{k_i-1}{n} + \frac{(k_i-1)(k_i-2)}{2n^2} - \dots \right] \right) $$
$$ k_{i+1} = k_i \left( \frac{k_i-1}{n} - \frac{(k_i-1)(k_i-2)}{2n^2} + O\left(\frac{k_i^3}{n^3}\right) \right) $$
$$ k_{i+1} = \frac{k_i(k_i-1)}{n} \left( 1 - \frac{k_i-2}{2n} + O\left(\frac{k_i^2}{n^2}\right) \right) $$
Since $k_i < n / \log n$, the term $k_i/n$ is $o(1)$.
Thus, $k_{i+1} \le \frac{k_i^2}{n}$ for sufficiently large $n$.
Let $k_i = n / (\log n)^{a_i}$ for $i \ge T_1$.
From $k_{T_1} < n / \log n$, we have $a_{T_1} > 1$.
$$ k_{i+1} \le \frac{1}{n} \left( \frac{n}{(\log n)^{a_i}} \right)^2 = \frac{1}{n} \frac{n^2}{(\log n)^{2a_i}} = \frac{n}{(\log n)^{2a_i}} $$
This implies $a_{i+1} \ge 2 a_i$.
The sequence $a_i$ grows at least as $a_{T_1+j} \ge 2^j a_{T_1} > 2^j$.
The process terminates when $k_i < 1$.
This happens when $n / (\log n)^{a_i} < 1$, or $n < (\log n)^{a_i}$.
Taking logarithms, we need $\log n < a_i \log \log n$, or $a_i > \frac{\log n}{\log \log n}$.
Let $T_2$ be the number of rounds in Phase 2. We need $a_{T_1+T_2} > \frac{\log n}{\log \log n}$.
We know $a_{T_1+T_2} \ge 2^{T_2}$.
We need $2^{T_2} > \frac{\log n}{\log \log n}$.
Taking $\log_2$ of both sides:
$$ T_2 > \log_2 \left( \frac{\log n}{\log \log n} \right) = \log_2 \log n - \log_2 \log \log n $$
Thus, $T_2 = O(\log \log n)$.
The total number of rounds is $T = T_1 + T_2 = O(\log \log n) + O(\log \log n) = O(\log \log n)$.
By applying standard concentration inequalities (e.g., Azuma-Hoeffding) at each step, it can be shown that $K_i$ tracks $k_i$ with high probability, so the total number of rounds is at most $c \log \log n$ for
some constant $c$ with probability $1 - o(1)$.
\end{proof}

\subsubsection{Claude}

\begin{proof}
Let $X_i$ denote the number of balls remaining after round $i$, with $X_0 = n$. We first establish a recurrence relation for the expected number of balls.

Consider round $i$ with $X_i = m$ balls. A ball survives round $i$ if and only if at least one other ball lands in the same bin. The probability that a specific ball lands in a particular bin is $\frac{1}{n}$. Given that a specific ball lands in some bin, the probability that at least one of the remaining $m-1$ balls also lands in that bin is $1 - (1 - \frac{1}{n})^{m-1}$.

Therefore, the probability that a ball survives round $i$ is $1 - (1 - \frac{1}{n})^{m-1}$, and we have
\[
\mathbb{E}[X_{i+1} \mid X_i = m] = m \cdot \left(1 - \left(1 - \frac{1}{n}\right)^{m-1}\right).
\]

For $m \ll n$, using the approximation $(1 - \frac{1}{n})^{m-1} \approx e^{-m/n}$, we obtain
\[
\mathbb{E}[X_{i+1} \mid X_i = m] \approx m(1 - e^{-m/n}).
\]

When $m = \Theta(n)$, we have $e^{-m/n} = \Theta(1)$, so $\mathbb{E}[X_{i+1} \mid X_i] = \Theta(n)$.

When $m = o(n)$, we have $e^{-m/n} \approx 1 - \frac{m}{n}$, giving
\[
\mathbb{E}[X_{i+1} \mid X_i = m] \approx m \cdot \frac{m}{n} = \frac{m^2}{n}.
\]

Let $t_0$ be the first round such that $X_{t_0} \leq \frac{n}{\log n}$. We claim that $t_0 = O(1)$ with high probability. Indeed, by Chernoff bounds, after the first round, with probability $1 - o(1)$, we have $X_1 = \Theta(n)$. Since $\mathbb{E}[X_{i+1} \mid X_i] = \Theta(X_i)$ when $X_i = \Theta(n)$, the sequence remains $\Theta(n)$ but decreases by a constant factor in expectation until it drops below the threshold, which takes $O(1)$ rounds.

For rounds $i \geq t_0$, assume $X_i \leq \frac{n}{\log n}$. Then
\[
\mathbb{E}[X_{i+1} \mid X_i] \leq \frac{X_i^2}{n}.
\]

Taking logarithms and applying Jensen's inequality, we have for $X_i \geq 2$,
\[
\mathbb{E}[\log X_{i+1} \mid X_i] \leq \log \mathbb{E}[X_{i+1} \mid X_i] \leq \log\left(\frac{X_i^2}{n}\right) = 2\log X_i - \log n.
\]

Let $Y_i = \log X_i$. Then $Y_{i+1} \leq 2Y_i - \log n + Z_i$ where $Z_i$ accounts for concentration deviations. By standard concentration arguments, $|Z_i| = O(\sqrt{\log n})$ with high probability when $X_i = \omega(\log n)$.

Solving the recurrence $Y_{i+1} \approx 2Y_i - \log n$, we obtain
\[
Y_{i+k} \approx 2^k Y_i - \log n(2^{k-1} + 2^{k-2} + \cdots + 1) = 2^k Y_i - \log n(2^k - 1).
\]

Starting from $Y_{t_0} = \log X_{t_0} \leq \log n - \log \log n$, the process reaches $Y_i \leq \log \log n$ when
\[
2^k(\log n - \log \log n) - \log n(2^k - 1) \leq \log \log n,
\]
which simplifies to $2^k \leq \frac{\log n + \log \log n}{\log \log n} = O\left(\frac{\log n}{\log \log n}\right)$.

Thus $k = O(\log \log n - \log \log \log n) = O(\log \log n)$.

Once $X_i = O(\log n)$, the probability that all balls land in distinct bins in the next round is at least a constant, so the process terminates in $O(1)$ additional rounds.

Combining the phases, the total number of rounds is at most $t_0 + O(\log \log n) + O(1) = O(\log \log n)$. With probability $1 - o(1)$, the number of rounds is at most $c \log \log n$ for a suitable constant $c$.
\end{proof}

\subsection{Problem 5}

\subsubsection{Problem Statement}

\begin{lemma}
Let $X$ be a random variable with expectation $\mu_X$ and standard deviation $\sigma_X$. For any $t \in \mathbb{R}^+$, we have the following bounds:
\begin{align*}
\Pr[X - \mu_X \ge t\sigma_X] &\le \frac{1}{1+t^2},
\end{align*}
and
\begin{align*}
\Pr[|X - \mu_X| \ge t\sigma_X] &\le \frac{2}{1+t^2}.
\end{align*}
\end{lemma}

\subsubsection{GPT}

\begin{proof}
Let $X$ be a random variable with expectation $\mu_X$ and standard deviation $\sigma_X$. Define the standardized variable
\begin{align*}
Z = \frac{X - \mu_X}{\sigma_X}.
\end{align*}
Then $\E[Z] = 0$ and $\E[Z^2] = 1$. By Markov's inequality, for any nonnegative random variable $Y$ and $a > 0$, we have $\Pr[Y \ge a] \le \frac{\E[Y]}{a}$. Applying this to $Y = Z^2$ and $a = t^2$, we obtain
\begin{align*}
\Pr[Z^2 \ge t^2] \le \frac{\E[Z^2]}{t^2} = \frac{1}{t^2}.
\end{align*}
Since $\{Z \ge t\} \subseteq \{Z^2 \ge t^2\}$, it follows that
\begin{align*}
\Pr[Z \ge t] \le \Pr[Z^2 \ge t^2] \le \frac{1}{t^2}.
\end{align*}
However, a sharper bound can be obtained by applying Cantelli's inequality, which states that for any random variable $Z$ with mean $0$ and variance $1$,
\begin{align*}
\Pr[Z \ge t] \le \frac{1}{1 + t^2}.
\end{align*}
Therefore,
\begin{align*}
\Pr[X - \mu_X \ge t\sigma_X] = \Pr[Z \ge t] \le \frac{1}{1 + t^2}.
\end{align*}
Similarly, since $\Pr[|Z| \ge t] = \Pr[Z \ge t] + \Pr[Z \le -t]$ and the same bound holds for $\Pr[Z \le -t]$, we obtain
\begin{align*}
\Pr[|Z| \ge t] \le \frac{2}{1 + t^2}.
\end{align*}
Hence,
\begin{align*}
\Pr[|X - \mu_X| \ge t\sigma_X] \le \frac{2}{1 + t^2}.
\end{align*}
This completes the proof.
\end{proof}

\subsubsection{Grok} 

\begin{proof}
Let $Y = X - \mu_X$. Then $\mathbb{E}[Y] = 0$ and $\mathrm{Var}(Y) = \sigma_X^2$. Since $t > 0$, consider the event $\{Y \ge t\sigma_X\}$. Define $Z = \frac{Y + t\sigma_X}{2t\sigma_X}$. Then $Z \ge 0$ almost surely, and on the event $\{Y \ge t\sigma_X\}$, we have $Z \ge 1$. Moreover,
\[
Z^2 = \frac{(Y + t\sigma_X)^2}{4t^2\sigma_X^2} = \frac{Y^2 + 2t\sigma_X Y + t^2\sigma_X^2}{4t^2\sigma_X^2} = \frac{Y^2}{4t^2\sigma_X^2} + \frac{1}{2t} \cdot \frac{Y}{\sigma_X} + \frac{1}{4}.
\]
Taking expectations,
\[
\mathbb{E}[Z^2] = \frac{\mathbb{E}[Y^2]}{4t^2\sigma_X^2} + \frac{1}{2t\sigma_X} \mathbb{E}[Y] + \frac{1}{4} = \frac{\sigma_X^2}{4t^2\sigma_X^2} + 0 + \frac{1}{4} = \frac{1}{4t^2} + \frac{1}{4} = \frac{1+t^2}{4t^2}.
\]
On the other hand, since $Z \ge 0$ and $Z \ge 1$ on $\{Y \ge t\sigma_X\}$,
\[
\mathbb{E}[Z^2] \ge \mathbb{E}[Z^2 \cdot \mathbf{1}_{\{Y \ge t\sigma_X\}}] \ge \mathbb{E}[\mathbf{1}_{\{Y \ge t\sigma_X\}}] = \Pr[Y \ge t\sigma_X].
\]
Thus,
\[
\Pr[Y \ge t\sigma_X] \le \frac{1+t^2}{4t^2}.
\]
To improve the bound, note that $Z^2 = \frac{1}{1+t^2} \left( \frac{Y^2}{\sigma_X^2} + t^2 \right)$ if and only if
\[
\frac{(Y + t\sigma_X)^2}{4t^2\sigma_X^2} = \frac{1}{1+t^2} \left( \frac{Y^2}{\sigma_X^2} + t^2 \right),
\]
which simplifies to $(1+t^2)(Y + t\sigma_X)^2 = 4t^2 (Y^2 + t^2 \sigma_X^2)$. Expanding both sides:
\[
(1+t^2)(Y^2 + 2t\sigma_X Y + t^2\sigma_X^2) = Y^2 + 2t\sigma_X Y + t^2\sigma_X^2 + t^2 Y^2 + 2t^3 \sigma_X Y + t^4 \sigma_X^2,
\]
\[
4t^2 Y^2 + 4t^4 \sigma_X^2.
\]
Equating coefficients, for $Y^2$: $1+t^2 = 4t^2$ implies $t^2 = 1$, a contradiction unless $t=1$. Thus, the equality does not hold in general, but we seek the maximum of the ratio $\frac{Z^2}{\frac{Y^2}{\sigma_X^2} + t^2}$.

Let $u = \frac{Y}{\sigma_X}$, so $\mathbb{E}[u] = 0$, $\mathbb{E}[u^2] = 1$. Then
\[
\frac{Z^2}{\frac{Y^2}{\sigma_X^2} + t^2} = \frac{(u + t)^2 / (4t^2)}{u^2 + t^2}.
\]
Define $f(u) = \frac{(u + t)^2}{4t^2 (u^2 + t^2)}$. We find $\sup_{u \in \mathbb{R}} f(u)$. Compute $f'(u) = 0$:
\[
f'(u) = \frac{2(u+t)(4t^2(u^2 + t^2)) - (u+t)^2 \cdot 4t^2 \cdot 2u}{[4t^2 (u^2 + t^2)]^2} \cdot \frac{1}{4t^2} = 0
\]
implies numerator zero: $2(u+t)(u^2 + t^2) - 4u (u+t)^2 = 0$, so $(u+t)[2(u^2 + t^2) - 4u(u+t)] = 0$. If $u \neq -t$,
\[
2u^2 + 2t^2 - 4u^2 - 4ut = 0 \implies -2u^2 - 4ut + 2t^2 = 0 \implies u^2 + 2ut - t^2 = 0.
\]
Solutions: $u = -t \pm \sqrt{2} t$. At $u = -t + \sqrt{2} t$,
\[
f(u) = \frac{(-t + \sqrt{2} t + t)^2}{4t^2 ((-t + \sqrt{2} t)^2 + t^2)} = \frac{(\sqrt{2} t)^2}{4t^2 (t^2 - 2\sqrt{2} t^2 + 2t^2)} = \frac{2}{4(t^2 - 2\sqrt{2} t^2 + 2t^2)} = \frac{2}{4(3 - 2\sqrt{2})t^2}.
\]
Since $3 - 2\sqrt{2} \approx 0.171$, but we evaluate exactly later. Actually, direct computation at critical points and boundaries shows the maximum is $\frac{1}{1+t^2}$ achieved in the limit or by specific construction, but to derive rigorously, consider the Cantelli form.

A standard approach uses the function $\phi(y) = (1 + t y)^2 / (4t^2)$ for $y \ge -1/t$. Then $\phi(y) (y^2 + t^2) = Z^2$ when $y = Y/(\sigma_X t)$. But the tightest bound comes from Markov on $Z^2$:
\[
\Pr[Y \ge t\sigma_X] = \Pr[Z \ge 1] \le \mathbb{E}[Z^2] = \frac{1+t^2}{4t^2},
\]
but this is not sharp. To get the desired bound, consider the random variable $W = Y + \lambda$ with $\lambda > 0$ chosen to optimize. By Markov,
\[
\Pr[Y \ge t\sigma_X] = \Pr[Y + \lambda \ge t\sigma_X + \lambda] \le \frac{\mathbb{E}[(Y + \lambda)^2]}{(t\sigma_X + \lambda)^2} = \frac{\sigma_X^2 + \lambda^2}{(t\sigma_X + \lambda)^2}.
\]
Let $\lambda = s \sigma_X$, $s > 0$. Then
\[
\Pr[Y \ge t\sigma_X] \le \frac{1 + s^2}{(t + s)^2}.
\]
Minimize $g(s) = \frac{1 + s^2}{(t + s)^2}$. Take derivative:
\[
g'(s) = \frac{2s (t+s)^2 - (1+s^2) \cdot 2(t+s)}{(t+s)^4} = \frac{2(t+s)[s(t+s) - (1+s^2)]}{(t+s)^4} = \frac{2[s t + s^2 - 1 - s^2]}{(t+s)^3} = \frac{2(st - 1)}{(t+s)^3}.
\]
Set $g'(s)=0$: $st - 1 = 0 \implies s = 1/t$. Since $t>0$, $s>0$. Second derivative test or check: for $s < 1/t$, $g'<0$; for $s>1/t$, $g'>0$, so minimum at $s=1/t$. Thus,
\[
\min g(s) = \frac{1 + (1/t)^2}{(t + 1/t)^2} = \frac{1 + 1/t^2}{(t^2 + 1)/t}^2 = \frac{(t^2 + 1)/t^2}{(t^2 + 1)^2 / t^2} = \frac{t^2 + 1}{t^2 + 1}^2 = \frac{1}{1+t^2}.
\]
Hence,
\[
\Pr[X - \mu_X \ge t\sigma_X] \le \frac{1}{1+t^2}.
\]
For the two-sided bound, apply the one-sided to $X$ and $-X$ (note $\mu_{-X} = -\mu_X$, $\sigma_{-X} = \sigma_X$):
\[
\Pr[X - \mu_X \le -t\sigma_X] = \Pr[-X - (-\mu_X) \ge t\sigma_X] \le \frac{1}{1+t^2}.
\]
Since the events $\{X - \mu_X \ge t\sigma_X\}$ and $\{X - \mu_X \le -t\sigma_X\}$ are disjoint,
\[
\Pr[|X - \mu_X| \ge t\sigma_X] \le \frac{2}{1+t^2}.
\]
\end{proof}

\subsubsection{Gemini}

\begin{proof}
Let $Y = X - \mu_X$. By definition, $E[Y] = E[X - \mu_X] = \mu_X - \mu_X = 0$ and $E[Y^2] = \text{Var}(X) = \sigma_X^2$.
Let $k = t\sigma_X$. Since $t \in \mathbb{R}^+$, we have $k \ge 0$. We first seek to prove $\Pr[Y \ge k] \le \frac{1}{1+t^2}$.

For any $u \ge 0$, we can observe that the event $\{Y \ge k\}$ implies $Y+u \ge k+u$, and thus $(Y+u)^2 \ge (k+u)^2$.
We can therefore write:
$$
\Pr[Y \ge k] = \Pr[Y+u \ge k+u] \le \Pr[(Y+u)^2 \ge (k+u)^2]
$$
The random variable $(Y+u)^2$ is non-negative. By Markov's inequality, for a non-negative random variable $Z$ and $a > 0$, $\Pr[Z \ge a] \le E[Z]/a$.
Applying this with $Z = (Y+u)^2$ and $a = (k+u)^2$ (which is strictly positive if $k > 0$ or $u > 0$), we get:
$$
\Pr[Y \ge k] \le \frac{E[(Y+u)^2]}{(k+u)^2}
$$
We expand the expectation term in the numerator:
$$
E[(Y+u)^2] = E[Y^2 + 2uY + u^2] = E[Y^2] + 2uE[Y] + u^2
$$
Substituting $E[Y]=0$ and $E[Y^2]=\sigma_X^2$, we have:
$$
E[(Y+u)^2] = \sigma_X^2 + 2u(0) + u^2 = \sigma_X^2 + u^2
$$
This gives the bound, which holds for any $u \ge 0$:
$$
\Pr[Y \ge k] \le \frac{\sigma_X^2 + u^2}{(k+u)^2}
$$
To obtain the tightest bound, we minimize the right-hand side with respect to $u$. Let $f(u) = \frac{\sigma_X^2 + u^2}{(k+u)^2}$. We find the minimum by setting the derivative $f'(u)$ to zero.
$$
f'(u) = \frac{(2u)(k+u)^2 - 2(k+u)(\sigma_X^2 + u^2)}{(k+u)^4} = \frac{2u(k+u) - 2(\sigma_X^2 + u^2)}{(k+u)^3}
$$
Setting the numerator to zero (assuming $k > 0$):
$$
u(k+u) - (\sigma_X^2 + u^2) = 0 \implies uk + u^2 - \sigma_X^2 - u^2 = 0 \implies uk = \sigma_X^2
$$
This yields the optimal value $u = \frac{\sigma_X^2}{k}$. Since $\sigma_X^2 \ge 0$ and $k > 0$, $u \ge 0$ as required.
Substituting this optimal $u$ back into our bound:
\begin{align*}
\Pr[Y \ge k] &\le \frac{\sigma_X^2 + (\sigma_X^2/k)^2}{(k + \sigma_X^2/k)^2}
= \frac{\sigma_X^2 (1 + \sigma_X^2/k^2)}{((k^2 + \sigma_X^2)/k)^2} \\
&= \frac{\sigma_X^2 (k^2 + \sigma_X^2)/k^2}{(k^2 + \sigma_X^2)^2/k^2}
= \frac{\sigma_X^2(k^2 + \sigma_X^2)}{(k^2 + \sigma_X^2)^2}
= \frac{\sigma_X^2}{k^2 + \sigma_X^2}
\end{align*}
Now, we replace $k$ with $t\sigma_X$:
$$
\Pr[X - \mu_X \ge t\sigma_X] = \Pr[Y \ge t\sigma_X] \le \frac{\sigma_X^2}{(t\sigma_X)^2 + \sigma_X^2} = \frac{\sigma_X^2}{t^2\sigma_X^2 + \sigma_X^2} = \frac{\sigma_X^2}{\sigma_X^2(t^2 + 1)} = \frac{1}{1+t^2}
$$
This proves the first inequality (known as Cantelli's inequality).

To prove the second inequality, we consider $\Pr[|Y| \ge k]$.
$$
\Pr[|Y| \ge k] = \Pr[Y \ge k \text{ or } Y \le -k]
$$
By the union bound (Boole's inequality):
$$
\Pr[|Y| \ge k] \le \Pr[Y \ge k] + \Pr[Y \le -k]
$$
We have already bounded the first term. For the second term, $\Pr[Y \le -k] = \Pr[-Y \ge k]$.
Let $Z = -Y$. The random variable $Z$ has mean $E[Z] = E[-Y] = -E[Y] = 0$ and variance $\text{Var}(Z) = \text{Var}(-Y) = (-1)^2 \text{Var}(Y) = \sigma_X^2$.
Since $Z$ has the same mean and variance as $Y$, we can apply the one-sided bound we just derived to $Z$:
$$
\Pr[Y \le -k] = \Pr[Z \ge k] \le \frac{\sigma_X^2}{k^2 + \sigma_X^2}
$$
Substituting $k = t\sigma_X$, this becomes:
$$
\Pr[Y \le -t\sigma_X] \le \frac{\sigma_X^2}{(t\sigma_X)^2 + \sigma_X^2} = \frac{1}{1+t^2}
$$
Combining the two bounds:
$$
\Pr[|X - \mu_X| \ge t\sigma_X] \le \Pr[Y \ge t\sigma_X] + \Pr[Y \le -t\sigma_X] \le \frac{1}{1+t^2} + \frac{1}{1+t^2} = \frac{2}{1+t^2}
$$
This completes the proof of the second inequality.
\end{proof}

\subsubsection{Claude}

\begin{proof}
We first establish the one-sided bound. Let $t \in \mathbb{R}^+$ be arbitrary. For any $\lambda > 0$, by the exponential Markov inequality, we have
\begin{align*}
\Pr[X - \mu_X \ge t\sigma_X] = \Pr[e^{\lambda(X - \mu_X)} \ge e^{\lambda t\sigma_X}] \le \frac{\mathbb{E}[e^{\lambda(X - \mu_X)}]}{e^{\lambda t\sigma_X}}.
\end{align*}
Alternatively, we apply Cantelli's inequality (the one-sided Chebyshev inequality). For any $a > 0$, we have
\begin{align*}
\Pr[X - \mu_X \ge a] \le \frac{\sigma_X^2}{\sigma_X^2 + a^2}.
\end{align*}
To prove Cantelli's inequality, let $Y = X - \mu_X$. Then $\mathbb{E}[Y] = 0$ and $\text{Var}(Y) = \sigma_X^2$. For any $a > 0$, consider the event $\{Y \ge a\}$. For any $s > 0$, we have
\begin{align*}
\Pr[Y \ge a] = \Pr[Y + s \ge a + s] = \Pr[(Y + s)^2 \ge (a + s)^2] \le \frac{\mathbb{E}[(Y + s)^2]}{(a + s)^2},
\end{align*}
where the inequality follows from Markov's inequality. Computing the numerator, we obtain
\begin{align*}
\mathbb{E}[(Y + s)^2] = \mathbb{E}[Y^2] + 2s\mathbb{E}[Y] + s^2 = \sigma_X^2 + s^2,
\end{align*}
since $\mathbb{E}[Y] = 0$. Therefore,
\begin{align*}
\Pr[Y \ge a] \le \frac{\sigma_X^2 + s^2}{(a + s)^2}.
\end{align*}
To minimize the right-hand side, we differentiate with respect to $s$ and set the derivative equal to zero. Let $f(s) = \frac{\sigma_X^2 + s^2}{(a + s)^2}$. Then
\begin{align*}
f'(s) = \frac{2s(a + s)^2 - 2(a + s)(\sigma_X^2 + s^2)}{(a + s)^4} = \frac{2(a + s)[s(a + s) - (\sigma_X^2 + s^2)]}{(a + s)^4} = \frac{2(sa - \sigma_X^2)}{(a + s)^3}.
\end{align*}
Setting $f'(s) = 0$ yields $s = \frac{\sigma_X^2}{a}$. Substituting this value, we obtain
\begin{align*}
\Pr[Y \ge a] \le \frac{\sigma_X^2 + \frac{\sigma_X^4}{a^2}}{(a + \frac{\sigma_X^2}{a})^2} = \frac{\frac{\sigma_X^2 a^2 + \sigma_X^4}{a^2}}{\frac{(a^2 + \sigma_X^2)^2}{a^2}} = \frac{\sigma_X^2(a^2 + \sigma_X^2)}{(a^2 + \sigma_X^2)^2} = \frac{\sigma_X^2}{a^2 + \sigma_X^2}.
\end{align*}
Setting $a = t\sigma_X$ in Cantelli's inequality, we obtain
\begin{align*}
\Pr[X - \mu_X \ge t\sigma_X] \le \frac{\sigma_X^2}{\sigma_X^2 + t^2\sigma_X^2} = \frac{\sigma_X^2}{\sigma_X^2(1 + t^2)} = \frac{1}{1 + t^2}.
\end{align*}
This establishes the first bound.

For the two-sided bound, we observe that
\begin{align*}
\Pr[|X - \mu_X| \ge t\sigma_X] = \Pr[X - \mu_X \ge t\sigma_X] + \Pr[X - \mu_X \le -t\sigma_X].
\end{align*}
By symmetry considerations and applying Cantelli's inequality to $-X$, we have
\begin{align*}
\Pr[X - \mu_X \le -t\sigma_X] = \Pr[\mu_X - X \ge t\sigma_X] \le \frac{\sigma_X^2}{\sigma_X^2 + t^2\sigma_X^2} = \frac{1}{1 + t^2}.
\end{align*}
Therefore,
\begin{align*}
\Pr[|X - \mu_X| \ge t\sigma_X] \le \frac{1}{1 + t^2} + \frac{1}{1 + t^2} = \frac{2}{1 + t^2}.
\end{align*}
This completes the proof.
\end{proof}

\subsection{Problem 6}

\subsubsection{Problem Statement}

\begin{lemma}
Let $Y$ be a non-negative integer-valued random variable with positive expectation. Then the following inequalities hold:
\begin{align*}
\Pr[Y = 0] \leq \frac{\E[Y^2] - \E[Y]^2}{\E[Y]^2},
\end{align*}
and
\begin{align*}
\frac{\E[Y]^2}{\E[Y^2]} \leq \Pr[Y \neq 0] \leq \E[Y].
\end{align*}
\end{lemma}

\subsubsection{GPT} 

\begin{proof}
Let $p = \Pr[Y \neq 0]$. Then we can write $\E[Y] = p \E[Y \mid Y \neq 0]$ and $\E[Y^2] = p \E[Y^2 \mid Y \neq 0]$. Since $Y$ is non-negative, it follows that $\E[Y^2 \mid Y \neq 0] \geq \E[Y \mid Y \neq 0]^2$. Therefore,
\begin{align*}
\E[Y^2] \geq p \E[Y \mid Y \neq 0]^2 = \frac{\E[Y]^2}{p}.
\end{align*}
Rearranging yields
\begin{align*}
p \geq \frac{\E[Y]^2}{\E[Y^2]}.
\end{align*}
Hence,
\begin{align*}
\Pr[Y \neq 0] \geq \frac{\E[Y]^2}{\E[Y^2]}.
\end{align*}

To obtain the upper bound, note that $Y \geq \mathbf{1}_{\{Y \neq 0\}}$, implying $\E[Y] \geq \Pr[Y \neq 0]$. Thus,
\begin{align*}
\Pr[Y \neq 0] \leq \E[Y].
\end{align*}

Combining these inequalities gives
\begin{align*}
\frac{\E[Y]^2}{\E[Y^2]} \leq \Pr[Y \neq 0] \leq \E[Y].
\end{align*}

To establish the bound on $\Pr[Y = 0]$, note that $\Pr[Y = 0] = 1 - \Pr[Y \neq 0]$. Using $\E[Y^2] \geq \frac{\E[Y]^2}{\Pr[Y \neq 0]}$ from above, we get
\begin{align*}
\Pr[Y \neq 0] \geq \frac{\E[Y]^2}{\E[Y^2]}.
\end{align*}
Hence,
\begin{align*}
\Pr[Y = 0] = 1 - \Pr[Y \neq 0] \leq 1 - \frac{\E[Y]^2}{\E[Y^2]} = \frac{\E[Y^2] - \E[Y]^2}{\E[Y^2]}.
\end{align*}
Since $\E[Y^2] \geq \E[Y]^2$, we can further note that
\begin{align*}
\frac{\E[Y^2] - \E[Y]^2}{\E[Y^2]} \leq \frac{\E[Y^2] - \E[Y]^2}{\E[Y]^2}.
\end{align*}
Therefore,
\begin{align*}
\Pr[Y = 0] \leq \frac{\E[Y^2] - \E[Y]^2}{\E[Y]^2}.
\end{align*}
This completes the proof.
\end{proof}

\subsubsection{Grok} 

\begin{proof}
Let $\mu = \E[Y]$ and $\sigma^2 = \E[Y^2] - \mu^2$. Since $Y$ is non-negative integer-valued, $\Pr[Y=0] = 1 - \Pr[Y \geq 1]$. Moreover, $\Pr[Y \geq 1] \leq \E[Y]/1 = \mu$ by Markov's inequality applied to the event $\{Y \geq 1\}$. Thus, $\Pr[Y \neq 0] \leq \mu$, which is the upper bound in the second inequality.

For the upper bound on $\Pr[Y=0]$, note that
\[
\E[Y^2] = \E[Y^2 \mid Y=0]\Pr[Y=0] + \E[Y^2 \mid Y \geq 1]\Pr[Y \geq 1] = \E[Y^2 \mid Y \geq 1]\Pr[Y \geq 1].
\]
Let $Z = Y \mid Y \geq 1$, so $\E[Z] = \mu / \Pr[Y \geq 1]$. By the Cauchy--Schwarz inequality,
\[
\mu = \E[Y \cdot 1] \leq \sqrt{\E[Y^2]} \sqrt{\E[1^2]} = \sqrt{\E[Y^2]},
\]
so $\mu^2 \leq \E[Y^2]$, or $\sigma^2 \geq 0$. Now,
\[
\E[Y^2 \mid Y \geq 1] = \frac{\E[Y^2]}{\Pr[Y \geq 1]} \geq \left(\frac{\E[Y \mid Y \geq 1]}{1}\right)^2 = \left(\frac{\mu}{\Pr[Y \geq 1]}\right)^2,
\]
where the inequality follows from Jensen's inequality applied to the convex function $x \mapsto x^2$ on $\{Y \geq 1\}$. Thus,
\[
\E[Y^2] \geq \frac{\mu^2}{\Pr[Y \geq 1]} \implies \Pr[Y \geq 1] \geq \frac{\mu^2}{\E[Y^2]} \implies \Pr[Y \neq 0] \geq \frac{\mu^2}{\E[Y^2]},
\]
which is the lower bound in the second inequality.

To prove the first inequality, consider the variance:
\[
\sigma^2 = \E[(Y - \mu)^2] = \E[Y^2] - \mu^2.
\]
Condition on $\{Y=0\}$ and $\{Y \geq 1\}$:
\[
\sigma^2 = \E[(Y - \mu)^2 \mid Y=0]\Pr[Y=0] + \E[(Y - \mu)^2 \mid Y \geq 1]\Pr[Y \geq 1].
\]
On $\{Y=0\}$, $(Y - \mu)^2 = \mu^2$, so the first term is $\mu^2 \Pr[Y=0]$. On $\{Y \geq 1\}$, $Y \geq 1$, so
\[
(Y - \mu)^2 \geq (1 - \mu)^2.
\]
Thus,
\[
\sigma^2 \geq \mu^2 \Pr[Y=0] + (1 - \mu)^2 \Pr[Y \geq 1].
\]
Since $\Pr[Y=0] + \Pr[Y \geq 1] = 1$ and $\Pr[Y \geq 1] = 1 - \Pr[Y=0]$, substitute:
\[
\sigma^2 \geq \mu^2 \Pr[Y=0] + (1 - \mu)^2 (1 - \Pr[Y=0]).
\]
Let $p = \Pr[Y=0]$. Then
\[
\sigma^2 \geq \mu^2 p + (1 - \mu)^2 (1 - p) = \mu^2 p + (1 - 2\mu + \mu^2)(1 - p) = \mu^2 p + 1 - 2\mu + \mu^2 - p + 2\mu p - \mu^2 p = 1 - 2\mu + \mu^2 + p(2\mu - 1).
\]
This bound is loose. Instead, note that
\[
\E[Y^2] = \sum_{k=1}^\infty k^2 \Pr[Y=k] \geq \sum_{k=1}^\infty 1 \cdot \Pr[Y=k] = \Pr[Y \geq 1],
\]
but more tightly, since $Y \geq 1$ on $\{Y \geq 1\}$ and $Y$ is integer-valued,
\[
\E[Y(Y-1) \mid Y \geq 1] \geq 0 \implies \E[Y^2 \mid Y \geq 1] \geq \E[Y \mid Y \geq 1].
\]
A direct approach uses the identity
\[
\E[Y^2] - \mu^2 = \E[Y(Y-1)] + \mu - \mu^2.
\]
Since $Y(Y-1) \geq 0$,
\[
\E[Y^2] - \mu^2 \geq \mu - \mu^2 = \mu(1 - \mu).
\]
Now, $\Pr[Y \geq 1] \leq \mu$, so $1 - \mu \leq 1 - \Pr[Y \geq 1] = \Pr[Y=0]$. Thus,
\[
\E[Y^2] - \mu^2 \geq \mu \Pr[Y=0] \implies \Pr[Y=0] \leq \frac{\E[Y^2] - \mu^2}{\mu^2},
\]
which is the desired inequality.
\end{proof}

\subsubsection{Gemini}

\begin{proof}
We are given that $Y$ is a non-negative integer-valued random variable and $\E[Y] > 0$.
Let $\mu = \E[Y]$ and $\sigma^2 = \text{Var}[Y] = \E[Y^2] - \E[Y]^2$.
By hypothesis, $\mu > 0$.

First, we prove $\Pr[Y = 0] \leq \frac{\E[Y^2] - \E[Y]^2}{\E[Y]^2}$.
The event $\{Y = 0\}$ is a subset of the event $\{|Y - \mu| \geq \mu\}$. This is because if $Y = 0$, then $|Y - \mu| = |0 - \mu| = \mu$ (since $\mu = \E[Y] \geq 0$), which satisfies $\mu \geq \mu$.
Thus, by the monotonicity of probability,
$$
\Pr[Y = 0] \leq \Pr[|Y - \mu| \geq \mu].
$$
Applying Chebyshev's inequality, $\Pr[|X - \E[X]| \geq k] \leq \frac{\text{Var}[X]}{k^2}$, with $X = Y$ and $k = \mu$, we have
$$
\Pr[|Y - \mu| \geq \mu] \leq \frac{\sigma^2}{\mu^2}.
$$
Combining these two inequalities gives the desired result:
$$
\Pr[Y = 0] \leq \frac{\sigma^2}{\mu^2} = \frac{\E[Y^2] - \E[Y]^2}{\E[Y]^2}.
$$

Next, we prove the two inequalities for $\Pr[Y \neq 0]$.
Let $I = I(Y \neq 0)$ be the indicator variable for the event $Y \neq 0$.
We first prove the lower bound $\frac{\E[Y]^2}{\E[Y^2]} \leq \Pr[Y \neq 0]$.
We can write $\E[Y]$ as $\E[Y \cdot I]$, since for $Y=0$, the term is $0$, and for $Y \neq 0$, $I=1$.
$$
\E[Y] = \E[Y \cdot I(Y \neq 0)].
$$
By the Cauchy-Schwarz inequality, $(\E[AB])^2 \leq \E[A^2]\E[B^2]$. Let $A = Y$ and $B = I$.
$$
(\E[Y \cdot I])^2 \leq \E[Y^2] \E[I^2].
$$
Substituting $\E[Y]$ on the left-hand side, and noting that $I^2 = I$ (because $I$ is an indicator variable), we get
$$
\E[Y]^2 \leq \E[Y^2] \E[I].
$$
By definition, $\E[I] = \Pr[Y \neq 0]$. Therefore,
$$
\E[Y]^2 \leq \E[Y^2] \Pr[Y \neq 0].
$$
Since $\E[Y] > 0$, we have $\E[Y]^2 > 0$. This implies $\E[Y^2] \geq \E[Y]^2 > 0$. We can thus divide by $\E[Y^2]$ without changing the inequality:
$$
\frac{\E[Y]^2}{\E[Y^2]} \leq \Pr[Y \neq 0].
$$

Finally, we prove the upper bound $\Pr[Y \neq 0] \leq \E[Y]$.
Since $Y$ is a non-negative integer-valued random variable, the event $Y \neq 0$ is equivalent to the event $Y \geq 1$.
The indicator variable $I(Y \neq 0)$ is $1$ if $Y \geq 1$ and $0$ if $Y = 0$.
In both possible cases, the inequality $I(Y \neq 0) \leq Y$ holds.
If $Y = 0$, $I(Y \neq 0) = 0 \leq 0 = Y$.
If $Y \neq 0$, $I(Y \neq 0) = 1 \leq Y$ (since $Y$ is an integer and $Y \geq 1$).
Taking the expectation of both sides of the inequality $I(Y \neq 0) \leq Y$, we obtain
$$
\E[I(Y \neq 0)] \leq \E[Y].
$$
As $\E[I(Y \neq 0)] = \Pr[Y \neq 0]$, this yields
$$
\Pr[Y \neq 0] \leq \E[Y].
$$
This completes the proof of all inequalities.
\end{proof}

\subsubsection{Claude} 

\begin{proof}
Let $Y$ be a non-negative integer-valued random variable with $\mathbb{E}[Y] > 0$.

We first prove the inequality $\Pr[Y = 0] \leq \frac{\mathbb{E}[Y^2] - \mathbb{E}[Y]^2}{\mathbb{E}[Y]^2}$.

By the law of total expectation, we can write
\begin{equation*}
\mathbb{E}[Y] = \mathbb{E}[Y \mid Y = 0] \Pr[Y = 0] + \mathbb{E}[Y \mid Y \neq 0] \Pr[Y \neq 0] = \mathbb{E}[Y \mid Y \neq 0] \Pr[Y \neq 0].
\end{equation*}
Since $Y$ is non-negative and integer-valued, when $Y \neq 0$ we have $Y \geq 1$. Therefore, $\mathbb{E}[Y \mid Y \neq 0] \geq 1$, which implies
\begin{equation*}
\mathbb{E}[Y] = \mathbb{E}[Y \mid Y \neq 0] \Pr[Y \neq 0] \geq \Pr[Y \neq 0].
\end{equation*}
Thus $\Pr[Y \neq 0] \leq \mathbb{E}[Y]$, and consequently $\Pr[Y = 0] = 1 - \Pr[Y \neq 0] \geq 1 - \mathbb{E}[Y]$.

From the equation $\mathbb{E}[Y] = \mathbb{E}[Y \mid Y \neq 0] \Pr[Y \neq 0]$, we obtain
\begin{equation*}
\mathbb{E}[Y \mid Y \neq 0] = \frac{\mathbb{E}[Y]}{\Pr[Y \neq 0]} = \frac{\mathbb{E}[Y]}{1 - \Pr[Y = 0]}.
\end{equation*}
Similarly, we have
\begin{equation*}
\mathbb{E}[Y^2] = \mathbb{E}[Y^2 \mid Y \neq 0] \Pr[Y \neq 0] = \mathbb{E}[Y^2 \mid Y \neq 0] (1 - \Pr[Y = 0]).
\end{equation*}
By the conditional Cauchy-Schwarz inequality, we have
\begin{equation*}
\mathbb{E}[Y \mid Y \neq 0]^2 \leq \mathbb{E}[Y^2 \mid Y \neq 0].
\end{equation*}
Substituting our expressions, we get
\begin{equation*}
\frac{\mathbb{E}[Y]^2}{(1 - \Pr[Y = 0])^2} \leq \frac{\mathbb{E}[Y^2]}{1 - \Pr[Y = 0]}.
\end{equation*}
Multiplying both sides by $(1 - \Pr[Y = 0])^2$ yields
\begin{equation*}
\mathbb{E}[Y]^2 \leq \mathbb{E}[Y^2] (1 - \Pr[Y = 0]).
\end{equation*}
Rearranging, we obtain
\begin{equation*}
\mathbb{E}[Y]^2 \leq \mathbb{E}[Y^2] - \mathbb{E}[Y^2] \Pr[Y = 0],
\end{equation*}
which gives
\begin{equation*}
\mathbb{E}[Y^2] \Pr[Y = 0] \leq \mathbb{E}[Y^2] - \mathbb{E}[Y]^2.
\end{equation*}
Since $\mathbb{E}[Y] > 0$ and $Y$ is non-negative, we must have $\Pr[Y \neq 0] > 0$, which implies $\mathbb{E}[Y^2 \mid Y \neq 0] > 0$. Therefore, $\mathbb{E}[Y^2] = \mathbb{E}[Y^2 \mid Y \neq 0] \Pr[Y \neq 0] > 0$. Thus we can divide by $\mathbb{E}[Y^2]$ to get
\begin{equation*}
\Pr[Y = 0] \leq \frac{\mathbb{E}[Y^2] - \mathbb{E}[Y]^2}{\mathbb{E}[Y^2]}.
\end{equation*}
Since $\mathbb{E}[Y^2] - \mathbb{E}[Y]^2 = \text{Var}(Y) \geq 0$ and $\mathbb{E}[Y]^2 > 0$, we also have
\begin{equation*}
\frac{\mathbb{E}[Y^2] - \mathbb{E}[Y]^2}{\mathbb{E}[Y^2]} = 1 - \frac{\mathbb{E}[Y]^2}{\mathbb{E}[Y^2]} \leq 1 - \frac{\mathbb{E}[Y]^2}{\mathbb{E}[Y^2]} \cdot \frac{\mathbb{E}[Y]^2}{\mathbb{E}[Y]^2} = \frac{\mathbb{E}[Y^2] - \mathbb{E}[Y]^2}{\mathbb{E}[Y]^2}.
\end{equation*}
To show the first inequality in the desired form, we note that
\begin{equation*}
\Pr[Y = 0] \leq \frac{\mathbb{E}[Y^2] - \mathbb{E}[Y]^2}{\mathbb{E}[Y^2]} = \frac{\text{Var}(Y)}{\mathbb{E}[Y^2]}.
\end{equation*}
From the inequality $\mathbb{E}[Y]^2 \leq \mathbb{E}[Y^2] (1 - \Pr[Y = 0])$ derived above, we have
\begin{equation*}
\Pr[Y = 0] \leq 1 - \frac{\mathbb{E}[Y]^2}{\mathbb{E}[Y^2]} = \frac{\mathbb{E}[Y^2] - \mathbb{E}[Y]^2}{\mathbb{E}[Y^2]}.
\end{equation*}
Multiplying numerator and denominator by $\mathbb{E}[Y^2]$ and noting that $\mathbb{E}[Y^2] \geq \mathbb{E}[Y]^2$, we obtain
\begin{equation*}
\Pr[Y = 0] \leq \frac{\mathbb{E}[Y^2] - \mathbb{E}[Y]^2}{\mathbb{E}[Y]^2} \cdot \frac{\mathbb{E}[Y]^2}{\mathbb{E}[Y^2]} \leq \frac{\mathbb{E}[Y^2] - \mathbb{E}[Y]^2}{\mathbb{E}[Y]^2}.
\end{equation*}

We now prove the second set of inequalities: $\frac{\mathbb{E}[Y]^2}{\mathbb{E}[Y^2]} \leq \Pr[Y \neq 0] \leq \mathbb{E}[Y]$.

We have already shown that $\Pr[Y \neq 0] \leq \mathbb{E}[Y]$.

For the lower bound, from the inequality $\mathbb{E}[Y \mid Y \neq 0]^2 \leq \mathbb{E}[Y^2 \mid Y \neq 0]$ and the relations
\begin{equation*}
\mathbb{E}[Y] = \mathbb{E}[Y \mid Y \neq 0] \Pr[Y \neq 0] \quad \text{and} \quad \mathbb{E}[Y^2] = \mathbb{E}[Y^2 \mid Y \neq 0] \Pr[Y \neq 0],
\end{equation*}
we have
\begin{equation*}
\frac{\mathbb{E}[Y]^2}{\Pr[Y \neq 0]^2} \leq \frac{\mathbb{E}[Y^2]}{\Pr[Y \neq 0]}.
\end{equation*}
Multiplying both sides by $\Pr[Y \neq 0]^2$ gives
\begin{equation*}
\mathbb{E}[Y]^2 \leq \mathbb{E}[Y^2] \Pr[Y \neq 0].
\end{equation*}
Therefore,
\begin{equation*}
\frac{\mathbb{E}[Y]^2}{\mathbb{E}[Y^2]} \leq \Pr[Y \neq 0].
\end{equation*}
This completes the proof.
\end{proof}

\subsection{Problem 7}

\subsubsection{Problem Statement}

\begin{theorem}
Let $a$ and $b$ be chosen independently and uniformly at random from $\mathbb{Z}_n = \{0, 1, 2, \ldots, n-1\}$, where $n$ is a prime. Suppose we generate $t$ pseudo-random numbers from $\mathbb{Z}_n$ by choosing $r_i = ai + b \bmod n$, for $1 \leq i \leq t$. For any $\epsilon \in [0,1]$, there exists a choice of the witness set $W \subset \mathbb{Z}_n$ such that $|W| \geq \epsilon n$ and the probability that none of the $r_i$'s lie in the set $W$ is at least $(1-\epsilon)^2/4$.
\end{theorem}

\subsubsection{GPT} 

\begin{proof}
Let $a$ and $b$ be chosen independently and uniformly at random from $\mathbb{Z}_n = \{0, 1, 2, \dots, n-1\}$, where $n$ is a prime. We define $r_i = ai + b \mod n$ for $1 \leq i \leq t$ as the pseudo-random numbers generated by the linear congruence. We are tasked with proving that for any $\epsilon \in [0, 1]$, there exists a witness set $W \subset \mathbb{Z}_n$ such that $|W| \geq \epsilon n$ and the probability that none of the $r_i$'s lie in the set $W$ is at least $(1 - \epsilon)^2 / 4t$.

We begin by noting that for each $r_i$, we have $r_i = ai + b \mod n$. Since $a$ and $b$ are chosen independently and uniformly at random, the sequence $\{r_1, r_2, \dots, r_t\}$ is a sequence of $t$ pseudo-random numbers in $\mathbb{Z}_n$.

Let $W$ be a subset of $\mathbb{Z}_n$ such that $|W| \geq \epsilon n$. We aim to show that there exists a choice of $W$ such that the probability that none of the $r_i$'s lie in $W$ is at least $(1 - \epsilon)^2 / 4t$. 

To do this, consider the probability that a single $r_i$ does not lie in $W$. Since $r_i$ is uniformly distributed over $\mathbb{Z}_n$, the probability that $r_i$ is not in $W$ is given by

\[
P(r_i \notin W) = 1 - \frac{|W|}{n}.
\]

Since $|W| \geq \epsilon n$, we have

\[
P(r_i \notin W) \geq 1 - \epsilon.
\]

Now, let $A_i$ denote the event that $r_i \notin W$, i.e., $A_i = \{r_i \notin W\}$. The events $A_1, A_2, \dots, A_t$ are independent because the $r_i$'s are generated independently.

The probability that none of the $r_i$'s lie in $W$ is the probability that all the events $A_1, A_2, \dots, A_t$ occur, which is the product of the individual probabilities:

\[
P(\text{none of the } r_i \text{ lie in } W) = \prod_{i=1}^{t} P(A_i) = \left( 1 - \epsilon \right)^t.
\]

Next, we consider the desired probability bound of $(1 - \epsilon)^2 / 4t$. To achieve this, we choose $t$ large enough so that

\[
\left( 1 - \epsilon \right)^t \geq \frac{(1 - \epsilon)^2}{4t}.
\]

This can be satisfied for sufficiently large $t$ because for small $\epsilon$, the exponential decay of $(1 - \epsilon)^t$ dominates, ensuring that the inequality holds. Hence, we can find a witness set $W$ such that $|W| \geq \epsilon n$ and the probability that none of the $r_i$'s lie in $W$ is at least $(1 - \epsilon)^2 / 4t$.

Therefore, the theorem is proven.
\end{proof}

\subsubsection{Grok}

\subsubsection{Gemini}

\subsubsection{Claude} 

\begin{proof}
We construct the witness set $W$ and analyze the probability that none of the pseudo-random numbers $r_i = ai + b \bmod n$ for $1 \leq i \leq t$ lie in $W$.

Let $\epsilon \in [0,1]$ be given. We define the witness set $W \subset \mathbb{Z}_n$ as follows. Choose $W$ to be any subset of $\mathbb{Z}_n$ with $|W| = \lceil \epsilon n \rceil$. Since $\lceil \epsilon n \rceil \geq \epsilon n$, we have $|W| \geq \epsilon n$ as required.

For any fixed $i$ with $1 \leq i \leq t$, we analyze the probability that $r_i \in W$. Since $a$ and $b$ are chosen independently and uniformly at random from $\mathbb{Z}_n$, and $n$ is prime, the value $r_i = ai + b \bmod n$ is uniformly distributed over $\mathbb{Z}_n$. To see this, note that for any fixed $y \in \mathbb{Z}_n$, there exists a unique pair $(a,b)$ satisfying $ai + b \equiv y \pmod{n}$ for each choice of $a$ (since $i \not\equiv 0 \pmod{n}$ when $1 \leq i \leq t < n$, and $n$ is prime, $i$ has a multiplicative inverse modulo $n$). Therefore,
\[
\Pr[r_i \in W] = \frac{|W|}{n} \leq \frac{\lceil \epsilon n \rceil}{n} \leq \frac{\epsilon n + 1}{n} = \epsilon + \frac{1}{n}.
\]

However, we need to be more careful since the events $\{r_i \in W\}$ for different values of $i$ are not independent. Instead, we consider the event $E$ that none of the $r_i$'s lie in $W$, that is, $E = \bigcap_{i=1}^{t} \{r_i \notin W\}$.

For any fixed pair $(a,b) \in \mathbb{Z}_n \times \mathbb{Z}_n$, the values $r_1, r_2, \ldots, r_t$ form an arithmetic progression modulo $n$ with first term $a + b$ and common difference $a$. The event $E$ occurs if and only if the set $\{r_1, r_2, \ldots, r_t\} \cap W = \emptyset$.

We compute $\Pr[E]$ by considering two cases based on whether $a = 0$ or $a \neq 0$.

If $a = 0$, then $r_i = b$ for all $i$. The probability that $a = 0$ is $\frac{1}{n}$, and given $a = 0$, the probability that $b \notin W$ is $1 - \frac{|W|}{n} \geq 1 - \epsilon$. Thus, the contribution from this case is $\frac{1}{n}(1 - \frac{|W|}{n}) \geq \frac{1}{n}(1 - \epsilon)$.

If $a \neq 0$, then since $n$ is prime, the sequence $\{ai + b : 1 \leq i \leq t\} \bmod n$ consists of $\min(t, n)$ distinct values. For $t \leq n$, these are $t$ distinct values. The probability that $a \neq 0$ is $\frac{n-1}{n}$. Given $a \neq 0$, we need none of the $t$ values to lie in $W$.

We now optimize the choice of $W$ to maximize $\Pr[E]$. Consider $W$ chosen as an interval of size $\lceil \epsilon n \rceil$. For any $a \neq 0$ and any $b$, the arithmetic progression $\{ai + b : 1 \leq i \leq t\}$ avoids $W$ if and only if the starting point $b + a$ and the progression lie outside $W$.

By a symmetry argument and careful optimization, we can choose $W$ such that the probability $\Pr[E]$ is maximized. Specifically, let $W$ be a single interval of consecutive elements in $\mathbb{Z}_n$. The probability that an arithmetic progression of length $t$ avoids an interval of length $|W|$ is approximately $(1 - \frac{|W|}{n})^t$ when the progression elements are well-distributed.

For the optimal choice of $W$ with $|W| = \lceil \epsilon n \rceil$, the probability that none of the $r_i$'s lie in $W$ can be lower bounded. Using the fact that $|W| \geq \epsilon n$ and careful analysis of the geometric arrangement, we obtain
\[
\Pr[E] \geq \frac{(1-\epsilon)^2}{4}.
\]

This bound follows from the observation that the probability is minimized when the arithmetic progressions are most likely to intersect $W$, and the factor $(1-\epsilon)^2/4$ captures the optimal trade-off between the size of $W$ and the avoidance probability.
\end{proof}

\subsection{Problem 8}

\subsubsection{Problem Statement}

\begin{theorem}[Four-Point Sampling]
Let $n$ be a prime and $\mathbb{Z}_n = \{0, 1, \ldots, n-1\}$. A four-point sampling scheme from $\mathbb{Z}_n$ selects four elements uniformly at random. For $t < n$ samples $r_1, \ldots, r_t$ obtained using this scheme, given $x \in L$, the probability that all $t$ attempts fail to discover a witness is upper bounded by
\[
\left(1 - \frac{1}{16}\right)^t = \left(\frac{15}{16}\right)^t.
\]
This improves upon the naive bound of $1/16$ that would be obtained using four samples in a single attempt.

Moreover, if $X_1, X_2, X_3, X_4$ are independent random variables uniformly distributed on $\mathbb{Z}_n$, then the fourth central moment of their sum satisfies
\[
\mathbb{E}\left[\left(X_1 + X_2 + X_3 + X_4 - \mathbb{E}[X_1 + X_2 + X_3 + X_4]\right)^4\right] \leq C n^4
\]
for some constant $C > 0$.
\end{theorem}

\subsubsection{GPT} 

\begin{proof}
Let $n$ be a prime and consider the group $\mathbb{Z}_n$. We are tasked with proving the theorem which states that for $t < n$, the probability that all $t$ attempts of four-point sampling from $\mathbb{Z}_n$ fail to discover a witness is bounded above by a bound that compares favorably with the bound of $1/16$.

 Step 1: Understanding the four-point sampling scheme
Four-point sampling involves selecting four distinct elements from $\mathbb{Z}_n$ uniformly at random, and checking whether these points yield a witness for a given $x \in L$. For each attempt, the selection of four points is independent, and the goal is to analyze the probability that no successful witness is discovered after $t$ attempts.

 Step 2: Probability that a single attempt fails
Each attempt of four-point sampling consists of selecting four distinct points in $\mathbb{Z}_n$ and evaluating whether a witness is found. The probability that a single attempt fails is a function of the structure of the group $\mathbb{Z}_n$ and the distribution of points sampled. Specifically, the probability that the four selected points fail to yield a witness can be computed by considering the specific conditions under which the four points do not satisfy the criteria for being a witness.

Since $\mathbb{Z}_n$ is a prime cyclic group, the number of possible sets of four points is given by the number of ways to select four distinct elements from $\mathbb{Z}_n$, which is $\binom{n}{4}$. The failure of a single attempt is therefore dependent on the particular configuration of the sampled points and the algebraic structure that governs the existence of a witness. 

By the properties of the four-point sampling scheme, the probability of failure for a single attempt is bounded by a value that is asymptotically comparable to $1/16$ for large $n$.

 Step 3: Probability of failure for all $t$ attempts
Since the $t$ attempts are independent, the probability that all $t$ attempts fail is simply the product of the individual probabilities of failure. Denote by $p_{\text{fail}}$ the probability that a single attempt fails. Then, the probability that all $t$ attempts fail is given by
\[
P_{\text{fail}}(t) = p_{\text{fail}}^t.
\]
Given that $p_{\text{fail}}$ is asymptotically bounded by $1/16$ for each attempt, we have
\[
P_{\text{fail}}(t) \leq \left(\frac{1}{16}\right)^t.
\]
For sufficiently small $t$ relative to $n$, this probability is bounded above by a value that compares favorably with $1/16$.

 Step 4: Fourth central moment of the sum of four-way independent random variables
Next, we consider the sum of four-way independent random variables $X_1, X_2, X_3, X_4$ representing the outcomes of the four-point sampling scheme. The fourth central moment of the sum is given by
\[
\mathbb{E}\left[\left(\sum_{i=1}^4 X_i\right)^4\right].
\]
By the independence of the $X_i$, this expectation can be expanded using the properties of central moments. Since each $X_i$ is identically distributed and the distribution is independent, the fourth moment of the sum can be written as
\[
\mathbb{E}\left[\left(\sum_{i=1}^4 X_i\right)^4\right] = 4! \cdot \mathbb{E}[X_1^4] + \text{lower order terms}.
\]
This sum is bounded by a constant $C$ that depends on the specific four-point sampling scheme, which ensures that the fourth central moment remains bounded.

Thus, we have shown that
\[
\mathbb{E}\left[\left(\sum_{i=1}^4 X_i\right)^4\right] \leq C,
\]
where $C$ is a constant that can be derived from the properties of the four-point sampling scheme.

 Conclusion
We have demonstrated that the probability that all $t$ attempts fail to discover a witness is bounded above by a value that compares favorably with the bound of $1/16$ and that the fourth central moment of the sum of four-way independent random variables is bounded by a constant $C$. This completes the proof.
\end{proof}

\subsubsection{Grok} 

\subsubsection{Gemini}

\begin{proof}
We prove the two parts of the theorem sequentially: the probability bound for the sampling scheme and the moment bound for the sum of random variables.

\textbf{Part 1: Probability Bound}
Let the four-point sampling scheme be defined as a randomized procedure that selects four elements from $\mathbb{Z}_n$. We consider $t$ samples $r_1, \ldots, r_t$ generated independently using this scheme. Let $E_k$ denote the event that the $k$-th sample $r_k$ fails to discover a witness for a given $x \in L$.

The theorem asserts an upper bound on the probability that all $t$ attempts fail. Let $p$ be the probability that a single execution of the four-point sampling scheme successfully discovers a witness. Consequently, the probability of failure for a single attempt is $1-p$. The statement implies that the success probability of this specific four-point scheme is at least $1/16$ (improving upon naive bounds). Thus, $p \geq 1/16$.

Since the $t$ samples are obtained independently, the probability that all $t$ attempts fail is the product of the individual failure probabilities:
\[
P\left(\bigcap_{k=1}^t E_k\right) = \prod_{k=1}^t P(E_k) = (1-p)^t.
\]
Substituting the bound $1-p \leq 1 - 1/16 = 15/16$, we obtain:
\[
P(\text{all fail}) \leq \left(\frac{15}{16}\right)^t.
\]

\textbf{Part 2: Fourth Central Moment Bound}
Let $X_1, X_2, X_3, X_4$ be independent random variables uniformly distributed on $\mathbb{Z}_n = \{0, 1, \ldots, n-1\}$. Let $S = \sum_{i=1}^4 X_i$. We aim to bound the fourth central moment $\mathbb{E}[(S - \mathbb{E}[S])^4]$.

Define the centered variables $Y_i = X_i - \mathbb{E}[X_i]$ for $i = 1, \ldots, 4$. Since the $X_i$ are independent and identically distributed, the $Y_i$ are independent and identically distributed with $\mathbb{E}[Y_i] = 0$. The centered sum is $S - \mathbb{E}[S] = \sum_{i=1}^4 Y_i$.

We expand the fourth power of the sum using the multinomial theorem:
\[
\mathbb{E}\left[\left(\sum_{i=1}^4 Y_i\right)^4\right] = \sum_{k_1+k_2+k_3+k_4=4} \binom{4}{k_1, k_2, k_3, k_4} \mathbb{E}\left[\prod_{i=1}^4 Y_i^{k_i}\right].
\]
By independence, $\mathbb{E}[\prod Y_i^{k_i}] = \prod \mathbb{E}[Y_i^{k_i}]$. Since $\mathbb{E}[Y_i] = 0$, any term containing a variable raised to the first power (i.e., any $k_i = 1$) vanishes. The non-vanishing terms correspond to partitions of the exponent 4 into parts of size at least 2. The possible cases are:
1.  One variable has power 4 (e.g., $k_1=4$, others 0). There are $\binom{4}{1}=4$ such terms, each with multinomial coefficient 1.
2.  Two variables have power 2 (e.g., $k_1=2, k_2=2$, others 0). There are $\binom{4}{2}=6$ such pairs, each with multinomial coefficient $\frac{4!}{2!2!0!0!} = 6$.

Thus, the expectation simplifies to:
\[
\mathbb{E}\left[\left(\sum_{i=1}^4 Y_i\right)^4\right] = 4\mathbb{E}[Y_1^4] + 36(\mathbb{E}[Y_1^2])^2.
\]
We now bound the central moments of the uniform distribution on $\mathbb{Z}_n$. The variance (second central moment) is:
\[
\mathbb{E}[Y_1^2] = \frac{n^2-1}{12} < \frac{n^2}{12}.
\]
The fourth central moment for a discrete uniform distribution on a set of size $n$ is given by:
\[
\mathbb{E}[Y_1^4] = \frac{(n^2-1)(3n^2-7)}{240} < \frac{3n^4}{240} = \frac{n^4}{80}.
\]
Substituting these bounds into the expansion:
\[
\mathbb{E}\left[\left(\sum_{i=1}^4 Y_i\right)^4\right] < 4\left(\frac{n^4}{80}\right) + 36\left(\frac{n^2}{12}\right)^2 = \frac{n^4}{20} + 36\left(\frac{n^4}{144}\right) = \frac{n^4}{20} + \frac{n^4}{4} = \frac{3}{10}n^4.
\]
Thus, there exists a constant $C$ (e.g., $C=1$) such that:
\[
\mathbb{E}\left[\left(X_1 + X_2 + X_3 + X_4 - \mathbb{E}[X_1 + X_2 + X_3 + X_4]\right)^4\right] \leq C n^4.
\]
\end{proof}

\subsubsection{Claude} 

\begin{proof}
We prove the two parts of the theorem separately.

\textbf{Part 1: Probability bound for witness discovery.}

Let $n$ be a prime and suppose we have a language $L$ with a witness relation. We assume that for each $x \in L$, a witness can be verified by checking certain properties of four-tuples from $\mathbb{Z}_n$. 

In a standard witness-finding scenario, we assume that a non-negligible fraction of four-tuples serve as valid witnesses. Specifically, we assume that at least a fraction $\alpha > 0$ of all possible four-tuples constitute valid witnesses for $x \in L$.

For a single four-point sample $(r_1, r_2, r_3, r_4)$ drawn uniformly at random from $\mathbb{Z}_n^4$, the probability of not finding a witness is at most $1 - \alpha$. In the context of this theorem, we work with the specific bound where the failure probability for a single attempt is at most $\frac{15}{16}$.

This bound arises naturally in several cryptographic and number-theoretic contexts. For instance, in primality testing or in verification schemes over $\mathbb{Z}_n$, one can show that if $x \in L$, then at least $\frac{1}{16}$ of all four-tuples serve as witnesses. This follows from group-theoretic arguments: if we consider the automorphism group or the multiplicative structure of $\mathbb{Z}_n$ when $n$ is prime, the density of witnesses is bounded below by analyzing the order of relevant subgroups.

Given this setup, for $t$ independent attempts with samples $r_1, \ldots, r_t$ (where each $r_i$ represents a four-tuple), the probability that all $t$ attempts fail to discover a witness is:
\[
\mathbb{P}(\text{all } t \text{ attempts fail}) = \prod_{i=1}^{t} \mathbb{P}(\text{attempt } i \text{ fails}) \leq \prod_{i=1}^{t} \frac{15}{16} = \left(\frac{15}{16}\right)^t.
\]

This exponential decay demonstrates that the probability of failure decreases rapidly with $t$, which is the desired improvement over using a single four-point sample (where the failure probability would be $\frac{15}{16}$).

\textbf{Part 2: Fourth central moment bound.}

Let $X_1, X_2, X_3, X_4$ be independent random variables uniformly distributed on $\mathbb{Z}_n = \{0, 1, \ldots, n-1\}$. 

First, we compute the expectation of each $X_i$. Since $X_i$ is uniform on $\{0, 1, \ldots, n-1\}$:
\[
\mathbb{E}[X_i] = \frac{1}{n} \sum_{k=0}^{n-1} k = \frac{1}{n} \cdot \frac{(n-1)n}{2} = \frac{n-1}{2}.
\]

Let $S = X_1 + X_2 + X_3 + X_4$. By linearity of expectation:
\[
\mathbb{E}[S] = \sum_{i=1}^{4} \mathbb{E}[X_i] = 4 \cdot \frac{n-1}{2} = 2(n-1).
\]

Define the centered random variable $Y = S - \mathbb{E}[S] = S - 2(n-1)$. We need to bound $\mathbb{E}[Y^4]$.

Since the $X_i$ are independent, we can write $Y = \sum_{i=1}^{4} (X_i - \mathbb{E}[X_i])$. Let $Y_i = X_i - \mathbb{E}[X_i] = X_i - \frac{n-1}{2}$.

By the multinomial expansion of $Y^4 = \left(\sum_{i=1}^{4} Y_i\right)^4$, we have:
\[
Y^4 = \sum_{i_1, i_2, i_3, i_4 = 1}^{4} Y_{i_1} Y_{i_2} Y_{i_3} Y_{i_4}.
\]

Taking expectations and using independence:
\[
\mathbb{E}[Y^4] = \sum_{i_1, i_2, i_3, i_4 = 1}^{4} \mathbb{E}[Y_{i_1}] \mathbb{E}[Y_{i_2}] \mathbb{E}[Y_{i_3}] \mathbb{E}[Y_{i_4}]
\]
when all indices are distinct. However, since $\mathbb{E}[Y_i] = 0$, only terms where indices repeat contribute.

The non-zero terms are: (1) all four indices the same, (2) two pairs of equal indices, (3) one index appears three times and another once, (4) one index appears twice and two others once each.

For case (1), there are 4 terms of the form $\mathbb{E}[Y_i^4]$.

For case (2), there are $\binom{4}{2} \cdot 1 = 6$ ways to choose two pairs, contributing $6 \cdot \mathbb{E}[Y_i^2]^2$.

For case (3), there are $4 \cdot 3 = 12$ terms contributing $\mathbb{E}[Y_i^3] \mathbb{E}[Y_j] = 0$.

For case (4), there are $\binom{4}{1} \binom{3}{2} = 12$ terms contributing zero.

Thus:
\[
\mathbb{E}[Y^4] = 4 \mathbb{E}[Y_1^4] + 6 \mathbb{E}[Y_1^2]^2.
\]

Now, $|Y_i| \leq \max\{|0 - \frac{n-1}{2}|, |n-1 - \frac{n-1}{2}|\} \leq \frac{n-1}{2} < \frac{n}{2}$. Therefore, $\mathbb{E}[Y_i^2] \leq \frac{n^2}{4}$ and $\mathbb{E}[Y_i^4] \leq \frac{n^4}{16}$.

Hence:
\[
\mathbb{E}[Y^4] \leq 4 \cdot \frac{n^4}{16} + 6 \cdot \frac{n^4}{16} = \frac{10 n^4}{16} = \frac{5n^4}{8}.
\]

Taking $C = \frac{5}{8}$, we obtain $\mathbb{E}[Y^4] \leq C n^4$.
\end{proof}

\subsection{Problem 9}

\subsubsection{Problem Statement}

\begin{theorem}[Karger-Motwani]
Let $S, T$ be two disjoint subsets of a universe $U$ such that $|S| = |T| = n$.

\textbf{(a)} Suppose we select a random set $R \subseteq U$ by independently sampling each element of $U$ with probability $p$. We say that the random sample $R$ is \emph{good} if the following two conditions hold: $R \cap S = \emptyset$ and $R \cap T \neq \emptyset$. Then for $p = 1/n$, the probability that $R$ is good is larger than some positive constant.

\textbf{(b)} Suppose now that the random set $R$ is chosen by sampling the elements of $U$ with only \emph{pairwise independence}. Then for a suitable choice of the value of $p$, the probability that $R$ is good is larger than some positive constant.
\end{theorem}

\subsubsection{GPT} 

\begin{proof}
\textbf{(a)} Let $R \subseteq U$ be a random set obtained by independently selecting each element of $U$ with probability $p$. We are interested in the probability that $R$ is \emph{good}, which requires two conditions:
1. $R \cap S = \emptyset$
2. $R \cap T \neq \emptyset$

First, consider the event that $R \cap S = \emptyset$. Since each element of $S$ is independently included in $R$ with probability $p$, the probability that a given element $x \in S$ is \emph{not} selected is $1 - p$. Therefore, the probability that none of the $n$ elements of $S$ are selected in $R$ is given by:

\[
\Pr[R \cap S = \emptyset] = (1 - p)^n
\]

Next, we consider the event that $R \cap T \neq \emptyset$. The probability that a given element $x \in T$ is \emph{not} selected is $1 - p$, so the probability that none of the $n$ elements of $T$ are selected in $R$ is:

\[
\Pr[R \cap T = \emptyset] = (1 - p)^n
\]

Thus, the probability that at least one element of $T$ is selected is:

\[
\Pr[R \cap T \neq \emptyset] = 1 - (1 - p)^n
\]

Since the events $R \cap S = \emptyset$ and $R \cap T \neq \emptyset$ are independent, the probability that both conditions hold simultaneously is the product of the probabilities:

\[
\Pr[\text{good event}] = \Pr[R \cap S = \emptyset] \cdot \Pr[R \cap T \neq \emptyset] = (1 - p)^n \cdot \left(1 - (1 - p)^n\right)
\]

Substituting $p = \frac{1}{n}$, we obtain:

\[
\Pr[\text{good event}] = \left(1 - \frac{1}{n}\right)^n \cdot \left(1 - \left(1 - \frac{1}{n}\right)^n\right)
\]

As $n \to \infty$, we know that $\left(1 - \frac{1}{n}\right)^n \to \frac{1}{e}$ and $1 - \left(1 - \frac{1}{n}\right)^n \to 1$. Therefore, for large $n$, the probability approaches:

\[
\Pr[\text{good event}] \to \frac{1}{e} \cdot 1 = \frac{1}{e}
\]

Since $\frac{1}{e}$ is a positive constant, the probability that $R$ is good is larger than some positive constant, as required.

\textbf{(b)} Now suppose that $R$ is chosen by pairwise independent sampling. In this case, each element $x \in U$ is included in $R$ independently with probability $p$, but only with pairwise independence between different elements.

We want to show that there exists a suitable choice of $p$ such that the probability that $R$ is good is larger than some positive constant. We will follow the same structure as in part (a), but with pairwise independence.

First, the probability that $R \cap S = \emptyset$ is still the same as in part (a), because pairwise independence still implies that the inclusion of any element in $S$ does not affect the inclusion of other elements in $S$ or $T$. Thus, we have:

\[
\Pr[R \cap S = \emptyset] = (1 - p)^n
\]

Next, the probability that $R \cap T \neq \emptyset$ is also the same as in part (a), since pairwise independence does not affect the inclusion probabilities of the elements of $T$. Therefore, we have:

\[
\Pr[R \cap T \neq \emptyset] = 1 - (1 - p)^n
\]

Again, since these events are independent, the probability that both conditions hold is the product of the probabilities:

\[
\Pr[\text{good event}] = (1 - p)^n \cdot \left(1 - (1 - p)^n\right)
\]

For an appropriate choice of $p$, say $p = \frac{1}{n}$, we obtain the same expression as in part (a):

\[
\Pr[\text{good event}] = \left(1 - \frac{1}{n}\right)^n \cdot \left(1 - \left(1 - \frac{1}{n}\right)^n\right)
\]

As $n \to \infty$, we again have:

\[
\Pr[\text{good event}] \to \frac{1}{e} \cdot 1 = \frac{1}{e}
\]

Thus, for a suitable choice of $p$, the probability that $R$ is good is larger than some positive constant, as required.

\end{proof}

\subsubsection{Grok} 

\begin{proof}
(a) The event that $R$ is good is the event that no element of $S$ is included in $R$ and at least one element of $T$ is included in $R$. Since the inclusions of elements of $U$ are independent, these events are independent of each other. Thus,
\[
\Pr[R \textrm{ is good}] = \Pr[R \cap S = \emptyset] \cdot \Pr[R \cap T \neq \emptyset] = (1-p)^{n} \bigl(1 - (1-p)^{n}\bigr).
\]
Substituting $p = 1/n$ yields
\[
\Pr[R \textrm{ is good}] = \Bigl(1 - \frac{1}{n}\Bigr)^{n} \Biggl[1 - \Bigl(1 - \frac{1}{n}\Bigr)^{n}\Biggr].
\]
It is well known that $\lim_{n \to \infty} (1 - 1/n)^{n} = e^{-1}$. Thus, the limit of the probability as $n \to \infty$ is
\[
\frac{1}{e} \Biggl(1 - \frac{1}{e}\Biggr) = \frac{1}{e} - \frac{1}{e^{2}} \approx 0.232 > 0.
\]
Direct computation shows that the probability is at least $3/16 = 0.1875$ for all $n \geq 2$ (with equality at $n=2$), and equals $0$ at $n=1$. Therefore, there exists a constant $c = 3/16 > 0$ such that the probability is at least $c$ for all $n \geq 2$. For $n=1$, the probability is $0$, but the result holds asymptotically and for sufficiently large $n$ with the stated constant.

(b) Let $I_u$ denote the indicator random variable that is $1$ if $u \in R$ and $0$ otherwise, for each $u \in U$. The family $\{I_u : u \in U\}$ consists of pairwise independent Bernoulli random variables with parameter $p$. Let $X = |R \cap S| = \sum_{s \in S} I_s$ and $Y = |R \cap T| = \sum_{t \in T} I_t$. Then $R$ is good if and only if $X = 0$ and $Y \geq 1$, so
\[
\Pr[R \textrm{ is good}] = \Pr[X = 0] - \Pr[X = 0, Y = 0] = \Pr[X = 0] - \Pr[R \cap (S \cup T) = \emptyset].
\]
Let $E_u = \{I_u = 1\}$ for $u \in S$. By the first two terms of the inclusion-exclusion principle (Bonferroni inequality),
\[
\Pr\biggl[\bigcup_{s \in S} E_s\biggr] \leq \sum_{s \in S} \Pr[E_s] = np,
\]
so
\[
\Pr[X = 0] = 1 - \Pr\biggl[\bigcup_{s \in S} E_s\biggr] \geq 1 - np.
\]
Similarly, letting $F_u = \{I_u = 1\}$ for $u \in S \cup T$,
\[
\Pr\biggl[\bigcup_{u \in S \cup T} F_u\biggr] \geq \sum_{u \in S \cup T} \Pr[F_u] - \sum_{\substack{u \neq v \\ u,v \in S \cup T}} \Pr[F_u \cap F_v].
\]
Pairwise independence implies $\Pr[F_u \cap F_v] = p^2$ for $u \neq v$, and there are $2n$ singletons and $\binom{2n}{2} = n(2n-1)$ pairs, so
\[
\Pr\biggl[\bigcup_{u \in S \cup T} F_u\biggr] \geq 2np - n(2n-1)p^2.
\]
Thus,
\[
\Pr[R \cap (S \cup T) = \emptyset] = 1 - \Pr\biggl[\bigcup_{u \in S \cup T} F_u\biggr] \leq 1 - \bigl(2np - n(2n-1)p^2\bigr) = 1 - 2np + n(2n-1)p^2.
\]
Therefore,
\[
\Pr[R \textrm{ is good}] \geq (1 - np) - \bigl(1 - 2np + n(2n-1)p^2\bigr) = np - n(2n-1)p^2 = np \bigl(1 - (2n-1)p\bigr).
\]
Choose $p = 1/(4n)$. Then $np = 1/4$ and $(2n-1)p = (2n-1)/(4n) = (2 - 1/n)/4 = 1/2 - 1/(4n)$, so
\[
1 - (2n-1)p = 1 - \frac{1}{2} + \frac{1}{4n} = \frac{1}{2} + \frac{1}{4n}.
\]
Hence,
\[
\Pr[R \textrm{ is good}] \geq \frac{1}{4} \Biggl(\frac{1}{2} + \frac{1}{4n}\Biggr) = \frac{1}{8} + \frac{1}{16n} \geq \frac{1}{8}
\]
for all $n \geq 1$. Thus, there exists a constant $c = 1/8 > 0$ such that the probability is at least $c$.
\end{proof}

\subsubsection{Gemini}

\begin{proof}
\textbf{(a)} Let $E_S$ be the event that $R \cap S = \emptyset$ and $E_T$ be the event that $R \cap T \neq \emptyset$. The event that $R$ is good is $E_S \cap E_T$. Since the elements of $U$ are sampled independently, and $S$ and $T$ are disjoint, the events $E_S$ and $E_T$ are independent. Thus, $\mathbb{P}[R \text{ is good}] = \mathbb{P}[E_S] \cdot \mathbb{P}[E_T]$.

For the event $E_S$, we require that for all $x \in S$, $x \notin R$. Since $|S| = n$ and each element is chosen with probability $p = 1/n$, we have $\mathbb{P}[E_S] = (1 - p)^n = (1 - \frac{1}{n})^n$.
For the event $E_T$, we require that it is not the case that $R \cap T = \emptyset$. Similarly, $\mathbb{P}[R \cap T = \emptyset] = (1 - p)^n = (1 - \frac{1}{n})^n$. Therefore, $\mathbb{P}[E_T] = 1 - (1 - \frac{1}{n})^n$.

Combining these, the probability that $R$ is good is given by $f(n) = (1 - \frac{1}{n})^n \left(1 - (1 - \frac{1}{n})^n\right)$.
As $n \to \infty$, $(1 - \frac{1}{n})^n$ approaches $e^{-1}$. Consequently, $\lim_{n \to \infty} f(n) = e^{-1}(1 - e^{-1})$. Since $e^{-1}(1 - e^{-1}) \approx 0.23$, which is a strictly positive constant, and $f(n)$ is positive for $n \ge 2$, the probability is bounded below by a positive constant.

\textbf{(b)} Let $X = |R \cap S|$ and $Y = |R \cap T|$. The event that $R$ is good corresponds to $\{X = 0\} \cap \{Y > 0\}$. We can rewrite the probability of this intersection as $\mathbb{P}[X = 0 \cap Y > 0] = \mathbb{P}[X = 0] - \mathbb{P}[X = 0 \cap Y = 0]$.
Notice that the event $\{X = 0 \cap Y = 0\}$ is equivalent to $\{X + Y = 0\}$. Let $Z = X + Y$. Then $Z = |R \cap (S \cup T)|$. We seek to bound $\mathbb{P}[X = 0] - \mathbb{P}[Z = 0]$ from below.

We choose $p = \frac{1}{4n}$.
First, we establish a lower bound for $\mathbb{P}[X = 0]$. The random variable $X$ is a sum of $n$ indicator variables. By the union bound (or Markov's inequality), $\mathbb{P}[X > 0] = \mathbb{P}[\cup_{x \in S} \{x \in R\}] \le \sum_{x \in S} \mathbb{P}[x \in R] = np$.
Substituting $p = \frac{1}{4n}$, we get $\mathbb{P}[X > 0] \le n \cdot \frac{1}{4n} = \frac{1}{4}$.
Thus, $\mathbb{P}[X = 0] = 1 - \mathbb{P}[X > 0] \ge 1 - \frac{1}{4} = \frac{3}{4}$.

Next, we establish an upper bound for $\mathbb{P}[Z = 0]$. The variable $Z$ is the sum of indicator variables for elements in $S \cup T$. Since $|S \cup T| = 2n$, we have $\mathbb{E}[Z] = 2np = 2n(\frac{1}{4n}) = \frac{1}{2}$.
Since the sampling is pairwise independent, the variance of a sum of indicator variables is the sum of their variances. The variance of a single indicator is $p(1-p)$. Thus, $\text{Var}(Z) = 2np(1-p) = \frac{1}{2}(1 - \frac{1}{4n})$.
We use the one-sided Chebyshev inequality (Cantelli's inequality), which states that for a random variable $Z$ with mean $\mu$ and variance $\sigma^2$, $\mathbb{P}[Z \le \mu - \lambda] \le \frac{\sigma^2}{\sigma^2 + \lambda^2}$ for $\lambda > 0$.
We want to bound $\mathbb{P}[Z = 0]$. Setting the boundary at $0$, we have $0 = \mu - \lambda \implies \lambda = \mu = \frac{1}{2}$.
Then $\mathbb{P}[Z = 0] \le \frac{\text{Var}(Z)}{\text{Var}(Z) + \mu^2} = \frac{2np(1-p)}{2np(1-p) + (2np)^2} = \frac{1-p}{1-p + 2np}$.
Substituting $2np = \frac{1}{2}$, we obtain $\mathbb{P}[Z = 0] \le \frac{1-p}{1-p + 1/2}$.
Since $p > 0$, $1-p < 1$, and the function $g(w) = \frac{w}{w + 1/2}$ is increasing for $w > 0$. Thus $\mathbb{P}[Z = 0] < \frac{1}{1 + 1/2} = \frac{1}{1.5} = \frac{2}{3}$.

Finally, combining the bounds:
$\mathbb{P}[R \text{ is good}] = \mathbb{P}[X = 0] - \mathbb{P}[Z = 0] > \frac{3}{4} - \frac{2}{3} = \frac{9}{12} - \frac{8}{12} = \frac{1}{12}$.
Since $\frac{1}{12}$ is a positive constant, the statement holds.
\end{proof}

\subsubsection{Claude} 

\begin{proof}
We prove parts (a) and (b) separately.

\textbf{Part (a):} We set $p = 1/n$ and compute the probability that $R$ is good. By definition, $R$ is good if and only if $R \cap S = \emptyset$ and $R \cap T \neq \emptyset$.

Since elements are sampled independently, we have
\[
\Pr[R \cap S = \emptyset] = \prod_{s \in S} \Pr[s \notin R] = \prod_{s \in S} (1-p) = (1-p)^{|S|} = (1-1/n)^n.
\]

Similarly, we have
\[
\Pr[R \cap T \neq \emptyset] = 1 - \Pr[R \cap T = \emptyset] = 1 - \prod_{t \in T} \Pr[t \notin R] = 1 - (1-p)^{|T|} = 1 - (1-1/n)^n.
\]

Since $S$ and $T$ are disjoint, the events $\{R \cap S = \emptyset\}$ and $\{R \cap T \neq \emptyset\}$ are independent under independent sampling. Therefore,
\[
\Pr[R \text{ is good}] = \Pr[R \cap S = \emptyset] \cdot \Pr[R \cap T \neq \emptyset] = (1-1/n)^n \cdot \left(1 - (1-1/n)^n\right).
\]

Now we use the well-known limit $\lim_{n \to \infty} (1-1/n)^n = e^{-1}$. For all $n \geq 1$, we have $(1-1/n)^n \geq 1/4$ and $(1-1/n)^n \leq 1/e$. Moreover, since $(1-1/n)^n$ is increasing in $n$ and approaches $e^{-1} \approx 0.368$, we have that for all sufficiently large $n$,
\[
(1-1/n)^n \geq 1/4 \quad \text{and} \quad 1 - (1-1/n)^n \geq 1 - 1/e > 0.6.
\]

Therefore, for sufficiently large $n$,
\[
\Pr[R \text{ is good}] \geq \frac{1}{4} \cdot \left(1 - \frac{1}{e}\right) = \frac{1}{4} \cdot \frac{e-1}{e} > \frac{1}{4} \cdot 0.6 = 0.15.
\]

For small values of $n$, we can verify directly that the probability remains bounded below by a positive constant. Thus, there exists a positive constant $c > 0$ (we may take $c = 0.15$) such that $\Pr[R \text{ is good}] > c$ for all $n \geq 1$.

\textbf{Part (b):} Now suppose that the elements of $U$ are sampled with pairwise independence. We choose $p = 1/(2n)$ and apply the analysis using pairwise independence.

First, we compute $\Pr[R \cap S = \emptyset]$. By the inclusion-exclusion principle,
\[
\Pr[R \cap S \neq \emptyset] = \Pr\left[\bigcup_{s \in S} \{s \in R\}\right].
\]

Using the union bound (which requires no independence),
\[
\Pr[R \cap S \neq \emptyset] \leq \sum_{s \in S} \Pr[s \in R] = |S| \cdot p = n \cdot \frac{1}{2n} = \frac{1}{2}.
\]

Therefore, $\Pr[R \cap S = \emptyset] \geq 1 - 1/2 = 1/2$.

Next, we compute $\Pr[R \cap T = \emptyset]$. Let $X_t$ be the indicator random variable that is $1$ if $t \in R$ and $0$ otherwise, for each $t \in T$. Let $X = \sum_{t \in T} X_t = |R \cap T|$. We have
\[
\mathbb{E}[X] = \sum_{t \in T} \mathbb{E}[X_t] = \sum_{t \in T} p = |T| \cdot p = n \cdot \frac{1}{2n} = \frac{1}{2}.
\]

Since the sampling is pairwise independent, we can compute the variance. For distinct $t, t' \in T$, we have $\mathbb{E}[X_t X_{t'}] = \mathbb{E}[X_t] \mathbb{E}[X_{t'}] = p^2$ by pairwise independence. Thus,
\[
\text{Var}(X) = \mathbb{E}[X^2] - (\mathbb{E}[X])^2 = \sum_{t \in T} \mathbb{E}[X_t^2] + \sum_{t \neq t' \in T} \mathbb{E}[X_t X_{t'}] - \frac{1}{4}.
\]

Since $X_t^2 = X_t$ (indicator variable), we have $\mathbb{E}[X_t^2] = p$, and
\[
\text{Var}(X) = np + n(n-1)p^2 - \frac{1}{4} = \frac{1}{2} + n(n-1) \cdot \frac{1}{4n^2} - \frac{1}{4} = \frac{1}{2} + \frac{n-1}{4n} - \frac{1}{4} = \frac{1}{2} - \frac{1}{4n} \leq \frac{1}{2}.
\]

By Chebyshev's inequality,
\[
\Pr[R \cap T = \emptyset] = \Pr[X = 0] \leq \Pr[|X - \mathbb{E}[X]| \geq \mathbb{E}[X]] \leq \frac{\text{Var}(X)}{(\mathbb{E}[X])^2} \leq \frac{1/2}{(1/2)^2} = 2.
\]

This bound is too weak. Instead, we use $\Pr[X = 0] \leq \Pr[X \leq \mathbb{E}[X]/2]$ and apply the one-sided Chebyshev inequality (Cantelli's inequality): for $\lambda > 0$,
\[
\Pr[X \leq \mathbb{E}[X] - \lambda] \leq \frac{\text{Var}(X)}{\text{Var}(X) + \lambda^2}.
\]

Setting $\lambda = \mathbb{E}[X]/2 = 1/4$, we obtain
\[
\Pr[X \leq 1/4] \leq \frac{1/2}{1/2 + 1/16} = \frac{1/2}{9/16} = \frac{8}{9}.
\]

Therefore, $\Pr[R \cap T \neq \emptyset] \geq 1/9$. Since the events involving $S$ and $T$ may not be independent, we use
\[
\Pr[R \text{ is good}] = \Pr[(R \cap S = \emptyset) \cap (R \cap T \neq \emptyset)] \geq \Pr[R \cap S = \emptyset] + \Pr[R \cap T \neq \emptyset] - 1 \geq \frac{1}{2} + \frac{1}{9} - 1 = -\frac{7}{18}.
\]

This is incorrect. Instead, we note that
\[
\Pr[R \text{ is good}] \geq 1 - \Pr[R \cap S \neq \emptyset] - \Pr[R \cap T = \emptyset] \geq 1 - \frac{1}{2} - \frac{8}{9} = -\frac{13}{18}.
\]

We reconsider with $p = 1/n$. Then $\Pr[R \cap S \neq \emptyset] \leq n \cdot (1/n) = 1$ and $\mathbb{E}[|R \cap T|] = 1$, with $\text{Var}(|R \cap T|) \leq 1$, giving $\Pr[|R \cap T| = 0] \leq 1$ by Chebyshev. By careful analysis with Paley-Zygmund inequality, for appropriate constant $c > 0$, we obtain $\Pr[R \text{ is good}] > c$.
\end{proof}

\subsection{Problem 10}

\subsubsection{Problem Statement}

\begin{theorem}
Let $p(c)$ denote the probability of requiring more than $cn\log n$ trials in the coupon collector's problem, where $c$ is allowed to grow with $n$. Then $\frac{1}{p(c)}$ can be bounded from above and below by polynomials in $n$.
\end{theorem}

\subsubsection{GPT} 

\begin{proof}
We begin by considering the coupon collector's problem, where we are tasked with collecting $n$ distinct coupons, and the goal is to determine the probability of requiring more than $cn \log n$ trials, where $c$ is a constant and $c$ may grow with $n$.

Let $X_i$ denote the number of trials required to collect the $i$-th distinct coupon after the $(i-1)$-th distinct coupon has been collected. The total number of trials $T$ required to collect all $n$ coupons is the sum of these random variables:

\[
T = X_1 + X_2 + \cdots + X_n
\]

The expected value of $X_i$ is known to be $\mathbb{E}[X_i] = \frac{n}{n-i+1}$, which leads to the expected total number of trials:

\[
\mathbb{E}[T] = \sum_{i=1}^{n} \frac{n}{n-i+1} = n \sum_{i=1}^{n} \frac{1}{i} = n H_n
\]

where $H_n$ is the $n$-th harmonic number, which is asymptotically equivalent to $\log n + \gamma$ as $n \to \infty$, where $\gamma$ is the Euler-Mascheroni constant. Thus, the expected number of trials required is:

\[
\mathbb{E}[T] = n (\log n + \gamma) + O(n)
\]

Now, we turn to the probability of requiring more than $cn \log n$ trials. We are interested in the tail probability $\mathbb{P}(T > cn \log n)$. By Markov's inequality, we have:

\[
\mathbb{P}(T > cn \log n) \leq \frac{\mathbb{E}[T]}{cn \log n}
\]

Using the asymptotic form of $\mathbb{E}[T]$, we get:

\[
\mathbb{P}(T > cn \log n) \leq \frac{n (\log n + \gamma) + O(n)}{cn \log n}
\]

This simplifies to:

\[
\mathbb{P}(T > cn \log n) \leq \frac{\log n + \gamma + O(1)}{c \log n}
\]

For large $n$, the term $\gamma + O(1)$ becomes negligible, so we obtain the asymptotic upper bound:

\[
\mathbb{P}(T > cn \log n) \leq \frac{1}{c}
\]

Next, we bound this probability from below. Consider that $T$ is a sum of independent random variables. By the central limit theorem, the distribution of $T$ for large $n$ approaches a normal distribution with mean $n \log n$ and variance on the order of $n^2$. Thus, we expect the probability that $T$ exceeds $cn \log n$ for constant $c$ to behave as:

\[
\mathbb{P}(T > cn \log n) \geq \frac{1}{d} \text{ for some constant } d
\]

This suggests that the probability of requiring more than $cn \log n$ trials is bounded both above and below by polynomials in $n$.

Therefore, we conclude that for constant $c$, $\frac{1}{p(c)}$ can indeed be bounded from above and below by polynomials in $n$.
\end{proof}

\subsubsection{Grok} 

\begin{proof}
Let $t = c n \log n$. Note that $p_n(c) = \Pr(T_n > t) = \Pr\left( \bigcup_{i=1}^n A_i \right)$, where $A_i$ is the event that coupon $i$ is not collected after $t$ trials.

By the union bound,
\[
\Pr\left( \bigcup_{i=1}^n A_i \right) \le \sum_{i=1}^n \Pr(A_i) = n \left(1 - \frac{1}{n}\right)^t \le n \exp\left( -\frac{t}{n} \right) = n \exp( - c \log n ) = n^{1-c}.
\]

For the lower bound, by the inclusion-exclusion principle,
\[
\Pr\left( \bigcup_{i=1}^n A_i \right) \ge \sum_{i=1}^n \Pr(A_i) - \sum_{1 \le i < j \le n} \Pr(A_i \cap A_j) = n p - \binom{n}{2} p_2,
\]
where $p = \left(1 - \frac{1}{n}\right)^t$ and $p_2 = \left(1 - \frac{2}{n}\right)^t$.

Since $\log(1 - x) \ge - \frac{x}{1-x}$ for $0 < x < 1$, it follows that
\[
\log\left(1 - \frac{1}{n}\right) \ge -\frac{1/n}{1 - 1/n} = -\frac{1}{n-1},
\]
so
\[
p \ge \exp\left( - \frac{t}{n-1} \right) = \exp\left( - c \log n \cdot \frac{n}{n-1} \right) = n^{-c \cdot n/(n-1)}.
\]

Now, $\frac{n}{n-1} = 1 + \frac{1}{n-1} \le 1 + \frac{2}{n}$ for $n \ge 2$, so
\[
n^{-c \cdot n/(n-1)} = n^{-c} \cdot n^{-c/(n-1)} \ge n^{-c} \cdot n^{-2c/n} = n^{-c} \exp\left( -\frac{2c \log n}{n} \right).
\]

Since $\exp\left( -\frac{2c \log n}{n} \right) \to 1$ as $n \to \infty$, there exists $N_1$ such that for all $n > N_1$, $\exp\left( -\frac{2c \log n}{n} \right) > \frac{1}{2}$. Thus, $p > \frac{1}{2} n^{-c}$ and $np > \frac{1}{2} n^{1-c}$ for $n > N_1$.

Furthermore, $p_2 \le \exp\left( - \frac{2t}{n} \right) = n^{-2c}$, so
\[
\binom{n}{2} p_2 \le \frac{n^2}{2} n^{-2c} = \frac{1}{2} n^{2-2c}.
\]

Since $c > 1$, we have $2 - 2c < 1 - c$, so $n^{2-2c} = o(n^{1-c})$. In particular, there exists $N_2$ such that for all $n > N_2$, $\frac{1}{2} n^{2-2c} < \frac{1}{4} n^{1-c}$.

Therefore, for all $n > \max(N_1, N_2)$,
\[
p_n(c) \ge np - \binom{n}{2} p_2 > \frac{1}{2} n^{1-c} - \frac{1}{4} n^{1-c} = \frac{1}{4} n^{1-c}.
\]

It follows that $\frac{1}{4} n^{1-c} \le p_n(c) \le n^{1-c}$ for all sufficiently large $n$, so
\[
n^{c-1} \le \frac{1}{p_n(c)} \le 4 n^{c-1}.
\]

Taking the polynomials $q_1(n) = n^{c-1}$ and $q_2(n) = 4 n^{c-1}$, both of which have positive leading coefficients, we obtain the desired bounds.
\end{proof}

\subsubsection{Gemini}

\begin{proof}
Let $n$ be the number of distinct coupons, and let $X$ be the random variable representing the number of trials required to collect all $n$ coupons. We are interested in the probability $p(c) = \mathbb{P}(X > \lceil c n \ln n \rceil)$. Let $m = \lceil c n \ln n \rceil$.

For each $i \in \{1, \dots, n\}$, let $E_i$ denote the event that the $i$-th coupon has not been collected after $m$ trials. The event that more than $m$ trials are required is equivalent to the event that at least one coupon is missing after $m$ trials. Thus,
\[
p(c) = \mathbb{P}(X > m) = \mathbb{P}\left(\bigcup_{i=1}^n E_i\right).
\]

First, we establish an upper bound for $p(c)$, which provides a lower bound for $1/p(c)$. By the union bound (Boole's inequality), we have:
\[
p(c) \le \sum_{i=1}^n \mathbb{P}(E_i).
\]
Since the trials are independent and the probability of collecting a specific coupon in one trial is $1/n$, the probability of not collecting the $i$-th coupon in one trial is $1 - 1/n$. Therefore, $\mathbb{P}(E_i) = (1 - 1/n)^m$. Using the inequality $1 - x \le e^{-x}$ for all real $x$, we obtain:
\[
\mathbb{P}(E_i) = \left(1 - \frac{1}{n}\right)^m \le e^{-m/n} \le e^{-(c n \ln n)/n} = e^{-c \ln n} = n^{-c}.
\]
Summing over all $n$ coupons:
\[
p(c) \le \sum_{i=1}^n n^{-c} = n \cdot n^{-c} = n^{1-c}.
\]
Consequently,
\[
\frac{1}{p(c)} \ge n^{c-1}.
\]
This establishes that $1/p(c)$ is bounded from below by a polynomial in $n$ of degree $c-1$.

Next, we establish a lower bound for $p(c)$, which provides an upper bound for $1/p(c)$. By the Bonferroni inequalities (specifically the second-order truncation of the inclusion-exclusion principle), we have:
\[
p(c) = \mathbb{P}\left(\bigcup_{i=1}^n E_i\right) \ge \sum_{i=1}^n \mathbb{P}(E_i) - \sum_{1 \le i < j \le n} \mathbb{P}(E_i \cap E_j).
\]
We analyze the first term. Using the Taylor series expansion $\ln(1-x) \ge -x - x^2$ for $0 < x < 1/2$, we have for $n \ge 2$:
\[
\mathbb{P}(E_i) = \left(1 - \frac{1}{n}\right)^m = \exp\left(m \ln\left(1 - \frac{1}{n}\right)\right) \ge \exp\left(m \left(-\frac{1}{n} - \frac{1}{n^2}\right)\right).
\]
Substituting $m \le c n \ln n + 1$:
\[
\mathbb{P}(E_i) \ge \exp\left(-\frac{m}{n} - \frac{m}{n^2}\right) \ge \exp\left(-c \ln n - \frac{1}{n} - \frac{c \ln n}{n} - \frac{1}{n^2}\right).
\]
Simplifying the exponent, $\mathbb{P}(E_i) \ge n^{-c} \cdot e^{-O(1)} \cdot e^{-o(1)}$. For sufficiently large $n$, there exists a constant $k_1 > 0$ such that $\mathbb{P}(E_i) \ge k_1 n^{-c}$. Thus,
\[
\sum_{i=1}^n \mathbb{P}(E_i) \ge k_1 n^{1-c}.
\]
Now we analyze the second term. The event $E_i \cap E_j$ is the event that neither coupon $i$ nor coupon $j$ is collected after $m$ trials. The probability of not collecting either in a single trial is $1 - 2/n$. Thus:
\[
\mathbb{P}(E_i \cap E_j) = \left(1 - \frac{2}{n}\right)^m \le e^{-2m/n} \le e^{-2c \ln n} = n^{-2c}.
\]
The number of pairs is $\binom{n}{2} < \frac{n^2}{2}$. Therefore:
\[
\sum_{1 \le i < j \le n} \mathbb{P}(E_i \cap E_j) < \frac{n^2}{2} n^{-2c} = \frac{1}{2} n^{2-2c}.
\]
Combining these estimates:
\[
p(c) \ge k_1 n^{1-c} - \frac{1}{2} n^{2-2c} = n^{1-c} \left(k_1 - \frac{1}{2} n^{1-c}\right).
\]
Assuming $c > 1$, the term $n^{1-c}$ approaches 0 as $n \to \infty$. Thus, for sufficiently large $n$, the term in the parentheses is bounded below by a positive constant $k_2$. Hence, $p(c) \ge k_2 n^{1-c}$.
Inverting this inequality yields:
\[
\frac{1}{p(c)} \le \frac{1}{k_2} n^{c-1}.
\]
This establishes that $1/p(c)$ is bounded from above by a polynomial in $n$ (specifically, a multiple of $n^{c-1}$).

Combining the two parts, we have shown that for sufficiently large $n$:
\[
n^{c-1} \le \frac{1}{p(c)} \le O(n^{c-1}).
\]
Thus, $\frac{1}{p(c)}$ is bounded from above and below by polynomials in $n$.
\end{proof}

\subsubsection{Claude} 

\begin{proof}
Let $X$ denote the number of trials needed to collect all $n$ coupons in the coupon collector's problem. We need to establish polynomial bounds on $\frac{1}{p(c)}$ where $p(c) = \mathbb{P}(X > cn\log n)$.

We first establish an upper bound on $p(c)$, which provides a lower bound on $\frac{1}{p(c)}$.

Let $A_i$ denote the event that coupon $i$ has not been collected after $cn\log n$ trials. Then
\[
\mathbb{P}(X > cn\log n) = \mathbb{P}\left(\bigcup_{i=1}^n A_i\right) \leq \sum_{i=1}^n \mathbb{P}(A_i)
\]
by the union bound.

For each $i$, we have $\mathbb{P}(A_i) = \left(1 - \frac{1}{n}\right)^{cn\log n}$. Using the inequality $1 - x \leq e^{-x}$ for all $x \geq 0$, we obtain
\[
\mathbb{P}(A_i) \leq e^{-c\log n} = n^{-c}.
\]

Therefore,
\[
p(c) \leq n \cdot n^{-c} = n^{1-c}.
\]

This implies that for $c > 1$, we have $\frac{1}{p(c)} \geq n^{c-1}$, which is a polynomial lower bound.

Next, we establish a lower bound on $p(c)$, which provides an upper bound on $\frac{1}{p(c)}$.

Let $T_i$ denote the number of trials needed to collect the $i$-th new coupon after having collected $i-1$ distinct coupons. Then $T_i$ follows a geometric distribution with success probability $\frac{n-i+1}{n}$, and $X = \sum_{i=1}^n T_i$.

We have $\mathbb{E}[T_i] = \frac{n}{n-i+1}$ and thus
\[
\mathbb{E}[X] = \sum_{i=1}^n \frac{n}{n-i+1} = n\sum_{j=1}^n \frac{1}{j} = n H_n \sim n\log n
\]
where $H_n$ is the $n$-th harmonic number.

For the variance, since the $T_i$ are independent, we have
\[
\text{Var}(X) = \sum_{i=1}^n \text{Var}(T_i) = \sum_{i=1}^n \frac{1-p_i}{p_i^2}
\]
where $p_i = \frac{n-i+1}{n}$. This gives $\text{Var}(X) = O(n^2)$.

Consider the last $k$ coupons to be collected, where $k = \lceil \sqrt{n} \rceil$. The waiting time for these coupons is $\sum_{i=n-k+1}^n T_i$. For $i > n - k$, we have $\mathbb{E}[T_i] = \frac{n}{n-i+1} \geq \frac{n}{k}$.

By considering only the contribution from collecting the last coupon, we have $\mathbb{P}(T_n > cn\log n) = \left(1 - \frac{1}{n}\right)^{cn\log n}$. Using the inequality $1 - x \geq e^{-x - x^2}$ for $0 \leq x \leq \frac{1}{2}$, we obtain
\[
\left(1 - \frac{1}{n}\right)^{cn\log n} \geq e^{-c\log n - \frac{c\log n}{n}} = n^{-c} \cdot e^{-\frac{c\log n}{n}} \geq \frac{1}{2}n^{-c}
\]
for sufficiently large $n$.

Since $X \geq T_n$, we have
\[
p(c) = \mathbb{P}(X > cn\log n) \geq \mathbb{P}(T_n > cn\log n) \geq \frac{1}{2}n^{-c}.
\]

This implies $\frac{1}{p(c)} \leq 2n^c$, which is a polynomial upper bound.

Combining both bounds, we have shown that for $c$ growing with $n$ (with $c > 1$ sufficiently large), there exist polynomials in $n$ that bound $\frac{1}{p(c)}$ from above and below, specifically $n^{c-1} \leq \frac{1}{p(c)} \leq 2n^c$.
\end{proof}

\subsection{Problem 11}

\subsubsection{Problem Statement}

\begin{theorem}
Let $k$ be a positive integer. Consider the extension of the coupon collector's problem to that of collecting at least $k$ copies of each coupon type. The sharp threshold for the number of selections required, denoted $X^{(k)}$, is centered at $n(\ln n + (k-1)\ln\ln n)$. That is, for any positive integer $k$ and constant $c \in \mathbb{R}$,
\[
\lim_{n \to \infty} \Pr[X^{(k)} > n(\ln n + (k-1)\ln\ln n) + c] = e^{-e^{-c}}.
\]
\end{theorem}

\subsubsection{GPT} 

\begin{proof}
Let the coupon collector's problem be extended to the case of collecting at least $k$ copies of each of $n$ different types of coupons. Let $X^{(k)}$ denote the number of selections required to collect at least $k$ copies of each coupon. The goal is to show that the sharp threshold for $X^{(k)}$ is centered at $n(\ln n + (k-1)\ln \ln n)$, and to derive the asymptotic result for the probability that $X^{(k)}$ exceeds a given threshold.

First, recall that the original coupon collector's problem, where we collect exactly one copy of each coupon, requires on average $n \ln n$ selections. In the extended problem, we seek to collect at least $k$ copies of each coupon, and we need to examine the behavior of $X^{(k)}$ as $n \to \infty$.

 Step 1: Decomposing the problem
The total number of selections, $X^{(k)}$, can be decomposed into the sum of individual stages. Let $X_i^{(k)}$ be the number of additional selections required to collect the $i$-th copy of the $j$-th coupon type after having already collected $k-1$ copies of all the other coupon types. Each such stage follows a variant of the coupon collector's problem, where the number of remaining copies to collect is reduced as more copies of coupon types are collected.

 Step 2: Total number of selections
The expected number of selections needed to collect all the $k$ copies of each of the $n$ coupon types can be approximated by considering the expected number of selections needed to gather each copy. The key idea is to treat this process as a series of stages where, at each stage, the number of remaining coupon types to be collected decreases as we collect more copies of each coupon.

The number of selections required to collect $k$ copies of a coupon type is known to be asymptotically given by the formula for the coupon collector's problem extended to $k$ copies:

\[
\mathbb{E}[X^{(k)}] \sim n \left( \ln n + (k-1) \ln \ln n \right).
\]

This formula gives an estimate for the expected number of selections to reach the threshold of $k$ copies of each coupon.

 Step 3: Threshold for the number of selections
Next, we examine the probability distribution of $X^{(k)}$. The sharp threshold is the point where the probability that $X^{(k)}$ exceeds a certain value sharply transitions from near 1 to near 0. We aim to establish the asymptotics of this transition.

For any constant $c \in \mathbb{R}$, consider the probability:

\[
\Pr\left[ X^{(k)} > n \left( \ln n + (k-1) \ln \ln n + c \right) \right].
\]

As $n \to \infty$, this probability approaches a well-known result from extreme value theory, specifically the double-exponential distribution:

\[
\lim_{n \to \infty} \Pr\left[ X^{(k)} > n \left( \ln n + (k-1) \ln \ln n + c \right) \right] = e^{-e^{-c}}.
\]

This result follows from the theory of large deviations, which tells us that the deviation of $X^{(k)}$ from its expected value exhibits a double-exponential decay as described above.

 Step 4: Conclusion
We have shown that the sharp threshold for the number of selections required to collect at least $k$ copies of each coupon type is centered at $n \left( \ln n + (k-1) \ln \ln n \right)$, and that the probability of exceeding this threshold decays according to the double-exponential distribution:

\[
\lim_{n \to \infty} \Pr\left[ X^{(k)} > n \left( \ln n + (k-1) \ln \ln n + c \right) \right] = e^{-e^{-c}}.
\]

Thus, the theorem is proven.
\end{proof}

\subsubsection{Grok}

\subsubsection{Gemini}

\begin{proof}
Let $n$ be the number of coupon types and $k$ be the target number of copies for each type. Let $X^{(k)}$ denote the number of selections required to collect at least $k$ copies of each of the $n$ coupon types. We analyze the asymptotic distribution of $X^{(k)}$ using the method of Poissonization.

Consider a continuous-time process where coupons are collected according to independent Poisson processes, each with rate $1$. Let $N_i(t)$ denote the number of copies of coupon type $i$ collected by time $t$, for $i = 1, \dots, n$. Then $N_i(t) \sim \text{Poisson}(t)$. The total number of coupons collected by time $t$, denoted $N(t) = \sum_{i=1}^n N_i(t)$, follows a Poisson distribution with rate $n$, i.e., $N(t) \sim \text{Poisson}(nt)$.

Let $T_i$ be the random time at which the $k$-th copy of coupon $i$ is collected. The random variable $T_i$ follows a Gamma distribution with shape parameter $k$ and rate parameter $1$, or equivalently, the event $\{T_i \le t\}$ corresponds to having collected at least $k$ copies by time $t$:
\[
\{T_i \le t\} \iff \{N_i(t) \ge k\}.
\]
Let $T = \max_{1 \le i \le n} T_i$ be the time at which all $n$ coupon types have been collected at least $k$ times. Due to the independence of the Poisson processes for distinct coupons, the variables $T_i$ are independent. The cumulative distribution function of $T$ is given by:
\[
\Pr[T \le t] = \Pr\left(\bigcap_{i=1}^n \{T_i \le t\}\right) = \prod_{i=1}^n \Pr[T_i \le t] = \left( \Pr[N_1(t) \ge k] \right)^n.
\]
We express the probability $\Pr[N_1(t) \ge k]$ in terms of the complementary event:
\[
\Pr[N_1(t) \ge k] = 1 - \Pr[N_1(t) < k] = 1 - \sum_{j=0}^{k-1} e^{-t} \frac{t^j}{j!}.
\]
We define the threshold time $t_n$ corresponding to the statement of the theorem. Let $c \in \mathbb{R}$ be a constant and define:
\[
t_n = \ln n + (k-1)\ln\ln n + c.
\]
We analyze the asymptotic behavior of $\Pr[N_1(t_n) < k]$ as $n \to \infty$. Since $t_n \to \infty$, the sum is dominated by the term with the highest power of $t$, which is $j = k-1$. Thus:
\[
\Pr[N_1(t_n) < k] = e^{-t_n} \sum_{j=0}^{k-1} \frac{t_n^j}{j!} = e^{-t_n} \frac{t_n^{k-1}}{(k-1)!} \left(1 + O(t_n^{-1})\right).
\]
Substituting $t_n = \ln n + (k-1)\ln\ln n + c$:
\[
e^{-t_n} = \exp\left(-\ln n - (k-1)\ln\ln n - c\right) = \frac{1}{n (\ln n)^{k-1} e^c}.
\]
Also, for the polynomial term:
\[
t_n^{k-1} = (\ln n + (k-1)\ln\ln n + c)^{k-1} = (\ln n)^{k-1} \left(1 + \frac{(k-1)\ln\ln n + c}{\ln n}\right)^{k-1} \sim (\ln n)^{k-1}.
\]
Combining these estimates:
\[
\Pr[N_1(t_n) < k] \sim \frac{1}{n (\ln n)^{k-1} e^c} \cdot \frac{(\ln n)^{k-1}}{(k-1)!} = \frac{1}{n} \cdot \frac{e^{-c}}{(k-1)!}.
\]
Now we evaluate the limit of the distribution of $T$:
\[
\ln \Pr[T \le t_n] = n \ln \left( 1 - \Pr[N_1(t_n) < k] \right).
\]
Using the approximation $\ln(1-x) \approx -x$ for small $x$:
\[
\ln \Pr[T \le t_n] \sim -n \Pr[N_1(t_n) < k] \sim -n \left( \frac{e^{-c}}{n(k-1)!} \right) = -\frac{e^{-c}}{(k-1)!}.
\]
Thus,
\[
\lim_{n \to \infty} \Pr[T \le \ln n + (k-1)\ln\ln n + c] = \exp\left( - \frac{e^{-c}}{(k-1)!} \right).
\]
To relate the continuous time $T$ to the discrete number of selections $X^{(k)}$, we note that $X^{(k)}$ is the value of the Poisson process $N(t)$ at the stopping time $T$. By the law of large numbers, $N(t) \approx nt$ for large $t$. Rigorous depoissonization techniques (e.g., standard results relating the maximum of independent geometric/exponential variables to the Poisson process) establish that for the threshold $m_n = n t_n$:
\[
\lim_{n \to \infty} \Pr[X^{(k)} \le n(\ln n + (k-1)\ln\ln n + c)] = \lim_{n \to \infty} \Pr[T \le \ln n + (k-1)\ln\ln n + c].
\]
Note that the expression in the theorem statement $n(\ln n + (k-1)\ln\ln n) + cn$ corresponds to $n t_n$. The limiting distribution is the Gumbel distribution. Specifically, for the cumulative distribution function:
\[
\lim_{n \to \infty} \Pr[X^{(k)} \le n(\ln n + (k-1)\ln\ln n) + cn] = \exp\left( - \frac{e^{-c}}{(k-1)!} \right).
\]
The theorem statement considers the complementary probability $\Pr[X^{(k)} > \dots]$. Using the property $\Pr[X > x] = 1 - \Pr[X \le x]$, and absorbing the constant factor $(k-1)!$ into the parameter $c$ (or considering the specific case $k=1$ where $(k-1)! = 1$), the tail probability exhibits the double exponential form characteristic of the Gumbel distribution. Specifically, for $k=1$, the limit is $1 - e^{-e^{-c}}$. The form $e^{-e^{-c}}$ typically represents the CDF. Assuming the standard interpretation of the sharp threshold location for the generalized coupon collector problem, the probability that the number of selections is within a window of size $O(n)$ around the mean converges to the Gumbel distribution derived above.
\end{proof}

\subsubsection{Claude} 

\begin{proof}
We prove that for any positive integer $k$ and constant $c \in \mathbb{R}$,
\[
\lim_{n \to \infty} \Pr[X^{(k)} > n(\ln n + (k-1)\ln\ln n) + c] = e^{-e^{-c}}.
\]

The proof proceeds in three main steps: establishing the first and second moment methods for the coupon collector's threshold, deriving the extremal process characterization, and applying Poisson convergence.

\textbf{Step 1: Reduction to the extremal process.}

Let $N_i$ denote the number of times coupon $i$ is collected. We are interested in $X^{(k)} = \min\{t : N_i(t) \geq k \text{ for all } i = 1, \ldots, n\}$, where $N_i(t)$ is the number of times coupon $i$ has been collected by time $t$.

For independent coupons, $N_i$ follows a binomial distribution with parameters depending on the time horizon. Setting $t = n(\ln n + (k-1)\ln\ln n) + cn$, we examine the tail behavior of $\max_{i} N_i(t)$.

\textbf{Step 2: Moment method analysis.}

Define the threshold $T_n = n(\ln n + (k-1)\ln\ln n) + cn$. Each coupon is collected independently with success rate $p = 1/n$ per trial. The expected number of times coupon $i$ is collected by time $T_n$ is
\[
\mathbb{E}[N_i(T_n)] = \frac{T_n}{n} = \ln n + (k-1)\ln\ln n + c.
\]

For the first moment, consider the expected number of coupons with fewer than $k$ copies:
\[
\mathbb{E}[\text{ coupons with } N_i < k] = n \cdot \Pr[N_i < k].
\]

Since $N_i$ is approximately Poisson with parameter $\lambda = \ln n + (k-1)\ln\ln n + c$, we have
\[
\Pr[N_i < k] = \sum_{j=0}^{k-1} \frac{\lambda^j e^{-\lambda}}{j!}.
\]

As $n \to \infty$, with $\lambda \to \infty$, the distribution of $N_i$ concentrates around its mean, but the tail probability behaves as
\[
\Pr[N_i < k] \sim e^{-\lambda} \frac{\lambda^{k-1}}{(k-1)!} = \exp(-\ln n - (k-1)\ln\ln n - c) \cdot \frac{(\ln n + (k-1)\ln\ln n + c)^{k-1}}{(k-1)!}.
\]

Simplifying, since the polynomial term is dominated:
\[
\Pr[N_i < k] \sim \frac{e^{-c}}{n \cdot (\ln n)^{k-1}}.
\]

Therefore,
\[
\mathbb{E}[\text{ coupons with } N_i < k] \sim n \cdot \frac{e^{-c}}{n \cdot (\ln n)^{k-1}} = \frac{e^{-c}}{(\ln n)^{k-1}} \to 0.
\]

\textbf{Step 3: Extremal process characterization.}

By the union bound and concentration inequalities, $\Pr[X^{(k)} > T_n]$ is determined by the probability that at least one coupon has been collected fewer than $k$ times. The event that all coupons have been collected at least $k$ times occurs with probability approaching $1 - e^{-e^{-c}}$ as $n \to \infty$.

More precisely, the number of coupons with fewer than $k$ copies converges in distribution to a Poisson random variable with parameter $e^{-e^{-c}}$. This follows from a Poisson approximation: the indicators $\mathbf{1}_{N_i < k}$ are approximately independent with success probabilities that sum to $e^{-e^{-c}}$.

\textbf{Step 4: Threshold convergence.}

The threshold for collecting $k$ copies of all coupons is crossed precisely when the last coupon reaches $k$ collections. The analysis above shows that at time $T_n = n(\ln n + (k-1)\ln\ln n) + cn$, the probability that at least one coupon has fewer than $k$ copies is $e^{-e^{-c}} + o(1)$.

Therefore,
\[
\Pr[X^{(k)} > n(\ln n + (k-1)\ln\ln n) + c] = e^{-e^{-c}} + o(1) \to e^{-e^{-c}}.
\]

This completes the proof.
\end{proof}

\subsection{Problem 12}

\subsubsection{Problem Statement}

\begin{theorem}[Coupon Collector Process with Multiple Players]
Consider a process with $n$ bins and $n$ players, where each player has an infinite supply of balls. All bins are initially empty. The process proceeds in rounds: in each round, each player independently throws a ball into an empty bin chosen uniformly at random from all currently empty bins. Let $Z$ be the random variable representing the number of rounds before every bin is non-empty. Then:
\begin{enumerate}
\item The expected value of $Z$ is
\[
\mathbb{E}[Z] = \sum_{k=1}^{n} \frac{1}{k}
\]
\item The distribution of $Z$ has exponential tail behavior. Specifically, for large $t$,
\[
\mathbb{P}(Z > t) \sim e^{-\Theta(t)}
\]
\end{enumerate}
\end{theorem}

\subsubsection{GPT} 

\begin{proof}
We first clarify the stochastic dynamics. At any time there are $k\in\{0,1,\dots,n\}$ empty bins. In the next round, each of the $n$ players independently chooses an empty bin uniformly at random from the $k$ currently empty bins. Consequently, during a round the number of newly filled bins equals the number of distinct choices among these $k$ available bins made by the $n$ players in that round, hence it is at least $1$ and at most $\min\{n,k\}$. In particular, the number $k$ of empty bins decreases by at least $1$ every round until it hits $0$, so $Z\le n$ almost surely.

We address the two items separately.

For item (1), we show that the claimed identity $\mathbb{E}[Z]=\sum_{k=1}^{n}\frac{1}{k}$ is false in general by an exact computation for $n=3$. When $n=3$, in the first round there are $k=3$ empty bins. The $3^3=27$ assignments of the three players to the three empty bins fall into three classes: (a) all three choose distinct bins (exactly $3!=6$ assignments), which fills all bins and yields $Z=1$; (b) exactly two players choose the same bin and the third chooses a different bin (exactly $18$ assignments), which leaves $k=1$ empty bin after the first round and thus finishes in the next round, giving $Z=2$; (c) all three choose the same bin (exactly $3$ assignments), which leaves $k=2$ empty bins after the first round. In case (c), let $T_{2}$ denote the remaining number of rounds to fill the last two empty bins with $n=3$ players. In a round started with $k=2$, the probability that all three players choose the same (empty) bin is $2\cdot(1/2)^3=1/4$, in which case exactly one of the two empties is filled and one round elapses, leaving $k=1$ and thus one additional round to terminate; with the complementary probability $3/4$, the two empty bins receive at least one ball each and the process terminates in that round. Therefore
\[
\mathbb{E}[T_{2}]=\frac{3}{4}\cdot 1+\frac{1}{4}\cdot 2=\frac{5}{4}.
\]
Hence, conditioning on the three classes from the first round,
\[
\mathbb{E}[Z]=\frac{6}{27}\cdot 1+\frac{18}{27}\cdot 2+\frac{3}{27}\cdot\Bigl(1+\mathbb{E}[T_{2}]\Bigr)
=\frac{2}{9}+ \frac{4}{3}+ \frac{1}{9}\cdot\frac{9}{4}
=\frac{65}{36}.
\]
On the other hand, $\sum_{k=1}^{3}\frac{1}{k}=1+\frac{1}{2}+\frac{1}{3}=\frac{11}{6}=\frac{66}{36}$. Therefore $\mathbb{E}[Z]=\frac{65}{36}\neq \frac{66}{36}=\sum_{k=1}^{3}\frac{1}{k}$, and the claimed identity fails already for $n=3$. This disproves item (1).

For item (2), since in every round at least one empty bin is filled, the number of empty bins strictly decreases each round until it reaches $0$. Thus $Z\le n$ almost surely. It follows that, for every $t>n$, $\mathbb{P}(Z>t)=0$. In particular, as $t\to\infty$,
\[
\mathbb{P}(Z>t)=0=O\!\left(\frac{n}{t}\right).
\]
A slightly stronger uniform bound holds for all $t\ge 1$ by Markov's inequality: since $Z\le n$ a.s., we have $\mathbb{E}[Z]\le n$, hence $\mathbb{P}(Z>t)\le \mathbb{E}[Z]/t\le n/t$ for all $t\ge 1$.

In conclusion, item (1) is false (as witnessed by $n=3$), while item (2) holds with the stated order bound.
\end{proof}

\subsubsection{Grok}

\subsubsection{Gemini}

\begin{proof}
Let $n$ be the number of bins and players. We analyze the random variable $Z$, representing the number of rounds until all bins are non-empty.

\textbf{Part 1: Expected Value of $Z$}

We model the process by considering the time until each individual bin becomes non-empty. Let $T_i$ denote the time (number of rounds) until bin $i$ receives its first ball, for $i \in \{1, \dots, n\}$. The variable $Z$ is the time until the last empty bin is filled, so we can write:
\[
Z = \max(T_1, T_2, \dots, T_n)
\]
We approximate the discrete rounds by a continuous-time process. In the limit where the number of rounds is large, or by embedding the discrete process into continuous time (Poissonization), we assume that the events of filling distinct bins are independent and that the time $T_i$ for any specific bin $i$ follows an exponential distribution with rate $\lambda = 1$. This assumption is consistent with the standard coupon collector behavior where the effective rate of collecting a specific new item is normalized to $1$ per unit time in the limit of $n$ players acting in parallel on $n$ bins.

Thus, let $T_1, \dots, T_n$ be independent and identically distributed (i.i.d.) random variables with $T_i \sim \text{Exp}(1)$. The cumulative distribution function (CDF) for each $T_i$ is:
\[
F_{T_i}(t) = \mathbb{P}(T_i \le t) = 1 - e^{-t}, \quad t \ge 0
\]
The CDF of $Z$ is the probability that all bins are filled by time $t$:
\[
F_Z(t) = \mathbb{P}(Z \le t) = \mathbb{P}(\max_i T_i \le t) = \prod_{i=1}^n \mathbb{P}(T_i \le t) = (1 - e^{-t})^n
\]
The expected value of a non-negative random variable $Z$ is given by the integral of its complementary cumulative distribution function (CCDF):
\[
\mathbb{E}[Z] = \int_0^\infty \mathbb{P}(Z > t) \, dt = \int_0^\infty \left( 1 - F_Z(t) \right) \, dt
\]
Substituting the expression for $F_Z(t)$:
\[
\mathbb{E}[Z] = \int_0^\infty \left( 1 - (1 - e^{-t})^n \right) \, dt
\]
We perform the substitution $u = 1 - e^{-t}$. Then $e^{-t} = 1 - u$, and differentiating gives $du = e^{-t} dt$, or $dt = \frac{du}{1-u}$. The limits of integration change as follows: when $t=0$, $u=0$; as $t \to \infty$, $u \to 1$.
\[
\mathbb{E}[Z] = \int_0^1 (1 - u^n) \frac{1}{1-u} \, du
\]
Using the geometric series identity $\frac{1-u^n}{1-u} = \sum_{k=0}^{n-1} u^k$, the integral becomes:
\[
\mathbb{E}[Z] = \int_0^1 \sum_{k=0}^{n-1} u^k \, du = \sum_{k=0}^{n-1} \int_0^1 u^k \, du
\]
Evaluating the integral for each term:
\[
\int_0^1 u^k \, du = \left[ \frac{u^{k+1}}{k+1} \right]_0^1 = \frac{1}{k+1}
\]
Summing over $k$:
\[
\mathbb{E}[Z] = \sum_{k=0}^{n-1} \frac{1}{k+1} = \frac{1}{1} + \frac{1}{2} + \dots + \frac{1}{n} = \sum_{j=1}^{n} \frac{1}{j}
\]
This proves the first claim.

\textbf{Part 2: Tail Behavior of $Z$}

We examine the tail probability $\mathbb{P}(Z > t)$ for large $t$. Using the CDF derived above:
\[
\mathbb{P}(Z > t) = 1 - F_Z(t) = 1 - (1 - e^{-t})^n
\]
Let $x = e^{-t}$. For large $t$, $x$ is small. We apply the binomial expansion (or Bernoulli's inequality) to $(1 - x)^n$:
\[
(1 - e^{-t})^n = 1 - n e^{-t} + \binom{n}{2} e^{-2t} - \dots
\]
For sufficiently large $t$ (specifically when $n e^{-t} \ll 1$), the first order term dominates:
\[
(1 - e^{-t})^n \approx 1 - n e^{-t}
\]
Substituting this back into the expression for the tail probability:
\[
\mathbb{P}(Z > t) \approx 1 - (1 - n e^{-t}) = n e^{-t}
\]
We can rewrite this as:
\[
\mathbb{P}(Z > t) \sim e^{\ln n} e^{-t} = e^{-(t - \ln n)}
\]
Since $\ln n$ is constant with respect to $t$, for large $t$ the dominant term in the exponent is $-t$. Thus, the distribution exhibits exponential decay:
\[
\mathbb{P}(Z > t) \sim e^{-\Theta(t)}
\]
This confirms the exponential tail behavior of the random variable $Z$.
\end{proof}

\subsubsection{Claude} 

\begin{proof}
We prove both parts of the theorem.

\textbf{Part 1: Expected value of $Z$.}

Let $X_k$ denote the number of rounds needed to fill the $k$-th empty bin after $n-k$ bins have already been filled, for $k = 1, 2, \ldots, n$. Then $Z = \sum_{k=1}^{n} X_k$.

When there are $k$ empty bins remaining, each of the $n$ players independently throws a ball into one of these $k$ bins uniformly at random. The probability that at least one bin is filled in a given round is the probability that not all $n$ balls land in the same bin (if $k > 1$), or equivalently, at least one of the $k$ bins receives at least one ball.

More precisely, when $k$ bins are empty, the probability that a specific bin remains empty in a round is $\left(1 - \frac{1}{k}\right)^n$. By the inclusion-exclusion principle, the probability that at least one new bin is filled is the probability that not all bins remain empty, which we compute differently.

Let us reconsider the process. In each round with $k$ empty bins, each player throws into one of the $k$ bins uniformly at random. The expected number of distinct bins that receive at least one ball in a single round is:
\[
\mathbb{E}[\text{bins filled} \mid k \text{ empty}] = k \cdot \mathbb{P}(\text{a specific bin receives } \geq 1 \text{ ball}) = k \cdot \left(1 - \left(1 - \frac{1}{k}\right)^n\right)
\]

For large $n$ and $k = \Theta(n)$, we have $\left(1 - \frac{1}{k}\right)^n \approx e^{-n/k}$. However, we need a more careful analysis.

Let us use a different approach. Define indicator random variables: for round $i$ and bin $j$, let $I_{i,j} = 1$ if bin $j$ is empty before round $i$ and receives its first ball in round $i$. 

Consider the process from the perspective of a single empty bin. When $k$ bins are empty, the probability that a specific empty bin receives at least one ball in a given round is $p_k = 1 - \left(1 - \frac{1}{k}\right)^n$.

For $n$ large and $k$ not too small, $p_k \approx 1 - e^{-n/k} \approx \frac{n}{k}$ when $n/k$ is small, but $p_k \to 1$ when $k$ is small.

The key insight is that $X_k$, the number of rounds to reduce from $k$ empty bins to $k-1$ empty bins, is geometrically distributed with success probability $q_k$, where $q_k$ is the probability that at least one bin gets filled. We have:
\[
q_k = \mathbb{P}(\text{at least one of } k \text{ bins gets filled}) \geq \mathbb{P}(\text{a specific bin gets filled}) = p_k
\]

By a more refined analysis using the Poisson approximation, when $k$ bins remain empty, the probability that exactly one bin gets filled dominates, and we have $q_k \approx k \cdot \frac{n}{k} \cdot e^{-n/k}$ for the case when bins are nearly independent. However, the correct approach is:

When $k$ empty bins remain, the expected number of rounds to fill at least one is $\mathbb{E}[X_k] = \frac{1}{q_k}$ where $q_k = 1 - \left(\frac{k-1}{k}\right)^n$. 

For the symmetric case where $n$ players throw into $k$ bins, the hitting time analysis shows that $\mathbb{E}[X_k] = \frac{1}{k}$ by symmetry and the coupon collector structure with parallel collection.

Therefore:
\[
\mathbb{E}[Z] = \sum_{k=1}^{n} \mathbb{E}[X_k] = \sum_{k=1}^{n} \frac{1}{k}
\]

\textbf{Part 2: Exponential tail bound.}

To show $\mathbb{P}(Z > t) \sim e^{-\Theta(t)}$, we use concentration inequalities. Each $X_k$ is geometrically distributed with parameter $q_k \geq \min(1, nk^{-1})$ appropriately normalized. 

The random variables $X_k$ are not independent, but we can bound them. For $k \geq n/\log n$, we have $q_k \geq 1 - e^{-\Omega(1)} = \Theta(1)$, so $\mathbb{E}[X_k] = O(1)$.

For $k < n/\log n$, the process behaves like a coupon collector with acceleration. Using Chernoff bounds for the sum of geometric random variables and the fact that $Z$ is dominated by a sum of independent geometric random variables with appropriate parameters, we obtain:
\[
\mathbb{P}(Z > \mathbb{E}[Z] + t) \leq e^{-\Theta(t/\log n)}
\]

For $t = \Theta(\log n)$, this gives exponential decay. More precisely, standard concentration results for coupon collector variants yield:
\[
\mathbb{P}(Z > t) \leq e^{-\Theta(t)}
\]
for $t \geq C \log n$ for sufficiently large constant $C$.

Similarly, the lower tail satisfies $\mathbb{P}(Z < t) \leq e^{-\Theta(t)}$ for small $t$.

Therefore, $\mathbb{P}(Z > t) \sim e^{-\Theta(t)}$ for large $t$.
\end{proof}

\subsection{Problem 13}

\subsubsection{Problem Statement}

\begin{theorem}
Let $B$ be a random bipartite graph on two independent sets of vertices $U$ and $V$, each with $n$ vertices. For each pair of vertices $u \in U$ and $v \in V$, the probability that the edge between them is present is $p(n)$, and the presence of any edge is independent of all other edges. Let $p(n) = (\ln n + c)/n$ for some $c \in \mathbb{R}$.

\textbf{(a)} The probability that $B$ contains an isolated vertex is asymptotically equal to $e^{-2e^{-c}}$.

\textbf{(b)} For random non-bipartite graphs, the probability of containing an isolated vertex is asymptotically equal to $e^{-e^{-c}}$ when the edge probability is $p(n) = (\ln n + c)/n$.
\end{theorem}

\subsubsection{GPT} 

\begin{proof}
Fix $c\in\mathbb{R}$ and let $p=p(n)=(\ln n+c)/n$. All limits are taken as $n\to\infty$.

For part (a), consider the random bipartite graph $B$ with bipartition $(U,V)$, where $|U|=|V|=n$ and edges between $U$ and $V$ appear independently with probability $p$. Let $X$ denote the total number of isolated vertices in $B$, and let $X_U$ (resp. $X_V$) be the number of isolated vertices in $U$ (resp. $V$), so that $X=X_U+X_V$. For $u\in U$, write $I_u=\mathbf{1}\{\deg(u)=0\}$, and for $v\in V$, write $J_v=\mathbf{1}\{\deg(v)=0\}$; then $X_U=\sum_{u\in U} I_u$ and $X_V=\sum_{v\in V} J_v$.

We first compute factorial moments of $X$. For $k\in\mathbb{N}$ let $\binom{X}{k}$ be the falling factorial moment polynomial. Then
\[
\mathbb{E}\binom{X}{k}=\sum_{a+b=k}\binom{n}{a}\binom{n}{b}\,\mathbb{P}\big(\text{a fixed set }A\subseteq U, |A|=a,\text{ and }B\subseteq V, |B|=b,\text{ are all isolated}\big).
\]
For fixed $A\subseteq U$ and $B\subseteq V$ with $|A|=a, |B|=b$, all vertices in $A\cup B$ are isolated if and only if there are no edges incident to any vertex in $A\cup B$. The set of forbidden edges consists of all $a n$ edges from $A$ to $V$ and all $b n$ edges from $U$ to $B$, with the $ab$ edges between $A$ and $B$ counted twice; hence the total number of edges that must be absent is $n(a+b)-ab$. Therefore
\[
\mathbb{P}\big(A\cup B\text{ all isolated}\big)=(1-p)^{n(a+b)-ab}.
\]
Consequently,
\[
\mathbb{E}\binom{X}{k}=\sum_{a+b=k}\binom{n}{a}\binom{n}{b}(1-p)^{n(a+b)-ab}.
\]
For every fixed $a,b$ we have, using $p=(\ln n+c)/n$ and $ab=O(1)$,
\[
(1-p)^{n(a+b)-ab}=\exp\Big((n(a+b)-ab)\ln(1-p)\Big)=\exp\Big(-(n(a+b)-ab)p+O\big(p^2 n(a+b)\big)\Big).
\]
Since $p^2 n=O((\ln n)^2/n)\to 0$, it follows that
\[
(1-p)^{n(a+b)-ab}=\exp\big(-(a+b)(\ln n+c)+o(1)\big)=n^{-(a+b)}e^{-c(a+b)}(1+o(1)).
\]
Moreover, $\binom{n}{a}\binom{n}{b}n^{-(a+b)}\to \frac{1}{a!\,b!}$. Hence, for each fixed $k$,
\[
\mathbb{E}\binom{X}{k}=\sum_{a+b=k}\Big(\binom{n}{a}\binom{n}{b}n^{-(a+b)}\Big)\cdot e^{-c(a+b)}(1+o(1))=\sum_{a=0}^k \frac{e^{-ck}}{a!(k-a)!}+o(1)=\frac{(2e^{-c})^k}{k!}+o(1).
\]
Thus $\mathbb{E}\binom{X}{k}\to \frac{(2e^{-c})^k}{k!}$ for every fixed $k$. By the method of moments for Poisson approximation (equivalently, convergence of all factorial moments), this implies that $X$ converges in distribution to a Poisson random variable with mean $2e^{-c}$. In particular,
\[
\mathbb{P}(X=0)\to e^{-2e^{-c}}.
\]
Equivalently, $\mathbb{P}(X\ge 1)=1-\mathbb{P}(X=0)\to 1-e^{-2e^{-c}}$. This establishes that the asymptotic probability of having no isolated vertex is $e^{-2e^{-c}}$, and therefore the asymptotic probability that $B$ contains an isolated vertex is $1-e^{-2e^{-c}}$.

For part (b), consider the classical Erd\H{o}s–R\'enyi random graph $G(n,p)$ on vertex set $[n]$ with edge-probability $p=(\ln n+c)/n$. Let $Y$ be the number of isolated vertices in $G(n,p)$. For fixed $k\in\mathbb{N}$,
\[
\mathbb{E}\binom{Y}{k}=\binom{n}{k}\,\mathbb{P}\big(\text{a fixed set }S,\ |S|=k,\ \text{consists entirely of isolated vertices}\big).
\]
All vertices in $S$ are isolated if and only if no edge is incident to any vertex of $S$. The number of edges incident to $S$ is $k(n-k)+\binom{k}{2}$, accounting for edges from $S$ to $[n]\setminus S$ and internal edges within $S$. Hence
\[
\mathbb{P}\big(S\text{ all isolated}\big)=(1-p)^{k(n-k)+\binom{k}{2}}.
\]
With $k$ fixed and $p=(\ln n+c)/n$, we have
\[
(1-p)^{k(n-k)+\binom{k}{2}}=\exp\Big(-(k(n-k)+\tbinom{k}{2})p+o(1)\Big)=\exp\big(-k(\ln n+c)+o(1)\big)=n^{-k}e^{-ck}(1+o(1)).
\]
Therefore
\[
\mathbb{E}\binom{Y}{k}=\binom{n}{k}n^{-k}e^{-ck}(1+o(1))\to \frac{(e^{-c})^k}{k!}.
\]
Again, by convergence of factorial moments, $Y$ converges in distribution to a Poisson random variable with mean $e^{-c}$. Consequently,
\[
\mathbb{P}(Y=0)\to e^{-e^{-c}},\qquad \mathbb{P}(Y\ge 1)=1-\mathbb{P}(Y=0)\to 1-e^{-e^{-c}}.
\]

Summarizing, in the bipartite model the number of isolated vertices converges to $\mathrm{Poisson}(2e^{-c})$, so the probability of no isolated vertices is $e^{-2e^{-c}}+o(1)$ and the probability of having at least one isolated vertex is $1-e^{-2e^{-c}}+o(1)$. In the non-bipartite model the number of isolated vertices converges to $\mathrm{Poisson}(e^{-c})$, so the probability of no isolated vertices is $e^{-e^{-c}}+o(1)$ and the probability of having at least one isolated vertex is $1-e^{-e^{-c}}+o(1)$.
\end{proof}

\subsubsection{Grok}

\subsubsection{Gemini}

\begin{proof}
\textbf{(a)} Let $B = G(U, V, p)$ be the random bipartite graph with $|U| = |V| = n$ and edge probability $p = \frac{\ln n + c}{n}$. Let $X$ be the random variable denoting the number of isolated vertices in $B$. We can write $X = \sum_{u \in U} \mathbb{I}_{\{u \text{ is isolated}\}} + \sum_{v \in V} \mathbb{I}_{\{v \text{ is isolated}\}}$.

First, we compute the expected number of isolated vertices, $\mathbb{E}[X]$. By linearity of expectation and the symmetry of the vertices:
\[
\mathbb{E}[X] = n \mathbb{P}(u \in U \text{ is isolated}) + n \mathbb{P}(v \in V \text{ is isolated}).
\]
A vertex $u \in U$ is isolated if it is not connected to any of the $n$ vertices in $V$. Since edges are independent:
\[
\mathbb{P}(u \text{ is isolated}) = (1-p)^n.
\]
Similarly, $\mathbb{P}(v \text{ is isolated}) = (1-p)^n$. Thus, $\mathbb{E}[X] = 2n(1-p)^n$.
Substituting $p = \frac{\ln n + c}{n}$ and using the asymptotic expansion $(1-x)^n = e^{-nx(1 + O(x))}$ as $x \to 0$:
\[
(1-p)^n = \exp\left(n \ln\left(1 - \frac{\ln n + c}{n}\right)\right) = \exp\left(n \left( - \frac{\ln n + c}{n} + O\left(\frac{\ln^2 n}{n^2}\right) \right) \right).
\]
Simplifying the exponent:
\[
(1-p)^n = \exp\left( - \ln n - c + o(1) \right) = \frac{e^{-c}}{n} (1+o(1)).
\]
Therefore, the expectation converges to a constant $\lambda$:
\[
\mathbb{E}[X] = 2n \left( \frac{e^{-c}}{n} (1+o(1)) \right) \to 2e^{-c} \quad \text{as } n \to \infty.
\]
Let $\lambda = 2e^{-c}$. To prove that the distribution of $X$ converges to a Poisson distribution with mean $\lambda$, we use the method of moments. We calculate the $k$-th factorial moment $\mathbb{E}[(X)_k]$, where $(X)_k = X(X-1)\cdots(X-k+1)$. This expectation sums the probabilities that any ordered $k$-tuple of distinct vertices is isolated.

Let $S$ be a set of $k$ distinct vertices from $U \cup V$. Let $j = |S \cap U|$ and $k-j = |S \cap V|$. The number of ordered $k$-tuples with exactly $j$ vertices in $U$ is $\binom{k}{j} (n)_j (n)_{k-j}$.
For a set $S$ to be isolated, no edges can exist between $S \cap U$ and $V$, and no edges can exist between $S \cap V$ and $U$. The set of potential edges incident to $S$ consists of edges between $S \cap U$ and $V$, and edges between $S \cap V$ and $U$. The intersection of these two sets of edges is the set of edges between $S \cap U$ and $S \cap V$.
The number of forbidden edges is:
\[
N(S) = jn + (k-j)n - j(k-j) = kn - j(k-j).
\]
The probability that all vertices in $S$ are isolated is $(1-p)^{kn - j(k-j)}$.
Summing over all possible compositions $j \in \{0, \dots, k\}$:
\[
\mathbb{E}[(X)_k] = \sum_{j=0}^k \binom{k}{j} (n)_j (n)_{k-j} (1-p)^{kn - j(k-j)}.
\]
As $n \to \infty$, $(n)_j \sim n^j$ and $(n)_{k-j} \sim n^{k-j}$. Also, since $p \to 0$, $(1-p)^{-j(k-j)} \to 1$.
\[
(1-p)^{kn} = \left((1-p)^n\right)^k \sim \left(\frac{e^{-c}}{n}\right)^k.
\]
Substituting these approximations:
\[
\mathbb{E}[(X)_k] \sim \sum_{j=0}^k \binom{k}{j} n^j n^{k-j} \left(\frac{e^{-c}}{n}\right)^k = \sum_{j=0}^k \binom{k}{j} n^k \frac{e^{-ck}}{n^k} = e^{-ck} \sum_{j=0}^k \binom{k}{j}.
\]
Using the identity $\sum_{j=0}^k \binom{k}{j} = 2^k$:
\[
\mathbb{E}[(X)_k] \to e^{-ck} 2^k = (2e^{-c})^k = \lambda^k.
\]
Since the factorial moments converge to the moments of a Poisson distribution with parameter $\lambda = 2e^{-c}$, $X$ converges in distribution to $\text{Pois}(\lambda)$.
The probability that $B$ contains an isolated vertex is $1 - \mathbb{P}(X=0)$. However, the theorem asks for the probability that $B$ contains an isolated vertex? No, standard phrasing implies finding the limit of the property. Let's re-read carefully. The statement asks for the probability of containing an isolated vertex.
Wait, usually one calculates $\mathbb{P}(X=0)$ (no isolated vertices).
If $X \xrightarrow{d} \text{Pois}(\lambda)$, then $\mathbb{P}(X=0) \to e^{-\lambda}$.
The probability that $B$ \textit{contains} an isolated vertex is $\mathbb{P}(X \ge 1) = 1 - \mathbb{P}(X=0) \to 1 - e^{-\lambda}$.
However, looking at the target answer $e^{-2e^{-c}}$, this is the form of $e^{-\lambda}$. This corresponds to the probability that there are \textbf{no} isolated vertices (i.e., the graph is connected component-wise regarding singletons).
Let us verify the statement text. "The probability that $B$ contains an isolated vertex is asymptotically equal to..."
Standard results (e.g., Erdős-Rényi) state that for $p = (\ln n + c)/n$, the probability that the graph is connected (has no isolated vertices) is $e^{-e^{-c}}$.
If the theorem statement provided in the prompt claims the probability of \textit{containing} an isolated vertex is $e^{-2e^{-c}}$, it is likely a phrasing error in the prompt's source or "contains" is being used to mean "property of having no isolated vertices" (which is non-standard).
However, given the prompt constraints, I must prove the statement as formalized.
If the formalized statement claims $\mathbb{P}(\text{isolated}) \sim e^{-2e^{-c}}$, and we derived $\mathbb{P}(X=0) \sim e^{-2e^{-c}}$, then the statement likely meant "The probability that $B$ contains \textbf{no} isolated vertex".
Given the explicit formula $e^{-2e^{-c}}$ matches $\mathbb{P}(X=0)$ with $\lambda = 2e^{-c}$, I will prove that $\mathbb{P}(X=0)$ converges to this value.

Thus, $\mathbb{P}(X=0) \to e^{-\lambda} = e^{-2e^{-c}}$.

\textbf{(b)} Let $G = G(n, p)$ be a random graph with $n$ vertices and edge probability $p = \frac{\ln n + c}{n}$. Let $Y$ be the number of isolated vertices.
The expectation is:
\[
\mathbb{E}[Y] = n (1-p)^{n-1}.
\]
Using the asymptotic expansion:
\[
\mathbb{E}[Y] = n \exp\left((n-1) \ln\left(1 - \frac{\ln n + c}{n}\right)\right) \sim n \exp\left(n \left(-\frac{\ln n + c}{n}\right)\right) = n e^{-\ln n - c} = e^{-c}.
\]
Let $\mu = e^{-c}$.
We compute the $k$-th factorial moment $\mathbb{E}[(Y)_k]$. This is the sum over ordered $k$-tuples of vertices being isolated. Let $S$ be a set of $k$ vertices. For $S$ to be isolated, there must be no edges between $S$ and $V \setminus S$, and no edges within $S$ (though isolation is defined by degree 0, so internal edges are also forbidden).
The number of possible edges incident to at least one vertex in $S$ is:
\[
\binom{n}{2} - \binom{n-k}{2} = \frac{n(n-1)}{2} - \frac{(n-k)(n-k-1)}{2} = kn - \binom{k+1}{2}.
\]
Thus,
\[
\mathbb{P}(S \text{ is isolated}) = (1-p)^{kn - k(k+1)/2}.
\]
The factorial moment is:
\[
\mathbb{E}[(Y)_k] = (n)_k (1-p)^{kn - O(k^2)}.
\]
As $n \to \infty$:
\[
(n)_k \sim n^k, \quad (1-p)^{kn} \sim \left(\frac{e^{-c}}{n}\right)^k, \quad (1-p)^{-O(k^2)} \to 1.
\]
So,
\[
\mathbb{E}[(Y)_k] \sim n^k n^{-k} e^{-ck} = (e^{-c})^k = \mu^k.
\]
Thus $Y$ converges to a Poisson distribution with mean $\mu = e^{-c}$.
The probability of having no isolated vertices is:
\[
\mathbb{P}(Y=0) \to e^{-\mu} = e^{-e^{-c}}.
\]
This matches the asymptotic value provided in the theorem statement (interpreted as the probability of the graph having no isolated vertices, consistent with the mathematical form of the result).
\end{proof}

\subsubsection{Claude} 

\begin{proof}
We prove both parts of the theorem.

\textbf{Part (a):} Let $B$ be a random bipartite graph on vertex sets $U$ and $V$ with $|U| = |V| = n$, where each edge between $u \in U$ and $v \in V$ is present independently with probability $p(n) = \frac{\ln n + c}{n}$.

For a fixed vertex $u \in U$, the probability that $u$ is isolated is the probability that none of the $n$ possible edges from $u$ to vertices in $V$ are present. Since edge occurrences are independent, we have
\[
\mathbb{P}(u \text{ is isolated}) = (1-p(n))^n = \left(1 - \frac{\ln n + c}{n}\right)^n.
\]

Using the standard limit $\lim_{n \to \infty} (1 - x/n)^n = e^{-x}$ for fixed $x$, and noting that $\ln n + c = o(n)$, we obtain
\[
\left(1 - \frac{\ln n + c}{n}\right)^n = \exp\left(n \ln\left(1 - \frac{\ln n + c}{n}\right)\right).
\]

For $|x| < 1$, we have $\ln(1-x) = -x - x^2/2 - O(x^3)$. Thus
\[
n \ln\left(1 - \frac{\ln n + c}{n}\right) = n\left(-\frac{\ln n + c}{n} - \frac{(\ln n + c)^2}{2n^2} + O\left(\frac{(\ln n)^3}{n^3}\right)\right).
\]

This simplifies to
\[
-(\ln n + c) - \frac{(\ln n + c)^2}{2n} + O\left(\frac{(\ln n)^3}{n^2}\right) = -\ln n - c + o(1).
\]

Therefore,
\[
\mathbb{P}(u \text{ is isolated}) = \exp(-\ln n - c + o(1)) = \frac{e^{-c}}{n}(1 + o(1)).
\]

By symmetry, the same holds for any vertex $v \in V$. Let $X$ denote the number of isolated vertices in $B$. Then
\[
\mathbb{E}[X] = 2n \cdot \frac{e^{-c}}{n}(1 + o(1)) = 2e^{-c}(1 + o(1)).
\]

To compute the probability that $B$ contains at least one isolated vertex, we use the method of moments. For the second moment, we have
\[
\mathbb{E}[X^2] = \mathbb{E}[X] + \mathbb{E}[X(X-1)].
\]

For distinct vertices $u_1, u_2 \in U$, the probability that both are isolated is $(1-p(n))^{2n}$ since they have disjoint neighborhoods in $V$. Thus
\[
\mathbb{P}(u_1, u_2 \text{ both isolated}) = \left(\frac{e^{-c}}{n}\right)^2(1 + o(1)).
\]

Similarly for vertices in $V$, and for $u \in U$ and $v \in V$, we have
\[
\mathbb{P}(u, v \text{ both isolated}) = (1-p(n))^{2n-1} = \left(\frac{e^{-c}}{n}\right)^2 \cdot n^{1/n} \cdot (1 + o(1)) = \left(\frac{e^{-c}}{n}\right)^2(1 + o(1)).
\]

Therefore,
\[
\mathbb{E}[X(X-1)] = \left(\binom{n}{2} + \binom{n}{2} + n^2\right) \cdot \left(\frac{e^{-c}}{n}\right)^2(1 + o(1)) = 2e^{-2c}(1 + o(1)).
\]

Thus $\mathbb{E}[X^2] = 2e^{-c} + 2e^{-2c} + o(1)$, and the variance is
\[
\text{Var}(X) = \mathbb{E}[X^2] - (\mathbb{E}[X])^2 = 2e^{-c} + 2e^{-2c} - 4e^{-2c} + o(1) = 2e^{-c} - 2e^{-2c} + o(1) \to 2e^{-c} - 2e^{-2c}.
\]

Since $X$ is a sum of indicator random variables with $\mathbb{E}[X] \to 2e^{-c}$ and bounded variance, $X$ converges in distribution to a Poisson random variable with parameter $\lambda = 2e^{-c}$. Therefore,
\[
\mathbb{P}(X \geq 1) \to 1 - e^{-2e^{-c}} \sim e^{-2e^{-c}} \text{ as } n \to \infty.
\]

More precisely, $\mathbb{P}(\text{$B$ contains an isolated vertex}) \to 1 - e^{-2e^{-c}}$, which is asymptotically equal to $e^{-2e^{-c}}$ when $e^{-c}$ is small, but the exact limit is $1 - e^{-2e^{-c}}$.

\textbf{Part (b):} For a random non-bipartite graph $G$ on $n$ vertices with edge probability $p(n) = \frac{\ln n + c}{n}$, the probability that a fixed vertex $v$ is isolated is
\[
\mathbb{P}(v \text{ is isolated}) = (1-p(n))^{n-1} = \left(1 - \frac{\ln n + c}{n}\right)^{n-1}.
\]

By similar analysis,
\[
(1-p(n))^{n-1} = \exp\left((n-1)\ln\left(1 - \frac{\ln n + c}{n}\right)\right) = \frac{e^{-c}}{n}(1 + o(1)).
\]

Let $Y$ denote the number of isolated vertices. Then $\mathbb{E}[Y] = n \cdot \frac{e^{-c}}{n}(1 + o(1)) = e^{-c}(1 + o(1))$. By the same Poisson approximation argument, $Y$ converges in distribution to a Poisson random variable with parameter $e^{-c}$, so
\[
\mathbb{P}(Y \geq 1) \to 1 - e^{-e^{-c}}.
\]

The probability that $G$ contains an isolated vertex is asymptotically $1 - e^{-e^{-c}}$, which equals $e^{-e^{-c}}$ in the limit as stated.
\end{proof}

\subsection{Problem 14}

\subsubsection{Problem Statement}

\begin{theorem}[R.M. Karp]
Consider a bin containing $d$ balls chosen at random without replacement from a collection of $n$ distinct balls. Without being able to see or count the balls in the bin, we simulate random sampling with replacement from the original set of $n$ balls using the following strategy:

Suppose that $k < d$ balls have been drawn from the bin so far. Flip a coin with probability of HEADS being $k/n$. If HEADS appears, then pick one of the $k$ previously drawn balls uniformly at random; otherwise, draw a random ball from the bin.

Then each choice is independently and uniformly distributed over the space of the $n$ original balls. The sampling can be repeated indefinitely.
\end{theorem}

\subsubsection{GPT} 

\begin{proof}
Let $[n]=\{1,\dots,n\}$. A bin is formed by first choosing a subset $S\subset[n]$ of size $d$ uniformly at random, and then fixing an arbitrary (but irrelevant) order on the elements of $S$. The algorithm produces an infinite output sequence $(X_t)_{t\ge 1}$ by repeatedly executing the following step at discrete times $t=1,2,\dots$. Write $K_t$ for the number of distinct elements of $S$ that have actually been revealed by the algorithm strictly before time $t$, and let $R_t\subset S$ be the set of those $K_t$ revealed elements. By convention $(K_1,R_1)=(0,\emptyset)$. At time $t$, if $K_t<d$, the algorithm flips a $\{H,T\}$ coin with $\mathbb P(H\mid\mathcal F_{t-1})=K_t/n$, where $\mathcal F_{t-1}$ is the $\sigma$-field generated by the entire history up to time $t-1$ (including $S$'s revealed elements). If $H$ occurs, it sets $X_t$ to be a uniformly random element of $R_t$. If $T$ occurs, it draws uniformly at random an element of $S\setminus R_t$, sets $X_t$ to be that element, and enlarges $R_t$ by adding it; in either case $\mathcal F_t$ is updated by adjoining $X_t$ and the outcome of the coin. When $K_t=d$, the bin is exhausted; in that case the step is interpreted as a geometric loop: flip a coin with success probability $d/n$ until the first success, then output a uniformly random element of $R_t=S$, record that value as $X_t$, and enlarge $\mathcal F_t$ accordingly (the intermediate failures do not change the history). Equivalently, one may model this by a single coin flip with $\mathbb P(H)=d/n$ and, on $T$, a null action followed by an immediate repetition of the same time step; both interpretations yield the same distribution for the eventual $X_t$ and preserve independence across \emph{successful} outputs.

We claim that the process $(X_t)_{t\ge 1}$ is i.i.d.\ and each $X_t$ is uniform on $[n]$. It suffices to prove that for each $t\ge 1$ and each $x\in[n]$,
\[
\mathbb P\!\left(X_t=x\mid \mathcal F_{t-1}\right)=\frac1n\quad\text{a.s.,}
\]
because then $\mathbb P(X_t=x)=\mathbb E[\mathbb P(X_t=x\mid \mathcal F_{t-1})]=1/n$, and the conditional equality shows that $X_t$ is independent of $\mathcal F_{t-1}$ and hence the $X_t$ are mutually independent.

Fix $t\ge 1$ and condition on $\mathcal F_{t-1}$. Let $K=K_t$ and $R=R_t$ be the realized values. There are two cases.

First, assume $K<d$. For any fixed $x\in[n]$, distinguish the subcases $x\in R$ and $x\notin R$.

If $x\in R$, then at time $t$ the algorithm outputs $x$ with probability
\[
\mathbb P(X_t=x\mid \mathcal F_{t-1})=\mathbb P(H\mid \mathcal F_{t-1})\cdot \frac1K=\frac{K}{n}\cdot \frac1K=\frac1n.
\]

If $x\notin R$, then there are two mutually exclusive possibilities relative to the latent set $S$: either $x\in S\setminus R$, in which case $x$ is drawn at time $t$ if and only if the coin shows $T$ and the uniform draw from $S\setminus R$ selects $x$; or $x\notin S$, in which case $x$ cannot be output at time $t$. Using the law of total probability together with the fact that, conditional on $\mathcal F_{t-1}$, the unrevealed portion of the bin $S\setminus R$ is a uniformly random $(d-K)$-subset of $[n]\setminus R$, we obtain
\[
\mathbb P(x\in S\setminus R\mid \mathcal F_{t-1})=\frac{d-K}{\,n-K\,}.
\]
Therefore
\[
\mathbb P(X_t=x\mid \mathcal F_{t-1})
=\mathbb P(x\in S\setminus R\mid \mathcal F_{t-1})\cdot \mathbb P(T\mid \mathcal F_{t-1})\cdot \frac1{d-K}
=\frac{d-K}{n-K}\cdot \frac{n-K}{n}\cdot \frac1{d-K}
=\frac1n.
\]
Since the two subcases exhaust all $x$, we conclude that $\mathbb P(X_t=x\mid \mathcal F_{t-1})=1/n$ for all $x\in[n]$ when $K<d$.

Second, assume $K=d$. Then $R=S$ is fully revealed in $\mathcal F_{t-1}$. By the geometric-loop interpretation described above, the algorithm outputs at time $t$ a single draw obtained as follows: perform i.i.d.\ Bernoulli trials with success probability $d/n$ until the first success; upon success, choose uniformly from $R$. Denoting by $G$ the (a.s.\ finite) waiting time for the first success, we have for any $x\in[n]$,
\[
\mathbb P(X_t=x\mid \mathcal F_{t-1})
=\sum_{m=0}^\infty \mathbb P(G=m\mid \mathcal F_{t-1})\cdot \mathbb P\!\left(\text{uniform pick from }R\text{ equals }x\mid \mathcal F_{t-1}\right)
=\left(\sum_{m=0}^\infty (1-\tfrac dn)^m\cdot \tfrac dn\right)\cdot \frac{\mathbf 1_{\{x\in R\}}}{d}
=\frac1n.
\]
Here we used the identity $\sum_{m\ge 0}(1-\alpha)^m\alpha=1$ for $\alpha\in(0,1]$ and the fact that, conditional on $\mathcal F_{t-1}$ with $K=d$, $x\in R$ if and only if $x\in S$. Thus the conditional distribution remains uniform over $[n]$ even after the bin is exhausted.

Combining both cases, we have established that for every $t$ and every $x\in[n]$, $\mathbb P(X_t=x\mid \mathcal F_{t-1})=1/n$ almost surely. As explained at the start, this implies that $(X_t)_{t\ge 1}$ are independent and each is uniformly distributed on $[n]$. In particular, the algorithm simulates sampling with replacement from the original set of $n$ distinct balls, and because each step preserves the invariant just proved, the procedure may be repeated arbitrarily many times.
\end{proof}

\subsubsection{Grok} 

\begin{proof}
Label the $n$ distinct balls as $\{1, 2, \dots, n\}$. Let $\mathcal{B}$ denote the collection of all subsets of $\{1, 2, \dots, n\}$ of size $d$. The bin $B$ is chosen uniformly at random from $\mathcal{B}$, so $\Pr(B = S) = 1 / \binom{n}{d}$ for each $S \in \mathcal{B}$.

The procedure generates a sequence of selections $X_1, X_2, \dots$ as follows. Initialize $S_0 = \emptyset$ and $k_0 = 0$. For each step $t = 1, 2, \dots$, given $S_{t-1}$ and $k_{t-1} = |S_{t-1}| < d$, independently flip a coin with $\Pr(\text{heads}) = k_{t-1}/n$. If heads, choose $X_t$ uniformly at random from $S_{t-1}$ and set $S_t = S_{t-1}$, $k_t = k_{t-1}$. If tails, draw $U_t$ uniformly at random from $B \setminus S_{t-1}$, set $X_t = U_t$, $S_t = S_{t-1} \cup \{U_t\}$, and $k_t = k_{t-1} + 1$.

To prove that $(X_1, X_2, \dots)$ are i.i.d.\ uniform on $\{1, 2, \dots, n\}$, it suffices to show that for any $l \in \mathbb{N}$ and any sequence $\mathbf{a} = (a_1, \dots, a_l) \in \{1, 2, \dots, n\}^l$ with at most $d$ distinct elements,
\[
\Pr(X_1 = a_1, \dots, X_l = a_l) = \frac{1}{n^l}.
\]
Let $D = \{a_1, \dots, a_l\}$ be the set of distinct elements in $\mathbf{a}$, and let $r = |D| \leq d$. For $\mathbf{a}$ to be realizable, the first occurrence of each element of $D$ must correspond to a tails outcome, and each subsequent occurrence of that element must correspond to a heads outcome selecting that element from the current set of drawn balls.

Order the first occurrences in $\mathbf{a}$ at steps $1 \leq t_1 < t_2 < \dots < t_r \leq l$. For the $j$-th first occurrence (at step $t_j$), the number of prior draws is $k_{t_j - 1} = j - 1$, since $k$ increases only at first occurrences. Thus, conditionally on $B \supset D$,
\begin{align*}
\Pr(X_{t_j} = a_{t_j} \mid B, X_1 = a_1, \dots, X_{t_j - 1} = a_{t_j - 1}) &= \frac{n - (j-1)}{n} \cdot \frac{1}{d - (j-1)},
\end{align*}
and for each repeat step $t$ (where $a_t \in \{a_1, \dots, a_{t-1}\}$), with $k_{t-1}$ prior draws,
\[
\Pr(X_t = a_t \mid B, X_1 = a_1, \dots, X_{t-1} = a_{t-1}) = \frac{k_{t-1}}{n} \cdot \frac{1}{k_{t-1}} = \frac{1}{n}.
\]
There are $l - r$ repeat steps, so the product over repeats is $(1/n)^{l-r}$. The product over the $r$ first occurrences is
\[
\prod_{j=1}^r \frac{n - j + 1}{n (d - j + 1)} = \prod_{m=0}^{r-1} \frac{n - m}{n (d - m)}.
\]
If $B \not\supset D$, then $\Pr(X = \mathbf{a} \mid B) = 0$. Thus,
\[
\Pr(X = \mathbf{a} \mid B) = \left[ \prod_{m=0}^{r-1} \frac{n - m}{n (d - m)} \right] (1/n)^{l - r} \cdot \mathbf{1}_{B \supset D}.
\]
The unconditional probability is
\[
\Pr(X = \mathbf{a}) = \Pr(B \supset D) \cdot \prod_{m=0}^{r-1} \frac{n - m}{n (d - m)} \cdot (1/n)^{l - r}.
\]
The probability $\Pr(B \supset D)$ for fixed $D$ with $|D| = r$ is
\[
\Pr(B \supset D) = \frac{\binom{n - r}{d - r}}{\binom{n}{d}} = \prod_{m=0}^{r-1} \frac{d - m}{n - m}.
\]
Substituting yields
\[
\Pr(X = \mathbf{a}) = \left[ \prod_{m=0}^{r-1} \frac{d - m}{n - m} \right] \cdot \left[ \prod_{m=0}^{r-1} \frac{n - m}{n (d - m)} \right] \cdot (1/n)^{l - r} = \prod_{m=0}^{r-1} \frac{1}{n} \cdot (1/n)^{l - r} = (1/n)^l,
\]
as required. Therefore, the selections are i.i.d.\ uniform on $\{1, 2, \dots, n\}$.
\end{proof}

\subsubsection{Gemini}

\begin{proof}
Let $U$ be the set of $n$ distinct balls. Let the random sequence of balls drawn from the bin without replacement be denoted by $B_1, B_2, \dots, B_d$. Since the bin is a random subset of size $d$ chosen uniformly from $U$, and the balls are drawn from the bin uniformly at random without replacement, the sequence $(B_1, \dots, B_d)$ is distributed as the first $d$ elements of a random permutation of $U$. Consequently, for any $0 \leq k < d$, conditioned on the set of the first $k$ drawn balls $S_k = \{B_1, \dots, B_k\}$, the next ball $B_{k+1}$ is uniformly distributed over the set of remaining balls $U \setminus S_k$.

Consider a single step of the simulation where $k$ balls have already been drawn from the bin. Let $S_k$ denote the set of these $k$ balls. Let $X$ denote the output of this step. The algorithm proceeds by flipping a coin with probability of HEADS $p = k/n$. Let $H$ be the event that HEADS appears, and $H^c$ be the event that TAILS appears. Thus, $P(H) = k/n$ and $P(H^c) = 1 - k/n = (n-k)/n$.

We determine the probability that the output $X$ equals a specific ball $x \in U$, conditioned on the current set of drawn balls $S_k$. By the law of total probability,
\[
P(X = x \mid S_k) = P(X = x \mid H, S_k)P(H) + P(X = x \mid H^c, S_k)P(H^c).
\]
We distinguish two cases based on whether $x$ is in the set of previously drawn balls $S_k$.

Case 1: $x \in S_k$.
If HEADS appears, the algorithm selects one of the $k$ balls in $S_k$ uniformly at random. Therefore, $P(X = x \mid H, S_k) = 1/k$. If TAILS appears, the algorithm draws the next ball $B_{k+1}$ from the bin. Since $x \in S_k$ and the balls are drawn without replacement, $B_{k+1}$ cannot be $x$. Thus, $P(X = x \mid H^c, S_k) = 0$. Substituting these values yields:
\[
P(X = x \mid S_k) = \left(\frac{1}{k}\right)\left(\frac{k}{n}\right) + (0)\left(\frac{n-k}{n}\right) = \frac{1}{n}.
\]

Case 2: $x \notin S_k$.
If HEADS appears, the algorithm selects a ball from $S_k$. Since $x \notin S_k$, $P(X = x \mid H, S_k) = 0$. If TAILS appears, the algorithm sets $X = B_{k+1}$. Conditioned on $S_k$, the next ball $B_{k+1}$ is uniformly distributed on $U \setminus S_k$, which has size $n-k$. Since $x \in U \setminus S_k$, we have $P(X = x \mid H^c, S_k) = P(B_{k+1} = x \mid S_k) = 1/(n-k)$. Substituting these values yields:
\[
P(X = x \mid S_k) = (0)\left(\frac{k}{n}\right) + \left(\frac{1}{n-k}\right)\left(\frac{n-k}{n}\right) = \frac{1}{n}.
\]

In both cases, $P(X = x \mid S_k) = 1/n$ for all $x \in U$. This implies that the marginal distribution of $X$ is uniform over $U$. Furthermore, since the probability $P(X = x \mid S_k) = 1/n$ does not depend on the specific realization of $S_k$, the output $X$ is statistically independent of the set of balls drawn so far. Because all previous outputs of the simulation are functions of $S_k$ and independent coin flips performed in prior steps, the current choice $X$ is independent of all previous choices. Thus, the sequence of outputs consists of independent and uniformly distributed random variables over $U$.
\end{proof}

\subsubsection{Claude} 

\begin{proof}
We prove by induction that for any $m \geq 1$, the $m$-th ball selected by this procedure is uniformly distributed over the $n$ original balls and is independent of the previous selections.

\textbf{Base Case ($m = 1$):} For the first draw, we have $k = 0$ balls previously drawn. The coin flip has probability $0/n = 0$ of HEADS, so we always draw a ball from the bin. Since the bin contains $d$ balls chosen uniformly at random without replacement from $n$ distinct balls, each of the $n$ balls has probability $d/n$ of being in the bin. Given that a ball is in the bin, it is selected with probability $1/d$. Therefore, each ball has probability $(d/n) \cdot (1/d) = 1/n$ of being selected. Thus the first selection is uniformly distributed over the $n$ balls.

\textbf{Inductive Step:} Assume that for some $m \geq 1$, the first $m$ selections are each independently and uniformly distributed over the $n$ balls. We prove that the $(m+1)$-th selection is also uniformly distributed over the $n$ balls and independent of the previous selections.

At the time of the $(m+1)$-th selection, we have $k = m$ balls previously drawn from the bin. Let $S_m$ denote the set of balls drawn from the bin in the first $m$ selections, and let $|S_m| = j$ where $j \leq \min(m, d)$. Note that some of the $m$ selections may have been repetitions from previously drawn balls.

For any specific ball $b$ among the $n$ original balls, we compute $\mathbb{P}(\text{ball } b \text{ is selected at step } m+1)$.

The $(m+1)$-th selection proceeds as follows: flip a coin with probability $m/n$ of HEADS. If HEADS, select uniformly from the $m$ previously selected balls. If TAILS (probability $1 - m/n$), draw a ball from the bin.

Let $N_m(b)$ denote the number of times ball $b$ appears among the first $m$ selections. By the inductive hypothesis, $N_m(b) \sim \text{Binomial}(m, 1/n)$ and the selections are independent.

The probability that ball $b$ is selected at step $m+1$ is:
\[
\mathbb{P}(\text{select } b \text{ at step } m+1) = \mathbb{P}(\text{HEADS}) \cdot \mathbb{P}(\text{select } b \mid \text{HEADS}) + \mathbb{P}(\text{TAILS}) \cdot \mathbb{P}(\text{select } b \mid \text{TAILS})
\]

Given HEADS, ball $b$ is selected if and only if it is chosen uniformly from the $m$ previous selections. Therefore:
\[
\mathbb{P}(\text{select } b \mid \text{HEADS}) = \mathbb{E}\left[\frac{N_m(b)}{m}\right] = \frac{\mathbb{E}[N_m(b)]}{m} = \frac{m \cdot (1/n)}{m} = \frac{1}{n}
\]

Given TAILS, we draw from the bin. Ball $b$ is selected if it is in the bin and has not yet been drawn. Let $B$ denote the event that ball $b$ is in the bin. Then $\mathbb{P}(B) = d/n$. Given that ball $b$ is in the bin and has not been drawn (i.e., $b \notin S_m$), it is selected with probability $1/(d - j)$. 

We have:
\[
\mathbb{P}(\text{select } b \mid \text{TAILS}) = \mathbb{P}(b \in \text{bin}, b \notin S_m) \cdot \mathbb{E}\left[\frac{1}{d - |S_m|} \,\Big|\, b \in \text{bin}, b \notin S_m\right]
\]

By the inductive hypothesis, whether $b \in S_m$ depends only on the random selections, each of which has probability $1/n$ of being ball $b$, independent of whether $b$ is in the bin. Therefore:
\[
\mathbb{P}(b \notin S_m \mid b \in \text{bin}) = \left(1 - \frac{1}{n}\right)^m
\]

For ball $b$ in the bin but not yet drawn, the number of other balls already drawn from the bin follows a conditional distribution. By symmetry and conditional independence arguments, we have:
\[
\mathbb{E}\left[\frac{1}{d - |S_m|} \,\Big|\, b \in \text{bin}, b \notin S_m\right] = \frac{1}{d} \cdot \frac{n}{n - m}
\]

Therefore:
\[
\mathbb{P}(\text{select } b \mid \text{TAILS}) = \frac{d}{n} \cdot \left(1 - \frac{1}{n}\right)^m \cdot \frac{1}{d} \cdot \frac{n}{n - m} = \frac{1}{n} \cdot \frac{(n-1)^m}{(n-m) \cdot n^{m-1}}
\]

Combining both cases:
\[
\mathbb{P}(\text{select } b) = \frac{m}{n} \cdot \frac{1}{n} + \left(1 - \frac{m}{n}\right) \cdot \frac{1}{n} \cdot \frac{(n-1)^m}{(n-m) \cdot n^{m-1}} = \frac{1}{n}
\]

By verifying that this equals $1/n$ through algebraic manipulation, and noting that the randomness in step $m+1$ (the coin flip and subsequent selection) is independent of the previous selections given the sufficient statistic of previous draws, we conclude that the $(m+1)$-th selection is uniformly distributed and independent of previous selections.

By induction, every selection is independently and uniformly distributed over the $n$ original balls, and the sampling can be repeated indefinitely as long as $m < n$ (or indefinitely in the limit).
\end{proof}

\subsection{Problem 15}

\subsubsection{Problem Statement}

\begin{theorem}[Angluin-Valiant]
Let $B$ denote a random bipartite graph with $n$ vertices in each of the vertex sets $U$ and $V$. Each possible edge, independently, is present with probability $p(n)$. Consider the following algorithm for constructing a perfect matching in such a random graph: Modify the Proposal Algorithm of Section 3.5 as follows. Each $u \in U$ can propose only to adjacent $v \in V$. A vertex $v \in V$ always accepts a proposal, and if a proposal causes a ``divorce,'' then the newly divorced $u \in U$ is the next to propose. The sampling procedure outlined in Problem 3.14 helps implement the Principle of Deferred Decisions. How small can you make the value of $p(n)$ and still have the algorithm succeed with high probability? The following fact concerning the degree $d(v)$ of a vertex $v$ in $B$ proves useful:
\[
\Pr[d(v) \leq (1-\beta)np] = O\left(e^{-\beta^2 np/2}\right).
\]
\end{theorem}

\subsubsection{GPT} 

\begin{proof}
We are tasked with analyzing the probability of success for the modified Proposal Algorithm in the random bipartite graph $B$ with $n$ vertices in each of the vertex sets $U$ and $V$. The edges of $B$ are present independently with probability $p(n)$, and the algorithm uses a procedure similar to the Proposal Algorithm with modifications as described in the statement. The goal is to determine how small the value of $p(n)$ can be such that the algorithm succeeds with high probability.

 Step 1: Definition of Success
For the algorithm to succeed, a perfect matching must be found. A perfect matching exists if and only if every vertex in $U$ is matched to a unique vertex in $V$. Since the algorithm proceeds by proposing edges, the key challenge is to ensure that every $u \in U$ is matched, which depends on the degrees of the vertices in $V$ and the matching procedure's ability to "stabilize" during the process of proposal and acceptance.

 Step 2: Degree Bound
The degree $d(v)$ of any vertex $v \in V$ is a random variable, and we are given the following bound on the probability that the degree of a vertex $v$ is less than or equal to $(1-\beta)np$:

\[
\Pr[d(v) \leq (1-\beta)np] = O\left(e^{-\beta^2 np / 2}\right).
\]

This bound suggests that for a sufficiently large $n$, the degree of each vertex in $V$ will likely be large enough to guarantee a high probability of finding a matching. Specifically, the probability that a vertex $v \in V$ has degree at most $(1-\beta)np$ decays exponentially as $n$ increases.

 Step 3: Ensuring a Large Enough Degree for Each Vertex in $V$
To ensure that the matching algorithm proceeds correctly, it is crucial that each vertex $v \in V$ has a sufficiently large degree. If $v$ has degree less than $(1-\beta)np$, it could fail to participate in a valid matching because it might not be able to receive a proposal from enough vertices in $U$. From the degree bound, we know that the probability of this event happening for any fixed vertex $v$ is:

\[
\Pr[d(v) \leq (1-\beta)np] = O\left(e^{-\beta^2 np / 2}\right).
\]

Therefore, the probability that all vertices in $V$ have degree at least $(1-\beta)np$ is given by:

\[
\Pr[\forall v \in V, d(v) \geq (1-\beta)np] = 1 - O\left(e^{-\beta^2 np / 2}\right).
\]

 Step 4: Algorithm Success and Matching Stability
To guarantee that the algorithm succeeds with high probability, we require that the vertices in $V$ have sufficient degree so that there are enough options for each vertex in $U$ to propose to, and the matching process stabilizes without encountering too many "divorces." This condition is satisfied when the probability that the degree of each vertex in $V$ is at least $(1-\beta)np$ is sufficiently high.

Thus, to achieve success with high probability, we need to ensure that the value of $p(n)$ is chosen such that:

\[
\Pr[\forall v \in V, d(v) \geq (1-\beta)np] = 1 - O\left(e^{-\beta^2 np / 2}\right) \quad \text{is high.}
\]

This implies that $p(n)$ must be large enough such that the expected degree of each vertex in $V$, given by $np(n)$, satisfies:

\[
np(n) \geq (1-\beta)np \quad \text{with high probability}.
\]

Solving for $p(n)$, we obtain:

\[
p(n) \geq \frac{(1-\beta)}{2}.
\]

 Step 5: Conclusion
By choosing $p(n)$ to be sufficiently large, specifically such that $p(n) \geq \frac{(1-\beta)}{2}$, the algorithm will succeed with high probability. This ensures that every vertex in $V$ has a sufficiently large degree to participate in the matching process, and the proposal algorithm will find a perfect matching. Therefore, the value of $p(n)$ can be made arbitrarily small, as long as it satisfies this condition.

\end{proof}

\subsubsection{Grok}

\subsubsection{Gemini}

\begin{proof}
Let $p(n) = \frac{c \ln n}{n}$ for a constant $c$ to be determined later. We analyze the algorithm's execution using the Principle of Deferred Decisions. The algorithm succeeds if a perfect matching is found before any vertex $u \in U$ exhausts its set of incident edges (neighbors).

Let $T$ denote the total number of proposals made by the algorithm until a perfect matching is constructed. A proposal consists of a vertex $u \in U$ selecting a neighbor $v \in V$. Due to the random nature of the graph $B$ and the deferred decision principle, the sequence of vertices in $V$ receiving proposals can be modeled as a sequence of independent samples drawn uniformly at random from $V$.

The process of constructing the matching corresponds to the Coupon Collector's Problem, where the goal is to "collect" all $n$ vertices of $V$. Let $M_k$ denote the size of the matching after $k$ successful proposals (proposals to vertices not currently matched). The probability that a random proposal increases the matching size from $j$ to $j+1$ is $p_j = \frac{n-j}{n}$. Let $X_j$ be the number of proposals required to increase the matching size from $j$ to $j+1$. Then $X_j$ follows a geometric distribution with parameter $p_j$. The total number of proposals is $T = \sum_{j=0}^{n-1} X_j$. The expected number of proposals is:
\[
\mathbb{E}[T] = \sum_{j=0}^{n-1} \frac{n}{n-j} = n \sum_{k=1}^{n} \frac{1}{k} = n H_n \approx n \ln n.
\]
Using standard tail bounds for the Coupon Collector problem, for any constant $k_1 > 0$, there exists a constant $c_1$ such that with high probability (specifically $1 - O(n^{-k_1})$):
\[
T \leq c_1 n \ln n.
\]

Next, we consider the load on a specific vertex $u \in U$. Let $L_u$ be the number of proposals made by vertex $u$. By the symmetry of the random graph and the algorithm, the total number of proposals $T$ is distributed uniformly among the $n$ vertices of $U$. Conditioned on $T$, the distribution of $L_u$ is binomial $B(T, 1/n)$. Assuming $T \leq c_1 n \ln n$, the expected load is $\mathbb{E}[L_u] \leq c_1 \ln n$. Applying the Chernoff bound for the binomial distribution, for any $\delta > 0$, the probability that $L_u$ deviates significantly from its mean decreases exponentially. Specifically, we can choose a constant $c_2 > c_1$ such that:
\[
\Pr[L_u \geq c_2 \ln n] \leq n^{-k_2}
\]
for some constant $k_2 > 1$. Taking a union bound over all $u \in U$, we have that $L_u \leq c_2 \ln n$ for all $u \in U$ with high probability.

Finally, we must ensure that each vertex $u$ has a sufficient degree $d(u)$ to support $L_u$ proposals. That is, we require $d(u) \geq L_u$. The expected degree of a vertex is $\mathbb{E}[d(u)] = np = c \ln n$. We use the provided concentration bound:
\[
\Pr[d(u) \leq (1-\beta)np] = O\left(e^{-\beta^2 np/2}\right).
\]
Substituting $np = c \ln n$:
\[
\Pr[d(u) \leq (1-\beta)c \ln n] = O\left(e^{-\beta^2 c (\ln n)/2}\right) = O\left(n^{-\beta^2 c / 2}\right).
\]
To ensure the algorithm succeeds with high probability, we select $c$ and $\beta$ such that the lower bound on the degree exceeds the upper bound on the load for all vertices. We require:
1. $(1-\beta)c \ln n \geq c_2 \ln n \implies (1-\beta)c \geq c_2$.
2. The failure probability $n^{-\beta^2 c / 2}$ is sufficiently small (e.g., $o(n^{-1})$) to apply the union bound over $n$ vertices.

By choosing $c$ sufficiently large (and $\beta$ appropriately small, e.g., $\beta = 1/2$), we can satisfy $(1-\beta)c > c_2$ and $\frac{\beta^2 c}{2} > 2$. Under these conditions, with high probability:
\[
d(u) \geq (1-\beta)c \ln n \geq c_2 \ln n \geq L_u \quad \forall u \in U.
\]
Thus, every vertex $u$ has enough incident edges to perform the necessary proposals. We conclude that we can make $p(n)$ as small as:
\[
p(n) = \frac{c \ln n}{n}
\]
for a suitable constant $c$, and the algorithm will succeed with high probability.
\end{proof}

\subsubsection{Claude} 

\begin{proof}
We analyze the modified Proposal Algorithm on the random bipartite graph $B$ with vertex sets $U$ and $V$, each of size $n$, where each edge appears independently with probability $p(n)$.

\textbf{Algorithm Analysis:} The algorithm maintains a partial matching that grows over time. Each unmatched vertex $u \in U$ proposes to an adjacent vertex $v \in V$ that has not yet rejected $u$. If $v$ is unmatched, it accepts. If $v$ is already matched to some $u'$, then $v$ divorces $u'$ and matches with $u$ instead, and $u'$ becomes the next proposer.

\textbf{Key Observation:} For the algorithm to succeed in finding a perfect matching, we require that throughout the execution, every unmatched vertex $u \in U$ has at least one neighbor in $V$ to which it has not yet proposed. Once all vertices in $U$ are matched, we have a perfect matching.

\textbf{Critical Event:} The algorithm fails if and only if some vertex $u \in U$ exhausts all its neighbors before becoming matched. Let us analyze when this can occur.

During the algorithm's execution, each vertex $u \in U$ proposes to its neighbors in $V$ sequentially. The worst case occurs when $u$ must propose to all its neighbors. Let $d(u)$ denote the degree of $u$ in $B$.

\textbf{Proposal Bound:} A vertex $u \in U$ can potentially need to propose to all its neighbors if those neighbors keep getting matched to other vertices. However, the total number of proposals made by all vertices in $U$ during the algorithm is at most $n$ times the maximum number of proposals any single vertex makes before all vertices are matched.

\textbf{Probabilistic Analysis:} For the algorithm to succeed with high probability, we need each vertex $u \in U$ to have sufficiently many neighbors. Specifically, we want $\Pr[d(u) < k]$ to be small for an appropriate threshold $k$.

Using the given tail bound, for any vertex $v$ (and by symmetry, for any vertex $u$), we have
\[
\Pr[d(v) \leq (1-\beta)np] = O\left(e^{-\beta^2 np/2}\right).
\]

Setting $\beta = 1/2$, we obtain
\[
\Pr[d(v) \leq np/2] = O\left(e^{-np/8}\right).
\]

For the algorithm to succeed, we need all $n$ vertices in $U$ to have degree at least $np/2$. By the union bound over all $n$ vertices in $U$, the probability that some vertex has degree less than $np/2$ is at most
\[
n \cdot O\left(e^{-np/8}\right) = O\left(n \cdot e^{-np/8}\right).
\]

\textbf{Threshold Determination:} For this probability to go to zero as $n \to \infty$ (i.e., for success with high probability), we require
\[
n \cdot e^{-np/8} \to 0.
\]

This is equivalent to requiring $e^{-np/8} = o(1/n)$, which means $np/8 = \omega(\log n)$, or equivalently
\[
p(n) = \omega\left(\frac{\log n}{n}\right).
\]

\textbf{Sufficiency Argument:} When $p(n) = c\frac{\log n}{n}$ for a sufficiently large constant $c$, each vertex in $U$ has degree at least $\Theta(\log n)$ with high probability. During the algorithm, even in the worst case where a vertex must propose many times, the number of proposals is bounded by properties of the stable matching structure. Since each vertex has $\Theta(\log n)$ neighbors and the total execution involves at most $n$ matchings and re-matchings, vertices do not exhaust their neighbor lists with high probability.

\textbf{Conclusion:} The algorithm succeeds with high probability when
\[
p(n) = \Theta\left(\frac{\log n}{n}\right).
\]

More precisely, for any $\epsilon > 0$, taking $p(n) = (1+\epsilon)\frac{\log n}{n}$ ensures that $np = (1+\epsilon)\log n$, making the failure probability $O(n \cdot e^{-c\log n}) = O(n^{1-c})$ for appropriate constant $c$, which vanishes as $n \to \infty$. Thus, the minimum value of $p(n)$ for which the algorithm succeeds with high probability is $p(n) = \Theta(\log n / n)$.
\end{proof}
\section{Problems in Chapter 4}\label{app:problems_chapter4}







\subsection{Problem 1}

\subsubsection{Problem Statement}

\begin{lemma}
Let $0 < a, \epsilon, \delta < 1$ be given constants. Suppose we are given a biased coin with $\Pr[\mathrm{HEADS}] = p$, where it is promised that $p \ge a$. Consider the following procedure: flip the coin $N$ times independently, and let $\wt{p}$ be any estimator of $p$ based on the outcomes of these $N$ flips. We ask for $N$ to be large enough so that 
\begin{align*}
    \Pr(|\wt{p} - p| > \epsilon p ) < \delta \quad\mathrm{for\ all}\quad p \ge a.
\end{align*}
Then there exists a smallest integer $N$ for which this guaranty can be achieved.
\end{lemma}

\subsubsection{GPT} 
\begin{proof}
Fix $0<a,\epsilon,\delta<1$. For each integer $N\ge1$ and each $p\ge a$, let
\[
E_{N,p} = \Pr\bigl(|\widetilde{p}-p|>\epsilon p\bigr),
\]
where $\widetilde{p}$ is any estimator based on $N$ independent coin flips with success probability $p$. For each $p\ge a$, the distribution of $\widetilde{p}$ depends only on $N$ and $p$, and hence $E_{N,p}$ is well-defined. For each $N$, define
\[
E_N = \sup_{p\ge a} E_{N,p}.
\]

Note that $E_N$ is nonincreasing in $N$. Indeed, consider two integers $N_1<N_2$ and fix $p\ge a$. One may regard the procedure with $N_1$ flips as the first $N_1$ of the $N_2$ flips; any estimator for $p$ based on $N_2$ samples can simulate an estimator based on $N_1$ samples by ignoring the last $N_2-N_1$ samples. Consequently, for each $p\ge a$, increasing the number of samples cannot increase the probability of deviation, and thus $E_{N_2,p}\le E_{N_1,p}$. Taking the supremum over all $p\ge a$ gives $E_{N_2}\le E_{N_1}$.

We next show that there exists at least one $N$ such that $E_N<\delta$. Consider the empirical mean estimator $\widehat{p}_N$, the average of the $N$ outcomes. For any $p\ge a$, apply a standard Chernoff bound to obtain
\[
\Pr\bigl(|\widehat{p}_N - p| > \epsilon p\bigr) \le 2\exp(-c\,\epsilon^2 p N)
\]
for some absolute constant $c>0$. Since $p\ge a$, this implies
\[
E_{N,p} \le 2\exp(-c\,\epsilon^2 a N).
\]
Hence
\[
E_N = \sup_{p\ge a} E_{N,p} \le 2\exp(-c\,\epsilon^2 a N).
\]
Choosing $N$ sufficiently large so that $2\exp(-c\,\epsilon^2 a N)<\delta$ ensures $E_N<\delta$. Therefore the set
\[
S = \{N\in\mathbb{N} : E_N<\delta\}
\]
is nonempty.

Finally, since $E_N$ is nonincreasing in $N$, $S$ is an upper tail of $\mathbb{N}$. Because $\mathbb{N}$ is well-ordered, every nonempty subset of $\mathbb{N}$ has a smallest element. Thus $S$ contains a smallest integer $N$. This $N$ is the smallest sample size for which
\[
\Pr(|\widetilde{p}-p|>\epsilon p)<\delta\quad\text{for all }p\ge a.
\]
\end{proof}

\subsubsection{Grok} 
\begin{proof}
Let $\wt{p} = \frac{X}{N}$, where $X \sim \mathrm{Bin}(N, p)$ is the number of heads in $N$ independent coin flips. The goal is to show that there exists a smallest positive integer $N$ such that $\Pr(|\wt{p} - p| > \epsilon p) < \delta$ for all $p \geq a$.

By the multiplicative Chernoff bound, for any $\eta > 0$,
\begin{align*}
\Pr(X > (1 + \eta) N p) &\leq \exp\left(-\frac{\eta^2 N p}{3}\right), \\
\Pr(X < (1 - \eta) N p) &\leq \exp\left(-\frac{\eta^2 N p}{2}\right).
\end{align*}
Setting $\eta = \epsilon$, and noting that $0 < \epsilon < 1$ ensures $1 - \epsilon > 0$, it follows that
\begin{align*}
\Pr(|\wt{p} - p| > \epsilon p) &= \Pr(X > (1 + \epsilon) N p) + \Pr(X < (1 - \epsilon) N p) \\
&\leq \exp\left(-\frac{\epsilon^2 N p}{3}\right) + \exp\left(-\frac{\epsilon^2 N p}{2}\right) \\
&\leq 2 \exp\left(-\frac{\epsilon^2 N p}{3}\right).
\end{align*}
Since $p \geq a > 0$, the bound is maximized when $p = a$, yielding
\[
\Pr(|\wt{p} - p| > \epsilon p) \leq 2 \exp\left(-\frac{\epsilon^2 N a}{3}\right)
\]
for all $p \geq a$. To ensure this is less than $\delta$, solve
\[
2 \exp\left(-\frac{\epsilon^2 N a}{3}\right) < \delta,
\]
which rearranges to
\[
N > \frac{3}{\epsilon^2 a} \ln\left(\frac{2}{\delta}\right).
\]
Let $N_0 = \left\lceil \frac{3}{\epsilon^2 a} \ln\left(\frac{2}{\delta}\right) \right\rceil + 1$. For all $N \geq N_0$, the inequality holds uniformly for all $p \geq a$. Thus, the set of positive integers $N$ satisfying the condition is non-empty and bounded below, so it has a smallest element.
\end{proof}

\subsubsection{Gemini}
\begin{proof}
Let $S$ be the set of all positive integers $N$ such that there exists an estimator $\wt{p}$ (a function of the outcomes of $N$ coin flips) satisfying the condition
\begin{align*}
    \Pr(|\wt{p} - p| > \epsilon p ) < \delta \quad \mathrm{for\ all}\quad p \ge a.
\end{align*}
We aim to show that $S$ is non-empty. If $S$ is non-empty, then as $S \subseteq \mathbb{Z}^+$, by the Well-Ordering Principle, $S$ must contain a smallest element.

Consider the standard empirical mean estimator $\wt{p}_N = \frac{1}{N} \sum_{i=1}^N X_i$, where $X_i \sim \mathrm{Bernoulli}(p)$ are the independent outcomes of the $N$ flips. Let $X = \sum_{i=1}^N X_i = N \wt{p}_N$. The expected value is $\mathbb{E}[X] = Np$.
The condition $|\wt{p}_N - p| > \epsilon p$ is equivalent to $|X - Np| > \epsilon Np$, or $|X - \mathbb{E}[X]| > \epsilon \mathbb{E}[X]$.

We apply a standard multiplicative Chernoff bound. For any $\epsilon \in (0, 1)$,
\begin{align*}
    \Pr(|X - \mathbb{E}[X]| > \epsilon \mathbb{E}[X]) \le 2 \exp\left( -\frac{\epsilon^2 \mathbb{E}[X]}{3} \right).
\end{align*}
Substituting $\mathbb{E}[X] = Np$, we have
\begin{align*}
    \Pr(|\wt{p}_N - p| > \epsilon p) \le 2 \exp\left( -\frac{\epsilon^2 Np}{3} \right).
\end{align*}
We are given that $p \ge a > 0$. Since the exponent is negative, the probability bound is monotonically decreasing in $p$. Thus, the bound is weakest (largest) when $p$ is at its minimum value, $a$.
\begin{align*}
    \Pr(|\wt{p}_N - p| > \epsilon p) \le 2 \exp\left( -\frac{\epsilon^2 Np}{3} \right) \le 2 \exp\left( -\frac{\epsilon^2 Na}{3} \right) \quad \mathrm{for\ all}\ p \ge a.
\end{align*}
We require this probability to be less than $\delta$:
\begin{align*}
    2 \exp\left( -\frac{\epsilon^2 Na}{3} \right) < \delta.
\end{align*}
Rearranging to solve for $N$:
\begin{align*}
    \exp\left( -\frac{\epsilon^2 Na}{3} \right) &< \frac{\delta}{2} \\
    -\frac{\epsilon^2 Na}{3} &< \ln\left( \frac{\delta}{2} \right) \\
    \frac{\epsilon^2 Na}{3} &> -\ln\left( \frac{\delta}{2} \right) = \ln\left( \frac{2}{\delta} \right) \\
    N &> \frac{3}{\epsilon^2 a} \ln\left( \frac{2}{\delta} \right).
\end{align*}
Since $0 < a, \epsilon, \delta < 1$, the quantity $C = \frac{3}{\epsilon^2 a} \ln\left( \frac{2}{\delta} \right)$ is a finite positive constant.
Let $N_0 = \left\lfloor C \right\rfloor + 1$. Then $N_0$ is a positive integer such that $N_0 > C$. For $N = N_0$, the estimator $\wt{p}_{N_0}$ satisfies the required condition.
Therefore, $N_0 \in S$, and the set $S$ is non-empty.

Since $S$ is a non-empty subset of the positive integers $\mathbb{Z}^+$, by the Well-Ordering Principle, $S$ must have a least element. This least element is the smallest integer $N$ for which the guarantee can be achieved.
\end{proof}

\subsubsection{Claude}
\begin{proof}
We prove the existence of a smallest integer $N$ satisfying the desired guarantee by establishing that the set of valid values of $N$ is nonempty and bounded below.

Let $X_1, X_2, \ldots$ be independent Bernoulli random variables with $\Pr[X_i = 1] = p$, representing the coin flips. For any positive integer $N$, define the estimator $\wt{p}_N = \frac{1}{N}\sum_{i=1}^N X_i$, the sample mean.

We first establish that there exists at least one value of $N$ for which the guarantee can be achieved. By the law of large numbers and concentration inequalities, specifically Hoeffding's inequality, we have for any $p \in [a,1]$ and any $N \ge 1$,
\begin{align*}
\Pr(|\wt{p}_N - p| > \epsilon p) \le 2\exp\left(-2N\epsilon^2 p^2\right) \le 2\exp\left(-2N\epsilon^2 a^2\right),
\end{align*}
where the second inequality follows from the monotonicity of the exponential function and the constraint $p \ge a$.

Observe that the right-hand side depends only on $N, \epsilon, a$ and is independent of $p$. Setting
\begin{align*}
2\exp\left(-2N\epsilon^2 a^2\right) < \delta,
\end{align*}
we obtain
\begin{align*}
N > \frac{\ln(2/\delta)}{2\epsilon^2 a^2}.
\end{align*}

Define $N_0 = \left\lceil \frac{\ln(2/\delta)}{2\epsilon^2 a^2} \right\rceil$. Then for all $N \ge N_0$ and for all $p \ge a$, the estimator $\wt{p}_N$ satisfies
\begin{align*}
\Pr(|\wt{p}_N - p| > \epsilon p) < \delta.
\end{align*}

This establishes that the set
\begin{align*}
\mathcal{S} = \left\{N \in \mathbb{N} : \text{there exists an estimator } \wt{p} \text{ such that } \Pr(|\wt{p} - p| > \epsilon p) < \delta \text{ for all } p \ge a\right\}
\end{align*}
is nonempty, since $N_0 \in \mathcal{S}$.

Furthermore, $\mathcal{S} \subseteq \mathbb{N}$ implies that $\mathcal{S}$ is bounded below by $1$. Since $\mathcal{S}$ is a nonempty subset of the natural numbers, by the well-ordering principle, $\mathcal{S}$ contains a smallest element.

Denote this smallest element by $N^*$. Then $N^*$ is the smallest integer for which the guarantee can be achieved, completing the proof.
\end{proof}

\subsection{Problem 2}

\subsubsection{Problem Statement}
\begin{lemma}
Let $X$ be a random variable, and let $E[X^{!k}]$ denote the $k$th factorial moment of $X$, where 
\begin{align*}
    X^{!k} = X(X-1)\cdots(X-k+1).
\end{align*}
Let $G_1$ be a random graph on $n$ vertices in which each edge is present independently with probability $p$, and let $G_2$ be a random graph on $n$ vertices chosen uniformly from all graphs with exactly $m$ edges. Let $X_n$ be the number of isolated vertices in $G_1$, and let $Y_n$ be the number of isolated vertices in $G_2$. Assume that
\begin{align*}
    p = \frac{\log n + c}{n}\quad \mathrm{and} \quad m = \frac{n(\log n + c)}{2}
\end{align*}
for a real value $c$. Then the factorial moments $E[X_n^{!k}]$ and $E[Y_n^{!k}]$ are asymptotically equal to $\lambda^k$, where $\lambda = e^{-c}$.
\end{lemma}

\subsubsection{GPT} 
\begin{proof}
Fix an integer $k\ge 1$. We treat $G_1$ and $G_2$ separately and show in each case
\[
E[\cdot^{!k}]\to \lambda^k,\qquad \lambda=e^{-c}.
\]

\paragraph{Case $G_1$ (binomial random graph).}
Let $X_n$ be the number of isolated vertices in $G_1\sim G(n,p)$ with $p=(\log n + c)/n$.
For a fixed ordered $k$--tuple of distinct vertices the event that all $k$ are isolated is that no edge incident to any of these vertices appears. The number of distinct edges incident to the $k$ vertices equals
\[
T \;=\; k(n-1)-\binom{k}{2}.
\]
Thus for any ordered $k$--tuple
\[
\Pr\{\text{all $k$ vertices isolated}\}=(1-p)^{T}.
\]
Hence
\[
E[X_n^{!k}]=n(n-1)\cdots(n-k+1)\,(1-p)^{T}.
\]
We estimate asymptotically. Use the expansion $\log(1-p)=-p+O(p^2)$ as $p\to0$. Since $p=(\log n + c)/n$ we have $p\to0$ and $p^2T=O\big((\log^2 n)/n^2\cdot n\big)=o(1)$. Therefore
\[
(1-p)^{T}=\exp\big(T\log(1-p)\big)=\exp\big(-pT+o(1)\big).
\]
Compute
\[
pT = p\big(k(n-1)-\tfrac{k(k-1)}{2}\big)=k(\log n + c)+o(1),
\]
since $p(n-1)=\log n + c + o(1)$ and $p\binom{k}{2}=O(p)=o(1)$. Hence
\[
(1-p)^{T}=n^{-k}e^{-ck}\big(1+o(1)\big).
\]
Also
\[
n(n-1)\cdots(n-k+1)=n^k\big(1+o(1)\big).
\]
Multiplying yields
\[
E[X_n^{!k}]=n^k\big(1+o(1)\big)\cdot n^{-k}e^{-ck}\big(1+o(1)\big)=e^{-ck}\big(1+o(1)\big)=\lambda^k\big(1+o(1)\big).
\]

\paragraph{Case $G_2$ (uniform $m$--edge graph).}
Let $M=\binom{n}{2}$ and $m=\dfrac{n(\log n + c)}{2}$. For a fixed ordered $k$--tuple of distinct vertices the number of graphs with $m$ edges in which all these $k$ vertices are isolated equals the number of ways to choose $m$ edges from the $M-T$ edges that do not touch the $k$ vertices. Thus
\[
\Pr\{\text{given ordered $k$--tuple all isolated}\}=\frac{\binom{M-T}{m}}{\binom{M}{m}}.
\]
Therefore
\[
E[Y_n^{!k}]=n(n-1)\cdots(n-k+1)\,\frac{\binom{M-T}{m}}{\binom{M}{m}}.
\]
We estimate the ratio. Write it as a product
\[
\frac{\binom{M-T}{m}}{\binom{M}{m}}=\prod_{i=0}^{m-1}\frac{M-T-i}{M-i}=\prod_{i=0}^{m-1}\Big(1-\frac{T}{M-i}\Big).
\]
Note $M=\Theta(n^2)$ and $T=k(n-1)-\binom{k}{2}=\Theta(n)$. Hence for $0\le i\le m-1$ we have $T/(M-i)=O(1/n)$ uniformly. Use $\log(1-x)=-x+O(x^2)$ for $x=o(1)$ to obtain
\[
\log\frac{\binom{M-T}{m}}{\binom{M}{m}}=\sum_{i=0}^{m-1}\log\Big(1-\frac{T}{M-i}\Big)
= -\sum_{i=0}^{m-1}\frac{T}{M-i} + O\Big(\sum_{i=0}^{m-1}\frac{T^2}{(M-i)^2}\Big).
\]
The error term is
\[
\sum_{i=0}^{m-1}\frac{T^2}{(M-i)^2}\le m\cdot O\Big(\frac{n^2}{n^4}\Big)=O\Big(\frac{m}{n^2}\Big)=o(1),
\]
since $m=O(n\log n)$. For the main sum note $M-i=M(1+o(1))$ uniformly in $i$ because $m\ll M$. Thus
\[
\sum_{i=0}^{m-1}\frac{T}{M-i}=\frac{mT}{M}\big(1+o(1)\big).
\]
Compute
\[
\frac{m}{M}=\frac{\frac{n(\log n + c)}{2}}{\binom{n}{2}}=\frac{\log n + c}{n}\big(1+o(1)\big)=p(1+o(1)).
\]
Therefore
\[
\log\frac{\binom{M-T}{m}}{\binom{M}{m}}=-pT+o(1).
\]
As in the $G_1$ case $pT=k(\log n + c)+o(1)$. Hence
\[
\frac{\binom{M-T}{m}}{\binom{M}{m}}=n^{-k}e^{-ck}\big(1+o(1)\big).
\]
Multiplying by $n(n-1)\cdots(n-k+1)=n^k(1+o(1))$ yields
\[
E[Y_n^{!k}]=e^{-ck}\big(1+o(1)\big)=\lambda^k\big(1+o(1)\big).
\]

Combining the two cases proves that for fixed $k$,
\[
E[X_n^{!k}]\to \lambda^k\quad\text{and}\quad E[Y_n^{!k}]\to \lambda^k,
\]
as $n\to\infty$, with $\lambda=e^{-c}$. This completes the proof.
\end{proof}

\subsubsection{Grok} 
\begin{proof}
Fix a positive integer $k$. Let $S$ be a fixed set of $k$ vertices, and let $P_1$ be the probability that all vertices in $S$ are isolated in $G_1$. The edges that must be absent are those with at least one endpoint in $S$, and there are exactly $\frac{k(2n-k-1)}{2}$ such possible edges. Thus,
\[
P_1 = (1-p)^{\frac{k(2n-k-1)}{2}}.
\]
Now,
\[
\log P_1 = \frac{k(2n-k-1)}{2} \log(1-p).
\]
Since $p = \frac{\log n + c}{n}$,
\[
\log(1-p) = -p - \frac{p^2}{2} - \frac{p^3}{3} - \cdots = -p \left(1 + O\left(\frac{\log n}{n}\right)\right),
\]
so
\[
\log P_1 = -\frac{k(2n-k-1)}{2} p \left(1 + O\left(\frac{\log n}{n}\right)\right) = -k(\log n + c) \left(1 + O\left(\frac{k + \log n}{n}\right)\right) = -k \log n - kc + O\left(\frac{k \log n}{n}\right).
\]
Hence,
\[
P_1 = n^{-k} e^{-kc} \exp\left(O\left(\frac{k \log n}{n}\right)\right) = n^{-k} e^{-kc} \left(1 + O\left(\frac{k \log n}{n}\right)\right).
\]
The $k$th factorial moment is
\[
E[X_n^{!k}] = k! \binom{n}{k} P_1 = n^k e^{-kc} \left(1 + O\left(\frac{k \log n}{n}\right)\right) \prod_{j=1}^{k-1} \left(1 - \frac{j}{n}\right) = e^{-kc} \left(1 + O\left(\frac{k^2 + k \log n}{n}\right)\right).
\]
Thus, $E[X_n^{!k}] \to e^{-kc} = \lambda^k$ as $n \to \infty$.

Now consider $G_2$. Let $B = \binom{n}{2}$ and $A = \binom{n-k}{2}$, so $D = B - A = \frac{k(2n-k-1)}{2}$. Let $P_2$ be the probability that all vertices in $S$ are isolated in $G_2$. Then
\[
P_2 = \frac{\binom{A}{m}}{\binom{B}{m}} = \prod_{j=0}^{m-1} \frac{A - j}{B - j} = \exp\left( \sum_{j=0}^{m-1} \log\left(1 - \frac{D}{B - j}\right) \right).
\]
Now,
\[
\log\left(1 - \frac{D}{B - j}\right) = -\frac{D}{B - j} - \frac{1}{2} \left(\frac{D}{B - j}\right)^2 - \cdots.
\]
Since $\frac{D}{B - j} = \frac{D}{B} \left(1 + \frac{j}{B} + O\left(\left(\frac{j}{B}\right)^2\right)\right)$ and $\frac{m}{B} = O\left(\frac{\log n}{n}\right)$,
\[
\sum_{j=0}^{m-1} \frac{D}{B - j} = \frac{D}{B} m + \frac{D}{B^2} \cdot \frac{m^2}{2} + O\left(\frac{D m^3}{B^3}\right) = \frac{D m}{B} + O\left(\frac{k (\log n)^2}{n}\right),
\]
and
\[
\sum_{j=0}^{m-1} \left(\frac{D}{B - j}\right)^2 = O\left(\frac{D^2 m}{B^2}\right) = O\left(\frac{k^2 \log n}{n}\right).
\]
Higher-order terms are smaller. Since $\frac{m}{B} = \frac{\log n + c}{n-1}$,
\[
\frac{D m}{B} = k(\log n + c) \left(1 + O\left(\frac{k + 1}{n}\right)\right),
\]
so
\[
P_2 = n^{-k} e^{-kc} \left(1 + O\left(\frac{k \log n}{n}\right)\right).
\]
Thus,
\[
E[Y_n^{!k}] = k! \binom{n}{k} P_2 \to \lambda^k
\]
as $n \to \infty$, similarly to the case for $X_n$.
\end{proof}

\subsubsection{Gemini}
\begin{proof}
Let $\lambda = e^{-c}$. We first prove the result for $G_1 = G(n, p)$, and then for $G_2 = G(n, m)$.

\textbf{Part 1: The model $G_1 = G(n, p)$}

Let $V = \{v_1, \dots, v_n\}$ be the set of vertices. For each $v \in V$, let $I_v$ be the indicator random variable that $v$ is an isolated vertex. Then $X_n = \sum_{v \in V} I_v$.
The $k$th factorial moment of $X_n$ is given by
\begin{align*}
    E[X_n^{!k}] = E\left[\sum_{(v_1, \dots, v_k) \text{ distinct}} I_{v_1} \cdots I_{v_k}\right] = \sum_{(v_1, \dots, v_k) \text{ distinct}} P(I_{v_1}=1, \dots, I_{v_k}=1)
\end{align*}
The number of terms in the summation is the number of ordered $k$-tuples of distinct vertices, which is $n(n-1)\cdots(n-k+1) = n^{!k}$.
For a fixed set $S = \{v_1, \dots, v_k\}$ of $k$ distinct vertices, the event $\{I_{v_1}=1, \dots, I_{v_k}=1\}$ occurs if and only if all vertices in $S$ are isolated. This means there are no edges among the vertices in $S$ and no edges between any vertex in $S$ and any vertex in $V \setminus S$.
The total number of potential edges with at least one endpoint in $S$ is the number of edges in $K_n$ minus the number of edges in $K_{n-k}$ (on $V \setminus S$), which is
\begin{align*}
    \binom{n}{2} - \binom{n-k}{2} = \frac{n(n-1)}{2} - \frac{(n-k)(n-k-1)}{2} = kn - \binom{k+1}{2} = kn - \frac{k(k+1)}{2}
\end{align*}
Since edges are present independently with probability $p$, the probability that none of these edges are present is
\begin{align*}
    P(I_{v_1}=1, \dots, I_{v_k}=1) = (1-p)^{kn - k(k+1)/2}
\end{align*}
This probability is the same for all $k$-tuples of distinct vertices. Thus,
\begin{align*}
    E[X_n^{!k}] = n^{!k} (1-p)^{kn - k(k+1)/2}
\end{align*}
As $n \to \infty$, for fixed $k$, $n^{!k} = n^k(1 - \frac{k(k-1)}{2n} + O(n^{-2})) = n^k(1+O(n^{-1}))$.
So, $E[X_n^{!k}] \sim n^k (1-p)^{kn - k(k+1)/2}$.
We are given $p = \frac{\log n + c}{n}$. Since $p \to 0$ as $n \to \infty$, we use the Taylor expansion $\log(1-p) = -p - \frac{p^2}{2} - O(p^3)$.
\begin{align*}
    \log(1-p) = -\frac{\log n + c}{n} - \frac{1}{2}\left(\frac{\log n + c}{n}\right)^2 - O\left(\frac{(\log n)^3}{n^3}\right) = -\frac{\log n + c}{n} - O\left(\frac{(\log n)^2}{n^2}\right)
\end{align*}
Now we analyze the exponent term:
\begin{align*}
    \left(kn - \frac{k(k+1)}{2}\right) \log(1-p) &= \left(kn - O(1)\right) \left(-\frac{\log n + c}{n} - O\left(\frac{(\log n)^2}{n^2}\right)\right) \\
    &= -k(\log n + c) - O\left(\frac{(\log n)^2}{n}\right) + O\left(\frac{\log n}{n}\right) + O\left(\frac{(\log n)^2}{n^2}\right) \\
    &= -k\log n - kc + o(1)
\end{align*}
Therefore,
\begin{align*}
    (1-p)^{kn - k(k+1)/2} = \exp(-k\log n - kc + o(1)) = e^{-k\log n} e^{-kc} e^{o(1)} = n^{-k} e^{-kc} (1+o(1))
\end{align*}
Substituting this back into the expression for $E[X_n^{!k}]$,
\begin{align*}
    E[X_n^{!k}] \sim n^k \left( n^{-k} e^{-kc} (1+o(1)) \right) = e^{-kc} (1+o(1)) = (e^{-c})^k (1+o(1))
\end{align*}
Thus, $\lim_{n \to \infty} E[X_n^{!k}] = (e^{-c})^k = \lambda^k$.

\textbf{Part 2: The model $G_2 = G(n, m)$}

Let $J_v$ be the indicator that $v$ is isolated in $G_2$. Then $Y_n = \sum_{v \in V} J_v$.
As before, $E[Y_n^{!k}] = n^{!k} P(J_{v_1}=1, \dots, J_{v_k}=1)$ for any distinct set $S = \{v_1, \dots, v_k\}$.
The total number of graphs on $n$ vertices with $m$ edges is $\binom{N}{m}$, where $N = \binom{n}{2}$.
The event $\{J_{v_1}=1, \dots, J_{v_k}=1\}$ occurs if all $m$ edges fall within the subgraph induced by $V \setminus S$. The number of possible edges in this subgraph is $N_k = \binom{n-k}{2}$.
Thus, the number of such graphs is $\binom{N_k}{m}$.
The probability of this event is
\begin{align*}
    P(J_S=1) = \frac{\binom{N_k}{m}}{\binom{N}{m}} = \frac{N_k! (N-m)!}{N! (N_k-m)!} = \prod_{i=0}^{m-1} \frac{N_k-i}{N-i}
\end{align*}
We analyze the ratio $\frac{N_k}{N}$. We have $N = \frac{n(n-1)}{2}$ and $N_k = \frac{(n-k)(n-k-1)}{2}$.
\begin{align*}
    \frac{N_k}{N} = \frac{(n-k)(n-k-1)}{n(n-1)} = \frac{n^2 - (2k+1)n + k(k+1)}{n^2 - n} = \frac{1 - (2k+1)/n + O(n^{-2})}{1 - 1/n} \\
    = \left(1 - \frac{2k+1}{n} + O(n^{-2})\right) \left(1 + \frac{1}{n} + O(n^{-2})\right) = 1 - \frac{2k}{n} + O(n^{-2})
\end{align*}
For $0 \le i < m$, we have $i = O(m) = O(n \log n)$, and $N = \Theta(n^2)$, $N_k = \Theta(n^2)$.
\begin{align*}
    \frac{N_k-i}{N-i} = \frac{N_k(1-i/N_k)}{N(1-i/N)} = \frac{N_k}{N} \left(1 - \frac{i}{N_k} + \frac{i}{N} + O\left(\frac{i^2}{N^2}\right) \right) \\
    = \frac{N_k}{N} \left(1 + i\left(\frac{1}{N} - \frac{1}{N_k}\right) + O\left(\frac{m^2}{n^4}\right) \right)
\end{align*}
Since $\frac{1}{N} - \frac{1}{N_k} = \frac{N_k-N}{N N_k} = \frac{(-kn + O(1))}{\Theta(n^4)} = O(n^{-3})$,
\begin{align*}
    \frac{N_k-i}{N-i} = \frac{N_k}{N} \left(1 + O(m n^{-3}) \right) = \frac{N_k}{N} \left(1 + O\left(\frac{\log n}{n^2}\right) \right)
\end{align*}
Then $P(J_S=1) = \prod_{i=0}^{m-1} \left[ \frac{N_k}{N} \left(1 + O\left(\frac{\log n}{n^2}\right) \right) \right] = \left(\frac{N_k}{N}\right)^m \left(1 + O\left(\frac{\log n}{n^2}\right) \right)^m$.
The second term is $\exp\left(m \log\left(1 + O\left(\frac{\log n}{n^2}\right)\right)\right) = \exp\left(O(n \log n) \cdot O\left(\frac{\log n}{n^2}\right)\right) = \exp\left(O\left(\frac{(\log n)^2}{n}\right)\right) = 1+o(1)$.
So, $P(J_S=1) \sim \left(\frac{N_k}{N}\right)^m \sim \left(1 - \frac{2k}{n}\right)^m$.
We are given $m = \frac{n(\log n + c)}{2}$.
\begin{align*}
    \log P(J_S=1) &\sim m \log\left(1 - \frac{2k}{n}\right) \\
    &= \frac{n(\log n + c)}{2} \left( -\frac{2k}{n} - \frac{1}{2}\left(\frac{2k}{n}\right)^2 - \dots \right) \\
    &= \frac{n(\log n + c)}{2} \left( -\frac{2k}{n} - O(n^{-2}) \right) \\
    &= -k(\log n + c) - O\left(\frac{\log n}{n}\right) = -k\log n - kc + o(1)
\end{align*}
Thus, $P(J_S=1) \sim \exp(-k\log n - kc) = n^{-k} e^{-kc}$.
Finally,
\begin{align*}
    E[Y_n^{!k}] = n^{!k} P(J_S=1) \sim n^k \left( n^{-k} e^{-kc} \right) = e^{-kc} = (e^{-c})^k = \lambda^k
\end{align*}
This completes the proof for $G_2$. Both models yield the same asymptotic limit $\lambda^k$ for the $k$th factorial moment.
\end{proof}

\subsubsection{Claude} 
\begin{proof}
We first compute the factorial moments for $X_n$, the number of isolated vertices in $G_1$.

For any vertex $v$ in $G_1$, the probability that $v$ is isolated is $(1-p)^{n-1}$ since $v$ must not be connected to any of the other $n-1$ vertices. For $k$ distinct vertices $v_1, \ldots, v_k$, these vertices are all isolated if and only if none of them is connected to any vertex outside this set, and no two of them are connected to each other. The number of edges that must be absent is $(n-k)k + \binom{k}{2}$. Therefore,
\begin{align*}
P(v_1, \ldots, v_k \text{ are all isolated}) = (1-p)^{k(n-k) + \binom{k}{2}}.
\end{align*}

Since there are $\binom{n}{k}$ ways to choose $k$ vertices, we have
\begin{align*}
E[X_n^{!k}] = n(n-1)\cdots(n-k+1) \cdot (1-p)^{k(n-k) + \binom{k}{2}}.
\end{align*}

Now we substitute $p = \frac{\log n + c}{n}$. We have
\begin{align*}
(1-p)^{n-k} &= \left(1 - \frac{\log n + c}{n}\right)^{n-k} \\
&= \left[\left(1 - \frac{\log n + c}{n}\right)^n\right]^{(n-k)/n} \\
&\to e^{-(\log n + c)} = \frac{e^{-c}}{n}
\end{align*}
as $n \to \infty$. More precisely,
\begin{align*}
(1-p)^{n-k} = \exp\left((n-k)\log\left(1 - \frac{\log n + c}{n}\right)\right) \sim \frac{e^{-c}}{n}.
\end{align*}

Therefore,
\begin{align*}
(1-p)^{k(n-k)} = \left[(1-p)^{n-k}\right]^k \sim \left(\frac{e^{-c}}{n}\right)^k = \frac{e^{-ck}}{n^k}.
\end{align*}

For the term $(1-p)^{\binom{k}{2}}$, since $\binom{k}{2} = O(1)$ and $p = O(\frac{\log n}{n}) \to 0$, we have
\begin{align*}
(1-p)^{\binom{k}{2}} \to 1 \quad \text{as } n \to \infty.
\end{align*}

Thus,
\begin{align*}
E[X_n^{!k}] = n(n-1)\cdots(n-k+1) \cdot (1-p)^{k(n-k) + \binom{k}{2}} \sim n^k \cdot \frac{e^{-ck}}{n^k} \cdot 1 = e^{-ck} = \lambda^k.
\end{align*}

Next, we compute the factorial moments for $Y_n$, the number of isolated vertices in $G_2$. For $k$ distinct vertices to all be isolated in a graph with exactly $m$ edges, all $m$ edges must lie among the remaining $n-k$ vertices. The number of such graphs is $\binom{\binom{n-k}{2}}{m}$, and the total number of graphs with $m$ edges is $\binom{\binom{n}{2}}{m}$. Therefore,
\begin{align*}
P(v_1, \ldots, v_k \text{ are all isolated in } G_2) = \frac{\binom{\binom{n-k}{2}}{m}}{\binom{\binom{n}{2}}{m}}.
\end{align*}

Thus,
\begin{align*}
E[Y_n^{!k}] = n(n-1)\cdots(n-k+1) \cdot \frac{\binom{\binom{n-k}{2}}{m}}{\binom{\binom{n}{2}}{m}}.
\end{align*}

We have
\begin{align*}
\frac{\binom{\binom{n-k}{2}}{m}}{\binom{\binom{n}{2}}{m}} = \frac{\binom{n-k}{2}(\binom{n-k}{2}-1)\cdots(\binom{n-k}{2}-m+1)}{\binom{n}{2}(\binom{n}{2}-1)\cdots(\binom{n}{2}-m+1)}.
\end{align*}

Since $\binom{n}{2} = \frac{n(n-1)}{2} \sim \frac{n^2}{2}$ and $\binom{n-k}{2} \sim \frac{(n-k)^2}{2}$, and $m = \frac{n(\log n + c)}{2} = o(n^2)$, each factor in the product satisfies
\begin{align*}
\frac{\binom{n-k}{2} - j}{\binom{n}{2} - j} \sim \frac{\binom{n-k}{2}}{\binom{n}{2}} = \frac{(n-k)^2}{n^2} \sim \left(1 - \frac{k}{n}\right)^2
\end{align*}
for $j = 0, 1, \ldots, m-1$. Therefore,
\begin{align*}
\frac{\binom{\binom{n-k}{2}}{m}}{\binom{\binom{n}{2}}{m}} \sim \left[\frac{(n-k)^2}{n^2}\right]^m = \left(1 - \frac{k}{n}\right)^{2m}.
\end{align*}

Substituting $m = \frac{n(\log n + c)}{2}$, we obtain
\begin{align*}
\left(1 - \frac{k}{n}\right)^{2m} = \left(1 - \frac{k}{n}\right)^{n(\log n + c)} \sim e^{-k(\log n + c)} = \frac{e^{-ck}}{n^k}.
\end{align*}

Hence,
\begin{align*}
E[Y_n^{!k}] \sim n^k \cdot \frac{e^{-ck}}{n^k} = e^{-ck} = \lambda^k.
\end{align*}

We conclude that both $E[X_n^{!k}]$ and $E[Y_n^{!k}]$ are asymptotically equal to $\lambda^k$ where $\lambda = e^{-c}$.
\end{proof}

\subsection{Problem 3}

\subsubsection{Problem Statement}
\begin{lemma}\label{lem:problem_4_1}
Let $0 < a, \epsilon, \delta < 1$ be given constants. Suppose we are given a biased coin with $\Pr[\mathrm{HEADS}] = p$, where it is promised that $p \ge a$. Consider the following procedure: flip the coin $N$ times independently, and let $\wt{p}$ be any estimator of $p$ based on the outcomes of these $N$ flips. We ask for $N$ to be large enough so that 
\begin{align*}
    \Pr(|\wt{p} - p| > \epsilon p ) < \delta \quad\mathrm{for\ all}\quad p \ge a.
\end{align*}
Then there exists a smallest integer $N$ for which this guaranty can be achieved.
\end{lemma}

\begin{lemma}\label{lem:problem_4_3}
For $\mu$ in the range $[1,\ln n]$, Lemma~\ref{lem:problem_4_1} yields a closed-form upper bound for $\Delta^{+}(\mu,1/n^{2})$, expressed as a function of $\mu$ and $n$, whose magnitude is within a constant factor of the optimal upper bound.
\end{lemma}

\subsubsection{GPT} 
\begin{proof}
Let $\mu\in[1,\ln n]$ and set $p=\mu/n$. We will prove an upper bound on $\Delta^{+}(\mu,1/n^{2})$ by exhibiting a concrete estimator with the required guarantee, and then prove a matching lower bound up to constant factors by a two-point testing argument.

Upper bound. Let $X_{1},\dots,X_{N}$ be i.i.d.\ $\operatorname{Bernoulli}(p)$ and let $\widehat p_{N}=\frac{1}{N}\sum_{i=1}^{N}X_{i}$ denote the empirical frequency. The multiplicative Chernoff bound (see, e.g., standard large-deviations bounds for sums of independent Bernoulli random variables) states that for any $\epsilon\in(0,1)$
\[
\Pr\big(|\widehat p_{N}-p|>\epsilon p\big)\le 2\exp\!\Big(-\frac{\epsilon^{2}}{3}pN\Big).
\]
Take $N=n$ and $\delta=1/n^{2}$. Substituting $p=\mu/n$ and requiring the right-hand side to be at most $\delta$ gives
\[
2\exp\!\Big(-\frac{\epsilon^{2}}{3}\mu\Big)\le \frac{1}{n^{2}}.
\]
Solving for $\epsilon$ yields
\[
\epsilon \le \sqrt{\frac{3\ln(2n^{2})}{\mu}}.
\]
Thus the empirical estimator with $N=n$ samples attains a worst-case relative error
\[
\Delta^{+}\!\big(\mu,1/n^{2}\big)\le \sqrt{\frac{3\ln(2n^{2})}{\mu}}.
\]
In particular there exists a constant $C_{1}>0$ such that for all $\mu\in[1,\ln n]$,
\[
\Delta^{+}\!\big(\mu,1/n^{2}\big)\le C_{1}\sqrt{\frac{\ln n}{\mu}}.
\]

Lower bound (up to constant factors). Fix $\mu\in[1,\ln n]$ and again set $p=\mu/n$. Let $q=(1+\epsilon)p$ where $\epsilon\in(0,1/2)$ will be chosen small. Consider the hypothesis testing problem of distinguishing the product law $\operatorname{Bernoulli}(p)^{N}$ from $\operatorname{Bernoulli}(q)^{N}$ based on $N$ samples. Any estimator $\widetilde p$ that guarantees $\Pr(|\widetilde p-p|>\epsilon p)\le \delta$ for all $p\ge a$ (in particular for $p$ and for $q$) yields a test between these two hypotheses with error probability at most $\delta$ (by returning the hypothesis whose parameter is closer to the estimator). Hence any procedure achieving the stated uniform guarantee implies a test between $p$ and $q$ with worst-case error probability at most $\delta$.

By standard inequalities relating testing error and Kullback--Leibler divergence (for example the Bretagnolle--Huber inequality or a simple form of the binary hypothesis testing lower bound), if the sum of type I and type II errors is at most $2\delta$ then the KL divergence between the two product measures must satisfy
\[
N\cdot D\big(\operatorname{Bern}(p)\,\big\|\,\operatorname{Bern}(q)\big)\;\gtrsim\;\ln\!\frac{1}{\delta},
\]
where $D(\cdot\|\cdot)$ denotes Kullback--Leibler divergence and the implicit constant is absolute. For the present two Bernoulli parameters one has the Taylor expansion
\[
D\big(\operatorname{Bern}(p)\,\big\|\,\operatorname{Bern}(q)\big)
= p\ln\frac{p}{q}+(1-p)\ln\frac{1-p}{1-q}
= \tfrac{1}{2}\,p\epsilon^{2}+O(p\epsilon^{3}),
\]
valid for $0<p\ll1$ and $0<\epsilon\ll1$. Using this expansion (and noting that in our regime $p=\mu/n\le \ln n/n\to 0$ so the remainder is negligible for small constant $\epsilon$) there exists an absolute constant $c_{2}>0$ such that for all sufficiently small $\epsilon$,
\[
D\big(\operatorname{Bern}(p)\,\big\|\,\operatorname{Bern}(q)\big)\le c_{2}\,p\epsilon^{2}.
\]
Combining with the testing lower bound yields the necessary condition
\[
N\cdot c_{2}\,p\epsilon^{2}\;\gtrsim\;\ln\!\frac{1}{\delta}.
\]
Take $\delta=1/n^{2}$ and $N=n$ (the same sample budget used in the upper bound). Substituting $p=\mu/n$ and rearranging gives
\[
\epsilon \;\gtrsim\; \sqrt{\frac{\ln(1/\delta)}{c_{2}pN}}
= \sqrt{\frac{2\ln n}{c_{2}\mu}}.
\]
Thus there is a constant $C_{2}>0$ such that any estimator using $N=O(n)$ samples must satisfy, in the worst case over $p=\mu/n$,
\[
\Delta^{+}\!\big(\mu,1/n^{2}\big)\ge C_{2}\sqrt{\frac{\ln n}{\mu}}.
\]

Combining the upper and lower bounds shows that for every $\mu\in[1,\ln n]$ there are positive constants $c,C$ (independent of $n,\mu$) for which
\[
c\sqrt{\frac{\ln n}{\mu}}\le \Delta^{+}\!\big(\mu,1/n^{2}\big)\le C\sqrt{\frac{\ln n}{\mu}}.
\]
Therefore Lemma~\ref{lem:problem_4_1} produces a closed-form upper bound for $\Delta^{+}(\mu,1/n^{2})$ whose magnitude is within a constant factor of the optimal upper bound, as claimed.
\end{proof}

\subsubsection{Grok} 
\begin{proof}
To derive the closed-form upper bound, apply Lemma~\ref{lem:problem_4_1} with $a = e^{-\mu}$ and $\delta = 1/n^2$. Consider the empirical estimator $\wt{p} = S/N$, where $S$ is the binomial random variable counting the number of heads.

By the multiplicative Chernoff bounds, for $0 < \epsilon < 1$,
\[
\Pr\left(\wt{p} > (1 + \epsilon) p \right) \le \exp\left( -\frac{\epsilon^2 p N}{3} \right),
\]
\[
\Pr\left(\wt{p} < (1 - \epsilon) p \right) \le \exp\left( -\frac{\epsilon^2 p N}{2} \right).
\]
Thus,
\[
\Pr\left( \left| \wt{p} - p \right| > \epsilon p \right) \le 2 \exp\left( -\frac{\epsilon^2 p N}{3} \right).
\]
This probability is less than $\delta$ for all $p \ge a$ if
\[
2 \exp\left( -\frac{\epsilon^2 a N}{3} \right) \le \delta,
\]
which rearranges to
\[
N \ge \frac{3}{\epsilon^2 a} \ln \frac{2}{\delta} = \frac{3 e^{\mu}}{\epsilon^2} \ln (2 n^2) = \frac{3 e^{\mu}}{\epsilon^2} (\ln 2 + 2 \ln n).
\]
Therefore, $\Delta^+(\mu, 1/n^2) \le \frac{3 e^{\mu}}{\epsilon^2} (\ln 2 + 2 \ln n)$.

For the lower bound, consider that any estimator must satisfy the condition at $p = a$. The large deviation rate for the lower tail of the binomial is approximately $\epsilon^2 p / 2$, leading to a lower bound of the form
\[
N \ge \frac{2}{\epsilon^2 a} \ln \frac{1}{\delta} = \frac{2 e^{\mu}}{\epsilon^2} (2 \ln n).
\]
Thus, the upper bound is within a constant factor of the optimal upper bound.
\end{proof}

\subsubsection{Gemini}
\begin{proof}
We assume the parameter $\mu$ corresponds to the probability $p$ defined in Lemma~\ref{lem:problem_4_1}, or a parameter linearly related to it. Let $\delta = 1/n^2$. According to Lemma~\ref{lem:problem_4_1}, we consider an estimator $\wt{p}$ of $p$ (where $p=\mu$) based on $N$ independent Bernoulli trials. The lemma asserts the existence of an integer $N$ sufficient to guarantee
\begin{align*}
    \Pr(|\wt{p} - \mu| > \epsilon \mu) < \delta.
\end{align*}
To derive the closed-form upper bound $\Delta^{+}(\mu, 1/n^2)$, we apply the standard multiplicative Chernoff bound. For a sum of independent Bernoulli variables, the tail probability for the relative error $\epsilon$ ($0 < \epsilon < 1$) is bounded by
\begin{align*}
    \Pr(|\wt{p} - \mu| > \epsilon \mu) \le 2 \exp\left(-\frac{\epsilon^2 N \mu}{3}\right).
\end{align*}
We equate this upper bound to the target failure probability $\delta = 1/n^2$ to solve for the required deviation width:
\begin{align*}
    2 \exp\left(-\frac{\epsilon^2 N \mu}{3}\right) = \frac{1}{n^2}.
\end{align*}
Taking the natural logarithm of both sides yields
\begin{align*}
    -\frac{\epsilon^2 N \mu}{3} &= \ln(1) - \ln(2n^2) \\
    \frac{\epsilon^2 N \mu}{3} &= \ln 2 + 2\ln n.
\end{align*}
Solving for $\epsilon^2$:
\begin{align*}
    \epsilon^2 = \frac{3(2\ln n + \ln 2)}{N \mu}.
\end{align*}
We define $\Delta^{+}(\mu, 1/n^2)$ as the absolute deviation threshold $\epsilon \mu$. Substituting the expression for $\epsilon$:
\begin{align*}
    \Delta^{+}(\mu, 1/n^2) &= \mu \cdot \sqrt{\frac{3(2\ln n + \ln 2)}{N \mu}} \\
    &= \sqrt{\frac{3\mu(2\ln n + \ln 2)}{N}}.
\end{align*}
To analyze the optimality, we compare this to the standard deviation-based bound derived from the Central Limit Theorem or Bernstein's inequality. The variance of a Bernoulli variable with mean $\mu$ is $\sigma^2 = \mu(1-\mu)$. The optimal non-asymptotic concentration bound is of the order
\begin{align*}
    O\left(\sqrt{\frac{\sigma^2 \ln(1/\delta)}{N}}\right) \approx O\left(\sqrt{\frac{\mu \ln(n^2)}{N}}\right) = O\left(\sqrt{\frac{2\mu \ln n}{N}}\right).
\end{align*}
The bound derived from Lemma~\ref{lem:problem_4_1} is dominated by the term $\sqrt{\frac{6\mu \ln n}{N}}$. Comparing the coefficients (specifically, the dependency on $\mu$ and $\ln n$), the derived bound differs from the optimal bound only by a constant multiplicative factor. Thus, the bound is order-optimal.
\end{proof}

\subsubsection{Claude} 
\begin{proof}
Let $\mu \in [1, \ln n]$ and fix $\delta = 1/n^2$. We apply Lemma~\ref{lem:problem_4_1} with parameters $a = \mu$, $\epsilon$ to be determined, and $\delta = 1/n^2$. Consider a biased coin with $\Pr[\mathrm{HEADS}] = p$ where $p \ge \mu$. Let $X_1, \ldots, X_N$ be independent Bernoulli random variables with $\Pr[X_i = 1] = p$, and define the empirical mean estimator $\wt{p} = \frac{1}{N}\sum_{i=1}^N X_i$. By the multiplicative Chernoff bound, for any $\epsilon \in (0,1)$, we have
\[
\Pr(|\wt{p} - p| > \epsilon p) \le 2\exp\left(-\frac{\epsilon^2 p N}{3}\right).
\]
To ensure $\Pr(|\wt{p} - p| > \epsilon p) < 1/n^2$ for all $p \ge \mu$, it suffices to require
\[
2\exp\left(-\frac{\epsilon^2 \mu N}{3}\right) \le \frac{1}{n^2},
\]
since the bound is monotone increasing as $p$ decreases to $\mu$. Taking logarithms, this inequality is equivalent to
\[
-\frac{\epsilon^2 \mu N}{3} \le -\ln(2n^2) = -\ln 2 - 2\ln n,
\]
which yields
\[
N \ge \frac{3(\ln 2 + 2\ln n)}{\epsilon^2 \mu}.
\]
Setting $\epsilon = 1/2$ as a fixed constant, we obtain
\[
N = \left\lceil \frac{12(\ln 2 + 2\ln n)}{\mu} \right\rceil.
\]
By Lemma~\ref{lem:problem_4_1}, this value of $N$ guarantees the desired concentration bound. Therefore, we have the closed-form upper bound
\[
\Delta^{+}(\mu, 1/n^2) \le \frac{12(\ln 2 + 2\ln n)}{\mu} + 1 = \frac{C\ln n}{\mu} + O\left(\frac{1}{\mu}\right),
\]
where $C = 24$ is an absolute constant. For $\mu \in [1, \ln n]$, the term $\ln n / \mu$ dominates, and we have
\[
\Delta^{+}(\mu, 1/n^2) = O\left(\frac{\ln n}{\mu}\right).
\]
To verify that this is within a constant factor of optimal, observe that any estimator requires at least $\Omega(\ln(1/\delta)/(\epsilon^2 \mu)) = \Omega(\ln n / \mu)$ samples by information-theoretic lower bounds for estimating a Bernoulli parameter with relative error $\epsilon$ and confidence $1 - \delta$. Thus, our upper bound matches the optimal bound up to a constant factor, completing the proof.
\end{proof}

\subsection{Problem 4}
\subsubsection{Problem Statement}
\begin{lemma}
Let $X_{1}, X_{2}, \cdots, X_{n}$ be independent geometrically distributed random variables each defined as the number of tosses of an unbiased coin up to and including the first HEADS, so that each $X_{i}$ has expectation $2$. Let $X=\sum_{i=1}^{n} X_{i}$, and let $\delta$ be a positive real constant. Using moment generating functions and the Chernoff technique, one obtains an upper bound on $\Pr[X>(1+\delta)(2n)]$ that is best possible up to constant factors.
\end{lemma}

\subsubsection{GPT} 
\begin{proof}
Let $X_{1},\dots,X_{n}$ be independent with $X_{i}\sim\operatorname{Geom}(1/2)$, where by convention $\Pr[X_{i}=k]=(1/2)^{k}$ for $k=1,2,\dots$. Then $\mathbb{E}[X_{i}]=2$ and $\operatorname{Var}(X_{i})=2$. Set $S_{n}=X=\sum_{i=1}^{n}X_{i}$ and fix $\delta>0$. We first derive a Chernoff-type upper bound via moment generating functions.

For $t<\ln 2$ the moment generating function of a single $X_{i}$ is
\[
M(t)\;=\;\mathbb{E}[e^{tX_{i}}]
=\sum_{k=1}^{\infty}e^{tk}\,(1/2)^{k}
=\frac{e^{t}}{2-e^{t}}.
\]
By independence, the MGF of $S_{n}$ is $M_{S_{n}}(t)=M(t)^{n}$. For any $t\in(-\infty,\ln 2)$ Markov's inequality gives
\[
\Pr\big(S_{n}>(1+\delta)2n\big)
=\Pr\big(e^{tS_{n}}>e^{t(1+\delta)2n}\big)
\le \frac{\mathbb{E}[e^{tS_{n}}]}{e^{t(1+\delta)2n}}
=\left(\frac{M(t)}{e^{2(1+\delta)t}}\right)^{n}.
\]
Define the base quantity
\[
r(t)\;=\;\frac{M(t)}{e^{2(1+\delta)t}}
=\frac{e^{t}}{(2-e^{t})}\,e^{-2(1+\delta)t}
=\frac{e^{-(1+2\delta)t}}{2-e^{t}}.
\]
To obtain the tightest bound of this form we minimize $r(t)$ over $t<\ln 2$. Equivalently we minimize $\ln r(t)$:
\[
\ln r(t)=-(1+2\delta)t-\ln(2-e^{t}).
\]
Differentiate and set to zero to find the stationary point:
\[
\frac{d}{dt}\ln r(t)=-(1+2\delta)+\frac{e^{t}}{2-e^{t}}=0
\quad\Longrightarrow\quad
\frac{e^{t}}{2-e^{t}}=1+2\delta.
\]
Solving yields
\[
e^{t^{*}}=\frac{1+2\delta}{1+\delta},
\]
which indeed satisfies $e^{t^{*}}<2$ for all $\delta>0$, so $t^{*}<\ln 2$ is admissible. Substituting $t^{*}$ into $r(t)$ gives
\[
2-e^{t^{*}}=\frac{1}{1+\delta},
\]
and hence
\[
r(t^{*})=(1+\delta)\,e^{-(1+2\delta)t^{*}}.
\]
Using $e^{-t^{*}}=(1+\delta)/(1+2\delta)$ we obtain the closed form
\[
r(t^{*})=(1+\delta)\Big(\frac{1+\delta}{1+2\delta}\Big)^{1+2\delta}
=\frac{(1+\delta)^{2+2\delta}}{(1+2\delta)^{1+2\delta}}.
\]
Therefore the Chernoff bound optimized over $t$ yields
\[
\Pr\big(S_{n}>(1+\delta)2n\big)\le \left(\frac{(1+\delta)^{2+2\delta}}{(1+2\delta)^{1+2\delta}}\right)^{n}.
\tag{1}
\]

We now show that the exponential rate in (1) is the correct rate (hence the bound is best possible up to constant factors in the exponent). Define the rate function
\[
I(x)=\sup_{t<\ln 2}\{tx-\ln M(t)\},
\]
which is the Legendre transform of $\ln M(t)$. By Chernoff's method the bound (1) is exactly $\exp(-nI(2(1+\delta)))$ up to subexponential factors, since
\[
I\big(2(1+\delta)\big)
=\sup_{t<\ln 2}\{2(1+\delta)t-\ln M(t)\}
=-\ln\!\Big(\inf_{t<\ln 2}\frac{M(t)}{e^{2(1+\delta)t}}\Big)
=-\ln r(t^{*}).
\]
Consequently the Chernoff upper bound delivers the exponential rate $I(2(1+\delta))$.

By Cram\'er's theorem (the classical large deviations principle for sums of i.i.d.\ random variables with finite MGF in a neighborhood of the origin) we have the exact asymptotic
\[
\lim_{n\to\infty}-\frac{1}{n}\ln\Pr\big(S_{n}>(1+\delta)2n\big)=I\big(2(1+\delta)\big).
\]
In particular there exist positive constants $c_{1},c_{2}$ (depending only on $\delta$) such that for all sufficiently large $n$,
\[
\exp\big(-c_{1}n\big)\le \Pr\big(S_{n}>(1+\delta)2n\big)\le \exp\big(-c_{2}n\big),
\]
and the precise exponent is given by $I(2(1+\delta))=-\ln r(t^{*})$. For small $\delta$ an expansion of $\ln r(t^{*})$ yields
\[
\ln r(t^{*})=-\delta^{2}+O(\delta^{3}),
\]
so that for $\delta$ bounded away from $0$ the probability in question decays as $\exp(-\Theta(n\delta^{2}))$. Thus the Chernoff bound (1) is asymptotically tight and therefore best possible up to constant factors in the exponent. This completes the proof.
\end{proof}

\subsubsection{Grok} 
\begin{proof}
Each $X_i$ is geometrically distributed with parameter $p=1/2$, so $\Pr(X_i = k) = (1/2)^k$ for integers $k \ge 1$, and $\mathbb{E}[X_i] = 2$. The moment generating function is
\[
\mathbb{E}[e^{t X_i}] = \sum_{k=1}^{\infty} e^{t k} (1/2)^k = \frac{e^t / 2}{1 - e^t / 2} = \frac{e^t}{2 - e^t},
\]
for $0 < t < \ln 2$. Hence, for $X = \sum_{i=1}^n X_i$,
\[
\mathbb{E}[e^{t X}] = \left( \frac{e^t}{2 - e^t} \right)^n.
\]
By the Chernoff technique, for any $0 < t < \ln 2$,
\[
\Pr\left( X > 2n(1 + \delta) \right) \le e^{-t \cdot 2n(1 + \delta)} \left( \frac{e^t}{2 - e^t} \right)^n.
\]
To minimize the right-hand side over $t$, consider the exponent divided by $n$,
\[
f(t) = -2(1 + \delta) t + \ln \left( \frac{e^t}{2 - e^t} \right) = -(1 + 2\delta) t - \ln(2 - e^t).
\]
The derivative is
\[
f'(t) = -(1 + 2\delta) + \frac{e^t}{2 - e^t}.
\]
Setting $f'(t) = 0$ yields $\frac{e^t}{2 - e^t} = 1 + 2\delta$, so $e^t = (1 + 2\delta)(2 - e^t)$, leading to
\[
e^t (1 + \delta) = 1 + 2\delta, \quad e^t = \frac{1 + 2\delta}{1 + \delta}.
\]
Since $\delta > 0$, $1 < \frac{1 + 2\delta}{1 + \delta} < 2$, so $0 < t < \ln 2$. Substituting gives
\[
f(t) = -(1 + 2\delta) \ln\left(1 + 2\delta\right) + (1 + 2\delta) \ln\left(1 + \delta\right) + \ln\left(1 + \delta\right) = -(1 + 2\delta) \ln(1 + 2\delta) + 2(1 + \delta) \ln(1 + \delta).
\]
Thus,
\[
\Pr\left( X > 2n(1 + \delta) \right) \le \left[ \frac{(1 + \delta)^{2(1 + \delta)}}{(1 + 2\delta)^{1 + 2\delta}} \right]^n.
\]
For the optimality, the large deviation rate for the upper tail is $\sup_{t} \left( t \cdot 2(1 + \delta) - \ln \left( \frac{e^t}{2 - e^t} \right) \right) = -f(t)$, where $t$ is the optimizing value. Hence, $\Pr\left( X > 2n(1 + \delta) \right) = \Theta\left( n^{-1/2} \right) e^{n f(t)}$ asymptotically, implying the bound is optimal up to constant factors when ignoring polynomial terms, or more precisely, the exponential decay rate is tight.
\end{proof}

\subsubsection{Gemini}
\begin{proof}
Let $X_i$ be a geometric random variable with parameter $p=1/2$, representing the number of tosses up to and including the first HEADS. The probability mass function is given by $\Pr[X_i = k] = (1/2)^k$ for $k \in \{1, 2, \dots\}$. The moment generating function (MGF) of a single variable $X_i$ is
\begin{align*}
    M_{X_i}(t) = \mathbb{E}[e^{tX_i}] = \sum_{k=1}^{\infty} e^{tk} \left(\frac{1}{2}\right)^k = \sum_{k=1}^{\infty} \left(\frac{e^t}{2}\right)^k.
\end{align*}
This geometric series converges for $e^t < 2$ (i.e., $t < \ln 2$), yielding
\begin{align*}
    M_{X_i}(t) = \frac{e^t/2}{1 - e^t/2} = \frac{e^t}{2 - e^t}.
\end{align*}
Since the random variables $X_1, \dots, X_n$ are independent, the MGF of their sum $X = \sum_{i=1}^n X_i$ is the product of their individual MGFs:
\begin{align*}
    M_X(t) = \prod_{i=1}^n M_{X_i}(t) = \left( \frac{e^t}{2 - e^t} \right)^n.
\end{align*}
We seek an upper bound on the tail probability $\Pr[X > (1+\delta)2n]$. By the Chernoff bound, for any $t \in (0, \ln 2)$,
\begin{align*}
    \Pr[X > (1+\delta)2n] \le \frac{M_X(t)}{e^{t(1+\delta)2n}} = e^{-2n(1+\delta)t} \left( \frac{e^t}{2 - e^t} \right)^n.
\end{align*}
To obtain the tightest bound, we minimize the right-hand side with respect to $t$. We define the exponent function $f(t) = -2(1+\delta)t + \ln\left( \frac{e^t}{2 - e^t} \right)$. Differentiating with respect to $t$ and setting to zero yields
\begin{align*}
    -2(1+\delta) + \frac{2-e^t}{e^t} \cdot \frac{d}{dt}\left(\frac{e^t}{2-e^t}\right) &= 0 \\
    -2(1+\delta) + \frac{2}{2-e^t} &= 0.
\end{align*}
Solving for $e^t$, we obtain
\begin{align*}
    \frac{1}{2-e^t} = 1+\delta \implies 2-e^t = \frac{1}{1+\delta} \implies e^{t^*} = \frac{1+2\delta}{1+\delta}.
\end{align*}
Substituting $e^{t^*}$ back into the MGF term:
\begin{align*}
    \frac{e^{t^*}}{2 - e^{t^*}} = \frac{\frac{1+2\delta}{1+\delta}}{2 - \frac{1+2\delta}{1+\delta}} = \frac{1+2\delta}{2(1+\delta) - (1+2\delta)} = 1+2\delta.
\end{align*}
Substituting $t^*$ and the MGF term back into the probability bound:
\begin{align*}
    \Pr[X > (1+\delta)2n] &\le \left( \frac{1+\delta}{1+2\delta} \right)^{2n(1+\delta)} (1+2\delta)^n \\
    &= \left[ (1+2\delta) \left( \frac{1+\delta}{1+2\delta} \right)^{2(1+\delta)} \right]^n.
\end{align*}
This expression constitutes the Chernoff upper bound. By the theory of Large Deviations, specifically Cramér's Theorem, the exponent obtained by optimizing the Chernoff bound corresponds to the Legendre transform of the cumulant generating function. Consequently, this decay rate is exact, meaning the upper bound is tight up to a sub-exponential factor (polynomial in $n$). Thus, the bound is best possible up to constant factors in the exponent.
\end{proof}

\subsubsection{Claude} 
\begin{proof}
Let $X_1, X_2, \ldots, X_n$ be independent geometrically distributed random variables where each $X_i$ represents the number of tosses of an unbiased coin up to and including the first HEADS. Then $\Pr[X_i = k] = (1/2)^k$ for $k \ge 1$, and $\mathbb{E}[X_i] = 2$. Define $X = \sum_{i=1}^n X_i$, so that $\mathbb{E}[X] = 2n$. Let $\delta > 0$ be given. We seek an upper bound on $\Pr[X > (1+\delta)(2n)]$ using moment generating functions and the Chernoff technique.

For any $t > 0$, by Markov's inequality applied to $e^{tX}$, we have
\[
\Pr[X > (1+\delta)(2n)] = \Pr[e^{tX} > e^{t(1+\delta)(2n)}] \le \frac{\mathbb{E}[e^{tX}]}{e^{t(1+\delta)(2n)}}.
\]
By independence of the $X_i$, the moment generating function of $X$ factors as
\[
\mathbb{E}[e^{tX}] = \mathbb{E}\left[e^{t\sum_{i=1}^n X_i}\right] = \prod_{i=1}^n \mathbb{E}[e^{tX_i}] = \left(\mathbb{E}[e^{tX_1}]\right)^n.
\]
For a geometric random variable $X_1$ with parameter $p = 1/2$, the moment generating function is
\[
\mathbb{E}[e^{tX_1}] = \sum_{k=1}^\infty e^{tk} \left(\frac{1}{2}\right)^k = \frac{e^t/2}{1 - e^t/2} = \frac{e^t}{2 - e^t},
\]
which is valid for $t < \ln 2$. Therefore,
\[
\mathbb{E}[e^{tX}] = \left(\frac{e^t}{2 - e^t}\right)^n.
\]
Substituting into the Chernoff bound, we obtain
\[
\Pr[X > (1+\delta)(2n)] \le \frac{1}{e^{t(1+\delta)(2n)}} \left(\frac{e^t}{2 - e^t}\right)^n = \left(\frac{e^t}{(2-e^t)e^{2t(1+\delta)}}\right)^n.
\]
To optimize this bound, we minimize the function
\[
f(t) = \frac{e^t}{(2-e^t)e^{2t(1+\delta)}} = \frac{e^{t(1-2(1+\delta))}}{2-e^t} = \frac{e^{-t(1+2\delta)}}{2-e^t}
\]
over $t \in (0, \ln 2)$. Taking the derivative and setting it equal to zero, we have
\[
f'(t) = \frac{-(1+2\delta)e^{-t(1+2\delta)}(2-e^t) + e^{-t(1+2\delta)}e^t}{(2-e^t)^2} = \frac{e^{-t(1+2\delta)}(e^t - (1+2\delta)(2-e^t))}{(2-e^t)^2} = 0.
\]
This yields the equation
\[
e^t = (1+2\delta)(2-e^t),
\]
which simplifies to
\[
e^t(2+2\delta) = 2(1+2\delta),
\]
giving
\[
e^t = \frac{1+2\delta}{1+\delta}.
\]
Substituting $e^t = \frac{1+2\delta}{1+\delta}$ back into $f(t)$, we compute
\[
2 - e^t = 2 - \frac{1+2\delta}{1+\delta} = \frac{2(1+\delta) - (1+2\delta)}{1+\delta} = \frac{1}{1+\delta},
\]
and
\[
e^{-t(1+2\delta)} = (e^t)^{-(1+2\delta)} = \left(\frac{1+2\delta}{1+\delta}\right)^{-(1+2\delta)} = \left(\frac{1+\delta}{1+2\delta}\right)^{1+2\delta}.
\]
Therefore,
\[
f(t) = \frac{(1+\delta)^{1+2\delta}}{(1+2\delta)^{1+2\delta}} \cdot (1+\delta) = \frac{(1+\delta)^{2+2\delta}}{(1+2\delta)^{1+2\delta}}.
\]
Thus, we obtain
\[
\Pr[X > (1+\delta)(2n)] \le \left(\frac{(1+\delta)^{1+\delta}}{(1+2\delta)^{(1+2\delta)/2}}\right)^{2n}.
\]
For $\delta$ bounded away from zero, this can be further simplified. Using the approximation $(1+\delta)^{1+\delta} \approx e^\delta$ and $(1+2\delta)^{(1+2\delta)/2} \approx e^\delta$ for small $\delta$, and the inequality $(1+\delta)^{1+\delta} \le e^\delta (1+\delta)$ for general $\delta$, we obtain
\[
\Pr[X > (1+\delta)(2n)] \le \exp\left(-c(\delta) n\right),
\]
where $c(\delta) = 2\left((1+2\delta)\ln(1+2\delta)/2 - (1+\delta)\ln(1+\delta)\right) > 0$ for $\delta > 0$. This bound is optimal up to constant factors in the exponent, as verified by the central limit theorem and large deviation theory for sums of independent random variables.
\end{proof}

\subsection{Problem 5}
\subsubsection{Problem Statement}
\begin{theorem}\label{thm:theorem_4_2}
Let $X_{1},X_{2},\cdots,X_{n}$ be independent Poisson trials such that, for $1\leq i\leq n$, $\mathrm{Pr}[X_{i}=1]=p_{i}$, where $0<p_{i}<1$. Then, for $X=\sum_{i=1}^{n}X_{i}$, $\mu=\mathbf{E}[X]=\sum_{i=1}^{n}p_{i}$, and $0<\delta\leq 1$,
\begin{align*}
    \Pr[X<(1-\delta)\mu]<\exp\left(-\mu\delta^{2}/2\right)
\end{align*}
\end{theorem}

\begin{lemma}
The result of Theorem~\ref{thm:theorem_4_2} gives a bound on the probability that the sum of Poisson trials deviates far below its expectation. Let $X_{1}, X_{2}, \ldots, X_{n}$ be independent geometrically distributed random variables defined as the number of tosses of an unbiased coin up to and including the first HEADS, so that each $X_{i}$ has expectation $2$, and let $X=\sum_{i=1}^{n} X_{i}$. For any positive real constant $\delta$, this theorem yields a bound on the probability that $X$ deviates above its expectation, giving an upper bound on $\Pr[X>(1+\delta)(2n)]$.
\end{lemma}

\subsubsection{GPT} 
\begin{proof}
Let $X_{1},\dots,X_{n}$ be independent with $X_{i}\sim\operatorname{Geom}(1/2)$, where $\Pr[X_{i}=k]=(1/2)^{k}$ for $k=1,2,\dots$, and set $X=\sum_{i=1}^{n}X_{i}$. Interpret $X$ as the total number of unbiased-coin tosses required to observe $n$ heads when tosses are performed sequentially: equivalently, perform independent unbiased coin tosses in one stream and stop when the $n$-th HEAD appears; the total number of tosses used equals $X$.

Fix $\delta>0$ and set $T=(1+\delta)2n$. Let $S$ denote the number of HEADS observed in the first $T$ tosses of the single-coin stream. By the sequential interpretation,
\[
\{X>T\}=\{ \text{fewer than }n\ \text{HEADS occur among the first }T\ \text{tosses}\}=\{S<n\}.
\]
Since the first $T$ tosses are independent with success probability $1/2$, we have $S\sim\mathrm{Bin}(T,1/2)$ and
\[
\mathbb{E}[S]=\frac{T}{2}=(1+\delta)n.
\]
Introduce
\[
\delta' \;=\; 1-\frac{n}{\mathbb{E}[S]}
=1-\frac{n}{(1+\delta)n}=\frac{\delta}{1+\delta}.
\]
Then $n=(1-\delta')\mathbb{E}[S]$, and therefore
\[
\Pr[X>T]=\Pr[S<n]=\Pr\big[S<(1-\delta')\mathbb{E}[S]\big].
\]
Applying Theorem~\ref{thm:theorem_4_2} to the binomial variable $S$ with mean $\mathbb{E}[S]=(1+\delta)n$ and parameter $\delta'=\delta/(1+\delta)\in(0,1]$ yields
\[
\Pr\big[S<(1-\delta')\mathbb{E}[S]\big]<\exp\!\Big(-\frac{\mathbb{E}[S]\delta'^{2}}{2}\Big).
\]
Substituting $\mathbb{E}[S]=(1+\delta)n$ and $\delta'=\delta/(1+\delta)$ gives
\[
\Pr[X>(1+\delta)2n]<\exp\!\Big(-\frac{(1+\delta)n\cdot(\delta/(1+\delta))^{2}}{2}\Big)
=\exp\!\Big(-\frac{n\delta^{2}}{2(1+\delta)}\Big).
\]
Thus Theorem~\ref{thm:theorem_4_2} furnishes the claimed upper bound on $\Pr[X>(1+\delta)2n]$, as required.
\end{proof}

\subsubsection{Grok} 
\begin{proof}
The random variable $X$ represents the number of independent fair coin tosses required to obtain exactly $n$ heads. For a positive real constant $\delta$, let $m = (1 + \delta) 2n$, assuming for simplicity that $m$ is an integer (the bound holds similarly otherwise, with minor adjustments to the constants).

Then,
\[
\Pr[X > (1 + \delta) 2n] = \Pr[X > m] = \Pr[S_m < n],
\]
where $S_m = \sum_{j=1}^m B_j$ and each $B_j$ is an independent Bernoulli random variable with $\Pr[B_j = 1] = 1/2$.

Let $\mu = \mathbb{E}[S_m] = m/2 = (1 + \delta) n$. Define $\delta' = \delta / (1 + \delta)$. Note that $0 < \delta' < 1$ for $\delta > 0$ and
\[
(1 - \delta') \mu = \left(1 - \frac{\delta}{1 + \delta}\right) (1 + \delta) n = n.
\]
By Theorem~\ref{thm:theorem_4_2},
\[
\Pr[S_m < (1 - \delta') \mu] < \exp\left(-\frac{\mu (\delta')^2}{2}\right).
\]
Substituting the values,
\[
\exp\left(-\frac{\mu (\delta')^2}{2}\right) = \exp\left(-\frac{(1 + \delta) n \cdot \left(\frac{\delta}{1 + \delta}\right)^2}{2}\right) = \exp\left(-\frac{n \delta^2}{2 (1 + \delta)}\right).
\]
Thus,
\[
\Pr[X > (1 + \delta) 2n] < \exp\left(-\frac{n \delta^2}{2 (1 + \delta)}\right).
\]
\end{proof}

\subsubsection{Gemini}
\begin{proof}
Let $X$ denote the number of coin tosses required to observe the $n$-th HEAD. The event $X > (1+\delta)2n$ is equivalent to the event that in the first $m = \lfloor (1+\delta)2n \rfloor$ tosses, strictly fewer than $n$ HEADS are observed.

Let $Y$ be the random variable representing the number of HEADS in $m$ independent tosses of an unbiased coin. $Y$ is the sum of $m$ independent Bernoulli trials (which are Poisson trials with $p_i = 1/2$). Thus, $Y \sim \mathrm{Bin}(m, 1/2)$. We established the equivalence:
\begin{align*}
    \Pr[X > (1+\delta)2n] \le \Pr[Y < n].
\end{align*}
We compute the expectation of $Y$, denoted as $\mu_Y$:
\begin{align*}
    \mu_Y = \mathbb{E}[Y] = \frac{m}{2} = \frac{\lfloor (1+\delta)2n \rfloor}{2} \le (1+\delta)n.
\end{align*}
To apply Theorem~\ref{thm:theorem_4_2}, we must express the event $Y < n$ in the form $Y < (1-\hat{\delta})\mu_Y$ for some $0 < \hat{\delta} \le 1$. We define $\hat{\delta}$ satisfying $n = (1-\hat{\delta})\mu_Y$. Rearranging for $\hat{\delta}$:
\begin{align*}
    1 - \hat{\delta} = \frac{n}{\mu_Y} \implies \hat{\delta} = 1 - \frac{n}{\mu_Y}.
\end{align*}
Since $\mu_Y \approx (1+\delta)n$, we have:
\begin{align*}
    \hat{\delta} \approx 1 - \frac{n}{(1+\delta)n} = 1 - \frac{1}{1+\delta} = \frac{\delta}{1+\delta}.
\end{align*}
Applying Theorem~\ref{thm:theorem_4_2} with expectation $\mu_Y$ and deviation parameter $\hat{\delta}$:
\begin{align*}
    \Pr[Y < n] = \Pr[Y < (1-\hat{\delta})\mu_Y] < \exp\left( -\frac{\mu_Y \hat{\delta}^2}{2} \right).
\end{align*}
Substituting $\mu_Y \approx (1+\delta)n$ and $\hat{\delta} \approx \frac{\delta}{1+\delta}$:
\begin{align*}
    -\frac{\mu_Y \hat{\delta}^2}{2} &= -\frac{(1+\delta)n}{2} \left( \frac{\delta}{1+\delta} \right)^2 \\
    &= -\frac{(1+\delta)n}{2} \frac{\delta^2}{(1+\delta)^2} \\
    &= -\frac{n \delta^2}{2(1+\delta)}.
\end{align*}
Thus, the upper bound on the probability that the sum of geometric variables deviates above its expectation is given by:
\begin{align*}
    \Pr[X > (1+\delta)2n] < \exp\left( -\frac{n \delta^2}{2(1+\delta)} \right).
\end{align*}
\end{proof}

\subsubsection{Claude} 
\begin{proof}
Let $X_1, X_2, \ldots, X_n$ be independent geometrically distributed random variables where each $X_i$ represents the number of tosses of an unbiased coin up to and including the first HEADS. Then $\Pr[X_i = k] = (1/2)^k$ for $k \ge 1$, and $\mathbb{E}[X_i] = 2$. Define $X = \sum_{i=1}^n X_i$, so that $\mathbb{E}[X] = 2n$. Let $\delta > 0$ be given, and we seek to bound $\Pr[X > (1+\delta)(2n)]$ using Theorem~\ref{thm:theorem_4_2}.

We first observe that while Theorem~\ref{thm:theorem_4_2} provides a lower tail bound for sums of Poisson trials, we can apply a coupling argument to obtain an upper tail bound for our geometric sum. For each $i \in \{1, 2, \ldots, n\}$, we can write $X_i = 1 + Y_i$ where $Y_i$ is a geometrically distributed random variable with parameter $1/2$ representing the number of TAILS before the first HEADS, so that $\Pr[Y_i = k] = (1/2)^{k+1}$ for $k \ge 0$ and $\mathbb{E}[Y_i] = 1$. Then
\[
X = \sum_{i=1}^n X_i = n + \sum_{i=1}^n Y_i.
\]
Define $Z = \sum_{i=1}^n Y_i$, so that $\mathbb{E}[Z] = n$ and $X = n + Z$. The event $X > (1+\delta)(2n)$ is equivalent to
\[
n + Z > (1+\delta)(2n) = 2n + 2\delta n,
\]
which simplifies to
\[
Z > n + 2\delta n = (1 + 2\delta)n.
\]

To apply Theorem~\ref{thm:theorem_4_2}, we consider the complement event. Note that for any random variable $W$ with expectation $\mu$, if we can bound the lower tail, we can sometimes bound the upper tail using symmetry or other techniques. However, geometric random variables are not symmetric, so we take a different approach.

Consider the sequence of coin flips generating all $X_i$. Each $X_i$ requires at least one flip (the final HEADS), so we have used $n$ flips with certainty. The remaining flips are distributed as follows: for each position, we continue flipping until we see HEADS. We can model this as a sequence of Bernoulli trials where we count the number of successes (HEADS) in a certain number of flips.

Alternatively, we apply a time-reversal argument. Consider $M = \lceil (1+\delta)(2n) \rceil$ coin flips. Let $Y_j$ be the indicator that flip $j$ is HEADS, so that $Y_j$ are independent Bernoulli random variables with $\Pr[Y_j = 1] = 1/2$. Let $S = \sum_{j=1}^M Y_j$ be the total number of HEADS in these $M$ flips. Then $\mathbb{E}[S] = M/2 = (1+\delta)n + O(1)$. The event $X > M$ occurs if and only if fewer than $n$ HEADS appear in the first $M$ flips, that is, $S < n$.

We have $\mathbb{E}[S] = M/2 \ge (1+\delta)n$, and we want to bound $\Pr[S < n]$. Note that
\[
n = \frac{M/2}{1+\delta} \cdot \frac{2(1+\delta)}{M} \cdot n \le \frac{M/2}{1+\delta} = \frac{\mathbb{E}[S]}{1+\delta}.
\]
For $M = 2(1+\delta)n$, we have $\mathbb{E}[S] = (1+\delta)n$ and $n = \mathbb{E}[S]/(1+\delta)$. Define $\delta' = \delta/(1+\delta)$, so that $n = (1-\delta')\mathbb{E}[S]$. Then by Theorem~\ref{thm:theorem_4_2},
\[
\Pr[S < n] = \Pr[S < (1-\delta')\mathbb{E}[S]] < \exp\left(-\mathbb{E}[S](\delta')^2/2\right) = \exp\left(-\frac{(1+\delta)n\delta^2}{2(1+\delta)^2}\right) = \exp\left(-\frac{n\delta^2}{2(1+\delta)}\right).
\]
Therefore,
\[
\Pr[X > (1+\delta)(2n)] \le \Pr[S < n] < \exp\left(-\frac{n\delta^2}{2(1+\delta)}\right).
\]
For $\delta \le 1$, we have $1/(1+\delta) \ge 1/2$, giving
\[
\Pr[X > (1+\delta)(2n)] < \exp\left(-\frac{n\delta^2}{4}\right),
\]
which provides the desired upper bound on the probability that $X$ deviates above its expectation $2n$.
\end{proof}

\subsection{Problem 6}
\subsubsection{Problem Statement}
\begin{lemma}
Let $Y_{1},\ldots,Y_{n}$ be independent Poisson trials satisfying $\Pr[Y_{i}=1]=p_{i}$ for each $i$, and let $Y=\sum_{i=1}^{n}Y_{i}$. Define $\mu=E[Y]=\sum_{i=1}^{n} p_{i}$ and $p=\mu/n$. Let $X$ be the sum of $n$ independent Bernoulli trials each taking the value $1$ with probability $p$. For any real numbers $a\ge \mu+1$ and $b\le \mu-1$, the deviations of $Y$ from its expectation satisfy
\[
\Pr[Y\ge a] \le \Pr[X\ge a]
\quad\text{and}\quad
\Pr[Y\le b] \le \Pr[X\le b].
\]
Thus the case in which all success probabilities are equal is extremal with respect to both upper and lower deviations from the mean.
\end{lemma}

\subsubsection{GPT} 
\begin{proof}
Let $p_{1},\dots,p_{n}\in(0,1)$ be given and write $Y=\sum_{i=1}^{n}Y_{i}$ with $Y_{i}\sim\operatorname{Bernoulli}(p_{i})$ independent. Set $\mu=\sum_{i=1}^{n}p_{i}$ and $p=\mu/n$. Let $X=\sum_{i=1}^{n}X_{i}$ where $X_{i}\sim\operatorname{Bernoulli}(p)$ are independent. Fix integers $a\ge \mu+1$ and $b\le\mu-1$. It suffices to prove the inequality for the upper deviation; the lower-deviation bound follows by applying the same argument to the complements $1-Y_{i}$ and $1-X_{i}$ (or by replacing $p_{i}$ by $1-p_{i}$ and $a$ by $n-a$). Thus we prove
\[
\Pr[Y\ge a]\le \Pr[X\ge a]\qquad\text{for every }a\ge\mu+1.
\]

For $z\in\mathbb{R}$ define the probability-generating polynomials
\[
G_{Y}(z)=\mathbb{E}[z^{Y}]=\prod_{i=1}^{n}\big(1-p_{i}+p_{i}z\big),
\qquad
G_{X}(z)=\mathbb{E}[z^{X}]=\big(1-p+pz\big)^{n}.
\]
Both $G_{Y}$ and $G_{X}$ are analytic in $z$ and satisfy for every $z\ge 1$ the Markov (Chernoff) bound
\[
\Pr[Y\ge a]\le \frac{G_{Y}(z)}{z^{a}},\qquad
\Pr[X\ge a]\le \frac{G_{X}(z)}{z^{a}}.
\]
The functions $u\mapsto\ln(1-u+uz)$ are convex on $u\in[0,1]$ for every fixed $z\ge 1$, since
\[
\frac{\partial^{2}}{\partial u^{2}}\ln(1-u+uz)
=\frac{(z-1)^{2}}{(1-u+uz)^{2}}>0\quad\text{for }z\neq 1.
\]
Hence, by Jensen's inequality applied to the convex function $u\mapsto\ln(1-u+uz)$,
\[
\frac{1}{n}\sum_{i=1}^{n}\ln\big(1-p_{i}+p_{i}z\big)\ge
\ln\big(1-p+pz\big),
\]
which after exponentiation yields
\[
G_{Y}(z)=\exp\Big(\sum_{i=1}^{n}\ln(1-p_{i}+p_{i}z)\Big)\ge
\exp\big(n\ln(1-p+pz)\big)=G_{X}(z)
\qquad\text{for every }z\ge 1.
\]
Thus for each fixed $z\ge 1$ we have $G_{Y}(z)\ge G_{X}(z)$, and consequently
\[
\frac{G_{Y}(z)}{z^{a}}\ge\frac{G_{X}(z)}{z^{a}}.
\tag{1}
\]

The Chernoff bound provides an upper bound on each tail probability by minimizing over $z\ge 1$:
\[
\Pr[Y\ge a]\le \inf_{z\ge 1}\frac{G_{Y}(z)}{z^{a}},\qquad
\Pr[X\ge a]\le \inf_{z\ge 1}\frac{G_{X}(z)}{z^{a}}.
\]
We will show that the minimizer for $G_{X}(z)/z^{a}$ can be chosen to be the same value of $z$ as for $G_{Y}(z)/z^{a}$, and then (1) yields the desired stochastic ordering of tails. To this end observe that for each fixed choice of probabilities the functions $z\mapsto G_{Y}(z)/z^{a}$ and $z\mapsto G_{X}(z)/z^{a}$ are continuous on $[1,\infty)$, tend to $+\infty$ as $z\to\infty$, and have strictly positive derivative at $z=1$ whenever $a>\mathbb{E}[\,\cdot\,]$ (indeed
$\frac{d}{dz}\big(G_{Y}(z)/z^{a}\big)\big|_{z=1}=\mathbb{E}[Y]-a<0$ when $a>\mathbb{E}[Y]$; the sign is reversed when comparing minima, so a unique minimizer $z^{*}\ge1$ exists for each). In particular, for $a>\mu$ both infima above are attained at finite points $z_{Y}^{*},z_{X}^{*}\ge1$ respectively. At those minimizers the usual Chernoff stationary condition holds:
\[
\frac{d}{dz}\Big(\ln G_{Y}(z)-a\ln z\Big)\Big|_{z=z_{Y}^{*}}=0,
\qquad
\frac{d}{dz}\Big(\ln G_{X}(z)-a\ln z\Big)\Big|_{z=z_{X}^{*}}=0.
\]
Write explicitly
\[
\frac{d}{dz}\ln G_{Y}(z)=\sum_{i=1}^{n}\frac{p_{i}}{1-p_{i}+p_{i}z},
\qquad
\frac{d}{dz}\ln G_{X}(z)=\frac{np}{1-p+pz}.
\]
Since the map $u\mapsto\frac{1}{1-u+uz}$ is convex in $u\in[0,1]$ for every $z\ge1$ (its second derivative is positive), Jensen's inequality yields
\[
\frac{1}{n}\sum_{i=1}^{n}\frac{p_{i}}{1-p_{i}+p_{i}z}\ge
\frac{p}{1-p+pz}\qquad\text{for every }z\ge1,
\]
and hence
\[
\frac{d}{dz}\ln G_{Y}(z)\ge \frac{d}{dz}\ln G_{X}(z)\qquad\text{for every }z\ge1.
\tag{2}
\]
The stationary conditions combined with (2) imply that the minimizer $z_{Y}^{*}$ for $G_{Y}(z)/z^{a}$ cannot be smaller than the minimizer $z_{X}^{*}$ for $G_{X}(z)/z^{a}$. Indeed, if $z_{Y}^{*}<z_{X}^{*}$ then at $z=z_{Y}^{*}$ we would have
\[
\frac{d}{dz}\ln G_{Y}(z_{Y}^{*})- \frac{a}{z_{Y}^{*}} \ge
\frac{d}{dz}\ln G_{X}(z_{Y}^{*})- \frac{a}{z_{Y}^{*}} >0,
\]
contradicting the stationarity of $z_{Y}^{*}$. Therefore $z_{Y}^{*}\ge z_{X}^{*}$.

Combining $G_{Y}(z)\ge G_{X}(z)$ for all $z\ge1$ (inequality (1)) with $z_{Y}^{*}\ge z_{X}^{*}$ and the fact that $z\mapsto G_{X}(z)/z^{a}$ is decreasing on $[1,z_{X}^{*}]$ and increasing on $[z_{X}^{*},\infty)$, we obtain
\[
\inf_{z\ge1}\frac{G_{Y}(z)}{z^{a}}=\frac{G_{Y}(z_{Y}^{*})}{(z_{Y}^{*})^{a}}
\ge \frac{G_{X}(z_{Y}^{*})}{(z_{Y}^{*})^{a}}
\ge \frac{G_{X}(z_{X}^{*})}{(z_{X}^{*})^{a}}=\inf_{z\ge1}\frac{G_{X}(z)}{z^{a}}.
\]
Thus
\[
\Pr[Y\ge a]\le \inf_{z\ge1}\frac{G_{Y}(z)}{z^{a}}
\le \inf_{z\ge1}\frac{G_{X}(z)}{z^{a}}
\le \Pr[X\ge a],
\]
and the desired upper-tail comparison follows. Replacing each $p_{i}$ by $1-p_{i}$ and $a$ by $n-a$ yields the stated lower-tail comparison
\[
\Pr[Y\le b]\le \Pr[X\le b]\qquad\text{for }b\le\mu-1.
\]
This completes the proof: for deviations of size at least $1$ from the mean the Poisson-binomial with equal success probability is extremal for both upper and lower tails.
\end{proof}

\subsubsection{Grok} 
\begin{proof}
The inequality for the lower tail follows from the inequality for the upper tail by applying the latter to the random variables $1-Y_i$, which are independent Poisson trials with success probabilities $1-p_i$. The mean is $n-\mu$, the average success probability is $1-p$, and $b\le \mu-1$ implies $n-b \ge n-\mu+1$, so the deviation is at least $1$ above the new mean.

It suffices to prove $\Pr[Y \ge a] \le \Pr[X \ge a]$ for $a \ge \mu+1$.

This probability is maximized when all $p_i$ are equal. To show this, suppose two success probabilities satisfy $p < q$. Let $r=(p+q)/2$ and let $U$ be the sum of the remaining $Y_k$. Then $\mu=\mathbb{E}[U]+2r$ and $a \ge \mathbb{E}[U]+2r+1$.

Let $V$ be the sum of Bernoulli random variables with parameters $p$ and $q$. The probability is
\begin{align*}
\Pr[Y \ge a] &= \Pr[U \ge a](1-p)(1-q) + \Pr[U \ge a-1](p+q-2pq) + \Pr[U \ge a-2]pq.
\end{align*}
Replacing $p$ and $q$ by $r$ gives
\begin{align*}
\Pr[Y' \ge a] &= \Pr[U \ge a](1-r)^2 + \Pr[U \ge a-1]\cdot 2r(1-r) + \Pr[U \ge a-2]r^2.
\end{align*}
The difference is
\begin{align*}
\Pr[Y' \ge a] - \Pr[Y \ge a] &= (r^2-pq)\Bigl(\Pr[U \ge a-2] - 2\Pr[U \ge a-1] + \Pr[U \ge a]\Bigr)\\
&= \frac{(q-p)^2}{4} \Bigl(\Pr[U = a-2] - \Pr[U = a-1]\Bigr).
\end{align*}
Since $a \ge \mu +1 = \mathbb{E}[U] + 2r +1$, we have $a-1 \ge \mathbb{E}[U] + 2r$. The condition ensures that $a-1$ and $a-2$ lie in the region where the probability mass function of $U$ is decreasing, so $\Pr[U = a-1] \le \Pr[U = a-2]$. Thus the difference is nonnegative, and strictly positive since $p < q$.

Averaging therefore does not decrease the upper tail probability (and strictly increases it unless $p=q$). Repeating the process preserves the mean and eventually makes all success probabilities equal, yielding a binomial random variable $X$ with parameters $n$ and $p=\mu/n$. Therefore $\Pr[Y \ge a] \le \Pr[X \ge a]$.

Thus the case in which all success probabilities are equal is extremal with respect to both upper and lower deviations from the mean.
\end{proof}

\subsubsection{Gemini}
\begin{proof}
We define the function $\Phi(p_1, \dots, p_n) = \Pr[Y \ge a]$. We seek to show that this function is maximized when $p_1 = \dots = p_n = p$ under the constraint $\sum p_i = n p = \mu$, provided $a \ge \mu + 1$.

Consider an arbitrary pair of indices $i, j$ such that $p_i \neq p_j$. We fix the other probabilities $\{p_k : k \neq i, j\}$ and denote $Z = \sum_{k \neq i, j} Y_k$. The random variable $Y$ can be written as $Y = Z + Y_i + Y_j$. Let $k = \lceil a \rceil$. Since $Y$ is integer-valued, $\Pr[Y \ge a] = \Pr[Y \ge k]$. The condition $a \ge \mu + 1$ implies $k \ge \mu + 1$.

We condition on $Z$. The sum $Y_i + Y_j$ takes values in $\{0, 1, 2\}$ with probabilities:
\begin{align*}
    \Pr[Y_i + Y_j = 2] &= p_i p_j, \\
    \Pr[Y_i + Y_j = 1] &= p_i(1-p_j) + p_j(1-p_i) = (p_i + p_j) - 2p_i p_j, \\
    \Pr[Y_i + Y_j = 0] &= (1-p_i)(1-p_j) = 1 - (p_i + p_j) + p_i p_j.
\end{align*}
Let $s = p_i + p_j$ and $x = p_i p_j$. We treat $s$ as constant (preserving the mean) and $x$ as variable. We express $\Pr[Y \ge k]$ in terms of $x$:
\begin{align*}
    \Pr[Y \ge k] &= \Pr[Z \ge k] \cdot (1 - s + x) \\
    &\quad + \Pr[Z \ge k-1] \cdot (s - 2x) \\
    &\quad + \Pr[Z \ge k-2] \cdot x.
\end{align*}
Collecting terms with $x$, the coefficient of $x$ is:
\begin{align*}
    C(x) &= \Pr[Z \ge k] - 2\Pr[Z \ge k-1] + \Pr[Z \ge k-2] \\
    &= (\Pr[Z \ge k-2] - \Pr[Z \ge k-1]) - (\Pr[Z \ge k-1] - \Pr[Z \ge k]) \\
    &= \Pr[Z = k-2] - \Pr[Z = k-1].
\end{align*}
To show that equalizing probabilities increases $\Pr[Y \ge k]$, we need $C(x) \ge 0$, which implies $\Pr[Z = k-2] \ge \Pr[Z = k-1]$.
Recall that $Z$ is a sum of independent Bernoulli trials with mean $\mu_Z = \mu - s$. The distribution of $Z$ is unimodal, with the mode satisfying $M \le \lceil \mu_Z \rceil$.
We are given $k \ge \mu + 1$. Since $s < 2$, we have $k-1 \ge \mu > \mu - s = \mu_Z$.
Thus, $k-1$ lies in the right tail of the distribution of $Z$, where the probability mass function is non-increasing. Consequently, $\Pr[Z = k-2] \ge \Pr[Z = k-1]$, meaning the coefficient of $x$ is non-negative.

Since the probability is a non-decreasing function of $x = p_i p_j$, and subject to $p_i + p_j = s$, the product $p_i p_j$ is maximized when $p_i = p_j$. Therefore, replacing distinct $p_i, p_j$ with their average $(p_i+p_j)/2$ increases (or maintains) the tail probability $\Pr[Y \ge a]$. By repeatedly averaging any unequal pairs, the distribution converges to the binomial case where all $p_i = p$. Thus,
\begin{align*}
    \Pr[Y \ge a] \le \Pr[X \ge a].
\end{align*}
The proof for the lower tail $\Pr[Y \le b]$ for $b \le \mu - 1$ proceeds symmetrically. The coefficient of $x$ becomes $\Pr[Z \le b] - 2\Pr[Z \le b-1] + \Pr[Z \le b-2]$, which simplifies to $\Pr[Z = b] - \Pr[Z = b-1]$. Since $b \le \mu - 1$, we are in the left tail where probabilities are non-decreasing as the index increases up to the mode, ensuring the relevant coefficient supports the inequality.
\end{proof}

\subsubsection{Claude} 
\begin{proof}
Let $Y_1, \ldots, Y_n$ be independent Poisson trials with $\Pr[Y_i = 1] = p_i$ for each $i \in \{1, \ldots, n\}$, and define $Y = \sum_{i=1}^n Y_i$. Let $\mu = \mathbb{E}[Y] = \sum_{i=1}^n p_i$ and $p = \mu/n$. Let $X_1, \ldots, X_n$ be independent Bernoulli trials each with $\Pr[X_i = 1] = p$, and define $X = \sum_{i=1}^n X_i$, so that $\mathbb{E}[X] = np = \mu$. We will prove that for any $a \ge \mu + 1$ and $b \le \mu - 1$, we have $\Pr[Y \ge a] \le \Pr[X \ge a]$ and $\Pr[Y \le b] \le \Pr[X \le b]$.

We employ the method of coupling and stochastic majorization. For any random variable $Z$ taking values in $\{0, 1, \ldots, n\}$ with $\mathbb{E}[Z] = \mu$, the probability generating function is defined by $G_Z(s) = \mathbb{E}[s^Z] = \sum_{k=0}^n \Pr[Z = k] s^k$ for $s \in [0,1]$. For the sum of independent Bernoulli trials, we have
\[
G_Y(s) = \prod_{i=1}^n \mathbb{E}[s^{Y_i}] = \prod_{i=1}^n (1 - p_i + p_i s),
\]
and
\[
G_X(s) = \prod_{i=1}^n \mathbb{E}[s^{X_i}] = \prod_{i=1}^n (1 - p + ps) = (1 - p + ps)^n.
\]

We establish the result by proving that $Y$ is stochastically dominated by $X$ in the sense of convex ordering. Specifically, we will show that for the symmetric case where all probabilities are equal, the tail probabilities are maximized.

Consider the log-generating function $\ln G_Y(s) = \sum_{i=1}^n \ln(1 - p_i + p_i s)$. For $s \ge 1$, the function $f(p) = \ln(1 - p + ps)$ is convex in $p$ when $s > 1$. To verify this, compute
\[
f'(p) = \frac{s - 1}{1 - p + ps} \quad \text{and} \quad f''(p) = \frac{(s-1)^2}{(1 - p + ps)^2} > 0.
\]
By Jensen's inequality applied to the convex function $f$, we have
\[
\sum_{i=1}^n f(p_i) \le n f\left(\frac{1}{n}\sum_{i=1}^n p_i\right) = n f(p),
\]
which yields $\ln G_Y(s) \le \ln G_X(s)$, or equivalently $G_Y(s) \le G_X(s)$ for all $s \ge 1$.

For any $a \ge \mu + 1$, we apply the Chernoff bound technique. For any $t > 0$, we have
\[
\Pr[Y \ge a] = \Pr[e^{tY} \ge e^{ta}] \le \frac{\mathbb{E}[e^{tY}]}{e^{ta}} = \frac{G_Y(e^t)}{e^{ta}}.
\]
Similarly, $\Pr[X \ge a] \le G_X(e^t)/e^{ta}$. Since $G_Y(e^t) \le G_X(e^t)$ for all $t > 0$ (as $e^t > 1$), we conclude that
\[
\Pr[Y \ge a] \le \Pr[X \ge a].
\]

For the lower tail, consider $b \le \mu - 1$. For $0 < s < 1$, the function $f(p) = \ln(1 - p + ps)$ satisfies $f''(p) = (1-s)^2/(1-p+ps)^2 > 0$, so $f$ remains convex. By Jensen's inequality, $G_Y(s) \le G_X(s)$ for all $s \in (0,1)$.

For any $t < 0$, we have
\[
\Pr[Y \le b] = \Pr[e^{tY} \ge e^{tb}] \le \frac{\mathbb{E}[e^{tY}]}{e^{tb}} = \frac{G_Y(e^t)}{e^{tb}}.
\]
Since $t < 0$, we have $e^t < 1$, and thus $G_Y(e^t) \le G_X(e^t)$. Moreover, since $b < \mu$, the optimal choice of $t < 0$ yields
\[
\Pr[Y \le b] \le \Pr[X \le b].
\]

Alternatively, we can apply a direct comparison using the following observation. By the theory of majorization, the distribution of $Y$ is majorized by the distribution of $X$ in the sense that the vector $(p_1, \ldots, p_n)$ is majorized by the vector $(p, \ldots, p)$ when both sum to $\mu$. This majorization implies that for any convex function $\phi$, we have $\mathbb{E}[\phi(Y)] \le \mathbb{E}[\phi(X)]$. Taking $\phi(z) = \mathbf{1}_{z \ge a}$ (which is increasing and thus induces the desired tail comparison through convex closure arguments) and $\phi(z) = \mathbf{1}_{z \le b}$ yields the desired inequalities for extreme deviations.

Therefore, the case where all success probabilities are equal is extremal with respect to both upper and lower tail deviations from the mean.
\end{proof}

\subsection{Problem 7}
\subsubsection{Problem Statement}
Due to W. Hoeffding [202].

\begin{lemma}
    A function $f:\R\to\R$ is said to be convex if for any $x_{1},x_{2}$ and $0\leq\lambda\leq1$, the following inequality is satisfied:
\[
f(\lambda x_{1}+(1-\lambda)x_{2})\leq\lambda f(x_{1})+(1-\lambda)f(x_{2}).
\]
Show that the function $f(x)=e^{tx}$ is convex for any $t>0$. What can you say when $t\leq 0$?
\end{lemma}
 
\begin{lemma}
    Let $Z$ be a random variable that assumes values in the interval $[0,1]$, and let $p=E[Z]$. Define the Bernoulli random variable $X$ such that $\Pr[X=1]=p$ and $\Pr[X=0]=1-p$. Show that for any convex function $f$, $E[f(Z)]\leq E[f(X)]$.
\end{lemma}

\begin{lemma}
    Let $Y_{1},\ldots,Y_{n}$ be independent and identically distributed random variables over $[0,1]$, and define $Y=\sum_{i=1}^{n}Y_{i}$. Using parts (a) and (b), derive upper and lower tail bounds for the random variable $Y$ using the Chernoff bound technique. In particular, show that
\begin{align*}
    \Pr[Y-E[Y]>\delta]\leq\exp(-2\delta^{2}/n).
\end{align*}
\end{lemma}

\begin{remark}
    While the results in this problem hold for continuous random variables, they may be a bit easier to prove in the case where $Z,Y_{1},\cdots,Y_{n}$ take on a discrete set of values in the interval $[0,1]$. Also, it should be easy to generalize this to distributions defined over arbitrary intervals $[l,h]$.
\end{remark}

\subsubsection{GPT} 
\begin{proof}
(a) For $t\in\mathbb{R}$ define $f(x)=e^{tx}$. Then
\[
f''(x)=t^{2}e^{tx}\ge 0\qquad\text{for all }x\in\mathbb{R}.
\]
Hence $f$ is convex on $\mathbb{R}$ for every real $t$. In the special case $t=0$ the function is constant, and for $t\neq0$ strict convexity holds on any interval where $e^{tx}$ is nonconstant.

(b) Let $Z$ be a random variable taking values in $[0,1]$ and set $p=\mathbb{E}[Z]$. Define the Bernoulli random variable $X$ by $\Pr[X=1]=p$ and $\Pr[X=0]=1-p$. Let $f$ be convex on $[0,1]$. For any $z\in[0,1]$ write $z=(1-z)\cdot 0+z\cdot 1$. By convexity,
\[
f(z)\le (1-z)f(0)+zf(1).
\]
Taking expectation with respect to $Z$ yields
\[
\mathbb{E}[f(Z)]\le \mathbb{E}[(1-Z)f(0)+Zf(1)]=(1-p)f(0)+pf(1)=\mathbb{E}[f(X)].
\]
Thus $\mathbb{E}[f(Z)]\le\mathbb{E}[f(X)]$ for every convex $f$.

(c) Let $Y_{1},\ldots,Y_{n}$ be i.i.d.\ random variables taking values in $[0,1]$ and set $Y=\sum_{i=1}^{n}Y_{i}$ and $\mu=\mathbb{E}[Y]=\sum_{i=1}^{n}\mathbb{E}[Y_{i}]$. Fix $\delta>0$. For any $t>0$ Markov's inequality applied to $e^{t(Y-\mu)}$ gives
\[
\Pr[Y-\mu>\delta]=\Pr\big(e^{t(Y-\mu)}>e^{t\delta}\big)\le e^{-t\delta}\,\mathbb{E}\big[e^{t(Y-\mu)}\big].
\]
By independence,
\[
\mathbb{E}\big[e^{t(Y-\mu)}\big]=\prod_{i=1}^{n}\mathbb{E}\big[e^{t(Y_{i}-\mathbb{E}[Y_{i}])}\big].
\]
For each $i$ the function $u\mapsto e^{t(u-\mathbb{E}[Y_{i}])}$ is convex on $[0,1]$, hence by part (b) the expectation of this convex function under $Y_{i}$ is bounded above by its expectation under a Bernoulli random variable $B_{i}$ with $\Pr[B_{i}=1]=\mathbb{E}[Y_{i}]$. Therefore
\[
\mathbb{E}\big[e^{t(Y_{i}-\mathbb{E}[Y_{i}])}\big]\le \mathbb{E}\big[e^{t(B_{i}-\mathbb{E}[Y_{i}])}\big].
\]
Consequently
\[
\mathbb{E}\big[e^{t(Y-\mu)}\big]\le \prod_{i=1}^{n}\mathbb{E}\big[e^{t(B_{i}-\mathbb{E}[Y_{i}])}\big].
\]
The right-hand side depends only on the means $\mathbb{E}[Y_{i}]$ and is at most the value obtained when all $\mathbb{E}[Y_{i}]$ are equal to the common mean $\mu/n$ (by the same convexity argument applied to each factor), so it suffices to bound the case of independent Bernoulli$(\mu/n)$ summands. Thus, without loss of generality, assume $Y_{i}\sim\operatorname{Bernoulli}(p)$ with $p=\mu/n$.

For a Bernoulli$(p)$ variable $B$ one has
\[
\mathbb{E}\big[e^{t(B-p)}\big]=(1-p)e^{-tp}+pe^{t(1-p)}.
\]
Hoeffding's lemma (see W.~Hoeffding [202]) states that for any random variable $W$ supported on an interval of length $1$ and with mean $0$,
\[
\mathbb{E}[e^{tW}]\le e^{t^{2}/8}\qquad\text{for all }t\in\mathbb{R}.
\]
Applying Hoeffding's lemma to each centered summand $Y_{i}-\mathbb{E}[Y_{i}]$ (or to the Bernoulli centerings above) yields
\[
\mathbb{E}\big[e^{t(Y_{i}-\mathbb{E}[Y_{i}])}\big]\le e^{t^{2}/8}\qquad\text{for each }i.
\]
Therefore
\[
\mathbb{E}\big[e^{t(Y-\mu)}\big]\le e^{nt^{2}/8},
\]
and the Chernoff bound becomes
\[
\Pr[Y-\mu>\delta]\le \exp\!\big(-t\delta+\tfrac{n t^{2}}{8}\big)\qquad\text{for all }t>0.
\]
Minimizing the quadratic exponent in $t$ gives the optimal choice $t^{*}=4\delta/n$, which upon substitution yields
\[
\Pr[Y-\mu>\delta]\le \exp\!\Big(-\frac{4\delta^{2}}{n}+\frac{n}{8}\cdot\frac{16\delta^{2}}{n^{2}}\Big)
=\exp\!\Big(-\frac{2\delta^{2}}{n}\Big).
\]
This proves the stated upper tail bound. By applying the same argument to the variables $1-Y_{i}$ (or replacing $t$ by $-t$) one obtains an analogous lower tail bound. The bound extends in the same way to independent but not necessarily identically distributed summands by applying Hoeffding's lemma to each centered summand with its own mean and then minimizing as above.
\end{proof}

\subsubsection{Grok} 
\begin{proof}
We first prove that the function $f(x) = e^{tx}$ is convex for any real number $t$. The second derivative of $f(x)$ is $f''(x) = t^2 e^{tx}$. Since $t^2 \ge 0$ and $e^{tx} > 0$ for all real $x$ and $t$, it follows that $f''(x) \ge 0$ for all $x$, which implies that $f(x)$ is convex. When $t = 0$, $f(x) = 1$ is a constant function, which is convex. For $t \le 0$, the function remains convex as the second derivative is still non-negative.

Next, let $Z$ be a random variable with values in $[0,1]$ and $\mathbb{E}[Z] = p$. Define the Bernoulli random variable $X$ with $\Pr(X = 1) = p$ and $\Pr(X = 0) = 1 - p$. For any convex function $f: \mathbb{R} \to \mathbb{R}$, since $f$ lies below its chords, for any $z \in [0,1]$,
\[
f(z) \le (1 - z) f(0) + z f(1).
\]
Taking expectations,
\[
\mathbb{E}[f(Z)] \le \mathbb{E}\left[ (1 - Z) f(0) + Z f(1) \right] = (1 - p) f(0) + p f(1) = \mathbb{E}[f(X)].
\]

Finally, let $Y_1, \dots, Y_n$ be independent random variables each taking values in $[0,1]$ with $\mathbb{E}[Y_i] = q$ for all $i$, and let $Y = \sum_{i=1}^n Y_i$, so $\mu = \mathbb{E}[Y] = n q$. For the upper tail, consider $t > 0$. By the previous part, $\mathbb{E}[e^{t Y_i}] \le 1 - q + q e^t$. Thus,
\[
\mathbb{E}[e^{t Y}] = \prod_{i=1}^n \mathbb{E}[e^{t Y_i}] \le (1 - q + q e^t)^n.
\]
By Markov's inequality,
\[
\Pr(Y > \mu + \delta) = \Pr(e^{t Y} > e^{t (\mu + \delta)}) \le e^{-t (\mu + \delta)} \mathbb{E}[e^{t Y}] \le e^{-t (\mu + \delta)} (1 - q + q e^t)^n.
\]
To minimize the bound, set $e^t = 1 + \frac{\delta}{\mu}$ (assuming $\delta > 0$), which gives $t = \ln\left(1 + \frac{\delta}{\mu}\right)$. However, to achieve the specific form, use Hoeffding's lemma: for $Z_i = Y_i - q \in [-q, 1 - q]$,
\[
\mathbb{E}[e^{t Z_i}] \le e^{t^2 / 8},
\]
leading to
\[
\Pr(Y - \mu > \delta) \le \exp\left( -\frac{2 \delta^2}{n} \right).
\]
For the lower tail, consider $t < 0$. Since $e^{t x}$ is convex for $t < 0$,
\[
\mathbb{E}[e^{t Y_i}] \le (1 - q) e^{t \cdot 0} + q e^{t \cdot 1} = 1 - q + q e^t,
\]
and
\[
\Pr(Y < \mu - \delta) \le e^{-t (\mu - \delta)} (1 - q + q e^t)^n.
\]
Optimizing similarly yields
\[
\Pr(Y - \mu < -\delta) \le \exp\left( -\frac{2 \delta^2}{n} \right).
\]
\end{proof}

\subsubsection{Gemini}
\begin{proof}[Proof of Lemma 1]
    A function $f: \mathbb{R} \to \mathbb{R}$ is convex if its domain is a convex set and its second derivative is non-negative everywhere. Let $f(x) = e^{tx}$. The first derivative with respect to $x$ is
    \begin{align*}
        f'(x) = \frac{d}{dx} e^{tx} = t e^{tx}.
    \end{align*}
    The second derivative is
    \begin{align*}
        f''(x) = \frac{d}{dx} (t e^{tx}) = t^2 e^{tx}.
    \end{align*}
    The exponential function $e^{tx}$ is strictly positive for all real numbers $x$ and $t$. The term $t^2$ is non-negative for all real numbers $t$. Therefore,
    \begin{align*}
        f''(x) = t^2 e^{tx} \ge 0 \quad \text{for all } x \in \mathbb{R}.
    \end{align*}
    Since the second derivative is non-negative, $f(x) = e^{tx}$ is a convex function for any $t > 0$. Furthermore, since the condition $t^2 \ge 0$ holds regardless of the sign of $t$, the function $f(x) = e^{tx}$ is also convex for $t \le 0$.
\end{proof}

\begin{proof}[Proof of Lemma 2]
    Since $f$ is a convex function on the interval $[0,1]$, for any $z \in [0,1]$, the value $f(z)$ lies on or below the chord connecting the points $(0, f(0))$ and $(1, f(1))$. The equation of this chord is given by
    \begin{align*}
        L(z) = f(0) + \frac{f(1) - f(0)}{1 - 0}(z - 0) = (1-z)f(0) + z f(1).
    \end{align*}
    By the definition of convexity on the interval $[0,1]$, we have the inequality
    \begin{align*}
        f(z) \le (1-z)f(0) + z f(1) \quad \text{for all } z \in [0,1].
    \end{align*}
    Since the random variable $Z$ assumes values in $[0,1]$, this inequality holds for $Z$ almost surely. Taking the expectation of both sides yields
    \begin{align*}
        E[f(Z)] &\le E[(1-Z)f(0) + Z f(1)] \\
        &= f(0)E[1-Z] + f(1)E[Z].
    \end{align*}
    Given $p = E[Z]$, we have $E[1-Z] = 1 - p$. Thus,
    \begin{align*}
        E[f(Z)] \le (1-p)f(0) + p f(1).
    \end{align*}
    Now consider the Bernoulli random variable $X$ where $\Pr[X=1]=p$ and $\Pr[X=0]=1-p$. The expectation of $f(X)$ is
    \begin{align*}
        E[f(X)] = \Pr[X=0]f(0) + \Pr[X=1]f(1) = (1-p)f(0) + p f(1).
    \end{align*}
    Comparing the two results, we conclude
    \begin{align*}
        E[f(Z)] \le E[f(X)].
    \end{align*}
\end{proof}

\begin{proof}[Proof of Lemma 3]
    Let $\mu = E[Y] = \sum_{i=1}^n E[Y_i]$. Since $Y_i$ are identically distributed, let $p = E[Y_i]$, so $\mu = np$. We apply the Chernoff bound technique. For any $t > 0$, using Markov's inequality on the exponential function:
    \begin{align*}
        \Pr[Y - \mu > \delta] &= \Pr[Y > \mu + \delta] \\
        &= \Pr[e^{tY} > e^{t(\mu + \delta)}] \\
        &\le e^{-t(\mu + \delta)} E[e^{tY}].
    \end{align*}
    Since $Y = \sum_{i=1}^n Y_i$ and the $Y_i$ are independent,
    \begin{align*}
        E[e^{tY}] = E\left[\prod_{i=1}^n e^{tY_i}\right] = \prod_{i=1}^n E[e^{tY_i}].
    \end{align*}
    From Lemma 1, the function $f(x) = e^{tx}$ is convex. From Lemma 2, for each $Y_i \in [0,1]$ with mean $p$, we have $E[e^{tY_i}] \le E[e^{tX_i}]$, where $X_i$ is a Bernoulli random variable with parameter $p$.
    \begin{align*}
        E[e^{tX_i}] = (1-p)e^{t \cdot 0} + p e^{t \cdot 1} = 1 - p + p e^t.
    \end{align*}
    Combining these, the bound becomes
    \begin{align*}
        \Pr[Y - \mu > \delta] &\le e^{-t(np + \delta)} \prod_{i=1}^n (1 - p + p e^t) \\
        &= e^{-t\delta} \left[ e^{-tp} (1 - p + p e^t) \right]^n.
    \end{align*}
    We analyze the term inside the bracket. Let $\phi(t) = \ln(e^{-tp}(1 - p + p e^t)) = -tp + \ln(1 - p + p e^t)$.
    We perform a Taylor expansion of $\phi(t)$ around $t=0$.
    First derivative:
    \begin{align*}
        \phi'(t) = -p + \frac{p e^t}{1 - p + p e^t}.
    \end{align*}
    Second derivative:
    \begin{align*}
        \phi''(t) = \frac{(1 - p + p e^t)(p e^t) - (p e^t)(p e^t)}{(1 - p + p e^t)^2} = \frac{p e^t (1 - p)}{(1 - p + p e^t)^2}.
    \end{align*}
    Let $u = p e^t / (1 - p + p e^t)$. Note that $0 < u < 1$. Then $\phi''(t) = u(1-u)$. The function $g(u) = u(1-u)$ is maximized at $u=1/2$, so $\phi''(t) \le 1/4$ for all $t$.
    Using Taylor's theorem with Lagrange remainder, $\phi(t) = \phi(0) + t \phi'(0) + \frac{t^2}{2} \phi''(\xi)$ for some $\xi \in (0, t)$.
    Observe $\phi(0) = 0$ and $\phi'(0) = -p + p = 0$. Thus, $\phi(t) \le \frac{t^2}{8}$.
    Substituting this back into the probability bound:
    \begin{align*}
        \Pr[Y - \mu > \delta] &\le e^{-t\delta} \left( e^{t^2/8} \right)^n = \exp\left( -t\delta + \frac{n t^2}{8} \right).
    \end{align*}
    To find the tightest bound, we minimize the exponent $h(t) = -t\delta + \frac{n t^2}{8}$ with respect to $t$.
    \begin{align*}
        h'(t) = -\delta + \frac{nt}{4} = 0 \implies t = \frac{4\delta}{n}.
    \end{align*}
    Substituting $t = 4\delta/n$ into the exponent:
    \begin{align*}
        h\left(\frac{4\delta}{n}\right) = -\frac{4\delta^2}{n} + \frac{n}{8} \left(\frac{16\delta^2}{n^2}\right) = -\frac{4\delta^2}{n} + \frac{2\delta^2}{n} = -\frac{2\delta^2}{n}.
    \end{align*}
    Therefore,
    \begin{align*}
        \Pr[Y - E[Y] > \delta] \le \exp\left( -\frac{2\delta^2}{n} \right).
    \end{align*}
    By symmetry, the same bound applies to the lower tail $\Pr[E[Y] - Y > \delta]$.
\end{proof}

\subsubsection{Claude} 
\begin{proof}[Proof of Lemma (a)]
Let $f(x) = e^{tx}$ for $t > 0$. To show that $f$ is convex, we verify the convexity condition. Let $x_1, x_2 \in \mathbb{R}$ and $\lambda \in [0,1]$. We compute the second derivative of $f$:
\[
f'(x) = te^{tx} \quad \text{and} \quad f''(x) = t^2 e^{tx}.
\]
Since $t > 0$, we have $f''(x) = t^2 e^{tx} > 0$ for all $x \in \mathbb{R}$. By the second derivative test, $f$ is strictly convex on $\mathbb{R}$.

Alternatively, we can verify the convexity condition directly. We need to show that
\[
f(\lambda x_1 + (1-\lambda)x_2) \le \lambda f(x_1) + (1-\lambda)f(x_2).
\]
The left-hand side is
\[
f(\lambda x_1 + (1-\lambda)x_2) = e^{t(\lambda x_1 + (1-\lambda)x_2)} = e^{t\lambda x_1} \cdot e^{t(1-\lambda)x_2} = (e^{tx_1})^\lambda \cdot (e^{tx_2})^{1-\lambda}.
\]
By the weighted AM-GM inequality applied to positive numbers $a = e^{tx_1}$ and $b = e^{tx_2}$, we have
\[
a^\lambda b^{1-\lambda} \le \lambda a + (1-\lambda)b,
\]
which gives
\[
(e^{tx_1})^\lambda (e^{tx_2})^{1-\lambda} \le \lambda e^{tx_1} + (1-\lambda)e^{tx_2}.
\]
Therefore, $f(\lambda x_1 + (1-\lambda)x_2) \le \lambda f(x_1) + (1-\lambda)f(x_2)$, establishing convexity.

For $t = 0$, we have $f(x) = e^0 = 1$, which is a constant function and hence both convex and concave. For $t < 0$, write $t = -s$ where $s > 0$. Then $f(x) = e^{-sx}$, and
\[
f''(x) = s^2 e^{-sx} > 0,
\]
so $f$ is still convex. Thus, $f(x) = e^{tx}$ is convex for all $t \in \mathbb{R}$.
\end{proof}

\begin{proof}[Proof of Lemma (b)]
Let $Z$ be a random variable taking values in $[0,1]$ with $p = \mathbb{E}[Z]$. Define the Bernoulli random variable $X$ such that $\Pr[X = 1] = p$ and $\Pr[X = 0] = 1-p$. Let $f$ be a convex function. We will show that $\mathbb{E}[f(Z)] \le \mathbb{E}[f(X)]$.

For any $z \in [0,1]$, we can write $z = z \cdot 1 + (1-z) \cdot 0$. By convexity of $f$, we have
\[
f(z) \le z f(1) + (1-z) f(0).
\]
This inequality holds pointwise for every value $z$ that $Z$ can take. Taking expectations on both sides, we obtain
\[
\mathbb{E}[f(Z)] \le \mathbb{E}[Z f(1) + (1-Z)f(0)] = f(1)\mathbb{E}[Z] + f(0)(1 - \mathbb{E}[Z]) = p f(1) + (1-p)f(0).
\]
On the other hand,
\[
\mathbb{E}[f(X)] = f(1) \Pr[X = 1] + f(0)\Pr[X = 0] = p f(1) + (1-p)f(0).
\]
Therefore, $\mathbb{E}[f(Z)] \le \mathbb{E}[f(X)]$, as desired.
\end{proof}

\begin{proof}[Proof of Lemma (c)]
Let $Y_1, \ldots, Y_n$ be independent and identically distributed random variables over $[0,1]$, and define $Y = \sum_{i=1}^n Y_i$. Let $\mu = \mathbb{E}[Y] = n\mathbb{E}[Y_1]$. We will derive the upper tail bound $\Pr[Y - \mathbb{E}[Y] > \delta] \le \exp(-2\delta^2/n)$ for $\delta > 0$.

By the Chernoff bound technique, for any $t > 0$, we have
\[
\Pr[Y - \mu > \delta] = \Pr[e^{t(Y-\mu)} > e^{t\delta}] \le \frac{\mathbb{E}[e^{t(Y-\mu)}]}{e^{t\delta}} = e^{-t\delta} \mathbb{E}[e^{tY}] e^{-t\mu}.
\]
By independence, $\mathbb{E}[e^{tY}] = \prod_{i=1}^n \mathbb{E}[e^{tY_i}] = (\mathbb{E}[e^{tY_1}])^n$. Therefore,
\[
\Pr[Y - \mu > \delta] \le e^{-t\delta - t\mu} (\mathbb{E}[e^{tY_1}])^n.
\]

Let $p = \mathbb{E}[Y_1]$. Define a Bernoulli random variable $X_1$ with $\Pr[X_1 = 1] = p$ and $\Pr[X_1 = 0] = 1-p$. By Lemma (b) with the convex function $f(x) = e^{tx}$ (which is convex for $t > 0$ by Lemma (a)), we have
\[
\mathbb{E}[e^{tY_1}] \le \mathbb{E}[e^{tX_1}] = pe^t + (1-p) = 1 + p(e^t - 1).
\]
Using the inequality $1 + x \le e^x$ for all $x \in \mathbb{R}$, we obtain
\[
\mathbb{E}[e^{tY_1}] \le e^{p(e^t - 1)}.
\]
Therefore,
\[
\Pr[Y - \mu > \delta] \le e^{-t\delta - tnp} e^{np(e^t - 1)} = e^{-t\delta + np(e^t - 1 - t)}.
\]

Define $\psi(t) = -t\delta + np(e^t - 1 - t)$. To minimize this bound, we take the derivative with respect to $t$ and set it equal to zero:
\[
\psi'(t) = -\delta + np(e^t - 1) = 0,
\]
which gives $e^t = 1 + \delta/(np)$. However, since $Y_i \in [0,1]$, we have a tighter bound using Hoeffding's lemma.

By Hoeffding's lemma, for a random variable $W \in [a,b]$ with mean $\mathbb{E}[W]$, we have
\[
\mathbb{E}[e^{t(W - \mathbb{E}[W])}] \le e^{t^2(b-a)^2/8}.
\]
Applying this to $Y_i \in [0,1]$, we obtain
\[
\mathbb{E}[e^{t(Y_i - p)}] \le e^{t^2/8}.
\]
Therefore,
\[
\mathbb{E}[e^{t(Y - \mu)}] = \prod_{i=1}^n \mathbb{E}[e^{t(Y_i - p)}] \le e^{nt^2/8}.
\]
Thus,
\[
\Pr[Y - \mu > \delta] \le e^{-t\delta + nt^2/8}.
\]
Minimizing over $t > 0$, we set the derivative equal to zero: $-\delta + nt/4 = 0$, giving $t = 4\delta/n$. Substituting back, we obtain
\[
\Pr[Y - \mu > \delta] \le \exp\left(-\frac{4\delta^2}{n} + \frac{n \cdot 16\delta^2}{8n^2}\right) = \exp\left(-\frac{4\delta^2}{n} + \frac{2\delta^2}{n}\right) = \exp\left(-\frac{2\delta^2}{n}\right).
\]
Therefore, $\Pr[Y - \mathbb{E}[Y] > \delta] \le \exp(-2\delta^2/n)$, as required.
\end{proof}

\subsection{Problem 8}
\subsubsection{Problem Statement}
\begin{lemma}
    Consider a BPP algorithm that has an error probability of $1/2-1/p(n)$, for some polynomially bounded function $p(n)$ of the input size $n$. Using the Chernoff bound on the tail of the binomial distribution, show that a polynomial number of independent repetitions of this algorithm suffice to reduce the error probability to $1/2^n$.
\end{lemma}

\subsubsection{GPT} 
\begin{proof}
Let the given BPP algorithm on inputs of length $n$ produce a (random) Boolean output and suppose its probability of producing the correct answer equals $1/2+\varepsilon(n)$ with $\varepsilon(n)=1/p(n)>0$, where $p(n)$ is a polynomial in $n$. Run the algorithm independently $m$ times and let $X_{1},\dots,X_{m}$ be the indicators of the event ``trial $i$ is correct''. The $X_{i}$ are independent Bernoulli random variables with
\[
\mathbb{E}[X_{i}]=\frac{1}{2}+\varepsilon(n).
\]
Let $S=\sum_{i=1}^{m}X_{i}$ and consider the majority rule: output the value that occurs in at least $m/2$ trials. The majority is incorrect exactly when $S\le m/2$. We estimate this probability using Hoeffding's (or additive Chernoff) inequality. Since $0\le X_{i}\le 1$ and the $X_{i}$ are independent,
\[
\Pr\big(S\le m/2\big)=\Pr\big(S-\mathbb{E}[S]\le m/2-\mathbb{E}[S]\big)
=\Pr\big(S-\mathbb{E}[S]\le -m\varepsilon(n)\big)
\le \exp\!\Big(-\frac{2(m\varepsilon(n))^{2}}{m}\Big)
=\exp\!\big(-2m\varepsilon(n)^{2}\big).
\]
Substituting $\varepsilon(n)=1/p(n)$ gives
\[
\Pr\big(\text{majority is incorrect}\big)\le \exp\!\Big(-\frac{2m}{p(n)^{2}}\Big).
\]
To make this failure probability at most $2^{-n}$ it suffices to choose $m$ so that
\[
\exp\!\Big(-\frac{2m}{p(n)^{2}}\Big)\le 2^{-n},
\]
equivalently
\[
m\ge \frac{p(n)^{2}n\ln 2}{2}.
\]
Since $p(n)$ is polynomial in $n$, the choice $m=\left\lceil \frac{p(n)^{2}n\ln 2}{2}\right\rceil$ is polynomial in $n$. Hence a polynomial number of independent repetitions followed by majority voting reduces the error probability to at most $2^{-n}$. This completes the proof.
\end{proof}

\subsubsection{Grok} 
\begin{proof}
Let $\gamma = 1/p(n)$. The given BPP algorithm outputs the correct answer with probability at least $1/2 + \gamma$ and errs with probability at most $1/2 - \gamma$. Run the algorithm $k$ times independently, where $k$ is odd, and output the majority vote.

Without loss of generality, assume the input is a yes-instance. Then each run accepts with probability $r \ge 1/2 + \gamma$. Let $S \sim \mathrm{Bin}(k, r)$ be the number of accepts, so $\mu = \mathbb{E}[S] \ge k(1/2 + \gamma)$. The amplified algorithm errs if $S \le (k-1)/2$.

Consider the worst case $r = 1/2 + \gamma$, so $\mu = k(1/2 + \gamma)$. Set $\delta = 2\gamma/(1 + 2\gamma)$, so $(1 - \delta)\mu = k/2$. By the Chernoff bound,
\[
\Pr\left(S \le \frac{k-1}{2}\right) \le \Pr\left(S < \frac{k}{2}\right) = \Pr\left(S < (1 - \delta)\mu\right) < \exp\left(-\frac{\mu \delta^2}{2}\right) = \exp\left(-\frac{k \gamma^2}{1 + 2\gamma}\right).
\]
The case for no-instances is symmetric.

To achieve error probability less than $1/2^n$, choose
\[
k = \left\lceil \frac{(1 + 2\gamma) (n \ln 2 + \ln 2)}{\gamma^2} \right\rceil.
\]
Then $\exp\left(-k \gamma^2 / (1 + 2\gamma)\right) < 1/2^{n+1} < 1/2^n$. Since $p(n)$ is polynomially bounded, there exist constants $c, d > 0$ such that $p(n) \le c n^d$ for sufficiently large $n$, so $\gamma \ge 1/(c n^d)$ and $k = O(n^{2d+1})$, which is polynomial in $n$.
\end{proof}

\subsubsection{Gemini}
\begin{proof}
Let $\mathcal{A}$ be a probabilistic algorithm for a language $L$. By the hypothesis, for any input string $x \in \{0,1\}^n$, the probability that $\mathcal{A}$ outputs the correct answer is at least $1/2 + 1/p(n)$. We define a new algorithm $\mathcal{A}'$ that runs $\mathcal{A}$ on input $x$ for $k$ independent repetitions and accepts the majority decision. Let $X_i$ be the indicator random variable for the event that the $i$-th execution of $\mathcal{A}$ returns the correct answer, for $i = 1, \dots, k$. The variables $X_i$ are independent Bernoulli trials with success probability $\Pr[X_i = 1] \ge 1/2 + 1/p(n)$.

Let $X = \sum_{i=1}^k X_i$ denote the total number of correct outputs. The expected value of the sum is $\mathbb{E}[X] \ge k(1/2 + 1/p(n))$. The algorithm $\mathcal{A}'$ outputs the incorrect answer only if the majority of the trials are incorrect, which corresponds to the event $X \le k/2$. To bound the probability of this event, we utilize the additive form of the Chernoff bound (Hoeffding's inequality). For independent variables bounded in $[0,1]$, the inequality states that $\Pr[X \le \mathbb{E}[X] - \lambda] \le e^{-2\lambda^2/k}$.

We set the threshold for error at $k/2$. The deviation from the expectation is $\lambda = \mathbb{E}[X] - k/2$. Substituting the lower bound for the expectation, we calculate the minimum deviation as $\lambda = k(1/2 + 1/p(n)) - k/2 = k/p(n)$. Applying the Chernoff bound, we obtain:
\[
\Pr[\text{Error}] = \Pr[X \le k/2] \le \exp\left(-\frac{2(k/p(n))^2}{k}\right) = \exp\left(-\frac{2k}{p(n)^2}\right).
\]
We require the error probability to be bounded by $2^{-n}$. Therefore, we set the inequality $\exp(-2k/p(n)^2) \le 2^{-n}$. Taking the natural logarithm of both sides yields $-2k/p(n)^2 \le -n \ln 2$, which is equivalent to $k \ge \frac{n (\ln 2) p(n)^2}{2}$. Since $p(n)$ is a polynomially bounded function, there exists a polynomial $q(n) = \lceil n p(n)^2 \rceil$ such that choosing $k = q(n)$ satisfies the condition. Thus, a polynomial number of repetitions suffices to reduce the error probability to $2^{-n}$.
\end{proof}

\subsubsection{Claude} 
\begin{proof}
Let $\mathcal{A}$ be a BPP algorithm with error probability $1/2 - 1/p(n)$ for some polynomially bounded function $p(n)$ of the input size $n$. This means that for any input of size $n$, the algorithm $\mathcal{A}$ produces the correct answer with probability at least $1/2 + 1/p(n)$.

Consider the following amplification procedure: run algorithm $\mathcal{A}$ independently $m$ times on the same input, and output the majority answer. For $i = 1, \ldots, m$, let $X_i$ be the indicator random variable that takes the value $1$ if the $i$-th execution of $\mathcal{A}$ produces the correct answer, and $0$ otherwise. Then $X_1, \ldots, X_m$ are independent Bernoulli random variables with $\Pr[X_i = 1] = p_i \ge 1/2 + 1/p(n)$ for each $i$.

Define $X = \sum_{i=1}^m X_i$ to be the total number of correct answers. The majority vote produces an incorrect answer if and only if $X < m/2$, that is, fewer than half of the executions produce the correct answer. We will bound the probability of this event using the Chernoff bound.

Let $\mu = \mathbb{E}[X] = \sum_{i=1}^m \mathbb{E}[X_i] = \sum_{i=1}^m p_i \ge m(1/2 + 1/p(n))$. The error event occurs when $X < m/2$, which can be written as
\[
X < \frac{m}{2} = m\left(\frac{1}{2} + \frac{1}{p(n)}\right) - \frac{m}{p(n)} \le \mu - \frac{m}{p(n)}.
\]
Define $\delta = m/(\mu p(n))$. Since $\mu \ge m(1/2 + 1/p(n))$, we have
\[
\delta = \frac{m}{\mu p(n)} \le \frac{m}{m(1/2 + 1/p(n))p(n)} = \frac{1}{p(n)/2 + 1} \le 1
\]
for sufficiently large $n$ such that $p(n) \ge 2$. The event $X < m/2$ satisfies
\[
X < \mu - \frac{m}{p(n)} = \mu\left(1 - \frac{m}{\mu p(n)}\right) = \mu(1 - \delta).
\]

By the Chernoff bound for the lower tail (Theorem~\ref{thm:theorem_4_2} or its equivalent form), for independent Poisson trials $X_1, \ldots, X_m$ with $X = \sum_{i=1}^m X_i$, $\mu = \mathbb{E}[X]$, and $0 < \delta \le 1$, we have
\[
\Pr[X < (1-\delta)\mu] < \exp\left(-\frac{\mu\delta^2}{2}\right).
\]
Substituting $\delta = m/(\mu p(n))$, we obtain
\[
\Pr\left[\text{error}\right] = \Pr\left[X < \frac{m}{2}\right] < \exp\left(-\frac{\mu}{2} \cdot \frac{m^2}{\mu^2 p(n)^2}\right) = \exp\left(-\frac{m^2}{2\mu p(n)^2}\right).
\]
Since $\mu \ge m(1/2 + 1/p(n)) \ge m/2$ for $p(n) \ge 1$, we have
\[
\Pr\left[\text{error}\right] < \exp\left(-\frac{m^2}{2 \cdot m(1 + 2/p(n)) \cdot p(n)^2}\right) = \exp\left(-\frac{m}{2p(n)^2(1 + 2/p(n))}\right).
\]
For sufficiently large $n$, we have $p(n) \ge 1$, so $1 + 2/p(n) \le 3$. Thus,
\[
\Pr\left[\text{error}\right] < \exp\left(-\frac{m}{6p(n)^2}\right).
\]

To ensure that the error probability is at most $1/2^n$, it suffices to require
\[
\exp\left(-\frac{m}{6p(n)^2}\right) \le \frac{1}{2^n} = \exp(-n\ln 2).
\]
Taking logarithms, this is equivalent to
\[
-\frac{m}{6p(n)^2} \le -n\ln 2,
\]
which gives
\[
m \ge 6p(n)^2 n \ln 2.
\]
Since $p(n)$ is polynomially bounded, there exists a constant $c > 0$ such that $p(n) \le n^c$ for all sufficiently large $n$. Therefore, choosing
\[
m = \lceil 6p(n)^2 n \ln 2 \rceil \le 6n^{2c+1} \ln 2 + 1 = O(n^{2c+1})
\]
ensures that the error probability of the majority vote is at most $1/2^n$. Since $m$ is polynomial in $n$, we conclude that a polynomial number of independent repetitions of algorithm $\mathcal{A}$ suffice to reduce the error probability to $1/2^n$.
\end{proof}

\subsection{Problem 9}
\subsubsection{Problem Statement}

\subsubsection{GPT} 

\subsubsection{Grok} 

\subsubsection{Gemini}

\subsubsection{Claude} 
\section{Problems in Chapter 5}\label{app:problems_chapter5}

\subsection{Problem 5.1}\label{sec:problem_05_01}

\subsubsection{Problem Statement}

\begin{theorem}
Let $X$ be a random variable with expectation $\mu$ such that the moment generating function $\mathbb{E}[e^{t|X|}]$ is finite for some $t>0$. The following inequalities hold:

\textbf{Chernoff Bound:}
\[
\Pr(|X|\ge \delta)\le \min_{t\ge 0}\frac{\mathbb{E}[e^{t|X|}]}{e^{t\delta}}.
\]

\textbf{$k$th-Moment Bound:}
\[
\Pr(|X|\ge \delta)\le\frac{\mathbb{E}[|X|^{k}]}{\delta^{k}}.
\]

For every $\delta>0$, there exists a value of $k$ for which the $k$th-moment bound is stricter than the Chernoff bound. Despite this apparent strength, the Chernoff bound remains preferable to the $k$th-moment bound in common applications.
\end{theorem}

\subsubsection{GPT} 

\begin{proof}
(a) We are tasked with showing that for each $\delta > 0$, there exists a choice of $k$ such that the $k$th-moment bound is stronger than the Chernoff bound. 

The Chernoff bound is given by
\[
\Pr[|X| \geq \delta] \leq \min_{t \geq 0} \frac{\mathbb{E}[e^{t|X|}]}{e^{t\delta}},
\]
and the $k$th-moment bound is given by
\[
\Pr[|X| \geq \delta] \leq \frac{\mathbb{E}[|X|^k]}{\delta^k}.
\]
To compare these bounds, we consider the Taylor expansion of the moment generating function $\mathbb{E}[e^{t|X|}]$ around $t = 0$. For small $t$, we have the expansion
\[
\mathbb{E}[e^{t|X|}] = 1 + t \mathbb{E}[|X|] + \frac{t^2}{2} \mathbb{E}[|X|^2] + \frac{t^3}{6} \mathbb{E}[|X|^3] + \cdots.
\]
For $t \geq 0$, this function is convex, and by Markov's inequality, we know that for any $t > 0$,
\[
\Pr[|X| \geq \delta] = \Pr[e^{t|X|} \geq e^{t\delta}] \leq \frac{\mathbb{E}[e^{t|X|}]}{e^{t\delta}}.
\]
Thus, the Chernoff bound is essentially derived from the moment generating function. Now, we seek to show that for each $\delta$, there exists a value of $k$ for which the $k$th-moment bound becomes stronger than the Chernoff bound.

To do this, consider the following. By the Taylor expansion, for sufficiently small $t$, the moment generating function $\mathbb{E}[e^{t|X|}]$ behaves like
\[
\mathbb{E}[e^{t|X|}] \approx 1 + t \mathbb{E}[|X|] + \frac{t^2}{2} \mathbb{E}[|X|^2] + \cdots.
\]
Let us now compare the tail probability from both bounds. From the $k$th-moment bound, we have
\[
\Pr[|X| \geq \delta] \leq \frac{\mathbb{E}[|X|^k]}{\delta^k}.
\]
The tail inequality from the Chernoff bound involves the moment generating function, but for sufficiently small $t$, we can approximate this using the lower order terms of the Taylor series, which is related to the moments of $|X|$. Hence, for large enough $k$, the $k$th-moment bound can become stronger because it relies on the moments of $|X|$, which grow at a rate comparable to or faster than the moment generating function for small $t$.

More precisely, for large $t$, the Chernoff bound becomes weaker due to the exponential growth of the denominator, while the $k$th-moment bound does not suffer from this growth as much. Therefore, for large $k$, the $k$th-moment bound will indeed be stronger than the Chernoff bound for each $\delta$.

(b) Despite the $k$th-moment bound potentially being stronger for large values of $k$, we generally prefer the Chernoff bound for the following reasons:

1. Tighter Tail Behavior: The Chernoff bound provides an exponential decay rate that is typically tighter than the polynomial decay of the $k$th-moment bound, especially for small $\delta$. The Chernoff bound is often more useful for bounding the tail probabilities of sums of random variables, where exponential decay provides a much sharper bound.

2. Generality and Applicability: The Chernoff bound applies to any random variable for which the moment generating function is finite at some positive $t$, whereas the $k$th-moment bound requires knowledge of higher moments, which might not always exist or be easily computable for all distributions.

3. Simplicity: The Chernoff bound is often easier to compute in practice, as it involves the moment generating function, which can be explicitly derived for many common distributions (e.g., Bernoulli, Poisson, Normal). In contrast, the $k$th-moment bound requires higher moments, which can be more difficult to compute or estimate.

In conclusion, while the $k$th-moment bound can sometimes be stronger, the Chernoff bound is generally preferred due to its exponential decay, simplicity, and broad applicability.
\end{proof}

\subsubsection{Grok} 

\begin{proof}
(a) Let $C = \min_{t>0} \frac{\mathbb{E}[\exp(tX)]}{e^{t\sigma}}$. Suppose, for the sake of contradiction, that $\mathbb{E}[X^k] \geq C \sigma^k$ for every integer $k \geq 0$. Let $M(t) = \mathbb{E}[\exp(tX)] = \sum_{k=0}^\infty \frac{t^k}{k!} \mathbb{E}[X^k]$ denote the moment generating function, which admits this Taylor expansion in some interval $(-\delta, \delta)$ with $\delta > 0$ since $M(t) < \infty$ for some $t > 0$. Then, for $0 < t < \delta$,
\[
M(t) = \sum_{k=0}^\infty \frac{t^k}{k!} \mathbb{E}[X^k] \geq \sum_{k=0}^\infty \frac{t^k}{k!} C \sigma^k = C e^{t \sigma}.
\]
In particular, since $M(t) < \infty$ for some $t > 0$, there exists $T > 0$ such that $M(t) < \infty$ for all $t \in (0, T]$. Thus, $M(t) \geq C e^{t \sigma}$ for all $t \in (0, T]$, which implies
\[
\frac{M(t)}{e^{t \sigma}} \geq C
\]
for all $t \in (0, T]$. Since $C = \inf_{t > 0} \frac{M(t)}{e^{t \sigma}} \leq \inf_{t \in (0, T]} \frac{M(t)}{e^{t \sigma}}$, it follows that $C \geq C$, so equality holds throughout: $\inf_{t \in (0, T]} \frac{M(t)}{e^{t \sigma}} = C$. In particular, there exists $t^* \in (0, T]$ achieving the infimum (by continuity of $t \mapsto M(t) e^{-t \sigma}$ on the compact interval $[ \epsilon, T]$ for small $\epsilon > 0$, and taking limit as $\epsilon \to 0^+$ if necessary; the value at $t^*$ is $C$). Thus, $M(t^*) = C e^{t^* \sigma}$. But then
\[
\sum_{k=0}^\infty \frac{(t^*)^k}{k!} \mathbb{E}[X^k] = C e^{t^* \sigma} = \sum_{k=0}^\infty \frac{(t^*)^k}{k!} C \sigma^k.
\]
Since $\frac{(t^*)^k}{k!} > 0$ for each $k \geq 0$ and $\mathbb{E}[X^k] \geq C \sigma^k$ for each $k \geq 0$, equality in the series implies $\mathbb{E}[X^k] = C \sigma^k$ for every $k \geq 0$. In particular, taking $k=0$ yields $\mathbb{E}[1] = 1 = C \cdot 1$, so $C = 1$. Thus, $\mathbb{E}[X^k] = \sigma^k$ for every $k \geq 0$. This implies $X = \sigma$ almost surely (as the moments uniquely determine the distribution for distributions with finite MGF), so $\mu = \mathbb{E}[X] = \sigma$, contradicting $\mu < \sigma$. Therefore, there must exist some integer $k \geq 0$ such that $\mathbb{E}[X^k] < C \sigma^k$, or equivalently, $\frac{\mathbb{E}[X^k]}{\sigma^k} < C$. For $k=0$, the inequality is $1 < C$, which is impossible since $C \leq 1$; thus, this $k \geq 1$.

(b) Although for each fixed $\sigma$ there exists $k$ (depending on $\sigma$) such that the $k$th-moment bound is strictly tighter than the Chernoff bound, the value of this $k$ generally depends on $\sigma$ and may grow large as $\sigma$ increases or as tighter tail probabilities are desired. Computing or bounding $\mathbb{E}[X^k]$ for such large $k$ can be computationally intensive or infeasible, especially for complex distributions or high-dimensional settings like sums of random variables. In contrast, the Chernoff bound requires only evaluating and minimizing the (often analytically tractable) moment generating function $M(t)$ over $t > 0$, providing a uniform method that scales well for deriving concentration inequalities (e.g., via subadditivity of $\log M(t)$ for independent sums) without needing explicit high-order moments.
\end{proof}

\subsubsection{Gemini}

\begin{proof}
\textbf{Part (a):}
We aim to show that for any fixed $\delta > 0$, there exists an integer $k \geq 0$ such that the $k$th-moment bound is no greater than the Chernoff bound. Let the Chernoff bound for a fixed $\delta$ be denoted by $C(\delta)$. By definition,
\[
C(\delta) = \inf_{t \geq 0} \frac{\mathbb{E}[e^{t|X|}]}{e^{t\delta}}.
\]
Let $t^* \geq 0$ be the value that minimizes this expression. If the infimum is not attained, the argument proceeds by considering a sequence of $t$ approaching the infimum, but for the sake of exposition, we assume the minimum is attained at $t^*$. Thus, $C(\delta) = e^{-t^*\delta} \mathbb{E}[e^{t^*|X|}]$.

We expand the moment generating function using the Taylor series for the exponential function, $e^x = \sum_{k=0}^\infty \frac{x^k}{k!}$. By the linearity of expectation and the monotone convergence theorem (applicable since $|X| \geq 0$ and the sum consists of non-negative terms), we have:
\[
\mathbb{E}[e^{t^*|X|}] = \mathbb{E}\left[ \sum_{k=0}^\infty \frac{(t^*|X|)^k}{k!} \right] = \sum_{k=0}^\infty \frac{(t^*)^k}{k!} \mathbb{E}[|X|^k].
\]
Substituting this expansion into the expression for $C(\delta)$, we obtain:
\[
C(\delta) = e^{-t^*\delta} \sum_{k=0}^\infty \frac{(t^*)^k}{k!} \mathbb{E}[|X|^k].
\]
We now manipulate the terms to identify the $k$th-moment bound, denoted $M_k(\delta) = \frac{\mathbb{E}[|X|^k]}{\delta^k}$. We multiply and divide the general term of the summation by $\delta^k$:
\[
C(\delta) = \sum_{k=0}^\infty \left( e^{-t^*\delta} \frac{(t^*)^k \delta^k}{k!} \right) \frac{\mathbb{E}[|X|^k]}{\delta^k} = \sum_{k=0}^\infty \left( e^{-t^*\delta} \frac{(t^*\delta)^k}{k!} \right) M_k(\delta).
\]
Let $w_k = e^{-t^*\delta} \frac{(t^*\delta)^k}{k!}$. We observe that the sequence $\{w_k\}_{k=0}^\infty$ corresponds to the probability mass function of a Poisson random variable with parameter $\lambda = t^*\delta$. Consequently, $w_k \geq 0$ for all $k$, and $\sum_{k=0}^\infty w_k = 1$.

The equation $C(\delta) = \sum_{k=0}^\infty w_k M_k(\delta)$ expresses the Chernoff bound as a convex combination (a weighted average) of the $k$th-moment bounds $M_k(\delta)$. A weighted average of a set of values cannot be strictly smaller than the minimum value in that set. Therefore,
\[
C(\delta) \geq \inf_{k \in \mathbb{N}_0} M_k(\delta).
\]
This inequality implies that there must exist some $k$ such that $M_k(\delta) \leq C(\delta)$. Thus, for any $\delta$, there is a choice of $k$ for which the $k$th-moment bound is at least as strong as the Chernoff bound.

\textbf{Part (b):}
Although the $k$th-moment bound can be tighter for a specific, optimal $k$, the Chernoff bound is generally preferred in analysis for several reasons. First, the optimal integer $k$ depends heavily on $\delta$. As $\delta$ increases, the $k$ required to minimize $\delta^{-k}\mathbb{E}[|X|^k]$ typically increases. Utilizing the moment bound requires computing or bounding $\mathbb{E}[|X|^k]$ for arbitrary $k$, which is often algebraically cumbersome or requires knowledge of the entire sequence of moments.

Second, the moment generating function $\mathbb{E}[e^{t|X|}]$ often admits a compact, closed-form expression (e.g., for Gaussian, Poisson, or sums of independent random variables) that encapsulates the behavior of all moments simultaneously.

Third, the optimization problem in the Chernoff bound involves minimizing a continuous function of $t$, which is amenable to standard calculus techniques. This yields a unified bound with explicit exponential decay (e.g., of the form $e^{-c\delta^2}$) valid for all $\delta$, without the need to manually select a discrete parameter $k$ for each specific deviation $\delta$.
\end{proof}

\subsubsection{Claude}

\begin{proof}
We prove parts (a) and (b) separately.

\textbf{Part (a):} We show that for each $\delta > 0$, there exists a choice of $k$ such that the $k$-th moment bound is stronger than the Chernoff bound.

First, recall the Taylor expansion of the moment generating function. For any random variable $X$ with finite moments of all orders, we have
\[
\mathbb{E}[e^{t|X|}] = \sum_{k=0}^{\infty} \frac{t^k}{k!} \mathbb{E}[|X|^k].
\]

The Chernoff bound gives
\[
\Pr[|X| \geq \delta] \leq \min_{t \geq 0} \frac{\mathbb{E}[e^{t|X|}]}{e^{t\delta}} = \min_{t \geq 0} e^{-t\delta} \sum_{k=0}^{\infty} \frac{t^k}{k!} \mathbb{E}[|X|^k].
\]

For a fixed $\delta > 0$, consider the optimization over $t \geq 0$. Let us denote
\[
C(\delta) := \min_{t \geq 0} e^{-t\delta} \sum_{k=0}^{\infty} \frac{t^k}{k!} \mathbb{E}[|X|^k].
\]

Now, for any particular value of $k$, the $k$-th moment bound gives
\[
\Pr[|X| \geq \delta] \leq \frac{\mathbb{E}[|X|^k]}{\delta^k}.
\]

To show that the $k$-th moment bound can be stronger, we use the probabilistic method interpretation. Consider the expansion
\[
e^{-t\delta} \sum_{k=0}^{\infty} \frac{t^k}{k!} \mathbb{E}[|X|^k] = \sum_{k=0}^{\infty} \frac{(t\delta)^k}{k!} e^{-t\delta} \cdot \frac{\mathbb{E}[|X|^k]}{\delta^k}.
\]

For any fixed $t > 0$, this is a weighted average of the bounds $\frac{\mathbb{E}[|X|^k]}{\delta^k}$ where the weights are $w_k(t) = \frac{(t\delta)^k}{k!} e^{-t\delta}$, which form a probability distribution (the Poisson distribution with parameter $t\delta$).

Since the Chernoff bound is the minimum over all $t \geq 0$, and for each $t$ it represents a weighted average of the moment bounds, there must exist at least one value of $k$ in the support of this average such that
\[
\frac{\mathbb{E}[|X|^k]}{\delta^k} \leq C(\delta).
\]

To make this more explicit, fix $\delta > 0$ and let $t^* \geq 0$ be the value achieving the minimum in the Chernoff bound. Then
\[
C(\delta) = \sum_{k=0}^{\infty} w_k(t^*) \cdot \frac{\mathbb{E}[|X|^k]}{\delta^k},
\]
where $w_k(t^*) = \frac{(t^*\delta)^k}{k!} e^{-t^*\delta}$ and $\sum_{k=0}^{\infty} w_k(t^*) = 1$.

Since this is a convex combination, by the pigeonhole principle, there exists at least one $k \geq 0$ such that
\[
\frac{\mathbb{E}[|X|^k]}{\delta^k} \leq C(\delta).
\]

Therefore, for this choice of $k$, the $k$-th moment bound is at least as strong as the Chernoff bound. In typical cases where the moments grow appropriately, we can find $k$ such that the inequality is strict, making the moment bound strictly stronger.

\textbf{Part (b):} Despite the result in part (a), we prefer the Chernoff bound for several practical reasons.

First, the Chernoff bound provides a uniform method that does not require prior knowledge of which $k$ to choose. For a given $\delta$, finding the optimal $k$ that makes the moment bound strongest requires computing or estimating multiple moments, which may be computationally expensive or analytically intractable.

Second, the Chernoff bound often provides better decay rates as $\delta$ increases. For large $\delta$, the Chernoff bound typically yields exponential decay $e^{-c\delta}$ for some constant $c > 0$, while any fixed $k$-th moment bound gives only polynomial decay $\delta^{-k}$. While we showed that for each fixed $\delta$ there exists a good choice of $k$, this optimal $k$ grows with $\delta$, and in practice we cannot compute arbitrarily high moments.

Third, the moment generating function framework underlying the Chernoff bound is more amenable to analysis of sums of independent random variables, where the MGF of the sum is the product of individual MGFs. This multiplicative property makes the Chernoff bound particularly powerful for concentration inequalities.

Finally, computing high-order moments $\mathbb{E}[|X|^k]$ is often difficult or requires strong assumptions about the distribution of $X$, whereas moment generating functions can sometimes be computed or bounded more easily, especially for common distributions or through subadditivity arguments.

Thus, while the moment bound can theoretically be stronger for appropriately chosen $k$, the Chernoff bound is preferred for its uniformity, exponential decay, computational tractability, and structural properties.
\end{proof}

\subsection{Problem 5.2}\label{sec:problem_05_02}

\subsubsection{Problem Statement}

\begin{theorem}
For any instance of $T_{2,k}$, there is a set of $2^k = \sqrt{n}$ leaves whose values certify the value of the root for that instance.
\end{theorem}

\subsubsection{GPT} 

\begin{proof}
Fix $k\ge 0$. Let $T_{2,k}$ denote the complete binary formula tree of height $2k$ whose internal nodes alternate $\wedge$ and $\vee$, with the root labelled $\wedge$ and leaves labelled by input literals. Thus $T_{2,0}$ is a single leaf, and $T_{2,k}$ has $n=2^{2k}=4^k$ leaves. For an assignment $x$ to the leaves, write $T_{2,k}(x)\in\{0,1\}$ for the value of the root. A set $S$ of leaves together with their values under $x$ is a certificate for $T_{2,k}(x)$ if every assignment $y$ agreeing with $x$ on $S$ satisfies $T_{2,k}(y)=T_{2,k}(x)$.

We prove by induction on $k$ that for every $x$ there exists a certificate of size at most $2^k$ for $T_{2,k}(x)$. For $k=0$, the tree is a single leaf; the singleton set consisting of that leaf is a certificate, so the claim holds since $2^0=1$.

Assume the claim holds for $k-1\ge 0$ and consider $T_{2,k}$. Let the root be $r$, its two children be $u$ and $v$ (both $\vee$-gates), and the four grandchildren be $a,b$ under $u$ and $c,d$ under $v$ (all $\wedge$-gates). Each of $a,b,c,d$ is the root of a copy of $T_{2,k-1}$.

If $T_{2,k}(x)=1$, then $u(x)=v(x)=1$. Since $u$ is a $\vee$-gate with value $1$, at least one of its children, say $a$, has $a(x)=1$; similarly at least one of the children of $v$, say $c$, has $c(x)=1$. By the induction hypothesis, there exist certificates $S_a$ and $S_c$ for the values $a(x)=1$ and $c(x)=1$ in their respective subtrees with $|S_a|\le 2^{k-1}$ and $|S_c|\le 2^{k-1}$. Any assignment $y$ agreeing with $x$ on $S_a\cup S_c$ forces $a(y)=1$ and $c(y)=1$, hence $u(y)=1$ and $v(y)=1$, and therefore $T_{2,k}(y)=1$. Thus $S:=S_a\cup S_c$ is a certificate for $T_{2,k}(x)=1$ of size $|S|\le 2^{k-1}+2^{k-1}=2^k$.

If $T_{2,k}(x)=0$, then at least one of $u$ or $v$ has value $0$. Without loss of generality, $u(x)=0$. Since $u$ is a $\vee$-gate with value $0$, both its children satisfy $a(x)=0$ and $b(x)=0$. By the induction hypothesis, there exist certificates $S_a$ and $S_b$ within the subtrees rooted at $a$ and $b$ certifying $a(x)=0$ and $b(x)=0$, with $|S_a|\le 2^{k-1}$ and $|S_b|\le 2^{k-1}$. Any assignment $y$ agreeing with $x$ on $S_a\cup S_b$ forces $a(y)=0$ and $b(y)=0$, whence $u(y)=0$, and consequently the root evaluates to $T_{2,k}(y)=0$. Hence $S:=S_a\cup S_b$ is a certificate for $T_{2,k}(x)=0$ of size $|S|\le 2^k$.

In both cases there exists a certificate of size at most $2^k$. Since taking supersets preserves the certificate property, there also exists a certificate of size exactly $2^k$ for $x$. As $n=4^k$, we have $2^k=\sqrt{n}$. Therefore, for any instance of $T_{2,k}$, there is a set of $2^k=\sqrt{n}$ leaves whose values certify the value of the root. This completes the induction and the proof.
\end{proof}

\subsubsection{Grok} 

\begin{proof}
We proceed by induction on $k$. For the base case $k=0$, $T_{2,0}$ consists of a single leaf (so $n=1$ and $\sqrt{n}=1$), and this leaf certifies the value of the root.

Now assume the statement holds for $k-1$, i.e., for any instance of $T_{2,k-1}$ (with $m=4^{k-1}$ leaves), there exists a set of at most $2^{k-1}=\sqrt{m}$ leaves certifying the root value. Consider an arbitrary instance of $T_{2,k}$, which has root gate $g$ with left child $\ell$ and right child $r$, where $\ell$ and $r$ are roots of instances of $T_{2,k-1}$ (so $n=4m$ and $\sqrt{n}=2^k$). Let $v_\ell$ and $v_r$ denote the root values of $\ell$ and $r$ under the given instance, and let $v_g \in \{0,1\}$ be the root value of $g$.

Without loss of generality, assume $g$ is an $\mathsf{OR}$ gate (the $\mathsf{AND}$ case is symmetric). If $v_g=1$, then $v_\ell=1$ or $v_r=1$ (or both). By the induction hypothesis, there exists a set $S_\ell$ of at most $2^{k-1}$ leaves in $\ell$ certifying $v_\ell=1$, and similarly $S_r$ for $v_r=1$. Select the subtree (say $\ell$) with $v_\ell=1$; then $S=S_\ell$ certifies $v_g=1$ (since $v_g \geq v_\ell=1$, and the value on $S_\ell$ forces $v_\ell=1$ independently of the other leaves in $\ell$, hence independently of all leaves outside $S$). Thus, $|S| \leq 2^{k-1} \leq 2^k$. If instead $v_g=0$, then $v_\ell=v_r=0$. By the induction hypothesis, there exist sets $S_\ell$ and $S_r$ each of size at most $2^{k-1}$ certifying these values; then $S=S_\ell \cup S_r$ certifies $v_g=0$ (since it forces $v_\ell=v_r=0$, hence $v_g=0$, independently of the remaining leaves), and $|S| \leq 2^k$.

For an $\mathsf{AND}$ root gate, if $v_g=0$ then (say) $v_\ell=0$, and we take $S=S_\ell$ with $|S| \leq 2^{k-1} \leq 2^k$ as above; if $v_g=1$ then $v_\ell=v_r=1$, and we take $S=S_\ell \cup S_r$ with $|S| \leq 2^k$. In all cases, the desired set exists.
\end{proof}

\subsubsection{Gemini}

\begin{proof}
Let $T_{2,k}$ denote a complete binary tree of height $2k$, where the levels alternate between OR and AND gates. Without loss of generality, assume the root is an OR gate. The total number of leaves in $T_{2,k}$ is $n = 2^{2k} = 4^k$. We proceed by induction on $k$ to show that for any assignment of values to the leaves, there exists a set of leaves of size $2^k = \sqrt{n}$ whose values suffice to determine the value of the root.

For the base case $k=1$, the tree has height $2$ and $n=4$ leaves. The root is an OR gate with two children, say $u$ and $v$, which are AND gates. Let the leaves of $u$ be $L_u = \{l_{u1}, l_{u2}\}$ and the leaves of $v$ be $L_v = \{l_{v1}, l_{v2}\}$. If the root evaluates to $1$, then at least one child, say $u$, evaluates to $1$. Since $u$ is an AND gate, both $l_{u1}$ and $l_{u2}$ must be $1$. The set $\{l_{u1}, l_{u2}\}$ has size $2 = 2^1$ and certifies the root is $1$. If the root evaluates to $0$, then both $u$ and $v$ must evaluate to $0$. Since $u$ is an AND gate evaluating to $0$, at least one leaf in $L_u$ is $0$. Similarly, at least one leaf in $L_v$ is $0$. The set consisting of these two leaves (one from $L_u$ and one from $L_v$) has size $2$ and certifies the root is $0$. Thus, the statement holds for $k=1$.

Assume the hypothesis holds for $k-1$. Consider an instance of $T_{2,k}$. The root $r$ is an OR gate with two children $c_1$ and $c_2$, which are AND gates. Each $c_i$ has two children $g_{i,1}$ and $g_{i,2}$. Each node $g_{i,j}$ is the root of a subtree isomorphic to $T_{2, k-1}$. By the inductive hypothesis, for each subtree rooted at $g_{i,j}$, there exists a certificate set of leaves $S_{i,j}$ of size $2^{k-1}$ that determines the value of $g_{i,j}$.

We consider two cases based on the value of the root $r$. First, assume $r$ evaluates to $1$. Since $r$ is an OR gate, at least one child, say $c_1$, evaluates to $1$. Since $c_1$ is an AND gate, both of its children $g_{1,1}$ and $g_{1,2}$ must evaluate to $1$. By the inductive hypothesis, there exist certificates $S_{1,1}$ and $S_{1,2}$, each of size $2^{k-1}$, certifying that $g_{1,1}=1$ and $g_{1,2}=1$ respectively. The union $S = S_{1,1} \cup S_{1,2}$ certifies that $c_1=1$, which in turn certifies $r=1$. The size of this set is $|S| = |S_{1,1}| + |S_{1,2}| = 2^{k-1} + 2^{k-1} = 2^k$.

Second, assume $r$ evaluates to $0$. Since $r$ is an OR gate, both children $c_1$ and $c_2$ must evaluate to $0$. Because $c_1$ is an AND gate evaluating to $0$, at least one of its children, say $g_{1,x}$ where $x \in \{1,2\}$, evaluates to $0$. Similarly, for $c_2$, at least one child $g_{2,y}$ where $y \in \{1,2\}$ evaluates to $0$. By the inductive hypothesis, there exists a certificate $S_{1,x}$ of size $2^{k-1}$ proving $g_{1,x}=0$ and a certificate $S_{2,y}$ of size $2^{k-1}$ proving $g_{2,y}=0$. The union $S = S_{1,x} \cup S_{2,y}$ certifies that both $c_1=0$ and $c_2=0$, which implies $r=0$. The size of this set is $|S| = |S_{1,x}| + |S_{2,y}| = 2^{k-1} + 2^{k-1} = 2^k$.

In both cases, we have constructed a certificate of size $2^k$. Since $n = 4^k$, we have $2^k = \sqrt{4^k} = \sqrt{n}$. This completes the proof.
\end{proof}

\subsubsection{Claude}

\begin{proof}
We prove that for any instance of $T_{2,k}$, there exists a set of $2^k = \sqrt{n}$ leaves whose values certify the value of the root.

Recall that $T_{2,k}$ is a complete binary tree of depth $2k$ with $n = 2^{2k}$ leaves. Thus $\sqrt{n} = 2^k$ as claimed. The tree has alternating levels of AND and OR gates (or vice versa, depending on the root type).

Let $r$ denote the root of $T_{2,k}$ and let $v(r)$ denote its value in the given instance. We distinguish two cases based on the value of the root.

\textbf{Case 1: $v(r) = 1$.}

Suppose without loss of generality that the root is an OR gate. Then at least one of the two children of the root must have value $1$. Let $c_1$ denote such a child. Since $c_1$ is at depth $1$ and has value $1$, and the tree alternates between OR and AND gates, $c_1$ is an AND gate.

For $c_1$ to have value $1$, both of its children must have value $1$. Let $c_2^{(1)}$ and $c_2^{(2)}$ denote these two children at depth $2$. These are OR gates with value $1$.

Continuing this reasoning inductively, at each level $i$ where $1 \leq i \leq k-1$, we select nodes with value $1$. When we encounter an OR gate with value $1$, we choose one child with value $1$. When we encounter an AND gate with value $1$, we must choose both children with value $1$. 

At depth $k$, the middle level of the tree, we reach nodes that are the roots of subtrees of depth $k$. By this construction, we have identified $2^{k-1}$ nodes at depth $k$, each with value $1$. Since these are OR gates (assuming the root is OR), for each such node, we select one child path leading to a leaf with value $1$. 

More precisely, each node at depth $k$ is the root of a complete binary tree of depth $k$. For each such node with value $1$, there exists at least one leaf in its subtree with value $1$ (by the semantics of OR and AND gates). We select one such leaf from each of the $2^{k-1}$ identified nodes at depth $k$.

However, we need to be more careful. At depth $k$, we have identified nodes whose combined subtrees contain $2^{k-1} \cdot 2^k = 2^{2k-1}$ leaves. We need to select $2^k$ leaves total.

Let us reconsider the structure more carefully. By the alternating structure and the fact that the root has value $1$, we can identify a witnessing path tree. Specifically, we construct a subtree $S$ of $T_{2,k}$ as follows: starting from the root, at each OR gate with value $1$, we include at least one child with value $1$; at each AND gate with value $1$, we include all children (which must both have value $1$). 

The key observation is that we can choose witnesses strategically. At depth $k$, we will have identified a collection of nodes. Since the tree has depth $2k$ and we are at depth $k$, each such node is the root of a subtree isomorphic to $T_{2,k-1}$ or a complete binary tree of depth $k$.

By induction on $k$, for a tree of depth $2k$ with root value $1$, we can select a set of $2^k$ leaves: we select witnesses by choosing one path through OR gates and all paths through AND gates until we reach depth $k$, obtaining $2^{k-1}$ nodes, and then recursively selecting $2$ witnesses from each of their subtrees, giving $2^{k-1} \cdot 2 = 2^k$ leaves in total.

\textbf{Case 2: $v(r) = 0$.}

Suppose the root is an OR gate with value $0$. Then both children must have value $0$. These are AND gates. For each AND gate with value $0$, at least one child has value $0$. We select one such child from each.

By the dual argument, at each level where we have OR gates with value $0$, both children must be $0$, so we select both. At AND gates with value $0$, at least one child is $0$, so we select one.

Proceeding to depth $k$, we identify $2^{k-1}$ nodes with value $0$. From each of their subtrees (of depth $k$), we select $2$ leaves whose values certify that the subtree root has value $0$. This gives us $2^{k-1} \cdot 2 = 2^k$ leaves.

In both cases, we have constructed a set of exactly $2^k = \sqrt{n}$ leaves whose values, together with the gate structure of $T_{2,k}$, certify the value $v(r)$ of the root.
\end{proof}

\subsection{Problem 5.3}\label{sec:problem_05_03}

\subsubsection{Problem Statement}

\begin{theorem}
Let $G$ be a graph on $n$ vertices with $nd/2$ edges. Consider the following probabilistic experiment for finding an independent set in $G$: Delete each vertex of $G$ (together with its incident edges) independently with probability $1 - 1/d$.

\begin{enumerate}
\item[(a)] The expected number of vertices that remain after the deletion process is $n/d$. The expected number of edges that remain after the deletion process is $n/(2d)$.

\item[(b)] From these, it follows that there exists an independent set with at least $n/(2d)$ vertices in any graph on $n$ vertices with $nd/2$ edges.

\item[(c)] Let $G$ be a $3$-regular graph. Suppose that we wish to turn this probabilistic experiment into a randomized algorithm as follows. We delete each vertex independently with probability $2/3$. For every edge that remains, delete one of its end-points. Then for any $\epsilon > 0$, the probability that this algorithm finds an independent set smaller than $n(1-\epsilon)/6$ can be bounded from above.
\end{enumerate}
\end{theorem}

\subsubsection{GPT} 

\begin{proof}
Let $G=(V,E)$ be a graph on $|V|=n$ vertices and $|E|=nd/2$ edges. In the deletion experiment, each vertex is kept independently with probability $1/d$.

For part (a), define for each $v\in V$ the indicator $X_v=\mathbf{1}_{\{\text{$v$ survives}\}}$, so $\mathbb{P}(X_v=1)=1/d$ and $X=\sum_{v\in V}X_v$ counts surviving vertices. By linearity of expectation, $\mathbb{E}[X]=\sum_{v\in V}\mathbb{E}[X_v]=n/d$. For each edge $e=\{u,v\}\in E$, let $Y_e=\mathbf{1}_{\{\text{$e$ survives}\}}$, where survival means both endpoints survive. Since the vertex deletions are independent, $\mathbb{P}(Y_e=1)=\mathbb{P}(X_u=1)\mathbb{P}(X_v=1)=(1/d)^2$. Let $Y=\sum_{e\in E}Y_e$ be the number of surviving edges. Then $\mathbb{E}[Y]=\sum_{e\in E}\mathbb{E}[Y_e]=|E|\cdot(1/d)^2=(nd/2)\cdot(1/d^2)=n/(2d^2)$, establishing the claimed expectations.

For part (b), after the first deletion step let $H$ be the surviving subgraph with $V(H)$ and $E(H)$ the random sets of surviving vertices and edges. From $H$ one can obtain an independent set by deleting a set of vertices that hits every remaining edge. In particular, selecting one endpoint from each edge of $E(H)$ gives a vertex cover of size at most $|E(H)|$, whence $H$ contains an independent set of size at least $|V(H)|-|E(H)|$. Taking expectations and using part (a) yields
\[
\mathbb{E}\big[\,|V(H)|-|E(H)|\,\big]=\mathbb{E}[|V(H)|]-\mathbb{E}[|E(H)|]=\frac{n}{d}-\frac{n}{2d^2}\ge \frac{n}{2d},
\]
since $d\ge 1$. Therefore there exists an outcome of the experiment, and hence an independent set in $G$, of size at least $n/(2d)$.

For part (c), suppose $G$ is $3$-regular, so $|E|=3n/2$. Run the algorithm that keeps each vertex independently with probability $1/3$ (equivalently, deletes with probability $2/3$), then deletes one endpoint from every surviving edge. Let $\{\xi_v\}_{v\in V}$ be the independent Bernoulli indicators with $\mathbb{P}(\xi_v=1)=1/3$ for survival in the first step. Set $V_1=\sum_{v\in V}\xi_v$ and $E_1=\sum_{\{u,v\}\in E}\xi_u\xi_v$, the numbers of surviving vertices and edges after the first step. As argued in part (b), deleting one endpoint from each surviving edge produces an independent set of size at least $V_1-E_1$. Hence if $S$ denotes the algorithm’s output size, then $S\ge V_1-E_1$ always.

We compute $\mu:=\mathbb{E}[V_1-E_1]$. Since $\mathbb{E}[V_1]=n\cdot(1/3)=n/3$ and, by independence across vertices,
\[
\mathbb{E}[E_1]=\sum_{\{u,v\}\in E}\mathbb{E}[\xi_u\xi_v]=|E|\cdot(1/3)^2=\frac{3n}{2}\cdot\frac{1}{9}=\frac{n}{6},
\]
it follows that $\mu=\mathbb{E}[V_1-E_1]=n/3-n/6=n/6$.

To bound deviations, consider the function $g(\xi):=V_1(\xi)-E_1(\xi)$ of the independent variables $\xi=(\xi_v)_{v\in V}$. If we resample a single coordinate $\xi_w$, then $V_1$ changes by at most $1$, and $E_1$ changes by at most $\deg_G(w)=3$ (only the edges incident to $w$ are affected). Therefore $g$ has bounded differences with constants $c_v\le 4$ for all $v$, i.e., changing any one $\xi_v$ alters $g$ by at most $4$. By McDiarmid’s inequality, for all $t>0$,
\[
\mathbb{P}\big(g(\xi)\le \mathbb{E}[g(\xi)]-t\big)\le \exp\!\Big(-\frac{2t^2}{\sum_{v\in V}c_v^2}\Big)\le \exp\!\Big(-\frac{2t^2}{16n}\Big)=\exp\!\Big(-\frac{t^2}{8n}\Big).
\]
Since $S\ge g(\xi)$, for any $\epsilon>0$ we obtain
\[
\mathbb{P}\Big(S<\frac{n(1-\epsilon)}{6}\Big)\le \mathbb{P}\Big(g(\xi)<\frac{n(1-\epsilon)}{6}\Big)=\mathbb{P}\Big(g(\xi)\le \mu-\frac{\epsilon n}{6}\Big)\le \exp\!\Big(-\frac{(\epsilon n/6)^2}{8n}\Big)=\exp\!\Big(-\frac{\epsilon^2 n}{288}\Big).
\]
This furnishes the required upper bound on the failure probability in terms of $n$ and $\epsilon$, completing the proof.
\end{proof}

\subsubsection{Grok} 

\begin{proof}
We proceed in three parts.

(a) Let $V$ denote the vertex set of $G$ with $|V|=n$, and let $E$ denote the edge set with $|E|=nd/2$. Each vertex $v\in V$ is deleted independently with probability $1-1/d$, and thus retained with probability $p=1/d$. Let $V'\subseteq V$ be the (random) set of retained vertices, and let $I_v$ be the indicator random variable that is $1$ if $v\in V'$ and $0$ otherwise. Then $\mathbb{E}[I_v]=p$ for each $v$, so by linearity of expectation,
\[
\mathbb{E}[|V'|]=\sum_{v\in V}\mathbb{E}[I_v]=np=n/d.
\]
Now let $E'\subseteq E$ be the (random) set of edges with both endpoints in $V'$. For each edge $e=\{u,v\}\in E$, let $J_e$ be the indicator that $e\in E'$, so $J_e=I_u I_v$. Then $\mathbb{E}[J_e]=p^2=1/d^2$, and thus
\[
\mathbb{E}[|E'|]=\sum_{e\in E}\mathbb{E}[J_e]=|E|\cdot p^2=(nd/2)\cdot(1/d^2)=n/(2d).
\]

(b) Consider a realization of the random induced subgraph $G[V']$ on vertex set $V'$. Let $S\subseteq V'$ be an independent set in $G[V']$ obtained by, for each edge in $E'$, deleting one of its endpoints from $V'$ (arbitrarily); then $|S|\geq|V'|-|E'|$, and $S$ is an independent set in $G$. Let $Z=|V'|-|E'|$. By linearity,
\[
\mathbb{E}[Z]=\mathbb{E}[|V'|]-\mathbb{E}[|E'|]=n/d-n/(2d)=n/(2d).
\]
Thus, there exists a realization with $Z\geq n/(2d)$, yielding an independent set in $G$ of size at least $n/(2d)$.

(c) Now let $G$ be $3$-regular, so $d=3$ and $|E|=3n/2$. We retain each vertex with probability $p=1/3$ (deleting with probability $2/3$). Let $X=|V'|$ and $Y=|E'|$, as before. Then $\mathbb{E}[X]=n/3$ and $\mathbb{E}[Y]=(3n/2)\cdot(1/9)=n/6$, so $\mathbb{E}[X-Y]=n/6$. For each remaining edge, we delete one endpoint (arbitrarily), yielding an independent set $S$ with $|S|\geq X-Y$. Let $\mu=n/6$. For any $\epsilon>0$, it suffices to bound $\Pr(X-Y<(1-\epsilon)\mu)$ from above.

To do so, we compute $\mathrm{Var}(X-Y)$. First, $X=\sum_{v\in V}I_v$ where $I_v$ is the retention indicator for $v$, so
\[
\mathrm{Var}(X)=np(1-p)=n\cdot(1/3)\cdot(2/3)=2n/9.
\]
For $Y=\sum_{e\in E}J_e$ with $J_e=I_u I_v$ for $e=\{u,v\}$, we have $\mathrm{Var}(J_e)=p^2(1-p^2)=(1/9)(8/9)=8/81$. Thus,
\[
\sum_{e\in E}\mathrm{Var}(J_e)=(3n/2)\cdot(8/81)=4n/27.
\]
For the covariances, $\mathrm{Cov}(J_e,J_f)=0$ if $e$ and $f$ are disjoint. If $e$ and $f$ share a vertex (say $u$, with $e=\{u,v\}$, $f=\{u,w\}$ and $v\neq w$), then $\mathrm{Cov}(J_e,J_f)=p^3(1-p)=(1/27)(2/3)=2/81$. There are $3n$ such pairs $\{e,f\}$ (since each of the $n$ vertices contributes $\binom{3}{2}=3$ pairs of incident edges). Thus,
\[
2\sum_{e<f}\mathrm{Cov}(J_e,J_f)=2\cdot(3n)\cdot(2/81)=4n/27,
\]
and $\mathrm{Var}(Y)=4n/27+4n/27=8n/27$.

Next, $\mathrm{Cov}(X,Y)=\sum_{v\in V}\sum_{e\ni v}\mathrm{Cov}(I_v,J_e)=\sum_{v\in V}\deg(v)p^2(1-p)=2|E|p^2(1-p)=3n\cdot(1/9)\cdot(2/3)=2n/9$. Therefore,
\[
\mathrm{Var}(X-Y)=\mathrm{Var}(X)+\mathrm{Var}(Y)-2\mathrm{Cov}(X,Y)=2n/9+8n/27-4n/9=(6n+8n-12n)/27=2n/27.
\]
By Chebyshev's inequality,
\[
\Pr(X-Y<(1-\epsilon)\mu)=\Pr(\mu-(X-Y)>\epsilon\mu)\leq\frac{\mathrm{Var}(X-Y)}{(\epsilon\mu)^2}=\frac{2n/27}{\epsilon^2(n/6)^2}=\frac{2n/27}{\epsilon^2 n^2/36}=\frac{72}{27\epsilon^2 n}=\frac{8}{3\epsilon^2 n}.
\]
Thus, the probability is at most $8/(3\epsilon^2 n)$, which is bounded from above for any $\epsilon>0$.
\end{proof}

\subsubsection{Gemini}

\begin{proof}
\textbf{(a)} Let $V$ be the set of vertices of $G$ with $|V|=n$, and let $E$ be the set of edges with $|E| = nd/2$. Let $0 < p \le 1$ be the probability that a vertex is retained. In the problem statement, $p = 1/d$. Let $V'$ be the random set of vertices remaining after the deletion process, and let $E'$ be the set of edges remaining. An edge $e = \{u, v\} \in E$ remains in $E'$ if and only if both $u \in V'$ and $v \in V'$.

For each vertex $v \in V$, let $X_v$ be the indicator random variable such that $X_v = 1$ if $v \in V'$ and $X_v = 0$ otherwise. Since each vertex is retained independently with probability $1/d$, we have $\mathbb{E}[X_v] = 1/d$. By the linearity of expectation, the expected number of remaining vertices is:
\[
\mathbb{E}[|V'|] = \mathbb{E}\left[\sum_{v \in V} X_v\right] = \sum_{v \in V} \mathbb{E}[X_v] = n \cdot \frac{1}{d} = \frac{n}{d}.
\]
For each edge $e \in E$, let $Y_e$ be the indicator random variable such that $Y_e = 1$ if $e \in E'$ and $Y_e = 0$ otherwise. For an edge $e=\{u,v\}$, the event $e \in E'$ occurs if and only if $X_u = 1$ and $X_v = 1$. Since vertex deletions are independent,
\[
\mathbb{E}[Y_e] = \mathbb{P}(u \in V' \land v \in V') = \mathbb{P}(u \in V')\mathbb{P}(v \in V') = \frac{1}{d} \cdot \frac{1}{d} = \frac{1}{d^2}.
\]
By the linearity of expectation, the expected number of remaining edges is:
\[
\mathbb{E}[|E'|] = \mathbb{E}\left[\sum_{e \in E} Y_e\right] = \sum_{e \in E} \mathbb{E}[Y_e] = |E| \cdot \frac{1}{d^2} = \frac{nd}{2} \cdot \frac{1}{d^2} = \frac{n}{2d}.
\]

\textbf{(b)} We construct an independent set $S$ from the remaining graph $G'=(V', E')$ as follows: for each edge remaining in $E'$, remove one arbitrary endpoint from $V'$. Let $S$ be the set of vertices remaining after this second deletion step. By construction, $S$ contains no edges, so $S$ is an independent set. The size of this set satisfies $|S| \ge |V'| - |E'|$.

Let $Z = |V'| - |E'|$ be a random variable representing a lower bound on the size of the independent set. Using the linearity of expectation and the results from part (a):
\[
\mathbb{E}[Z] = \mathbb{E}[|V'|] - \mathbb{E}[|E'|] = \frac{n}{d} - \frac{n}{2d} = \frac{n}{2d}.
\]
Since the expected value of the size of the independent set constructed by this process is at least $n/(2d)$, there must exist at least one outcome of the probabilistic experiment where the size of the independent set is at least $n/(2d)$. Therefore, every graph on $n$ vertices with $nd/2$ edges contains an independent set of size at least $n/(2d)$.

\textbf{(c)} Let $G$ be a $3$-regular graph. Here $d=3$. The retention probability is $p = 1/3$ (since the deletion probability is $2/3$). Let $Z$ be the random variable defined in part (b), representing the size of the independent set found by the algorithm. We have:
\[
\mathbb{E}[Z] = \frac{n}{3} - \frac{n}{6} = \frac{n}{6}.
\]
We wish to bound the probability $\mathbb{P}(Z < (1-\epsilon)\frac{n}{6})$. We apply Chebyshev's inequality, which requires an upper bound on the variance of $Z$. We write $Z$ as a sum of indicator variables:
\[
Z = \sum_{v \in V} X_v - \sum_{e \in E} Y_e.
\]
The variance is given by:
\[
\text{Var}(Z) = \sum_{i} \text{Var}(A_i) + \sum_{i \neq j} \text{Cov}(A_i, A_j),
\]
where the terms $A_i$ range over the variables $\{X_v\}_{v \in V}$ and $\{-Y_e\}_{e \in E}$. Note that $X_v$ and $X_u$ are independent for $u \neq v$, so their covariance is zero. Similarly, $Y_e$ and $Y_f$ are independent if the edges $e$ and $f$ do not share a vertex. Furthermore, $X_v$ and $Y_e$ are independent if $v$ is not an endpoint of $e$.

Since $G$ is $3$-regular, the dependency graph of these random variables has bounded degree. Specifically, each $X_v$ is dependent only on $Y_e$ where $v \in e$ (3 edges). Each $Y_e$ (where $e=\{u,v\}$) depends on $X_u, X_v$ and $Y_f$ where $f$ is incident to $u$ or $v$. The number of such incident edges is bounded by a constant (at most $2(d-1) = 4$ other edges). Consequently, the number of non-zero covariance terms is bounded by $C n$ for some constant $C$ depending only on $d$. Since $|A_i| \le 1$, all covariance terms are bounded by constants. Thus, $\text{Var}(Z) = O(n)$. Let $\text{Var}(Z) \le k n$ for some constant $k$.

Applying Chebyshev's inequality for the random variable $Z$:
\[
\mathbb{P}\left(Z < (1-\epsilon)\mathbb{E}[Z]\right) \le \mathbb{P}\left(|Z - \mathbb{E}[Z]| \ge \epsilon \mathbb{E}[Z]\right) \le \frac{\text{Var}(Z)}{(\epsilon \mathbb{E}[Z])^2}.
\]
Substituting the expectation $\mathbb{E}[Z] = n/6$ and the variance bound:
\[
\mathbb{P}\left(Z < (1-\epsilon)\frac{n}{6}\right) \le \frac{kn}{\epsilon^2 (n/6)^2} = \frac{36k}{\epsilon^2 n}.
\]
This probability is bounded from above by $O(1/n)$, which approaches $0$ as $n \to \infty$. Thus, for any $\epsilon > 0$, the probability that the algorithm finds an independent set smaller than $n(1-\epsilon)/6$ can be bounded by a term inversely proportional to $n$.
\end{proof}

\subsubsection{Claude}

\begin{proof}
We prove each part in turn.

\textbf{Part (a):} Let $G = (V, E)$ be a graph on $n$ vertices with $|E| = nd/2$ edges. In the probabilistic experiment, each vertex $v \in V$ is deleted independently with probability $1 - 1/d$, and thus remains with probability $1/d$.

For the expected number of remaining vertices, let $X_v$ be the indicator random variable that vertex $v$ remains after deletion. Then $X_v = 1$ with probability $1/d$ and $X_v = 0$ with probability $1 - 1/d$. The total number of remaining vertices is $X = \sum_{v \in V} X_v$. By linearity of expectation,
\[
\mathbb{E}[X] = \sum_{v \in V} \mathbb{E}[X_v] = \sum_{v \in V} \frac{1}{d} = \frac{n}{d}.
\]

For the expected number of remaining edges, let $Y_e$ be the indicator random variable that edge $e = \{u, v\} \in E$ remains after deletion. An edge remains if and only if both of its endpoints remain. Since vertices are deleted independently, we have
\[
\mathbb{E}[Y_e] = \mathbb{P}(\text{both } u \text{ and } v \text{ remain}) = \frac{1}{d} \cdot \frac{1}{d} = \frac{1}{d^2}.
\]
The total number of remaining edges is $Y = \sum_{e \in E} Y_e$. By linearity of expectation,
\[
\mathbb{E}[Y] = \sum_{e \in E} \mathbb{E}[Y_e] = \sum_{e \in E} \frac{1}{d^2} = |E| \cdot \frac{1}{d^2} = \frac{nd}{2} \cdot \frac{1}{d^2} = \frac{n}{2d}.
\]

\textbf{Part (b):} Consider the randomized algorithm from part (a). Let $S$ be the set of vertices that remain after the deletion process, and let $E'$ be the set of edges among vertices in $S$. From part (a), we have $\mathbb{E}[|S|] = n/d$ and $\mathbb{E}[|E'|] = n/(2d)$.

Now, from the remaining graph induced by $S$, we can obtain an independent set by deleting one endpoint from each remaining edge. This gives us an independent set of size at least $|S| - |E'|$.

The expected size of this independent set is
\[
\mathbb{E}[|S| - |E'|] = \mathbb{E}[|S|] - \mathbb{E}[|E'|] = \frac{n}{d} - \frac{n}{2d} = \frac{n}{2d}.
\]

Since the expected size of the independent set obtained by this random process is $n/(2d)$, there must exist a particular outcome of the random process (and hence a particular subset of vertices) that yields an independent set of size at least $n/(2d)$. Therefore, there exists an independent set with at least $n/(2d)$ vertices in any graph on $n$ vertices with $nd/2$ edges.

\textbf{Part (c):} Let $G$ be a $3$-regular graph on $n$ vertices. Since $G$ is $3$-regular, it has $3n/2$ edges. In the randomized algorithm, each vertex is deleted independently with probability $2/3$, so each vertex remains with probability $1/3$.

Let $S$ be the set of remaining vertices after the first phase. Then $\mathbb{E}[|S|] = n/3$.

For the edges, an edge remains after the first phase if both endpoints remain. The probability that an edge remains is $(1/3)^2 = 1/9$. Let $E'$ be the set of remaining edges. Then
\[
\mathbb{E}[|E'|] = \frac{3n}{2} \cdot \frac{1}{9} = \frac{n}{6}.
\]

In the second phase, we delete one endpoint from each remaining edge, obtaining an independent set of size $|S| - |E'|$. The expected size of this independent set is
\[
\mathbb{E}[|S| - |E'|] = \frac{n}{3} - \frac{n}{6} = \frac{n}{6}.
\]

For any $\epsilon > 0$, by Markov's inequality applied to the non-negative random variable $|E'|$, we have
\[
\mathbb{P}(|E'| \geq t) \leq \frac{\mathbb{E}[|E'|]}{t} = \frac{n/6}{t}.
\]

The size of the independent set is $|S| - |E'|$. We want to bound the probability that this is smaller than $n(1-\epsilon)/6$. This occurs when $|S| - |E'| < n(1-\epsilon)/6$, which is equivalent to $|E'| > |S| - n(1-\epsilon)/6$.

By Markov's inequality on $|E'|$ with $t = n\epsilon/6$, we get
\[
\mathbb{P}\left(|E'| \geq \frac{n\epsilon}{6}\right) \leq \frac{n/6}{n\epsilon/6} = \frac{1}{\epsilon}.
\]

When $|E'| < n\epsilon/6$ and $|S| \geq n/3 - n\epsilon/12$ (which occurs with high probability for small $\epsilon$), the independent set has size at least $n/3 - n\epsilon/12 - n\epsilon/6 > n(1-\epsilon)/6$ for sufficiently small $\epsilon$. Thus, the probability that the algorithm finds an independent set smaller than $n(1-\epsilon)/6$ can be bounded from above by expressions involving $1/\epsilon$ and concentration inequalities.
\end{proof}

\subsection{Problem 5.4}\label{sec:problem_05_04}

\subsubsection{Problem Statement}

\begin{definition}
A function $f: \mathbb{R} \to \mathbb{R}$ is said to be \emph{concave} if for any $x_1, x_2$ and $0 \leq \lambda \leq 1$, the following inequality is satisfied:
\[
f(\lambda x_1 + (1-\lambda)x_2) \geq \lambda f(x_1) + (1-\lambda)f(x_2).
\]
\end{definition}

\begin{theorem}[Part (a)]
Suppose that $f$ is a concave function and $g$ is a linear function such that $g(0) \leq f(0)$ and $g(1) \leq f(1)$. Then for any $x$ in the interval $[0,1]$, $g(x) \leq f(x)$.
\end{theorem}

\begin{theorem}[Part (b)]
The function $f(x) = 1 - (1-x/k)^k$ is concave for any $k > 0$.
\end{theorem}

\begin{theorem}[Part (c)]
Let $f(x) = 1 - (1-x/k)^k$ and $g(x) = (1-(1-1/k)^k)x$. Then $f(x) \geq g(x)$ for positive $k$ and $0 \leq x \leq 1$.
\end{theorem}

\subsubsection{GPT} 

\begin{proof}
\textbf{Part (a):}

We are given that $f$ is concave and $g$ is a linear function such that $g(0) \leq f(0)$ and $g(1) \leq f(1)$. We are tasked with proving that for any $x \in [0, 1]$, $g(x) \leq f(x)$.

Since $g$ is linear, we can express it as $g(x) = ax + b$ for some constants $a$ and $b$. Given the conditions $g(0) \leq f(0)$ and $g(1) \leq f(1)$, we have:
\[
g(0) = b \quad \text{and} \quad g(1) = a + b.
\]
Thus, the inequalities become:
\[
b \leq f(0) \quad \text{and} \quad a + b \leq f(1).
\]
Now, consider any $x \in [0, 1]$. We will apply the definition of concavity to the function $f$. For the concave function $f$ and the linear function $g$, we have:
\[
f(\lambda x_1 + (1-\lambda) x_2) \geq \lambda f(x_1) + (1-\lambda) f(x_2)
\]
for any $x_1, x_2 \in \mathbb{R}$ and $0 \leq \lambda \leq 1$. Setting $x_1 = 0$ and $x_2 = 1$, and taking $\lambda = x$ (since $x \in [0, 1]$), we get:
\[
f(x) \geq x f(1) + (1 - x) f(0).
\]
Now, observe that:
\[
g(x) = ax + b = x(a + b) + b.
\]
Using the inequalities $b \leq f(0)$ and $a + b \leq f(1)$, we have:
\[
g(x) = x(a + b) + b \leq x f(1) + (1 - x) f(0) = f(x).
\]
Thus, $g(x) \leq f(x)$ for all $x \in [0, 1]$.

\textbf{Part (b):}

We are tasked with proving that the function $f(x) = 1 - (1 - x/k)^k$ is concave for any $k > 0$.

To prove concavity, we need to show that for any $x_1, x_2 \in [0, 1]$ and $0 \leq \lambda \leq 1$, the following inequality holds:
\[
f(\lambda x_1 + (1 - \lambda) x_2) \geq \lambda f(x_1) + (1 - \lambda) f(x_2).
\]
Let us begin by analyzing the second derivative of $f(x)$. The function is given by:
\[
f(x) = 1 - \left( 1 - \frac{x}{k} \right)^k.
\]
First, compute the first derivative:
\[
f'(x) = k \left( 1 - \frac{x}{k} \right)^{k-1} \cdot \frac{1}{k} = \left( 1 - \frac{x}{k} \right)^{k-1}.
\]
Next, compute the second derivative:
\[
f''(x) = (k - 1) \left( 1 - \frac{x}{k} \right)^{k-2} \cdot \left( -\frac{1}{k} \right).
\]
Thus,
\[
f''(x) = -\frac{k - 1}{k} \left( 1 - \frac{x}{k} \right)^{k-2}.
\]
Since $0 \leq x \leq 1$, we have $0 \leq 1 - \frac{x}{k} \leq 1$, which implies that $f''(x) \leq 0$. Therefore, $f(x)$ is concave for all $x \in [0, 1]$ and for any $k > 0$.

\textbf{Part (c):}

We are given that $f(x) = 1 - (1 - x/k)^k$ and $g(x) = (1 - (1 - 1/k)^k) x$, and we are tasked with proving that $f(x) \geq g(x)$ for $0 \leq x \leq 1$ and for any positive $k$.

First, observe that $g(x)$ is a linear function of the form:
\[
g(x) = Cx,
\]
where $C = 1 - (1 - 1/k)^k$ is a constant. Thus, we need to prove that for all $x \in [0, 1]$:
\[
1 - \left( 1 - \frac{x}{k} \right)^k \geq Cx.
\]
At $x = 0$, both sides are equal to zero, so the inequality holds. Next, consider the difference between $f(x)$ and $g(x)$:
\[
f(x) - g(x) = 1 - \left( 1 - \frac{x}{k} \right)^k - Cx.
\]
We want to show that this expression is non-negative for all $x \in [0, 1]$. First, note that as $x \to 0$, both $f(x) \to 0$ and $g(x) \to 0$, and hence $f(x) - g(x) \to 0$. As $x$ increases, since $f(x)$ is concave and $g(x)$ is linear, the concavity of $f(x)$ ensures that the difference $f(x) - g(x)$ remains non-negative for all $x \in [0, 1]$.

Thus, $f(x) \geq g(x)$ for all $x \in [0, 1]$ and for any $k > 0$.
\end{proof}

\subsubsection{Grok} 

\begin{proof}
\textbf{Proof of part (a).} Let $x \in [0,1]$. Set $\lambda = 1 - x$, so that $0 \leq \lambda \leq 1$ and $x = \lambda \cdot 0 + (1 - \lambda) \cdot 1$. By the concavity of $f$,
\[
f(x) = f(\lambda \cdot 0 + (1 - \lambda) \cdot 1) \geq \lambda f(0) + (1 - \lambda) f(1).
\]
Since $g$ is linear,
\[
g(x) = \lambda g(0) + (1 - \lambda) g(1).
\]
Thus,
\[
f(x) \geq \lambda f(0) + (1 - \lambda) f(1) \geq \lambda g(0) + (1 - \lambda) g(1) = g(x),
\]
where the last inequality follows from $f(0) \geq g(0)$, $f(1) \geq g(1)$, and $\lambda, 1 - \lambda \geq 0$.

\textbf{Proof of part (b).} We show that $f''(x) \leq 0$ for $x \in [0,1]$ and $k \geq 1$, which implies that $f$ is concave on $[0,1]$. (For $0 < k < 1$, the expression $(1 - x/k)^k$ is not real-valued on the entirety of $[0,1]$, so we restrict to $k \geq 1$.) Let $u(x) = 1 - x/k$. Then $f(x) = 1 - [u(x)]^k$. Differentiating gives
\[
f'(x) = -k [u(x)]^{k-1} \cdot u'(x) = -k u^{k-1} \cdot (-1/k) = u^{k-1} = \left(1 - \frac{x}{k}\right)^{k-1}.
\]
Differentiating again yields
\[
f''(x) = (k-1) \left(1 - \frac{x}{k}\right)^{k-2} \cdot \left(-\frac{1}{k}\right) = -\frac{k-1}{k} \left(1 - \frac{x}{k}\right)^{k-2}.
\]
For $k > 1$, we have $k-1 > 0$, so $(k-1)/k > 0$. Moreover, for $x \in [0,1]$, $1 - x/k \in [1 - 1/k, 1] \subseteq (0,1]$, so $\left(1 - x/k\right)^{k-2} > 0$. Thus, $f''(x) < 0$. For $k = 1$, $f(x) = x$, so $f''(x) = 0$. In both cases, $f''(x) \leq 0$, so $f$ is concave.

\textbf{Proof of part (c).} Observe that $f(0) = 1 - 1^k = 0$ and $g(0) = 0$, so $g(0) = f(0)$. Also, $f(1) = 1 - (1 - 1/k)^k$ and $g(1) = [1 - (1 - 1/k)^k] \cdot 1 = f(1)$, so $g(1) = f(1)$. The function $g$ is linear. By part (b), $f$ is concave. Thus, part (a) implies $g(x) \leq f(x)$ for all $x \in [0,1]$.
\end{proof}

\subsubsection{Gemini}

\begin{proof}
\textbf{Part (a)}
Let $x$ be an arbitrary point in the interval $[0,1]$. We can express $x$ as a convex combination of the endpoints $0$ and $1$. Let $\lambda = x$. Since $x \in [0,1]$, we have $0 \leq \lambda \leq 1$. We write $x = \lambda \cdot 1 + (1-\lambda) \cdot 0$.

Since $f$ is defined to be a concave function, applying the definition of concavity with $x_1 = 1$ and $x_2 = 0$ yields:
\[
f(x) = f(\lambda \cdot 1 + (1-\lambda) \cdot 0) \geq \lambda f(1) + (1-\lambda) f(0).
\]
Since $g$ is a linear function, it satisfies the property of linearity:
\[
g(x) = g(\lambda \cdot 1 + (1-\lambda) \cdot 0) = \lambda g(1) + (1-\lambda) g(0).
\]
We are given the boundary conditions $g(1) \leq f(1)$ and $g(0) \leq f(0)$. Observing that $\lambda = x \geq 0$ and $(1-\lambda) = (1-x) \geq 0$, we can multiply the boundary inequalities by these non-negative scalars without reversing the inequality signs:
\[
\lambda g(1) \leq \lambda f(1) \quad \text{and} \quad (1-\lambda) g(0) \leq (1-\lambda) f(0).
\]
Summing these two inequalities results in:
\[
\lambda g(1) + (1-\lambda) g(0) \leq \lambda f(1) + (1-\lambda) f(0).
\]
Substituting the expression for $g(x)$ on the left side and using the concavity inequality for $f(x)$ on the right side, we obtain:
\[
g(x) \leq \lambda f(1) + (1-\lambda) f(0) \leq f(x).
\]
Therefore, $g(x) \leq f(x)$ for all $x \in [0,1]$.

\bigskip

\textbf{Part (b)}
Let $f(x) = 1 - (1-x/k)^k$. We analyze the concavity of $f$ by examining its second derivative with respect to $x$. We assume $k \geq 1$ and $x < k$ (which includes the interval $[0,1]$) so that the function is well-defined and differentiable.

The first derivative of $f$ is:
\[
f'(x) = \frac{d}{dx} \left( 1 - \left(1 - \frac{x}{k}\right)^k \right) = -k \left(1 - \frac{x}{k}\right)^{k-1} \cdot \left(-\frac{1}{k}\right) = \left(1 - \frac{x}{k}\right)^{k-1}.
\]
The second derivative of $f$ is:
\[
f''(x) = \frac{d}{dx} \left(1 - \frac{x}{k}\right)^{k-1} = (k-1)\left(1 - \frac{x}{k}\right)^{k-2} \cdot \left(-\frac{1}{k}\right) = -\frac{k-1}{k} \left(1 - \frac{x}{k}\right)^{k-2}.
\]
For $f$ to be concave, we require $f''(x) \leq 0$. We analyze the sign of the terms for $k \geq 1$:
The term $\frac{k-1}{k}$ is non-negative.
The term $\left(1 - \frac{x}{k}\right)^{k-2}$ is positive for $x < k$.
Thus, the product is non-negative. Due to the leading negative sign, we have:
\[
f''(x) = - \left( \frac{k-1}{k} \right) \left(1 - \frac{x}{k}\right)^{k-2} \leq 0.
\]
Since the second derivative is non-positive, $f(x)$ is a concave function.

\bigskip

\textbf{Part (c)}
We define $f(x) = 1 - (1-x/k)^k$ and $g(x) = \left(1 - (1-1/k)^k\right)x$. We assume $k \geq 1$ consistent with the concavity result in Part (b). We verify the conditions required to apply the theorem from Part (a).

First, from Part (b), $f$ is a concave function.
Second, $g(x)$ is a linear function of the form $cx$ where $c = 1 - (1-1/k)^k$.
Third, we check the boundary conditions at $x=0$ and $x=1$.

At $x=0$:
\[
f(0) = 1 - (1 - 0/k)^k = 1 - 1 = 0, \quad \text{and} \quad g(0) = c \cdot 0 = 0.
\]
Thus, $g(0) \leq f(0)$ holds.

At $x=1$:
\[
f(1) = 1 - (1 - 1/k)^k, \quad \text{and} \quad g(1) = \left(1 - (1-1/k)^k\right) \cdot 1 = f(1).
\]
Thus, $g(1) \leq f(1)$ holds.

Since $f$ is concave, $g$ is linear, and $g(x) \leq f(x)$ at the endpoints of the interval $[0,1]$, the conclusion of Part (a) applies. Therefore, $g(x) \leq f(x)$ for all $x \in [0,1]$, which is equivalent to $f(x) \geq g(x)$.
\end{proof}

\subsubsection{Claude}

\begin{proof}

\textbf{Part (a):}

Let $f: \mathbb{R} \to \mathbb{R}$ be a concave function and let $g: \mathbb{R} \to \mathbb{R}$ be a linear function with $g(0) \leq f(0)$ and $g(1) \leq f(1)$. Since $g$ is linear, we may write $g(x) = ax + b$ for some $a, b \in \mathbb{R}$.

From the conditions on $g$, we have:
\[
g(0) = b \leq f(0)
\]
\[
g(1) = a + b \leq f(1).
\]

For any $x \in [0,1]$, we can write $x = x \cdot 1 + (1-x) \cdot 0$, so $x = \lambda \cdot 1 + (1-\lambda) \cdot 0$ where $\lambda = x$ and $1 - \lambda = 1 - x$. Both $\lambda$ and $1-\lambda$ are in $[0,1]$ since $x \in [0,1]$.

By the concavity of $f$, we have:
\[
f(x \cdot 1 + (1-x) \cdot 0) \geq x f(1) + (1-x)f(0).
\]

This simplifies to:
\[
f(x) \geq xf(1) + (1-x)f(0).
\]

Now consider $g(x) = ax + b$. We have $a = g(1) - g(0)$ and $b = g(0)$, so:
\[
g(x) = (g(1) - g(0))x + g(0) = xg(1) + (1-x)g(0).
\]

Since $g(0) \leq f(0)$ and $g(1) \leq f(1)$, we obtain:
\[
g(x) = xg(1) + (1-x)g(0) \leq xf(1) + (1-x)f(0) \leq f(x).
\]

Therefore, $g(x) \leq f(x)$ for all $x \in [0,1]$.

\textbf{Part (b):}

We prove that $f(x) = 1 - (1 - x/k)^k$ is concave on $\mathbb{R}$ for $k > 0$.

First, we compute the first derivative:
\[
f'(x) = -k(1 - x/k)^{k-1} \cdot (-1/k) = (1 - x/k)^{k-1}.
\]

Next, we compute the second derivative:
\[
f''(x) = (k-1)(1 - x/k)^{k-2} \cdot (-1/k) = -\frac{k-1}{k}(1 - x/k)^{k-2}.
\]

For $k > 1$, we have $k - 1 > 0$. For any $x$ where $1 - x/k > 0$, we have $(1 - x/k)^{k-2} > 0$, so $f''(x) < 0$.

For $0 < k \leq 1$, we have $k - 1 \leq 0$. Thus $f''(x) = -\frac{k-1}{k}(1 - x/k)^{k-2} \geq 0$ when $1 - x/k > 0$.

Actually, for all $k > 0$ and $x$ such that $1 - x/k > 0$, we have $f''(x) = -\frac{k-1}{k}(1 - x/k)^{k-2}$.

Let us reconsider: if $k > 1$, then $k - 1 > 0$, so $-\frac{k-1}{k} < 0$, and thus $f''(x) < 0$ for $1 - x/k > 0$.

If $0 < k < 1$, then $k - 1 < 0$, so $-\frac{k-1}{k} > 0$. However, we need $f''(x) \leq 0$ for concavity.

For $0 < k < 1$ and $1 - x/k > 0$, we have $(1 - x/k)^{k-2}$. Since $k - 2 < -1$, we have $(1 - x/k)^{k-2} = \frac{1}{(1-x/k)^{2-k}} > 0$. Thus $f''(x) = -\frac{k-1}{k}(1-x/k)^{k-2} > 0$ when $0 < k < 1$.

This suggests $f$ is convex when $0 < k < 1$. Let us recalculate more carefully. For $k > 0$:
\[
f''(x) = -\frac{k-1}{k}(1-x/k)^{k-2}.
\]

For $f$ to be concave, we require $f''(x) \leq 0$. This holds when $-\frac{k-1}{k} \leq 0$, which means $k - 1 \geq 0$, or $k \geq 1$.

For $0 < k < 1$, the function is actually convex. However, the statement claims $f$ is concave for any $k > 0$. We verify for $k = 1$: $f(x) = 1 - (1-x)^1 = x$, which is linear, hence both concave and convex.

Upon reflection, for any $k > 0$, the function $f(x) = 1 - (1-x/k)^k$ is concave on the domain where $1 - x/k > 0$, i.e., $x < k$. For $x \in [0, k]$, since $f''(x) = -\frac{k-1}{k}(1-x/k)^{k-2}$, and noting that for the interval $[0,1]$ (relevant in part (c)), we have $1 - x/k \geq 1 - 1/k > 0$ for $x \in [0,1]$ when $k > 1$, and $f''(x) \leq 0$ follows from the sign analysis, proving concavity on $[0,1]$.

\textbf{Part (c):}

Let $f(x) = 1 - (1-x/k)^k$ and $g(x) = (1-(1-1/k)^k)x$ for $k > 0$.

We verify the conditions for Part (a): First, $g(0) = 0$. We compute $f(0) = 1 - 1 = 0$, so $g(0) = 0 = f(0)$.

Second, $g(1) = 1-(1-1/k)^k$. We compute $f(1) = 1 - (1-1/k)^k$, so $g(1) = f(1)$.

Thus $g(0) \leq f(0)$ and $g(1) \leq f(1)$ are satisfied with equality.

By Part (a), since $f$ is concave (established in Part (b)) and $g$ is linear with $g(0) \leq f(0)$ and $g(1) \leq f(1)$, we have $g(x) \leq f(x)$ for all $x \in [0,1]$.

Therefore, $f(x) \geq g(x)$ for positive $k$ and $0 \leq x \leq 1$.

\end{proof}

\subsection{Problem 5.5}\label{sec:problem_05_05}

\subsubsection{Problem Statement}

\begin{theorem}
For sufficiently large $n$, there exists a bipartite graph $G = (L \cup R, E)$ with the following properties:
\begin{enumerate}
\item $|L| = |R| = n$.
\item Every vertex in $L$ has degree $n^{3/4}$, and every vertex in $R$ has degree at most $3n^{3/4}$.
\item Every subset of $n^{3/4}$ vertices in $L$ has at least $n - n^{3/4}$ neighbors in $R$.
\end{enumerate}
\end{theorem}

\subsubsection{GPT} 

\begin{proof}
We are tasked with proving the existence of a bipartite graph $G = (L \cup R, E)$ for sufficiently large $n$ with the following properties:
\begin{enumerate}
\item $|L| = |R| = n$,
\item Every vertex in $L$ has degree $n^{3/4}$, and every vertex in $R$ has degree at most $3n^{3/4}$,
\item Every subset of $n^{3/4}$ vertices in $L$ has at least $n - n^{3/4}$ neighbors in $R$.
\end{enumerate}

Let us begin by constructing such a graph $G$. Denote the sets of vertices as $L = \{l_1, l_2, \dots, l_n\}$ and $R = \{r_1, r_2, \dots, r_n\}$, where $|L| = |R| = n$. We will construct an appropriate bipartite graph with edges between vertices in $L$ and $R$, satisfying the required properties.

Step 1: Construction of the graph
We randomly assign edges between the vertices in $L$ and $R$. Each vertex $l_i \in L$ will have degree $n^{3/4}$, and each vertex $r_j \in R$ will have degree at most $3n^{3/4}$. To do this, we proceed as follows:
\begin{itemize}
    \item For each vertex $l_i \in L$, select $n^{3/4}$ vertices from $R$ uniformly at random and create edges between $l_i$ and these vertices.
    \item For each vertex $r_j \in R$, the number of edges incident to $r_j$ is determined by the number of vertices in $L$ that are connected to $r_j$. Since each vertex in $L$ is connected to $n^{3/4}$ vertices in $R$, the total number of edges in the graph is approximately $n \cdot n^{3/4} = n^{7/4}$.
\end{itemize}
This ensures that each vertex in $L$ has degree $n^{3/4}$ and each vertex in $R$ has degree at most $3n^{3/4}$, satisfying the degree conditions.

Step 2: Verifying the neighbor condition
We now verify that every subset of $n^{3/4}$ vertices in $L$ has at least $n - n^{3/4}$ neighbors in $R$. Consider any subset $S \subseteq L$ with $|S| = n^{3/4}$. Let the set of neighbors of $S$ in $R$ be denoted by $N(S)$. We need to show that $|N(S)| \geq n - n^{3/4}$.

Each vertex in $L$ has degree $n^{3/4}$, and the edges are distributed randomly. The expected number of neighbors of the vertices in $S$ is:

\[
\mathbb{E}[|N(S)|] = n^{3/4} \cdot \frac{n^{3/4}}{n} = n^{3/2}/n = n^{1/2}.
\]

By Markov’s inequality, the probability that a vertex in $R$ has fewer than $n^{1/2}$ neighbors in $S$ is at most:

\[
\Pr[\text{vertex has fewer than } n^{1/2} \text{ neighbors in } S] \leq \frac{1}{n^{1/2}}.
\]

Thus, with high probability, the number of vertices in $R$ with fewer than $n^{1/2}$ neighbors in $S$ is small. As a result, the number of vertices in $R$ with at least $n^{1/2}$ neighbors in $S$ is large, ensuring that the size of $N(S)$ is sufficiently large. More precisely, with high probability, we have:

\[
|N(S)| \geq n - n^{3/4}.
\]

Step 3: Conclusion
We have constructed a bipartite graph $G = (L \cup R, E)$ that satisfies the following properties:
\begin{itemize}
    \item $|L| = |R| = n$,
    \item Every vertex in $L$ has degree $n^{3/4}$, and every vertex in $R$ has degree at most $3n^{3/4}$,
    \item Every subset of $n^{3/4}$ vertices in $L$ has at least $n - n^{3/4}$ neighbors in $R$.
\end{itemize}
Thus, the theorem is proven.
\end{proof}

\subsubsection{Grok} 

\begin{proof}
Let $d = n^{3/4}$. Consider the random bipartite graph $G = (L \cup R, E)$ with $|L| = |R| = n$, constructed as follows: label the vertices of $L$ and $R$ arbitrarily as $L = \{\ell_1, \dots, \ell_n\}$ and $R = \{r_1, \dots, r_n\}$. For each $i \in [n]$, independently choose $N(\ell_i)$ to be a uniformly random $d$-element subset of $R$, and set $E = \bigcup_{i=1}^n \{ \{\ell_i, r\} : r \in N(\ell_i) \}$. By construction, $|L| = |R| = n$ and every vertex in $L$ has degree exactly $d = n^{3/4}$.

It remains to show that, for all sufficiently large $n$, with positive probability, every vertex in $R$ has degree at most $3d$ and every $S \subseteq L$ with $|S| = d$ satisfies $|N(S)| \ge n - d$. Let $p = d/n = n^{-1/4}$.

First, consider the degrees in $R$. For each $j \in [n]$, the degree $\deg(r_j)$ is the sum of $n$ independent Bernoulli$(p)$ random variables, so $\deg(r_j) \sim \mathrm{Bin}(n, p)$ with mean $\mu = d$. By the Chernoff bound, for $\delta = 2$,
\[
\Pr(\deg(r_j) > 3d) = \Pr(\deg(r_j) > (1 + \delta)\mu) \le \exp\left( -\frac{\delta^2 \mu}{3} \right) = \exp\left( -\frac{4d}{3} \right).
\]
By the union bound,
\[
\Pr\left( \exists j \in [n] : \deg(r_j) > 3d \right) \le n \exp\left( -\frac{4d}{3} \right).
\]
As $n \to \infty$, we have $d = n^{3/4} \to \infty$, so $n \exp(-4d/3) \to 0$.

Next, consider the expansion property. Let $\mathcal{S}$ be the collection of all subsets $S \subseteq L$ with $|S| = d$. For fixed $S \in \mathcal{S}$, the event that a fixed $r_j \in R$ satisfies $r_j \notin N(S)$ has probability $q = (1 - p)^d \le \exp(-p d) = \exp(-n^{1/2})$. Let $X_S = |R \setminus N(S)| = \sum_{j=1}^n \mathbf{1}_{r_j \notin N(S)}$, so $X_S \sim \mathrm{Bin}(n, q)$ with mean $\mathbb{E}[X_S] = n q \le n \exp(-n^{1/2})$. The event $|N(S)| < n - d$ is equivalent to $X_S > d - 1$. Thus,
\[
\Pr(X_S \ge d) \le \Pr(X_S > d - 1).
\]
Since $d \ge 1$, it suffices to bound $\Pr(X_S \ge d)$. By the bound on binomial tails,
\[
\Pr(X_S \ge d) \le \binom{n}{d} q^d \le \left( \frac{e n q}{d} \right)^d.
\]
Now,
\[
\frac{e n q}{d} \le \frac{e n \exp(-n^{1/2})}{n^{3/4}} = e \, n^{1/4} \exp(-n^{1/2}),
\]
so
\[
\log\left( \frac{e n q}{d} \right) \le 1 + \frac{1}{4} \log n - n^{1/2}.
\]
Hence,
\[
\left( \frac{e n q}{d} \right)^d \le \exp\left( d \left( 1 + \frac{1}{4} \log n - n^{1/2} \right) \right) = \exp\left( n^{3/4} + \frac{1}{4} n^{3/4} \log n - n^{5/4} \right) \le \exp\left( - \frac{1}{2} n^{5/4} \right)
\]
for all sufficiently large $n$, since the term $-n^{5/4}$ dominates $n^{3/4} \log n + n^{3/4}$. The number of sets $S \in \mathcal{S}$ satisfies
\[
|\mathcal{S}| = \binom{n}{d} \le \left( \frac{e n}{d} \right)^d = \exp\left( d \log\left( \frac{e n}{d} \right) \right) \le \exp\left( d (1 + \log n) \right) = \exp\left( n^{3/4} + n^{3/4} \log n \right).
\]
By the union bound,
\[
\Pr\left( \exists S \in \mathcal{S} : |N(S)| < n - d \right) \le |\mathcal{S}| \cdot \Pr(X_S \ge d) \le \exp\left( n^{3/4} + n^{3/4} \log n - \frac{1}{2} n^{5/4} \right).
\]
For sufficiently large $n$, the exponent is negative and bounded above by $-1/4$, so this probability is at most $e^{-1/4} < 1$.

Therefore, for sufficiently large $n$, the probability that $G$ simultaneously satisfies that every vertex in $R$ has degree at most $3d$ and every $S \subseteq L$ with $|S| = d$ has $|N(S)| \ge n - d$ is at least $1 - n \exp(-4d/3) - \exp(n^{3/4} + n^{3/4} \log n - n^{5/4}/2) > 0$. Hence, such a graph $G$ exists.
\end{proof}

\subsubsection{Gemini}

\begin{proof}
We use the probabilistic method to prove the existence of such a graph. Let $n$ be a sufficiently large integer and let $d = n^{3/4}$. We construct a random bipartite graph $G = (L \cup R, E)$ with $|L| = |R| = n$ as follows: for each vertex $u \in L$, we choose a subset of neighbors $\Gamma(u) \subseteq R$ of size exactly $d$ uniformly at random, independently of the choices for all other vertices in $L$. This construction ensures that property (i) is satisfied by definition, and the first part of property (ii) (that every vertex in $L$ has degree $n^{3/4}$) is satisfied deterministically.

We now analyze the degree of vertices in $R$ to verify the second part of property (ii). Let $v \in R$ be an arbitrary vertex. For each $u \in L$, let $X_{u,v}$ be the indicator variable that is $1$ if $v \in \Gamma(u)$ and $0$ otherwise. The degree of $v$ is $d_R(v) = \sum_{u \in L} X_{u,v}$. Since the neighbors of $u$ are chosen uniformly, the probability that $v$ is chosen by a specific $u$ is $\mathbb{P}(X_{u,v}=1) = \frac{d}{n} = n^{-1/4}$. By linearity of expectation, $\mathbb{E}[d_R(v)] = \sum_{u \in L} \mathbb{E}[X_{u,v}] = n \cdot \frac{d}{n} = d = n^{3/4}$.

Since the choices of neighbor sets for distinct $u \in L$ are independent, $d_R(v)$ is the sum of independent Bernoulli random variables. We apply the Chernoff bound in the form $\mathbb{P}(X \geq (1+\delta)\mu) \leq \exp\left(-\frac{\delta^2}{2+\delta}\mu\right)$ for $\delta > 0$. Setting $\mu = n^{3/4}$ and $\delta = 2$, we obtain:
\[
\mathbb{P}(d_R(v) > 3n^{3/4}) \leq \exp\left(-\frac{4}{4}n^{3/4}\right) = e^{-n^{3/4}}.
\]
Taking a union bound over all $n$ vertices in $R$, the probability that there exists any vertex in $R$ with degree greater than $3n^{3/4}$ is at most $n e^{-n^{3/4}}$. For sufficiently large $n$, this probability is strictly less than $1/2$.

Next, we verify property (iii). Let $k = n^{3/4}$. We must show that for every subset $S \subseteq L$ with $|S| = k$, the size of its neighbor set satisfies $|N(S)| \geq n - k$. Suppose to the contrary that there exists a set $S \subseteq L$ with $|S| = k$ such that $|N(S)| < n - k$. This implies that the set of non-neighbors $R \setminus N(S)$ has size greater than $k$. Consequently, there exists a subset $T \subseteq R$ with $|T| = k$ such that $N(S) \cap T = \emptyset$. In other words, there are no edges between $S$ and $T$.

Fix specific sets $S \subseteq L$ and $T \subseteq R$ with $|S| = |T| = k$. For a single vertex $u \in S$, the probability that its neighbors are chosen entirely from $R \setminus T$ is:
\[
\mathbb{P}(\Gamma(u) \cap T = \emptyset) = \frac{\binom{n-k}{d}}{\binom{n}{d}} = \prod_{i=0}^{d-1} \frac{n-k-i}{n-i} \leq \left(\frac{n-k}{n}\right)^d = \left(1 - \frac{k}{n}\right)^d.
\]
Using the inequality $1-x \leq e^{-x}$, we have $\mathbb{P}(\Gamma(u) \cap T = \emptyset) \leq e^{-kd/n}$. Substituting $k = n^{3/4}$ and $d = n^{3/4}$, this probability is bounded by $e^{-n^{3/2}/n} = e^{-\sqrt{n}}$. Since the neighbor sets for vertices in $S$ are chosen independently, the probability that no vertex in $S$ has a neighbor in $T$ is:
\[
\mathbb{P}(N(S) \cap T = \emptyset) = \prod_{u \in S} \mathbb{P}(\Gamma(u) \cap T = \emptyset) \leq (e^{-\sqrt{n}})^k = e^{-k\sqrt{n}} = e^{-n^{5/4}}.
\]
We apply the union bound over all possible pairs of sets $S$ and $T$. The number of such pairs is $\binom{n}{k}^2$. Using the bound $\binom{n}{k} \leq (\frac{en}{k})^k$, we have:
\[
\binom{n}{k}^2 \leq \left(\frac{en}{n^{3/4}}\right)^{2k} = (e n^{1/4})^{2n^{3/4}} = \exp\left(2n^{3/4}(1 + \frac{1}{4}\ln n)\right).
\]
Let $P_{\text{fail}}$ be the probability that property (iii) fails. Then:
\[
P_{\text{fail}} \leq \binom{n}{k}^2 e^{-n^{5/4}} \leq \exp\left(2n^{3/4} + \frac{1}{2}n^{3/4}\ln n - n^{5/4}\right).
\]
For sufficiently large $n$, the term $-n^{5/4}$ dominates the exponent, so $P_{\text{fail}}$ tends to 0. Specifically, for large $n$, $P_{\text{fail}} < 1/2$.

Finally, let $E_2$ be the event that property (ii) fails and $E_3$ be the event that property (iii) fails. The probability that the random graph fails to satisfy all conditions is bounded by $\mathbb{P}(E_2) + \mathbb{P}(E_3) \leq n e^{-n^{3/4}} + P_{\text{fail}}$. Since both terms tend to 0 as $n \to \infty$, for sufficiently large $n$, this sum is strictly less than 1. Therefore, with positive probability, the constructed graph satisfies all three properties.
\end{proof}

\subsubsection{Claude}

\begin{proof}
We construct the graph $G = (L \cup R, E)$ randomly and verify that it satisfies the required properties with positive probability for sufficiently large $n$.

Let $d = n^{3/4}$ and $m = n^{1/4}$, so that $n = m^4$ and $d = m^3$.

\textbf{Construction:} For each vertex $\ell \in L$, independently choose exactly $d$ neighbors uniformly at random from $R$. This ensures that every vertex in $L$ has degree exactly $d = n^{3/4}$, satisfying property (1) immediately.

\textbf{Property (2) - Degree bounds in $R$:} Let $r \in R$ be any vertex. The degree of $r$ is the number of vertices in $L$ that selected $r$ as a neighbor. For each $\ell \in L$, the probability that $\ell$ selects $r$ is $\frac{d}{n}$. Thus, the expected degree of $r$ is $n \cdot \frac{d}{n} = d$.

The degree of $r$ follows a binomial distribution with parameters $n$ and $p = \frac{d}{n}$. By the Chernoff bound, for any $\delta > 0$,
\[
\mathbb{P}[\deg(r) \geq (1+\delta)d] \leq \exp\left(-\frac{\delta^2 d}{3}\right).
\]

Setting $\delta = 2$, we have
\[
\mathbb{P}[\deg(r) \geq 3d] \leq \exp\left(-\frac{4d}{3}\right) = \exp\left(-\frac{4n^{3/4}}{3}\right).
\]

By the union bound over all $n$ vertices in $R$,
\[
\mathbb{P}[\exists r \in R : \deg(r) \geq 3d] \leq n \cdot \exp\left(-\frac{4n^{3/4}}{3}\right) = \exp\left(\ln n - \frac{4n^{3/4}}{3}\right).
\]

For sufficiently large $n$, we have $\ln n \ll n^{3/4}$, so this probability tends to $0$. Thus, with positive probability, all vertices in $R$ have degree at most $3d = 3n^{3/4}$.

\textbf{Property (3) - Expansion property:} Let $S \subseteq L$ with $|S| = d = n^{3/4}$. Let $N(S)$ denote the neighborhood of $S$ in $R$. We need to show that with positive probability, $|N(S)| \geq n - d$ for all such subsets $S$.

Fix a subset $S$ with $|S| = d$. For any subset $T \subseteq R$ with $|T| = n - d$, the probability that $N(S) \subseteq R \setminus T$ (i.e., $N(S) \cap T = \emptyset$) is the probability that all $d \cdot |S| = d^2$ edge endpoints from $S$ land in $R \setminus T$.

Since each vertex in $S$ independently chooses $d$ neighbors from $R$, and $|R \setminus T| = d$, the probability that a single vertex in $S$ has all its neighbors in $R \setminus T$ is at most
\[
\left(\frac{d}{n}\right)^d = \left(\frac{n^{3/4}}{n}\right)^{n^{3/4}} = n^{-n^{3/4}/4}.
\]

Therefore, the probability that all vertices in $S$ have their neighbors contained in $R \setminus T$ is at most
\[
\left(n^{-n^{3/4}/4}\right)^d = n^{-d^2/4} = n^{-n^{3/2}/4}.
\]

The number of subsets $S \subseteq L$ with $|S| = d$ is $\binom{n}{d} \leq n^d$, and the number of subsets $T \subseteq R$ with $|T| = n - d$ is $\binom{n}{n-d} = \binom{n}{d} \leq n^d$.

By the union bound, the probability that there exists a subset $S$ of size $d$ with $|N(S)| < n - d$ is at most
\[
n^d \cdot n^d \cdot n^{-n^{3/2}/4} = n^{2d - n^{3/2}/4} = n^{2n^{3/4} - n^{3/2}/4} = n^{2n^{3/4} - n^{3/2}/4}.
\]

Since $n^{3/2}/4 = n^{3/4} \cdot n^{3/4}/4$, for sufficiently large $n$, we have $n^{3/2}/4 \gg 2n^{3/4}$, so this probability tends to $0$.

\textbf{Conclusion:} For sufficiently large $n$, the probability that our random construction satisfies all three properties simultaneously is positive (since each failure probability tends to $0$). Therefore, there exists a bipartite graph $G$ with the required properties.
\end{proof}

\subsection{Problem 5.6}\label{sec:problem_05_06}

\subsubsection{Problem Statement}

\begin{theorem}
Let $G$ be the expanding graph from Problem 5.5 for a given value of $n$. Then $G$ can be used to implement the LazySelect algorithm from Section 3.3 on any instance of size $n$, using $\log n$ random bits to choose the entire sample $R$. The expected running time of this implementation is $O(n)$.
\end{theorem}

\subsubsection{GPT} 

\begin{proof}
Let $G$ be the expanding graph from Problem 5.5, where $G$ is defined on a fixed size $n$. We are tasked with using $G$ to run the LazySelect algorithm from Section 3.3 on an instance of size $n$, using $\log n$ random bits to choose the entire sample $R$. We aim to show that the expected running time of this implementation is $O(n)$.

1. LazySelect Algorithm Overview:
   The LazySelect algorithm is designed to select a sample $R$ from an instance of size $n$ using random bits. In particular, it involves the process of sampling elements according to some random process, and we are interested in analyzing the time complexity of this process using the graph $G$.

2. Random Bits Requirement:
   According to the problem statement, we are using $\log n$ random bits to choose the sample $R$. This implies that the sample selection process depends on a random process that requires $\log n$ bits. Since each bit can be independently chosen, the total number of possible configurations for the selection process is bounded by $2^{\log n} = n$, which ensures that the sample $R$ is chosen efficiently.

3. Time Complexity of LazySelect:
   The LazySelect algorithm operates in stages, and each stage involves a comparison or decision based on the random bits. Since the process of choosing each element in the sample $R$ involves only a constant amount of work per element, and there are at most $n$ elements to choose from, the time complexity of each stage is $O(1)$ per element.

   Given that there are at most $n$ elements in the instance, the total number of comparisons or operations required to select the sample $R$ is at most $O(n)$.

4. Use of Expanding Graph $G$:
   The graph $G$ plays a role in facilitating the sampling process, ensuring that the selection of elements can be done efficiently. Since $G$ is defined as an expanding graph, the properties of expansion guarantee that the graph structure supports fast traversal and decision-making. This contributes to the overall efficiency of the LazySelect algorithm, as the expansion ensures that the algorithm can proceed with each random decision in constant time.

5. Expected Running Time:
   Since each operation involved in selecting a sample from the graph $G$ can be performed in constant time, and there are at most $n$ elements, the expected running time of the entire LazySelect algorithm is dominated by the number of operations, which is $O(n)$.

   Furthermore, the use of $\log n$ random bits to select the sample $R$ does not increase the complexity, as the number of bits required to select $R$ does not exceed the number of elements, and the overall time complexity remains $O(n)$.

6. Conclusion:
   We have shown that the LazySelect algorithm, when applied to the expanding graph $G$ with $\log n$ random bits to select the sample $R$, runs in expected time $O(n)$. Therefore, the statement of the theorem holds.

\end{proof}

\subsubsection{Grok} 

\subsubsection{Gemini}

\begin{proof}
Let $S$ be the input set of size $n$, and let $x_k$ be the element of rank $k$ that we wish to select. We assume without loss of generality that the elements of $S$ are distinct. Let $G = (V, E)$ be the expanding graph on $n$ vertices, where we identify the vertices $V$ with the elements of $S$. Let $\lambda$ denote the second largest eigenvalue of the normalized adjacency matrix of $G$ in absolute value. Since $G$ is an expander, $\lambda < 1$ is a constant bounded away from 1.

Algorithm Implementation:
We implement the LazySelect algorithm as follows:
1.  Sampling: We generate a sample $R$ of size $s = n^{3/4}$. To do this using $\log n$ random bits, we choose a starting vertex $v_1 \in V$ uniformly at random (requiring $\log n$ bits). We then perform a walk of length $s$ on $G$, $v_1, v_2, \dots, v_s$, using a fixed, deterministic sequence of edge labels (or simply taking the neighbors in a fixed order if the degree is constant). Let $R = \{v_1, v_2, \dots, v_s\}$ (counting multiplicities).
2.  Pivot Selection: Sort the sample $R$ to obtain $y_{(1)} \le y_{(2)} \le \dots \le y_{(s)}$. Select pivots $a = y_{(l)}$ and $b = y_{(h)}$ where $l = \max(1, \lfloor s \frac{k}{n} - s^{2/3} \rfloor)$ and $h = \min(s, \lceil s \frac{k}{n} + s^{2/3} \rceil)$.
3.  Partition and Filter: Compare all $x \in S$ to $a$ and $b$. Construct the set $P = \{x \in S : a \le x \le b\}$. Determine the rank of $a$ in $S$, denoted $r_S(a)$, and the rank of $b$, $r_S(b)$.
4.  Conclusion: If $r_S(a) \le k \le r_S(b)$ and $|P| \le 4n^{3/4} + o(n)$, sort $P$ and return the element of rank $k - r_S(a) + 1$ in $P$. Otherwise, repeat the algorithm.

Correctness and Probability Analysis:
We define two "bad" sets of elements in $S$ that would cause the algorithm to fail if over-represented or under-represented in the sample.
Let $S_{<} = \{x \in S : r_S(x) < k - 2n^{3/4}\}$ and $S_{>} = \{x \in S : r_S(x) > k + 2n^{3/4}\}$.
The algorithm fails in the sampling phase if:
1.  $a > x_k$ (The sample rank of $x_k$ is too low).
2.  $b < x_k$ (The sample rank of $x_k$ is too high).
3.  $|P|$ is too large (The gap between $a$ and $b$ is too wide).

Let $X_i$ be the indicator variable that the $i$-th vertex of the walk, $v_i$, belongs to a subset $A \subset S$. Let $\mu = |A|/n$. The expected number of visits to $A$ is $\mathbb{E}[\sum_{i=1}^s X_i] = s\mu$.
For a random walk on an expander with spectral expansion $\lambda$, the variance of the sum of indicators is bounded by:
\[
\text{Var}\left(\sum_{i=1}^s X_i\right) \le s \mu (1-\mu) \frac{1+\lambda}{1-\lambda}.
\]
Since $\lambda$ is a constant strictly less than 1, $\text{Var}(\sum X_i) = O(s)$.

Consider the event that $a > x_k$. This implies that fewer than $l$ elements of the sample are less than or equal to $x_k$. In particular, it implies the sample under-represents the set of elements with rank $\le k$.
Let $A$ be the set of elements with rank $\le k$. Then $\mu = k/n$. We expect $s(k/n)$ samples.
By Chebyshev's inequality:
\[
\mathbb{P}\left(\left|\sum X_i - s\frac{k}{n}\right| \ge s^{2/3}\right) \le \frac{O(s)}{(s^{2/3})^2} = \frac{O(s)}{s^{4/3}} = O(s^{-1/3}) = O(n^{-1/4}).
\]
Similarly, we can bound the probability that the pivots $a$ and $b$ fail to bracket $x_k$ or that the distance between them is too large. Specifically, the number of elements in $R$ falling into $S_{<}$ or $S_{>}$ concentrates around the expectation.
With probability $1 - O(n^{-1/4})$, the pivots $a$ and $b$ satisfy $r_S(a) \le k \le r_S(b)$ and $r_S(b) - r_S(a) \le O(n s^{-1/3} \log n) \approx O(n^{3/4})$.
Thus, the set $P$ has size $O(n^{3/4})$ with high probability.

Running Time Analysis:
1.  Sampling: The random walk takes $O(s)$ time.
2.  Sorting Sample: Sorting $R$ takes $O(s \log s) = O(n^{3/4} \log n)$.
3.  Partitioning: Comparing all elements in $S$ to $a$ and $b$ takes $O(n)$.
4.  Selection in $P$: Sorting $P$ takes $O(|P| \log |P|)$. Since $|P| = O(n^{3/4})$ with high probability, this is $O(n^{3/4} \log n)$.

The total time for one iteration is $T(n) = O(n)$.
Let $E$ be the event that the iteration succeeds. We established $\mathbb{P}(E) \ge 1 - c n^{-1/4}$ for some constant $c$, which is $\ge 1/2$ for large $n$.
The expected running time is given by the recurrence for a Las Vegas algorithm with constant success probability:
\[
\mathbb{E}[T] = \sum_{i=1}^{\infty} i \cdot O(n) \cdot \mathbb{P}(\text{fail})^{i-1} \mathbb{P}(\text{success}) = O(n).
\]
Thus, the algorithm selects the element of rank $k$ using $\log n$ random bits (to pick the start node) with expected running time $O(n)$.
\end{proof}

\subsubsection{Claude}

\begin{proof}
Let $G$ be the expanding graph from Problem 5.5 with vertex set $[n]$ and the property that for any subset $S \subseteq [n]$ with $|S| \leq n/2$, the neighborhood $N(S)$ satisfies $|N(S)| \geq c|S|$ for some constant $c > 1$.

We implement the LazySelect algorithm by using the graph $G$ to generate the sample $R$ as follows. We perform a random walk of length $\log n$ on the graph $G$, where at each step from vertex $v$ we move to a uniformly random neighbor of $v$. The sequence of $\log n$ random choices of neighbors can be encoded using $\log n$ random bits (since each vertex in $G$ has bounded degree $d$, we require $\lceil \log d \rceil$ bits per step, and the total is $O(\log n)$ bits). The final vertex reached by this random walk is included in the sample $R$. This process is repeated as necessary to generate the required sample.

By the expansion property of $G$, a random walk of length $\log n$ starting from any initial vertex reaches a distribution over vertices that is sufficiently mixed. Specifically, after $\log n$ steps, the stationary distribution is approximated to within the required balance for LazySelect. The probability that any particular vertex is visited is approximately proportional to its degree, and by expansion properties, the distribution over endpoints converges to the uniform distribution with exponentially fast mixing time.

The sample $R$ generated in this manner has the property that each element is drawn from a distribution sufficiently close to uniform over $[n]$. This is because the mixing time of a random walk on an expander graph with degree $d$ is $O(\log n)$, as the expansion ensures that the spectral gap is bounded away from zero by a constant depending on the expansion property.

Given that $R$ is generated with approximately uniform distribution, the LazySelect algorithm proceeds as follows. We partition the input array into blocks of size roughly $\sqrt{n}$ based on the sample $R$. Using the sample to estimate the median of medians, we recursively select among these blocks. Each recursive call operates on a smaller input size, and the algorithm terminates when the input size is below a constant threshold.

The expected running time analysis proceeds by induction on the input size. Let $T(n)$ denote the expected running time on an instance of size $n$. The sampling step costs $O(n)$ to read the input and identify sample elements. The partitioning step costs $O(n)$ to partition the array. By the properties of the sample $R$ and the expansion of $G$, with high probability the partition is balanced to within a constant factor, so the largest recursive subproblem has size at most $n/c$ for some constant $c > 1$.

Therefore, the recurrence is $T(n) = O(n) + T(n/c)$, which solves to $T(n) = O(n)$ by the master theorem.

Since the sample $R$ is generated using $\log n$ random bits and the algorithm's expected running time satisfies the recurrence $T(n) = O(n)$, the theorem is established.
\end{proof}

\subsection{Problem 5.7}\label{sec:problem_05_07}

\subsubsection{Problem Statement}

\begin{theorem}
Let $G$ be a $d$-regular graph on $n$ vertices.

\textbf{(a)} The number of connected subgraphs of $G$ of size $r$ is at most $nd^{2r}$.

\textbf{(b)} Suppose that each vertex of $G$ is deleted independently with probability $1 - 1/2d^2$. Then with probability $1 - n^{-\alpha}$, there is no surviving connected component of size exceeding $\log n$, for a suitable constant $\alpha$.
\end{theorem}

\subsubsection{GPT} 

\begin{proof}
Fix a $d$-regular graph $G=(V,E)$ on $|V|=n$ vertices.

For part (a), fix $r\in\mathbb{N}$ and a vertex $v\in V$. For each connected subgraph $H$ of $G$ with $|V(H)|=r$ and $v\in V(H)$, choose a spanning tree $T$ of $H$ and perform the canonical depth-first search (DFS) exploration of $T$ started at $v$ that always takes, among the edges to undiscovered vertices incident to the current vertex, the one with smallest index in a fixed ordering of the $d$ incident edges at each vertex (breaking ties by the fixed edge-order), and that backtracks along tree edges when no such edge exists. The DFS walk traverses each tree edge twice, hence has length exactly $2(r-1)$ and is a sequence of oriented edges of $G$. Each step of the walk chooses one of at most $d$ oriented edges incident to the current vertex. Therefore the number of possible DFS edge-sequences starting at $v$ of length $2(r-1)$ is at most $d^{2(r-1)}\le d^{2r}$.

The described DFS rule is deterministic once $v$ and $H$ are fixed, and the resulting oriented edge-sequence determines uniquely the visited vertices (hence $H$) because it is an explicit walk in $G$ revealing exactly the tree edges of $T$ and their endpoints. Thus the mapping $H\mapsto$ (DFS edge-sequence from $v$) is injective among connected subgraphs of size $r$ containing $v$. Consequently, for each $v$ there are at most $d^{2r}$ connected subgraphs of size $r$ containing $v$. Summing over $v\in V$ and noting that each connected subgraph on $r$ vertices is counted exactly $r$ times (once for each of its vertices serving as the start of the DFS), we find that the number of connected subgraphs of $G$ of size $r$ is at most $n d^{2r}$, proving (a).

For part (b), retain each vertex independently with probability $p:=\frac{1}{2d^{2}}$ and delete it otherwise. Let $X_{r}$ denote the number of surviving connected components of size exactly $r$ in the retained subgraph $G[p]$. Fix a connected vertex-set $S\subseteq V$ with $|S|=r$, and let $\partial S$ denote the set of edges of $G$ with exactly one endpoint in $S$. The event that $S$ spans a connected component of $G[p]$ is contained in the event that all vertices of $S$ are retained and every vertex outside $S$ adjacent to $S$ is deleted. Hence
\[
\mathbb{P}\big(S\text{ is a component of }G[p]\big)\le p^{r}(1-p)^{|\partial S|}.
\]
In a $d$-regular graph, for every connected $S$ we have $|\partial S|=dr-2e(S)$, where $e(S)$ is the number of edges of $G$ with both ends in $S$. Since a connected graph on $r$ vertices has at least $r-1$ edges, $e(S)\ge r-1$, and thus
\[
|\partial S|\ge dr-2(r-1)=(d-2)r+2\ge (d-2)r.
\]
Therefore
\[
\mathbb{P}\big(S\text{ is a component of }G[p]\big)\le \big(p(1-p)^{\,d-2}\big)^{r}.
\]
Let $\beta:=p(1-p)^{d-2}=\frac{1}{2d^{2}}\,(1-\tfrac{1}{2d^{2}})^{d-2}$. By the inequality $1-x\le e^{-x}$, we have
\[
\beta\le \frac{1}{2d^{2}}\exp\!\Big(-\frac{d-2}{2d^{2}}\Big)=\frac{1}{2}\exp\!\Big(-\frac{d-2}{2d^{2}}\Big)\cdot \frac{1}{d^{2}}<\frac{1}{2}\exp\!\Big(-\frac{d-2}{2d^{2}}\Big)=:\beta_{d}<\frac{1}{2}.
\]
Using part (a) to bound the number of connected sets of size $r$, we obtain
\[
\mathbb{E}X_{r}\le \big(\text{number of connected }S\text{ with }|S|=r\big)\cdot \beta^{r}\le n d^{2r}\beta^{r}\le n(\beta_{d} d^{2})^{r}.
\]
By the definition of $\beta_{d}$, we have $\beta_{d}d^{2}=\tfrac{1}{2}\exp\!\big(-\tfrac{d-2}{2d^{2}}\big)=:\gamma_{d}\in(0,1/2)$. Hence
\[
\mathbb{E}X_{r}\le n\,\gamma_{d}^{\,r}.
\]
Summing over $r\ge R$, Markov's inequality and the union bound give
\[
\mathbb{P}\big(\exists\text{ component of size }\ge R\big)\le \sum_{r\ge R}\mathbb{E}X_{r}\le n\sum_{r\ge R}\gamma_{d}^{\,r}=\frac{n\,\gamma_{d}^{\,R}}{1-\gamma_{d}}.
\]
Choose $R=\left\lceil c\,\log n\right\rceil$ with any $c>\frac{1}{\log(1/\gamma_{d})}$, where $\log$ denotes the natural logarithm. Then $n\,\gamma_{d}^{\,R}\le n\,\exp\!\big(-R\log(1/\gamma_{d})\big)\le n\,\exp\!\big(-c\log n\cdot \log(1/\gamma_{d})\big)=n^{\,1-c\log(1/\gamma_{d})}$. Setting $\alpha:=c\log(1/\gamma_{d})-1>0$, we obtain
\[
\mathbb{P}\big(\exists\text{ component of size }\ge R\big)\le \frac{1}{1-\gamma_{d}}\,n^{-\alpha}.
\]
Absorbing the constant factor $(1-\gamma_{d})^{-1}$ into a (smaller) positive exponent if desired, this shows that with probability at least $1-n^{-\alpha}$ there is no surviving component of size exceeding $R\le C\log n$ for a fixed constant $C$ depending only on $d$. In particular, by fixing the base of the logarithm so that $C=1$, we obtain the stated form: with probability $1-n^{-\alpha}$ there is no surviving connected component of size exceeding $\log n$. This completes the proof of (b).
\end{proof}

\subsubsection{Grok} 

\subsubsection{Gemini}

\begin{proof}
\textbf{Part (a)}
Let $\mathcal{S}_r$ denote the set of connected subgraphs of $G$ with exactly $r$ vertices. We seek to bound $|\mathcal{S}_r|$. Every connected subgraph $H \in \mathcal{S}_r$ contains a spanning tree $T$ with $r$ vertices and $r-1$ edges. Therefore, the number of connected subgraphs of size $r$ is bounded from above by the number of subtrees of size $r$ in $G$.

We count the number of rooted trees of size $r$ embedded in $G$. Let $v \in V(G)$ be the root. A rooted tree $T$ rooted at $v$ can be uniquely associated with a depth-first search (DFS) walk starting at $v$. This walk traverses every edge of $T$ exactly twice (once in each direction), resulting in a walk of length $2(r-1)$. Let the sequence of vertices in the walk be $w_0, w_1, \dots, w_{2r-2}$, where $w_0 = v$.

Since $G$ is $d$-regular, for any vertex $u \in V(G)$, there are exactly $d$ neighbors. Thus, at each step $i$ of the walk ($0 \le i < 2r-2$), there are at most $d$ choices for the next vertex $w_{i+1}$. The total number of such walks starting at a fixed vertex $v$ is therefore bounded by $d^{2r-2}$.

Summing over all $n$ possible choices for the root vertex $v$, the total number of rooted trees of size $r$ in $G$ is at most $n d^{2r-2}$. Since any unrooted tree of size $r$ can be rooted at any of its $r$ vertices, the number of unrooted trees (and thus the number of connected subgraphs) is at most $n d^{2r-2}$.
Since $d \ge 1$ and $r \ge 1$, we have $d^{2r-2} \le d^{2r}$. Thus, the number of connected subgraphs of size $r$ is at most $nd^{2r}$.

\textbf{Part (b)}
Let $V$ be the set of vertices of $G$. Let $V' \subseteq V$ be the set of surviving vertices. According to the problem statement, each vertex is deleted with probability $1 - \frac{1}{2d^2}$, which implies that each vertex is retained in $V'$ independently with probability $p = \frac{1}{2d^2}$.

Let $E$ be the event that there exists a surviving connected component of size exceeding $\log n$. If such a component exists, it must contain a connected subgraph of size $k = \lfloor \log n \rfloor + 1$. Let $\mathcal{S}_k$ be the set of all connected subgraphs of $G$ of size $k$. From Part (a), we have $|\mathcal{S}_k| \le n d^{2k}$.

For a fixed subgraph $H \in \mathcal{S}_k$, let $A_H$ be the event that $H$ survives, i.e., $V(H) \subseteq V'$. Since vertices are retained independently,
\[
\mathbb{P}(A_H) = p^{|V(H)|} = \left(\frac{1}{2d^2}\right)^k = \frac{1}{2^k d^{2k}}.
\]
Let $X$ be the random variable representing the number of surviving connected subgraphs of size $k$. We can write $X = \sum_{H \in \mathcal{S}_k} \mathbb{I}(A_H)$, where $\mathbb{I}$ is the indicator function. By linearity of expectation:
\[
\mathbb{E}[X] = \sum_{H \in \mathcal{S}_k} \mathbb{P}(A_H) = |\mathcal{S}_k| \cdot \frac{1}{2^k d^{2k}}.
\]
Using the bound from Part (a):
\[
\mathbb{E}[X] \le n d^{2k} \cdot \frac{1}{2^k d^{2k}} = n 2^{-k}.
\]
By Markov's inequality, the probability that there exists at least one surviving connected subgraph of size $k$ is:
\[
\mathbb{P}(X \ge 1) \le \mathbb{E}[X] \le n 2^{-k}.
\]
We require this probability to be at most $n^{-\alpha}$ for a suitable constant $\alpha$. Substituting $k > \log n$:
\[
\mathbb{P}(E) \le n 2^{-\log n}.
\]
Let the logarithm in the statement be defined with a base $b$ such that $\log_b n \ge (1+\alpha)\log_2 n$. Then $2^{-\log n} \le 2^{-(1+\alpha)\log_2 n} = n^{-(1+\alpha)}$. Consequently:
\[
\mathbb{P}(E) \le n \cdot n^{-(1+\alpha)} = n^{-\alpha}.
\]
Thus, with probability $1 - n^{-\alpha}$, there is no surviving connected component of size exceeding $\log n$.
\end{proof}

\subsubsection{Claude}

\begin{proof}
We prove both parts of the theorem.

\textbf{Part (a):} We need to show that the number of connected subgraphs of $G$ of size $r$ is at most $nd^{2r}$.

To count connected subgraphs of size $r$, we use the following approach. First, we choose a root vertex $v$ in the subgraph. There are at most $n$ choices for $v$. 

Next, we construct a connected subgraph containing $v$ by building a spanning tree of the subgraph. Starting from $v$, we sequentially add $r-1$ additional vertices to form a tree structure. At each step, when we have already selected $i$ vertices (where $1 \leq i < r$), we can add one more vertex by choosing an edge from one of the $i$ already-selected vertices to a new vertex. Since $G$ is $d$-regular, each vertex has degree at most $d$, so there are at most $d$ choices for the next vertex from any given vertex already in our structure. This gives at most $id$ choices at step $i$.

Therefore, the number of ways to construct a connected subgraph containing $v$ with $r$ vertices is at most
\[
\prod_{i=1}^{r-1} id \leq \prod_{i=1}^{r-1} rd = (rd)^{r-1} \leq r^{r-1}d^{r-1}.
\]

However, we can bound this more simply. At each of the $r-1$ steps, we choose one of the previously selected vertices (at most $r$ choices) and then one of its neighbors (at most $d$ choices). This gives at most $(rd)^{r-1}$ labeled sequences.

Alternatively, we can bound the number of connected subgraphs more directly. A connected subgraph on $r$ vertices contains at least $r-1$ edges (to form a spanning tree) and at most $\binom{r}{2}$ edges. We can construct any connected subgraph by first choosing a spanning tree, then adding additional edges. The number of spanning trees on $r$ labeled vertices is at most $r^{r-2}$ by Cayley's formula, but we can bound this more simply.

More directly, we count as follows: choose a root vertex $v$ ($n$ choices). Then perform a breadth-first or depth-first search to enumerate connected subgraphs. At each step of adding a vertex to our subgraph, we extend from one of the vertices already included. Each vertex has degree at most $d$, and we add $r-1$ vertices total. The number of such constructions is at most $d \cdot d \cdot \ldots \cdot d = d^{r-1}$ for each choice of which existing vertex to extend from at each step. Since we can extend from any of up to $r$ vertices at each step, we get at most $(rd)^{r-1} \leq r^{r}d^{r-1}$.

Using a cleaner bound: we root the connected subgraph at some vertex $v$ ($n$ choices), then construct it by adding vertices one at a time via edges. Each of the $r-1$ additions involves choosing an edge incident to the current component. Since the component has at most $r$ vertices and each vertex has degree at most $d$, there are at most $rd$ edges to choose from at each step. This gives $n \cdot (rd)^{r-1} \leq n \cdot r^{r-1}d^{r-1}$.

For $r \geq 2$, we have $r^{r-1} \leq r^r \leq (2r)^r \leq d^{r+1}$ for sufficiently large $d$ relative to $r$. More simply, $r^{r-1} \leq 2^{r-1} r^{r-1} \leq (2r)^{r-1}$, and for the bound to work, we note that $n(rd)^{r-1} \leq nd^{2r-1} \leq nd^{2r}$ when $r \geq 1$.

Therefore, the number of connected subgraphs of size $r$ is at most $nd^{2r}$.

\textbf{Part (b):} Each vertex survives with probability $p = \frac{1}{2d^2}$. We need to show that with high probability, there is no surviving connected component of size exceeding $\log n$.

Let $s = \lceil \log n \rceil + 1$. We will show that the probability of having a surviving connected component of size at least $s$ is at most $n^{-\alpha}$ for some constant $\alpha > 0$.

By part (a), the number of connected subgraphs of size $r$ in $G$ is at most $nd^{2r}$. For such a subgraph to survive in the random deletion process, all $r$ of its vertices must survive. Since vertices are deleted independently with probability $1-p$, the probability that all $r$ vertices survive is $p^r = \left(\frac{1}{2d^2}\right)^r$.

By the union bound, the probability that there exists a surviving connected component of size at least $s$ is at most
\[
\sum_{r=s}^{n} nd^{2r} \cdot \left(\frac{1}{2d^2}\right)^r = \sum_{r=s}^{n} nd^{2r} \cdot \frac{1}{(2d^2)^r} = \sum_{r=s}^{n} n \cdot \frac{d^{2r}}{2^r d^{2r}} = \sum_{r=s}^{n} \frac{n}{2^r}.
\]

This sum is bounded by
\[
\sum_{r=s}^{n} \frac{n}{2^r} \leq \frac{n}{2^s} \sum_{r=0}^{\infty} \frac{1}{2^r} = \frac{n}{2^s} \cdot 2 = \frac{2n}{2^s}.
\]

With $s = \lceil \log n \rceil + 1$, we have $2^s \geq 2^{\log n + 1} = 2n$. Therefore,
\[
\frac{2n}{2^s} \leq \frac{2n}{2n} = 1.
\]

For a better bound, taking $s = \lceil \log n \rceil + \lceil \alpha \log n \rceil$ for $\alpha > 1$, we have $2^s \geq 2^{(1+\alpha)\log n} = n^{1+\alpha}$, giving
\[
\frac{2n}{2^s} \leq \frac{2n}{n^{1+\alpha}} = \frac{2}{n^{\alpha}} \leq n^{-\alpha'}
\]
for $\alpha' = \alpha - o(1)$. Choosing $\alpha$ sufficiently large gives the desired bound with $s = O(\log n)$.
\end{proof}

\subsection{Problem 5.8}\label{sec:problem_05_08}

\subsubsection{Problem Statement}

\begin{theorem}[Lovász Local Lemma]
Let $G(V,E)$ be a dependency graph for events $\mathcal{E}_1, \ldots, \mathcal{E}_n$ in a probability space. Suppose that there exist $x_i \in [0,1]$ for $1 \leq i \leq n$ such that
\[
\Pr[\mathcal{E}_i] \leq x_i \prod_{(i,j) \in E} (1 - x_j).
\]
Then
\[
\Pr\left[\bigcap_{i=1}^n \overline{\mathcal{E}_i}\right] \geq \prod_{i=1}^n (1 - x_i).
\]
\end{theorem}

\subsubsection{GPT} 

\subsubsection{Grok} 

\begin{proof}
The proof proceeds by induction on $n$, the number of events.

For the base case $n=1$, the hypothesis gives $\Pr[\mathcal{E}_1] \leq x_1 \prod_{(1,j) \in E} (1 - x_j) = x_1$, since the product is over the empty set. Thus,
\[
\Pr[\overline{\mathcal{E}}_1] = 1 - \Pr[\mathcal{E}_1] \geq 1 - x_1 = \prod_{i=1}^1 (1 - x_i),
\]
as required.

Now assume the statement holds for any collection of $n-1$ events satisfying the hypothesis for some $x_i \in [0,1]$. Consider $n$ events $\mathcal{E}_1, \ldots, \mathcal{E}_n$ satisfying the hypothesis. Let $A = \bigcap_{i=1}^{n-1} \overline{\mathcal{E}}_i$. Then
\[
\Pr\left[\bigcap_{i=1}^n \overline{\mathcal{E}}_i\right] = \Pr[A \cap \overline{\mathcal{E}}_n] = \Pr[A] \cdot \Pr[\overline{\mathcal{E}}_n \mid A] = \Pr[A] \cdot \bigl(1 - \Pr[\mathcal{E}_n \mid A]\bigr).
\]
By the induction hypothesis applied to the first $n-1$ events, $\Pr[A] \geq \prod_{i=1}^{n-1} (1 - x_i)$. It remains to show that $\Pr[\mathcal{E}_n \mid A] \leq x_n$.

Let $D = N(n) = \{ j \in \{1, \ldots, n-1\} : (n,j) \in E \}$ be the set of neighbors of $n$ in the dependency graph, and let $C = \bigcap_{j \in D} \overline{\mathcal{E}}_j$. The events $\mathcal{E}_j$ for $j \notin D \cup \{n\}$ are independent of the $\sigma$-algebra generated by $\{\mathcal{E}_n\} \cup \{\mathcal{E}_j : j \in D\}$. Thus, $A = C \cap B$ where $B = \bigcap_{i=1, i \notin D}^{n-1} \overline{\mathcal{E}}_i$ is independent of $\sigma(\{\mathcal{E}_n\} \cup \{\mathcal{E}_j : j \in D\})$. It follows that
\[
\Pr[\mathcal{E}_n \mid A] = \Pr[\mathcal{E}_n \mid C].
\]
Now,
\[
\Pr[\mathcal{E}_n \mid C] = \frac{\Pr[\mathcal{E}_n \cap C]}{\Pr[C]} \leq \frac{\Pr[\mathcal{E}_n]}{\Pr[C]},
\]
since $\mathcal{E}_n \cap C \subseteq \mathcal{E}_n$.

To apply the induction hypothesis to the events $\{\mathcal{E}_j : j \in D\}$, observe that $|D| < n$. For each $k \in D$, the induced dependency graph on $D$ has neighbor set $N_D(k) = N(k) \cap D \subseteq N(k)$. Thus,
\[
\prod_{j \in N_D(k)} (1 - x_j) \geq \prod_{j \in N(k)} (1 - x_j),
\]
since each factor $1 - x_j \in [0,1]$ and the product over a subset has fewer (or equal) factors. The hypothesis for the full set therefore implies
\[
\Pr[\mathcal{E}_k] \leq x_k \prod_{j \in N(k)} (1 - x_j) \leq x_k \prod_{j \in N_D(k)} (1 - x_j)
\]
for each $k \in D$. By the induction hypothesis,
\[
\Pr[C] = \Pr\left[\bigcap_{j \in D} \overline{\mathcal{E}}_j\right] \geq \prod_{j \in D} (1 - x_j).
\]
The hypothesis for $\mathcal{E}_n$ gives $\Pr[\mathcal{E}_n] \leq x_n \prod_{j \in D} (1 - x_j)$. Therefore,
\[
\Pr[\mathcal{E}_n \mid C] \leq \frac{x_n \prod_{j \in D} (1 - x_j)}{\prod_{j \in D} (1 - x_j)} = x_n.
\]
Hence, $\Pr[\mathcal{E}_n \mid A] \leq x_n$, so
\[
\Pr[\overline{\mathcal{E}}_n \mid A] \geq 1 - x_n.
\]
Combining the bounds yields
\[
\Pr\left[\bigcap_{i=1}^n \overline{\mathcal{E}}_i\right] \geq \prod_{i=1}^{n-1} (1 - x_i) \cdot (1 - x_n) = \prod_{i=1}^n (1 - x_i).
\]
By induction, the lemma holds for all $n$.
\end{proof}

\subsubsection{Gemini}

\subsubsection{Claude}

\begin{proof}
We prove this by induction on $n$, showing that for any subset $S \subseteq \{1, \ldots, n\}$,
\[
\Pr\left[\bigcap_{i \in S} \overline{\mathcal{E}_i}\right] \geq \prod_{i \in S} (1 - x_i).
\]

The base case $S = \emptyset$ is trivial since $\Pr[\Omega] = 1 = \prod_{i \in \emptyset}(1-x_i)$.

For the inductive step, let $S \subseteq \{1, \ldots, n\}$ be nonempty and assume the statement holds for all proper subsets of $S$. Fix $k \in S$ and let $T = S \setminus \{k\}$. We need to show
\[
\Pr\left[\overline{\mathcal{E}_k} \cap \bigcap_{i \in T} \overline{\mathcal{E}_i}\right] \geq (1 - x_k) \prod_{i \in T} (1 - x_i).
\]

By the multiplication rule for conditional probability,
\[
\Pr\left[\overline{\mathcal{E}_k} \cap \bigcap_{i \in T} \overline{\mathcal{E}_i}\right] = \Pr\left[\bigcap_{i \in T} \overline{\mathcal{E}_i}\right] \cdot \Pr\left[\overline{\mathcal{E}_k} \,\bigg|\, \bigcap_{i \in T} \overline{\mathcal{E}_i}\right].
\]

By the inductive hypothesis applied to $T$,
\[
\Pr\left[\bigcap_{i \in T} \overline{\mathcal{E}_i}\right] \geq \prod_{i \in T} (1 - x_i).
\]

It remains to show that
\[
\Pr\left[\overline{\mathcal{E}_k} \,\bigg|\, \bigcap_{i \in T} \overline{\mathcal{E}_i}\right] \geq 1 - x_k.
\]

Let $N(k) = \{j : (k,j) \in E\}$ denote the neighbors of $k$ in the dependency graph $G$. By the definition of the dependency graph, $\mathcal{E}_k$ is independent of the events $\{\mathcal{E}_i : i \in T \setminus N(k)\}$. Therefore,
\[
\Pr\left[\mathcal{E}_k \,\bigg|\, \bigcap_{i \in T} \overline{\mathcal{E}_i}\right] = \Pr\left[\mathcal{E}_k \,\bigg|\, \bigcap_{j \in T \cap N(k)} \overline{\mathcal{E}_j}\right].
\]

For any $j \in N(k)$, define the event $A_j = \bigcap_{\ell \in T \cap N(k), \ell \neq j} \overline{\mathcal{E}_\ell}$. Then
\[
\Pr\left[\mathcal{E}_k \,\bigg|\, \bigcap_{j \in T \cap N(k)} \overline{\mathcal{E}_j}\right] \leq \frac{\Pr[\mathcal{E}_k]}{\Pr\left[\bigcap_{j \in T \cap N(k)} \overline{\mathcal{E}_j} \,\big|\, \mathcal{E}_k\right]}.
\]

However, we use a different approach. By the law of total probability and properties of conditional expectation, we have
\[
\Pr[\mathcal{E}_k] \geq \Pr\left[\mathcal{E}_k \cap \bigcap_{j \in T \cap N(k)} \overline{\mathcal{E}_j}\right].
\]

Using the multiplication rule,
\[
\Pr\left[\mathcal{E}_k \cap \bigcap_{j \in T \cap N(k)} \overline{\mathcal{E}_j}\right] = \Pr\left[\bigcap_{j \in T \cap N(k)} \overline{\mathcal{E}_j}\right] \cdot \Pr\left[\mathcal{E}_k \,\bigg|\, \bigcap_{j \in T \cap N(k)} \overline{\mathcal{E}_j}\right].
\]

By the inductive hypothesis on $T \cap N(k) \subseteq T$,
\[
\Pr\left[\bigcap_{j \in T \cap N(k)} \overline{\mathcal{E}_j}\right] \geq \prod_{j \in T \cap N(k)} (1 - x_j).
\]

Thus,
\[
\Pr\left[\mathcal{E}_k \,\bigg|\, \bigcap_{j \in T \cap N(k)} \overline{\mathcal{E}_j}\right] \leq \frac{\Pr[\mathcal{E}_k]}{\prod_{j \in T \cap N(k)} (1 - x_j)}.
\]

By the hypothesis of the lemma,
\[
\Pr[\mathcal{E}_i] \leq x_i \prod_{(i,j) \in E} (1 - x_j),
\]
so for $i = k$,
\[
\Pr[\mathcal{E}_k] \leq x_k \prod_{j \in N(k)} (1 - x_j).
\]

Therefore,
\[
\Pr\left[\mathcal{E}_k \,\bigg|\, \bigcap_{j \in T \cap N(k)} \overline{\mathcal{E}_j}\right] \leq \frac{x_k \prod_{j \in N(k)} (1 - x_j)}{\prod_{j \in T \cap N(k)} (1 - x_j)} = x_k \prod_{j \in N(k) \setminus T} (1 - x_j) \leq x_k.
\]

Hence,
\[
\Pr\left[\overline{\mathcal{E}_k} \,\bigg|\, \bigcap_{i \in T} \overline{\mathcal{E}_i}\right] \geq 1 - x_k.
\]

Combining these inequalities completes the inductive step and the proof.
\end{proof}

\subsection{Problem 5.9}\label{sec:problem_05_09}

\subsubsection{Problem Statement}

\begin{theorem}
Let $\mathcal{E}_1, \ldots, \mathcal{E}_n$ be events in a probability space, with $\Pr[\mathcal{E}_i] \leq p$ for all $i$. If each event is mutually independent of all other events except for at most $d$, and if $ep(d+1) \leq 1$, then $\Pr[\bigcap_{i=1}^n \overline{\mathcal{E}_i}] > 0$.
\end{theorem}

\subsubsection{GPT} 

\subsubsection{Grok} 

\begin{proof}
To prove the corollary, we first establish a more general asymmetric version of the lemma, from which the symmetric case follows immediately.

\begin{lemma}[Asymmetric Lov\'{a}sz Local Lemma]
Let $\mathcal{E}_1, \dots, \mathcal{E}_n$ be events in a probability space such that there exists an assignment of real numbers $x_i \in [0,1)$ for $i=1,\dots,n$ satisfying
\[
\Pr[\mathcal{E}_i] \leq x_i \prod_{j: \mathcal{E}_j \text{ depends on } \mathcal{E}_i} (1 - x_j)
\]
for each $i=1,\dots,n$, where the product is over all $j \neq i$ such that $\mathcal{E}_i$ and $\mathcal{E}_j$ are not mutually independent. Then
\[
\Pr\left[ \bigcap_{i=1}^n \overline{\mathcal{E}_i} \right] > 0.
\]
\end{lemma}

\begin{proof}[Proof of the lemma]
We proceed by induction on the size of subsets $S \subseteq \{1,\dots,n\}$ to prove that for every $i \notin S$,
\[
\Pr\left[ \mathcal{E}_i \;\middle|\; \bigcap_{j \in S} \overline{\mathcal{E}_j} \right] \leq x_i.
\]
This immediately implies
\[
\Pr\left[ \overline{\mathcal{E}_i} \;\middle|\; \bigcap_{j \in S} \overline{\mathcal{E}_j} \right] \geq 1 - x_i.
\]

For the base case $S = \emptyset$, the claim reduces to $\Pr[\mathcal{E}_i] \leq x_i$, which holds by the assumption on the $x_k$ (taking the empty product to be $1$).

Now suppose the claim holds for all subsets of size less than $|S|$, and fix $S$ with $|S| = m \geq 1$ and $i \notin S$. Partition $S = S_1 \cup S_2$, where $S_1$ consists of those indices in $S$ such that $\mathcal{E}_i$ depends on $\mathcal{E}_j$ for $j \in S_1$, and $S_2 = S \setminus S_1$. By the definition of dependence, $\mathcal{E}_i$ is mutually independent of the events $\{\mathcal{E}_j : j \in S_2\}$. Thus,
\[
\Pr\left[ \mathcal{E}_i \;\middle|\; \bigcap_{j \in S_2} \overline{\mathcal{E}_j} \right] = \Pr[\mathcal{E}_i].
\]
By Bayes' rule,
\[
\Pr\left[ \mathcal{E}_i \;\middle|\; \bigcap_{j \in S} \overline{\mathcal{E}_j} \right] = \frac{\Pr\left[ \mathcal{E}_i \cap \bigcap_{j \in S_1} \overline{\mathcal{E}_j} \;\middle|\; \bigwedge_{j \in S_2} \overline{\mathcal{E}_j} \right]}{\Pr\left[ \bigcap_{j \in S_1} \overline{\mathcal{E}_j} \;\middle|\; \bigwedge_{j \in S_2} \overline{\mathcal{E}_j} \right]}.
\]
For the numerator, independence of $\mathcal{E}_i$ from the events in $S_2$ and the assumption on $x_i$ yield
\[
\Pr\left[ \mathcal{E}_i \cap \bigcap_{j \in S_1} \overline{\mathcal{E}_j} \;\middle|\; \bigwedge_{j \in S_2} \overline{\mathcal{E}_j} \right] \leq \Pr[\mathcal{E}_i] \leq x_i \prod_{k: \mathcal{E}_k \text{ depends on } \mathcal{E}_i} (1 - x_k).
\]
For the denominator, enumerate $S_1 = \{j_1, \dots, j_l\}$. Then
\[
\Pr\left[ \bigcap_{t=1}^l \overline{\mathcal{E}}_{j_t} \;\middle|\; \bigwedge_{j \in S_2} \overline{\mathcal{E}_j} \right] = \prod_{t=1}^l \Pr\left[ \overline{\mathcal{E}}_{j_t} \;\middle|\; \bigwedge_{u=t+1}^l \overline{\mathcal{E}}_{j_u} \wedge \bigwedge_{j \in S_2} \overline{\mathcal{E}_j} \right].
\]
Each conditioning set in the product has size strictly less than $|S|$, so the inductive hypothesis applies to yield
\[
\Pr\left[ \overline{\mathcal{E}}_{j_t} \;\middle|\; \cdots \right] \geq 1 - x_{j_t}
\]
for each $t$. Thus,
\[
\Pr\left[ \bigcap_{j \in S_1} \overline{\mathcal{E}_j} \;\middle|\; \bigwedge_{j \in S_2} \overline{\mathcal{E}_j} \right] \geq \prod_{j \in S_1} (1 - x_j).
\]
Combining these bounds gives
\[
\Pr\left[ \mathcal{E}_i \;\middle|\; \bigcap_{j \in S} \overline{\mathcal{E}_j} \right] \leq \frac{x_i \prod_{k: \mathcal{E}_k \text{ depends on } \mathcal{E}_i} (1 - x_k)}{\prod_{j \in S_1} (1 - x_j)} = x_i \prod_{\substack{k: \mathcal{E}_k \text{ depends on } \mathcal{E}_i \\ k \notin S_1}} (1 - x_k) \leq x_i,
\]
since each factor $1 - x_k \leq 1$. This completes the induction.

To derive the global probability, fix an arbitrary ordering of the events, say $1, \dots, n$. Then
\[
\Pr\left[ \bigcap_{i=1}^n \overline{\mathcal{E}_i} \right] = \prod_{i=1}^n \Pr\left[ \overline{\mathcal{E}_i} \;\middle|\; \bigcap_{k=i+1}^n \overline{\mathcal{E}_k} \right] \geq \prod_{i=1}^n (1 - x_i) > 0,
\]
since each $x_i < 1$.
\end{proof}

It remains to verify that the conditions of the corollary imply the hypothesis of the lemma. Set $x_i = \frac{1}{d+1}$ for all $i=1,\dots,n$. Then, since each $\mathcal{E}_i$ depends on at most $d$ other events, the right-hand side of the hypothesis is
\[
\frac{1}{d+1} \prod_{j: \mathcal{E}_j \text{ depends on } \mathcal{E}_i} \left(1 - \frac{1}{d+1}\right) = \frac{1}{d+1} \left( \frac{d}{d+1} \right)^d.
\]
The hypothesis holds if
\[
p \leq \frac{1}{d+1} \left( \frac{d}{d+1} \right)^d.
\]
By the given condition $ep(d+1) \leq 1$, we have $p \leq \frac{1}{e(d+1)}$. It therefore suffices to show that
\[
\frac{1}{e(d+1)} \leq \frac{1}{d+1} \left( \frac{d}{d+1} \right)^d,
\]
or equivalently,
\[
\left( \frac{d}{d+1} \right)^d \geq \frac{1}{e}.
\]
This inequality holds for all integers $d \geq 0$, since the function $f(x) = \left(1 - \frac{1}{x+1}\right)^x$ is increasing in $x$ and $\lim_{x \to \infty} f(x) = e^{-1}$, with $f(0) = 1 > e^{-1}$. Thus, the lemma applies, and $\Pr\left[ \bigcap_{i=1}^n \overline{\mathcal{E}_i} \right] > 0$.
\end{proof}

\subsubsection{Gemini}

\subsubsection{Claude}

\begin{proof}
We proceed by applying the Lovász Local Lemma. Let $\mathcal{E}_1, \ldots, \mathcal{E}_n$ be events in a probability space such that $\Pr[\mathcal{E}_i] \leq p$ for all $i \in \{1, \ldots, n\}$. Assume that each event $\mathcal{E}_i$ is mutually independent of all other events except for at most $d$ events.

We construct a dependency graph $G = (V, E)$ where $V = \{1, \ldots, n\}$ and $(i, j) \in E$ if and only if $\mathcal{E}_i$ and $\mathcal{E}_j$ are not mutually independent. By hypothesis, each vertex $i$ has degree at most $d$ in this dependency graph.

To apply the symmetric version of the Lovász Local Lemma, we need to verify that there exists $x \in (0, 1)$ such that for all $i \in \{1, \ldots, n\}$, we have
\[
\Pr[\mathcal{E}_i] \leq x \prod_{j \sim i} (1 - x),
\]
where $j \sim i$ denotes that $j$ is a neighbor of $i$ in the dependency graph.

Since each event has at most $d$ neighbors, we have $\prod_{j \sim i} (1 - x) \geq (1 - x)^d$ for all $i$. Thus, it suffices to find $x \in (0, 1)$ such that
\[
p \leq x(1 - x)^d.
\]

Consider $x = \frac{1}{d+1}$. We compute
\[
x(1 - x)^d = \frac{1}{d+1} \left(1 - \frac{1}{d+1}\right)^d = \frac{1}{d+1} \left(\frac{d}{d+1}\right)^d.
\]

Using the inequality $(1 - \frac{1}{d+1})^d = \left(\frac{d}{d+1}\right)^d \geq e^{-1}$, which follows from the standard inequality $(1 - \frac{1}{k})^k \geq \frac{1}{e}$ for all positive integers $k$, we obtain
\[
x(1 - x)^d \geq \frac{1}{d+1} \cdot \frac{1}{e} = \frac{1}{e(d+1)}.
\]

Given that $ep(d+1) \leq 1$, we have $p \leq \frac{1}{e(d+1)}$. Therefore,
\[
p \leq \frac{1}{e(d+1)} \leq x(1 - x)^d,
\]
which verifies the condition for the Lovász Local Lemma with $x = \frac{1}{d+1}$.

By the Lovász Local Lemma, we conclude that
\[
\Pr\left[\bigcap_{i=1}^n \overline{\mathcal{E}_i}\right] \geq \prod_{i=1}^n (1 - x) = \left(1 - \frac{1}{d+1}\right)^n = \left(\frac{d}{d+1}\right)^n > 0.
\]

Therefore, $\Pr[\bigcap_{i=1}^n \overline{\mathcal{E}_i}] > 0$, as required.
\end{proof}

\subsection{Problem 5.10}\label{sec:problem_05_10}

\subsubsection{Problem Statement}

\begin{theorem}[J. Naor]
For a graph $G(V, E)$, and any $T \subseteq V$, define the cut function $c(T)$ as the number of edges in $E$ which have exactly one end-point in $T$. For a suitably small function $f(n)$ and large enough even integer $n$, there exists a graph $G(V, E)$ with $|V| = n$ such that for every subset $T \subseteq V$ of size $n/2$,
\[
\left| c(T) - \frac{n^2}{8} \right| \leq f(n).
\]
How small can you make the function $f(n)$?
\end{theorem}

\subsubsection{GPT} 

\begin{proof}
Fix an even integer $n$ and consider the Erd\H{o}s--R\'enyi random graph $G \sim \mathcal{G}(n,1/2)$ on the vertex set $V=[n]$, where each edge is included independently with probability $1/2$. For any $T \subseteq V$ with $|T|=n/2$, the number of potential edges across the cut $(T,V\setminus T)$ is $m:=|T||V\setminus T|=(n/2)(n/2)=n^2/4$. By definition, $c(T)$ is the number of realized edges across this cut, hence $c(T)\sim \mathrm{Bin}(m,1/2)$ and $\mathbb{E}[c(T)]=m/2=n^2/8$.

Let $t>0$. By Hoeffding's inequality for sums of independent $\{0,1\}$ variables, for each fixed $T$,
\[
\Pr\!\left(\left|c(T)-\frac{m}{2}\right|\ge t\right)\le 2\exp\!\left(-\frac{2t^{2}}{m}\right)=2\exp\!\left(-\frac{8t^{2}}{n^{2}}\right).
\]
There are $\binom{n}{n/2}$ subsets $T$ of size $n/2$. By the union bound, the probability that there exists a bisection $T$ with $|c(T)-m/2|\ge t$ satisfies
\[
\Pr\!\left(\exists\, T\subseteq V,\ |T|=\frac{n}{2}:\ \left|c(T)-\frac{n^{2}}{8}\right|\ge t\right)
\le \binom{n}{n/2}\cdot 2\exp\!\left(-\frac{8t^{2}}{n^{2}}\right).
\]
Using the crude bound $\binom{n}{n/2}\le 2^{n}$, we obtain
\[
\Pr\!\left(\exists\, T:\ \left|c(T)-\frac{n^{2}}{8}\right|\ge t\right)
\le 2\exp\!\left(n\ln 2-\frac{8t^{2}}{n^{2}}\right).
\]
Choose any constant $\alpha>\sqrt{\ln 2/8}$ and set $t=\alpha n^{3/2}$. Then
\[
n\ln 2-\frac{8t^{2}}{n^{2}}
= n\ln 2 - 8\alpha^{2}n
= -\left(8\alpha^{2}-\ln 2\right)n,
\]
which is strictly negative. Hence,
\[
\Pr\!\left(\forall\, T\subseteq V,\ |T|=\frac{n}{2}:\ \left|c(T)-\frac{n^{2}}{8}\right|< \alpha n^{3/2}\right)
\ge 1-2\exp\!\left(-\left(8\alpha^{2}-\ln 2\right)n\right).
\]
The right-hand side tends to $1$ as $n\to\infty$. In particular, for all sufficiently large even $n$, there exists a choice of $G$ for which the displayed event holds simultaneously for every $T$ of size $n/2$. Defining $f(n):=\alpha n^{3/2}$, we conclude that for such a graph $G$,
\[
\left|c(T)-\frac{n^{2}}{8}\right|\le f(n)\quad\text{for all }T\subseteq V,\ |T|=\frac{n}{2}.
\]
Since $\alpha> \sqrt{\ln 2/8}$ was arbitrary, $f(n)$ can be taken to be any function of the form $\alpha n^{3/2}$ with this property. This establishes the theorem.
\end{proof}

\subsubsection{Grok} 

\subsubsection{Gemini}

\begin{proof}
We employ the probabilistic method to establish the existence of such a graph and determine the asymptotic behavior of $f(n)$. Let $G(V, E)$ be a random graph drawn from the Erd\H{o}s-R\'enyi distribution $G(n, 1/2)$, where $|V|=n$ and each pair of distinct vertices is connected by an edge independently with probability $p = 1/2$.

Let $T \subseteq V$ be a fixed subset of vertices with $|T| = n/2$. The size of the cut defined by $T$, denoted by $c(T)$, is the number of edges with exactly one endpoint in $T$ and the other in $V \setminus T$. The total number of possible edges crossing this cut is given by:
\[
N = |T| \cdot |V \setminus T| = \frac{n}{2} \cdot \frac{n}{2} = \frac{n^2}{4}.
\]
Let $X_T$ be the random variable representing $c(T)$. We can write $X_T$ as a sum of $N$ independent Bernoulli random variables $Y_e$, where $Y_e = 1$ if the edge $e$ exists and $0$ otherwise, for each potential edge $e$ crossing the cut. The expected value of $X_T$ is:
\[
\mathbb{E}[X_T] = \sum_{e} \mathbb{E}[Y_e] = N \cdot \frac{1}{2} = \frac{n^2}{8}.
\]
Since $X_T$ is the sum of independent random variables bounded by $[0, 1]$, we apply Hoeffding's Inequality. For any $\lambda > 0$:
\[
\mathbb{P}\left( \left| X_T - \mathbb{E}[X_T] \right| \geq \lambda \right) \leq 2 \exp\left( -\frac{2\lambda^2}{N} \right) = 2 \exp\left( -\frac{2\lambda^2}{n^2/4} \right) = 2 \exp\left( -\frac{8\lambda^2}{n^2} \right).
\]
We seek to show that with positive probability, the condition $|c(T) - n^2/8| \leq f(n)$ holds for \emph{all} subsets $T$ of size $n/2$. Let $\mathcal{T}$ be the collection of all subsets of $V$ of size $n/2$. The size of this collection is:
\[
|\mathcal{T}| = \binom{n}{n/2}.
\]
Let $A_T$ be the event that $\left| c(T) - \frac{n^2}{8} \right| > f(n)$. We apply the union bound over all $T \in \mathcal{T}$:
\[
\mathbb{P}\left( \exists T \in \mathcal{T} : \left| c(T) - \frac{n^2}{8} \right| > f(n) \right) \leq \sum_{T \in \mathcal{T}} \mathbb{P}(A_T) \leq \binom{n}{n/2} \cdot 2 \exp\left( -\frac{8 f(n)^2}{n^2} \right).
\]
Using the bound $\binom{n}{n/2} < 2^n$, we have:
\[
\mathbb{P}\left( \exists T : \left| c(T) - \frac{n^2}{8} \right| > f(n) \right) < 2^{n+1} \exp\left( -\frac{8 f(n)^2}{n^2} \right).
\]
To prove the existence of a graph satisfying the theorem, it suffices to show that this probability is strictly less than 1. We set the right-hand side to be less than 1 and solve for $f(n)$:
\[
2^{n+1} \exp\left( -\frac{8 f(n)^2}{n^2} \right) < 1 \iff (n+1) \ln 2 < \frac{8 f(n)^2}{n^2}.
\]
Rearranging for $f(n)$:
\[
f(n)^2 > \frac{\ln 2}{8} n^2 (n+1).
\]
For large $n$, we may choose $f(n)$ such that:
\[
f(n) \approx \sqrt{\frac{\ln 2}{8}} n^{3/2}.
\]
Specifically, setting $f(n) = \sqrt{\frac{\ln 2}{8}} n^{3/2} (1+\epsilon)$ for any $\epsilon > 0$ ensures the probability of a "bad" graph is strictly less than 1. Thus, there exists a graph where the deviation of any cut of size $n/2$ from its expected value is bounded by $O(n^{3/2})$.

Therefore, one can make the function $f(n)$ as small as $C n^{3/2}$ for a constant $C = \sqrt{\frac{\ln 2}{8}}$.
\end{proof}

\subsubsection{Claude}

\begin{proof}
We establish that $f(n) = O(n^{3/2})$ is achievable, which is the best known bound for this problem.

\textbf{Construction.} We use a random graph construction. Let $G(n, 1/2)$ denote the Erdős-Rényi random graph on $n$ vertices where each edge appears independently with probability $1/2$.

\textbf{Expected value calculation.} For any fixed subset $T \subseteq V$ with $|T| = n/2$, the cut function $c(T)$ counts edges with exactly one endpoint in $T$. The number of such potential edges is $|T| \cdot |V \setminus T| = (n/2) \cdot (n/2) = n^2/4$.

Since each edge appears independently with probability $1/2$, we have
\[
\mathbb{E}[c(T)] = \frac{n^2}{4} \cdot \frac{1}{2} = \frac{n^2}{8}.
\]

\textbf{Variance analysis.} For a fixed set $T$ with $|T| = n/2$, let $X_{ij}$ be the indicator random variable for edge $\{i,j\}$ being in the cut, where $i \in T$ and $j \in V \setminus T$. Then $c(T) = \sum_{i \in T, j \in V \setminus T} X_{ij}$.

Since the $X_{ij}$ are independent Bernoulli$(1/2)$ random variables, we have
\[
\text{Var}(c(T)) = \sum_{i \in T, j \in V \setminus T} \text{Var}(X_{ij}) = \frac{n^2}{4} \cdot \frac{1}{4} = \frac{n^2}{16}.
\]

Therefore, $\text{Var}(c(T)) = O(n^2)$ and the standard deviation is $\sigma(c(T)) = O(n)$.

\textbf{Concentration bound.} By Chebyshev's inequality, for any $t > 0$,
\[
\mathbb{P}\left(\left|c(T) - \frac{n^2}{8}\right| > t\right) \leq \frac{\text{Var}(c(T))}{t^2} = \frac{n^2/16}{t^2}.
\]

Setting $t = Cn^{3/2}$ for a sufficiently large constant $C$, we get
\[
\mathbb{P}\left(\left|c(T) - \frac{n^2}{8}\right| > Cn^{3/2}\right) \leq \frac{n^2/16}{C^2 n^3} = \frac{1}{16C^2 n}.
\]

\textbf{Union bound.} The number of subsets $T \subseteq V$ with $|T| = n/2$ is $\binom{n}{n/2}$. By Stirling's approximation, $\binom{n}{n/2} \leq 2^n$ for all $n$.

By the union bound, the probability that there exists a subset $T$ of size $n/2$ with $\left|c(T) - \frac{n^2}{8}\right| > Cn^{3/2}$ is at most
\[
\binom{n}{n/2} \cdot \frac{1}{16C^2 n} \leq \frac{2^n}{16C^2 n}.
\]

For sufficiently large $C$ (say $C = \sqrt{2^n/(8n)}$), this probability becomes less than 1, which means there exists a graph $G$ in which every subset $T$ of size $n/2$ satisfies $\left|c(T) - \frac{n^2}{8}\right| \leq Cn^{3/2}$.

However, we can obtain a better bound using more refined concentration inequalities. By applying Hoeffding's inequality or Chernoff bounds, we can show that with positive probability, all cuts deviate by at most $O(n^{3/2} \sqrt{\log n})$ from the expected value $n^2/8$.

\textbf{Conclusion.} Therefore, there exists a graph $G(V,E)$ with $|V| = n$ such that for every subset $T \subseteq V$ of size $n/2$, we have
\[
\left|c(T) - \frac{n^2}{8}\right| \leq f(n)
\]
where $f(n) = O(n^{3/2})$. This bound is essentially tight, as it is known from extremal combinatorics that $f(n) = \Omega(n^{3/2})$ is necessary.
\end{proof}

\subsection{Problem 5.11}\label{sec:problem_05_11}

\subsubsection{Problem Statement}

\begin{theorem}
In this problem, we complete establishing the properties of $\hat{P}(a)$ leading to Theorem 5.15.

(a) For a node $a$ at the $i$th level of the computation tree, $\hat{P}(a)$ is of the form $N(a)/2^{n-i}$, where $N(a)$ is a sum of binomial coefficients. For any node $a$ with children $c$ and $d$,
\[
\min\{\hat{P}(c), \hat{P}(d)\} \leq \hat{P}(a),
\]
and for any node $a$, we can compute $\hat{P}(a)$ in time polynomial in $n$.

(b) Give an upper bound on the running time of the deterministic algorithm.
\end{theorem}

\subsubsection{GPT} 

\begin{proof}
Let the computation tree be the full binary tree of depth $n$, where a node at level $i$ corresponds to a partial assignment $x_{1:i}\in\{0,1\}^{i}$ to the first $i$ random bits, and its two children are obtained by fixing the $(i{+}1)$-st bit to $0$ (left child $c$) or $1$ (right child $d$). Let $\Phi:\{0,1\}^{n}\to\{0,1\}$ be the Boolean indicator of the property relevant to the underlying randomized construction that is derandomized in Theorem~5.15, and define, for a node $a$ at level $i$ with prefix $x_{1:i}$,
\[
\hat P(a)\;:=\;\mathbb{E}\!\left[\Phi(X)\,\middle|\,X_{1:i}=x_{1:i}\right]
\;=\;\frac{1}{2^{\,n-i}}\sum_{u\in\{0,1\}^{n-i}}\Phi\!\bigl(x_{1:i}\,\Vert\,u\bigr),
\]
where $\Vert$ denotes concatenation and $X$ is uniformly distributed on $\{0,1\}^{n}$.

We first prove the representation claim in (a). Fix a node $a$ at level $i$ and write $m:=n-i$. For $j\in\{0,\dots,m\}$ let
\[
S_{j}(a)\;:=\;\bigl\{u\in\{0,1\}^{m}:\;|u|=j\bigr\},\qquad N_{j}(a)\;:=\;\sum_{u\in S_{j}(a)}\Phi\!\bigl(x_{1:i}\,\Vert\,u\bigr),
\]
where $|u|$ denotes the Hamming weight of $u$. Then
\[
\sum_{u\in\{0,1\}^{m}}\Phi\!\bigl(x_{1:i}\,\Vert\,u\bigr)\;=\;\sum_{j=0}^{m}\; \sum_{u\in S_{j}(a)}\Phi\!\bigl(x_{1:i}\,\Vert\,u\bigr)\;=\;\sum_{j=0}^{m} N_{j}(a).
\]
By definition $0\le N_{j}(a)\le |S_{j}(a)|=\binom{m}{j}$, and, crucially, $N_{j}(a)$ depends only on the structure of $\Phi$ restricted to strings of weight $j$ in the last $m$ coordinates with the prefix fixed to $x_{1:i}$. In particular, whenever $\Phi$ restricted to the suffix depends only on the Hamming weight (as is the case for estimators obtained from counting or thresholding arguments that are invariant under permutations of the remaining coordinates), there exists a subset $J(a)\subseteq\{0,\dots,m\}$ such that $\Phi(x_{1:i}\,\Vert\,u)$ equals $1$ if and only if $|u|\in J(a)$. In that case we have the exact count
\[
\sum_{u\in\{0,1\}^{m}}\Phi\!\bigl(x_{1:i}\,\Vert\,u\bigr)\;=\;\sum_{j\in J(a)} \binom{m}{j}.
\]
More generally, if $\Phi$ is a finite Boolean combination of a constant number of weight-threshold predicates on the last $m$ coordinates (which covers the usual estimators used en route to Theorem~5.15), inclusion--exclusion yields an expression of the form
\[
\sum_{u\in\{0,1\}^{m}}\Phi\!\bigl(x_{1:i}\,\Vert\,u\bigr)\;=\;\sum_{t=1}^{T(a)} \alpha_{t}(a)\,\binom{m}{r_{t}(a)}
\]
for some $T(a)\le \mathrm{poly}(n)$, integers $\alpha_{t}(a)$, and integers $r_{t}(a)\in\{0,\dots,m\}$. In either case we obtain
\[
\hat P(a)\;=\;\frac{N(a)}{2^{\,m}}\;=\;\frac{N(a)}{2^{\,n-i}}\qquad\text{with}\qquad N(a)=\sum_{t=1}^{T(a)} \alpha_{t}(a)\,\binom{n-i}{r_{t}(a)},
\]
i.e., $N(a)$ is a (finite) sum of binomial coefficients with integer weights. This proves the first assertion in (a).

Next, let $a$ have children $c$ and $d$ corresponding to fixing $X_{i+1}=0$ and $X_{i+1}=1$, respectively. By the law of total expectation and the fact that $X_{i+1}$ is unbiased and independent of $X_{i+2:n}$,
\[
\hat P(a)\;=\;\mathbb{E}\!\left[\Phi(X)\,\middle|\,X_{1:i}=x_{1:i}\right]\;=\;\tfrac12\,\mathbb{E}\!\left[\Phi(X)\,\middle|\,X_{1:i+1}=x_{1:i}0\right]\;+\;\tfrac12\,\mathbb{E}\!\left[\Phi(X)\,\middle|\,X_{1:i+1}=x_{1:i}1\right]
\]
\[
=\;\tfrac12\bigl(\hat P(c)+\hat P(d)\bigr).
\]
Therefore $\min\{\hat P(c),\hat P(d)\}\le \hat P(a)\le \max\{\hat P(c),\hat P(d)\}$, and in particular $\min\{\hat P(c),\hat P(d)\}\le \hat P(a)$, as required.

For the computability claim in (a), precompute the table of all binomial coefficients $\binom{m}{r}$ for $0\le m\le n$ and $0\le r\le m$ using Pascal's recurrence in time $O(n^{2})$ and space $O(n^{2})$. The description of $N(a)$ involves at most $T(a)\le \mathrm{poly}(n)$ terms, and thus the value of $\hat P(a)=N(a)/2^{n-i}$ can be evaluated with $T(a)$ table lookups and $O(T(a))$ arithmetic operations, i.e., in time polynomial in $n$. Hence, for any node $a$, $\hat P(a)$ is computable in time $\mathrm{poly}(n)$.

We now prove (b), the upper bound on the running time of the deterministic algorithm obtained by the method of conditional expectations. Starting at the root, at level $i$ we evaluate $\hat P(c)$ and $\hat P(d)$ for the two children of the current node and proceed to the child minimizing $\hat P(\cdot)$. By the averaging identity established above, this choice guarantees that the value never increases along the path and therefore preserves the invariant needed by the derandomization. The algorithm performs exactly $n$ steps and at each step requires the computation of two values of the form $\hat P(\cdot)$. With the precomputed Pascal table, each evaluation takes $O(T(\cdot))\le \mathrm{poly}(n)$ time. Consequently, the total running time is
\[
O\bigl(n\cdot \mathrm{poly}(n)\bigr)\;+\;O(n^{2})\quad\text{(for the one-time precomputation of binomial coefficients)}\;=\;\mathrm{poly}(n).
\]
Under the concrete representation $N(a)=\sum_{t=1}^{T(a)}\alpha_{t}(a)\binom{n-i}{r_{t}(a)}$ with $T(a)=O(n)$, each node evaluation costs $O(n)$ time, and the overall running time is $O(n^{2})$. This proves the asserted polynomial upper bound for the deterministic algorithm.
\end{proof}

\subsubsection{Grok} 

\subsubsection{Gemini}

\begin{proof}
We prove the two parts of the theorem sequentially.

\textbf{Part (a):}
Let $a$ be a node at level $i$ of the computation tree, where $0 \leq i \leq n$. The subtree rooted at $a$ represents the set of all possible extensions of the partial assignment (or path) defined by $a$ to a full leaf node at depth $n$. The depth of this subtree is $n-i$. Consequently, there are $2^{n-i}$ leaf nodes in the subtree rooted at $a$.

Let $S$ denote the set of leaf nodes in the computation tree that satisfy the search criteria (e.g., satisfying assignments or successful walks). We define $\hat{P}(a)$ as the conditional probability that a uniform random walk starting at $a$ reaches a leaf in $S$. This is equivalent to the fraction of leaves in the subtree rooted at $a$ that belong to $S$. Let $N(a)$ be the number of such successful leaves in the subtree of $a$. Then,
\[
\hat{P}(a) = \frac{N(a)}{2^{n-i}}.
\]
The quantity $N(a)$ represents a count of paths satisfying specific conditions (typically related to Hamming distances in such algorithms). Combinatorially, this count is expressed as a sum of binomial coefficients, satisfying the form requirement stated in the theorem.

Now, consider a node $a$ with children $c$ and $d$. In a binary computation tree, the children $c$ and $d$ are at level $i+1$. The total number of leaves in the subtree of $a$ is the sum of the leaves in the subtrees of $c$ and $d$. Thus, $N(a) = N(c) + N(d)$. Substituting the probability expressions:
\[
\hat{P}(a) \cdot 2^{n-i} = \hat{P}(c) \cdot 2^{n-(i+1)} + \hat{P}(d) \cdot 2^{n-(i+1)}.
\]
Dividing by $2^{n-i}$, we obtain the recurrence relation:
\[
\hat{P}(a) = \frac{1}{2} \hat{P}(c) + \frac{1}{2} \hat{P}(d).
\]
This equation states that $\hat{P}(a)$ is the arithmetic mean of $\hat{P}(c)$ and $\hat{P}(d)$. For any two real numbers $x, y$, the minimum is less than or equal to the average. Specifically, assuming without loss of generality that $\hat{P}(c) \leq \hat{P}(d)$, we have:
\[
\hat{P}(c) \leq \frac{\hat{P}(c) + \hat{P}(d)}{2} = \hat{P}(a).
\]
Therefore,
\[
\min\{\hat{P}(c), \hat{P}(d)\} \leq \hat{P}(a).
\]
Regarding the computational complexity, $N(a)$ is a sum of binomial coefficients of the form $\sum_{k} \binom{m}{k}$. The value of a binomial coefficient $\binom{m}{k}$ can be computed in time polynomial in $m$ (and thus in $n$). Since the sum contains at most $n+1$ terms, the total time to compute $N(a)$, and subsequently $\hat{P}(a)$, is polynomial in $n$.

\textbf{Part (b):}
We consider a deterministic algorithm that traverses the computation tree from the root to a leaf. The algorithm utilizes the method of conditional expectations (or a similar greedy strategy based on the probabilities derived in Part (a)).

At any node $a$, the algorithm computes $\hat{P}(c)$ and $\hat{P}(d)$ for its children and moves to the child with the higher probability. Let $a_0, a_1, \dots, a_n$ be the sequence of nodes visited, where $a_0$ is the root and $a_n$ is a leaf. The algorithm performs the following steps at each level $i < n$:
1. Compute $\hat{P}(c)$ and $\hat{P}(d)$ for the children of the current node $a_i$.
2. Select $a_{i+1} \in \{c, d\}$ such that $\hat{P}(a_{i+1}) = \max\{\hat{P}(c), \hat{P}(d)\}$.

From the recurrence $\hat{P}(a) = \frac{1}{2}(\hat{P}(c) + \hat{P}(d))$, it follows that $\max\{\hat{P}(c), \hat{P}(d)\} \geq \hat{P}(a)$. Thus, the probability of success is non-decreasing along the path chosen by the algorithm.

The running time of this algorithm is determined by the cost of the operations at each node and the depth of the tree.
\begin{itemize}
    \item The depth of the recursion is $n$.
    \item At each step, the algorithm computes $\hat{P}$ for two nodes. As established in Part (a), computing $\hat{P}(a)$ takes time polynomial in $n$, denoted by $T_{poly}(n)$.
\end{itemize}
The total running time $T(n)$ is bounded by:
\[
T(n) \leq n \cdot 2 \cdot T_{poly}(n).
\]
Since the product of a polynomial and $n$ is still a polynomial in $n$, there exists a polynomial upper bound on the running time of the deterministic algorithm.
\end{proof}

\subsubsection{Claude}

\begin{proof}
We prove parts (a) and (b) separately.

\textbf{Part (a):} We first establish that $\hat{P}(a)$ has the form $N(a)/2^{n-i}$ where $N(a)$ is a sum of binomial coefficients. 

Consider a node $a$ at level $i$ of the computation tree. By the definition of $\hat{P}(a)$ in the context of randomized algorithms on binary trees, $\hat{P}(a)$ represents the probability that node $a$ outputs the correct answer when evaluated using uniform random bits. At level $i$, there are $n-i$ random bits yet to be chosen. The computation tree has depth $n$, so each path from $a$ to a leaf is determined by $n-i$ binary choices.

For each such path, the outcome is determined by a specific assignment of the $n-i$ remaining random bits. The probability $\hat{P}(a)$ is computed as the fraction of the $2^{n-i}$ possible random bit assignments that lead to a correct output. Since we are counting discrete outcomes among $2^{n-i}$ equally likely possibilities, $\hat{P}(a) = N(a)/2^{n-i}$ where $N(a)$ is the number of correct outcomes.

To see that $N(a)$ is a sum of binomial coefficients, observe that the computation tree structure induces a recursive decomposition. At each level, the number of paths leading to correct outcomes can be expressed in terms of combinations of random bit assignments. Specifically, if we track the number of correct paths as a function of the Hamming weight of the random bit string, we obtain expressions of the form $N(a) = \sum_{j \in S} \binom{n-i}{j}$ for some set $S$ depending on the structure at node $a$.

Now we prove the inequality $\min\{\hat{P}(c), \hat{P}(d)\} \leq \hat{P}(a)$ where $c$ and $d$ are children of $a$. Suppose $a$ is at level $i$, so $c$ and $d$ are at level $i+1$. The node $a$ makes a deterministic choice based on one random bit: if the bit is $0$, computation proceeds to child $c$; if the bit is $1$, computation proceeds to child $d$. Therefore,
\[
\hat{P}(a) = \frac{1}{2}\hat{P}(c) + \frac{1}{2}\hat{P}(d).
\]

Since $\hat{P}(a)$ is the average of $\hat{P}(c)$ and $\hat{P}(d)$, we have
\[
\hat{P}(a) \geq \min\{\hat{P}(c), \hat{P}(d)\}.
\]

This follows from the fact that the average of two numbers is at least the minimum of those two numbers.

To show that $\hat{P}(a)$ can be computed in polynomial time in $n$, we note that $N(a)$ is a sum of at most $n-i+1$ binomial coefficients, each of the form $\binom{n-i}{j}$ for $j \in \{0, 1, \ldots, n-i\}$. Each binomial coefficient $\binom{n-i}{j}$ can be computed in $O((n-i)^2)$ time using the standard recurrence relation or in $O(n-i)$ time using Pascal's triangle. Since there are at most $O(n)$ terms and each term requires $O(n)$ operations, the total time to compute $N(a)$ is $O(n^2)$. The division by $2^{n-i}$ requires $O(n)$ time. Thus, $\hat{P}(a)$ can be computed in time polynomial in $n$.

\textbf{Part (b):} We now provide an upper bound on the running time of the deterministic algorithm.

The deterministic algorithm works by traversing the computation tree and computing $\hat{P}(a)$ at each node. The key observation from part (a) is that we can use the inequality $\min\{\hat{P}(c), \hat{P}(d)\} \leq \hat{P}(a)$ to prune the search. Specifically, at each node $a$, we compute $\hat{P}(c)$ and $\hat{P}(d)$ for the two children and then follow the child with the larger value of $\hat{P}$.

The computation tree has depth $n$, so the algorithm makes at most $n$ decisions. At each level $i$, the algorithm is at a single node (following the greedy choice), and it must compute $\hat{P}$ for the two children at level $i+1$. Each computation of $\hat{P}$ takes $O(n^2)$ time as shown in part (a). Since we compute $\hat{P}$ for at most two children at each of the $n$ levels, the total number of $\hat{P}$ computations is $O(n)$.

Therefore, the total running time of the deterministic algorithm is
\[
O(n) \times O(n^2) = O(n^3).
\]

This gives an upper bound of $O(n^3)$ on the running time of the deterministic algorithm, which is polynomial in $n$.
\end{proof}

\subsection{Problem 5.12}\label{sec:problem_05_12}

\subsubsection{Problem Statement}

\begin{theorem}[Derandomization of RandAuto via Conditional Probabilities]
Let $A$ be a randomized automaton (RandAuto) that uses $r$ random bits and accepts a string $w$ with probability $p(w)$. The method of conditional probabilities can derandomize $A$ as follows:

For each random bit $b_i$ ($i = 1, \ldots, r$), compute the conditional probabilities:
\begin{align*}
p_0^{(i)} &= \Pr[A \text{ accepts } w \mid b_1, \ldots, b_{i-1}, b_i = 0] \\
p_1^{(i)} &= \Pr[A \text{ accepts } w \mid b_1, \ldots, b_{i-1}, b_i = 1]
\end{align*}

At step $i$, choose $b_i = 0$ if $p_0^{(i)} \geq p_1^{(i)}$, and $b_i = 1$ otherwise. This ensures that the conditional probability of acceptance is maintained:
$$\Pr[A \text{ accepts } w \mid b_1, \ldots, b_i] \geq \Pr[A \text{ accepts } w \mid b_1, \ldots, b_{i-1}]$$

After fixing all $r$ bits deterministically using this method, the resulting deterministic automaton accepts $w$ if and only if the original randomized automaton accepted $w$ with probability at least $\max\{p_0^{(r)}, p_1^{(r)}\} \geq p(w)$.
\end{theorem}

\subsubsection{GPT} 

\begin{proof}
We begin by analyzing the randomized algorithm $A$ that uses $r$ random bits and outputs $y \in \{0, 1\}$ with probability $p = \Pr[A(x, R) = 1]$, where $R = (r_1, r_2, \ldots, r_r)$ represents the sequence of random bits, and $x$ is the input. The goal is to derandomize $A$ using the method of conditional probabilities, producing a deterministic algorithm that outputs $y$ with a probability at least $p$.

We proceed iteratively to assign values to the random bits, ensuring that the conditional probability of success is maximized at each step.

At each stage $i$, we have already determined the values of $r_1, r_2, \ldots, r_{i-1}$. The task is to choose $r_i$ in such a way that the conditional probability of success is as high as possible, given the values of the preceding random bits.

We define two conditional probabilities:
\[
p_i^{(0)} = \Pr[A(x, R) = 1 \mid r_1, \ldots, r_{i-1}, r_i = 0]
\]
and
\[
p_i^{(1)} = \Pr[A(x, R) = 1 \mid r_1, \ldots, r_{i-1}, r_i = 1].
\]

For each $i = 1, 2, \ldots, r$, we choose $r_i$ to be the value that maximizes the conditional probability of success. Specifically, we set
\[
r_i = b \quad \text{where} \quad b = \arg \max\{p_i^{(0)}, p_i^{(1)}\}.
\]
This choice ensures that at each step $i$, the conditional probability of success does not decrease, i.e., the algorithm is making the best possible decision for $r_i$ given the previous bits.

We now show that the final assignment of the bits $r_1, r_2, \ldots, r_r$ results in a deterministic algorithm with a success probability at least $p$. Let $r = (r_1, r_2, \ldots, r_r)$ denote the sequence of bit assignments chosen by the algorithm. The final probability of success is given by
\[
\Pr[A(x, r_1, \ldots, r_r) = 1] = \prod_{i=1}^r \Pr[A(x, r_1, \ldots, r_{i-1}, r_i) = 1 \mid r_1, \ldots, r_{i-1}].
\]
Since we always choose $r_i$ such that the conditional probability is maximized at each step, it follows that
\[
\Pr[A(x, r_1, \ldots, r_r) = 1] \geq p = \Pr[A(x, R) = 1].
\]
Therefore, the method of conditional probabilities guarantees that the final deterministic assignment of the bits produces a success probability at least as large as the original randomized algorithm.

This completes the proof.
\end{proof}

\subsubsection{Grok} 

\subsubsection{Gemini}

\begin{proof}
Let $X$ be the random variable representing the output of the randomized algorithm, denoted as $X = \text{Output}(R_1, \ldots, R_m)$, where $R_1, \ldots, R_m$ are independent random variables uniformly distributed over $\{0,1\}$. We define a sequence of conditional expectations $E_0, E_1, \ldots, E_m$ corresponding to the stepwise fixing of the random bits. Let $E_0 = \mathbb{E}[X]$ be the initial unconditional expectation. For each $k \in \{1, \ldots, m\}$, let $E_k$ be the conditional expectation of the output given the first $k$ bits chosen by the deterministic algorithm, denoted $r_1, \ldots, r_k$. That is,
\[
E_k = \mathbb{E}[X \mid R_1 = r_1, \ldots, R_k = r_k].
\]
Consider the $i$-th step of the algorithm, where the bits $r_1, \ldots, r_{i-1}$ have already been determined. By the Law of Total Expectation, the conditional expectation $E_{i-1}$ can be expressed as the weighted average of the conditional expectations given the next bit $R_i$:
\[
E_{i-1} = \mathbb{E}[X \mid r_1, \ldots, r_{i-1}] = \sum_{b \in \{0,1\}} \mathbb{P}(R_i = b) \cdot \mathbb{E}[X \mid r_1, \ldots, r_{i-1}, R_i = b].
\]
Since the random bits are uniform, $\mathbb{P}(R_i = 0) = \mathbb{P}(R_i = 1) = \frac{1}{2}$. Therefore,
\[
E_{i-1} = \frac{1}{2} \mathbb{E}[X \mid r_1, \ldots, r_{i-1}, R_i = 0] + \frac{1}{2} \mathbb{E}[X \mid r_1, \ldots, r_{i-1}, R_i = 1].
\]
A fundamental property of real numbers is that the maximum of a set is greater than or equal to the average of that set. Specifically, for any real numbers $x, y$, $\max\{x, y\} \geq \frac{x+y}{2}$. Applying this inequality to the expression for $E_{i-1}$, we obtain
\[
\max_{b \in \{0,1\}} \mathbb{E}[X \mid r_1, \ldots, r_{i-1}, R_i = b] \geq E_{i-1}.
\]
The algorithm is defined to choose $r_i$ such that it maximizes this conditional expectation. Explicitly, $r_i = \arg\max_{b \in \{0,1\}} \mathbb{E}[X \mid r_1, \ldots, r_{i-1}, R_i = b]$. Consequently, the conditional expectation at step $i$ satisfies
\[
E_i = \mathbb{E}[X \mid r_1, \ldots, r_{i-1}, r_i] = \max_{b \in \{0,1\}} \mathbb{E}[X \mid r_1, \ldots, r_{i-1}, R_i = b] \geq E_{i-1}.
\]
This inequality $E_i \geq E_{i-1}$ holds for all $i = 1, \ldots, m$. By chaining these inequalities, we establish that
\[
E_m \geq E_{m-1} \geq \cdots \geq E_1 \geq E_0.
\]
At step $m$, all random bits have been fixed to the deterministic values $r_1, \ldots, r_m$. Therefore, the expectation collapses to the deterministic output value:
\[
E_m = \mathbb{E}[X \mid R_1 = r_1, \ldots, R_m = r_m] = \text{Output}(r_1, \ldots, r_m).
\]
Since $E_0 = \mathbb{E}[\text{Output}]$ by definition, the chain of inequalities implies
\[
\text{Output}(r_1, \ldots, r_m) \geq \mathbb{E}[\text{Output}].
\]
Thus, the deterministic assignment constructed by the method of conditional probabilities yields an output value no less than the expected value of the randomized algorithm.
\end{proof}

\subsubsection{Claude}

\begin{proof}
We prove the theorem by establishing that the greedy selection strategy maintains a non-decreasing sequence of conditional probabilities, and that the final deterministic choice achieves the desired acceptance property.

Let $A$ be a randomized automaton using $r$ random bits $b_1, \ldots, b_r$, and let $w$ be an input string. Define $p(w) = \Pr[A \text{ accepts } w]$ as the acceptance probability over all random bit sequences.

\textbf{Step 1: Initial condition.} At step $i = 0$, before any bits are fixed, we have
$$\Pr[A \text{ accepts } w] = p(w).$$

\textbf{Step 2: Law of total probability.} For any $i \in \{1, \ldots, r\}$, given that bits $b_1, \ldots, b_{i-1}$ have been fixed, the law of total probability yields
$$\Pr[A \text{ accepts } w \mid b_1, \ldots, b_{i-1}] = \frac{1}{2} p_0^{(i)} + \frac{1}{2} p_1^{(i)},$$
where $p_0^{(i)} = \Pr[A \text{ accepts } w \mid b_1, \ldots, b_{i-1}, b_i = 0]$ and $p_1^{(i)} = \Pr[A \text{ accepts } w \mid b_1, \ldots, b_{i-1}, b_i = 1]$.

\textbf{Step 3: Greedy choice preserves probability.} By the pigeonhole principle, at least one of $p_0^{(i)}$ or $p_1^{(i)}$ must satisfy
$$\max\{p_0^{(i)}, p_1^{(i)}\} \geq \frac{1}{2}(p_0^{(i)} + p_1^{(i)}) = \Pr[A \text{ accepts } w \mid b_1, \ldots, b_{i-1}].$$

The algorithm chooses $b_i = 0$ if $p_0^{(i)} \geq p_1^{(i)}$, and $b_i = 1$ otherwise. This selection ensures
$$\Pr[A \text{ accepts } w \mid b_1, \ldots, b_i] = \max\{p_0^{(i)}, p_1^{(i)}\} \geq \Pr[A \text{ accepts } w \mid b_1, \ldots, b_{i-1}].$$

\textbf{Step 4: Inductive argument.} We proceed by induction on $i$. The base case $i = 0$ gives $\Pr[A \text{ accepts } w] = p(w)$. Assume that after fixing bits $b_1, \ldots, b_{i-1}$, we have
$$\Pr[A \text{ accepts } w \mid b_1, \ldots, b_{i-1}] \geq p(w).$$

By Step 3, the choice of $b_i$ satisfies
$$\Pr[A \text{ accepts } w \mid b_1, \ldots, b_i] \geq \Pr[A \text{ accepts } w \mid b_1, \ldots, b_{i-1}] \geq p(w).$$

By induction, this holds for all $i \in \{1, \ldots, r\}$.

\textbf{Step 5: Terminal behavior.} After fixing all $r$ bits deterministically, the conditional probability becomes deterministic. Specifically,
$$\Pr[A \text{ accepts } w \mid b_1, \ldots, b_r] \in \{0, 1\}.$$

From Step 4, we have
$$\Pr[A \text{ accepts } w \mid b_1, \ldots, b_r] \geq p(w).$$

Since the left-hand side is in $\{0, 1\}$ and the inequality holds, if $p(w) > 0$, then necessarily
$$\Pr[A \text{ accepts } w \mid b_1, \ldots, b_r] = 1,$$
meaning the deterministic automaton (with fixed bits $b_1, \ldots, b_r$) accepts $w$.

Conversely, if $p(w) = 0$, then $\Pr[A \text{ accepts } w \mid b_1, \ldots, b_r] \geq 0$ is automatically satisfied, and the deterministic automaton may or may not accept.

\textbf{Step 6: Final acceptance criterion.} The deterministic automaton accepts $w$ if and only if $\Pr[A \text{ accepts } w \mid b_1, \ldots, b_r] = 1$. By the monotonicity established in Step 4, this occurs whenever $p(w) > 0$, and the final probability satisfies
$$\Pr[A \text{ accepts } w \mid b_1, \ldots, b_r] \geq \max\{p_0^{(r)}, p_1^{(r)}\} \geq p(w).$$

Thus, the method of conditional probabilities successfully derandomizes $A$ while preserving the acceptance property: the resulting deterministic automaton accepts $w$ with certainty if the original randomized automaton accepted $w$ with positive probability.
\end{proof}

\subsection{Problem 5.13}\label{sec:problem_05_13}

\subsubsection{Problem Statement}

\begin{theorem}
For any undirected graph $G(V, E)$ with $n$ vertices and $m$ edges, there is a partition of the vertex set $V$ into two sets $A$ and $B$ such that
\[
|\{(u,v) \in E \mid u \in A \text{ and } v \in B\}| \geq m/2.
\]
\end{theorem}

\subsubsection{GPT} 

\subsubsection{Grok} 

\begin{proof}
Consider a random partition of the vertex set $V$ into two sets $A$ and $B$, where each vertex is assigned to $A$ or $B$ independently and uniformly at random (i.e., with probability $1/2$ for each set).

For each edge $e = \{u, v\} \in E$, define the indicator random variable $I_e$ such that $I_e = 1$ if $u \in A$ and $v \in B$, or $u \in B$ and $v \in A$ (i.e., if $e$ crosses the cut $(A, B)$), and $I_e = 0$ otherwise. The probability that $I_e = 1$ is
\[
\Pr[I_e = 1] = \Pr[u \in A, v \in B] + \Pr[u \in B, v \in A] = \frac{1}{2} \cdot \frac{1}{2} + \frac{1}{2} \cdot \frac{1}{2} = \frac{1}{2}.
\]
Let $X = \sum_{e \in E} I_e$ be the total number of edges crossing the cut $(A, B)$. By linearity of expectation,
\[
\mathbb{E}[X] = \sum_{e \in E} \mathbb{E}[I_e] = \sum_{e \in E} \Pr[I_e = 1] = m \cdot \frac{1}{2} = \frac{m}{2}.
\]
Since $\mathbb{E}[X] = \frac{m}{2}$, there must exist at least one partition $(A, B)$ such that the number of crossing edges is at least $\frac{m}{2}$, as required.
\end{proof}

\subsubsection{Gemini}

\subsubsection{Claude}

\begin{proof}
We prove this theorem by a probabilistic argument. Consider a random partition of $V$ into two sets $A$ and $B$ where each vertex $v \in V$ is assigned independently to $A$ with probability $1/2$ and to $B$ with probability $1/2$.

For each edge $e = (u,v) \in E$, let $X_e$ be the indicator random variable defined by
\[
X_e = \begin{cases}
1 & \text{if } u \in A \text{ and } v \in B, \text{ or } u \in B \text{ and } v \in A \\
0 & \text{otherwise.}
\end{cases}
\]

Note that $X_e = 1$ if and only if the endpoints of edge $e$ are in different sets of the partition. We compute the probability that edge $e$ crosses the partition. Since $u$ and $v$ are assigned independently, we have four equally likely outcomes for the pair $(u,v)$: both in $A$, both in $B$, $u$ in $A$ and $v$ in $B$, or $u$ in $B$ and $v$ in $A$. Thus,
\[
\mathbb{P}(X_e = 1) = \mathbb{P}(u \in A \text{ and } v \in B) + \mathbb{P}(u \in B \text{ and } v \in A) = \frac{1}{2} \cdot \frac{1}{2} + \frac{1}{2} \cdot \frac{1}{2} = \frac{1}{2}.
\]

Therefore, $\mathbb{E}[X_e] = \frac{1}{2}$ for each edge $e \in E$.

Let $X = \sum_{e \in E} X_e$ denote the total number of edges crossing the partition. By linearity of expectation,
\[
\mathbb{E}[X] = \mathbb{E}\left[\sum_{e \in E} X_e\right] = \sum_{e \in E} \mathbb{E}[X_e] = \sum_{e \in E} \frac{1}{2} = \frac{m}{2}.
\]

Since the expected number of edges crossing the partition is $m/2$, there must exist at least one partition of $V$ into sets $A$ and $B$ for which the number of crossing edges is at least $m/2$. This is because if all partitions had fewer than $m/2$ crossing edges, the expected value would be strictly less than $m/2$, which contradicts our calculation.

More precisely, since the expectation is the average value over all possible partitions, and there are $2^n$ possible partitions (each vertex can go to either $A$ or $B$), at least one partition must achieve a value at least as large as the expectation. Therefore, there exists a partition $(A, B)$ such that
\[
|\{(u,v) \in E \mid u \in A \text{ and } v \in B\}| \geq \frac{m}{2}.
\]

This completes the proof.
\end{proof}

\subsection{Problem 5.14}\label{sec:problem_05_14}

\subsubsection{Problem Statement}

\begin{definition}
An \emph{$(n,m)$-safe set instance} consists of a universe $U$ of size $n$, a safe set $S \subseteq U$, and $m$ target sets $T_1, \ldots, T_m \subseteq U$ such that
\begin{itemize}
\item $|S| = |T_1| = \cdots = |T_m|$,
\item and, for $1 \leq i \leq m$, $S \cap T_i = \emptyset$.
\end{itemize}
\end{definition}

\begin{definition}
An \emph{isolator} for a safe set instance is a set $I \subseteq U$ that intersects all the target sets but not the safe set. An \emph{$(n,m)$-universal isolating family} $\mathcal{F}$ is a collection of subsets of $U$ such that $\mathcal{F}$ contains an isolator for any $(n,m)$-safe set instance.
\end{definition}

\begin{theorem}[Karger-Motwani]
There exists an $(n,m)$-universal isolating family $\mathcal{F}$ such that $|\mathcal{F}|$ is polynomially bounded in $n$ and $m$.
\end{theorem}

\subsubsection{GPT} 

\begin{proof}
Let $U$ be a universe of size $n$, and let $T_1, \dots, T_m$ be $m$ target sets of size $k$, with the property that $S \cap T_i = \emptyset$ for all $1 \leq i \leq m$. We aim to construct an $(n,m)$-universal isolating family $\mathcal{F}$ such that for any $(n,m)$-safe set instance, there exists an isolator in $\mathcal{F}$.

We start by considering the collection of all subsets of $U$ of size at most $k$. Let $\mathcal{F}$ be the family of all subsets of $U$ of size at most $k$ that do not intersect the safe set $S$. Since the size of each element in $\mathcal{F}$ is at most $k$ and the universe $U$ has size $n$, there are at most $\binom{n}{k}$ possible subsets of $U$. Therefore, $|\mathcal{F}| \leq \binom{n}{k}$.

For each target set $T_i$, we want to ensure that $\mathcal{F}$ contains a set that intersects $T_i$ but does not intersect $S$. This is a key property of isolators. The total number of ways to select a subset of size at most $k$ that intersects a specific target set $T_i$ is bounded by $\binom{n}{k}$. Since there are $m$ target sets, the total number of sets needed to isolate each target set is at most $m \cdot \binom{n}{k}$.

Thus, the size of $\mathcal{F}$ is polynomially bounded in $n$ and $m$, as $\mathcal{F}$ contains subsets of size at most $k$, and the number of such subsets is at most $n^k$, which is polynomial in both $n$ and $m$.

Therefore, the family $\mathcal{F}$ is an $(n,m)$-universal isolating family, and the proof is complete.
\end{proof}

\subsubsection{Grok} 

\subsubsection{Gemini}

\begin{proof}
Let $U$ be the universe of size $n$. We construct the $(n,m)$-universal isolating family $\mathcal{F}$ by taking the union of subfamilies $\mathcal{F}_k$ for each possible set size $k$, where $1 \leq k \leq n/2$. Specifically, let $\mathcal{F} = \bigcup_{k=1}^{\lfloor n/2 \rfloor} \mathcal{F}_k$. We will use the probabilistic method to show that for each $k$, there exists a family $\mathcal{F}_k$ of size polynomial in $n$ and $m$ that contains an isolator for every safe set instance where the sets have size $k$.

Fix an integer $k$ with $1 \leq k \leq n/2$. Consider a specific safe set instance $(S, T_1, \ldots, T_m)$ where $|S| = |T_j| = k$ for all $j$, and $S \cap T_j = \emptyset$. We define a random subset $I \subseteq U$ by including each element $u \in U$ in $I$ independently with probability $p_k$. We choose the probability $p_k$ such that $(1-p_k)^k = \frac{1}{2m}$. This implies $p_k = 1 - (2m)^{-1/k}$. Note that since $m \geq 1$ and $k \geq 1$, $p_k$ is a valid probability in $[0,1)$.

We analyze the probability that this random set $I$ is an isolator for the fixed instance. An isolator must satisfy two conditions: $I \cap S = \emptyset$ and $I \cap T_j \neq \emptyset$ for all $j \in \{1, \ldots, m\}$. Let $E_S$ be the event that $I \cap S = \emptyset$, and let $E_{T_j}$ be the event that $I \cap T_j \neq \emptyset$.
The probability of $E_S$ is given by:
\[
\Pr[E_S] = (1-p_k)^{|S|} = (1-p_k)^k = \frac{1}{2m}.
\]
For any target set $T_j$, the probability that $I$ fails to intersect it is:
\[
\Pr[\overline{E_{T_j}}] = \Pr[I \cap T_j = \emptyset] = (1-p_k)^{|T_j|} = (1-p_k)^k = \frac{1}{2m}.
\]
We require $I$ to intersect all $m$ target sets. By the union bound, the probability that $I$ fails to intersect at least one target set is:
\[
\Pr\left[\bigcup_{j=1}^m \overline{E_{T_j}}\right] \leq \sum_{j=1}^m \Pr[\overline{E_{T_j}}] = m \cdot \frac{1}{2m} = \frac{1}{2}.
\]
Thus, the probability that $I$ intersects all target sets is:
\[
\Pr\left[\bigcap_{j=1}^m E_{T_j}\right] = 1 - \Pr\left[\bigcup_{j=1}^m \overline{E_{T_j}}\right] \geq 1 - \frac{1}{2} = \frac{1}{2}.
\]
Since $S$ is disjoint from every $T_j$, the event $E_S$ depends only on the random choices for elements in $S$, while the event $\bigcap E_{T_j}$ depends only on the random choices for elements in $\bigcup T_j$. Since $S \cap (\bigcup T_j) = \emptyset$, these events are independent. Therefore, the probability that $I$ is a valid isolator is:
\[
\Pr[I \text{ is an isolator}] = \Pr\left[E_S \cap \bigcap_{j=1}^m E_{T_j}\right] = \Pr[E_S] \cdot \Pr\left[\bigcap_{j=1}^m E_{T_j}\right] \geq \frac{1}{2m} \cdot \frac{1}{2} = \frac{1}{4m}.
\]

Now, let $\mathcal{F}_k$ be a family of $L$ sets $I_1, \ldots, I_L$, where each set is chosen independently according to the distribution defined by $p_k$. The probability that none of the sets in $\mathcal{F}_k$ is an isolator for our fixed instance is at most:
\[
\left(1 - \frac{1}{4m}\right)^L \leq e^{-L/(4m)}.
\]
Let $\mathcal{U}_k$ be the set of all possible $(n,m)$-safe set instances with set size $k$. The number of such instances is bounded by the number of ways to choose $m+1$ subsets of $U$, which is less than $(2^n)^{m+1} = 2^{n(m+1)}$. By the union bound over all instances in $\mathcal{U}_k$, the probability that $\mathcal{F}_k$ fails to contain an isolator for at least one instance is bounded by:
\[
|\mathcal{U}_k| \cdot e^{-L/(4m)} < 2^{n(m+1)} e^{-L/(4m)}.
\]
To guarantee the existence of a valid family $\mathcal{F}_k$, we ensure this failure probability is strictly less than 1. Taking logarithms, we require:
\[
n(m+1) \ln 2 - \frac{L}{4m} < 0 \implies L > 4m n (m+1) \ln 2.
\]
We can choose $L = 4n(m+1)^2$, which is a polynomial in $n$ and $m$. Since the probability of failure is less than 1, there exists a deterministic family $\mathcal{F}_k$ of size $O(nm^2)$ that contains an isolator for every instance of size $k$.

Finally, the total universal isolating family $\mathcal{F}$ is the union of these families for all $k \in \{1, \ldots, \lfloor n/2 \rfloor\}$. The total size of $\mathcal{F}$ is:
\[
|\mathcal{F}| = \sum_{k=1}^{\lfloor n/2 \rfloor} |\mathcal{F}_k| \leq \frac{n}{2} \cdot O(nm^2) = O(n^2 m^2).
\]
This size is polynomially bounded in $n$ and $m$. Thus, such a universal isolating family exists.
\end{proof}

\subsubsection{Claude}

\begin{proof}
We construct an explicit $(n,m)$-universal isolating family and bound its size polynomially in $n$ and $m$.

For each integer $k$ with $1 \leq k \leq n$, define the family $\mathcal{F}_k$ to be the collection of all subsets of $U$ of size exactly $k$. Let $\mathcal{F} = \bigcup_{k=1}^{n} \mathcal{F}_k$.

First, observe that $|\mathcal{F}| = \sum_{k=1}^{n} \binom{n}{k} = 2^n - 1$, which is exponential in $n$. This construction is too large, so we refine our approach.

Instead, we use a probabilistic argument combined with the union bound. For each integer $k$ with $1 \leq k \leq n$, we construct a random family $\mathcal{F}_k$ by independently including each subset of $U$ of size $k$ with probability $p_k$, where $p_k$ will be determined.

Consider an arbitrary $(n,m)$-safe set instance with safe set $S$ and target sets $T_1, \ldots, T_m$. Let $s = |S| = |T_1| = \cdots = |T_m|$. A set $I \subseteq U$ is an isolator for this instance if and only if $I \cap S = \emptyset$ and $I \cap T_i \neq \emptyset$ for all $i \in \{1, \ldots, m\}$.

For a fixed instance and fixed size $k$, the number of subsets of size $k$ that avoid $S$ is $\binom{n-s}{k}$. Among these, the number that also avoid some target set $T_i$ is at most $\binom{n-s}{k} - \binom{n-2s}{k}$ (since $|T_i| = s$ and $T_i \cap S = \emptyset$).

The probability that a random subset $I$ of size $k$ (chosen uniformly from all $k$-subsets of $U$) satisfies $I \cap S = \emptyset$ is $\frac{\binom{n-s}{k}}{\binom{n}{k}}$. Given that $I \cap S = \emptyset$, the conditional probability that $I \cap T_i \neq \emptyset$ is $1 - \frac{\binom{n-2s}{k}}{\binom{n-s}{k}}$ for each $i$.

We now use a deterministic construction based on the method of conditional expectations. Define $\mathcal{F}$ to be the collection of all sets of the form $\{u_1, u_2, \ldots, u_t\}$ where $t = O(\log(nm))$ and each $u_i \in U$. More precisely, for each $k \in \{1, \ldots, n\}$, we include in $\mathcal{F}$ a carefully chosen subcollection of $k$-subsets.

By a covering argument, for any $(n,m)$-safe set instance with $|S| = s$, there exists an optimal size $k^* \in \{s+1, \ldots, n-s\}$ such that sets of size $k^*$ have the best chance of being isolators. 

For each $k$ and each possible safe set size $s$, we can construct a family of size $O(\log m \cdot \binom{n}{k})$ that covers all instances with that specific $s$ and optimal size $k$. However, this is still too large.

The key insight is to use a hitting set construction. For any $(n,m)$-safe set instance, define the family of ``bad events'' as: $B_0 = \{I : I \cap S \neq \emptyset\}$ and $B_i = \{I : I \cap T_i = \emptyset\}$ for $i = 1, \ldots, m$. An isolator must avoid $B_0$ and all $B_i$.

By applying the probabilistic method with appropriate derandomization (specifically, using $k$-wise independence with $k = O(\log(nm))$), we can construct a family $\mathcal{F}$ of size at most $n^{O(\log m)} \cdot \mathrm{poly}(m)$, which is polynomial in $n$ and $m$ when $m$ is treated as polynomially bounded.

More precisely, using the method of limited independence hash functions, we construct $\mathcal{F}$ as follows: Let $\mathcal{H}$ be a family of $O(\log(nm))$-wise independent hash functions from $U$ to $\{0,1\}$. For each $h \in \mathcal{H}$ and each threshold $\tau \in [0,1]$ discretized to $O(nm)$ values, include in $\mathcal{F}$ the set $I_{h,\tau} = \{u \in U : h(u) = 1 \text{ with bias } \tau\}$.

The size of such a family is $|\mathcal{H}| \cdot O(nm) = \mathrm{poly}(n,m)$, and by the analysis of limited independence, this family contains an isolator for every $(n,m)$-safe set instance.

Therefore, there exists an $(n,m)$-universal isolating family of polynomial size in $n$ and $m$.
\end{proof}
\section{Problems in Chapter 6}\label{app:problems_chapter6}

\subsection{Problem 1}

\subsubsection{Problem Statement}
\begin{theorem}
Consider a random walk on the infinite line where at each step, the particle is at one of the integer points and moves to one of the two neighboring points equiprobably. Let $X_n$ denote the position of the particle after $n$ steps, with $X_0 = 0$. Then the expected distance from the origin satisfies
\[
\mathbb{E}[|X_n|] = \Theta(\sqrt{n}).
\]
\end{theorem}

\subsubsection{GPT}
\begin{proof}
Let $X_n$ denote the position of the particle after $n$ steps, where at each step, the particle moves from its current position to one of the two neighboring integer points with equal probability $\frac{1}{2}$. Thus, $X_n$ is the sum of $n$ independent random variables, each taking values $+1$ or $-1$ with probability $\frac{1}{2}$, i.e., $X_n = \sum_{i=1}^n \xi_i$, where $\xi_i \in \{-1, 1\}$ and $\mathbb{P}(\xi_i = \pm 1) = \frac{1}{2}$.

We are tasked with showing that the expected distance from the origin satisfies $\mathbb{E}[|X_n|] = \Theta(\sqrt{n})$.

First, observe that the expected value and variance of $X_n$ can be computed as follows. Since the $\xi_i$'s are independent and identically distributed with $\mathbb{E}[\xi_i] = 0$ and $\text{Var}(\xi_i) = 1$, we have:
\[
\mathbb{E}[X_n] = \mathbb{E}\left[\sum_{i=1}^n \xi_i\right] = \sum_{i=1}^n \mathbb{E}[\xi_i] = 0,
\]
\[
\text{Var}(X_n) = \text{Var}\left(\sum_{i=1}^n \xi_i\right) = \sum_{i=1}^n \text{Var}(\xi_i) = n.
\]
Thus, the standard deviation of $X_n$ is $\sqrt{\text{Var}(X_n)} = \sqrt{n}$.

Next, consider the expected absolute value of $X_n$. We will use the following result for the expected absolute value of a sum of independent random variables: if $X$ is a random variable with mean 0 and variance $\sigma^2$, then
\[
\mathbb{E}[|X|] = \Theta(\sqrt{\text{Var}(X)}).
\]
Since $X_n$ is a sum of $n$ independent random variables with mean 0 and variance $n$, we have
\[
\mathbb{E}[|X_n|] = \Theta(\sqrt{n}).
\]
Therefore, the expected distance from the origin satisfies
\[
\mathbb{E}[|X_n|] = \Theta(\sqrt{n}),
\]
as required.
\end{proof}

\subsubsection{Grok}

\subsubsection{Gemini}
\begin{proof}
Let the steps of the random walk be denoted by a sequence of independent and identically distributed random variables $\xi_1, \xi_2, \dots, \xi_n$, where the particle moves to a neighbor with equal probability. Thus, $P(\xi_i = 1) = P(\xi_i = -1) = \frac{1}{2}$. The position of the particle after $n$ steps is given by the sum $X_n = \sum_{i=1}^n \xi_i$. We aim to show that $\mathbb{E}[|X_n|] = \Theta(\sqrt{n})$, which requires establishing positive constants $c_1, c_2$ such that $c_1 \sqrt{n} \le \mathbb{E}[|X_n|] \le c_2 \sqrt{n}$ for all $n \ge 1$.

First, we establish the upper bound. Consider the function $f(x) = x^2$, which is convex. By Jensen's inequality, we have $(\mathbb{E}[|X_n|])^2 \le \mathbb{E}[|X_n|^2] = \mathbb{E}[X_n^2]$. We compute the second moment of $X_n$ using the linearity of expectation and the independence of the steps:
\[
\mathbb{E}[X_n^2] = \mathbb{E}\left[\left(\sum_{i=1}^n \xi_i\right)^2\right] = \sum_{i=1}^n \mathbb{E}[\xi_i^2] + \sum_{i \neq j} \mathbb{E}[\xi_i \xi_j].
\]
Since $\xi_i \in \{-1, 1\}$, we have $\xi_i^2 = 1$ with probability 1, so $\mathbb{E}[\xi_i^2] = 1$. For $i \neq j$, the independence of $\xi_i$ and $\xi_j$ implies $\mathbb{E}[\xi_i \xi_j] = \mathbb{E}[\xi_i]\mathbb{E}[\xi_j] = 0 \cdot 0 = 0$. Therefore,
\[
\mathbb{E}[X_n^2] = \sum_{i=1}^n 1 = n.
\]
Substituting this back into the inequality derived from Jensen's inequality yields
\[
\mathbb{E}[|X_n|] \le \sqrt{\mathbb{E}[X_n^2]} = \sqrt{n}.
\]
This establishes the upper bound with $c_2 = 1$.

Next, we establish the lower bound. We utilize the relationship between the first, second, and fourth moments. By Hölder's inequality applied to the random variable $Y = |X_n|$ with conjugate exponents $p=3$ and $q=3/2$, we have
\[
\mathbb{E}[Y^2] = \mathbb{E}[Y^{2/3} Y^{4/3}] \le (\mathbb{E}[Y])^{2/3} (\mathbb{E}[Y^4])^{1/3}.
\]
Cubing both sides and rearranging terms provides the lower bound:
\[
(\mathbb{E}[Y^2])^3 \le (\mathbb{E}[Y])^2 \mathbb{E}[Y^4] \implies \mathbb{E}[|X_n|] \ge \frac{(\mathbb{E}[X_n^2])^{3/2}}{(\mathbb{E}[X_n^4])^{1/2}}.
\]
We have already determined $\mathbb{E}[X_n^2] = n$. We now compute the fourth moment $\mathbb{E}[X_n^4]$:
\[
\mathbb{E}[X_n^4] = \mathbb{E}\left[\left(\sum_{i=1}^n \xi_i\right)^4\right] = \sum_{i,j,k,l} \mathbb{E}[\xi_i \xi_j \xi_k \xi_l].
\]
Due to the fact that $\mathbb{E}[\xi_i] = \mathbb{E}[\xi_i^3] = 0$, the only terms in the summation that do not vanish are those where the indices are matched in pairs or all four indices are identical.
The terms where $i=j=k=l$ contribute $\sum_{i=1}^n \mathbb{E}[\xi_i^4] = n$.
The terms where the indices form two distinct pairs (e.g., $i=j \neq k=l$) correspond to choosing 2 distinct indices out of $n$ in $\binom{n}{2}$ ways, and arranging the powers 2 and 2 in $\binom{4}{2}=6$ ways. Thus, this contribution is $6 \binom{n}{2} \mathbb{E}[\xi_i^2]\mathbb{E}[\xi_k^2] = 3n(n-1)$.
Summing these gives:
\[
\mathbb{E}[X_n^4] = n + 3n^2 - 3n = 3n^2 - 2n.
\]
Substituting the moments into the lower bound inequality:
\[
\mathbb{E}[|X_n|] \ge \frac{n^{3/2}}{\sqrt{3n^2 - 2n}} = \frac{n\sqrt{n}}{n\sqrt{3 - \frac{2}{n}}} = \frac{\sqrt{n}}{\sqrt{3 - \frac{2}{n}}}.
\]
For all $n \ge 1$, we have $3 - \frac{2}{n} < 3$, which implies $\frac{1}{\sqrt{3 - \frac{2}{n}}} > \frac{1}{\sqrt{3}}$. Therefore,
\[
\mathbb{E}[|X_n|] > \frac{1}{\sqrt{3}}\sqrt{n}.
\]
This establishes the lower bound with $c_1 = \frac{1}{\sqrt{3}}$.
Combining the upper and lower bounds, we conclude that $\mathbb{E}[|X_n|] = \Theta(\sqrt{n})$.
\end{proof}

\subsubsection{Claude}
\begin{proof}
We will prove that $\mathbb{E}[|X_n|] = \Theta(\sqrt{n})$ by establishing both upper and lower bounds of the form $c_1\sqrt{n} \leq \mathbb{E}[|X_n|] \leq c_2\sqrt{n}$ for suitable constants $c_1, c_2 > 0$ and sufficiently large $n$.

First, we observe that $X_n = \sum_{i=1}^{n} Y_i$ where $Y_i$ are independent random variables with $\mathbb{P}(Y_i = 1) = \mathbb{P}(Y_i = -1) = \frac{1}{2}$. Thus $\mathbb{E}[Y_i] = 0$ and $\text{Var}(Y_i) = 1$.

By independence, we have $\mathbb{E}[X_n] = 0$ and $\text{Var}(X_n) = \sum_{i=1}^{n} \text{Var}(Y_i) = n$. Therefore $\mathbb{E}[X_n^2] = \text{Var}(X_n) = n$.

\textbf{Upper Bound:} By the Cauchy-Schwarz inequality,
\[
\mathbb{E}[|X_n|] = \mathbb{E}[|X_n| \cdot 1] \leq \sqrt{\mathbb{E}[X_n^2]} \cdot \sqrt{\mathbb{E}[1^2]} = \sqrt{n}.
\]
This establishes the upper bound $\mathbb{E}[|X_n|] = O(\sqrt{n})$.

\textbf{Lower Bound:} We use the local central limit theorem. For the symmetric random walk, we have
\[
\mathbb{P}(X_n = k) = \binom{n}{\frac{n+k}{2}} \cdot 2^{-n}
\]
when $n$ and $k$ have the same parity and $|k| \leq n$, and $\mathbb{P}(X_n = k) = 0$ otherwise.

By Stirling's approximation, for $|k| = o(n)$, we have
\[
\mathbb{P}(X_n = k) \sim \sqrt{\frac{2}{\pi n}} \exp\left(-\frac{k^2}{2n}\right).
\]

More precisely, there exist constants $c_3, c_4 > 0$ such that for all $n$ sufficiently large and $|k| \leq \sqrt{n}/2$,
\[
c_3 \cdot \frac{1}{\sqrt{n}} \exp\left(-\frac{k^2}{2n}\right) \leq \mathbb{P}(X_n = k) \leq c_4 \cdot \frac{1}{\sqrt{n}} \exp\left(-\frac{k^2}{2n}\right).
\]

Now we compute
\[
\mathbb{E}[|X_n|] = \sum_{k=-n}^{n} |k| \mathbb{P}(X_n = k) = 2\sum_{k=0}^{n} k \mathbb{P}(X_n = k)
\]
by symmetry. We split this sum as
\[
\mathbb{E}[|X_n|] = 2\sum_{k=0}^{\lfloor\sqrt{n}/2\rfloor} k \mathbb{P}(X_n = k) + 2\sum_{k=\lfloor\sqrt{n}/2\rfloor + 1}^{n} k \mathbb{P}(X_n = k).
\]

For the first sum, using the lower bound from the local CLT,
\[
\sum_{k=0}^{\lfloor\sqrt{n}/2\rfloor} k \mathbb{P}(X_n = k) \geq c_3 \cdot \frac{1}{\sqrt{n}} \sum_{k=0}^{\lfloor\sqrt{n}/2\rfloor} k \exp\left(-\frac{k^2}{2n}\right).
\]

For $k \leq \sqrt{n}/2$, we have $\exp(-k^2/(2n)) \geq \exp(-1/8)$. Therefore
\[
\sum_{k=0}^{\lfloor\sqrt{n}/2\rfloor} k \exp\left(-\frac{k^2}{2n}\right) \geq e^{-1/8} \sum_{k=0}^{\lfloor\sqrt{n}/2\rfloor} k \geq e^{-1/8} \cdot \frac{(\lfloor\sqrt{n}/2\rfloor)^2}{2} \geq \frac{e^{-1/8}}{32} \cdot n.
\]

Thus the first sum contributes at least $c_3 \cdot e^{-1/8} \cdot n/(32\sqrt{n}) = c_5\sqrt{n}$ for some constant $c_5 > 0$.

The second sum is non-negative, so we obtain
\[
\mathbb{E}[|X_n|] \geq 2c_5\sqrt{n} = c_1\sqrt{n}
\]
for some constant $c_1 > 0$ and all sufficiently large $n$.

Combining the upper and lower bounds, we conclude that $\mathbb{E}[|X_n|] = \Theta(\sqrt{n})$.
\end{proof}

\subsection{Problem 2}

\subsubsection{Problem Statement}
\begin{theorem}
There exist satisfiable 2-SAT formulas with $n$ variables such that the expected time for the randomized algorithm from Section 6.1 to find a satisfying truth assignment is $\Omega(n^2)$.
\end{theorem}

\subsubsection{GPT}
\begin{proof}
We are tasked with showing that there exist satisfiable 2-SAT formulas with $n$ variables such that the expected time for the randomized algorithm for 2-SAT, as described in Section 6.1, to find a satisfying truth assignment is $\Omega(n^2)$.

Let $F$ be a satisfiable 2-SAT formula with $n$ variables. We assume that the formula is in conjunctive normal form (CNF) with clauses of the form $(x \vee y)$ or $(\neg x \vee y)$, where $x, y$ are literals of the variables. For the sake of simplicity, let us consider the case where the formula is structured such that the clauses are constructed in a way that the randomized algorithm has to perform a substantial amount of work.

Step 1: Randomized Algorithm Description
The randomized algorithm for 2-SAT works by repeatedly selecting a random assignment for the variables and checking whether it satisfies the formula. If the assignment does not satisfy the formula, a new assignment is generated. This process continues until a satisfying assignment is found.

We will now analyze the expected time for the algorithm to find a satisfying assignment for a specific class of satisfiable 2-SAT formulas.

Step 2: Construction of a Specific 2-SAT Formula
Consider a formula $F$ consisting of $n$ clauses where each clause involves exactly two variables, and the variables are connected in a cycle. Specifically, let the formula be such that the clauses form a 2-SAT formula with a long dependency chain between variables. For example, consider a formula where the clauses are of the form:
\[
(x_1 \vee x_2), (x_2 \vee x_3), \dots, (x_{n-1} \vee x_n), (x_n \vee x_1).
\]
This structure ensures that there is a cycle in the variables' dependencies.

Step 3: Expected Time Analysis
For the randomized algorithm, each trial consists of randomly assigning truth values to the variables. At each step, the algorithm assigns values to the variables independently and checks whether the assignment satisfies the formula.

If the formula is structured in such a way that each variable depends on its neighbors (as in the cycle described above), then the chance of the assignment satisfying a given clause depends on the truth values of the two involved variables. For a given clause, the probability of it being satisfied in one trial is $\frac{3}{4}$, since there are four possible assignments for two variables, three of which satisfy the clause.

Thus, the probability that all $n$ clauses are satisfied in a single trial is at most $\left( \frac{3}{4} \right)^n$.

Step 4: Lower Bound on Expected Time
Let $T$ denote the random variable representing the number of trials needed to find a satisfying assignment. Since each trial succeeds with probability $\left( \frac{3}{4} \right)^n$, the expected number of trials to find a satisfying assignment is given by:
\[
\mathbb{E}[T] = \frac{1}{\left( \frac{3}{4} \right)^n} = \left( \frac{4}{3} \right)^n.
\]

Step 5: Analysis of the Expected Time Complexity
We now examine the time complexity of the randomized algorithm. The expected number of trials is $\left( \frac{4}{3} \right)^n$. Each trial involves checking the satisfaction of all $n$ clauses, which takes $O(n)$ time. Therefore, the expected time for the algorithm to find a satisfying assignment is:
\[
\mathbb{E}[\text{time}] = O(n) \cdot \left( \frac{4}{3} \right)^n.
\]
For large $n$, we observe that this expected time grows exponentially. Specifically, it grows at a rate of $\Omega(n^2)$, since for sufficiently large $n$, we have:
\[
\left( \frac{4}{3} \right)^n = \Omega(n^2).
\]

Thus, we conclude that there exist satisfiable 2-SAT formulas with $n$ variables for which the expected time for the randomized algorithm to find a satisfying truth assignment is $\Omega(n^2)$.
\end{proof}

\subsubsection{Grok}

\subsubsection{Gemini}
\begin{proof}
Let $n$ be a positive integer. We construct a satisfiable 2-SAT formula $\phi$ over the set of variables $V = \{x_1, x_2, \dots, x_n\}$ such that the expected number of steps to find a satisfying assignment is $\Omega(n^2)$.

Let $\phi$ be defined by the following set of clauses:
\[
\phi = (x_1 \lor x_1) \land \bigwedge_{i=1}^{n-1} (\bar{x}_i \lor x_{i+1}).
\]
The unique satisfying assignment for $\phi$ is $S = (1, 1, \dots, 1)$, where $x_i = 1$ for all $i$.

Consider the execution of the randomized algorithm starting from the initial assignment $A_0 = (0, 0, \dots, 0)$. We define a specific set of assignments $\mathcal{A} = \{A_k \mid 0 \le k \le n\}$, where $A_k$ is the assignment with the first $k$ variables set to $1$ and the remaining $n-k$ variables set to $0$. That is, for $A_k$, $x_j = 1$ if $j \le k$ and $x_j = 0$ if $j > k$. Note that $A_0$ is the starting assignment and $A_n = S$ is the satisfying assignment.

We analyze the behavior of the algorithm when the current assignment is $A_k \in \mathcal{A}$.

Case 1: $k=0$. The current assignment is $A_0 = (0, \dots, 0)$. The clause $(x_1 \lor x_1)$ evaluates to $(0 \lor 0)$, which is unsatisfied. For any $i \in \{1, \dots, n-1\}$, the clause $(\bar{x}_i \lor x_{i+1})$ evaluates to $(\bar{0} \lor 0) = (1 \lor 0) = 1$, which is satisfied. Thus, $(x_1 \lor x_1)$ is the unique unsatisfied clause. The algorithm selects this clause and flips the literal $x_1$. The new assignment is $A_1$. This transition occurs with probability $1$.

Case 2: $0 < k < n$. The current assignment is $A_k$.
The clause $(x_1 \lor x_1)$ is satisfied since $x_1 = 1$.
For $i < k$, the clause $(\bar{x}_i \lor x_{i+1})$ is $(\bar{1} \lor 1) = 1$, satisfied.
For $i > k$, the clause $(\bar{x}_i \lor x_{i+1})$ is $(\bar{0} \lor 0) = 1$, satisfied.
For $i = k$, the clause $(\bar{x}_k \lor x_{k+1})$ is $(\bar{1} \lor 0) = 0$, unsatisfied.
Thus, $(\bar{x}_k \lor x_{k+1})$ is the unique unsatisfied clause. The algorithm selects this clause. The literals in the clause are $\bar{x}_k$ and $x_{k+1}$.
The algorithm chooses one literal uniformly at random to flip:
\begin{itemize}
    \item With probability $1/2$, it flips $x_k$ from $1$ to $0$. The resulting assignment is $A_{k-1}$.
    \item With probability $1/2$, it flips $x_{k+1}$ from $0$ to $1$. The resulting assignment is $A_{k+1}$.
\end{itemize}

Case 3: $k=n$. The assignment is $A_n = S$. All clauses are satisfied, and the algorithm terminates.

Let $E_k$ denote the expected number of steps to reach the satisfying assignment $A_n$ starting from $A_k$. Based on the transitions derived above, the values $E_k$ satisfy the following system of linear equations:
\[
\begin{aligned}
E_n &= 0 \\
E_k &= 1 + \frac{1}{2}E_{k-1} + \frac{1}{2}E_{k+1} \quad \text{for } 1 \le k \le n-1 \\
E_0 &= 1 + E_1
\end{aligned}
\]
We verify that $E_k = n^2 - k^2$ is the solution to this recurrence.
For the boundary condition at $n$:
\[
E_n = n^2 - n^2 = 0.
\]
For the internal recurrence $1 \le k \le n-1$:
\[
\begin{aligned}
\frac{1}{2}E_{k-1} + \frac{1}{2}E_{k+1} &= \frac{1}{2}(n^2 - (k-1)^2) + \frac{1}{2}(n^2 - (k+1)^2) \\
&= n^2 - \frac{1}{2}(k^2 - 2k + 1 + k^2 + 2k + 1) \\
&= n^2 - (k^2 + 1) \\
&= (n^2 - k^2) - 1 \\
&= E_k - 1.
\end{aligned}
\]
Rearranging gives $E_k = 1 + \frac{1}{2}E_{k-1} + \frac{1}{2}E_{k+1}$.
For the boundary condition at $0$:
\[
1 + E_1 = 1 + (n^2 - 1^2) = n^2 = E_0.
\]
The solution is valid. The expected number of steps starting from $A_0$ is $E_0 = n^2$.
Therefore, there exists a satisfiable 2-SAT formula with $n$ variables and an initial assignment such that the expected time to find a satisfying assignment is exactly $n^2$, which is $\Omega(n^2)$.
\end{proof}

\subsubsection{Claude}
\begin{proof}
We construct a family of 2-SAT formulas that force the randomized algorithm to take $\Omega(n^2)$ expected time. The randomized algorithm from Section 6.1 works as follows: start with an arbitrary truth assignment, and while the assignment is not satisfying, pick an arbitrary unsatisfied clause and flip one of its variables uniformly at random.

Consider the 2-SAT formula $\phi_n$ on variables $x_1, x_2, \ldots, x_n$ consisting of the following clauses:
\[
(x_1 \vee x_2), \quad (x_2 \vee x_3), \quad (x_3 \vee x_4), \quad \ldots, \quad (x_{n-1} \vee x_n)
\]
This formula is clearly satisfiable; for instance, setting all variables to true satisfies all clauses. The unique minimal satisfying assignment (up to symmetry) is the all-true assignment.

Now suppose the algorithm starts with the all-false assignment $(0, 0, \ldots, 0)$. We measure the progress of the algorithm by defining the potential function $\Phi(t)$ to be the number of variables set to true at time $t$. The goal is to reach $\Phi = n$.

At each step, if the current assignment has exactly $k$ variables set to true (where $0 \leq k < n$), then there exists at least one unsatisfied clause. Specifically, if variables $x_1, \ldots, x_k$ are true and $x_{k+1}, \ldots, x_n$ are false, then the clause $(x_k \vee x_{k+1})$ is unsatisfied (assuming $k < n$).

When the algorithm picks an unsatisfied clause, it must involve at least one variable at the boundary between true and false variables. Consider the case where the leftmost $k$ variables are true and the remaining $n - k$ variables are false. The unsatisfied clause $(x_k \vee x_{k+1})$ will be chosen with some probability. When this clause is chosen, the algorithm flips either $x_k$ or $x_{k+1}$ with probability $\frac{1}{2}$ each.

If $x_{k+1}$ is flipped (from false to true), then $\Phi$ increases by 1. If $x_k$ is flipped (from true to false), then $\Phi$ decreases by 1. Thus, from a state with potential $k$, the algorithm performs a random walk where it moves to $k+1$ with probability at least $\frac{1}{2}$ and to $k-1$ with probability at least $\frac{1}{2}$.

More precisely, the process can be modeled as a biased random walk on $\{0, 1, 2, \ldots, n\}$ where from state $k$ (with $0 < k < n$), we move to $k+1$ with probability $p \geq \frac{1}{2}$ and to $k-1$ with probability $q \geq \frac{1}{2}$, where $p + q$ may be less than 1 if other moves are possible, but the key observation is that forward and backward moves occur with equal probability when the unsatisfied clause at the boundary is selected.

For a symmetric random walk starting at position 0 and aiming to reach position $n$, the expected number of steps is well-known to be $\Theta(n^2)$. Specifically, let $E_k$ denote the expected number of steps to reach $n$ starting from position $k$. We have the recurrence relation:
\[
E_k = 1 + \frac{1}{2} E_{k+1} + \frac{1}{2} E_{k-1}
\]
for $0 < k < n$, with boundary conditions $E_n = 0$ and $E_0 = 1 + E_1$.

Solving this recurrence, we obtain $E_k = (n - k) \cdot n - \frac{(n-k)^2}{2}$ for the symmetric case. In particular, $E_0 = n^2 - \frac{n^2}{2} = \frac{n^2}{2}$.

Even if the random walk is not perfectly symmetric due to other possible moves, the expected time remains $\Omega(n^2)$ as long as the probabilities of moving forward and backward are bounded away from 0 and 1 by constants.

Therefore, there exist satisfiable 2-SAT formulas (such as the chain formula $\phi_n$ described above) for which the expected time for the randomized algorithm to find a satisfying assignment is $\Omega(n^2)$.
\end{proof}

\subsection{Problem 3}

\subsubsection{Problem Statement}
\begin{theorem}
Consider a 1-dimensional random walk with a reflecting barrier, defined as follows. For each natural number $i$, there is a state $i$. At state $0$, with probability $1$ the walk will move to state $1$. At every other state $i > 0$, the walk will move to state $i+1$ with probability $\rho$ and to state $i-1$ with probability $1-\rho$. For the resulting Markov chain, the following hold:

\begin{enumerate}
\item For $\rho > \frac{1}{2}$, each state is transient.
\item For $\rho = \frac{1}{2}$, each state is null persistent.
\item For $\rho < \frac{1}{2}$, each state is non-null persistent.
\end{enumerate}
\end{theorem}

\subsubsection{GPT}
\begin{proof}
We consider a 1-dimensional random walk with a reflecting barrier as described in the theorem. Denote the state of the walk at time $n$ by $X_n \in \mathbb{N}$, where the state at time $0$ is $X_0 = 0$. At each time step, the walk moves as follows:
- From state 0, the walk moves to state 1 with probability 1.
- For $i > 0$, the walk moves to state $i + 1$ with probability $p$ and to state $i - 1$ with probability $1 - p$.

We will now prove each part of the theorem.

\textbf{Part (a): $p > \frac{1}{2}$, each state is transient}

Let $p > \frac{1}{2}$. To show that each state is transient, we need to prove that the random walk will eventually leave any state with probability 1.

The random walk can be viewed as a Markov chain, where the states are the natural numbers $\{0, 1, 2, \dots\}$. The transition probabilities between these states are as defined above. We now investigate the expected number of visits to each state. 

First, observe that the transition probabilities from any state $i > 0$ satisfy:

\[
P(X_{n+1} = i+1 \mid X_n = i) = p \quad \text{and} \quad P(X_{n+1} = i-1 \mid X_n = i) = 1 - p.
\]

Since $p > \frac{1}{2}$, the random walk has a tendency to move right (to higher states) rather than left (to lower states). Intuitively, this bias towards moving right implies that the walk will tend to increase in state index over time, making each state transient. Formally, the drift of the chain is positive, i.e., the expected change in state is positive at each step.

For a random walk with a positive drift, it is known that the chain is transient. Thus, since $p > \frac{1}{2}$, the chain will eventually escape any state with probability 1, proving that each state is transient.

\textbf{Part (b): $p = \frac{1}{2}$, each state is null persistent}

Next, consider the case where $p = \frac{1}{2}$. In this case, the random walk is symmetric, meaning that from any state $i > 0$, the probability of moving to the left is equal to the probability of moving to the right:

\[
P(X_{n+1} = i+1 \mid X_n = i) = \frac{1}{2} \quad \text{and} \quad P(X_{n+1} = i-1 \mid X_n = i) = \frac{1}{2}.
\]

For symmetric random walks, it is well-known that the chain is null persistent, meaning that each state is visited infinitely often with probability 1, but the chain does not return to any state with probability 1 (i.e., it is not recurrent). More precisely, for a symmetric random walk on the integers, the expected number of visits to any state is infinite, but the probability of ever returning to a specific state is 1. Hence, each state is null persistent.

\textbf{Part (c): $p < \frac{1}{2}$, each state is non-null persistent}

Finally, consider the case where $p < \frac{1}{2}$. In this case, the random walk has a tendency to move left (to lower states) rather than right (to higher states). Specifically, for $i > 0$, we have:

\[
P(X_{n+1} = i-1 \mid X_n = i) = 1 - p \quad \text{and} \quad P(X_{n+1} = i+1 \mid X_n = i) = p.
\]

Since $1 - p > p$, the random walk has a negative drift, and thus the walk tends to move toward the origin. This negative drift implies that the random walk is persistent in the sense that, starting from any state $i$, there is a positive probability of returning to state 0 and continuing to visit each state infinitely often.

In more formal terms, for a random walk with a negative drift, the walk will return to the origin with probability 1. This ensures that the chain is non-null persistent, meaning that each state is visited infinitely often with probability 1, and the chain is recurrent.

Thus, for $p < \frac{1}{2}$, each state is non-null persistent.

This completes the proof of the theorem.
\end{proof}

\subsubsection{Grok}

\subsubsection{Gemini}
\begin{proof}
Let $S = \{0, 1, 2, \dots\}$ be the state space of the Markov chain $\{X_n\}_{n \ge 0}$. The transition probabilities are given by $P_{0,1} = 1$, and for $i \ge 1$, $P_{i, i+1} = p$ and $P_{i, i-1} = 1-p$. Let $q = 1-p$. Since $p > 0$ and $q > 0$ (implied by the cases $p \neq 0, 1$), it is possible to reach any state from any other state. Thus, the chain is irreducible. Consequently, all states belong to the same class and share the same recurrence and transience properties. We analyze the classification by examining state $0$.

Let $f_{00} = \mathbb{P}(X_n = 0 \text{ for some } n \ge 1 \mid X_0 = 0)$ be the return probability to state $0$. Since $P_{0,1} = 1$, we have $f_{00} = \mathbb{P}(\text{ever reach } 0 \mid X_0 = 1)$. Let $h_i$ denote the probability of ever reaching state $0$ starting from state $i$, i.e., $h_i = \mathbb{P}(\exists n \ge 0, X_n = 0 \mid X_0 = i)$. We have the boundary condition $h_0 = 1$. For $i \ge 1$, conditioning on the first step yields the difference equation:
\[
h_i = p h_{i+1} + q h_{i-1}.
\]
This is a linear homogeneous recurrence relation with constant coefficients. The characteristic equation is $p r^2 - r + q = 0$. The roots are:
\[
r = \frac{1 \pm \sqrt{1 - 4pq}}{2p} = \frac{1 \pm \sqrt{1 - 4p(1-p)}}{2p} = \frac{1 \pm \sqrt{(2p-1)^2}}{2p} = \frac{1 \pm |2p-1|}{2p}.
\]
The two roots are $r_1 = 1$ and $r_2 = \frac{q}{p}$. The general solution for $h_i$ depends on whether $r_1 = r_2$. We seek the minimal non-negative solution such that $0 \le h_i \le 1$.

Consider case (a): $p > \frac{1}{2}$.
Here $p > q$, so $\frac{q}{p} < 1$. The roots are distinct. The general solution is $h_i = A \cdot 1^i + B \cdot \left(\frac{q}{p}\right)^i$. Since we require $0 \le h_i \le 1$ and $\lim_{i \to \infty} h_i = 0$ (as the drift is positive, the walk tends to infinity), we must have $A=0$. Using $h_0 = 1$, we find $B=1$. Thus, $h_i = \left(\frac{q}{p}\right)^i$.
The return probability is $f_{00} = h_1 = \frac{q}{p}$. Since $p > \frac{1}{2}$, $q < p$, implying $f_{00} < 1$. Therefore, state $0$ is transient. By irreducibility, all states are transient.

Consider cases (b) and (c): $p \le \frac{1}{2}$.
Here $q \ge p$, so $\frac{q}{p} \ge 1$.
If $p < \frac{1}{2}$, the roots are distinct with $r_2 > 1$. The general solution is $h_i = A + B(\frac{q}{p})^i$. For $h_i$ to be bounded in $[0,1]$, we must have $B=0$. With $h_0=1$, we get $A=1$, so $h_i = 1$ for all $i$.
If $p = \frac{1}{2}$, we have a repeated root $r=1$. The general solution is $h_i = A + Bi$. For boundedness, $B=0$. With $h_0=1$, $h_i=1$.
In both cases, $f_{00} = h_1 = 1$. Thus, state $0$ is recurrent (persistent). By irreducibility, all states are persistent.

To distinguish between null and non-null persistence, we examine the existence of a stationary distribution $\pi = (\pi_0, \pi_1, \dots)$. The chain is a birth-death process, which is reversible. The detailed balance equations are $\pi_i P_{i,j} = \pi_j P_{j,i}$.
For the transition between $0$ and $1$:
\[
\pi_0 P_{0,1} = \pi_1 P_{1,0} \implies \pi_0 \cdot 1 = \pi_1 \cdot q \implies \pi_1 = \frac{1}{q} \pi_0.
\]
For transitions between $i$ and $i+1$ where $i \ge 1$:
\[
\pi_i P_{i, i+1} = \pi_{i+1} P_{i+1, i} \implies \pi_i \cdot p = \pi_{i+1} \cdot q \implies \pi_{i+1} = \frac{p}{q} \pi_i.
\]
Let $\rho = \frac{p}{q}$. By induction, for $k \ge 1$:
\[
\pi_k = \rho^{k-1} \pi_1 = \rho^{k-1} \frac{1}{q} \pi_0.
\]
A stationary distribution exists if and only if $\sum_{k=0}^\infty \pi_k < \infty$.
\[
\sum_{k=0}^\infty \pi_k = \pi_0 + \sum_{k=1}^\infty \frac{1}{q} \rho^{k-1} \pi_0 = \pi_0 \left( 1 + \frac{1}{q} \sum_{j=0}^\infty \rho^j \right).
\]

Consider case (b): $p = \frac{1}{2}$.
Then $q = \frac{1}{2}$ and $\rho = 1$. The sum becomes $\pi_0 (1 + 2 \sum_{j=0}^\infty 1)$, which diverges. Thus, no stationary distribution exists. Since the states are persistent but possess no stationary distribution, they are null persistent.

Consider case (c): $p < \frac{1}{2}$.
Then $p < q$, so $\rho = \frac{p}{q} < 1$. The geometric series $\sum_{j=0}^\infty \rho^j$ converges to $\frac{1}{1-\rho}$.
The total sum is finite:
\[
\sum_{k=0}^\infty \pi_k = \pi_0 \left( 1 + \frac{1}{q} \frac{1}{1 - p/q} \right) = \pi_0 \left( 1 + \frac{1}{q-p} \right) < \infty.
\]
We can normalize $\pi$ by choosing $\pi_0$ such that the sum is $1$. Since a stationary distribution exists, the recurrent states are non-null persistent (positive recurrent).

Summary:
(a) $p > 1/2 \implies$ transient.
(b) $p = 1/2 \implies$ null persistent.
(c) $p < 1/2 \implies$ non-null persistent.
\end{proof}

\subsubsection{Claude}
\begin{proof}
We analyze the recurrence and transience properties of this reflecting random walk by studying the probabilities of returning to the origin.

\textbf{Preliminary Setup:} Let $P_{ij}$ denote the probability of reaching state $j$ starting from state $i$, and let $f_{ii}$ denote the probability of ever returning to state $i$ starting from state $i$. A state is recurrent if $f_{ii} = 1$ and transient if $f_{ii} < 1$. Among recurrent states, a state is null recurrent if the expected return time is infinite, and positive recurrent (non-null persistent) if the expected return time is finite.

Since the chain is irreducible (all states communicate), all states have the same classification. Therefore, it suffices to analyze state $0$.

\textbf{Analysis of Return Probability:} Consider the probability $P_{10}$ of reaching state $0$ starting from state $1$. For $i \geq 1$, let $P_i = P_{i0}$ be the probability of reaching $0$ from state $i$. We have the recursion:
\[
P_i = \rho P_{i+1} + (1-\rho) P_{i-1} \quad \text{for } i \geq 2
\]
with boundary conditions $P_0 = 1$ and $\lim_{i \to \infty} P_i = 0$ (for transient or null recurrent cases) or $P_i = 1$ for all $i$ (for positive recurrent cases).

The characteristic equation is $\rho r^2 - r + (1-\rho) = 0$, which factors as $\rho r^2 - r + 1 - \rho = 0$. Dividing by $\rho$ (when $\rho \neq 0$), we get $r^2 - \frac{1}{\rho}r + \frac{1-\rho}{\rho} = 0$. The solutions are:
\[
r = \frac{1 \pm \sqrt{1 - 4\rho(1-\rho)}}{2\rho} = \frac{1 \pm \sqrt{(2\rho - 1)^2}}{2\rho} = \frac{1 \pm |2\rho - 1|}{2\rho}
\]

When $\rho > \frac{1}{2}$, we have $r_1 = 1$ and $r_2 = \frac{1-\rho}{\rho} < 1$. The general solution is $P_i = A + B\left(\frac{1-\rho}{\rho}\right)^i$. Using $P_0 = 1$ gives $A + B = 1$. For the solution to remain bounded as $i \to \infty$, we need $A = 0$ and $B = 1$ (but this violates $P_0 = 1$). Instead, the bounded solution with $\lim_{i \to \infty} P_i = 0$ gives $A = 0$, so $P_i = \left(\frac{1-\rho}{\rho}\right)^i$. Thus $P_1 = \frac{1-\rho}{\rho} < 1$. Since $f_{00} = P_1 = \frac{1-\rho}{\rho} < 1$, state $0$ is transient.

When $\rho = \frac{1}{2}$, the characteristic equation becomes $r^2 - 2r + 1 = 0$, giving the repeated root $r = 1$. The general solution is $P_i = A + Bi$. Using $P_0 = 1$ gives $A = 1$. For a bounded solution, we need $B = 0$, yielding $P_i = 1$ for all $i$. Thus $f_{00} = P_1 = 1$, so state $0$ is recurrent.

When $\rho < \frac{1}{2}$, we have $r_1 = 1$ and $r_2 = \frac{1-\rho}{\rho} > 1$. The general solution is $P_i = A + B\left(\frac{1-\rho}{\rho}\right)^i$. For the solution to remain in $[0,1]$, we must have $B = 0$, giving $P_i = A$. From $P_0 = 1$, we get $A = 1$, so $P_i = 1$ for all $i$. Thus $f_{00} = 1$, and state $0$ is recurrent.

\textbf{Classification of Recurrent States:} For $\rho = \frac{1}{2}$, the chain is a symmetric random walk with a reflecting barrier. The expected return time to $0$ starting from $0$ is $\mu_0 = 1 + \sum_{i=1}^{\infty} 2P_{i0} \cdot i$. For the symmetric case, this sum diverges, making the state null recurrent.

For $\rho < \frac{1}{2}$, there exists a stationary distribution. Let $\pi_i$ denote the stationary probability of state $i$. The balance equations give $\pi_0 = (1-\rho)\pi_1$ and for $i \geq 1$, $\pi_i = \rho \pi_{i-1} + (1-\rho)\pi_{i+1}$. Solving these equations yields $\pi_i = \pi_0 \left(\frac{\rho}{1-\rho}\right)^i$. Since $\rho < \frac{1}{2}$, we have $\frac{\rho}{1-\rho} < 1$, so $\sum_{i=0}^{\infty} \pi_i = \pi_0 \frac{1}{1 - \frac{\rho}{1-\rho}} = \pi_0 \frac{1-\rho}{1-2\rho} < \infty$. Normalizing gives a proper stationary distribution, implying positive recurrence (non-null persistence).

Therefore, all three cases are established.
\end{proof}

\subsection{Problem 4}

\subsubsection{Problem Statement}
\begin{problem}
Consider a Markov chain with the states $0, 1, \ldots, N$. This Markov chain induces a sequence of random variables $X_0, X_1, \ldots$, each of which takes an integer value between $0$ and $N$, i.e., $X_t$ is the state at time $t$. Suppose this sequence of random variables forms a martingale.

\textbf{(a)} A state $q$ is said to be an \emph{absorbing state} if the transition probability $P_{qq} = 1$. Identify all the absorbing states and the transient states of this Markov chain.

\textbf{(b)} Given that the initial state of this Markov chain is $i$, compute the probability of being absorbed into each of the absorbing states.
\end{problem}

\subsubsection{GPT}
\begin{proof}
Let the Markov chain be denoted by $\{X_t\}_{t \geq 0}$ with state space $\{0, 1, \ldots, N\}$ and transition matrix $P = \{P_{ij}\}$, where $P_{ij}$ represents the transition probability from state $i$ to state $j$. Given that the sequence of random variables $X_t$ forms a martingale, we know that for all $t \geq 0$,

\[
\mathbb{E}[X_{t+1} \mid X_t] = X_t.
\]

This martingale property will play a crucial role in analyzing the absorption probabilities.

\textbf{(a) Absorbing and Transient States:}

A state $q \in \{0, 1, \ldots, N\}$ is an \emph{absorbing state} if the transition probability from $q$ to itself is 1, i.e., $P_{qq} = 1$. In other words, once the Markov chain enters state $q$, it remains in state $q$ forever. The set of all absorbing states is denoted by $\mathcal{A} = \{q \in \{0, 1, \ldots, N\} : P_{qq} = 1\}$. The set of transient states, which are states that are not absorbing, is denoted by $\mathcal{T} = \{0, 1, \ldots, N\} \setminus \mathcal{A}$.

\textbf{(b) Absorption Probabilities:}

We now focus on computing the probability of absorption into an absorbing state $q \in \mathcal{A}$, given that the initial state is $X_0 = i$. We define the absorption probability from state $i$ to state $q$ as

\[
h_i(q) = \mathbb{P}(\lim_{t \to \infty} X_t = q \mid X_0 = i).
\]

For $i \in \mathcal{T}$, the probability $h_i(q)$ satisfies the following system of equations:

\[
h_i(q) = \sum_{j=0}^{N} P_{ij} h_j(q) \quad \text{for} \quad i \in \mathcal{T}.
\]

This is because at each time step, the chain can transition to any state $j$ with probability $P_{ij}$, and then the absorption probability from $j$ to $q$ is given by $h_j(q)$.

For $i \in \mathcal{A}$, the probability is straightforward. If $i = q \in \mathcal{A}$, then clearly

\[
h_q(q) = 1,
\]

since once the chain reaches state $q$, it remains there with probability 1.

Thus, the system of equations for the absorption probabilities is:

\[
h_i(q) = \sum_{j=0}^{N} P_{ij} h_j(q) \quad \text{for} \quad i \in \mathcal{T}, \quad \text{with} \quad h_q(q) = 1 \quad \text{for} \quad q \in \mathcal{A}.
\]

This is a system of linear equations that can be solved using standard methods for solving systems of linear equations, such as Gaussian elimination or matrix inversion. The solution to this system gives the absorption probabilities for all transient states $i \in \mathcal{T}$.

In summary, for a Markov chain with absorbing states, the absorption probabilities are governed by the system of equations given above. The solution provides the desired probabilities of absorption into each absorbing state $q \in \mathcal{A}$, conditional on the initial state $X_0 = i$.
\end{proof}

\subsubsection{Grok}
\begin{proof}
(a) The transition probabilities of the Markov chain are defined as follows: $P_{00} = 1$, $P_{NN} = 1$, and for each $i = 1, \dots, N-1$, $P_{i,i-1} = \frac{1}{2}$ and $P_{i,i+1} = \frac{1}{2}$. These probabilities ensure that $\{X_t\}_{t \geq 0}$ is a martingale, since for $i = 0$ and $i = N$, $\mathbb{E}[X_{t+1} \mid X_t = i] = i$, and for $i = 1, \dots, N-1$,
\[
\mathbb{E}[X_{t+1} \mid X_t = i] = \frac{1}{2}(i-1) + \frac{1}{2}(i+1) = i.
\]
A state $q$ is absorbing if $P_{qq} = 1$. Thus, the absorbing states are $0$ and $N$. The remaining states $\{1, \dots, N-1\}$ are transient, since from any such state $i$, there is a positive probability of reaching either $0$ or $N$ in finite time, and once absorbed, the chain remains there.

(b) Let $\pi_i^{(0)}$ denote the probability of absorption in state $0$ starting from $X_0 = i$, and let $\pi_i^{(N)} = 1 - \pi_i^{(0)}$ denote the probability of absorption in state $N$. The boundary conditions are $\pi_0^{(0)} = 1$ and $\pi_N^{(0)} = 0$. For $i = 1, \dots, N-1$, the probabilities satisfy the system of linear equations
\[
\pi_i^{(0)} = \frac{1}{2} \pi_{i-1}^{(0)} + \frac{1}{2} \pi_{i+1}^{(0)},
\]
obtained by conditioning on the first step. This is a second-order linear homogeneous recurrence relation with constant coefficients. The characteristic equation is $r^2 - 2r + 1 = 0$, or $(r-1)^2 = 0$, which has a repeated root $r = 1$. Thus, the general solution is
\[
\pi_i^{(0)} = a + b i
\]
for constants $a, b \in \mathbb{R}$. Applying the boundary conditions yields $a = 1$ and $b = -\frac{1}{N}$, so
\[
\pi_i^{(0)} = 1 - \frac{i}{N}, \quad \pi_i^{(N)} = \frac{i}{N}.
\]
Alternatively, since $\{X_t\}_{t \geq 0}$ is a martingale and the state space is finite, the absorption time $\tau = \min\{t \geq 0 : X_t \in \{0, N\}\}$ is finite almost surely. By the optional stopping theorem,
\[
\mathbb{E}[X_\tau \mid X_0 = i] = i.
\]
At time $\tau$, $X_\tau = 0$ with probability $\pi_i^{(0)}$ and $X_\tau = N$ with probability $\pi_i^{(N)}$, so
\[
0 \cdot \pi_i^{(0)} + N \cdot \pi_i^{(N)} = i \implies \pi_i^{(N)} = \frac{i}{N}, \quad \pi_i^{(0)} = 1 - \frac{i}{N}.
\]
This confirms the solution.
\end{proof}

\subsubsection{Gemini}
\begin{proof}
\textbf{Part (a): Classification of States}

We first determine the absorbing states by utilizing the martingale property and the boundedness of the state space. By definition, the sequence $X_t$ is a martingale, satisfying the property $\mathbb{E}[X_{t+1} \mid X_t = j] = j$ for all $j \in \{0, 1, \ldots, N\}$.

Consider the state $j=0$. The martingale property implies $\sum_{k=0}^N k P_{0k} = 0$. Since the state space consists of non-negative integers, $k \ge 0$ and $P_{0k} \ge 0$ for all $k$. The sum of non-negative terms equals zero if and only if each term is zero. Thus, $k P_{0k} = 0$ for all $k$. For $k > 0$, this necessitates $P_{0k} = 0$. Since $\sum_{k=0}^N P_{0k} = 1$, it follows that $P_{00} = 1$. Therefore, $0$ is an absorbing state.

Consider the state $j=N$. The martingale property implies $\mathbb{E}[X_{t+1} \mid X_t = N] = N$, which can be written as $\sum_{k=0}^N k P_{Nk} = N$. Subtracting $N$ from both sides, we obtain $\sum_{k=0}^N (k - N) P_{Nk} = 0$. Since the state space is bounded by $N$, we have $k \le N$, implying $(k - N) \le 0$. Thus, the sum consists of non-positive terms. For the sum to be zero, each term must be zero. Consequently, $(k-N)P_{Nk} = 0$ for all $k$. For any $k < N$, we must have $P_{Nk} = 0$. Since $\sum_{k=0}^N P_{Nk} = 1$, it follows that $P_{NN} = 1$. Therefore, $N$ is an absorbing state.

For the interior states $j \in \{1, \ldots, N-1\}$, in the context of the Markov Chain Absorption Problem where the process is not degenerate (i.e., not the identity martingale on the interior), the chain eventually exits the interior. By the Martingale Convergence Theorem, since $X_t$ is bounded, it converges almost surely to a limit random variable $X_\infty$. Since the state space is discrete, $X_t$ must eventually become constant. The only states where the chain can remain constant indefinitely are the absorbing states. Thus, the set of absorbing states is $\{0, N\}$, and the set of transient states is $\{1, \ldots, N-1\}$.

\textbf{Part (b): Absorption Probabilities}

Let $T$ be the time of absorption, defined as the stopping time $T = \min \{t \ge 0 : X_t \in \{0, N\}\}$. Given that the states $\{1, \ldots, N-1\}$ are transient, $P(T < \infty) = 1$. We seek to compute the probability of being absorbed into state $N$ starting from state $i$, denoted as $P(X_T = N \mid X_0 = i)$, and the probability of being absorbed into state $0$, denoted as $P(X_T = 0 \mid X_0 = i)$.

Since $X_t$ is a martingale and the state space is bounded (specifically $0 \le X_t \le N$ for all $t$), the conditions for Doob's Optional Stopping Theorem are satisfied. The theorem states that $\mathbb{E}[X_T] = \mathbb{E}[X_0]$.

Given the initial condition $X_0 = i$, we have $\mathbb{E}[X_0] = i$. The random variable $X_T$ takes values only in the set of absorbing states $\{0, N\}$. We can express the expectation of $X_T$ as:
\[
\mathbb{E}[X_T] = 0 \cdot P(X_T = 0 \mid X_0 = i) + N \cdot P(X_T = N \mid X_0 = i).
\]
Let $\pi_N = P(X_T = N \mid X_0 = i)$. The equation simplifies to:
\[
\mathbb{E}[X_T] = N \pi_N.
\]
By the Optional Stopping Theorem, we equate the expectations:
\[
i = N \pi_N.
\]
Solving for $\pi_N$, we find the probability of absorption into state $N$ is:
\[
P(X_T = N \mid X_0 = i) = \frac{i}{N}.
\]
Since the process must be absorbed into either $0$ or $N$ almost surely, the probability of absorption into state $0$ is:
\[
P(X_T = 0 \mid X_0 = i) = 1 - \frac{i}{N} = \frac{N-i}{N}.
\]
\end{proof}

\subsubsection{Claude}
\begin{proof}
We consider a Markov chain on the state space $\{0, 1, \ldots, N\}$ such that the sequence of random variables $(X_t)_{t \geq 0}$ forms a martingale with respect to its natural filtration.

\textbf{Part (a):} We identify all absorbing and transient states.

Since $(X_t)_{t \geq 0}$ is a martingale, we have the fundamental property that for all $t \geq 0$,
\[
\mathbb{E}[X_{t+1} \mid X_0, X_1, \ldots, X_t] = X_t.
\]

By the Markov property, this reduces to
\[
\mathbb{E}[X_{t+1} \mid X_t] = X_t.
\]

For any state $i \in \{0, 1, \ldots, N\}$, we have
\[
\mathbb{E}[X_{t+1} \mid X_t = i] = \sum_{j=0}^{N} j \cdot P_{ij} = i,
\]
where $P_{ij}$ denotes the transition probability from state $i$ to state $j$.

Now, consider the boundary states $0$ and $N$. For state $0$, we have
\[
\sum_{j=0}^{N} j \cdot P_{0j} = 0.
\]

Since $P_{0j} \geq 0$ for all $j$ and $\sum_{j=0}^{N} P_{0j} = 1$, and since $j \geq 0$ for all $j$, the only way this sum can equal $0$ is if $P_{0j} = 0$ for all $j > 0$. Therefore, $P_{00} = 1$, and state $0$ is absorbing.

Similarly, for state $N$, we have
\[
\sum_{j=0}^{N} j \cdot P_{Nj} = N.
\]

This can be rewritten as
\[
\sum_{j=0}^{N} j \cdot P_{Nj} = N \sum_{j=0}^{N} P_{Nj} = N,
\]
which gives
\[
\sum_{j=0}^{N} (N - j) P_{Nj} = 0.
\]

Since $P_{Nj} \geq 0$ and $N - j \geq 0$ for all $j \leq N$, with equality $N - j = 0$ only when $j = N$, we must have $P_{Nj} = 0$ for all $j < N$. Therefore, $P_{NN} = 1$, and state $N$ is absorbing.

For any interior state $i \in \{1, 2, \ldots, N-1\}$, the martingale condition gives
\[
\sum_{j=0}^{N} j \cdot P_{ij} = i.
\]

If state $i$ were absorbing, then $P_{ii} = 1$, which would immediately satisfy the martingale condition. However, we claim that no interior state can be absorbing in a non-trivial martingale on this bounded state space. If an interior state $i$ were absorbing, the chain would never reach the boundary states $0$ or $N$ from state $i$, contradicting the irreducibility structure implied by the martingale property on a bounded interval. More rigorously, since the state space is finite and connected through the martingale dynamics, and we have already established that $0$ and $N$ are absorbing, any interior state must be transient, as the chain will eventually be absorbed at one of the boundaries.

Therefore, the absorbing states are $0$ and $N$, and all states $1, 2, \ldots, N-1$ are transient.

\textbf{Part (b):} We compute the absorption probabilities.

Let $u_i$ denote the probability of being absorbed at state $N$ given that the initial state is $i$, for $i \in \{0, 1, \ldots, N\}$. By definition, $u_0 = 0$ and $u_N = 1$.

Since $(X_t)_{t \geq 0}$ is a martingale and the state space is finite, the Optional Stopping Theorem applies to the stopping time $\tau = \inf\{t \geq 0 : X_t \in \{0, N\}\}$. Since the chain must eventually be absorbed at either $0$ or $N$, we have $\mathbb{P}(\tau < \infty) = 1$.

By the Optional Stopping Theorem,
\[
\mathbb{E}[X_\tau \mid X_0 = i] = \mathbb{E}[X_0 \mid X_0 = i] = i.
\]

At absorption, $X_\tau \in \{0, N\}$. Thus,
\[
\mathbb{E}[X_\tau \mid X_0 = i] = 0 \cdot \mathbb{P}(X_\tau = 0 \mid X_0 = i) + N \cdot \mathbb{P}(X_\tau = N \mid X_0 = i) = N \cdot u_i.
\]

Therefore,
\[
N \cdot u_i = i,
\]
which gives
\[
u_i = \frac{i}{N}.
\]

The probability of being absorbed at state $0$ given initial state $i$ is
\[
1 - u_i = 1 - \frac{i}{N} = \frac{N - i}{N}.
\]

Therefore, starting from state $i$, the probability of absorption at state $N$ is $\frac{i}{N}$, and the probability of absorption at state $0$ is $\frac{N-i}{N}$.
\end{proof}

\subsection{Problem 5}

\subsubsection{Problem Statement}
\begin{theorem}[McDiarmid]
Let $G$ be a 3-colorable graph. Consider the following algorithm for coloring the vertices of $G$ with 2 colors so that no triangle of $G$ is monochromatic. The algorithm begins with an arbitrary 2-coloring of $G$. While there is a monochromatic triangle in $G$, it chooses one such triangle, and changes the color of a randomly chosen vertex of that triangle. Then the expected number of recoloring steps before the algorithm finds a 2-coloring with the desired property has an upper bound that depends only on the number of vertices in $G$.
\end{theorem}

\subsubsection{GPT}
\begin{proof}
Fix a proper $3$-coloring $\varphi:V(G)\to\{1,2,3\}$ of $G$, which exists by hypothesis. For a $2$-coloring $\psi:V(G)\to\{0,1\}$ and a $3$-set $S=\{x,y,z\}\subseteq V(G)$, call $S$ \emph{$\varphi$-rainbow} if $\varphi(x),\varphi(y),\varphi(z)$ are pairwise distinct. Define the potential
\[
F(\psi)=\bigl|\{S\subseteq V(G): |S|=3,\; S \text{ is $\varphi$-rainbow and }\psi(x)=\psi(y)=\psi(z)\}\bigr|.
\]
Every triangle of $G$ is $\varphi$-rainbow (because its three vertices are pairwise adjacent and thus receive three distinct $\varphi$-colors), hence every monochromatic triangle of $G$ is counted by $F(\psi)$. Consequently, if $F(\psi)=0$, then $G$ has no monochromatic triangle under $\psi$. Also $F(\psi)\le \binom{n}{3}$, since it counts a subset of all $3$-subsets of $V(G)$.

Consider the random process described in the statement, starting from an arbitrary $2$-coloring $\psi_0$, and let $\psi_t$ denote the coloring after $t$ recoloring steps (so $\psi_{t+1}$ is obtained from $\psi_t$ by choosing some monochromatic triangle $\tau_t=\{u_t,v_t,w_t\}$ of $G$ under $\psi_t$, choosing a vertex $X_t\in\tau_t$ uniformly at random, and flipping its color). Let $\mathcal{F}_t$ be the filtration generated by $\psi_0,\dots,\psi_t$.

For a vertex $v$ with $\varphi(v)=i\in\{1,2,3\}$, let $P(v)$ denote the family of unordered pairs $\{x,y\}$ with $\varphi(x),\varphi(y)\in\{1,2,3\}\setminus\{i\}$ and $\varphi(x)\neq\varphi(y)$. Equivalently, $P(v)$ is the set of pairs $(x,y)$ such that $\{v,x,y\}$ is $\varphi$-rainbow. Write
\[
A_\psi(v)=\bigl|\{\{x,y\}\in P(v): \psi(x)=\psi(y)=\psi(v)\}\bigr|.
\]
Then each $\varphi$-rainbow triple $S=\{x,y,z\}$ is counted exactly once in $A_\psi(x)+A_\psi(y)+A_\psi(z)$ for each of its three vertices, hence
\[
F(\psi)=\tfrac{1}{3}\sum_{v\in V(G)} A_\psi(v).
\]

Fix $t$ and condition on $\mathcal{F}_t$. Let $\tau_t=\{u,v,w\}$ be the monochromatic triangle chosen at time $t$ under $\psi_t$, so $\psi_t(u)=\psi_t(v)=\psi_t(w)$. Because $\tau_t$ is a triangle of $G$, it is $\varphi$-rainbow. For any vertex $x\notin \{u,v,w\}$, the value $A_\psi(x)$ depends on $\psi$ only through the color of $x$ and of vertices paired with $x$ in $P(x)$. Thus, when passing from $\psi_t$ to $\psi_{t+1}$ by flipping a uniformly random $X_t\in\{u,v,w\}$, the only terms $A_\psi(\cdot)$ that can change are those at $X_t$ and at the vertices that form a pair with $X_t$ in $P(X_t)$. However, $F(\psi)$ aggregates these terms via the identity above, and it will be convenient to track the change directly at the level of $\varphi$-rainbow triples containing $X_t$.

For a fixed vertex $x\in\{u,v,w\}$, denote by $\mathcal{R}(x)=\{\{x,y,z\}: \{x,y,z\}\text{ is $\varphi$-rainbow}\}$ the set of $\varphi$-rainbow triples containing $x$. Flipping the color of $x$ toggles the monochromatic status of each triple in $\mathcal{R}(x)$, because the other two vertices in any such triple retain their colors. Therefore, for this $x$,
\[
\Delta_x F := F(\text{flip $x$ in }\psi_t)-F(\psi_t)
= \bigl|\mathcal{R}(x)\bigr| - 2\,\#\{\{x,y,z\}\in \mathcal{R}(x): \psi_t(x)=\psi_t(y)=\psi_t(z)\}.
\]
In particular, since $\{u,v,w\}\in\mathcal{R}(x)$ and $\psi_t(u)=\psi_t(v)=\psi_t(w)$, the contribution of the chosen triangle $\tau_t$ to the last count is $1$, so
\[
\Delta_x F \le \bigl|\mathcal{R}(x)\bigr| - 2 - 2\,\#\bigl(\mathcal{R}(x)\setminus\{\{u,v,w\}\}\text{ monochromatic under }\psi_t\bigr).
\]
Now take the expectation over the uniform random choice $X_t\in\{u,v,w\}$. Using linearity of expectation and symmetry among $u,v,w$ for the selected triangle, we obtain
\[
\mathbb{E}\bigl[F(\psi_{t+1})-F(\psi_t)\mid \mathcal{F}_t\bigr]
=\tfrac{1}{3}\sum_{x\in\{u,v,w\}} \Delta_x F.
\]
The key observation is that, for triples different from $\tau_t$, toggling events induced by the three possible flips balance out in expectation: for any $\varphi$-rainbow triple $S\neq \tau_t$, either $S$ is disjoint from $\tau_t$ and never changes, or $S$ meets $\tau_t$ in exactly one vertex $x$, in which case $S$ toggles if and only if $X_t=x$, which happens with probability $1/3$. Hence, summing over all such $S$, the expected increase in their number equals the expected decrease, and their net expected contribution to $\tfrac{1}{3}\sum_{x\in\{u,v,w\}} \Delta_x F$ is $0$. By contrast, the triangle $\tau_t$ itself is monochromatic under $\psi_t$ and ceases to be monochromatic after any of the three flips, contributing exactly $-1$ to $\Delta_x F$ for each $x\in\{u,v,w\}$. Therefore
\[
\mathbb{E}\bigl[F(\psi_{t+1})-F(\psi_t)\mid \mathcal{F}_t\bigr]\le -1.
\]

Define the process $M_t=F(\psi_t)+t$. The inequality above shows that $(M_t)_{t\ge 0}$ is a supermartingale with respect to $(\mathcal{F}_t)_{t\ge 0}$. Let $T$ be the (a.s. finite) stopping time when the algorithm first reaches a coloring with no monochromatic triangle of $G$. At time $T$ we have $F(\psi_T)=0$. By the optional stopping theorem for supermartingales with bounded increments (note that $|F(\psi_{t+1})-F(\psi_t)|\le \binom{n}{3}$, so the hypotheses are satisfied), we obtain
\[
\mathbb{E}[M_T]\le \mathbb{E}[M_0]=F(\psi_0).
\]
Since $M_T=F(\psi_T)+T=T$, it follows that
\[
\mathbb{E}[T]\le F(\psi_0)\le \binom{n}{3}.
\]
This proves that the expected number of recoloring steps before the algorithm finds a $2$-coloring with no monochromatic triangle in $G$ is at most $\binom{n}{3}$.
\end{proof}

\subsubsection{Grok}

\subsubsection{Gemini}
\begin{proof}
Let $V$ be the set of vertices of $G$ with $|V| = n$. Since $G$ is 3-colorable, let $\chi: V \to \{1, 2, 3\}$ be a fixed proper 3-coloring of $G$. Let $V_1, V_2, V_3$ be the color classes induced by $\chi$.

We define a set of "target" 2-colorings $\mathcal{Z}$. A 2-coloring $Z: V \to \{0, 1\}$ is in $\mathcal{Z}$ if it is formed by merging two of the color classes $V_1, V_2, V_3$ into one color and assigning the remaining class to the other color. Specifically, for each distinct $i, j, k \in \{1, 2, 3\}$, we define two colorings:
\[
Z_{i,0}(v) = \begin{cases} 0 & \text{if } v \in V_i \\ 1 & \text{if } v \in V_j \cup V_k \end{cases}
\quad \text{and} \quad
Z_{i,1}(v) = \begin{cases} 1 & \text{if } v \in V_i \\ 0 & \text{if } v \in V_j \cup V_k \end{cases}
\]
There are 3 choices for $i$, yielding $|\mathcal{Z}| = 6$. Note that any $Z \in \mathcal{Z}$ contains no monochromatic triangles, as every triangle in $G$ has one vertex in each of $V_1, V_2, V_3$, and thus receives two distinct colors under $Z$.

Let $C_t$ denote the coloring of $G$ at step $t$ of the algorithm. For any $Z \in \mathcal{Z}$, let $A(C_t, Z)$ be the number of vertices where $C_t$ and $Z$ agree:
\[
A(C_t, Z) = \sum_{v \in V} \mathbb{1}(C_t(v) = Z(v)).
\]
We define the potential function $\Phi(C_t)$ as:
\[
\Phi(C_t) = \sum_{Z \in \mathcal{Z}} 2^{A(C_t, Z)}.
\]
Since $0 \le A(C_t, Z) \le n$, the potential is bounded by $0 < \Phi(C_t) \le 6 \cdot 2^n$.

Consider a step $t$ where a monochromatic triangle $T = \{u, v, w\}$ exists in $C_t$. Assume without loss of generality that $C_t(u) = C_t(v) = C_t(w) = 0$. Since $\chi$ is a proper coloring, the vertices $u, v, w$ belong to distinct color classes of $\chi$. Let $u \in V_1, v \in V_2, w \in V_3$.

We partition $\mathcal{Z}$ into two sets based on their restriction to $T$:
\begin{enumerate}
    \item $\mathcal{Z}_{\text{near}}$: Targets that assign color 0 to exactly two vertices of $T$. For example, $Z_{1,1}$ assigns 1 to $V_1$ (vertex $u$) and 0 to $V_2 \cup V_3$ (vertices $v, w$). The restriction is $(1, 0, 0)$. Since $C_t$ is $(0, 0, 0)$ on $T$, there are 2 agreements on $T$. There are exactly 3 such targets (one for each choice of the singleton class receiving color 1).
    \item $\mathcal{Z}_{\text{far}}$: Targets that assign color 0 to exactly one vertex of $T$. For example, $Z_{1,0}$ assigns 0 to $V_1$ (vertex $u$) and 1 to $V_2 \cup V_3$ (vertices $v, w$). The restriction is $(0, 1, 1)$. Since $C_t$ is $(0, 0, 0)$ on $T$, there is 1 agreement on $T$. There are exactly 3 such targets.
\end{enumerate}

The algorithm selects a vertex $x \in \{u, v, w\}$ uniformly at random and flips its color from 0 to 1. Let $C_{t+1}$ be the new coloring. We analyze the expected change in $2^{A(C, Z)}$.

Case 1: $Z \in \mathcal{Z}_{\text{near}}$. $Z$ agrees with $C_t$ on 2 vertices of $T$ (the 0s) and disagrees on 1 (the 1).
\begin{itemize}
    \item With probability $1/3$, we flip the vertex where $C_t$ and $Z$ disagree ($0 \to 1$). The agreement increases by 1.
    \item With probability $2/3$, we flip a vertex where $C_t$ and $Z$ agree ($0 \to 1$). The agreement decreases by 1.
\end{itemize}
\[
\mathbb{E}[2^{A(C_{t+1}, Z)} \mid C_t] = 2^{A(C_t, Z)} \left( \frac{1}{3} \cdot 2^1 + \frac{2}{3} \cdot 2^{-1} \right) = 2^{A(C_t, Z)} \left( \frac{2}{3} + \frac{1}{3} \right) = 2^{A(C_t, Z)}.
\]

Case 2: $Z \in \mathcal{Z}_{\text{far}}$. $Z$ agrees with $C_t$ on 1 vertex of $T$ and disagrees on 2.
\begin{itemize}
    \item With probability $2/3$, we flip a vertex where $C_t$ and $Z$ disagree ($0 \to 1$). The agreement increases by 1.
    \item With probability $1/3$, we flip the vertex where $C_t$ and $Z$ agree ($0 \to 1$). The agreement decreases by 1.
\end{itemize}
\[
\mathbb{E}[2^{A(C_{t+1}, Z)} \mid C_t] = 2^{A(C_t, Z)} \left( \frac{2}{3} \cdot 2^1 + \frac{1}{3} \cdot 2^{-1} \right) = 2^{A(C_t, Z)} \left( \frac{4}{3} + \frac{1}{6} \right) = 2^{A(C_t, Z)} \cdot \frac{3}{2}.
\]

Now we sum over all $Z \in \mathcal{Z}$:
\[
\mathbb{E}[\Phi(C_{t+1}) \mid C_t] = \sum_{Z \in \mathcal{Z}_{\text{near}}} 2^{A(C_t, Z)} + \sum_{Z \in \mathcal{Z}_{\text{far}}} \frac{3}{2} 2^{A(C_t, Z)} = \Phi(C_t) + \frac{1}{2} \sum_{Z \in \mathcal{Z}_{\text{far}}} 2^{A(C_t, Z)}.
\]
Since $A(C_t, Z) \ge 0$, we have $2^{A(C_t, Z)} \ge 1$. As $|\mathcal{Z}_{\text{far}}| = 3$, the drift is bounded by:
\[
\mathbb{E}[\Phi(C_{t+1}) - \Phi(C_t) \mid C_t] = \frac{1}{2} \sum_{Z \in \mathcal{Z}_{\text{far}}} 2^{A(C_t, Z)} \ge \frac{1}{2} \cdot 3 \cdot 1 = \frac{3}{2}.
\]
The potential $\Phi$ is a submartingale with strictly positive drift. Let $\tau$ be the number of steps until a valid 2-coloring is found. Since the process stops when no monochromatic triangle exists, and $\Phi$ is bounded above by $M = 6 \cdot 2^n$, we can bound the expected stopping time. By the optional stopping theorem (or standard additive drift arguments), if $\mathbb{E}[\Delta \Phi] \ge \delta$ and $\Phi_{\max} \le M$, then $\mathbb{E}[\tau] \le \frac{M - \Phi_0}{\delta}$.
Here, $\delta = 1.5$. Thus,
\[
\mathbb{E}[\tau] \le \frac{6 \cdot 2^n}{1.5} = 4 \cdot 2^n.
\]
This value is finite for any finite graph $G$. Thus, the expected number of recoloring steps is upper bounded.
\end{proof}

\subsubsection{Claude}
\begin{proof}
Let $G = (V, E)$ be a 3-colorable graph with $n = |V|$ vertices. Since $G$ is 3-colorable, there exists a proper 3-coloring $\chi: V \to \{1, 2, 3\}$ such that no edge connects two vertices of the same color, and consequently, no triangle in $G$ is monochromatic under $\chi$.

We define a potential function $\Phi$ to measure the progress of the algorithm. For any 2-coloring $c: V \to \{A, B\}$ of $G$, let $\Phi(c)$ be the number of vertices that are colored differently in $c$ compared to a fixed reference coloring derived from $\chi$. Specifically, we partition the three color classes of $\chi$ by assigning colors $\{1, 2\}$ to color $A$ and color $\{3\}$ to color $B$ in our reference 2-coloring $c_0$. Define
\[
\Phi(c) = |\{v \in V : c(v) \neq c_0(v)\}|.
\]

Note that $0 \le \Phi(c) \le n$ for any 2-coloring $c$, and $\Phi(c_0) = 0$. Crucially, the reference coloring $c_0$ has no monochromatic triangles because $\chi$ is a proper 3-coloring of $G$.

Now consider the algorithm's execution. At each step, if the current coloring $c$ contains a monochromatic triangle $T = \{u, v, w\}$, the algorithm selects one vertex uniformly at random from $T$ and recolors it. We analyze the expected change in the potential function.

Let $c'$ denote the coloring after one recoloring step from $c$. We need to show that if $\Phi(c) > 0$, then $\mathbb{E}[\Phi(c') - \Phi(c) \mid c] \le -\delta$ for some constant $\delta > 0$ that depends only on $n$.

Consider a monochromatic triangle $T = \{u, v, w\}$ in the current coloring $c$. Since $c_0$ has no monochromatic triangles and $T$ is monochromatic in $c$, at least one vertex of $T$ must be colored incorrectly in $c$ (that is, differently from $c_0$). In fact, since all three vertices have the same color in $c$ but not all three have the same color in $c_0$, at least one vertex of $T$ satisfies $c(v) \neq c_0(v)$.

When the algorithm recolors a uniformly random vertex from $T$:
\begin{itemize}
\item If the chosen vertex $v$ satisfies $c(v) \neq c_0(v)$, then with probability $1/2$, the new color matches $c_0(v)$, decreasing $\Phi$ by 1. With probability $1/2$, $\Phi$ remains unchanged (if the new color still differs from $c_0(v)$).
\item If the chosen vertex $v$ satisfies $c(v) = c_0(v)$, then recoloring it increases $\Phi$ by 1.
\end{itemize}

Let $k \in \{1, 2, 3\}$ denote the number of vertices in $T$ for which $c(v) \neq c_0(v)$. Since at least one vertex must differ, we have $k \ge 1$. The expected change in $\Phi$ when recoloring a random vertex from $T$ is:
\[
\mathbb{E}[\Phi(c') - \Phi(c) \mid c, T] = \frac{k}{3} \cdot \left(-\frac{1}{2}\right) + \frac{3-k}{3} \cdot 1 = \frac{3 - k - k/2}{3} = \frac{6 - 3k}{6} = 1 - \frac{k}{2}.
\]

Since $k \ge 1$, we have $\mathbb{E}[\Phi(c') - \Phi(c) \mid c, T] \le 1 - \frac{1}{2} = \frac{1}{2}$.

However, to establish that the algorithm terminates in expected polynomial time, we use a more refined argument. The key observation is that the potential function $\Phi$ performs a drift analysis. While individual steps may increase $\Phi$, the expected drift is negative when we are not at the target configuration $c_0$.

By the optional stopping theorem and martingale analysis, the expected number of steps until $\Phi$ reaches 0 (or equivalently, until no monochromatic triangles remain) is bounded by the initial potential divided by the minimum expected decrease per step. Since $\Phi(c) \le n$ initially and each step involving a monochromatic triangle has a bounded expected change, the expected number of recoloring steps is at most $O(n)$.

More precisely, standard results on random walks with drift show that the expected hitting time of 0 starting from any initial configuration is bounded by a function depending only on $n$, completing the proof.
\end{proof}

\subsection{Problem 6}

\subsubsection{Problem Statement}
\begin{definition}
An $n \times n$ matrix $P$ is said to be \textbf{stochastic} if all its entries are non-negative and for each row $i$, $\sum_{j} P_{ij} = 1$. It is said to be \textbf{doubly stochastic} if, in addition, $\sum_{i} P_{ij} = 1$.
\end{definition}

\begin{theorem}
For any stochastic matrix $P$, there exists an $n$-dimensional vector $\boldsymbol{\pi}$ with non-negative entries such that $\sum_{i} \pi_i = 1$ and $\boldsymbol{\pi}^T P = \boldsymbol{\pi}^T$.
\end{theorem}

\begin{theorem}
Suppose that the transition probability matrix $P$ for a Markov chain is doubly stochastic. Then the stationary distribution for this Markov chain is necessarily the uniform distribution.
\end{theorem}

\subsubsection{GPT}
\begin{proof}
(a) Let $P$ be an $n\times n$ stochastic matrix, so $P\mathbf{1}=\mathbf{1}$, where $\mathbf{1}\in\mathbb{R}^n$ denotes the column vector of all ones. Consider the induced operator norm $\|\cdot\|_{\infty}$ on $\mathbb{R}^n$; since each row of $P$ sums to $1$ and has nonnegative entries, we have $\|P\|_{\infty}=1$. Hence the spectral radius satisfies $\rho(P)\le \|P\|_{\infty}=1$. On the other hand, $P\mathbf{1}=\mathbf{1}$ shows that $1$ is an eigenvalue of $P$, so $\rho(P)\ge 1$. Therefore $\rho(P)=1$.

By the Perron–Frobenius theorem for nonnegative matrices, the spectral radius $\rho(P)$ is an eigenvalue of $P^T$, and there exists a nonzero vector $v\in\mathbb{R}^n$ with $v\ge 0$ such that $P^T v=\rho(P)\,v=v$. Define $\pi^T:=\dfrac{v^T}{\mathbf{1}^T v}$, which is well-defined because $v\ge 0$ and $v\neq 0$ imply $\mathbf{1}^T v>0$. Then $\pi\ge 0$, $\sum_{i=1}^n \pi_i = \mathbf{1}^T \pi^T = 1$, and
\[
\pi P = \frac{v^T}{\mathbf{1}^T v} P = \frac{(P^T v)^T}{\mathbf{1}^T v} = \frac{v^T}{\mathbf{1}^T v} = \pi.
\]
Thus there exists a probability vector $\pi$ with $\pi P=\pi$.

(b) Now suppose $P$ is doubly stochastic, i.e., its entries are nonnegative, each row sums to $1$, and each column also sums to $1$. Let $\mu:=\frac{1}{n}\mathbf{1}^T$ be the uniform distribution (a row vector). Since the $j$-th column of $P$ sums to $1$, we have $(\mathbf{1}^T P)_j=\sum_{i=1}^n P_{ij}=1$ for each $j$, hence $\mathbf{1}^T P=\mathbf{1}^T$. Therefore
\[
\mu P=\frac{1}{n}\mathbf{1}^T P=\frac{1}{n}\mathbf{1}^T=\mu,
\]
so the uniform distribution is stationary. Moreover, if the Markov chain with transition matrix $P$ is irreducible, then the stationary distribution is unique; consequently, in that case, the stationary distribution is necessarily the uniform distribution.
\end{proof}

\subsubsection{Grok}
\begin{proof}
(a) Let $S = \{\pi \in \mathbb{R}^n : \pi_i \geq 0 \ \forall \, i=1,\dots,n, \ \sum_{i=1}^n \pi_i = 1\}$ be the standard $(n-1)$-simplex in $\mathbb{R}^n$. The set $S$ is a nonempty compact convex subset of $\mathbb{R}^n$. Define the map $T: S \to S$ by $T(\pi) = \pi P$. To verify that $T$ maps $S$ into itself, observe that for any $\pi \in S$ and $j=1,\dots,n$,
\[
(T(\pi))_j = \sum_{i=1}^n \pi_i P_{ij} \geq 0,
\]
since $\pi_i \geq 0$ and $P_{ij} \geq 0$ for all $i,j$. Moreover,
\[
\sum_{j=1}^n (T(\pi))_j = \sum_{j=1}^n \sum_{i=1}^n \pi_i P_{ij} = \sum_{i=1}^n \pi_i \sum_{j=1}^n P_{ij} = \sum_{i=1}^n \pi_i \cdot 1 = 1.
\]
Thus, $T(\pi) \in S$. The map $T$ is continuous, as it is affine. By the Brouwer fixed point theorem, there exists $\pi \in S$ such that $T(\pi) = \pi$, that is, $\pi P = \pi$.

(b) Let $\pi$ be the uniform distribution, so $\pi_i = 1/n$ for all $i=1,\dots,n$. Then, for each $j=1,\dots,n$,
\[
(\pi P)_j = \sum_{i=1}^n \pi_i P_{ij} = \sum_{i=1}^n \frac{1}{n} P_{ij} = \frac{1}{n} \sum_{i=1}^n P_{ij} = \frac{1}{n} \cdot 1 = \frac{1}{n} = \pi_j.
\]
Thus, $\pi P = \pi$, so $\pi$ is a stationary distribution. By part (a), a stationary distribution exists, and here it is the uniform distribution.
\end{proof}

\subsubsection{Gemini}
\begin{proof}
We provide the proofs for parts (a) and (b) sequentially.

\noindent \textbf{(a)} Let $\mathcal{S}$ denote the standard simplex in $\mathbb{R}^n$, defined as:
\[
\mathcal{S} = \left\{ x \in \mathbb{R}^n \mid x_i \geq 0 \text{ for all } i \in \{1, \ldots, n\}, \text{ and } \sum_{i=1}^n x_i = 1 \right\}.
\]
The set $\mathcal{S}$ is a closed and bounded subset of $\mathbb{R}^n$, and is therefore compact. Additionally, $\mathcal{S}$ is convex. We define a map $T: \mathcal{S} \to \mathbb{R}^n$ by the linear transformation $T(x) = xP$. We first demonstrate that $T$ maps $\mathcal{S}$ into itself. Let $x \in \mathcal{S}$ and let $y = T(x) = xP$. The $j$-th component of $y$ is given by:
\[
y_j = \sum_{i=1}^n x_i P_{ij}.
\]
Since $x \in \mathcal{S}$, we have $x_i \geq 0$ for all $i$. By the definition of a stochastic matrix, $P_{ij} \geq 0$ for all $i, j$. Consequently, $y_j \geq 0$ for all $j$. Next, we compute the sum of the components of $y$:
\[
\sum_{j=1}^n y_j = \sum_{j=1}^n \sum_{i=1}^n x_i P_{ij} = \sum_{i=1}^n x_i \left( \sum_{j=1}^n P_{ij} \right).
\]
Since $P$ is a stochastic matrix, the sum of the entries in each row is 1, i.e., $\sum_{j=1}^n P_{ij} = 1$ for every $i$. Substituting this into the summation yields:
\[
\sum_{j=1}^n y_j = \sum_{i=1}^n x_i (1) = \sum_{i=1}^n x_i = 1.
\]
Thus, $y \in \mathcal{S}$, and $T$ maps $\mathcal{S}$ into $\mathcal{S}$. Since $T$ is a linear map, it is continuous. By the Brouwer Fixed Point Theorem, every continuous function from a compact, convex set into itself has a fixed point. Therefore, there exists a vector $\pi \in \mathcal{S}$ such that $\pi P = \pi$. This vector $\pi$ has non-negative entries that sum to 1, satisfying the existence claim.

\noindent \textbf{(b)} Let $P$ be a doubly stochastic matrix. We consider the uniform distribution vector $\pi$ defined by $\pi_k = \frac{1}{n}$ for all $k \in \{1, \ldots, n\}$. We verify that this vector satisfies the conditions for a stationary distribution. First, the entries are non-negative and sum to unity:
\[
\sum_{k=1}^n \pi_k = \sum_{k=1}^n \frac{1}{n} = n \cdot \frac{1}{n} = 1.
\]
Next, we verify the invariance condition $\pi P = \pi$. Consider the $j$-th component of the vector $\pi P$:
\[
(\pi P)_j = \sum_{i=1}^n \pi_i P_{ij} = \sum_{i=1}^n \frac{1}{n} P_{ij} = \frac{1}{n} \sum_{i=1}^n P_{ij}.
\]
By the definition of a doubly stochastic matrix, the sum of the entries in any column $j$ is equal to 1. Therefore, $\sum_{i=1}^n P_{ij} = 1$. Substituting this into the equation above, we obtain:
\[
(\pi P)_j = \frac{1}{n} (1) = \frac{1}{n} = \pi_j.
\]
This equality holds for all $j \in \{1, \ldots, n\}$, which implies $\pi P = \pi$. Thus, the uniform distribution is a stationary distribution for the Markov chain.
\end{proof}

\subsubsection{Claude}
\begin{proof}
We prove the second theorem assuming the first theorem as a lemma.

Let $P$ be an $n \times n$ doubly stochastic matrix, which is the transition probability matrix for a Markov chain. By the first theorem, since $P$ is stochastic, there exists a vector $\boldsymbol{\pi} = (\pi_1, \pi_2, \ldots, \pi_n)^T$ with $\pi_i \geq 0$ for all $i$, $\sum_{i=1}^{n} \pi_i = 1$, and $\boldsymbol{\pi}^T P = \boldsymbol{\pi}^T$.

We claim that $\boldsymbol{\pi} = \left(\frac{1}{n}, \frac{1}{n}, \ldots, \frac{1}{n}\right)^T$ is the unique stationary distribution.

First, we verify that the uniform distribution is indeed a stationary distribution. Let $\mathbf{u} = \left(\frac{1}{n}, \frac{1}{n}, \ldots, \frac{1}{n}\right)^T$. We need to show that $\mathbf{u}^T P = \mathbf{u}^T$.

For any column $j \in \{1, 2, \ldots, n\}$, the $j$-th component of $\mathbf{u}^T P$ is given by
\[
(\mathbf{u}^T P)_j = \sum_{i=1}^{n} u_i P_{ij} = \sum_{i=1}^{n} \frac{1}{n} P_{ij} = \frac{1}{n} \sum_{i=1}^{n} P_{ij}.
\]

Since $P$ is doubly stochastic, we have $\sum_{i=1}^{n} P_{ij} = 1$ for each column $j$. Therefore,
\[
(\mathbf{u}^T P)_j = \frac{1}{n} \cdot 1 = \frac{1}{n} = u_j.
\]

This holds for all $j$, so $\mathbf{u}^T P = \mathbf{u}^T$, confirming that $\mathbf{u}$ is a stationary distribution.

Next, we show uniqueness. Suppose $\boldsymbol{\pi}$ is any stationary distribution, so $\boldsymbol{\pi}^T P = \boldsymbol{\pi}^T$ with $\pi_i \geq 0$ and $\sum_{i=1}^{n} \pi_i = 1$. Taking the sum over all components, we have
\[
\sum_{j=1}^{n} (\boldsymbol{\pi}^T P)_j = \sum_{j=1}^{n} \pi_j = 1.
\]

On the other hand,
\[
\sum_{j=1}^{n} (\boldsymbol{\pi}^T P)_j = \sum_{j=1}^{n} \sum_{i=1}^{n} \pi_i P_{ij} = \sum_{i=1}^{n} \pi_i \sum_{j=1}^{n} P_{ij} = \sum_{i=1}^{n} \pi_i \cdot 1 = \sum_{i=1}^{n} \pi_i = 1,
\]
where we used the fact that $P$ is stochastic, so $\sum_{j=1}^{n} P_{ij} = 1$ for each row $i$.

Now, for an irreducible and aperiodic Markov chain with doubly stochastic transition matrix, the stationary distribution is unique. In general, for any finite irreducible Markov chain, the stationary distribution is unique, and we have shown that $\mathbf{u}$ is a stationary distribution. Therefore, $\boldsymbol{\pi} = \mathbf{u}$, which is the uniform distribution.

Thus, the stationary distribution for a Markov chain with doubly stochastic transition matrix is necessarily the uniform distribution $\boldsymbol{\pi} = \left(\frac{1}{n}, \frac{1}{n}, \ldots, \frac{1}{n}\right)^T$.
\end{proof}

\subsection{Problem 7}

\subsubsection{Problem Statement}
\begin{theorem}
Consider a random walk on a graph $G$ with $m$ edges that have positive real costs $c_{ij} > 0$ for each edge $(i,j)$, where $c_{ij} = c_{ji}$ and $c_{ii} = 0$. The transition probabilities are given by
\[
P_{ij} = \frac{1/c_{ij}}{\sum_k 1/c_{ik}}.
\]
Let $S_{uv}$ denote the expected total cost incurred by a random walk that begins at vertex $u$ and terminates upon returning to $u$ after having visited $v$ at least once. Then
\[
S_{uv} = 2m R_{uv},
\]
where $R_{uv}$ is the effective resistance between node $u$ and node $v$ in an electrical network whose underlying graph is $G$, and where the branch resistance between $i$ and $j$ is $c_{ij}$.
\end{theorem}

\subsubsection{GPT}
\begin{proof}
We begin by noting that the expected total cost $S_{uv}$ incurred by a random walk starting at vertex $u$ and terminating upon returning to $u$ after visiting $v$ at least once can be expressed in terms of the effective resistance $R_{uv}$ in an electrical network with branch resistances $c_{ij}$ corresponding to the edges $(i, j) \in E$.

Step 1: Relationship between the random walk and electrical network
Consider the random walk described by the transition probabilities $P_{ij}$, which are given by

\[
P_{ij} = \frac{1/c_{ij}}{\sum_{k \in N(i)} 1/c_{ik}},
\]
where $N(i)$ is the set of neighbors of vertex $i$. The random walk has the property that at each step, it moves from vertex $i$ to a neighboring vertex $j$ with probability proportional to the inverse of the edge resistance $c_{ij}$. This defines a random walk on a weighted graph, where the weights are the reciprocals of the edge resistances.

Now, recall that the effective resistance $R_{uv}$ between two nodes $u$ and $v$ in an electrical network with resistances $c_{ij}$ is defined as the potential difference between $u$ and $v$ in the steady state, when a unit current is injected at $u$ and withdrawn from $v$, with all other vertices grounded. The effective resistance can be calculated via the solution of the system of Kirchhoff’s laws or equivalently via the use of the Green’s function for the graph.

Step 2: Connection to the expected cost of the random walk
The expected total cost $S_{uv}$ of a random walk starting at vertex $u$ and ending at $u$ after visiting $v$ at least once corresponds to the expected time spent in the network before returning to the starting vertex. This expected time is directly related to the effective resistance in the network because, in an electrical network, the effective resistance between two nodes quantifies the total amount of "work" needed to transfer a unit charge between these nodes, analogous to the total time taken for a random walk to return to the starting point.

Step 3: Proof using known results from electrical network theory
From electrical network theory, it is known that the expected total cost $S_{uv}$ in a random walk on a graph is related to the effective resistance $R_{uv}$ by the following relationship:

\[
S_{uv} = 2m R_{uv},
\]
where $m$ is the total number of edges in the graph. This result can be derived using the relationship between the Green’s function for the random walk and the effective resistance, as well as the fact that the Green’s function is related to the resistance of a pair of nodes in an electrical network.

The factor of 2 arises from the fact that the random walk is symmetric and the expected return time involves traversing the graph twice — once from $u$ to $v$ and once back from $v$ to $u$. Thus, the total expected cost is proportional to twice the effective resistance, with the constant of proportionality being related to the number of edges in the graph.

Step 4: Conclusion
Therefore, we conclude that the expected total cost $S_{uv}$ of the random walk is exactly twice the number of edges $m$ multiplied by the effective resistance $R_{uv}$ between the vertices $u$ and $v$:

\[
S_{uv} = 2m R_{uv}.
\]
This completes the proof.
\end{proof}

\subsubsection{Grok}

\subsubsection{Gemini}
\begin{proof}
Let $G=(V,E)$ be the graph with $n = |V|$ vertices and $m = |E|$ edges. Let the edge costs be $c_{ij}$ for $(i,j) \in E$. We define the conductance of an edge as $w_{ij} = 1/c_{ij}$. The transition probabilities are given by
\[
P_{ij} = \frac{w_{ij}}{C_i}, \quad \text{where } C_i = \sum_{k \sim i} w_{ik}.
\]
This defines a reversible Markov chain on $V$ associated with the electrical network where edge $(i,j)$ has resistance $c_{ij}$ (and conductance $w_{ij}$). The Laplacian matrix $L$ of this network is defined as
\[
L_{ij} = \begin{cases} C_i & \text{if } i = j, \\ -w_{ij} & \text{if } (i,j) \in E, \\ 0 & \text{otherwise.} \end{cases}
\]
Let $E_i^{(v)}$ denote the expected cost to reach vertex $v$ starting from vertex $i$. By definition, $E_v^{(v)} = 0$. For $i \neq v$, we condition on the first step of the random walk:
\[
E_i^{(v)} = \sum_{j \sim i} P_{ij} \left( c_{ij} + E_j^{(v)} \right).
\]
Substituting $P_{ij} = w_{ij}/C_i$:
\[
E_i^{(v)} = \sum_{j \sim i} \frac{w_{ij}}{C_i} c_{ij} + \sum_{j \sim i} \frac{w_{ij}}{C_i} E_j^{(v)}.
\]
Since $w_{ij} = 1/c_{ij}$, we have $w_{ij} c_{ij} = 1$. The first term becomes:
\[
\sum_{j \sim i} \frac{w_{ij}}{C_i} c_{ij} = \frac{1}{C_i} \sum_{j \sim i} 1 = \frac{d_i}{C_i},
\]
where $d_i$ is the degree of vertex $i$ (the number of edges incident to $i$). Thus, for $i \neq v$:
\[
E_i^{(v)} = \frac{d_i}{C_i} + \sum_{j \sim i} \frac{w_{ij}}{C_i} E_j^{(v)}.
\]
Multiplying by $C_i$:
\[
C_i E_i^{(v)} - \sum_{j \sim i} w_{ij} E_j^{(v)} = d_i.
\]
The left-hand side is exactly the $i$-th component of the matrix-vector product $L \mathbf{E}^{(v)}$, where $\mathbf{E}^{(v)}$ is the vector with components $E_k^{(v)}$. Therefore:
\[
(L \mathbf{E}^{(v)})_i = d_i \quad \text{for all } i \neq v.
\]
Since the rows of the Laplacian sum to zero ($\mathbf{1}^T L = \mathbf{0}^T$), the sum of the components of $L \mathbf{E}^{(v)}$ must be zero. This allows us to determine the value at index $v$:
\[
(L \mathbf{E}^{(v)})_v = - \sum_{i \neq v} (L \mathbf{E}^{(v)})_i = - \sum_{i \neq v} d_i.
\]
Let $\mathbf{d}$ be the vector of degrees $(d_1, \dots, d_n)^T$. The sum of all degrees is $\sum_i d_i = 2m$. Thus, $\sum_{i \neq v} d_i = 2m - d_v$. Substituting this back:
\[
(L \mathbf{E}^{(v)})_v = -(2m - d_v) = d_v - 2m.
\]
We can now write the equation for the full vector $\mathbf{E}^{(v)}$ as:
\[
L \mathbf{E}^{(v)} = \mathbf{d} - 2m \mathbf{e}_v,
\]
where $\mathbf{e}_v$ is the standard basis vector with a 1 at index $v$ and 0 elsewhere.

Let $L^+$ denote the Moore-Penrose pseudoinverse of $L$. The general solution to $L \mathbf{x} = \mathbf{y}$ (where $\mathbf{y} \perp \mathbf{1}$) is $\mathbf{x} = L^+ \mathbf{y} + c \mathbf{1}$ for some constant $c$. Note that $\mathbf{d} - 2m \mathbf{e}_v$ is orthogonal to $\mathbf{1}$ because $\sum d_i - 2m = 0$. Thus:
\[
\mathbf{E}^{(v)} = L^+ (\mathbf{d} - 2m \mathbf{e}_v) + c_v \mathbf{1}.
\]
Using the boundary condition $E_v^{(v)} = 0$:
\[
0 = (L^+ \mathbf{d})_v - 2m L^+_{vv} + c_v \implies c_v = 2m L^+_{vv} - (L^+ \mathbf{d})_v.
\]
The expected cost from $u$ to $v$ is the $u$-th component $E_u^{(v)}$:
\begin{align*}
E_u^{(v)} &= (L^+ \mathbf{d})_u - 2m L^+_{uv} + c_v \\
&= (L^+ \mathbf{d})_u - 2m L^+_{uv} + 2m L^+_{vv} - (L^+ \mathbf{d})_v.
\end{align*}
Similarly, let $E_v^{(u)}$ be the expected cost to reach $u$ starting from $v$. By symmetry of the derivation (swapping $u$ and $v$):
\[
E_v^{(u)} = (L^+ \mathbf{d})_v - 2m L^+_{vu} + 2m L^+_{uu} - (L^+ \mathbf{d})_u.
\]
The total expected cost of the commute tour $S_{uv}$ is the sum of the expected cost to go from $u$ to $v$ and the expected cost to go from $v$ to $u$:
\[
S_{uv} = E_u^{(v)} + E_v^{(u)}.
\]
Substituting the expressions derived above:
\begin{align*}
S_{uv} &= \left[ (L^+ \mathbf{d})_u - 2m L^+_{uv} + 2m L^+_{vv} - (L^+ \mathbf{d})_v \right] \\
&\quad + \left[ (L^+ \mathbf{d})_v - 2m L^+_{vu} + 2m L^+_{uu} - (L^+ \mathbf{d})_u \right].
\end{align*}
Observe that the terms involving $\mathbf{d}$ cancel out: $(L^+ \mathbf{d})_u - (L^+ \mathbf{d})_u = 0$ and $-(L^+ \mathbf{d})_v + (L^+ \mathbf{d})_v = 0$. Since $L$ is symmetric, $L^+$ is symmetric, so $L^+_{vu} = L^+_{uv}$. The expression simplifies to:
\[
S_{uv} = 2m L^+_{vv} + 2m L^+_{uu} - 2m L^+_{uv} - 2m L^+_{uv} = 2m \left( L^+_{uu} + L^+_{vv} - 2 L^+_{uv} \right).
\]
The effective resistance $R_{uv}$ between nodes $u$ and $v$ in the electrical network is given by the standard formula involving the pseudoinverse of the Laplacian:
\[
R_{uv} = (\mathbf{e}_u - \mathbf{e}_v)^T L^+ (\mathbf{e}_u - \mathbf{e}_v) = L^+_{uu} + L^+_{vv} - 2 L^+_{uv}.
\]
Therefore, we conclude:
\[
S_{uv} = 2m R_{uv}.
\]
\end{proof}

\subsubsection{Claude}
\begin{proof}
We establish the relationship between the expected cost $S_{uv}$ of a random walk and the effective resistance $R_{uv}$ in the corresponding electrical network.

Let $d_i = \sum_k 1/c_{ik}$ denote the weighted degree at vertex $i$. The transition probability from vertex $i$ to vertex $j$ is $P_{ij} = \frac{1/c_{ij}}{d_i}$ for $i \neq j$ and $P_{ii} = 0$.

Define the hitting time $H_{uv}$ as the expected number of steps for a random walk starting at $u$ to first reach $v$. Let $C_{uv}$ denote the expected total cost incurred during this hitting process. We can express $C_{uv}$ in terms of the transition probabilities:
\[
C_{uv} = \sum_{j \neq u} P_{uj} c_{uj} + \sum_{j \neq u} P_{uj} C_{jv}.
\]

For the commute time, we have $S_{uv} = C_{uv} + C_{vu}$, where $C_{uv}$ is the expected cost from $u$ to first reach $v$, and $C_{vu}$ is the expected cost from $v$ to return to $u$.

Consider the electrical network interpretation. In the electrical network with conductances $g_{ij} = 1/c_{ij}$ on edges and resistances $c_{ij}$, we apply Kirchhoff's laws. When a unit current is injected at node $u$ and extracted at node $v$, the effective resistance is defined as $R_{uv} = \phi_u - \phi_v$, where $\phi_i$ denotes the potential at node $i$.

By Ohm's law and current conservation, the potential function $\phi$ satisfies:
\[
\sum_{j \sim i} \frac{\phi_i - \phi_j}{c_{ij}} = I_i,
\]
where $I_i$ is the external current at node $i$ (with $I_u = 1$, $I_v = -1$, and $I_i = 0$ for $i \neq u, v$).

We now relate the expected cost to the effective resistance. For any vertex $i \neq v$, the expected cost $C_{iv}$ satisfies the recurrence:
\[
C_{iv} = \sum_{j \neq i} P_{ij} c_{ij} + \sum_{j \neq i} P_{ij} C_{jv} = \sum_{j \neq i} \frac{c_{ij}}{d_i c_{ij}} + \sum_{j \neq i} \frac{1}{d_i c_{ij}} C_{jv}.
\]

This simplifies to:
\[
d_i C_{iv} = \sum_{j \neq i} \frac{1}{c_{ij}} + \sum_{j \neq i} \frac{C_{jv}}{c_{ij}}.
\]

Multiplying by $c_{ij}$ and summing over neighbors, we obtain:
\[
\sum_{j \sim i} \frac{C_{iv} - C_{jv}}{c_{ij}} = 1 \text{ for } i \neq v.
\]

Setting $C_{vv} = 0$, this equation matches the form of Kirchhoff's current law with unit currents flowing from each vertex toward $v$.

For the complete electrical network with total conductance $\sum_{(i,j) \in E} 1/c_{ij} = \sum_{(i,j) \in E} g_{ij}$, the sum over all edges gives $\sum_i d_i = 2\sum_{(i,j) \in E} 1/c_{ij}$. Since the graph has $m$ edges, we have $\sum_i d_i = 2m/\overline{c}$ where appropriate weighting applies.

By the theory of resistive networks and random walks, the commute cost satisfies:
\[
S_{uv} = C_{uv} + C_{vu} = (\phi_u - \phi_v) \cdot \sum_{(i,j) \in E} \frac{2}{c_{ij}}.
\]

The total conductance of the network is $G_{\text{total}} = \sum_{(i,j) \in E} \frac{1}{c_{ij}}$. By symmetry and the connection between random walks and electrical networks, specifically using the commute time identity for resistive networks:
\[
S_{uv} = R_{uv} \cdot \sum_{(i,j) \in E} \frac{2}{c_{ij}} = R_{uv} \cdot 2\sum_{(i,j) \in E} g_{ij}.
\]

Since each edge contributes equally to the conductance sum and there are $m$ edges:
\[
\sum_{(i,j) \in E} g_{ij} = m \cdot \overline{g},
\]
where the normalization from the random walk corresponds to total edge count.

By Foster's theorem and the relationship between commute times and effective resistance in weighted graphs, the expected commute cost equals:
\[
S_{uv} = 2m R_{uv},
\]
where the factor $2m$ arises from the total edge weight in the conductance normalization.
\end{proof}

\subsection{Problem 8}

\subsubsection{Problem Statement}
\begin{theorem}
Let $G = (V, E)$ be a connected graph with $n$ vertices and $m$ edges. Let $(u, v)$ be any edge in $G$. For the simple random walk on $G$, the edge $(u, v)$ is a bridge if and only if
\[
h_{uv} + h_{vu} = 2m
\]
where $h_{uv}$ denotes the expected number of steps in a simple random walk starting at $u$ until the walk traverses the edge $(u, v)$ for the first time, and $h_{vu}$ denotes the expected number of steps starting at $v$ until the walk traverses the edge $(v, u)$ for the first time.
\end{theorem}

\subsubsection{GPT}
\begin{proof}
Let $G=(V,E)$ be a connected undirected graph with $|V|=n$ and $|E|=m$, and let $e=(u,v)\in E$. Denote by $h_{xy}$ the expected hitting time of $y$ starting from $x$ for the simple random walk on $G$. It is a classical identity (the commute-time identity) that for all $x,y\in V$ one has
\[
h_{xy}+h_{yx}=2m\,R_{\mathrm{eff}}(x,y),
\]
where $R_{\mathrm{eff}}(x,y)$ is the effective resistance between $x$ and $y$ in the electrical network obtained from $G$ by placing a unit resistor on each edge.

Applying this identity with $(x,y)=(u,v)$, we reduce the theorem to showing that $R_{\mathrm{eff}}(u,v)=1$ if and only if $(u,v)$ is a bridge. Consider the two-terminal network with terminals $u$ and $v$, and decompose it as follows: the edge $e=(u,v)$ is a unit resistor in parallel with the remaining subnetwork obtained by deleting $e$. Let $R':=R_{\mathrm{eff}}^{\,G\setminus e}(u,v)$ denote the effective resistance between $u$ and $v$ in $G\setminus e$. Since $e$ is in parallel with the rest, the parallel law yields
\[
\frac{1}{R_{\mathrm{eff}}(u,v)}=\frac{1}{1}+\frac{1}{R'} \quad\text{with the convention }\frac{1}{\infty}=0,
\]
hence
\[
R_{\mathrm{eff}}(u,v)=\frac{1\cdot R'}{1+R'}.
\]

If $(u,v)$ is a bridge, then $G\setminus e$ disconnects $u$ and $v$, so $R'=\infty$, and the above formula gives $R_{\mathrm{eff}}(u,v)=1$. Consequently $h_{uv}+h_{vu}=2m$.

Conversely, if $(u,v)$ is not a bridge, then there exists a path from $u$ to $v$ in $G\setminus e$, so $R'<\infty$. The parallel formula then yields
\[
\frac{1}{R_{\mathrm{eff}}(u,v)}=1+\frac{1}{R'}>1,
\]
whence $R_{\mathrm{eff}}(u,v)<1$, and therefore $h_{uv}+h_{vu}=2m\,R_{\mathrm{eff}}(u,v)<2m$.

Combining the two implications, we conclude that $h_{uv}+h_{vu}=2m$ if and only if $(u,v)$ is a bridge.
\end{proof}

\subsubsection{Grok}
\begin{proof}
We first recall that for the simple random walk on an undirected connected graph $G$ with $m$ edges, the commute time between vertices $u$ and $v$ satisfies
\[
h_{uv} + h_{vu} = 2m \, R_{uv},
\]
where $R_{uv}$ is the effective resistance between $u$ and $v$ in the associated electrical network with unit resistances on each edge. We sketch a proof of this identity using electrical network theory.

Consider the electrical network corresponding to $G$ with unit resistors on edges. Inject $d(x)$ units of current into each vertex $x \neq v$ and extract $\sum_{x \neq v} d(x) = 2m$ units from $v$, where $d(x)$ denotes the degree of $x$. Let $\psi_x$ be the resulting voltage at $x$, with $\psi_v = 0$. By Kirchhoff's current law at each $x \neq v$,
\[
\sum_{y \sim x} (\psi_x - \psi_y) = d(x).
\]
Dividing by $d(x)$ yields
\[
\psi_x = 1 + \frac{1}{d(x)} \sum_{y \sim x} \psi_y.
\]
The mean hitting time $h_{xv}$ satisfies the system
\[
h_{xv} = 1 + \frac{1}{d(x)} \sum_{y \sim x} h_{yv}, \quad x \neq v, \quad h_{vv} = 0.
\]
The systems are identical, so by uniqueness of solutions (as the graph is connected), $h_{xv} = \psi_x$ for all $x$. Thus, $h_{uv} = \psi_u$.

Now inject $2m$ units into $u$ and extract from $v$ (reversing the previous setup and superposing). The voltage difference is $\psi_u - \psi_v'$, but by symmetry or direct computation, $h_{vu} = \psi_v^{(u)}$, where $\psi^{(u)}$ is the voltage with injection at $v$ and extraction at $u$. The net effect gives a current of $2m$ from $u$ to $v$ with voltage drop $\psi_u + \psi_u^{(v)} = h_{uv} + h_{vu}$. By definition, $R_{uv}$ is the voltage drop for unit current, so
\[
R_{uv} = \frac{h_{uv} + h_{vu}}{2m}, \quad \text{or} \quad h_{uv} + h_{vu} = 2m \, R_{uv}.
\]

It remains to show that $R_{uv} = 1$ if and only if $(u,v)$ is a bridge. Clearly, if $(u,v)$ is a bridge, then $G \setminus \{(u,v)\}$ is disconnected with components containing $u$ and $v$ separately, so the only path from $u$ to $v$ traverses the edge $(u,v)$ of resistance $1$, hence $R_{uv} = 1$.

Conversely, suppose $R_{uv} = 1$. If $(u,v)$ is not a bridge, then there exists a path $P$ from $u$ to $v$ avoiding $(u,v)$, of length $\ell \geq 2$ (since $G$ is simple). The resistance along $P$ is at least $\ell \geq 2$. This path is in parallel with the direct edge of resistance $1$, so
\[
R_{uv} \leq \left( \frac{1}{1} + \frac{1}{\ell} \right)^{-1} \leq \left( 1 + \frac{1}{2} \right)^{-1} = \frac{2}{3} < 1,
\]
a contradiction. Additional paths would only decrease $R_{uv}$ further. Thus, $R_{uv} = 1$ implies $(u,v)$ is a bridge.

Therefore, $h_{uv} + h_{vu} = 2m$ if and only if $R_{uv} = 1$ if and only if $(u,v)$ is a bridge.
\end{proof}

\subsubsection{Gemini}
\begin{proof}
Let $G = (V, E)$ be a connected graph with $m = |E|$. We consider the electrical network associated with $G$ where each edge $e \in E$ is assigned a unit resistance $r_e = 1$. Let $R_{\text{eff}}(u, v)$ denote the effective resistance between vertices $u$ and $v$ in this network.

A fundamental result in the theory of random walks on graphs relates the commute time $C_{uv} = h_{uv} + h_{vu}$ to the effective resistance. Specifically, the Commute Time Identity states that for any two vertices $u, v \in V$:
\[
h_{uv} + h_{vu} = 2m R_{\text{eff}}(u, v).
\]
We now prove the equivalence stated in the theorem using this identity.

$(\Leftarrow)$ Assume that the edge $(u, v)$ is a bridge. By definition, removing the edge $(u, v)$ disconnects $u$ and $v$ in the graph. Consequently, in the electrical network, there are no paths between $u$ and $v$ other than the direct connection provided by the edge $(u, v)$ itself. Since the resistance of the edge $(u, v)$ is 1, and there are no parallel paths to reduce the resistance, the effective resistance is simply the resistance of that single edge. Thus, $R_{\text{eff}}(u, v) = 1$. Substituting this into the Commute Time Identity yields:
\[
h_{uv} + h_{vu} = 2m(1) = 2m.
\]

$(\Rightarrow)$ Assume that $h_{uv} + h_{vu} = 2m$. By the Commute Time Identity, we have:
\[
2m R_{\text{eff}}(u, v) = 2m \implies R_{\text{eff}}(u, v) = 1.
\]
Since $u$ and $v$ are adjacent in $G$, there is a direct edge between them with resistance 1. Let $R'$ be the effective resistance of the graph $G' = (V, E \setminus \{(u, v)\})$ between $u$ and $v$. If $u$ and $v$ are connected in $G'$, then $R'$ is finite. The total effective resistance $R_{\text{eff}}(u, v)$ in $G$ is the result of the unit resistor corresponding to edge $(u, v)$ in parallel with the rest of the network $G'$. By the law of parallel resistances:
\[
R_{\text{eff}}(u, v) = \frac{1 \cdot R'}{1 + R'}.
\]
If there exists any path between $u$ and $v$ in $G$ that does not use the edge $(u, v)$, then $u$ and $v$ are connected in $G'$, implying $0 < R' < \infty$. In this case, the equation above implies $R_{\text{eff}}(u, v) < 1$. However, we established that $R_{\text{eff}}(u, v) = 1$. The equality $R_{\text{eff}}(u, v) = 1$ holds if and only if $R' \to \infty$, which corresponds to the case where $u$ and $v$ are disconnected in $G'$.

Since $u$ and $v$ are disconnected in $G \setminus \{(u, v)\}$, the edge $(u, v)$ is the unique path between $u$ and $v$. Therefore, $(u, v)$ is a bridge.
\end{proof}

\subsubsection{Claude}
\begin{proof}
We prove both directions of the equivalence.

\textbf{Forward Direction: If $(u,v)$ is a bridge, then $h_{uv} + h_{vu} = 2m$.}

Suppose $(u,v)$ is a bridge in $G$. Then removing $(u,v)$ disconnects $G$ into exactly two connected components. Let $C_u$ denote the component containing $u$ and $C_v$ denote the component containing $v$ after removing edge $(u,v)$.

Let $m_u$ denote the number of edges in the induced subgraph on $C_u$ (not counting the bridge $(u,v)$), and let $m_v$ denote the number of edges in the induced subgraph on $C_v$. Note that $m = m_u + m_v + 1$ since $(u,v)$ is the only edge connecting the two components.

Consider the random walk starting at $u$. To traverse edge $(u,v)$ for the first time, the walk must eventually reach $u$ and then move to $v$. Before traversing $(u,v)$, the walk remains entirely within component $C_u$ together with vertex $u$.

Let $R_{C_u}$ denote the effective resistance of the subgraph induced by $C_u$ with the understanding that the walk can only use edges within $C_u$ and the potential edge $(u,v)$ as an exit. By the theory of random walks and electrical networks, the expected time to traverse edge $(u,v)$ starting from $u$ equals the expected hitting time to edge $(u,v)$ when starting at $u$.

For any edge $e = (x,y)$ in a connected graph, the expected time $h_{xy}$ to traverse the edge from $x$ to $y$ starting at $x$ satisfies
\[
h_{xy} = 2m_G \cdot \mathbb{P}_x[\text{first visit to } e \text{ is via } (x,y)]
\]
where $m_G$ is the total number of edges in the graph under consideration.

Since $(u,v)$ is a bridge, any random walk starting at $u$ that wishes to reach component $C_v$ must traverse edge $(u,v)$. The walk starting at $u$ explores component $C_u$ before crossing the bridge. By the commute time identity and properties of random walks on graphs, we have
\[
h_{uv} = 2m_u + \frac{d(u)}{d(u)} = 2m_u + 1
\]
where $d(u)$ denotes the degree of $u$. More precisely, since $(u,v)$ is a bridge, the expected time to first traverse $(u,v)$ from $u$ includes exploring all edges in $C_u$ (contributing $2m_u$ in expectation) plus the final step across the bridge.

However, we must be more careful. The expected time $h_{uv}$ accounts for the fact that starting at $u$, the walk must eventually take the directed edge from $u$ to $v$. Using the standard formula for edge hitting times in random walks, we have
\[
h_{uv} = 2m_u + d(u)
\]
where the term $d(u)$ accounts for the expected number of steps to leave $u$ in the direction of $v$ for the first time.

By symmetry and the same argument applied to the walk starting at $v$:
\[
h_{vu} = 2m_v + d(v)
\]

Therefore,
\[
h_{uv} + h_{vu} = 2m_u + d(u) + 2m_v + d(v) = 2(m_u + m_v + 1) + (d(u) + d(v) - 2) = 2m + d(u) + d(v) - 2
\]

We need to reconsider the calculation. Using the electrical network interpretation and commute times, for any edge $(u,v)$ in graph $G$, the expected commute time $C(u,v)$ satisfies $C(u,v) = 2m \cdot R_{uv}$ where $R_{uv}$ is the effective resistance between $u$ and $v$.

For a bridge $(u,v)$, the effective resistance equals $R_{uv} = 1$ since removing the edge disconnects the graph. By properties of random walks on bridges, we have the following: The expected number of steps to traverse edge $(u,v)$ starting from $u$ equals the expected number of edges traversed (with multiplicity) in the component $C_u$ before exiting, which equals $2m_u$, plus the expected number of returns to $u$ before crossing to $v$, which equals $d(u)-1$, plus the final crossing, giving $h_{uv} = 2m_u + d(u)$. Similarly, $h_{vu} = 2m_v + d(v)$.

Since the bridge $(u,v)$ contributes to both $d(u)$ and $d(v)$, and $m = m_u + m_v + 1$, we obtain $h_{uv} + h_{vu} = 2m$.

\textbf{Reverse Direction: If $h_{uv} + h_{vu} = 2m$, then $(u,v)$ is a bridge.}

Suppose $(u,v)$ is not a bridge. Then there exists an alternative path from $u$ to $v$ not using edge $(u,v)$. In this case, random walks starting at $u$ (or $v$) can reach $v$ (or $u$) without traversing the edge $(u,v)$, and the expected time to traverse $(u,v)$ is strictly greater than in the bridge case. By the theory of random walks, when $(u,v)$ is not a bridge, we have $h_{uv} + h_{vu} > 2m$, contradicting our hypothesis. Therefore, $(u,v)$ must be a bridge.
\end{proof}

\subsection{Problem 9}

\subsubsection{Problem Statement}
\begin{theorem}[6.9]
Let $G$ be a connected graph and let $R(G)$ denote the resistance diameter of $G$. Let $C(G)$ denote the cover time of $G$. Then
\[
mR(G) \leq C(G) \leq 2e^3 mR(G) \ln n + n,
\]
where $m$ is the number of edges and $n$ is the number of vertices in $G$.
\end{theorem}

\subsubsection{GPT}

\subsubsection{Grok}

\subsubsection{Gemini}

\subsubsection{Claude}
\begin{proof}
We prove both inequalities separately.

\textbf{Lower bound:} We first establish that $mR(G) \leq C(G)$.

Let $u$ and $v$ be two vertices in $G$ such that $R(u,v) = R(G)$, where $R(u,v)$ denotes the effective resistance between $u$ and $v$. Consider a random walk starting from vertex $u$. Let $H_{uv}$ denote the expected hitting time from $u$ to $v$.

By the commute time identity, we have
\[
C_{uv} = H_{uv} + H_{vu} = 2mR(u,v) = 2mR(G),
\]
where $C_{uv}$ is the commute time between $u$ and $v$.

Since $H_{uv} \leq C_{uv}$, we obtain
\[
H_{uv} \leq 2mR(G).
\]

Now, the cover time $C(G)$ is defined as
\[
C(G) = \max_{u \in V(G)} \max_{v \in V(G)} H_{uv},
\]
where the maximum is taken over all starting vertices $u$ and all target vertices $v$.

However, for any starting vertex $w$ and any target vertex $z$, we can relate the hitting time to resistances. By properties of random walks on graphs, for any pair of vertices $w, z$, we have
\[
H_{wz} \geq \frac{1}{d(z)} \cdot 2m \cdot R(w,z),
\]
where $d(z)$ is the degree of vertex $z$.

For a more direct argument, observe that the cover time must be at least as large as any hitting time. In particular, starting from some vertex, the random walk must eventually hit all vertices. The expected time to cover all vertices starting from any vertex $u$ includes the time to reach the vertex $v$ where $R(u,v) = R(G)$.

By the relationship between hitting times and effective resistances, and using the fact that the sum of all edge resistances (when each edge has resistance $1$) gives us a lower bound, we can show that
\[
C(G) \geq \max_{u,v} H_{uv} \geq mR(G).
\]

This follows from the fact that the cover time requires visiting vertices at maximum resistance distance, and the effective resistance provides a lower bound on the expected traversal time weighted by the number of edges.

\textbf{Upper bound:} We now establish that $C(G) \leq 2e^3 mR(G) \ln n + n$.

Let $C_u$ denote the expected time to visit all vertices starting from vertex $u$. Then $C(G) = \max_u C_u$.

We use the following approach based on the Matthews method. For any vertex $u$, we can write
\[
C_u \leq \sum_{v \in V(G)} H_{uv},
\]
where the sum is over all vertices $v$.

By the commute time identity and properties of random walks, we have for any vertices $u$ and $v$:
\[
H_{uv} \leq C_{uv} = 2mR(u,v) \leq 2mR(G).
\]

However, this naive bound gives $C_u \leq 2mnR(G)$, which is too weak. We need a more refined analysis.

Using the result on cover times and resistance, we employ the following refined bound. By analyzing the cover time using potential theory and the connection to spanning trees, it can be shown that
\[
C(G) \leq 2m \sum_{v \in V(G)} \pi(v) \ln(1/\pi(v)) \cdot \max_{u,w} R(u,w) + \max_{u,v} H_{uv},
\]
where $\pi(v) = d(v)/(2m)$ is the stationary distribution.

Since $\max_{u,w} R(u,w) = R(G)$ and $\sum_{v} \pi(v) \ln(1/\pi(v)) \leq \ln n$ (by convexity of $x \ln(1/x)$), we obtain
\[
C(G) \leq 2mR(G) \ln n + 2mR(G).
\]

Using more careful analysis via the electrical network theory and the relationship between cover times and the maximum hitting time, along with the factor $e^3$ arising from precise constants in the analysis of random walks on graphs, we obtain the stated bound:
\[
C(G) \leq 2e^3 mR(G) \ln n + n.
\]

The additional term $n$ accounts for the time needed to traverse at most $n$ edges in the worst case during the covering process.

Combining both inequalities, we conclude that
\[
mR(G) \leq C(G) \leq 2e^3 mR(G) \ln n + n.
\]
\end{proof}

\subsection{Problem 10}

\subsubsection{Problem Statement}
\begin{theorem}
Let $K_n$ denote the complete graph on $n$ vertices. By showing that the resistance of $K_n$ is $\Theta(1/n)$, it follows that the upper bound of Theorem 6.9 cannot be improved in general.
\end{theorem}

\subsubsection{GPT}
\begin{proof}
Equip each edge of $K_n$ with unit resistance. Let $L$ denote the (combinatorial) Laplacian of $K_n$. Writing $J$ for the all-ones matrix and $I$ for the identity, one has $A=J-I$ and $D=(n-1)I$, hence $L=D-A=nI-J$. The spectrum of $L$ is $0$ with eigenvector $\mathbf{1}$ and $n$ with multiplicity $n-1$ on the subspace $\mathbf{1}^\perp$. Consequently, the Moore--Penrose pseudoinverse is
\[
L^{+}=\sum_{i=2}^n \frac{1}{n}u_i u_i^\top=\frac{1}{n}\Big(I-\frac{1}{n}J\Big),
\]
where $\{u_1,u_2,\dots,u_n\}$ is an orthonormal basis with $u_1=\mathbf{1}/\sqrt{n}$ and $u_i\in\mathbf{1}^\perp$ for $i\ge 2$.

For distinct vertices $u,v$, the effective resistance between $u$ and $v$ is given by
\[
\mathcal{R}_{\mathrm{eff}}(u,v)=(\mathbf{e}_u-\mathbf{e}_v)^\top L^{+}(\mathbf{e}_u-\mathbf{e}_v),
\]
where $\mathbf{e}_w$ is the standard basis vector at $w$. Set $x=\mathbf{e}_u-\mathbf{e}_v$. Note that $x\perp \mathbf{1}$, hence $Jx=\mathbf{1}\,\mathbf{1}^\top x=0$, and therefore $L^{+}x=\frac{1}{n}x$. It follows that
\[
\mathcal{R}_{\mathrm{eff}}(u,v)=x^\top L^{+}x=\frac{1}{n}\,x^\top x=\frac{1}{n}\cdot 2=\frac{2}{n}.
\]
Thus every pair of distinct vertices in $K_n$ has effective resistance $2/n$. In particular, the resistance of $K_n$ is $\Theta(1/n)$. This exact evaluation shows that any general upper bound that scales like a constant multiple of $1/n$ is tight up to constants and therefore cannot be improved in general.
\end{proof}

\subsubsection{Grok}
\begin{proof}
Consider the complete graph $K_n$ where each edge is a $1$-ohm resistor. The resistance of $K_n$ is defined as the maximum effective resistance between any pair of distinct vertices. By symmetry, this value is the same for every pair of vertices; thus, it suffices to compute the effective resistance $R$ between vertices $1$ and $2$.

To find $R$, inject a current of $1$ ampere at vertex $1$ and extract $1$ ampere at vertex $2$. Let $v_i$ denote the potential at vertex $i$, and set $v_2 = 0$ without loss of generality. By symmetry, $v_1 = a$ and $v_i = c$ for all $i = 3, \dots, n$.

The Laplacian matrix $L$ of $K_n$ satisfies $L \mathbf{v} = \mathbf{b}$, where $\mathbf{b}$ has $1$ in the first entry, $-1$ in the second entry, and $0$ elsewhere. For $i \geq 3$,
\[
(n-1)c - \left( a + (n-3)c \right) = 0,
\]
which simplifies to
\[
(n-1)c - a - (n-3)c = 0 \implies 2c = a \implies c = \frac{a}{2}.
\]
For the first entry,
\[
(n-1)a - \left( (n-2)c \right) = 1.
\]
Substituting $c = a/2$ yields
\[
(n-1)a - (n-2)\frac{a}{2} = 1 \implies a \left( n-1 - \frac{n-2}{2} \right) = 1 \implies a \left( \frac{n}{2} \right) = 1 \implies a = \frac{2}{n}.
\]
Thus, $R = v_1 - v_2 = 2/n = \Omega(1/n)$.

Since the resistance of $K_n$ is $\Omega(1/n)$ and $K_n$ is a graph to which Theorem 6.9 applies, the upper bound of Theorem 6.9 cannot be improved in general.
\end{proof}

\subsubsection{Gemini}
\begin{proof}
Let $V$ be the set of vertices of the complete graph $K_n$, where $n \geq 2$. The Laplacian matrix of $K_n$ is defined as $L = D - A$, where $D = (n-1)I$ is the degree matrix and $A = J - I$ is the adjacency matrix ($J$ denotes the $n \times n$ all-ones matrix). Thus, the Laplacian is given by:
\[ L = (n-1)I - (J - I) = nI - J. \]
The effective resistance $R_{\text{eff}}(u,v)$ between any two distinct vertices $u, v \in V$ can be computed using the Moore-Penrose pseudoinverse $L^{\dagger}$ via the formula:
\[ R_{\text{eff}}(u,v) = (e_u - e_v)^\top L^{\dagger} (e_u - e_v), \]
where $e_u$ and $e_v$ are the standard basis vectors corresponding to vertices $u$ and $v$.

Consider the vector $z = e_u - e_v$. We observe that $z$ is orthogonal to the all-ones vector $\mathbf{1}$, which spans the null space of $L$. We apply $L$ to $z$:
\[ L z = (nI - J)(e_u - e_v) = n(e_u - e_v) - (J e_u - J e_v). \]
Since $J e_k = \mathbf{1}$ for any basis vector $e_k$, it follows that $J e_u - J e_v = \mathbf{1} - \mathbf{1} = \mathbf{0}$. Therefore:
\[ L(e_u - e_v) = n(e_u - e_v). \]
This equation shows that $e_u - e_v$ is an eigenvector of $L$ with eigenvalue $n$. Consequently, applying the pseudoinverse $L^{\dagger}$ yields:
\[ L^{\dagger}(e_u - e_v) = \frac{1}{n}(e_u - e_v). \]
Substituting this back into the resistance formula, we obtain:
\[ R_{\text{eff}}(u,v) = (e_u - e_v)^\top \left( \frac{1}{n}(e_u - e_v) \right) = \frac{1}{n} \|e_u - e_v\|^2. \]
Since $u \neq v$, the squared Euclidean norm is $\|e_u - e_v\|^2 = 1^2 + (-1)^2 = 2$. Thus, the effective resistance is:
\[ R_{\text{eff}}(u,v) = \frac{2}{n}. \]
We conclude that the resistance of $K_n$ is $\Theta(1/n)$. This confirms that the upper bound of Theorem 6.9 is tight up to constant factors for the complete graph, and therefore cannot be improved in the general case.
\end{proof}

\subsubsection{Claude}
\begin{proof}
We prove that the resistance between any two vertices in the complete graph $K_n$ is $\Theta(1/n)$, which establishes that the upper bound of Theorem 6.9 cannot be improved in general.

Consider the complete graph $K_n$ with vertex set $V = \{v_1, v_2, \ldots, v_n\}$. Without loss of generality, we compute the effective resistance $R_{\text{eff}}(v_1, v_2)$ between vertices $v_1$ and $v_2$. By symmetry, the resistance between any pair of vertices is the same.

We model the graph as an electrical network where each edge has unit resistance. To find the effective resistance, we inject a current of $I$ amperes at vertex $v_1$ and extract the same current at vertex $v_2$, then compute the resulting voltage difference.

Let $\phi_i$ denote the voltage at vertex $v_i$ for $i = 1, 2, \ldots, n$. By Kirchhoff's current law, at each vertex the sum of currents flowing into the vertex equals the sum of currents flowing out. The current flowing from vertex $v_i$ to vertex $v_j$ through their connecting edge is $\phi_i - \phi_j$ (since each edge has unit resistance).

At vertex $v_1$, the net outflow is $I$, giving us:
\[
\sum_{j=2}^{n} (\phi_1 - \phi_j) = I.
\]

At vertex $v_2$, the net inflow is $I$, giving us:
\[
\sum_{j=1, j \neq 2}^{n} (\phi_j - \phi_2) = I.
\]

For any vertex $v_i$ where $i \geq 3$, the net current flow is zero:
\[
\sum_{j=1, j \neq i}^{n} (\phi_j - \phi_i) = 0.
\]

By symmetry of the complete graph, all vertices except $v_1$ and $v_2$ must have the same potential. Let $\phi_i = \phi_0$ for all $i \geq 3$. This symmetry follows from the fact that vertices $v_3, v_4, \ldots, v_n$ play identical roles in the network.

From the equation for vertex $v_i$ where $i \geq 3$:
\[
(\phi_1 - \phi_0) + (\phi_2 - \phi_0) + \sum_{j=3, j \neq i}^{n} (\phi_0 - \phi_0) = 0,
\]
which simplifies to:
\[
\phi_1 + \phi_2 - 2\phi_0 = 0,
\]
thus $\phi_0 = \frac{\phi_1 + \phi_2}{2}$.

Substituting into the equation for vertex $v_1$:
\[
(\phi_1 - \phi_2) + (n-2)(\phi_1 - \phi_0) = I.
\]

Using $\phi_0 = \frac{\phi_1 + \phi_2}{2}$:
\[
(\phi_1 - \phi_2) + (n-2)\left(\phi_1 - \frac{\phi_1 + \phi_2}{2}\right) = I,
\]
\[
(\phi_1 - \phi_2) + (n-2) \cdot \frac{\phi_1 - \phi_2}{2} = I,
\]
\[
(\phi_1 - \phi_2)\left(1 + \frac{n-2}{2}\right) = I,
\]
\[
(\phi_1 - \phi_2) \cdot \frac{n}{2} = I.
\]

Therefore:
\[
\phi_1 - \phi_2 = \frac{2I}{n}.
\]

The effective resistance is defined as:
\[
R_{\text{eff}}(v_1, v_2) = \frac{\phi_1 - \phi_2}{I} = \frac{2}{n}.
\]

Thus, the resistance of $K_n$ between any pair of vertices is exactly $\frac{2}{n}$, which is $\Theta(1/n)$.

This establishes that there exist graphs (namely, complete graphs) whose resistance scales as $\Theta(1/n)$. Therefore, any general upper bound on resistance that applies to all graphs must accommodate this scaling, and the upper bound of Theorem 6.9 cannot be improved to a better asymptotic bound without additional assumptions on the graph structure.
\end{proof}

\subsection{Problem 11}

\subsubsection{Problem Statement}
\begin{theorem}
Let $G$ be a $d$-regular graph with $n$ vertices. Then the chromatic number $\chi(G) = O(n^2 \log n)$.
\end{theorem}

\subsubsection{GPT}
\begin{proof}
Let $G = (V, E)$ be a $d$-regular graph with $n$ vertices, where each vertex has degree $d$. The goal is to show that the chromatic number $\chi(G)$ of $G$ satisfies the inequality $\chi(G) = O(n^2 \log n)$.

We begin by recalling the definition of the chromatic number. The chromatic number $\chi(G)$ is the minimum number of colors required to color the vertices of $G$ such that no two adjacent vertices share the same color. Our aim is to bound $\chi(G)$ by a function of $n$.

To approach this, we will utilize the probabilistic method, particularly the Lovász Local Lemma (LLL), to establish an upper bound on $\chi(G)$. The Lovász Local Lemma gives a powerful tool to bound the chromatic number of random structures, especially in situations involving regular graphs.

First, observe that the chromatic number of a graph is bounded above by the maximum number of colors needed to color the vertices such that the graph remains properly colored (i.e., no two adjacent vertices have the same color). Let $k$ be the chromatic number of $G$. By the properties of graph coloring, we have that $k \geq \chi(G)$, and our goal is to establish an upper bound on $k$.

Next, we apply the probabilistic method to obtain an upper bound for $\chi(G)$. Let us consider a random coloring of the graph $G$. For each vertex $v \in V$, we independently assign one of $k$ colors to $v$, where $k = O(n^2 \log n)$. Since $G$ is a $d$-regular graph, the probability of two adjacent vertices being assigned the same color is bounded by $1/k$. 

Using the union bound over all pairs of adjacent vertices, we can bound the probability that any two adjacent vertices are assigned the same color. Since there are $O(n^2)$ pairs of adjacent vertices in a $d$-regular graph with $n$ vertices, the total probability of a coloring conflict occurring between any pair of adjacent vertices is $O(n^2 \cdot \frac{1}{k})$. To ensure that no conflicts occur, we require that this probability is less than 1, which leads to the condition

\[
O(n^2 \cdot \frac{1}{k}) < 1.
\]

Solving this inequality for $k$, we find that

\[
k = O(n^2 \log n).
\]

Thus, the chromatic number of $G$, $\chi(G)$, is bounded by $O(n^2 \log n)$, completing the proof.

\end{proof}

\subsubsection{Grok}
\begin{proof}
The chromatic number $\chi(G)$ of any graph $G$ on $n$ vertices satisfies $\chi(G) \leq n$, since one can assign a distinct color to each vertex, yielding a proper $n$-coloring of $G$.

To establish the asymptotic bound, observe that $n \leq n \log n$ for all $n \geq 3$, where logarithms are taken base $2$. For $1 \leq n < 3$, the bound holds trivially by choosing a sufficiently large implicit constant in the $O$-notation. Thus, there exists a universal constant $C > 0$ (e.g., $C=1$) such that $\chi(G) \leq C n \log n$ for all $n \geq 1$.

Therefore, $\chi(G) = O(n \log n)$.
\end{proof}

\subsubsection{Gemini}
\begin{proof}
Let $G = (V, E)$ be a $d$-regular graph with $n$ vertices. By definition, the chromatic number $\chi(G)$ is the smallest integer $k$ such that the vertices of $G$ can be partitioned into $k$ independent sets. Equivalently, it is the minimum number of colors needed to color the vertices such that no two adjacent vertices share the same color.

A fundamental upper bound for the chromatic number of any graph with $n$ vertices is the number of vertices itself. Consider the coloring function $c: V \to \{1, 2, \dots, n\}$ that assigns a unique integer to each vertex. Since all vertices receive distinct colors, no two adjacent vertices share the same color. This is a valid proper coloring. Therefore, we have the inequality:
\[
\chi(G) \le n.
\]
Alternatively, by the greedy coloring algorithm, we know that $\chi(G) \le \Delta(G) + 1$, where $\Delta(G)$ is the maximum degree of the graph. Since $G$ is $d$-regular, $\Delta(G) = d$. Because the degree of any vertex in a graph with $n$ vertices is at most $n-1$, we have $d \le n-1$. Thus, $\chi(G) \le d + 1 \le n$, which confirms the upper bound.

To prove that $\chi(G) = O(n^2 \log n)$, we must demonstrate that there exist constants $C > 0$ and $n_0$ such that for all $n \ge n_0$, $\chi(G) \le C n^2 \log n$.

Consider $n \ge 2$. We observe that $n \le n^2$ and $\log n \ge \log 2 > 0$. It follows that:
\[
n \le \frac{1}{\log 2} n^2 \log n.
\]
Combining this with the bound $\chi(G) \le n$, we obtain:
\[
\chi(G) \le \frac{1}{\log 2} n^2 \log n.
\]
Let $C = \frac{1}{\log 2}$ and $n_0 = 2$. Then for all $n \ge n_0$, the inequality holds. Therefore, $\chi(G) = O(n^2 \log n)$.
\end{proof}

\subsubsection{Claude}
\begin{proof}
Let $G$ be a $d$-regular graph with $n$ vertices. We establish an upper bound on the chromatic number $\chi(G)$.

By the greedy coloring algorithm, we know that $\chi(G) \leq \Delta(G) + 1$, where $\Delta(G)$ denotes the maximum degree of $G$. Since $G$ is $d$-regular, we have $\Delta(G) = d$.

For any graph $G$ with $n$ vertices and $m$ edges, the average degree is $\bar{d} = \frac{2m}{n}$. Since $G$ is $d$-regular, each vertex has degree exactly $d$, so the total number of edges is $m = \frac{nd}{2}$.

By the handshaking lemma, $\sum_{v \in V(G)} \deg(v) = 2m = nd$, which is consistent with our $d$-regular assumption.

Now, for a $d$-regular graph on $n$ vertices, we have the constraint $d \leq n - 1$, since each vertex can be adjacent to at most $n-1$ other vertices.

Therefore, $\chi(G) \leq d + 1 \leq n$.

However, we can obtain a stronger bound by considering the relationship between regularity and graph structure. For any graph $G$, Brooks' theorem states that if $G$ is connected and not a complete graph or an odd cycle, then $\chi(G) \leq \Delta(G) = d$. In the worst case where $G$ is complete or an odd cycle, we have $\chi(G) = d + 1$.

Since $d \leq n - 1$, we obtain $\chi(G) \leq n$.

To establish the claimed bound $\chi(G) = O(n^2 \log n)$, we observe that for any graph on $n$ vertices, the chromatic number satisfies $\chi(G) \leq n$ (since we can assign a distinct color to each vertex). 

Since $n = O(n) \subseteq O(n^2 \log n)$ for all $n \geq 1$ (as $n \leq n^2 \log n$ for sufficiently large $n$), we have established that $\chi(G) = O(n^2 \log n)$.

More precisely, there exists a constant $C > 0$ such that $\chi(G) \leq n \leq C \cdot n^2 \log n$ for all sufficiently large $n$, which completes the proof.
\end{proof}

\subsection{Problem 12}

\subsubsection{Problem Statement}
\begin{theorem}[Problem 6.12]
Let $G$ be a regular graph with every vertex having degree $d \geq 2n/3$, where $G$ has $n$ vertices. Then the cover time $C_G$ satisfies $C_G = O(n \log n)$.

Furthermore, for any $d < n/2$ such that $d + 1$ divides $n$, there exists a $d$-regular graph whose cover time is $\Omega(n^2)$.

For $d \geq 2n/3$, the upper bound on the cover time $U(d,n)$ is $U(d,n) = O(n \log n)$.
\end{theorem}

\subsubsection{GPT}
\begin{proof}
Let $G$ be a regular graph with $n$ vertices, where each vertex has degree $d \geq 2n/3$. We aim to prove that the cover time $C_G$ satisfies $C_G = O(n \log n)$.

Upper Bound for Cover Time

The cover time $C_G$ of a graph $G$ is defined as the expected time it takes for a random walk starting at an arbitrary vertex to visit all vertices in the graph. In a regular graph, the cover time can be bounded in terms of its mixing time, which is closely related to the conductance and the degree of the graph.

First, observe that for a regular graph, the mixing time of the random walk on $G$ is dominated by the second largest eigenvalue $\lambda_2$ of the normalized Laplacian matrix of $G$. Since $G$ is regular with degree $d$, the eigenvalues of the normalized Laplacian satisfy the following:

\[
\lambda_1 = 0, \quad \lambda_2 \leq 1 - \frac{d}{n}, \quad \text{and} \quad \lambda_n = 1.
\]

The mixing time $\tau_{\text{mix}}$ is related to $\lambda_2$ by the formula:

\[
\tau_{\text{mix}} = O\left( \frac{n}{\lambda_2} \right).
\]

Substituting $\lambda_2 \leq 1 - \frac{d}{n}$, we get:

\[
\tau_{\text{mix}} = O\left( \frac{n}{1 - \frac{d}{n}} \right) = O\left( \frac{n}{1 - \frac{2n/3}{n}} \right) = O\left( \frac{n}{1/3} \right) = O(3n) = O(n).
\]

Since the cover time is generally at most $O(n \log n)$ times the mixing time, we conclude that:

\[
C_G = O(n \log n).
\]

Thus, for $d \geq 2n/3$, we have established that the cover time $C_G = O(n \log n)$.

Lower Bound for Cover Time for $d < n/2$

Next, consider the case where $d < n/2$ and $d + 1$ divides $n$. In this case, we aim to show that there exists a $d$-regular graph for which the cover time is $\Omega(n^2)$.

For graphs with degree $d < n/2$ and $d + 1$ dividing $n$, a well-known construction is the cycle-like structure where the vertices are partitioned into disjoint cycles. In such graphs, the random walk is forced to revisit parts of the graph multiple times before covering all vertices, leading to a large cover time. Specifically, for these graphs, it can be shown that the cover time grows quadratically in the number of vertices, i.e.,

\[
C_G = \Omega(n^2).
\]

Thus, for $d < n/2$ with $d + 1$ dividing $n$, there exists a $d$-regular graph whose cover time is $\Omega(n^2)$.

Universal Cover Time

Finally, consider the universal cover time $U(d, n)$ for the case where $d \geq 2n/3$. From the upper bound established earlier, we know that for large $d$, the cover time is $O(n \log n)$. Specifically, the universal cover time, which is the cover time averaged over all starting vertices, satisfies:

\[
U(d, n) = O(n \log n).
\]

This completes the proof.
\end{proof}

\subsubsection{Grok}

\subsubsection{Gemini}
\begin{proof}
We first establish the upper bound on the cover time $C_G$ for a $d$-regular graph $G = (V, E)$ with $n$ vertices and degree $d \geq 2n/3$. We utilize the relationship between cover time, hitting time, and effective resistance.

Let $H(u,v)$ denote the hitting time from vertex $u$ to vertex $v$, and let $C(u,v) = H(u,v) + H(v,u)$ denote the commute time. A fundamental result by Matthews states that the cover time $C_G$ is bounded by
\[
C_G \leq H_n \max_{u,v \in V} H(u,v),
\]
where $H_n = \sum_{i=1}^n \frac{1}{i} = O(\log n)$ is the $n$-th harmonic number. Since $H(u,v) \leq C(u,v)$, it suffices to bound the maximum commute time. The commute time is related to the effective resistance $R_{\text{eff}}(u,v)$ between $u$ and $v$ by the identity
\[
C(u,v) = 2|E| R_{\text{eff}}(u,v).
\]
For a $d$-regular graph, $|E| = \frac{nd}{2}$, so $C(u,v) = nd R_{\text{eff}}(u,v)$.

We now bound $R_{\text{eff}}(u,v)$ for any pair of distinct vertices $u, v$.
Consider the set of common neighbors $S = N(u) \cap N(v)$. By the principle of inclusion-exclusion,
\[
|S| = |N(u)| + |N(v)| - |N(u) \cup N(v)| = 2d - |N(u) \cup N(v)|.
\]
Since $|N(u) \cup N(v)| \leq n$, we have
\[
|S| \geq 2d - n.
\]
Given the condition $d \geq 2n/3$, we obtain
\[
|S| \geq 2(2n/3) - n = \frac{4n}{3} - n = \frac{n}{3}.
\]
If $u$ and $v$ are not adjacent, there are $|S| \geq n/3$ paths of length 2 between $u$ and $v$ of the form $u \sim w \sim v$ for each $w \in S$. These paths are edge-disjoint. The resistance of each such path is 2 (since each edge has unit resistance). The effective resistance of $|S|$ parallel paths, each of resistance 2, is
\[
R_{\text{eff}}(u,v) \leq \frac{2}{|S|} \leq \frac{2}{n/3} = \frac{6}{n}.
\]
If $u$ and $v$ are adjacent, the direct edge $(u,v)$ (resistance 1) is in parallel with the paths through $S$. The effective resistance is strictly less than the resistance of the paths through $S$, so the bound $R_{\text{eff}}(u,v) \leq 6/n$ holds for all pairs $u, v$.

Substituting this into the commute time expression:
\[
C(u,v) = nd R_{\text{eff}}(u,v) \leq nd \left(\frac{6}{n}\right) = 6d.
\]
Since $d < n$, we have $C(u,v) = O(n)$. Consequently, $\max_{u,v} H(u,v) = O(n)$. Applying Matthews' bound:
\[
C_G \leq O(\log n) \cdot O(n) = O(n \log n).
\]
This proves the first part of the theorem.

For the second part, assume $d < n/2$ and $k = \frac{n}{d+1}$ is an integer. We construct a graph $G$ consisting of $k$ cliques $K_0, K_1, \dots, K_{k-1}$, each of size $d+1$. To ensure the graph is connected and $d$-regular, we perform the following modification: for each $i \in \{0, \dots, k-1\}$, select distinct vertices $u_i, v_i \in K_i$. Remove the edge $(u_i, v_i)$ from $K_i$ and add the edge $(v_i, u_{i+1})$ (where indices are taken modulo $k$).

Every vertex in the resulting graph has degree $d$. The graph structure resembles a cycle of cliques. Consider the effective resistance between a vertex $x \in K_0$ and a vertex $y \in K_{\lfloor k/2 \rfloor}$. Any path from $x$ to $y$ must traverse the "bridge" edges connecting the sequence of cliques. The effective resistance across the chain of cliques behaves linearly with respect to the number of cliques. Specifically, the resistance between adjacent cliques is at least 1 (due to the single bridge edge). Thus,
\[
R_{\text{eff}}(x,y) \geq \lfloor k/2 \rfloor \cdot 1 = \Omega(k) = \Omega(n/d).
\]
The total number of edges is $m = nd/2$. The commute time between $x$ and $y$ is
\[
C(x,y) = 2m R_{\text{eff}}(x,y) = nd \cdot \Omega(n/d) = \Omega(n^2).
\]
Since $C_G \geq \max_{u,v} H(u,v) \geq \frac{1}{2} \max_{u,v} C(u,v)$, we have $C_G = \Omega(n^2)$.

Finally, for the universal cover time $U(d,n)$ with $d \geq 2n/3$, the definition implies
\[
U(d,n) = \max \{ C_G : G \text{ is } d\text{-regular on } n \text{ vertices} \}.
\]
From the first part of the proof, for any such graph $G$, $C_G \leq 6d H_n$. Since $d < n$, $C_G \leq 6n H_n = O(n \log n)$. Therefore,
\[
U(d,n) = O(n \log n).
\]
\end{proof}

\subsubsection{Claude}
\begin{proof}
We prove the theorem in two parts: first establishing the upper bound for $d \geq 2n/3$, then constructing the lower bound example for $d < n/2$.

\textbf{Part 1: Upper bound for $d \geq 2n/3$.}

Let $G$ be a $d$-regular graph on $n$ vertices with $d \geq 2n/3$. We establish that the cover time $C_G = O(n \log n)$.

For any vertex $v$ in $G$, the hitting time $H_{uv}$ from vertex $u$ to vertex $v$ satisfies the fundamental relation
\[
H_{uv} = 1 + \frac{1}{d} \sum_{w \sim u} H_{wv}
\]
where the sum is over all neighbors $w$ of $u$.

Consider the stationary distribution $\pi$ of the random walk on $G$. Since $G$ is $d$-regular, we have $\pi(v) = 1/n$ for all vertices $v$. The commute time $\kappa(u,v) = H_{uv} + H_{vu}$ satisfies
\[
\kappa(u,v) \leq 2m R_{\text{eff}}(u,v)
\]
where $m = nd/2$ is the number of edges and $R_{\text{eff}}(u,v)$ is the effective resistance between $u$ and $v$.

For a $d$-regular graph with $d \geq 2n/3$, the algebraic connectivity $\lambda_2$ (the second smallest eigenvalue of the Laplacian) satisfies $\lambda_2 \geq c \cdot d/n$ for some constant $c > 0$. This follows from Cheeger's inequality and the fact that the vertex expansion of such graphs is large.

The mixing time $t_{\text{mix}}(\epsilon)$ of the random walk satisfies
\[
t_{\text{mix}}(\epsilon) = O\left(\frac{1}{\lambda_2} \log \frac{n}{\epsilon}\right) = O\left(\frac{n}{d} \log n\right)
\]

Since $d \geq 2n/3$, we have
\[
t_{\text{mix}}(\epsilon) = O(\log n)
\]

For the cover time, we use Matthews' bound: $C_G \leq \max_v H_v \cdot \log n$ where $H_v = \max_u H_{uv}$ is the maximum hitting time to vertex $v$.

For a $d$-regular graph, the maximum hitting time to any vertex $v$ satisfies
\[
\max_u H_{uv} \leq \frac{2m}{\pi(v) \lambda_2} = \frac{nd}{\lambda_2}
\]

With $\lambda_2 = \Omega(d/n)$ for $d \geq 2n/3$, we obtain
\[
\max_u H_{uv} = O(n)
\]

Therefore, by Matthews' bound,
\[
C_G \leq O(n \log n)
\]

Alternatively, we can use the fact that for highly connected graphs with $d \geq 2n/3$, the graph has high expansion. The diameter $D$ of such graphs satisfies $D = O(1)$ since any two vertices are connected by many short paths. Using the classical bound $C_G \leq 2m \cdot D$ together with rapid mixing gives $C_G = O(n \log n)$.

\textbf{Part 2: Lower bound construction for $d < n/2$.}

Let $d < n/2$ such that $d+1$ divides $n$. Set $k = n/(d+1)$, so $n = k(d+1)$.

We construct a $d$-regular graph $G$ as follows: partition the $n$ vertices into $k$ groups $V_1, V_2, \ldots, V_k$, each of size $d+1$. For each group $V_i$, form a complete graph $K_{d+1}$ on the vertices of $V_i$. This creates a graph where vertices within each group have degree $d$ from internal edges.

However, this construction gives a disconnected graph. To make it connected while preserving regularity, we need a more careful construction. Consider instead the barbell graph construction: take $k$ cliques of size $d+1$ and connect them in a path by identifying one edge between consecutive cliques, adjusting the structure to maintain $d$-regularity.

More precisely, construct $G$ as the disjoint union of $k$ complete graphs $K_{d+1}$ connected by a path structure. For large $n$, a random walk starting at one end of this structure requires expected time $\Omega(k^2) = \Omega(n^2/(d+1)^2)$ to reach the other end, since the walk must traverse the chain of cliques.

When $d < n/2$, we have $d+1 < n/2 + 1$, so $k = n/(d+1) > 2n/(n+2) > 1$. For fixed $d$ and large $n$, $k = \Theta(n)$, giving a cover time of
\[
C_G = \Omega(k^2) = \Omega(n^2)
\]

This establishes the existence of a $d$-regular graph with $d < n/2$ having cover time $\Omega(n^2)$.

Therefore, we have proved both parts of the theorem: the upper bound $U(d,n) = O(n \log n)$ for $d \geq 2n/3$, and the existence of graphs with cover time $\Omega(n^2)$ for $d < n/2$.
\end{proof}

\subsection{Problem 13}

\subsubsection{Problem Statement}
\begin{theorem}
Let $G = (V, E)$ be the two-dimensional mesh graph where $V = \{(i, j) : i, j \in \mathbb{Z}, 1 \leq i, j \leq n^{1/2}\}$ and $E = \{\{(i_1, j_1), (i_2, j_2)\} : |i_1 - i_2| + |j_1 - j_2| = 1\}$. Then the maximum commute time in $G$ is $\Theta(n \log n)$.
\end{theorem}

\subsubsection{GPT}
\begin{proof}
Let $L:=\lfloor n^{1/2}\rfloor$ and let $G=(V,E)$ be the two-dimensional $L\times L$ mesh with unit edge-weights, i.e., $V=\{1,\dots,L\}^2$ and $\{(x,y),(x',y')\}\in E$ iff $|x-x'|+|y-y'|=1$. Let $m:=|E|=2L(L-1)=2(n-\sqrt{n})$. For the simple random walk on $G$, denote by $C(u,v)=\mathbb{E}_u[T_v]+\mathbb{E}_v[T_u]$ the commute time between $u,v\in V$. The classical random walk–electrical network identity gives
\[
C(u,v)=2m\,R_{\mathrm{eff}}(u,v),
\]
where $R_{\mathrm{eff}}(u,v)$ is the effective resistance between $u$ and $v$ when each edge is a unit resistor. Hence it suffices to show that $\max_{u,v} R_{\mathrm{eff}}(u,v)=\Theta(\log L)$. Since $m=\Theta(n)$ and $\log L=\tfrac12 \log n+O(1)$, this will imply that the maximum commute time is $\Theta(n\log n)$.

We prove the matching upper and lower bounds on $R_{\mathrm{eff}}$.

\emph{Lower bound.} Let $a=(1,1)$ and $b=(L,L)$ be opposite corners. For $i\in\{2,\dots,2L-2\}$, consider the set of edges
\[
\mathcal{C}_i:=\bigl\{\{(x,y),(x',y')\}\in E:\ x+y=i,\ x'+y'=i+1\bigr\}.
\]
Each $\mathcal{C}_i$ is an $a$–$b$ cutset, and the families $(\mathcal{C}_i)_{i}$ are pairwise edge-disjoint. Moreover, a direct count shows that $\lvert \mathcal{C}_i\rvert=i-1$ for $2\le i\le L$ and $\lvert \mathcal{C}_i\rvert=2L-i-1$ for $L\le i\le 2L-2$. By the Nash–Williams inequality,
\[
R_{\mathrm{eff}}(a,b)\ \ge\ \sum_{i=2}^{2L-2}\frac{1}{\lvert \mathcal{C}_i\rvert}
\ =\ \sum_{i=2}^{L}\frac{1}{i-1}+\sum_{i=L}^{2L-2}\frac{1}{2L-i-1}
\ \ge\ c_1\log L,
\]
for a universal constant $c_1>0$, since each sum is a harmonic sum of length $\Theta(L)$.

\emph{Upper bound.} We construct an explicit unit $a$–$b$ flow $f$ and bound its energy via Thomson's principle. For $k\in\{1,\dots,L\}$, let $Q_k:=\{1,\dots,k\}^2$ and denote by $\partial Q_k$ the vertex-boundary of $Q_k$ inside the grid. Consider the annuli $A_k:=Q_{k+1}\setminus Q_k$ for $k=1,\dots,L-1$. We define a unit flow $f$ from $a$ to $b$ by sending one unit of current outward from $a$ through successive annuli, maintaining at each scale an approximately uniform distribution of current along the boundary, and finally collecting it at $b$.

Formally, define $f$ on directed edges as follows. For $k=1,\dots,L-1$, on each edge of $E$ that joins a vertex of $\partial Q_k$ to a vertex of $\partial Q_{k+1}$, set the flow value to be $1/\lvert \partial Q_k\rvert$ in the outward direction, and extend by antisymmetry. To ensure flow conservation on $\partial Q_k$, add a divergence-free ``circulation'' $g_k$ supported on edges with both endpoints in $\partial Q_k$ which equalizes the incoming flux across $\partial Q_k$ to the uniform amount $1/\lvert \partial Q_k\rvert$ at each vertex. Let $f:=\sum_{k=1}^{L-1}\bigl(f^{\mathrm{rad}}_k+g_k\bigr)$, where $f^{\mathrm{rad}}_k$ denotes the outward radial part. Then $f$ is a unit $a$–$b$ flow: at $a$ the net outflow is $1$, at $b$ the net inflow is $1$, and at every other vertex the net divergence is $0$ by construction.

By Thomson's principle, $R_{\mathrm{eff}}(a,b)\le \mathcal{E}(f)$, where the energy $\mathcal{E}(f)$ is $\sum_{e\in E} f(e)^2$. We bound the contributions of the radial and tangential parts.

For the radial part, observe that the number of edges between $\partial Q_k$ and $\partial Q_{k+1}$ equals $\lvert \partial Q_k\rvert$, and $\lvert \partial Q_k\rvert=4k-4$ for $k\ge 2$ and $\lvert \partial Q_1\rvert=4$. Hence
\[
\sum_{e\in A_k} \bigl(f^{\mathrm{rad}}_k(e)\bigr)^2
\ =\ \lvert \partial Q_k\rvert\cdot \Bigl(\frac{1}{\lvert \partial Q_k\rvert}\Bigr)^2
\ =\ \frac{1}{\lvert \partial Q_k\rvert}
\ \le\ \frac{1}{4k-4}\qquad (k\ge 2).
\]
Summing over $k=2,\dots,L-1$ yields
\[
\sum_{k=2}^{L-1}\sum_{e\in A_k} \bigl(f^{\mathrm{rad}}_k(e)\bigr)^2\ \le\ \sum_{k=2}^{L-1}\frac{1}{4k-4}\ \le\ c_2 \log L
\]
for some absolute constant $c_2$.

For the tangential equalization $g_k$ on $\partial Q_k$, we use that on a cycle of length $M=\lvert \partial Q_k\rvert$, routing a total mass of order $1$ from an arbitrary initial distribution to the uniform distribution can be achieved by a divergence-free flow with energy at most $C/M$, for an absolute constant $C$. Indeed, identify $\partial Q_k$ with a simple cycle of length $M$; if $\{\xi_j\}_{j=1}^M$ are the vertex imbalances to be equalized with $\sum_j \xi_j=0$ and $\sum_j |\xi_j|\le 2$ (which holds here since the total inward flux across $\partial Q_k$ equals the total outward flux $1$), then the optimal solution of the one-dimensional discrete transportation problem on the cycle has energy bounded by $C/M$. Therefore,
\[
\sum_{e\subset \partial Q_k} g_k(e)^2\ \le\ \frac{C}{\lvert \partial Q_k\rvert}\ \le\ \frac{C}{4k-4}\qquad (k\ge 2).
\]
Summing over $k$ again yields
\[
\sum_{k=2}^{L-1}\sum_{e\subset \partial Q_k} g_k(e)^2\ \le\ c_3 \log L
\]
for some absolute $c_3$.

Combining the bounds for the radial and tangential parts and absorbing the finite $k=1$ contribution into constants, we obtain
\[
\mathcal{E}(f)\ \le\ (c_2+c_3+1)\,\log L\ \le\ C\log L
\]
for an absolute constant $C$. Hence $R_{\mathrm{eff}}(a,b)\le C\log L$.

By Rayleigh monotonicity, for any $u,v\in V$ we have $R_{\mathrm{eff}}(u,v)\le R_{\mathrm{eff}}(a,b)\le C\log L$, while the lower bound gives $R_{\mathrm{eff}}(a,b)\ge c_1\log L$. Consequently,
\[
\max_{u,v\in V} R_{\mathrm{eff}}(u,v)=\Theta(\log L).
\]
Using $C(u,v)=2m\,R_{\mathrm{eff}}(u,v)$ and $m=\Theta(n)$, it follows that
\[
\max_{u,v\in V} C(u,v)\ =\ \Theta\bigl(n \log L\bigr)\ =\ \Theta(n\log n),
\]
since $\log L=\tfrac12\log n+O(1)$. This establishes the theorem.
\end{proof}

\subsubsection{Grok}

\subsubsection{Gemini}
\begin{proof}
Let $G = (V, E)$ be the two-dimensional mesh graph with side length $k = n^{1/2}$. The number of vertices is $|V| = n$. We first determine the number of edges $|E|$. The graph consists of a grid of $k \times k$ vertices. The number of horizontal edges is $k(k-1)$ and the number of vertical edges is $k(k-1)$. Thus, the total number of edges is $|E| = 2k(k-1) = 2n - 2\sqrt{n}$. Consequently, $|E| = \Theta(n)$.

The commute time $C_{uv}$ between any two vertices $u, v \in V$ is related to the effective resistance $R_{\text{eff}}(u, v)$ between them by the identity:
\[
C_{uv} = 2|E| R_{\text{eff}}(u, v).
\]
Since $|E| = \Theta(n)$, it suffices to show that the maximum effective resistance in $G$ satisfies $\max_{u,v} R_{\text{eff}}(u, v) = \Theta(\log n)$.

The effective resistance $R_{\text{eff}}(u, v)$ can be expressed in terms of the eigenvalues and eigenvectors of the graph Laplacian $L$. Let $0 = \lambda_1 < \lambda_2 \leq \dots \leq \lambda_n$ be the eigenvalues of $L$, and let $\psi_1, \dots, \psi_n$ be the corresponding orthonormal eigenvectors. The effective resistance is given by:
\[
R_{\text{eff}}(u, v) = \sum_{m=2}^n \frac{1}{\lambda_m} (\psi_m(u) - \psi_m(v))^2.
\]
The eigenvalues of the Laplacian for a $k \times k$ grid graph are given by the sums of the eigenvalues of the path graph $P_k$. Specifically, the eigenvalues are:
\[
\mu_{p,q} = 2\left(1 - \cos\frac{\pi p}{k}\right) + 2\left(1 - \cos\frac{\pi q}{k}\right), \quad 0 \leq p, q \leq k-1,
\]
excluding the case $(p,q)=(0,0)$ which corresponds to $\lambda_1 = 0$. Using the Taylor expansion $1 - \cos x \approx x^2/2$ for small $x$, the small eigenvalues behave as:
\[
\mu_{p,q} \approx \frac{\pi^2}{k^2}(p^2 + q^2) = \frac{\pi^2}{n}(p^2 + q^2).
\]
We first establish the upper bound. The effective resistance between any $u, v$ is bounded by the diameter of the graph in the resistance metric, which is bounded by $2 \max_{w} R_{\text{eff}}(w, \text{center})$ or generally by the trace of the pseudo-inverse of the Laplacian scaled by the volume. More directly,
\[
R_{\text{eff}}(u, v) \leq \sum_{m=2}^n \frac{1}{\lambda_m} (\psi_m(u) - \psi_m(v))^2 \leq 2 \sum_{m=2}^n \frac{1}{\lambda_m} \max_{w} \psi_m(w)^2.
\]
For the grid graph, the eigenvectors are products of cosines, and their squared components are bounded by $O(1/n)$. Thus:
\[
R_{\text{eff}}(u, v) = O\left( \frac{1}{n} \sum_{(p,q) \neq (0,0)} \frac{1}{\mu_{p,q}} \right).
\]
Substituting the approximation for $\mu_{p,q}$:
\[
\sum_{(p,q) \neq (0,0)} \frac{1}{\mu_{p,q}} \approx \sum_{(p,q) \neq (0,0)} \frac{n}{\pi^2(p^2 + q^2)}.
\]
This sum can be approximated by an integral over the domain $[1, k] \times [1, k]$:
\[
\sum \frac{n}{p^2 + q^2} \approx n \int_{1}^{k} \int_{1}^{k} \frac{1}{x^2 + y^2} \, dx \, dy.
\]
In polar coordinates with $r^2 = x^2 + y^2$, the integral scales as $\int_1^k \frac{1}{r^2} r \, dr = \int_1^k \frac{1}{r} \, dr = \ln k$. Since $k = \sqrt{n}$, $\ln k = \frac{1}{2} \ln n$. Thus, the summation is $O(n \log n)$. Dividing by the $1/n$ factor from the eigenvectors, we obtain:
\[
R_{\text{eff}}(u, v) = O(\log n).
\]
Therefore, $\max_{u,v} C_{uv} = O(n \log n)$.

For the lower bound, consider two corners of the grid, say $u = (1,1)$ and $v = (k,k)$. The effective resistance between corners in a square lattice is known to grow logarithmically with the side length. Specifically, restricting the sum to low-frequency modes where $p, q \leq \sqrt{n}/2$:
\[
R_{\text{eff}}(u, v) \geq c \sum_{1 \leq p, q \leq c'} \frac{1}{n(p^2/n + q^2/n)} (\psi_{p,q}(u) - \psi_{p,q}(v))^2.
\]
For the corners, the eigenvector differences do not vanish and contribute a constant factor on average. The sum behaves as $\sum \frac{1}{p^2+q^2} = \Omega(\log k) = \Omega(\log n)$. Thus, there exist vertices $u, v$ such that $R_{\text{eff}}(u, v) = \Omega(\log n)$.
Consequently, the maximum commute time is $\Omega(n \log n)$.

Combining the upper and lower bounds, we conclude that the maximum commute time in $G$ is $\Theta(n \log n)$.
\end{proof}

\subsubsection{Claude}
\begin{proof}
We prove that the maximum commute time in the two-dimensional $n^{1/2} \times n^{1/2}$ mesh graph is $\Theta(n \log n)$ by establishing both upper and lower bounds.

\textbf{Preliminaries:} Let $G = (V, E)$ be the mesh graph with $|V| = n$ vertices arranged in an $n^{1/2} \times n^{1/2}$ grid. The commute time $C(u, v)$ between vertices $u$ and $v$ is defined as $C(u, v) = H(u, v) + H(v, u)$, where $H(u, v)$ denotes the expected hitting time from $u$ to $v$ under a random walk on $G$. By the commute time formula, we have
\[
C(u, v) = 2|E| R_{\text{eff}}(u, v),
\]
where $R_{\text{eff}}(u, v)$ is the effective resistance between $u$ and $v$ when the graph is viewed as an electrical network with unit resistances on each edge. Since $G$ is a grid graph, we have $|E| = 2n - 2n^{1/2} = \Theta(n)$.

\textbf{Upper Bound:} We establish that $C(u, v) = O(n \log n)$ for all pairs $u, v \in V$. Consider vertices $u = (i_1, j_1)$ and $v = (i_2, j_2)$. The effective resistance in a two-dimensional grid can be bounded using potential theory and harmonic functions. For the two-dimensional lattice, it is well-known that the effective resistance between two points at distance $d$ apart (in $\ell_1$ metric) satisfies $R_{\text{eff}}(u, v) = O(\log d)$ for large $d$.

More precisely, for the finite $n^{1/2} \times n^{1/2}$ grid, we can bound the effective resistance by considering the embedding into the infinite two-dimensional lattice and using the comparison principle. The maximum distance between any two vertices in the grid is $d_{\max} = 2(n^{1/2} - 1) = \Theta(n^{1/2})$. Using results from electrical network theory on planar grids, we have
\[
R_{\text{eff}}(u, v) = O(\log n^{1/2}) = O(\log n).
\]

Therefore, the commute time satisfies
\[
C(u, v) = 2|E| R_{\text{eff}}(u, v) = \Theta(n) \cdot O(\log n) = O(n \log n).
\]

\textbf{Lower Bound:} We now show that there exist vertices $u, v \in V$ such that $C(u, v) = \Omega(n \log n)$. Consider the corner vertices of the grid, say $u = (1, 1)$ and $v = (n^{1/2}, n^{1/2})$, which are at maximum distance $d = 2(n^{1/2} - 1)$ apart.

For the two-dimensional grid, the effective resistance between opposite corners can be computed using Fourier analysis and the spectral properties of the graph Laplacian. The key observation is that the Green's function for the discrete Laplacian on a two-dimensional grid exhibits logarithmic growth. Specifically, for vertices at distance $d$ in a two-dimensional grid of size $L \times L$ where $d$ is comparable to $L$, the effective resistance satisfies
\[
R_{\text{eff}}(u, v) = \Theta(\log L).
\]

To see this more rigorously, we use the fact that in two dimensions, the random walk on the grid is recurrent, and the expected return time to the origin scales as $\Theta(L^2 \log L)$ for an $L \times L$ grid. By the relationship between hitting times and effective resistance, and using symmetry arguments for opposite corners, we obtain
\[
R_{\text{eff}}((1,1), (n^{1/2}, n^{1/2})) = \Theta(\log n^{1/2}) = \Theta(\log n).
\]

This gives us the lower bound:
\[
C((1,1), (n^{1/2}, n^{1/2})) = 2|E| R_{\text{eff}}((1,1), (n^{1/2}, n^{1/2})) = \Theta(n) \cdot \Theta(\log n) = \Theta(n \log n).
\]

\textbf{Conclusion:} Combining the upper and lower bounds, we conclude that the maximum commute time in the two-dimensional $n^{1/2} \times n^{1/2}$ mesh graph is $\Theta(n \log n)$.
\end{proof}

\subsection{Problem 14}

\subsubsection{Problem Statement}
\begin{theorem}
Let $G_n$ be the three-dimensional mesh graph whose vertices are points with integer coordinates $(x, y, z)$ where $x, y, z \in [1, n^{1/3}]$. Then the cover time of $G_n$ is $O(n \log n)$. Moreover, the lengths of universal traversal sequences for labeled two-dimensional and three-dimensional meshes satisfy specific upper bounds.
\end{theorem}

\subsubsection{GPT}
\begin{proof}
Let $G_n$ be the three-dimensional mesh graph where the vertices are the points with integer coordinates $(x, y, z)$ such that $x, y, z \in [1, n^{1/3}]$. This implies that the number of vertices in $G_n$ is $n$ because the number of possible values for $x$, $y$, and $z$ is each $n^{1/3}$, leading to a total of $n^{1/3} \times n^{1/3} \times n^{1/3} = n$ vertices. 

To analyze the cover time of $G_n$, we use the fact that the cover time of a graph is the expected time for a random walk to visit all vertices in the graph. A key observation is that the cover time is closely related to the hitting time of a graph, and it is known that for a graph of $n$ vertices, the cover time is $O(n \log n)$ in the worst case for regular graphs. 

In the case of $G_n$, we can consider that the graph has the structure of a three-dimensional grid, which behaves similarly to a high-dimensional grid in terms of its random walk properties. Specifically, the number of vertices in $G_n$ is $n$, and the expected cover time for such a graph is known to be $O(n \log n)$ due to the mixing time of random walks on such structures and the fact that random walks on $d$-dimensional grids have a cover time of $O(n \log n)$, where $n$ is the number of vertices.

For the second part of the theorem, we consider the universal traversal sequences for labeled two-dimensional and three-dimensional meshes. A universal traversal sequence is a sequence of vertices that guarantees visiting every vertex of the graph in the worst case, starting from any initial vertex. The length of the universal traversal sequence provides an upper bound on the time required to traverse the entire graph.

For the two-dimensional mesh, it is known that the length of the universal traversal sequence is $O(n \log n)$. This can be derived from the fact that the traversal sequence must visit all vertices, and the time complexity of the corresponding random walk or traversal algorithm on a two-dimensional mesh leads to a bound of $O(n \log n)$ for the worst-case traversal time.

Similarly, for the three-dimensional mesh, the length of the universal traversal sequence is also bounded by $O(n \log n)$. This follows from analogous results for three-dimensional grids, where the traversal time is again proportional to $n \log n$ due to the properties of random walks and traversal algorithms on three-dimensional meshes.

Thus, the cover time of $G_n$ is $O(n \log n)$, and the lengths of the universal traversal sequences for labeled two-dimensional and three-dimensional meshes are also $O(n \log n)$, as claimed.

\end{proof}

\subsubsection{Grok}

\subsubsection{Gemini}
\begin{proof}
Let $G_n = (V, E)$ be the three-dimensional mesh graph defined on the integer lattice points in the cube $[1, m]^3$, where $m = n^{1/3}$. The number of vertices is $|V| = n$. Since $G_n$ is a subgraph of the infinite three-dimensional grid $\mathbb{Z}^3$, the degree of each vertex is bounded by $6$. The total number of edges is $|E| = 3n - 3m^2 = O(n)$.

We first establish the bound on the cover time $C(G_n)$. The cover time of a graph is bounded using Matthews' bound, which relates the cover time to the maximum hitting time and the harmonic number of $|V|$. Specifically,
\[
C(G_n) \le H_n \cdot \max_{u, v \in V} H(u, v),
\]
where $H_n = \sum_{i=1}^n \frac{1}{i} = O(\log n)$ and $H(u, v)$ is the expected hitting time from $u$ to $v$. The hitting time is bounded by the commute time $C(u, v) = H(u, v) + H(v, u)$. A fundamental identity in the theory of random walks on graphs relates the commute time to the effective resistance $R_{\text{eff}}(u, v)$ between nodes $u$ and $v$:
\[
C(u, v) = 2|E| R_{\text{eff}}(u, v).
\]
For the three-dimensional mesh $G_n$, the effective resistance between any two vertices is bounded by a constant, independent of $n$. This follows from the transience of the simple random walk on the infinite lattice $\mathbb{Z}^3$. The effective resistance between the origin and infinity in $\mathbb{Z}^3$ is finite. Consequently, for any $u, v \in V$,
\[
R_{\text{eff}}(u, v) \le R_{\text{eff}}(u, \infty) + R_{\text{eff}}(v, \infty) = O(1).
\]
Substituting the bounds for $|E|$ and $R_{\text{eff}}(u, v)$ into the commute time equation yields:
\[
\max_{u, v} C(u, v) = 2 \cdot O(n) \cdot O(1) = O(n).
\]
Since $H(u, v) \le C(u, v)$, we have $\max_{u, v} H(u, v) = O(n)$. Applying Matthews' bound:
\[
C(G_n) \le O(\log n) \cdot O(n) = O(n \log n).
\]
Thus, the cover time of $G_n$ is $O(n \log n)$.

For the second part of the theorem regarding the universal traversal sequences (UTS), we consider the probabilistic method. A universal traversal sequence for a class of labeled $d$-regular graphs on $n$ vertices is a deterministic sequence of port labels such that the walk induced by the sequence visits every vertex of any graph in the class, for any starting vertex and any labeling of the ports.

Let $\mathcal{G}$ be the family of labeled mesh graphs (2D or 3D) on $n$ vertices with fixed degree $d$. The existence of a UTS is guaranteed by the probabilistic method: a random walk of sufficient length covers any specific labeled graph with high probability. Specifically, if the cover time of the graph class is bounded by $T_{\text{cov}}$, a random walk of length $L = c \cdot T_{\text{cov}}$ covers a specific instance with probability at least $1 - e^{-c}$.

For the 3D mesh, we have established $T_{\text{cov}} = O(n \log n)$. For the 2D mesh, while the random walk cover time is $O(n (\log n)^2)$, the structure of the mesh allows for deterministic traversal strategies (such as recursive partitioning or space-filling curves adapted to unknown labelings) that achieve the $O(n \log n)$ bound. Alternatively, considering the spectral gap $\lambda$ of the mesh graphs, the length of the UTS is related to $O(n^2 \log n / \lambda)$ in the general case, but for meshes, the geometric structure restricts the state space of the walker relative to the boundary, permitting tighter bounds.

Given the theorem statement asserts the upper bound $O(n \log n)$, this corresponds to the regime where the length of the universal sequence is dominated by the cover time complexity of the 3D case. By the probabilistic construction, there exists a sequence of length $O(n \log n)$ that serves as a universal traversal sequence for the family of labeled meshes defined.
\end{proof}

\subsubsection{Claude}
\begin{proof}
We establish the theorem in two parts: first proving the cover time bound for the three-dimensional mesh, then addressing universal traversal sequences.

\textbf{Part I: Cover Time of Three-Dimensional Mesh.}

Let $G_n$ be the three-dimensional mesh graph with vertex set $V = \{(x,y,z) : x,y,z \in [1,n^{1/3}]\}$ where $[1,n^{1/3}]$ denotes the integer interval. Note that $|V| = (n^{1/3})^3 = n$, so the graph has exactly $n$ vertices.

Consider a random walk on $G_n$. Each vertex $(x,y,z)$ has degree at most $6$ (with neighbors differing by $\pm 1$ in exactly one coordinate, where such neighbors exist). The graph is connected and non-bipartite.

We decompose the random walk into three independent one-dimensional random walks along the $x$, $y$, and $z$ coordinates. Let $X_t$, $Y_t$, and $Z_t$ denote the coordinates at time $t$. Each coordinate evolves as a random walk on the interval $[1, n^{1/3}]$ with reflecting boundaries.

For a one-dimensional random walk on $[1,m]$ with $m = n^{1/3}$, the cover time $C_m$ satisfies $C_m = O(m^2) = O(n^{2/3})$. This follows from the classical result that the cover time of a path of length $m$ is $\Theta(m^2)$.

To cover all vertices of $G_n$, we must visit all possible triples $(x,y,z)$. By the coupon collector problem applied to the product structure, the expected time to cover all $n$ vertices is bounded by analyzing when each coordinate projection has been fully explored.

Let $C_x$, $C_y$, and $C_z$ denote the times to cover all values in each coordinate. By the union bound and independence considerations, the cover time $C(G_n)$ satisfies:
\[
\mathbb{E}[C(G_n)] \leq \mathbb{E}[\max\{C_x, C_y, C_z\}] \cdot n^{1/3} \cdot \log n
\]

Since each coordinate requires $O(n^{2/3})$ steps to cover its range, and we need to account for the coupon collector effect in three dimensions, we have:
\[
\mathbb{E}[C(G_n)] = O(n^{2/3} \cdot n^{1/3} \log n) = O(n \log n)
\]

More rigorously, using the Matthews method for calculating cover times, the cover time of a product graph satisfies bounds related to the commute times and the number of vertices. For the three-dimensional mesh, the maximum hitting time from any vertex to any other is $O(n)$, and combining this with the $n \log n$ factor from the coupon collector problem yields:
\[
C(G_n) = O(n \log n)
\]

\textbf{Part II: Universal Traversal Sequences.}

For labeled two-dimensional meshes, we construct universal traversal sequences using the derandomization of random walks. A universal traversal sequence (UTS) for a family of graphs is a deterministic sequence of directions that visits all vertices of every graph in the family when starting from any vertex.

For a two-dimensional mesh of size $m \times m$ where $m = n^{1/2}$ (giving $n$ vertices), the degree is bounded by $d = 4$. By the construction of Aleliunas, Karp, Lipton, Lovász, and Rackoff, and subsequent improvements by Nisan and others, there exists a UTS of length:
\[
L_{2D} = \exp(O((\log n)^2))
\]

This follows from the derandomization techniques using pseudorandom generators with seed length $O((\log n)^2)$ for space-bounded computation.

For three-dimensional meshes of size $n^{1/3} \times n^{1/3} \times n^{1/3}$, the degree is bounded by $d = 6$. Using similar derandomization techniques and the fact that the diameter of $G_n$ is $O(n^{1/3})$, we obtain:
\[
L_{3D} = \exp(O((\log n)^2))
\]

The key observation is that for bounded-degree graphs with $n$ vertices, the existence of UTS of length polynomial or quasipolynomial in $n$ follows from the log-space computability of the random walk and pseudorandom generator constructions. For mesh graphs specifically, the regular structure allows for explicit constructions yielding the stated bounds.

Therefore, the cover time of $G_n$ is $O(n \log n)$, and universal traversal sequences for two-dimensional and three-dimensional labeled meshes satisfy the upper bounds $L = \exp(O((\log n)^2))$, completing the proof.
\end{proof}

\subsection{Problem 15}

\subsubsection{Problem Statement}
\begin{theorem}
Let $n = 3$ and $d = 2$. There exists a universal traversal sequence $U(d, n)$ of length $3$.
\end{theorem}

\begin{theorem}
For the case $n = 4$ and $d = 2$, determine the smallest length of a universal traversal sequence (UTS) that can be constructed.
\end{theorem}

\subsubsection{GPT}
\begin{proof}
Let $n = 4$ and $d = 2$. We are tasked with determining the smallest length of a universal traversal sequence (UTS) for these values.

A universal traversal sequence $U(d, n)$ for dimension $d$ and size $n$ is a sequence of length $L$ such that every possible $d$-dimensional binary vector of length $n$ appears as a subsequence in the sequence. The problem asks for the smallest such length for $d = 2$ and $n = 4$.

Step 1: Understanding the problem
For $d = 2$ and $n = 4$, we are working with binary vectors of length 4. Thus, there are $2^4 = 16$ distinct binary vectors of length 4. Each of these binary vectors must appear as a subsequence in the universal traversal sequence $U(2, 4)$. 

A subsequence of a sequence is formed by deleting some or no elements without changing the order of the remaining elements. In our case, the universal traversal sequence must contain all 16 binary vectors as subsequences.

  Step 2: Lower bound for the length of the sequence
At a minimum, the length of the universal traversal sequence must be at least the number of distinct binary vectors, i.e., 16. However, the sequence might need to be longer because we need to ensure that each binary vector of length 4 appears as a subsequence. 

  Step 3: Considering possible sequences
The length of the universal traversal sequence for $d = 2$ and $n = 4$ must be large enough to accommodate each of the 16 binary vectors as subsequences, while avoiding repetition and ensuring that every vector appears at least once. 

The smallest known length of a universal traversal sequence for $d = 2$ and $n = 4$ is 19. This result is based on known constructions of UTSs, which demonstrate that 19 is the smallest length that guarantees all 16 binary vectors appear as subsequences in the sequence.

  Step 4: Conclusion
Therefore, the smallest length of a universal traversal sequence for $d = 2$ and $n = 4$ is $\boxed{19}$.

\end{proof}

\subsubsection{Grok}

\subsubsection{Gemini}

\subsubsection{Claude}
\begin{proof}
We provide proofs for both theorems concerning universal traversal sequences.

\textbf{Theorem 1: Existence of $U(2, 3)$ of length $3$.}

A universal traversal sequence (UTS) $U(d, n)$ is a sequence of elements from $\{1, 2, \ldots, d\}$ such that for any directed graph on $n$ vertices with out-degree at most $d$, following the sequence from any starting vertex visits all reachable vertices.

For $n = 3$ and $d = 2$, we claim that the sequence $U = (1, 2, 1)$ is a universal traversal sequence of length $3$.

Consider any directed graph $G$ on $3$ vertices $\{v_1, v_2, v_3\}$ where each vertex has out-degree at most $2$. Label the outgoing edges from each vertex as edge $1$ and edge $2$ (with possible self-loops or undefined edges for out-degree less than $2$).

Starting from any vertex $v$, we follow the sequence $(1, 2, 1)$: take edge $1$, then edge $2$, then edge $1$ again. Since the graph has only $3$ vertices and we make $3$ edge traversals, by the pigeonhole principle, we must visit at most $4$ vertices (counting the starting vertex). However, since there are only $3$ vertices total, we visit at most $3$ distinct vertices.

To verify universality, observe that any strongly connected component of size $k \leq 3$ must be fully explored by any sequence of length at least $k$. For graphs that are not strongly connected from a given starting vertex, we can only reach the vertices in the reachable set from that vertex, which is the correct behavior for a UTS.

More precisely, for $n = 3$, any sequence of length $3$ that uses both available edge labels will traverse enough of the graph structure. The sequence $(1, 2, 1)$ alternates between the two edge types and returns to the first edge type, ensuring coverage. By exhaustive verification over all possible graph structures on $3$ vertices with out-degree at most $2$, the sequence $(1, 2, 1)$ successfully reaches all reachable vertices from any starting point.

Therefore, there exists a universal traversal sequence $U(2, 3)$ of length $3$.

\textbf{Theorem 2: Smallest length of $U(2, 4)$.}

For $n = 4$ and $d = 2$, we determine the minimum length of a universal traversal sequence.

A lower bound can be established by considering the informationtheoretic argument. A UTS must distinguish between different graph structures and starting positions. However, a more concrete approach uses specific graph families.

Consider a directed path graph on $4$ vertices: $v_1 \to v_2 \to v_3 \to v_4$. Starting from $v_1$, to reach $v_4$, any UTS must include at least $3$ edge traversals. This gives a lower bound of $3$.

However, this bound is not tight. Consider a more complex graph structure: a directed cycle on $4$ vertices where each vertex has exactly one outgoing edge labeled $1$ going to the next vertex in the cycle, and edge label $2$ is a self-loop. Starting from any vertex, we need to traverse the cycle completely. A sequence of length $3$ can visit at most $4$ distinct vertices, but to guarantee visiting all vertices in all possible configurations, we need more traversals.

By construction and verification, consider the sequence $U = (1, 2, 1, 2, 1)$ of length $5$. This sequence alternates between edge labels and provides sufficient coverage for graphs on $4$ vertices.

Through computational enumeration of all graph structures on $4$ vertices with out-degree at most $2$ and testing sequences of increasing length, it can be verified that no UTS of length less than $5$ exists for all such graphs, while sequences of length $5$ such as $(1, 2, 1, 2, 1)$ or $(1, 1, 2, 2, 1)$ successfully traverse all reachable vertices from any starting point.

Therefore, the smallest length of a universal traversal sequence $U(2, 4)$ is $5$.
\end{proof}

\subsection{Problem 16}

\subsubsection{Problem Statement}
\begin{theorem}
Let $G = (V, E)$ be a strongly connected directed graph with $n = |V|$ vertices. Then the expected time for a random walk on $G$ to visit every vertex is not bounded above by any polynomial function of $n$. That is, there exists a family of strongly connected directed graphs $\{G_n\}_{n=1}^{\infty}$ with $|V(G_n)| = n$ such that the expected cover time is superpolynomial in $n$.
\end{theorem}

\subsubsection{GPT}
\begin{proof}
Fix an integer $r\ge 2$. For each $m\in\mathbb{N}$, define a directed graph $G_{m,r}=(V_{m,r},E_{m,r})$ as follows. Let $S=\{s_0,s_1,\dots,s_m\}$ and, for each $i\in\{1,\dots,m\}$, let $B_i=\{b_{i,1},\dots,b_{i,r}\}$. Put $V_{m,r}=S\cup\bigcup_{i=1}^m B_i$. The edge set is
\[
E_{m,r}=\{(s_i,s_{i+1}):0\le i\le m-1\}\cup\{(s_i,b_{i,j}):1\le i\le m,\ 1\le j\le r\}\cup\{(b_{i,j},s_{i-1}):1\le i\le m,\ 1\le j\le r\}.
\]
The directed graph $G_{m,r}$ is strongly connected: for $0\le i<j\le m$ there is a directed path $s_i\to s_{i+1}\to\cdots\to s_j$, and for $1\le j\le i\le m$ there is a directed path $s_i\to b_{i,1}\to s_{i-1}\to\cdots\to s_j$; every $b_{i,j}$ reaches $s_{i-1}$ in one step and then proceeds as above, while each $s_\ell$ reaches any $b_{i,j}$ by going to $s_i$ first and then taking $(s_i,b_{i,j})$. Note that $|V_{m,r}|=(m+1)+rm=m(r+1)+1$.

Consider the simple random walk on $G_{m,r}$ that, at a vertex, chooses uniformly among outgoing edges. Let $\{X_t\}_{t\ge 0}$ be the walk and define the subset $S$ of ``spine'' vertices. Observe that from $s_i$ with $1\le i\le m-1$, the walk chooses $s_{i+1}$ with probability $1/(r+1)$ and moves to some $b_{i,j}$ with probability $r/(r+1)$, whence it deterministically proceeds to $s_{i-1}$. From $s_0$ the walk deterministically moves to $s_1$, and from $s_m$ it moves to some $b_{m,j}$ and then to $s_{m-1}$. It is therefore natural to consider the embedded chain on the spine obtained by observing the walk only at hitting times of $S$. Formally, let $\tau_0=0$ and, for $k\ge 0$, set $\tau_{k+1}=\inf\{t>\tau_k: X_t\in S\}$. Then $Y_k=X_{\tau_k}$ is a Markov chain on $\{s_0,\dots,s_m\}$ with transition probabilities
\[
\mathbb{P}(Y_{k+1}=s_{i+1}\mid Y_k=s_i)=\frac{1}{r+1},\qquad
\mathbb{P}(Y_{k+1}=s_{i-1}\mid Y_k=s_i)=\frac{r}{r+1}
\]
for $1\le i\le m-1$, together with $\mathbb{P}(Y_{k+1}=s_1\mid Y_k=s_0)=1$ and $\mathbb{P}(Y_{k+1}=s_{m-1}\mid Y_k=s_m)=1$. Thus, on the spine, $Y$ is a biased birth–death chain on $\{0,1,\dots,m\}$ with constant forward probability $p=1/(r+1)$ and backward probability $q=r/(r+1)$.

Let $\pi=\mathbb{P}_1(\text{$Y$ hits $m$ before $0$})$. For a birth–death chain with constant bias $p<q$ on $\{0,1,\dots,m\}$, the classical gambler’s ruin formula yields
\[
\pi=\frac{1-(q/p)}{1-(q/p)^m}=\frac{1-r}{1-r^m}=\frac{r-1}{r^m-1}.
\]
Since $r\ge 2$ and $m\ge 1$, we have $r^m-1\le r^m$ and hence
\[
\pi=\frac{r-1}{r^m-1}\le \frac{r-1}{r^m}\le \frac{1}{2}\,r^{-m+1}\le 2\,r^{-m}.
\]
Return to the original walk $X$. Starting from $s_0$, the walk moves to $s_1$ and then executes an excursion of the embedded chain $Y$ in $\{0,1,\dots,m\}$ until it returns to $0$ or reaches $m$. By the strong Markov property, these excursions are i.i.d., and the probability that a given excursion reaches $m$ before $0$ equals $\pi$. Let $N$ be the (a.s. finite) number of such excursions needed before the first success (i.e., before $s_m$ is reached). Then $N$ is geometric with mean $\mathbb{E}[N]=1/\pi$. Each excursion takes at least one step of the original walk $X$, so if $T_m=\inf\{t\ge 0: X_t=s_m\}$ denotes the hitting time of $s_m$, we have the lower bound
\[
\mathbb{E}_{s_0}[T_m]\ge \mathbb{E}[N]\ge \frac{1}{\pi}\ge \frac{1}{2}\,r^{m}.
\]
Since any cover time is at least the time to hit $s_m$, the expected cover time $\mathbb{E}_{s_0}[C(G_{m,r})]$ on $G_{m,r}$ satisfies
\[
\mathbb{E}_{s_0}[C(G_{m,r})]\ge \mathbb{E}_{s_0}[T_m]\ge \frac{1}{2}\,r^{m}.
\]
Relate $m$ to the total number of vertices $n=|V_{m,r}|$: from $n=m(r+1)+1$ we have $m=(n-1)/(r+1)$. Therefore
\[
\mathbb{E}_{s_0}[C(G_{m,r})]\ \ge\ \frac{1}{2}\,r^{(n-1)/(r+1)}\ =\ \exp\!\Big(\tfrac{\ln r}{r+1}\,n+O(1)\Big).
\]
In particular, for any fixed $r\ge 2$ and for all $k\in\mathbb{N}$, there exists $n_0=n_0(k,r)$ such that for all $n\ge n_0$,
\[
\mathbb{E}_{s_0}[C(G_{m,r})]\ \ge\ \exp\!\Big(\tfrac{\ln r}{r+1}\,n\Big)\ >\ n^{k}.
\]
This proves that the expected cover time on strongly connected directed graphs is not bounded above by any polynomial in the number of vertices. Taking, for instance, $r=2$ and defining $G_n:=G_{m,2}$ with $m=(n-1)/3$ for $n\equiv 1\pmod 3$, and extending arbitrarily to all $n$ by padding with a constant-size strongly connected gadget, yields a family $\{G_n\}_{n\ge 1}$ with $|V(G_n)|=n$ and $\mathbb{E}[C_n]=\omega(n^k)$ for every $k\in\mathbb{N}$. 
\end{proof}

\subsubsection{Grok}

\subsubsection{Gemini}
\begin{proof}
To prove the statement, it suffices to construct a family of strongly connected directed graphs $\{G_n\}_{n \geq 2}$ such that the expected cover time of $G_n$ grows exponentially with the number of vertices $n$. Since an exponential function grows faster than any polynomial function, this demonstrates that the expected cover time is not bounded above by any polynomial in $n$.

Let $G_n = (V, E)$ be a directed graph with vertex set $V = \{1, 2, \dots, n\}$. We define the set of directed edges $E$ as follows: for each vertex $i$ such that $1 \le i \le n-1$, there are two outgoing edges, $(i, i+1)$ and $(i, 1)$. For the vertex $n$, there is a single outgoing edge $(n, 1)$. The graph is strongly connected because there exists a directed path $1 \to 2 \to \dots \to n \to 1$ covering all vertices.

Consider a simple random walk on $G_n$. The transition probabilities $P_{u,v}$ are defined by $1/d_{\text{out}}(u)$ if $(u,v) \in E$ and $0$ otherwise. For $1 \le i \le n-1$, the out-degree is $d_{\text{out}}(i) = 2$. Thus, the transition probabilities are $P_{i, i+1} = 1/2$ and $P_{i, 1} = 1/2$. For vertex $n$, the out-degree is $d_{\text{out}}(n) = 1$, so $P_{n, 1} = 1$.

Let $h_i$ denote the expected hitting time to reach vertex $n$ starting from vertex $i$. The expected cover time starting from vertex $1$ is bounded from below by the expected time to reach vertex $n$, i.e., $\mathbb{E}[\text{cover time}] \ge h_1$. By the definition of expected hitting times, we have the boundary condition $h_n = 0$. For the internal vertices $1 \le i \le n-1$, the expected hitting time satisfies the recurrence relation:
\[
h_i = 1 + \sum_{j \in V} P_{i,j} h_j = 1 + \frac{1}{2}h_{i+1} + \frac{1}{2}h_1.
\]
We first analyze the equation for $i=1$:
\[
h_1 = 1 + \frac{1}{2}h_2 + \frac{1}{2}h_1.
\]
Subtracting $\frac{1}{2}h_1$ from both sides yields $\frac{1}{2}h_1 = 1 + \frac{1}{2}h_2$, which simplifies to:
\[
h_2 = h_1 - 2.
\]
Now consider the general recurrence for $1 \le i \le n-2$. Multiplying the relation $h_i = 1 + \frac{1}{2}h_{i+1} + \frac{1}{2}h_1$ by 2 gives $2h_i = 2 + h_{i+1} + h_1$. Solving for $h_{i+1}$, we obtain:
\[
h_{i+1} = 2h_i - h_1 - 2.
\]
We claim that for all $k \in \{1, \dots, n\}$, the solution to this recurrence is given by:
\[
h_k = h_1 - (2^k - 2).
\]
We prove this claim by induction on $k$. For the base case $k=1$, the formula gives $h_1 - (2^1 - 2) = h_1$, which is trivially true. For $k=2$, the formula gives $h_1 - (2^2 - 2) = h_1 - 2$, which matches the derived relation $h_2 = h_1 - 2$. Assume the hypothesis holds for some $k < n$. Then for $k+1$, we substitute the inductive hypothesis into the recurrence relation:
\[
h_{k+1} = 2(h_1 - (2^k - 2)) - h_1 - 2 = 2h_1 - 2^{k+1} + 4 - h_1 - 2 = h_1 - (2^{k+1} - 2).
\]
Thus, the formula holds for all $k \le n$. Specifically, for $k=n$, we have:
\[
h_n = h_1 - (2^n - 2).
\]
Using the boundary condition $h_n = 0$, we solve for $h_1$:
\[
0 = h_1 - (2^n - 2) \implies h_1 = 2^n - 2.
\]
The expected time to reach vertex $n$ starting from vertex $1$ is $2^n - 2$. Since the expected cover time must be at least the expected time to visit the last unvisited vertex, the expected cover time is $\Omega(2^n)$. For any polynomial function $p(n)$, there exists an $N$ such that for all $n > N$, $2^n - 2 > p(n)$. Therefore, the expected cover time is not bounded above by any polynomial function of $n$.
\end{proof}

\subsubsection{Claude}
\begin{proof}
We construct a family of strongly connected directed graphs $\{G_n\}_{n=1}^{\infty}$ for which the expected cover time is superpolynomial in $n = |V(G_n)|$.

For each positive integer $k$, define $G_k$ as follows. Let $n = 2k$ and construct a graph with vertex set $V(G_k) = \{v_0, v_1, \ldots, v_{k-1}, u_1, u_2, \ldots, u_k\}$. The edge set $E(G_k)$ is defined by:
\begin{itemize}
\item From $v_0$, there are two outgoing edges: one to $v_1$ (with probability $\frac{1}{2}$) and one to $u_1$ (with probability $\frac{1}{2}$).
\item For each $i \in \{1, 2, \ldots, k-1\}$, vertex $v_i$ has two outgoing edges: one to $v_{i+1}$ (probability $\frac{1}{2}$) and one returning to $v_0$ (probability $\frac{1}{2}$).
\item For each $j \in \{1, 2, \ldots, k-1\}$, vertex $u_j$ has one outgoing edge to $u_{j+1}$ (probability $1$).
\item Vertex $u_k$ has one outgoing edge returning to $v_0$ (probability $1$).
\end{itemize}

We verify that $G_k$ is strongly connected. From any vertex $v_i$ or $u_j$, there exists a directed path to $v_0$ (either directly or through the chain). From $v_0$, we can reach any $v_i$ by taking the edge to $v_1$ repeatedly and following the chain, and we can reach any $u_j$ by first reaching $u_1$ and then following the chain to $u_k$.

Now we analyze the expected cover time $C(G_k)$, defined as the expected number of steps for a random walk starting at $v_0$ to visit all vertices at least once.

To cover all vertices, the walk must visit $v_k$ (the last vertex in the $v$-chain). Consider the probability of reaching $v_k$ from $v_0$ before returning to $v_0$. Starting from $v_0$, with probability $\frac{1}{2}$ we go to $v_1$. From $v_1$, we must reach $v_2$ before returning to $v_0$, which occurs with probability $\frac{1}{2}$. Continuing this analysis, the probability of reaching $v_k$ from $v_0$ without returning to $v_0$ is:
\[
P(\text{reach } v_k \text{ from } v_0) = \left(\frac{1}{2}\right)^k = \frac{1}{2^k}.
\]

Each attempt to reach $v_k$ from $v_0$ is independent. The number of attempts needed follows a geometric distribution with success probability $p = \frac{1}{2^k}$. Therefore, the expected number of attempts is:
\[
\mathbb{E}[\text{number of attempts}] = \frac{1}{p} = 2^k.
\]

Each failed attempt returns to $v_0$ in expected time $O(k)$ (since the walk must either go through the $u$-chain of length $k$ or return from some $v_i$ to $v_0$ in at most $k$ steps). Thus, the expected cover time satisfies:
\[
C(G_k) \geq 2^k \cdot \Omega(1) = \Omega(2^k).
\]

Since $n = 2k$, we have $k = \frac{n}{2}$, and therefore:
\[
C(G_k) = \Omega(2^{n/2}) = \Omega\left((\sqrt{2})^n\right).
\]

This is superpolynomial in $n$. For any polynomial $p(n) = n^d$ with fixed degree $d$, we have:
\[
\lim_{n \to \infty} \frac{2^{n/2}}{n^d} = \infty.
\]

Therefore, the family $\{G_k\}_{k=1}^{\infty}$ demonstrates that the expected cover time for strongly connected directed graphs cannot be bounded above by any polynomial function of $n$.
\end{proof}

\subsection{Problem 17}

\subsubsection{Problem Statement}
\begin{theorem}
Any probabilistic, log-space, polynomial-time Turing machine can be simulated by a deterministic, non-uniform, log-space, polynomial-time Turing machine.
\end{theorem}

\begin{proof}[Proof sketch]
Let $M$ be a probabilistic Turing machine that runs in polynomial time $p(n)$ and uses logarithmic space $O(\log n)$. For each input length $n$, there are at most $2^{O(\log n)} = n^{O(1)}$ possible configurations of $M$ during its computation. Since $M$ runs in time $p(n)$ and makes random choices, the computation can be viewed as a tree of depth $p(n)$ where each node corresponds to a configuration.

We construct a deterministic non-uniform Turing machine $M'$ as follows: For each input length $n$, the advice string encodes the best random choice at each decision point that maximizes the probability of $M$ accepting. Since there are at most $p(n)$ random choices and each configuration uses $O(\log n)$ space, the advice string has length polynomial in $n$.

The deterministic machine $M'$ uses this advice to simulate $M$ by following the predetermined random choices. $M'$ uses the same logarithmic space as $M$ and runs in the same polynomial time $p(n)$, thus achieving the desired simulation.
\end{proof}

\subsubsection{GPT}
\begin{proof}
Let $M$ be a probabilistic, log-space, polynomial-time Turing machine. We aim to show that there exists a deterministic, non-uniform, log-space, polynomial-time Turing machine that can simulate $M$.

First, recall the definitions of the relevant classes:

- A probabilistic Turing machine is a Turing machine that has access to a source of random bits, and its computation may branch based on these random choices.
- A log-space machine is one whose tape space usage is bounded by $O(\log n)$, where $n$ is the size of the input.
- A polynomial-time machine is one whose running time is bounded by a polynomial function of the input size.

Step 1: Simulation of Randomness

We begin by addressing the randomness used by the probabilistic machine $M$. Suppose $M$ uses $k$ random bits for each computation step. A non-uniform deterministic machine can simulate the randomness of $M$ by hardcoding the possible random strings into its transition function.

Let $M'$ be the deterministic, non-uniform Turing machine that simulates $M$. $M'$ is allowed to use polynomial-sized advice to encode all possible random strings that $M$ might use in its computation. Specifically, for an input of size $n$, the advice string will contain all possible strings of length $k = O(\log n)$, since the number of random bits used is at most logarithmic in the size of the input.

Thus, $M'$ can use the advice string to simulate all possible random choices made by $M$. For each branch of the computation that $M$ might take, $M'$ will simulate the corresponding branch deterministically by selecting the appropriate random string from the advice.

Step 2: Log-Space Simulation

Next, we show that $M'$ can simulate the log-space requirements of $M$. Since $M$ uses only $O(\log n)$ space, $M'$ will also use $O(\log n)$ space. Specifically, $M'$ will simulate $M$'s tape and internal states while keeping track of the random string (from the advice) that $M$ would have chosen. The advice itself is stored in polynomial space, but the simulation itself operates within $O(\log n)$ space.

Step 3: Polynomial-Time Simulation

We now argue that the simulation of $M$ by $M'$ runs in polynomial time. Since $M$ operates in polynomial time, its computation is bounded by $O(n^d)$ for some constant $d$. The number of random bits used by $M$ is at most $O(\log n)$, and the advice string has polynomial size. Each step of the simulation by $M'$ involves processing the input, accessing the advice, and simulating the computation of $M$, which can be done in polynomial time. Therefore, the overall running time of $M'$ is bounded by a polynomial in $n$, the size of the input.

   Conclusion

Thus, we have constructed a deterministic, non-uniform, log-space, polynomial-time Turing machine $M'$ that simulates a probabilistic, log-space, polynomial-time Turing machine $M$. This completes the proof.
\end{proof}

\subsubsection{Grok}

\subsubsection{Gemini}
\begin{proof}
Let $L$ be a language decided by a probabilistic Turing machine $\mathcal{M}$ operating in space $O(\log n)$ and time $poly(n)$ with bounded error probability. Specifically, we assume that for any input $x$, $\mathcal{M}$ outputs the correct answer with probability at least $2/3$. We aim to show that $L$ can be decided by a deterministic Turing machine with non-uniform advice of polynomial length, operating in logarithmic space.

First, we apply error amplification. Let $n$ denote the length of the input $x$. We construct a probabilistic machine $\mathcal{M}'$ that simulates $\mathcal{M}$ on input $x$ for $t(n)$ independent trials and outputs the majority vote of the results. By the Chernoff bound, we can choose $t(n) = O(n)$ such that the error probability of $\mathcal{M}'$ on any specific input $x$ is strictly less than $2^{-n}$. Since $\mathcal{M}$ runs in polynomial time, let $p(n)$ be the number of random bits used by $\mathcal{M}$ in a single run. The total number of random bits used by $\mathcal{M}'$ is $m(n) = t(n) \cdot p(n)$, which is a polynomial in $n$. Furthermore, $\mathcal{M}'$ operates in $O(\log n)$ space by reusing the work tape of $\mathcal{M}$ for each trial and maintaining a counter for the majority vote, which requires $O(\log n)$ bits.

Next, we employ the probabilistic method to demonstrate the existence of a suitable advice string. Fix the input length $n$. Let $R = \{0,1\}^{m(n)}$ be the set of all possible random strings of length $m(n)$. For a fixed input $x \in \{0,1\}^n$, let $B_x \subset R$ be the set of "bad" random strings for which $\mathcal{M}'(x, r)$ yields the incorrect result. From the error amplification step, we have $|B_x| < 2^{m(n)} \cdot 2^{-n}$.

We now consider the set of random strings that lead to an incorrect output for \emph{at least one} input $x$ of length $n$. Let $B = \bigcup_{x \in \{0,1\}^n} B_x$. By the union bound, the cardinality of $B$ satisfies
\[
|B| \leq \sum_{x \in \{0,1\}^n} |B_x| < 2^n \cdot (2^{m(n)} \cdot 2^{-n}) = 2^{m(n)}.
\]
Since $|B| < 2^{m(n)}$, the set $R \setminus B$ is non-empty. Therefore, there exists at least one string $r_n \in \{0,1\}^{m(n)}$ such that for all $x \in \{0,1\}^n$, $\mathcal{M}'(x, r_n)$ outputs the correct decision for whether $x \in L$.

We define a deterministic non-uniform Turing machine $\mathcal{D}$ that decides $L$. The machine $\mathcal{D}$ receives an advice sequence $\{a_n\}_{n \in \mathbb{N}}$, where the advice for input length $n$ is defined as $a_n = r_n$. The length of the advice is $|a_n| = m(n)$, which is polynomial in $n$. On input $x$ of length $n$, $\mathcal{D}$ simulates the execution of $\mathcal{M}'$ on $x$. Whenever $\mathcal{M}'$ queries the $i$-th bit of its random tape, $\mathcal{D}$ reads the $i$-th bit of the advice string $a_n$ from the advice tape.

Finally, we analyze the complexity of $\mathcal{D}$. The simulation is deterministic given the advice. The space required by $\mathcal{D}$ consists of the work space of $\mathcal{M}'$ and the space needed to track the position of the head on the advice tape. The work space of $\mathcal{M}'$ is $O(\log n)$. Since the advice length $m(n)$ is polynomial in $n$, the index of the advice bit requires $O(\log m(n)) = O(\log(poly(n))) = O(\log n)$ bits. Thus, the total space complexity of $\mathcal{D}$ is $O(\log n)$. The time complexity corresponds to the running time of $\mathcal{M}'$, which is polynomial in $n$. Consequently, $\mathcal{D}$ is a deterministic, non-uniform, log-space, polynomial-time Turing machine that decides $L$.
\end{proof}

\subsubsection{Claude}
\begin{proof}
Let $M = (Q, \Sigma, \Gamma, \delta, q_0, q_{\text{acc}}, q_{\text{rej}})$ be a probabilistic Turing machine that operates in polynomial time $p(n)$ and logarithmic space $s(n) = O(\log n)$, where $p$ and $s$ are polynomials. By definition, $\delta: Q \times \Gamma \to Q \times \Gamma \times \{L, R\} \times \{0, 1\}$ is the transition function, where the last component represents a random bit.

Fix an input length $n$. At each step of computation, the state of $M$ is determined by the current state $q \in Q$, the head position, and the contents of the work tape. Since $M$ uses space $O(\log n)$, the work tape can store at most $O(\log n)$ cells. The number of distinct configurations is therefore bounded by $|Q| \cdot p(n) \cdot |\Gamma|^{O(\log n)} \leq n^{c}$ for some constant $c$.

The computation of $M$ on input $x$ of length $n$ can be viewed as a computation tree $T_n$ of depth at most $p(n)$, where each node represents a configuration and each edge corresponds to one of two possible random choices (resulting from the random bit output by $\delta$). The tree has at most $2^{p(n)}$ leaves corresponding to accepting or rejecting outcomes.

By the definition of probabilistic polynomial-time computation, there exists a positive constant $k$ such that $M$ accepts with probability at least $\frac{1}{2}$ (or at most $\frac{1}{2}$, depending on the problem class). This means that within the computation tree, the majority of leaves lead to acceptance.

We now construct a deterministic non-uniform Turing machine $M'$ with advice string $a_n$ for each input length $n$. The advice string $a_n$ encodes, for each possible configuration of $M$ and each step $1 \leq t \leq p(n)$, which random bit choice (0 or 1) leads to acceptance. More precisely, $a_n$ contains a binary string of length at most $p(n) \cdot \log n^{c} = O(p(n) \log n)$, which is polynomial in $n$.

The machine $M'$ operates as follows: On input $x$ of length $n$, $M'$ simulates $M$ step by step, using the advice string $a_n$ to determine which random choice to make at each decision point. Specifically, at each step where $M$ would make a random choice, $M'$ consults $a_n$ to determine the predetermined bit and follows that branch deterministically.

Since $M'$ follows the predetermined choices encoded in $a_n$, it computes the same sequence of configurations as one specific execution path of $M$. The space usage of $M'$ is identical to that of $M$, namely $O(\log n)$, because $M'$ only needs to maintain the work tape of $M$ and the advice is read-only. The time complexity remains $p(n)$ as $M'$ simulates each step of $M$ in constant additional time.

By the probabilistic method and the properties of $M$, we can choose the advice string $a_n$ such that for each input length $n$, the deterministic execution of $M'$ follows a path that computes the correct answer. Specifically, we set $a_n$ to encode the choices corresponding to a path in $T_n$ that leads to the correct output (acceptance or rejection according to the language accepted by $M$).

Therefore, $M'$ is a deterministic, non-uniform, log-space, polynomial-time Turing machine that simulates $M$, completing the proof.
\end{proof}

\subsection{Problem 18}

\subsubsection{Problem Statement}
\begin{theorem}[Due to D. Zuckerman]
Let $G(V, E)$ be a graph with $n$ vertices such that for some constant $\alpha > 0$ and every set $S \subseteq V$ with $n/2$ vertices,
\[
|\{w \in V \mid \exists v \in S, (v, w) \in E\}| \geq \frac{n}{2} + \alpha n.
\]
For any positive integer $k$, let $W_1, \ldots, W_k$ be subsets of $V$ of size at least $(1 - \alpha)n$ each. Then there exists a path $(v_1, \ldots, v_k)$ in $G$ such that, for $1 \leq i \leq k$, $v_i \in W_i$.
\end{theorem}

\subsubsection{GPT}
\begin{proof}
Let $G(V, E)$ be a graph with $n$ vertices such that for some constant $\alpha > 0$ and every set $S \subseteq V$ with $|S| = n/2$, we have

\[
|\{w \in V \mid \exists v \in S, (v, w) \in E\}| \geq \frac{n}{2} + \alpha n.
\]

Let $k$ be a positive integer and let $W_1, W_2, \ldots, W_k$ be subsets of $V$ such that for each $1 \leq i \leq k$, we have $|W_i| \geq (1 - \alpha)n$. We wish to prove that there exists a path $(v_1, v_2, \ldots, v_k)$ in $G$ such that for each $1 \leq i \leq k$, $v_i \in W_i$.

We proceed by induction on $k$.

\textbf{Base Case:} $k = 1$. In this case, the statement is trivially true, as we can choose any vertex $v_1 \in W_1$. The path consists of just this single vertex.

\textbf{Inductive Step:} Assume the statement holds for some $k \geq 1$. We now show that there exists a path $(v_1, v_2, \ldots, v_{k+1})$ such that $v_i \in W_i$ for all $1 \leq i \leq k+1$.

Consider the sets $W_1, W_2, \ldots, W_{k+1}$ and let us define a sequence of sets $S_1, S_2, \ldots, S_k$ where each $S_i$ is the set of vertices in $V$ that are adjacent to some vertex in $W_i$. By the assumption of the theorem, for each $S_i \subseteq V$ with $|S_i| = n/2$, we have

\[
|\{w \in V \mid \exists v \in S_i, (v, w) \in E\}| \geq \frac{n}{2} + \alpha n.
\]

This condition guarantees that the sets $W_1, W_2, \ldots, W_{k+1}$ can be connected in a sequential path by appropriately selecting vertices from each set. Therefore, by applying the result of the induction hypothesis, the path $(v_1, v_2, \ldots, v_k, v_{k+1})$ can be constructed such that for each $1 \leq i \leq k+1$, we have $v_i \in W_i$.

Thus, by induction, the statement holds for all $k \geq 1$. This completes the proof.

\end{proof}

\subsubsection{Grok}

\subsubsection{Gemini}
\begin{proof}
Let $G(V, E)$ be a graph with $n$ vertices satisfying the expansion property that for every $S \subseteq V$ with $|S| = n/2$, $|N(S)| \geq n/2 + \alpha n$, where $N(S) = \{w \in V \mid \exists v \in S, (v, w) \in E\}$. Note that since $N(S) \subseteq V$, we must have $n/2 + \alpha n \leq n$, which implies $\alpha \leq 1/2$. Consequently, for any $i$, the size of the set $W_i$ satisfies $|W_i| \geq (1 - \alpha)n \geq n/2$.

We define $R_i$ to be the set of vertices in $W_i$ that can be reached by a path $(v_1, \ldots, v_i)$ such that $v_j \in W_j$ for all $1 \leq j \leq i$. Formally, let $R_1 = W_1$, and for $i > 1$, let $R_i = \{v \in W_i \mid \exists u \in R_{i-1} \text{ such that } (u, v) \in E\} = N(R_{i-1}) \cap W_i$. We will prove by induction on $i$ that $|R_i| \geq n/2$ for all $1 \leq i \leq k$.

For the base case $i=1$, we have $R_1 = W_1$. By the hypothesis of the theorem, $|W_1| \geq (1 - \alpha)n$. As established above, $\alpha \leq 1/2$, so $|R_1| \geq n/2$.

For the inductive step, assume that $|R_{i-1}| \geq n/2$ for some $i > 1$. We wish to show that $|R_i| \geq n/2$. Since $|R_{i-1}| \geq n/2$, there exists a subset $S \subseteq R_{i-1}$ such that $|S| = n/2$. By the expansion property of the graph $G$, we have $|N(S)| \geq n/2 + \alpha n$. Since $S \subseteq R_{i-1}$, it follows that $N(S) \subseteq N(R_{i-1})$, and therefore $|N(R_{i-1})| \geq n/2 + \alpha n$.

The set $R_i$ is defined as the intersection of the neighborhood of the reachable set from the previous step and the target set $W_i$. Using the principle of inclusion-exclusion, we have $|R_i| = |N(R_{i-1}) \cap W_i| = |N(R_{i-1})| + |W_i| - |N(R_{i-1}) \cup W_i|$. Since $N(R_{i-1}) \cup W_i \subseteq V$, we know that $|N(R_{i-1}) \cup W_i| \leq n$. Substituting the lower bounds for $|N(R_{i-1})|$ and $|W_i|$, we obtain:
\[
|R_i| \geq \left(\frac{n}{2} + \alpha n\right) + (1 - \alpha)n - n.
\]
Simplifying the right-hand side yields:
\[
|R_i| \geq \frac{n}{2} + \alpha n + n - \alpha n - n = \frac{n}{2}.
\]
Thus, the inductive step holds, and $|R_i| \geq n/2$ for all $1 \leq i \leq k$.

Since $|R_k| \geq n/2$ and $n \geq 1$, the set $R_k$ is non-empty. Let $v_k$ be an element of $R_k$. By the definition of the sets $R_i$, the existence of $v_k \in R_k$ implies the existence of a vertex $v_{k-1} \in R_{k-1}$ such that $(v_{k-1}, v_k) \in E$. By backward induction, for each $j$ from $k$ down to $2$, the existence of $v_j \in R_j$ implies the existence of $v_{j-1} \in R_{j-1}$ such that $(v_{j-1}, v_j) \in E$. This constructs a sequence $(v_1, \ldots, v_k)$ such that $(v_j, v_{j+1}) \in E$ for all $1 \leq j < k$, and $v_j \in W_j$ for all $1 \leq j \leq k$. Therefore, the required path exists.
\end{proof}

\subsubsection{Claude}
\begin{proof}
We prove the theorem by induction on $k$.

\textbf{Base case:} For $k = 1$, we need to show that $W_1$ is nonempty. Since $|W_1| \geq (1 - \alpha)n > 0$, there exists $v_1 \in W_1$, which trivially forms a path of length $1$.

\textbf{Inductive step:} Assume the statement holds for some $k \geq 1$. We prove it for $k + 1$.

Let $W_1, \ldots, W_k, W_{k+1}$ be subsets of $V$ with $|W_i| \geq (1 - \alpha)n$ for each $i \in \{1, \ldots, k+1\}$. By the inductive hypothesis, there exists a path $(v_1, \ldots, v_k)$ in $G$ such that $v_i \in W_i$ for $1 \leq i \leq k$.

We will show that there exists $v_{k+1} \in W_{k+1}$ such that $(v_k, v_{k+1}) \in E$, which extends the path to $(v_1, \ldots, v_k, v_{k+1})$.

Define $S = \{v_k\}$ and consider any set $T \subseteq V$ with $|T| = n/2$ such that $v_k \in T$. By the expansion property of the graph, the neighborhood of $T$ satisfies
\[
|\{w \in V \mid \exists v \in T, (v, w) \in E\}| \geq \frac{n}{2} + \alpha n.
\]

Let $N(v_k) = \{w \in V \mid (v_k, w) \in E\}$ denote the neighborhood of $v_k$. We need to establish a lower bound on $|N(v_k)|$.

Consider the set $T = \{v_k\} \cup T'$ where $T'$ is any subset of $V \setminus \{v_k\}$ of size $n/2 - 1$ (if $n$ is odd, take $\lfloor n/2 \rfloor$ or $\lceil n/2 \rceil$ vertices appropriately). Then $|T| = n/2$, and by the expansion property,
\[
|\{w \in V \mid \exists v \in T, (v, w) \in E\}| \geq \frac{n}{2} + \alpha n.
\]

Since the neighborhood of $T$ includes $N(v_k)$, and the neighborhood of $T'$ can include at most $n$ vertices, we have that $N(v_k)$ must have sufficient size. More precisely, for any set $S$ of size $n/2$ containing $v_k$, the neighborhood of $S$ has size at least $n/2 + \alpha n$.

Now we count: $|W_{k+1}| \geq (1 - \alpha)n$ and $|V \setminus N(v_k)| \leq n - |N(v_k)|$. We need to show that $W_{k+1} \cap N(v_k) \neq \emptyset$.

Consider the contrapositive: suppose $W_{k+1} \cap N(v_k) = \emptyset$. Then $W_{k+1} \subseteq V \setminus N(v_k)$.

Let $S$ be any set of size $n/2$ containing $v_k$. The expansion property guarantees that the set of vertices adjacent to at least one vertex in $S$ has size at least $n/2 + \alpha n$. In particular, taking $S$ to be $\{v_k\}$ together with $(n/2 - 1)$ other vertices, we see that the union of neighborhoods of vertices in $S$ has size at least $n/2 + \alpha n$.

Since $|W_{k+1}| \geq (1 - \alpha)n$, we have $|V \setminus W_{k+1}| \leq n - (1 - \alpha)n = \alpha n$.

If $W_{k+1} \cap N(v_k) = \emptyset$, then $N(v_k) \subseteq V \setminus W_{k+1}$, so $|N(v_k)| \leq \alpha n$.

But consider the set $S$ consisting of $v_k$ and $n/2 - 1$ vertices from $W_{k+1}$. The neighborhood of $S$ includes $N(v_k)$ and is disjoint from $W_{k+1}$ only for $N(v_k)$. By the expansion property, $|\text{Neighborhood}(S)| \geq n/2 + \alpha n$. Since vertices in $S \setminus \{v_k\} \subseteq W_{k+1}$ and assuming their neighborhoods are small outside $W_{k+1}$ leads to a contradiction with the bound $(1-\alpha)n + \alpha n < n/2 + \alpha n$ when $\alpha$ is sufficiently related to the structure.

More directly: since $|V \setminus W_{k+1}| \leq \alpha n$ and the expansion property guarantees large neighborhoods, we must have $|N(v_k) \cap W_{k+1}| \geq |N(v_k)| - \alpha n > 0$ for appropriate neighborhood size, giving $N(v_k) \cap W_{k+1} \neq \emptyset$.

Therefore, there exists $v_{k+1} \in W_{k+1} \cap N(v_k)$, and $(v_1, \ldots, v_k, v_{k+1})$ is the desired path.
\end{proof}

\subsection{Problem 19}

\subsubsection{Problem Statement}
\begin{theorem}[Courant-Fisher equalities]
Let $A$ be an $n \times n$ symmetric matrix with real entries, and let $e_1$ denote the eigenvector corresponding to the first eigenvalue $\lambda_1$. Then:

(1) $\lambda_1 = \max\{x^T A x : \|x\| = 1\}$, where the max is taken over $x$ such that $\|x\| = 1$.

(2) $\lambda_n = \min\{x^T A x : \|x\| = 1\}$, where the min is taken over $x$ such that $\|x\| = 1$.

(3) $\lambda_2 = \max\{x^T A x : \|x\| = 1 \text{ and } x^T e_1 = 0\}$, where the max is taken over $x$ such that $\|x\| = 1$ and $x^T e_1 = 0$.
\end{theorem}

\subsubsection{GPT}
\begin{proof}
Let $A$ be an $n \times n$ symmetric matrix with real entries. Since $A$ is symmetric, it has an orthonormal basis of eigenvectors, and all its eigenvalues are real. We denote the eigenvalues of $A$ by $\lambda_1 \geq \lambda_2 \geq \dots \geq \lambda_n$, and the corresponding eigenvectors by $e_1, e_2, \dots, e_n$, where $\|e_i\| = 1$ for all $i$.

(1) To prove that $\lambda_1 = \max\{x^T A x : \|x\| = 1\}$, we first recall the Rayleigh quotient, which for any vector $x \in \mathbb{R}^n$ is given by:

\[
R(x) = \frac{x^T A x}{x^T x}.
\]

For any unit vector $x$ (i.e., $\|x\| = 1$), we have $R(x) = x^T A x$. The key idea is that the Rayleigh quotient achieves its maximum value at the eigenvector corresponding to the largest eigenvalue. Specifically, since $A$ is symmetric, we can write $x$ as a linear combination of the eigenvectors of $A$, i.e.,

\[
x = \sum_{i=1}^n \alpha_i e_i.
\]

Because the eigenvectors $e_1, e_2, \dots, e_n$ form an orthonormal basis, we have

\[
x^T A x = \sum_{i=1}^n \alpha_i^2 \lambda_i.
\]

Given that $\|x\| = 1$, we also have

\[
x^T x = \sum_{i=1}^n \alpha_i^2 = 1.
\]

Thus, the Rayleigh quotient becomes

\[
R(x) = \sum_{i=1}^n \alpha_i^2 \lambda_i.
\]

Since $\lambda_1 \geq \lambda_2 \geq \dots \geq \lambda_n$, it follows that the maximum value of $R(x)$ occurs when $\alpha_1 = 1$ and all other $\alpha_i = 0$ for $i \geq 2$. In this case, we have

\[
R(x) = \lambda_1.
\]

Therefore, we conclude that

\[
\lambda_1 = \max\{x^T A x : \|x\| = 1\}.
\]

(2) Similarly, to prove that $\lambda_n = \min\{x^T A x : \|x\| = 1\}$, we observe that the Rayleigh quotient $R(x) = x^T A x$ is minimized when $x$ is aligned with the eigenvector corresponding to the smallest eigenvalue $\lambda_n$. Following the same reasoning as in part (1), we see that the minimum value of $R(x)$ occurs when $\alpha_n = 1$ and all other $\alpha_i = 0$ for $i \leq n-1$. In this case, we have

\[
R(x) = \lambda_n.
\]

Thus, we conclude that

\[
\lambda_n = \min\{x^T A x : \|x\| = 1\}.
\]

(3) To prove that $\lambda_2 = \max\{x^T A x : \|x\| = 1 \text{ and } x^T e_1 = 0\}$, we again consider the Rayleigh quotient. This time, we restrict the vector $x$ to the subspace orthogonal to $e_1$, meaning that $x^T e_1 = 0$. The subspace orthogonal to $e_1$ is spanned by the eigenvectors $e_2, e_3, \dots, e_n$, and any vector $x$ satisfying $x^T e_1 = 0$ can be written as

\[
x = \sum_{i=2}^n \alpha_i e_i.
\]

Since $x^T x = 1$ and $x^T e_1 = 0$, we have

\[
\sum_{i=2}^n \alpha_i^2 = 1.
\]

The Rayleigh quotient for such an $x$ is

\[
R(x) = \sum_{i=2}^n \alpha_i^2 \lambda_i.
\]

The maximum value of this sum occurs when $\alpha_2 = 1$ and all other $\alpha_i = 0$ for $i \geq 3$, which gives

\[
R(x) = \lambda_2.
\]

Therefore, we conclude that

\[
\lambda_2 = \max\{x^T A x : \|x\| = 1 \text{ and } x^T e_1 = 0\}.
\]

This completes the proof of the Courant-Fisher equalities.
\end{proof}

\subsubsection{Grok}
\begin{proof}
Since $A$ is a real symmetric matrix, it is orthogonally diagonalizable. That is, there exists an orthonormal basis $\{e_1, \dots, e_n\}$ of $\mathbb{R}^n$ consisting of eigenvectors of $A$ with corresponding real eigenvalues $\lambda_1 \geq \lambda_2 \geq \dots \geq \lambda_n$, so that $A e_i = \lambda_i e_i$ for each $i=1,\dots,n$.

To prove (1), let $x \in \mathbb{R}^n$ satisfy $\|x\|=1$. Then $x = \sum_{i=1}^n \alpha_i e_i$ for some $\alpha_1, \dots, \alpha_n \in \mathbb{R}$ with $\sum_{i=1}^n \alpha_i^2 = 1$. Thus,
\[
x^T A x = x^T \left( \sum_{i=1}^n \alpha_i \lambda_i e_i \right) = \sum_{i=1}^n \alpha_i^2 \lambda_i \leq \lambda_1 \sum_{i=1}^n \alpha_i^2 = \lambda_1,
\]
with equality if and only if $\alpha_i = 0$ for all $i$ such that $\lambda_i < \lambda_1$, i.e., when $x = \pm e_1$. Therefore,
\[
\lambda_1 = \max\{ x^T A x : \|x\|=1 \}.
\]

To prove (2), let $x \in \mathbb{R}^n$ satisfy $\|x\|=1$. As above,
\[
x^T A x = \sum_{i=1}^n \alpha_i^2 \lambda_i \geq \lambda_n \sum_{i=1}^n \alpha_i^2 = \lambda_n,
\]
with equality if and only if $\alpha_i = 0$ for all $i$ such that $\lambda_i > \lambda_n$, i.e., when $x = \pm e_n$. Therefore,
\[
\lambda_n = \min\{ x^T A x : \|x\|=1 \}.
\]

To prove (3), let $x \in \mathbb{R}^n$ satisfy $\|x\|=1$ and $x^T e_1 = 0$. Then $x = \sum_{i=2}^n \alpha_i e_i$ for some $\alpha_2, \dots, \alpha_n \in \mathbb{R}$ with $\sum_{i=2}^n \alpha_i^2 = 1$.
Thus,
\[
x^T A x = \sum_{i=2}^n \alpha_i^2 \lambda_i \leq \lambda_2 \sum_{i=2}^n \alpha_i^2 = \lambda_2,
\]
with equality if and only if $\alpha_i = 0$ for all $i \geq 3$ such that $\lambda_i < \lambda_2$, i.e., when $x = \pm e_2$. Therefore,
\[
\lambda_2 = \max\{ x^T A x : \|x\|=1 \text{ and } x^T e_1 = 0 \}.
\]
\end{proof}

\subsubsection{Gemini}
\begin{proof}
Since $A$ is a real symmetric matrix, by the Spectral Theorem, there exists an orthonormal basis of $\mathbb{R}^n$ consisting of eigenvectors of $A$. Let $\{u_1, u_2, \dots, u_n\}$ denote these orthonormal eigenvectors corresponding to the eigenvalues $\lambda_1, \lambda_2, \dots, \lambda_n$. Without loss of generality, we assume the eigenvalues are ordered such that $\lambda_1 \ge \lambda_2 \ge \dots \ge \lambda_n$. Per the theorem statement, we identify $u_1 = e_1$.

Any vector $x \in \mathbb{R}^n$ can be expressed as a linear combination of the basis vectors:
\[
x = \sum_{i=1}^n c_i u_i, \quad \text{where } c_i = x^T u_i.
\]
The condition $\|x\| = 1$ implies
\[
\|x\|^2 = x^T x = \left(\sum_{i=1}^n c_i u_i\right)^T \left(\sum_{j=1}^n c_j u_j\right) = \sum_{i=1}^n c_i^2 = 1,
\]
due to the orthonormality of the $u_i$. The quadratic form $x^T A x$ can be expanded as:
\[
x^T A x = \left(\sum_{i=1}^n c_i u_i\right)^T \left(\sum_{j=1}^n c_j \lambda_j u_j\right) = \sum_{i=1}^n \sum_{j=1}^n c_i c_j \lambda_j \delta_{ij} = \sum_{i=1}^n \lambda_i c_i^2.
\]

To prove statement (1), we consider the maximization of the Rayleigh quotient $R(x) = x^T A x$ over the unit sphere. Using the expansion derived above:
\[
x^T A x = \sum_{i=1}^n \lambda_i c_i^2 \le \sum_{i=1}^n \lambda_1 c_i^2 = \lambda_1 \sum_{i=1}^n c_i^2 = \lambda_1.
\]
This establishes $\lambda_1$ as an upper bound. To show this bound is attained, let $x = u_1 = e_1$. Then $\|x\|=1$ and
\[
e_1^T A e_1 = u_1^T (\lambda_1 u_1) = \lambda_1 u_1^T u_1 = \lambda_1.
\]
Thus, $\lambda_1 = \max\{x^T A x : \|x\| = 1\}$.

To prove statement (2), we consider the minimization of $x^T A x$ over the unit sphere. Using the inequality $\lambda_i \ge \lambda_n$ for all $i$:
\[
x^T A x = \sum_{i=1}^n \lambda_i c_i^2 \ge \sum_{i=1}^n \lambda_n c_i^2 = \lambda_n \sum_{i=1}^n c_i^2 = \lambda_n.
\]
This establishes $\lambda_n$ as a lower bound. To show this bound is attained, let $x = u_n$. Then $\|x\|=1$ and $u_n^T A u_n = \lambda_n$. Thus, $\lambda_n = \min\{x^T A x : \|x\| = 1\}$.

To prove statement (3), we maximize $x^T A x$ subject to the constraints $\|x\| = 1$ and $x^T e_1 = 0$. Recall that $e_1 = u_1$. The orthogonality constraint implies:
\[
x^T u_1 = \left(\sum_{i=1}^n c_i u_i\right)^T u_1 = c_1 = 0.
\]
Consequently, any feasible $x$ lies in the subspace spanned by $\{u_2, \dots, u_n\}$, so $x = \sum_{i=2}^n c_i u_i$. The normalization condition becomes $\sum_{i=2}^n c_i^2 = 1$. The quadratic form reduces to:
\[
x^T A x = \sum_{i=2}^n \lambda_i c_i^2.
\]
Since the eigenvalues are ordered $\lambda_1 \ge \lambda_2 \ge \dots \ge \lambda_n$, we have $\lambda_i \le \lambda_2$ for all $i \ge 2$. Therefore:
\[
x^T A x = \sum_{i=2}^n \lambda_i c_i^2 \le \sum_{i=2}^n \lambda_2 c_i^2 = \lambda_2 \sum_{i=2}^n c_i^2 = \lambda_2.
\]
This establishes $\lambda_2$ as an upper bound under the given constraints. To show this bound is attained, let $x = u_2$. Since $\{u_i\}$ is an orthonormal basis, $\|u_2\|=1$ and $u_2^T e_1 = u_2^T u_1 = 0$. Furthermore, $u_2^T A u_2 = \lambda_2$. Thus, $\lambda_2 = \max\{x^T A x : \|x\| = 1 \text{ and } x^T e_1 = 0\}$.
\end{proof}

\subsubsection{Claude}
\begin{proof}
We begin by ordering the eigenvalues of $A$ as $\lambda_1 \geq \lambda_2 \geq \cdots \geq \lambda_n$, with corresponding orthonormal eigenvectors $e_1, e_2, \ldots, e_n$. Since $A$ is a real symmetric matrix, by the spectral theorem, these eigenvectors form an orthonormal basis for $\mathbb{R}^n$.

\textbf{Proof of (1):} We first show that $\lambda_1 \geq x^T A x$ for all $x$ with $\|x\| = 1$. Let $x \in \mathbb{R}^n$ with $\|x\| = 1$. Since $\{e_1, \ldots, e_n\}$ is an orthonormal basis, we can write $x = \sum_{i=1}^{n} \alpha_i e_i$ where $\alpha_i = x^T e_i$. Note that $\sum_{i=1}^{n} \alpha_i^2 = \|x\|^2 = 1$. Then
\[
x^T A x = x^T \left(\sum_{i=1}^{n} \alpha_i A e_i\right) = x^T \left(\sum_{i=1}^{n} \alpha_i \lambda_i e_i\right) = \sum_{i=1}^{n} \alpha_i \lambda_i (x^T e_i) = \sum_{i=1}^{n} \alpha_i^2 \lambda_i.
\]
Since $\lambda_1 \geq \lambda_i$ for all $i$ and $\sum_{i=1}^{n} \alpha_i^2 = 1$, we have
\[
x^T A x = \sum_{i=1}^{n} \alpha_i^2 \lambda_i \leq \sum_{i=1}^{n} \alpha_i^2 \lambda_1 = \lambda_1.
\]
Moreover, equality holds when $x = e_1$, since $e_1^T A e_1 = e_1^T \lambda_1 e_1 = \lambda_1 \|e_1\|^2 = \lambda_1$. Therefore, $\lambda_1 = \max\{x^T A x : \|x\| = 1\}$.

\textbf{Proof of (2):} By a symmetric argument, for any $x$ with $\|x\| = 1$, writing $x = \sum_{i=1}^{n} \alpha_i e_i$ with $\sum_{i=1}^{n} \alpha_i^2 = 1$, we have
\[
x^T A x = \sum_{i=1}^{n} \alpha_i^2 \lambda_i \geq \sum_{i=1}^{n} \alpha_i^2 \lambda_n = \lambda_n,
\]
since $\lambda_n \leq \lambda_i$ for all $i$. Equality holds when $x = e_n$, as $e_n^T A e_n = \lambda_n$. Therefore, $\lambda_n = \min\{x^T A x : \|x\| = 1\}$.

\textbf{Proof of (3):} We show that $\lambda_2 = \max\{x^T A x : \|x\| = 1 \text{ and } x^T e_1 = 0\}$. Let $x \in \mathbb{R}^n$ with $\|x\| = 1$ and $x^T e_1 = 0$. Writing $x = \sum_{i=1}^{n} \alpha_i e_i$, the constraint $x^T e_1 = 0$ implies $\alpha_1 = 0$. Thus,
\[
x^T A x = \sum_{i=1}^{n} \alpha_i^2 \lambda_i = \sum_{i=2}^{n} \alpha_i^2 \lambda_i \leq \sum_{i=2}^{n} \alpha_i^2 \lambda_2 = \lambda_2 \sum_{i=2}^{n} \alpha_i^2.
\]
Since $\|x\|^2 = \sum_{i=1}^{n} \alpha_i^2 = \sum_{i=2}^{n} \alpha_i^2 = 1$, we obtain $x^T A x \leq \lambda_2$. Equality is achieved when $x = e_2$, since $e_2^T e_1 = 0$ by orthonormality, $\|e_2\| = 1$, and $e_2^T A e_2 = \lambda_2$. Therefore, $\lambda_2 = \max\{x^T A x : \|x\| = 1 \text{ and } x^T e_1 = 0\}$.
\end{proof}

\subsection{Problem 20}

\subsubsection{Problem Statement}
\begin{theorem}
Let $G(V, E)$ be a connected, $d$-regular, undirected (multi)graph with $n$ vertices. Then for the adjacency matrix $A(G)$, we have $\lambda_1 = d$ and $\mathbf{e}_1 = \frac{1}{\sqrt{n}}(1, 1, \ldots, 1)$.
\end{theorem}

\subsubsection{GPT}
\begin{proof}
Let $A=A(G)\in\mathbb{R}^{n\times n}$ be the adjacency matrix of $G$. Since $G$ is undirected (possibly a multigraph), $A$ is symmetric with nonnegative entries and $A_{ij}=A_{ji}$ counts the number of edges between $i$ and $j$. The $d$-regularity means that every row sum of $A$ equals $d$, i.e., for each $i$ we have $\sum_{j=1}^n A_{ij}=d$. Hence, writing $\mathbf{1}=(1,\ldots,1)^\top$, we obtain $A\mathbf{1}=d\mathbf{1}$, so $d$ is an eigenvalue of $A$ with a strictly positive eigenvector $\mathbf{1}$.

Because $G$ is connected, $A$ is irreducible as a nonnegative matrix. By the Perron–Frobenius theorem, the spectral radius $\rho(A)$ is an eigenvalue of $A$, it is simple, and it has an eigenvector with strictly positive entries; moreover, any eigenvector with strictly positive entries is associated with $\rho(A)$. Since $A\mathbf{1}=d\mathbf{1}$ with $\mathbf{1}\!>\!0$, we conclude $\rho(A)=d$.

We now show that all eigenvalues $\lambda$ of $A$ satisfy $|\lambda|\le d$. Let $\|\cdot\|_1$ be the vector $1$-norm and the induced matrix norm. For any $x\in\mathbb{R}^n$, using nonnegativity and that each row of $A$ sums to $d$, we have
\[
\|Ax\|_1=\sum_{i=1}^n \left|\sum_{j=1}^n A_{ij}x_j\right|\le \sum_{i=1}^n \sum_{j=1}^n A_{ij}|x_j|=\sum_{j=1}^n |x_j|\Big(\sum_{i=1}^n A_{ij}\Big)=d\|x\|_1,
\]
whence $\|A\|_1\le d$. Since the spectral radius is bounded by any induced norm, $|\lambda|\le \rho(A)\le \|A\|_1\le d$ for every eigenvalue $\lambda$. As $d$ itself is an eigenvalue, it follows that $\rho(A)=d$ and hence $\max\{\lambda:\lambda\text{ eigenvalue of }A\}=d$.

Because $A$ is symmetric, all eigenvalues are real and we may order them as $\lambda_1\ge\lambda_2\ge\cdots\ge\lambda_n$. The preceding paragraph gives $\lambda_1=\rho(A)=d$. By the uniqueness (up to scaling) of the strictly positive Perron–Frobenius eigenvector for irreducible nonnegative matrices, any eigenvector corresponding to $\lambda_1$ is a positive multiple of $\mathbf{1}$. Normalizing yields $\mathbf{e}_1=\frac{1}{\sqrt{n}}(1,1,\ldots,1)$. This proves that $\lambda_1=d$ and $\mathbf{e}_1=\frac{1}{\sqrt{n}}(1,\ldots,1)$.
\end{proof}

\subsubsection{Grok}
\begin{proof}
The adjacency matrix $A = A(G)$ is real and symmetric, since $G$ is undirected. Thus, all eigenvalues of $A$ are real and can be ordered as $\lambda_1 \geq \lambda_2 \geq \cdots \geq \lambda_n$, where $\lambda_1$ is the spectral radius $\rho(A)$. Let $\mathbf{j} = (1,1,\dots,1)^T \in \mathbb{R}^n$ denote the all-ones vector. For each row index $i \in V$, the $i$-th entry of $A\mathbf{j}$ is $\sum_{k \in V} A_{ik} = d$, since $G$ is $d$-regular. Hence, $A\mathbf{j} = d\mathbf{j}$, so $d$ is an eigenvalue of $A$ with corresponding eigenvector $\mathbf{j}$. It follows that $\lambda_1 \geq d$.

On the other hand, the $\ell^\infty$-norm of $A$ satisfies $\|A\|_\infty = \max_{i \in V} \sum_{k \in V} |A_{ik}| = d$, since $A$ is entrywise nonnegative and $G$ is $d$-regular. For any matrix norm subordinate to the vector $\ell^\infty$-norm, $\rho(A) \leq \|A\|$. Thus, $\lambda_1 = \rho(A) \leq \|A\|_\infty = d$. Combining this with the previous inequality yields $\lambda_1 = d$.

Since $G$ is connected, the underlying simple graph of $G$ is connected, so $A$ is a nonnegative irreducible matrix. By the Perron--Frobenius theorem for nonnegative irreducible matrices, the eigenvalue $\lambda_1 = d$ is positive and simple (i.e., of multiplicity one), and it admits a unique (up to positive scalar multiple) positive eigenvector $\mathbf{u} > \mathbf{0}$. The vector $\mathbf{j}$ is positive and satisfies $A\mathbf{j} = d\mathbf{j}$, so $\mathbf{u}$ must be a positive scalar multiple of $\mathbf{j}$. The $\ell^2$-normalized Perron vector is then $\mathbf{e}_1 = \frac{1}{\|\mathbf{j}\|_2} \mathbf{j} = \frac{1}{\sqrt{n}} (1,1,\dots,1)^T$.
\end{proof}

\subsubsection{Gemini}
\begin{proof}
Let $A$ be the adjacency matrix of the connected, $d$-regular, undirected graph $G$ with $n$ vertices. Let $\mathbf{j} = (1, 1, \ldots, 1)^T \in \mathbb{R}^n$ denote the all-ones vector. We first demonstrate that $d$ is an eigenvalue of $A$ with eigenvector $\mathbf{j}$. Consider the $i$-th component of the vector $A\mathbf{j}$. By the definition of matrix multiplication,
\[
(A\mathbf{j})_i = \sum_{k=1}^n A_{ik} j_k = \sum_{k=1}^n A_{ik} \cdot 1 = \sum_{k=1}^n A_{ik}.
\]
Since $G$ is $d$-regular, the sum of the entries in any row $i$ of the adjacency matrix is equal to the degree of vertex $i$, which is $d$. Therefore, $\sum_{k=1}^n A_{ik} = d$ for all $i \in \{1, \ldots, n\}$. It follows that $(A\mathbf{j})_i = d$ for all $i$, which can be written in vector form as $A\mathbf{j} = d\mathbf{j}$. Thus, $d$ is an eigenvalue of $A$.

Next, we show that $d$ is the largest eigenvalue, denoted $\lambda_1$. Let $\lambda$ be an arbitrary eigenvalue of $A$ with a corresponding non-zero eigenvector $\mathbf{x} = (x_1, \ldots, x_n)^T$. Let $x_m$ be a component of $\mathbf{x}$ with the maximum absolute value, such that $|x_m| \ge |x_k|$ for all $k$. The eigenvalue equation $A\mathbf{x} = \lambda \mathbf{x}$ implies
\[
\lambda x_m = \sum_{k=1}^n A_{mk} x_k.
\]
Taking the absolute value of both sides and applying the triangle inequality yields
\[
|\lambda| |x_m| = \left| \sum_{k=1}^n A_{mk} x_k \right| \le \sum_{k=1}^n A_{mk} |x_k|.
\]
Since $|x_k| \le |x_m|$ and $A_{mk} \ge 0$, we have
\[
\sum_{k=1}^n A_{mk} |x_k| \le \sum_{k=1}^n A_{mk} |x_m| = |x_m| \sum_{k=1}^n A_{mk}.
\]
Using the $d$-regularity of $G$, $\sum_{k=1}^n A_{mk} = d$. Therefore, $|\lambda| |x_m| \le d |x_m|$. Since $\mathbf{x} \neq \mathbf{0}$, we must have $|x_m| > 0$, which implies $|\lambda| \le d$. Consequently, the spectral radius of $A$ is $d$, so $\lambda_1 = d$.

Finally, we establish the form of the normalized eigenvector $\mathbf{e}_1$. Since $G$ is connected, the adjacency matrix $A$ is irreducible. By the Perron-Frobenius theorem for non-negative irreducible matrices, the eigenvalue equal to the spectral radius (which is $d$) has multiplicity 1, and there exists a unique positive eigenvector corresponding to it (up to scaling). We have already shown that the positive vector $\mathbf{j}$ is an eigenvector for $\lambda_1 = d$. To find $\mathbf{e}_1$, we normalize $\mathbf{j}$ with respect to the Euclidean norm:
\[
\|\mathbf{j}\|_2 = \sqrt{\sum_{i=1}^n 1^2} = \sqrt{n}.
\]
Thus, the normalized eigenvector is
\[
\mathbf{e}_1 = \frac{1}{\|\mathbf{j}\|_2}\mathbf{j} = \frac{1}{\sqrt{n}}(1, 1, \ldots, 1).
\]
\end{proof}

\subsubsection{Claude}
\begin{proof}
We prove that $\lambda_1 = d$ is an eigenvalue of the adjacency matrix $A(G)$ with corresponding eigenvector $\mathbf{e}_1 = \frac{1}{\sqrt{n}}(1, 1, \ldots, 1)$, and that $\lambda_1$ is the largest eigenvalue.

First, we verify that $\mathbf{e}_1 = \frac{1}{\sqrt{n}}(1, 1, \ldots, 1)$ is an eigenvector with eigenvalue $d$. Let $A = A(G)$ denote the adjacency matrix of $G$. For any vertex $i \in V$, the $i$-th entry of $A\mathbf{e}_1$ is given by
\[
(A\mathbf{e}_1)_i = \sum_{j=1}^{n} A_{ij} \cdot \frac{1}{\sqrt{n}} = \frac{1}{\sqrt{n}} \sum_{j=1}^{n} A_{ij}.
\]
Since $G$ is $d$-regular, every vertex has degree exactly $d$. Therefore, $\sum_{j=1}^{n} A_{ij} = d$ for all $i$. This gives us
\[
(A\mathbf{e}_1)_i = \frac{d}{\sqrt{n}} = d \cdot \frac{1}{\sqrt{n}} = d \cdot (\mathbf{e}_1)_i.
\]
Hence $A\mathbf{e}_1 = d\mathbf{e}_1$, which establishes that $d$ is an eigenvalue with eigenvector $\mathbf{e}_1$. Note that $\mathbf{e}_1$ is normalized since $\|\mathbf{e}_1\|^2 = \sum_{i=1}^{n} \frac{1}{n} = 1$.

Next, we show that $\lambda_1 = d$ is the largest eigenvalue. Since $A$ is a real symmetric matrix, all its eigenvalues are real. Let $\lambda$ be any eigenvalue of $A$ with corresponding eigenvector $\mathbf{v} = (v_1, v_2, \ldots, v_n)^T$ where $\|\mathbf{v}\| = 1$. Then $A\mathbf{v} = \lambda\mathbf{v}$, which gives
\[
\lambda = \lambda \|\mathbf{v}\|^2 = \lambda \mathbf{v}^T \mathbf{v} = \mathbf{v}^T (A\mathbf{v}) = \mathbf{v}^T A \mathbf{v} = \sum_{i=1}^{n} \sum_{j=1}^{n} v_i A_{ij} v_j.
\]
Let $k = \arg\max_{i} |v_i|$, so that $|v_k| = \max_{i} |v_i|$. Since $\|\mathbf{v}\| = 1$, we have $|v_k| \leq 1$. From the $k$-th component of $A\mathbf{v} = \lambda\mathbf{v}$, we obtain
\[
\lambda v_k = (A\mathbf{v})_k = \sum_{j=1}^{n} A_{kj} v_j.
\]
Taking absolute values and using the triangle inequality,
\[
|\lambda| \cdot |v_k| = \left|\sum_{j=1}^{n} A_{kj} v_j\right| \leq \sum_{j=1}^{n} A_{kj} |v_j| \leq \sum_{j=1}^{n} A_{kj} |v_k| = |v_k| \sum_{j=1}^{n} A_{kj} = d|v_k|.
\]
If $v_k \neq 0$ (which must hold since $|v_k| = \max_i |v_i|$ and $\|\mathbf{v}\| = 1$), we can divide by $|v_k|$ to obtain $|\lambda| \leq d$. This shows that all eigenvalues satisfy $|\lambda| \leq d$.

Since we have shown that $d$ is an eigenvalue and that all eigenvalues satisfy $|\lambda| \leq d$, it follows that $d$ is the largest eigenvalue. By convention, we order eigenvalues as $\lambda_1 \geq \lambda_2 \geq \cdots \geq \lambda_n$, so $\lambda_1 = d$.
\end{proof}

\subsection{Problem 21}

\subsubsection{Problem Statement}
\begin{theorem}
Let $G(V, E)$ be a connected, $d$-regular, undirected (multi)graph. Then for the adjacency matrix $A(G)$, each eigenvalue $\lambda_i$ has absolute value bounded by $d$, i.e., $|\lambda_i| \leq d$ for all eigenvalues $\lambda_i$ of $A(G)$.
\end{theorem}

\subsubsection{GPT}
\begin{proof}
Let $G=(V,E)$ be a connected, $d$-regular, undirected (multi)graph with adjacency matrix $A$. Since $A$ is real and symmetric, all eigenvalues of $A$ are real. For any vector $x\in\mathbb{R}^{V}$, write
\[
x^{\top}Ax=\sum_{u,v\in V}A_{uv}x_u x_v=\sum_{\{u,v\}\in E}2\,x_u x_v,
\]
where each (multi)edge $\{u,v\}$ contributes $A_{uv}$ times and the graph is undirected so $A_{uv}=A_{vu}$. Using the elementary inequality $2|ab|\le a^{2}+b^{2}$, we obtain
\[
\bigl|x^{\top}Ax\bigr|=\left|\sum_{\{u,v\}\in E}2\,x_u x_v\right|\le \sum_{\{u,v\}\in E}\bigl(x_u^{2}+x_v^{2}\bigr)=\sum_{w\in V}\deg(w)\,x_w^{2}=d\sum_{w\in V}x_w^{2}=d\,\|x\|_{2}^{2}.
\]
Hence, for every nonzero $x$,
\[
\left|\frac{x^{\top}Ax}{x^{\top}x}\right|\le d.
\]
In particular, if $v\neq 0$ is an eigenvector with $Av=\lambda v$, then $\lambda=\dfrac{v^{\top}Av}{v^{\top}v}$, and therefore $|\lambda|\le d$. Since this holds for every eigenvalue $\lambda$ of $A$, we conclude that $|\lambda_i|\le d$ for all eigenvalues $\lambda_i$ of $A$. Moreover, because $G$ is $d$-regular, $A\mathbf{1}=d\,\mathbf{1}$, so $d$ itself is an eigenvalue, showing the bound is tight. This completes the proof.
\end{proof}

\subsubsection{Grok}
\begin{proof}
Since $G$ is undirected, the adjacency matrix $A = A(G)$ is real symmetric. Thus, $A$ is diagonalizable over $\mathbb{R}$ and all eigenvalues of $A$ are real.

For each $v \in V$, the row sum $\sum_{u \in V} A_{vu} = d$, since $G$ is $d$-regular. Moreover, $A_{vu} \geq 0$ for all $u,v \in V$. Therefore, the matrix $\ell^\infty$-norm satisfies
\[
\|A\|_\infty = \max_{v \in V} \sum_{u \in V} |A_{vu}| = d.
\]
It is a standard fact that the spectral radius $\rho(A) := \max \{ |\lambda| : \lambda \text{ eigenvalue of } A \}$ satisfies $\rho(A) \leq \|A\|$ for any induced matrix norm $\|\cdot\|$. In particular,
\[
\rho(A) \leq \|A\|_\infty = d.
\]
Let $\mathbf{j}$ denote the all-ones vector in $\mathbb{R}^V$. Then $A\mathbf{j} = d\mathbf{j}$, so $d$ is an eigenvalue of $A$. Hence, $\rho(A) = d$, and every eigenvalue $\lambda$ of $A$ satisfies $|\lambda| \leq d$.
\end{proof}

\subsubsection{Gemini}
\begin{proof}
Let $A$ be the adjacency matrix of the graph $G$. Let $\lambda$ be an arbitrary eigenvalue of $A$, and let $\mathbf{x} = (x_1, x_2, \dots, x_n)^T$ be a corresponding eigenvector. By definition, $\mathbf{x}$ is a non-zero vector satisfying the equation $A\mathbf{x} = \lambda \mathbf{x}$.

Let $k$ be an index corresponding to a component of $\mathbf{x}$ with the maximum absolute value. That is, choose $k \in \{1, \dots, n\}$ such that $|x_k| \geq |x_j|$ for all $j \in \{1, \dots, n\}$. Since $\mathbf{x} \neq \mathbf{0}$, it follows that $|x_k| > 0$.

Consider the $k$-th component of the eigenvalue equation $A\mathbf{x} = \lambda \mathbf{x}$. This can be written as:
\[
\lambda x_k = \sum_{j=1}^n A_{kj} x_j
\]
Taking the absolute value of both sides, we obtain:
\[
|\lambda| |x_k| = \left| \sum_{j=1}^n A_{kj} x_j \right|
\]
By applying the triangle inequality, we have:
\[
\left| \sum_{j=1}^n A_{kj} x_j \right| \leq \sum_{j=1}^n |A_{kj} x_j| = \sum_{j=1}^n A_{kj} |x_j|
\]
where the last equality holds because the entries of the adjacency matrix $A_{kj}$ are non-negative integers representing the number of edges between vertices $k$ and $j$.

Since $|x_k| \geq |x_j|$ for all $j$, we can bound the sum from above:
\[
\sum_{j=1}^n A_{kj} |x_j| \leq \sum_{j=1}^n A_{kj} |x_k| = |x_k| \sum_{j=1}^n A_{kj}
\]
Because $G$ is a $d$-regular graph, the degree of every vertex is $d$. The sum of the entries in the $k$-th row of the adjacency matrix corresponds to the degree of vertex $k$. Therefore:
\[
\sum_{j=1}^n A_{kj} = \deg(k) = d
\]
Substituting this back into our inequality chain yields:
\[
|\lambda| |x_k| \leq |x_k| d
\]
Since $|x_k| > 0$, we may divide both sides by $|x_k|$ to conclude that $|\lambda| \leq d$. Thus, the absolute value of every eigenvalue of $A(G)$ is bounded by $d$.
\end{proof}

\subsubsection{Claude}
\begin{proof}
Let $G(V,E)$ be a connected, $d$-regular, undirected multigraph with $n = |V|$ vertices, and let $A(G)$ denote its adjacency matrix. We prove that every eigenvalue $\lambda$ of $A(G)$ satisfies $|\lambda| \leq d$.

Let $\lambda$ be an arbitrary eigenvalue of $A(G)$ with corresponding eigenvector $v = (v_1, v_2, \ldots, v_n)^T \neq 0$. Then $Av = \lambda v$.

Let $k \in \{1, 2, \ldots, n\}$ be an index such that $|v_k| = \max_{i=1,\ldots,n} |v_i|$. Since $v \neq 0$, we have $|v_k| > 0$.

From the eigenvalue equation $Av = \lambda v$, the $k$-th component gives us
\[
(Av)_k = \sum_{j=1}^{n} A_{kj} v_j = \lambda v_k.
\]

Since $G$ is $d$-regular, vertex $k$ has degree $d$, which means $\sum_{j=1}^{n} A_{kj} = d$. Note that $A_{kj}$ represents the number of edges between vertices $k$ and $j$ (allowing for multiple edges).

Taking absolute values of both sides of $(Av)_k = \lambda v_k$, we obtain
\[
|\lambda| \cdot |v_k| = |\lambda v_k| = \left|\sum_{j=1}^{n} A_{kj} v_j\right|.
\]

By the triangle inequality, we have
\[
\left|\sum_{j=1}^{n} A_{kj} v_j\right| \leq \sum_{j=1}^{n} A_{kj} |v_j|.
\]

Since $|v_j| \leq |v_k|$ for all $j$ by our choice of $k$, we get
\[
\sum_{j=1}^{n} A_{kj} |v_j| \leq \sum_{j=1}^{n} A_{kj} |v_k| = |v_k| \sum_{j=1}^{n} A_{kj} = |v_k| \cdot d.
\]

Combining these inequalities, we obtain
\[
|\lambda| \cdot |v_k| \leq |v_k| \cdot d.
\]

Since $|v_k| > 0$, we can divide both sides by $|v_k|$ to get
\[
|\lambda| \leq d.
\]

Since $\lambda$ was an arbitrary eigenvalue of $A(G)$, we conclude that every eigenvalue $\lambda_i$ of $A(G)$ satisfies $|\lambda_i| \leq d$.
\end{proof}

\subsection{Problem 22}

\subsubsection{Problem Statement}
\begin{theorem}
Let $G$ be a connected graph with maximum eigenvalue $\lambda_1$. Then $G$ is bipartite if and only if $-\lambda_1$ is also an eigenvalue of $G$.
\end{theorem}

\subsubsection{GPT}
\begin{proof}
Let $A$ be the adjacency matrix of $G$ and let $\lambda_1=\rho(A)$ denote its spectral radius. Since $G$ is connected, $A$ is nonnegative and irreducible. 

Assume first that $G$ is bipartite with vertex partition $V(G)=X\cup Y$. After ordering the vertices so that those in $X$ come first, one has
\[
A=\begin{pmatrix} 0 & B \\ B^{\!\top} & 0 \end{pmatrix}.
\]
If $\begin{pmatrix} u \\ v \end{pmatrix}$ is an eigenvector of $A$ with eigenvalue $\lambda$, then
\[
\begin{pmatrix} 0 & B \\ B^{\!\top} & 0 \end{pmatrix}\begin{pmatrix} u \\ -v \end{pmatrix}
=\begin{pmatrix} -Bv \\ Bu \end{pmatrix}
=-\lambda\begin{pmatrix} u \\ -v \end{pmatrix}.
\]
Hence the spectrum of $A$ is symmetric about the origin: $\lambda$ is an eigenvalue if and only if $-\lambda$ is. In particular, since $\lambda_1=\rho(A)$ is an eigenvalue, so is $-\lambda_1$.

Conversely, suppose that $-\lambda_1$ is an eigenvalue of $A$. By the Perron–Frobenius theory for nonnegative irreducible matrices, there exists a positive eigenvector $x>0$ with $Ax=\lambda_1 x$, and the peripheral spectrum of $A$ (those eigenvalues $\mu$ with $|\mu|=\rho(A)$) has the following structure: there is an integer $h\ge 1$, called the index of imprimitivity of $A$, such that
\[
\{\mu\in\sigma(A):|\mu|=\rho(A)\}=\{\lambda_1 \omega^k: k=0,1,\dots,h-1\},
\]
where $\omega=e^{2\pi i/h}$; moreover, $h$ equals the greatest common divisor of the lengths of all closed walks in the directed graph of $A$ (equivalently, of all cycles in $G$). These statements are standard consequences of the Frobenius normal form for irreducible nonnegative matrices and the cyclicity of the peripheral spectrum.

Since $-\lambda_1$ belongs to the peripheral spectrum, there exists $k$ with $0\le k\le h-1$ such that $\lambda_1 \omega^k=-\lambda_1$, hence $\omega^k=-1$. This forces $\omega=e^{2\pi i/h}$ to have even order and yields $h=2$. Therefore the index of imprimitivity of $A$ is $2$, which means that the gcd of the lengths of all closed walks in $G$ equals $2$. In particular, $G$ contains no closed walk of odd length; equivalently, $G$ has no odd cycle. A connected graph with no odd cycle is bipartite. Hence $G$ is bipartite.

Combining both directions, $G$ is bipartite if and only if $-\lambda_1$ is an eigenvalue of $G$.
\end{proof}

\subsubsection{Grok}
\begin{proof}
We prove the two implications separately.

($\Rightarrow$) Suppose $G$ is bipartite. Since $G$ is connected, there exists a bipartition $V = X \sqcup Y$ of the vertex set such that every edge has one endpoint in $X$ and the other in $Y$. Let $A$ be the adjacency matrix of $G$, and let $\lambda$ be the largest eigenvalue of $A$, which is positive by the Perron--Frobenius theorem. Let $u$ be a corresponding eigenvector with $u > 0$ (entrywise) and $\|u\|_2 = 1$. Define a vector $w$ by
\[
w_i =
\begin{cases}
u_i & \text{if } i \in X, \\
-u_i & \text{if } i \in Y.
\end{cases}
\]
We claim that $Aw = -\lambda w$. To verify this, fix a vertex $i \in X$. Then all neighbors $j$ of $i$ lie in $Y$, so
\[
(Aw)_i = \sum_{j \sim i} w_j = \sum_{j \sim i} (-u_j) = -\sum_{j \sim i} u_j = - (Au)_i = -\lambda u_i = (-\lambda) w_i,
\]
where the third equality holds since $u$ is an eigenvector. Now fix $i \in Y$. All neighbors $j$ of $i$ lie in $X$, so
\[
(Aw)_i = \sum_{j \sim i} w_j = \sum_{j \sim i} u_j = (Au)_i = \lambda u_i = \lambda (-(-u_i)) = (-\lambda) (-u_i) = (-\lambda) w_i,
\]
again using that $u$ is an eigenvector. Thus, $Aw = -\lambda w$, so $-\lambda$ is an eigenvalue of $A$.

($\Leftarrow$) Conversely, suppose $-\lambda$ is an eigenvalue of $A$. Let $v$ be a corresponding eigenvector with $\|v\|_2 = 1$. Since $\lambda > 0$ (as $G$ is connected and has at least one edge) and $\lambda \neq -\lambda$, the eigenspaces for $\lambda$ and $-\lambda$ are orthogonal. Let $u > 0$ be the normalized Perron--Frobenius eigenvector for $\lambda$, so $\langle u, v \rangle = 0$. In particular, $v$ is not nonnegative (or nonpositive), so $v$ takes both positive and negative values. Define $w$ entrywise by $w_i = |v_i|$ for each vertex $i$. Then $w \geq 0$, $w \not\equiv 0$, and $\|w\|_2 = 1$. For each vertex $i$,
\[
(Aw)_i = \sum_{j \sim i} |v_j| \geq \left| \sum_{j \sim i} v_j \right| = | (Av)_i | = | (-\lambda) v_i | = \lambda |v_i| = \lambda w_i,
\]
with equality if and only if $v_j$ has the same sign for all $j \sim i$ (i.e., all nonzero $v_j$ for $j \sim i$ have the same sign). Thus, $Aw \geq \lambda w$ entrywise. The Rayleigh quotient then yields
\[
\lambda = \frac{u^T A u}{u^T u} \geq \frac{w^T A w}{w^T w} = w^T A w = \sum_i w_i (Aw)_i \geq \sum_i w_i (\lambda w_i) = \lambda w^T w = \lambda.
\]
Hence, $w^T A w = \lambda$ and $Aw \geq \lambda w$ entrywise imply $w^T (Aw - \lambda w) = 0$. Since $w \geq 0$ and $Aw - \lambda w \geq 0$ entrywise, it follows that $(Aw - \lambda w)_i = 0$ for all $i$ with $w_i > 0$. For indices $i$ with $w_i = 0$, we also have $(Aw)_i \geq 0 = \lambda w_i$, but the equality $w^T A w = \lambda w^T w$ forces $(Aw)_i = 0 = \lambda w_i$ at such indices as well (otherwise the sum would strictly exceed $\lambda$). Thus, $Aw = \lambda w$ entrywise. Since $w \geq 0$ and $w \not\equiv 0$, the Perron--Frobenius theorem implies $w = u > 0$ (as the Perron eigenspace is one-dimensional). In particular, $w_i > 0$ for all $i$, so $v_i \neq 0$ for all $i$.

Now, since $Aw = \lambda w$, equality holds in the triangle inequality at every $i$, so for each $i$, all neighbors $j$ of $i$ satisfy $\operatorname{sgn}(v_j) = \sigma$ for some fixed $\sigma$ (independent of $j$). Moreover, since $\sum_{j \sim i} v_j = -\lambda v_i$, we have $\operatorname{sgn}\left( \sum_{j \sim i} v_j \right) = -\operatorname{sgn}(v_i)$. But $\sum_{j \sim i} v_j = \sigma \sum_{j \sim i} |v_j|$, so $\operatorname{sgn}\left( \sum_{j \sim i} v_j \right) = \sigma$. Thus, $\sigma = -\operatorname{sgn}(v_i)$, meaning every neighbor $j$ of $i$ satisfies $\operatorname{sgn}(v_j) = -\operatorname{sgn}(v_i)$. Therefore, there are no edges between vertices of the same sign under $v$, so $G$ is bipartite with respect to the partition $\{ i : v_i > 0 \} \sqcup \{ i : v_i < 0 \}$. Since $G$ is connected, this bipartition is unique up to swapping the parts.
\end{proof}

\subsubsection{Gemini}
\begin{proof}
Let $G = (V, E)$ be a connected graph with adjacency matrix $A$. Let the eigenvalues of $A$ be ordered $\lambda_1 \ge \lambda_2 \ge \dots \ge \lambda_n$. Since $G$ is connected, $A$ is irreducible. By the Perron-Frobenius theorem, the spectral radius $\rho(A) = \lambda_1$ is a simple eigenvalue corresponding to a strictly positive eigenvector $x$, so $x_v > 0$ for all $v \in V$.

First, we prove the forward implication. Assume $G$ is bipartite. Then there exists a partition of the vertex set $V = V_1 \cup V_2$ such that $V_1 \cap V_2 = \emptyset$ and every edge in $E$ connects a vertex in $V_1$ to a vertex in $V_2$. Let $x$ be the eigenvector corresponding to $\lambda_1$, satisfying $Ax = \lambda_1 x$. Define a vector $y$ with components $y_v = x_v$ if $v \in V_1$ and $y_v = -x_v$ if $v \in V_2$. Consider the $i$-th component of $Ay$. If $i \in V_1$, then all neighbors $j$ of $i$ belong to $V_2$. Thus,
\[
(Ay)_i = \sum_{j \sim i} y_j = \sum_{j \sim i} (-x_j) = - \sum_{j \sim i} x_j = -(Ax)_i = -\lambda_1 x_i = -\lambda_1 y_i.
\]
Similarly, if $i \in V_2$, all neighbors $j$ belong to $V_1$. Thus,
\[
(Ay)_i = \sum_{j \sim i} y_j = \sum_{j \sim i} x_j = (Ax)_i = \lambda_1 x_i = \lambda_1 (-y_i) = -\lambda_1 y_i.
\]
In both cases, $(Ay)_i = -\lambda_1 y_i$, so $Ay = -\lambda_1 y$. Since $x$ is non-zero, $y$ is non-zero, implying $-\lambda_1$ is an eigenvalue of $A$.

Conversely, assume $-\lambda_1$ is an eigenvalue of $A$. Let $y \neq 0$ be an eigenvector such that $Ay = -\lambda_1 y$. Taking the absolute value of the $i$-th component, we have
\[
|\lambda_1 y_i| = |-\lambda_1 y_i| = \left| \sum_{j=1}^n A_{ij} y_j \right| \le \sum_{j=1}^n A_{ij} |y_j| = (A|y|)_i.
\]
This inequality can be written in vector form as $\lambda_1 |y| \le A|y|$, where $|y|$ denotes the vector of absolute values $(|y_1|, \dots, |y_n|)^T$. Let $x$ be the strictly positive Perron eigenvector satisfying $x^T A = \lambda_1 x^T$. Multiplying the inequality from the left by $x^T$, we obtain
\[
\lambda_1 x^T |y| \le x^T A |y| = \lambda_1 x^T |y|.
\]
Since the left and right sides are identical, the inequality must be an equality. Thus, $x^T (A|y| - \lambda_1 |y|) = 0$. Since $x$ is strictly positive and $A|y| - \lambda_1 |y| \ge 0$, it follows that $A|y| = \lambda_1 |y|$. Because $G$ is connected and $\lambda_1$ is simple, $|y|$ must be a scalar multiple of $x$. Since $x > 0$ and $y \neq 0$, we have $|y_i| > 0$ for all $i$.

The equality in the triangle inequality $\left| \sum_{j} A_{ij} y_j \right| = \sum_{j} A_{ij} |y_j|$ holds if and only if all non-zero terms $A_{ij} y_j$ have the same argument (sign). Since $Ay = -\lambda_1 y$, for any vertex $i$, we have $\sum_{j \sim i} y_j = -\lambda_1 y_i$. This implies that for all neighbors $j$ of $i$, $y_j$ must have the sign of $-\lambda_1 y_i$. Since $\lambda_1 > 0$, $y_j$ and $y_i$ must have opposite signs for all adjacent $i, j$. We partition $V$ into $V_1 = \{i \in V : y_i > 0\}$ and $V_2 = \{i \in V : y_i < 0\}$. Since $|y_i| > 0$, every vertex belongs to exactly one of these sets. If $i \in V_1$, then $y_i > 0$, which implies $y_j < 0$ for all neighbors $j$, so all neighbors are in $V_2$. Similarly, if $i \in V_2$, all neighbors are in $V_1$. Thus, $G$ is bipartite.
\end{proof}

\subsubsection{Claude}
\begin{proof}
We prove both directions of the equivalence.

($\Rightarrow$) Assume $G$ is bipartite. Let $A$ be the adjacency matrix of $G$ with respect to some ordering of the vertices. Since $G$ is bipartite, we can partition the vertex set $V(G)$ into two disjoint independent sets $V_1$ and $V_2$ such that every edge connects a vertex in $V_1$ to a vertex in $V_2$. By reordering the vertices appropriately, we can write the adjacency matrix in the block form
\[
A = \begin{pmatrix} O & B \\ B^T & O \end{pmatrix}
\]
where $O$ denotes zero blocks and $B$ is a matrix with entries in $\{0,1\}$ representing edges between $V_1$ and $V_2$.

Let $\mathbf{v} = (v_1, v_2, \ldots, v_n)^T$ be an eigenvector corresponding to the maximum eigenvalue $\lambda_1$, so $A\mathbf{v} = \lambda_1 \mathbf{v}$. We write $\mathbf{v}$ in block form as $\mathbf{v} = (\mathbf{v}_1^T, \mathbf{v}_2^T)^T$ where $\mathbf{v}_1$ corresponds to vertices in $V_1$ and $\mathbf{v}_2$ corresponds to vertices in $V_2$. Define $\mathbf{w} = (\mathbf{v}_1^T, -\mathbf{v}_2^T)^T$.

We compute
\[
A\mathbf{w} = \begin{pmatrix} O & B \\ B^T & O \end{pmatrix} \begin{pmatrix} \mathbf{v}_1 \\ -\mathbf{v}_2 \end{pmatrix} = \begin{pmatrix} -B\mathbf{v}_2 \\ B^T\mathbf{v}_1 \end{pmatrix}.
\]

Since $A\mathbf{v} = \lambda_1 \mathbf{v}$, we have
\[
\begin{pmatrix} B\mathbf{v}_2 \\ B^T\mathbf{v}_1 \end{pmatrix} = \begin{pmatrix} \lambda_1 \mathbf{v}_1 \\ \lambda_1 \mathbf{v}_2 \end{pmatrix}.
\]

Therefore,
\[
A\mathbf{w} = \begin{pmatrix} -B\mathbf{v}_2 \\ B^T\mathbf{v}_1 \end{pmatrix} = \begin{pmatrix} -\lambda_1 \mathbf{v}_1 \\ \lambda_1 \mathbf{v}_2 \end{pmatrix} = -\lambda_1 \begin{pmatrix} \mathbf{v}_1 \\ -\mathbf{v}_2 \end{pmatrix} = -\lambda_1 \mathbf{w}.
\]

Since $\mathbf{v} \neq \mathbf{0}$, we have $\mathbf{w} \neq \mathbf{0}$, so $-\lambda_1$ is an eigenvalue of $A$.

($\Leftarrow$) Assume $-\lambda_1$ is an eigenvalue of $G$. Let $\mathbf{u}$ be an eigenvector corresponding to $\lambda_1$ and $\mathbf{v}$ be an eigenvector corresponding to $-\lambda_1$. Define a function $f: V(G) \to \mathbb{R}$ by $f(i) = u_i v_i$ for each vertex $i$.

For any edge $(i,j) \in E(G)$, we have $(A\mathbf{u})_i = \lambda_1 u_i$ and $(A\mathbf{v})_i = -\lambda_1 v_i$. This gives us
\[
\sum_{j \sim i} u_j = \lambda_1 u_i \quad \text{and} \quad \sum_{j \sim i} v_j = -\lambda_1 v_i.
\]

Multiplying the first equation by $v_i$ and the second by $u_i$ and adding, we obtain
\[
v_i \sum_{j \sim i} u_j + u_i \sum_{j \sim i} v_j = \lambda_1 u_i v_i - \lambda_1 u_i v_i = 0.
\]

This simplifies to
\[
\sum_{j \sim i} (u_j v_i + u_i v_j) = 0.
\]

Since $G$ is connected and $\lambda_1 > 0$ is simple, we can choose $\mathbf{u}$ to have all positive entries (by the Perron-Frobenius theorem). For each edge $(i,j)$, we have $u_j v_i + u_i v_j = 0$, which implies $\frac{v_i}{u_i} = -\frac{v_j}{u_j}$. Since $G$ is connected, the ratio $\frac{v_i}{u_i}$ alternates in sign along any path. Define $V_1 = \{i : \frac{v_i}{u_i} > 0\}$ and $V_2 = \{i : \frac{v_i}{u_i} < 0\}$. By the alternating sign property, every edge connects a vertex in $V_1$ to a vertex in $V_2$, so $G$ is bipartite.
\end{proof}

\subsection{Problem 23}

\subsubsection{Problem Statement}
\begin{theorem}
A graph $G$ is bipartite if and only if for every eigenvalue $\lambda$ of $G$, there is an eigenvalue $-\lambda$ of the same multiplicity.
\end{theorem}

\subsubsection{GPT}
\begin{proof}
We will prove the theorem by proving both directions of the equivalence.

\textbf{(1) If $G$ is bipartite, then for every eigenvalue $\lambda$ of $G$, there is an eigenvalue $-\lambda$ of the same multiplicity.}

Let $G = (V, E)$ be a bipartite graph, with bipartition $V = V_1 \cup V_2$ such that every edge in $G$ connects a vertex in $V_1$ to a vertex in $V_2$. We will show that for every eigenvalue $\lambda$ of the adjacency matrix $A$ of $G$, there exists an eigenvalue $-\lambda$ of the same multiplicity.

Since $G$ is bipartite, the adjacency matrix $A$ satisfies the property that $A$ has the block structure
\[
A = \begin{pmatrix} 
0 & B \\
B^T & 0
\end{pmatrix}
\]
where $B$ is the adjacency matrix of the subgraph induced by $V_1$ and $V_2$. The matrix $A$ is symmetric, and therefore its eigenvalues are real.

Let $\lambda$ be an eigenvalue of $A$ with eigenvector $v = (v_1, v_2)^T$ where $v_1 \in \mathbb{R}^{|V_1|}$ and $v_2 \in \mathbb{R}^{|V_2|}$. Then, we have
\[
A \begin{pmatrix} v_1 \\ v_2 \end{pmatrix} = \lambda \begin{pmatrix} v_1 \\ v_2 \end{pmatrix},
\]
which expands to the system of equations:
\[
\begin{pmatrix} 
0 & B \\
B^T & 0
\end{pmatrix} \begin{pmatrix} v_1 \\ v_2 \end{pmatrix} = \lambda \begin{pmatrix} v_1 \\ v_2 \end{pmatrix},
\]
yielding the system:
\[
\begin{aligned}
B v_2 &= \lambda v_1, \\
B^T v_1 &= \lambda v_2.
\end{aligned}
\]
Multiplying the first equation by $B^T$ and the second equation by $B$, we get
\[
\begin{aligned}
B^T B v_1 &= \lambda B^T v_2, \\
B B^T v_2 &= \lambda B v_1.
\end{aligned}
\]
Since $B^T B$ and $B B^T$ are both symmetric positive semi-definite matrices, we conclude that the eigenvalues of $A$ come in pairs of opposite signs, i.e., for each eigenvalue $\lambda$ of $A$, there exists an eigenvalue $-\lambda$ of the same multiplicity.

\textbf{(2) If for every eigenvalue $\lambda$ of $G$, there is an eigenvalue $-\lambda$ of the same multiplicity, then $G$ is bipartite.}

Suppose that for every eigenvalue $\lambda$ of the adjacency matrix $A$ of a graph $G$, there is an eigenvalue $-\lambda$ of the same multiplicity. We will show that $G$ is bipartite.

Let $A$ be the adjacency matrix of $G$. Since the eigenvalues of $A$ come in pairs of opposite signs, we know that $A$ is a symmetric matrix with purely real eigenvalues. Furthermore, the spectrum of $A$ is symmetric about zero, implying that there exists a partition of the vertex set of $G$ into two disjoint subsets such that no two vertices within the same subset are adjacent. This partition corresponds to the two parts of a bipartite graph.

To see this, consider the eigendecomposition of $A$. If $\lambda$ is an eigenvalue of $A$, the corresponding eigenvector $v$ satisfies $A v = \lambda v$. If $\lambda$ has opposite eigenvalue $-\lambda$, the eigenvector corresponding to $-\lambda$ is also an eigenvector of $A$, and the corresponding vertices in the graph are connected by edges. The symmetry of the eigenvalues implies that the graph can be divided into two disjoint subsets, where edges only exist between vertices in different subsets, establishing that $G$ is bipartite.

Thus, we have shown that if the eigenvalues of $A$ come in pairs of opposite signs, then $G$ is bipartite.

This completes the proof.
\end{proof}

\subsubsection{Grok}
\begin{proof}
Let $A$ denote the adjacency matrix of $G$. The eigenvalues in question are those of $A$.

($\Rightarrow$) Suppose $G$ is bipartite with bipartition $V(G)=U\sqcup W$. Without loss of generality, assume $|U|\leq|W|$. Order the vertices so that the rows and columns indexed by $U$ precede those indexed by $W$. Then
\[
A=\begin{pmatrix}O&B\\B^\top&O\end{pmatrix},
\]
where $O$ is the zero matrix of appropriate dimensions and $B$ is the $|U|\times|W|$ \emph{bipartite adjacency matrix} with $B_{ij}=1$ if $\{u_i,w_j\}\in E(G)$ and $B_{ij}=0$ otherwise.

Let $\sigma_1,\dots,\sigma_r\geq 0$ be the positive singular values of $B$, where $r=\operatorname{rank}(B)$, and let $\sigma_{r+1}=\dots=\sigma_{|U|}=0$. The eigenvalues of $BB^\top$ (and hence of $B^\top B$) are $\sigma_1^2,\dots,\sigma_{|U|}^2$. The nonzero eigenvalues of $A$ are thus $\{\pm\sigma_1,\dots,\pm\sigma_r\}$, each with multiplicity one, while $0$ is an eigenvalue of $A$ with multiplicity $|V(G)|-2r$. It follows that if $\lambda$ is an eigenvalue of $A$, then so is $-\lambda$, and these have equal multiplicity.

($\Leftarrow$) Suppose that whenever $\lambda$ is an eigenvalue of $A$, so is $-\lambda$ with the same multiplicity. For any $n\geq 1$, let $p_n=\operatorname{tr}(A^n)=\sum_i\lambda_i^n$, where $\lambda_1,\dots,\lambda_{|V(G)|}$ are the eigenvalues of $A$ (counted with multiplicity). If $n=2k+1$ is odd, then $p_n=\sum_i\lambda_i^{2k+1}=0$, since the eigenvalues pair as $\{\lambda,-\lambda\}$ (with $\lambda\neq 0$) contributing $\lambda^{2k+1}+(-\lambda)^{2k+1}=0$ and $0^{2k+1}=0$. Thus, $\operatorname{tr}(A^{2k+1})=0$ for all $k\geq 0$.

On the other hand, $(A^n)_{v,v}$ counts the number of closed walks of length $n$ in $G$ starting and ending at $v$. Hence, $p_n$ equals the total number of closed walks of length $n$ in $G$. In particular, there are no closed walks of odd length in $G$. Suppose toward a contradiction that $G$ is not bipartite. Then $G$ contains an odd cycle $C$ of length $2k+1\geq 3$. Traversing $C$ in either direction from any vertex $v\in V(C)$ yields a closed walk of length $2k+1$ starting at $v$. Thus, $(A^{2k+1})_{v,v}\geq 2>0$ for each $v\in V(C)$, so $p_{2k+1}\geq 2|V(C)|>0$, a contradiction. Therefore, $G$ is bipartite.
\end{proof}

\subsubsection{Gemini}
\begin{proof}
Let $G$ be a graph with $n$ vertices and adjacency matrix $A$. Let the spectrum of $G$ be the multiset of eigenvalues $\{\lambda_1, \lambda_2, \dots, \lambda_n\}$.

$(\Rightarrow)$ Suppose $G$ is bipartite. We show that the spectrum is symmetric with respect to the origin. Since $G$ is bipartite, the vertex set $V(G)$ can be partitioned into two disjoint sets $V_1$ and $V_2$ such that every edge connects a vertex in $V_1$ to a vertex in $V_2$. By ordering the vertices in $V_1$ first, the adjacency matrix $A$ takes the block form:
\[
A = \begin{pmatrix}
0 & B \\
B^T & 0
\end{pmatrix},
\]
where $B$ is a $|V_1| \times |V_2|$ matrix. Let $\lambda$ be an eigenvalue of $A$ with eigenvector $v$. We partition $v$ conformally with $A$ as $v = \begin{pmatrix} x \\ y \end{pmatrix}$, where $x \in \mathbb{R}^{|V_1|}$ and $y \in \mathbb{R}^{|V_2|}$. The equation $Av = \lambda v$ yields:
\[
\begin{pmatrix}
0 & B \\
B^T & 0
\end{pmatrix}
\begin{pmatrix}
x \\
y
\end{pmatrix}
=
\begin{pmatrix}
By \\
B^T x
\end{pmatrix}
=
\lambda
\begin{pmatrix}
x \\
y
\end{pmatrix}.
\]
This implies the system of equations $By = \lambda x$ and $B^T x = \lambda y$. Now consider the vector $u = \begin{pmatrix} x \\ -y \end{pmatrix}$. Applying $A$ to $u$, we obtain:
\[
A u = \begin{pmatrix}
0 & B \\
B^T & 0
\end{pmatrix}
\begin{pmatrix}
x \\
-y
\end{pmatrix}
=
\begin{pmatrix}
-By \\
B^T x
\end{pmatrix}.
\]
Substituting the relations derived from the eigenvector $v$, we have $-By = -(\lambda x) = -\lambda x$ and $B^T x = \lambda y = -\lambda (-y)$. Thus:
\[
A u = \begin{pmatrix}
-\lambda x \\
-\lambda (-y)
\end{pmatrix}
= -\lambda \begin{pmatrix}
x \\
-y
\end{pmatrix}
= -\lambda u.
\]
Therefore, if $\lambda$ is an eigenvalue, $-\lambda$ is also an eigenvalue. The linear mapping $\begin{pmatrix} x \\ y \end{pmatrix} \mapsto \begin{pmatrix} x \\ -y \end{pmatrix}$ is an involution that maps the eigenspace $E_\lambda$ isomorphically onto the eigenspace $E_{-\lambda}$. Consequently, $\dim(E_\lambda) = \dim(E_{-\lambda})$. Since $A$ is a real symmetric matrix, the geometric multiplicity equals the algebraic multiplicity. Thus, $\lambda$ and $-\lambda$ have the same multiplicity.

$(\Leftarrow)$ Conversely, suppose that for every eigenvalue $\lambda$ of $G$, there is an eigenvalue $-\lambda$ of the same multiplicity. We show that $G$ is bipartite.
Recall that the trace of the $k$-th power of the adjacency matrix is given by the sum of the $k$-th powers of the eigenvalues:
\[
\text{Tr}(A^k) = \sum_{i=1}^n \lambda_i^k.
\]
Let $k$ be an odd integer. By the hypothesis, the non-zero eigenvalues occur in pairs $(\lambda, -\lambda)$. For any such pair, the sum of their $k$-th powers is $\lambda^k + (-\lambda)^k = \lambda^k - \lambda^k = 0$. The zero eigenvalues contribute $0$. Therefore, for every odd integer $k$,
\[
\text{Tr}(A^k) = 0.
\]
Combinatorially, $\text{Tr}(A^k)$ represents the total number of closed walks of length $k$ in $G$. Specifically, $\text{Tr}(A^k) = \sum_{v \in V(G)} (A^k)_{vv}$, where $(A^k)_{vv}$ is the number of closed walks of length $k$ starting and ending at vertex $v$. Since $(A^k)_{vv} \ge 0$, the condition $\text{Tr}(A^k) = 0$ implies that $(A^k)_{vv} = 0$ for all $v \in V(G)$ and all odd $k$.
Consequently, $G$ contains no closed walks of odd length. A graph contains a closed walk of odd length if and only if it contains an odd cycle. Thus, $G$ contains no odd cycles. A graph is bipartite if and only if it contains no odd cycles. Therefore, $G$ is bipartite.
\end{proof}

\subsubsection{Claude}
\begin{proof}
We prove both directions of the biconditeness.

($\Rightarrow$) Suppose $G$ is bipartite with bipartition $V(G) = X \cup Y$ where $X$ and $Y$ are independent sets. Let $A$ be the adjacency matrix of $G$. We can order the vertices so that all vertices in $X$ come before all vertices in $Y$. Then $A$ has the block form
\[
A = \begin{pmatrix} 0 & B \\ B^T & 0 \end{pmatrix}
\]
where $B$ is an $|X| \times |Y|$ matrix, since there are no edges within $X$ or within $Y$.

Let $\lambda$ be an eigenvalue of $A$ with eigenvector $v = \begin{pmatrix} x \\ y \end{pmatrix}$ where $x \in \mathbb{R}^{|X|}$ and $y \in \mathbb{R}^{|Y|}$. Then
\[
\begin{pmatrix} 0 & B \\ B^T & 0 \end{pmatrix} \begin{pmatrix} x \\ y \end{pmatrix} = \lambda \begin{pmatrix} x \\ y \end{pmatrix}
\]
which gives us $By = \lambda x$ and $B^T x = \lambda y$.

Define $w = \begin{pmatrix} x \\ -y \end{pmatrix}$. Then
\[
Aw = \begin{pmatrix} 0 & B \\ B^T & 0 \end{pmatrix} \begin{pmatrix} x \\ -y \end{pmatrix} = \begin{pmatrix} -By \\ B^T x \end{pmatrix} = \begin{pmatrix} -\lambda x \\ \lambda y \end{pmatrix} = -\lambda \begin{pmatrix} x \\ -y \end{pmatrix} = -\lambda w
\]

Therefore, $w$ is an eigenvector for eigenvalue $-\lambda$. Moreover, the map $v \mapsto w$ defined by $\begin{pmatrix} x \\ y \end{pmatrix} \mapsto \begin{pmatrix} x \\ -y \end{pmatrix}$ is a linear isomorphism from the eigenspace of $\lambda$ to the eigenspace of $-\lambda$. Thus $\lambda$ and $-\lambda$ have the same multiplicity.

($\Leftarrow$) Suppose that for every eigenvalue $\lambda$ of $G$, there is an eigenvalue $-\lambda$ with the same multiplicity. We prove that $G$ is bipartite by showing that $G$ contains no odd cycles.

Suppose for contradiction that $G$ contains an odd cycle $C$ of length $2k+1$ for some $k \geq 1$. The adjacency matrix of $C$ has eigenvalues $2\cos\left(\frac{2\pi j}{2k+1}\right)$ for $j = 0, 1, \ldots, 2k$.

Consider the eigenvalue corresponding to $j = 0$, which is $\lambda_0 = 2\cos(0) = 2$. For this eigenvalue to satisfy our hypothesis, we need an eigenvalue equal to $-2 = 2\cos(\pi)$. This would require $\frac{2\pi j}{2k+1} = \pi$ for some integer $j \in \{0, 1, \ldots, 2k\}$, which gives $j = \frac{2k+1}{2}$. But $\frac{2k+1}{2}$ is not an integer since $2k+1$ is odd. Therefore, the eigenvalue $-2$ does not appear among the eigenvalues of $C$.

Since $C$ is a subgraph of $G$, and the eigenvalues of $G$ do not satisfy the symmetry property required by graphs containing odd cycles, we conclude that $G$ cannot contain any odd cycles. By the fundamental characterization of bipartite graphs, a graph is bipartite if and only if it contains no odd cycles. Therefore, $G$ is bipartite.
\end{proof}

\subsection{Problem 24}

\subsubsection{Problem Statement}
\begin{definition}[Total Variation Distance]
Let $S$ denote the set of states of the Markov chain under consideration. Let $\mathbf{q}^{(t)}$ denote the state probability vector at time $t$ and let $\boldsymbol{\pi}$ denote the stationary distribution. The \textbf{total variation distance} is defined as
\[
\bar{\Delta}(t) = \max_{T \subseteq S} \left| \sum_{i \in T} q_i^{(t)} - \sum_{i \in T} \pi_i \right|.
\]
\end{definition}

\begin{definition}[$L_1$ Distance]
The $L_1$ distance is defined as
\[
\|\mathbf{q}^{(t)} - \boldsymbol{\pi}\|_1 = \sum_i |q_i^{(t)} - \pi_i|.
\]
\end{definition}

\begin{problem}
(a) Determine the relation between the $L_1$ distance and the total variation distance.

(b) Suppose that the relative pointwise distance is bounded by $\epsilon$ at time $t$. Give the tightest bound you can on the total variation distance at time $t$.

(c) Suppose that the total variation distance at time $t$ is bounded by $\epsilon$. What can you say about the relative pointwise distance at time $t$?
\end{problem}

\begin{definition}[Relative Pointwise Distance]
Let $\mathbf{q}^{(t)}$ denote the state probability vector of a Markov chain defined by $Q$ at time $t \geq 0$, given any initial distribution $\mathbf{q}^{(0)}$. Let $\boldsymbol{\pi}$ denote the stationary distribution of $Q$. The \textbf{relative pointwise distance} (r.p.d.) of the Markov chain at time $t$ is a measure of deviation from the limit and is defined as
\[
\Delta(t) = \max_i \frac{|q_i^{(t)} - \pi_i|}{\pi_i}.
\]
\end{definition}

\subsubsection{GPT}

\subsubsection{Grok}
\begin{proof}
(a) We claim that $\bar{\Delta}(t)=\frac{1}{2}\|\mathbf{q}^{(t)}-\boldsymbol{\pi}\|_1$. To see this, note that for any probability distributions $\mathbf{q}$ and $\boldsymbol{\pi}$,
\[
\|\mathbf{q}-\boldsymbol{\pi}\|_1=\sum_{i\in S}|q_i-\pi_i|=2\max_{T\subseteq S}\sum_{i\in T}(q_i-\pi_i)^+,
\]
where $x^+= \max\{x,0\}$. Let $T^*=\{i\in S:q_i>\pi_i\}$. Then
\[
\max_{T\subseteq S}\left|\sum_{i\in T}(q_i-\pi_i)\right|=\max_{T\subseteq S}\sum_{i\in T}(q_i-\pi_i)^+=\sum_{i\in T^*}(q_i-\pi_i)^+=\frac{1}{2}\|\mathbf{q}-\boldsymbol{\pi}\|_1,
\]
since $\sum_{i\in S}(q_i-\pi_i)=0$ implies $\sum_{i\in S}(q_i-\pi_i)^+=\sum_{i\in S}(\pi_i-q_i)^+$. Thus, $\bar{\Delta}(t)=\frac{1}{2}\|\mathbf{q}^{(t)}-\boldsymbol{\pi}\|_1$.

(b) Assume $\Delta(t)\leq\epsilon$. Then $|q_i^{(t)}-\pi_i|\leq\epsilon\pi_i$ for all $i\in S$. It follows that
\[
\|\mathbf{q}^{(t)}-\boldsymbol{\pi}\|_1=\sum_{i\in S}|q_i^{(t)}-\pi_i|\leq\epsilon\sum_{i\in S}\pi_i=\epsilon.
\]
By part (a),
\[
\bar{\Delta}(t)=\frac{1}{2}\|\mathbf{q}^{(t)}-\boldsymbol{\pi}\|_1\leq\frac{\epsilon}{2}.
\]
To show tightness, consider a two-state Markov chain with states $\{1,2\}$ and stationary distribution $\boldsymbol{\pi}=(\frac{1}{2},\frac{1}{2})$. Let $\mathbf{q}^{(t)}=(\frac{1}{2}+\delta,\frac{1}{2}-\delta)$ for $0<\delta\leq\frac{1}{2}$. Then
\[
\Delta(t)=\max\left\{\frac{|\delta|}{\frac{1}{2}},\frac{|\delta|}{\frac{1}{2}}\right\}=2\delta,
\]
and
\[
\bar{\Delta}(t)=\max\left\{\left|(\frac{1}{2}+\delta)-\frac{1}{2}\right|,\left|\frac{1}{2}-\left(\frac{1}{2}+\delta\right)\right|\right\}=\delta=\frac{\Delta(t)}{2}.
\]
Thus, the bound $\bar{\Delta}(t)\leq\frac{\epsilon}{2}$ is tight.

(c) Assume $\bar{\Delta}(t)\leq\epsilon$. For any $i\in S$, the singleton set $T=\{i\}$ satisfies
\[
|q_i^{(t)}-\pi_i|=\left|\sum_{j\in\{i\}}(q_j^{(t)}-\pi_j)\right|\leq\max_{T\subseteq S}\left|\sum_{j\in T}(q_j^{(t)}-\pi_j)\right|=\bar{\Delta}(t)\leq\epsilon.
\]
Thus, $|q_i^{(t)}-\pi_i|\leq\epsilon$ for all $i\in S$, and
\[
\Delta(t)=\max_{i\in S}\frac{|q_i^{(t)}-\pi_i|}{\pi_i}\leq\max_{i\in S}\frac{\epsilon}{\pi_i}=\frac{\epsilon}{\min_{i\in S}\pi_i}.
\]
To show tightness, let $i^*\in\arg\min_{j\in S}\pi_j$ and consider the set $T=\{i^*\}$. Achieving $|q_{i^*}^{(t)}-\pi_{i^*}|=\epsilon$ while maintaining $\mathbf{q}^{(t)}$ as a probability vector is possible by adjusting mass on other states (e.g., via a reversible chain with appropriate transition probabilities). Then $\bar{\Delta}(t)\geq|q_{i^*}^{(t)}-\pi_{i^*}|=\epsilon$, and
\[
\Delta(t)\geq\frac{|q_{i^*}^{(t)}-\pi_{i^*}|}{\pi_{i^*}}=\frac{\epsilon}{\min_{i\in S}\pi_i}.
\]
Thus, the bound $\Delta(t)\leq\frac{\epsilon}{\min_{i\in S}\pi_i}$ is tight.
\end{proof}

\subsubsection{Gemini}

\subsubsection{Claude}
\begin{proof}
We provide complete proofs for all three parts of the problem.

\textbf{Part (a): Relation between $L_1$ distance and total variation distance.}

We claim that $\bar{\Delta}(t) = \frac{1}{2}\|\mathbf{q}^{(t)} - \boldsymbol{\pi}\|_1$.

First, we establish that $\bar{\Delta}(t) \leq \frac{1}{2}\|\mathbf{q}^{(t)} - \boldsymbol{\pi}\|_1$. For any subset $T \subseteq S$, we have
\[
\left| \sum_{i \in T} q_i^{(t)} - \sum_{i \in T} \pi_i \right| = \left| \sum_{i \in T} (q_i^{(t)} - \pi_i) \right| \leq \sum_{i \in T} |q_i^{(t)} - \pi_i| \leq \sum_{i \in S} |q_i^{(t)} - \pi_i| = \|\mathbf{q}^{(t)} - \boldsymbol{\pi}\|_1.
\]
Taking the maximum over all $T \subseteq S$ yields $\bar{\Delta}(t) \leq \|\mathbf{q}^{(t)} - \boldsymbol{\pi}\|_1$.

Next, we show that this bound is tight by constructing a specific set $T$ that achieves $\bar{\Delta}(t) = \frac{1}{2}\|\mathbf{q}^{(t)} - \boldsymbol{\pi}\|_1$. Define $T^* = \{i \in S : q_i^{(t)} - \pi_i > 0\}$, the set of states where $q_i^{(t)}$ exceeds $\pi_i$. Then
\[
\sum_{i \in T^*} (q_i^{(t)} - \pi_i) = \sum_{i \in T^*} |q_i^{(t)} - \pi_i|.
\]
Since $\sum_{i \in S} q_i^{(t)} = \sum_{i \in S} \pi_i = 1$, we have
\[
\sum_{i \in T^*} (q_i^{(t)} - \pi_i) = -\sum_{i \in S \setminus T^*} (q_i^{(t)} - \pi_i) = \sum_{i \in S \setminus T^*} (\pi_i - q_i^{(t)}) = \sum_{i \in S \setminus T^*} |q_i^{(t)} - \pi_i|.
\]
Therefore,
\[
\|\mathbf{q}^{(t)} - \boldsymbol{\pi}\|_1 = \sum_{i \in T^*} |q_i^{(t)} - \pi_i| + \sum_{i \in S \setminus T^*} |q_i^{(t)} - \pi_i| = 2\sum_{i \in T^*} (q_i^{(t)} - \pi_i).
\]
Since $\bar{\Delta}(t) \geq \sum_{i \in T^*} (q_i^{(t)} - \pi_i) = \frac{1}{2}\|\mathbf{q}^{(t)} - \boldsymbol{\pi}\|_1$ and we already showed $\bar{\Delta}(t) \leq \|\mathbf{q}^{(t)} - \boldsymbol{\pi}\|_1$, we conclude that $\bar{\Delta}(t) = \frac{1}{2}\|\mathbf{q}^{(t)} - \boldsymbol{\pi}\|_1$.

\textbf{Part (b): Bound on total variation distance given relative pointwise distance bound.}

Suppose $\Delta(t) \leq \epsilon$. Then for all $i \in S$, we have $|q_i^{(t)} - \pi_i| \leq \epsilon \pi_i$. Therefore,
\[
\|\mathbf{q}^{(t)} - \boldsymbol{\pi}\|_1 = \sum_{i \in S} |q_i^{(t)} - \pi_i| \leq \sum_{i \in S} \epsilon \pi_i = \epsilon \sum_{i \in S} \pi_i = \epsilon.
\]
Using the relation from part (a), we obtain
\[
\bar{\Delta}(t) = \frac{1}{2}\|\mathbf{q}^{(t)} - \boldsymbol{\pi}\|_1 \leq \frac{\epsilon}{2}.
\]
This bound is tight. To see this, consider a two-state Markov chain with $S = \{1, 2\}$ and stationary distribution $\pi_1 = \pi_2 = \frac{1}{2}$. Let $q_1^{(t)} = \frac{1}{2}(1 + \epsilon)$ and $q_2^{(t)} = \frac{1}{2}(1 - \epsilon)$. Then $\Delta(t) = \max\left\{\frac{|\frac{\epsilon}{2}|}{\frac{1}{2}}, \frac{|\frac{\epsilon}{2}|}{\frac{1}{2}}\right\} = \epsilon$ and $\bar{\Delta}(t) = \max\left\{\frac{\epsilon}{2}, \frac{\epsilon}{2}\right\} = \frac{\epsilon}{2}$.

\textbf{Part (c): What can be said about relative pointwise distance given total variation distance bound.}

Suppose $\bar{\Delta}(t) \leq \epsilon$. From part (a), this implies $\|\mathbf{q}^{(t)} - \boldsymbol{\pi}\|_1 \leq 2\epsilon$.

However, this does not provide a uniform bound on the relative pointwise distance $\Delta(t)$. For any fixed $\epsilon > 0$ and any arbitrarily large $M > 0$, we can construct examples where $\bar{\Delta}(t) \leq \epsilon$ but $\Delta(t) \geq M$.

Consider a Markov chain with states $S = \{1, 2, \ldots, n\}$ where $\pi_1 = 1 - \delta$, $\pi_i = \frac{\delta}{n-1}$ for $i \geq 2$, with $\delta$ small. Let $q_1^{(t)} = 1 - \delta - \alpha$ and $q_2^{(t)} = \frac{\delta}{n-1} + \alpha$, with $q_i^{(t)} = \frac{\delta}{n-1}$ for $i \geq 3$, where $\alpha$ is small. Then $\|\mathbf{q}^{(t)} - \boldsymbol{\pi}\|_1 = 2\alpha$, so $\bar{\Delta}(t) = \alpha$. However, $\Delta(t) \geq \frac{|q_2^{(t)} - \pi_2|}{\pi_2} = \frac{\alpha}{\delta/(n-1)} = \frac{\alpha(n-1)}{\delta}$. By choosing $\delta$ sufficiently small relative to $\alpha$, we can make $\Delta(t)$ arbitrarily large while keeping $\bar{\Delta}(t) = \alpha = \epsilon$ fixed.

Therefore, a bound on the total variation distance does not imply any finite bound on the relative pointwise distance.
\end{proof}

\subsection{Problem 25}

\subsubsection{Problem Statement}
\begin{theorem}[6.21]
Let $Q$ be the transition matrix of the aperiodic random walk on a $(d,\lambda)$-expander $G$ with $\lambda \leq d - \epsilon$. Then, for any initial distribution $\boldsymbol{\psi}^{(0)}$, the relative pointwise distance is bounded as follows:
\[
\Delta(t) \leq n^{1.5}(\lambda_2^*)^t \leq n^{1.5}\left(1 - \frac{\epsilon}{2d}\right)^t.
\]
\end{theorem}

\subsubsection{GPT}

\subsubsection{Grok}
\begin{proof}
Let $\mathbf{p}^{(0)}$ denote the initial probability column vector corresponding to $\boldsymbol{\psi}^{(0)}$, and let $\mathbf{p}^{(t)} = Q^t \mathbf{p}^{(0)}$ for $t \geq 0$. The uniform stationary distribution is $\boldsymbol{\pi} = n^{-1} \mathbf{1}$. The relative pointwise distance is
\[
\Delta(t) = \max_{v \in V} \left| \frac{p^{(t)}(v)}{\pi(v)} - 1 \right| = n \left\| \mathbf{p}^{(t)} - \boldsymbol{\pi} \right\|_\infty.
\]

Since $G$ is undirected and $d$-regular, the adjacency matrix $A$ is symmetric, and the transition matrix $Q$ of the aperiodic (lazy) random walk is $Q = \frac{1}{2} I + \frac{1}{2} D^{-1} A = \frac{1}{2} I + \frac{1}{2d} A$, where $D = d I$. Thus, $Q$ is symmetric and diagonalizable with orthonormal eigenvector basis $\{\phi_j\}_{j=1}^n$, where $\phi_1 = n^{-1/2} \mathbf{1}$ satisfies $Q \phi_1 = \phi_1$, and $Q \phi_j = \eta_j \phi_j$ for $j = 2, \dots, n$ with $\lambda_2^* = \max_{j=2,\dots,n} |\eta_j|$.

The eigenvalues of $A$ satisfy $\mu_1 = d$ and $|\mu_j| \leq \lambda \leq d - \epsilon$ for $j = 2, \dots, n$. Thus, the eigenvalues of $Q$ satisfy
\[
\eta_j = \frac{1}{2} + \frac{\mu_j}{2d}, \quad j = 2, \dots, n.
\]
It follows that
\[
\eta_j \leq \frac{1}{2} + \frac{\lambda}{2d} \leq \frac{1}{2} + \frac{d - \epsilon}{2d} = 1 - \frac{\epsilon}{2d}
\]
and
\[
\eta_j \geq \frac{1}{2} - \frac{\lambda}{2d} \geq \frac{1}{2} - \frac{d - \epsilon}{2d} = \frac{\epsilon}{2d} \geq 0.
\]
Therefore, $|\eta_j| \leq 1 - \epsilon/(2d)$ for $j = 2, \dots, n$, so $\lambda_2^* \leq 1 - \epsilon/(2d)$.

Now,
\[
\mathbf{p}^{(t)} = \sum_{j=1}^n \langle \mathbf{p}^{(0)}, \phi_j \rangle \eta_j^t \phi_j,
\]
and hence
\[
\mathbf{p}^{(t)} - \boldsymbol{\pi} = \sum_{j=2}^n \langle \mathbf{p}^{(0)}, \phi_j \rangle \eta_j^t \phi_j.
\]
The $\ell_2$-norm satisfies
\[
\left\| \mathbf{p}^{(t)} - \boldsymbol{\pi} \right\|_2^2 = \sum_{j=2}^n \left| \langle \mathbf{p}^{(0)}, \phi_j \rangle \right|^2 |\eta_j|^{2t} \leq (\lambda_2^*)^{2t} \sum_{j=2}^n \left| \langle \mathbf{p}^{(0)}, \phi_j \rangle \right|^2 = (\lambda_2^*)^{2t} \left( \left\| \mathbf{p}^{(0)} \right\|_2^2 - \left| \langle \mathbf{p}^{(0)}, \phi_1 \rangle \right|^2 \right).
\]
Since $\sum_v p^{(0)}(v) = 1$, we have $\left| \langle \mathbf{p}^{(0)}, \phi_1 \rangle \right| = n^{-1/2}$ and $\left\| \mathbf{p}^{(0)} \right\|_2^2 = \sum_v [p^{(0)}(v)]^2 \leq 1$, so
\[
\left\| \mathbf{p}^{(t)} - \boldsymbol{\pi} \right\|_2^2 \leq (\lambda_2^*)^{2t} (1 - n^{-1}) \leq (\lambda_2^*)^{2t}.
\]
Thus, $\left\| \mathbf{p}^{(t)} - \boldsymbol{\pi} \right\|_2 \leq (\lambda_2^*)^t$.

By the $\ell_2$--$\ell_1$ norm inequality,
\[
\left\| \mathbf{p}^{(t)} - \boldsymbol{\pi} \right\|_1 = \sum_{v \in V} \left| p^{(t)}(v) - \pi(v) \right| \leq \sqrt{n} \left\| \mathbf{p}^{(t)} - \boldsymbol{\pi} \right\|_2 \leq \sqrt{n} \, (\lambda_2^*)^t,
\]
where the first inequality follows from Cauchy--Schwarz: $\left( \sum_v |w_v| \right)^2 \leq n \sum_v w_v^2$ for $w_v = p^{(t)}(v) - \pi(v)$. Moreover,
\[
\left\| \mathbf{p}^{(t)} - \boldsymbol{\pi} \right\|_\infty \leq \left\| \mathbf{p}^{(t)} - \boldsymbol{\pi} \right\|_1 \leq \sqrt{n} \, (\lambda_2^*)^t.
\]
Therefore,
\[
\Delta(t) = n \left\| \mathbf{p}^{(t)} - \boldsymbol{\pi} \right\|_\infty \leq n^{3/2} (\lambda_2^*)^t \leq n^{3/2} \left(1 - \frac{\epsilon}{2d}\right)^t.
\]
\end{proof}

\subsubsection{Gemini}

\subsubsection{Claude}
\begin{proof}
Let $G = (V, E)$ be a $(d, \lambda)$-expander with $n$ vertices, and let $Q$ be the transition matrix of an aperiodic random walk on $G$. We are given that $\lambda \leq d - \epsilon$ for some $\epsilon > 0$.

First, we recall that for a $(d, \lambda)$-expander, the eigenvalues of the adjacency matrix $A$ satisfy $\lambda_1 = d$ (the largest eigenvalue corresponding to the uniform distribution) and all other eigenvalues $\lambda_i$ satisfy $|\lambda_i| \leq \lambda$ for $i = 2, 3, \ldots, n$.

For the transition matrix $Q = \frac{1}{d}A$, the eigenvalues are $\mu_i = \frac{\lambda_i}{d}$. In particular, $\mu_1 = 1$ (corresponding to the stationary distribution $\boldsymbol{\pi} = \frac{1}{n}\mathbf{1}$), and for $i \geq 2$, we have $|\mu_i| \leq \frac{\lambda}{d}$.

Define $\lambda_2^* = \max_{i \geq 2} |\mu_i| = \frac{\lambda}{d}$. This is the absolute value of the second-largest eigenvalue in magnitude.

The relative pointwise distance at time $t$ is defined as
\[
\Delta(t) = \max_v \left|\frac{\psi_v^{(t)}}{\pi_v} - 1\right|,
\]
where $\boldsymbol{\psi}^{(t)} = \boldsymbol{\psi}^{(0)}Q^t$ is the distribution at time $t$ and $\boldsymbol{\pi}$ is the stationary distribution.

We can write $\boldsymbol{\psi}^{(0)} = \boldsymbol{\pi} + \boldsymbol{\delta}^{(0)}$ where $\boldsymbol{\delta}^{(0)}$ is orthogonal to $\boldsymbol{\pi}$ (with respect to the inner product weighted by $\pi$). Then
\[
\boldsymbol{\psi}^{(t)} = \boldsymbol{\pi} + \boldsymbol{\delta}^{(0)}Q^t.
\]

Expanding $\boldsymbol{\delta}^{(0)}$ in the eigenbasis of $Q$, we have $\boldsymbol{\delta}^{(0)} = \sum_{i=2}^n c_i \boldsymbol{v}_i$ where $\boldsymbol{v}_i$ are the eigenvectors corresponding to eigenvalues $\mu_i$. Therefore,
\[
\boldsymbol{\delta}^{(0)}Q^t = \sum_{i=2}^n c_i \mu_i^t \boldsymbol{v}_i.
\]

For any vertex $v$, we have
\[
\frac{\psi_v^{(t)}}{\pi_v} - 1 = \frac{\delta_v^{(t)}}{\pi_v} = \frac{1}{\pi_v}\sum_{i=2}^n c_i \mu_i^t v_{i,v},
\]
where $v_{i,v}$ denotes the $v$-th component of eigenvector $\boldsymbol{v}_i$.

Taking absolute values and using the triangle inequality,
\[
\left|\frac{\psi_v^{(t)}}{\pi_v} - 1\right| \leq \frac{1}{\pi_v}\sum_{i=2}^n |c_i| |\mu_i|^t |v_{i,v}|.
\]

Since $|\mu_i| \leq \lambda_2^*$ for all $i \geq 2$, we have
\[
\left|\frac{\psi_v^{(t)}}{\pi_v} - 1\right| \leq \frac{(\lambda_2^*)^t}{\pi_v}\sum_{i=2}^n |c_i| |v_{i,v}|.
\]

Using the Cauchy-Schwarz inequality and the fact that the eigenvectors form an orthonormal basis,
\[
\sum_{i=2}^n |c_i| |v_{i,v}| \leq \left(\sum_{i=2}^n c_i^2\right)^{1/2}\left(\sum_{i=2}^n v_{i,v}^2\right)^{1/2}.
\]

Since $\pi_v = \frac{1}{n}$ for the uniform distribution and the $\ell^2$ norms are bounded, we have $\sum_{i=2}^n c_i^2 \leq \|\boldsymbol{\delta}^{(0)}\|_2^2 \leq 1$ and $\sum_{i=2}^n v_{i,v}^2 \leq 1$. By a more careful analysis using spectral properties and the structure of expanders, the bound
\[
\max_v \frac{1}{\pi_v}\sum_{i=2}^n |c_i| |v_{i,v}| \leq n^{1.5}
\]
can be established.

Therefore, we obtain
\[
\Delta(t) \leq n^{1.5}(\lambda_2^*)^t = n^{1.5}\left(\frac{\lambda}{d}\right)^t.
\]

Finally, using the assumption $\lambda \leq d - \epsilon$, we have
\[
\frac{\lambda}{d} \leq \frac{d - \epsilon}{d} = 1 - \frac{\epsilon}{d} \leq 1 - \frac{\epsilon}{2d}
\]
for sufficiently large $d$, which gives
\[
\Delta(t) \leq n^{1.5}\left(1 - \frac{\epsilon}{2d}\right)^t.
\]
\end{proof}

\subsection{Problem 26}

\subsubsection{Problem Statement}
\begin{theorem}
Let $G$ be $d$-regular, and define the matrix $Q = (I + A(G)/d)/2$. If the $i$th eigenvalue of $A(G)$ is $\lambda_i$, then the $i$th eigenvalue of $Q$ equals $(1 + \lambda_i/d)/2$.
\end{theorem}

\subsubsection{GPT}
\begin{proof}
Let $G$ be a $d$-regular graph with adjacency matrix $A:=A(G)$, and define $Q:=\frac12\!\left(I+\frac{1}{d}A\right)$. Fix an eigenvalue $\lambda_i$ of $A$ and choose a nonzero vector $v$ such that $Av=\lambda_i v$. Then
\[
Qv=\frac12\!\left(I+\frac{1}{d}A\right)v=\frac12\!\left(v+\frac{\lambda_i}{d}v\right)=\frac{1+\lambda_i/d}{2}\,v,
\]
so $v$ is an eigenvector of $Q$ with eigenvalue $\mu_i:=\frac{1+\lambda_i/d}{2}$. This proves that every eigenvalue $\lambda_i$ of $A$ produces an eigenvalue $\mu_i$ of $Q$ via the affine relation $\mu_i=\frac{1+\lambda_i/d}{2}$, with the same eigenvector.

For completeness, we note the converse: if $Qw=\mu w$ for some nonzero $w$, then
\[
\left(\frac12 I+\frac{1}{2d}A\right)w=\mu w \quad\Longrightarrow\quad Aw=d(2\mu-1)w,
\]
so $w$ is an eigenvector of $A$ with eigenvalue $\lambda=d(2\mu-1)$. Hence the spectra of $A$ and $Q$ are in bijective correspondence under the map $\lambda\mapsto \frac{1+\lambda/d}{2}$ (counting algebraic multiplicities). In particular, labeling the eigenvalues of $A$ as $(\lambda_i)_i$, the corresponding eigenvalues of $Q$ are exactly $\left(\frac{1+\lambda_i/d}{2}\right)_i$, which proves the claim.
\end{proof}

\subsubsection{Grok}
\begin{proof}
Let $A = A(G)$ denote the adjacency matrix of $G$. Suppose $\lambda_i$ is an eigenvalue of $A$ with corresponding eigenvector $v_i \neq 0$, so that $A v_i = \lambda_i v_i$. Then,
\[
Q v_i = \frac{I + A}{2} v_i = \frac{1}{2} I v_i + \frac{1}{2} A v_i = \frac{1}{2} v_i + \frac{1}{2} \lambda_i v_i = \left( \frac{1 + \lambda_i}{2} \right) v_i.
\]
Thus, $\frac{1 + \lambda_i}{2}$ is an eigenvalue of $Q$ with corresponding eigenvector $v_i$. Since the eigenvalues of $A$ and $Q$ are indexed in the same order (corresponding to the same eigenvectors), the $i$th eigenvalue of $Q$ is $\frac{1 + \lambda_i}{2}$.
\end{proof}

\subsubsection{Gemini}
\begin{proof}
Let $A(G)$ be the adjacency matrix of the $d$-regular graph $G$, and let $I$ denote the identity matrix of the same dimension. Let $\lambda_i$ denote the $i$th eigenvalue of $A(G)$, and let $v_i$ be a corresponding eigenvector. By definition, these satisfy the equation
\[
A(G)v_i = \lambda_i v_i.
\]
The matrix $Q$ is defined by the expression
\[
Q = \frac{1}{2}\left(I + \frac{1}{d}A(G)\right).
\]
To determine the eigenvalues of $Q$, we apply the matrix $Q$ to the eigenvector $v_i$. Using the linearity of matrix multiplication, we compute
\[
Q v_i = \left[ \frac{1}{2}\left(I + \frac{1}{d}A(G)\right) \right] v_i = \frac{1}{2}\left(I v_i + \frac{1}{d}A(G)v_i\right).
\]
Since $I v_i = v_i$ and $A(G)v_i = \lambda_i v_i$, we substitute these terms into the equation to obtain
\[
Q v_i = \frac{1}{2}\left(v_i + \frac{1}{d}\lambda_i v_i\right).
\]
Factoring out the vector $v_i$ yields
\[
Q v_i = \left[ \frac{1}{2}\left(1 + \frac{\lambda_i}{d}\right) \right] v_i.
\]
This equation shows that $v_i$ is an eigenvector of $Q$ and the scalar coefficient corresponds to the eigenvalue. Thus, the eigenvalue of $Q$ corresponding to $v_i$ is
\[
\mu_i = \frac{1 + \lambda_i/d}{2}.
\]
Since this derivation holds for any index $i$, we conclude that the $i$th eigenvalue of $Q$ is $(1 + \lambda_i/d)/2$.
\end{proof}

\subsubsection{Claude}
\begin{proof}
Let $G$ be a $d$-regular graph with adjacency matrix $A(G)$. We define $Q = (I + A(G)/d)/2 = \frac{1}{2}I + \frac{1}{2d}A(G)$.

Since $G$ is $d$-regular, the adjacency matrix $A(G)$ is a real symmetric matrix, and therefore $A(G)$ admits a complete orthonormal basis of eigenvectors. Let $\{\mathbf{v}_1, \mathbf{v}_2, \ldots, \mathbf{v}_n\}$ denote an orthonormal basis of eigenvectors of $A(G)$ corresponding to eigenvalues $\lambda_1, \lambda_2, \ldots, \lambda_n$ respectively, where $n$ is the number of vertices in $G$. Thus, for each $i \in \{1, 2, \ldots, n\}$, we have
\[
A(G)\mathbf{v}_i = \lambda_i \mathbf{v}_i.
\]

We now compute the action of $Q$ on the eigenvector $\mathbf{v}_i$. We have
\[
Q\mathbf{v}_i = \left(\frac{1}{2}I + \frac{1}{2d}A(G)\right)\mathbf{v}_i = \frac{1}{2}I\mathbf{v}_i + \frac{1}{2d}A(G)\mathbf{v}_i.
\]

Since $I\mathbf{v}_i = \mathbf{v}_i$ and $A(G)\mathbf{v}_i = \lambda_i\mathbf{v}_i$, we obtain
\[
Q\mathbf{v}_i = \frac{1}{2}\mathbf{v}_i + \frac{1}{2d}\lambda_i\mathbf{v}_i = \left(\frac{1}{2} + \frac{\lambda_i}{2d}\right)\mathbf{v}_i = \frac{1 + \lambda_i/d}{2}\mathbf{v}_i.
\]

This shows that $\mathbf{v}_i$ is an eigenvector of $Q$ with corresponding eigenvalue $\frac{1 + \lambda_i/d}{2}$.

Since the eigenvectors $\{\mathbf{v}_1, \mathbf{v}_2, \ldots, \mathbf{v}_n\}$ form a complete orthonormal basis, and we have shown that each $\mathbf{v}_i$ is an eigenvector of $Q$ with eigenvalue $\frac{1 + \lambda_i/d}{2}$, it follows that the $i$th eigenvalue of $Q$ equals $\frac{1 + \lambda_i/d}{2}$ for each $i \in \{1, 2, \ldots, n\}$.
\end{proof}

\subsection{Problem 27}

\subsubsection{Problem Statement}
\begin{theorem}[Alon and Chung]
Let $G(X, Y, E)$ be a $d$-regular, connected, bipartite (multi)graph. Then for any sets $S \subseteq X$ and $T \subseteq Y$, the number of edges connecting $S$ and $T$ is at least
$$\frac{\lambda_1 |S||T|}{n} - \lambda_2\sqrt{|S||T|},$$
where $n = |X| = |Y|$ and $\lambda_1, \lambda_2$ are the largest and second-largest eigenvalues of the adjacency matrix of $G$.
\end{theorem}

\subsubsection{GPT}
\begin{proof}
Let $G(X, Y, E)$ be a $d$-regular, connected, bipartite (multi)graph, where $|X| = |Y| = n$ and the adjacency matrix of $G$ has eigenvalues $\lambda_1 = d \geq \lambda_2 \geq \dots \geq \lambda_n$. We are tasked with proving the following lower bound on the number of edges connecting a subset $S \subseteq X$ and $T \subseteq Y$:

\[
|E(S, T)| \geq \frac{\lambda_1 |S||T|}{n} - \lambda_2 \sqrt{|S||T|}.
\]

Let us consider the characteristic vectors of the sets $S$ and $T$. Denote by $\mathbf{1}_S$ the indicator vector of the set $S$ in $\mathbb{R}^n$, i.e., $\mathbf{1}_S \in \mathbb{R}^n$ where the $i$-th entry is 1 if $x_i \in S$ and 0 otherwise. Similarly, define $\mathbf{1}_T \in \mathbb{R}^n$ for the set $T \subseteq Y$.

Define the adjacency matrix $A$ of the bipartite graph $G$ such that $A$ is an $n \times n$ matrix with entries $A_{ij} = 1$ if there is an edge between vertex $x_i \in X$ and vertex $y_j \in Y$, and $0$ otherwise. The number of edges between $S$ and $T$, denoted $|E(S, T)|$, is given by the inner product of the indicator vectors $\mathbf{1}_S$ and $A \mathbf{1}_T$:

\[
|E(S, T)| = \langle \mathbf{1}_S, A \mathbf{1}_T \rangle = \sum_{x_i \in S, y_j \in T} A_{ij}.
\]

Next, we use the spectral decomposition of the adjacency matrix $A$. Since $A$ is a bipartite graph, the eigenvalues of $A$ come in pairs of opposite signs. Let the eigenvalues of $A$ be $\lambda_1 = d$ (the largest eigenvalue) and $\lambda_2, \dots, \lambda_n$ be the remaining eigenvalues. The spectral theorem guarantees that for any vectors $\mathbf{u}, \mathbf{v} \in \mathbb{R}^n$,

\[
\langle \mathbf{u}, A \mathbf{v} \rangle = \sum_{i=1}^n \lambda_i \langle \mathbf{u}, \mathbf{v}_i \rangle,
\]
where $\mathbf{v}_i$ are the eigenvectors of $A$.

We apply the Rayleigh quotient inequality to obtain a bound on the eigenvalues of the adjacency matrix. Since $A$ is symmetric, we know:

\[
\frac{\langle \mathbf{u}, A \mathbf{v} \rangle}{\|\mathbf{v}\|} \leq \lambda_1.
\]

After simplifying the expression for $E(S, T)$, the bound follows and it gives the result we wanted to show.
\end{proof}

\subsubsection{Grok}
\begin{proof}
Let $n=|X|=|Y|$. Consider the biadjacency matrix $B\in\mathbb{R}^{X\times Y}$ of $G$, where the rows are indexed by $X$ and the columns by $Y$. For $x\in X$ and $y\in Y$, the entry $B_{xy}$ is the multiplicity of the edge between $x$ and $y$. Since $G$ is $d$-regular, $B\mathbf{1}_Y=d\mathbf{1}_X$ and $B^\top\mathbf{1}_X=d\mathbf{1}_Y$, where $\mathbf{1}_Z$ denotes the all-ones vector in $\mathbb{R}^Z$ for $Z\subseteq X\cup Y$.

Let $\chi_S\in\mathbb{R}^X$ be the characteristic vector of $S$, so that $(\chi_S)_x=1$ if $x\in S$ and $0$ otherwise. Similarly, let $\chi_T\in\mathbb{R}^Y$ be the characteristic vector of $T$. The number of edges between $S$ and $T$ is then
\[
e(S,T)=\chi_S^\top B\chi_T.
\]

Decompose $\chi_S$ with respect to the eigenspace of the trivial eigenvalue: the projection of $\chi_S$ onto $\operatorname{span}\{\mathbf{1}_X\}$ is $\frac{|\chi_S^\top\mathbf{1}_X|}{\|\mathbf{1}_X\|^2}\mathbf{1}_X=\frac{|S|}{n}\mathbf{1}_X$. Let $\psi_S=\chi_S-\frac{|S|}{n}\mathbf{1}_X$; then $\psi_S\perp\mathbf{1}_X$ and $\|\psi_S\|^2=|\chi_S^\top\chi_S|-\left(\frac{|S|}{n}\right)^2\|\mathbf{1}_X\|^2=|S|-\frac{|S|^2}{n}=|S|\left(1-\frac{|S|}{n}\right)\leq|S|$. Similarly, decompose $\chi_T=\frac{|T|}{n}\mathbf{1}_Y+\psi_T$ with $\psi_T\perp\mathbf{1}_Y$ and $\|\psi_T\|^2\leq|T|$.

Now substitute the decompositions:
\begin{align*}
e(S,T)&=\left(\frac{|S|}{n}\mathbf{1}_X+\psi_S\right)^\top B\left(\frac{|T|}{n}\mathbf{1}_Y+\psi_T\right)\\
&=\frac{|S||T|}{n^2}\mathbf{1}_X^\top B\mathbf{1}_Y+\frac{|S|}{n}\mathbf{1}_X^\top B\psi_T+\frac{|T|}{n}\psi_S^\top B\mathbf{1}_Y+\psi_S^\top B\psi_T.
\end{align*}
The first term is $\frac{|S||T|}{n^2}\mathbf{1}_X^\top(d\mathbf{1}_X)=d\frac{|S||T|}{n}$. For the second term,
\[
\mathbf{1}_X^\top B\psi_T=(B^\top\mathbf{1}_X)^\top\psi_T=(d\mathbf{1}_Y)^\top\psi_T=d\mathbf{1}_Y^\top\psi_T=0,
\]
since $\psi_T\perp\mathbf{1}_Y$. Similarly, the third term is
\[
\psi_S^\top B\mathbf{1}_Y=(B^\top\psi_S)^\top\mathbf{1}_Y=\psi_S^\top(B\mathbf{1}_Y)=\psi_S^\top(d\mathbf{1}_X)=d\psi_S^\top\mathbf{1}_X=0,
\]
since $\psi_S\perp\mathbf{1}_X$. Thus,
\[
e(S,T)=\frac{d|S||T|}{n}+\psi_S^\top B\psi_T.
\]

It remains to bound $|\psi_S^\top B\psi_T|$. Consider the adjacency matrix $A$ of $G$, which is the $(2n)\times(2n)$ block matrix
\[
A=\begin{pmatrix}0&B\\B^\top&0\end{pmatrix}.
\]
The eigenvalues of $A$ include $d$ (with eigenvector proportional to $(\mathbf{1}_X,\mathbf{1}_Y)^\top$) and $-d$ (with eigenvector proportional to $(\mathbf{1}_X,-\mathbf{1}_Y)^\top$). Let $\lambda_2$ be the second-largest absolute eigenvalue of $A$, so $|\lambda|\leq\lambda_2<d$ for all other eigenvalues $\lambda$ of $A$.

The eigenspaces corresponding to eigenvalues other than $\pm d$ are contained in the orthogonal complement of $\operatorname{span}\{\mathbf{1}_X\}\oplus\operatorname{span}\{\mathbf{1}_Y\}$ in $\mathbb{R}^X\oplus\mathbb{R}^Y$. For any eigenvector $v=(u,w)^\top$ of $A$ with eigenvalue $\lambda$ and $|\lambda|\leq\lambda_2$, we have $Bu=\lambda w$ and $B^\top w=\lambda u$. Thus, for $\phi\in\mathbf{1}_X^\perp$ and $\eta\in\mathbf{1}_Y^\perp$,
\[
|\phi^\top B\eta|\leq\lambda_2\|\phi\|\|\eta\|,
\]
since the restriction of $B$ (viewed as an operator $\mathbf{1}_Y^\perp\to\mathbf{1}_X^\perp$) has operator norm at most $\lambda_2$. Applying this bound with $\phi=\psi_S$ and $\eta=\psi_T$ yields
\[
|\psi_S^\top B\psi_T|\leq\lambda_2\|\psi_S\|\|\psi_T\|\leq\lambda_2\sqrt{|S||T|}.
\]
Therefore,
\[
e(S,T)\geq\frac{d|S||T|}{n}-\lambda_2\sqrt{|S||T|},
\]
as required.
\end{proof}

\subsubsection{Gemini}
\begin{proof}
Let $A$ be the adjacency matrix of the $d$-regular, connected, bipartite graph $G$ on $n$ vertices. Let the eigenvalues of $A$ be ordered as $\lambda_1 \ge \lambda_2 \ge \dots \ge \lambda_n$. Since $G$ is $d$-regular, $\lambda_1 = d$. Since $G$ is bipartite, the spectrum of $A$ is symmetric about 0, which implies $\lambda_n = -\lambda_1 = -d$. Furthermore, since $G$ is connected, the eigenvalue $d$ has multiplicity 1, so $\lambda_1 > \lambda_2$.

Let $\{v_1, v_2, \dots, v_n\}$ be an orthonormal basis of eigenvectors corresponding to $\lambda_1, \dots, \lambda_n$. For a $d$-regular connected bipartite graph with partition sets $X$ and $Y$ (where $|X| = |Y| = n/2$), the first and last eigenvectors are given by:
\[
v_1 = \frac{1}{\sqrt{n}} \mathbf{1} = \frac{1}{\sqrt{n}} (\mathbf{1}_X + \mathbf{1}_Y)
\]
\[
v_n = \frac{1}{\sqrt{n}} (\mathbf{1}_X - \mathbf{1}_Y)
\]
where $\mathbf{1}_X$ and $\mathbf{1}_Y$ are the characteristic vectors of the sets $X$ and $Y$ respectively.

Let $S \subseteq X$ and $T \subseteq Y$ be arbitrary sets. Let $\mathbf{1}_S$ and $\mathbf{1}_T$ be their characteristic vectors. The number of edges between $S$ and $T$ is given by the bilinear form:
\[
e(S, T) = \mathbf{1}_S^T A \mathbf{1}_T.
\]
We expand $\mathbf{1}_S$ and $\mathbf{1}_T$ in the basis of eigenvectors:
\[
\mathbf{1}_S = \sum_{i=1}^n \alpha_i v_i, \quad \text{where } \alpha_i = \langle \mathbf{1}_S, v_i \rangle,
\]
\[
\mathbf{1}_T = \sum_{i=1}^n \beta_i v_i, \quad \text{where } \beta_i = \langle \mathbf{1}_T, v_i \rangle.
\]
Substituting these expansions into the expression for $e(S, T)$ yields:
\[
e(S, T) = \left( \sum_{i=1}^n \alpha_i v_i \right)^T A \left( \sum_{j=1}^n \beta_j v_j \right) = \sum_{i=1}^n \lambda_i \alpha_i \beta_i.
\]
We calculate the coefficients for indices $i=1$ and $i=n$ explicitly. Since $S \subseteq X$ and $T \subseteq Y$:
\[
\alpha_1 = \langle \mathbf{1}_S, v_1 \rangle = \frac{1}{\sqrt{n}} |S|, \quad \beta_1 = \langle \mathbf{1}_T, v_1 \rangle = \frac{1}{\sqrt{n}} |T|.
\]
\[
\alpha_n = \langle \mathbf{1}_S, v_n \rangle = \frac{1}{\sqrt{n}} \sum_{u \in S} 1 = \frac{|S|}{\sqrt{n}}.
\]
\[
\beta_n = \langle \mathbf{1}_T, v_n \rangle = \frac{1}{\sqrt{n}} \sum_{u \in T} (-1) = -\frac{|T|}{\sqrt{n}}.
\]
Now we separate the terms for $i=1$ and $i=n$ from the sum:
\[
e(S, T) = \lambda_1 \alpha_1 \beta_1 + \lambda_n \alpha_n \beta_n + \sum_{i=2}^{n-1} \lambda_i \alpha_i \beta_i.
\]
Substituting the values $\lambda_1 = d$, $\lambda_n = -d$, and the calculated coefficients:
\[
\lambda_1 \alpha_1 \beta_1 = d \left( \frac{|S|}{\sqrt{n}} \right) \left( \frac{|T|}{\sqrt{n}} \right) = \frac{d |S| |T|}{n}.
\]
\[
\lambda_n \alpha_n \beta_n = (-d) \left( \frac{|S|}{\sqrt{n}} \right) \left( -\frac{|T|}{\sqrt{n}} \right) = \frac{d |S| |T|}{n}.
\]
Thus, the main contribution is:
\[
e(S, T) = \frac{2d |S| |T|}{n} + \sum_{i=2}^{n-1} \lambda_i \alpha_i \beta_i.
\]
We now bound the absolute value of the remaining sum (the error term). Due to the symmetry of the spectrum of the bipartite graph, the eigenvalues satisfy $\lambda_i = -\lambda_{n-i+1}$. Therefore, for all $i \in \{2, \dots, n-1\}$, $|\lambda_i| \le \lambda_2$. Applying the Triangle Inequality and the Cauchy-Schwarz Inequality:
\[
\left| \sum_{i=2}^{n-1} \lambda_i \alpha_i \beta_i \right| \le \max_{i \in \{2, \dots, n-1\}} |\lambda_i| \sum_{i=2}^{n-1} |\alpha_i \beta_i| \le \lambda_2 \sqrt{\sum_{i=2}^{n-1} \alpha_i^2} \sqrt{\sum_{i=2}^{n-1} \beta_i^2}.
\]
By Parseval's identity, $\sum_{i=1}^n \alpha_i^2 = \|\mathbf{1}_S\|^2 = |S|$ and $\sum_{i=1}^n \beta_i^2 = \|\mathbf{1}_T\|^2 = |T|$. Therefore:
\[
\sum_{i=2}^{n-1} \alpha_i^2 = |S| - \alpha_1^2 - \alpha_n^2 = |S| - \frac{|S|^2}{n} - \frac{|S|^2}{n} < |S|,
\]
\[
\sum_{i=2}^{n-1} \beta_i^2 = |T| - \beta_1^2 - \beta_n^2 = |T| - \frac{|T|^2}{n} - \frac{|T|^2}{n} < |T|.
\]
Substituting these bounds back into the error term inequality:
\[
\left| \sum_{i=2}^{n-1} \lambda_i \alpha_i \beta_i \right| < \lambda_2 \sqrt{|S| |T|}.
\]
This implies:
\[
\sum_{i=2}^{n-1} \lambda_i \alpha_i \beta_i \ge - \lambda_2 \sqrt{|S| |T|}.
\]
Combining this with the main term derived earlier:
\[
e(S, T) \ge \frac{2d |S| |T|}{n} - \lambda_2 \sqrt{|S| |T|}.
\]
Since $\lambda_1 = d$ and the term $\frac{\lambda_1 |S| |T|}{n}$ is strictly positive, it follows that $\frac{2\lambda_1 |S| |T|}{n} \ge \frac{\lambda_1 |S| |T|}{n}$. Therefore:
\[
e(S, T) \ge \frac{\lambda_1 |S| |T|}{n} - \lambda_2 \sqrt{|S| |T|}.
\]
\end{proof}

\subsubsection{Claude}
\begin{proof}
Let $G(X, Y, E)$ be a $d$-regular, connected, bipartite multigraph with $|X| = |Y| = n$. Let $A$ denote the adjacency matrix of $G$, and let $\lambda_1 \geq \lambda_2 \geq \cdots \geq \lambda_{2n}$ be the eigenvalues of $A$. Since $G$ is $d$-regular and bipartite, we have $\lambda_1 = d$ and $\lambda_{2n} = -d$. By convention, $\lambda_2$ denotes the second-largest eigenvalue.

For any sets $S \subseteq X$ and $T \subseteq Y$, define the characteristic vectors $\mathbf{1}_S \in \mathbb{R}^{2n}$ and $\mathbf{1}_T \in \mathbb{R}^{2n}$ where $(\mathbf{1}_S)_i = 1$ if vertex $i$ is in $S$ and $0$ otherwise, and similarly for $\mathbf{1}_T$. The number of edges between $S$ and $T$ is given by $e(S, T) = \mathbf{1}_S^T A \mathbf{1}_T$.

Let $\mathbf{j}$ denote the all-ones vector in $\mathbb{R}^{2n}$. Since $G$ is $d$-regular, we have $A\mathbf{j} = d\mathbf{j}$, so $\mathbf{j}$ is an eigenvector with eigenvalue $\lambda_1 = d$. We decompose the characteristic vectors as
$$\mathbf{1}_S = \alpha_S \mathbf{j} + \mathbf{u}_S \quad \text{and} \quad \mathbf{1}_T = \alpha_T \mathbf{j} + \mathbf{u}_T,$$
where $\mathbf{u}_S$ and $\mathbf{u}_T$ are orthogonal to $\mathbf{j}$.

Since $\mathbf{j}^T \mathbf{1}_S = |S|$ and $\|\mathbf{j}\|^2 = 2n$, we have $\alpha_S = \frac{|S|}{2n}$. Similarly, $\alpha_T = \frac{|T|}{2n}$.

Now we compute:
\begin{align*}
e(S, T) &= \mathbf{1}_S^T A \mathbf{1}_T \\
&= (\alpha_S \mathbf{j} + \mathbf{u}_S)^T A (\alpha_T \mathbf{j} + \mathbf{u}_T) \\
&= \alpha_S \alpha_T \mathbf{j}^T A \mathbf{j} + \alpha_S \mathbf{j}^T A \mathbf{u}_T + \alpha_T \mathbf{u}_S^T A \mathbf{j} + \mathbf{u}_S^T A \mathbf{u}_T \\
&= \alpha_S \alpha_T d \|\mathbf{j}\|^2 + \alpha_S d \mathbf{j}^T \mathbf{u}_T + \alpha_T d \mathbf{u}_S^T \mathbf{j} + \mathbf{u}_S^T A \mathbf{u}_T.
\end{align*}

Since $\mathbf{u}_S \perp \mathbf{j}$ and $\mathbf{u}_T \perp \mathbf{j}$, the middle two terms vanish. Thus:
$$e(S, T) = \alpha_S \alpha_T d (2n) + \mathbf{u}_S^T A \mathbf{u}_T = \frac{|S||T|}{2n} \cdot d + \mathbf{u}_S^T A \mathbf{u}_T.$$

Since $G$ is $d$-regular and bipartite, the eigenvalues satisfy $|\lambda_i| \leq d$ for all $i$, and for $i \geq 2$, we have $|\lambda_i| \leq \lambda_2$ where $\lambda_2$ is the second-largest eigenvalue. Let $\mathbf{v}_1, \mathbf{v}_2, \ldots, \mathbf{v}_{2n}$ be an orthonormal basis of eigenvectors with $A\mathbf{v}_i = \lambda_i \mathbf{v}_i$ and $\mathbf{v}_1 = \frac{\mathbf{j}}{\|\mathbf{j}\|}$.

Expanding $\mathbf{u}_S$ and $\mathbf{u}_T$ in this basis with $\mathbf{u}_S = \sum_{i=2}^{2n} \beta_i \mathbf{v}_i$ and $\mathbf{u}_T = \sum_{i=2}^{2n} \gamma_i \mathbf{v}_i$, we obtain:
$$\mathbf{u}_S^T A \mathbf{u}_T = \sum_{i=2}^{2n} \lambda_i \beta_i \gamma_i \geq -\lambda_2 \sum_{i=2}^{2n} |\beta_i \gamma_i| \geq -\lambda_2 \|\mathbf{u}_S\| \|\mathbf{u}_T\|,$$
where the last inequality follows from the Cauchy-Schwarz inequality.

Now, $\|\mathbf{1}_S\|^2 = |S|$, so $|S| = \alpha_S^2 \|\mathbf{j}\|^2 + \|\mathbf{u}_S\|^2 = \frac{|S|^2}{2n} + \|\mathbf{u}_S\|^2$, giving $\|\mathbf{u}_S\|^2 = |S|\left(1 - \frac{|S|}{2n}\right) \leq |S|$. Similarly, $\|\mathbf{u}_T\| \leq \sqrt{|T|}$.

Therefore:
$$e(S, T) \geq \frac{d|S||T|}{2n} - \lambda_2 \sqrt{|S||T|}.$$

For a bipartite graph, $\lambda_1 = d$. Thus:
$$e(S, T) \geq \frac{\lambda_1 |S||T|}{2n} - \lambda_2\sqrt{|S||T|}.$$

Since the edges only go between $X$ and $Y$, and observing that for the bipartite structure we have $n = |X| = |Y|$, the factor becomes $\frac{\lambda_1|S||T|}{n}$ in the standard normalization, yielding:
$$e(S, T) \geq \frac{\lambda_1 |S||T|}{n} - \lambda_2\sqrt{|S||T|}.$$
\end{proof}

\subsection{Problem 28}

\subsubsection{Problem Statement}
\begin{theorem}
Let $G$ be an $(n, d, c)$-expander. There exist constants $\beta, \delta > 0$ such that for any ``bad'' set of vertices $B$ of cardinality at most $\delta n$, the following property holds: the probability that, starting from a vertex chosen uniformly at random, a random walk of length $\ell$ does not visit any vertex outside of $B$ is at most $\exp(-\beta \ell)$.

\medskip

What properties of $G$ are essential for this proof?
\end{theorem}

\subsubsection{GPT}
\begin{proof}
Let $G$ be an $(n,d,c)$-expander and let $B \subseteq V(G)$ be a ``bad'' set of vertices such that $|B| \leq \delta n$ for some constant $\delta > 0$. We are required to show that there exist constants $\beta, \delta' > 0$ such that the probability that a random walk of length $\ell$ starting from a uniformly chosen vertex does not visit any vertex outside of $B$ is at most $\exp(-\beta \ell)$.

    Step 1: Defining the problem and setup
Let $v_0$ be a vertex chosen uniformly at random from $V(G)$ and let $v_1, v_2, \dots, v_\ell$ be the vertices visited during a random walk of length $\ell$ starting from $v_0$. We are interested in the event that the random walk remains entirely within the set $B$, i.e., for all $i \in \{1, 2, \dots, \ell\}$, we have $v_i \in B$.

Define the indicator random variable $X_i$ for each step $i$ as follows:
\[
X_i = \begin{cases}
1 & \text{if } v_i \in B, \\
0 & \text{if } v_i \notin B.
\end{cases}
\]
We aim to bound the probability that $X_1 = X_2 = \dots = X_\ell = 1$, meaning that all the vertices visited by the random walk lie within the set $B$.

    Step 2: Expander graph properties and volume argument
Since $G$ is an $(n,d,c)$-expander, for any set $S \subseteq V(G)$ with $|S| \leq \delta n$, the number of edges leaving $S$ is at least $c|S|d$. In particular, this means that if a random walk is at a vertex in $B$, the probability of moving to a vertex outside of $B$ is at least some positive constant. More formally, for any $v \in B$,
\[
\Pr[v_1 \notin B \mid v_0 = v] \geq \frac{c d}{n}.
\]

Thus, the probability that a random walk of length $\ell$ stays within the set $B$ can be bounded by iterating over the probabilities of staying within $B$ at each step. In particular, the probability that the random walk of length $\ell$ does not exit $B$ is at most the product of the probabilities that the random walk stays within $B$ at each step. This yields the following bound:
\[
\Pr[v_1, v_2, \dots, v_\ell \in B] \leq \left(1 - \frac{c d}{n}\right)^\ell.
\]

    Step 3: Exponentiation and final bound
Using the approximation $(1 - x) \approx e^{-x}$ for small $x$, we can rewrite the above expression as:
\[
\left(1 - \frac{c d}{n}\right)^\ell \approx \exp\left(-\frac{c d \ell}{n}\right).
\]
Thus, the probability that the random walk stays entirely within $B$ is at most $\exp\left(-\frac{c d \ell}{n}\right)$.

    Step 4: Conclusion
We conclude that there exist constants $\beta > 0$ and $\delta' > 0$ such that for any bad set $B$ with $|B| \leq \delta' n$, the probability that a random walk of length $\ell$ starting from a uniformly chosen vertex does not visit any vertex outside of $B$ is at most $\exp(-\beta \ell)$. The constant $\beta$ is related to the expansion parameters $c$ and $d$, and the set size $\delta'$ is chosen sufficiently small to ensure that the lower bound on edge expansion applies.

\end{proof}

\subsubsection{Grok}

\subsubsection{Gemini}
\begin{proof}
Let $G = (V, E)$ be an $(n, d, c)$-expander. Let $A$ be the adjacency matrix of $G$, and let $M = \frac{1}{d}A$ be the normalized transition matrix of the random walk. Since $G$ is an expander, the second largest eigenvalue of $M$ in absolute value, denoted by $\lambda$, satisfies $\lambda < 1$. This spectral gap is a direct consequence of the expansion property (via Cheeger's inequality or the Alon-Milman theorem).

Let $B \subset V$ be a subset of vertices with cardinality $|B| \le \delta n$, where $\delta > 0$ is a constant to be determined. Let $P_B$ be the diagonal projection matrix such that $(P_B)_{ii} = 1$ if $i \in B$ and $0$ otherwise. The operator $M_B = P_B M P_B$ represents the transition probabilities of the random walk restricted to the set $B$.

Consider a random walk $X_0, X_1, \dots, X_\ell$ of length $\ell$. The initial vertex $X_0$ is chosen uniformly at random from $V$, so the initial probability distribution is given by the vector $u = \frac{1}{n}\mathbf{1}$, where $\mathbf{1}$ is the all-ones vector.

The probability that the random walk remains entirely within the set $B$ for all steps $t = 0, \dots, \ell$ is given by:
\[
p_{walk} = \mathbb{P}[\forall t \in \{0, \dots, \ell\}, X_t \in B] = \left\| (P_B M P_B)^\ell P_B u \right\|_1.
\]
Since the entries of the vector are non-negative, the $\ell_1$ norm is equal to the inner product with $\mathbf{1}$. However, it is more convenient to work with the $\ell_2$ norm. Using the Cauchy-Schwarz inequality and the fact that the support of the resulting vector is contained in $B$:
\[
p_{walk} = \langle \mathbf{1}_B, (M_B)^\ell u \rangle \le \|\mathbf{1}_B\|_2 \| (M_B)^\ell u \|_2,
\]
where $\mathbf{1}_B$ is the indicator vector of the set $B$.
We compute the norms:
\[
\|\mathbf{1}_B\|_2 = \sqrt{|B|} \quad \text{and} \quad \|u\|_2 = \frac{1}{n}\|\mathbf{1}\|_2 = \frac{\sqrt{n}}{n} = \frac{1}{\sqrt{n}}.
\]
Thus,
\[
p_{walk} \le \sqrt{|B|} \| M_B \|_2^\ell \frac{1}{\sqrt{n}} = \sqrt{\frac{|B|}{n}} \| M_B \|_2^\ell \le \sqrt{\delta} \| M_B \|_2^\ell.
\]
We now bound the spectral norm $\| M_B \|_2$. Let $x \in \mathbb{R}^n$ be an arbitrary vector supported on $B$ (i.e., $P_B x = x$) with $\|x\|_2 = 1$. We decompose $x$ into a component parallel to the uniform distribution and a component orthogonal to it:
\[
x = \alpha \mathbf{u} + z,
\]
where $\mathbf{u} = \frac{1}{\sqrt{n}}\mathbf{1}$, $\alpha = \langle x, \mathbf{u} \rangle$, and $z \perp \mathbf{u}$.
Since $x$ is supported on $B$, by the Cauchy-Schwarz inequality:
\[
|\alpha| = |\langle x, \mathbf{u} \rangle| = \left| \sum_{i \in B} x_i \frac{1}{\sqrt{n}} \right| \le \frac{1}{\sqrt{n}} \sqrt{|B|} \|x\|_2 = \sqrt{\frac{|B|}{n}} \le \sqrt{\delta}.
\]
Applying the matrix $M$:
\[
Mx = \alpha M \mathbf{u} + M z = \alpha \mathbf{u} + M z.
\]
Since $\mathbf{u}$ is the eigenvector corresponding to eigenvalue 1, and $z$ is orthogonal to $\mathbf{u}$, $\|Mz\|_2 \le \lambda \|z\|_2 \le \lambda$. Therefore,
\[
\| M_B x \|_2 = \| P_B M x \|_2 \le \| M x \|_2 \le \|\alpha \mathbf{u}\|_2 + \|M z\|_2 = |\alpha| + \|M z\|_2 \le \sqrt{\delta} + \lambda.
\]
Thus, the operator norm is bounded by $\| M_B \|_2 \le \lambda + \sqrt{\delta}$.

Substituting this back into the probability bound:
\[
p_{walk} \le \sqrt{\delta} (\lambda + \sqrt{\delta})^\ell.
\]
Since $G$ is a fixed expander, $\lambda < 1$ is a constant. We choose $\delta > 0$ sufficiently small such that two conditions are met:
1. $\sqrt{\delta} \le 1$ (which is trivial for small $\delta$).
2. $\lambda + \sqrt{\delta} \le \exp(-\delta)$.

To see that such a $\delta$ exists, consider the function $f(y) = \lambda + \sqrt{y} - e^{-y}$. As $y \to 0^+$, $\lambda + \sqrt{y} \to \lambda < 1$, while $e^{-y} \to 1$. Since $\lambda < 1$, there exists a sufficiently small $y = \delta$ such that $\lambda + \sqrt{\delta} < 1 - \delta < e^{-\delta}$.

With this choice of $\delta$, we have:
\[
p_{walk} \le \sqrt{\delta} \exp(-\delta \ell).
\]
For $\ell \ge 1$, and assuming without loss of generality that $\sqrt{\delta} \le 1$ (or absorbing the constant factor by adjusting $\delta$ or considering sufficiently large $\ell$), the probability is bounded by $\exp(-\delta \ell)$.
\end{proof}

\subsubsection{Claude}
\begin{proof}
We establish the exponential decay of the probability that a random walk of length $\ell$ starting from a uniformly random vertex remains within a bad set $B$ of size at most $\delta n$, where $\delta > 0$ is a sufficiently small constant to be determined.

Let $P$ denote the transition matrix of the random walk on $G$, where $P_{ij} = 1/d$ if $(i,j)$ is an edge and $P_{ij} = 0$ otherwise. Let $\pi$ denote the uniform distribution over vertices, so $\pi(v) = 1/n$ for all $v \in V(G)$. Since $G$ is $d$-regular, the uniform distribution is stationary: $\pi P = \pi$.

For any set $S \subseteq V(G)$, let $\mathbf{1}_S$ denote the indicator vector of $S$, and let $\pi_S$ denote the uniform distribution restricted to $S$, normalized so that $\pi_S(v) = 1/|S|$ for $v \in S$ and $\pi_S(v) = 0$ otherwise.

The probability that a random walk of length $\ell$ starting from a uniformly random vertex stays within $B$ is
\[
p_\ell = \sum_{v \in V} \pi(v) \cdot \mathbb{P}[\text{walk of length } \ell \text{ from } v \text{ stays in } B] = \langle \pi, P^\ell \mathbf{1}_B \rangle = \langle \pi, \mathbf{1}_B^T P^\ell \mathbf{1}_B \rangle.
\]

By the symmetry of $P$ with respect to the uniform distribution, we have
\[
p_\ell = \frac{1}{n} \sum_{v \in V} (P^\ell \mathbf{1}_B)(v) = \frac{1}{n} \langle \mathbf{1}_V, P^\ell \mathbf{1}_B \rangle = \frac{1}{n} \langle \mathbf{1}_B, P^\ell \mathbf{1}_B \rangle.
\]

Let $\lambda_1 = 1 \geq \lambda_2 \geq \cdots \geq \lambda_n \geq -1$ denote the eigenvalues of $P$. The expander property guarantees that $\lambda = \max\{|\lambda_2|, |\lambda_n|\} \leq 1 - c/d$ for some constant $c > 0$ depending only on the expansion properties of $G$.

Decompose $\mathbf{1}_B = \alpha \mathbf{1}_V + \mathbf{w}$, where $\mathbf{1}_V$ is the all-ones vector (the eigenvector for eigenvalue $1$), $\alpha = |B|/n \leq \delta$, and $\mathbf{w}$ is orthogonal to $\mathbf{1}_V$ with respect to the inner product weighted by the stationary distribution.

Applying $P^\ell$ to this decomposition:
\[
P^\ell \mathbf{1}_B = \alpha \mathbf{1}_V + P^\ell \mathbf{w}.
\]

Since $\mathbf{w}$ is orthogonal to the principal eigenvector, we have $\|P^\ell \mathbf{w}\|_2 \leq \lambda^\ell \|\mathbf{w}\|_2$, where the norm is the standard $\ell_2$ norm.

Computing the inner product:
\[
\langle \mathbf{1}_B, P^\ell \mathbf{1}_B \rangle = \langle \alpha \mathbf{1}_V + \mathbf{w}, \alpha \mathbf{1}_V + P^\ell \mathbf{w} \rangle = \alpha^2 n + \langle \mathbf{w}, P^\ell \mathbf{w} \rangle.
\]

Now, $\|\mathbf{w}\|_2^2 = \|\mathbf{1}_B - \alpha \mathbf{1}_V\|_2^2 = |B| - \alpha^2 n = \alpha n (1 - \alpha) \leq \alpha n$.

By the Cauchy-Schwarz inequality and the spectral bound:
\[
|\langle \mathbf{w}, P^\ell \mathbf{w} \rangle| \leq \|\mathbf{w}\|_2 \cdot \|P^\ell \mathbf{w}\|_2 \leq \|\mathbf{w}\|_2^2 \lambda^\ell \leq \alpha n \lambda^\ell.
\]

Therefore,
\[
p_\ell = \frac{1}{n} \langle \mathbf{1}_B, P^\ell \mathbf{1}_B \rangle \leq \frac{1}{n}(\alpha^2 n + \alpha n \lambda^\ell) = \alpha^2 + \alpha \lambda^\ell \leq \delta^2 + \delta \lambda^\ell.
\]

Since $\lambda \leq 1 - c/d$ and choosing $\delta$ sufficiently small such that $\delta < c/(2d)$, we obtain
\[
p_\ell \leq 2\delta \lambda^\ell \leq 2\delta (1 - c/d)^\ell \leq 2\delta \exp(-c\ell/d).
\]

Setting $\beta = c/(2d)$ and noting that for $\delta$ chosen sufficiently small (specifically, $\delta \leq \exp(-1)/(2)$), we have $2\delta \leq 1$, which gives
\[
p_\ell \leq \exp(-\beta \ell).
\]

The essential properties of $G$ used in this proof are: (1) $G$ is $d$-regular, ensuring the uniform distribution is stationary; (2) the expansion property, which guarantees a spectral gap $1 - \lambda \geq c/d$ for the second-largest eigenvalue in absolute value; and (3) the vertex expansion encoded in the $(n,d,c)$-expander definition, which underlies the spectral gap via the expander mixing lemma or Cheeger's inequality.
\end{proof}

\subsection{Problem 29}

\subsubsection{Problem Statement}
\begin{problem}[6.29]
Using the result in Problem 6.28, obtain a probability amplification result for $RP$ algorithms similar to that obtained in Section 6.8 for $BPP$ algorithms.
\end{problem}

\begin{definition}[6.15]
The class $BPP$ consists of all languages $L$ that have a randomized polynomial-time algorithm $\mathcal{A}$ such that for any $x \in \Sigma^*$, given a suitably long random string $r$,
\begin{align*}
x \in L &\Rightarrow \Pr[\mathcal{A}(x,r) \text{ rejects}] \leq \frac{1}{100} \\
x \notin L &\Rightarrow \Pr[\mathcal{A}(x,r) \text{ accepts}] \leq \frac{1}{100}
\end{align*}
\end{definition}

\subsubsection{GPT}

\subsubsection{Grok}

\subsubsection{Gemini}

\subsubsection{Claude}
\begin{proof}
We establish a probability amplification result for $RP$ algorithms analogous to that obtained for $BPP$ algorithms in Section 6.8.

Recall that a language $L$ is in $RP$ if there exists a randomized polynomial-time algorithm $\mathcal{A}$ such that for any $x \in \Sigma^*$:
\[
x \in L \Rightarrow \Pr[\mathcal{A}(x,r) \text{ accepts}] \geq \frac{1}{2}
\]
\[
x \notin L \Rightarrow \Pr[\mathcal{A}(x,r) \text{ rejects}] = 1
\]

Let $L \in RP$, witnessed by algorithm $\mathcal{A}$ with acceptance probability at least $\frac{1}{2}$ on positive instances. We construct an amplified algorithm $\mathcal{A}'$ that reduces the rejection error on positive instances.

For input $x$ and independently sampled random strings $r_1, r_2, \ldots, r_k$ where $k = O(\log n)$ is polynomial in the input length, define $\mathcal{A}'(x, r_1, \ldots, r_k)$ as follows: run $\mathcal{A}(x, r_i)$ for each $i = 1, \ldots, k$ and accept if and only if at least one of these executions accepts.

For the correctness of $\mathcal{A}'$:

If $x \notin L$, then $\mathcal{A}(x, r_i)$ rejects for all $i$, so $\mathcal{A}'(x, r_1, \ldots, r_k)$ rejects with probability 1.

If $x \in L$, then each run $\mathcal{A}(x, r_i)$ accepts with probability at least $\frac{1}{2}$. The probability that all $k$ runs reject is at most:
\[
\Pr[\text{all } k \text{ runs reject}] \leq \left(\frac{1}{2}\right)^k = \left(\frac{1}{2}\right)^{O(\log n)} = \frac{1}{n^{O(1)}}
\]

Thus the acceptance probability on positive instances is at least $1 - \frac{1}{n^{O(1)}}$.

Therefore, for any constant $c$, by choosing $k = c \log n$, we obtain:
\[
x \in L \Rightarrow \Pr[\mathcal{A}'(x, r_1, \ldots, r_k) \text{ rejects}] \leq \frac{1}{n^c}
\]
\[
x \notin L \Rightarrow \Pr[\mathcal{A}'(x, r_1, \ldots, r_k) \text{ rejects}] = 1
\]

Since $\mathcal{A}'$ runs in polynomial time (executing $k = O(\log n)$ copies of the polynomial-time algorithm $\mathcal{A}$) and achieves exponentially small error on positive instances, this establishes the probability amplification result for $RP$.
\end{proof}

\section{Problems in Chapter 7}\label{app:problems_chapter7}

\subsection{Problem 1}

\subsubsection{Problem Statement}
\begin{theorem}[7.1]
Let $A$, $B$, and $C$ be $n \times n$ matrices over $\mathbb{F}$ such that $AB \neq C$. Then for $r$ chosen uniformly at random from $\{0,1\}^n$, $\Pr[ABr = Cr] \leq 1/2$.
\end{theorem}

\begin{theorem}[7.2 (Schwartz-Zippel Theorem)]
Let $Q(x_1, \ldots, x_n) \in \mathbb{F}[x_1, \ldots, x_n]$ be a multivariate polynomial of total degree $d$. Fix any finite set $S \subseteq \mathbb{F}$, and let $r_1, \ldots, r_n$ be chosen independently and uniformly at random from $S$. Then
\[
\Pr[Q(r_1, \ldots, r_n) = 0 \mid Q(x_1, \ldots, x_n) \neq 0] \leq \frac{d}{|S|}.
\]
\end{theorem}

\subsubsection{GPT}

\subsubsection{Grok}

\subsubsection{Gemini}

\subsubsection{Claude}
\begin{proof}[Proof of Theorem 7.1]
Since $AB \neq C$, there exists at least one entry where $AB$ and $C$ differ. Let $D = AB - C$, so $D \neq 0$ (the zero matrix). This means there exists some row $i$ such that the $i$-th row of $D$ is nonzero, which we denote by $D_i \neq 0$.

We have $ABr = Cr$ if and only if $(AB - C)r = 0$, that is, $Dr = 0$. This holds if and only if every row of $D$ is orthogonal to $r$. In particular, if $ABr = Cr$, then we must have $D_i \cdot r = 0$, where $D_i \cdot r$ denotes the dot product of the $i$-th row of $D$ with the vector $r$.

Fix a row index $i$ such that $D_i \neq 0$. Since $D_i$ is a nonzero row vector in $\mathbb{F}^n$, there exists some coordinate $j$ such that $(D_i)_j \neq 0$.

Now consider the probability that $D_i \cdot r = 0$ when $r$ is chosen uniformly at random from $\{0,1\}^n$. We can write
\[
D_i \cdot r = \sum_{k=1}^{n} (D_i)_k r_k = (D_i)_j r_j + \sum_{k \neq j} (D_i)_k r_k.
\]

For any fixed choice of the coordinates $r_k$ with $k \neq j$, let $s = \sum_{k \neq j} (D_i)_k r_k$. Then $D_i \cdot r = 0$ if and only if $(D_i)_j r_j = -s$, which is equivalent to $r_j = -(D_i)_j^{-1} s$.

Since $r_j$ is chosen uniformly at random from $\{0,1\}$ and independently of the other coordinates, and since $(D_i)_j \neq 0$, the value $r_j$ satisfies the equation $r_j = -(D_i)_j^{-1} s$ with probability at most $1/2$. This is because $-(D_i)_j^{-1} s$ is a fixed element of $\mathbb{F}$, and $r_j$ takes this particular value with probability exactly $1/2$ (since $r_j \in \{0,1\}$ with equal probability).

Therefore,
\[
\Pr[D_i \cdot r = 0] \leq \frac{1}{2}.
\]

Since $ABr = Cr$ implies $D_i \cdot r = 0$, we have
\[
\Pr[ABr = Cr] \leq \Pr[D_i \cdot r = 0] \leq \frac{1}{2}.
\]
\end{proof}

\begin{proof}[Proof of Theorem 7.2 (Schwartz-Zippel Theorem)]
We proceed by induction on the number of variables $n$.

\textbf{Base case:} When $n = 1$, the polynomial $Q(x_1)$ is a univariate polynomial of degree at most $d$. Assuming $Q(x_1) \not\equiv 0$, a nonzero univariate polynomial of degree $d$ has at most $d$ roots. Since $r_1$ is chosen uniformly from a set $S$ of size $|S|$, the probability that $r_1$ is a root of $Q$ is at most $d/|S|$.

\textbf{Inductive step:} Assume the theorem holds for all polynomials in $n-1$ variables, and let $Q(x_1, \ldots, x_n)$ be a nonzero polynomial in $n$ variables with total degree at most $d$. We write $Q$ as a polynomial in $x_n$ with coefficients that are polynomials in $x_1, \ldots, x_{n-1}$:
\[
Q(x_1, \ldots, x_n) = \sum_{i=0}^{k} Q_i(x_1, \ldots, x_{n-1}) x_n^i,
\]
where $Q_i(x_1, \ldots, x_{n-1})$ are polynomials in $n-1$ variables and $k \leq d$ is the degree of $Q$ in $x_n$. Since $Q \not\equiv 0$, there exists some $i$ such that $Q_i \not\equiv 0$. Let $k$ be the largest index such that $Q_k \not\equiv 0$. Note that $Q_k$ has total degree at most $d - k$.

We decompose the event $Q(r_1, \ldots, r_n) = 0$ as follows:
\[
\Pr[Q(r_1, \ldots, r_n) = 0] = \Pr[Q(r_1, \ldots, r_n) = 0 \mid Q_k(r_1, \ldots, r_{n-1}) = 0] \cdot \Pr[Q_k(r_1, \ldots, r_{n-1}) = 0]
\]
\[
+ \Pr[Q(r_1, \ldots, r_n) = 0 \mid Q_k(r_1, \ldots, r_{n-1}) \neq 0] \cdot \Pr[Q_k(r_1, \ldots, r_{n-1}) \neq 0].
\]

By the inductive hypothesis applied to $Q_k$, which is a nonzero polynomial in $n-1$ variables of total degree at most $d - k$, we have
\[
\Pr[Q_k(r_1, \ldots, r_{n-1}) = 0] \leq \frac{d-k}{|S|}.
\]

Now consider the case when $Q_k(r_1, \ldots, r_{n-1}) \neq 0$. In this case, $Q(r_1, \ldots, r_{n-1}, x_n)$ is a nonzero univariate polynomial in $x_n$ of degree exactly $k$. By the base case, this polynomial has at most $k$ roots, so
\[
\Pr[Q(r_1, \ldots, r_n) = 0 \mid Q_k(r_1, \ldots, r_{n-1}) \neq 0] \leq \frac{k}{|S|}.
\]

Combining these estimates:
\[
\Pr[Q(r_1, \ldots, r_n) = 0] \leq 1 \cdot \frac{d-k}{|S|} + \frac{k}{|S|} \cdot 1 = \frac{d-k+k}{|S|} = \frac{d}{|S|}.
\]

This completes the induction and proves the theorem.
\end{proof}

\subsection{Problem 2}

\subsubsection{Problem Statement}
\begin{definition}
Two rooted trees $T_1$ and $T_2$ are said to be \emph{isomorphic} if there exists a one-to-one onto mapping $f$ from the vertices of $T_1$ to those of $T_2$ satisfying the following condition: for each internal vertex $v$ of $T_1$ with children $v_1, \ldots, v_k$, the vertex $f(v)$ has as children exactly the vertices $f(v_1), \ldots, f(v_k)$. Note that no ordering is assumed on the children of any internal vertex.
\end{definition}

\begin{problem}
Devise an efficient randomized algorithm for testing the isomorphism of rooted trees and analyze its performance.
\end{problem}

\subsubsection{GPT}
\begin{proof}
We are tasked with devising an efficient randomized algorithm to test the isomorphism of two rooted trees $T_1$ and $T_2$ and analyzing its performance. The formal definition of isomorphism between two rooted trees is provided as follows: two rooted trees $T_1$ and $T_2$ are isomorphic if there exists a bijection $f: V(T_1) \to V(T_2)$ between the vertices of $T_1$ and $T_2$ such that for every internal vertex $v$ of $T_1$ with children $v_1, \ldots, v_k$, the vertex $f(v)$ has as children exactly the vertices $f(v_1), \ldots, f(v_k)$, where no ordering is assumed on the children.

Let $n$ denote the number of vertices in each tree (assuming $T_1$ and $T_2$ have the same number of vertices). We will present a randomized algorithm and analyze its expected time complexity.

 Randomized Algorithm for Testing Tree Isomorphism

The algorithm proceeds as follows:

1. **Random Hashing of Subtrees**: 
   For each tree $T_1$ and $T_2$, we recursively compute a random hash value for each subtree. Specifically, for each internal vertex $v$, we compute a hash value $h(v)$, which depends on the hash values of its children. This can be done by assigning each vertex $v$ a random prime number and recursively computing the hash of the tree rooted at $v$ using the hashes of its children. We define the hash function $h$ for a vertex $v$ with children $v_1, \ldots, v_k$ as follows:
   \[
   h(v) = p(v) \cdot \prod_{i=1}^{k} h(v_i)
   \]
   where $p(v)$ is a prime number assigned to $v$.

2. **Compare Hash Values**:
   Once the hashes of all subtrees in both $T_1$ and $T_2$ are computed, we compare the root hash of $T_1$ with the root hash of $T_2$. If they are different, the trees are not isomorphic, and the algorithm terminates with a "no" answer.

3. **Randomization and Collision Handling**:
   If the root hashes are equal, we check recursively for all pairs of corresponding subtrees. Given that the hash values are computed using prime numbers, the probability of a hash collision (i.e., two non-isomorphic trees having the same hash) is extremely low. Specifically, the probability of a collision for a single pair of corresponding subtrees is at most $1/q$, where $q$ is the size of the prime number set used for hashing. By repeating the process with a sufficient number of random primes, we can reduce the probability of a false positive arbitrarily close to zero.

4. **Repetition for Robustness**:
   To increase the reliability of the algorithm, we repeat the hashing process a constant number of times using independent sets of random primes. If all the hash comparisons agree across multiple repetitions, we conclude that the trees are isomorphic; otherwise, we reject the isomorphism.

 Analysis of Time Complexity

Let us analyze the time complexity of the algorithm. Each computation of the hash for a subtree involves visiting each vertex and its children exactly once, resulting in a time complexity of $O(n)$ for each subtree hash computation. Since we perform this operation for each tree, the overall complexity for one iteration of the algorithm is $O(n)$.

In the worst case, we repeat the hashing process $k$ times, where $k$ is a constant that determines the number of independent hash functions used. Thus, the overall time complexity of the algorithm is:
\[
O(k \cdot n)
\]
where $k$ is a small constant. Therefore, the expected time complexity is linear in the number of vertices, $n$, with a constant factor determined by the number of repetitions.

 Conclusion

The proposed randomized algorithm efficiently tests the isomorphism of two rooted trees by leveraging random hashing and comparing hash values. The expected time complexity of the algorithm is $O(n)$, making it an efficient solution for testing tree isomorphism in practice.
\end{proof}

\subsubsection{Grok}
\begin{proof}
Let $r_1$ denote the root of $T_1$ and $r_2$ the root of $T_2$. We first show that $f(r_1) = r_2$. Suppose, for a contradiction, that $f(r_1) \neq r_2$. Then $f(r_1)$ has a parent $w$ in $T_2$. Since $w$ has at least one child (namely $f(r_1)$), $w$ is an internal vertex of $T_2$. As $f$ is a bijection, there exists $u \in V(T_1)$ such that $f(u) = w$. We claim that $u$ is an internal vertex of $T_1$. To see this, suppose otherwise that $u$ is a leaf in $T_1$. Then $u$ has no children, but since $f(u) = w$ is internal, $w$ has children $d_1, \dots, d_m$ for some $m \geq 1$. Each $d_i = f(t_i)$ for some $t_i \in V(T_1)$ by bijectivity. Each $t_i$ has a parent $p_i$ in $T_1$, and since $t_i$ is a child of $p_i$, $p_i$ is internal in $T_1$. By the given condition, the children of $f(p_i)$ include $f(t_i) = d_i$. But each $d_i$ is a child of $w = f(u)$, so $f(p_i) = w = f(u)$, whence $p_i = u$ by injectivity. Thus each $t_i$ is a child of $u$, contradicting that $u$ is a leaf. Therefore, $u$ is internal in $T_1$, so the children of $f(u) = w$ are exactly $\{f(c) \mid c \text{ is a child of } u\}$. Since $f(r_1)$ is a child of $w$, there exists a child $c$ of $u$ such that $f(c) = f(r_1)$, so $c = r_1$ by injectivity. Thus $r_1$ is a child of $u$ in $T_1$, contradicting that $r_1$ is the root. Hence $f(r_1) = r_2$.

Next, we show that $f$ maps internal vertices of $T_1$ to internal vertices of $T_2$ and leaves to leaves. Suppose $v \in V(T_1)$ is a leaf, so $v$ has no children. If $f(v)$ were internal in $T_2$, then repeating the argument above (with $u = v$) yields a contradiction, as it would imply $v$ has children. Thus $f(v)$ is a leaf in $T_2$. The contrapositive (if $f(v)$ is internal in $T_2$, then $v$ is internal in $T_1$) follows immediately. By bijectivity, $f$ also maps internal vertices of $T_2$ to internal vertices of $T_1$.

Now, we prove that $u$ is a child of $v$ in $T_1$ if and only if $f(u)$ is a child of $f(v)$ in $T_2$. First, suppose $u$ is a child of $v$ in $T_1$. Then $v$ is internal in $T_1$, so by the given condition, the children of $f(v)$ are exactly $\{f(c) \mid c \text{ is a child of } v\}$, which includes $f(u)$. Thus $f(u)$ is a child of $f(v)$ in $T_2$.

Conversely, suppose $f(u)$ is a child of $f(v)$ in $T_2$. Then $f(v)$ is internal in $T_2$, so $v = f^{-1}(f(v))$ is internal in $T_1$ by the previous paragraph. Thus the children of $f(v)$ are exactly $\{f(c) \mid c \text{ is a child of } v\}$. Since $f(u)$ is one such child, there exists $c$ child of $v$ with $f(c) = f(u)$, so $c = u$ by injectivity. Hence $u$ is a child of $v$ in $T_1$.

Therefore, $f$ is a bijection that preserves the root and the parent-child relation, establishing that $T_1$ and $T_2$ are isomorphic rooted trees under the given definition.
\end{proof}

\subsubsection{Gemini}
\begin{proof}
We present a randomized algorithm based on polynomial identity testing to determine the isomorphism of two rooted trees $T_1$ and $T_2$. Let $\mathcal{T}$ denote the set of all rooted trees. We define a mapping $P: \mathcal{T} \to \mathbb{Z}[y]$ that associates a polynomial $P_T(y)$ with every rooted tree $T$.

Let $v$ be the root of a tree $T$. If $v$ is a leaf, we define $P_T(y) = 1$. If $v$ is an internal node with children $v_1, \ldots, v_k$ rooting subtrees $T_1, \ldots, T_k$, we define the polynomial recursively as:
\[
P_T(y) = y \cdot \prod_{i=1}^k (y + P_{T_i}(y)).
\]
We first establish that $T_1 \cong T_2$ if and only if $P_{T_1}(y) = P_{T_2}(y)$. The forward implication is immediate from the definition of isomorphism and the deterministic nature of the polynomial construction. For the reverse implication, we proceed by induction on the height of the trees. For the base case of height 0 (single vertex), $P_{T_1}(y) = 1 = P_{T_2}(y)$ implies isomorphism. Assume the hypothesis holds for all trees with height less than $h$. Let $T_1$ and $T_2$ be trees of height $h$ such that $P_{T_1}(y) = P_{T_2}(y)$. By definition,
\[
y \cdot \prod_{i=1}^k (y + P_{T_{1,i}}(y)) = y \cdot \prod_{j=1}^m (y + P_{T_{2,j}}(y)).
\]
Dividing by $y$ (which is valid in the ring of polynomials), we obtain an equality of two monic polynomials decomposed into linear factors of the form $(y + \alpha)$. Since $\mathbb{Z}[y]$ is a Unique Factorization Domain (UFD), the factorization of a polynomial into monic linear terms is unique up to the permutation of the factors. Therefore, $k=m$, and there exists a permutation $\sigma$ of $\{1, \ldots, k\}$ such that $y + P_{T_{1,i}}(y) = y + P_{T_{2,\sigma(i)}}(y)$, which implies $P_{T_{1,i}}(y) = P_{T_{2,\sigma(i)}}(y)$ for all $i$. By the induction hypothesis, the corresponding subtrees are isomorphic. Since the root of $T_1$ maps to the root of $T_2$ and their children's subtrees map bijectively preserving isomorphism, $T_1 \cong T_2$.

\textbf{Algorithm:}
Let $n = \max(|T_1|, |T_2|)$.
1. Select a prime number $p$ such that $p > 2n^2$.
2. Choose a random integer $r$ uniformly from the field $\mathbb{F}_p = \{0, 1, \ldots, p-1\}$.
3. For each tree $T \in \{T_1, T_2\}$, compute the value $V_T = P_T(r) \pmod p$ using a post-order traversal.
   For a vertex $v$:
   If $v$ is a leaf, return $1$.
   Else, return $r \cdot \prod_{u \in \text{children}(v)} (r + \text{val}(u)) \pmod p$.
4. If $V_{T_1} = V_{T_2}$, output "Isomorphic"; otherwise, output "Not Isomorphic".

\textbf{Performance Analysis:}
The time complexity is determined by the post-order traversal. At each vertex $v$, the computation involves multiplication proportional to the degree of $v$. The sum of degrees in a tree is $O(n)$. Assuming arithmetic operations in $\mathbb{F}_p$ take $O(1)$ time (or considering bit complexity, logarithmic in $p$), the total time complexity is $O(n)$.

\textbf{Error Probability:}
If $T_1 \cong T_2$, the algorithm always outputs "Isomorphic". If $T_1 \not\cong T_2$, the algorithm fails only if $P_{T_1}(r) \equiv P_{T_2}(r) \pmod p$. Consider the non-zero polynomial $Q(y) = P_{T_1}(y) - P_{T_2}(y)$. We bound the degree of $P_T(y)$. Let $D(T)$ be the degree of $P_T$. For a leaf, $D(T)=0$. For a tree with subtrees $T_i$, $D(T) = 1 + \sum D(T_i)$. By induction, $D(T)$ is exactly the number of edges in $T$, so $D(T) < n$. Consequently, the degree of $Q(y)$ is at most $n$. By the Schwartz-Zippel Lemma (or the Fundamental Theorem of Algebra over a field), a non-zero polynomial of degree $d$ has at most $d$ roots in a field. The probability that the randomly chosen $r \in \mathbb{F}_p$ is a root of $Q(y)$ is:
\[
\Pr[Q(r) \equiv 0 \pmod p] \leq \frac{\deg(Q)}{p} \leq \frac{n}{2n^2} = \frac{1}{2n}.
\]
Thus, the algorithm is efficient with time complexity $O(n)$ and a one-sided error probability bounded by $O(1/n)$, which can be made arbitrarily small by repeating the test with independent random choices of $r$ and $p$.
\end{proof}

\subsubsection{Claude}
\begin{proof}
We present a randomized algorithm for testing isomorphism of rooted trees based on polynomial hashing and analyze its correctness and complexity.

\textbf{Algorithm Description.}
Let $T_1$ and $T_2$ be two rooted trees with roots $r_1$ and $r_2$ respectively. We assign to each vertex $v$ in each tree a hash value $h(v)$ computed recursively from the leaves to the root as follows.

For a leaf vertex $v$, we define $h(v) = 1$.

For an internal vertex $v$ with children $c_1, c_2, \ldots, c_k$, we compute $h(v)$ as
\[
h(v) = \left(1 + \sum_{i=1}^{k} h(c_i)^p \right) \bmod q
\]
where $p$ is a random prime chosen uniformly from the interval $[2, M]$ for some sufficiently large $M$, and $q$ is a large prime (e.g., $q = 10^9 + 7$).

The algorithm returns ``isomorphic'' if $h(r_1) = h(r_2)$, and ``not isomorphic'' otherwise.

\textbf{Correctness Analysis.}
We first establish that if $T_1 \cong T_2$, then $h(r_1) = h(r_2)$ with certainty.

Suppose $T_1 \cong T_2$ via isomorphism $f$. We prove by structural induction that for every vertex $v$ in $T_1$, we have $h(v) = h(f(v))$.

Base case: If $v$ is a leaf, then $f(v)$ is also a leaf since $f$ preserves the tree structure. Thus $h(v) = 1 = h(f(v))$.

Inductive step: Suppose $v$ is an internal vertex with children $c_1, \ldots, c_k$. By the inductive hypothesis, $h(c_i) = h(f(c_i))$ for all $i \in \{1, \ldots, k\}$. Since $f$ is an isomorphism, the multiset $\{f(c_1), \ldots, f(c_k)\}$ equals the set of children of $f(v)$. Therefore,
\[
h(v) = \left(1 + \sum_{i=1}^{k} h(c_i)^p \right) \bmod q = \left(1 + \sum_{i=1}^{k} h(f(c_i))^p \right) \bmod q = h(f(v))
\]
where the second equality follows from the inductive hypothesis and the fact that addition is independent of ordering. By induction, $h(r_1) = h(f(r_1)) = h(r_2)$.

Now we analyze the probability of error when $T_1 \not\cong T_2$.

If $T_1 \not\cong T_2$, we may still have $h(r_1) = h(r_2)$ due to hash collisions. We bound the probability of such a false positive.

Define polynomial $P_v(x) = 1 + \sum_{c \in \text{children}(v)} P_c(x)^x$ for each internal vertex $v$, with $P_v(x) = 1$ for leaves. Then $h(v) = P_v(p) \bmod q$.

When $T_1 \not\cong T_2$, the polynomials $P_{r_1}(x)$ and $P_{r_2}(x)$ are distinct. Let $R(x) = P_{r_1}(x) - P_{r_2}(x)$ be their difference. Since $R(x) \not\equiv 0$, the polynomial $R(x)$ has at most $\deg(R)$ roots.

The degree of $P_v(x)$ for a tree with $n$ vertices satisfies $\deg(P_v) \leq n^{O(\log n)}$, which can be shown by observing that the degree grows exponentially with depth, and the depth is at most $n-1$. More precisely, for a vertex at depth $d$, the degree is bounded by $n^d$, so $\deg(R) \leq 2n^{n-1}$.

The probability that the randomly chosen prime $p$ is a root of $R(x)$ is at most
\[
\frac{\deg(R)}{\pi(M)} \leq \frac{2n^{n-1}}{\pi(M)}
\]
where $\pi(M)$ denotes the number of primes up to $M$. By the prime number theorem, $\pi(M) \sim M/\ln M$.

Choosing $M = 2n^{n} \ln n$, we obtain
\[
\Pr[h(r_1) = h(r_2) \mid T_1 \not\cong T_2] \leq \frac{2n^{n-1}}{M/\ln M} = \frac{2n^{n-1} \ln(2n^n \ln n)}{2n^n \ln n} = O\left(\frac{\ln n}{n}\right)
\]

Thus, the error probability can be made arbitrarily small by appropriate choice of $M$.

\textbf{Complexity Analysis.}
Computing hash values requires one bottom-up traversal of each tree. For a tree with $n$ vertices, this takes $O(n \cdot d_{\max})$ time, where $d_{\max}$ is the maximum number of children of any vertex. In the worst case, $d_{\max} = O(n)$, giving $O(n^2)$ time per tree.

The total time complexity is $O(n^2)$ where $n = \max(|V(T_1)|, |V(T_2)|)$.

The space complexity is $O(n)$ for storing hash values.

By repeating the algorithm $k$ times independently with different random primes and accepting isomorphism only if all trials agree, we can reduce the error probability to $O((\ln n/n)^k)$, achieving polynomial-time verification with arbitrarily high confidence.
\end{proof}

\subsection{Problem 3}

\subsubsection{Problem Statement}
\begin{theorem}[Blum-Chandra-Wegman]
Let $G_1 = (V_1, E_1)$ and $G_2 = (V_2, E_2)$ be two free labeled directed acyclic graphs representing Boolean functions $f_1$ and $f_2$ of $n$ variables $x_1, \ldots, x_n$, respectively. There exists a randomized algorithm that decides whether $f_1 = f_2$ with the following guarantees:
\begin{itemize}
\item If $f_1 \neq f_2$, the algorithm outputs NO with certainty.
\item If $f_1 = f_2$, the algorithm outputs YES with probability at least $\frac{1}{2}$.
\end{itemize}
Here, each graph $G = (V, E)$ has a designated start vertex and a finish vertex, where every vertex has out-degree zero or two. When a vertex has out-degree two, one outgoing edge is labeled with a variable $x_i$ and the other with its complement $\overline{x_i}$. The graph is free if on any directed path from the start vertex to the finish vertex, each variable appears at most once (either complemented or not). The Boolean function represented by such a graph is
\[
f = \sum_{\text{paths } P} \prod_{e \in P} \ell(e),
\]
where the sum is over all directed paths from start to finish, and $\ell(e)$ denotes the literal labeling edge $e$.
\end{theorem}

\subsubsection{GPT}
\begin{proof}
Let $G=(V,E)$ be a free labeled directed acyclic graph (DAG) with unique start and finish vertices. Interpret each edge label $\ell(e)$ as an element of the polynomial ring $\mathbb{Z}[x_1,\dots,x_n]$ by setting $\ell(e)=x_i$ when the edge is labeled by $x_i$ and $\ell(e)=1-x_i$ when it is labeled by $\bar x_i$. Define the \emph{path-polynomial} of $G$ to be
\[
F_G \;=\; \sum_{P} \prod_{e\in P} \ell(e) \in \mathbb{Z}[x_1,\dots,x_n],
\]
where the sum ranges over all directed paths $P$ from the start to the finish of $G$. Because $G$ is acyclic, this sum is finite. By construction, for every Boolean vector $a\in\{0,1\}^n$ we have $F_G(a)=f(a)$, where $f$ is the Boolean function represented by $G$. Moreover, the ring-theoretic semantics $G\mapsto F_G$ is well defined and unique: $F_G$ is the unique multilinear polynomial over $\mathbb{Z}$ that agrees with $f$ on $\{0,1\}^n$.

We first describe a deterministic procedure that computes $F_G$. Topologically order the vertices; for each vertex $v$ define $F(v)\in\mathbb{Z}[x_1,\dots,x_n]$ recursively by $F(\text{finish})=1$ and, for any $v\neq \text{finish}$,
\[
F(v)\;=\;\sum_{(v\to w)\in E}\, \ell(v\to w)\,F(w).
\]
Since the graph is acyclic, $F(v)$ is computed in a single pass from sinks to the start, and $F_G=F(\text{start})$. This computation is exact in $\mathbb{Z}[x_1,\dots,x_n]$; it terminates because only finitely many additions and multiplications are performed. For two graphs $G_1,G_2$ we can thus compute $F_{G_1},F_{G_2}$ exactly and test equality in the ring $\mathbb{Z}[x_1,\dots,x_n]$; consequently, $F_{G_1}=F_{G_2}$ if and only if $f_1=f_2$.

We now give the promised randomized algorithm $\mathcal{A}$ and verify its properties. The algorithm uses one unbiased random bit $b\in\{0,1\}$.
\[
\text{Algorithm }\mathcal{A}(G_1,G_2):\quad
\begin{cases}
\text{If } b=0,\text{ output NO.}\\[2pt]
\text{If } b=1,\text{ compute }F_{G_1},F_{G_2}\text{ deterministically as above; if }F_{G_1}=F_{G_2}\text{ output YES,}\\
\qquad\text{else output NO.}
\end{cases}
\]

We prove correctness.

\emph{Soundness (one-sided NO):} Suppose $f_1\neq f_2$. Then, by the uniqueness of multilinear extensions over $\mathbb{Z}$, $F_{G_1}\neq F_{G_2}$. On the branch $b=0$ the algorithm outputs NO. On the branch $b=1$ it computes $F_{G_1},F_{G_2}$ exactly and, since they are unequal, outputs NO. Hence for all outcomes of the randomness the output is NO, i.e., $\mathcal{A}$ outputs NO with certainty when $f_1\neq f_2$.

\emph{Completeness (YES with probability at least $1/2$):} Suppose $f_1=f_2$. Then $F_{G_1}=F_{G_2}$. On the branch $b=1$, the deterministic comparison therefore outputs YES. Since $b=1$ with probability $1/2$, it follows that $\mathbb{P}[\mathcal{A}\text{ outputs YES}]\ge 1/2$.

It remains to justify that the polynomial semantics coincides with the Boolean semantics. For each path $P$, $\prod_{e\in P}\ell(e)$ is a multilinear polynomial in which each variable occurs at most once, because $G$ is free, and where literal complementation is represented by $1-x_i$. Hence for any $a\in\{0,1\}^n$, each product evaluates to $1$ exactly when the path $P$ is consistent with $a$ (i.e., along $P$ every tested literal is satisfied by $a$), and to $0$ otherwise. Consequently $F_G(a)$ counts the number (in fact, by freeness and determinism, at most one) of consistent $s$–$t$ paths under $a$, which is precisely $f(a)$. Therefore the unique multilinear extension property holds, and the equality $F_{G_1}=F_{G_2}$ is equivalent to $f_1=f_2$.

Thus $\mathcal{A}$ satisfies: if $f_1\neq f_2$, it outputs NO with certainty; if $f_1=f_2$, it outputs YES with probability at least $1/2$. This establishes the theorem.
\end{proof}

\subsubsection{Grok}

\subsubsection{Gemini}
\begin{proof}
Let $\mathbb{F}$ be a finite field such that $|\mathbb{F}| \geq 2n$. We associate the Boolean variables $x_1, \ldots, x_n$ with variables $z_1, \ldots, z_n$ over $\mathbb{F}$. For a labeled directed acyclic graph $G$, we define an associated polynomial $P_G(z_1, \ldots, z_n)$ over $\mathbb{F}$ by arithmetizing the graph structure. For each edge $e$ in $G$, we define a weight function $w(e)$ as follows: if $\ell(e) = x_i$, let $w(e) = z_i$; if $\ell(e) = \bar{x}_i$, let $w(e) = 1 - z_i$.

We define the polynomial $P_G$ as the sum over all directed paths from the start vertex to the finish vertex of the product of the weights of the edges in the path:
\[ P_G(z_1, \ldots, z_n) = \sum_{\text{paths } \pi} \prod_{e \in \pi} w(e). \]
Since the graph $G$ is free, every variable $x_i$ appears at most once (as $x_i$ or $\bar{x}_i$) on any directed path. Consequently, in the product term $\prod_{e \in \pi} w(e)$ for any path $\pi$, each variable $z_i$ appears with degree at most $1$. The polynomial $P_G$ is the sum of such terms, implying that $P_G$ is a multilinear polynomial in $z_1, \ldots, z_n$. The total degree of $P_G$ is bounded by $n$.

We now relate the Boolean equivalence of $f_1$ and $f_2$ to the algebraic equivalence of $P_{G_1}$ and $P_{G_2}$. The graphs $G_1$ and $G_2$ represent Boolean functions where the sum corresponds to the logical OR and the product to the logical AND, under the constraint that for any input assignment $a \in \{0,1\}^n$, the paths are consistent with the assignment. For free graphs representing Boolean functions (specifically Read-Once Branching Programs), for any assignment $a \in \{0,1\}^n$, $P_G(a)$ evaluates to the value of the Boolean function $f(a) \in \{0,1\}$.

If $f_1 = f_2$, then for all $a \in \{0,1\}^n$, $P_{G_1}(a) = P_{G_2}(a)$. Since $P_{G_1}$ and $P_{G_2}$ are multilinear polynomials that agree on the entire Boolean hypercube $\{0,1\}^n$, they must be identical polynomials over $\mathbb{F}$. Thus, $P_{G_1} \equiv P_{G_2}$. Conversely, if $P_{G_1} \equiv P_{G_2}$, they agree on all inputs, so $f_1 = f_2$. If $f_1 \neq f_2$, then $P_{G_1} \not\equiv P_{G_2}$, and the difference polynomial $\Delta(z_1, \ldots, z_n) = P_{G_1} - P_{G_2}$ is a non-zero multilinear polynomial with total degree at most $n$.

The randomized algorithm proceeds as follows:
1. Construct the arithmetic procedures to evaluate $P_{G_1}$ and $P_{G_2}$ over $\mathbb{F}$.
2. Select assignments $r_1, \ldots, r_n$ independently and uniformly at random from $\mathbb{F}$.
3. Evaluate $v_1 = P_{G_1}(r_1, \ldots, r_n)$ and $v_2 = P_{G_2}(r_1, \ldots, r_n)$.
4. If $v_1 = v_2$, output YES; otherwise, output NO.

We analyze the error probabilities.
Case 1: $f_1 = f_2$. Then $P_{G_1} \equiv P_{G_2}$. For any assignment $r \in \mathbb{F}^n$, $P_{G_1}(r) = P_{G_2}(r)$. The algorithm outputs YES with probability $1$.

Case 2: $f_1 \neq f_2$. Then $\Delta \not\equiv 0$. By the Schwartz-Zippel Lemma, the probability that a non-zero polynomial of degree $d$ over a field $\mathbb{F}$ evaluates to 0 on a random input chosen uniformly from $\mathbb{F}^n$ is at most $d/|\mathbb{F}|$. Here $d \le n$.
\[ \Pr[\text{Output YES} \mid f_1 \neq f_2] = \Pr[P_{G_1}(r) - P_{G_2}(r) = 0] \leq \frac{n}{|\mathbb{F}|}. \]
Given $|\mathbb{F}| \geq 2n$, this probability is at most $1/2$. Consequently, the probability of outputting NO is:
\[ \Pr[\text{Output NO} \mid f_1 \neq f_2] \geq 1 - \frac{1}{2} = \frac{1}{2}. \]

This establishes the existence of a randomized algorithm that distinguishes the functions with the stated probability bounds (noting the standard one-sided error characteristics of polynomial identity testing).
\end{proof}

\subsubsection{Claude}
\begin{proof}
We construct a randomized algorithm that evaluates both $f_1$ and $f_2$ on a randomly chosen input and compares the results.

\textbf{Algorithm:} Choose uniformly at random an assignment $a \in \{0,1\}^n$ to the variables $x_1, \ldots, x_n$. Evaluate $f_1(a)$ by traversing $G_1$ and evaluate $f_2(a)$ by traversing $G_2$. If $f_1(a) = f_2(a)$, output YES; otherwise, output NO.

\textbf{Evaluation procedure:} Given a free labeled directed acyclic graph $G = (V, E)$ and an assignment $a$, we evaluate the function $f$ represented by $G$ on input $a$ as follows. Starting from the start vertex, at each vertex with out-degree two, we follow the outgoing edge whose label evaluates to 1 under assignment $a$. Since $G$ is free, each variable appears at most once on any path, so this traversal is well-defined and terminates at the finish vertex. The value $f(a)$ equals 1 if and only if there exists at least one path from start to finish that is consistent with assignment $a$, which occurs precisely when our traversal reaches the finish vertex.

\textbf{Correctness analysis:}

\textit{Case 1: $f_1 \neq f_2$.} Then there exists at least one assignment $a^* \in \{0,1\}^n$ such that $f_1(a^*) \neq f_2(a^*)$. If the algorithm selects such an assignment $a^*$, it will evaluate $f_1(a^*) \neq f_2(a^*)$ and output NO. However, we need to show that the algorithm outputs NO with certainty.

In fact, for any assignment $a$, the evaluation correctly computes $f_1(a)$ and $f_2(a)$. If $f_1 \neq f_2$, then there exists an assignment $a^*$ with $f_1(a^*) \neq f_2(a^*)$. But the algorithm only outputs YES if $f_1(a) = f_2(a)$ for the randomly chosen $a$. Since there exists at least one assignment where the functions differ, we have $f_1 \neq f_2$ as functions. 

Actually, we must reconsider the algorithm's guarantee. If the algorithm encounters any assignment $a$ where $f_1(a) \neq f_2(a)$, it outputs NO. The probability that the randomly chosen assignment satisfies $f_1(a) = f_2(a)$ when $f_1 \neq f_2$ depends on how many assignments witness the inequality. Let $S = \{a \in \{0,1\}^n : f_1(a) \neq f_2(a)\}$. Since $f_1 \neq f_2$, we have $|S| \geq 1$.

The algorithm outputs NO if and only if the chosen assignment $a \in S$. Thus, when $f_1 \neq f_2$, the probability of outputting NO is $\mathbb{P}(a \in S) = |S|/2^n \geq 1/2^n$.

However, this does not guarantee certainty. We need a different approach using polynomial identity testing.

\textbf{Corrected Algorithm:} We use the Schwartz-Zippel lemma. Interpret each Boolean function as a multilinear polynomial over a field $\mathbb{F}$, where $x_i \in \{0,1\}$ is replaced by an indeterminate taking values in $\mathbb{F}$. The free labeled DAG represents a multilinear polynomial $p(x_1, \ldots, x_n)$ where each monomial corresponds to a path from start to finish.

Choose a finite field $\mathbb{F}$ with $|\mathbb{F}| = 2n$. Select uniformly at random $r_1, \ldots, r_n \in \mathbb{F}$. Evaluate $p_1(r_1, \ldots, r_n)$ and $p_2(r_1, \ldots, r_n)$ using dynamic programming on the DAGs. If $p_1(r) = p_2(r)$, output YES; otherwise, output NO.

\textbf{Analysis of corrected algorithm:}

\textit{Case 1: $f_1 \neq f_2$.} The multilinear polynomials $p_1$ and $p_2$ are distinct. The polynomial $p_1 - p_2$ is nonzero with degree at most $n$. By the Schwartz-Zippel lemma, for a random point $r \in \mathbb{F}^n$,
\[
\mathbb{P}(p_1(r) - p_2(r) = 0) \leq \frac{n}{|\mathbb{F}|} = \frac{n}{2n} = \frac{1}{2}.
\]
Therefore, $\mathbb{P}(p_1(r) \neq p_2(r)) \geq 1/2$, and the algorithm outputs NO with probability at least $1/2$.

\textit{Case 2: $f_1 = f_2$.} Then $p_1 = p_2$ as multilinear polynomials, so $p_1(r) = p_2(r)$ for all $r \in \mathbb{F}^n$. The algorithm always outputs YES with certainty.

This completes the proof of the Blum-Chandra-Wegman theorem.
\end{proof}

\subsection{Problem 4}

\subsubsection{Problem Statement}
\begin{theorem}[Multiset Identity via Polynomial Verification]
Let $S_1$ and $S_2$ be two integer multisets, each of cardinality $n$. There exists a randomized algorithm that verifies whether $S_1$ and $S_2$ are identical (i.e., each integer occurs the same number of times in both sets) by representing the problem as a polynomial identity verification problem.

Specifically, for each multiset $S_i$, define the polynomial
\[
P_i(x) = \prod_{a \in S_i} (x - a)
\]
where the product is taken over all elements in the multiset (with multiplicity). Then $S_1$ and $S_2$ are identical as multisets if and only if $P_1(x) = P_2(x)$ as polynomials.

The randomized algorithm verifies this identity by evaluating both polynomials at a randomly chosen point and checking equality, which can be done efficiently. The deterministic approach of sorting both multisets requires $O(n \log n)$ time.
\end{theorem}

\subsubsection{GPT}
\begin{proof}
Let $S_1$ and $S_2$ be integer multisets each of cardinality $n$. Denote by $S_i=\{a_{i,1},a_{i,2},\dots,a_{i,n}\}$ where elements appear with their multiplicities (indices order arbitrary and repetitions allowed). Define the polynomials
\[
P_i(x)=\prod_{j=1}^n (x-a_{i,j})\in\mathbb{Z}[x],\qquad i\in\{1,2\}.
\]
We first show the algebraic equivalence: $S_1$ and $S_2$ are identical as multisets if and only if $P_1(x)=P_2(x)$ as polynomials.

($\Rightarrow$) If $S_1$ and $S_2$ are identical then after appropriate re-indexing $a_{1,j}=a_{2,j}$ for all $j$, hence $P_1(x)=\prod_{j=1}^n (x-a_{1,j})=\prod_{j=1}^n (x-a_{2,j})=P_2(x)$.

($\Leftarrow$) Conversely, suppose $P_1(x)=P_2(x)$. Factor each polynomial over $\mathbb{C}$ (or over an algebraic closure); since coefficients are integers the factorization into linear factors is unique up to order. The multiset of roots of $P_1$ with multiplicities equals the multiset of roots of $P_2$ with multiplicities. But the multiset of roots of $P_i$ is precisely $S_i$. Therefore $S_1=S_2$ as multisets. This establishes the equivalence.

We now give two algorithmic approaches and analyse correctness and running time.

Deterministic sorting algorithm: Sort the two lists (representations) of elements of $S_1$ and $S_2$ using any comparison-based sorting algorithm (e.g. mergesort or heapsort). Sorting each list of $n$ integers by comparisons requires $O(n\log n)$ comparisons in the comparison model; after both lists are sorted, a single linear scan of length $n$ compares corresponding entries and decides equality. Hence in the comparison model the time complexity is $O(n\log n)$.

If the machine model counts bit operations on machine words large enough to hold integers up to magnitude $M$ (where $M\ge \max\{|a|:a\in S_1\cup S_2\}$), each comparison of two integers costs $O(\log M)$ bit operations in the standard RAM/bit complexity model, so the total bit complexity of sorting is $O(n\log n\cdot \log M)=O(n\log n\log M)$. More refined bounds are available (for example integer sorting algorithms with better dependence on $M$), but the stated $O(n\log n)$ comparison-time bound and the $O(n\log n\log M)$ bit-operation bound suffice for the claim in the theorem.

Randomized polynomial-evaluation algorithm: The equivalence established above reduces the multiset equality test to polynomial identity testing for the polynomial $Q(x):=P_1(x)-P_2(x)\in\mathbb{Z}[x]$. Note that $\deg Q\le n$ and $Q$ is the zero polynomial if and only if $S_1=S_2$.

To test whether $Q$ is the zero polynomial we may evaluate $Q$ at a randomly chosen point $r$ in a suitably large finite field and check whether $Q(r)=0$. We apply the Schwartz--Zippel lemma in the following form: if $F$ is a field, $Q\in F[x]$ is a nonzero polynomial of total degree $d$, and $S\subset F$ is a finite subset, then
\[
\Pr_{r\in_R S}[Q(r)=0]\le \frac{d}{|S|},
\]
where the probability is uniform over choices of $r$ from $S$.

Choose a prime $p$ and work in the finite field $\mathbb{F}_p$. We must choose $p$ large enough so that arithmetic modulo $p$ correctly reflects the polynomial identity over the integers with small collision probability. Let $B$ be a bound on the absolute values of the coefficients of $Q$ (one trivial bound is $B\le\binom{n}{\lfloor n/2\rfloor} M^n$, but any sufficiently large explicit bound depending on $n$ and $M$ can be used). Choose a prime $p$ with $p>2B$ so that reduction modulo $p$ is injective on the set of integer coefficients of $Q$; then $Q$ is the zero polynomial over $\mathbb{Z}$ if and only if its reduction $\overline{Q}\in\mathbb{F}_p[x]$ is the zero polynomial. If one prefers to avoid explicit coefficient bounds one may instead choose a random evaluation point $r$ uniformly from a set $S\subset\mathbb{F}_p$ of size $|S|>n/\delta$ where $\delta$ is a desired upper bound on the failure probability; a standard convenient choice is to pick $p$ a random or fixed prime with $p>2n/\delta$ and choose $r$ uniformly from $\{0,1,\dots,p-1\}$.

Assume $p$ and $S\subset\mathbb{F}_p$ are chosen so that $|S|>n/\delta$. If $Q$ is the zero polynomial then $Q(r)=0$ for every $r\in S$. If $Q$ is nonzero then by Schwartz--Zippel
\[
\Pr_{r\in_R S}[Q(r)=0]\le \frac{\deg Q}{|S|}\le\frac{n}{|S|}<\delta.
\]
Thus evaluating $Q$ at a uniformly random $r\in S$ yields a Monte Carlo algorithm with one-sided error: if it outputs ``equal'' (i.e. $Q(r)=0$) there is probability at most $\delta$ that the multisets differ; if it outputs ``different'' (i.e. $Q(r)\neq 0$) then the multisets certainly differ.

Cost of polynomial evaluation: Computing $Q(r)=P_1(r)-P_2(r)$ requires computing each product $P_i(r)=\prod_{j=1}^n (r-a_{i,j})$ in $\mathbb{F}_p$. Each product can be evaluated by $n-1$ multiplications and $n$ subtractions modulo $p$ (or $n$ multiplications if one first computes all $r-a_{i,j}$ and then multiplies them). Thus the number of field operations is $O(n)$ for computing both $P_1(r)$ and $P_2(r)$ and one subtraction to form $Q(r)$. Working modulo $p$ each field operation costs $O(1)$ time in the unit-cost RAM model that treats $O(\log p)$-bit machine-word operations as constant time. Consequently, under that arithmetic model the randomized algorithm runs in $O(n)$ time with failure probability at most $\delta$.

Bit-complexity caveat: If one measures bit operations (the usual Turing/bit-complexity model) then each arithmetic operation modulo $p$ costs $O(\log p)$ bit operations. With $p$ chosen of size polynomial in $n$ and $M$ (or chosen as the first prime larger than a polynomial bound), the bit-cost to evaluate $P_i(r)$ is $O(n\log p)$. Thus the randomized approach achieves $O(n)$ field operations and $O(n\log p)$ bit operations; choosing $p$ of size $O(\log(nM/\delta))$ yields a failure probability $\le\delta$ while keeping bit-costs near $O(n\log(nM/\delta))$. In particular, when the machine model permits constant-time arithmetic on words of $\Theta(\log p)$ bits, the randomized algorithm runs in linear time $O(n)$.

Putting the pieces together: The deterministic comparison-based algorithm sorts and then compares in $O(n\log n)$ time (with the stated bit-operation cost when counting integer-bit costs). The randomized polynomial-evaluation algorithm uses the polynomial representation $P_i(x)=\prod_{a\in S_i}(x-a)$, reduces the identity test to checking whether $Q(x)=P_1(x)-P_2(x)$ is the zero polynomial, applies the Schwartz--Zippel lemma to control the one-sided error by choosing the random evaluation point from a sufficiently large set, and evaluates both polynomials at the chosen point in $O(n)$ field operations. Therefore the multiset identity problem admits both the deterministic sorting-based solution with time $O(n\log n)$ and the randomized polynomial-evaluation solution with $O(n)$ field operations and failure probability at most $\delta$, subject to the arithmetic/bit-cost considerations discussed above. This completes the proof.
\end{proof}

\subsubsection{Grok}

\subsubsection{Gemini}
\begin{proof}
We establish the theorem in three parts: the equivalence of multiset identity and polynomial identity, the correctness and complexity of the randomized algorithm, and the comparative analysis with the sorting-based approach.

First, we prove the Polynomial Identity Approach. Let $S_1 = \{a_1, \dots, a_n\}$ and $S_2 = \{b_1, \dots, b_n\}$ be multisets of integers. We define the associated polynomials $P_1(x) = \prod_{i=1}^n (x - a_i)$ and $P_2(x) = \prod_{i=1}^n (x - b_i)$. These are monic polynomials of degree $n$ in the polynomial ring $\mathbb{Z}[x]$. If $S_1$ and $S_2$ are identical as multisets, there exists a permutation $\sigma$ of $\{1, \dots, n\}$ such that $a_i = b_{\sigma(i)}$ for all $i$. Consequently, the factors of $P_1(x)$ and $P_2(x)$ are identical, implying $P_1(x) \equiv P_2(x)$. Conversely, assume $P_1(x) \equiv P_2(x)$. Since $\mathbb{Z}$ is a unique factorization domain, the polynomial ring $\mathbb{Z}[x]$ is also a unique factorization domain. The factorization of a monic polynomial into linear factors of the form $(x - c)$ is unique up to the ordering of the factors. Therefore, the multiset of roots of $P_1(x)$, which is $S_1$, must be identical to the multiset of roots of $P_2(x)$, which is $S_2$. Thus, $S_1 = S_2$ if and only if $P_1(x) \equiv P_2(x)$.

Next, we analyze the Randomized Algorithm. The algorithm selects a random value $r$ from a finite domain $D \subset \mathbb{Z}$ (or a finite field $\mathbb{F}_q$) and checks if $P_1(r) = P_2(r)$. Let $Q(x) = P_1(x) - P_2(x)$. If $S_1 = S_2$, then $Q(x) \equiv 0$, and $P_1(r) = P_2(r)$ for all $r$. If $S_1 \neq S_2$, then $Q(x)$ is a non-zero polynomial. Since $\deg(P_1) = \deg(P_2) = n$, the degree of $Q(x)$ satisfies $\deg(Q) \le n$. By the Fundamental Theorem of Algebra, a non-zero polynomial of degree $d$ has at most $d$ roots. Therefore, there are at most $n$ values of $r$ such that $Q(r) = 0$. If $r$ is chosen uniformly at random from $D$, the probability of a false positive (collision) is $\Pr[P_1(r) = P_2(r) \mid S_1 \neq S_2] = \Pr[Q(r) = 0] \le \frac{n}{|D|}$. By choosing $|D| \gg n$, this error probability can be made arbitrarily small. Regarding time complexity, evaluating $P_i(r) = \prod_{a \in S_i} (r - a)$ involves $n$ subtractions and $n-1$ multiplications. Assuming arithmetic operations take $O(1)$ time (e.g., within a fixed word size or modular arithmetic), the evaluation takes $O(n)$ time.

Finally, we justify the Comparison with Sorting. The deterministic algorithm sorts both $S_1$ and $S_2$ to produce sorted sequences $A'$ and $B'$, then verifies if $A'[i] = B'[i]$ for all $1 \le i \le n$. The lower bound for comparison-based sorting is $\Omega(n \log n)$. Thus, the deterministic complexity is dominated by the sorting step, yielding $O(n \log n)$. The choice of algorithm relies on the factors stated in the theorem. The computational model dictates feasibility: the randomized algorithm requires arithmetic operations ($+, -, \times$), while sorting requires an ordering relation ($<$). The size of integers affects the cost of arithmetic operations; if the integers are large, multiprecision arithmetic may increase the cost of the $O(n)$ evaluation beyond the cost of comparisons. The error probability is inherent to the randomized approach (Monte Carlo), whereas sorting is exact. Lastly, the constant factors in random number generation and modular arithmetic versus the optimized constants in sorting routines determine practical performance for specific $n$.
\end{proof}

\subsubsection{Claude}
\begin{proof}
We prove this theorem in three parts: (1) the polynomial characterization of multiset identity, (2) the correctness of the randomized verification algorithm, and (3) the efficiency analysis.

\textbf{Part 1: Polynomial Characterization.}

First, we establish that $S_1$ and $S_2$ are identical as multisets if and only if $P_1(x) = P_2(x)$ as polynomials.

($\Rightarrow$) Suppose $S_1$ and $S_2$ are identical as multisets. Then every element $a$ appears with the same multiplicity in both $S_1$ and $S_2$. Let $m_a$ denote the multiplicity of element $a$ in both multisets. We can write
\[
P_1(x) = \prod_{a \in \text{distinct elements}} (x-a)^{m_a} = \prod_{a \in \text{distinct elements}} (x-a)^{m_a} = P_2(x).
\]
Thus $P_1(x) = P_2(x)$ as polynomials.

($\Leftarrow$) Suppose $P_1(x) = P_2(x)$ as polynomials. Both $P_1(x)$ and $P_2(x)$ are monic polynomials of degree $n$ (since both multisets have cardinality $n$). By the fundamental theorem of algebra, a polynomial of degree $n$ is uniquely determined by its roots with multiplicities. Since $P_1(x) = P_2(x)$, they have identical roots with identical multiplicities. The roots of $P_i(x)$ are precisely the elements of $S_i$ (counted with multiplicity). Therefore, $S_1$ and $S_2$ contain the same elements with the same multiplicities, which means $S_1$ and $S_2$ are identical as multisets.

\textbf{Part 2: Randomized Algorithm Correctness.}

The randomized algorithm works as follows: Choose a random point $r$ from a sufficiently large finite set (e.g., $\{1, 2, \ldots, M\}$ where $M = cn^2$ for some constant $c > 1$), evaluate $P_1(r)$ and $P_2(r)$, and accept if and only if $P_1(r) = P_2(r)$.

If $S_1 = S_2$, then $P_1(x) = P_2(x)$ by Part 1, so $P_1(r) = P_2(r)$ for all $r$, and the algorithm always accepts correctly.

If $S_1 \neq S_2$, then $P_1(x) \neq P_2(x)$ by Part 1. Define $Q(x) = P_1(x) - P_2(x)$. Then $Q(x)$ is a non-zero polynomial of degree at most $n$. By the Schwartz-Zippel lemma, a non-zero polynomial of degree $d$ has at most $d$ roots. Therefore, $Q(x)$ has at most $n$ roots in any field. When we choose $r$ uniformly at random from $\{1, 2, \ldots, M\}$, the probability that $Q(r) = 0$ (i.e., $P_1(r) = P_2(r)$ when $P_1 \neq P_2$) is at most $\frac{n}{M} = \frac{n}{cn^2} = \frac{1}{cn}$. Thus, the probability of a false positive (accepting when $S_1 \neq S_2$) is at most $\frac{1}{cn}$, which can be made arbitrarily small by choosing $c$ sufficiently large.

\textbf{Part 3: Efficiency Analysis.}

The randomized algorithm can evaluate $P_1(r)$ and $P_2(r)$ in $O(n)$ time by computing $\prod_{a \in S_1}(r-a)$ and $\prod_{a \in S_2}(r-a)$ directly. This requires $O(n)$ multiplications and subtractions. The comparison $P_1(r) = P_2(r)$ takes $O(1)$ time. Therefore, the total time complexity of the randomized algorithm is $O(n)$.

In contrast, the deterministic approach of sorting both multisets and comparing them element-by-element requires $O(n \log n)$ time for the sorting step, which dominates the linear-time comparison step.

This completes the proof that the randomized polynomial evaluation algorithm correctly verifies multiset identity with high probability in $O(n)$ time, compared to the deterministic $O(n \log n)$ sorting approach.
\end{proof}

\subsection{Problem 5}

\subsubsection{Problem Statement}
\begin{theorem}[J. Naor]
Let $A$ and $B$ be two $n \times n$ matrices over a field $\mathbb{Z}_2$. The matrices $A$ and $B$ are said to be similar if there exists a non-singular matrix $T$ such that $TAT^{-1} = B$. There exists a randomized algorithm for testing the similarity of the matrices $A$ and $B$.
\end{theorem}

\subsubsection{GPT}
\begin{proof}
We are tasked with proving the existence of a randomized algorithm to test the similarity of two $n \times n$ matrices $A$ and $B$ over the field $\mathbb{Z}_2$. Recall that $A$ and $B$ are similar if there exists a non-singular matrix $T$ such that 

\[
TAT^{-1} = B.
\]

We will prove this result by demonstrating that the problem of matrix similarity can be approached through the analysis of the characteristic polynomial and the use of probabilistic methods.

1. **Characteristic Polynomial Approach:**
   If $A$ and $B$ are similar, they share the same characteristic polynomial. The characteristic polynomial of a matrix $M$ is defined as 

   \[
   p_M(\lambda) = \det(\lambda I - M),
   \]

   where $I$ is the identity matrix of appropriate size. Since $A$ and $B$ are similar, there exists a non-singular matrix $T$ such that $TAT^{-1} = B$. The characteristic polynomial of $A$ is given by 

   \[
   p_A(\lambda) = \det(\lambda I - A),
   \]

   and similarly, for $B$, we have

   \[
   p_B(\lambda) = \det(\lambda I - B).
   \]

   Since $T$ is non-singular, we have 

   \[
   p_B(\lambda) = \det(\lambda I - B) = \det(\lambda I - TAT^{-1}) = \det(T(\lambda I - A)T^{-1}) = \det(T)\det(\lambda I - A)\det(T^{-1}) = \det(\lambda I - A).
   \]

   Therefore, $p_A(\lambda) = p_B(\lambda)$, which implies that $A$ and $B$ have the same characteristic polynomial.

2. **Randomized Algorithm:**
   A known fact from random matrix theory is that for any two matrices $A$ and $B$ over a finite field, if they are not similar, then with high probability, a randomly chosen vector will distinguish them. Specifically, the existence of a non-singular matrix $T$ such that $TAT^{-1} = B$ can be verified by testing the similarity of their characteristic polynomials in the context of random vectors.

   Let us outline the steps of the randomized algorithm:
   - Choose a random vector $v \in \mathbb{Z}_2^n$.
   - Compute $Av$ and $Bv$.
   - If $Av = Bv$, then there is a possibility that $A$ and $B$ are similar. Otherwise, $A$ and $B$ are not similar.
   - Repeat this process a number of times to increase the probability of correct determination.

   This procedure works because if $A$ and $B$ are similar, they act the same on most vectors. If they are not similar, a random vector will almost certainly not be in the same eigenspace for both matrices. Therefore, with a sufficiently large number of tests, the algorithm will correctly identify whether $A$ and $B$ are similar with high probability.

Thus, we have shown that there exists a randomized algorithm for testing the similarity of matrices $A$ and $B$.
\end{proof}

\subsubsection{Grok}

\subsubsection{Gemini}
\begin{proof}
Let $A$ and $B$ be $n \times n$ matrices over the field $\mathbb{F} = \mathbb{Z}_2$. By definition, $A$ and $B$ are similar if and only if there exists a non-singular matrix $T \in M_n(\mathbb{F})$ such that $TA = BT$. This condition can be separated into a linear constraint and a non-linear inequality.

First, consider the linear equation $TA - BT = 0$. This is a homogeneous system of linear equations in the $n^2$ entries of $T$. Let $\mathcal{V} \subseteq M_n(\mathbb{F})$ denote the vector space of all solutions to this system:
\[
\mathcal{V} = \{ T \in M_n(\mathbb{F}) \mid TA - BT = 0 \}.
\]
We can compute a basis for $\mathcal{V}$ using Gaussian elimination. Let $\{M_1, M_2, \dots, M_k\}$ be a basis for $\mathcal{V}$, where $k = \dim(\mathcal{V})$. Any matrix $T$ satisfying $TA = BT$ can be written as a linear combination $T(x_1, \dots, x_k) = \sum_{i=1}^k x_i M_i$, where $x_1, \dots, x_k \in \mathbb{F}$.

The matrices $A$ and $B$ are similar if and only if there exists a tuple $(c_1, \dots, c_k) \in \mathbb{F}^k$ such that the resulting matrix $T = \sum_{i=1}^k c_i M_i$ is non-singular. Define the multivariate polynomial $P(x_1, \dots, x_k)$ over $\mathbb{F}$ as the determinant of the generic solution:
\[
P(x_1, \dots, x_k) = \det\left( \sum_{i=1}^k x_i M_i \right).
\]
The degree of $P$ is at most $n$. The problem of testing similarity reduces to determining whether $P(x_1, \dots, x_k)$ is the zero polynomial.

It is a standard result in linear algebra that two matrices over a field $\mathbb{F}$ are similar if and only if they are similar over any extension field $\mathbb{K} \supseteq \mathbb{F}$. Specifically, similarity is characterized by the equality of the invariant factors (or the Smith Normal Form of $xI - A$ and $xI - B$), which is invariant under field extension. Therefore, $A \sim B$ if and only if $P(x_1, \dots, x_k)$ is not identically zero over the algebraic closure of $\mathbb{F}$.

We propose the following randomized algorithm:
1. Construct the system of linear equations corresponding to $TA - BT = 0$ and compute a basis $\{M_1, \dots, M_k\}$ for the solution space.
2. Construct the finite extension field $\mathbb{K} = \mathbb{F}_{2^m}$ such that $|\mathbb{K}| \geq 2n$.
3. Select $k$ elements $r_1, \dots, r_k$ independently and uniformly at random from $\mathbb{K}$.
4. Compute the matrix $T_{rand} = \sum_{i=1}^k r_i M_i$ performing arithmetic in $\mathbb{K}$.
5. Compute $D = \det(T_{rand})$. If $D \neq 0$, output "Similar". If $D = 0$, output "Not Similar".

To prove the correctness, we apply the Schwartz-Zippel Lemma. If $A$ and $B$ are not similar, then $P(x_1, \dots, x_k)$ is the zero polynomial, and $\det(T_{rand})$ will be $0$ with probability $1$. If $A$ and $B$ are similar, then $P(x_1, \dots, x_k)$ is a non-zero polynomial of total degree at most $n$. The probability that a random assignment from $\mathbb{K}^k$ is a root of $P$ is bounded by:
\[
\Pr[P(r_1, \dots, r_k) = 0] \leq \frac{\deg(P)}{|\mathbb{K}|} \leq \frac{n}{2n} = \frac{1}{2}.
\]
Thus, if the algorithm outputs "Similar", the matrices are indeed similar (similarity over $\mathbb{K}$ implies similarity over $\mathbb{Z}_2$). If the matrices are similar, the algorithm outputs "Similar" with probability at least $1/2$. The error probability can be made arbitrarily small by repeating the test or choosing a larger extension field.
\end{proof}

\subsubsection{Claude}
\begin{proof}
We present a randomized algorithm for testing similarity of matrices $A$ and $B$ over $\mathbb{Z}_2$ and prove its correctness.

The algorithm is based on the following key observation: if $A$ and $B$ are similar matrices, then they have the same minimal polynomial. Moreover, for any vector $v$, if we consider the Krylov sequence generated by $A$ starting from $v$, namely $v, Av, A^2v, \ldots$, this sequence satisfies the same linear recurrence relation as the corresponding sequence $v, Bv, B^2v, \ldots$ if and only if $A$ and $B$ are similar.

More precisely, we utilize the fact that two matrices $A$ and $B$ are similar if and only if they have the same characteristic polynomial and the same minimal polynomial, which in turn can be verified probabilistically.

The randomized algorithm proceeds as follows. Choose a random vector $v \in \mathbb{Z}_2^n$ uniformly at random. Compute the sequences $S_A = (v, Av, A^2v, \ldots, A^{2n-1}v)$ and $S_B = (v, Bv, B^2v, \ldots, B^{2n-1}v)$. For each sequence, find the shortest linear recurrence relation satisfied by the sequence. If the linear recurrence relations are identical, output ``similar''; otherwise, output ``not similar''.

We now prove the correctness of this algorithm. First, suppose $A$ and $B$ are similar, so there exists a non-singular matrix $T$ such that $B = TAT^{-1}$. Then for any vector $v$ and any non-negative integer $k$, we have
\[
B^k v = (TAT^{-1})^k v = TA^k T^{-1} v.
\]
Let $w = T^{-1}v$. Then $B^k v = TA^k w$. The sequence $S_B$ satisfies a linear recurrence with coefficients $c_0, c_1, \ldots, c_{m-1}$ if and only if
\[
\sum_{i=0}^{m-1} c_i B^{k+i} v = 0 \text{ for all } k \geq 0.
\]
This is equivalent to
\[
\sum_{i=0}^{m-1} c_i TA^{k+i} w = T\left(\sum_{i=0}^{m-1} c_i A^{k+i} w\right) = 0.
\]
Since $T$ is non-singular, this holds if and only if $\sum_{i=0}^{m-1} c_i A^{k+i} w = 0$ for all $k \geq 0$. Thus the sequence $S_A$ computed with vector $v$ satisfies the same linear recurrence as the sequence computed by $A$ starting from $w = T^{-1}v$. However, since the minimal polynomial of the sequence depends only on the matrix and the cyclic subspace generated by the starting vector, and similar matrices have identical cyclic subspace structures under the transformation $T$, the algorithm will find matching recurrence relations. Therefore, if $A$ and $B$ are similar, the algorithm always outputs ``similar''.

Conversely, suppose $A$ and $B$ are not similar. We claim that with high probability over the random choice of $v$, the sequences $S_A$ and $S_B$ will have different minimal linear recurrences. The key observation is that if $A$ and $B$ are not similar, then they must differ in their rational canonical forms. The rational canonical form is completely determined by the invariant factors of the matrix, and these can be detected by the minimal polynomials of the cyclic subspaces.

For a random vector $v \in \mathbb{Z}_2^n$, the probability that $v$ lies in a proper invariant subspace is at most $2^{n-1}/2^n = 1/2$. Since $A$ and $B$ have different rational canonical forms, there exists some invariant factor that differs. A random vector $v$ will, with probability at least $1/2$, have a component in the generalized eigenspace corresponding to this differing invariant factor. When this occurs, the minimal polynomial of $v$ under $A$ will differ from the minimal polynomial of $v$ under $B$, and consequently the linear recurrence relations will differ.

More rigorously, if $A$ and $B$ are not similar over $\mathbb{Z}_2$, then there exists a polynomial $p(x) \in \mathbb{Z}_2[x]$ such that $\dim(\ker p(A)) \neq \dim(\ker p(B))$. For a uniformly random vector $v$, the probability that the algorithm detects this difference is at least $1 - 1/2 = 1/2$. By repeating the algorithm with independent random vectors, we can make the error probability arbitrarily small.

Therefore, the algorithm correctly determines similarity with bounded error probability, establishing the existence of a randomized algorithm for testing matrix similarity over $\mathbb{Z}_2$.
\end{proof}

\subsection{Problem 6}

\subsubsection{Problem Statement}
\begin{theorem}
Let $Q(x_1, x_2, \ldots, x_n)$ be a multivariate polynomial over a field $\mathbb{Z}_2$ with the degree sequence $(d_1, d_2, \ldots, d_n)$. A degree sequence is defined as follows: let $d_1$ be the maximum exponent of $x_1$ in $Q$, and $Q_1(x_2, \ldots, x_n)$ be the coefficient of $x_1^{d_1}$ in $Q$; then, let $d_2$ be the maximum exponent of $x_2$ in $Q_1$, and $Q_2(x_3, \ldots, x_n)$ be the coefficient of $x_2^{d_2}$ in $Q_1$; and, so on.

Let $\mathbf{S}_1, \mathbf{S}_2, \ldots, \mathbf{S}_n \subseteq \mathbb{Z}_2$ be arbitrary subsets. For $r_i \in \mathbf{S}_i$ chosen independently and uniformly at random, show that
\[
\Pr[Q(r_1, r_2, \ldots, r_n) = 0 \mid Q \not\equiv 0] \leq \left(\frac{d_1}{|\mathbf{S}_1|} + \frac{d_2}{|\mathbf{S}_2|} + \cdots + \frac{d_n}{|\mathbf{S}_n|}\right).
\]
\end{theorem}

\subsubsection{GPT}
\begin{proof}
Let $Q(x_1, x_2, \ldots, x_n)$ be a multivariate polynomial over the finite field $\mathbb{Z}_2$, with the degree sequence $(d_1, d_2, \ldots, d_n)$. We aim to show that for arbitrary subsets $S_1, S_2, \ldots, S_n \subseteq \mathbb{Z}_2$, and for $r_i \in S_i$ chosen independently and uniformly at random, the probability that $Q(r_1, r_2, \ldots, r_n) = 0$ given that $Q \not\equiv 0$ is bounded by

\[
\Pr[Q(r_1, r_2, \ldots, r_n) = 0 \mid Q \not\equiv 0] \leq \left(\frac{d_1}{|S_1|} + \frac{d_2}{|S_2|} + \cdots + \frac{d_n}{|S_n|}\right).
\]

We proceed by induction on the number of variables $n$.

**Base Case** ($n = 1$):

For $n = 1$, the polynomial $Q(x_1)$ is a univariate polynomial over $\mathbb{Z}_2$. Suppose the degree of $Q(x_1)$ is $d_1$. Then, we know that $Q(x_1)$ can be written as

\[
Q(x_1) = a_d x_1^d + a_{d-1} x_1^{d-1} + \cdots + a_1 x_1 + a_0,
\]
where $a_d \neq 0$ and $d = d_1$ is the highest degree of $x_1$. We choose $r_1 \in S_1$ uniformly at random. The probability that $Q(r_1) = 0$ is equivalent to the probability that a random element $r_1 \in S_1$ satisfies the equation

\[
Q(r_1) = a_d r_1^d + a_{d-1} r_1^{d-1} + \cdots + a_1 r_1 + a_0 = 0.
\]

Since $Q(x_1)$ is a polynomial of degree $d_1$, there are at most $d_1$ roots in $\mathbb{Z}_2$ (counting multiplicities). Therefore, the probability that a randomly chosen $r_1 \in S_1$ satisfies $Q(r_1) = 0$ is at most $\frac{d_1}{|S_1|}$, as there are at most $d_1$ solutions in $\mathbb{Z}_2$ to the equation $Q(r_1) = 0$.

Thus, the base case holds:

\[
\Pr[Q(r_1) = 0 \mid Q \not\equiv 0] \leq \frac{d_1}{|S_1|}.
\]

**Inductive Step**:

Now assume the theorem holds for polynomials in $n - 1$ variables. We will show that it holds for polynomials in $n$ variables.

Let $Q(x_1, x_2, \ldots, x_n)$ be a multivariate polynomial over $\mathbb{Z}_2$ with degree sequence $(d_1, d_2, \ldots, d_n)$. We can write $Q(x_1, x_2, \ldots, x_n)$ as

\[
Q(x_1, x_2, \ldots, x_n) = \sum_{i_1=0}^{d_1} \sum_{i_2=0}^{d_2} \cdots \sum_{i_n=0}^{d_n} a_{i_1, i_2, \ldots, i_n} x_1^{i_1} x_2^{i_2} \cdots x_n^{i_n},
\]
where the coefficients $a_{i_1, i_2, \ldots, i_n}$ are in $\mathbb{Z}_2$.

We define the polynomial $Q_1(x_2, \ldots, x_n)$ as the coefficient of $x_1^{d_1}$ in $Q(x_1, x_2, \ldots, x_n)$. Thus, $Q_1(x_2, \ldots, x_n)$ is a polynomial in $n - 1$ variables with degree sequence $(d_2, d_3, \ldots, d_n)$. By the inductive hypothesis, we know that

\[
\Pr[Q_1(r_2, \ldots, r_n) = 0 \mid Q_1 \not\equiv 0] \leq \frac{d_2}{|S_2|} + \cdots + \frac{d_n}{|S_n|}.
\]

Next, let $r_1 \in S_1$ be chosen uniformly at random. The probability that $Q(r_1, r_2, \ldots, r_n) = 0$ can be written as

\[
\Pr[Q(r_1, r_2, \ldots, r_n) = 0].
\]

Conditioning on $r_2, r_3, \ldots, r_n$, we get

\[
\Pr[Q(r_1, r_2, \ldots, r_n) = 0] = \Pr[Q_1(r_2, \ldots, r_n) = 0 \mid r_2, \ldots, r_n] \cdot \Pr[r_1 \in S_1 \mid Q_1(r_2, \ldots, r_n) = 0].
\]

The first term is the probability that $Q_1(r_2, \ldots, r_n) = 0$, and the second term is the probability that $Q(r_1, r_2, \ldots, r_n) = 0$ for a fixed $r_2, \ldots, r_n$. Since $Q_1$ is a polynomial in $n - 1$ variables, we know from the inductive hypothesis that

\[
\Pr[Q_1(r_2, \ldots, r_n) = 0] \leq \frac{d_2}{|S_2|} + \cdots + \frac{d_n}{|S_n|}.
\]

The second term is simply the probability that $r_1 \in S_1$ satisfies the equation $Q_1(r_2, \ldots, r_n) = 0$. Since $Q_1$ is a polynomial in $x_1$, the number of solutions to $Q_1(r_1, r_2, \ldots, r_n) = 0$ is at most $d_1$. Thus, the probability that $r_1 \in S_1$ satisfies $Q_1(r_2, \ldots, r_n) = 0$ is at most $\frac{d_1}{|S_1|}$.

Combining these results, we obtain

\[
\Pr[Q(r_1, r_2, \ldots, r_n) = 0 \mid Q \not\equiv 0] \leq \frac{d_1}{|S_1|} + \frac{d_2}{|S_2|} + \cdots + \frac{d_n}{|S_n|}.
\]

Thus, the theorem holds by induction.
\end{proof}

\subsubsection{Grok}

\subsubsection{Gemini}
\begin{proof}
We prove the theorem by induction on $n$, the number of variables.

Base Case: Let $n=1$. The polynomial is $Q(x_1)$. According to the definition of the degree sequence, $d_1$ is the maximum exponent of $x_1$ in $Q$. Since $Q \not\equiv 0$ over the field $\mathbb{Z}_2$, $Q(x_1)$ is a non-zero univariate polynomial of degree exactly $d_1$. By the fundamental theorem of algebra, a non-zero univariate polynomial of degree $d_1$ over a field has at most $d_1$ roots. Since $r_1$ is chosen uniformly at random from the subset $S_1$, the probability that $Q(r_1) = 0$ is bounded by the ratio of the number of roots in $S_1$ to the size of $S_1$:
\[
\Pr[Q(r_1) = 0] \leq \frac{d_1}{|S_1|}.
\]
This establishes the base case.

Inductive Step: Assume the theorem holds for any multivariate polynomial in $n-1$ variables. Consider the polynomial $Q(x_1, \ldots, x_n)$ with degree sequence $(d_1, \ldots, d_n)$. We express $Q$ as a polynomial in the variable $x_1$ with coefficients in the ring $\mathbb{Z}_2[x_2, \ldots, x_n]$:
\[
Q(x_1, \ldots, x_n) = \sum_{j=0}^{d_1} x_1^j C_j(x_2, \ldots, x_n).
\]
By the definition of the degree sequence provided in the theorem statement, $d_1$ is the maximum exponent of $x_1$, and $Q_1(x_2, \ldots, x_n)$ is the coefficient of $x_1^{d_1}$. Therefore, $C_{d_1}(x_2, \ldots, x_n) = Q_1(x_2, \ldots, x_n)$. Since $Q \not\equiv 0$ and $d_1$ is the maximum degree of $x_1$, the leading coefficient polynomial $Q_1$ is not identically zero. Furthermore, the definition implies that the sequence $(d_2, \ldots, d_n)$ corresponds to the degree sequence of $Q_1$ with respect to the variables $x_2, \ldots, x_n$.

Let $E$ be the event $Q(r_1, \ldots, r_n) = 0$. Let $E_1$ be the event $Q_1(r_2, \ldots, r_n) = 0$. We apply the Law of Total Probability:
\[
\Pr[E] = \Pr[E \mid E_1]\Pr[E_1] + \Pr[E \mid \neg E_1]\Pr[\neg E_1].
\]
Using the trivial bounds $\Pr[E \mid E_1] \leq 1$ and $\Pr[\neg E_1] \leq 1$, we obtain:
\[
\Pr[E] \leq \Pr[E_1] + \Pr[E \mid \neg E_1].
\]
We first bound $\Pr[E_1]$. The polynomial $Q_1(x_2, \ldots, x_n)$ is a non-zero polynomial in $n-1$ variables with degree sequence $(d_2, \ldots, d_n)$. The values $r_2, \ldots, r_n$ are chosen independently and uniformly from $S_2, \ldots, S_n$. By the inductive hypothesis:
\[
\Pr[E_1] = \Pr[Q_1(r_2, \ldots, r_n) = 0] \leq \sum_{i=2}^n \frac{d_i}{|S_i|}.
\]
Next, we bound $\Pr[E \mid \neg E_1]$. The condition $\neg E_1$ implies that we have fixed values $\alpha_2, \ldots, \alpha_n$ for $r_2, \ldots, r_n$ such that $Q_1(\alpha_2, \ldots, \alpha_n) \neq 0$. Consider the univariate polynomial obtained by substituting these values into $Q$:
\[
q(x_1) = Q(x_1, \alpha_2, \ldots, \alpha_n) = x_1^{d_1} Q_1(\alpha_2, \ldots, \alpha_n) + \sum_{j=0}^{d_1-1} x_1^j C_j(\alpha_2, \ldots, \alpha_n).
\]
Since the coefficient of $x_1^{d_1}$ is non-zero, $q(x_1)$ is a univariate polynomial of degree exactly $d_1$. The probability that $r_1 \in S_1$ is a root of $q(x_1)$ is at most $d_1/|S_1|$. Thus:
\[
\Pr[E \mid \neg E_1] \leq \frac{d_1}{|S_1|}.
\]
Combining the bounds for $\Pr[E_1]$ and $\Pr[E \mid \neg E_1]$, we have:
\[
\Pr[Q(r_1, \ldots, r_n) = 0] \leq \left(\sum_{i=2}^n \frac{d_i}{|S_i|}\right) + \frac{d_1}{|S_1|} = \sum_{i=1}^n \frac{d_i}{|S_i|}.
\]
This completes the proof.
\end{proof}

\subsubsection{Claude}
\begin{proof}
We prove the theorem by induction on $n$, the number of variables.

\textbf{Base Case ($n = 1$):} Consider $Q(x_1)$, a univariate polynomial over $\mathbb{Z}_2$ with $Q \not\equiv 0$. Let $d_1$ be the degree of $Q(x_1)$. Since $Q \not\equiv 0$, we have $d_1 \geq 1$ and the coefficient of $x_1^{d_1}$ is nonzero in $\mathbb{Z}_2$, hence equals 1.

A polynomial of degree $d_1$ over $\mathbb{Z}_2$ has at most $d_1$ roots. Therefore, among all choices of $r_1 \in \mathbf{S}_1$, at most $d_1$ values satisfy $Q(r_1) = 0$. Since $r_1$ is chosen uniformly at random from $\mathbf{S}_1$, we have
\[
\Pr[Q(r_1) = 0] \leq \frac{d_1}{|\mathbf{S}_1|}.
\]

\textbf{Inductive Step:} Assume the theorem holds for all polynomials in $n-1$ variables. Consider $Q(x_1, x_2, \ldots, x_n)$ over $\mathbb{Z}_2$ with $Q \not\equiv 0$ and degree sequence $(d_1, d_2, \ldots, d_n)$.

Write $Q$ as
\[
Q(x_1, x_2, \ldots, x_n) = x_1^{d_1} Q_1(x_2, \ldots, x_n) + R(x_1, x_2, \ldots, x_n),
\]
where $Q_1(x_2, \ldots, x_n) \not\equiv 0$ (by definition of the degree sequence) and $R$ contains only terms with $x_1$-degree strictly less than $d_1$.

By the law of total probability,
\[
\Pr[Q(r_1, \ldots, r_n) = 0] = \Pr[Q_1(r_2, \ldots, r_n) = 0] \cdot \Pr[Q(r_1, \ldots, r_n) = 0 \mid Q_1(r_2, \ldots, r_n) = 0]
\]
\[
+ \Pr[Q_1(r_2, \ldots, r_n) \neq 0] \cdot \Pr[Q(r_1, \ldots, r_n) = 0 \mid Q_1(r_2, \ldots, r_n) \neq 0].
\]

Since both conditional probabilities are at most 1, we have
\[
\Pr[Q(r_1, \ldots, r_n) = 0] \leq \Pr[Q_1(r_2, \ldots, r_n) = 0] + \Pr[Q_1(r_2, \ldots, r_n) \neq 0] \cdot \Pr[Q(r_1, \ldots, r_n) = 0 \mid Q_1(r_2, \ldots, r_n) \neq 0].
\]

For the first term, since $Q_1(x_2, \ldots, x_n) \not\equiv 0$ is a polynomial in $n-1$ variables with degree sequence $(d_2, \ldots, d_n)$, by the inductive hypothesis,
\[
\Pr[Q_1(r_2, \ldots, r_n) = 0] \leq \frac{d_2}{|\mathbf{S}_2|} + \cdots + \frac{d_n}{|\mathbf{S}_n|}.
\]

For the second term, condition on the event $Q_1(r_2, \ldots, r_n) \neq 0$. Given fixed values $r_2, \ldots, r_n$ such that $Q_1(r_2, \ldots, r_n) \neq 0$, the polynomial
\[
Q(x_1, r_2, \ldots, r_n) = x_1^{d_1} Q_1(r_2, \ldots, r_n) + R(x_1, r_2, \ldots, r_n)
\]
is a univariate polynomial in $x_1$ of degree exactly $d_1$ (since the leading coefficient $Q_1(r_2, \ldots, r_n) \neq 0$). By the base case, this univariate polynomial has at most $d_1$ roots in $\mathbb{Z}_2$. Therefore,
\[
\Pr[Q(r_1, r_2, \ldots, r_n) = 0 \mid Q_1(r_2, \ldots, r_n) \neq 0] \leq \frac{d_1}{|\mathbf{S}_1|}.
\]

Combining these estimates,
\[
\Pr[Q(r_1, \ldots, r_n) = 0] \leq \left(\frac{d_2}{|\mathbf{S}_2|} + \cdots + \frac{d_n}{|\mathbf{S}_n|}\right) + \frac{d_1}{|\mathbf{S}_1|} = \frac{d_1}{|\mathbf{S}_1|} + \frac{d_2}{|\mathbf{S}_2|} + \cdots + \frac{d_n}{|\mathbf{S}_n|}.
\]

By induction, the theorem holds for all $n \geq 1$.
\end{proof}

\subsection{Problem 7}

\subsubsection{Problem Statement}
\begin{theorem}[Edmonds' Theorem]
Let $G(U,V,E)$ be a bipartite graph, and let $A$ be the corresponding matrix of indeterminates as defined in Section 7.3, where
\[
A_{ij} = \begin{cases}
x_{ij} & \text{if } (u_i, v_j) \in E \\
0 & \text{if } (u_i, v_j) \notin E
\end{cases}
\]
Define the multivariate polynomial $Q(x_{11}, x_{12}, \ldots, x_{mn})$ as being equal to $\det(A)$. Then, $G$ has a perfect matching if and only if $Q \not\equiv 0$.

Furthermore, the size of a maximum matching in $G$ is exactly equal to the rank of the matrix $A$.
\end{theorem}

\subsubsection{GPT}

\subsubsection{Grok}

\subsubsection{Gemini}

\subsubsection{Claude}
\begin{proof}
We prove both parts of the theorem.

\textbf{Part 1:} We show that $G$ has a perfect matching if and only if $Q \not\equiv 0$.

Recall that the determinant of $A$ can be expressed via the Leibniz formula as
\[
\det(A) = \sum_{\sigma \in S_n} \text{sign}(\sigma) \prod_{i=1}^{n} A_{i,\sigma(i)},
\]
where $S_n$ denotes the symmetric group on $\{1, 2, \ldots, n\}$ and we assume $|U| = |V| = n$ (a necessary condition for a perfect matching to exist).

$(\Rightarrow)$ Suppose $G$ has a perfect matching $M = \{(u_1, v_{\pi(1)}), (u_2, v_{\pi(2)}), \ldots, (u_n, v_{\pi(n)})\}$ for some permutation $\pi \in S_n$. Since each edge $(u_i, v_{\pi(i)}) \in E$, we have $A_{i,\pi(i)} = x_{i,\pi(i)}$ for all $i \in \{1, \ldots, n\}$. The term in the determinant expansion corresponding to $\pi$ is $\text{sign}(\pi) \prod_{i=1}^{n} x_{i,\pi(i)}$. This is a monomial in the indeterminates $\{x_{ij} : (u_i, v_j) \in E\}$. Since each permutation $\sigma \in S_n$ produces a distinct monomial $\prod_{i=1}^{n} x_{i,\sigma(i)}$ (as the indices uniquely determine the monomial), no cancellation can occur between terms arising from different permutations. Therefore, the monomial $\prod_{i=1}^{n} x_{i,\pi(i)}$ appears in $Q$ with coefficient $\text{sign}(\pi) \neq 0$, which implies $Q \not\equiv 0$.

$(\Leftarrow)$ Suppose $Q \not\equiv 0$. Then at least one term in the determinant expansion is nonzero, meaning there exists some permutation $\sigma \in S_n$ such that $\prod_{i=1}^{n} A_{i,\sigma(i)} \neq 0$. Since $A_{i,\sigma(i)} \in \{0, x_{i,\sigma(i)}\}$ and the product is nonzero, we must have $A_{i,\sigma(i)} = x_{i,\sigma(i)} \neq 0$ for all $i \in \{1, \ldots, n\}$. By definition of $A$, this means $(u_i, v_{\sigma(i)}) \in E$ for all $i$. Since $\sigma$ is a permutation, the edges $\{(u_i, v_{\sigma(i)}) : i = 1, \ldots, n\}$ form a perfect matching in $G$.

\textbf{Part 2:} We show that the size of a maximum matching in $G$ equals $\text{rank}(A)$.

Let $k$ denote the size of a maximum matching in $G$, and let $r = \text{rank}(A)$.

First, we show $r \geq k$. Let $M$ be a maximum matching of size $k$, with matched vertices $U' = \{u_{i_1}, \ldots, u_{i_k}\} \subseteq U$ and $V' = \{v_{j_1}, \ldots, v_{j_k}\} \subseteq V$. Consider the induced bipartite subgraph $G' = G[U', V']$ with corresponding $k \times k$ submatrix $A'$ of $A$, where $A'$ is obtained by selecting rows $\{i_1, \ldots, i_k\}$ and columns $\{j_1, \ldots, j_k\}$. The subgraph $G'$ has a perfect matching (namely $M$ restricted to these vertices). By Part 1, $\det(A') \not\equiv 0$. Since the entries of $A'$ are indeterminates or zeros, $\det(A') \not\equiv 0$ implies that $A'$ has rank $k$ when viewed as a matrix over the field of rational functions. Therefore, $A$ contains a $k \times k$ submatrix of full rank, which implies $\text{rank}(A) \geq k$.

Next, we show $k \geq r$. Suppose $\text{rank}(A) = r$. Then there exists an $r \times r$ submatrix $A''$ of $A$ with $\det(A'') \not\equiv 0$. Let this submatrix correspond to rows $\{i_1, \ldots, i_r\}$ and columns $\{j_1, \ldots, j_r\}$, and let $U'' = \{u_{i_1}, \ldots, u_{i_r}\}$ and $V'' = \{v_{j_1}, \ldots, v_{j_r}\}$. The submatrix $A''$ is precisely the matrix of indeterminates for the induced subgraph $G'' = G[U'', V'']$. By Part 1, since $\det(A'') \not\equiv 0$, the graph $G''$ has a perfect matching. This perfect matching in $G''$ is a matching of size $r$ in $G$. Hence $k \geq r$.

Combining both inequalities, we conclude $k = r$, i.e., the size of a maximum matching in $G$ equals $\text{rank}(A)$.
\end{proof}

\subsection{Problem 8}

\subsubsection{Problem Statement}
\begin{theorem}[Tutte's Theorem, Generalized]
Let $G = (V, E)$ be a graph where $V = \{v_1, \ldots, v_n\}$. A skew-symmetric matrix $A$ is defined to be a matrix in which for all $i$ and $j$, $A_{ij} = -A_{ji}$. Let $A$ be the $n \times n$ skew-symmetric matrix obtained from $G(V, E)$ as follows. A distinct indeterminate $x_{ij}$ is associated with the edge $(v_i, v_j)$ where $i < j$, and the corresponding matrix entries are given by $A_{ij} = x_{ij}$ and $A_{ji} = -x_{ij}$; more succinctly,
\[
A_{ij} = \begin{cases}
x_{ij} & \text{if } (v_i, v_j) \in E \text{ and } i < j \\
-x_{ji} & \text{if } (v_i, v_j) \in E \text{ and } i > j \\
0 & \text{otherwise}
\end{cases}
\]
This matrix is called the \emph{Tutte matrix} of the graph $G$. Define the multivariate polynomial $Q(x_1, x_2, \ldots, x_m)$ as being equal to $\det(A)$. Then $G$ has a perfect matching if and only if $Q \not\equiv 0$.
\end{theorem}

\begin{theorem}[Edmonds' Theorem]
Let $A$ be the $n \times n$ matrix obtained from $G(U, V, E)$ as follows:
\[
A_{ij} = \begin{cases}
x_{ij} & \text{if } (u_i, v_j) \in E \\
0 & \text{if } (u_i, v_j) \notin E
\end{cases}
\]
Define the multivariate polynomial $Q(x_{11}, x_{12}, \ldots, x_{mn})$ as being equal to $\det(A)$. Then $G$ has a perfect matching if and only if $Q \not\equiv 0$.
\end{theorem}

\subsubsection{GPT}

\subsubsection{Grok}

\subsubsection{Gemini}

\subsubsection{Claude}
\begin{proof}[Proof of Tutte's Theorem]
We prove both directions of the equivalence.

($\Leftarrow$) Suppose $G$ has a perfect matching $M$. We show that $Q \not\equiv 0$ by exhibiting a term in the determinant expansion that does not vanish.

Let $M = \{e_1, e_2, \ldots, e_{n/2}\}$ where each edge $e_k = (v_{i_k}, v_{j_k})$ with $i_k < j_k$. Since $M$ is a perfect matching, each vertex appears in exactly one edge of $M$. We can write the vertex set as pairs: $V = \{v_{i_1}, v_{j_1}, v_{i_2}, v_{j_2}, \ldots, v_{i_{n/2}}, v_{j_{n/2}}\}$ where the indices form a permutation $\sigma$ of $\{1, 2, \ldots, n\}$.

By the Leibniz formula for determinants, we have
\[
\det(A) = \sum_{\pi \in S_n} \text{sgn}(\pi) \prod_{i=1}^{n} A_{i,\pi(i)}
\]

Consider the permutation $\pi$ defined by $\pi(i_k) = j_k$ and $\pi(j_k) = i_k$ for $k = 1, 2, \ldots, n/2$. This permutation corresponds to the perfect matching $M$. The term in the determinant corresponding to $\pi$ is
\[
\text{sgn}(\pi) \prod_{i=1}^{n} A_{i,\pi(i)} = \text{sgn}(\pi) \prod_{k=1}^{n/2} A_{i_k,j_k} \cdot A_{j_k,i_k}
\]

For each edge $(v_{i_k}, v_{j_k}) \in M$ with $i_k < j_k$, we have $A_{i_k,j_k} = x_{i_k j_k}$ and $A_{j_k,i_k} = -x_{i_k j_k}$. Therefore,
\[
A_{i_k,j_k} \cdot A_{j_k,i_k} = x_{i_k j_k} \cdot (-x_{i_k j_k}) = -x_{i_k j_k}^2
\]

Thus, the term becomes
\[
\text{sgn}(\pi) \prod_{k=1}^{n/2} (-x_{i_k j_k}^2) = \text{sgn}(\pi) \cdot (-1)^{n/2} \prod_{k=1}^{n/2} x_{i_k j_k}^2
\]

This term is a nonzero monomial in the polynomial $Q$. Since the indeterminates are distinct for different edges, and this monomial involves only the variables corresponding to edges in $M$, it cannot be canceled by any other term in the determinant expansion (as different permutations yield different monomials). Therefore, $Q \not\equiv 0$.

($\Rightarrow$) Suppose $Q \not\equiv 0$. We show that $G$ has a perfect matching.

First, note that if $n$ is odd, then $\det(A) = 0$ for any skew-symmetric matrix $A$ of odd dimension. To see this, observe that $\det(A) = \det(A^T) = \det(-A) = (-1)^n \det(A)$. If $n$ is odd, this gives $\det(A) = -\det(A)$, implying $\det(A) = 0$. Therefore, if $Q \not\equiv 0$, we must have $n$ even.

Since $Q \not\equiv 0$, there exists an assignment of values to the indeterminates such that $\det(A) \neq 0$. Consider the expansion
\[
\det(A) = \sum_{\pi \in S_n} \text{sgn}(\pi) \prod_{i=1}^{n} A_{i,\pi(i)}
\]

A term in this sum is nonzero only if for every $i$, either $(v_i, v_{\pi(i)}) \in E$ or $A_{i,\pi(i)} = 0$ is not required. Since $Q \not\equiv 0$, there exists at least one permutation $\pi$ such that $\prod_{i=1}^{n} A_{i,\pi(i)}$ is a nonzero monomial.

For this product to be nonzero, we need $A_{i,\pi(i)} \neq 0$ for all $i$. This means $(v_i, v_{\pi(i)}) \in E$ for all $i$. Moreover, since $A$ is skew-symmetric with zeros on the diagonal, we have $\pi(i) \neq i$ for all $i$.

Furthermore, the structure of $A$ implies that if $\pi(i) = j$ and $A_{i,j} \neq 0$, then $A_{j,i} = -A_{i,j} \neq 0$, so $(v_j, v_i) \in E$ as well. The fact that $Q \not\equiv 0$ as a polynomial means there exists a permutation $\pi$ contributing a nonzero monomial.

For such a permutation to contribute to $\det(A)$ with a square-free monomial pattern (which is necessary given the structure of the Tutte matrix), the permutation must consist entirely of transpositions $(i_k, j_k)$ where $i_k < j_k$ and $(v_{i_k}, v_{j_k}) \in E$. These transpositions form a perfect matching of $G$.

Therefore, $G$ has a perfect matching.
\end{proof}

\begin{proof}[Proof of Edmonds' Theorem]
We prove both directions of the equivalence for the bipartite graph $G = (U, V, E)$ where $U = \{u_1, \ldots, u_n\}$ and $V = \{v_1, \ldots, v_n\}$.

($\Leftarrow$) Suppose $G$ has a perfect matching $M$. Then $M$ consists of $n$ edges, each connecting a vertex in $U$ to a vertex in $V$, and each vertex appears in exactly one edge of $M$.

The matching $M$ defines a bijection $\pi: \{1, \ldots, n\} \to \{1, \ldots, n\}$ where $(u_i, v_{\pi(i)}) \in M$ for each $i$. By the Leibniz formula,
\[
\det(A) = \sum_{\sigma \in S_n} \text{sgn}(\sigma) \prod_{i=1}^{n} A_{i,\sigma(i)}
\]

Consider the permutation $\pi$ corresponding to the matching $M$. The term in the determinant is
\[
\text{sgn}(\pi) \prod_{i=1}^{n} A_{i,\pi(i)} = \text{sgn}(\pi) \prod_{i=1}^{n} x_{i,\pi(i)}
\]

Since $(u_i, v_{\pi(i)}) \in E$ for all $i$, we have $A_{i,\pi(i)} = x_{i,\pi(i)}$ for all $i$. Therefore, this term equals
\[
\text{sgn}(\pi) \prod_{i=1}^{n} x_{i,\pi(i)}
\]

This is a nonzero monomial in the polynomial $Q$. Since the indeterminates $x_{ij}$ are distinct for different pairs $(i,j)$, and this monomial involves the specific variables $x_{i,\pi(i)}$, it cannot be canceled by terms from other permutations (which involve different sets of variables). Hence, $Q \not\equiv 0$.

($\Rightarrow$) Suppose $Q \not\equiv 0$. We show that $G$ has a perfect matching.

Since $Q = \det(A) \not\equiv 0$ as a polynomial in the indeterminates $x_{ij}$, there exists at least one permutation $\sigma \in S_n$ such that $\prod_{i=1}^{n} A_{i,\sigma(i)}$ is a nonzero monomial.

For this product to be nonzero, we need $A_{i,\sigma(i)} \neq 0$ for all $i \in \{1, \ldots, n\}$. By the definition of $A$, this means $(u_i, v_{\sigma(i)}) \in E$ for all $i$.

This permutation $\sigma$ defines a bijection from $U$ to $V$ such that $(u_i, v_{\sigma(i)}) \in E$ for all $i$. The set $M = \{(u_i, v_{\sigma(i)}) : i = 1, \ldots, n\}$ forms a perfect matching in $G$, since:
\begin{itemize}
\item Each edge in $M$ is in $E$ by construction.
\item Each vertex $u_i \in U$ is incident to exactly one edge $(u_i, v_{\sigma(i)}) \in M$.
\item Each vertex $v_j \in V$ is incident to exactly one edge $(u_{\sigma^{-1}(j)}, v_j) \in M$ since $\sigma$ is a bijection.
\end{itemize}

Therefore, $G$ has a perfect matching.
\end{proof}

\subsection{Problem 9}

\subsubsection{Problem Statement}
\begin{theorem}[Rabin--Vazirani]
Let $G(V, E)$ be a non-bipartite graph and let $T$ be the Tutte matrix of $G$. Then
\[
\operatorname{rank}(T) = 2 \cdot \nu(G),
\]
where $\nu(G)$ denotes the size of a maximum matching in $G$.
\end{theorem}

\begin{lemma}
Let $A$ be an $n \times n$ skew-symmetric matrix of rank $r$. For any two sets $S, T \subseteq \{1, \ldots, n\}$, denote by $A_{ST}$ the sub-matrix of $A$ obtained by including only the rows with indices in $S$ and columns with indices in $T$. Then, for any two sets $S, T \subseteq \{1, \ldots, n\}$ of size $r$,
\[
\det(A_{SS}) \times \det(A_{TT}) = \det(A_{ST}) \times \det(A_{TS}).
\]
\end{lemma}

\subsubsection{GPT}

\subsubsection{Grok}

\subsubsection{Gemini}

\subsubsection{Claude}
\begin{proof}
We prove the lemma by induction on $n$, the size of the matrix $A$.

\textbf{Base case:} If $n = r$, then $S = T = \{1, \ldots, n\}$, and the statement becomes $\det(A) \times \det(A) = \det(A) \times \det(A)$, which is trivially true.

\textbf{Inductive step:} Assume the statement holds for all skew-symmetric matrices of size less than $n$, and let $A$ be an $n \times n$ skew-symmetric matrix of rank $r < n$.

Since $\operatorname{rank}(A) = r < n$, there exists a non-empty set $K \subseteq \{1, \ldots, n\}$ such that the rows indexed by $K$ are in the kernel of $A^T = -A$. Choose $k \in K$.

For any set $U \subseteq \{1, \ldots, n\}$, we can expand the determinant $\det(A_{UU})$ along row $k$ (if $k \in U$). Since $A$ is skew-symmetric, we have $A_{ii} = 0$ for all $i$, so the diagonal entry corresponding to row $k$ vanishes. Thus:
\[
\det(A_{UU}) = \sum_{j \in U, j \neq k} (-1)^{\sigma(k,j)} A_{kj} \det(A_{U \setminus \{k,j\}, U \setminus \{k,j\}}),
\]
where $\sigma(k,j)$ denotes the appropriate sign based on the positions of $k$ and $j$ in the ordered set $U$.

Now, since row $k$ is in the left kernel of $A$, we have $\sum_{j=1}^{n} A_{jk} B_j = 0$ for any vector $(B_1, \ldots, B_n)^T$ in the column space of $A$. This implies that for any subset $U$ containing $k$, if we consider the cofactor expansion, the contribution from row $k$ must vanish in a certain sense when $k \in K$.

More precisely, for $k \in K$, we have that $A_{k,\cdot}$ (the $k$-th row of $A$) is a linear combination of the other rows. Since $A$ is skew-symmetric, this means the $k$-th column is also a linear combination of the other columns.

Consider the matrix $A'$ obtained by deleting row $k$ and column $k$ from $A$. Then $A'$ is an $(n-1) \times (n-1)$ skew-symmetric matrix with $\operatorname{rank}(A') = r$ (since removing a row and column from the kernel does not decrease the rank by more than 2, and in this case the rank remains $r$ because $k$ was in the kernel).

For sets $S, T \subseteq \{1, \ldots, n\}$ of size $r$, we consider two cases:

\textbf{Case 1:} $k \notin S$ and $k \notin T$. Then $S, T \subseteq \{1, \ldots, n\} \setminus \{k\}$, and we can apply the inductive hypothesis directly to the matrix $A'$ (obtained by deleting row and column $k$), giving us:
\[
\det(A'_{SS}) \times \det(A'_{TT}) = \det(A'_{ST}) \times \det(A'_{TS}),
\]
which is exactly $\det(A_{SS}) \times \det(A_{TT}) = \det(A_{ST}) \times \det(A_{TS})$.

\textbf{Case 2:} $k \in S$ or $k \in T$ (or both). 

If $k \in S$, then since row $k$ is in the kernel and $|S| = r = \operatorname{rank}(A)$, the rows of $A$ indexed by $S$ must be linearly dependent. Therefore, $\det(A_{SS}) = 0$.

Similarly, if $k \in T$, then $\det(A_{TT}) = 0$.

If $k \in S$, then the rows indexed by $S$ in $A_{ST}$ are linearly dependent (since row $k$ is in the kernel), so $\det(A_{ST}) = 0$.

If $k \in T$, then the columns indexed by $T$ in $A_{ST}$ are linearly dependent, so $\det(A_{ST}) = 0$.

By the same reasoning, $\det(A_{TS}) = 0$ if $k \in S$ or $k \in T$.

Therefore, in Case 2, both sides of the equation equal zero, and the statement holds.

By induction, the lemma is proved for all $n \geq r$.
\end{proof}

\subsection{Problem 10}

\subsubsection{Problem Statement}
\begin{problem}
Given a randomized algorithm for testing the existence of a perfect matching in a graph $G$, describe how you would actually construct such a matching. Assuming that you use the randomized testing algorithm from Problem 7.9, compare the running time of your approach with the deterministic algorithm perfect matching mentioned in the Notes section.
\end{problem}

\subsubsection{GPT}

\subsubsection{Grok}

\subsubsection{Gemini}

\subsubsection{Claude}
\begin{proof}
We describe a method to construct a perfect matching using a randomized algorithm for testing the existence of a perfect matching, then analyze its running time.

\textbf{Construction Algorithm:}

Given a graph $G = (V, E)$ with $n$ vertices and a randomized algorithm $\mathcal{A}$ that tests for the existence of a perfect matching, we construct a perfect matching as follows.

For each edge $e \in E$, we test whether $G$ has a perfect matching that includes $e$. This is equivalent to testing whether $G \setminus \{e\}$ has a perfect matching that uses $e$, or equivalently, whether the graph $G' = G / e$ (the graph obtained by contracting $e$ and removing the endpoints of $e$) has a perfect matching on the remaining $n-2$ vertices.

More precisely, let $e = \{u, v\}$. We remove $u$ and $v$ from $G$ and test whether the resulting graph $G - \{u, v\}$ has a perfect matching. If it does, then edge $e$ can be part of a perfect matching in $G$.

The algorithm proceeds as follows:

Initialize $M = \emptyset$ and $G_0 = G$.

For $i = 1$ to $n/2$:

$\quad$ For each edge $e = \{u, v\} \in E(G_{i-1})$:

$\quad\quad$ Let $G' = G_{i-1} - \{u, v\}$.

$\quad\quad$ Use $\mathcal{A}$ to test if $G'$ has a perfect matching.

$\quad\quad$ If yes, add $e$ to $M$, set $G_i = G'$, and break.

Return $M$.

\textbf{Correctness:}

At each iteration $i$, if $G_{i-1}$ has a perfect matching, then there exists at least one edge $e$ such that $G_{i-1} - \{u, v\}$ (where $e = \{u, v\}$) has a perfect matching. This is because any perfect matching of $G_{i-1}$ must include some edge $e$, and removing the endpoints of $e$ leaves a graph that has a perfect matching on the remaining vertices. By testing all edges, we find such an edge and add it to $M$. After $n/2$ iterations, we have constructed a perfect matching.

\textbf{Running Time Analysis:}

Suppose the randomized testing algorithm from Problem 7.9 uses the Tutte matrix approach with polynomial identity testing. The standard approach involves constructing the Tutte matrix $T$ where $T_{ij} = x_{ij}$ if $\{i, j\} \in E$ and $T_{ij} = 0$ otherwise, with $T_{ji} = -x_{ij}$. The graph has a perfect matching if and only if $\det(T) \not\equiv 0$ as a polynomial.

Using the Schwartz-Zippel lemma, we can test whether $\det(T) \not\equiv 0$ by evaluating it at a random point. Computing the determinant takes $O(n^\omega)$ time, where $\omega < 2.373$ is the matrix multiplication exponent. The algorithm succeeds with high probability (say, probability at least $1 - 1/n$ with appropriate field size).

For the construction algorithm:

At iteration $i$, we have $|E(G_{i-1})| \leq m$ edges to test, where $m = |E|$. Each test requires running the randomized algorithm once, taking $O(n^\omega)$ time. Thus, each iteration takes $O(m \cdot n^\omega)$ time.

We perform $n/2$ iterations, so the total running time is $O(n \cdot m \cdot n^\omega) = O(m \cdot n^{\omega+1})$.

For a graph with $m$ edges, this gives running time $O(m \cdot n^{\omega+1})$.

\textbf{Comparison with Deterministic Algorithm:}

The best known deterministic algorithms for finding a perfect matching (such as those based on Edmonds' blossom algorithm or more recent approaches) run in time $O(n^{\omega})$ or $O(m \cdot n)$ for sparse graphs.

Our randomized construction algorithm runs in time $O(m \cdot n^{\omega+1})$, which is a factor of $O(m \cdot n)$ slower than the $O(n^{\omega})$ deterministic algorithm. For dense graphs where $m = \Theta(n^2)$, this becomes $O(n^{\omega+3})$, which is significantly slower.

The key inefficiency is that we must test up to $m$ edges at each of the $n/2$ iterations, rather than using the structural information that deterministic algorithms exploit. However, the algorithm is simple and uses only the black-box randomized testing procedure, demonstrating that construction can be achieved from testing with polynomial overhead.
\end{proof}

\subsection{Problem 11}

\subsubsection{Problem Statement}
\begin{theorem}
Let $G = (V, E)$ be a graph, and suppose we have a randomized algorithm for testing the existence of a perfect matching in $G$. Then we can construct a maximum matching in $G$ as follows:

For each edge $e \in E$, use the randomized algorithm to test whether $G - e$ has a perfect matching. If $G - e$ does not have a perfect matching, then $e$ belongs to every perfect matching of $G$, and we include $e$ in our maximum matching. After identifying all such edges that must be in every perfect matching, we remove them along with their incident vertices to obtain a reduced graph. We repeat this process on the reduced graph until no perfect matching exists. The set of edges identified in this manner forms a maximum matching of $G$.
\end{theorem}

\subsubsection{GPT}
\begin{proof}
Let $G = (V, E)$ be a graph, and suppose we have a randomized algorithm for testing the existence of a perfect matching in $G$. We aim to show that we can construct a maximum matching in $G$ using this algorithm.

First, recall that a perfect matching in a graph $G$ is a matching that covers every vertex of $G$, meaning that every vertex is incident to exactly one edge in the matching. A maximum matching, on the other hand, is a matching that contains the largest possible number of edges, i.e., no matching in $G$ can contain more edges.

The strategy is to iteratively find augmenting paths and use the randomized algorithm to test for the existence of a perfect matching in progressively smaller subgraphs.

Let $M$ denote a matching in $G$. Initially, we set $M$ to be the empty matching, $M = \emptyset$. We then perform the following steps:

1. **Check for perfect matching in $G$:**
   Use the randomized algorithm to check if a perfect matching exists in the current graph $G$.

2. **If a perfect matching exists:**
   If the randomized algorithm reports that a perfect matching exists, then $G$ already contains a perfect matching, and we terminate with the current matching $M$ as a maximum matching.

3. **If a perfect matching does not exist:**
   If the randomized algorithm reports that no perfect matching exists, we proceed to find an augmenting path. An augmenting path is a path whose endpoints are free vertices (not matched) and whose edges alternate between being in the current matching and not being in the current matching. We use the standard algorithmic technique for finding augmenting paths, which runs in polynomial time.

4. **Augment the matching:**
   Once we find an augmenting path, we add it to $M$, thereby increasing the size of the matching. After augmentation, we repeat the check for a perfect matching in the updated graph.

5. **Repeat the process:**
   Continue this process of testing for a perfect matching and augmenting the matching until no more augmenting paths can be found. At this point, $M$ is a maximum matching in $G$.

Since the randomized algorithm is used only to test the existence of a perfect matching and not to construct the matching itself, each iteration of the process either confirms that the current matching is maximum or provides an augmenting path that increases the matching size. Thus, after a finite number of steps, we obtain a maximum matching in $G$.

Therefore, we have constructed a maximum matching in $G$ using the randomized algorithm for testing the existence of a perfect matching.
\end{proof}

\subsubsection{Grok}

\subsubsection{Gemini}
\begin{proof}
Let $\mathcal{A}$ denote the randomized algorithm that decides whether a given graph contains a perfect matching. We assume, without loss of generality, that $\mathcal{A}$ outputs correct answers with high probability, which can be amplified to be arbitrarily close to 1 by repeated independent trials. We construct a maximum matching in $G = (V, E)$ by reducing the search problem to the decision problem solved by $\mathcal{A}$. Let $n = |V|$.

First, we determine the size of the maximum matching in $G$. For any integer $k$ such that $0 \le k \le \lfloor n/2 \rfloor$, we construct an auxiliary graph $G_k$ as follows. Let $U_k = \{u_1, \dots, u_{n-2k}\}$ be a set of $n-2k$ new vertices. We define the vertex set of $G_k$ as $V(G_k) = V \cup U_k$. The edge set $E(G_k)$ consists of all edges in $E$ plus edges connecting every vertex in $U_k$ to every vertex in $V$. Formally, $E(G_k) = E \cup \{(u, v) \mid u \in U_k, v \in V\}$.

We claim that $G$ contains a matching of size $k$ if and only if $G_k$ contains a perfect matching. Suppose $M$ is a matching of size $k$ in $G$. The set $M$ covers $2k$ vertices in $V$, leaving $n - 2k$ vertices in $V$ unmatched. Let $V_{unmatched} \subseteq V$ be this set of unmatched vertices. Since $|V_{unmatched}| = |U_k| = n - 2k$, and every vertex in $U_k$ is connected to every vertex in $V$, we can construct a bijection between $U_k$ and $V_{unmatched}$. The union of $M$ and the edges defined by this bijection forms a perfect matching in $G_k$. Conversely, suppose $M'$ is a perfect matching in $G_k$. Since $U_k$ is an independent set, every vertex in $U_k$ must be matched to a vertex in $V$. Let $M'_{U}$ be the set of edges in $M'$ incident to $U_k$. Then $|M'_{U}| = n - 2k$. The remaining edges $M = M' \setminus M'_{U}$ must connect vertices in $V$ to vertices in $V$. The number of such edges is $\frac{1}{2}(|V(G_k)|) - (n-2k) = \frac{1}{2}(2n - 2k) - (n - 2k) = (n - k) - (n - 2k) = k$. Thus, $M$ is a matching of size $k$ in $G$.

Using the algorithm $\mathcal{A}$, we can test whether $G_k$ has a perfect matching for any $k$. Since the size of a maximum matching lies in the range $[0, \lfloor n/2 \rfloor]$, we can find the size of the maximum matching, denoted by $k^*$, using binary search or linear search by querying $\mathcal{A}$ on the graphs $G_k$.

Once $k^*$ is determined, we construct the actual matching edges. We focus on the graph $H = G_{k^*}$, which is known to possess a perfect matching. We iterate through the edges of $H$ to isolate a perfect matching. Let $F$ be the set of edges in $H$. We initialize a set $P = \emptyset$ to store the perfect matching edges and maintain the current graph $H$. While $F$ is not empty, we select an edge $e = (u, v) \in F$. We query $\mathcal{A}$ to determine if the graph $H - \{e\}$ (the graph $H$ with the edge $e$ removed but vertices retained) has a perfect matching.

If $\mathcal{A}$ outputs "Yes" for $H - \{e\}$, then the edge $e$ is not essential for the existence of a perfect matching in the current graph. We remove $e$ from $H$ (update $F \leftarrow F \setminus \{e\}$) and proceed. If $\mathcal{A}$ outputs "No", then every perfect matching in the current graph $H$ must contain the edge $e$. Therefore, we include $e$ in our matching $P$ (update $P \leftarrow P \cup \{e\}$), and we remove the vertices $u$ and $v$ along with all incident edges from $H$. We then continue the process with the reduced graph.

By the end of this procedure, the set $P$ constitutes a perfect matching of the auxiliary graph $G_{k^*}$. Finally, we obtain the maximum matching $M^*$ for the original graph $G$ by restricting $P$ to the edges of $G$: $M^* = P \cap E$. By the correspondence established earlier, $|M^*| = k^*$, and thus $M^*$ is a maximum matching in $G$.
\end{proof}

\subsubsection{Claude}
\begin{proof}
We prove the correctness of the algorithm by establishing that the procedure correctly identifies a maximum matching in $G$.

Let $M$ denote the set of edges selected by the algorithm. We must prove that $M$ is a maximum matching of $G$.

First, we establish that $M$ is indeed a matching. Observe that at each iteration, when we identify an edge $e$ that must belong to every perfect matching of the current graph, we immediately remove $e$ and its incident vertices from the graph before proceeding. Since each edge removal also removes its two endpoints, no two edges selected in this manner share a common vertex. Therefore, $M$ is a matching.

Next, we prove that $M$ is maximum. We proceed by analyzing the algorithm's behavior and showing that it produces an optimal solution.

Consider the algorithm's execution. At any stage, let $G'$ denote the current reduced graph. The algorithm terminates when $G'$ has no perfect matching. We claim that at termination, $G'$ has no augmenting paths with respect to $M$.

To see this, suppose for contradiction that at termination there exists an augmenting path $P$ in $G'$ with respect to $M$. An augmenting path is a path that alternates between edges not in $M$ and edges in $M$, beginning and ending with edges not in $M$, and whose endpoints are unmatched by $M$. The existence of such a path would allow us to construct a larger matching $M' = M \triangle P$ (the symmetric difference), contradicting the maximality we wish to establish.

However, we must examine this more carefully. Since the algorithm removes edges and vertices incrementally, we need to verify that the final matching $M$ cannot be improved.

Let us establish the key invariant: at each stage of the algorithm, if the current graph $G'$ has a perfect matching, then any maximum matching of the original graph $G$ restricted to the remaining vertices of $G'$ corresponds to a perfect matching of $G'$.

Suppose the algorithm identifies an edge $e = \{u, v\}$ such that $G' - e$ has no perfect matching, but $G'$ has a perfect matching. This means that every perfect matching of $G'$ must contain $e$. By the maximality of perfect matchings, $e$ must be in every maximum matching of $G$ that covers the vertices currently in $G'$.

When we include $e$ in $M$ and remove $u$, $v$, and all incident edges from $G'$, we obtain a graph $G''$. If $G'$ had a perfect matching, then $G''$ has a perfect matching (obtained by removing $e$ from any perfect matching of $G'$).

We continue this process until the reduced graph has no perfect matching. At this point, let the final reduced graph be $G_f$ and let $V_f$ denote its vertex set.

Now we verify that $M$ is maximum. The edges in $M$ match all vertices in $V \setminus V_f$. For any maximum matching $M^*$ of $G$, consider the vertices covered by $M^*$. If $|M^*| > |M|$, then $M^*$ covers at least two more vertices than $M$ does. However, by our construction, every edge we added to $M$ was essential for any perfect matching of the graph at that stage. This means that in the original graph $G$, restricted to the vertices that were present when we selected those edges, those edges must appear in any maximum matching covering those vertices.

Consider the graph $G_f$ at termination. Since $G_f$ has no perfect matching, any matching of $G_f$ leaves at least one vertex unmatched. By König's theorem and the theory of matchings, the size of a maximum matching in $G$ equals $|M| + $ the size of a maximum matching in $G_f$. Since we identified all forced edges (those in every perfect matching), and since $G_f$ has no perfect matching, we have extracted the maximum possible matching.

To formalize this final step: the number of vertices in $V \setminus V_f$ is $2|M|$ since each edge in $M$ matches exactly two vertices and these vertices are removed. The maximum matching in $G_f$ has size at most $\lfloor |V_f|/2 \rfloor$. Since $G_f$ has no perfect matching, if $|V_f| > 0$, then the maximum matching in $G_f$ has size strictly less than $|V_f|/2$. The total size of a maximum matching in $G$ is at most $|M| + \lfloor |V_f|/2 \rfloor$. 

By the algorithm's construction and the fact that we only stopped when no perfect matching exists in the reduced graph, we have extracted the maximum number of edges possible. Any maximum matching $M^*$ must agree with $M$ on all the forced edges we identified, and thus $|M^*| = |M|$.

Therefore, $M$ is a maximum matching of $G$.
\end{proof}

\subsection{Problem 12}

\subsubsection{Problem Statement}
\begin{theorem}[Karp-Rabin Fingerprint Function Properties]
Let $M: \{0,1\}^* \to \mathbb{Z}^{2 \times 2}$ be the fingerprint function defined by:
\begin{itemize}
\item $M(\epsilon) = \begin{pmatrix} 1 & 0 \\ 0 & 1 \end{pmatrix}$
\item $M(0) = \begin{pmatrix} 1 & 0 \\ 1 & 1 \end{pmatrix}$
\item $M(1) = \begin{pmatrix} 1 & 1 \\ 0 & 1 \end{pmatrix}$
\item $M(xy) = M(x) \times M(y)$ for non-empty strings $x, y \in \{0,1\}^*$
\end{itemize}
Then $M$ satisfies the following properties:
\begin{enumerate}
\item $M(x)$ is well-defined for all $x \in \{0,1\}^*$.
\item $M(x) = M(y) \implies x = y$.
\item For $x \in \{0,1\}^n$, the entries in $M(x)$ are bounded by the Fibonacci number $F_n$.
\end{enumerate}
\end{theorem}

\subsubsection{GPT}
\begin{proof}
Define $M:\{0,1\}^*\to \mathrm{Mat}_{2\times 2}(\mathbb{Z})$ by $M(\varepsilon)=I_2$, $M(0)=\begin{bmatrix}1&0\\ 1&1\end{bmatrix}$, $M(1)=\begin{bmatrix}1&1\\ 0&1\end{bmatrix}$, and $M(xy)=M(x)M(y)$ for nonempty $x,y$. Since matrix multiplication is associative, the recursion on the length of the string is unambiguous: for any $x=b_1\cdots b_n$ with $b_i\in\{0,1\}$, the value $M(x)=M(b_1)\cdots M(b_n)$ is independent of the parenthesization, hence $M$ is well defined on all of $\{0,1\}^*$. Moreover, each $M(b)$ has nonnegative integer entries and determinant $1$, and these properties are preserved under multiplication; thus every $M(x)$ has nonnegative integer entries and determinant $1$.

We next prove injectivity. Write $M(x)=\begin{bmatrix}a&b\\ c&d\end{bmatrix}$ and let the two column vectors be $u=\begin{bmatrix}a\\ c\end{bmatrix}$ and $v=\begin{bmatrix}b\\ d\end{bmatrix}$. The right-append recurrences, valid for every $x$, are
\[
M(x0)=M(x)M(0)=\begin{bmatrix}a+b&b\\ c+d&d\end{bmatrix},\qquad
M(x1)=M(x)M(1)=\begin{bmatrix}a&a+b\\ c&c+d\end{bmatrix}.
\]
Hence, appending $0$ replaces the first column by $u+v$ and leaves the second column $v$ unchanged, while appending $1$ leaves the first column $u$ unchanged and replaces the second column by $u+v$. From these recurrences we derive the following column-dominance dichotomy. If $x\neq \varepsilon$, then exactly one of the two (mutually exclusive) componentwise inequalities holds:
\[
\text{either } u\ge v \text{ (i.e., } a\ge b \text{ and } c\ge d\text{), or } v\ge u \text{ (i.e., } b\ge a \text{ and } d\ge c\text{),}
\]
and in either case at least one of the two scalar inequalities is strict. This is proved by induction on $|x|$. It is obvious for $|x|=1$ from the explicit forms of $M(0)$ and $M(1)$. If it holds for $x$, then for $x0$ one has $(u',v')=(u+v,v)$, so $u'\ge v'$ with at least one strict inequality because $x\neq \varepsilon$ implies $u\neq 0$ or $v\neq 0$; similarly, for $x1$ one has $(u',v')=(u,u+v)$, and then $v'\ge u'$ with at least one strict inequality. The mutual exclusivity follows since both $u\ge v$ and $v\ge u$ would force $u=v$, which only occurs for $M(\varepsilon)=I_2$.

This dichotomy furnishes a reversible step. Suppose $A=\begin{bmatrix}a&b\\ c&d\end{bmatrix}=M(z)$ with $z\neq \varepsilon$. If $u\ge v$, then $z$ ends in $0$, and
\[
M(z')=M(z0^{-1})=\begin{bmatrix}a-b&b\\ c-d&d\end{bmatrix}\quad\text{has nonnegative integer entries and}\quad M(z')M(0)=A,
\]
because $a\ge b$ and $c\ge d$; if $v\ge u$, then $z$ ends in $1$, and
\[
M(z')=M(z1^{-1})=\begin{bmatrix}a&b-a\\ c&d-c\end{bmatrix}\quad\text{has nonnegative integer entries and}\quad M(z')M(1)=A,
\]
because $b\ge a$ and $d\ge c$. In both cases, $\det M(z')=\det A=1$, and the sum of the entries strictly decreases, so the reversal terminates.

We now prove injectivity by strong induction on $n=|x|+|y|$. The case $n=0$ is trivial. Assume $M(x)=M(y)$ with $|x|+|y|=n>0$. If $x=\varepsilon$ or $y=\varepsilon$, the equality with nonnegative entries forces the other word also to be empty, hence $x=y$. Otherwise, apply the dichotomy to $A=M(x)=M(y)$. If $u\ge v$, both $x$ and $y$ end in $0$, and the reversible step yields $M(x')=M(y')$ with $|x'|+|y'|=n-2$, hence $x'=y'$ by the induction hypothesis, so $x=y0=y0$. The case $v\ge u$ is analogous and implies both end in $1$; again the induction hypothesis yields $x=y$. Therefore, $M$ is injective.

Finally, we prove the Fibonacci bound. Let $(F_n)_{n\ge 1}$ be the Fibonacci sequence defined by $F_1=1$, $F_2=2$, and $F_{n+1}=F_n+F_{n-1}$ for $n\ge 2$. For $x\in\{0,1\}^*$ write $M(x)=\begin{bmatrix}a&b\\ c&d\end{bmatrix}$ and set $s_1=a+c$ and $s_2=b+d$ (the column sums). From the right-append recurrences, one obtains
\[
(s_1,s_2)\mapsto (s_1+s_2,s_2)\quad\text{when appending }0,\qquad (s_1,s_2)\mapsto (s_1,s_1+s_2)\quad\text{when appending }1,
\]
with the initial value $(s_1,s_2)=(1,1)$ at $x=\varepsilon$. A simple induction on the length shows that for every $x$ of length $n\ge 1$, the unordered pair $\{s_1,s_2\}$ equals $\{F_{n-1},F_n\}$, whence $\max\{s_1,s_2\}=F_n$. Since each entry of $M(x)$ is a nonnegative summand of one of the two column sums, every entry is bounded above by $\max\{s_1,s_2\}=F_n$. This proves the stated bound.

All three assertions are thus established: $M$ is well defined, injective, and its values on words of length $n$ have entries bounded above by $F_n$.
\end{proof}

\subsubsection{Grok}

\subsubsection{Gemini}
\begin{proof}
We define the matrices $L = M(0) = \begin{pmatrix} 1 & 0 \\ 1 & 1 \end{pmatrix}$ and $R = M(1) = \begin{pmatrix} 1 & 1 \\ 0 & 1 \end{pmatrix}$. The identity matrix is denoted by $I$.

\noindent \textbf{1. Well-definedness.}
The set $\{0,1\}^*$ is the free monoid generated by the alphabet $\Sigma = \{0,1\}$. The function $M$ is defined by mapping the generators $0 \mapsto L$ and $1 \mapsto R$, and the empty string $\epsilon \mapsto I$. The recursive definition $M(xy) = M(x)M(y)$ extends this mapping to the entire set $\{0,1\}^*$ as a monoid homomorphism from $\{0,1\}^*$ to the multiplicative monoid of $2 \times 2$ integer matrices. Since matrix multiplication is associative ($A(BC) = (AB)C$), the mapping is consistent and uniquely determined for any finite string $x$. Thus, $M(x)$ is well-defined for all $x \in \{0,1\}^*$.

\noindent \textbf{2. Injectivity.}
We show that $M$ is injective by constructing a left-inverse. First, observe that $\det(L) = \det(R) = 1$. Since $\det(I)=1$ and the determinant is multiplicative, $\det(M(x)) = 1$ for all $x$. Furthermore, since $L$ and $R$ contain only non-negative entries, $M(x)$ contains non-negative entries for all $x$.
Let $A = M(x) = \begin{pmatrix} a & b \\ c & d \end{pmatrix}$.
Consider the column sums of $A$. Let $C_1 = a+c$ and $C_2 = b+d$.
If $x = \epsilon$, $A=I$, so $C_1 = 1, C_2 = 1$.
If $x = w0$, then $A = M(w)L = \begin{pmatrix} a' & b' \\ c' & d' \end{pmatrix} \begin{pmatrix} 1 & 0 \\ 1 & 1 \end{pmatrix} = \begin{pmatrix} a'+b' & b' \\ c'+d' & d' \end{pmatrix}$.
Here, $C_1 = (a'+b') + (c'+d') = (a'+c') + (b'+d') = C_1' + C_2'$ and $C_2 = C_2'$. Since $M(w)$ is non-singular and has non-negative entries, $C_1' + C_2' > C_2'$ (as the columns cannot be zero vectors). Thus, $C_1 > C_2$.
If $x = w1$, then $A = M(w)R = \begin{pmatrix} a' & b' \\ c' & d' \end{pmatrix} \begin{pmatrix} 1 & 1 \\ 0 & 1 \end{pmatrix} = \begin{pmatrix} a' & a'+b' \\ c' & c'+d' \end{pmatrix}$.
Here, $C_1 = C_1'$ and $C_2 = C_1' + C_2'$. Thus, $C_2 > C_1$.

We define a reconstruction procedure for any image matrix $A$ in the range of $M$:
If $A = I$, then $x = \epsilon$.
If the first column sum is strictly greater than the second, the last symbol of $x$ must be $0$. We compute $A' = A L^{-1}$ and recurse.
If the second column sum is strictly greater than the first, the last symbol of $x$ must be $1$. We compute $A' = A R^{-1}$ and recurse.
Since $\det(A)=1$, the case $C_1 = C_2$ only occurs for $A=I$. This procedure uniquely recovers the string $x$ from right to left. Therefore, $M(x) = M(y)$ implies $x = y$.

\noindent \textbf{3. Fibonacci Bounds.}
Let $F_n$ denote the Fibonacci sequence defined by $F_0=0, F_1=1$, and $F_{k} = F_{k-1} + F_{k-2}$ for $k \ge 2$. We prove that for any string $x$ of length $n$, the entries $m_{ij}$ of $M(x)$ satisfy $m_{ij} \le F_{n+1}$. We prove a stronger hypothesis by induction on $n$:
Let $P(n)$ be the statement: For any $x \in \{0,1\}^n$, if $M(x) = \begin{pmatrix} a & b \\ c & d \end{pmatrix}$, then all entries $a,b,c,d \le F_{n+1}$, and the row sums satisfy $a+b \le F_{n+2}$ and $c+d \le F_{n+2}$.

\textit{Base Case ($n=0$):} $x = \epsilon$, $M(\epsilon) = \begin{pmatrix} 1 & 0 \\ 0 & 1 \end{pmatrix}$.
Entries are $\{0,1\}$. $F_{0+1} = F_1 = 1$. Condition holds.
Row sums are $1$. $F_{0+2} = F_2 = 1$. Condition holds.

\textit{Inductive Step:} Assume $P(n)$ holds. Let $y \in \{0,1\}^{n+1}$. Then $y = x0$ or $y = x1$ for some $x \in \{0,1\}^n$. Let $M(x) = \begin{pmatrix} a & b \\ c & d \end{pmatrix}$. By hypothesis, $a,b,c,d \le F_{n+1}$ and $a+b, c+d \le F_{n+2}$.

Case 1: $y = x0$. Then $M(y) = M(x)L = \begin{pmatrix} a+b & b \\ c+d & d \end{pmatrix}$.
The entries are $a+b, b, c+d, d$.
From $P(n)$, we have $b, d \le F_{n+1} \le F_{n+2}$ and $a+b, c+d \le F_{n+2}$. Thus, the entry bound for $n+1$ (which is $F_{n+2}$) is satisfied.
Now consider row sums of $M(y)$:
Row 1: $(a+b) + b = a + 2b$.
Row 2: $(c+d) + d = c + 2d$.
We know $a+b \le F_{n+2}$ and $b \le F_{n+1}$. Summing these inequalities:
$(a+b) + b \le F_{n+2} + F_{n+1} = F_{n+3}$.
Similarly, $c+2d \le F_{n+3}$.
Thus, the row sum bound for $n+1$ (which is $F_{n+3}$) is satisfied.

Case 2: $y = x1$. Then $M(y) = M(x)R = \begin{pmatrix} a & a+b \\ c & c+d \end{pmatrix}$.
The entries are $a, a+b, c, c+d$.
From $P(n)$, $a, c \le F_{n+1} \le F_{n+2}$ and $a+b, c+d \le F_{n+2}$. The entry bound is satisfied.
Row sums of $M(y)$:
Row 1: $a + (a+b) = 2a + b$.
Row 2: $c + (c+d) = 2c + d$.
We know $a+b \le F_{n+2}$ and $a \le F_{n+1}$.
Summing: $(a+b) + a \le F_{n+2} + F_{n+1} = F_{n+3}$.
Similarly, $2c+d \le F_{n+3}$.
The row sum bound is satisfied.

By induction, $P(n)$ holds for all $n \ge 0$. Therefore, for any $x \in \{0,1\}^n$, the entries of $M(x)$ are bounded by $F_{n+1}$.
\end{proof}

\subsubsection{Claude}
\begin{proof}
We prove each property in turn.

\textbf{Property 1: $M(x)$ is well-defined for all $x \in \{0,1\}^*$.}

We proceed by structural induction on the length of $x$.

\textit{Base case:} For $x = \epsilon$, we have $M(\epsilon) = \begin{pmatrix} 1 & 0 \\ 0 & 1 \end{pmatrix}$ by definition, which is well-defined.

\textit{Inductive step:} Assume $M(x)$ is well-defined for all strings of length at most $n$. Consider a string $w$ of length $n+1$. We can write $w = xy$ where $x \in \{0,1\}^n$ and $y \in \{0,1\}$. By the inductive hypothesis, $M(x)$ is well-defined. Since $y \in \{0,1\}$, either $y = 0$ or $y = 1$, and $M(0)$ and $M(1)$ are defined by the given rules. By the multiplication rule, $M(w) = M(xy) = M(x) \times M(y)$, which is the product of two well-defined $2 \times 2$ matrices over $\mathbb{Z}$, hence well-defined. Therefore, by induction, $M(x)$ is well-defined for all $x \in \{0,1\}^*$.

\textbf{Property 2: $M(x) = M(y) \implies x = y$.}

We prove this by showing that $M$ is injective. First, observe that
\[
M(0) = \begin{pmatrix} 1 & 0 \\ 1 & 1 \end{pmatrix} \neq \begin{pmatrix} 1 & 1 \\ 0 & 1 \end{pmatrix} = M(1).
\]

Next, we establish key properties of the matrices. Let $M(x) = \begin{pmatrix} a & b \\ c & d \end{pmatrix}$. We compute:
\[
M(x0) = M(x) \times M(0) = \begin{pmatrix} a & b \\ c & d \end{pmatrix} \begin{pmatrix} 1 & 0 \\ 1 & 1 \end{pmatrix} = \begin{pmatrix} a+b & b \\ c+d & d \end{pmatrix},
\]
\[
M(x1) = M(x) \times M(1) = \begin{pmatrix} a & b \\ c & d \end{pmatrix} \begin{pmatrix} 1 & 1 \\ 0 & 1 \end{pmatrix} = \begin{pmatrix} a & a+b \\ c & c+d \end{pmatrix}.
\]

We prove injectivity by strong induction on the sum of the lengths of $x$ and $y$. 

\textit{Base case:} If $|x| = |y| = 0$, then $x = y = \epsilon$ and the result holds trivially.

\textit{Inductive step:} Suppose $M(x) = M(y)$ where $x, y \in \{0,1\}^*$ are non-empty. If $|x| \neq |y|$, say $|x| < |y|$, we derive a contradiction. By induction on length, we can show that all entries of $M(w)$ grow with $|w|$, specifically that for non-empty $w$, the sum of entries of $M(w)$ is at least $2 + |w|$. This gives a contradiction when $|x| \neq |y|$.

Now assume $|x| = |y| = n \geq 1$. Write $x = x'x_n$ and $y = y'y_n$ where $x_n, y_n \in \{0,1\}$. Then $M(x) = M(x') \times M(x_n)$ and $M(y) = M(y') \times M(y_n)$.

From the formulas above, observe that $M(w0)$ has the form $\begin{pmatrix} * & b \\ * & d \end{pmatrix}$ where the $(1,2)$ entry equals the $(2,2)$ entry of $M(w)$, while $M(w1)$ has the form $\begin{pmatrix} a & * \\ c & * \end{pmatrix}$ where the $(1,1)$ entry equals the $(1,1)$ entry of $M(w)$.

If $x_n = 0$ and $y_n = 1$, then comparing $M(x) = M(y)$ gives $M(x')(1,2) = M(x')(2,2)$ and $M(y')(1,1) = M(y')(1,1)$, but the structure of the matrices shows this leads to contradictions in the pattern of entries.

By careful analysis of the matrix entries and using the specific forms of $M(0)$ and $M(1)$, we can show that $x_n = y_n$. Then $M(x') \times M(x_n) = M(y') \times M(y_n)$ implies $M(x') = M(y')$ by right-cancellation (which holds since $M(0)$ and $M(1)$ are invertible over $\mathbb{Q}$). By the inductive hypothesis, $x' = y'$, so $x = y$.

\textbf{Property 3: For $x \in \{0,1\}^n$, the entries in $M(x)$ are bounded by $F_n$.}

We prove by induction on $n$ that for all $x \in \{0,1\}^n$, each entry of $M(x)$ is at most $F_{n+2}$, where $F_k$ denotes the $k$-th Fibonacci number with $F_0 = 0, F_1 = 1$.

\textit{Base case:} For $n = 0$, $x = \epsilon$ and $M(\epsilon) = \begin{pmatrix} 1 & 0 \\ 0 & 1 \end{pmatrix}$. All entries are at most $1 = F_2$, so the claim holds.

\textit{Inductive step:} Assume the claim holds for strings of length $n$. Let $x \in \{0,1\}^{n+1}$. Write $x = yz$ where $y \in \{0,1\}^n$ and $z \in \{0,1\}$. By the inductive hypothesis, all entries of $M(y)$ are at most $F_{n+2}$. 

If $z = 0$, then $M(x) = M(y) \times M(0) = \begin{pmatrix} a+b & b \\ c+d & d \end{pmatrix}$ where $M(y) = \begin{pmatrix} a & b \\ c & d \end{pmatrix}$. Each entry is at most $F_{n+2} + F_{n+2} = 2F_{n+2} \leq F_{n+3}$.

If $z = 1$, then $M(x) = M(y) \times M(1) = \begin{pmatrix} a & a+b \\ c & c+d \end{pmatrix}$. Similarly, each entry is at most $2F_{n+2} \leq F_{n+3}$.

Therefore, by induction, all entries of $M(x)$ for $x \in \{0,1\}^n$ are bounded by $F_{n+2} \leq F_n$ for sufficiently large interpretation of the Fibonacci indexing.
\end{proof}

\subsection{Problem 13}

\subsubsection{Problem Statement}
\begin{theorem}[Two-Dimensional Pattern Matching]
Let $X$ be an $n \times n$ matrix (text) and $Y$ be an $m \times m$ matrix (pattern). A pattern match occurs if $Y$ appears as a contiguous sub-matrix of $X$. 

To apply the randomized fingerprint algorithm, we convert the matrix $Y$ into an $m^2$-bit vector using row-major format. The possible occurrences of $Y$ in $X$ are the $m^2$-bit vectors $X(j)$ obtained by taking all $(n-m+1)^2$ sub-matrices of $X$ in row-major form.

For each position $j$, let $f(Y)$ and $f(X(j))$ denote the fingerprints of $Y$ and $X(j)$ respectively, computed by evaluating the corresponding polynomials at a randomly chosen point $r \in \{1, 2, \ldots, M\}$ for some sufficiently large $M$.

The error probability that $Y \neq X(j)$ but $f(Y) = f(X(j))$ is bounded by
\[
\Pr[f(Y) = f(X(j)) \mid Y \neq X(j)] \leq \frac{m^2}{M}.
\]

Furthermore, the fingerprints $f(X(j))$ for all positions $j$ can be computed incrementally with small additional cost per position beyond the initial computation.
\end{theorem}

\subsubsection{GPT}
\begin{proof}
Fix integers $n\ge m\ge 1$. View $X=\bigl(x_{i,j}\bigr)_{0\le i,j\le n-1}$ and $Y=\bigl(y_{i,j}\bigr)_{0\le i,j\le m-1}$ over an alphabet $\Sigma\subseteq\{0,1\}$ and let the row-major linearization of an $m\times m$ block $Z$ be the vector $\mathrm{vec}(Z)=(z_{0,0},z_{0,1},\dots,z_{0,m-1},z_{1,0},\dots,z_{m-1,m-1})\in\{0,1\}^{m^2}$. Choose a prime $p$ with $p>2m^2$ and pick $r\in\mathbb{F}_p^\times$ uniformly at random. Define the fingerprint $f:\{0,1\}^{m^2}\to\mathbb{F}_p$ by
\[
f\bigl((v_0,\dots,v_{m^2-1})\bigr)\;=\;\sum_{t=0}^{m^2-1} v_t\, r^{t}\pmod p.
\]
For each placement $j=(s,t)$ with $0\le s,t\le n-m$, let $X(j)$ denote the $m\times m$ submatrix of $X$ whose upper-left corner is $(s,t)$, linearized by row-major order into an $m^2$-bit vector, and let $f(X(j))$ be its fingerprint. A reported match at $j$ occurs when $f\bigl(\mathrm{vec}(Y)\bigr)=f\bigl(X(j)\bigr)$.

To bound the false-positive probability at a fixed $j$, suppose $X(j)\ne Y$. Let $U=\mathrm{vec}(Y)$ and $V=\mathrm{vec}(X(j))$, so $U\ne V$. Consider the polynomial $P(z)=\sum_{t=0}^{m^2-1}(U_t-V_t) z^{t}\in\mathbb{F}_p[z]$. Then $P$ is nonzero and $\deg P\le m^2-1$. The event $f(U)=f(V)$ is exactly $P(r)\equiv 0\pmod p$. A nonzero polynomial over a field has at most its degree many roots, hence
\[
\Pr_{r\leftarrow \mathbb{F}_p^\times}\bigl[f(U)=f(V)\bigr]\;=\;\Pr[P(r)=0]\;\le\;\frac{\deg P}{p-1}\;\le\;\frac{m^2-1}{p-1}\;\le\;\frac{1}{2},
\]
where the last inequality uses $p>2m^2$. This proves $\Pr[\text{false positive at }j]\le 1/2$.

It remains to show that all fingerprints $f(X(j))$ for the $(n-m+1)^2$ placements can be computed in total time $O(n^2)$ with $O(1)$ amortized time per position via incremental updates. Write $r_m=r^{m}\in\mathbb{F}_p$. For each row index $i$ and each horizontal offset $t$ with $0\le i\le n-1$ and $0\le t\le n-m$, define the horizontal window hash
\[
H_{i,t}\;=\;\sum_{b=0}^{m-1} x_{i,t+b}\, r^{\,b}\pmod p.
\]
For fixed $i$, the sequence $\{H_{i,t}\}_{t=0}^{n-m}$ can be computed in $O(n)$ time by one initialization and $O(1)$ rolling updates: compute $H_{i,0}$ in $O(m)$ time and then use
\[
H_{i,t+1}\;=\;(H_{i,t}-x_{i,t}\, r^{m-1})\, r\;+\;x_{i,t+m}\pmod p
\]
for $t=0,\dots,n-m-1$. Doing this for all $n$ rows costs $O(n^2)$ time.

Next, for each placement $(s,t)$ with $0\le s\le n-m$ and $0\le t\le n-m$, define the vertical aggregation
\[
F_{s,t}\;=\;\sum_{a=0}^{m-1} H_{s+a,t}\, r_m^{\,a}\pmod p.
\]
For fixed $t$, the sequence $\{F_{s,t}\}_{s=0}^{n-m}$ can be computed in $O(n)$ time by one initialization and $O(1)$ vertical rolling updates: compute $F_{0,t}$ in $O(m)$ time and then use
\[
F_{s+1,t}\;=\;(F_{s,t}-H_{s,t}\, r_m^{\,m-1})\, r_m\;+\;H_{s+m,t}\pmod p
\]
for $s=0,\dots,n-m-1$. Performing this for each $t=0,\dots,n-m$ incurs $O(n(n-m+1))=O(n^2)$ time.

It remains to verify correctness, namely $F_{s,t}=f\bigl(X(s,t)\bigr)$ for all $s,t$. Let $X(s,t)$ be the $m\times m$ block with entries $x_{s+a,t+b}$, $0\le a,b<m$. Its row-major linearization places $x_{s+a,t+b}$ at position $am+b$. Then
\[
f\bigl(X(s,t)\bigr)\;=\;\sum_{a=0}^{m-1}\sum_{b=0}^{m-1} x_{s+a,t+b}\, r^{am+b}
\;=\;\sum_{a=0}^{m-1}\left(\sum_{b=0}^{m-1} x_{s+a,t+b}\, r^{b}\right) r^{am}
\;=\;\sum_{a=0}^{m-1} H_{s+a,t}\, r_m^{\,a}
\;=\;F_{s,t}\pmod p.
\]
Thus the two-dimensional rolling scheme computes exactly the desired fingerprints.

The total work is the sum of all horizontal passes $O(n^2)$ and all vertical passes $O(n^2)$, which is $O(n^2)$. Since there are $(n-m+1)^2= \Theta(n^2)$ positions, the amortized cost per fingerprint is $O(1)$. Together with the collision bound above, this establishes the theorem.
\end{proof}

\subsubsection{Grok}
\begin{proof}
Consider an arbitrary ordering $v_1, v_2, \dots, v_n$ of the vertices of $G$. We assign colors to the vertices sequentially according to this ordering using the greedy coloring algorithm: for each $i = 1, 2, \dots, n$, assign to $v_i$ the smallest positive integer $c_i$ such that $c_i \neq c_j$ for all $j < i$ with $\{v_i, v_j\} \in E$.

To show that at most $\Delta(G) + 1$ colors suffice, fix any $i$ and let $N_i$ be the set of neighbors of $v_i$ among $v_1, \dots, v_{i-1}$. Then $|N_i| \leq \deg(v_i) \leq \Delta(G)$. The colors $\{c_j : v_j \in N_i\}$ are at most $|N_i|$ distinct positive integers, so there exists some integer $k$ with $1 \leq k \leq \Delta(G) + 1$ that is not among them. Thus, we may choose $c_i = k$, ensuring $c_i \leq \Delta(G) + 1$.

The resulting coloring is proper because if $\{v_i, v_j\} \in E$ with $i < j$, then $c_i \neq c_j$ by the choice of $c_j$. Therefore, $\chi(G) \leq \Delta(G) + 1$.
\end{proof}

\subsubsection{Gemini}
\begin{proof}
Let $X$ be an $n \times n$ matrix with entries $x_{i,j} \in \{0,1\}$ and $Y$ be an $m \times m$ pattern matrix with entries $y_{i,j} \in \{0,1\}$. We define the row-major vector representation and the associated fingerprinting mechanism to establish the theorem.

\textbf{1. Vectorization and Fingerprint Definition}

We map the two-dimensional pattern $Y$ to an integer value $V(Y)$ by concatenating its rows. Formally, let the entries of $Y$ be read in row-major order to form a bit vector of length $m^2$. Interpreting this vector as the binary representation of an integer, we define:
\[
V(Y) = \sum_{i=0}^{m-1} \sum_{k=0}^{m-1} y_{i,k} \cdot 2^{m(m-1-i) + (m-1-k)}.
\]
Similarly, let $X(r,c)$ denote the $m \times m$ sub-matrix of $X$ with top-left corner at $(r,c)$, where $0 \le r, c \le n-m$. The integer representation of this sub-matrix is:
\[
V(X(r,c)) = \sum_{i=0}^{m-1} \sum_{k=0}^{m-1} x_{r+i, c+k} \cdot 2^{m(m-1-i) + (m-1-k)}.
\]
The fingerprint of a matrix $M$ is defined as $\phi(M) = V(M) \pmod p$, where $p$ is a randomly chosen prime.

\textbf{2. Incremental Fingerprint Computation}

We show that $\phi(X(r, c+1))$ can be computed from $\phi(X(r,c))$ efficiently. Let $R_i(r,c)$ be the integer value of the $i$-th row of the sub-matrix $X(r,c)$. That is, $R_i(r,c) = \sum_{k=0}^{m-1} x_{r+i, c+k} \cdot 2^{m-1-k}$.
We can rewrite $V(X(r,c))$ as:
\[
V(X(r,c)) = \sum_{i=0}^{m-1} R_i(r,c) \cdot 2^{m(m-1-i)}.
\]
When shifting the window horizontally from $(r,c)$ to $(r, c+1)$, each row shifts by one position. The update rule for a single row $i$ is the standard one-dimensional rolling hash update:
\[
R_i(r, c+1) = \left( (R_i(r,c) - x_{r+i, c} \cdot 2^{m-1}) \cdot 2 + x_{r+i, c+m} \right).
\]
Substituting this into the expression for $V(X(r, c+1))$:
\[
V(X(r, c+1)) = \sum_{i=0}^{m-1} \left[ (R_i(r,c) - x_{r+i, c} \cdot 2^{m-1}) \cdot 2 + x_{r+i, c+m} \right] \cdot 2^{m(m-1-i)}.
\]
Since the operations are linear, we can compute $\phi(X(r, c+1))$ modulo $p$ using the existing fingerprint $\phi(X(r,c))$ and the column entries entering and leaving the window. Specifically:
\[
\phi(X(r, c+1)) \equiv \sum_{i=0}^{m-1} 2^{m(m-1-i)} \left( 2 R_i(r,c) - x_{r+i, c} \cdot 2^m + x_{r+i, c+m} \right) \pmod p.
\]
This update requires $O(m)$ arithmetic operations. Since the total size of the pattern is $m^2$, $O(m)$ represents a small incremental cost relative to recomputing the fingerprint from scratch ($O(m^2)$). A similar derivation holds for vertical shifts or by iterating through all positions in row-major order.

\textbf{3. Error Probability Analysis}

A pattern match is declared at position $(r,c)$ if $\phi(X(r,c)) = \phi(Y)$. If $X(r,c) = Y$, this is a true positive. If $X(r,c) \neq Y$ but their fingerprints are equal, a false positive occurs. This implies:
\[
V(X(r,c)) \equiv V(Y) \pmod p \implies V(X(r,c)) - V(Y) = k \cdot p
\]
for some non-zero integer $k$. Let $\Delta = |V(X(r,c)) - V(Y)|$. Since $X(r,c)$ and $Y$ are distinct $m^2$-bit integers, $0 < \Delta < 2^{m^2}$.

The number of distinct prime factors of $\Delta$ is bounded by $\log_2 \Delta < m^2$. Following the analysis of the one-dimensional case (Rabin-Karp), we choose a random prime $p$ from a range $[1, M]$. The probability that $p$ divides $\Delta$ is bounded by the number of prime factors of $\Delta$ divided by the number of primes in the range, denoted $\pi(M)$.

By the Prime Number Theorem, $\pi(M) \approx M/\ln M$. By selecting $M$ sufficiently large (e.g., $M \approx m^2 p_{target}$), the probability of a false positive is bounded by $\frac{m^2}{\pi(M)}$. As stated in the theorem, this follows the one-dimensional analysis where the error probability is controlled by the choice of the prime $p$, satisfying the bound of at most $\frac{1}{p}$ (interpreting $p$ here as a parameter related to the size of the prime pool relative to the pattern size).

Thus, the two-dimensional pattern matching problem reduces to comparing $m^2$-bit integers via modular fingerprints with efficient updates and bounded error probability.
\end{proof}

\subsubsection{Claude}
\begin{proof}
We establish the theorem in three parts: the polynomial representation and fingerprint definition, the error probability bound, and the incremental computation property.

\textbf{Part 1: Polynomial Representation and Fingerprint Construction.}

Let $Y$ be an $m \times m$ pattern matrix and $X$ be an $n \times n$ text matrix. We convert $Y$ into an $m^2$-bit vector $\mathbf{y} = (y_0, y_1, \ldots, y_{m^2-1})$ using row-major ordering, where $y_i \in \{0,1\}$ represents the bit at position $i$ in the linearized pattern.

We associate with $\mathbf{y}$ the polynomial $P_Y(z) = \sum_{i=0}^{m^2-1} y_i z^i$ over the integers. For a randomly chosen point $r \in \{1, 2, \ldots, M\}$, the fingerprint of $Y$ is defined as $f(Y) = P_Y(r) \bmod M$.

Similarly, for each position $j = (j_1, j_2)$ with $0 \leq j_1, j_2 \leq n-m$, we extract the $m \times m$ sub-matrix $X(j)$ starting at position $(j_1, j_2)$ in $X$. Converting $X(j)$ to row-major form yields a vector $\mathbf{x}_j = (x_{j,0}, x_{j,1}, \ldots, x_{j,m^2-1})$, with associated polynomial $P_{X(j)}(z) = \sum_{i=0}^{m^2-1} x_{j,i} z^i$ and fingerprint $f(X(j)) = P_{X(j)}(r) \bmod M$.

\textbf{Part 2: Error Probability Bound.}

Assume $Y \neq X(j)$ for some position $j$. Then the vectors $\mathbf{y}$ and $\mathbf{x}_j$ differ in at least one coordinate, say position $k$, where $y_k \neq x_{j,k}$. Define the difference polynomial $D(z) = P_Y(z) - P_{X(j)}(z) = \sum_{i=0}^{m^2-1} (y_i - x_{j,i}) z^i$.

Since $y_k \neq x_{j,k}$, we have $y_k - x_{j,k} \neq 0$, and thus $D(z)$ is a non-zero polynomial. Moreover, since each $y_i, x_{j,i} \in \{0,1\}$, we have $|y_i - x_{j,i}| \leq 1$ for all $i$, which implies $D(z)$ is a non-zero polynomial of degree at most $m^2 - 1$.

The event $f(Y) = f(X(j))$ occurs if and only if $P_Y(r) \equiv P_{X(j)}(r) \pmod{M}$, which is equivalent to $D(r) \equiv 0 \pmod{M}$. This means $M$ divides $D(r)$.

By the Fundamental Theorem of Algebra applied to polynomials over the integers, a non-zero polynomial of degree $d$ has at most $d$ integer roots. However, we are working modulo $M$, so we use the following key fact: if $D(z)$ is a non-zero polynomial of degree at most $d$ with integer coefficients, then $D(r) \equiv 0 \pmod{M}$ for at most $d$ values of $r \in \{1, 2, \ldots, M\}$, provided $M$ is chosen appropriately (e.g., $M$ prime or $M$ larger than the coefficients).

More precisely, since $D(z)$ has degree at most $m^2 - 1$ and is non-zero, the polynomial $D(z)$ has at most $m^2 - 1$ roots in any field. When working modulo a prime $M$ (or more generally when $M$ is sufficiently large), the number of values $r \in \{1, 2, \ldots, M\}$ such that $D(r) \equiv 0 \pmod{M}$ is at most $m^2 - 1$ (by the Schwartz-Zippel lemma or basic polynomial root bounds).

Therefore, the probability that $D(r) \equiv 0 \pmod{M}$ when $r$ is chosen uniformly at random from $\{1, 2, \ldots, M\}$ is at most $\frac{m^2 - 1}{M} < \frac{m^2}{M}$.

Thus, we have established that
\[
\Pr[f(Y) = f(X(j)) \mid Y \neq X(j)] = \Pr[D(r) \equiv 0 \pmod{M}] \leq \frac{m^2}{M}.
\]

\textbf{Part 3: Incremental Computation of Fingerprints.}

We now show that the fingerprints $f(X(j))$ for all $(n-m+1)^2$ positions can be computed efficiently using incremental updates.

First, consider horizontal shifts. For position $j = (j_1, j_2)$, the sub-matrix $X(j)$ consists of rows $j_1$ through $j_1 + m - 1$ and columns $j_2$ through $j_2 + m - 1$. When we shift horizontally to position $j' = (j_1, j_2 + 1)$, the new sub-matrix $X(j')$ shares rows $j_1$ through $j_1 + m - 1$ but shifts columns by one position.

The relationship between the polynomials is: $P_{X(j')}(z) = \frac{1}{z}(P_{X(j)}(z) - C_{old}(z) + z^{m-1} C_{new}(z))$, where $C_{old}(z) = \sum_{k=0}^{m-1} x_{j_1+k,j_2} z^{km}$ represents the leftmost column being removed, and $C_{new}(z) = \sum_{k=0}^{m-1} x_{j_1+k,j_2+m} z^{km}$ represents the rightmost column being added.

This yields $f(X(j')) = r^{-1}(f(X(j)) - C_{old}(r) + r^{m-1} C_{new}(r)) \bmod M$, computable in $O(m)$ time per horizontal shift.

Similarly, for vertical shifts from position $j = (j_1, j_2)$ to $j'' = (j_1 + 1, j_2)$, we remove the top row and add the bottom row. The polynomial relationship is: $P_{X(j'')}(z) = \frac{1}{z^m}(P_{X(j)}(z) - R_{old}(z) + z^{m(m-1)} R_{new}(z))$, where $R_{old}(z) = \sum_{k=0}^{m-1} x_{j_1,j_2+k} z^k$ and $R_{new}(z) = \sum_{k=0}^{m-1} x_{j_1+m,j_2+k} z^k$.

This gives $f(X(j'')) = r^{-m}(f(X(j)) - R_{old}(r) + r^{m(m-1)} R_{new}(r)) \bmod M$, also computable in $O(m)$ time.

Computing the initial fingerprint $f(X(0,0))$ requires $O(m^2)$ time. Subsequently, using the incremental updates, all $(n-m+1)^2$ fingerprints can be computed in total time $O(m^2 + (n-m+1)^2 \cdot m) = O(n^2 m)$, which represents only $O(m)$ additional cost per position beyond the initial computation.

This completes the proof of all three claims in the theorem.
\end{proof}

\subsection{Problem 14}

\subsubsection{Problem Statement}
\begin{theorem}
The following relations hold directly from the definition of $\mathit{IP}$:
\begin{enumerate}
\item[(a)] $\mathit{NP} \subseteq \mathit{IP}$.
\item[(b)] If the definition of $\mathit{IP}$ is modified to require that the probability of error be zero, then the resulting complexity class would be exactly the class $\mathit{NP}$.
\item[(c)] $\mathrm{co}\text{-}\mathit{RP} \subseteq \mathit{IP}$.
\end{enumerate}
\end{theorem}

\subsubsection{GPT}
\begin{proof}
We prove each part of the theorem separately.

\textbf{(a) Proof that $\mathit{NP} \subseteq \mathit{IP}$:}

Let $L \in \mathit{NP}$. By the definition of $\mathit{NP}$, there exists a polynomial-time verifier $V$ for a language $L$ such that for every $x \in L$, there exists a witness $w$ with $V(x, w) = 1$, and for every $x \notin L$ and for all $w$, $V(x, w) = 0$ with probability at least $2/3$.

We will construct an interactive proof system for $L$. The verifier $V$ can be thought of as a sequence of polynomial-time queries, where the prover sends messages to the verifier based on a random choice. Specifically, the verifier asks for the witness $w$ in an interactive protocol, and the prover responds with $w$ or a related message. The verifier then checks whether the responses are consistent with the properties of the polynomial-time verifier.

By the properties of $\mathit{NP}$, there is an interactive proof protocol that accepts with high probability when $x \in L$ and rejects with high probability when $x \notin L$. This implies that $L \in \mathit{IP}$, so $\mathit{NP} \subseteq \mathit{IP}$.

\textbf{(b) Proof that modifying the definition of $\mathit{IP}$ to require zero probability of error results in $\mathit{NP}$:}

Let $\mathit{IP}_0$ denote the class of languages decidable by interactive proof systems with zero probability of error. We will show that $\mathit{IP}_0 = \mathit{NP}$.

First, we show that $\mathit{NP} \subseteq \mathit{IP}_0$. Let $L \in \mathit{NP}$. By the definition of $\mathit{NP}$, there exists a polynomial-time verifier $V$ and a polynomial-size witness $w$. An interactive proof system can be constructed where the prover sends $w$ to the verifier, and the verifier checks the correctness of the witness with zero error probability. Specifically, we can use a zero-error protocol to guarantee that if $x \in L$, the verifier accepts with certainty, and if $x \notin L$, the verifier rejects with certainty. Therefore, $L \in \mathit{IP}_0$.

Next, we show that $\mathit{IP}_0 \subseteq \mathit{NP}$. If a language $L$ belongs to $\mathit{IP}_0$, there is an interactive proof system with zero probability of error. Since the verifier has zero error probability and performs polynomial-time computations, the interactive proof system can be simulated by a deterministic polynomial-time algorithm that simply checks the validity of the messages exchanged. The class $\mathit{NP}$ is precisely the class of languages decidable by a verifier using a polynomial-time procedure, and hence, $L \in \mathit{NP}$.

Thus, $\mathit{IP}_0 = \mathit{NP}$.

\textbf{(c) Proof that $\mathit{co}\text{-}\mathit{RP} \subseteq \mathit{IP}$:}

Let $L \in \mathit{co}\text{-}\mathit{RP}$. By definition, there exists a probabilistic polynomial-time verifier $V$ such that for all $x \in L$, the verifier $V$ accepts with probability at least $2/3$, and for all $x \notin L$, the verifier $V$ rejects with probability at least $2/3$.

We will construct an interactive proof system for $L$. The protocol works as follows: the prover will generate a witness for $x \notin L$ and send it to the verifier, who will perform a series of checks based on the probabilistic rejection behavior of the $\mathit{co}\text{-}\mathit{RP}$ verifier. If $x \in L$, the verifier will always reject, and if $x \notin L$, the verifier will accept with probability at least $2/3$. Since we can simulate this rejection and acceptance behavior within an interactive protocol, $L$ can be decided by an interactive proof system, implying that $L \in \mathit{IP}$.

Thus, $\mathit{co}\text{-}\mathit{RP} \subseteq \mathit{IP}$.

\end{proof}

\subsubsection{Grok}
\begin{proof}
(a) Let $L \in NP$. Then there exists a deterministic polynomial-time verifier $V_{NP}$ such that $x \in L$ if and only if there exists a witness $w \in \{0,1\}^{p(|x|)}$ for some polynomial $p$ with $V_{NP}(x,w)=1$. To show $L \in IP$, define an interactive verifier $V$ that, upon receiving the first message $m_1$ from the prover (intended as the witness $w$), computes $V_{NP}(x,m_1)$ and accepts if and only if $V_{NP}(x,m_1)=1$. The verifier sends no further messages, making this a single-round protocol. (If the definition requires the verifier to initiate, it may send a dummy message first, which the prover ignores.)

For completeness, if $x \in L$, there exists a witness $w$ with $V_{NP}(x,w)=1$. The prover $P$ simply sends $m_1 = w$. Then $V$ accepts deterministically, so $\Pr[\langle V(x),P\rangle =1]=1 \ge 2/3$.

For soundness, if $x \notin L$, then for all $w$, $V_{NP}(x,w)=0$. Thus, for any prover $P$ sending any $m_1$, $V$ rejects deterministically, so $\Pr[\langle V(x),P\rangle =1]=0 \le 1/3$ for all $P$.

Hence, $L \in IP$.

(b) First, $NP \subseteq IP_0$, where $IP_0$ is the modified class requiring completeness probability $1$ and soundness probability $0$. This follows immediately from part (a), as the construction there achieves error probabilities of $0$ (stronger than required for $IP_0$).

Conversely, suppose $L \in IP_0$. Then there exists a probabilistic polynomial-time verifier $V$ such that:
\[
\forall x \in L, \ \exists P, \ \Pr[\langle V(x),P\rangle =1]=1,
\]
\[
\forall x \notin L, \ \forall P, \ \Pr[\langle V(x),P\rangle =1]=0.
\]
The probability is over $V$'s internal random coins $r \in \{0,1\}^{\rho(|x|)}$ for some polynomial $\rho$. Fix a particular random string, say $r_0 =$ the all-zero string of length $\rho(|x|)$. Define a deterministic verifier $\tilde{V}$ that simulates $V$ but uses the fixed coins $r_0$ for all random choices. Thus, $\tilde{V}$ is a deterministic polynomial-time machine, and the interaction $\langle \tilde{V}(x),P\rangle$ is fully determined by $P$'s responses to $\tilde{V}$'s (now deterministic) messages.

For completeness in $IP_0$, if $x \in L$, there exists $P$ such that $\Pr_r[\langle V(x),P\rangle =1]=1$, so in particular $\langle V(x),P\rangle =1$ when using $r=r_0$, i.e., $\langle \tilde{V}(x),P\rangle =1$.

For soundness in $IP_0$, if $x \notin L$, then for all $P$, $\Pr_r[\langle V(x),P\rangle =1]=0$, so in particular $\langle V(x),P\rangle =0$ for $r=r_0$, i.e., $\langle \tilde{V}(x),P\rangle =0$ for all $P$.

Now, since $\tilde{V}$ is deterministic, an accepting interaction is a transcript consisting of alternating messages where the verifier's messages are computed deterministically by $\tilde{V}$ from $x$ and the prior history (using $r_0$), the prover's messages are arbitrary, and $\tilde{V}$ outputs $1$ at the end. Let the number of rounds be at most $q(|x|)$ for polynomial $q$, with each message of length at most $q(|x|)$. An accepting transcript is thus specified by the sequence of at most $q(|x|)$ prover messages, each of length $O(q(|x|))$, forming a witness $w$ of length $O(q(|x|)^2) = O(\mathrm{poly}(|x|))$.

Define an NP verifier $V_{NP}(x,w)$ that parses $w$ as the proposed prover messages $m_1,\dots,m_k$, simulates the interaction by computing $\tilde{V}$'s messages using $r_0$ and inserting the $m_i$ for the prover, and accepts if the simulation outputs $1$ and the transcript has the correct length and format.

For $x \in L$, there exists $P$ such that $\langle \tilde{V}(x),P\rangle =1$, so letting $w$ be the sequence of messages sent by $P$ in this interaction yields $V_{NP}(x,w)=1$.

For $x \notin L$, for all $P$, $\langle \tilde{V}(x),P\rangle =0$, so for all $w$ (corresponding to any $P$'s messages), $V_{NP}(x,w)=0$.

Thus, $L \in NP$, so $IP_0 = NP$.

(c) Let $L \in coRP$. Then $\overline{L} \in RP$, so there exists a probabilistic polynomial-time machine $M$ such that:
\[
\forall x \in \overline{L}, \ \Pr[M(x)=1] \ge 2/3,
\]
\[
\forall x \notin \overline{L} \ (i.e.,\ x \in L), \ \Pr[M(x)=1] = 0.
\]
Equivalently, defining acceptance for $L$ as $M(x)=0$:
\[
\forall x \in L, \ \Pr[M(x)=0] = 1,
\]
\[
\forall x \notin L, \ \Pr[M(x)=0] \le 1/3.
\]
To show $L \in IP$, define an interactive verifier $V$ that ignores all prover messages, runs $M(x)$ internally using its own randomness, and accepts if and only if $M(x)=0$. The protocol may consist of dummy rounds where $V$ sends arbitrary messages and ignores responses, but effectively, $V$ performs no real interaction.

For completeness, if $x \in L$, any prover $P$ (e.g., a dummy prover) works, and $\Pr[\langle V(x),P\rangle =1] = \Pr[M(x)=0]=1 \ge 2/3$, since $V$ ignores $P$.

For soundness, if $x \notin L$, then for all $P$, $\Pr[\langle V(x),P\rangle =1] = \Pr[M(x)=0] \le 1/3$, again since $V$ ignores $P$.

Hence, $L \in IP$.
\end{proof}

\subsubsection{Gemini}
\begin{proof}
We prove the three statements sequentially based on the definitions of the complexity classes $\mathit{NP}$, $\mathit{IP}$, and $\mathit{co}\text{-}\mathit{RP}$.

\textbf{(a)} We show that $\mathit{NP} \subseteq \mathit{IP}$.
Let $L \in \mathit{NP}$. By definition, there exists a polynomial-time deterministic verifier $V_{det}$ and a polynomial $p$ such that for all $x$:
\[
x \in L \iff \exists y \in \{0,1\}^{p(|x|)} \text{ such that } V_{det}(x, y) = 1.
\]
We construct an interactive proof system $(P, V)$ for $L$ as follows:
\begin{enumerate}
    \item The Prover $P$, having unbounded computational power, computes a certificate $y$ such that $V_{det}(x, y) = 1$ (if one exists) and sends $y$ to $V$.
    \item The Verifier $V$ receives $y$ and runs $V_{det}(x, y)$. If $V_{det}$ outputs 1, $V$ accepts; otherwise, $V$ rejects.
\end{enumerate}
We analyze the completeness and soundness:
\begin{itemize}
    \item \textbf{Completeness:} If $x \in L$, there exists a certificate $y$. The Prover sends this $y$. $V$ runs $V_{det}(x, y)$, which returns 1. Thus, $\Pr[(P, V)(x) = \text{accept}] = 1 \ge 2/3$.
    \item \textbf{Soundness:} If $x \notin L$, for all strings $y$, $V_{det}(x, y) = 0$. Regardless of the strategy of any prover $P^*$, the verifier will reject. Thus, $\Pr[(P^*, V)(x) = \text{accept}] = 0 \le 1/3$.
\end{itemize}
Therefore, $L \in \mathit{IP}$.

\textbf{(b)} Let $\mathit{IP}_0$ denote the class $\mathit{IP}$ modified to require zero error probability. We show $\mathit{IP}_0 = \mathit{NP}$.
First, from part (a), we observed that the protocol constructed for any $L \in \mathit{NP}$ has acceptance probability 1 if $x \in L$ and 0 if $x \notin L$. Thus, $\mathit{NP} \subseteq \mathit{IP}_0$.

Conversely, we show $\mathit{IP}_0 \subseteq \mathit{NP}$. Let $L \in \mathit{IP}_0$. There exists a probabilistic polynomial-time verifier $V$ such that:
\begin{itemize}
    \item If $x \in L$, $\exists P$ such that $\Pr[(P, V)(x) = \text{accept}] = 1$.
    \item If $x \notin L$, $\forall P^*$, $\Pr[(P^*, V)(x) = \text{accept}] = 0$.
\end{itemize}
Let $r$ denote the random string used by $V$. Since $V$ runs in polynomial time, $|r|$ is bounded by a polynomial in $|x|$.
If $x \in L$, there exists a prover $P$ such that $V$ accepts for \emph{all} random strings $r$ (since the probability is 1). Specifically, $V$ accepts for the all-zero string $r_0 = 00\dots0$.
If $x \notin L$, for any prover $P^*$, the probability of acceptance is 0, which implies $V$ accepts for \emph{no} random string $r$. Specifically, $V$ rejects for $r_0$.

We construct a non-deterministic polynomial-time decider for $L$ as follows: Fix the random tape of $V$ to be $r_0$. The interaction between $P$ and $V$ with fixed randomness is deterministic. The "certificate" is the transcript of the messages sent by the Prover. An NP machine can guess the Prover's messages and verify that $V(x, r_0, \text{transcript})$ accepts.
Since $x \in L \iff \exists \text{ transcript } t \text{ s.t. } V(x, r_0, t) = \text{accept}$, it follows that $L \in \mathit{NP}$.

\textbf{(c)} We show that $\mathit{co}\text{-}\mathit{RP} \subseteq \mathit{IP}$.
Let $L \in \mathit{co}\text{-}\mathit{RP}$. By definition, there exists a probabilistic polynomial-time algorithm $A$ such that:
\begin{itemize}
    \item If $x \in L$, $\Pr[A(x) = \text{accept}] = 1$.
    \item If $x \notin L$, $\Pr[A(x) = \text{accept}] \le \frac{1}{2}$.
\end{itemize}
We construct an interactive proof system where the Prover does nothing (the interaction is empty), and the Verifier $V$ simply runs an amplified version of $A$.
$V$ runs $A(x)$ twice independently. $V$ accepts if and only if both runs output "accept".
\begin{itemize}
    \item \textbf{Completeness:} If $x \in L$, $\Pr[A(x) = \text{accept}] = 1$. The probability that $V$ accepts is $1 \cdot 1 = 1 \ge 2/3$.
    \item \textbf{Soundness:} If $x \notin L$, $\Pr[A(x) = \text{accept}] \le 1/2$. The probability that $V$ accepts is $\le (1/2) \cdot (1/2) = 1/4$. Since $1/4 \le 1/3$, the soundness condition is satisfied.
\end{itemize}
Thus, $L \in \mathit{IP}$.
\end{proof}

\subsubsection{Claude}
\begin{proof}
We prove each part separately.

\textbf{Part (a): $\mathit{NP} \subseteq \mathit{IP}$.}

Let $L \in \mathit{NP}$. Then there exists a polynomial-time deterministic Turing machine $M$ and a polynomial $p$ such that for all $x$, we have $x \in L$ if and only if there exists a witness $w$ with $|w| \leq p(|x|)$ such that $M(x, w) = 1$.

We construct an interactive proof system for $L$ as follows. The prover $P$ and verifier $V$ engage in a single-message protocol:

On input $x$, if $x \in L$, the prover sends a witness $w$ with $|w| \leq p(|x|)$ such that $M(x, w) = 1$. The verifier receives $w$ and runs $M(x, w)$. If $M(x, w) = 1$, the verifier accepts; otherwise, the verifier rejects.

We verify the completeness and soundness conditions:

\textit{Completeness:} If $x \in L$, then there exists a witness $w$ such that $M(x, w) = 1$. The honest prover sends this $w$, and the verifier accepts with probability $1 \geq 2/3$.

\textit{Soundness:} If $x \notin L$, then for all strings $w$, we have $M(x, w) = 0$. Thus, no matter what message the prover sends, the verifier will reject with probability $1 \geq 2/3$.

The verifier runs in polynomial time since $M$ is polynomial-time. Therefore, $L \in \mathit{IP}$, which establishes $\mathit{NP} \subseteq \mathit{IP}$.

\textbf{Part (b): If the definition of $\mathit{IP}$ requires zero error probability, the resulting class equals $\mathit{NP}$.}

Let $\mathit{IP}_0$ denote the class of languages having interactive proof systems with zero error probability, that is, completeness probability $1$ and soundness error probability $0$.

First, we show $\mathit{NP} \subseteq \mathit{IP}_0$. From part (a), the protocol constructed for any $L \in \mathit{NP}$ already has completeness probability $1$ and soundness error probability $0$. Therefore, $\mathit{NP} \subseteq \mathit{IP}_0$.

Next, we show $\mathit{IP}_0 \subseteq \mathit{NP}$. Let $L \in \mathit{IP}_0$. Then there exists an interactive proof system $(P, V)$ with the following properties: if $x \in L$, then $\Pr[V \text{ accepts after interacting with } P] = 1$, and if $x \notin L$, then for all prover strategies $P^*$, we have $\Pr[V \text{ accepts after interacting with } P^*] = 0$.

Consider the NP-verification procedure: given input $x$ and witness $w$ consisting of the complete transcript of messages exchanged between prover and verifier (including all random coins used by the verifier), verify that (1) the transcript is consistent with the verifier's algorithm using the specified random coins, and (2) the verifier accepts at the end of this transcript.

For $x \in L$, by completeness with probability $1$, there exists a transcript (corresponding to the interaction with the honest prover $P$ and some particular random coins for $V$) that causes $V$ to accept. This transcript serves as a valid witness.

For $x \notin L$, by soundness with error probability $0$, no prover strategy can cause $V$ to accept with positive probability. Therefore, for all choices of random coins and all prover messages, the verifier rejects. Hence, no valid witness exists.

Since the verifier runs in polynomial time and the number of rounds is polynomially bounded, the transcript has polynomial length, and verification can be done in polynomial time. Therefore, $L \in \mathit{NP}$, which gives $\mathit{IP}_0 \subseteq \mathit{NP}$.

Combining both inclusions, we conclude $\mathit{IP}_0 = \mathit{NP}$.

\textbf{Part (c): $\mathrm{co}\text{-}\mathit{RP} \subseteq \mathit{IP}$.}

Let $L \in \mathrm{co}\text{-}\mathit{RP}$. Then $\overline{L} \in \mathit{RP}$, which means there exists a polynomial-time probabilistic Turing machine $M$ such that:

(i) If $x \in \overline{L}$ (equivalently, $x \notin L$), then $\Pr[M(x) = 1] \geq 1/2$.

(ii) If $x \notin \overline{L}$ (equivalently, $x \in L$), then $\Pr[M(x) = 1] = 0$.

We construct an interactive proof system for $L$ as follows. On input $x$, the verifier ignores any messages from the prover and independently runs the randomized algorithm $M(x)$. If $M(x) = 0$, the verifier accepts; if $M(x) = 1$, the verifier rejects.

We verify the completeness and soundness conditions:

\textit{Completeness:} If $x \in L$, then by property (ii), $\Pr[M(x) = 1] = 0$, so $\Pr[M(x) = 0] = 1$. Thus, the verifier accepts with probability $1 \geq 2/3$.

\textit{Soundness:} If $x \notin L$, then $x \in \overline{L}$, and by property (i), $\Pr[M(x) = 1] \geq 1/2$. Thus, the verifier rejects with probability at least $1/2$, which means the verifier accepts with probability at most $1/2 < 2/3$. The soundness error is at most $1/2 < 1/3$.

The verifier runs in polynomial time since $M$ is polynomial-time. Therefore, $L \in \mathit{IP}$, which establishes $\mathrm{co}\text{-}\mathit{RP} \subseteq \mathit{IP}$.
\end{proof}

\subsection{Problem 15}

\subsubsection{Problem Statement}
\begin{lemma}[7.9]
The partial sum polynomials have the following properties:

1. $f_0 = \#f$.

2. $f_n(x_1, \ldots, x_n) = f(x_1, \ldots, x_n)$.

3. For $1 \leq i \leq n$, $f_{i-1}(x_1, \ldots, x_{i-1}) = f_i(x_1, \ldots, x_{i-1}, 0) + f_i(x_1, \ldots, x_{i-1}, 1)$.
\end{lemma}

\subsubsection{GPT}

\subsubsection{Grok}

\subsubsection{Gemini}

\subsubsection{Claude}
\begin{proof}
We prove each property in turn.

\textbf{Property 1:} We need to show that $f_0 = \#f$.

By definition, $f_0$ is the polynomial obtained by summing $f$ over all possible assignments to all $n$ variables. That is,
\[
f_0 = \sum_{x_1 \in \{0,1\}} \sum_{x_2 \in \{0,1\}} \cdots \sum_{x_n \in \{0,1\}} f(x_1, x_2, \ldots, x_n).
\]
This sum counts the number of satisfying assignments of $f$, which is precisely $\#f$ by definition. Therefore, $f_0 = \#f$.

\textbf{Property 2:} We need to show that $f_n(x_1, \ldots, x_n) = f(x_1, \ldots, x_n)$.

By definition, $f_n$ is the partial sum polynomial where no variables have been summed out yet. Since we have not performed any summations over the variables $x_1, \ldots, x_n$, the polynomial $f_n$ is identical to the original polynomial $f$. Therefore, $f_n(x_1, \ldots, x_n) = f(x_1, \ldots, x_n)$.

\textbf{Property 3:} We need to show that for $1 \leq i \leq n$,
\[
f_{i-1}(x_1, \ldots, x_{i-1}) = f_i(x_1, \ldots, x_{i-1}, 0) + f_i(x_1, \ldots, x_{i-1}, 1).
\]

By definition, $f_{i-1}$ is obtained from $f_i$ by summing over the $i$-th variable $x_i$:
\[
f_{i-1}(x_1, \ldots, x_{i-1}) = \sum_{x_i \in \{0,1\}} f_i(x_1, \ldots, x_{i-1}, x_i).
\]
Expanding this sum explicitly over the two values in $\{0,1\}$, we have
\[
f_{i-1}(x_1, \ldots, x_{i-1}) = f_i(x_1, \ldots, x_{i-1}, 0) + f_i(x_1, \ldots, x_{i-1}, 1),
\]
which is exactly what we wanted to show.

This completes the proof of all three properties.
\end{proof}

\subsection{Problem 16}

\subsubsection{Problem Statement}
\begin{theorem}[Papadimitriou]
Let $\mathit{PSPACE}$ be the class of all languages whose membership can be decided using space polynomial in the input size, with no explicit constraint on the running time. Then $\mathit{IP} \subseteq \mathit{PSPACE}$.
\end{theorem}

\subsubsection{GPT}
\begin{proof}
Let $L$ be a language in $IP$. By definition there exists a probabilistic polynomial-time verifier $V$ and a (computationally unbounded) prover $P$ such that for every input $x\in\{0,1\}^n$ the following hold: (completeness) if $x\in L$ then there exists a prover strategy $P$ that makes $V$ accept with probability at least $2/3$; (soundness) if $x\notin L$ then for every prover strategy $P'$ the verifier accepts with probability at most $1/3$. We fix such a verifier $V$ and consider the class of all possible prover strategies. The verifier runs in time polynomial in $n$ and uses at most $t(n)$ rounds of interaction for some polynomial $t(n)$. During the interaction the verifier may toss at most $q(n)$ random bits for some polynomial $q(n)$; equivalently the verifier's randomness can be viewed as a string $r\in\{0,1\}^{q(n)}$ chosen uniformly at random at the start of the protocol. The messages exchanged have length bounded by some polynomial $p(n)$ and each message of the prover may depend only on the input $x$, the verifier's previous messages, and the history of the interaction so far.

We will show that membership in $L$ can be decided by a deterministic Turing machine that uses space polynomial in $n$. The main idea is to view the interaction as a finite game tree of depth at most $t(n)$ in which nodes at even depths are verifier-coin (chance) nodes (where the verifier's random bits are revealed) or deterministic verifier-message nodes and nodes at odd depths are prover-choice nodes. Each leaf of the tree is labeled by $\{0,1\}$ indicating whether the verifier accepts given that complete transcript. The prover's goal is to maximize the probability of acceptance (over the verifier's randomness); thus the value of the root is the maximum, over all prover strategies, of the probability that the verifier accepts. We compute this maximal acceptance probability exactly (as a rational number with denominator a power of two) by a depth-first recursive evaluation of the game tree, taking maxima at prover nodes and averages at chance nodes. We then compare the resulting value with the threshold $1/2$ (or with $2/3$ and $1/3$ separated by an inverse-polynomial gap; using amplification if necessary), and thereby decide whether $x\in L$.

We make this argument formal and supply a space bound. Fix an input $x$ of length $n$. Let $t=t(n)$ be the number of rounds, $q=q(n)$ the number of random bits the verifier uses in total, and $p=p(n)$ the bound on message lengths; all are polynomials in $n$. Encode a partial interaction (a history) by a string $h$ of length at most $t\cdot p + q$ which records the messages exchanged so far together with the prefix of the randomness revealed so far. The set of histories has size at most exponential in $n$, but each individual history can be written using $O(p(n)t(n)+q(n))= \operatorname{poly}(n)$ bits. For a history $h$ we write $V(x,h)$ for the deterministic action that the verifier takes at that history (which may include revealing further random bits, sending a message, or performing the final accept/reject check when the history is complete). For a complete history $h$ (a leaf) we write $\mathrm{acc}(x,h)\in\{0,1\}$ for the verifier's decision.

Define the value function $v(h)$ to be the maximal probability of acceptance conditioned on the history $h$, where the probability is over the remaining random coins of $V$ and the maximization is over all prover strategies for the remaining interaction. More precisely: if $h$ is a complete history (leaf) then $v(h)=\mathrm{acc}(x,h)$. If the next move after history $h$ is a prover move (prover chooses a message $m$ from a set $M(h)$ of size at most $2^{p(n)}$), then
\[
v(h)=\max_{m\in M(h)} v(h\circ m),
\]
where $h\circ m$ denotes the history extended by message $m$. If the next move after $h$ is a coin-toss (verifier reveals one or more random bits; equivalently there is a finite set $R(h)\subseteq\{0,1\}^\ell$ of possible random outcomes of size $2^{\ell}$ for some $\ell\le q(n)$) then
\[
v(h)=\frac{1}{|R(h)|}\sum_{r\in R(h)} v(h\circ r).
\]

Observe that the root value $v(\epsilon)$ (where $\epsilon$ is the empty history) equals the maximal acceptance probability of $V$ on input $x$ against an optimal prover. The protocol accepts $x$ (i.e. $x\in L$) iff $v(\epsilon)\ge 2/3$ (and rejects iff $v(\epsilon)\le 1/3$); standard probability amplification can ensure the completeness/soundness gap is at least an inverse-exponential if needed, but the standard $2/3$ vs $1/3$ suffices for our space arguments because we can compute $v(\epsilon)$ exactly as a rational number with denominator $2^{q(n)}$.

We now describe a deterministic algorithm that computes $v(\epsilon)$ using space polynomial in $n$. The algorithm performs a depth-first recursion that, on input a history $h$, computes $v(h)$ as follows. If $h$ is a leaf then it outputs $\mathrm{acc}(x,h)$ (which can be computed in space polynomial in $n$ because the verifier's final check is polynomial-time). If the next move after $h$ is a prover move with choices $M(h)$, the algorithm enumerates the elements of $M(h)$ one by one (there are at most $2^{p(n)}$ of them), recursively computes $v(h\circ m)$ for each $m\in M(h)$, and returns the maximum of these values. If the next move after $h$ is a random move with outcomes $R(h)$, the algorithm recursively computes $v(h\circ r)$ for each $r\in R(h)$ and returns the average $\frac{1}{|R(h)|}\sum_{r\in R(h)} v(h\circ r)$.

Crucial space bound: the recursion depth is at most $t(n)$ which is polynomial in $n$. At any point the algorithm stores a single current history $h$ (of length polynomial in $n$), a constant number of machine registers (counters and indices) of size $O(\log |M(h)|)$ or $O(\log |R(h)|)$ which are at most polynomial bits, and the value being computed for the current branch represented as a rational number with denominator dividing $2^{q(n)}$. We argue that all rational numbers $v(h)$ that occur have the form $\frac{a}{2^{q(n)}}$ for some integer $0\le a\le 2^{q(n)}$, and hence can be represented using $q(n)+1$ bits. Indeed, at leaves $v(h)\in\{0,1\}$ fits this form, and averaging over $2^\ell$ equally likely coin outcomes preserves the property, while taking a maximum over a finite set preserves it as well. Thus each value $v(h)$ can be stored using $O(q(n))=\operatorname{poly}(n)$ bits. When evaluating the average $\frac{1}{|R(h)|}\sum_{r} v(h\circ r)$ the algorithm can accumulate the integer numerator by adding the integer numerators of the children (each of which fits in $q(n)$ bits) using $O(q(n))$ space and then divide by $|R(h)|$ (which is a power of two) by adjusting the exponent; all these arithmetic operations require at most polynomial space. Enumerating children one at a time and updating a running maximum or running sum ensures that the algorithm never needs to store more than polynomial-space information simultaneously apart from the recursion stack. The recursion stack itself has depth at most $t(n)$ and at each level stores the history and a constant number of local variables, thus contributing at most $t(n)\cdot \operatorname{poly}(n)=\operatorname{poly}(n)$ space in total. Therefore the entire computation uses space bounded by some polynomial in $n$.

It follows that there exists a deterministic Turing machine $M$ that, on input $x$, computes the rational number $v(\epsilon)$ using space $\operatorname{poly}(n)$. Finally, $M$ compares $v(\epsilon)$ with the threshold $1/2$ (or with $2/3$ and $1/3$ after amplification); since $v(\epsilon)$ is computed exactly as a rational with polynomially many bits, this comparison can be carried out in polynomial space. Using completeness and soundness of the interactive proof, $M$ accepts exactly those $x$ with $v(\epsilon)\ge 2/3$, i.e. exactly those $x\in L$. Hence $L$ is decidable in deterministic polynomial space. Because $L$ was an arbitrary language in $IP$, we conclude $IP\subseteq\mathit{PSPACE}$.
\end{proof}

\subsubsection{Grok}
\begin{proof}
We prove the theorem in two directions: $\mathsf{IP} \subseteq \mathsf{PSPACE}$ and $\mathsf{PSPACE} \subseteq \mathsf{IP}$.

First, we show $\mathsf{IP} \subseteq \mathsf{PSPACE}$. Let $L \in \mathsf{IP}$ and let $\langle P, V \rangle$ be an interactive proof system for $L$ with verifier $V$ running in time $t(n)$. For $x \in \{0,1\}^n$, the acceptance probability is
\[
a(x) = \max_{P} \Pr_{r \sim \{0,1\}^{t(n)}} [V(x,r) \text{ accepts in interaction with } P].
\]
To decide if $x \in L$, it suffices to compute $a(x)$ and check if $a(x) \geq 2/3$. We can compute $a(x)$ using a recursive evaluation of the interaction tree. The interaction consists of polynomially many rounds, each with polynomially many possible messages from the prover and polynomially many random bits for the verifier. At each node corresponding to the prover's move, we take the maximum over possible messages; at each node corresponding to the verifier's move, we take the average over the random coins. The recursion depth is polynomial, and each level uses polynomial space to store the stack and current messages/coins. Thus, the entire computation uses polynomial space, so $L \in \mathsf{PSPACE}$.

For the converse, it suffices to show $\mathsf{TQBF} \in \mathsf{IP}$, where $\mathsf{TQBF}$ is the language of true quantified Boolean formulas, since $\mathsf{TQBF}$ is $\mathsf{PSPACE}$-complete. Let $\phi$ be a quantified Boolean formula with $n = |\phi|$ variables $x_1, \dots, x_n$ quantified in order $Q_1 x_1 \dots Q_n x_n \psi(x_1, \dots, x_n)$, where each $Q_i \in \{\forall, \exists\}$ and $\psi$ is quantifier-free in conjunctive normal form.

A formula is \emph{simple} if every variable occurrence is separated from its quantification point by at most one universal quantifier. 

\begin{lemma}
Any QBF $\phi$ can be transformed in logarithmic space to an equivalent simple QBF $\phi'$.
\end{lemma}

\begin{proof}
Suppose a variable $x$ quantified by $Q$ before $\forall y$ is used in $\psi$ after $\forall y$. Replace the subformula after $\forall y$ by $\forall y \exists x' (x \leftrightarrow x') \land \psi(x')$, where $x'$ is fresh. This preserves truth value and increases size by $O(1)$. Each variable is renamed at most $n$ times (once per later universal), so total size increase is $O(n^2)$, done in logspace.
\end{proof}

Henceforth assume $\phi$ is simple with negations pushed inward, so $\phi = Q_1 x_1 \dots Q_n x_n \psi(x_1, \dots, x_n)$ where $\psi$ uses only positive literals.

Define the \emph{arithmetization} $f(\alpha)$ recursively over $\mathbb{Z}$ for Boolean formulas $\alpha$, interpreting $0/1$ as false/true:
\[
f(x_i) = x_i, \quad f(\neg x_i) = 1 - x_i,
\]
\[
f(\alpha \land \beta) = f(\alpha) f(\beta), \quad f(\alpha \lor \beta) = f(\alpha) + f(\beta) - f(\alpha)f(\beta),
\]
\[
f(\forall x_i \, \alpha) = f(\alpha[0/x_i]) f(\alpha[1/x_i]), \quad f(\exists x_i \, \alpha) = f(\alpha[0/x_i]) + f(\alpha[1/x_i]).
\]
(Note: the $\lor$ uses inclusion-exclusion for $0/1$ values.) Then $f(\phi) \in \mathbb{Z}$ and $\phi \in \mathsf{TQBF}$ if and only if $f(\phi) > 0$. Moreover, $\deg f(\phi) \leq 2^n$, but due to simplicity, the degree in each $x_i$ is at most $2^{O(n)}$ controlled by later universals.

\begin{lemma}
If $f(\phi) \neq 0$, there exists a prime $p$ with $2^n < p < 2^{3n}$ such that $f(\phi) \not\equiv 0 \pmod{p}$.
\end{lemma}

\begin{proof}
By Bertrand's postulate, there are primes between $2^k$ and $2^{k+1}$ for $k \geq 1$. Since $|f(\phi)| \leq 2^{O(n^2)}$ (exponential tower but bounded), it has at most $O(n^2)$ prime factors, so some prime in $(2^n, 2^{3n})$ avoids the factors.
\end{proof}

The protocol works modulo such a prime $p$: the prover $\mathsf{M}$ convinces verifier $\mathsf{A}$ (running in poly time) that $f(\phi) \not\equiv 0 \pmod{p}$. All computations are over $\mathbb{F}_p$. Due to simplicity, at each stage, the current polynomial in the outermost variable has degree $\leq d = 2^{O(n)}$.

The protocol proceeds by iteratively reducing quantifiers via random evaluations, using a variant of the sum-check protocol. We define the arithmetization $f(\phi)$ by inductively replacing the quantifiers with arithmetic operations.

Base case: At step $k=0$, the current claim is $g_0 \equiv f(\phi) \pmod{p}$, where $g_0$ is a constant polynomial. The prover $\mathsf{M}$ sends the scalar $g_0$ and a prime $p$ to the verifier. The verifier $\mathsf{A}$ then checks the primality of $p$ (which can be done in polynomial time) and verifies that $2^n < p < 2^{3n}$.

For $k=1$ to $n$: Assume current $g_{k-1}(r_1, \dots, r_{k-1}) \equiv f_k(r_1, \dots, r_{k-1}) \pmod{p}$, where $f_k$ is the arithmetization after evaluating first $k-1$ variables at $r_1, \dots, r_{k-1}$, and $Q_k x_k$ is the next quantifier. Then
\[
g_{k-1}(r_1, \dots, r_{k-1}) = Q_k \sum_{a_k=0}^1 g_k(r_1, \dots, r_{k-1}, a_k),
\]
where $Q_k$ is product or sum. $\mathsf{M}$ sends the univariate $g_k(z) \in \mathbb{F}_p[z]$, degree $\leq d$. $\mathsf{A}$ checks:
\[
g_{k-1}(r_1, \dots, r_{k-1}) \stackrel{?}{=} \sum_{a_k=0}^1 g_k(a_k) \quad (\text{if } \exists) \quad \text{or} \quad \prod_{a_k=0}^1 g_k(a_k) \quad (\text{if } \forall).
\]
If not, reject. Then $\mathsf{A}$ picks $r_k \stackrel{\$}{\leftarrow} \mathbb{F}_p$, computes $v_k = g_k(r_k)$, sends $r_k$ to $\mathsf{M}$.

At step $n+1$, the final $g_n(r_1, \dots, r_n) \equiv f(\psi(r_1, \dots, r_n)) \pmod{p}$. $\mathsf{M}$ sends nothing; $\mathsf{A}$ computes $f(\psi(r_1, \dots, r_n))$ directly (poly time, since $\psi$ is poly size) and checks equality to $v_n$.

For completeness: If $\phi \in \mathsf{TQBF}$, then $f(\phi) > 0$, so choose such $p$; $\mathsf{M}$ always sends the true polynomials $g_k$, so all checks pass, and final evaluation matches. Acceptance probability 1.

For soundness: If $\phi \notin \mathsf{TQBF}$, then $f(\phi) = 0 \equiv g_0 \pmod{p}$ always, but to cheat, $\mathsf{M}$ must lie at some step. Suppose at step $k$, $g_k \not\equiv f_k \pmod{p}$. Then $g_k - f_k$ is nonzero univariate degree $\leq d < p$, so at most $d < p/2$ roots. Thus, $\Pr[r_k : g_k(r_k) = f_k(r_k)] \leq d/p < 1/2$. If the first lie is at step $k$, then previous evaluations were correct, so the check at step $k$ passes only if $g_k(a_k) = f_k(a_k)$ for $a_k=0,1$, but the random $r_k$ detects mismatch w.p. $>1/2$. Subsequent steps may cheat, but the error amplifies negligibly. By union bound over $n$ steps, overall cheating success probability $< n/2 <1/2$ for large $n$. By repetition (parallel or sequential), error reduces to negligible.

Thus, $\mathsf{TQBF} \in \mathsf{IP}$, so $\mathsf{PSPACE} \subseteq \mathsf{IP}$.

\end{proof}

\subsubsection{Gemini}
\begin{proof}
Let $L \subseteq \{0,1\}^*$ be a language in $\mathit{IP}$. By the definition of $\mathit{IP}$, there exists an interactive proof system consisting of a probabilistic polynomial-time verifier $V$ and a prover $P$ satisfying the completeness and soundness conditions. Specifically, for an input $x$ of length $n$, if $x \in L$, there exists a prover strategy such that the acceptance probability is at least $2/3$, and if $x \notin L$, for any prover strategy, the acceptance probability is at most $1/3$. We construct a deterministic algorithm that decides membership in $L$ using polynomial space.

Fix an input $x \in \{0,1\}^n$. Let $m = m(n)$ be the number of random bits used by $V$, and let $k = k(n)$ be the total number of messages exchanged (rounds). Since $V$ runs in polynomial time, both $m$ and $k$ are bounded by polynomials in $n$. Let $r \in \{0,1\}^m$ denote the random string used by $V$. The interaction produces a transcript $h = (\mu_1, \mu_2, \dots, \mu_k)$, where each $\mu_i$ is a message sent by either $P$ or $V$.

We define a value function $Val(h)$ for any partial transcript $h$. $Val(h)$ represents the maximum number of random strings $r \in \{0,1\}^m$ such that the verifier accepts, subject to the constraint that $r$ is consistent with the verifier's messages in $h$, and assuming the prover plays optimally in all future rounds. The function is defined recursively as follows.

First, consider the base case where $h$ is a complete transcript of the interaction. The verifier's decision is a deterministic function of $x$, $r$, and $h$. We define:
\[
Val(h) = \sum_{r \in \{0,1\}^m} \mathbb{I}[V(x, r, h) \text{ accepts} \land \text{Consistent}(x, r, h)]
\]
Here, $\mathbb{I}[\cdot]$ is the indicator function, and $\text{Consistent}(x, r, h)$ is a predicate that evaluates to $1$ if the verifier's messages in $h$ match the output of $V$ given input $x$, random string $r$, and the preceding prover messages in $h$; otherwise, it evaluates to $0$.

Next, consider the recursive step where $h$ is a partial transcript.
If it is the prover's turn to send the next message, the prover chooses a message $y$ to maximize the acceptance probability. Let $\mathcal{Y}$ be the set of all possible messages of polynomial length. Then:
\[
Val(h) = \max_{y \in \mathcal{Y}} Val(h, y)
\]
If it is the verifier's turn to send the next message, the set of consistent random strings is partitioned based on the message $q$ generated by $V$. Let $\mathcal{Q}$ be the set of possible verifier messages. Then:
\[
Val(h) = \sum_{q \in \mathcal{Q}} Val(h, q)
\]
This summation holds because for a fixed $r$ and history $h$, the verifier's next message is uniquely determined. Thus, the set of accepting $r$'s consistent with $h$ is the disjoint union of the sets of accepting $r$'s consistent with $(h, q)$ over all $q$.

The maximum probability of acceptance for the input $x$ is given by $p_{\max} = Val(\emptyset) / 2^m$. If $p_{\max} \ge 2/3$, then $x \in L$; otherwise, $x \notin L$.

We can compute $Val(\emptyset)$ using a recursive algorithm that traverses the game tree defined by the recurrence relations.
The depth of the recursion is $k$, which is polynomial in $n$.
At the base case, the algorithm iterates over $2^m$ random strings. While this takes exponential time, it requires only polynomial space to store the current $r$ and the count.
At each internal node, the algorithm iterates over the set of possible messages ($\mathcal{Y}$ or $\mathcal{Q}$). The size of these sets is exponential in the message length, but the space required is only that needed to store the current message, the return value of the recursive call, and the running maximum or sum. Since the message length and the value $Val(h)$ (which is bounded by $2^m$) require polynomial bits, the space used per stack frame is polynomial.
The total space complexity is proportional to the depth of the recursion times the space per frame, i.e., $O(k \cdot \text{poly}(n)) = \text{poly}(n)$.

Since the algorithm decides $L$ using space polynomial in the input size, we conclude that $L \in \mathit{PSPACE}$. Consequently, $\mathit{IP} \subseteq \mathit{PSPACE}$.
\end{proof}

\subsubsection{Claude}
\begin{proof}
Let $L \in \mathit{IP}$. We must show that $L \in \mathit{PSPACE}$. By definition of $\mathit{IP}$, there exists an interactive proof system $(P, V)$ where $V$ is a probabilistic polynomial-time verifier such that for some polynomial $p$, the interaction consists of at most $p(n)$ rounds for inputs of length $n$, and:
\begin{itemize}
\item (Completeness) If $x \in L$, then $\Pr[V \text{ accepts after interacting with } P] \geq \frac{2}{3}$
\item (Soundness) If $x \notin L$, then for all provers $P^*$, $\Pr[V \text{ accepts after interacting with } P^*] \leq \frac{1}{3}$
\end{itemize}

We construct a $\mathit{PSPACE}$ algorithm to decide $L$ as follows. The key observation is that we can compute the maximum acceptance probability over all possible prover strategies using a recursive algorithm that explores the game tree of the interactive proof.

Let $x$ be the input of length $n$. The interaction between prover and verifier can be viewed as a game tree where:

At each node corresponding to the verifier's turn, the verifier sends a message according to its (polynomial-time computable) strategy based on the conversation history and its random coins.

At each node corresponding to the prover's turn, the prover sends a message (we consider the optimal prover strategy that maximizes acceptance probability).

For a given sequence of random coin tosses $r$ for the verifier, let $\text{acc}(h, r)$ denote whether the verifier accepts given conversation history $h$ and random string $r$. We can compute the maximum acceptance probability $p_{\max}$ that any prover can achieve by the following recursive procedure:

For a history $h$ at the prover's turn, let $\alpha(h, r)$ be the maximum probability that the verifier accepts given history $h$ and remaining random coins $r$. This is:
\[
\alpha(h, r) = \max_{m} \beta(h \circ m, r)
\]
where the maximum is over all possible messages $m$ the prover could send, and $h \circ m$ denotes the history extended with message $m$.

For a history $h$ at the verifier's turn, let $\beta(h, r)$ be the maximum acceptance probability. The verifier's next message is determined by $h$ and the random coins $r$, so:
\[
\beta(h, r) = \alpha(h \circ m_V(h, r), r)
\]
where $m_V(h, r)$ is the message the verifier sends given history $h$ and random coins $r$.

At a terminal history (after all $p(n)$ rounds), we have:
\[
\alpha(h, r) = \begin{cases} 1 & \text{if } V \text{ accepts} \\ 0 & \text{otherwise} \end{cases}
\]

The maximum acceptance probability is:
\[
p_{\max} = \mathbb{E}_{r}[\alpha(\epsilon, r)] = \frac{1}{2^{q(n)}} \sum_{r \in \{0,1\}^{q(n)}} \alpha(\epsilon, r)
\]
where $q(n)$ is a polynomial bounding the number of random bits used by $V$, and $\epsilon$ denotes the empty history.

Now we analyze the space complexity. The recursion depth is at most $p(n)$ (the number of rounds). At each level, we need to store:

The current history $h$, which has length at most $p(n) \cdot q(n)$ for some polynomial $q$ (polynomial number of rounds, polynomial-sized messages per round).

The random string $r$ of length $q(n)$.

When computing $\max_m \beta(h \circ m, r)$, we can iterate through all possible messages $m$ (polynomial number of them, each of polynomial size) and recursively compute $\beta(h \circ m, r)$ for each, keeping track of the maximum. Crucially, we can reuse the same space for each recursive call.

Therefore, the total space required is polynomial in $n$: we need space for the recursion stack of depth $p(n)$, with each stack frame storing polynomial-sized data.

The algorithm decides membership in $L$ as follows: compute $p_{\max}$ using the above procedure. Accept if and only if $p_{\max} > \frac{1}{2}$ (choosing any threshold strictly between $\frac{1}{3}$ and $\frac{2}{3}$).

By the completeness and soundness conditions, if $x \in L$ then $p_{\max} \geq \frac{2}{3} > \frac{1}{2}$, and if $x \notin L$ then $p_{\max} \leq \frac{1}{3} < \frac{1}{2}$.

Therefore, this $\mathit{PSPACE}$ algorithm correctly decides $L$, establishing that $L \in \mathit{PSPACE}$.

Since $L$ was an arbitrary language in $\mathit{IP}$, we conclude that $\mathit{IP} \subseteq \mathit{PSPACE}$.
\end{proof}

\subsection{Problem 17}

\subsubsection{Problem Statement}
\begin{theorem}[Shamir \cite{Shamir92}]
A quantified Boolean formula (QBF) is a Boolean formula $\Phi$ of the form
\[
(Q_1 x_1)(Q_2 x_2) \cdots (Q_n x_n) F(x_1, x_2, \ldots, x_n),
\]
where each $x_i$ is a Boolean variable, each $Q_i$ is either the universal ($\forall$) or the existential ($\exists$) quantifier, and $F$ is a quantifier-free Boolean formula. It is known that $\mathrm{QBF} \in \mathrm{PSPACE}$-complete. By devising an interactive proof system for QBF, show that $\mathrm{PSPACE} \subseteq \mathrm{IP}$.
\end{theorem}

\subsubsection{GPT}
\begin{proof}
Let $\Phi = (Q_1 x_1)(Q_2 x_2) \cdots (Q_n x_n) F(x_1, x_2, \ldots, x_n)$ be a quantified Boolean formula, where each $x_i$ is a Boolean variable, each $Q_i$ is either the universal ($\forall$) or existential ($\exists$) quantifier, and $F(x_1, x_2, \ldots, x_n)$ is a quantifier-free Boolean formula. The problem of deciding the satisfiability of $\Phi$ is known to be $\mathrm{PSPACE}$-complete. We aim to show that $\mathrm{PSPACE} \subseteq \mathrm{IP}$ by constructing an interactive proof system for QBF.

Let us define the following game between a prover and a verifier:

1. **Input**: The verifier receives the formula $\Phi = (Q_1 x_1)(Q_2 x_2) \cdots (Q_n x_n) F(x_1, x_2, \ldots, x_n)$ as input.
2. **Prover's Strategy**: The prover's strategy is to guess an assignment for the existentially quantified variables and send it to the verifier.
3. **Verifier's Response**: The verifier then uses the universal quantifiers to check whether the assignment provided by the prover leads to a satisfying assignment for the entire formula. Specifically:
   - For each universally quantified variable $x_i$, the verifier can check all possible assignments.
   - For each existentially quantified variable $x_i$, the prover sends an assignment to the verifier.

The verifier checks whether the assignment provided by the prover leads to a satisfying truth assignment for the formula $F(x_1, x_2, \ldots, x_n)$. If such an assignment exists, the prover can convince the verifier with high probability.

The interactive proof system proceeds as follows:

- **Step 1**: The verifier sends a random challenge to the prover, which corresponds to one of the existentially quantified variables, say $x_i$. The verifier queries the prover for the value of $x_i$.
- **Step 2**: The prover responds with the value of $x_i$. The verifier then checks the consistency of this response with the current assignment and continues to the next round.
- **Step 3**: The process repeats for all existentially quantified variables. After each round, the verifier checks if the current assignment leads to a valid truth assignment for $F(x_1, x_2, \ldots, x_n)$, taking into account the universal quantifiers.

Since each universal quantifier corresponds to a forced choice for the verifier, the verifier can check the validity of the current assignment by iterating through all possible assignments for the universal quantifiers. This can be done efficiently in polynomial space, which establishes that the entire process can be conducted in $\mathrm{PSPACE}$.

Thus, by constructing this interactive proof system for QBF, we have shown that $\mathrm{PSPACE} \subseteq \mathrm{IP}$.

\end{proof}

\subsubsection{Grok}

\subsubsection{Gemini}
\begin{proof}
Since $\mathrm{TQBF}$ is $\mathrm{PSPACE}$-complete, it suffices to show that $\mathrm{TQBF} \in \mathrm{IP}$. Let $\Phi$ be a Quantified Boolean Formula of the form $\Phi = (Q_1 x_1)(Q_2 x_2) \cdots (Q_n x_n) F(x_1, \ldots, x_n)$, where $Q_i \in \{\forall, \exists\}$ and $F$ is a quantifier-free Boolean formula. We assume without loss of generality that the variables are quantified in the order of their indices.

We arithmetize the formula over a finite field $\mathbb{F}$. We choose a prime $p$ such that $|\mathbb{F}| = p > 2^{n^3}$ to ensure the soundness error is negligible. We map the Boolean values $\{T, F\}$ to $\{1, 0\} \subset \mathbb{F}$. The Boolean connectives in $F$ are replaced by polynomials: $x \wedge y \mapsto xy$, $\neg x \mapsto 1-x$, and $x \vee y \mapsto 1 - (1-x)(1-y)$. This yields a multivariate polynomial $P(x_1, \ldots, x_n)$ which agrees with $F$ on Boolean inputs.

The value of the QBF $\Phi$ is determined by evaluating the quantified expression. We replace the quantifiers with operators on polynomials. The existential quantifier $\exists x_i$ is replaced by the summation operator $\mathcal{S}_{x_i} g = g|_{x_i=0} + g|_{x_i=1}$. The universal quantifier $\forall x_i$ is replaced by the product operator $\mathcal{P}_{x_i} g = (g|_{x_i=0}) \cdot (g|_{x_i=1})$.

A direct application of these operators may result in a polynomial of doubly exponential degree. To maintain low degree, we introduce a linearization operator $L_{x_i}$ defined by $L_{x_i} g(x_1, \ldots, x_n) = x_i \cdot g(\ldots, 1, \ldots) + (1-x_i) \cdot g(\ldots, 0, \ldots)$. Note that for $x_i \in \{0,1\}$, $L_{x_i} g = g$, and $L_{x_i} g$ is linear in $x_i$. We insert linearization operators for all variables $x_1, \ldots, x_n$ after each quantifier block. The value of $\Phi$ is thus equivalent to the value of the expression:
\[
V = Q'_1 Q'_2 \cdots Q'_m P(x_1, \ldots, x_n),
\]
where the sequence $Q'_1, \ldots, Q'_m$ consists of the arithmetic quantifiers ($\mathcal{S}_{x_i}$ or $\mathcal{P}_{x_i}$) interleaved with the linearization operators $L_{x_j}$. The total number of operators $m$ is polynomial in $n$. Let $g_0 = P$ and let $g_k$ be the polynomial resulting from applying the $k$-th operator (from right to left) to $g_{k-1}$. The Prover claims that the value of the fully quantified formula is some value $v \in \mathbb{F}$.

The protocol proceeds in rounds $k=m$ down to $1$. In each round $k$, the Prover claims that $g_k(r_1, \ldots, r_d) = \beta$, where $r_1, \ldots$ are random values chosen by the Verifier in previous rounds. The Verifier must reduce this claim to a claim about $g_{k-1}$.

Case 1: The operator is $\mathcal{S}_{x_i}$ (Summation). The claim is $\sum_{x_i \in \{0,1\}} g_{k-1}(\ldots, x_i, \ldots) = \beta$. The Prover sends a univariate polynomial $s(x)$ claimed to be $g_{k-1}(\ldots, x, \ldots)$ with all other variables fixed to the current random values. The Verifier checks if $s(0) + s(1) = \beta$. If true, the Verifier picks a random element $r \in \mathbb{F}$, computes $\beta' = s(r)$, and the protocol proceeds to verify $g_{k-1}(\ldots, r, \ldots) = \beta'$.

Case 2: The operator is $\mathcal{P}_{x_i}$ (Product). The claim is $\prod_{x_i \in \{0,1\}} g_{k-1}(\ldots, x_i, \ldots) = \beta$. The Prover sends a univariate polynomial $s(x)$ claimed to be $g_{k-1}(\ldots, x, \ldots)$. The Verifier checks if $s(0) \cdot s(1) = \beta$. If true, the Verifier picks a random $r \in \mathbb{F}$, sets $\beta' = s(r)$, and requires the Prover to verify $g_{k-1}(\ldots, r, \ldots) = \beta'$.

Case 3: The operator is $L_{x_i}$ (Linearization). The claim involves $L_{x_i} g_{k-1}$. Since $L_{x_i}$ is linear in $x_i$ and defined by values at $0$ and $1$, this step is analogous to the summation step but with the specific functional form of $L$. The Prover sends the univariate polynomial representing the restriction of $g_{k-1}$ to the line defined by the previous random assignments. The Verifier checks consistency with the definition of $L$ and updates the target value at a random point $r$.

At the end of the protocol (after removing all operators), the Verifier checks the value of the polynomial $P$ at the randomly chosen point $(r_1, \ldots, r_n)$ by direct evaluation. If the computed value matches the current claim $\beta$, the Verifier accepts; otherwise, they reject.

Completeness: If $\Phi$ is true (value non-zero), an honest Prover can supply the correct polynomials $s(x)$ at each step, and the Verifier will accept with probability 1.

Soundness: If the Prover's claim is false, then at the current step, the true polynomial $h(x)$ and the Prover's polynomial $s(x)$ are distinct. Since the degree of these polynomials is bounded by a constant (due to linearization) or $O(n)$ (for the initial $P$), the Schwartz-Zippel lemma implies they agree on at most $d$ points. The probability that the Verifier chooses an $r$ such that $s(r) = h(r)$ is at most $d/|\mathbb{F}|$. Summing over $m$ rounds, the total probability of error is bounded by $m \cdot \frac{\text{poly}(n)}{|\mathbb{F}|}$. By choosing $|\mathbb{F}|$ exponentially large, this probability is negligible (e.g., $< 1/3$).

Thus, there exists an interactive proof system for TQBF, implying $\mathrm{PSPACE} \subseteq \mathrm{IP}$.
\end{proof}

\subsubsection{Claude}
\begin{proof}
We prove that $\mathrm{PSPACE} \subseteq \mathrm{IP}$ by showing that every language in PSPACE has an interactive proof system. Since QBF is PSPACE-complete, it suffices to construct an interactive proof system for QBF.

Let $\Phi = (Q_1 x_1)(Q_2 x_2) \cdots (Q_n x_n) F(x_1, x_2, \ldots, x_n)$ be a quantified Boolean formula where each $Q_i \in \{\forall, \exists\}$ and $F$ is a quantifier-free Boolean formula in conjunctive normal form.

We first arithmetize the formula. Replace each Boolean variable $x_i \in \{0,1\}$ with an arithmetic variable taking values in $\{0,1\}$, replace $\land$ with multiplication, replace $\lor$ with addition modulo appropriate bounds, and replace $\neg x$ with $1-x$. More precisely, we construct a polynomial $\phi(x_1, \ldots, x_n)$ over a finite field $\mathbb{F}_p$ where $p$ is a prime larger than $2^n \cdot |F|$, such that for all $(a_1, \ldots, a_n) \in \{0,1\}^n$, we have $\phi(a_1, \ldots, a_n) \neq 0$ if and only if $F(a_1, \ldots, a_n) = \mathrm{true}$.

The key observation is that a quantified Boolean formula can be evaluated recursively. For a formula $\Psi = (\exists x) \psi(x, \bar{y})$, we have $\Psi(\bar{y}) = \psi(0, \bar{y}) \lor \psi(1, \bar{y})$, which arithmetizes to $\psi_{\mathrm{arith}}(0, \bar{y}) + \psi_{\mathrm{arith}}(1, \bar{y}) - \psi_{\mathrm{arith}}(0, \bar{y}) \cdot \psi_{\mathrm{arith}}(1, \bar{y})$. For a formula $\Psi = (\forall x) \psi(x, \bar{y})$, we have $\Psi(\bar{y}) = \psi(0, \bar{y}) \land \psi(1, \bar{y})$, which arithmetizes to $\psi_{\mathrm{arith}}(0, \bar{y}) \cdot \psi_{\mathrm{arith}}(1, \bar{y})$.

We now describe the interactive proof system. The prover claims that $\Phi$ is true, which means the arithmetized formula evaluates to a nonzero value. The protocol proceeds in $n$ rounds, processing one quantifier at a time.

At the beginning of round $i$, both parties maintain a polynomial $\phi_i(x_i, x_{i+1}, \ldots, x_n)$ representing the subformula $(Q_i x_i)(Q_{i+1} x_{i+1}) \cdots (Q_n x_n) F(x_1, \ldots, x_n)$ where variables $x_1, \ldots, x_{i-1}$ have been assigned specific values.

In round $i$, the prover sends a univariate polynomial $g_i(z)$ claimed to equal $\phi_i(z, x_{i+1}, \ldots, x_n)$ when $x_{i+1}, \ldots, x_n$ range over $\{0,1\}$ according to the quantifiers. Specifically, if $Q_i = \exists$, then $g_i(z)$ should satisfy: there exists an assignment to $(x_{i+1}, \ldots, x_n) \in \{0,1\}^{n-i}$ making the formula true if and only if $g_i(0) \neq 0$ or $g_i(1) \neq 0$. If $Q_i = \forall$, analogous conditions hold.

The verifier checks that $g_i$ is consistent with the previous round. If $i = 1$, the verifier checks whether the claimed truth value matches $g_1(0)$ or $g_1(1)$ being nonzero (for $\exists$) or both being nonzero (for $\forall$). The verifier then picks a random challenge $r_i \in \mathbb{F}_p$ and asks the prover to continue with $\phi_{i+1}$ being $\phi_i$ evaluated at $x_i = r_i$.

After $n$ rounds, all variables have been assigned values $r_1, \ldots, r_n$. The verifier can now directly evaluate $F(r_1, \ldots, r_n)$ by computing the arithmetization of $F$ at these values. The verifier accepts if this final value matches the value claimed by the prover in the last round.

For completeness, if $\Phi$ is true, an honest prover can convince the verifier by sending the correct polynomials at each step. The final evaluation will match because all intermediate claims were correct.

For soundness, suppose $\Phi$ is false. At some round $i$, the prover must send a polynomial $g_i$ that does not correctly represent $\phi_i(z, x_{i+1}, \ldots, x_n)$ for the quantified variables. The true polynomial $g_i^*(z)$ differs from $g_i(z)$. Since they are distinct polynomials of degree at most $d = \mathrm{poly}(n)$, they agree on at most $d$ points. The verifier's random choice $r_i$ causes the prover to be caught with probability at least $1 - d/p$. By the union bound over all $n$ rounds and choosing $p$ sufficiently large (polynomial in $n$ suffices), the verifier catches a cheating prover with probability at least $2/3$.

The verifier runs in polynomial time: it performs $n$ rounds, each involving polynomial evaluation and random sampling. The verifier uses only polynomial space to store the current partial assignment and verify the final formula evaluation.

Since QBF is PSPACE-complete and we have constructed an interactive proof system for QBF, we conclude that every language in PSPACE has an interactive proof system, establishing $\mathrm{PSPACE} \subseteq \mathrm{IP}$.
\end{proof}

\subsection{Problem 18}

\subsubsection{Problem Statement}
\begin{theorem}[Exercise 7.18]
Define the complexity class $\mathit{MIP}$ as the generalization of $\mathit{IP}$ where the verifier has access to two provers and the provers are not allowed to communicate with each other once the verifier starts executing. Then $\mathit{MIP} = \mathit{PCP}$.
\end{theorem}

\subsubsection{GPT}
\begin{proof}
We will prove that $\mathit{MIP} = \mathit{PCP}$ by showing both inclusions: $\mathit{MIP} \subseteq \mathit{PCP}$ and $\mathit{PCP} \subseteq \mathit{MIP}$.

\textbf{1. Proving that $\mathit{MIP} \subseteq \mathit{PCP}$:}

Let $L \in \mathit{MIP}$, i.e., there exists a two-prover interactive proof system for $L$. This means that there exists a verifier $V$ with access to two provers $P_1$ and $P_2$, such that:

\begin{itemize}
    \item $V$ sends a message to $P_1$ and $P_2$.
    \item Each prover replies with a message to $V$.
    \item $V$ accepts or rejects based on the responses from the provers, following a probabilistic procedure.
\end{itemize}

The key property is that the two provers do not communicate once $V$ starts executing. We can transform this two-prover interactive proof system into a proof system with a single prover, which will allow us to show that $L \in \mathit{PCP}$.

To do so, we introduce a single prover $P$ who is capable of sending a message that encodes the information for both provers. The verifier $V$ will now perform a sequence of random queries to this single prover, analogous to the queries made in the original $\mathit{MIP}$ protocol. 

This process can be formalized as follows:
\begin{itemize}
    \item The verifier $V$ sends a random challenge to $P$.
    \item $P$ then responds with a sequence of messages corresponding to the responses that $P_1$ and $P_2$ would have given.
    \item $V$ checks the responses according to the rules of the original $\mathit{MIP}$ protocol and decides whether to accept or reject.
\end{itemize}

This transformation shows that a two-prover protocol in $\mathit{MIP}$ can be converted into a probabilistically checkable proof in $\mathit{PCP}$, hence $\mathit{MIP} \subseteq \mathit{PCP}$.

\textbf{2. Proving that $\mathit{PCP} \subseteq \mathit{MIP}$:}

Now we show that if a language $L$ belongs to $\mathit{PCP}$, then $L$ also belongs to $\mathit{MIP}$.

Let $L \in \mathit{PCP}$, i.e., there exists a probabilistically checkable proof system for $L$. This means that there exists a verifier $V$ and a single prover $P$ such that:

\begin{itemize}
    \item The verifier $V$ sends random challenges to the prover $P$.
    \item The prover responds with a proof that $V$ can probabilistically verify.
    \item $V$ accepts or rejects based on the responses.
\end{itemize}

We will now convert this $\mathit{PCP}$ protocol into an $\mathit{MIP}$ protocol. The key idea is to simulate the $\mathit{PCP}$ verifier with two provers, where the second prover $P_2$ will act as a "helper" to $P_1$, ensuring that the prover's responses are consistent and verifiable.

To do this, we introduce a second prover $P_2$ who is given the same proof as $P_1$, but with the added role of assisting $P_1$ in answering queries from the verifier. The process is as follows:

\begin{itemize}
    \item The verifier $V$ sends a random challenge to both $P_1$ and $P_2$.
    \item The provers respond with messages to $V$.
    \item $V$ verifies the correctness of the responses using the same probabilistic procedure as in the original $\mathit{PCP}$ protocol.
    \item If $V$ detects any inconsistency between the responses from $P_1$ and $P_2$, it rejects.
\end{itemize}

This simulation shows that any $\mathit{PCP}$ protocol can be converted into a two-prover interactive proof system, hence $\mathit{PCP} \subseteq \mathit{MIP}$.

\textbf{Conclusion:}

Since we have shown both inclusions, $\mathit{MIP} \subseteq \mathit{PCP}$ and $\mathit{PCP} \subseteq \mathit{MIP}$, it follows that $\mathit{MIP} = \mathit{PCP}$.
\end{proof}

\subsubsection{Grok}

\subsubsection{Gemini}
\begin{proof}
We prove the equality $\mathit{MIP} = \mathit{PCP}$ by establishing inclusion in both directions. We assume the standard complexity-theoretic definitions where $\mathit{MIP}$ denotes the class of languages accepted by a polynomial-time verifier with access to two non-communicating provers, and $\mathit{PCP}$ denotes the class of languages accepted by a probabilistic polynomial-time verifier with random access to a proof string (specifically $\mathit{PCP}(\text{poly}(n), \text{poly}(n))$).

First, we show that $\mathit{MIP} \subseteq \mathit{PCP}$. Let $L \in \mathit{MIP}$. By definition, there exists a probabilistic polynomial-time verifier $V$ and two provers $P_1$ and $P_2$ such that if $x \in L$, $V$ accepts with probability $1$, and if $x \notin L$, for any strategies $P_1^*, P_2^*$, $V$ accepts with probability at most $\epsilon < 1$. On input $x$ and random string $r$, $V$ generates queries $q_1(x, r)$ and $q_2(x, r)$ for the respective provers. The provers return answers $a_1 = P_1(q_1)$ and $a_2 = P_2(q_2)$. The verifier then evaluates a predicate $D(x, r, a_1, a_2)$.

To construct a $\mathit{PCP}$ system for $L$, we define the proof string $\Pi$ to be the concatenation of the truth tables of the strategies of $P_1$ and $P_2$. Specifically, $\Pi = \Pi_1 \circ \Pi_2$, where $\Pi_1[q] = P_1(q)$ and $\Pi_2[q] = P_2(q)$ for all possible queries $q$. The $\mathit{PCP}$ verifier $V'$ simulates the $\mathit{MIP}$ verifier $V$. On input $x$, $V'$ samples a random string $r$, computes the queries $q_1$ and $q_2$, and queries the proof string $\Pi$ at the locations corresponding to $q_1$ in $\Pi_1$ and $q_2$ in $\Pi_2$. $V'$ accepts if and only if $V$ accepts given the retrieved answers. Since $V$ runs in polynomial time, the query lengths are polynomial in $|x|$, implying the proof string $\Pi$ is of exponential length, which is permissible for $\mathit{PCP}$. The completeness and soundness properties of the $\mathit{MIP}$ system transfer directly to the $\mathit{PCP}$ system. Thus, $L \in \mathit{PCP}$.

Conversely, we show that $\mathit{PCP} \subseteq \mathit{MIP}$. Let $L \in \mathit{PCP}$. There exists a verifier $V$ that, on input $x$ and random string $r$, generates a sequence of queries $Q = (i_1, \dots, i_k)$ to a proof string $\pi$, receives answers $A = (a_1, \dots, a_k)$, and accepts according to a predicate $D(x, r, A)$. Here $k$ is polynomial in $|x|$. We construct an $\mathit{MIP}$ protocol with two provers, $P_1$ and $P_2$.

The $\mathit{MIP}$ verifier $V_{MIP}$ operates as follows:
1. $V_{MIP}$ simulates $V$ to generate the query set $Q = (i_1, \dots, i_k)$.
2. $V_{MIP}$ sends the entire set $Q$ to prover $P_1$.
3. $P_1$ responds with a sequence of answers $A' = (b_1, \dots, b_k)$.
4. $V_{MIP}$ selects an index $j \in \{1, \dots, k\}$ uniformly at random.
5. $V_{MIP}$ sends the single query $i_j$ to prover $P_2$.
6. $P_2$ responds with a value $b'$.
7. $V_{MIP}$ accepts if and only if $b_j = b'$ and $D(x, r, A') = 1$.

We analyze the completeness and soundness. If $x \in L$, there exists a proof string $\pi$ such that $V$ accepts with probability $1$. Let $P_1$ and $P_2$ answer queries consistently with $\pi$. Specifically, $P_1(Q) = (\pi[i_1], \dots, \pi[i_k])$ and $P_2(q) = \pi[q]$. In this case, $b_j = \pi[i_j] = b'$, and the predicate holds, so $V_{MIP}$ accepts with probability $1$.

If $x \notin L$, we must show that for any pair of provers $(P_1^*, P_2^*)$, the acceptance probability is bounded away from $1$. Since $P_2^*$ receives only a single query $q$ and does not see $Q$, its strategy defines a fixed function (or "effective proof string") $\tilde{\pi}$ such that $P_2^*(q) = \tilde{\pi}[q]$. Prover $P_1^*$ receives $Q$ and produces answers $A' = (b_1, \dots, b_k)$. For $V_{MIP}$ to accept, two conditions must be met: the predicate $D(x, r, A')$ must be true, and the consistency check $b_j = P_2^*(i_j)$ must pass.

Consider the event that $P_1^*$ answers honestly according to $\tilde{\pi}$ (i.e., $b_m = \tilde{\pi}[i_m]$ for all $m \in \{1, \dots, k\}$). In this case, the acceptance probability is exactly the probability that the original PCP verifier accepts the proof string $\tilde{\pi}$, which is bounded by the soundness error of the PCP system (say, $1/2$).
Now consider the event that $P_1^*$ attempts to cheat by providing answers $A'$ such that $D(x, r, A')$ is true, but $A'$ is inconsistent with $\tilde{\pi}$. This implies there exists at least one index $m$ such that $b_m \neq \tilde{\pi}[i_m]$. Since $V_{MIP}$ chooses the check index $j$ uniformly at random from $\{1, \dots, k\}$, the probability that the verifier detects this inconsistency is
\[ \Pr[b_j \neq P_2^*(i_j)] = \Pr[b_j \neq \tilde{\pi}[i_j]] \ge \frac{1}{k}. \]
Thus, if the provers deviate from the strategy defined by $\tilde{\pi}$, they are caught with probability at least $1/k$. By repeating this protocol in parallel or sequentially a polynomial number of times, the error probability can be reduced to any desired constant $\epsilon$. Therefore, $L \in \mathit{MIP}$.

Having shown both inclusions, we conclude that $\mathit{MIP} = \mathit{PCP}$.
\end{proof}

\subsubsection{Claude}
\begin{proof}
We prove that $\mathit{MIP} = \mathit{NEXP}$ and $\mathit{PCP}(\mathrm{poly}, \mathrm{poly}) = \mathit{NEXP}$, which establishes the equality $\mathit{MIP} = \mathit{PCP}$.

\textbf{Part 1: $\mathit{MIP} \subseteq \mathit{NEXP}$.}

Let $L \in \mathit{MIP}$ with verifier $V$ running in time $p(n)$ for polynomial $p$. We construct a nondeterministic exponential-time algorithm for $L$. On input $x$ of length $n$, the algorithm nondeterministically guesses the complete transcript of the interaction between $V$ and the two provers $P_1$ and $P_2$. Since $V$ runs in polynomial time, the number of rounds is polynomially bounded, and each message has polynomial length. The total transcript has size at most $2^{p(n)}$ (accounting for all possible random choices of $V$). The nondeterministic algorithm can guess this transcript and verify in exponential time that it represents a valid interaction where $V$ accepts. For $x \in L$, there exist provers that make $V$ accept with high probability, so an accepting transcript exists. For $x \notin L$, no transcript (even with cheating provers) can make $V$ accept with probability greater than $1/2$, so we reject such transcripts. This shows $L \in \mathit{NEXP}$.

\textbf{Part 2: $\mathit{NEXP} \subseteq \mathit{MIP}$.}

Let $L \in \mathit{NEXP}$ be decided by a nondeterministic Turing machine $M$ in time $2^{n^c}$ for some constant $c$. We design a multi-prover interactive proof for $L$. The key idea is to use the two provers to verify an exponentially large tableau representing the computation of $M$ on input $x$.

The computation tableau is a $2^{n^c} \times 2^{n^c}$ matrix $T$ where entry $T[i,j]$ represents the $j$-th cell of the tape at time step $i$. The verifier cannot read the entire tableau but can use the two non-communicating provers to probabilistically verify its consistency.

The protocol works as follows. The provers claim that there exists an accepting computation, which corresponds to a valid tableau $T$ with $T[2^{n^c}, \cdot]$ in an accepting state. The verifier performs the following checks:

(1) \textit{Initial configuration check:} The verifier asks both provers independently for random positions in row $0$ and checks that they correspond to the correct initial configuration with input $x$.

(2) \textit{Local consistency check:} The verifier chooses a random time step $i \in [2^{n^c}]$ and random position $j \in [2^{n^c}]$. It asks both provers independently for the $3 \times 3$ neighborhood around position $(i,j)$ and $(i+1,j)$. If the provers provide inconsistent answers, the verifier rejects. If consistent, the verifier checks that the transition from time $i$ to time $i+1$ follows the rules of $M$.

(3) \textit{Accepting configuration check:} The verifier asks both provers independently for random positions in the final row $2^{n^c}$ and checks that they indicate an accepting state.

Since the provers cannot communicate after the protocol starts, if they provide a tableau that is inconsistent or invalid, the verifier will detect this with high probability by querying random positions. If the provers always give consistent answers, then by repeating the checks polynomially many times, the verifier can be convinced with high probability that a valid accepting computation exists. This shows $L \in \mathit{MIP}$.

\textbf{Part 3: $\mathit{PCP}(\mathrm{poly}, \mathrm{poly}) = \mathit{NEXP}$.}

By the PCP theorem and its generalizations, we have $\mathit{NP} = \mathit{PCP}(\log n, O(1))$. The analogous result for $\mathit{NEXP}$ states that $\mathit{NEXP} = \mathit{PCP}(\mathrm{poly}, \mathrm{poly})$, meaning languages in $\mathit{NEXP}$ have probabilistically checkable proofs with polynomial-length proofs and polynomial query complexity.

\textbf{Part 4: Establishing the equivalence $\mathit{MIP} = \mathit{PCP}$.}

To show $\mathit{MIP} \subseteq \mathit{PCP}(\mathrm{poly}, \mathrm{poly})$: Given an MIP protocol for language $L$, we convert it to a PCP. The PCP proof string consists of the complete strategy tables for both provers $P_1$ and $P_2$, specifying their responses to all possible query sequences. Since each prover receives polynomially many bits of queries and responds with polynomially many bits, each strategy table has size $2^{\mathrm{poly}(n)}$. The PCP verifier simulates the MIP verifier, using its random coins to generate queries and reading the appropriate positions from the proof string (strategy tables) to simulate prover responses. This gives $\mathit{MIP} \subseteq \mathit{PCP}(\mathrm{poly}, \mathrm{poly})$.

To show $\mathit{PCP}(\mathrm{poly}, \mathrm{poly}) \subseteq \mathit{MIP}$: Given a PCP verifier for language $L$, we construct an MIP protocol. The two provers each hold a copy of the PCP proof string. When the MIP verifier needs to query position $i$ of the proof, it randomly chooses one of the two provers and asks for the bit at position $i$. To verify consistency, the verifier sometimes asks both provers for the same position and checks that they agree. Since the provers cannot communicate, if they provide inconsistent proof strings, they will be caught with high probability. This gives $\mathit{PCP}(\mathrm{poly}, \mathrm{poly}) \subseteq \mathit{MIP}$.

Combining Parts 1-4, we have $\mathit{MIP} = \mathit{NEXP} = \mathit{PCP}(\mathrm{poly}, \mathrm{poly})$, which establishes $\mathit{MIP} = \mathit{PCP}$ where $\mathit{PCP}$ denotes the class $\mathit{PCP}(\mathrm{poly}, \mathrm{poly})$.
\end{proof}

\subsection{Problem 19}

\subsubsection{Problem Statement}
\begin{theorem}
Prove the following relations directly from the definition of $\mathit{PCP}$, i.e., without invoking any results regarding $\mathit{PCP}$ stated in this chapter.

(a) Show that $P = \mathit{PCP}[0, 0]$.

(b) Show that $NP = \mathit{PCP}[0, poly(n)]$.

(c) Show that $\mathrm{co}\text{-}RP = \mathit{PCP}[poly(n), 0]$.
\end{theorem}

\subsubsection{GPT}
\begin{proof}
We will prove each part of the theorem directly from the definition of $\mathit{PCP}$, without invoking any external results.

\textbf{(a) $P = \mathit{PCP}[0, 0]$}: 

By the definition of the class $\mathit{PCP}[0, 0]$, a language $L$ is in $\mathit{PCP}[0, 0]$ if there exists a probabilistic polynomial-time verifier that accepts a pair $(x, \pi)$, where $x$ is an input and $\pi$ is a proof (or witness), and the verifier always accepts when $x \in L$ and rejects when $x \notin L$, with no randomness used in the verification process. 

For a language $L$ to be in $P$, there must exist a deterministic polynomial-time algorithm that decides $L$. Clearly, any language in $P$ can be verified by a probabilistic verifier in zero rounds of interaction and zero randomness, because the verifier can deterministically decide the language in polynomial time, thus making the acceptance criteria trivially satisfied. 

Hence, $P \subseteq \mathit{PCP}[0, 0]$.

Conversely, if a language $L$ is in $\mathit{PCP}[0, 0]$, then by the definition of $\mathit{PCP}[0, 0]$, there exists a verifier that accepts a pair $(x, \pi)$ in polynomial time, and the verifier makes no probabilistic decisions. This means the verification process can be deterministically computed in polynomial time, implying that $L$ can be decided by a deterministic polynomial-time algorithm. Thus, $L \in P$.

Therefore, $P = \mathit{PCP}[0, 0]$.

\textbf{(b) $NP = \mathit{PCP}[0, \mathit{poly}(n)]$}: 

By the definition of $\mathit{PCP}[0, \mathit{poly}(n)]$, a language $L$ is in $\mathit{PCP}[0, \mathit{poly}(n)]$ if there exists a probabilistic polynomial-time verifier that accepts a pair $(x, \pi)$, where $\pi$ is a proof of polynomial length, such that for all $x \in L$, there exists a proof $\pi$ that the verifier accepts with high probability, and for all $x \notin L$, the verifier rejects every proof with high probability.

On the other hand, a language $L$ is in $NP$ if there exists a nondeterministic polynomial-time verifier that accepts a pair $(x, \pi)$, where $\pi$ is a polynomial-length certificate, such that for all $x \in L$, there exists a certificate $\pi$ that the verifier accepts, and for all $x \notin L$, no certificate $\pi$ is accepted.

It is well known that $NP$ can be viewed as a subclass of $\mathit{PCP}[0, \mathit{poly}(n)]$ because the nondeterministic verifier can be simulated by a probabilistic verifier with a polynomial-length proof. Specifically, the verifier in the $NP$ setting will always accept when $x \in L$, and for $x \notin L$, the probabilistic verifier will reject all proofs with high probability. Thus, $NP \subseteq \mathit{PCP}[0, \mathit{poly}(n)]$.

Conversely, any language in $\mathit{PCP}[0, \mathit{poly}(n)]$ can be decided by a nondeterministic polynomial-time machine. This follows from the fact that a probabilistic verifier in $\mathit{PCP}[0, \mathit{poly}(n)]$ can be simulated by a nondeterministic machine that guesses a proof $\pi$ and verifies it in polynomial time. Therefore, $\mathit{PCP}[0, \mathit{poly}(n)] \subseteq NP$.

Hence, $NP = \mathit{PCP}[0, \mathit{poly}(n)]$.

\textbf{(c) $\text{co-}RP = \mathit{PCP}[\mathit{poly}(n), 0]$}: 

By the definition of $\mathit{PCP}[\mathit{poly}(n), 0]$, a language $L$ is in $\mathit{PCP}[\mathit{poly}(n), 0]$ if there exists a probabilistic polynomial-time verifier that accepts a pair $(x, \pi)$, where $\pi$ is a proof of polynomial length, such that for all $x \in L$, there exists a proof $\pi$ that the verifier accepts with high probability, and for all $x \notin L$, the verifier rejects all proofs with high probability, with no randomness used in the verification process.

For $\text{co-}RP$, a language $L$ is in $\text{co-}RP$ if there exists a probabilistic polynomial-time verifier such that for all $x \in L$, the verifier rejects with high probability, and for all $x \notin L$, there exists a proof $\pi$ that the verifier accepts with high probability. The verifier uses randomness, but this randomness can be made independent of the input.

It is known that the class $\text{co-}RP$ can be viewed as a special case of $\mathit{PCP}[\mathit{poly}(n), 0]$. Specifically, a $\text{co-}RP$ verifier that accepts $x \notin L$ with high probability and rejects $x \in L$ with high probability can be simulated by a verifier in $\mathit{PCP}[\mathit{poly}(n), 0]$ by using a polynomial-length proof and a probabilistic verifier with zero rounds of interaction. 

Conversely, if a language $L$ is in $\mathit{PCP}[\mathit{poly}(n), 0]$, then the corresponding verifier accepts with high probability for $x \in L$ and rejects with high probability for $x \notin L$, making it a $\text{co-}RP$ verifier.

Thus, $\text{co-}RP = \mathit{PCP}[\mathit{poly}(n), 0]$.

\end{proof}

\subsubsection{Grok}
\begin{proof}
We establish each equality in turn.

First, we show that $P = PCP[0,0]$.

($\subseteq$) Suppose $L \in P$. Then there exists a deterministic polynomial-time Turing machine $M$ such that for all $x \in \{0,1\}^*$, $x \in L$ if and only if $M(x) = 1$. Define a verifier $V$ for $L$ that, on input $x$, simulates $M(x)$ using no random bits and queries no bits of any proof $\pi$. If $M(x) = 1$, then $V$ accepts with probability $1$; otherwise, $V$ rejects with probability $1$. The running time of $V$ is polynomial in $|x|$. Thus, $L \in PCP[0,0]$.

($\supseteq$) Suppose $L \in PCP[0,0]$. Then there exists a probabilistic verifier $V$ running in time polynomial in $n = |x|$ such that $V$ uses $O(1)$ random bits (in fact, $0$ random bits, hence deterministic) and queries $O(1)$ proof bits (in fact, $0$ proof bits). Thus, $V$ is a deterministic polynomial-time algorithm that decides membership in $L$ without reference to any proof. Hence, $L \in P$.

Next, we show that $IP = PCP[0, \poly(n)]$.

($\subseteq$) Suppose $L \in IP$. Then there exists a deterministic polynomial-time Turing machine $M$ and a polynomial $p$ such that for all $x \in \{0,1\}^*$, $x \in L$ if and only if there exists $y \in \{0,1\}^{p(|x|)}$ with $M(x,y) = 1$. Define a verifier $V$ for $L$ that, on input $x$ and proof $\pi \in \{0,1\}^{p(n)}$, simulates $M(x, \pi)$ using no random bits but querying all $p(n)$ bits of $\pi$. If $M(x, \pi) = 1$, then $V$ accepts deterministically; otherwise, $V$ rejects deterministically. The running time of $V$ is polynomial in $n$. Thus, if $x \in L$, there exists $\pi$ such that $V$ accepts with probability $1$; if $x \notin L$, then for all $\pi$, $V$ rejects with probability $1$. Hence, $L \in PCP[0, \poly(n)]$.

($\supseteq$) Suppose $L \in PCP[0, \poly(n)]$. Then there exists a probabilistic verifier $V$ running in time polynomial in $n = |x|$ such that $V$ uses $O(1)$ random bits (hence deterministic) and queries at most $q(n)$ bits of the proof $\pi$, where $q(n) = O(\poly(n))$. Since $V$ uses no random bits, the positions queried by $V$ on input $x$ are fixed, say positions $i_1, \dots, i_{q(n)}$. To show $L \in IP$, we construct a nondeterministic polynomial-time Turing machine $N$ that, on input $x$, nondeterministically guesses a string $y \in \{0,1\}^{q(n)}$ and simulates $V(x, y')$, where $y'$ is the proof $\pi$ with $\pi_{i_j} = y_j$ for $j = 1, \dots, q(n)$ and all other bits of $\pi$ arbitrary (e.g., $0$). If this simulation accepts, then $N$ accepts; otherwise, $N$ rejects. The running time of $N$ is polynomial in $n$. If $x \in L$, there exists $\pi$ such that $V$ accepts deterministically, so choosing $y$ to match $\pi$ on the queried positions causes $N$ to accept. If $x \notin L$, then for all $\pi$, $V$ rejects deterministically, so for all $y$, $N$ rejects. Hence, $L \in IP$.

Finally, we show that $coRP = PCP[\poly(n), 0]$.

($\subseteq$) Suppose $L \in coRP$. Then $\overline{L} \in RP$, so there exists a probabilistic polynomial-time Turing machine $M$ such that for all $x \in \{0,1\}^*$, if $x \in \overline{L}$ then $\Pr[M(x) = 1] \ge 2/3$, and if $x \in L$ then $\Pr[M(x) = 1] = 0$. Define a verifier $V$ for $L$ that, on input $x$, simulates $M(x)$ but accepts if and only if $M(x)$ rejects, using the same random bits as $M$ and querying no proof bits. The running time of $V$ is polynomial in $n = |x|$. If $x \in L$, then $\Pr[M(x) = 1] = 0$, so $\Pr[M(x) = 0] = 1$ and thus $\Pr[V(x) \textrm{ accepts}] = 1$. If $x \notin L$ (i.e., $x \in \overline{L}$), then $\Pr[M(x) = 1] \ge 2/3$, so $\Pr[M(x) = 0] \le 1/3$ and thus $\Pr[V(x) \textrm{ accepts}] \le 1/3$. Hence, $L \in PCP[\poly(n), 0]$.

($\supseteq$) Suppose $L \in PCP[\poly(n), 0]$. Then there exists a probabilistic verifier $V$ running in time polynomial in $n = |x|$ such that $V$ uses $r(n) = O(\poly(n))$ random bits and queries $O(1)$ proof bits (in fact, $0$ proof bits, so independent of any proof $\pi$). To show $L \in coRP$, we construct a probabilistic polynomial-time Turing machine $M$ that, on input $x$, simulates $V(x, \pi, \cdot)$ for arbitrary $\pi$ (e.g., the all-zero string, since no bits are queried) but accepts if and only if $V$ rejects, using the same random bits. The running time of $M$ is polynomial in $n$. If $x \notin L$, then $\Pr[V(x) \textrm{ accepts}] \le 1/3$, so $\Pr[V(x) \textrm{ rejects}] \ge 2/3$ and thus $\Pr[M(x) = 1] \ge 2/3$. If $x \in L$, then $\Pr[V(x) \textrm{ accepts}] = 1$, so $\Pr[V(x) \textrm{ rejects}] = 0$ and thus $\Pr[M(x) = 1] = 0$. Hence, $L \in coRP$.
\end{proof}

\subsubsection{Gemini}
\begin{proof}
We proceed by proving each equality directly from the definition of the class $\mathit{PCP}[r(n), q(n)]$. Recall that a language $L$ is in $\mathit{PCP}[r(n), q(n)]$ if there exists a probabilistic polynomial-time verifier $V$ that, on input $x$ of length $n$, uses $O(r(n))$ random bits and queries $O(q(n))$ bits from a proof oracle $\pi$, satisfying:
(1) Completeness: If $x \in L$, there exists a proof $\pi$ such that $\Pr[V^\pi(x) = 1] = 1$.
(2) Soundness: If $x \notin L$, for every proof $\pi$, $\Pr[V^\pi(x) = 1] \le 1/2$.

First, we show that $P = \mathit{PCP}[0, 0]$.
Assume $L \in P$. There exists a deterministic polynomial-time machine $M$ that decides $L$. We construct a PCP verifier $V$ that, on input $x$, simulates $M(x)$. $V$ uses 0 random bits and makes 0 queries to the proof $\pi$. If $M(x)$ accepts, $V$ accepts; otherwise, $V$ rejects. If $x \in L$, $M(x)$ accepts, so $\Pr[V^\pi(x)=1] = 1$. If $x \notin L$, $M(x)$ rejects, so for any $\pi$, $\Pr[V^\pi(x)=1] = 0 \le 1/2$. Thus $L \in \mathit{PCP}[0, 0]$.
Conversely, assume $L \in \mathit{PCP}[0, 0]$. Let $V$ be the associated verifier. Since $V$ uses 0 random bits, it is deterministic. Since $V$ makes 0 queries, its output depends only on $x$ and is independent of $\pi$. Let $A(x)$ be the output of $V$ on input $x$. If $x \in L$, there exists $\pi$ such that $A(x) = 1$, which implies $A(x)=1$. If $x \notin L$, for all $\pi$, the probability $V$ accepts is $\le 1/2$. Since $V$ is deterministic, the acceptance probability is either 0 or 1, so it must be 0. Thus $A(x)=0$. Since $V$ runs in polynomial time, $L$ is decidable in deterministic polynomial time, so $L \in P$.

Second, we show that $NP = \mathit{PCP}[0, \mathrm{poly}(n)]$.
Assume $L \in NP$. By definition, there exists a polynomial $p(n)$ and a deterministic polynomial-time verification machine $M$ such that $x \in L$ if and only if there exists a certificate $y \in \{0,1\}^{p(n)}$ such that $M(x, y) = 1$. We construct a PCP verifier $V$ that interprets the proof oracle $\pi$ as the certificate $y$. $V$ uses 0 random bits. On input $x$, $V$ queries the first $p(n)$ bits of $\pi$ to obtain a string $y$ and outputs $M(x, y)$. If $x \in L$, there exists a certificate $y$ such that $M(x, y)=1$. Setting $\pi$ such that its prefix is $y$, $V^\pi(x)$ accepts with probability 1. If $x \notin L$, for all $y$, $M(x, y)=0$. Thus for all $\pi$, $V^\pi(x)$ accepts with probability 0 $\le 1/2$. Thus $L \in \mathit{PCP}[0, \mathrm{poly}(n)]$.
Conversely, assume $L \in \mathit{PCP}[0, \mathrm{poly}(n)]$. Let $V$ be the verifier using 0 random bits and $q(n)$ queries, where $q(n)$ is a polynomial. Since $V$ uses no randomness, it is a deterministic algorithm. On input $x$, the indices of the bits $V$ queries in $\pi$ are determined deterministically by $x$ and the values of previously queried bits. Since the total number of queries is polynomial, $V$ depends on at most a polynomial number of bits from $\pi$. Let $y$ represent the sequence of bits queried by $V$. We can define a deterministic polynomial-time verifier $M(x, y)$ that simulates $V$. If $x \in L$, there exists $\pi$ such that $V^\pi(x)$ accepts (probability 1). This corresponds to the existence of a witness $y$ such that $M(x, y)=1$. If $x \notin L$, for all $\pi$, $V^\pi(x)$ accepts with probability $\le 1/2$. Since $V$ is deterministic, it must accept with probability 0. Thus for all witnesses $y$, $M(x, y)=0$. This is precisely the definition of $NP$.

Third, we show that $\mathrm{co}\text{-}RP = \mathit{PCP}[\mathrm{poly}(n), 0]$.
Recall that $L \in \mathrm{co}\text{-}RP$ if there exists a probabilistic polynomial-time machine $M$ such that if $x \in L$, $\Pr[M(x)=1] = 1$, and if $x \notin L$, $\Pr[M(x)=1] \le 1/2$.
Assume $L \in \mathrm{co}\text{-}RP$. Let $M$ be the machine deciding $L$. Construct a PCP verifier $V$ that simulates $M(x)$. $V$ uses the same random bits as $M$ (polynomial in $n$) and makes 0 queries to $\pi$. The behavior of $V$ is independent of $\pi$. If $x \in L$, $\Pr[V^\pi(x)=1] = \Pr[M(x)=1] = 1$. If $x \notin L$, for any $\pi$, $\Pr[V^\pi(x)=1] = \Pr[M(x)=1] \le 1/2$. Thus $L \in \mathit{PCP}[\mathrm{poly}(n), 0]$.
Conversely, assume $L \in \mathit{PCP}[\mathrm{poly}(n), 0]$. Let $V$ be the verifier. $V$ makes 0 queries, so for any input $x$ and any random string $r$, the computation $V(x; r)$ does not depend on $\pi$. Therefore, $\Pr[V^\pi(x)=1]$ is constant for all $\pi$. Let this probability be $p_x$. By the completeness condition, if $x \in L$, there exists $\pi$ such that $p_x = 1$, which implies $\Pr[V(x)=1] = 1$. By the soundness condition, if $x \notin L$, for all $\pi$, $p_x \le 1/2$, which implies $\Pr[V(x)=1] \le 1/2$. Since $V$ runs in probabilistic polynomial time, this machine $V$ (ignoring $\pi$) satisfies the definition of a $\mathrm{co}\text{-}RP$ algorithm. Thus $L \in \mathrm{co}\text{-}RP$.
\end{proof}

\subsubsection{Claude}
\begin{proof}
Recall that $\mathit{PCP}[r(n), q(n)]$ denotes the class of languages $L$ for which there exists a probabilistic polynomial-time verifier $V$ that, on input $x$ of length $n$, uses at most $r(n)$ random bits, queries at most $q(n)$ bits of a proof string $\pi$, and satisfies: (i) if $x \in L$, then there exists a proof $\pi$ such that $V$ accepts with probability $1$; (ii) if $x \notin L$, then for all proofs $\pi$, $V$ accepts with probability at most $1/2$.

\textbf{Part (a): $P = \mathit{PCP}[0, 0]$.}

First, we show $P \subseteq \mathit{PCP}[0, 0]$. Let $L \in P$, and let $M$ be a deterministic polynomial-time Turing machine that decides $L$. We construct a $\mathit{PCP}$ verifier $V$ as follows: on input $x$, the verifier $V$ ignores the proof string entirely (queries $0$ bits), uses no randomness ($0$ random bits), and simulates $M$ on input $x$. The verifier accepts if and only if $M$ accepts $x$. Since $M$ runs in polynomial time, so does $V$. If $x \in L$, then $M$ accepts, so $V$ accepts with probability $1$ (regardless of the proof). If $x \notin L$, then $M$ rejects, so $V$ rejects with probability $1$ (regardless of the proof). Thus $L \in \mathit{PCP}[0, 0]$.

Next, we show $\mathit{PCP}[0, 0] \subseteq P$. Let $L \in \mathit{PCP}[0, 0]$ with verifier $V$. Since $V$ uses $0$ random bits, its computation is deterministic. Since $V$ queries $0$ bits of the proof, its decision is independent of the proof string. Therefore, on any input $x$, either $V$ always accepts (for all proofs) or $V$ always rejects (for all proofs). We construct a deterministic polynomial-time algorithm: on input $x$, simulate $V(x)$ with an arbitrary proof (say the empty string). If $V$ accepts, accept; otherwise, reject. If $x \in L$, then there exists a proof causing $V$ to accept, but since $V$ is independent of the proof, $V$ always accepts, so we accept. If $x \notin L$, then for all proofs $V$ accepts with probability at most $1/2$, but since $V$ is deterministic and independent of the proof, if $V$ accepts it does so with probability $1$, contradiction. Therefore $V$ must reject, and we correctly reject. Thus $L \in P$.

\textbf{Part (b): $NP = \mathit{PCP}[0, \mathit{poly}(n)]$.}

First, we show $NP \subseteq \mathit{PCP}[0, \mathit{poly}(n)]$. Let $L \in NP$ with polynomial-time verifier $M$ and polynomial witness length $p(n)$. We construct a $\mathit{PCP}$ verifier $V$ as follows: on input $x$ of length $n$, the verifier $V$ uses no randomness, reads the entire proof string $\pi$ (at most $p(n)$ bits), and simulates $M(x, \pi)$. The verifier accepts if and only if $M$ accepts. If $x \in L$, there exists a witness $w$ such that $M(x, w)$ accepts, so using $\pi = w$, the verifier $V$ accepts with probability $1$. If $x \notin L$, for all strings $w$, $M(x, w)$ rejects, so for all proofs $\pi$, $V$ rejects, accepting with probability $0 < 1/2$. Thus $L \in \mathit{PCP}[0, \mathit{poly}(n)]$.

Next, we show $\mathit{PCP}[0, \mathit{poly}(n)] \subseteq NP$. Let $L \in \mathit{PCP}[0, \mathit{poly}(n)]$ with verifier $V$. Since $V$ uses $0$ random bits, it is deterministic. Since $V$ queries at most $q(n) = \mathit{poly}(n)$ bits, we can provide these bits as a witness. We construct an $NP$ verifier $M$ as follows: on input $x$ and witness $w$ (encoding the queried bits and their positions), verify that $w$ is consistent with some proof string, then simulate $V$ using the bits from $w$ to answer queries. Accept if $V$ accepts. If $x \in L$, there exists a proof $\pi$ such that $V$ accepts deterministically, so we can encode the relevant bits of $\pi$ as a witness $w$ and $M$ accepts. If $x \notin L$, for all proofs $\pi$, $V$ rejects (since $V$ is deterministic and the acceptance probability is at most $1/2 < 1$), so for all witnesses $w$, $M$ rejects. Thus $L \in NP$.

\textbf{Part (c): $\mathrm{co}\text{-}RP = \mathit{PCP}[\mathit{poly}(n), 0]$.}

First, we show $\mathrm{co}\text{-}RP \subseteq \mathit{PCP}[\mathit{poly}(n), 0]$. Let $L \in \mathrm{co}\text{-}RP$, so $\overline{L} \in RP$. There exists a probabilistic polynomial-time algorithm $A$ such that: if $x \in \overline{L}$, then $A$ accepts with probability at least $1/2$; if $x \notin \overline{L}$ (i.e., $x \in L$), then $A$ rejects with probability $1$. We construct a $\mathit{PCP}$ verifier $V$ as follows: on input $x$, the verifier $V$ uses polynomially many random bits to simulate $A(x)$, queries $0$ bits of the proof, and accepts if $A$ rejects. If $x \in L$, then $A(x)$ rejects with probability $1$, so $V$ accepts with probability $1$ (for any proof). If $x \notin L$, then $A(x)$ accepts with probability at least $1/2$, so $V$ rejects with probability at least $1/2$, accepting with probability at most $1/2$ (for all proofs). Thus $L \in \mathit{PCP}[\mathit{poly}(n), 0]$.

Next, we show $\mathit{PCP}[\mathit{poly}(n), 0] \subseteq \mathrm{co}\text{-}RP$. Let $L \in \mathit{PCP}[\mathit{poly}(n), 0]$ with verifier $V$. Since $V$ queries $0$ bits of the proof, its decision is independent of the proof. We construct an $RP$ algorithm for $\overline{L}$: on input $x$, simulate $V(x)$ (with an arbitrary proof), and accept if $V$ rejects. If $x \in \overline{L}$, then for all proofs $\pi$, $V$ accepts with probability at most $1/2$. Since $V$ is independent of the proof, $V$ accepts with probability at most $1/2$, so $V$ rejects with probability at least $1/2$, and we accept with probability at least $1/2$. If $x \in L$, there exists a proof such that $V$ accepts with probability $1$. Since $V$ is independent of the proof, $V$ always accepts with probability $1$, so we reject with probability $1$. Thus $\overline{L} \in RP$, so $L \in \mathrm{co}\text{-}RP$.
\end{proof}

\subsection{Problem 20}

\subsubsection{Problem Statement}
\begin{theorem}[Arora-Safra]
$\mathit{PCP}[\log n, 1] \subseteq \mathit{NP}$.
\end{theorem}

\subsubsection{GPT}
\begin{proof}
We are given the statement that $\mathit{PCP}[\log n, 1] \subseteq \mathit{NP}$, and we aim to prove this inclusion.

Let us start by recalling the definitions of the complexity classes involved:

1. **$\mathit{PCP}[\log n, 1]$** refers to the class of decision problems for which there is a probabilistic polynomial-time verifier with two properties:
   - The verifier queries at most $\log n$ bits of the proof.
   - The verifier accepts the proof with high probability if the input is a "yes" instance, and rejects with high probability if the input is a "no" instance.

2. **$\mathit{NP}$** is the class of decision problems for which there exists a polynomial-time verifier that can decide whether a given candidate solution (proof) is correct.

We are tasked with showing that every problem in $\mathit{PCP}[\log n, 1]$ is also in $\mathit{NP}$.

Let $L \in \mathit{PCP}[\log n, 1]$ be a decision problem. By definition, there exists a probabilistic polynomial-time verifier $V$ for $L$ such that:
- For a "yes" instance $x \in L$, there exists a proof $y$ such that $V(x, y) = \text{accept}$ with probability at least $2/3$.
- For a "no" instance $x \notin L$, for all proofs $y$, $V(x, y) = \text{reject}$ with probability at least $2/3$.

To show that $L \in \mathit{NP}$, we need to demonstrate the existence of a polynomial-time verifier for $L$. 

The key observation here is that the verifier in $\mathit{PCP}[\log n, 1]$ queries at most $\log n$ bits of the proof. This implies that the verifier only needs a constant number of queries to the proof. In particular, the number of queries does not grow with the size of the input, and the verification process can be performed in polynomial time since the verifier itself runs in polynomial time.

Now, construct a polynomial-time verifier $V'$ for $L$ that operates as follows:
- Given an input $x$, the verifier $V'$ receives a proof $y$.
- The verifier $V'$ checks the consistency of the proof by simulating the verification process of the $\mathit{PCP}$ verifier $V$.
- Since $V$ queries at most $\log n$ bits of the proof, the verifier $V'$ will simulate $V$ by making only a constant number of queries to $y$ and running in polynomial time in the size of $x$.

Thus, the verifier $V'$ runs in polynomial time, and the correctness of the verifier follows from the correctness of $V$. Therefore, $L \in \mathit{NP}$.

Since we have shown that any problem in $\mathit{PCP}[\log n, 1]$ can be verified in polynomial time, we conclude that $\mathit{PCP}[\log n, 1] \subseteq \mathit{NP}$.
\end{proof}

\subsubsection{Grok}
\begin{proof}
We first prove that $\mathrm{PCP}(\log n, 1) \subseteq \mathrm{NP}$. Let $L \in \mathrm{PCP}(\log n, 1)$. Then there exists a probabilistic polynomial-time verifier $V$ and polynomials $p, q$ such that for every input $x \in \{0,1\}^n$:
\begin{itemize}
\item If $x \in L$, there exists a proof $\pi \in \{0,1\}^{p(n)}$ such that
  \[
  \Pr_{\rho \in \{0,1\}^{q(n)}} \left[ V(x, \pi, \rho) = 1 \right] \geq \frac{2}{3},
  \]
  where $q(n) = O(\log n)$.
\item If $x \notin L$, then for all $\pi \in \{0,1\}^{p(n)}$,
  \[
  \Pr_{\rho \in \{0,1\}^{q(n)}} \left[ V(x, \pi, \rho) = 1 \right] \leq \frac{1}{3}.
  \]
\end{itemize}
Moreover, $V$ makes at most $1$ query to $\pi$.

To show $L \in \mathrm{NP}$, we construct a nondeterministic polynomial-time verifier $\tilde{V}$. On input $(x, \pi)$, $\tilde{V}$ does the following:
\begin{enumerate}
\item Enumerate all $2^{q(n)} \leq n^{O(1)}$ possible random strings $\rho \in \{0,1\}^{q(n)}$.
\item For each $\rho$, simulate $V(x, \pi, \rho)$. Since $V$ runs in polynomial time and makes only $1$ query to $\pi$, each simulation takes polynomial time in $n$. The position queried by $V$ given $\rho$ is some $j(\rho) \in [p(n)]$, so $\tilde{V}$ retrieves $\pi_{j(\rho)}$ and computes the acceptance predicate of $V$.
\item Count the number $k$ of $\rho$ for which $V(x, \pi, \rho) = 1$.
\item Accept if $k \geq (2/3) \cdot 2^{q(n)}$; otherwise reject.
\end{enumerate}
The total running time of $\tilde{V}$ is $O(2^{q(n)} \cdot \mathrm{poly}(n)) = n^{O(1)}$, since $q(n) = O(\log n)$.

Now, if $x \in L$, there exists $\pi$ such that at least $(2/3) \cdot 2^{q(n)}$ simulations accept, so $\tilde{V}(x, \pi) = 1$. If $x \notin L$, for every $\pi$, at most $(1/3) \cdot 2^{q(n)}$ simulations accept, so $\tilde{V}(x, \pi) = 0$ for all $\pi$. Thus, $L \in \mathrm{NP}$.

For the converse inclusion $\mathrm{NP} \subseteq \mathrm{PCP}(\log n, 1)$, we invoke the PCP Theorem (Arora--Safra~\cite{AS92}, Arora--Lund--Motwani--Szegedy~\cite{ALM+92}), which states that every language in $\mathrm{NP}$ has a probabilistically checkable proof system with $O(\log n)$ random bits and $1$ query, completeness $2/3$, and soundness $1/3$.
\end{proof}

\subsubsection{Gemini}
\begin{proof}
Let $L \subseteq \{0,1\}^*$ be an arbitrary language in $\mathrm{PCP}[r(n), q(n)]$, where $r(n) = O(\log n)$ and $q(n) = O(1)$. By the definition of the class $\mathrm{PCP}$, there exists a probabilistic polynomial-time verifier $V$ that, on input $x$ of length $n$, utilizes a random string $\rho$ of length $r(n)$ and queries a proof oracle $\pi$ at most $q(n)$ times. Let $c$ and $k$ be constants such that $r(n) \le c \log n$ and $q(n) \le k$.

The verifier $V$ satisfies the following properties:
1. (Completeness) If $x \in L$, there exists a proof $\pi$ such that $\Pr_{\rho \in \{0,1\}^{r(n)}}[V^\pi(x, \rho) = 1] = 1$.
2. (Soundness) If $x \notin L$, for every proof $\pi$, $\Pr_{\rho \in \{0,1\}^{r(n)}}[V^\pi(x, \rho) = 1] \le \frac{1}{2}$.

To show that $L \in \mathrm{NP}$, we construct a nondeterministic polynomial-time machine $M$ that decides $L$. The machine $M$ exploits the fact that the number of random strings is polynomial in $n$. Let $\Omega = \{0,1\}^{r(n)}$ denote the set of all possible random strings. The size of this sample space is $|\Omega| = 2^{r(n)} \le 2^{c \log n} = n^c$, which is polynomial in $n$.

For a fixed input $x$ and a fixed random string $\rho \in \Omega$, the verifier $V$ determines a sequence of query positions $Q_\rho = \{i_1, \dots, i_m\}$ where $m \le k$, and a boolean function $f_\rho: \{0,1\}^m \to \{0,1\}$ representing the acceptance predicate. The verifier accepts if and only if $f_\rho(\pi(i_1), \dots, \pi(i_m)) = 1$.

Let $\mathcal{I} = \bigcup_{\rho \in \Omega} Q_\rho$ be the set of all proof indices queried by $V$ across all possible random strings. Since $|Q_\rho| \le k$ for all $\rho$, the size of the union is bounded by $|\mathcal{I}| \le \sum_{\rho \in \Omega} |Q_\rho| \le n^c \cdot k$. Thus, the number of relevant bits in the proof $\pi$ is polynomial in $n$.

The nondeterministic machine $M$ operates as follows on input $x$:
First, $M$ nondeterministically guesses a binary assignment $A: \mathcal{I} \to \{0,1\}$. This assignment $A$ represents a candidate portion of the proof oracle $\pi$ restricted to the indices in $\mathcal{I}$. The size of this witness is $|\mathcal{I}|$, which is $O(n^c)$.

Second, $M$ deterministically verifies the assignment $A$ by simulating $V$ on all possible random strings. $M$ iterates through every $\rho \in \Omega$. For each $\rho$, $M$ computes the query set $Q_\rho$ and the predicate $f_\rho$. $M$ then retrieves the values corresponding to $Q_\rho$ from the guessed assignment $A$ and evaluates $f_\rho$. If $f_\rho$ evaluates to 0 (reject) for any $\rho$, $M$ halts and rejects. If $f_\rho$ evaluates to 1 for all $\rho \in \Omega$, $M$ accepts.

We analyze the running time of $M$. Guessing the assignment $A$ takes time proportional to $|\mathcal{I}| = O(n^c)$. The verification loop runs $|\Omega| = n^c$ times. Inside the loop, simulating $V$ to generate queries and evaluating the predicate takes time polynomial in $n$ (since $V$ is a polynomial-time machine). Therefore, the total running time of $M$ is polynomial in $n$.

Finally, we verify the correctness of $M$. If $x \in L$, there exists a proof $\pi$ such that $V^\pi(x, \rho) = 1$ for all $\rho \in \Omega$. There exists a computation path of $M$ where it guesses the assignment $A$ consistent with $\pi$ on $\mathcal{I}$. On this path, $M$ will verify that the predicate holds for all $\rho$ and accept. Conversely, if $x \notin L$, then for any proof $\pi$ (and consequently any assignment $A$), the probability that $V$ accepts is at most $1/2$. This implies there exists at least one $\rho \in \Omega$ such that $V$ rejects. Therefore, for any guessed assignment $A$, $M$ will encounter a $\rho$ where the predicate evaluates to 0, causing $M$ to reject. Thus, $M$ accepts $x$ if and only if $x \in L$.

We conclude that $L \in \mathrm{NP}$, and therefore $\mathrm{PCP}[\log n, 1] \subseteq \mathrm{NP}$.
\end{proof}

\subsubsection{Claude}
\begin{proof}
We need to show that any language $L$ in $\mathit{PCP}[\log n, 1]$ is also in $\mathit{NP}$. 

Recall that $\mathit{PCP}[\log n, 1]$ denotes the class of languages that have probabilistically checkable proofs where the verifier uses $O(\log n)$ random bits and makes $O(1)$ queries to the proof string.

Let $L \in \mathit{PCP}[\log n, 1]$. Then there exists a probabilistic polynomial-time verifier $V$ such that for any input $x$ of length $n$:
\begin{itemize}
\item[(Completeness)] If $x \in L$, then there exists a proof $\pi$ such that $\Pr[V^{\pi}(x) \text{ accepts}] = 1$.
\item[(Soundness)] If $x \notin L$, then for all proofs $\pi$, we have $\Pr[V^{\pi}(x) \text{ accepts}] \leq 1/2$.
\end{itemize}

Since $V$ uses $O(\log n)$ random bits, there are at most $2^{O(\log n)} = \mathrm{poly}(n)$ possible random strings that $V$ can use. Let us denote these random strings as $r_1, r_2, \ldots, r_m$ where $m = \mathrm{poly}(n)$.

Since $V$ makes $O(1)$ queries to the proof $\pi$, and each query accesses a bit at some position determined by the input $x$ and the random string, the total number of positions in $\pi$ that could possibly be queried across all random strings is at most $m \cdot O(1) = \mathrm{poly}(n)$.

We now construct an NP verifier for $L$. The certificate will consist of the values of $\pi$ at all positions that could potentially be queried by $V$ for any of the $\mathrm{poly}(n)$ possible random strings. Since there are at most $\mathrm{poly}(n)$ such positions, the certificate has polynomial size.

The NP verifier works as follows: Given input $x$ and certificate $c$ (encoding the relevant bits of a purported proof $\pi$), the verifier deterministically simulates $V^{\pi}(x)$ for every possible random string $r_i$. For each simulation, it checks whether $V$ accepts using the bits provided in $c$. The verifier accepts if and only if $V$ accepts for all random strings $r_1, \ldots, r_m$.

This verification can be done in polynomial time since there are polynomially many random strings and each simulation of $V$ runs in polynomial time.

For correctness, observe that if $x \in L$, then by completeness there exists a proof $\pi$ such that $V^{\pi}(x)$ accepts with probability $1$. This means $V^{\pi}(x)$ accepts for every random string. The certificate encoding the relevant bits of this $\pi$ will cause the NP verifier to accept.

Conversely, if $x \notin L$, then by soundness, for any proof $\pi$, the verifier $V^{\pi}(x)$ accepts with probability at most $1/2$. This means that for at least half of the random strings, $V^{\pi}(x)$ rejects. Therefore, no certificate can make the NP verifier accept for all random strings, and the NP verifier will reject.

Thus, $L \in \mathit{NP}$, which completes the proof that $\mathit{PCP}[\log n, 1] \subseteq \mathit{NP}$.
\end{proof}

\subsection{Problem 21}

\subsubsection{Problem Statement}
\begin{lemma}[7.11]
If $F(A) = \text{FALSE}$, then $\Pr[f(a) = 0] = 1/2$.
\end{lemma}

\subsubsection{GPT}

\subsubsection{Grok}

\subsubsection{Gemini}

\subsubsection{Claude}
\begin{proof}
We proceed by direct analysis of the conditional probability. Assume $F(A) = \text{FALSE}$. We need to show that $\Pr[f(a) = 0] = 1/2$.

Since $F(A) = \text{FALSE}$, by the definition of the function $F$, the input $A$ does not satisfy the condition that would make $F(A) = \text{TRUE}$. This means that the evaluation of $f$ at point $a$ is not determined by any constraint imposed by $F(A)$ being true.

Let us consider the probability space over which $f$ is defined. Since $F(A) = \text{FALSE}$, the value $f(a)$ is determined independently of the constraint that $F$ would impose if it were true. Therefore, $f(a)$ takes values according to its unconstrained distribution.

By the properties of the function $f$ and its domain, when no constraint from $F$ applies, the function $f$ at any point $a$ is equally likely to take the value $0$ or $1$ (assuming $f$ is a Boolean-valued function). This follows from the symmetry of the underlying probability distribution when $F(A) = \text{FALSE}$.

More formally, let $\Omega$ denote the sample space and let $\mathcal{E} = \{F(A) = \text{FALSE}\}$ be the event that $F(A)$ evaluates to false. We compute:
\[
\Pr[f(a) = 0 \mid F(A) = \text{FALSE}] = \frac{\Pr[f(a) = 0 \cap F(A) = \text{FALSE}]}{\Pr[F(A) = \text{FALSE}]}.
\]

Under the condition $F(A) = \text{FALSE}$, the function $f$ at point $a$ is uniformly distributed over $\{0, 1\}$. This is because $F(A) = \text{FALSE}$ imposes no bias on the value of $f(a)$, and by the design of $f$, all values in its range are equally probable when unconstrained.

Therefore, we have:
\[
\Pr[f(a) = 0 \mid F(A) = \text{FALSE}] = \frac{1}{2}.
\]

Since we are given that $F(A) = \text{FALSE}$ holds, this conditional probability is exactly the probability we seek. Thus:
\[
\Pr[f(a) = 0] = \frac{1}{2}.
\]
\end{proof}

\subsection{Problem 22}

\subsubsection{Problem Statement}
\begin{theorem}
Let $f: \mathbb{Z}_p^n \to \mathbb{Z}_2$. Then $f$ is linear if and only if for all $\mathbf{a}, \mathbf{b} \in \mathbb{Z}_p^n$, we have $f(\mathbf{a}) + f(\mathbf{b}) = f(\mathbf{a} + \mathbf{b})$.
\end{theorem}

\subsubsection{GPT}
\begin{proof}
We will prove the theorem by proving both directions of the statement.

\textbf{(1) Forward Direction:} Suppose that $f : \mathbb{Z}_2^n \to \mathbb{Z}_2$ is a linear function. We need to show that for all $\mathbf{a}, \mathbf{b} \in \mathbb{Z}_2^n$, we have
\[
f(\mathbf{a}) + f(\mathbf{b}) = f(\mathbf{a} + \mathbf{b}).
\]
Since $f$ is linear, by the definition of linearity, it satisfies the properties of additivity and homogeneity. Specifically, for any $\mathbf{a}, \mathbf{b} \in \mathbb{Z}_2^n$, the additivity property of $f$ implies that
\[
f(\mathbf{a} + \mathbf{b}) = f(\mathbf{a}) + f(\mathbf{b}).
\]
Thus, we have shown that if $f$ is linear, then for all $\mathbf{a}, \mathbf{b} \in \mathbb{Z}_2^n$, the equation
\[
f(\mathbf{a}) + f(\mathbf{b}) = f(\mathbf{a} + \mathbf{b})
\]
holds.

\textbf{(2) Reverse Direction:} Suppose that for all $\mathbf{a}, \mathbf{b} \in \mathbb{Z}_2^n$, we have
\[
f(\mathbf{a}) + f(\mathbf{b}) = f(\mathbf{a} + \mathbf{b}).
\]
We need to show that $f$ is linear, i.e., that $f$ satisfies both additivity and homogeneity.

\textbf{Additivity:} From the hypothesis, for any $\mathbf{a}, \mathbf{b} \in \mathbb{Z}_2^n$, we already have
\[
f(\mathbf{a}) + f(\mathbf{b}) = f(\mathbf{a} + \mathbf{b}),
\]
which directly establishes the additivity property of $f$.

\textbf{Homogeneity:} To prove homogeneity, we need to show that for any $\mathbf{a} \in \mathbb{Z}_2^n$ and scalar $c \in \mathbb{Z}_2$, we have
\[
f(c \mathbf{a}) = c f(\mathbf{a}).
\]
First, let $c = 0$. Then for any $\mathbf{a} \in \mathbb{Z}_2^n$, we have
\[
f(0 \mathbf{a}) = f(\mathbf{0}) = 0,
\]
and
\[
0 f(\mathbf{a}) = 0.
\]
Thus, the homogeneity condition holds for $c = 0$.

Next, let $c = 1$. Then for any $\mathbf{a} \in \mathbb{Z}_2^n$, we have
\[
f(1 \mathbf{a}) = f(\mathbf{a}),
\]
and
\[
1 f(\mathbf{a}) = f(\mathbf{a}).
\]
Thus, the homogeneity condition holds for $c = 1$.

Therefore, we have shown that $f$ satisfies both additivity and homogeneity, and hence $f$ is a linear function.

\end{proof}

\subsubsection{Grok}
\begin{proof}
We prove the two directions separately.

($\Rightarrow$) Suppose $f$ is linear, i.e., $f(k\mathbf{v}) = k f(\mathbf{v})$ for all $k \in \mathbb{Z}$ and $\mathbf{v} \in \mathbb{Z}^2$, and $f(\mathbf{u} + \mathbf{w}) = f(\mathbf{u}) + f(\mathbf{w})$ for all $\mathbf{u}, \mathbf{w} \in \mathbb{Z}^2$. The additivity condition follows immediately from the definition of linearity.

($\Leftarrow$) Suppose $f(\mathbf{a} + \mathbf{b}) = f(\mathbf{a}) + f(\mathbf{b})$ for all $\mathbf{a}, \mathbf{b} \in \mathbb{Z}^2$. We must show that $f(k\mathbf{v}) = k f(\mathbf{v})$ for all $k \in \mathbb{Z}$ and $\mathbf{v} \in \mathbb{Z}^2$.

First, observe that $f(\mathbf{0}) = f(\mathbf{0} + \mathbf{0}) = f(\mathbf{0}) + f(\mathbf{0})$, which implies $f(\mathbf{0}) = 0$. The case $k = 0$ now follows, and the case $k = 1$ is immediate.

Now fix $\mathbf{v} \in \mathbb{Z}^2$ and proceed by induction to show $f(n\mathbf{v}) = n f(\mathbf{v})$ for all $n \in \mathbb{N}$. The base case $n = 1$ holds. Assume the statement holds for some $n \geq 1$. Then
\[
f((n+1)\mathbf{v}) = f(n\mathbf{v} + \mathbf{v}) = f(n\mathbf{v}) + f(\mathbf{v}) = n f(\mathbf{v}) + f(\mathbf{v}) = (n+1) f(\mathbf{v}),
\]
so the statement holds for $n+1$ by induction.

For negative scalars, let $k = -m$ where $m \in \mathbb{N}$. Then $f(\mathbf{v}) + f(-\mathbf{v}) = f(\mathbf{0}) = 0$, so $f(-\mathbf{v}) = -f(\mathbf{v})$. Thus,
\[
f(-m\mathbf{v}) = f(-(m\mathbf{v})) = -f(m\mathbf{v}) = -m f(\mathbf{v}) = k f(\mathbf{v}).
\]
This completes the proof.
\end{proof}

\subsubsection{Gemini}
\begin{proof}
We must prove that a function $f: \mathbb{Z}_2^n \to \mathbb{Z}_2$ is linear if and only if it satisfies the additivity property $f(\mathbf{a} + \mathbf{b}) = f(\mathbf{a}) + f(\mathbf{b})$ for all $\mathbf{a}, \mathbf{b} \in \mathbb{Z}_2^n$.

$(\Rightarrow)$ Suppose $f$ is linear. By the definition of a linear map between vector spaces over a field $F$, $f$ must satisfy two conditions: additivity ($f(\mathbf{u} + \mathbf{v}) = f(\mathbf{u}) + f(\mathbf{v})$) and homogeneity ($f(c\mathbf{u}) = c f(\mathbf{u})$) for all vectors $\mathbf{u}, \mathbf{v}$ and scalars $c$. Therefore, if $f$ is linear, the condition $f(\mathbf{a} + \mathbf{b}) = f(\mathbf{a}) + f(\mathbf{b})$ holds for all $\mathbf{a}, \mathbf{b} \in \mathbb{Z}_2^n$ by definition.

$(\Leftarrow)$ Conversely, assume that $f(\mathbf{a} + \mathbf{b}) = f(\mathbf{a}) + f(\mathbf{b})$ for all $\mathbf{a}, \mathbf{b} \in \mathbb{Z}_2^n$. To prove that $f$ is linear, we must show that it preserves scalar multiplication. That is, we must show that $f(c\mathbf{a}) = c f(\mathbf{a})$ for all $c \in \mathbb{Z}_2$ and $\mathbf{a} \in \mathbb{Z}_2^n$. Since the underlying field is $\mathbb{Z}_2$, the set of scalars is $\{0, 1\}$. We consider the two possible cases for the scalar $c$.

Case 1: Let $c = 1$. Then $f(1 \cdot \mathbf{a}) = f(\mathbf{a})$. Similarly, $1 \cdot f(\mathbf{a}) = f(\mathbf{a})$. Thus, $f(1 \cdot \mathbf{a}) = 1 \cdot f(\mathbf{a})$ holds trivially.

Case 2: Let $c = 0$. We must show that $f(0 \cdot \mathbf{a}) = 0 \cdot f(\mathbf{a})$. Note that $0 \cdot \mathbf{a} = \mathbf{0}$, where $\mathbf{0}$ is the zero vector in $\mathbb{Z}_2^n$, and $0 \cdot f(\mathbf{a}) = 0$ in $\mathbb{Z}_2$. Thus, we must show $f(\mathbf{0}) = 0$. Using the assumed additivity property with $\mathbf{a} = \mathbf{0}$ and $\mathbf{b} = \mathbf{0}$, we have:
\[
f(\mathbf{0} + \mathbf{0}) = f(\mathbf{0}) + f(\mathbf{0}).
\]
Since $\mathbf{0} + \mathbf{0} = \mathbf{0}$, this simplifies to:
\[
f(\mathbf{0}) = f(\mathbf{0}) + f(\mathbf{0}).
\]
Subtracting $f(\mathbf{0})$ from both sides (noting that in $\mathbb{Z}_2$, subtraction is equivalent to addition, but the algebraic cancellation holds in any group), we obtain $0 = f(\mathbf{0})$. Therefore, $f(0 \cdot \mathbf{a}) = f(\mathbf{0}) = 0 = 0 \cdot f(\mathbf{a})$.

Since $f$ satisfies additivity by assumption and homogeneity for all scalars in $\mathbb{Z}_2$, $f$ is a linear map.
\end{proof}

\subsubsection{Claude}
\begin{proof}
We prove both directions of the equivalence.

($\Rightarrow$) Suppose $f: \mathbb{Z}_p^n \to \mathbb{Z}_2$ is linear. By definition, a linear function satisfies $f(\mathbf{a} + \mathbf{b}) = f(\mathbf{a}) + f(\mathbf{b})$ for all $\mathbf{a}, \mathbf{b} \in \mathbb{Z}_p^n$, where the addition on the right side is in $\mathbb{Z}_2$. This is precisely the required condition.

($\Leftarrow$) Suppose $f: \mathbb{Z}_p^n \to \mathbb{Z}_2$ satisfies $f(\mathbf{a}) + f(\mathbf{b}) = f(\mathbf{a} + \mathbf{b})$ for all $\mathbf{a}, \mathbf{b} \in \mathbb{Z}_p^n$. We must show that $f$ is linear, which means we must verify that $f$ is a group homomorphism from $(\mathbb{Z}_p^n, +)$ to $(\mathbb{Z}_2, +)$.

By hypothesis, we have $f(\mathbf{a} + \mathbf{b}) = f(\mathbf{a}) + f(\mathbf{b})$ for all $\mathbf{a}, \mathbf{b} \in \mathbb{Z}_p^n$, where the addition on the left is in $\mathbb{Z}_p^n$ and the addition on the right is in $\mathbb{Z}_2$. This is exactly the homomorphism property.

We verify that $f$ preserves the identity: Setting $\mathbf{a} = \mathbf{b} = \mathbf{0}$, we have
\[
f(\mathbf{0}) = f(\mathbf{0} + \mathbf{0}) = f(\mathbf{0}) + f(\mathbf{0}).
\]
In $\mathbb{Z}_2$, this implies $f(\mathbf{0}) = 2f(\mathbf{0}) = 0$, so $f$ maps the zero vector to the additive identity in $\mathbb{Z}_2$.

For any $\mathbf{a} \in \mathbb{Z}_p^n$, we have $\mathbf{a} + (-\mathbf{a}) = \mathbf{0}$, so
\[
f(\mathbf{0}) = f(\mathbf{a} + (-\mathbf{a})) = f(\mathbf{a}) + f(-\mathbf{a}).
\]
Since $f(\mathbf{0}) = 0$ and we are working in $\mathbb{Z}_2$, we have $f(-\mathbf{a}) = -f(\mathbf{a})$ in $\mathbb{Z}_2$. In $\mathbb{Z}_2$, every element is its own inverse, so this is automatically satisfied.

Therefore, $f$ is a group homomorphism from $(\mathbb{Z}_p^n, +)$ to $(\mathbb{Z}_2, +)$, which is the definition of a linear function from $\mathbb{Z}_p^n$ to $\mathbb{Z}_2$.
\end{proof}

\subsection{Problem 23}

\subsubsection{Problem Statement}
\begin{theorem}
Let $\mathcal{I}$ be a set and $\mathcal{O}$ be a metric space with distance measure $\Delta$ as defined in Definition 7.6.

(a) The distance measure $\Delta$ satisfies the triangle inequality: for all functions $f, g, h : \mathcal{I} \to \mathcal{O}$,
\[
\Delta(f, h) \leq \Delta(f, g) + \Delta(g, h).
\]

(b) For a class of functions $F = \{f : \mathcal{I} \to \mathcal{O}\}$, define $\Delta_{min}(F)$ as the minimum distance between any two functions in $F$. Then for any function $g$ (not necessarily in $F$), there is at most one function from $F$ at distance $\Delta_{min}(F)/2$ or less from $g$.

(c) Suppose that $F$ is the set of all linear functions from $\mathbb{Z}_2^n$ to $\mathbb{Z}_2$. What is $\Delta_{min}(F)$?
\end{theorem}

\subsubsection{GPT}
\begin{proof}
\textbf{(a)} To prove that $\Delta$ satisfies the triangle inequality, we must show that for all functions $f, g, h : \mathcal{I} \to \mathcal{O}$, the following inequality holds:
\[
\Delta(f, h) \leq \Delta(f, g) + \Delta(g, h).
\]
By the definition of the distance measure $\Delta$, we know that for any two functions $f$ and $g$, the distance $\Delta(f, g)$ is a non-negative real number representing the "distance" between the two functions in the space of functions. The triangle inequality is a fundamental property of a distance measure, which intuitively states that the shortest path between two points (in our case, functions) should not exceed the sum of the distances along an intermediate path. 

Thus, by the properties of the distance measure $\Delta$, we can directly apply the triangle inequality for functions to obtain the desired result:
\[
\Delta(f, h) \leq \Delta(f, g) + \Delta(g, h).
\]
This completes the proof for part (a).

\textbf{(b)} Let $F = \{f : \mathcal{I} \to \mathcal{O}\}$ be a class of functions, and define $\Delta_{min}(F)$ as the minimum distance between any two functions in $F$. Specifically, 
\[
\Delta_{min}(F) = \min_{f, g \in F, f \neq g} \Delta(f, g).
\]
We are tasked with proving that for any function $g : \mathcal{I} \to \mathcal{O}$ (not necessarily in $F$), there is at most one function $f \in F$ such that $\Delta(f, g) \leq \frac{\Delta_{min}(F)}{2}$.

Suppose, for the sake of contradiction, that there exist two distinct functions $f_1, f_2 \in F$ such that
\[
\Delta(f_1, g) \leq \frac{\Delta_{min}(F)}{2} \quad \text{and} \quad \Delta(f_2, g) \leq \frac{\Delta_{min}(F)}{2}.
\]
By the triangle inequality, we have:
\[
\Delta(f_1, f_2) \leq \Delta(f_1, g) + \Delta(g, f_2).
\]
Substituting the assumed bounds for $\Delta(f_1, g)$ and $\Delta(f_2, g)$:
\[
\Delta(f_1, f_2) \leq \frac{\Delta_{min}(F)}{2} + \frac{\Delta_{min}(F)}{2} = \Delta_{min}(F).
\]
However, by the definition of $\Delta_{min}(F)$, we know that for any distinct functions $f_1, f_2 \in F$, it holds that
\[
\Delta(f_1, f_2) \geq \Delta_{min}(F).
\]
Therefore, we arrive at the contradiction:
\[
\Delta(f_1, f_2) \leq \Delta_{min}(F) \quad \text{and} \quad \Delta(f_1, f_2) \geq \Delta_{min}(F),
\]
which implies that $\Delta(f_1, f_2) = \Delta_{min}(F)$. This contradiction shows that it is impossible for there to be two distinct functions $f_1$ and $f_2$ in $F$ such that both $\Delta(f_1, g) \leq \frac{\Delta_{min}(F)}{2}$ and $\Delta(f_2, g) \leq \frac{\Delta_{min}(F)}{2}$. Hence, there is at most one function $f \in F$ such that $\Delta(f, g) \leq \frac{\Delta_{min}(F)}{2}$.

\textbf{(c)} Now, suppose that $F$ is the set of all linear functions from $\mathbb{Z}_2^n$ to $\mathbb{Z}_2$. We are tasked with finding the value of $\Delta_{min}(F)$.

The functions in $F$ are the linear functions from $\mathbb{Z}_2^n$ to $\mathbb{Z}_2$, which can be identified with the set of all affine maps of the form $f(x) = A x + b$, where $A$ is an $n \times n$ binary matrix and $b \in \mathbb{Z}_2^n$ is a binary vector.

To determine $\Delta_{min}(F)$, we must compute the minimum distance between any two distinct linear functions in $F$. Given that the distance $\Delta(f, g)$ measures how different two functions are, it follows that the minimum distance between any two distinct linear functions is $1$. This is because the distance between two distinct linear functions in $F$ is the minimum number of inputs at which the two functions differ, and since we are working over the binary field $\mathbb{Z}_2$, the smallest non-zero difference between two distinct linear functions is exactly $1$.

Therefore, we conclude that
\[
\Delta_{min}(F) = 1.
\]
This completes the proof for part (c).

\end{proof}

\subsubsection{Grok}
\begin{proof}
The set $F$ consists of all linear functions $f: \mathbb{R}^2 \to \mathbb{R}$, which can be represented as $f(\mathbf{x}) = \mathbf{a} \cdot \mathbf{x}$ for some $\mathbf{a} \in \mathbb{R}^2$, where $\cdot$ denotes the standard dot product.

Assuming the distance measure $\Delta$ defined in Definition 7.7 is the supremum norm over the unit ball, i.e.,
\[
\Delta(f,g) = \sup_{\|\mathbf{x}\| \leq 1} |f(\mathbf{x}) - g(\mathbf{x})|,
\]
where $\|\cdot\|$ is the Euclidean norm, then for $f(\mathbf{x}) = \mathbf{a} \cdot \mathbf{x}$ and $g(\mathbf{x}) = \mathbf{b} \cdot \mathbf{x}$,
\[
\Delta(f,g) = \sup_{\|\mathbf{x}\| \leq 1} |(\mathbf{a} - \mathbf{b}) \cdot \mathbf{x}| = \|\mathbf{a} - \mathbf{b}\|_2,
\]
by the definition of the dual norm (which coincides with the Euclidean norm on $\mathbb{R}^2$).

Thus, $\Delta_{\min}(F) = \inf \{ \Delta(f,g) \mid f,g \in F, f \neq g \} = \inf \{ \|\mathbf{a} - \mathbf{b}\|_2 \mid \mathbf{a} \neq \mathbf{b} \} = 0$,
since for any $\varepsilon > 0$, there exist distinct $\mathbf{a}, \mathbf{b}$ with $\|\mathbf{a} - \mathbf{b}\|_2 < \varepsilon$ (e.g., $\mathbf{b} = \mathbf{a} + \varepsilon \mathbf{e}_1 / 2$, where $\mathbf{e}_1 = (1,0)$).

The infimum is $0$ but not attained, as $\Delta(f,g) > 0$ for $f \neq g$.
\end{proof}

\subsubsection{Gemini}
\begin{proof}
\textbf{(a)} Let the distance measure $\Delta(f, g)$ be defined as the normalized Hamming distance (the probability that the functions disagree on a uniformly random input):
\[
\Delta(f, g) = \frac{|\{x \in \mathcal{I} : f(x) \neq g(x)\}|}{|\mathcal{I}|}.
\]
Consider any three functions $f, g, h: \mathcal{I} \to \mathcal{O}$. For any input $x \in \mathcal{I}$, if $f(x) \neq h(x)$, then it must be the case that $f(x) \neq g(x)$ or $g(x) \neq h(x)$. If $f(x) = g(x)$ and $g(x) = h(x)$, then $f(x) = h(x)$. This implies the following inclusion of sets:
\[
\{x : f(x) \neq h(x)\} \subseteq \{x : f(x) \neq g(x)\} \cup \{x : g(x) \neq h(x)\}.
\]
By the subadditivity of the size of sets ($|A \cup B| \leq |A| + |B|$), we have:
\[
|\{x : f(x) \neq h(x)\}| \leq |\{x : f(x) \neq g(x)\}| + |\{x : g(x) \neq h(x)\}|.
\]
Dividing both sides by $|\mathcal{I}|$, we obtain the triangle inequality:
\[
\Delta(f, h) \leq \Delta(f, g) + \Delta(g, h).
\]

\textbf{(b)} We proceed by contradiction. Suppose that there exist two distinct functions $f_1, f_2 \in F$ (so $f_1 \neq f_2$) such that both are at distance strictly less than $\Delta_{\min}(F)/2$ from $g$. That is:
\[
\Delta(f_1, g) < \frac{\Delta_{\min}(F)}{2} \quad \text{and} \quad \Delta(f_2, g) < \frac{\Delta_{\min}(F)}{2}.
\]
By the triangle inequality established in part (a):
\[
\Delta(f_1, f_2) \leq \Delta(f_1, g) + \Delta(g, f_2).
\]
Using the symmetry of the distance measure, $\Delta(g, f_2) = \Delta(f_2, g)$. Substituting the assumed upper bounds:
\[
\Delta(f_1, f_2) < \frac{\Delta_{\min}(F)}{2} + \frac{\Delta_{\min}(F)}{2} = \Delta_{\min}(F).
\]
However, by the definition of $\Delta_{\min}(F)$, the distance between any two distinct functions in $F$ must be at least $\Delta_{\min}(F)$. Thus, we must have $\Delta(f_1, f_2) \geq \Delta_{\min}(F)$. The inequality $\Delta(f_1, f_2) < \Delta_{\min}(F)$ derived above is a contradiction. Therefore, there is at most one function in $F$ at distance strictly less than $\Delta_{\min}(F)/2$ from $g$.

\textbf{(c)} Let $F$ be the set of all linear functions from $\mathbb{Z}_2^n$ to $\mathbb{Z}_2$. Each function $f_u \in F$ is indexed by a vector $u \in \mathbb{Z}_2^n$ and defined by $f_u(x) = \langle u, x \rangle = \sum_{i=1}^n u_i x_i \pmod 2$.
The distance between two distinct functions $f_u, f_v \in F$ (where $u \neq v$) is given by:
\[
\Delta(f_u, f_v) = \Pr_{x \in \mathbb{Z}_2^n} [f_u(x) \neq f_v(x)] = \Pr_{x \in \mathbb{Z}_2^n} [\langle u, x \rangle \neq \langle v, x \rangle].
\]
By linearity, $\langle u, x \rangle - \langle v, x \rangle = \langle u - v, x \rangle$. Let $w = u - v$. Since $u \neq v$, $w$ is a non-zero vector in $\mathbb{Z}_2^n$. The condition $f_u(x) \neq f_v(x)$ is equivalent to $\langle w, x \rangle = 1$ (in $\mathbb{Z}_2$).
Since $w \neq 0$, there exists some index $j$ such that $w_j = 1$. The equation $\langle w, x \rangle = 1$ can be rewritten as:
\[
x_j = 1 - \sum_{i \neq j} w_i x_i \pmod 2.
\]
For every possible assignment of the $n-1$ variables $\{x_i\}_{i \neq j}$, there is exactly one value for $x_j$ that satisfies the equation. There are $2^{n-1}$ such assignments. Thus, the number of inputs $x$ for which the functions disagree is exactly $2^{n-1}$.
The total number of inputs is $2^n$. Therefore, the distance is:
\[
\Delta(f_u, f_v) = \frac{2^{n-1}}{2^n} = \frac{1}{2}.
\]
Since the distance between any pair of distinct linear functions is exactly $1/2$, the minimum distance is:
\[
\Delta_{\min}(F) = \frac{1}{2}.
\]
\end{proof}

\subsubsection{Claude}
\begin{proof}
We prove each part in turn.

\textbf{Part (a):} We must show that $\Delta(f, h) \leq \Delta(f, g) + \Delta(g, h)$ for all functions $f, g, h : \mathcal{I} \to \mathcal{O}$.

By Definition 7.6, the distance measure $\Delta$ is defined as
\[
\Delta(f, h) = \sup_{i \in \mathcal{I}} d(f(i), h(i)),
\]
where $d$ is the metric on $\mathcal{O}$.

For any $i \in \mathcal{I}$, since $d$ is a metric on $\mathcal{O}$, it satisfies the triangle inequality:
\[
d(f(i), h(i)) \leq d(f(i), g(i)) + d(g(i), h(i)).
\]

Since this holds for all $i \in \mathcal{I}$, we have
\[
d(f(i), h(i)) \leq d(f(i), g(i)) + d(g(i), h(i)) \leq \sup_{j \in \mathcal{I}} d(f(j), g(j)) + \sup_{k \in \mathcal{I}} d(g(k), h(k)) = \Delta(f, g) + \Delta(g, h).
\]

Taking the supremum over all $i \in \mathcal{I}$ on the left side, we obtain
\[
\Delta(f, h) = \sup_{i \in \mathcal{I}} d(f(i), h(i)) \leq \Delta(f, g) + \Delta(g, h).
\]

\textbf{Part (b):} We prove that for any function $g : \mathcal{I} \to \mathcal{O}$ (not necessarily in $F$), there is at most one function from $F$ at distance $\Delta_{min}(F)/2$ or less from $g$.

Suppose, for the sake of contradiction, that there exist two distinct functions $f_1, f_2 \in F$ such that $\Delta(f_1, g) \leq \Delta_{min}(F)/2$ and $\Delta(f_2, g) \leq \Delta_{min}(F)/2$.

By the triangle inequality from part (a), we have
\[
\Delta(f_1, f_2) \leq \Delta(f_1, g) + \Delta(g, f_2) = \Delta(f_1, g) + \Delta(f_2, g) \leq \frac{\Delta_{min}(F)}{2} + \frac{\Delta_{min}(F)}{2} = \Delta_{min}(F).
\]

Since $f_1$ and $f_2$ are distinct functions in $F$, by definition of $\Delta_{min}(F)$ as the minimum distance between any two functions in $F$, we must have $\Delta(f_1, f_2) \geq \Delta_{min}(F)$.

Combining these inequalities, we obtain $\Delta(f_1, f_2) = \Delta_{min}(F)$.

However, the inequality $\Delta(f_1, f_2) \leq \Delta(f_1, g) + \Delta(f_2, g)$ becomes an equality only under specific conditions. For this equality to hold with both $\Delta(f_1, g) \leq \Delta_{min}(F)/2$ and $\Delta(f_2, g) \leq \Delta_{min}(F)/2$, we would need the supremum to be achieved at the same point(s) for both distances with perfect alignment, and both distances would need to equal exactly $\Delta_{min}(F)/2$. 

More rigorously, if $\Delta(f_1, f_2) = \Delta_{min}(F)$ and both $\Delta(f_1, g) \leq \Delta_{min}(F)/2$ and $\Delta(f_2, g) \leq \Delta_{min}(F)/2$, then the triangle inequality forces $\Delta(f_1, g) + \Delta(f_2, g) \geq \Delta_{min}(F)$. This requires both inequalities to be equalities: $\Delta(f_1, g) = \Delta(f_2, g) = \Delta_{min}(F)/2$. But even in this case, the geometric constraints of the metric space structure prevent two distinct points from being equidistant at exactly half the minimum separation unless they violate the minimality condition. Thus, we have a contradiction, and there can be at most one function from $F$ at distance $\Delta_{min}(F)/2$ or less from $g$.

\textbf{Part (c):} Let $F$ be the set of all linear functions from $\mathbb{Z}_2^n$ to $\mathbb{Z}_2$.

Any linear function $f : \mathbb{Z}_2^n \to \mathbb{Z}_2$ can be written as $f(x) = a_1 x_1 + a_2 x_2 + \cdots + a_n x_n \pmod{2}$ for some coefficients $a_1, \ldots, a_n \in \mathbb{Z}_2$.

For two distinct linear functions $f$ and $g$ with coefficient vectors $(a_1, \ldots, a_n)$ and $(b_1, \ldots, b_n)$ respectively, their difference $f - g$ is also a non-zero linear function with coefficient vector $(a_1 - b_1, \ldots, a_n - b_n) \pmod{2}$.

Let $h(x) = (f - g)(x) = c_1 x_1 + \cdots + c_n x_n \pmod{2}$ where not all $c_i$ are zero. The distance $\Delta(f, g)$ equals the number of inputs $x \in \mathbb{Z}_2^n$ for which $h(x) = 1$ divided by $|\mathbb{Z}_2^n| = 2^n$, assuming the standard normalized Hamming distance.

For a non-zero linear function $h : \mathbb{Z}_2^n \to \mathbb{Z}_2$, exactly half of all inputs map to $1$ and half map to $0$. This is because the kernel of $h$ is a subspace of codimension $1$, containing exactly $2^{n-1}$ elements.

Therefore, $\Delta(f, g) = \frac{2^{n-1}}{2^n} = \frac{1}{2}$ for any two distinct linear functions $f, g \in F$.

Thus, $\Delta_{min}(F) = \frac{1}{2}$.
\end{proof}

\subsection{Problem 24}

\subsubsection{Problem Statement}
\begin{lemma}[7.12]
Let $\hat{G}$ be a given function and let $\delta > 0$. Define the function $G$ such that for each $x$,
\[
G(x) = \mathrm{majority}_y[\hat{G}(x+y) - \hat{G}(y)],
\]
where the ``majority'' denotes the value occurring most often over all choices of $y$, breaking ties arbitrarily.

Then:

(a) For all $x$, and for $y$ chosen uniformly at random,
\[
\Pr[G(x) = \hat{G}(x+y) - \hat{G}(y)] \geq 1 - \delta.
\]

(b) The functions $G$ and $\hat{G}$ are $\delta$-close.

(c) $G$ is a linear function.

(d) $G$ is uniquely defined.
\end{lemma}

\subsubsection{GPT}
\begin{proof}
We will prove each part of the lemma sequentially.

\textbf{(a) Proof of the first statement:}

For each $x$, we are given that the function $G(x)$ is defined as the majority over the values of $\hat{G}(x+y) - \hat{G}(y)$ for each possible $y$. We are interested in the probability that for a uniformly chosen $y$, $G(x)$ equals $\hat{G}(x+y) - \hat{G}(y)$. 

The function $G(x)$ is defined as the majority value among $\hat{G}(x+y) - \hat{G}(y)$ for all $y$. Therefore, the probability that $G(x)$ equals $\hat{G}(x+y) - \hat{G}(y)$ is at least $1 - \delta$, since $G(x)$ takes the majority value and the probability that a randomly chosen value $y$ differs from the majority is at most $\delta$. Hence, we have:

\[
\Pr[G(x) = \hat{G}(x+y) - \hat{G}(y)] \geq 1 - \delta.
\]

\textbf{(b) Proof of the second statement:}

We need to show that the functions $G$ and $\hat{G}$ are $\delta$-close. For $G(x)$ to be $\delta$-close to $\hat{G}$, we must show that for all $x$:

\[
|G(x) - \hat{G}(x)| \leq \delta.
\]

By definition, $G(x)$ is the majority of the values $\hat{G}(x+y) - \hat{G}(y)$ over all choices of $y$. Since $G(x)$ is the majority, the values of $\hat{G}(x+y) - \hat{G}(y)$ are close to $G(x)$ with high probability. More precisely, the deviation of $G(x)$ from $\hat{G}(x+y) - \hat{G}(y)$ for a random $y$ is at most $\delta$, so:

\[
|G(x) - \hat{G}(x)| \leq \delta.
\]

Thus, $G$ and $\hat{G}$ are $\delta$-close.

\textbf{(c) Proof of the third statement:}

We now prove that $G(x)$ is a linear function. The function $G(x)$ is defined as the majority of the differences $\hat{G}(x+y) - \hat{G}(y)$. Since $\hat{G}(x+y) - \hat{G}(y)$ depends only on $x$ and $y$, and $G(x)$ is defined as a majority of these differences, the linearity of $G(x)$ follows naturally from the linear structure of the difference operation. Specifically, for any two values $x_1$ and $x_2$:

\[
G(x_1 + x_2) = \text{majority}_y[\hat{G}(x_1 + x_2 + y) - \hat{G}(y)].
\]

By the properties of the majority operation and the linearity of the differences $\hat{G}(x+y) - \hat{G}(y)$, it follows that $G(x)$ is a linear function of $x$. 

\textbf{(d) Proof of the fourth statement:}

Finally, we prove that $G(x)$ is uniquely defined. Since $G(x)$ is defined as the majority value of $\hat{G}(x+y) - \hat{G}(y)$ for all $y$, and the majority value is determined by a well-defined process, the value of $G(x)$ is uniquely determined by the given definition. Any tie in the majority operation is resolved arbitrarily, but this does not affect the uniqueness of $G(x)$ since the majority function guarantees that only one value is chosen in the case of no tie. Therefore, $G(x)$ is uniquely defined.

\end{proof}

\subsubsection{Grok}

\subsubsection{Gemini}
\begin{proof}
We assume the standard hypothesis for linearity testing: let $\epsilon = \Pr_{x,y}[\hat{G}(x+y) \neq \hat{G}(x) + \hat{G}(y)]$. We establish the theorem with $\delta = 2\epsilon$, assuming $\epsilon < 1/6$.

\textbf{Proof of (a).} Fix an arbitrary $x$. Let $y$ and $z$ be chosen independently and uniformly at random from the domain. We analyze the probability that the ``votes'' provided by $y$ and $z$ for the value of $G(x)$ agree. Define the random variable $V_y = \hat{G}(x+y) - \hat{G}(y)$. We consider the collision probability $\Pr_{y,z}[V_y = V_z]$.
\[
\Pr_{y,z}[V_y = V_z] = \Pr_{y,z}[\hat{G}(x+y) - \hat{G}(y) = \hat{G}(x+z) - \hat{G}(z)].
\]
Rearranging the terms inside the probability, this equality holds if and only if $\hat{G}(x+y) + \hat{G}(z) = \hat{G}(x+z) + \hat{G}(y)$. To bound the probability that this fails, we add and subtract $\hat{G}(x+y+z)$:
\[
\hat{G}(x+y) + \hat{G}(z) \stackrel{?}{=} \hat{G}(x+y+z) \quad \text{and} \quad \hat{G}(x+z) + \hat{G}(y) \stackrel{?}{=} \hat{G}(x+y+z).
\]
By the definition of $\epsilon$, $\Pr_{y,z}[\hat{G}(x+y) + \hat{G}(z) \neq \hat{G}(x+y+z)] = \epsilon$ and $\Pr_{y,z}[\hat{G}(x+z) + \hat{G}(y) \neq \hat{G}(x+y+z)] = \epsilon$. By the union bound, the probability that either equality fails is at most $2\epsilon$. Therefore,
\[
\Pr_{y,z}[V_y = V_z] \geq 1 - 2\epsilon.
\]
Let $p_v = \Pr_y[V_y = v]$. Then $\Pr_{y,z}[V_y = V_z] = \sum_v p_v^2$. We have $\sum_v p_v^2 \geq 1 - 2\epsilon$. Let $v^*$ be the value with the maximum probability (the majority value). Then $\max_v p_v \geq \sum_v p_v^2 \geq 1 - 2\epsilon$.
By definition, $G(x)$ is assigned the majority value $v^*$. Thus, for any $x$,
\[
\Pr_y[G(x) = \hat{G}(x+y) - \hat{G}(y)] = \max_v p_v \geq 1 - 2\epsilon = 1 - \delta.
\]

\textbf{Proof of (b).} We wish to show that $\Pr_x[G(x) = \hat{G}(x)] \geq 1 - \delta$. Let $E_x$ be the event (over random $y$) that $\hat{G}(x+y) - \hat{G}(y) \neq \hat{G}(x)$, which is equivalent to $\hat{G}(x+y) \neq \hat{G}(x) + \hat{G}(y)$. By the definition of $\epsilon$, $\mathbb{E}_x[\Pr_y[E_x]] = \Pr_{x,y}[E_x] = \epsilon$.
By Markov's inequality,
\[
\Pr_x \left[ \Pr_y[E_x] \geq \frac{1}{2} \right] \leq \frac{\mathbb{E}_x[\Pr_y[E_x]]}{1/2} = 2\epsilon = \delta.
\]
Let $S = \{x \mid \Pr_y[\hat{G}(x+y) - \hat{G}(y) = \hat{G}(x)] > 1/2\}$. For any $x \in S$, the value $\hat{G}(x)$ occurs with probability strictly greater than $1/2$. Since $G(x)$ is defined as the majority value, and the majority weight is unique when it exceeds $1/2$, it must be that $G(x) = \hat{G}(x)$ for all $x \in S$.
The measure of the complement $S^c$ is at most $\delta$. Therefore, $\Pr_x[G(x) = \hat{G}(x)] \geq 1 - \delta$.

\textbf{Proof of (c).} We show that $G$ satisfies the linearity condition $G(x) + G(y) = G(x+y)$ for all $x, y$. Fix arbitrary $x, y$. We utilize the method of auxiliary variables. Let $z$ be chosen uniformly at random. Based on part (a), we have the following high-probability events:
1. $G(x) = \hat{G}(x+z) - \hat{G}(z)$ with probability at least $1 - \delta$.
2. $G(y) = \hat{G}((x+y)+z) - \hat{G}(x+z)$ with probability at least $1 - \delta$. (Here we applied part (a) to $y$ with the random shift $x+z$, which is uniformly distributed).
3. $G(x+y) = \hat{G}(x+y+z) - \hat{G}(z)$ with probability at least $1 - \delta$.
By the union bound, the probability that all three equalities hold simultaneously for a random $z$ is at least $1 - 3\delta$. Assuming $\delta < 1/3$, there exists at least one $z$ for which all three hold. Summing the expressions for $G(x)$ and $G(y)$:
\[
G(x) + G(y) = (\hat{G}(x+z) - \hat{G}(z)) + (\hat{G}(x+y+z) - \hat{G}(x+z)) = \hat{G}(x+y+z) - \hat{G}(z).
\]
The right-hand side is exactly the expression for $G(x+y)$. Thus, $G(x) + G(y) = G(x+y)$. Since this holds for any $x, y$, $G$ is linear.

\textbf{Proof of (d).} Suppose there exists another linear function $G'$ such that $G'$ and $\hat{G}$ are $\delta$-close. By the triangle inequality, the distance between $G$ and $G'$ is at most $\mathrm{dist}(G, \hat{G}) + \mathrm{dist}(\hat{G}, G') \leq \delta + \delta = 2\delta$.
Since $G$ and $G'$ are both linear, their difference $D(x) = G(x) - G'(x)$ is also a linear function. A non-zero linear function takes the value 0 on exactly $1/|\mathrm{Image}|$ fraction of the domain (or at most $1/2$ for groups of order 2). Thus, if $G \neq G'$, they must differ on at least $1/2$ of the inputs.
However, we established they differ on at most $2\delta$ fraction of inputs. Assuming $\delta < 1/4$, we have $2\delta < 1/2$, which implies $G$ and $G'$ cannot be distinct. Thus, $G$ is uniquely defined.
\end{proof}

\subsubsection{Claude}
\begin{proof}
We prove each part of the lemma in turn.

\textbf{Part (a):} We must show that for all $x$, and for $y$ chosen uniformly at random, $\Pr[G(x) = \hat{G}(x+y) - \hat{G}(y)] \geq 1 - \delta$.

By definition, $G(x) = \mathrm{majority}_y[\hat{G}(x+y) - \hat{G}(y)]$ is the value that occurs most frequently over all choices of $y$. Let $v^* = G(x)$ denote this majority value for a fixed $x$.

Since $v^*$ is the value occurring most often, the set $S = \{y : \hat{G}(x+y) - \hat{G}(y) = v^*\}$ must satisfy $|S| > |T|$ for any set $T = \{y : \hat{G}(x+y) - \hat{G}(y) = v\}$ where $v \neq v^*$. 

In particular, if $v^*$ is the majority value, then the probability that a uniformly random $y$ yields $\hat{G}(x+y) - \hat{G}(y) = v^*$ must be at least $1/2$. However, the hypothesis of the lemma implicitly assumes (given the conclusion we must prove) that this probability is at least $1 - \delta$. 

More precisely, the statement assumes that for each $x$, the distribution of values $\hat{G}(x+y) - \hat{G}(y)$ over uniformly random $y$ is such that some value occurs with probability at least $1 - \delta$, and this is precisely the majority value $G(x)$. Therefore, $\Pr_y[G(x) = \hat{G}(x+y) - \hat{G}(y)] \geq 1 - \delta$.

\textbf{Part (b):} We show that $G$ and $\hat{G}$ are $\delta$-close, meaning $\Pr_x[G(x) = \hat{G}(x)] \geq 1 - \delta$.

Setting $y = 0$ in part (a), we have for all $x$:
\[
\Pr_y[G(x) = \hat{G}(x+y) - \hat{G}(y)] \geq 1 - \delta.
\]

When $y = 0$, we obtain $G(x) = \hat{G}(x) - \hat{G}(0)$. Let $c = \hat{G}(0)$ be a constant. Then $G(x) = \hat{G}(x) - c$ for all $x$ where the majority value at $x$ equals $\hat{G}(x) - \hat{G}(0)$.

Since part (a) holds for all $x$, and by averaging over $x$ chosen uniformly at random, we have:
\[
\Pr_{x,y}[G(x) = \hat{G}(x+y) - \hat{G}(y)] \geq 1 - \delta.
\]

By a symmetry argument (replacing $x$ with $x-y$), this is equivalent to $\Pr_x[G(x) = \hat{G}(x) - c] \geq 1 - \delta$ for some constant $c$. Thus $G$ and $\hat{G}$ differ by at most a constant on a set of measure at least $1 - \delta$, establishing that they are $\delta$-close.

\textbf{Part (c):} We prove that $G$ is a linear function, i.e., $G(x + z) = G(x) + G(z)$ for all $x, z$.

From part (a), for any $x$ and $z$, we have with high probability (at least $1 - \delta$ over random $y$):
\[
G(x) = \hat{G}(x+y) - \hat{G}(y), \quad G(z) = \hat{G}(z+y) - \hat{G}(y), \quad G(x+z) = \hat{G}(x+z+y) - \hat{G}(y).
\]

Adding the first two equations:
\[
G(x) + G(z) = \hat{G}(x+y) - \hat{G}(y) + \hat{G}(z+y) - \hat{G}(y) = \hat{G}(x+y) + \hat{G}(z+y) - 2\hat{G}(y).
\]

From the third equation: $G(x+z) = \hat{G}(x+z+y) - \hat{G}(y)$.

By the union bound, all three equations hold simultaneously with probability at least $1 - 3\delta$ over choice of $y$. When they all hold, and by the functional equation property implied by the majority operation, we must have $G(x+z) = G(x) + G(z)$.

Since this holds with probability approaching 1 (for sufficiently small $\delta$), and $G$ is deterministic, we conclude $G(x+z) = G(x) + G(z)$ for all $x, z$. Therefore, $G$ is a linear function.

\textbf{Part (d):} We show that $G$ is uniquely defined.

By definition, $G(x)$ is the value that occurs most frequently in the multiset $\{\hat{G}(x+y) - \hat{G}(y) : y \in \mathrm{domain}\}$. The majority value is unique (up to tie-breaking) if one value occurs strictly more than half the time, which is guaranteed by the assumption that part (a) holds with probability at least $1 - \delta$ where $\delta < 1/2$. 

With $\delta < 1/2$, the majority value occurs with probability at least $1 - \delta > 1/2$, making it the unique majority. Therefore, $G(x)$ is uniquely determined for each $x$ (up to arbitrary but consistent tie-breaking rules).
\end{proof}

\subsection{Problem 25}

\subsubsection{Problem Statement}
\begin{lemma}[7.13]
Let $r, s \in \mathbb{Z}_n^*$ be chosen independently and uniformly at random. Suppose that $b \neq a \circ a$, then
\[
\Pr[r^{-1}(a \circ s) s \neq r^{-1}bs] \geq \frac{1}{4}.
\]
\end{lemma}

\subsubsection{GPT}

\subsubsection{Grok}

\subsubsection{Gemini}

\subsubsection{Claude}
\begin{proof}
We prove the contrapositive. Suppose that $\Pr[r^{-1}(a \circ s) s = r^{-1}bs] > \frac{3}{4}$. We will show that $b = a \circ a$.

The equation $r^{-1}(a \circ s) s = r^{-1}bs$ holds if and only if $(a \circ s) s = bs$, which is equivalent to $a \circ s = b$.

For uniformly random $s \in \mathbb{Z}_n^*$, let us compute $\Pr_s[a \circ s = b]$ for fixed $a, b \in \mathbb{Z}_n^*$.

Since $r$ and $s$ are chosen independently and uniformly at random from $\mathbb{Z}_n^*$, and the event $r^{-1}(a \circ s) s = r^{-1}bs$ is independent of $r$ (as multiplying both sides by $r$ shows it depends only on whether $a \circ s = b$), we have:
\[
\Pr[r^{-1}(a \circ s) s = r^{-1}bs] = \Pr_s[a \circ s = b].
\]

By our assumption, $\Pr_s[a \circ s = b] > \frac{3}{4}$.

Now, for any fixed $a, b \in \mathbb{Z}_n^*$, the map $s \mapsto a \circ s$ is a bijection from $\mathbb{Z}_n^*$ to $\mathbb{Z}_n^*$ (assuming $\circ$ is a group operation on $\mathbb{Z}_n^*$). Therefore, for any fixed value $b$, there is exactly one value of $s$ such that $a \circ s = b$.

This implies $\Pr_s[a \circ s = b] = \frac{1}{|\mathbb{Z}_n^*|}$.

However, if $\Pr_s[a \circ s = b] > \frac{3}{4}$, and if $|\mathbb{Z}_n^*| \geq 2$, we would have $\frac{1}{|\mathbb{Z}_n^*|} \leq \frac{1}{2} < \frac{3}{4}$, which is a contradiction.

The resolution is that the event must have higher probability due to the structure of the problem. Since $a \circ s$ takes each value in $\mathbb{Z}_n^*$ exactly once as $s$ varies over $\mathbb{Z}_n^*$, and we need the probability to exceed $\frac{3}{4}$, we must reconsider the problem.

Actually, the key observation is that if $b = a \circ a$, then $a \circ s = b$ has solution $s = a$, and the event $r^{-1}(a \circ s)s = r^{-1}bs$ becomes $r^{-1}(a \circ a) \cdot a = r^{-1}(a \circ a) \cdot s$ for $s = a$, which always holds.

More carefully: if $b \neq a \circ a$, then for uniformly random $s$, the value $a \circ s$ equals $b$ with probability $\frac{1}{|\mathbb{Z}_n^*|}$. For $|\mathbb{Z}_n^*| \geq 4$, this probability is at most $\frac{1}{4}$.

Therefore, $\Pr[r^{-1}(a \circ s) s = r^{-1}bs] = \Pr[a \circ s = b] \leq \frac{1}{4}$ when $b \neq a \circ a$ and $|\mathbb{Z}_n^*| \geq 4$.

Thus, $\Pr[r^{-1}(a \circ s) s \neq r^{-1}bs] \geq 1 - \frac{1}{4} = \frac{3}{4}$.

For the stated bound of $\frac{1}{4}$, we have that when $b \neq a \circ a$, the inequality holds with probability at least $\frac{3}{4} \geq \frac{1}{4}$.
\end{proof}

\subsection{Problem 26}

\subsubsection{Problem Statement}
\begin{lemma}[7.13]
Let $r, s \in \mathbb{Z}_q^*$ be chosen independently and uniformly at random. Suppose that $b \neq a \circ a$, then
\[
\Pr[r^T(a \circ a) s \neq r^T bs] \geq \frac{1}{4}.
\]
\end{lemma}

\subsubsection{GPT}

\subsubsection{Grok}

\subsubsection{Gemini}

\subsubsection{Claude}
\begin{proof}
We need to show that if $b \neq a \circ a$, then $\Pr[r^T(a \circ a) s \neq r^T bs] \geq \frac{1}{4}$ where $r, s \in \mathbb{Z}_q^*$ are chosen independently and uniformly at random.

Since $b \neq a \circ a$, there exists at least one index $(i,j)$ such that $(a \circ a)_{ij} \neq b_{ij}$. Fix such an index $(i_0, j_0)$ and denote $\delta = (a \circ a)_{i_0j_0} - b_{i_0j_0} \neq 0$ in $\mathbb{Z}_q$.

We can write
\[
r^T(a \circ a)s - r^Tbs = r^T((a \circ a) - b)s = \sum_{i,j} r_i((a \circ a)_{ij} - b_{ij})s_j.
\]

Isolating the term corresponding to $(i_0, j_0)$, we have
\[
r^T(a \circ a)s - r^Tbs = r_{i_0}\delta s_{j_0} + \sum_{(i,j) \neq (i_0,j_0)} r_i((a \circ a)_{ij} - b_{ij})s_j.
\]

Let $C = \sum_{(i,j) \neq (i_0,j_0)} r_i((a \circ a)_{ij} - b_{ij})s_j$. We want to find $\Pr[r_{i_0}\delta s_{j_0} + C \neq 0]$.

We condition on all random variables except $r_{i_0}$ and $s_{j_0}$. Fix all values $r_i$ for $i \neq i_0$ and all values $s_j$ for $j \neq j_0$. This determines the value of $C$.

Now we have $r_{i_0}\delta s_{j_0} + C = 0$ if and only if $r_{i_0} s_{j_0} = -C/\delta$, where the division is in $\mathbb{Z}_q$ (valid since $\delta \neq 0$ and $q$ is prime, so $\mathbb{Z}_q$ is a field).

Since $r_{i_0}$ and $s_{j_0}$ are chosen independently and uniformly at random from $\mathbb{Z}_q^* = \{1, 2, \ldots, q-1\}$, the product $r_{i_0} s_{j_0}$ takes each value in $\mathbb{Z}_q^*$ with equal probability. 

To see this, for any fixed $s_{j_0} \in \mathbb{Z}_q^*$, as $r_{i_0}$ ranges over $\mathbb{Z}_q^*$, the product $r_{i_0} s_{j_0}$ also ranges over all of $\mathbb{Z}_q^*$ (since multiplication by $s_{j_0}$ is a bijection on $\mathbb{Z}_q^*$). Thus $r_{i_0} s_{j_0}$ is uniformly distributed over $\mathbb{Z}_q^*$.

Now we consider two cases:

Case 1: If $-C/\delta \in \mathbb{Z}_q^*$, then $\Pr[r_{i_0} s_{j_0} = -C/\delta \mid C] = \frac{1}{q-1}$, so 
\[
\Pr[r_{i_0}\delta s_{j_0} + C \neq 0 \mid C] = 1 - \frac{1}{q-1} = \frac{q-2}{q-1}.
\]

Case 2: If $-C/\delta = 0$ (i.e., $C = 0$), then $r_{i_0} s_{j_0} \neq 0$ always holds since $r_{i_0}, s_{j_0} \in \mathbb{Z}_q^*$, so 
\[
\Pr[r_{i_0}\delta s_{j_0} + C \neq 0 \mid C] = 1.
\]

In both cases, $\Pr[r_{i_0}\delta s_{j_0} + C \neq 0 \mid C] \geq \frac{q-2}{q-1}$.

Therefore,
\[
\Pr[r^T(a \circ a)s \neq r^Tbs] = \mathbb{E}_C[\Pr[r_{i_0}\delta s_{j_0} + C \neq 0 \mid C]] \geq \frac{q-2}{q-1}.
\]

For $q \geq 3$ (which holds for any cryptographic setting), we have $\frac{q-2}{q-1} = 1 - \frac{1}{q-1} \geq 1 - \frac{1}{2} = \frac{1}{2} > \frac{1}{4}$.

Thus $\Pr[r^T(a \circ a)s \neq r^Tbs] \geq \frac{1}{4}$ as required.
\end{proof}

\section{Problems in Chapter 8}\label{app:problems_chapter8}

\subsection{Problem 1}

\subsubsection{Problem Statement}
\begin{lemma}[8.4]
For all $p, t \geq 0$, $D_t^i = H_p + H_t - H_{p+t}$.
\end{lemma}

\subsubsection{GPT}
\begin{proof}
We are given the lemma to prove:

\[
D_t^t = H_p + H_t - H_{p+t}
\]
for all \( p, t \geq 0 \), where \( D_t^t \), \( H_p \), \( H_t \), and \( H_{p+t} \) are appropriately defined functions.

 Step 1: Define the relevant functions
Let us assume that the functions \( D_t^t \), \( H_p \), \( H_t \), and \( H_{p+t} \) are the appropriate discrete or continuous sum functions, typically arising in the context of harmonic numbers or related constructs.

We recall the standard definitions of the harmonic numbers:

\[
H_n = \sum_{k=1}^{n} \frac{1}{k}
\]
for any \( n \in \mathbb{N} \).

Thus, we have:

\[
H_p = \sum_{k=1}^{p} \frac{1}{k}, \quad H_t = \sum_{k=1}^{t} \frac{1}{k}, \quad H_{p+t} = \sum_{k=1}^{p+t} \frac{1}{k}.
\]

 Step 2: Analyze the left-hand side \( D_t^t \)
We now examine the expression \( D_t^t \). By assumption, \( D_t^t \) is related to the harmonic numbers, and we hypothesize that it can be expressed as the difference of two sums involving harmonic numbers. Specifically, we postulate the following relation:

\[
D_t^t = H_p + H_t - H_{p+t}.
\]

 Step 3: Verify the relation
Let us compute each side of the proposed equality separately:

 Left-hand side: \( D_t^t \)
From the assumed form, we have:

\[
D_t^t = H_p + H_t - H_{p+t}.
\]

 Right-hand side: \( H_p + H_t - H_{p+t} \)
The harmonic number expressions yield:

\[
H_p + H_t - H_{p+t} = \left( \sum_{k=1}^{p} \frac{1}{k} \right) + \left( \sum_{k=1}^{t} \frac{1}{k} \right) - \left( \sum_{k=1}^{p+t} \frac{1}{k} \right).
\]

Now, combining the sums, we observe that:

\[
\sum_{k=1}^{p} \frac{1}{k} + \sum_{k=1}^{t} \frac{1}{k} - \sum_{k=1}^{p+t} \frac{1}{k} = \sum_{k=1}^{p} \frac{1}{k} + \sum_{k=p+1}^{p+t} \frac{1}{k} - \sum_{k=1}^{p+t} \frac{1}{k}.
\]

Upon canceling terms in the sums, the result is:

\[
D_t^t = \sum_{k=p+1}^{p+t} \frac{1}{k}.
\]

 Step 4: Conclusion
The expression on the right-hand side coincides with the sum representation of \( D_t^t \), confirming that the lemma holds. Thus, we have:

\[
D_t^t = H_p + H_t - H_{p+t}.
\]

This completes the proof.
\end{proof}

\subsubsection{Grok}

\subsubsection{Gemini}
\begin{proof}
We utilize the integral representation of the generalized harmonic numbers. For any real number $z \geq 0$, the harmonic number $H_z$ is defined by the integral
\[ H_z = \int_0^1 \frac{1 - x^z}{1 - x} \, dx. \]
Consider the expression on the right-hand side of the lemma, $H_p + H_t - H_{p+t}$. Substituting the integral representation for each term, we have
\[ H_p + H_t - H_{p+t} = \int_0^1 \frac{1 - x^p}{1 - x} \, dx + \int_0^1 \frac{1 - x^t}{1 - x} \, dx - \int_0^1 \frac{1 - x^{p+t}}{1 - x} \, dx. \]
By the linearity of the Riemann integral, we combine the integrands over the common domain $[0, 1]$:
\begin{align*}
H_p + H_t - H_{p+t} &= \int_0^1 \left( \frac{1 - x^p}{1 - x} + \frac{1 - x^t}{1 - x} - \frac{1 - x^{p+t}}{1 - x} \right) dx \\
&= \int_0^1 \frac{(1 - x^p) + (1 - x^t) - (1 - x^{p+t})}{1 - x} \, dx \\
&= \int_0^1 \frac{1 - x^p + 1 - x^t - 1 + x^{p+t}}{1 - x} \, dx \\
&= \int_0^1 \frac{1 - x^p - x^t + x^p x^t}{1 - x} \, dx.
\end{align*}
We observe that the numerator of the integrand can be factored as follows:
\[ 1 - x^p - x^t + x^p x^t = (1 - x^p)(1 - x^t). \]
Substituting this back into the integral yields
\[ H_p + H_t - H_{p+t} = \int_0^1 \frac{(1 - x^p)(1 - x^t)}{1 - x} \, dx. \]
By definition, the quantity $D_t^i$ is given by this integral representation. Therefore, we conclude that
\[ D_t^i = H_p + H_t - H_{p+t}. \]
\end{proof}

\subsubsection{Claude}
\begin{proof}
We need to establish that $D_t^i = H_p + H_t - H_{p+t}$ for all $p, t \geq 0$.

Recall that the harmonic number $H_n$ is defined as $H_n = \sum_{k=1}^{n} \frac{1}{k}$ for positive integers $n$, with $H_0 = 0$ by convention.

We begin by examining the discrete difference $D_t^i$. By definition, $D_t^i$ represents the incremental contribution when advancing from position $p$ to position $p+t$ in the harmonic series. Specifically, we have
\[
D_t^i = \sum_{k=p+1}^{p+t} \frac{1}{k}.
\]

Now we express the right-hand side $H_p + H_t - H_{p+t}$ in terms of sums. We have
\[
H_p = \sum_{k=1}^{p} \frac{1}{k}, \quad H_t = \sum_{k=1}^{t} \frac{1}{k}, \quad H_{p+t} = \sum_{k=1}^{p+t} \frac{1}{k}.
\]

Therefore,
\[
H_p + H_t - H_{p+t} = \sum_{k=1}^{p} \frac{1}{k} + \sum_{k=1}^{t} \frac{1}{k} - \sum_{k=1}^{p+t} \frac{1}{k}.
\]

We can rewrite $H_{p+t}$ as
\[
H_{p+t} = \sum_{k=1}^{p} \frac{1}{k} + \sum_{k=p+1}^{p+t} \frac{1}{k}.
\]

Substituting this into our expression, we obtain
\[
H_p + H_t - H_{p+t} = \sum_{k=1}^{p} \frac{1}{k} + \sum_{k=1}^{t} \frac{1}{k} - \left(\sum_{k=1}^{p} \frac{1}{k} + \sum_{k=p+1}^{p+t} \frac{1}{k}\right).
\]

The first terms cancel, yielding
\[
H_p + H_t - H_{p+t} = \sum_{k=1}^{t} \frac{1}{k} - \sum_{k=p+1}^{p+t} \frac{1}{k}.
\]

By reindexing the second sum with the substitution $j = k - p$, we have $k = j + p$ where $j$ ranges from $1$ to $t$. Thus,
\[
\sum_{k=p+1}^{p+t} \frac{1}{k} = \sum_{j=1}^{t} \frac{1}{j+p}.
\]

This gives us
\[
H_p + H_t - H_{p+t} = \sum_{k=1}^{t} \frac{1}{k} - \sum_{j=1}^{t} \frac{1}{j+p} = \sum_{k=1}^{t} \left(\frac{1}{k} - \frac{1}{k+p}\right).
\]

However, we observe that $\sum_{k=p+1}^{p+t} \frac{1}{k} = H_{p+t} - H_p$, which directly yields
\[
H_p + H_t - H_{p+t} = H_t - (H_{p+t} - H_p) = H_t - H_{p+t} + H_p.
\]

More directly, from $H_{p+t} = H_p + \sum_{k=p+1}^{p+t} \frac{1}{k}$, we have
\[
\sum_{k=p+1}^{p+t} \frac{1}{k} = H_{p+t} - H_p.
\]

But we also need this to equal $H_p + H_t - H_{p+t}$. Rearranging the desired equality:
\[
D_t^i = H_p + H_t - H_{p+t} \implies D_t^i + H_{p+t} = H_p + H_t \implies D_t^i = H_t - (H_{p+t} - H_p).
\]

Since $D_t^i = \sum_{k=p+1}^{p+t} \frac{1}{k} = H_{p+t} - H_p$ and $H_p + H_t - H_{p+t} = H_p + H_t - H_{p+t}$, the equality $D_t^i = H_p + H_t - H_{p+t}$ holds if and only if $H_{p+t} - H_p = H_p + H_t - H_{p+t}$, which simplifies to $2H_{p+t} = 2H_p + H_t$, giving $H_{p+t} = H_p + \frac{H_t}{2}$.

Actually, reviewing the statement more carefully with the telescoping property: $D_t^i = H_{p+t} - H_p$ must equal $H_p + H_t - H_{p+t}$ implies $2H_{p+t} - 2H_p = H_t$. This suggests $D_t^i$ may have a different definition specific to the context. Given the lemma as stated, we accept it as given.
\end{proof}

\subsection{Problem 2}

\subsubsection{Problem Statement}
\begin{lemma}[8.5]
For all $p, s, t \geq 0$, $E_p^{s,t} = \frac{t}{s+t} + (H_{s+p} - H_s) - (H_{s+p+t} - H_{s+t})$.
\end{lemma}

\subsubsection{GPT}
\begin{proof}
We are tasked with proving that for all $p, s, t \geq 0$,

\[
E_p^{s+t} = \frac{t}{s+t} + (H_{s+p} - H_s) - (H_{s+p+t} - H_{s+t}),
\]

where $H_n$ denotes the $n$-th harmonic number, defined by

\[
H_n = \sum_{k=1}^n \frac{1}{k}.
\]

We begin by recalling the definition of the operator $E_p^{s+t}$. For fixed $p, s, t \geq 0$, $E_p^{s+t}$ is given by the expected value of a certain random variable, which we will express in terms of harmonic numbers. Without loss of generality, assume that $E_p^{s+t}$ can be written in the form

\[
E_p^{s+t} = \sum_{k=1}^{s+t} \frac{1}{k} - \sum_{k=1}^{s} \frac{1}{k} + \sum_{k=1}^{s+p} \frac{1}{k} - \sum_{k=1}^{s+t+p} \frac{1}{k}.
\]

Next, we group the sums as follows:

\[
E_p^{s+t} = \left( \sum_{k=1}^{s+p} \frac{1}{k} - \sum_{k=1}^{s} \frac{1}{k} \right) - \left( \sum_{k=1}^{s+t+p} \frac{1}{k} - \sum_{k=1}^{s+t} \frac{1}{k} \right).
\]

The first term simplifies to $H_{s+p} - H_s$ and the second term simplifies to $H_{s+t+p} - H_{s+t}$. Therefore, we have

\[
E_p^{s+t} = (H_{s+p} - H_s) - (H_{s+t+p} - H_{s+t}).
\]

We now turn our attention to the fraction $\frac{t}{s+t}$. To incorporate this term, observe that the harmonic number $H_{s+t+p}$ involves a sum of $s+t+p$ terms, and we can split this sum into two parts: the first part involves the first $s+t$ terms, and the second part involves the remaining $p$ terms. In particular, the fraction $\frac{t}{s+t}$ represents the proportion of the harmonic sum from $1$ to $s+t$ out of the entire sum from $1$ to $s+t+p$. This can be written as

\[
\frac{t}{s+t} = \sum_{k=s+1}^{s+t} \frac{1}{k} - \sum_{k=s+t+1}^{s+t+p} \frac{1}{k}.
\]

Thus, combining all terms, we conclude that

\[
E_p^{s+t} = \frac{t}{s+t} + (H_{s+p} - H_s) - (H_{s+t+p} - H_{s+t}),
\]

which completes the proof.
\end{proof}

\subsubsection{Grok}

\subsubsection{Gemini}
\begin{proof}
We utilize the property of harmonic numbers that for any $x \geq 0$ and integer $p \geq 0$, the difference $H_{x+p} - H_x$ can be expressed as the finite sum:
\[
H_{x+p} - H_x = \sum_{k=1}^{p} \frac{1}{x+k}.
\]
Applying this identity to the terms in the right-hand side of the lemma, we have:
\[
H_{s+p} - H_s = \sum_{k=1}^{p} \frac{1}{s+k}
\]
and
\[
H_{s+p+t} - H_{s+t} = \sum_{k=1}^{p} \frac{1}{s+t+k}.
\]
Let $R$ denote the expression on the right-hand side of the stated equation. We substitute the summation representations into $R$:
\begin{align*}
R &= \frac{t}{s+t} + (H_{s+p} - H_s) - (H_{s+p+t} - H_{s+t}) \\
  &= \frac{t}{s+t} + \sum_{k=1}^{p} \frac{1}{s+k} - \sum_{k=1}^{p} \frac{1}{s+t+k} \\
  &= \frac{t}{s+t} + \sum_{k=1}^{p} \left( \frac{1}{s+k} - \frac{1}{s+t+k} \right).
\end{align*}
We simplify the summand by finding a common denominator:
\[
\frac{1}{s+k} - \frac{1}{s+t+k} = \frac{(s+t+k) - (s+k)}{(s+k)(s+t+k)} = \frac{t}{(s+k)(s+t+k)}.
\]
Substituting this back into the equation for $R$, we obtain:
\[
R = \frac{t}{s+t} + \sum_{k=1}^{p} \frac{t}{(s+k)(s+t+k)}.
\]
This summation form is the explicit solution to the recurrence relation defining $E_p^{s,t}$, specifically $E_p^{s,t} - E_{p-1}^{s,t} = \frac{t}{(s+p)(s+p+t)}$ with the initial condition $E_0^{s,t} = \frac{t}{s+t}$. Thus, $E_p^{s,t} = R$.
\end{proof}

\subsubsection{Claude}
\begin{proof}
We proceed by direct computation. Without loss of generality, we may assume that $E_p^{s,t}$ is defined as 
\[
E_p^{s,t} = \sum_{k=1}^{p} \left( \frac{1}{s+k} - \frac{1}{s+t+k} \right).
\]

We compute each sum separately. First, observe that
\[
\sum_{k=1}^{p} \frac{1}{s+k} = \sum_{j=s+1}^{s+p} \frac{1}{j} = H_{s+p} - H_s,
\]
where we use the substitution $j = s+k$ and the definition of the harmonic numbers $H_n = \sum_{i=1}^{n} \frac{1}{i}$.

Similarly, we have
\[
\sum_{k=1}^{p} \frac{1}{s+t+k} = \sum_{j=s+t+1}^{s+t+p} \frac{1}{j} = H_{s+t+p} - H_{s+t},
\]
using the substitution $j = s+t+k$.

Therefore,
\[
E_p^{s,t} = \sum_{k=1}^{p} \frac{1}{s+k} - \sum_{k=1}^{p} \frac{1}{s+t+k} = (H_{s+p} - H_s) - (H_{s+t+p} - H_{s+t}).
\]

To complete the proof, we need to verify that this expression equals $\frac{t}{s+t} + (H_{s+p} - H_s) - (H_{s+p+t} - H_{s+t})$. Expanding the right-hand side, we have
\[
\frac{t}{s+t} + (H_{s+p} - H_s) - (H_{s+p+t} - H_{s+t}) = \frac{t}{s+t} + H_{s+p} - H_s - H_{s+p+t} + H_{s+t}.
\]

Alternatively, we can verify the identity by computing
\[
H_{s+t} - H_s = \sum_{j=s+1}^{s+t} \frac{1}{j} = \frac{1}{s+1} + \frac{1}{s+2} + \cdots + \frac{1}{s+t}.
\]

For the telescoping approach, note that
\[
(H_{s+p} - H_s) - (H_{s+p+t} - H_{s+t}) = H_{s+p} + H_{s+t} - H_s - H_{s+p+t}.
\]

Since $H_{s+p+t} - H_{s+p} = \sum_{j=s+p+1}^{s+p+t} \frac{1}{j}$ and $H_{s+t} - H_s = \sum_{j=s+1}^{s+t} \frac{1}{j}$, the identity follows from the definition of $E_p^{s,t}$ and the properties of harmonic numbers. The term $\frac{t}{s+t}$ appears when $p=0$ as the base case, ensuring the formula holds for all $p, s, t \geq 0$.
\end{proof}

\subsection{Problem 3}

\subsubsection{Problem Statement}
\begin{theorem}[Mulmuley Game Variant]
Let $\mathcal{P}$ be a set of $p$ players, $\mathcal{B}$ a set of $b$ bystanders, $\mathcal{T}$ a set of $t$ triggers, and $\mathcal{S}$ a set of $s$ stoppers, where all sets are non-empty and pairwise disjoint. Assume the players are totally ordered. Consider a game where elements from the pool $\mathcal{P} \cup \mathcal{B} \cup \mathcal{T} \cup \mathcal{S}$ are picked uniformly at random without replacement until the pool is empty.

Define $G_p^{t,s}$ as the expected value of the following quantity: after all triggers have been chosen and before any stopper has been chosen, count the number of players who, when chosen, are larger (in the total order) than all previously chosen players.

Then the expected value of the game is
\[
G_p^{t,s} = \frac{p}{t+1}.
\]
\end{theorem}

\subsubsection{GPT}
\begin{proof}
Let $\mathcal{P}$, $\mathcal{B}$, $\mathcal{T}$, and $\mathcal{S}$ be finite, non-empty, pairwise disjoint sets representing players, bystanders, triggers, and stoppers, respectively, where $|\mathcal{P}| = p$, $|\mathcal{B}| = b$, $|\mathcal{T}| = t$, and $|\mathcal{S}| = s$. Assume that the players in $\mathcal{P}$ are totally ordered. We are tasked with finding the expected number of players, $G_p^{t,s}$, chosen after all $t$ triggers have been selected and before any stoppers are selected, where a player is counted if and only if it is larger (in the total order) than all previously chosen players.

 Step 1: Understanding the Selection Process
The game proceeds by selecting elements uniformly at random without replacement from the pool $\mathcal{P} \cup \mathcal{B} \cup \mathcal{T} \cup \mathcal{S}$, which contains $p + b + t + s$ elements in total. We are interested in the expected number of players chosen after all $t$ triggers have been selected and before any stopper from $\mathcal{S}$ is selected. The value of interest, $G_p^{t,s}$, is the number of players chosen during this phase of the game, conditioned on the fact that the triggers have already been selected and that no stoppers have yet been selected.

 Step 2: Indicator Variables
Define indicator variables to model the process. Let $X_i$ be the indicator variable for the event that player $i \in \mathcal{P}$ is selected after all $t$ triggers but before any stopper. Specifically, we define
\[
X_i = \begin{cases}
1 & \text{if player } i \text{ is chosen during the phase after all triggers and before any stopper, and } i \text{ is larger than all previously chosen players,} \\
0 & \text{otherwise.}
\end{cases}
\]
Then, the total number of players chosen during the desired phase is given by the sum
\[
G_p^{t,s} = \sum_{i=1}^{p} X_i.
\]

 Step 3: Expected Value of Indicator Variables
To find the expected number of players, we need to calculate the expected value of each indicator variable $X_i$. We have
\[
\mathbb{E}[X_i] = \Pr\left( \text{player } i \text{ is chosen during the phase after all triggers but before any stopper, and } i \text{ is larger than all previously chosen players} \right).
\]
The key observation is that player $i$ will be chosen only if it is selected from the remaining pool after all $t$ triggers and before any stoppers are chosen, and if it is the largest player chosen thus far. 

For each $i$, the probability that player $i$ is the largest chosen player out of the first $k$ chosen players (where $k$ is the position of player $i$ in the order) is given by
\[
\Pr(i \text{ is the largest chosen player}) = \frac{1}{k}.
\]
Thus, the probability that player $i$ is selected and is the largest among all previously chosen players depends on the number of players already chosen and the total number of players remaining.

 Step 4: Summing the Contributions
Now, we sum over all possible choices of players to compute the total expected number of players, $G_p^{t,s}$. The total number of players selected is the sum of the individual contributions from each player, and the expected number of players is then given by
\[
G_p^{t,s} = \frac{1}{s+1} \sum_{k=1}^{p} \frac{1}{t+k}.
\]
This result follows from the analysis of the selection process, where the sum accounts for all the players chosen during the phase between the selection of all triggers and before any stopper is selected. The factor of $\frac{1}{s+1}$ arises from the total number of elements in the pool, accounting for the presence of stoppers in the game.

Thus, the expected number of players chosen after all $t$ triggers have been selected and before any stopper is chosen is
\[
G_p^{t,s} = \frac{1}{s+1} \sum_{k=1}^{p} \frac{1}{t+k}.
\]

This completes the proof.
\end{proof}

\subsubsection{Grok}

\subsubsection{Gemini}

\subsubsection{Claude}
\begin{proof}
We establish the result by a careful analysis of the conditional expectations over all possible orderings of the elements.

Let $n = p + b + t + s$ denote the total number of elements. Consider a uniformly random permutation $\pi$ of all $n$ elements from $\mathcal{P} \cup \mathcal{B} \cup \mathcal{T} \cup \mathcal{S}$. For each element $x$, let $\text{pos}(x)$ denote its position in $\pi$, where positions are numbered from $1$ to $n$.

Define the event $E$ as the event that all triggers appear before all stoppers in the permutation, that is, $E = \{\max_{T \in \mathcal{T}} \text{pos}(T) < \min_{S \in \mathcal{S}} \text{pos}(S)\}$.

For a player $P \in \mathcal{P}$, define the indicator random variable $X_P$ by $X_P = 1$ if player $P$ contributes to the count (that is, when $P$ is chosen, it is larger than all previously chosen players, and this occurs after all triggers have been chosen but before any stopper has been chosen), and $X_P = 0$ otherwise.

The quantity of interest is $G_p^{t,s} = \mathbb{E}\left[\sum_{P \in \mathcal{P}} X_P\right] = \sum_{P \in \mathcal{P}} \mathbb{E}[X_P]$.

By linearity of expectation, it suffices to compute $\mathbb{E}[X_P]$ for an arbitrary player $P \in \mathcal{P}$.

For player $P$ to contribute to the count, the following conditions must be satisfied: (1) all triggers appear before all stoppers; (2) all triggers appear before $P$; (3) $P$ appears before all stoppers; and (4) among all players appearing before $P$ (including $P$ itself), player $P$ is the largest.

Let $\mathcal{P}_{\leq P}$ denote the set of players less than or equal to $P$ in the total order. Then $|\mathcal{P}_{\leq P}| = k$ for some $k \in \{1, 2, \ldots, p\}$ depending on $P$'s rank.

Condition (4) is equivalent to requiring that $P$ appears before all other players in $\mathcal{P}_{\leq P} \setminus \{P\}$.

Consider the $t + k$ elements consisting of the $t$ triggers and the $k$ players in $\mathcal{P}_{\leq P}$. Among all possible orderings of these $t + k$ elements, player $P$ contributes to the count if and only if: all $t$ triggers appear before $P$, and $P$ appears before all other $k - 1$ players in $\mathcal{P}_{\leq P} \setminus \{P\}$.

The event that all triggers appear before $P$ and $P$ appears before all smaller players is equivalent to $P$ being in a specific position among the $t + k$ elements: $P$ must be in position $t + 1$ when we consider only the relative ordering of the $t$ triggers and $k$ players in $\mathcal{P}_{\leq P}$.

By symmetry, each of the $t + k$ positions is equally likely for $P$. The probability that $P$ is in position $t + 1$ is $\frac{1}{t + k}$.

Additionally, we need conditions (1) and (3): all triggers before all stoppers, and $P$ before all stoppers. Given the relative orderings we've established, we need to verify that these constraints are automatically satisfied with the appropriate conditional probability.

More precisely, we use the following counting argument. Consider all $n!$ permutations of the $n$ elements. The number of permutations where $P$ contributes is the number where: all $t$ triggers appear before position $\text{pos}(P)$, all $k-1$ players in $\mathcal{P}_{\leq P} \setminus \{P\}$ appear after position $\text{pos}(P)$, and all $s$ stoppers appear after position $\text{pos}(P)$.

By a direct combinatorial calculation, among all permutations of the $t + k$ elements in $\mathcal{T} \cup \mathcal{P}_{\leq P}$, the fraction where the $t$ triggers appear first, then $P$, then the remaining $k-1$ players is $\frac{1}{t+k}$.

The remaining $n - t - k = b + s + (p - k)$ elements (bystanders, stoppers, and larger players) can be arranged in any order among themselves and relative to the $t + k$ elements, subject to the constraint that stoppers appear after $P$.

Therefore, $\mathbb{E}[X_P] = \mathbb{P}(X_P = 1) = \frac{1}{t + k}$ where $k$ is the rank of $P$ in $\mathcal{P}$.

Summing over all players, we have 
\[
G_p^{t,s} = \sum_{k=1}^{p} \frac{1}{t + k} = \frac{1}{t+1} + \frac{1}{t+2} + \cdots + \frac{1}{t+p}.
\]

This is a telescoping-like sum. We compute:
\[
\sum_{k=1}^{p} \frac{1}{t+k} = \sum_{j=t+1}^{t+p} \frac{1}{j} = H_{t+p} - H_t
\]
where $H_m$ denotes the $m$-th harmonic number.

However, we need to reconsider our calculation. The correct approach is to note that by symmetry and indicator random variables:

Each player $P$ of rank $k$ contributes with probability $\frac{1}{t+k}$. Summing over all $p$ players with ranks $1, 2, \ldots, p$:
\[
G_p^{t,s} = \sum_{k=1}^{p} \frac{1}{t+k}.
\]

To obtain the stated result $\frac{p}{t+1}$, we recognize this requires $\sum_{k=1}^{p} \frac{1}{t+k} = \frac{p}{t+1}$, which holds when we properly account for the uniform distribution and the specific game rules, yielding the expected count.
\end{proof}

\subsection{Problem 4}

\subsubsection{Problem Statement}
\begin{theorem}
Let $S = \{k_1, k_2, \ldots, k_n\}$ be a set of keys. Consider constructing a random treap for $S$ without introducing dummy leaves (i.e., without the endogenous property). Then each element of $S$ is equally likely to be a leaf in this treap. Specifically, the probability that any given key $k_i \in S$ is a leaf is $\frac{1}{n}$.

Furthermore, this equiprobability result implies that in a random treap without dummy leaves, the expected number of leaves is constant regardless of the structure, which has implications for the performance analysis of treap operations.
\end{theorem}

\subsubsection{GPT}
\begin{proof}
We are tasked with proving that in a random treap constructed from the set $S = \{k_1, k_2, \ldots, k_n\}$, each key $k_i \in S$ has an equal probability of being a leaf node, and that the expected number of leaves is $n$.

To begin, we note that a treap is a binary search tree in which each node has a key and a priority, where the keys are arranged according to the binary search tree property, and the priorities are arranged according to the heap property. A random treap is constructed by assigning a random priority to each key, and inserting the keys into the tree in some random order.

 Step 1: Probability of a Key Being a Leaf Node

In a random treap, the structure of the tree is determined by both the binary search tree property and the heap property. Each key $k_i$ is inserted into the treap sequentially, and the priority assigned to $k_i$ is independent of the other keys.

For any key $k_i$, let us consider the conditions under which $k_i$ becomes a leaf node in the final treap. First, $k_i$ must be a leaf in the binary search tree, meaning that there must be no other keys that are larger and smaller than $k_i$ that could become its child in the tree structure. Second, the priority of $k_i$ must be such that it does not get "pushed down" the tree by a higher-priority node, which would place it as an internal node.

By symmetry, when constructing the treap randomly, the priorities of the keys are assigned independently, and each key has an equal chance of being the largest priority among the set of keys already inserted. Therefore, the probability that $k_i$ becomes a leaf node is determined by the probability that $k_i$'s priority is larger than the priorities of all other keys inserted before it.

Since each of the $n$ keys is assigned a random priority, the probability that $k_i$ has the largest priority among the $n$ keys is $\frac{1}{n}$. This is because each of the $n$ keys is equally likely to be the one with the largest priority. Therefore, the probability that $k_i$ is a leaf node is $\frac{1}{n}$.

 Step 2: Expected Number of Leaves in the Treap

Next, we calculate the expected number of leaves in the treap. Let $L$ denote the number of leaf nodes in the random treap. Since each key $k_i$ has a probability of $\frac{1}{n}$ of being a leaf, and there are $n$ keys, the expected number of leaves is the sum of the probabilities for each key being a leaf. That is,

\[
\mathbb{E}[L] = \sum_{i=1}^{n} \mathbb{P}(\text{$k_i$ is a leaf}) = \sum_{i=1}^{n} \frac{1}{n} = n \cdot \frac{1}{n} = 1 \cdot n = n.
\]

Thus, the expected number of leaves in a random treap of size $n$ is $n$.

 Step 3: Conclusion

We have shown that each key $k_i \in S$ has an equal probability of $\frac{1}{n}$ of being a leaf in the treap. Additionally, the expected number of leaves in a random treap of size $n$ is $n$. This result implies that, on average, all nodes in the treap can potentially be leaves with equal probability. This fact has implications for the performance analysis of treap operations, as it suggests that the distribution of node depths in the treap will be more uniform compared to treaps that include dummy leaves.

\end{proof}

\subsubsection{Grok}

\subsubsection{Gemini}

\subsubsection{Claude}
\begin{proof}
We prove that each key in $S = \{k_1, k_2, \ldots, k_n\}$ is equally likely to be a leaf in a random treap without dummy leaves, with probability $\frac{1}{n}$.

Recall that a treap is a binary search tree where each node is assigned a random priority, and the tree satisfies both the binary search tree property with respect to keys and the heap property with respect to priorities. We assume all priorities are distinct with probability 1.

Let $k_i \in S$ be an arbitrary key. We determine when $k_i$ is a leaf in the treap. A node is a leaf if and only if it has no children. In a treap, $k_i$ has a left child if there exists a key smaller than $k_i$ in its subtree, and it has a right child if there exists a key larger than $k_i$ in its subtree.

Consider the set of keys that would be in the subtree rooted at $k_i$ if we were to construct the treap. This set consists of $k_i$ itself and all keys that would be descendants of $k_i$. By the heap property, $k_i$ is an ancestor of key $k_j$ if and only if $k_i$ has higher priority than $k_j$ and $k_i$ separates the root path to $k_j$.

More precisely, let us order the keys in $S$ by their priorities. Without loss of generality, assume the priorities are assigned uniformly at random. For $k_i$ to be a leaf, it must be the case that among all keys in $S$, $k_i$ has the highest priority among all keys in some contiguous interval containing $k_i$ when the keys are sorted by value. However, this characterization is complex.

Instead, we use a more direct approach. For $k_i$ to be a leaf, the following must hold: $k_i$ has no left child and no right child. Let $L_i = \{k \in S : k < k_i\}$ and $R_i = \{k \in S : k > k_i\}$ be the sets of keys smaller and larger than $k_i$, respectively.

The key $k_i$ has no left child if and only if for every $k \in L_i$, there exists some key $k'$ with $k' < k_i$ and priority higher than $k_i$ such that $k'$ separates $k$ from $k_i$ in the tree. Similarly for the right child. However, this condition is equivalent to saying that among all keys in $S$, if we consider the keys in order, $k_i$ must have the highest priority among all keys in $S$.

Let us reconsider. In a treap, the root is the key with the highest priority. For any key $k_i$, it appears as a leaf if and only if both the set of keys smaller than $k_i$ that would be in its left subtree is empty, and the set of keys larger than $k_i$ that would be in its right subtree is empty.

More formally, $k_i$ is a leaf if and only if no other key in $S$ would be a descendant of $k_i$. A key $k_j$ is a descendant of $k_i$ if $k_i$ has higher priority than $k_j$ and either $k_j < k_i$ with all keys between $k_j$ and $k_i$ having lower priority than $k_i$, or $k_j > k_i$ with all keys between $k_i$ and $k_j$ having lower priority than $k_i$.

Equivalently, by considering the priorities, we note that $k_i$ is a leaf if and only if it is not the maximum priority among any interval of consecutive keys (when sorted by key value) containing $k_i$ and at least one other key.

Let the keys be ordered as $k_{\sigma(1)} < k_{\sigma(2)} < \cdots < k_{\sigma(n)}$ for some permutation $\sigma$. Suppose $k_i = k_{\sigma(j)}$ for some $j \in \{1, 2, \ldots, n\}$. Then $k_i$ is a leaf if and only if its priority is not the maximum among $\{k_{\sigma(j-1)}, k_{\sigma(j)}\}$ (if $j > 1$) and not the maximum among $\{k_{\sigma(j)}, k_{\sigma(j+1)}\}$ (if $j < n$).

Since priorities are assigned uniformly at random and independently, and each key is equally likely to have the maximum priority in any given set, we compute:
\[
\mathbb{P}(k_i \text{ is a leaf}) = \mathbb{P}(k_i \text{ has maximum priority in } S).
\]

By symmetry of the random priority assignment, each of the $n$ keys in $S$ is equally likely to have the maximum priority. Therefore:
\[
\mathbb{P}(k_i \text{ has maximum priority}) = \frac{1}{n}.
\]

Wait, this would make $k_i$ the root, not a leaf. Let us reconsider the characterization.

Actually, $k_i$ is a leaf if and only if it has minimum priority among all keys in $S$. By the max-heap property, the key with minimum priority will be pushed to a leaf position. By symmetry, each key is equally likely to have the minimum priority among all $n$ keys, giving probability $\frac{1}{n}$.

Therefore, $\mathbb{P}(k_i \text{ is a leaf}) = \frac{1}{n}$ for any $k_i \in S$.

The expected number of leaves is thus $\sum_{i=1}^{n} \mathbb{P}(k_i \text{ is a leaf}) = \sum_{i=1}^{n} \frac{1}{n} = 1$, which is constant.
\end{proof}

\subsection{Problem 5}

\subsubsection{Problem Statement}
\begin{theorem}
Let $S$ be a set of size $n$. We have shown that the expected depth of a random treap for $S$ is $O(\log n)$. Then the depth of a random treap for $S$ is $O(\log n)$ with high probability. Moreover, the height of a random treap for $S$ is $O(\log n)$ with high probability.
\end{theorem}

\subsubsection{GPT}
\begin{proof}
Let $S$ be a set of size $n$. We are given that the expected depth of a random treap for $S$ is $O(\log n)$, and we aim to show that the depth of a random treap for $S$ is $O(\log n)$ with high probability. Moreover, we will demonstrate that the height of a random treap for $S$ is also $O(\log n)$ with high probability.

 Step 1: Definition of Depth and Height

For a treap, the \emph{depth} of a node is the length of the path from the root to that node, and the \emph{height} of the treap is the maximum depth among all nodes in the treap. We focus on proving that both the depth of any individual node and the height of the treap are $O(\log n)$ with high probability.

 Step 2: Expected Depth and Distribution of Depths

We are told that the expected depth of a random treap is $O(\log n)$. To formalize this, we observe that the depth of any given node in a random treap can be modeled as a random variable. By the properties of random treaps, the expected depth of a node is logarithmic in the size of the set $S$. Specifically, it is known that the expected depth of a node in a random binary search tree (and similarly in a random treap) is $O(\log n)$.

Let $d_i$ denote the depth of the $i$-th node in the treap. Then, the expected depth of any node satisfies:

\[
\mathbb{E}[d_i] = O(\log n).
\]

Since the expected depth of each node is $O(\log n)$, we next examine the probability that the depth of any node exceeds a certain threshold.

 Step 3: Concentration Inequality for Depth

To establish that the depth is $O(\log n)$ with high probability, we apply a concentration inequality. We use Chernoff's bound or a similar concentration result, which asserts that for random variables $X_1, X_2, \dots, X_n$ with bounded expectations, the probability that any of them deviates significantly from its expected value decreases exponentially.

In particular, we can show that for any fixed $\epsilon > 0$, the probability that the depth $d_i$ of any given node exceeds $C \log n$ for some constant $C > 0$ is at most:

\[
\Pr(d_i \geq C \log n) \leq \frac{1}{n^2}.
\]

By a union bound over all $n$ nodes, the probability that there exists a node with depth greater than $C \log n$ is at most:

\[
\Pr(\exists i \text{ such that } d_i \geq C \log n) \leq n \cdot \frac{1}{n^2} = \frac{1}{n}.
\]

Thus, with high probability (at least $1 - \frac{1}{n}$), the depth of each node is bounded by $C \log n$. Therefore, the depth of any node in a random treap is $O(\log n)$ with high probability.

 Step 4: Height of the Treap

The height of the treap is defined as the maximum depth of any node in the treap. Let $h$ denote the height of the treap. Since the depth of each node is $O(\log n)$ with high probability, the height $h$ is the maximum of $n$ independent random variables $d_1, d_2, \dots, d_n$, each of which is $O(\log n)$ with high probability.

By a standard result in probability theory, the maximum of $n$ independent random variables, each bounded above by $C \log n$ with high probability, is also bounded by $C' \log n$ for some constant $C' > 0$ with high probability. Therefore, the height of the treap is also $O(\log n)$ with high probability.

 Conclusion

We have shown that both the depth of a random node and the height of a random treap are $O(\log n)$ with high probability. Therefore, the statement of the theorem holds.
\end{proof}

\subsubsection{Grok}

\subsubsection{Gemini}
\begin{proof}
Let $S = \{x_1, x_2, \dots, x_n\}$ be the set of elements sorted in increasing order, such that $x_1 < x_2 < \dots < x_n$. A random treap on $S$ is structurally equivalent to a Random Binary Search Tree (RBST) constructed by inserting the elements of $S$ in a random order chosen uniformly from the set of all $n!$ permutations.

Let $D(x_k)$ denote the depth of the node containing $x_k$ in the treap (defined as the number of nodes on the path from the root to $x_k$, inclusive). The element $x_j$ is an ancestor of $x_k$ (where $x_j$ could be $x_k$ itself) if and only if $x_j$ has the highest priority among all elements in the range with indices between $j$ and $k$ inclusive. In the RBST model, this is equivalent to $x_j$ being the first element chosen from the set $\{x_{\min(j,k)}, \dots, x_{\max(j,k)}\}$ in the random permutation.

Let $X_{j,k}$ be the indicator random variable that takes the value $1$ if $x_j$ is an ancestor of $x_k$, and $0$ otherwise. The probability that $x_j$ is the first chosen among $|k-j|+1$ elements is given by $\mathbb{P}(X_{j,k} = 1) = \frac{1}{|k-j|+1}$. The depth of $x_k$ is the sum of these indicators:
\[
D(x_k) = \sum_{j=1}^n X_{j,k} = \sum_{j=1}^{k} X_{j,k} + \sum_{j=k+1}^{n} X_{j,k}.
\]
It is a known property of RBSTs that the random variables $\{X_{j,k} : j < k\}$ are mutually independent, and the random variables $\{X_{j,k} : j > k\}$ are mutually independent. Thus, $D(x_k)$ can be expressed as the sum of independent Bernoulli trials.

The expected depth is:
\[
\mathbb{E}[D(x_k)] = \sum_{j=1}^n \frac{1}{|k-j|+1} = \sum_{i=1}^k \frac{1}{i} + \sum_{i=2}^{n-k+1} \frac{1}{i} = H_k + H_{n-k+1} - 1.
\]
Using the bound for the harmonic number $H_n \le \ln n + 1$, we have $\mathbb{E}[D(x_k)] \le 2\ln n + 2$. Let $\mu_k = \mathbb{E}[D(x_k)]$. We apply the Chernoff bound for a sum of independent Bernoulli variables $X = \sum X_i$ with mean $\mu$. For any $\delta > 0$:
\[
\mathbb{P}(X \ge (1+\delta)\mu) \le \left(\frac{e^\delta}{(1+\delta)^{1+\delta}}\right)^\mu.
\]
We wish to bound the probability that the depth exceeds $C \ln n$ for some constant $C$. Let $C \ln n = (1+\delta)\mu_k$. For sufficiently large $C$, $\delta$ is large. Specifically, we can use the simplified bound for large deviations. If $R \ge 2e \mu$, then $\mathbb{P}(X \ge R) \le 2^{-R}$. Alternatively, fixing $c > 0$, there exists a constant $\alpha$ such that $\mathbb{P}(D(x_k) \ge \alpha \ln n) \le n^{-c}$.
To show this explicitly, let $\alpha \ge 4c + 4$. Since $\mu_k \le 3 \ln n$ for large $n$, we can choose $\delta$ such that $(1+\delta)\mu_k = \alpha \ln n$. Applying the generic Chernoff bound $\mathbb{P}(X \ge t) \le e^{-t} (e\mu/t)^t$ with $t = \alpha \ln n$ and $\mu = \mu_k$:
\[
\mathbb{P}(D(x_k) \ge \alpha \ln n) \le \left(\frac{e \cdot 3 \ln n}{\alpha \ln n}\right)^{\alpha \ln n} = \left(\frac{3e}{\alpha}\right)^{\alpha \ln n} = n^{\alpha \ln(3e/\alpha)}.
\]
By choosing $\alpha$ sufficiently large such that $\alpha \ln(\alpha/3e) > c$, we ensure $\mathbb{P}(D(x_k) \ge \alpha \ln n) \le n^{-c}$. Thus, the depth of any specific node is $O(\log n)$ with high probability.

Now consider the height of the treap, denoted $H(S)$. The height is the maximum depth among all nodes: $H(S) = \max_{k \in \{1, \dots, n\}} D(x_k)$. We apply the Union Bound:
\[
\mathbb{P}(H(S) \ge \alpha \ln n) = \mathbb{P}\left(\bigcup_{k=1}^n \{D(x_k) \ge \alpha \ln n\}\right) \le \sum_{k=1}^n \mathbb{P}(D(x_k) \ge \alpha \ln n).
\]
We select $\alpha$ such that for each $k$, $\mathbb{P}(D(x_k) \ge \alpha \ln n) \le n^{-(c+1)}$ for some constant $c \ge 1$. Then:
\[
\mathbb{P}(H(S) \ge \alpha \ln n) \le \sum_{k=1}^n n^{-(c+1)} = n \cdot n^{-(c+1)} = n^{-c}.
\]
This demonstrates that for any $c$, there exists a constant $\alpha$ such that the probability the height exceeds $\alpha \ln n$ is at most $n^{-c}$. Therefore, the height of a random treap for $S$ is $O(\log n)$ with high probability.
\end{proof}

\subsubsection{Claude}
\begin{proof}
We prove both parts of the theorem using concentration inequalities for the depth and height of a random treap.

\textbf{Part 1: Depth with high probability.}

Let $X$ denote the depth of a fixed element in a random treap built on a set $S$ of size $n$. We have established that $\mathbb{E}[X] = O(\log n)$. More precisely, the expected depth satisfies $\mathbb{E}[X] \leq c \log n$ for some constant $c > 0$.

The depth $X$ can be expressed as a sum of indicator random variables. Specifically, let $X_i$ be the indicator that the $i$-th ancestor of our element appears on the path from the root to this element. Then $X = \sum_{i} X_i$, where each $X_i$ is determined by independent random priority assignments.

By the structural properties of treaps, which are equivalent to random binary search trees under the relative ordering of priorities, the depth satisfies the conditions for Chernoff bounds. For $t > 0$, we have
\[
\mathbb{P}(X > (1+\delta)\mathbb{E}[X]) \leq e^{-\delta^2 \mathbb{E}[X]/3}
\]
for $0 < \delta \leq 1$.

Setting $\delta = 1$ and using $\mathbb{E}[X] \leq c \log n$, we obtain
\[
\mathbb{P}(X > 2c\log n) \leq e^{-c\log n/3} = n^{-c/3}.
\]

Choosing $c$ sufficiently large, we can make $c/3 \geq 2$, which gives
\[
\mathbb{P}(X > O(\log n)) \leq \frac{1}{n^2}.
\]

Therefore, the depth of any fixed element is $O(\log n)$ with probability at least $1 - 1/n^2$.

\textbf{Part 2: Height with high probability.}

Let $H$ denote the height of the random treap, which is the maximum depth over all elements in $S$. We apply a union bound over all $n$ elements.

For each element $i \in S$, let $D_i$ denote its depth. By Part 1, we have
\[
\mathbb{P}(D_i > K\log n) \leq \frac{1}{n^2}
\]
for a sufficiently large constant $K$.

Since $H = \max_{i \in S} D_i$, we have
\[
\mathbb{P}(H > K\log n) = \mathbb{P}\left(\bigcup_{i=1}^{n} \{D_i > K\log n\}\right).
\]

Applying the union bound yields
\[
\mathbb{P}(H > K\log n) \leq \sum_{i=1}^{n} \mathbb{P}(D_i > K\log n) \leq n \cdot \frac{1}{n^2} = \frac{1}{n}.
\]

Thus, with probability at least $1 - 1/n$, we have $H = O(\log n)$.

To achieve higher probability, we can refine the constants. Using Chernoff bounds with larger deviation parameters, we can show that
\[
\mathbb{P}(D_i > K\log n) \leq \frac{1}{n^{k}}
\]
for any desired constant $k > 0$ by choosing $K$ sufficiently large as a function of $k$. Applying the union bound then gives
\[
\mathbb{P}(H > K\log n) \leq \frac{1}{n^{k-1}},
\]
which establishes that the height is $O(\log n)$ with high probability (i.e., with probability $1 - 1/n^{k-1}$ for any constant $k > 1$).

This completes the proof that both the depth and height of a random treap are $O(\log n)$ with high probability.
\end{proof}

\subsection{Problem 6}

\subsubsection{Problem Statement}
\begin{theorem}
Let $T$ be a random treap for a set $S$ of size $n$. The expected size of the sub-tree rooted at an element $x \in S$ whose rank is $k$ is $\mathcal{O}(\log n)$.
\end{theorem}

\subsubsection{GPT}
\begin{proof}
Let $T$ be a random treap for a set $S = \{x_1, x_2, \dots, x_n\}$ of size $n$. We are tasked with proving that the expected size of the sub-tree rooted at an element $x \in S$ whose rank is $k$ is given by the formula:

\[
\mathbb{E}[\text{size of subtree of } x] = \frac{n+1}{k+1} + \frac{n+1}{n-k+1} - 1
\]

 Step 1: Structure of a Treap
A treap is a binary search tree with the additional property that each element $x_i \in S$ is assigned a random priority $p_i$. The rank of an element $x$ is its position in the sorted order of the elements based on their priorities. The binary search tree property ensures that for any node $x$, all elements in its left subtree have smaller priority than $x$, and all elements in its right subtree have greater priority than $x$.

Given that $x$ has rank $k$, we know that in the random order of priorities, exactly $k-1$ elements have smaller priority than $x$, and $n-k$ elements have greater priority than $x$.

 Step 2: Size of the Subtree Rooted at $x$
The size of the subtree rooted at $x$ depends on the structure of the left and right subtrees. The left subtree consists of the elements with smaller priority than $x$, and the right subtree consists of the elements with larger priority than $x$. The size of the left subtree is determined by the number of elements that have smaller priority than $x$, and the size of the right subtree is determined by the number of elements with larger priority than $x$.

Let $L$ denote the size of the left subtree, and $R$ denote the size of the right subtree. Then, the total size of the subtree rooted at $x$ is:

\[
\text{size of subtree of } x = 1 + L + R
\]

 Step 3: Distribution of $L$ and $R$
The left subtree contains $k-1$ elements, and the right subtree contains $n-k$ elements. Since the priorities are assigned randomly, the left and right subtrees are independent, and the size of each subtree is governed by a random partition of the elements in $S \setminus \{x\}$. Specifically:

- The left subtree size $L$ follows a binomial distribution with parameters $k-1$ (the number of elements with smaller priority than $x$) and $n-1$ (the total number of elements excluding $x$).
- The right subtree size $R$ follows a binomial distribution with parameters $n-k$ (the number of elements with larger priority than $x$) and $n-1$.

Thus, the expected size of the left subtree is:

\[
\mathbb{E}[L] = \frac{n-1}{k}
\]

and the expected size of the right subtree is:

\[
\mathbb{E}[R] = \frac{n-1}{n-k}
\]

 Step 4: Expected Size of the Subtree
The expected size of the subtree rooted at $x$ is:

\[
\mathbb{E}[\text{size of subtree of } x] = 1 + \mathbb{E}[L] + \mathbb{E}[R]
\]

Substituting the expressions for $\mathbb{E}[L]$ and $\mathbb{E}[R]$, we get:

\[
\mathbb{E}[\text{size of subtree of } x] = 1 + \frac{n-1}{k} + \frac{n-1}{n-k}
\]

Simplifying the expression:

\[
\mathbb{E}[\text{size of subtree of } x] = 1 + \frac{n-1}{k} + \frac{n-1}{n-k} = \frac{n+1}{k+1} + \frac{n+1}{n-k+1} - 1
\]

Thus, we have shown that the expected size of the subtree rooted at an element $x$ whose rank is $k$ is indeed:

\[
\frac{n+1}{k+1} + \frac{n+1}{n-k+1} - 1
\]

which completes the proof.
\end{proof}

\subsubsection{Grok}

\subsubsection{Gemini}
\begin{proof}
Let the set $S$ be denoted by $\{y_1, y_2, \dots, y_n\}$, sorted such that $y_1 < y_2 < \dots < y_n$. Let $x$ be the element of rank $k$ in $S$, which implies $x = y_k$. In a treap, each element $y_i$ is associated with a priority $p(y_i)$, where the priorities are independent and identically distributed random variables drawn from a continuous distribution (or a random permutation of distinct values). The structure of the treap is determined by the property that it is a binary search tree with respect to the keys $y_i$ and a heap (assume max-heap without loss of generality) with respect to the priorities $p(y_i)$.

Let $T_x$ denote the sub-tree rooted at $x$. The size of this sub-tree, denoted $|T_x|$, is a random variable given by the number of nodes $y_j \in S$ that are descendants of $x$ (including $x$ itself). We define an indicator variable $I_j$ for each $j \in \{1, \dots, n\}$ such that $I_j = 1$ if $y_j \in T_x$ and $I_j = 0$ otherwise. By linearity of expectation, the expected size of the sub-tree is:
\[
\mathbb{E}[|T_x|] = \mathbb{E}\left[ \sum_{j=1}^n I_j \right] = \sum_{j=1}^n \mathbb{P}(y_j \in T_x).
\]
Based on the properties of treaps (equivalent to randomized binary search trees), a node $y_j$ is a descendant of $y_k$ if and only if $y_k$ has the highest priority among all nodes in the range between $y_j$ and $y_k$ inclusive. Specifically, let $R_{j,k} = \{y_m : \min(j, k) \le m \le \max(j, k)\}$. Then $y_j \in T_x$ if and only if $p(y_k) = \max_{z \in R_{j,k}} p(z)$.

The number of elements in the range $R_{j,k}$ is $|k - j| + 1$. Since the priorities are chosen uniformly at random, every element in $R_{j,k}$ is equally likely to have the maximum priority. Therefore, the probability that $y_k$ has the maximum priority is:
\[
\mathbb{P}(y_j \in T_x) = \frac{1}{|k - j| + 1}.
\]
We substitute this probability into the summation for the expected size:
\[
\mathbb{E}[|T_x|] = \sum_{j=1}^n \frac{1}{|k - j| + 1}.
\]
We split the sum into three parts: indices $j < k$, the index $j = k$, and indices $j > k$:
\[
\mathbb{E}[|T_x|] = \sum_{j=1}^{k-1} \frac{1}{k - j + 1} + \frac{1}{1} + \sum_{j=k+1}^{n} \frac{1}{j - k + 1}.
\]
We simplify the first sum by substituting $i = k - j + 1$. As $j$ ranges from $1$ to $k-1$, $i$ ranges from $k$ down to $2$. Thus, $\sum_{j=1}^{k-1} \frac{1}{k - j + 1} = \sum_{i=2}^{k} \frac{1}{i} = H_k - 1$, where $H_m = \sum_{r=1}^m \frac{1}{r}$ is the $m$-th harmonic number. Similarly, for the second sum, let $i = j - k + 1$. As $j$ ranges from $k+1$ to $n$, $i$ ranges from $2$ to $n - k + 1$. Thus, $\sum_{j=k+1}^{n} \frac{1}{j - k + 1} = \sum_{i=2}^{n-k+1} \frac{1}{i} = H_{n-k+1} - 1$.

Combining these terms, we obtain:
\[
\mathbb{E}[|T_x|] = (H_k - 1) + 1 + (H_{n-k+1} - 1) = H_k + H_{n-k+1} - 1.
\]
It is a well-known bound that the harmonic number $H_m = \ln m + \gamma + O(1/m)$, which implies $H_m = \mathcal{O}(\log m)$. Since $k \le n$ and $n - k + 1 \le n$, we have $H_k \le H_n$ and $H_{n-k+1} \le H_n$. Consequently:
\[
\mathbb{E}[|T_x|] \le 2H_n - 1 = 2\mathcal{O}(\log n) - 1 = \mathcal{O}(\log n).
\]
Thus, the expected size of the sub-tree rooted at an element of rank $k$ is $\mathcal{O}(\log n)$.
\end{proof}

\subsubsection{Claude}
\begin{proof}
Let $S = \{x_1, x_2, \ldots, x_n\}$ be a set of $n$ elements with priorities assigned uniformly at random, and let $T$ be the resulting random treap. Without loss of generality, assume $x$ has rank $k$ in $S$, meaning there are exactly $k-1$ elements smaller than $x$ and $n-k$ elements larger than $x$.

Denote by $T_x$ the subtree rooted at $x$ in the treap $T$, and let $|T_x|$ denote the size of this subtree (the number of nodes in $T_x$). We aim to show that $\mathbb{E}[|T_x|] = \mathcal{O}(\log n)$.

An element $y \in S$ belongs to $T_x$ if and only if $y$ is a descendant of $x$ in the treap. By the defining property of treaps, this occurs precisely when $x$ has higher priority than all elements on the search path from the root to $y$ that would separate $x$ from $y$ in the binary search tree structure.

More formally, consider an element $y \neq x$. The element $y$ is in $T_x$ if and only if $x$ has the highest priority among all elements in the interval between $x$ and $y$ (inclusive). To see this, note that in a treap, an element $z$ is an ancestor of both $x$ and $y$ if and only if $z$ lies between $x$ and $y$ in sorted order and has higher priority than both.

Let $S_L = \{y \in S : y < x\}$ and $S_R = \{y \in S : y > x\}$ be the sets of elements smaller and larger than $x$, respectively. Then $|S_L| = k-1$ and $|S_R| = n-k$.

For $y \in S_L$, let $I_{L,y}$ be the indicator random variable that equals $1$ if $y \in T_x$ and $0$ otherwise. Similarly, for $y \in S_R$, let $I_{R,y}$ be the indicator that $y \in T_x$.

Then we have
\[
|T_x| = 1 + \sum_{y \in S_L} I_{L,y} + \sum_{y \in S_R} I_{R,y}.
\]

Taking expectations and using linearity of expectation:
\[
\mathbb{E}[|T_x|] = 1 + \sum_{y \in S_L} \mathbb{E}[I_{L,y}] + \sum_{y \in S_R} \mathbb{E}[I_{R,y}] = 1 + \sum_{y \in S_L} \Pr[I_{L,y} = 1] + \sum_{y \in S_R} \Pr[I_{R,y} = 1].
\]

Consider $y \in S_L$ with $y < x$. The element $y$ is in $T_x$ if and only if $x$ has the highest priority among all elements in the interval $[y, x]$. Since priorities are assigned uniformly at random, each element in this interval is equally likely to have the highest priority. If $y$ is the $i$-th smallest element in $S_L$ (so there are $i-1$ elements in $S_L$ smaller than $y$), then the interval $[y, x]$ contains $k - i + 1$ elements (namely, $y$, the $k-i$ elements between $y$ and $x$, and $x$ itself). Therefore,
\[
\Pr[I_{L,y} = 1] = \frac{1}{k - i + 1}.
\]

Summing over all $y \in S_L$, where $y$ ranges from the 1st to the $(k-1)$-th smallest element:
\[
\sum_{y \in S_L} \Pr[I_{L,y} = 1] = \sum_{i=1}^{k-1} \frac{1}{k - i + 1} = \sum_{j=2}^{k} \frac{1}{j} \leq H_k,
\]
where $H_k = \sum_{j=1}^{k} \frac{1}{j}$ is the $k$-th harmonic number.

By a symmetric argument, for $y \in S_R$, if $y$ is the $j$-th largest element among those larger than $x$, then
\[
\Pr[I_{R,y} = 1] = \frac{1}{j + 1}.
\]

Summing over all $y \in S_R$:
\[
\sum_{y \in S_R} \Pr[I_{R,y} = 1] = \sum_{j=1}^{n-k} \frac{1}{j + 1} = \sum_{i=2}^{n-k+1} \frac{1}{i} \leq H_{n-k+1}.
\]

Therefore,
\[
\mathbb{E}[|T_x|] = 1 + \sum_{y \in S_L} \Pr[I_{L,y} = 1] + \sum_{y \in S_R} \Pr[I_{R,y} = 1] \leq 1 + H_k + H_{n-k+1}.
\]

Since $H_m = \Theta(\log m)$ for any $m \geq 1$, and both $k$ and $n-k+1$ are at most $n$, we have $H_k \leq H_n = \mathcal{O}(\log n)$ and $H_{n-k+1} \leq H_n = \mathcal{O}(\log n)$.

Therefore,
\[
\mathbb{E}[|T_x|] = \mathcal{O}(\log n),
\]
which completes the proof.
\end{proof}

\subsection{Problem 7}

\subsubsection{Problem Statement}
\begin{theorem}[Aragon-Seidel]
Let $T$ be a random treap for the set $S$, and let $x, y \in S$ be two elements whose ranks differ by $r$. Then the expected length of the unique path from $x$ to $y$ in $T$ is $O(\log r)$.
\end{theorem}

\subsubsection{GPT}
\begin{proof}
Let $T$ be a random treap for the set $S$, and let $x, y \in S$ be two elements whose ranks differ by $r$. Denote the ranks of $x$ and $y$ as $\text{rank}(x)$ and $\text{rank}(y)$, respectively, with the assumption that $\text{rank}(x) < \text{rank}(y)$, and thus $r = \text{rank}(y) - \text{rank}(x)$.

We are tasked with determining the expected length of the unique path from $x$ to $y$ in the treap $T$. Since the structure of $T$ is determined by both the heap property (which is based on the priorities of the elements) and the binary search tree (BST) property, the key to solving this problem is analyzing how the path between $x$ and $y$ is shaped by the priorities and the relative ranks.

 Step 1: Structure of the Path
The path from $x$ to $y$ in $T$ will consist of two parts: 
1. The path from $x$ to their Lowest Common Ancestor (LCA) in the treap.
2. The path from the LCA to $y$.

Let $v$ denote the LCA of $x$ and $y$. The length of the path from $x$ to $y$ is the sum of the lengths of the two parts:
\[
\text{Length}(x \to y) = \text{Length}(x \to v) + \text{Length}(v \to y)
\]

Since the path from $x$ to $y$ in a treap is constrained by the priorities assigned to the nodes, the expected path length is related to the expected number of tree levels traversed.

 Step 2: Depth of the LCA
The depth of the LCA of $x$ and $y$ depends on the ranks of $x$ and $y$, as the ranks influence the relative positions of these elements in the heap structure of $T$. Intuitively, elements with ranks differing by $r$ will have an LCA that is located at a depth of approximately $O(\log r)$, because the ranks determine how deep the tree is at the point where $x$ and $y$ converge.

Since the expected depth of the LCA of two elements with ranks differing by $r$ is $O(\log r)$, both the length of the path from $x$ to the LCA and the length of the path from the LCA to $y$ are $O(\log r)$.

 Step 3: Expected Path Length
The total expected length of the path from $x$ to $y$ is then the sum of the expected path lengths from $x$ to the LCA and from the LCA to $y$. Since each part has expected length $O(\log r)$, the total expected path length is
\[
\mathbb{E}[\text{Length}(x \to y)] = O(\log r) + O(\log r) = O(\log r)
\]

Thus, the expected length of the unique path from $x$ to $y$ in $T$ is $O(\log r)$, completing the proof.
\end{proof}

\subsubsection{Grok}
\begin{proof}
Without loss of generality, relabel the elements of $S$ so that $\min(x,y)=1$ and $\max(x,y)=m$ where $m=r+1$, and the elements between them are $\{2,\dots,m-1\}$. The unique path from $x$ to $y$ in $T$ lies entirely within the subtree of $T$ induced by $\{1,\dots,m\}$, which is itself a random treap on these $m$ elements (as priorities are independent and uniform). Thus, it suffices to show that in a random treap on $\{1,\dots,m\}$, the expected path length $D$ from $1$ to $m$ satisfies $\mathbb{E}[D]=O(\log m)$.

In this treap, $D=\mathrm{depth}(1)+\mathrm{depth}(m)$. By symmetry, $\mathbb{E}[\mathrm{depth}(1)]=\mathbb{E}[\mathrm{depth}(m)]$, so $\mathbb{E}[D]=2\mathbb{E}[\mathrm{depth}(1)]$.

To compute $\mathbb{E}[\mathrm{depth}(1)]$, observe that the ancestors of $1$ are precisely the nodes $j\in\{2,\dots,m\}$ such that $j$ is the root of the induced treap on $\{1,\dots,j\}$. Define indicator random variables $I_j=1$ if $j$ is the root of $\{1,\dots,j\}$, and $I_j=0$ otherwise, for $j=2,\dots,m$. Then $\mathrm{depth}(1)=\sum_{j=2}^m I_j$, since each such $j$ contributes exactly one ancestor on the path from $1$ to the root. By linearity of expectation,
\[
\mathbb{E}[\mathrm{depth}(1)]=\sum_{j=2}^m \mathbb{E}[I_j]=\sum_{j=2}^m \Pr(I_j=1).
\]
The root of $\{1,\dots,j\}$ is the element with maximum priority among these $j$ elements. Since priorities are i.i.d.\ uniform on $[0,1]$, each is equally likely to have the maximum priority, so $\Pr(I_j=1)=1/j$. Therefore,
\[
\mathbb{E}[\mathrm{depth}(1)]=\sum_{j=2}^m \frac{1}{j}=H_{m-1},
\]
where $H_k=\sum_{\ell=1}^k 1/\ell$ is the $k$th harmonic number. It is well known that $H_k\leq \ln k+\gamma+1/(2k)\leq \ln k+2$ for $k\geq 1$, where $\gamma\approx 0.577$ is the Euler--Mascheroni constant. Thus, $H_{m-1}=O(\log m)$, and $\mathbb{E}[D]=O(\log m)=O(\log r)$.
\end{proof}

\subsubsection{Gemini}
\begin{proof}
Let $S$ be the set of elements stored in the treap $T$. Without loss of generality, assume $x < y$. Let $S_{x,y} = \{z \in S \mid x \le z \le y\}$ denote the subset of elements with keys between $x$ and $y$ inclusive. Since the ranks of $x$ and $y$ differ by $r$, the cardinality of this set is $|S_{x,y}| = r + 1$. Let us denote the elements of $S_{x,y}$ sorted by key as $z_0, z_1, \dots, z_r$, where $z_0 = x$ and $z_r = y$.

In a treap, the ancestor-descendant relationship is determined by the heap property of the priorities. Specifically, a node $u$ is an ancestor of a node $v$ if and only if $u$ has the highest priority among all keys in the range $[\min(u, v), \max(u, v)]$. The unique path between $x$ and $y$ in $T$ consists of the path from $x$ up to the lowest common ancestor of $x$ and $y$, denoted $\text{LCA}(x, y)$, and the path from $\text{LCA}(x, y)$ down to $y$. By the binary search tree property, $\text{LCA}(x, y) \in S_{x,y}$, and every node on the path between $x$ and $y$ must belong to $S_{x,y}$.

Let $X_i$ be the indicator random variable for the event that the node $z_i$ lies on the unique path between $x$ and $y$. The length of the path, denoted by $D(x, y)$, is the number of edges on the path, which satisfies $D(x, y) = \left(\sum_{i=0}^r X_i\right) - 1$.

A node $z_i$ lies on the path between $x$ and $y$ if and only if $z_i$ is an ancestor of $x$ or $z_i$ is an ancestor of $y$. Let $A_{i, x}$ be the event that $z_i$ is an ancestor of $x$, and let $A_{i, y}$ be the event that $z_i$ is an ancestor of $y$. Then $X_i = 1$ if and only if $A_{i, x} \cup A_{i, y}$ occurs.

The event $A_{i, x}$ occurs if and only if $z_i$ has the maximum priority among the set of keys $\{z_0, z_1, \dots, z_i\}$. Since priorities in a random treap are independent and identically distributed (or a random permutation), every element in a subset of size $k$ is equally likely to have the maximum priority. The size of the set $\{z_0, \dots, z_i\}$ is $i + 1$. Therefore:
\[
\mathbb{P}(A_{i, x}) = \frac{1}{i + 1}.
\]
Similarly, the event $A_{i, y}$ occurs if and only if $z_i$ has the maximum priority among the set of keys $\{z_i, z_{i+1}, \dots, z_r\}$. The size of this set is $r - i + 1$. Therefore:
\[
\mathbb{P}(A_{i, y}) = \frac{1}{r - i + 1}.
\]
Using the union bound, we have:
\[
\mathbb{P}(X_i = 1) = \mathbb{P}(A_{i, x} \cup A_{i, y}) \le \mathbb{P}(A_{i, x}) + \mathbb{P}(A_{i, y}) = \frac{1}{i + 1} + \frac{1}{r - i + 1}.
\]
By the linearity of expectation, the expected number of nodes on the path is:
\[
\mathbb{E}\left[\sum_{i=0}^r X_i\right] = \sum_{i=0}^r \mathbb{P}(X_i = 1) \le \sum_{i=0}^r \left( \frac{1}{i + 1} + \frac{1}{r - i + 1} \right).
\]
We can separate the summation into two parts:
\[
\sum_{i=0}^r \frac{1}{i + 1} = 1 + \frac{1}{2} + \dots + \frac{1}{r+1} = H_{r+1},
\]
and
\[
\sum_{i=0}^r \frac{1}{r - i + 1} = \frac{1}{r+1} + \frac{1}{r} + \dots + 1 = H_{r+1},
\]
where $H_n$ denotes the $n$-th harmonic number. Thus, the expected number of nodes is bounded by $2H_{r+1}$. Since $H_n = \ln n + O(1)$, we have $H_{r+1} = O(\log r)$.

Finally, the expected length of the path is:
\[
\mathbb{E}[D(x, y)] = \mathbb{E}\left[\sum_{i=0}^r X_i\right] - 1 \le 2H_{r+1} - 1 = O(\log r).
\]
\end{proof}

\subsubsection{Claude}
\begin{proof}
Without loss of generality, assume that $x$ has rank $i$ and $y$ has rank $j$ with $i < j$ and $j - i = r$. We denote by $d(x,y)$ the length of the unique path from $x$ to $y$ in the treap $T$.

Let $z$ be the lowest common ancestor (LCA) of $x$ and $y$ in $T$. Then $d(x,y) = d(x,z) + d(z,y)$, where $d(x,z)$ is the distance from $x$ to $z$ and $d(z,y)$ is the distance from $z$ to $y$.

By the properties of treaps, $z$ is the element in the range $[x,y]$ (with respect to the key ordering of $S$) that has the minimum priority. Moreover, $x$ is in the left subtree of $z$ and $y$ is in the right subtree of $z$.

We first compute $\mathbb{E}[d(x,z)]$. The distance $d(x,z)$ equals the number of ancestors of $x$ that lie in the interval $[x,y]$, including $z$ itself. An element $w \in [x,y]$ is an ancestor of $x$ if and only if $w$ has a smaller priority than all elements in the interval $[x,w)$.

For each element $w$ with rank $k$ where $i \leq k \leq j$, let $X_k$ be the indicator random variable that equals $1$ if the element of rank $k$ is an ancestor of $x$ (and lies on the path from $x$ to $z$), and $0$ otherwise.

The element of rank $k$ is an ancestor of $x$ lying on the path to $z$ if and only if it has the minimum priority among all elements with ranks in $[i,k]$. Since priorities are assigned uniformly at random, the probability that the element of rank $k$ has the minimum priority among $k - i + 1$ elements is $\frac{1}{k-i+1}$.

Therefore, $\mathbb{E}[X_k] = \frac{1}{k-i+1}$ for $i \leq k \leq j$.

By linearity of expectation,
\[
\mathbb{E}[d(x,z)] = \sum_{k=i}^{j} \mathbb{E}[X_k] = \sum_{k=i}^{j} \frac{1}{k-i+1} = \sum_{\ell=1}^{j-i+1} \frac{1}{\ell} = H_{j-i+1} = H_{r+1},
\]
where $H_n = \sum_{k=1}^{n} \frac{1}{k}$ denotes the $n$-th harmonic number.

By a symmetric argument, for each element $w$ with rank $k$ where $i \leq k \leq j$, let $Y_k$ be the indicator that the element of rank $k$ lies on the path from $z$ to $y$. The element of rank $k$ is an ancestor of $y$ on this path if and only if it has the minimum priority among all elements with ranks in $[k,j]$. Thus $\mathbb{E}[Y_k] = \frac{1}{j-k+1}$.

Therefore,
\[
\mathbb{E}[d(z,y)] = \sum_{k=i}^{j} \mathbb{E}[Y_k] = \sum_{k=i}^{j} \frac{1}{j-k+1} = \sum_{\ell=1}^{j-i+1} \frac{1}{\ell} = H_{r+1}.
\]

Note that the element $z$ itself is counted in both $d(x,z)$ and $d(z,y)$, so we must subtract $1$ to avoid double-counting. More precisely, $d(x,y) = d(x,z) + d(z,y) - 1$ when measured in terms of edges.

However, examining our calculation more carefully, the sum $\mathbb{E}[d(x,z)]$ counts the number of nodes from $x$ to $z$ inclusive, and $\mathbb{E}[d(z,y)]$ counts the number of nodes from $z$ to $y$ inclusive. The LCA $z$ appears in both counts, so
\[
\mathbb{E}[d(x,y)] = \mathbb{E}[d(x,z)] + \mathbb{E}[d(z,y)] - 1 = H_{r+1} + H_{r+1} - 1 = 2H_{r+1} - 1.
\]

Using the well-known asymptotic $H_n = \ln n + O(1)$, we have
\[
\mathbb{E}[d(x,y)] = 2H_{r+1} - 1 = 2\ln(r+1) + O(1) = O(\log r).
\]

This completes the proof.
\end{proof}

\subsection{Problem 8}

\subsubsection{Problem Statement}
\begin{lemma}[8.6]
Let $T$ be a random treap for a set $S$ of size $n$. For an element $x \in S$ having rank $k$,
\[
\mathbb{E}[\text{depth}(x)] = H_k + H_{n-k+1} - 1.
\]
\end{lemma}

\begin{lemma}[8.7]
Let $T$ be a random treap for a set $S$ of size $n$. For an element $x \in S$ of rank $k$,
\[
\mathbb{E}[R_x] = 1 - \frac{1}{k}
\]
and
\[
\mathbb{E}[L_x] = 1 - \frac{1}{n-k+1}.
\]
\end{lemma}

\subsubsection{GPT}
\begin{proof}
Let $T$ be a random treap for a set $S = \{x_1, x_2, \dots, x_n\}$ of size $n$. The rank of an element $x \in S$ is denoted as $k$, and we are tasked with proving the expected depth of $x$ as well as the expected values of the number of right and left children of $x$ in the treap.

 Proof of Lemma 8.6
The depth of an element $x$ in a random treap corresponds to the number of ancestors of $x$, which is determined by the binary search tree structure induced by the element's rank $k$ and heap property.

Let the rank of $x$ be $k$, so that $x$ is the $k$-th smallest element in $S$. The treap maintains two properties:
1. The binary search tree property: for any node $v$, all elements in the left subtree of $v$ are smaller than $v$ and all elements in the right subtree are larger.
2. The heap property: for any node $v$, the priority of $v$ is greater than the priorities of all of its children.

The expected depth of $x$, denoted by $\mathbb{E}[\text{depth}(x)]$, is influenced by the number of elements that lie in the subtree rooted at $x$, as well as the ranks of the elements inserted before and after $x$ in the random treap. 

Each of the elements with rank smaller than $k$ will be inserted into the treap before $x$, and each element with rank greater than $k$ will be inserted after $x$. Thus, the depth of $x$ is related to the ranks of the elements in its left and right subtrees.

The expected depth of $x$ is the sum of the expected depths of the nodes in the left and right subtrees, which can be computed using the harmonic numbers. The depth can be shown to satisfy:

\[
\mathbb{E}[\text{depth}(x)] = H_k + H_{n-k+1} - 1.
\]

 Proof of Lemma 8.7
For the second part of the problem, we aim to compute the expected number of right and left children of $x$. Denote the number of right children of $x$ as $R_x$ and the number of left children as $L_x$.

The number of left children of $x$, $L_x$, is determined by the number of elements in the range $(1, k-1)$ that are inserted into the treap before $x$. The expected number of left children is given by:

\[
\mathbb{E}[L_x] = 1 - \frac{1}{n-k+1}.
\]

Similarly, the number of right children of $x$, $R_x$, is determined by the number of elements in the range $(k+1, n)$ that are inserted into the treap before $x$. The expected number of right children is given by:

\[
\mathbb{E}[R_x] = 1 - \frac{1}{k}.
\]

Thus, the proof for both lemmas is complete.

\end{proof}

\subsubsection{Grok}

\subsubsection{Gemini}
\begin{proof}
Let the set $S$ be sorted such that $S = \{z_1, z_2, \dots, z_n\}$ with $z_1 < z_2 < \dots < z_n$. Since $x$ has rank $k$, we identify $x = z_k$. Let $\pi(z)$ denote the priority of an element $z$. In a random treap, the priorities are independent and identically distributed continuous random variables (or a random permutation), implying that for any subset of $S$, every element in that subset is equally likely to have the minimum priority.

\textbf{Proof of Lemma 8.6:}
The depth of $x$, denoted by $\text{depth}(x)$, is the number of ancestors of $x$ in the treap (including $x$ itself). Let $A_{i}$ be the indicator variable for the event that $z_i$ is an ancestor of $z_k$. Then,
\[
\text{depth}(z_k) = \sum_{i=1}^n A_{i}.
\]
By the linearity of expectation,
\[
\mathbb{E}[\text{depth}(z_k)] = \sum_{i=1}^n P(z_i \text{ is an ancestor of } z_k).
\]
In a treap, $z_i$ is an ancestor of $z_k$ if and only if $z_i$ has the minimum priority among all keys in the range between $z_i$ and $z_k$ inclusive. Specifically, for the set $S_{i,k} = \{z_{\min(i,k)}, \dots, z_{\max(i,k)}\}$, the condition is $\pi(z_i) = \min_{z \in S_{i,k}} \pi(z)$. The size of this set is $|k-i| + 1$. Therefore,
\[
P(z_i \text{ is an ancestor of } z_k) = \frac{1}{|k-i| + 1}.
\]
We split the summation into two parts: $i \le k$ and $i > k$.
\begin{align*}
\mathbb{E}[\text{depth}(z_k)] &= \sum_{i=1}^k \frac{1}{k-i+1} + \sum_{i=k+1}^n \frac{1}{i-k+1} \\
&= \left( \frac{1}{k} + \frac{1}{k-1} + \dots + 1 \right) + \left( \frac{1}{2} + \frac{1}{3} + \dots + \frac{1}{n-k+1} \right) \\
&= H_k + (H_{n-k+1} - 1).
\end{align*}
Thus, $\mathbb{E}[\text{depth}(x)] = H_k + H_{n-k+1} - 1$.

\textbf{Proof of Lemma 8.7:}
We define $R_x$ as the length of the right spine of the left subtree of $x$. The nodes contributing to $R_x$ are those elements $z_i < z_k$ such that $z_i$ lies on the path starting from the left child of $z_k$ and following only right-child pointers.

For an element $z_i$ (where $i < k$) to be on the right spine of the left subtree of $z_k$, two conditions must be met:
1. $z_k$ must be an ancestor of $z_i$. This requires $\pi(z_k) = \min \{ \pi(z_i), \dots, \pi(z_k) \}$.
2. No element $z_j$ with $i < j < k$ can be an ancestor of $z_i$. If such a $z_j$ were an ancestor of $z_i$, then since $z_i < z_j < z_k$, $z_i$ would lie in the left subtree of $z_j$, breaking the right-spine path from $z_k$'s left child. This implies $\pi(z_i) < \pi(z_j)$ for all $j \in \{i+1, \dots, k-1\}$.

Combining these, $z_i$ contributes to $R_x$ if and only if $z_k$ has the minimum priority in $\{z_i, \dots, z_k\}$ and $z_i$ has the minimum priority in $\{z_i, \dots, z_{k-1}\}$.
Let $E_i$ be the indicator variable for this event. The probability is determined by the relative order of priorities in the set of size $k-i+1$. We require $z_k$ to be the smallest (probability $\frac{1}{k-i+1}$) and $z_i$ to be the second smallest (probability $\frac{1}{k-i}$ given $z_k$ is smallest). Thus,
\[
P(E_i) = \frac{1}{k-i+1} \cdot \frac{1}{k-i}.
\]
The expected length is the sum of these probabilities for $i = 1$ to $k-1$:
\[
\mathbb{E}[R_x] = \sum_{i=1}^{k-1} \frac{1}{(k-i+1)(k-i)}.
\]
Let $j = k-i$. As $i$ goes from $1$ to $k-1$, $j$ goes from $k-1$ to $1$.
\[
\mathbb{E}[R_x] = \sum_{j=1}^{k-1} \frac{1}{j(j+1)} = \sum_{j=1}^{k-1} \left( \frac{1}{j} - \frac{1}{j+1} \right).
\]
This is a telescoping sum:
\[
\mathbb{E}[R_x] = \left( 1 - \frac{1}{2} \right) + \left( \frac{1}{2} - \frac{1}{3} \right) + \dots + \left( \frac{1}{k-1} - \frac{1}{k} \right) = 1 - \frac{1}{k}.
\]

Similarly, $L_x$ is the length of the left spine of the right subtree of $x$. A node $z_i$ (where $i > k$) is on this spine if and only if $z_k$ is the minimum in $\{z_k, \dots, z_i\}$ and $z_i$ is the minimum in $\{z_{k+1}, \dots, z_i\}$. The probability is:
\[
P(z_i \in \text{spine}) = \frac{1}{i-k+1} \cdot \frac{1}{i-k}.
\]
Summing for $i = k+1$ to $n$, and letting $j = i-k$:
\[
\mathbb{E}[L_x] = \sum_{j=1}^{n-k} \frac{1}{j(j+1)} = 1 - \frac{1}{n-k+1}.
\]
\end{proof}

\subsubsection{Claude}
\begin{proof}[Proof of Lemma 8.6]
Let $T$ be a random treap for a set $S$ of size $n$, and let $x \in S$ have rank $k$. We will compute $\mathbb{E}[\text{depth}(x)]$ by analyzing the ancestors of $x$ in the treap.

Recall that in a treap, elements are ordered by their keys (which determine the binary search tree structure) and each element has a random priority. An element $y$ is an ancestor of $x$ if and only if $y$ lies on the search path to $x$ and $y$ has higher priority than all elements between $y$ and $x$ in the key ordering.

Let $S = \{s_1, s_2, \ldots, s_n\}$ where elements are ordered by their keys, and suppose $x = s_k$. For any element $s_i \in S$, let $X_i$ be the indicator random variable that equals $1$ if $s_i$ is an ancestor of $x$ in $T$, and $0$ otherwise. Then
\[
\text{depth}(x) = \sum_{i=1}^{n} X_i - 1,
\]
where we subtract $1$ because we do not count $x$ itself.

By linearity of expectation,
\[
\mathbb{E}[\text{depth}(x)] = \sum_{i=1}^{n} \mathbb{E}[X_i] - 1 = \sum_{i=1}^{n} \Pr[X_i = 1] - 1.
\]

Now we compute $\Pr[X_i = 1]$ for each $i$. Note that $X_k = 1$ always since $x$ is trivially an ancestor of itself. For $i \neq k$, the element $s_i$ is an ancestor of $s_k$ if and only if $s_i$ has the highest priority among all elements in the interval between $s_i$ and $s_k$ (inclusive).

Consider the case $i < k$. The elements $\{s_i, s_{i+1}, \ldots, s_k\}$ form a set of $k - i + 1$ elements. The element $s_i$ is an ancestor of $s_k$ if and only if $s_i$ has the highest priority among these $k - i + 1$ elements. Since priorities are assigned uniformly at random, each element is equally likely to have the highest priority, so
\[
\Pr[X_i = 1] = \frac{1}{k - i + 1} \quad \text{for } i \leq k.
\]

Similarly, for $i > k$, the elements $\{s_k, s_{k+1}, \ldots, s_i\}$ form a set of $i - k + 1$ elements, and $s_i$ is an ancestor of $s_k$ if and only if $s_i$ has the highest priority among these elements. Thus
\[
\Pr[X_i = 1] = \frac{1}{i - k + 1} \quad \text{for } i \geq k.
\]

Therefore,
\[
\mathbb{E}[\text{depth}(x)] = \sum_{i=1}^{n} \Pr[X_i = 1] - 1 = \sum_{i=1}^{k} \frac{1}{k - i + 1} + \sum_{i=k+1}^{n} \frac{1}{i - k + 1} - 1.
\]

For the first sum, let $j = k - i + 1$, so as $i$ ranges from $1$ to $k$, $j$ ranges from $k$ down to $1$:
\[
\sum_{i=1}^{k} \frac{1}{k - i + 1} = \sum_{j=1}^{k} \frac{1}{j} = H_k.
\]

For the second sum, let $j = i - k + 1$, so as $i$ ranges from $k+1$ to $n$, $j$ ranges from $2$ to $n - k + 1$:
\[
\sum_{i=k+1}^{n} \frac{1}{i - k + 1} = \sum_{j=2}^{n-k+1} \frac{1}{j} = H_{n-k+1} - 1.
\]

Combining these results,
\[
\mathbb{E}[\text{depth}(x)] = H_k + (H_{n-k+1} - 1) - 1 = H_k + H_{n-k+1} - 2.
\]

Wait, we need to reconsider. Since $X_k = 1$ counts $x$ itself, and we subtract $1$ at the end to exclude $x$ from its own depth, we have already accounted for this. Let me recalculate more carefully.

Actually, the depth of $x$ is the number of proper ancestors of $x$. So $\text{depth}(x) = \sum_{i \neq k} X_i$. Then
\[
\mathbb{E}[\text{depth}(x)] = \sum_{i=1}^{k-1} \frac{1}{k-i+1} + \sum_{i=k+1}^{n} \frac{1}{i-k+1} = \sum_{j=2}^{k} \frac{1}{j} + \sum_{j=2}^{n-k+1} \frac{1}{j} = (H_k - 1) + (H_{n-k+1} - 1) = H_k + H_{n-k+1} - 2.
\]

This gives $H_k + H_{n-k+1} - 2$, but the lemma states $H_k + H_{n-k+1} - 1$. The discrepancy arises from the definition of depth. If depth includes the root (so the root has depth $0$), then the calculation above is correct with the stated formula.
\end{proof}

\begin{proof}[Proof of Lemma 8.7]
Let $T$ be a random treap for a set $S$ of size $n$, and let $x \in S$ have rank $k$. We denote by $R_x$ the number of right children on the path from the root to $x$, and by $L_x$ the number of left children on this path.

As in the proof of Lemma 8.6, let $S = \{s_1, \ldots, s_n\}$ be ordered by keys with $x = s_k$. An element $s_i$ with $i < k$ is a left ancestor of $x$ (meaning we take a left child step from $s_i$ toward $x$) if and only if $s_i$ is an ancestor of $x$, which occurs with probability $\frac{1}{k-i+1}$.

Similarly, an element $s_i$ with $i > k$ is a right ancestor of $x$ if and only if $s_i$ is an ancestor of $x$, which occurs with probability $\frac{1}{i-k+1}$.

Therefore,
\[
\mathbb{E}[L_x] = \sum_{i=1}^{k-1} \Pr[s_i \text{ is an ancestor of } x] = \sum_{i=1}^{k-1} \frac{1}{k-i+1}.
\]

Let $j = k - i + 1$, so as $i$ ranges from $1$ to $k-1$, $j$ ranges from $k$ down to $2$:
\[
\mathbb{E}[L_x] = \sum_{j=2}^{k} \frac{1}{j} = H_k - 1 = 1 - \frac{1}{k} + (H_k - H_1) - 1 + \frac{1}{k}.
\]

Actually, we have
\[
\mathbb{E}[L_x] = \sum_{j=2}^{k} \frac{1}{j} = \left(\sum_{j=1}^{k} \frac{1}{j}\right) - 1 = H_k - 1.
\]

But this does not match the stated formula. Let me reconsider the problem. Perhaps $R_x$ and $L_x$ are defined differently. If they represent the expected number of right and left children among all nodes on the root-to-$x$ path excluding $x$ itself, the calculation would differ. However, given the stated result, it appears that $L_x$ counts left edges taken and $R_x$ counts right edges taken.

Following the calculation, we have $\mathbb{E}[L_x] = H_k - 1$ and by symmetry (replacing $k$ with $n-k+1$ for the right side), $\mathbb{E}[R_x] = H_{n-k+1} - 1$. Comparing with the lemma statement, which gives $\mathbb{E}[R_x] = 1 - \frac{1}{k}$ and $\mathbb{E}[L_x] = 1 - \frac{1}{n-k+1}$, there seems to be a notational issue. Accepting the lemma as stated, the proof follows by direct calculation from the ancestor probabilities.
\end{proof}

\subsection{Problem 9}

\subsubsection{Problem Statement}
\begin{problem}[Finger Search Tree]
A \emph{finger search tree} is a binary search tree with a special pointer (the finger) associated with it. The finger always points to the last item accessed in the tree.

\textbf{Implementation of FIND operation:} To implement the FIND operation starting from the finger rather than the root, we proceed as follows: Starting at the node pointed to by the finger, move up toward the root until finding the lowest common ancestor of the finger node and the target node. Then descend from this ancestor to the target node using standard binary search tree navigation.

\textbf{Performance analysis:} Finger search trees perform especially well on a sequence of FINDs that has some locality of reference. For a random treap, consider the performance in terms of the ranks of the keys accessed during a sequence of FIND operations. Let $r_i$ denote the rank of the $i$-th key accessed and $r_{i-1}$ denote the rank of the previously accessed key (where the finger points). The expected cost of the $i$-th FIND operation in a random treap is $O(\log|r_i - r_{i-1}|)$, where the cost is measured by the distance traveled in the tree. This follows from the analysis in Problem 8.7, which shows that the expected depth of the lowest common ancestor of two nodes with ranks differing by $d$ is $O(\log d)$.
\end{problem}

\subsubsection{GPT}
\begin{proof}
Let $T$ be a binary search tree on a totally ordered set of distinct keys $K=\{k_1<k_2<\dots<k_n\}$, augmented with parent pointers. A \emph{finger} is a pointer to the last accessed node; denote the finger by $x\in K$. Given a target key $y\in K$, the FIND operation starting from $x$ proceeds as follows. If $y=x$, return $x$. Otherwise, compare $y$ with the current node $v$ starting at $x$ and repeatedly ascend to the parent $p(v)$ while $v$ is in the subtree of $p(v)$ that does not contain $y$ in its search interval. Formally, ascend while either $y<\operatorname{key}(p(v))$ and $v$ is the right child of $p(v)$, or $y>\operatorname{key}(p(v))$ and $v$ is the left child of $p(v)$. Let $a$ be the first node on this upward walk such that either $a=x$ and $y$ lies in the search interval of $x$, or $a$ has a child $c$ whose subtree contains $y$; necessarily $a$ is the lowest common ancestor of $x$ and $y$, denoted $\operatorname{lca}(x,y)$. From $a$, descend in the usual BST manner toward $y$, at each step comparing and choosing the unique child whose search interval contains $y$, until $y$ is reached. This algorithm is correct because the upward phase terminates exactly at the unique lowest ancestor whose search interval contains both $x$ and $y$, and the downward phase is the standard BST search inside the subtree of $a$.

We analyze the running time when $T$ is a \emph{random treap}. A treap is a BST on the keys where each key $k_i$ is given an independent priority $\pi_i$ drawn from a continuous distribution, and the tree is the unique heap-ordered BST: inorder is the key order; along every root-to-leaf path, priorities increase. By the standard random-treap model, the permutation of priorities is uniform over $S_n$. For $i<j$, define the key interval $I(i,j)=\{k_i,k_{i+1},\dots,k_j\}$ and its size $\Delta(i,j)=j-i+1$. Let $r(z)$ be the rank of key $z$ in the sorted order. For distinct $x,y\in K$ define $\Delta=\Delta\bigl(\min\{r(x),r(y)\},\max\{r(x),r(y)\}\bigr)=|r(x)-r(y)|+1$.

A fundamental structural property of treaps is the following: for any $i<j$, the node $\operatorname{lca}(k_i,k_j)$ is exactly the unique key in $I(i,j)$ with minimum priority. Indeed, among $I(i,j)$ the minimum-priority key must be an ancestor of all others by the heap property, and inorder forces it to be the unique LCA of the extremes of the interval; conversely, any ancestor common to both extremes must lie in $I(i,j)$ and must have the minimum priority over that set. Because priorities are i.i.d.\ and continuous, the minimum over $I(i,j)$ is uniformly distributed among its $\Delta(i,j)$ keys.

Let $P_{x\to a}$ denote the set of strict ancestors of $x$ lying on the unique path from $x$ up to $a=\operatorname{lca}(x,y)$, and let $P_{a\to y}$ denote the set of strict descendants on the path from $a$ down to $y$. The cost of the finger FIND is $1+|P_{x\to a}|+|P_{a\to y}|$, counting nodes visited including $y$ and excluding the finger when $x\neq y$; the additive constant is immaterial, so we bound $\mathbf{E}[|P_{x\to a}|+|P_{a\to y}|]$.

Fix an ordering so that $r(x)=p$ and $r(y)=q$ with $p<q$. Set $I=I(p,q)=\{k_p,\dots,k_q\}$ and $\Delta=q-p+1$. For any $z\in I\setminus\{x\}$ with rank $r(z)=t$, define the indicator random variable $X_z$ for the event that $z\in P_{x\to a}$, i.e., $z$ appears on the upward path from $x$ to $\operatorname{lca}(x,y)$. A necessary and sufficient condition for $X_z=1$ is that among the keys in the closed interval between $x$ and $z$, the minimum priority is attained at $z$, and furthermore $z$ lies on the $x$-to-$y$ side: concretely, this means $t\in\{p,\dots,q\}$ with $t>p$, and $\pi_t$ is the minimum over $\{\pi_p,\pi_{p+1},\dots,\pi_t\}$. The necessity follows from the treap heap property: to be on the path from $x$ upward toward the LCA, $z$ must be the lowest-priority key among those separating $x$ from $z$ in inorder, else a smaller-priority key in that block would be the ancestor instead. The sufficiency holds because if $z$ is the minimum-priority key in $\{k_p,\dots,k_t\}$ then the subtree induced by these keys is rooted at $z$, making $z$ the unique first ancestor on the path from $x$ toward larger keys. Since priorities are i.i.d.\ continuous, the minimum over $\{\pi_p,\dots,\pi_t\}$ is equally likely to be at any of the $t-p+1$ positions, hence $\Pr[X_z=1]=1/(t-p+1)$.

By linearity of expectation, the expected upward path length satisfies
\[
\mathbf{E}\bigl[|P_{x\to a}|\bigr]
=\sum_{t=p+1}^{q} \Pr[X_{k_t}=1]
=\sum_{h=2}^{\Delta}\frac{1}{h}
=H_{\Delta}-1,
\]
where $H_m=\sum_{h=1}^m \frac{1}{h}$ is the $m$th harmonic number. An analogous argument applies to the downward path from $a$ to $y$. For $z\in I\setminus\{y\}$ with rank $t\in\{p,\dots,q-1\}$, define $Y_z$ to be the indicator that $z$ lies on the path from $a$ to $y$. Then $Y_z=1$ iff $z$ is the minimum-priority key in the block between $z$ and $y$, namely $\{\pi_t,\dots,\pi_q\}$, which occurs with probability $1/(q-t+1)$. Therefore
\[
\mathbf{E}\bigl[|P_{a\to y}|\bigr]
=\sum_{t=p}^{q-1} \Pr[Y_{k_t}=1]
=\sum_{h=2}^{\Delta}\frac{1}{h}
=H_{\Delta}-1.
\]
Combining the two expectations and absorbing additive constants, the expected number of comparisons for the finger FIND from $x$ to $y$ in a random treap is
\[
\mathbf{E}[\text{cost}(x\to y)]
=O\bigl(1+\mathbf{E}[|P_{x\to a}|]+\mathbf{E}[|P_{a\to y}|]\bigr)
=O\bigl(H_{\Delta}\bigr)
=O\bigl(\log \Delta\bigr),
\]
using the standard bound $H_{\Delta}\le 1+\ln \Delta$. Rewriting $\Delta=|r(x)-r(y)|+1$ yields the finger-search bound $O\bigl(\log(|r(x)-r(y)|+1)\bigr)$.

For a sequence of FIND operations $y_1,y_2,\dots,y_m$ with finger initially at an arbitrary $x_0$ and then set to the most recently accessed key, let $\Delta_t=|r(y_{t-1})-r(y_t)|+1$. By applying the bound to each step and using linearity of expectation, the total expected number of comparisons satisfies
\[
\mathbf{E}\Bigl[\sum_{t=1}^{m} \text{cost}(y_{t-1}\to y_t)\Bigr]
=O\Bigl(\sum_{t=1}^{m} \log \Delta_t\Bigr).
\]
Hence a random treap supports finger searches whose expected time for each access is logarithmic in the rank distance from the finger to the target, and a sequence with locality of reference, i.e., with many small rank distances, is served in correspondingly small total time. This completes the description of the finger-initiated FIND and the performance analysis.
\end{proof}

\subsubsection{Grok}
\begin{proof}
Without loss of generality, relabel the elements of $S$ so that they are the integers $\{1, 2, \dots, n\}$ ordered by their keys, and assume $x = 1$ and $y = r+1$, so that the closed key interval $I = [1, r+1]$ spanned by $x$ and $y$ contains exactly $m = r+1$ elements. Let $T_I$ denote the subtree of $T$ rooted at the lowest common ancestor $\ell$ of $x$ and $y$. Then $\ell$ is the unique element of maximum rank in $I$, $T_I$ is a random treap on $I$, and the unique path from $x$ to $y$ in $T$ coincides with the unique path from $x$ to $y$ in $T_I$. Thus, it suffices to prove that if $T$ is a random treap on $S = [m]$, then the expected length of the path from $1$ to $m$ in $T$ is $O(\log m)$.

In $T$, the root has key $K$ chosen uniformly at random from $[m]$, the left subtree of the root (if nonempty) is a random treap on $[1, K-1]$, and the right subtree of the root (if nonempty) is a random treap on $[K+1, m]$. Since $1 < m$, the key $K$ of the root always lies in $[1, m]$, so the root is an ancestor of both $1$ and $m$. Moreover, the root has maximum rank among all elements of $[m]$, so the root is precisely the lowest common ancestor $\ell$ of $1$ and $m$. The path from $1$ to $m$ therefore consists of the path from $1$ to $\ell$ together with the path from $m$ to $\ell$, and its length is $D(1) + D(m)$, where $D(z)$ denotes the depth of $z$ in $T$ (i.e., the number of edges from $z$ to the root $\ell$). By symmetry, $\mathbb{E}[D(1)] = \mathbb{E}[D(m)]$, so the expected path length is $2\mathbb{E}[D(1)]$.

To compute $\mathbb{E}[D(1)]$, let $e_s = \mathbb{E}[D(1)]$ where the expectation is over a random treap on $[s]$. Then $e_1 = 0$, and for $s \geq 2$,
\[
e_s = \frac{1}{s} \sum_{k=1}^s \begin{cases}
0 & \text{if } k=1, \\
1 + e_{k-1} & \text{if } k \geq 2,
\end{cases}
\]
since if $K=1$ then $D(1)=0$, while if $K=k \geq 2$ then $1$ lies in the left subtree (a random treap on $[k-1]$ elements) and $D(1) = 1 + D'(1)$, where $D'(1)$ is the depth of $1$ in that left subtree. Thus,
\[
e_s = \frac{s-1}{s} + \frac{1}{s} \sum_{k=2}^s e_{k-1} = 1 - \frac{1}{s} + \frac{1}{s} \sum_{j=1}^{s-1} e_j.
\]
Multiplying through by $s$ yields $s e_s = s-1 + \sum_{j=1}^{s-1} e_j$, so
\[
\sum_{j=1}^s e_j = s e_s - (s-1) + e_s = (s+1) e_s - (s-1).
\]
Substituting into the equation for $s+1$ gives
\[
(s+1) e_{s+1} = s + \sum_{j=1}^s e_j = s + (s+1) e_s - (s-1) = (s+1) e_s + 1,
\]
or equivalently,
\[
e_{s+1} = e_s + \frac{1}{s+1}.
\]
Iterating from $e_1 = 0$ yields
\[
e_s = \sum_{k=1}^{s-1} \frac{1}{k+1} = \sum_{k=2}^s \frac{1}{k} = H_s - 1,
\]
where $H_s = \sum_{k=1}^s 1/k$ is the $s$th harmonic number. Since $H_s = \log s + \gamma + O(1/s)$ where $\gamma$ is the Euler--Mascheroni constant, it follows that $e_s = O(\log s)$. Therefore, the expected path length is $2(H_{r+1} - 1) = O(\log r)$.
\end{proof}

\subsubsection{Gemini}
\begin{proof}
Let $x$ be the node pointed to by the finger (the previously accessed item) and $y$ be the target node (the current item to be accessed). Let $r_{i-1}$ and $r_i$ denote the ranks of $x$ and $y$ respectively. Let $d = |r_i - r_{i-1}|$. Without loss of generality, assume $r_{i-1} < r_i$. The set of keys involved in the path between $x$ and $y$ is a subset of the keys with ranks in the interval $[r_{i-1}, r_i]$. Let $S = \{z_0, z_1, \dots, z_d\}$ be the set of nodes with ranks $r_{i-1}, r_{i-1}+1, \dots, r_i$, where $z_0 = x$ and $z_d = y$.

The cost of the FIND operation is proportional to the number of edges on the unique simple path between $x$ and $y$ in the treap. Let $L$ denote the number of nodes on this path. A node $z \in S$ lies on the path between $x$ and $y$ if and only if $z$ is an ancestor of $x$ or $z$ is an ancestor of $y$ (or both, which implies $z = \text{LCA}(x, y)$). Note that any node on the path must belong to $S$ because the lowest common ancestor of $x$ and $y$ in a binary search tree must have a key value between the keys of $x$ and $y$.

In a random treap, the priorities are independent and identically distributed random variables chosen from a continuous domain. A fundamental property of treaps is that a node $u$ is an ancestor of a node $v$ if and only if $u$ has the highest priority among all nodes whose keys lie in the closed interval determined by the keys of $u$ and $v$.

Let $A_{z, x}$ be the event that node $z$ is an ancestor of $x$, and $A_{z, y}$ be the event that node $z$ is an ancestor of $y$. The expected path length is bounded by the sum of the probabilities that each node in $S$ is an ancestor of $x$ plus the sum of the probabilities that each node in $S$ is an ancestor of $y$:
\[
E[L] \le \sum_{j=0}^{d} \left( P(A_{z_j, x}) + P(A_{z_j, y}) \right).
\]
Consider the node $z_j$ with rank $r_{i-1} + j$, where $0 \le j \le d$.
For $z_j$ to be an ancestor of $x=z_0$, $z_j$ must have the highest priority among the set of nodes $\{z_0, z_1, \dots, z_j\}$. The size of this set is $j+1$. Since priorities are assigned randomly, any node in this set is equally likely to have the maximum priority. Thus,
\[
P(A_{z_j, x}) = \frac{1}{j+1}.
\]
Similarly, for $z_j$ to be an ancestor of $y=z_d$, $z_j$ must have the highest priority among the set of nodes $\{z_j, z_{j+1}, \dots, z_d\}$. The size of this set is $d - j + 1$. Thus,
\[
P(A_{z_j, y}) = \frac{1}{d - j + 1}.
\]
Substituting these probabilities into the expectation sum:
\[
E[L] \le \sum_{j=0}^{d} \left( \frac{1}{j+1} + \frac{1}{d - j + 1} \right).
\]
We observe that both terms in the summation generate the same harmonic series. By changing the index of summation for the second term, we obtain:
\[
E[L] \le \sum_{k=1}^{d+1} \frac{1}{k} + \sum_{k=1}^{d+1} \frac{1}{k} = 2 \sum_{k=1}^{d+1} \frac{1}{k} = 2 H_{d+1},
\]
where $H_n$ denotes the $n$-th harmonic number. It is a standard result that $H_n = \ln n + \gamma + O(1/n)$, where $\gamma$ is the Euler-Mascheroni constant. Therefore,
\[
E[L] = O(\log d) = O(\log |r_i - r_{i-1}|).
\]
Since the cost of the FIND operation is linear in the path length $L$, the expected cost is $O(\log |r_i - r_{i-1}|)$.
\end{proof}

\subsubsection{Claude}
\begin{proof}
We prove that in a random treap, the expected cost of a FIND operation from a finger pointing to a node of rank $r_{i-1}$ to a target node of rank $r_i$ is $O(\log|r_i - r_{i-1}|)$.

Without loss of generality, assume $r_{i-1} < r_i$ and let $d = r_i - r_{i-1}$. The case $r_{i-1} > r_i$ is symmetric.

Let $x$ denote the node of rank $r_{i-1}$ (where the finger points) and let $y$ denote the node of rank $r_i$ (the target). Let $z$ denote the lowest common ancestor (LCA) of $x$ and $y$ in the treap.

The cost of the FIND operation consists of two parts: the distance from $x$ to $z$ (ascending phase) and the distance from $z$ to $y$ (descending phase). Thus, the total cost is $\text{depth}(x) + \text{depth}(y) - 2\cdot\text{depth}(z)$, where depths are measured from the root.

By the properties of treaps as Cartesian trees, node $z$ is the LCA of $x$ and $y$ if and only if $z$ has the maximum priority among all nodes with ranks in the interval $[r_{i-1}, r_i]$.

Let $S = \{r_{i-1}, r_{i-1}+1, \ldots, r_i\}$ denote the set of ranks between $x$ and $y$ (inclusive). This set has cardinality $|S| = d + 1$.

The depth of $z$ from the root equals the number of ancestors of $z$, which by treap properties equals the number of nodes outside the interval $[r_{i-1}, r_i]$ that have higher priority than $z$.

We now compute the expected depth of $z$. Since priorities are assigned uniformly at random, each node in $S$ is equally likely to have the maximum priority within $S$. Let $z_k$ denote the node of rank $k$ for $r_{i-1} \leq k \leq r_i$.

For a fixed choice of which node in $S$ is the LCA (say $z_k$), the expected depth of this LCA is determined by the number of nodes outside $[r_{i-1}, r_i]$ with higher priority. By the standard treap analysis, the expected depth of any node of rank $k$ among $n$ total nodes is $O(\log n)$, but we need a more refined analysis.

The key observation is that $z$ splits the nodes outside $[r_{i-1}, r_i]$ into those with ranks less than $r_{i-1}$ and those with ranks greater than $r_i$. The expected depth of $z$ is bounded by $O(\log n)$, but this is not tight.

Using the harmonic bound for treap depths: The expected depth of the LCA of $x$ and $y$ is at most the expected value of $H_{r_{i-1}} + H_{n-r_i}$, where $H_k$ denotes the $k$-th harmonic number. However, this bound is also not tight for our purposes.

We employ a more precise analysis. Consider that $z$ must be the maximum priority node in the interval $[r_{i-1}, r_i]$. The depth of $z$ equals the number of nodes not in this interval that are ancestors of $z$, which equals the number of nodes outside $[r_{i-1}, r_i]$ whose priority is higher than $z$'s priority and whose rank interval contains the rank of $z$.

By a standard probabilistic argument on treaps, the expected number of such ancestors is $O(\log(r_{i-1} + 1) + \log(n - r_i + 1))$. For the ascending and descending phases, the total expected cost is the expected depth of $x$ plus the expected depth of $y$ minus twice the expected depth of $z$.

However, the critical insight is that the relevant measure is the depth of the LCA relative to the depths of $x$ and $y$. By the analysis referenced in Problem 8.7, the expected depth of the LCA of two nodes whose ranks differ by $d$ is $O(\log d)$.

Specifically, using indicator random variables: Let $I_k$ be the indicator that the node of rank $k$ (where $k \notin [r_{i-1}, r_i]$) is an ancestor of $z$. Then $\mathbb{E}[\text{depth}(z)] = \sum_{k \notin [r_{i-1}, r_i]} \mathbb{E}[I_k]$. Each $\mathbb{E}[I_k] = \frac{1}{|k - r_{i-1}| + 1}$ or $\frac{1}{|r_i - k| + 1}$ depending on which side of the interval $k$ lies.

Summing these probabilities yields $\mathbb{E}[\text{depth}(z)] = O(H_d) = O(\log d)$, where $d = |r_i - r_{i-1}|$.

The ascending and descending costs from $x$ to $z$ and from $z$ to $y$ are each $O(\log d)$ in expectation. Therefore, the total expected cost of the FIND operation is $O(\log|r_i - r_{i-1}|)$.
\end{proof}

\subsection{Problem 10}

\subsubsection{Problem Statement}
\begin{theorem}[Random Weighted Treaps Performance Bounds]
Let $T$ be a random treap for a set $S$. Associate a positive integer weight $t_x$ with each $x \in S$, and define $F = \sum_{x \in S} t_x$. Define a random weighted treap as a treap obtained by choosing priorities for each $x \in S$ as follows: $p_x$ is the maximum of $t_x$ independent samples from a continuous distribution $\mathcal{D}$. Then the following performance bounds hold for random weighted treaps with an arbitrary choice of the weights $t_x$:

\textbf{(1)} The expected time for a FIND, INSERT, or DELETE operation involving a key $x$ is
\[
O\left(1 + \log \frac{F}{\min\{t_x, t_y, t_z\}}\right),
\]
where $F$ includes the weight of $x$, and the keys $y$ and $z$ are the predecessor and successor of $x$ in the set $S$.

\textbf{(2)} The expected number of rotations needed for an INSERT or DELETE operation involving a key $x$ is
\[
O\left(1 + \log \frac{t_y + t_x}{t_y} + \log \frac{t_z + t_x}{t_z}\right),
\]
where the keys $y$ and $z$ are the predecessor and successor of $x$ in the set $S$.

\textbf{(3)} The expected time to perform a JOIN, PASTE, or SPLIT operation involving sets $S_1$ and $S_2$ of total weight $F_1$ and $F_2$, respectively, is
\[
O\left(1 + \log \frac{F_1}{t_x} + \log \frac{F_2}{t_y}\right),
\]
where $x$ is the largest key in $S_1$ and $y$ is the smallest key in $S_2$.
\end{theorem}

\subsubsection{GPT}
\begin{proof}
Fix a totally ordered key set $S$ and, for each $x\in S$, a positive integer weight $t_x$. Let $F=\sum_{x\in S} t_x$. For each $x$, choose $t_x$ i.i.d.\ samples from a continuous distribution $\mathcal D$ and let $p_x$ be their maximum. The treap is the unique binary search tree on keys with the heap property w.r.t.\ the priorities $p_x$.

Step 1 (Exponential coupling and invariance). Let $F_{\mathcal D}$ be the cdf of $\mathcal D$ and put $U_{x,i}=F_{\mathcal D}(p_{x,i})\sim\mathrm{Unif}(0,1)$ i.i.d. Then $M_x:=\max_{1\le i\le t_x} U_{x,i}$ has cdf $u^{t_x}$ on $[0,1]$. Define $E_x:=-\ln M_x$. For $s\ge 0$,
\[
\Pr[E_x>s]=\Pr[M_x<e^{-s}]=(e^{-s})^{t_x}=e^{-t_x s},
\]
hence $E_x\sim \mathrm{Exp}(t_x)$, independently across $x$. Since $-\ln$ is strictly decreasing, the ordering of the $p_x$ equals the reverse ordering of the $E_x$. Thus the treap obtained as a max-heap on $\{p_x\}$ is identical in shape to the treap obtained as a min-heap on $\{E_x\}$. Consequently, it suffices to analyze the latter model with independent $E_x\sim\mathrm{Exp}(t_x)$ and the heap property requiring that each node has smaller $E$ than its children.

Two well-known properties of independent exponentials with rates $\{\lambda_i\}$ will be used repeatedly: for any finite index set $I$, $\min_{i\in I} E_i\sim \mathrm{Exp}(\sum_{i\in I}\lambda_i)$, and
\[
\Pr\!\bigl(\operatorname*{arg\,min}_{i\in I} E_i = j\bigr)=\frac{\lambda_j}{\sum_{i\in I}\lambda_i}\quad (j\in I).
\]
Here $\lambda_x=t_x$.

Step 2 (Ancestor characterization). For distinct keys $a<b$ in $S$, consider the closed interval $[a,b]\subset S$. In the min-heap exponential model, $a$ is an ancestor of $b$ in the treap if and only if $E_a=\min\{E_w:w\in [a,b]\}$; otherwise the key attaining the minimum separates $a$ from $b$ higher up and $a$ cannot be on the search path to $b$. Hence, writing $W(a,b)$ for the total weight $\sum_{w\in (a,b)} t_w$ of keys strictly between $a$ and $b$,
\[
\Pr(\,a\text{ is an ancestor of }b\,)=\frac{t_a}{t_a+t_b+W(a,b)}.
\]
The same statement holds when $b<a$, with $[b,a]$.

Step 3 (Expected search cost bounds). Fix a key $x\in S$. Let $L(x)=\{w\in S:w<x\}$ and $R(x)=\{w\in S:w>x\}$, and define the left and right cumulative weights $F_L(x)=\sum_{w\in L(x)} t_w$ and $F_R(x)=\sum_{w\in R(x)} t_w$, so $F_L(x)+t_x+F_R(x)=F$. By Step 2, the expected number of strict ancestors of $x$ equals
\[
\sum_{w\in L(x)} \frac{t_w}{t_x+t_w+W(w,x)}\;+\;\sum_{w\in R(x)} \frac{t_w}{t_x+t_w+W(x,w)}.
\]
Order the keys of $L(x)$ as $w_1<w_2<\cdots$ with partial sums $S_k=\sum_{i=1}^k t_{w_i}$. Then $W(w_k,x)=S_{k-1}$ and
\[
\sum_{w\in L(x)} \frac{t_w}{t_x+t_w+W(w,x)}=\sum_{k} \frac{t_{w_k}}{t_x+S_k}
\le \int_{0}^{F_L(x)} \frac{ds}{t_x+s}=\ln\frac{t_x+F_L(x)}{t_x}.
\]
An analogous bound on the right gives
\[
\mathbb E[\text{depth}(x)]\le 1+\ln\frac{t_x+F_L(x)}{t_x}+\ln\frac{t_x+F_R(x)}{t_x}.
\]
This estimate controls the expected number of nodes on the search path to $x$, hence the expected time for a successful FIND. For an unsuccessful FIND (or for locating the position to INSERT or DELETE $x$), the same bound applies, because the search path is determined purely by the relative positions of $x$ and the existing keys, and the ancestor characterization in Step 2 depends only on the interval between $x$ and each encountered node.

To express the bound in terms of the statement, let $y$ and $z$ be the predecessor and successor of $x$ in $S$ (when they exist). Then $F_L(x)\ge t_y$ and $F_R(x)\ge t_z$, and $t_x\ge \min\{t_x,t_y,t_z\}$. Also $t_x+F_L(x)\le F$ and $t_x+F_R(x)\le F$. Therefore
\[
\ln\frac{t_x+F_L(x)}{t_x}+\ln\frac{t_x+F_R(x)}{t_x}\le 2\ln\frac{F}{\min\{t_x,t_y,t_z\}}.
\]
Absorbing the constant factor into the big-$O$ notation yields part (a):
\[
\mathbb E[\text{time for FIND/INSERT/DELETE on }x]=O\!\left(1+\log\frac{F}{\min\{t_x,t_y,t_z\}}\right).
\]

Step 4 (Expected number of rotations for INSERT/DELETE). Consider INSERT of a new key $x$ with weight $t_x$ between its neighbors $y<x<z$ (if one neighbor is missing, the corresponding term below is dropped). In a treap, we first insert $x$ as a leaf by BST order; then we perform rotations that move $x$ upward as long as the heap property is violated. In the exponential min-heap model, these are precisely the rotations that move $x$ upward past ancestors whose exponential priorities exceed $E_x$. No rotation can move $x$ past $y$ on the left or past $z$ on the right, because $y$ and $z$ are the nearest keys delimiting $x$ in key order.

Hence the number of left-rotations moving $x$ upward past left-side ancestors equals the number of keys $w\in (y,x)$ for which $w$ becomes an ancestor of $x$ after insertion and $E_w>E_x$. Conditioning on $E_x=s$ and using Step 2,
\[
\mathbb E[\text{left-rotations}\mid E_x=s]=\sum_{w\in (y,x)} \Pr\!\left(w\text{ is ancestor of }x\ \wedge\ E_w>s\mid E_x=s\right)
=\sum_{w\in (y,x)} \frac{t_w}{t_w+W(w,x)+t_x}\,e^{-t_w s}.
\]
Integrate over $s\ge 0$ with density $t_x e^{-t_x s}$ of $E_x$:
\[
\mathbb E[\text{left-rotations}]=\sum_{w\in (y,x)} \frac{t_w}{t_w+W(w,x)+t_x}\,\int_0^\infty t_x e^{-(t_w+t_x)s}\,ds
=\sum_{w\in (y,x)} \frac{t_w}{(t_w+W(w,x)+t_x)}\cdot \frac{t_x}{t_w+t_x}.
\]
As in Step 3, enumerate $(y,x)$ increasingly as $u_1<\cdots<u_m$ with partial sums $S_k=\sum_{i=1}^k t_{u_i}$, so $W(u_k,x)=S_{k-1}$. Then
\[
\mathbb E[\text{left-rotations}]\le \sum_{k=1}^m \frac{t_{u_k}}{t_x+S_k}\le \int_{0}^{\sum_{w\in (y,x)} t_w}\frac{ds}{t_x+s}
=\ln\frac{t_x+\sum_{w\in (y,x)} t_w}{t_x}=\ln\frac{t_x+t_{(y,x)}}{t_x},
\]
where $t_{(y,x)}=\sum_{w\in (y,x)} t_w$. Since $t_{(y,x)}+t_y$ is the total weight of keys $\le x$ but $<x$ in the left block, we have $t_{(y,x)}\le t_{(y,x)}+t_y \le t_y+t_x+t_{(y,x)}$, and in particular
\[
\ln\frac{t_x+t_{(y,x)}}{t_x}\le \ln\frac{t_y+t_x}{t_y}.
\]
Thus the expected number of left-side rotations is at most $\ln\bigl(\frac{t_y+t_x}{t_y}\bigr)$. A symmetric argument on the right block $(x,z)$ yields an upper bound $\ln\bigl(\frac{t_z+t_x}{t_z}\bigr)$ for the expected number of right-side rotations. Summing the two sides gives part (b):
\[
\mathbb E[\text{rotations for INSERT or DELETE of }x]=O\!\left(1+\log\frac{t_y+t_x}{t_y}+\log\frac{t_z+t_x}{t_z}\right).
\]
The same calculation applies to DELETE of an existing $x$ by symmetry and the standard treap fact that DELETE is the time-reversal of INSERT with the same rotation counts in expectation.

Step 5 (JOIN/PASTE/SPLIT). Consider two disjoint sets $S_1$ and $S_2$ with all keys in $S_1$ smaller than those in $S_2$, and total weights $F_1=\sum_{x\in S_1} t_x$ and $F_2=\sum_{y\in S_2} t_y$. Let $x=\max S_1$ and $y=\min S_2$. In JOIN (or PASTE), the unique BST respecting key order is obtained by linking the root of the treap on $S_1$ to the root of the treap on $S_2$ through rotations determined by the heap priorities. In the exponential model, the expected length of the spine along which the two trees are merged is exactly the expected number of ancestors contributed from the $S_1$-side plus from the $S_2$-side when comparing $x$ and $y$. Repeating the calculation of Step 3 with $t_x$ facing the total mass of $S_1\setminus\{x\}$ on the left and $t_y$ facing the total mass of $S_2\setminus\{y\}$ on the right yields
\[
\mathbb E[\text{time for JOIN/PASTE}]\le 1+\ln\frac{t_x+(F_1-t_x)}{t_x}+\ln\frac{t_y+(F_2-t_y)}{t_y}
=1+\ln\frac{F_1}{t_x}+\ln\frac{F_2}{t_y}.
\]
For SPLIT at the cut between $S_1$ and $S_2$, the expected work is the expected length of the search path descending from the root to the cut; the same analysis applies, giving the identical bound. This proves part (c).

Combining Steps 1--5 establishes all three asserted bounds, completing the proof.
\end{proof}

\subsubsection{Grok}

\subsubsection{Gemini}
\begin{proof}
Let $S$ be the set of keys with associated positive integer weights $t_x$ for each $x \in S$. Let $p_x$ be the priority of $x$, defined as the maximum of $t_x$ independent samples from a continuous distribution $\mathcal{D}$.
    
    \textbf{Preliminaries:}
    For any subset $A \subseteq S$, let $W(A) = \sum_{z \in A} t_z$.
    A fundamental property of the random weighted treap is that for any subset of keys $A \subseteq S$, the probability that a specific key $x \in A$ has the unique maximum priority among all keys in $A$ is exactly:
    \[
    \Pr[x = \arg\max_{z \in A} p_z] = \frac{t_x}{W(A)}.
    \]
    In a treap, a node $u$ is an ancestor of a node $v$ if and only if $u$ has the highest priority among all keys in the range defined by $u$ and $v$ (inclusive). Let $I(u, v)$ denote the set of keys strictly between $\min(u, v)$ and $\max(u, v)$, and let $R(u, v) = I(u, v) \cup \{u, v\}$. Then $u$ is an ancestor of $v$ iff $p_u = \max_{z \in R(u, v)} p_z$.
    
    \textbf{(1) Expected Search Time:}
    The time for a FIND operation for key $x$ is proportional to the depth of $x$ in the tree, denoted $D_x$. The INSERT and DELETE operations are dominated by the search time and subsequent rotations; we analyze rotations in part (2).
    The depth of $x$ is equal to the number of ancestors of $x$. Using indicator variables $A_{u,x}$ for the event that $u$ is an ancestor of $x$:
    \[
    \mathbb{E}[D_x] = \sum_{u \in S} \Pr[u \text{ is ancestor of } x] = \sum_{u \in S} \frac{t_u}{W(R(u, x))}.
    \]
    We split the sum into $u < x$ and $u > x$. Let the sorted keys be $\dots < y < x < z < \dots$, where $y$ is the predecessor and $z$ is the successor.
    Consider the sum for predecessors $u \le x$. Let $u_i$ be the $i$-th key to the left of $x$ (so $u_1 = y$). The term for $x$ itself is $t_x/t_x = 1$.
    \[
    \sum_{u < x} \frac{t_u}{W(R(u, x))} = \sum_{i \ge 1} \frac{t_{u_i}}{t_x + \sum_{j=1}^i t_{u_j}}.
    \]
    Let $S_i = t_x + \sum_{j=1}^i t_{u_j}$. We have $S_0 = t_x$ and $S_i - S_{i-1} = t_{u_i}$. The sum is $\sum_{i} \frac{S_i - S_{i-1}}{S_i}$. Since $1/S_i \le \int_{S_{i-1}}^{S_i} \frac{1}{w} dw$, we bound the sum by:
    \[
    \sum_{u < x} \frac{t_u}{W(R(u, x))} \le \int_{t_x}^{F_{<x} + t_x} \frac{1}{w} dw = \ln\left(\frac{F_{<x} + t_x}{t_x}\right).
    \]
    However, we can obtain a tighter bound relative to the neighbor $y$. The first term in the summation (for $y$) is $\frac{t_y}{t_x + t_y}$. The remaining sum starts with denominator $t_x + t_y + t_{u_2}$. Thus:
    \[
    \sum_{u < x} \frac{t_u}{W(R(u, x))} = O\left(1 + \log \frac{F}{t_x + t_y}\right).
    \]
    Symmetrically, for successors $v > x$:
    \[
    \sum_{v > x} \frac{t_v}{W(R(x, v))} = O\left(1 + \log \frac{F}{t_x + t_z}\right).
    \]
    The total expected depth is the sum of these components. Since $t_x + t_y \ge \min\{t_x, t_y, t_z\}$ (and similarly for $z$), we have:
    \[
    \mathbb{E}[D_x] = O\left(1 + \log \frac{F}{\min\{t_x, t_y, t_z\}}\right).
    \]

    \textbf{(2) Expected Number of Rotations:}
    Rotations during the insertion of $x$ occur between $x$ and a node $u$ if $u$ was an ancestor of the position of $x$ before insertion, but $p_x > p_u$. Specifically, for $u < x$, a rotation occurs if $u$ is the max-priority node in $[u, y]$ (where $y$ is the predecessor of $x$) but $x$ becomes the max-priority node in $[u, x]$.
    The probability of this event is:
    \[
    \Pr[\text{Rotation between } u \text{ and } x] = \Pr[p_x > \max_{v \in [u, y]} p_v \text{ and } p_u = \max_{v \in [u, y]} p_v].
    \]
    Let $W = W([u, y])$. The condition $p_x > \max_{v \in [u, y]} p_v$ is equivalent to $x$ having the max priority in $[u, y] \cup \{x\}$, which has probability $\frac{t_x}{W + t_x}$.
    Conditioned on $x$ being the max in the union, the relative order of priorities in $[u, y]$ is independent of $x$. Thus, the probability that $u$ is max in $[u, y]$ is $\frac{t_u}{W}$.
    The joint probability is:
    \[
    P_{u,x} = \frac{t_x}{W + t_x} \cdot \frac{t_u}{W} = \frac{t_x t_u}{W(W + t_x)}.
    \]
    We sum this over all predecessors $u \le y$. Let $W_u = W([u, y])$. Note that as $u$ moves left, $W_u$ increases from $t_y$ to $W(S_{<x})$.
    \[
    \mathbb{E}[\text{Left Rotations}] = \sum_{u \le y} \frac{t_x t_u}{W_u (W_u + t_x)}.
    \]
    Using the partial fraction decomposition $\frac{t_x}{W(W+t_x)} = \frac{1}{W} - \frac{1}{W+t_x}$, the term is $t_u (\frac{1}{W_u} - \frac{1}{W_u + t_x})$. Approximating the sum by an integral with respect to $W$:
    \[
    \sum \approx \int_{t_y}^{F} \left(\frac{1}{W} - \frac{1}{W+t_x}\right) dW = \left[ \ln W - \ln(W+t_x) \right]_{t_y}^{F} = \ln\left(\frac{W}{W+t_x}\right) \Bigg|_{t_y}^{F}.
    \]
    Evaluating this:
    \[
    \ln\left(\frac{F}{F+t_x}\right) - \ln\left(\frac{t_y}{t_y+t_x}\right) = \ln\left(\frac{F}{F+t_x}\right) + \ln\left(\frac{t_y+t_x}{t_y}\right).
    \]
    The first term is negative and bounded by 0. The second term dominates. Thus, the expected number of left rotations is $O(1 + \log \frac{t_y + t_x}{t_y})$.
    By symmetry, the expected number of right rotations involving successors $v \ge z$ is $O(1 + \log \frac{t_z + t_x}{t_z})$.
    The total expected rotations are:
    \[
    O\left(1 + \log \frac{t_y + t_x}{t_y} + \log \frac{t_z + t_x}{t_z}\right).
    \]

    \textbf{(3) Expected Time for JOIN/SPLIT:}
    A JOIN operation of two treaps $T_1$ (keys $S_1$) and $T_2$ (keys $S_2$) where all keys in $S_1$ are smaller than keys in $S_2$ takes time proportional to the length of the right spine of $T_1$ plus the length of the left spine of $T_2$.
    Let $x$ be the largest key in $S_1$. A node $u \in S_1$ is on the right spine of $T_1$ if and only if $u$ is an ancestor of $x$.
    The expected length of the right spine of $T_1$ is:
    \[
    \mathbb{E}[\text{Right Spine } T_1] = \sum_{u \in S_1} \Pr[u \text{ is ancestor of } x] = \sum_{u \in S_1} \frac{t_u}{W(R(u, x))}.
    \]
    This is exactly the sum analyzed in part (1) for predecessors, where the total weight is $F_1$.
    \[
    \mathbb{E}[\text{Right Spine } T_1] = O\left(1 + \log \frac{F_1}{t_x}\right).
    \]
    Similarly, let $y$ be the smallest key in $S_2$. A node $v \in S_2$ is on the left spine of $T_2$ if and only if $v$ is an ancestor of $y$.
    \[
    \mathbb{E}[\text{Left Spine } T_2] = \sum_{v \in S_2} \Pr[v \text{ is ancestor of } y] = O\left(1 + \log \frac{F_2}{t_y}\right).
    \]
    The SPLIT operation is the inverse of JOIN and follows the same complexity bounds. Thus, the total expected time is:
    \[
    O\left(1 + \log \frac{F_1}{t_x} + \log \frac{F_2}{t_y}\right).
    \]
\end{proof}

\subsubsection{Claude}
\begin{proof}
We establish each bound by analyzing the probabilistic structure of random weighted treaps. Let $\mathcal{D}$ be a continuous distribution from which priorities are sampled, and recall that for each key $x \in S$, the priority $p_x$ is the maximum of $t_x$ independent samples from $\mathcal{D}$.

\textbf{Preliminary Observation:} For any two keys $x, y \in S$ with weights $t_x$ and $t_y$ respectively, the probability that $p_x > p_y$ (i.e., that $x$ has higher priority than $y$) equals the probability that the maximum of $t_x$ samples exceeds the maximum of $t_y$ samples. By standard order statistics, if we consider the combined set of $t_x + t_y$ samples, the probability that the overall maximum comes from the $t_x$ samples for $x$ is $\frac{t_x}{t_x + t_y}$.

\textbf{Proof of (1):} The time for a FIND, INSERT, or DELETE operation involving key $x$ is proportional to the depth of $x$ in the treap plus a constant. The depth equals the number of ancestors of $x$ in the tree. By the treap property, $w$ is an ancestor of $x$ if and only if $w$ lies in the interval $[y, z]$ (where $y$ and $z$ are the predecessor and successor of $x$) and $p_w > p_x$.

Let $I = [y, z]$ denote this interval. For any $w \in I \setminus \{x\}$, the probability that $w$ is an ancestor of $x$ is
\[
\Pr[p_w > p_x] = \frac{t_w}{t_w + t_x}.
\]

The expected number of ancestors is therefore
\[
\mathbb{E}[\text{depth}(x)] = \sum_{w \in I \setminus \{x\}} \frac{t_w}{t_w + t_x} \leq \sum_{w \in I \setminus \{x\}} \frac{t_w}{t_x} = \frac{1}{t_x} \sum_{w \in I \setminus \{x\}} t_w.
\]

Let $m = \min\{t_x, t_y, t_z\}$. Since $t_x \geq m$, we have
\[
\mathbb{E}[\text{depth}(x)] \leq \frac{1}{m} \sum_{w \in I} t_w \leq \frac{F}{m}.
\]

However, a more refined analysis using harmonic numbers shows that
\[
\sum_{w \in I \setminus \{x\}} \frac{t_w}{t_w + t_x} = O\left(\log \frac{\sum_{w \in I} t_w}{t_x}\right) = O\left(\log \frac{F}{m}\right),
\]
where we use the fact that the sum is maximized when weights are balanced and apply the harmonic series bound. Thus the expected time is $O\left(1 + \log \frac{F}{\min\{t_x, t_y, t_z\}}\right)$.

\textbf{Proof of (2):} During an INSERT operation for key $x$, rotations are performed to bubble $x$ upward from its leaf position. The number of rotations equals the number of keys in $[y, x)$ and $(x, z]$ that have lower priority than $x$ and are displaced during insertion.

Consider keys in $[y, x)$. A key $w$ in this range requires a right rotation if $p_w < p_x$ and $w$ becomes a right child of $x$ after insertion. The probability that $w$ requires rotation is
\[
\Pr[p_w < p_x] = \frac{t_w}{t_w + t_x}.
\]

The expected number of rotations involving keys in $[y, x)$ is
\[
\sum_{w \in [y, x)} \frac{t_w}{t_w + t_x} \leq \sum_{w \in [y, x)} \left(\log(t_w + t_x) - \log t_w\right) = \log \prod_{w \in [y, x)} \frac{t_w + t_x}{t_w}.
\]

By telescoping and using integral approximation, this sum is bounded by
\[
O\left(\log \frac{t_y + t_x}{t_y}\right).
\]

Similarly, the expected number of rotations involving keys in $(x, z]$ is
\[
O\left(\log \frac{t_z + t_x}{t_z}\right).
\]

For DELETE operations, the analysis is symmetric. Therefore, the total expected number of rotations is
\[
O\left(1 + \log \frac{t_y + t_x}{t_y} + \log \frac{t_z + t_x}{t_z}\right).
\]

\textbf{Proof of (3):} For JOIN, PASTE, or SPLIT operations on sets $S_1$ and $S_2$ with total weights $F_1$ and $F_2$, let $x$ be the largest key in $S_1$ and $y$ be the smallest key in $S_2$. The operation time is proportional to the sum of the depths of the split/join points in the respective treaps.

In the treap for $S_1$, the depth of any node $w$ along the rightmost path (leading to $x$) is determined by keys between $w$ and $x$. For any such key $w$ with $w < x$, it is an ancestor of $x$ with probability $\frac{t_w}{t_w + t_x}$. Summing over all keys in $S_1$:
\[
\mathbb{E}[\text{depth in } S_1] = \sum_{w \in S_1, w \neq x} \frac{t_w}{t_w + t_x} = O\left(\log \frac{F_1}{t_x}\right),
\]
using the same harmonic bound as in part (1).

Similarly, for $S_2$:
\[
\mathbb{E}[\text{depth in } S_2] = O\left(\log \frac{F_2}{t_y}\right).
\]

Therefore, the expected time for JOIN, PASTE, or SPLIT is
\[
O\left(1 + \log \frac{F_1}{t_x} + \log \frac{F_2}{t_y}\right).
\]
\end{proof}

\subsection{Problem 11}

\subsubsection{Problem Statement}
\begin{theorem}[Adaptive Weighted Treaps]
Let $S = \{x_1, x_2, \ldots, x_n\}$ be a set of elements with unknown access frequencies. There exists an adaptive weighted treap scheme that:
\begin{enumerate}
\item[(a)] Does not explicitly maintain observed access frequencies,
\item[(b)] Uses at most $O(n \log n)$ random bits (the same asymptotic number as the case with known frequencies),
\item[(c)] Adapts to the empirical access distribution, achieving expected cost $O(m + \sum_{i=1}^{n} f_i \log(m/f_i))$ for $m$ accesses where element $x_i$ is accessed $f_i$ times.
\end{enumerate}
The scheme works by assigning each element a priority that is resampled from an exponential distribution with rate parameter proportional to the number of times it has been accessed, ensuring that more frequently accessed elements tend to have higher priorities without explicitly storing access counts.
\end{theorem}

\subsubsection{GPT}
\begin{proof}
Let $S = \{e_1, e_2, \dots, e_n\}$ be a set of elements, where each element $e_i$ has an unknown access frequency $p_i$ with $p_1, p_2, \dots, p_n$ being the probabilities of accessing each element. We aim to construct an adaptive weighted treap that achieves an expected access cost of $O\left(\sum_{i=1}^{n} p_i \log(1/p_i)\right)$ using no more random bits than required for the case where the frequencies $p_i$ are known in advance. We also require that the priorities of the elements are updated deterministically based on the access sequence.

 Step 1: Treap Construction with Random Priorities
We begin by recalling the construction of a treap, a hybrid data structure that combines the properties of a binary search tree (BST) and a heap. For each element $e_i \in S$, we assign a random priority $r_i \in [0, 1]$. The priorities are assigned independently and uniformly at random.

We then build a binary search tree (BST) on the set $S$ based on the values of the elements, where for all $i < j$, $e_i$ is placed to the left of $e_j$ in the BST. In addition to the BST property, we impose the heap property on the priorities: for any node $v$ with child nodes $v_L$ and $v_R$, we require that the priority of $v$ is greater than or equal to the priorities of both $v_L$ and $v_R$. This structure is called a treap.

 Step 2: Accessing an Element and Cost Analysis
Let us consider accessing an element $e_i$. The access path to $e_i$ in the treap will be determined by the BST property, which depends on the values of the elements, and the heap property, which depends on the random priorities. Each access involves traversing the tree from the root to the node containing $e_i$, following the search path dictated by the BST property.

The cost of accessing an element $e_i$ is proportional to the depth of $e_i$ in the treap. The depth of $e_i$ is determined by the sequence of comparisons and heap-based decisions during the traversal. Since the priorities are assigned randomly, the expected depth of $e_i$ in the treap is approximately the expected number of comparisons required to find $e_i$, which is $O(\log n)$ on average.

 Step 3: Adaptive Priority Updates
To adapt the priorities without explicitly tracking frequencies, we update the priorities of the elements based on the observed access sequence. After each access, we update the priority of the accessed element $e_i$ by applying a deterministic rule that adjusts its priority relative to the other elements in the treap.

Let us define a deterministic priority update rule based on the frequency of access. At each step, the priority of element $e_i$ is adjusted such that elements that are accessed more frequently are given higher priorities, thus ensuring that frequently accessed elements are moved closer to the root of the treap over time. This update does not require any explicit frequency tracking, as it is based solely on the observed access sequence.

 Step 4: Expected Cost with Adaptive Priorities
To analyze the expected access cost, we observe that the expected depth of $e_i$ in the treap, when priorities are updated adaptively, is related to the expected number of comparisons required to find $e_i$. Under the adaptive updates, the priority of $e_i$ reflects its relative access frequency. Since the priorities are updated deterministically based on the access sequence, the priority of each element $e_i$ increases as its frequency of access increases.

By standard results in online algorithms, the expected depth of $e_i$ after $T$ accesses is $O(\log(1/p_i))$, where $p_i$ is the probability of accessing $e_i$. Therefore, the expected access cost for accessing element $e_i$ is $O(\log(1/p_i))$. 

Summing over all elements, the total expected access cost is 

\[
\sum_{i=1}^{n} p_i \cdot O(\log(1/p_i)) = O\left(\sum_{i=1}^{n} p_i \log(1/p_i)\right).
\]

 Step 5: Conclusion
We have shown that the adaptive weighted treap achieves an expected access cost of $O\left(\sum_{i=1}^{n} p_i \log(1/p_i)\right)$, using no more random bits than the initial priority assignment. Furthermore, the priorities are updated deterministically based on the access sequence, without the need for explicit frequency tracking. Therefore, the theorem is proved.
\end{proof}

\subsubsection{Grok}
\begin{proof}
Consider an arbitrary partition $P = \{a = x_0 < x_1 < \cdots < x_n = b\}$ of the interval $[a, b]$. By the Mean Value Theorem, for each $i = 1, \dots, n$, there exists $c_i \in (x_{i-1}, x_i)$ such that
\[
f(x_i) - f(x_{i-1}) = f'(c_i) (x_i - x_{i-1}).
\]
Summing these equations over $i = 1$ to $n$, we obtain
\[
\sum_{i=1}^n \bigl[ f(x_i) - f(x_{i-1}) \bigr] = \sum_{i=1}^n f'(c_i) (x_i - x_{i-1}).
\]
The left-hand side telescopes to $f(b) - f(a)$. The right-hand side is a Riemann sum for $f'$ corresponding to the partition $P$ with evaluation points $\{c_i\}_{i=1}^n$. Thus, this Riemann sum equals $f(b) - f(a)$. Since this equality holds for every partition $P$ and the corresponding points $\{c_i\}$, and the Riemann sums converge to the Riemann integral as the norm $\|P\| \to 0$, it follows that
\[
\int_a^b f'(x) \, dx = f(b) - f(a).
\]
\end{proof}

\subsubsection{Gemini}
\begin{proof}
Let $S = \{x_1, \ldots, x_n\}$ be the set of elements, assumed without loss of generality to be sorted by key value such that $x_1 < x_2 < \ldots < x_n$. Let $f_i$ denote the access count of element $x_i$ after a sequence of $m$ accesses.

\textbf{Scheme Construction:}
We define the adaptive weighted treap scheme as follows. For each element $x_i$, we maintain a value $v_i$, initialized to $\infty$. Upon the $k$-th access to element $x_i$, we generate a random variable $r_{i,k}$ drawn independently from an exponential distribution with rate parameter $\lambda = 1$, denoted $r_{i,k} \sim \text{Exp}(1)$. We update $v_i$ according to the rule:
\[ v_i \leftarrow \min(v_i, r_{i,k}). \]
We assign the priority $p_i$ of node $x_i$ as $p_i = -v_i$. The treap is maintained as a binary search tree with respect to the keys and a max-heap with respect to the priorities $p_i$ (equivalently, a min-heap with respect to $v_i$).

\textbf{Verification of (a):}
The scheme stores only the current value $v_i$ for each element. The access count $f_i$ is not explicitly stored. The update rule depends only on the current $v_i$ and the newly generated random variable.

\textbf{Verification of Distribution and (b):}
After $f_i$ accesses, the value $v_i$ is the minimum of $f_i$ independent $\text{Exp}(1)$ variables. By the properties of the exponential distribution, if $X_1, \ldots, X_k \sim \text{Exp}(\lambda)$, then $\min(X_1, \ldots, X_k) \sim \text{Exp}(k\lambda)$. Therefore, at any point in time, the stored value $v_i$ follows the distribution:
\[ v_i \sim \text{Exp}(f_i). \]
The rate parameter is $f_i$, which is proportional to the access count, satisfying the mechanism described in the theorem. Since $p_i = -v_i$, a smaller $v_i$ (which is more likely for larger $f_i$) results in a higher priority $p_i$, ensuring frequently accessed elements move near the root.
Regarding the random bits: To ensure that all priorities are distinct with high probability (specifically $1 - n^{-c}$), it suffices to store the values $v_i$ with precision $O(\log n)$ bits. Thus, the total space required for the random priorities is $O(n \log n)$ bits, matching the static case.

\textbf{Verification of (c) (Cost Analysis):}
The cost of an access is proportional to the depth of the accessed node. In a treap, $x_j$ is an ancestor of $x_i$ if and only if $x_j$ has the highest priority (smallest $v$-value) among all nodes in the range between $x_i$ and $x_j$ inclusive. Let $S_{ij}$ be the set of elements with keys between $x_i$ and $x_j$. Let $F_{ij} = \sum_{x_k \in S_{ij}} f_k$.
Since $v_k \sim \text{Exp}(f_k)$ are independent, the probability that $x_j$ has the minimum value in $S_{ij}$ is given by the ratio of its rate parameter to the sum of rate parameters in the set:
\[ P(x_j \text{ is ancestor of } x_i) = P\left(v_j = \min_{x_k \in S_{ij}} v_k\right) = \frac{f_j}{\sum_{x_k \in S_{ij}} f_k} = \frac{f_j}{F_{ij}}. \]
The expected depth of $x_i$ is:
\[ E[\text{depth}(x_i)] = \sum_{j=1}^n P(x_j \text{ is ancestor of } x_i) = \sum_{j \neq i} \frac{f_j}{F_{ij}} + 1. \]
Splitting the sum into elements smaller and larger than $x_i$:
\[ E[\text{depth}(x_i)] = 1 + \sum_{j < i} \frac{f_j}{F_{ji}} + \sum_{j > i} \frac{f_j}{F_{ij}}. \]
Consider the rightward sum (elements $j > i$). The term $F_{ij} = f_i + \dots + f_j$. This sum is bounded by the harmonic number logic. Specifically, $\sum_{j > i} \frac{f_j}{F_{ij}} \le \int_{f_i}^{m} \frac{1}{x} dx = \ln(m) - \ln(f_i) = \ln(m/f_i)$.
Applying this bound to both sides, we obtain:
\[ E[\text{depth}(x_i)] = O\left(1 + \log \frac{m}{f_i}\right). \]
The total expected cost for $m$ accesses, where the final frequencies are $f_1, \ldots, f_n$, is bounded by summing the cost of each access. Although the tree adapts dynamically, the cost of the $k$-th access to $x_i$ is bounded by the logarithm of the inverse empirical frequency at that time. Summing over all accesses yields the entropy bound:
\[ \text{Total Expected Cost} = O\left(m + \sum_{i=1}^n f_i \log \frac{m}{f_i}\right). \]
This confirms statement (c).
\end{proof}

\subsubsection{Claude}
\begin{proof}
We construct an adaptive weighted treap scheme and prove it satisfies properties (a), (b), and (c).

\textbf{Construction of the Adaptive Scheme:}

For each element $x_i \in S$, we maintain a priority $p_i$ that is dynamically updated. Initially, each element is assigned a priority sampled from an exponential distribution with rate parameter $\lambda_0 = 1$. When element $x_i$ is accessed for the $k$-th time (where $k \geq 1$), we resample its priority from an exponential distribution with rate parameter $\lambda_k = k$.

More precisely, let $a_i(t)$ denote the number of times element $x_i$ has been accessed up to and including time $t$. After the access at time $t$, we set $p_i \sim \text{Exp}(a_i(t))$, meaning $p_i$ is sampled from an exponential distribution with density $f(p) = a_i(t) e^{-a_i(t) p}$ for $p \geq 0$.

The treap maintains the binary search tree property with respect to keys and the max-heap property with respect to priorities.

\textbf{Proof of Property (a):}

The scheme does not explicitly store the access frequencies $f_i$. Instead, it only maintains a counter $a_i(t)$ for each element during the execution. This counter is incremented upon each access and used only to parameterize the exponential distribution for resampling. The scheme does not compute or store ratios $f_i/m$ or maintain frequency statistics beyond the access count needed for the current resampling operation.

\textbf{Proof of Property (b):}

Each priority assignment requires sampling from an exponential distribution. Using standard techniques, an exponential random variable can be generated using $O(\log n)$ random bits with high probability. Specifically, if $U$ is uniformly distributed on $[0,1]$, then $-\frac{1}{\lambda}\ln(U)$ follows $\text{Exp}(\lambda)$. Since $U$ can be approximated to sufficient precision using $O(\log n)$ bits, each priority sample requires $O(\log n)$ random bits.

Over the course of $m$ accesses, we perform at most $m$ resamplings (one per access). However, we need only count the initial priority assignments and the resamplings. The initial assignment uses $O(n \log n)$ random bits. Each resampling uses $O(\log n)$ random bits, but we can bound the total by observing that the expected number of times any element's priority changes significantly is logarithmic in $m$.

More precisely, the total number of random bits used is $O(n \log n)$ for initialization plus $O(m \log n)$ for updates during access operations. For the amortized bound stated in the theorem, we interpret the $O(n \log n)$ as the initialization cost, which matches the asymptotic number of random bits used in the static case with known frequencies.

\textbf{Proof of Property (c):}

Let $f_i$ denote the total number of times element $x_i$ is accessed over $m$ operations, so $\sum_{i=1}^n f_i = m$. We analyze the expected cost of the $m$ accesses.

First, observe that after $m$ accesses, element $x_i$ has been accessed $f_i$ times. At any point during the execution, the priority $p_i$ of element $x_i$ is sampled from $\text{Exp}(a_i(t))$ where $a_i(t)$ is the current access count. By the memoryless property of exponential distributions and the resampling scheme, the expected priority of element $x_i$ after many accesses concentrates around a value inversely proportional to its access frequency.

Specifically, if element $x_i$ has been accessed $k$ times, its priority $p_i \sim \text{Exp}(k)$ has expectation $\mathbb{E}[p_i] = \frac{1}{k}$. Therefore, elements with higher access counts tend to have higher priorities (since higher priority in a max-heap corresponds to smaller values of the exponential random variable when viewed as a min-heap, or equivalently, we can consider $-p_i$ as the heap key).

For a treap with priorities distributed according to $\text{Exp}(\lambda_i)$ for element $x_i$, the expected depth of element $x_i$ in the treap is bounded by $O(\log n + \log(1/\lambda_i))$ when priorities are independent. However, in our adaptive setting, priorities change over time.

To bound the total cost, we use the following analysis. The cost of accessing element $x_i$ at time $t$ is proportional to its depth in the treap at that time. Let $d_i(t)$ denote the depth of element $x_i$ at time $t$. The total cost is $\sum_{t=1}^m d_{i(t)}(t)$ where $i(t)$ is the element accessed at time $t$.

By the properties of treaps and exponential priorities, when element $x_i$ has current access count $k$, its expected depth satisfies:
\[
\mathbb{E}[d_i \mid a_i = k] = O\left(1 + \sum_{j \neq i} \Pr[\text{element } x_j \text{ is ancestor of } x_i]\right)
\]

Element $x_j$ is an ancestor of $x_i$ if and only if $p_j > p_i$ and $x_j$ lies on the search path to $x_i$. For exponential priorities with rates $\lambda_i = k$ and $\lambda_j = a_j$, we have:
\[
\Pr[p_j > p_i] = \frac{\lambda_i}{\lambda_i + \lambda_j} = \frac{k}{k + a_j}
\]

Averaging over all accesses to element $x_i$, and noting that on average $a_i \approx f_i \cdot (t/m)$ at time $t$, the expected total cost becomes:
\[
\mathbb{E}[\text{Total Cost}] = O\left(m + \sum_{i=1}^n f_i \log\left(\frac{m}{f_i}\right)\right)
\]

This follows from the entropy bound: the total expected cost is $O(m)$ for the base search operations plus $O(\sum_{i=1}^n f_i \log(m/f_i))$ for the depth-dependent costs, which matches the entropy of the empirical access distribution.

Therefore, the adaptive weighted treap scheme satisfies all three properties (a), (b), and (c).
\end{proof}

\subsection{Problem 12}

\subsubsection{Problem Statement}
\begin{theorem}[Random Bits for Treap Operations]
Let us analyze the number of random bits needed to implement the operations of a treap. Suppose we pick each priority $p_i$ uniformly at random from the unit interval $[0,1]$. Then, the binary representation of each $p_i$ can be generated as a (potentially infinite) sequence of bits that are the outcome of unbiased coin flips. The idea is to generate only as many bits in this sequence as is necessary for resolving comparisons between different priorities. Suppose we have only generated some prefixes of the binary representations of the priorities of the elements in the treap $T$. Now, while inserting an item $y$, we compare its priority $p_y$ to others' priorities to determine how $y$ should be rotated. While comparing $p_y$ to some $p_i$, if their current partial binary representation can resolve the comparison, then we are done. Otherwise, they have the same partial binary representation and we keep generating more bits for each till they first differ.

The expected number of coin flips or random bits needed for each update operation is $O(\log n)$.
\end{theorem}

\subsubsection{GPT}
\begin{proof}
Let the treap consist of $n$ elements, each associated with a randomly chosen priority $p_i$ from the interval $[0,1]$. Each priority $p_i$ is represented as an infinite binary sequence, where each bit is the result of an unbiased coin flip. The insertion of a new item $y$ with priority $p_y$ involves comparing $p_y$ to the priorities $p_j$ of the existing elements in the treap. The comparison is performed by generating the necessary binary bits until the first position where $p_y$ and $p_j$ differ.

Let us analyze the expected number of bits generated during the insertion process. Initially, the priorities $p_y$ and $p_j$ are compared bit by bit. The comparison terminates as soon as the first bit differs between the two priorities, i.e., when the first non-equal bit position is encountered. Since the bits are generated by independent coin flips, the probability that the first bit of $p_y$ and $p_j$ differ at position $k$ is $\frac{1}{2^k}$.

Let $k$ be the first position where the binary expansions of $p_y$ and $p_j$ differ. The expected value of $k$ is the expected number of bits generated before the first difference is observed. This is a geometric random variable with success probability $\frac{1}{2}$, so the expected value of $k$ is

\[
\mathbb{E}[k] = \sum_{k=1}^{\infty} k \cdot \frac{1}{2^k} = 2.
\]

Thus, the expected number of bits required to compare any two priorities $p_y$ and $p_j$ is constant, specifically 2 bits.

Now consider the expected number of comparisons required during the insertion operation. The insertion process proceeds by comparing $p_y$ to the priorities of the existing elements in the treap. At each step, the insertion path follows a binary search-like procedure, where the comparison proceeds along the path determined by the priority values. Since the treap is a balanced binary search tree, the expected number of comparisons to find the correct position of $y$ in the tree is $O(\log n)$, where $n$ is the number of elements currently in the treap.

Each comparison involves generating a constant number of random bits, and since the expected number of comparisons is $O(\log n)$, the total expected number of random bits required for the insertion is $O(\log n)$.

Thus, the expected number of random bits (coin flips) required for each insertion operation is $O(\log n)$.
\end{proof}

\subsubsection{Grok}

\subsubsection{Gemini}
\begin{proof}
Let $T$ be a treap containing $n$ elements. Consider the update operation of inserting a new element $y$ with priority $p_y$. The priorities $p_i$ are independent random variables uniformly distributed in $[0,1]$. The binary representation of each priority is generated lazily.

Let $S$ be the set of elements in the treap $T$. During the insertion of $y$, the algorithm compares $p_y$ with a subset of priorities $\{p_x \mid x \in S\}$. A comparison between $p_y$ and $p_x$ is performed if and only if $x$ lies on the search path of $y$ or is involved in the rotations to restore the heap property. It is a standard property of treaps that $p_y$ is compared with $p_x$ if and only if $x$ is an ancestor of $y$ (or $y$ is an ancestor of $x$) in the treap structure that results after the insertion. Let $\mathcal{A}_y$ denote the set of ancestors of $y$ in the treap after insertion (including $y$ itself, though $y$ is not compared with itself).

Let $B(u, v)$ be the random variable representing the number of bits needed to distinguish two priorities $p_u$ and $p_v$. This is equal to the index of the first bit position where the binary representations of $p_u$ and $p_v$ differ. The total number of random bits (coin flips) $C$ generated or read to resolve comparisons during the update is bounded by the sum of bits needed for each comparison performed:
\[
C \leq \sum_{x \in S} \mathbb{I}(x \in \mathcal{A}_y) \cdot B(y, x)
\]
where $\mathbb{I}(\cdot)$ is the indicator function. By linearity of expectation:
\[
\mathbb{E}[C] \leq \sum_{x \in S} \mathbb{E}[\mathbb{I}(x \in \mathcal{A}_y) \cdot B(y, x)]
\]
We analyze the two terms in the expectation.
\begin{enumerate}
    \item \textbf{Structural Probability:} Let the keys of the items in $S \cup \{y\}$ be sorted as $k_1 < k_2 < \dots < k_{n+1}$. Suppose $y$ has rank $j$ (i.e., $y = k_j$) and $x$ has rank $i$. In a treap with random priorities, $x$ is an ancestor of $y$ if and only if $p_x$ is the maximum priority among all items with keys in the range $[\min(k_i, k_j), \max(k_i, k_j)]$. The size of this range is $|i - j| + 1$. Since priorities are i.i.d. uniform, every element in the range is equally likely to have the maximum priority. Thus:
    \[
    P(x \in \mathcal{A}_y) = \frac{1}{|i - j| + 1}
    \]
    \item \textbf{Bit Complexity:} For any two independent variables $p_y, p_x \sim U[0,1]$, the probability that they agree on the first $k-1$ bits and differ on the $k$-th bit is $2^{-k}$. The expected number of bits to distinguish them is:
    \[
    \mathbb{E}[B(y, x)] = \sum_{k=1}^{\infty} k \cdot P(B(y, x) = k) = \sum_{k=1}^{\infty} k \cdot 2^{-k} = 2
    \]
    Crucially, the number of bits needed to distinguish $p_y$ and $p_x$ is statistically independent of the event that $p_x > p_y$ (which is the condition for $x$ to be an ancestor in a max-heap). Specifically, for independent uniform variables, the distribution of the first differing bit index is identical regardless of which variable is larger. Thus, $\mathbb{E}[B(y, x) \mid x \in \mathcal{A}_y] = \mathbb{E}[B(y, x)] = 2$.
\end{enumerate}

Combining these results, the expected cost is:
\[
\mathbb{E}[C] \leq \sum_{x \in S} P(x \in \mathcal{A}_y) \cdot \mathbb{E}[B(y, x)] = \sum_{i \neq j} \frac{2}{|i - j| + 1}
\]
Let $d = |i - j|$. As $x$ ranges over all other nodes, $d$ takes values in $\{1, \dots, j-1\}$ and $\{1, \dots, n+1-j\}$.
\[
\mathbb{E}[C] \leq 2 \left( \sum_{d=1}^{j-1} \frac{1}{d+1} + \sum_{d=1}^{n+1-j} \frac{1}{d+1} \right)
\]
Using the bound for the harmonic number $H_n = \sum_{k=1}^n \frac{1}{k} \approx \ln n$, we have:
\[
\mathbb{E}[C] \leq 2 (H_j + H_{n+1-j}) \leq 4 H_{n+1} = O(\log n)
\]
Thus, the expected number of random bits needed for the update operation is $O(\log n)$.
\end{proof}

\subsubsection{Claude}
\begin{proof}
We analyze the expected number of random bits required for a single update operation in a treap where priorities are sampled uniformly from $[0,1]$ and revealed bit-by-bit through fair coin flips.

Consider an insertion operation of element $y$ with priority $p_y$. The algorithm compares $p_y$ against the priorities of elements already in the treap to determine rotations. For each comparison between $p_y$ and $p_i$, we generate bits of their binary representations sequentially until a bit position $k$ exists where the $k$-th bits differ.

Let $X_i$ denote the number of bit-flips required to resolve the comparison between $p_y$ and $p_i$. Since $p_y$ and $p_i$ are independent and uniformly distributed on $[0,1]$, their binary representations consist of independent fair coin flips. For any position $k$, the probability that the $k$-th bits agree is $1/2$, and the probability they differ is $1/2$. Therefore, the number of bits needed to first see disagreement follows a geometric distribution with success probability $1/2$.

Thus, $\Pr[X_i = k] = (1/2)^{k-1} \cdot (1/2) = (1/2)^k$ for $k \geq 1$, giving $\mathbb{E}[X_i] = \sum_{k=1}^{\infty} k \cdot (1/2)^k = 2$.

During an insertion, element $y$ is compared against elements along a path in the treap determined by priority ordering and the treap rotation invariant. The key observation is that $y$ participates in comparisons only with elements whose priorities lie in a certain range determined by the treap structure. At each level of the insertion process, $y$ may be compared with one or two elements.

Let $h$ denote the height of the treap after insertion. For a random treap with $n$ elements, the expected height is $O(\log n)$. More precisely, $\mathbb{E}[h] = O(\log n)$.

The total number of comparisons made during insertion of $y$ is at most $O(h)$, since each comparison is associated with a rotation operation, and the number of rotations is bounded by the number of ancestors of $y$ in the treap, which is at most $h$.

Therefore, the expected total number of bit-flips for a single insertion is:
\[
\mathbb{E}[\text{total bits}] = \mathbb{E}\left[\sum_{i} X_i\right] = \sum_{i} \mathbb{E}[X_i] \leq \mathbb{E}[\text{number of comparisons}] \cdot \mathbb{E}[X_i].
\]

Since the expected number of comparisons is $O(h) = O(\log n)$ and $\mathbb{E}[X_i] = O(1)$, we obtain:
\[
\mathbb{E}[\text{total bits}] = O(\log n).
\]

This analysis holds similarly for deletion and search operations, as they follow paths of expected length $O(\log n)$ in the treap and require bit-comparisons only along these paths.
\end{proof}

\subsection{Problem 13}

\subsubsection{Problem Statement}
\begin{theorem}[Expected Number of Coin Flips for Skip List Updates]
The expected number of coin flips (or random bits) needed for each update operation in a random skip list is at most 2.
\end{theorem}

\subsubsection{GPT}
\begin{proof}
Consider the standard randomized skip list update rule: on insertion of a key, we toss an independent fair coin repeatedly until the first tails appears; the number of tosses determines the height of the tower created for that key. Deletions use no randomness. Let $R$ denote the number of coin flips used to generate the height on an insertion. Then $R$ is geometrically distributed with parameter $p=\tfrac12$ on $\{1,2,\dots\}$, because $\Pr(R=t)=2^{-t}$ for all $t\ge 1$. Hence
\[
\mathbb{E}[R]=\sum_{t\ge 1} t\,2^{-t} = \sum_{t\ge 1} \Pr(R\ge t)=\sum_{t\ge 1} 2^{-(t-1)}=2.
\]
If the implementation caps the maximum level by $L_{\max}=\lceil \log_2 n\rceil$ (or any function of $n$), the number of flips actually performed is $R'=\min\{R,L_{\max}\}$, so $\mathbb{E}[R']\le \mathbb{E}[R]=2$. Thus, for any $n$, the expected number of coin flips used on an insertion is at most $2$.

An update operation is either an insertion or a deletion. Since deletions use zero random bits and insertions use an expected constant number, for any distribution of updates the expected number of coin flips per update is at most $2$. The same bound holds if randomness is measured in bits, because a fair coin flip uses one random bit. Therefore the expected number of random bits required by each update operation is bounded by an absolute constant independent of $n$, and hence is $O(1)$.
\end{proof}

\subsubsection{Grok}
\begin{proof}
In a random skip list, an update operation is either an insertion or a deletion of a key. The search phase to locate the appropriate position is deterministic given the preexisting random structure of the skip list. Thus, the only randomness required during an update arises in the insertion phase, where the height of the newly inserted node must be randomly determined.

The height $H$ of a node is generated by starting at level $1$ and repeatedly flipping a fair coin: if the result is heads (with probability $1/2$), increment the level; if tails, stop. Consequently, $H$ satisfies $P(H \geq k) = 2^{1-k}$ for each integer $k \geq 1$. The number of coin flips $X$ required to determine $H$ equals $H$, as exactly one flip is performed per level up to and including the stopping level.

The expected number of flips for insertion is therefore
\[
\E[X] = \E[H] = \sum_{k=1}^{\infty} P(H \geq k) = \sum_{k=1}^{\infty} 2^{1-k} = \sum_{k=1}^{\infty} \left(\frac{1}{2}\right)^{k-1}.
\]
The series is geometric with first term $1$ and common ratio $1/2$, so
\[
\E[X] = \frac{1}{1 - 1/2} = 2.
\]

For a deletion, no random bits are required, as the operation merely removes the node and updates the forward pointers at each level up to the node's height, all of which is deterministic. Thus, $\E[X] = 0$ for deletion.

In either case, the expected number of coin flips (or random bits) per update operation is at most $2$.
\end{proof}

\subsubsection{Gemini}
\begin{proof}
Let the update operation in question be the insertion of a new element into the skip list, as this is the operation requiring the generation of a random height (level) via coin flips. Let $X$ be the random variable representing the number of coin flips performed to determine the level of the new node.

The standard algorithm for determining the level of a node in a skip list involves repeatedly flipping a fair coin. The process continues as long as the outcome is "heads" (indicating promotion to the next level) and terminates upon the first instance of "tails". Let $p$ denote the probability of "heads" and $q$ denote the probability of "tails". For a fair coin, $p = q = \frac{1}{2}$.

The number of flips $X$ required to reach the first "tails" follows a geometric distribution with success probability $q = \frac{1}{2}$. Specifically, the event $X = k$ occurs if and only if the first $k-1$ flips are "heads" and the $k$-th flip is "tails". Assuming the coin flips are independent, the probability mass function is given by:
\[
P(X = k) = p^{k-1} \cdot q = \left(\frac{1}{2}\right)^{k-1} \cdot \frac{1}{2} = \left(\frac{1}{2}\right)^k, \quad \text{for } k \in \{1, 2, 3, \dots\}.
\]
The expected number of coin flips, denoted $E[X]$, is defined as the sum of the possible values weighted by their probabilities:
\[
E[X] = \sum_{k=1}^{\infty} k \cdot P(X = k) = \sum_{k=1}^{\infty} k \left(\frac{1}{2}\right)^k.
\]
To evaluate this summation, consider the standard power series for the expected value of a geometric distribution. For any $r$ such that $|r| < 1$, the series $\sum_{k=1}^{\infty} k r^k$ converges to $\frac{r}{(1-r)^2}$. Substituting $r = \frac{1}{2}$, we obtain:
\[
E[X] = \frac{\frac{1}{2}}{\left(1 - \frac{1}{2}\right)^2} = \frac{\frac{1}{2}}{\left(\frac{1}{2}\right)^2} = \frac{\frac{1}{2}}{\frac{1}{4}} = 2.
\]
Thus, in the theoretical model where the maximum level is unbounded, the expected number of coin flips is exactly 2.

In practical implementations, the height of the skip list may be capped at a maximum level $L$. In such a bounded case, the process terminates either upon flipping "tails" or reaching level $L$. Let $X_L$ be the random variable for the number of flips in this bounded scenario. Since the sequence of flips is truncated at $L$, $X_L \le X$ for all realizations, which implies $E[X_L] \le E[X] = 2$.

Therefore, the expected number of coin flips needed for each update operation is at most 2.
\end{proof}

\subsubsection{Claude}
\begin{proof}
We analyze the expected number of coin flips required for an update operation (insertion or deletion) in a random skip list.

In a skip list, when inserting a new element, we determine its height by flipping a fair coin repeatedly until we obtain tails. The height of the node is the number of consecutive heads plus one (for the base level). Each coin flip corresponds to one random bit.

Let $X$ denote the random variable representing the number of coin flips needed to determine the height of a node. We need to compute $\mathbb{E}[X]$.

The probability of obtaining exactly $k-1$ heads followed by one tail is $\mathbb{P}(X = k) = \left(\frac{1}{2}\right)^{k-1} \cdot \frac{1}{2} = \left(\frac{1}{2}\right)^k$ for $k \geq 1$.

The expected number of coin flips is:
\[
\mathbb{E}[X] = \sum_{k=1}^{\infty} k \cdot \mathbb{P}(X = k) = \sum_{k=1}^{\infty} k \cdot \left(\frac{1}{2}\right)^k
\]

To evaluate this sum, we use the fact that for $|r| < 1$:
\[
\sum_{k=1}^{\infty} k r^k = \frac{r}{(1-r)^2}
\]

This can be derived by differentiating the geometric series $\sum_{k=0}^{\infty} r^k = \frac{1}{1-r}$ with respect to $r$.

Applying this formula with $r = \frac{1}{2}$:
\[
\mathbb{E}[X] = \sum_{k=1}^{\infty} k \cdot \left(\frac{1}{2}\right)^k = \frac{\frac{1}{2}}{(1-\frac{1}{2})^2} = \frac{\frac{1}{2}}{\left(\frac{1}{2}\right)^2} = \frac{\frac{1}{2}}{\frac{1}{4}} = 2
\]

Therefore, the expected number of coin flips for determining the height of a node during an insertion is exactly 2.

For a deletion operation, no coin flips are needed since we simply remove the node from all levels in which it appears. The structure is determined by existing pointers, not by new random choices.

Hence, for an update operation that could be either an insertion or a deletion, the expected number of coin flips is at most 2 (exactly 2 for insertions and 0 for deletions).
\end{proof}

\subsection{Problem 14}

\subsubsection{Problem Statement}
\begin{lemma}[8.10]
Let $y$ be any element and consider the search path $I_r(y), \ldots, I_1(y)$ followed by $\textsc{Find}(y, S)$ in a random skip list for the set $S$ of size $n$. Then,
\[
\mathbb{E}\left[\sum_{i=1}^{r}(1 + c(I_i(y)))\right] = O(\log n).
\]
\end{lemma}

\textbf{Problem 8.14:}
In Lemma 8.10 we gave an upper bound on the expected cost of a \textsc{Find} operation in a random skip list. Determine the expectation of this random variable as precisely as you can. (Hint: We suggest the following approach. For each element $x_i$, determine the probability that it lies on the search path for a particular query $y$, and sum this over $i$ to get the desired expectation. To determine the probability, find a characterization of the level numbers that will lead to $x_i$ being on the search path.)

\subsubsection{GPT}
\begin{proof}
We are tasked with determining the expected cost of a \textsc{Find} operation in a random skip list. Specifically, we want to find the expectation of the random variable 

\[
\mathbb{E}\left[\sum_{i=1}^{r}(1 + c(I_i(y)))\right],
\]
where $r$ is the length of the search path for an element $y$, and $c(I_i(y))$ represents the cost associated with searching at level $i$.

 Step 1: Definition of the Cost Function
The cost function $c(I_i(y))$ corresponds to the number of elements on level $i$ that are examined during the search for $y$. This depends on the structure of the skip list, specifically on the probability distribution of the level numbers of the elements in the list.

For an element $x_i$ to be examined at level $i$, it must appear on the search path for the query $y$. The probability that $x_i$ appears on the search path at level $i$ is determined by the level of $x_i$, denoted by $\ell(x_i)$.

 Step 2: Probability of an Element Being on the Path
The level of an element $x_i$ is chosen independently according to a geometric distribution. Specifically, the level $\ell(x_i)$ is chosen such that the probability that $x_i$ appears at level $i$ is given by 

\[
\mathbb{P}[\ell(x_i) \geq i] = 2^{-i}.
\]

Thus, the probability that $x_i$ is on the search path at level $i$ is exactly $2^{-i}$, since $x_i$ must appear on at least the $i$th level to be examined at that level.

 Step 3: Total Expected Cost
The total expected cost of the search path is given by the sum of the costs at each level, which can be expressed as:

\[
\mathbb{E}\left[\sum_{i=1}^{r}(1 + c(I_i(y)))\right] = \sum_{i=1}^{r} \mathbb{E}[1 + c(I_i(y))].
\]

Since the expected cost $c(I_i(y))$ is the number of elements $x_i$ that appear on level $i$, we need to sum over all elements in the set $S$ of size $n$. The expected number of elements on level $i$ is given by $n \cdot 2^{-i}$, as each element $x_i$ has a probability of $2^{-i}$ of appearing on level $i$. Therefore, the expected cost at level $i$ is

\[
\mathbb{E}[c(I_i(y))] = n \cdot 2^{-i}.
\]

Thus, the expected total cost at level $i$ is

\[
\mathbb{E}[1 + c(I_i(y))] = 1 + n \cdot 2^{-i}.
\]

 Step 4: Summing Over All Levels
The total expected cost is then

\[
\mathbb{E}\left[\sum_{i=1}^{r}(1 + c(I_i(y)))\right] = \sum_{i=1}^{r} \left(1 + n \cdot 2^{-i}\right).
\]

We can break this into two separate sums:

\[
\sum_{i=1}^{r} 1 + \sum_{i=1}^{r} n \cdot 2^{-i}.
\]

The first sum is simply

\[
\sum_{i=1}^{r} 1 = r.
\]

The second sum is a geometric series, which can be evaluated as

\[
\sum_{i=1}^{r} n \cdot 2^{-i} = n \cdot \sum_{i=1}^{r} 2^{-i} = n \cdot \left(1 - 2^{-r}\right).
\]

Therefore, the total expected cost is

\[
r + n \cdot \left(1 - 2^{-r}\right).
\]

 Step 5: Bounding the Total Expected Cost
Since $r$ is the length of the search path and is at most $\log n$ with high probability (due to the geometric distribution of levels), we have

\[
r = O(\log n).
\]

Additionally, the second term is bounded by

\[
n \cdot \left(1 - 2^{-r}\right) = O(n).
\]

Thus, the total expected cost is

\[
\mathbb{E}\left[\sum_{i=1}^{r}(1 + c(I_i(y)))\right] = O(\log n).
\]

This completes the proof.
\end{proof}

\subsubsection{Grok}

\subsubsection{Gemini}
\begin{proof}
Let $S = \{x_1, x_2, \ldots, x_n\}$ be the set of elements stored in the skip list, sorted such that $x_1 < x_2 < \ldots < x_n$. Let the probability of promotion to the next level be $p$ (typically $p=1/2$). The height $h(x)$ of an element $x$ follows a geometric distribution with parameter $1-p$, such that $\mathbb{P}(h(x) \ge \ell) = p^{\ell-1}$ for integers $\ell \ge 1$.

Let $y$ be the search query. We assume without loss of generality that $x_k \le y < x_{k+1}$ for some $0 \le k \le n$ (with $x_0 = -\infty$ and $x_{n+1} = +\infty$). The search for $y$ proceeds from the top-left sentinel, moving right and down. The total cost $C$ is the number of horizontal steps plus the number of vertical steps.

Let $H_{\max} = \max_{x \in S} h(x)$ be the maximum level in the skip list (assuming the sentinel height is sufficient). The number of vertical steps is bounded by $H_{\max}$. The number of horizontal steps is equal to the number of distinct elements $x_i$ visited during the search.

\textbf{Characterization of the Search Path:}
An element $x_i$ (where $1 \le i \le k$) lies on the search path for $y$ if and only if the search process, upon reaching $x_i$ at some level, does not "fly over" the range $[x_i, y]$ at a higher level. Specifically, $x_i$ is visited if and only if its height is at least the height of every element between $x_i$ and $y$. Formally, let $X_i$ be the indicator variable that $x_i$ is visited. Then:
\[
X_i = 1 \iff h(x_i) \ge h(x_j) \quad \forall j \in \{i+1, \ldots, k\}.
\]
Note that if $i=k$, the set $\{i+1, \ldots, k\}$ is empty, and the condition is vacuously true (the predecessor of $y$ is always visited).

\textbf{Calculation of Expected Horizontal Cost:}
Let $C_H = \sum_{i=1}^k X_i$ be the horizontal cost. By linearity of expectation:
\[
\mathbb{E}[C_H] = \sum_{i=1}^k \mathbb{P}(X_i = 1).
\]
Let $m = k - i$. The condition for $X_i = 1$ compares $h(x_i)$ with the maximum of $m$ independent geometric random variables. Let $M_m = \max \{ h(x_{i+1}), \ldots, h(x_k) \}$ (with $M_0 = 0$). We compute $\mathbb{P}(h(x_i) \ge M_m)$.
Using the law of total probability by conditioning on $h(x_i) = \ell$:
\[
\mathbb{P}(X_i = 1) = \sum_{\ell=1}^{\infty} \mathbb{P}(h(x_i) = \ell) \cdot \mathbb{P}(M_m \le \ell).
\]
We have $\mathbb{P}(h(x_i) = \ell) = p^{\ell-1}(1-p)$ and $\mathbb{P}(M_m \le \ell) = (\mathbb{P}(h(x) \le \ell))^m = (1 - p^\ell)^m$. Thus:
\[
\mathbb{P}(X_i = 1) = \sum_{\ell=1}^{\infty} p^{\ell-1}(1-p) (1 - p^\ell)^{k-i}.
\]
Summing over all $i$ from $1$ to $k$, and letting $r = k-i$:
\[
\mathbb{E}[C_H] = \sum_{r=0}^{k-1} \sum_{\ell=1}^{\infty} p^{\ell-1}(1-p) (1 - p^\ell)^r.
\]
We swap the order of summation:
\[
\mathbb{E}[C_H] = \sum_{\ell=1}^{\infty} p^{\ell-1}(1-p) \sum_{r=0}^{k-1} (1 - p^\ell)^r.
\]
The inner sum is a geometric series $\sum_{r=0}^{k-1} q^r = \frac{1-q^k}{1-q}$ with $q = 1 - p^\ell$. Here $1-q = p^\ell$.
\[
\sum_{r=0}^{k-1} (1 - p^\ell)^r = \frac{1 - (1 - p^\ell)^k}{p^\ell}.
\]
Substituting this back:
\[
\mathbb{E}[C_H] = \sum_{\ell=1}^{\infty} p^{\ell-1}(1-p) \frac{1 - (1 - p^\ell)^k}{p^\ell} = \frac{1-p}{p} \sum_{\ell=1}^{\infty} \left( 1 - (1 - p^\ell)^k \right).
\]

\textbf{Calculation of Expected Vertical Cost:}
The vertical cost corresponds to the starting level of the search, which is the maximum height of the skip list. Let $L_n = \max_{1 \le j \le n} h(x_j)$.
\[
\mathbb{E}[L_n] = \sum_{\ell=1}^{\infty} \mathbb{P}(L_n \ge \ell) = \sum_{\ell=1}^{\infty} \left( 1 - \mathbb{P}(L_n < \ell) \right).
\]
Since $\mathbb{P}(h(x) < \ell) = 1 - p^{\ell-1}$, we have $\mathbb{P}(L_n < \ell) = (1 - p^{\ell-1})^n$.
\[
\mathbb{E}[L_n] = \sum_{\ell=1}^{\infty} \left( 1 - (1 - p^{\ell-1})^n \right).
\]

\textbf{Total Expected Cost:}
The total expected cost $\mathbb{E}[C]$ is the sum of the expected horizontal and vertical costs:
\[
\mathbb{E}[C] = \sum_{\ell=1}^{\infty} \left( 1 - (1 - p^{\ell-1})^n \right) + \frac{1-p}{p} \sum_{\ell=1}^{\infty} \left( 1 - (1 - p^\ell)^k \right).
\]
To interpret this, recall that $\sum_{\ell=1}^{\infty} (1 - (1 - p^\ell)^N) \approx \log_{1/p} N + \gamma$. Thus, the expectation is precisely:
\[
\mathbb{E}[C] \approx \log_{1/p} n + \frac{1-p}{p} \log_{1/p} k.
\]
For the standard case $p=1/2$, the factor $\frac{1-p}{p} = 1$, and the expected cost is approximately $\log_2 n + \log_2 k$. Since $k \le n$, this confirms the $O(\log n)$ bound.
\end{proof}

\subsubsection{Claude}
\begin{proof}
We determine the expected cost of a \textsc{Find} operation in a random skip list as precisely as possible by analyzing the probability that each element lies on the search path.

Let $S = \{x_1, x_2, \ldots, x_n\}$ be the set of elements in sorted order, and let $y$ be the query element. The search path consists of nodes $I_r(y), I_r-1(y), \ldots, I_1(y)$ where $I_i(y)$ denotes the node at level $i$ visited during the search. The cost is given by $\sum_{i=1}^{r}(1 + c(I_i(y)))$, where $c(I_i(y))$ is the number of forward moves at level $i$.

For each element $x_j \in S$, let $X_j$ be the indicator random variable that equals 1 if $x_j$ lies on the search path for $y$, and 0 otherwise. Then the total number of nodes on the search path (excluding sentinel nodes and vertical moves) is $\sum_{j=1}^{n} X_j$. The expected cost is related to the expected length of the search path.

We first characterize when $x_j$ appears on the search path. Consider the interval determined by $y$: if $y < x_1$, we consider the elements less than $x_1$; if $y > x_n$, we consider elements greater than $x_n$; otherwise, $y$ lies between $x_j$ and $x_{j+1}$ for some $j$.

Without loss of generality, assume $x_j \leq y < x_{j+1}$. An element $x_i$ with $i \leq j$ appears on the search path if and only if $x_i$ is the rightmost element to the left of $y$ at some level $\ell$ during the search. This occurs when $x_i$ has height at least $\ell$, and all elements $x_{i+1}, x_{i+2}, \ldots, x_j$ have height strictly less than $\ell$.

Let $h(x_i)$ denote the height of element $x_i$ in the skip list. Recall that $h(x_i)$ is geometrically distributed: $\mathbb{P}[h(x_i) = k] = (1/2)^k$ for $k \geq 1$.

For $i \leq j$, element $x_i$ is on the search path if and only if $h(x_i) > h(x_k)$ for all $k$ with $i < k \leq j$. That is, $x_i$ has the maximum height among $\{x_i, x_{i+1}, \ldots, x_j\}$.

Thus, $\mathbb{P}[X_i = 1] = \mathbb{P}[h(x_i) = \max\{h(x_i), h(x_{i+1}), \ldots, h(x_j)\}]$.

Since the heights are independent and identically distributed, by symmetry, each of the $j - i + 1$ elements in $\{x_i, x_{i+1}, \ldots, x_j\}$ is equally likely to have the maximum height (with ties broken appropriately). Therefore,
\[
\mathbb{P}[X_i = 1] = \frac{1}{j - i + 1}.
\]

The expected number of elements on the search path (to the left of $y$) is
\[
\mathbb{E}\left[\sum_{i=1}^{j} X_i\right] = \sum_{i=1}^{j} \mathbb{P}[X_i = 1] = \sum_{i=1}^{j} \frac{1}{j - i + 1} = \sum_{k=1}^{j} \frac{1}{k} = H_j,
\]
where $H_j$ is the $j$-th harmonic number.

Similarly, elements to the right of $y$ contribute to the search path. By a symmetric argument, if $y$ lies between $x_j$ and $x_{j+1}$, elements $x_{j+1}, x_{j+2}, \ldots, x_n$ contribute expected cost $H_{n-j}$.

However, we must account for the vertical structure. Each time we visit a node at level $\ell$, we incur cost $1$ for that level plus the number of forward moves. The total expected cost is dominated by the expected number of nodes visited.

More precisely, the expected cost of the search is
\[
\mathbb{E}\left[\sum_{i=1}^{r}(1 + c(I_i(y)))\right] = \mathbb{E}[\text{number of nodes visited}] + \mathbb{E}[r],
\]
where $r$ is the maximum height of any element.

Since $\mathbb{E}[r] = O(\log n)$ with high probability (the maximum height is $O(\log n)$), and the expected number of nodes visited horizontally is $H_j + H_{n-j} \leq 2H_n = 2\ln n + O(1)$, we obtain
\[
\mathbb{E}\left[\sum_{i=1}^{r}(1 + c(I_i(y)))\right] = 2\ln n + O(\log n) = 2\ln n + O(1),
\]
using the fact that $H_n = \ln n + \gamma + O(1/n)$, where $\gamma$ is the Euler-Mascheroni constant.

Therefore, the precise expectation is $2\ln n + O(1)$, or equivalently $(2\ln 2)\log_2 n + O(1) \approx 1.386\log_2 n + O(1)$.
\end{proof}

\subsection{Problem 15}

\subsubsection{Problem Statement}
\begin{theorem}
Let $S$ be a random skip list with $n$ elements. The expected cost of a FIND operation is $O(\log n)$. Moreover, using a Chernoff-type bound for the sum of geometrically distributed random variables, the cost of FIND is bounded by $O(\log n)$ with high probability. Similarly, the costs of INSERT and DELETE operations are bounded by $O(\log n)$ with high probability.
\end{theorem}

\subsubsection{GPT}
\begin{proof}
We are given that $S$ is a random skip list with $n$ elements, and the expected cost of a FIND operation is $O(\log n)$. We are tasked with proving that the cost of a FIND operation is bounded by $O(\log n)$ with high probability, using a Chernoff-type bound for the sum of geometrically distributed random variables. Additionally, we will prove a similar probability bound for the INSERT and DELETE operations.

 Part 1: FIND Operation

Let the cost of a FIND operation in the skip list be represented as the sum of the levels of the randomly selected nodes during the search. The levels of the nodes are independent random variables, with the level of each node being geometrically distributed. Specifically, the level of each node is given by a geometric random variable with success probability $p = 1/2$, i.e., the level $L$ of a node is distributed as $P(L = k) = (1/2)^{k}$ for $k \geq 0$.

Let the total cost of a FIND operation be the sum of the levels of the nodes encountered during the search. Since the expected level of a node is $E[L] = \sum_{k=0}^{\infty} k \cdot P(L = k) = 2$, the expected cost of a FIND operation is proportional to the number of nodes visited, which is $O(\log n)$.

We now apply a Chernoff-type bound to show that the cost of the FIND operation is bounded by $O(\log n)$ with high probability. Let $X_1, X_2, \dots, X_m$ be the independent random variables representing the levels of the nodes encountered during the search, where $m$ is the number of nodes visited in the search. The total cost of the FIND operation is then the sum $S = X_1 + X_2 + \dots + X_m$.

The expected total cost is given by:
\[
E[S] = E[X_1 + X_2 + \dots + X_m] = m \cdot E[X_1] = m \cdot 2 = 2m.
\]

Now, we apply the Chernoff bound. For any $\delta > 0$, the Chernoff bound states that:
\[
P(S \geq (1 + \delta) E[S]) \leq \exp\left(-\frac{\delta^2 E[S]}{2(1 + \delta)}\right).
\]
Since $m = O(\log n)$, we have $E[S] = O(\log n)$. Therefore, for any constant $\delta > 0$, the probability that the cost of the FIND operation exceeds $(1 + \delta) O(\log n)$ decays exponentially, proving that the cost is bounded by $O(\log n)$ with high probability.

 Part 2: INSERT Operation

For the INSERT operation, we consider the cost of inserting a new element into the skip list. The cost of the insertion is the number of levels of the nodes that are encountered during the search, plus the time to insert the new element, which is a constant. The number of nodes visited during the search is at most $O(\log n)$ with high probability, as shown in the FIND operation part. The insertion operation requires flipping a coin for each level of the new node, which is also a geometric random variable.

Let the total cost of an INSERT operation be the sum of the levels of the nodes encountered during the search, plus the cost of inserting the new element. The total cost of the INSERT operation is then $O(\log n)$ with high probability, as the geometric distribution of the node levels ensures that the expected cost remains $O(\log n)$, and the Chernoff bound applied to the sum of the random variables ensures that the probability of exceeding this bound decays exponentially.

 Part 3: DELETE Operation

For the DELETE operation, we consider the cost of deleting a node from the skip list. The DELETE operation first requires finding the node, which costs $O(\log n)$ with high probability by the results from the FIND operation. After the node is found, the deletion itself involves removing the node from all levels it is present in, which is a constant-time operation for each level. Since the number of levels a node is present in is also geometrically distributed, the total cost of the DELETE operation is the sum of the number of levels of the node being deleted.

The expected cost of the DELETE operation is $O(\log n)$, as the number of levels a node is present in is $O(\log n)$ with high probability. Applying the Chernoff bound to the sum of the geometrically distributed random variables representing the levels of the node being deleted shows that the cost is bounded by $O(\log n)$ with high probability.

Thus, we conclude that the cost of the FIND, INSERT, and DELETE operations in the skip list is bounded by $O(\log n)$ with high probability.
\end{proof}

\subsubsection{Grok}

\subsubsection{Gemini}
\begin{proof}
Let $S$ be a skip list containing $n$ elements. We assume the standard construction where an element at level $i$ is promoted to level $i+1$ with fixed probability $p$ (typically $p=1/2$), independently of other elements. Let $L$ denote the maximum level (height) of the skip list.

First, we establish a bound on the height $L$. The probability that a specific element reaches level $k$ is $p^k$. By the union bound, the probability that any of the $n$ elements reaches level $k$ is at most $n p^k$. Choosing $k = c \log_{1/p} n$ for some constant $c > 1$, we have $\mathbb{P}(L > c \log_{1/p} n) \le n p^{c \log_{1/p} n} = n \cdot n^{-c} = n^{1-c}$. Thus, $L = O(\log n)$ with high probability.

To analyze the cost of a FIND operation for a target element $x$, we analyze the search path in reverse, starting from $x$ at the bottom level (level 0) and moving towards the start node at the top level. At any node $u$ on level $i$ in this reverse path, we move "up" to level $i+1$ if $u$ was promoted to level $i+1$. Otherwise, we move "left" to the predecessor of $u$ at level $i$. Since promotions are independent, each step in the reverse path is an "up" move with probability $p$ and a "left" move with probability $1-p$, provided we have not reached the maximum level. The process terminates when we reach the maximum level of the structure or the start node.

Let $X$ be the random variable representing the length of this search path. The number of "up" moves is bounded by the height $L$. The total path length corresponds to the number of trials required to achieve $L$ successes (up moves) in a sequence of independent Bernoulli trials with success probability $p$. Thus, $X$ can be modeled as the sum of $L$ independent geometric random variables $G_1, \dots, G_L$, where $\mathbb{P}(G_i = k) = (1-p)^{k-1}p$.

The expected cost is $\mathbb{E}[X] = \mathbb{E}[\sum_{i=1}^L G_i] = \sum_{i=1}^L \mathbb{E}[G_i] = \frac{L}{p}$. Since $\mathbb{E}[L] = O(\log n)$, we have $\mathbb{E}[X] = O(\log n)$.

To prove the high probability bound, we utilize a Chernoff bound for the sum of geometric random variables. Let $X = \sum_{i=1}^m G_i$ where $m = c \log_{1/p} n$ is an upper bound on the height with high probability. We wish to bound $\mathbb{P}(X > (1+\delta)\mu)$ where $\mu = \mathbb{E}[X] = m/p$. For any $\delta > 0$, the tail bound is given by:
\[
\mathbb{P}(X \ge (1+\delta)\mu) \le \exp\left(-\frac{\delta^2}{2(1+\delta)}\mu\right).
\]
Substituting $\mu = \frac{c \log_{1/p} n}{p} = \Theta(\log n)$, we obtain:
\[
\mathbb{P}(X \ge (1+\delta)\mu) \le \exp\left(-\Omega(\log n)\right) = n^{-\Omega(1)}.
\]
By choosing the constant $c$ and $\delta$ appropriately, we ensure that the cost exceeds $O(\log n)$ with probability at most $n^{-\alpha}$ for any desired $\alpha$. Combining this with the low probability that the height $L$ exceeds $O(\log n)$, the total cost of FIND is $O(\log n)$ with high probability.

Finally, we consider the INSERT and DELETE operations. An INSERT operation requires a FIND to locate the predecessors at each level, followed by pointer updates. The number of pointer updates is equal to the level of the new node, which is a geometric random variable with expected value $1/(1-p) = O(1)$ and is $O(\log n)$ with high probability. Similarly, a DELETE operation requires a FIND to locate the element, followed by pointer updates proportional to the element's level. Since the search cost dominates the update cost, and the search cost is $O(\log n)$ with high probability, the costs of INSERT and DELETE are also $O(\log n)$ with high probability.
\end{proof}

\subsubsection{Claude}
\begin{proof}
We prove the theorem in several parts, establishing first the expected cost analysis and then the high probability bounds.

\textbf{Part 1: Structure of Skip Lists.}

Recall that a skip list is a probabilistic data structure consisting of multiple levels. The bottom level (level 0) contains all $n$ elements in sorted order. Each element at level $i$ appears at level $i+1$ with probability $p = 1/2$ independently. The maximum level is bounded by $L = O(\log n)$ with high probability.

\textbf{Part 2: Expected Cost of FIND Operation.}

A FIND operation begins at the top-left corner and traverses rightward and downward until finding the target or determining its absence. Let $C(n)$ denote the expected cost (number of comparisons) for a FIND operation in a skip list with $n$ elements.

Consider the search path in reverse, starting from the target position at level 0 and moving backward to the starting position. At each step moving backward from position $(i,j)$ (level $i$, position $j$), we either move up to $(i+1,j)$ or move left to $(i,j-1)$. Since each element appears at level $i+1$ with probability $1/2$ independently, when moving backward, we move up with probability $1/2$ and left with probability $1/2$.

Let $U$ denote the number of up moves and $L$ denote the number of left moves in the reverse path. The total cost is $C = U + L$. The up moves terminate when we reach the maximum level, which is $O(\log n)$ with high probability. Since $U = O(\log n)$, we analyze $L$ conditioned on reaching level $i$.

At each level $i$, the expected number of left moves is bounded by the expected number of steps in a random walk until moving up, which has geometric distribution with parameter $p = 1/2$. Thus, the expected number of left moves at each level is $\mathbb{E}[\text{left moves at level } i] = 1/p = 2$.

Since there are at most $L = O(\log n)$ levels, the expected total cost is
\[
\mathbb{E}[C] = \mathbb{E}[U] + \mathbb{E}[L] \leq O(\log n) + \sum_{i=0}^{L} 2 = O(\log n) + O(\log n) = O(\log n).
\]

\textbf{Part 3: High Probability Bounds Using Chernoff-Type Analysis.}

To establish high probability bounds, we analyze the random variables more carefully. Let $X_i$ denote the number of left moves at level $i$, where each $X_i$ follows a geometric distribution with parameter $p = 1/2$. The total number of left moves is $L = \sum_{i=0}^{h} X_i$, where $h$ is the height reached.

For geometric random variables with parameter $p$, we have $\mathbb{E}[X_i] = 1/p = 2$. The sum of geometric random variables can be analyzed using concentration inequalities. Specifically, for independent geometric random variables with parameter $p = 1/2$, the sum $S = \sum_{i=1}^{k} X_i$ satisfies
\[
\mathbb{P}[S \geq t] \leq \exp\left(-\frac{(t - 2k)^2}{4k}\right) \text{ for } t > 2k.
\]

With $k = O(\log n)$ levels and setting $t = c \log n$ for sufficiently large constant $c$, we obtain
\[
\mathbb{P}[L \geq c \log n] \leq \exp\left(-\frac{(c\log n - O(\log n))^2}{O(\log n)}\right) = \exp(-\Omega(\log^2 n / \log n)) = \exp(-\Omega(\log n)) = n^{-\Omega(1)}.
\]

The number of up moves $U$ equals the maximum level reached, which satisfies $\mathbb{P}[U \geq c \log n] = (1/2)^{c\log n} = n^{-c}$ for any constant $c > 0$, since each element reaches level $\ell$ with probability $2^{-\ell}$.

By union bound, the total cost $C = U + L$ satisfies
\[
\mathbb{P}[C \geq O(\log n)] \geq 1 - \mathbb{P}[U \geq c_1 \log n] - \mathbb{P}[L \geq c_2 \log n] \geq 1 - n^{-\Omega(1)}.
\]

Therefore, the FIND operation costs $O(\log n)$ with high probability $1 - n^{-\Omega(1)}$.

\textbf{Part 4: INSERT and DELETE Operations.}

The INSERT operation first performs a FIND to locate the insertion position, which costs $O(\log n)$ with high probability as established above. After finding the position, the new element is inserted at level 0 and promoted to higher levels with probability $1/2$ independently at each level. The expected number of promotions is $\sum_{i=0}^{\infty} (1/2)^i = 2$, and by the same concentration analysis as for the FIND operation, the number of levels reached is $O(\log n)$ with high probability. Thus, INSERT costs $O(\log n)$ with high probability.

The DELETE operation similarly begins with a FIND to locate the element to delete, costing $O(\log n)$ with high probability. Removing the element from all levels where it appears requires visiting at most $O(\log n)$ levels with high probability, each requiring $O(1)$ time. Therefore, DELETE costs $O(\log n)$ with high probability.

This completes the proof that FIND, INSERT, and DELETE operations all have expected cost $O(\log n)$ and cost $O(\log n)$ with high probability $1 - n^{-\Omega(1)}$.
\end{proof}

\subsection{Problem 16}

\subsubsection{Problem Statement}
\begin{theorem}[High Probability Bound on Space Requirement of Random Skip List]
Let $S$ be a set of size $n$. The space requirement of a random skip list for $S$ is $O(n)$ with high probability.
\end{theorem}

\subsubsection{GPT}
\begin{proof}
Fix a finite set $S$ with $\lvert S\rvert=n$. In the standard random skip list construction, to each $x\in S$ we assign an independent nonnegative integer height $H_x$ with tail $\Pr[H_x\ge t]=2^{-t}$ for all integers $t\ge 0$, that is, $\Pr[H_x=t]=2^{-(t+1)}$. The node for $x$ appears on levels $0,1,\dots,H_x$, so the total number of data nodes is
\[
T=\sum_{x\in S}(H_x+1).
\]
Sentinel/header nodes contribute at most $\max_{x\in S}H_x+1$ additional nodes. We first prove a linear upper bound for $T$ with exponentially high probability, and then observe that the sentinel contribution is negligible.

Let $Y_x:=H_x+1$. Then the $\{Y_x\}_{x\in S}$ are i.i.d., $Y_x\in\{1,2,\dots\}$, and $\mathbb{E}Y_x=2$. For $\lambda\in(0,\ln 2)$ the moment generating function of $Y_x$ is finite and equals
\[
\mathbb{E}\!\left[e^{\lambda Y_x}\right]=\sum_{t=0}^\infty e^{\lambda(t+1)}2^{-(t+1)}=\frac{e^{\lambda}}{2-e^{\lambda}}.
\]
By Markov's inequality, for any $a>0$ and any $\lambda\in(0,\ln 2)$,
\[
\Pr\!\left[T\ge an\right]=\Pr\!\left[e^{\lambda T}\ge e^{\lambda an}\right]\le e^{-\lambda an}\,\mathbb{E}\!\left[e^{\lambda T}\right]
= e^{-\lambda an}\prod_{x\in S}\mathbb{E}\!\left[e^{\lambda Y_x}\right]
= \left(e^{-\lambda a}\,\frac{e^{\lambda}}{2-e^{\lambda}}\right)^{n}.
\]
Choose $\lambda=\ln(3/2)$, which satisfies $0<\lambda<\ln 2$, and choose $a=3$. Then
\[
e^{-\lambda a}\,\frac{e^{\lambda}}{2-e^{\lambda}}
= \frac{3}{(3/2)^{2+1}(2-3/2)}
= \frac{3}{(27/8)\cdot(1/2)}
= \frac{8}{9}<1,
\]
hence
\[
\Pr\!\left[T\ge 3n\right]\le \left(\frac{8}{9}\right)^{n}.
\]
In particular, for every fixed $c>0$ there exists $N(c)$ such that for all $n\ge N(c)$ one has $\left(\frac{8}{9}\right)^{n}\le n^{-c}$. Consequently,
\[
\Pr\!\left[T=O(n)\right]\ge 1-n^{-c}.
\]

It remains to account for the sentinel/header nodes. Let $H_{\max}:=\max_{x\in S}H_x$. By a union bound and the tail of $H_x$, for any integer $L\ge 0$,
\[
\Pr\!\left[H_{\max}\ge L\right]\le n\,2^{-L}.
\]
Taking $L=\lceil 3\log_2 n\rceil$ gives $\Pr\!\left[H_{\max}\ge L\right]\le n\cdot n^{-3}=n^{-2}$, so with probability at least $1-n^{-2}$ we have $H_{\max}\le 3\log_2 n$, and thus the number of sentinel/header nodes is $O(\log n)$, which is absorbed by the linear bound on $T$.

Combining the two bounds and applying a union bound on the failure events, we conclude that with probability at least $1-n^{-c}-n^{-2}\ge 1-2n^{-c}$ for all sufficiently large $n$, the total number of nodes in the skip list, including sentinels, is at most $C n$ for some absolute constant $C$. Since pointers per node are bounded by its level plus a constant and the total number of nodes is $O(n)$ with high probability, the overall space required by the random skip list is $O(n)$ with high probability, as claimed.
\end{proof}

\subsubsection{Grok}
\begin{proof}
Consider a random skip list built on a set $S$ of size $n$. The skip list consists of $n$ elements ordered by their keys, and each element $i=1,\dots,n$ is assigned an independent random height $H_i$ with
\[
\Pr(H_i = k) = \frac{1}{2^k}, \quad k=1,2,\dots.
\]
The space requirement of the skip list is the total number of node instances across all levels, given by
\[
X = \sum_{i=1}^n H_i.
\]
The expected height of each element is
\[
\mathbb{E}[H_i] = \sum_{k=1}^\infty \Pr(H_i \ge k) = \sum_{k=1}^\infty \frac{1}{2^{k-1}} = \sum_{\ell=0}^\infty \frac{1}{2^\ell} = 2,
\]
where $\ell = k-1$. Thus, $\mathbb{E}[X] = 2n$. Let $\mu = 2n$.

The variance of $H_i$ is computed as follows. The distribution of $H_i$ is geometric starting at $1$ with success probability $1/2$, so
\[
Var(H_i) = \frac{1/2}{(1/2)^2} = 2.
\]
Since the $H_i$ are independent,
\[
Var(X) = n \cdot 2 = 2n.
\]

By Chebyshev's inequality, for any $\lambda > 0$,
\[
\Pr(X > \mu + \lambda) \le \frac{Var(X)}{\lambda^2} = \frac{2n}{\lambda^2}.
\]
Choose $c = 4$ and set $\lambda = (c - 2)n = 2n$. Then
\[
\Pr(X > 4n) \le \frac{2n}{(2n)^2} = \frac{2n}{4n^2} = \frac{1}{2n} \le \frac{1}{n}.
\]
Thus, the constant $c = 4 > 0$ satisfies the required condition.
\end{proof}

\subsubsection{Gemini}
\begin{proof}
Let $S$ be a set of $n$ elements stored in a skip list. We define the space requirement of the skip list as the total number of nodes across all levels. Let $p$ denote the probability that an element at level $i$ is promoted to level $i+1$, where $0 < p < 1$. Typically $p = 1/2$. Let $q = 1-p$ be the probability that the tower for a given element stops at a specific level.

For each element $x_j \in S$ (where $j \in \{1, \dots, n\}$), let $H_j$ be the random variable representing the height of the tower corresponding to $x_j$. The height $H_j$ corresponds to the number of levels in which $x_j$ appears. The construction of the skip list implies that each level of a tower is generated by an independent Bernoulli trial with success probability $p$ (promotion) and failure probability $q$ (stopping). Thus, $H_j$ follows a geometric distribution with parameter $q$, supported on $\{1, 2, \dots\}$. The probability mass function is given by $P(H_j = k) = p^{k-1}q$ for $k \ge 1$.

The total space requirement, denoted by the random variable $X$, is the sum of the heights of all elements:
\[ X = \sum_{j=1}^n H_j. \]
Since the heights $H_j$ are independent and identically distributed geometric random variables, the sum $X$ follows a negative binomial distribution. Specifically, $X$ represents the number of trials required to observe $n$ successes (stops) in a sequence of independent Bernoulli trials with success probability $q$.

To bound $X$ with high probability, we analyze the probability that $X$ exceeds a constant multiple of $n$. Let $k$ be an integer such that $k > n$. The event $X \ge k$ is equivalent to the event that in the first $k-1$ independent Bernoulli trials, there are fewer than $n$ successes. Let $Y_{k-1}$ be a random variable following the binomial distribution $B(k-1, q)$, representing the number of successes in $k-1$ trials. We have the equivalence:
\[ P(X \ge k) = P(Y_{k-1} < n). \]
We choose $k = c n$ for some constant $c$ to be determined later. We aim to bound $P(Y_{cn-1} < n)$. For large $n$, we approximate $cn-1 \approx cn$. Let $Y \sim B(cn, q)$. The expected value of $Y$ is $\mu = E[Y] = cnq$.

We apply the multiplicative Chernoff bound for the lower tail of a binomial distribution. For any $0 < \delta < 1$:
\[ P(Y \le (1-\delta)\mu) \le \exp\left(-\frac{\delta^2 \mu}{2}\right). \]
We set the condition that the threshold $n$ corresponds to $(1-\delta)\mu$. Thus, $n = (1-\delta)cnq$. Solving for $\delta$, we require $1-\delta = \frac{1}{cq}$, which implies $\delta = 1 - \frac{1}{cq}$. For $\delta$ to be in the range $(0, 1)$, we must choose $c$ such that $cq > 1$, or $c > 1/q$. Let us choose $c = \frac{2}{q}$. Substituting this into the expression for $\delta$:
\[ \delta = 1 - \frac{1}{(2/q)q} = 1 - \frac{1}{2} = \frac{1}{2}. \]
With this choice of $c$, the expected value is $\mu = \frac{2}{q} n q = 2n$. The deviation is from the mean $2n$ down to $n$. Applying the Chernoff bound:
\[ P(X \ge \frac{2}{q}n) = P(Y < n) = P\left(Y < \left(1-\frac{1}{2}\right)2n\right) \le \exp\left(-\frac{(1/2)^2 (2n)}{2}\right) = \exp\left(-\frac{n}{4}\right). \]
The probability that the space requirement exceeds $\frac{2}{q}n$ is bounded by $e^{-n/4}$. For any $\alpha > 0$, there exists an $n_0$ such that for all $n > n_0$, $e^{-n/4} < n^{-\alpha}$. Thus, $P(X = O(n)) \ge 1 - n^{-\alpha}$ for arbitrarily large $\alpha$.

Therefore, the space requirement of a random skip list is $O(n)$ with high probability.
\end{proof}

\subsubsection{Claude}
\begin{proof}
We prove that the space requirement of a random skip list for a set $S$ of size $n$ is $O(n)$ with high probability.

Recall that a skip list is a probabilistic data structure consisting of multiple levels. Each element in $S$ appears at level $0$, and for each element at level $i$, it appears at level $i+1$ with probability $p = 1/2$ independently. The space requirement is determined by the total number of nodes across all levels.

Let $X_i$ denote the indicator random variable that is $1$ if element $i$ appears at a given level $\ell \geq 1$, and $0$ otherwise. For a fixed element and level $\ell \geq 1$, we have $\mathbb{P}(X_i = 1) = p^\ell = (1/2)^\ell$ since the element must be promoted $\ell$ times to reach level $\ell$.

Let $N_\ell$ denote the number of nodes at level $\ell$. Then $N_\ell = \sum_{i=1}^{n} X_i$ where the $X_i$ are independent Bernoulli random variables with parameter $p^\ell = (1/2)^\ell$. We have $\mathbb{E}[N_\ell] = n \cdot (1/2)^\ell = n/2^\ell$.

The total number of nodes in the skip list is $N = n + \sum_{\ell=1}^{\infty} N_\ell$, where the $n$ accounts for all elements at level $0$.

Taking expectations, we obtain
\[
\mathbb{E}[N] = n + \sum_{\ell=1}^{\infty} \mathbb{E}[N_\ell] = n + \sum_{\ell=1}^{\infty} \frac{n}{2^\ell} = n + n \sum_{\ell=1}^{\infty} \frac{1}{2^\ell} = n + n \cdot 1 = 2n.
\]

Now we prove concentration around this expectation. For any fixed level $\ell$, by the Chernoff bound for the sum of independent Bernoulli random variables, for any $\delta > 0$, we have
\[
\mathbb{P}(N_\ell \geq (1+\delta)\mathbb{E}[N_\ell]) \leq \exp\left(-\frac{\delta^2 \mathbb{E}[N_\ell]}{2+\delta}\right).
\]

Let $L = c\log_2 n$ for some constant $c > 1$ to be determined. For levels $\ell > L$, we have $\mathbb{E}[N_\ell] = n/2^\ell < n/n^c = n^{1-c}$. By Markov's inequality, for $\ell > L$,
\[
\mathbb{P}(N_\ell \geq 1) \leq \mathbb{E}[N_\ell] = \frac{n}{2^\ell} < \frac{1}{n^{c-1}}.
\]

By the union bound over all levels $\ell > L$,
\[
\mathbb{P}\left(\exists \ell > L : N_\ell \geq 1\right) \leq \sum_{\ell=L+1}^{\infty} \frac{n}{2^\ell} \leq \frac{n}{2^L} \cdot \frac{1}{1-1/2} = \frac{2n}{2^{c\log_2 n}} = \frac{2}{n^{c-1}}.
\]

For $c = 3$, this probability is at most $2/n^2$.

For levels $\ell \leq L$, we bound $\sum_{\ell=1}^{L} N_\ell$. Setting $\delta = 1$, by the Chernoff bound applied to each level,
\[
\mathbb{P}(N_\ell \geq 2\mathbb{E}[N_\ell]) \leq \exp\left(-\frac{\mathbb{E}[N_\ell]}{3}\right) = \exp\left(-\frac{n}{3 \cdot 2^\ell}\right).
\]

By the union bound over levels $1 \leq \ell \leq L$,
\[
\mathbb{P}\left(\exists \ell \leq L : N_\ell \geq 2\mathbb{E}[N_\ell]\right) \leq \sum_{\ell=1}^{L} \exp\left(-\frac{n}{3 \cdot 2^\ell}\right) \leq L \cdot \exp\left(-\frac{n}{3 \cdot 2^L}\right) \leq c\log n \cdot \exp\left(-\frac{1}{3n^{c-1}}\right).
\]

For $c = 3$ and sufficiently large $n$, this is at most $3\log n \cdot \exp(-1/(3n^2))$, which is negligible.

Therefore, with probability at least $1 - O(1/n^2)$, we have $N_\ell = 0$ for all $\ell > L$ and $N_\ell \leq 2\mathbb{E}[N_\ell]$ for all $\ell \leq L$. In this case,
\[
N = n + \sum_{\ell=1}^{L} N_\ell \leq n + \sum_{\ell=1}^{L} 2\mathbb{E}[N_\ell] = n + 2\sum_{\ell=1}^{L} \frac{n}{2^\ell} \leq n + 2n\sum_{\ell=1}^{\infty} \frac{1}{2^\ell} = n + 2n = 3n.
\]

Thus, the space requirement is $O(n)$ with high probability $1 - O(1/n^2)$.
\end{proof}

\subsection{Problem 17}

\subsubsection{Problem Statement}
\begin{problem}
Consider a skip list where elements from level $L_i$ are sampled with probability $p$ (where $0 < p < 1$) to determine the next level $L_{i+1}$, instead of the standard probability $\frac{1}{2}$.

\begin{enumerate}
\item[(a)] Determine the expectation of the number of levels $r$, and prove a high probability bound on the value of $r$.

\item[(b)] Determine as precisely as possible the expected cost of each operation in this skip list.

\item[(c)] Discuss the relation between the choice of the value $p$ and the performance of the skip list in practice.
\end{enumerate}
\end{problem}

\subsubsection{GPT}
\begin{proof}

\textbf{Part (a):} Determining the expectation of the number of levels $r$ and proving a high probability bound.

Let $r$ be the number of levels in the skip list. At each level $L_i$, the probability of an element being promoted to the next level $L_{i+1}$ is $p$, where $0 < p < 1$. We want to determine the expected value of $r$, which is the level at which the first non-promoted element appears, and then bound this value with high probability.

The probability that an element is promoted to level $i$ is $p^i$. Therefore, the probability that an element is promoted to level $r$ and no higher is $p^{r-1} \cdot (1-p)$. The number of levels $r$ is the smallest integer such that the element is no longer promoted, i.e., it fails to be promoted to level $r$.

The expected number of levels can be determined by recognizing that $r$ follows a geometric distribution with success probability $1 - p$. The expectation of the number of levels is given by:

\[
\mathbb{E}[r] = \frac{1}{1 - p}
\]

Next, we bound $r$ with high probability. To do this, consider the event that the number of levels exceeds some value $k$. The probability that $r \geq k$ is the probability that the element is promoted to at least $k$ levels, which is $p^{k-1}$:

\[
\mathbb{P}(r \geq k) = p^{k-1}
\]

By Markov’s inequality, we can bound the probability that $r$ exceeds $k$ with the following inequality:

\[
\mathbb{P}(r \geq k) \leq \frac{\mathbb{E}[r]}{k}
\]

Thus, with high probability, we have that $r$ is bounded by:

\[
r = O\left(\frac{1}{1 - p}\right)
\]

\textbf{Part (b):} Determining the expected cost of each operation.

The cost of each operation is determined by the number of levels $r$ that an element is traversed during the search, insertion, or deletion operation. Since the search or insertion operation traverses the list from the topmost level down to level $r$, the cost of the operation is proportional to the number of levels, which is $r$.

Since $r$ follows a geometric distribution, the expected cost of the operation is the expected value of $r$. From part (a), we know that:

\[
\mathbb{E}[\text{cost}] = \mathbb{E}[r] = \frac{1}{1 - p}
\]

Thus, the expected cost of each operation in this skip list is:

\[
\boxed{\frac{1}{1 - p}}
\]

\textbf{Part (c):} Discussing the relation between the choice of the value $p$ and the performance of the skip list in practice.

The choice of $p$ has a direct impact on the performance of the skip list. As $p$ increases, the probability of an element being promoted to the next level increases, which results in fewer levels on average. This reduces the expected cost of search, insertion, and deletion operations. However, when $p$ approaches 1, the skip list becomes essentially a linked list, as all elements are promoted to all levels, and the performance of the skip list degrades to that of a simple list.

On the other hand, if $p$ is too small, the skip list becomes sparse, with very few elements promoted to higher levels, and the number of levels increases, leading to higher costs for search, insertion, and deletion operations. Therefore, choosing an appropriate value of $p$ is crucial for achieving a balance between space efficiency (in terms of levels) and time efficiency (in terms of operation costs).

In practice, a value of $p$ near 1/2 is often chosen, as this provides a good balance between reducing the number of levels and maintaining efficient operations.

\end{proof}

\subsubsection{Grok}
\begin{proof}
(a) Expected Number of Levels and High-Probability Bound

Consider a skip list with $n$ elements, where each element $i=1,\dots,n$ has a height $H_i$ that is independently geometrically distributed with success probability $1-p$, so $P(H_i = k) = (1-p)p^{k-1}$ for integers $k \geq 1$. The survival function is
\[
P(H_i \geq k) = \sum_{j=k}^\infty (1-p)p^{j-1} = p^{k-1},
\]
which follows by telescoping the geometric series.

The number of levels $r$ is the maximum height: $r = \max_{i=1}^n H_i$. Thus,
\[
P(r < k) = P(H_1 < k, \dots, H_n < k) = [P(H_1 < k)]^n = [1 - P(H_1 \geq k)]^n = [1 - p^{k-1}]^n,
\]
and so
\[
P(r \geq k) = 1 - [1 - p^{k-1}]^n.
\]
Since $r$ is a positive integer-valued random variable,
\[
E[r] = \sum_{k=1}^\infty P(r \geq k) = \sum_{k=1}^\infty \bigl[1 - (1 - p^{k-1})^n \bigr].
\]
This expression is exact.

For the approximation, observe that for large $n$, the terms $P(r \geq k)$ are close to $1$ for $k \ll \log_{1/p} n$ and decay rapidly for $k \gg \log_{1/p} n$. Specifically, let $h = \log_{1/p} n$, so $p^h = 1/n$. For $k-1 \leq h - c$ with constant $c>0$, $p^{k-1} \geq p^{h-c} = n^{-1} p^{-c} \gg 1/n$, so $(1 - p^{k-1})^n \approx e^{-n p^{k-1}} \ll 1$ and $P(r \geq k) \approx 1$. For $k-1 \geq h + c$, $n p^{k-1} \leq n p^{h+c} = p^c \ll 1$, so $P(r \geq k) \approx n p^{k-1}$ by the approximation $1 - (1-x)^n \approx n x$ for small $x$. The sum is thus approximately the number of terms where $P(r \geq k) \approx 1$, which is $\Theta(\log_{1/p} n)$, plus a tail $\sum_{k > h} n p^{k-1} = O(1)$. Hence, $E[r] = \log_{1/p} n + O(1)$.

For the high-probability bound, apply the union bound:
\[
P(r \geq k) = P\biggl( \bigcup_{i=1}^n \{H_i \geq k\} \biggr) \leq \sum_{i=1}^n P(H_i \geq k) = n p^{k-1}.
\]
Fix $0 < \delta < 1$ and set $k = \lceil \log_{1/p} (n / \delta) + 1 \rceil$. Then $k-1 \geq \log_{1/p} (n / \delta)$, so $p^{k-1} \leq \delta / n$ and $n p^{k-1} \leq \delta$. Thus, $P(r \geq k) \leq \delta$, or equivalently, $P(r < k) \geq 1 - \delta$. To verify the form, solve $n p^{k-1} = \delta$ exactly: $p^{k-1} = \delta / n$, so $k-1 = \log_p (\delta / n) = -\log_{1/p} (n / \delta)$ and $k = 1 + \log_{1/p} (n / \delta)$. The ceiling ensures the inequality holds, yielding $r \leq \log_{1/p} n + \log_{1/p} (1/\delta) + 1$ with probability at least $1 - \delta$.

(b) Expected Cost of Each Operation

The cost of a search, insert, or delete operation is proportional to the number of nodes visited along the search path. We analyze the forward traversal starting from the header node, which is assumed to extend infinitely upward for analytical convenience (the finite maximum height $r$ affects only lower-order terms). At level $\ell \geq 0$, the horizontal distance (in nodes at level $\ell-1$) between consecutive nodes present at level $\ell$ follows a $\text{Geometric}(p)$ distribution, as each intervening node is absent at level $\ell$ independently with probability $1-p$. The expected horizontal steps per level traversed is thus $1/p$.

A search path traverses a level $\ell$ if it reaches that level, which occurs with probability $P(r \geq \ell)$. The total expected path length is at most
\[
E\biggl[ \sum_{\ell=0}^{r} \frac{1}{p} \biggr] = \frac{1}{p} E[r+1] = \frac{1}{p} (\log_{1/p} n + O(1)).
\]
A tighter analysis accounts for the infinite header tail explicitly. Consider the expected cost as the sum of the cost to ``climb'' from level $0$ to approximately height $h = \log_{1/p} n$ plus the traversal cost in the tail above $h$.

The climbing cost follows from the recurrence for vertical progress: to advance vertically by one level, the expected horizontal cost is $1/p$, and the number of vertical steps is approximately $h$, yielding $(1/p) h = (1/p) \log_{1/p} n$. More rigorously, the expected number of levels descended (equal to the number traversed) is $E[r+1] \approx \log_{1/p} n + O(1)$, so the climbing cost is $(1/p) (\log_{1/p} n + O(1))$.

For the tail above height $h$, the probability of traversing level $\ell > h$ is approximately $P(r \geq \ell) \approx n p^{\ell-1} = p^{\ell-1-h}$, since $n p^{h} \approx 1$. The expected tail cost is
\[
\sum_{\ell = h+1}^\infty \frac{1}{p} \cdot P(\text{traverse level } \ell) \approx \sum_{\ell = h+1}^\infty \frac{1}{p} \cdot p^{\ell-1-h} = \frac{1}{p} p^{-h} \sum_{m=0}^\infty p^{m} = \frac{1}{p} p^{-h} \cdot \frac{1}{1-p},
\]
where $m = \ell - h - 1$. But $p^{-h} = n$, wait no: $h = \log_{1/p} n$, so $p^h = n^{-1}$ and $p^{-h} = n$. However, the approximation $P(r \geq \ell) \approx n p^{\ell-1}$ for $\ell > h$ gives
\[
\sum_{\ell=0}^\infty \frac{1}{p} n p^{\ell-1} = \frac{n}{p} p^{-1} \sum_{\ell=0}^\infty p^{\ell} = \frac{n}{p^2} \cdot \frac{1}{1-p},
\]
but this overcounts the main body. Instead, the full expected path length in the infinite model is the sum over all levels $\ell \geq 0$ of $(1/p) P(\text{path reaches level } \ell)$. In the backward analysis (equivalently capturing the forward infinite tail), the expected number of steps is
\[
\sum_{j=1}^\infty P(\text{$j$-th node to the right of search key is examined}) \leq \sum_{j=1}^\infty \sum_{\ell=0}^\infty p^\ell = \sum_{j=1}^\infty \frac{1}{1-p} = \infty,
\]
but refined, the probability that the $j$-th right node is examined at level $\ell$ is bounded, yielding the standard result. The precise bound, combining the main body and tail, is
\[
E[\text{path length}] \leq \frac{1}{p} \bigl( \log_{1/p} n + 1 \bigr) + \frac{1}{1-p}.
\]
The $+1$ absorbs constants from $E[r+1] \leq \log_{1/p} n + O(1)$.

For insert and delete, the cost is the search cost plus $O(1)$ pointer updates per level in the tower of the inserted/deleted node, whose expected height is
\[
E[H_i] = \sum_{k=1}^\infty k (1-p) p^{k-1} = \frac{1}{1-p}.
\]
Thus, the additional cost is $O(1/(1-p))$, and the total expected cost matches that of search up to constants.

(c) Relation Between $p$ and Performance

The expected search time is $\Theta\bigl( (1/p) \log_{1/p} n \bigr)$. To minimize asymptotically, let $b = 1/p > 1$ and consider $f(b) = b \log_b n = b (\ln n / \ln b)$. The derivative is
\[
f'(b) = \frac{\ln n}{\ln b} + b \cdot \frac{\ln n}{(\ln b)^2} \cdot \biggl( -\frac{1}{b} \biggr) = \frac{\ln n}{\ln b} \biggl( 1 - \frac{1}{\ln b} \biggr).
\]
Setting $f'(b) = 0$ gives $\ln b = 1$, so $b = e$ and $p = 1/e \approx 0.3679$. The second derivative $f''(b) > 0$ at this point confirms a minimum.

The space usage is proportional to the total number of pointers: each of the $n$ nodes has expected height $1/(1-p)$, so total space is $\Theta\bigl( n / (1-p) \bigr)$, which diverges as $p \to 1^-$.

For $p=1/2$, the height term is $\log_2 n \approx 1.442 \log_{1/p} n$ (since $\log_{1/(1/2)} n = \log_2 n$), horizontal steps $1/p=2$, and space factor $1/(1-1/2)=2$. For $p=1/4$, height $\log_4 n = (1/2) \log_2 n$, horizontal $4$, space $1/(3/4)=4/3 \approx 1.333$. The tradeoff balances time and space, with practical $p \in [0.25, 0.5]$ optimizing cache and overheads beyond asymptotics.
\end{proof}

\subsubsection{Gemini}
\begin{proof}
\noindent \textbf{(a) Expectation and High Probability Bound of the Number of Levels}

Let $n$ be the number of elements in the skip list. We define the levels $0, 1, 2, \dots$, where level $0$ contains all $n$ elements. For any element $x$ at level $i$, it is promoted to level $i+1$ with probability $p$, independently of other elements. Let $H_x$ be the height of element $x$, defined as the maximum level index containing $x$. The random variable $H_x$ follows a geometric distribution (shifted by 1 depending on definition, here we assume $H_x \ge 0$) such that $P(H_x \ge k) = p^k$ for integers $k \ge 0$.

The number of levels in the skip list, denoted by $r$, is the maximum height among all elements: $r = \max_{x} H_x$.
To determine the expectation $E[r]$, we use the tail sum formula for expectation:
\[
E[r] = \sum_{k=1}^{\infty} P(r \ge k).
\]
Using the union bound, we have:
\[
P(r \ge k) = P\left(\bigcup_{x} \{H_x \ge k\}\right) \le \sum_{x} P(H_x \ge k) = n p^k.
\]
Let $L = \log_{1/p} n$. We split the summation for $E[r]$ at $\lceil L \rceil$:
\[
E[r] = \sum_{k=1}^{\lceil L \rceil} P(r \ge k) + \sum_{k=\lceil L \rceil + 1}^{\infty} P(r \ge k).
\]
Since probabilities are at most 1, the first term is bounded by $\lceil L \rceil$. For the second term, we use the union bound:
\[
\sum_{k=\lceil L \rceil + 1}^{\infty} P(r \ge k) \le \sum_{k=\lceil L \rceil + 1}^{\infty} n p^k = n p^{\lceil L \rceil + 1} \sum_{j=0}^{\infty} p^j = n p^{\lceil L \rceil + 1} \frac{1}{1-p}.
\]
Since $p^{\lceil L \rceil} \le p^L = 1/n$, we have $n p^{\lceil L \rceil + 1} \le n (1/n) p = p$. Thus:
\[
E[r] \le \lceil L \rceil + \frac{p}{1-p} \le \log_{1/p} n + 1 + \frac{p}{1-p} = \log_{1/p} n + O(1).
\]
Therefore, $E[r] = O(\log_{1/p} n)$.

To prove a high probability bound, consider $k = c \log_{1/p} n$ for some constant $c > 1$.
\[
P(r \ge c \log_{1/p} n) \le n p^{c \log_{1/p} n} = n (p^{\log_{1/p} n})^c = n (1/n)^c = n^{1-c}.
\]
For any $\alpha > 0$, choosing $c = \alpha + 1$ yields $P(r \ge (\alpha+1) \log_{1/p} n) \le n^{-\alpha}$. Thus, $r = O(\log_{1/p} n)$ with high probability.

\vspace{1em}
\noindent \textbf{(b) Expected Cost of Operations}

The cost of operations (search, insert, delete) is dominated by the search path length. We analyze the search path using the standard backwards analysis. Consider the path from the target element at the bottom level back to the starting sentinel at the top level. In the forward direction, the search moves right or down. In the reverse direction, we move left or up.

Let the current position in the reverse path be at level $i$. We move "up" to level $i+1$ if the current element was promoted to level $i+1$. We move "left" otherwise.
The probability of moving "up" is $p$, and the probability of moving "left" is $1-p$. This assumes we are not at the top level yet.
Let $C(k)$ be the expected number of steps to reach the top level from level $k$. We can model this as a recurrence or a sum of geometric variables.
The number of steps required to ascend one level follows a geometric distribution with success probability $p$. The expected number of steps to climb one level is $1/p$.
Since the height of the skip list is $r$, the total expected cost $E[C]$ is approximately the expected height multiplied by the expected steps per level:
\[
E[C] \approx E[r] \cdot \frac{1}{p}.
\]
Substituting the result from (a):
\[
E[C] \approx \frac{1}{p} \log_{1/p} n = \frac{1}{p} \frac{\ln n}{\ln(1/p)}.
\]
More formally, let $X_i$ be the number of additional comparisons performed at level $i$ before moving up to level $i+1$. $X_i$ follows a geometric distribution with parameter $p$, so $E[X_i] = 1/p$. The total expected cost is bounded by $\sum_{i=0}^{r-1} E[X_i]$. Since $r$ is a random variable, a rigorous bound considers the infinite series weighted by the probability of the list having that height, but asymptotically the cost is:
\[
\text{Expected Cost} = O\left( \frac{\log n}{p \log(1/p)} \right).
\]

\vspace{1em}
\noindent \textbf{(c) Relation between $p$ and Performance}

The choice of $p$ affects both the space complexity and the time complexity.

\noindent \textit{Space Complexity:} The expected number of pointers per element is given by the sum $\sum_{i=0}^\infty p^i = \frac{1}{1-p}$.
The total expected space is $S(n) = \frac{n}{1-p}$.
\begin{itemize}
    \item As $p \to 0$, space approaches $n$ (minimal overhead).
    \item As $p \to 1$, space approaches $\infty$.
\end{itemize}

\noindent \textit{Time Complexity:} From part (b), the expected search time is proportional to $T(p) = \frac{1}{p \ln(1/p)}$.
To find the optimal $p$ for speed, we minimize $T(p)$, which is equivalent to maximizing the denominator $f(p) = p \ln(1/p) = -p \ln p$.
Taking the derivative with respect to $p$:
\[
f'(p) = -\ln p - 1.
\]
Setting $f'(p) = 0$ yields $\ln p = -1$, or $p = e^{-1} \approx 0.368$.
\begin{itemize}
    \item If $p$ is very small, the height $\log_{1/p} n$ is small, but the scan length per level $1/p$ is large, increasing total time.
    \item If $p$ is very large (close to 1), the scan length per level is small, but the height $\log_{1/p} n$ becomes very large, increasing total time.
    \item The theoretical optimal search performance is achieved at $p = 1/e$.
\end{itemize}

In practice, $p=1/2$ is commonly used because it offers a good trade-off: the space overhead is $2n$ (reasonable), and the time factor is $\frac{1}{0.5 \ln 2} \approx 2.88 \ln n$. For $p=1/4$, the space overhead drops to $1.33n$, while the time factor becomes $\frac{1}{0.25 \ln 4} = \frac{1}{0.5 \ln 2}$, which is the same as $p=1/2$ in the constant factor of the leading term, though lower order terms differ. Generally, smaller $p$ (e.g., $1/4$) saves space with minimal impact on speed, while $p=1/2$ is slightly faster on average due to lower variance in level heights.
\end{proof}

\subsubsection{Claude}
\begin{proof}
We analyze a skip list with $n$ elements where each element at level $L_i$ is promoted to level $L_{i+1}$ with probability $p \in (0,1)$.

\textbf{Part (a): Expected number of levels and high probability bound.}

Let $r$ denote the maximum level (height) of the skip list. For a skip list with $n$ elements, an element reaches level $i$ with probability $p^i$. The expected number of elements at level $i$ is $np^i$.

The maximum level $r$ is determined by the highest level containing at least one element. We have $r = \max\{i : \text{at least one element reaches level } i\}$.

To find $\mathbb{E}[r]$, we use the fact that the probability an element reaches level $i$ or higher is $p^i$. The probability that at least one of the $n$ elements reaches level $i$ is $1 - (1-p^i)^n$.

Therefore, the expected maximum level is
\[
\mathbb{E}[r] = \sum_{i=0}^{\infty} \mathbb{P}(r \geq i) = \sum_{i=0}^{\infty} \left(1 - (1-p^i)^n\right).
\]

For large $n$, when $p^i \ll 1$, we have $(1-p^i)^n \approx e^{-np^i}$. The sum is dominated by values of $i$ where $np^i \approx 1$, which gives $i \approx \log_p(1/n) = \frac{\log n}{|\log p|}$.

More precisely, we establish that $\mathbb{E}[r] = O\left(\frac{\log n}{\log(1/p)}\right)$.

For the high probability bound, let $c > 1$ be a constant. Consider level $i_0 = \left\lceil \frac{c \log n}{\log(1/p)} \right\rceil$. The probability that any element reaches level $i_0$ is
\[
\mathbb{P}(r \geq i_0) \leq n \cdot p^{i_0} \leq n \cdot p^{\frac{c \log n}{\log(1/p)}} = n \cdot \left(p^{\frac{1}{\log(1/p)}}\right)^{c \log n} = n \cdot e^{-c \log n} = n^{1-c}.
\]

Therefore, with high probability $1 - n^{1-c}$, we have
\[
r \leq \frac{c \log n}{\log(1/p)}.
\]

Conversely, for $c' < 1$, at level $i_1 = \left\lfloor \frac{c' \log n}{\log(1/p)} \right\rfloor$, the expected number of elements is $np^{i_1} \geq n^{1-c'} \to \infty$, so with high probability $r \geq \frac{c' \log n}{\log(1/p)}$.

Thus, $r = \Theta\left(\frac{\log n}{\log(1/p)}\right)$ with high probability.

\textbf{Part (b): Expected cost of operations.}

We analyze the search operation, from which insertion and deletion costs follow.

The search path traverses the skip list from the top level downward and rightward. At each step, we either move right (if the next element is smaller than our target) or drop down one level.

Let $C(i)$ denote the expected number of steps traversed at level $i$ during a search. By the structure of skip lists, at level $i$, we expect to traverse approximately $1/p$ nodes before dropping down, since each node at level $i$ has probability $p$ of appearing at level $i+1$.

More precisely, the expected number of rightward moves at level $i$ before dropping down is $1/p$. Since there are $O(r) = O\left(\frac{\log n}{\log(1/p)}\right)$ levels, the total expected cost is
\[
\mathbb{E}[\text{search cost}] = \sum_{i=0}^{r-1} \frac{1}{p} + O(r) = O\left(r \cdot \frac{1}{p}\right) = O\left(\frac{\log n}{p \log(1/p)}\right).
\]

Simplifying, we obtain
\[
\mathbb{E}[\text{search cost}] = O\left(\frac{\log n}{p \log(1/p)}\right).
\]

For insertion, we first perform a search (cost $O\left(\frac{\log n}{p \log(1/p)}\right)$), then insert the element and promote it with geometric distribution parameter $p$, taking $O(1/(1-p))$ expected time. Thus,
\[
\mathbb{E}[\text{insertion cost}] = O\left(\frac{\log n}{p \log(1/p)} + \frac{1}{1-p}\right).
\]

For deletion, we search for the element and remove it from all levels, giving the same asymptotic cost as insertion.

\textbf{Part (c): Relation between $p$ and performance.}

The expected search cost is $O\left(\frac{\log n}{p \log(1/p)}\right)$. To minimize this, we differentiate with respect to $p$:

Let $f(p) = \frac{1}{p \log(1/p)} = \frac{1}{-p \log p}$. Setting $f'(p) = 0$:
\[
f'(p) = \frac{-(-\log p - 1)}{p^2 \log^2 p} = \frac{\log p + 1}{p^2 \log^2 p} = 0,
\]
which gives $\log p = -1$, hence $p = e^{-1} = 1/e \approx 0.368$.

However, in practice, $p = 1/2$ is commonly used because:

\textbf{(1)} The function $f(p) = \frac{1}{p \log(1/p)}$ is relatively flat near its minimum for $p \in [1/4, 1/2]$.

\textbf{(2)} Using $p = 1/2$ simplifies implementation, as random bit generation is efficient.

\textbf{(3)} Space-time tradeoff: Smaller $p$ reduces height (space) but increases horizontal traversal (time). Larger $p$ increases height but reduces horizontal cost. At $p = 1/e$, the cost is approximately $1.44 \log n$, while at $p = 1/2$, it is approximately $2 \log n$, a modest increase.

\textbf{(4)} For very small $p$, the height becomes large ($O(\log n / \log(1/p))$), and for $p$ close to $1$, horizontal traversal dominates.

In practice, $p = 1/4$ and $p = 1/2$ provide good performance, with $p = 1/2$ being the most common choice due to simplicity and adequate performance characteristics.
\end{proof}

\subsection{Problem 18}

\subsubsection{Problem Statement}
\begin{problem}[8.18]
Formulate and prove results similar to those in Problems 8.7 and 8.9 for random skip lists.
\end{problem}

\subsubsection{GPT}
\begin{proof}
Let us consider the problem of formulating and proving results similar to those in Problems 8.7 and 8.9 for random skip lists. We proceed step by step to prove the desired results.

First, recall the definition of a random skip list. A skip list is a probabilistic data structure that allows fast search, insertion, and deletion operations. A skip list consists of a series of linked lists, where each list is a subsequence of the one below it, with each node having a random level, determined by a coin toss, indicating its position in the higher-level lists.

 Step 1: Definition of the Random Skip List Structure
Let $n$ be the number of elements in the skip list. Each element is represented by a node, and each node has a height determined by an independent random coin toss. More precisely, the height of a node is chosen according to the following distribution:
\[
P(\text{height of a node} = k) = \frac{1}{2^k} \quad \text{for } k = 0, 1, 2, \dots
\]
This means that the height of each node is geometrically distributed with parameter $\frac{1}{2}$.

 Step 2: Analysis of the Number of Levels
The number of levels $L$ in the skip list is determined by the height of the highest node. Given that the height of each node is determined by a geometric distribution, the expected height of a node is:
\[
E[\text{height}] = \sum_{k=0}^{\infty} k \cdot P(\text{height} = k) = \sum_{k=0}^{\infty} k \cdot \frac{1}{2^k}.
\]
The series is a standard geometric series, and it can be computed as:
\[
E[\text{height}] = 2.
\]
Thus, the expected number of levels in the skip list is approximately 2.

 Step 3: Search Time Complexity
The time complexity of searching in a random skip list is determined by the number of levels that need to be traversed. Since the expected height of a node is $2$, the number of comparisons required to search for a node is proportional to the height of the skip list, which is $O(\log n)$. Therefore, the expected time complexity for a search operation in the skip list is:
\[
O(\log n).
\]

 Step 4: Insertion and Deletion Time Complexity
Insertion and deletion in a random skip list also depend on the number of levels. In the worst case, an insertion or deletion involves traversing all levels of the skip list. Since the height of the skip list is $O(\log n)$, the time complexity of insertion and deletion is:
\[
O(\log n).
\]

 Step 5: Proof of Correctness
We now prove the correctness of the above results. Let $T$ be the time complexity of a search, insertion, or deletion operation. For each operation, the expected time is dominated by the height of the skip list, which is $O(\log n)$ as shown in the previous steps. Since the height of the skip list is determined by the random coin tosses, the probability distribution of the height leads to an expected logarithmic time complexity for each operation.

Therefore, we have shown that both search and update operations in a random skip list have an expected time complexity of $O(\log n)$. This completes the proof of the desired result.
\end{proof}

\subsubsection{Grok}
\begin{proof}
We first prove the bound on the expected maximum level $\mathbb{E}[L]$ and the tail probability for $L$.

The height $H_i$ of the $i$-th element satisfies $\Pr(H_i \ge k) = 2^{1-k}$ for $k = 1, 2, \dots$, since each promotion occurs independently with probability $p = 1/2$, so the probability of reaching level $k$ is $(1/2)^{k-1}$. The number of levels $L$ is the maximum height, i.e., $L = \max_{1 \le i \le n} H_i$. Thus,
\[
\Pr(L \ge k) = 1 - \Pr(H_1 < k, \dots, H_n < k) = 1 - [1 - 2^{1-k}]^n \le 1 - (1 - n \cdot 2^{1-k}) = n \cdot 2^{1-k},
\]
where the inequality follows from $1 - x \ge e^{-x/(1-x)}$ for $x < 1$, but more simply, since $1 - x^m \le m x$ for $0 < x < 1$ and $m \ge 1$.

Therefore,
\[
\mathbb{E}[L] = \sum_{k=1}^\infty \Pr(L \ge k) \le \sum_{k=1}^\infty \min\left(1, n \cdot 2^{1-k}\right).
\]
The sum splits at the point where $n \cdot 2^{1-k} \approx 1$, i.e., $k \approx \log_2 n + 1$. Specifically, let $k_0 = \lfloor \log_2 n \rfloor + 2$; then for $k \le k_0$, $\Pr(L \ge k) \le 1$, contributing at most $k_0 \le \log_2 n + 2$, and for $k > k_0$, $\Pr(L \ge k) \le n \cdot 2^{1-k} < 2^{-1} = 1/2$, and the tail sum is
\[
\sum_{k = k_0 + 1}^\infty n \cdot 2^{1-k} = n \cdot 2^{1 - (k_0 + 1)} \sum_{j=0}^\infty 2^{-j} = n \cdot 2^{1 - (k_0 + 1)} \cdot 2 = n \cdot 2^{-k_0} \le n \cdot 2^{-(\log_2 n + 1)} = n \cdot 2^{-1} n^{-1} = 1/2.
\]
Thus, $\mathbb{E}[L] \le \log_2 n + 2 + 1/2 < \log_2 n + 3$, but tightening the split yields the claimed $\mathbb{E}[L] \le \log_2 n + 2$.

For the tail bound, for any $c > 1$ and integer $m = \lceil c \log_2 n \rceil$,
\[
\Pr(L > c \log_2 n) = \Pr(L \ge m + 1) \le n \cdot 2^{1 - (m+1)} \le n \cdot 2^{1 - (c \log_2 n + 1)} = n \cdot 2^{- c \log_2 n} = n \cdot n^{-c} = n^{1-c},
\]
since $2^{\log_2 n} = n$.

We now prove that the expected search time $T$ satisfies $\mathbb{E}[T] = O(\log n)$. In a skip list, the search for a key proceeds from the header at the top level $L$, descending level by level. At each level $\ell$ (from $L$ down to $1$), starting from a node $x$ with key less than the target, the search moves rightward along level $\ell$ until reaching a node $y$ such that the next node at level $\ell$ has key greater than or equal to the target or $y$ has no pointer at level $\ell - 1$. The number of rightward steps $X_\ell$ at level $\ell$ is geometric: each step succeeds (descends or stops) with probability at least $1/2$, since the probability that the current node has height exactly $\ell$ (thus descends) is $1/2$ (Pr(height $\ge \ell$) = $2^{1-\ell}$, Pr(height $\ge \ell + 1) = 2^{-\ell}$, so Pr(exactly $\ell$) = $2^{1-\ell} - 2^{-\ell} = 2^{-\ell}$), but more precisely, conditional on being at a node at level $\ell$, the next node is equally likely to allow descent or continuation, yielding $\mathbb{E}[X_\ell \mid \text{reaching level } \ell] \le 2$. The total time $T$ is the sum of horizontal moves over all levels visited, plus $O(L)$ for vertical moves. Since the search visits exactly $L$ levels (descending once per level) and at most $L$ horizontal moves per level in the worst case, but expected horizontal moves per visited level is at most $2$, we have
\[
\mathbb{E}[T] \le \mathbb{E}\left[ \sum_{\ell=1}^L (X_\ell + 1) \right] = \sum_{\ell=1}^\infty \mathbb{E}[X_\ell \mathbf{1}_{L \ge \ell}] + \mathbb{E}[L] \le 2 \mathbb{E}[L] + \mathbb{E}[L] = 3 \mathbb{E}[L] = O(\log n),
\]
where the interchange follows by Tonelli's theorem (nonnegativity), and $\mathbb{E}[X_\ell \mathbf{1}_{L \ge \ell}] \le 2 \Pr(L \ge \ell)$ since $\mathbb{E}[X_\ell \mid L \ge \ell] \le 2$.
\end{proof}

\subsubsection{Gemini}
\begin{proof}
We formulate and prove two primary results regarding the asymptotic properties of random skip lists with $n$ elements and promotion probability $p$ (where $0 < p < 1$). These results correspond to the height bound and the search path length bound, analogous to the results for Treaps.

\textbf{Result 1: The height of a random skip list is $O(\log n)$ with high probability.}

Let $H$ denote the height of the skip list, defined as the maximum level reached by any of the $n$ elements. Let $L_i$ be the random variable representing the level of element $i$, for $i \in \{1, \dots, n\}$. By the construction of the skip list, the level of an element follows a geometric distribution (minus 1, depending on indexing, but asymptotically equivalent). Specifically, the probability that an element reaches at least level $k$ is given by $\mathbb{P}(L_i \ge k) = p^k$.

We apply the Union Bound to estimate the probability that the height $H = \max_i L_i$ exceeds a value $k$.
\[
\mathbb{P}(H \ge k) = \mathbb{P}\left(\bigcup_{i=1}^n \{L_i \ge k\}\right) \le \sum_{i=1}^n \mathbb{P}(L_i \ge k) = n p^k.
\]
Let $c$ be an arbitrary constant and choose $k = c \log_{1/p} n$. Substituting this into the inequality, we obtain:
\[
\mathbb{P}(H \ge c \log_{1/p} n) \le n \cdot p^{c \log_{1/p} n} = n \cdot n^{-c} = n^{1-c}.
\]
For any $c > 1$, the probability that the height exceeds $c \log_{1/p} n$ tends to 0 polynomially as $n \to \infty$. Thus, $H = O(\log n)$ with high probability.

\textbf{Result 2: The length of the search path for any element is $O(\log n)$ with high probability.}

Let $x$ be an arbitrary element in the skip list. We analyze the length of the search path to find $x$ using the "backward analysis" technique. Consider the path traced in reverse, starting from element $x$ at the lowest level (level 0) and moving towards the start node (the top-left sentinel). In the forward search, we move right until the next element is greater than or equal to the target, then move down. In the reverse process, at any node $u$ on level $\ell$:
1. If $u$ was reached from the level above in the forward search, we move \textbf{up} in the reverse search. This corresponds to the node $u$ existing at level $\ell+1$. This occurs with probability $p$.
2. If $u$ was reached from the left in the forward search, we move \textbf{left} in the reverse search. This corresponds to the node $u$ not existing at level $\ell+1$. This occurs with probability $1-p$.

The reverse path terminates when we reach the top-left sentinel. Since the height of the list is $H$, we must perform exactly $H$ "up" moves to reach the top level. The total length of the search path, denoted by $W$, is the number of steps (left or up) required to achieve $H$ "up" moves.

Let us bound the path length by considering a slightly larger process. We established in Result 1 that $H \le c \log_{1/p} n$ with high probability. Let $K = c \log_{1/p} n$. We bound the number of trials $T$ required to obtain $K$ successes in a sequence of independent Bernoulli trials with success probability $p$. The search cost is stochastically dominated by $T$.

We utilize the Chernoff bound for the Binomial distribution. The event $T > m$ is equivalent to the event that in $m$ trials, there are fewer than $K$ successes. Let $Y \sim \text{Binomial}(m, p)$. Then $\mathbb{P}(T > m) = \mathbb{P}(Y < K)$.
We choose $m = \frac{2}{p}K = \frac{2c}{p} \log_{1/p} n$. The expected number of successes is $\mu = \mathbb{E}[Y] = mp = 2K$.
Using the Chernoff bound form $\mathbb{P}(Y < (1-\delta)\mu) \le e^{-\mu\delta^2/2}$ with $\delta = 1/2$:
\[
\mathbb{P}(Y < K) = \mathbb{P}\left(Y < \left(1 - \frac{1}{2}\right)2K\right) \le \exp\left(-\frac{2K(1/2)^2}{2}\right) = \exp\left(-\frac{K}{4}\right).
\]
Substituting $K = c \log_{1/p} n = \frac{c \ln n}{\ln(1/p)}$:
\[
\mathbb{P}(T > m) \le \exp\left(-\frac{c \ln n}{4 \ln(1/p)}\right) = n^{-\frac{c}{4 \ln(1/p)}}.
\]
By choosing the constant $c$ sufficiently large, we can make this probability bounded by $n^{-\alpha}$ for any desired $\alpha \ge 1$.
Combining this with the high probability bound on the height $H$, we conclude that the search path length $W$ is $O(\log n)$ with high probability.
\end{proof}

\subsubsection{Claude}
\begin{proof}
We formulate and prove results for random skip lists analogous to Problems 8.7 and 8.9, which concern the expected behavior of data structures under random operations.

\textbf{Problem 8.7 Analogue for Skip Lists:} Let $S$ be a skip list containing $n$ elements, where each element is promoted to the next higher level with probability $p = 1/2$ independently. We prove that the expected number of levels is $O(\log n)$ and the expected search time is $O(\log n)$.

\textbf{Claim 1:} The expected height $H$ of a skip list with $n$ elements satisfies $\mathbb{E}[H] = O(\log n)$.

Let $X_i$ be the indicator random variable that element $i$ reaches level $\ell$ or higher. Then $\mathbb{P}(X_i = 1) = (1/2)^\ell$ since promotion occurs independently with probability $1/2$ at each level. The total number of elements at level $\ell$ or higher is $\sum_{i=1}^{n} X_i$, and by linearity of expectation, $\mathbb{E}\left[\sum_{i=1}^{n} X_i\right] = n \cdot (1/2)^\ell$.

The skip list has height at least $\ell$ if and only if at least one element reaches level $\ell$. For $\ell = c \log_2 n$ where $c > 1$, we have $\mathbb{E}[\text{number at level } \ell] = n \cdot (1/2)^{c \log_2 n} = n \cdot n^{-c} = n^{1-c}$. By Markov's inequality, $\mathbb{P}(H \geq c \log_2 n) \leq n^{1-c}$, which approaches 0 as $n \to \infty$ for $c > 1$. Therefore, $\mathbb{E}[H] = O(\log n)$.

\textbf{Claim 2:} The expected search time in a skip list with $n$ elements is $O(\log n)$.

A search in a skip list proceeds by starting at the highest level and moving right until we would overshoot the target, then dropping down one level. At each level $\ell$, the expected number of steps to the right before dropping down is at most $1/p = 2$ (since each element is promoted with probability $p = 1/2$, the expected gap between promoted elements is $1/p$).

Let $C(\ell)$ denote the expected search cost starting from level $\ell$. We have the recurrence $C(\ell) \leq 1/p + C(\ell - 1) = 2 + C(\ell-1)$ for $\ell > 0$, with $C(0) = O(1/p) = O(1)$ since at the bottom level we traverse at most $1/p$ expected nodes. Since the height is $O(\log n)$ in expectation, we have $\mathbb{E}[\text{search time}] \leq 2 \cdot \mathbb{E}[H] + O(1) = O(\log n)$.

\textbf{Problem 8.9 Analogue for Skip Lists:} We analyze the expected space complexity and prove concentration results.

\textbf{Claim 3:} The expected total number of pointers in a skip list with $n$ elements is $O(n)$.

Each element at level 0 appears with certainty. Each element is promoted to level $\ell + 1$ with probability $(1/2)^{\ell+1}$. The expected number of nodes (counting multiplicities across levels) is:
\[
\mathbb{E}[\text{total nodes}] = \sum_{\ell=0}^{\infty} n \cdot (1/2)^\ell = n \sum_{\ell=0}^{\infty} (1/2)^\ell = n \cdot \frac{1}{1 - 1/2} = 2n.
\]
Since each node at level $\ell$ has one forward pointer, the expected total number of pointers is $O(n)$.

\textbf{Claim 4:} With high probability, the height $H$ satisfies $H \leq c \log n$ for sufficiently large constant $c$.

Let $\ell_0 = c \log_2 n$ for $c = 3$. The probability that any specific element reaches level $\ell_0$ is $(1/2)^{\ell_0} = n^{-3}$. By union bound over all $n$ elements:
\[
\mathbb{P}(H \geq 3 \log_2 n) \leq n \cdot \mathbb{P}(\text{element 1 reaches level } 3\log_2 n) = n \cdot n^{-3} = n^{-2}.
\]
Thus, $\mathbb{P}(H \leq 3 \log_2 n) \geq 1 - n^{-2}$, giving concentration around $O(\log n)$.

\textbf{Claim 5:} The search cost concentrates around its expectation $O(\log n)$ with high probability.

The search path can be viewed as a backward random walk from the target position. At each step, we either move left (with probability bounded) or move down. Since the height is $O(\log n)$ with high probability and the expected horizontal distance at each level is $O(1)$, the total search cost is $O(\log n)$ with high probability. More precisely, using Chernoff bounds on the sum of independent geometric random variables (horizontal moves at each level), the search cost exceeds $c \log n$ with probability at most $e^{-\Omega(\log n)} = n^{-\Omega(1)}$ for sufficiently large constant $c$.

These results establish that random skip lists have $O(\log n)$ expected height, $O(\log n)$ expected search time, $O(n)$ expected space, and these bounds hold with high probability.
\end{proof}

\subsection{Problem 19}

\subsubsection{Problem Statement}
\begin{problem}[8.19]
Consider the scenario described in Problem 8.10 for random treaps. Adapt the random skip list structure to prove similar results, and compare the bounds obtained in the two cases.
\end{problem}

\subsubsection{GPT}
\begin{proof}
We begin by adapting the random skip list structure to prove results analogous to those in Problem 8.10 for random treaps. The structure of a random skip list is similar to that of a treap, but it uses multiple levels of pointers in a probabilistic manner, where each element is promoted to higher levels with some probability.

Let $n$ denote the number of elements in the skip list, and let the height of the skip list, denoted $h$, be a random variable depending on the number of elements and the promotion probabilities. For each node, we define the probability of being promoted to level $i$ as $p_i = \frac{1}{2^i}$, where $i \geq 0$. Thus, the height $h$ of a node is a random variable with a geometric distribution, where the probability that a node has height greater than or equal to $i$ is given by:

\[
P(\text{height} \geq i) = \frac{1}{2^i}.
\]

This distribution implies that, with high probability, the height of any node is $O(\log n)$.

Now, we analyze the search operation. To search for an element in a skip list, we begin at the highest level and traverse downwards, using the pointers at each level to skip over elements. The expected number of levels traversed is $O(\log n)$, since the probability of a node being promoted to level $i$ decreases exponentially with $i$. Therefore, the expected number of comparisons made during a search operation is also $O(\log n)$.

Next, we consider the insertion operation. Inserting an element into a skip list involves determining its position at each level by performing a search and updating the pointers. Since the search takes $O(\log n)$ time and the insertion requires only a constant amount of work at each level, the expected time for insertion is $O(\log n)$.

Similarly, the deletion operation involves searching for the node, removing it, and adjusting the pointers. The expected time complexity for deletion is also $O(\log n)$, since it involves a search and pointer adjustments at each level.

In comparison to the treap, the random skip list has similar expected time complexities for search, insertion, and deletion operations. In a treap, the heap property ensures that the expected height of a node is $O(\log n)$, and the search, insertion, and deletion operations all take $O(\log n)$ time on average. In a random skip list, the height of each node is also $O(\log n)$ with high probability, and the expected time complexity for search, insertion, and deletion is similarly $O(\log n)$.

Thus, we conclude that the random skip list and the random treap exhibit similar expected time complexities for the fundamental operations. Both structures provide logarithmic bounds on search, insertion, and deletion operations, though the constants involved may differ depending on the specific implementation details.

\end{proof}

\subsubsection{Grok}
\begin{proof}
Consider a skip list containing $n$ elements, where each element $x_j$ ($j=1,\dots,n$) is assigned a random height $h_j$ independently according to the geometric distribution
\[
\Pr(h_j \geq i) = \left(\frac{1}{2}\right)^{i-1}, \quad i \geq 1.
\]
Thus,
\[
\Pr(h_j = i) = \frac{1}{2^i}, \quad i \geq 1.
\]
The height $H$ of the skip list is $H = \max_{1 \leq j \leq n} h_j$.

To bound $\E[H]$, note that
\[
\E[H] = \sum_{k=1}^\infty \Pr(H \geq k).
\]
By the union bound,
\[
\Pr(H \geq k) = \Pr\left( \bigcup_{j=1}^n \{h_j \geq k\} \right) \leq n \Pr(h_1 \geq k) = n \cdot \frac{1}{2^{k-1}}.
\]
Hence,
\[
\E[H] \leq \sum_{k=1}^\infty \min\left(1, n \cdot \frac{1}{2^{k-1}}\right).
\]
Let $\ell = \lfloor \log_2 n \rfloor + 2$. For $k \leq \ell$, $n / 2^{k-1} \geq 1$, so there are at most $\ell \leq \log_2 n + 2$ terms equal to $1$, contributing at most $\log_2 n + 2$. For $k > \ell$,
\[
\sum_{k=\ell+1}^\infty n \cdot \frac{1}{2^{k-1}} = n \cdot \frac{1}{2^{\ell-1}} \sum_{j=1}^\infty \frac{1}{2^j} = n \cdot \frac{1}{2^{\ell-1}} \cdot 1 = \frac{n}{2^{\ell-1}}.
\]
Since $\ell - 1 = \lfloor \log_2 n \rfloor + 1 \geq \log_2 n$, we have $2^{\ell-1} \geq 2n$, so $n / 2^{\ell-1} \leq 1/2$. Therefore,
\[
\E[H] \leq \log_2 n + 2 + \frac{1}{2} = O(\log n).
\]

Now consider the expected search time, which is proportional to the number of nodes visited during a search. The search begins at the header node at level $H$ and proceeds downward level by level. At each level $i$ ($0 \leq i < H$), starting from the leftmost node at that level, the search moves rightward along level-$i$ forward pointers, comparing keys, until reaching a node whose successor at level $i$ has a key greater than the target key (or the end); it then drops to level $i-1$.

At level $i$, the nodes present are those with $h_j > i$. However, the key observation is that the search at level $i$ only encounters nodes with height exactly $i+1$ (except possibly the final node where it drops). This is because any node with height $> i+1$ would have been traversed at a higher level, and the search does not backtrack.

Thus, at level $i$, the search traverses a sequence of nodes with height exactly $i+1$ until it encounters the first node with height $> i+1$, at which point it drops down.

Let $N_i$ be the number of nodes with height exactly $i+1$ traversed at level $i$ before dropping. Conditional on being at level $i$, the next node encountered has height $> i+1$ with probability $1/2$ (independent of previous, by the memoryless property of the geometric distribution), and height $= i+1$ with probability $1/2$. Thus, $N_i$ is geometrically distributed:
\[
\Pr(N_i = k) = \left(\frac{1}{2}\right)^{k} \cdot \frac{1}{2} = \frac{1}{2^{k+1}}, \quad k = 0,1,2,\dots.
\]
The expected number of such nodes is
\[
\E[N_i] = \sum_{k=0}^\infty k \cdot \frac{1}{2^{k+1}} = \frac{1}{2} \sum_{k=0}^\infty k \left(\frac{1}{2}\right)^k = \frac{1}{2} \cdot \frac{1/2}{(1-1/2)^2} = \frac{1}{2} \cdot \frac{1/2}{1/4} = \frac{1}{2} \cdot 2 = 1.
\]
Adding the final node (where the drop occurs), the expected number of nodes visited at level $i$ is $\E[N_i + 1] = 2$.

The search traverses exactly $H$ levels (from $H-1$ down to $0$). Thus, the total expected number of nodes visited is
\[
\E\left[ \sum_{i=0}^{H-1} (N_i + 1) \right] = \E\left[ H \cdot 2 \right] = 2 \E[H] = O(\log n).
\]
Each node visit takes $O(1)$ time, so the expected search time is $O(\log n)$.

The constant factors differ from those of the random treap due to the layered structure of skip lists, yielding a smaller leading constant; the tight bound on the expected search path length is $\ln n + O(1)$.
\end{proof}

\subsubsection{Gemini}
\begin{proof}
Let $S = \{x_1, x_2, \dots, x_n\}$ be a set of $n$ keys sorted in increasing order, inserted into a treap where each key $x_i$ is assigned a priority $p_i$ chosen independently and uniformly at random from the interval $[0,1]$. The structure of the treap is determined by the property that it is a binary search tree with respect to the keys and a heap with respect to the priorities.

We analyze the length of the search path for a key $x_k$ of rank $k$ in $S$. Let $D_k$ denote the depth of node $x_k$, defined as the number of nodes on the path from the root to $x_k$ (inclusive). A node $x_j$ is an ancestor of $x_k$ if and only if $x_j$ has the highest priority among all keys in the range between $x_j$ and $x_k$ (inclusive). Specifically, for $j < k$, $x_j$ is an ancestor of $x_k$ if $p_j = \max \{p_j, p_{j+1}, \dots, p_k\}$. For $j > k$, $x_j$ is an ancestor of $x_k$ if $p_j = \max \{p_k, p_{k+1}, \dots, p_j\}$.

Let $X_{j,k}$ be the indicator variable that takes the value $1$ if $x_j$ is an ancestor of $x_k$, and $0$ otherwise. Due to the random assignment of priorities, any element in a subset of size $m$ is equally likely to have the maximum priority. Thus, the probability that $x_j$ is an ancestor of $x_k$ is given by:
\[
\mathbb{P}[X_{j,k} = 1] = \frac{1}{|j-k| + 1}.
\]
The depth $D_k$ is the sum of these indicator variables:
\[
D_k = \sum_{j=1}^n X_{j,k} = \sum_{j=1}^{k} X_{j,k} + \sum_{j=k+1}^{n} X_{j,k}.
\]
The expectation of $D_k$ is:
\[
\mathbb{E}[D_k] = \sum_{j=1}^{k} \frac{1}{k-j+1} + \sum_{j=k+1}^{n} \frac{1}{j-k+1} = H_k + (H_{n-k+1} - 1),
\]
where $H_m = \sum_{i=1}^m \frac{1}{i}$ is the $m$-th harmonic number. Using the approximation $H_m \approx \ln m + \gamma$, we have:
\[
\mathbb{E}[D_k] \approx \ln k + \ln(n-k+1) \approx 2 \ln n.
\]
This establishes that the expected search path length in a random treap is approximately $2 \ln n$.

To obtain high probability bounds, we adapt the structural analysis used for random skip lists. In a skip list, the search path length is bounded by a sum of geometric random variables (representing the number of horizontal steps at each level) and the height (logarithmic in $n$). Similarly, $D_k$ is a sum of independent Bernoulli trials $X_{j,k}$. We apply the Chernoff bound. Let $\mu = \mathbb{E}[D_k]$. For any $\delta > 0$:
\[
\mathbb{P}[D_k > (1+\delta)\mu] < \left( \frac{e^\delta}{(1+\delta)^{1+\delta}} \right)^\mu.
\]
By choosing $\delta$ such that $(1+\delta)\mu = c \ln n$ for a sufficiently large constant $c$, we obtain $\mathbb{P}[D_k > c \ln n] \leq n^{-\alpha}$ for some $\alpha > 0$. Thus, $D_k = O(\ln n)$ with high probability.

We now compare this to the bounds for a random skip list with parameter $p$ (probability of promotion). The expected search path length $L_{SL}$ in a skip list is bounded by:
\[
\mathbb{E}[L_{SL}] \leq \frac{1}{p} \log_{1/p} n = \frac{1}{p \ln(1/p)} \ln n.
\]
For the standard case where $p = 1/2$, the expected length is:
\[
\mathbb{E}[L_{SL}] \leq 2 \log_2 n = \frac{2}{\ln 2} \ln n \approx 2.885 \ln n.
\]
Comparing the coefficients of $\ln n$:
\[
C_{\text{Treap}} = 2 \quad \text{and} \quad C_{\text{SkipList}} = \frac{2}{\ln 2} \approx 2.885.
\]
The ratio of the expected bounds is:
\[
\frac{\mathbb{E}[L_{SL}]}{\mathbb{E}[D_k]} \approx \frac{2/\ln 2}{2} = \frac{1}{\ln 2} \approx 1.44.
\]
Consequently, while both structures guarantee logarithmic search paths $O(\ln n)$ with high probability, the random treap provides a tighter bound on the expected search path length by a constant factor of approximately $1.44$ compared to a standard skip list with $p=1/2$.
\end{proof}

\subsubsection{Claude}
\begin{proof}
We adapt the random skip list structure to obtain results analogous to those for random treaps in Problem 8.10, then compare the bounds.

\textbf{Setup and Structure:} A skip list with $n$ elements consists of multiple levels, where each element appears at level $i$ with probability $2^{-i}$ independently. Let $X_n$ denote the maximum level (height) of a skip list with $n$ elements, and let $S_n$ denote the total space (number of nodes across all levels).

\textbf{Analysis of Maximum Level:} For the height $X_n$, we first establish that the probability an element reaches level $k$ is $2^{-k}$. The maximum level is at least $k$ if and only if at least one of the $n$ elements reaches level $k$ or higher. Thus:
\[
\mathbb{P}(X_n \geq k) = 1 - (1 - 2^{-k})^n
\]

For $k = \log_2 n + c$ where $c$ is a constant, we have:
\[
\mathbb{P}(X_n \geq \log_2 n + c) = 1 - (1 - 2^{-\log_2 n - c})^n = 1 - (1 - 2^{-c}/n)^n
\]

As $n \to \infty$, using the limit $(1 - x/n)^n \to e^{-x}$, we obtain:
\[
\mathbb{P}(X_n \geq \log_2 n + c) \to 1 - e^{-2^{-c}}
\]

This shows that $X_n = \log_2 n + O(1)$ with high probability. More precisely, $\mathbb{E}[X_n] = O(\log n)$.

To prove this rigorously, note that:
\[
\mathbb{E}[X_n] = \sum_{k=0}^{\infty} \mathbb{P}(X_n \geq k) \leq \sum_{k=0}^{\lfloor \log_2 n \rfloor} 1 + \sum_{k > \log_2 n} n \cdot 2^{-k}
\]

The first sum contributes $O(\log n)$, and the second sum is:
\[
\sum_{k > \log_2 n} n \cdot 2^{-k} = n \cdot 2^{-\log_2 n - 1} \sum_{j=0}^{\infty} 2^{-j} = n \cdot \frac{1}{2n} \cdot 2 = 1
\]

Therefore, $\mathbb{E}[X_n] = O(\log n)$.

\textbf{Analysis of Total Space:} For the space $S_n$, each of the $n$ elements appears at level $i$ with probability $2^{-i}$ independently. Thus:
\[
\mathbb{E}[S_n] = \sum_{i=0}^{\infty} n \cdot 2^{-i} = n \sum_{i=0}^{\infty} 2^{-i} = 2n
\]

This shows that the expected total space is $\mathbb{E}[S_n] = O(n)$. By concentration inequalities (specifically, Chernoff bounds applied to the sum of independent indicators), we can show that $S_n = O(n)$ with high probability.

\textbf{Comparison with Random Treaps:} In random treaps with $n$ nodes, the height $H_n$ satisfies $\mathbb{E}[H_n] = O(\log n)$ by the analysis of binary search trees under random priorities. Specifically, the expected height is at most $4.31 \log n$ asymptotically, derived from the fact that treaps have the same distribution as randomly built binary search trees.

For skip lists, we have shown $\mathbb{E}[X_n] = O(\log n)$ with the leading constant being 1 in the base-2 logarithm (i.e., $\mathbb{E}[X_n] \sim \log_2 n$).

Converting to natural logarithms: $\log_2 n = \ln n / \ln 2 \approx 1.443 \ln n$, so $\mathbb{E}[X_n] \approx 1.443 \ln n$, which is better than the treap bound of approximately $4.31 \ln n$.

For space complexity, both structures use $O(n)$ expected space. The skip list uses exactly $2n$ expected pointers in total, while a treap uses exactly $n$ nodes with two pointers each (plus parent pointers if needed), giving $2n$ to $3n$ pointers.

\textbf{Concentration Results:} For skip lists, using McDiarmid's inequality or direct Chernoff bounds, we can show:
\[
\mathbb{P}(|X_n - \mathbb{E}[X_n]| \geq t) \leq 2e^{-\Theta(t)}
\]
and
\[
\mathbb{P}(|S_n - 2n| \geq tn) \leq 2e^{-\Theta(tn)}
\]

Similar concentration results hold for treaps using the theory of random binary search trees.

\textbf{Conclusion:} The skip list structure provides height $O(\log n)$ with high probability, with better constant factors than treaps (approximately $1.44\ln n$ versus $4.31\ln n$), and uses $O(n)$ expected space with sharp concentration around $2n$. Both structures achieve optimal asymptotic bounds, but skip lists offer superior practical constants for height.
\end{proof}

\subsection{Problem 20}

\subsubsection{Problem Statement}
\begin{problem}[8.20]
(Due to M.N.~Wegman and J.L.~Carter~\cite{wegman-carter}; see also M.~Blum and S.~Kannan~\cite{blum-kannan}) Consider the problem of deciding whether two integer multisets $S_1$ and $S_2$ are identical in the sense that each integer occurs the same number of times in both sets. This problem can be solved by sorting the two sets in $O(n\log n)$ time, where $n$ is the cardinality of the multisets. In Problem~7.4, we considered applying the randomized techniques for verifying polynomial identities to the solution of the multiset identity problem. Suggest a randomized algorithm for solving this problem using universal hash functions. Compare your solution with the randomized algorithm suggested in Problem~7.4.
\end{problem}

\subsubsection{GPT}
\begin{proof}
We are tasked with proposing a randomized algorithm to solve the multiset identity problem using universal hash functions, and then comparing this solution with the algorithm suggested in Problem 7.4.

Let $S_1$ and $S_2$ be two integer multisets of cardinality $n$. We seek to determine if the multisets $S_1$ and $S_2$ are identical, meaning that for each integer $x$, the multiplicity of $x$ in $S_1$ is equal to its multiplicity in $S_2$.

To solve this problem using universal hash functions, we will utilize a randomized algorithm based on hashing the elements of the multisets. The steps are as follows:

1. **Hashing the elements:**
   We choose a universal hash function $h: \mathbb{Z} \to \{0, 1, \dots, m-1\}$, where $m$ is a large prime number. A universal hash function ensures that for any two distinct integers $x$ and $y$, the probability that $h(x) = h(y)$ is at most $1/m$. The function $h$ will serve to map the elements of the multisets to a finite range.

2. **Compute the hash values of elements in $S_1$ and $S_2$:**
   We then compute the hash values of all elements in $S_1$ and $S_2$, storing the number of occurrences of each hash value. Specifically, let $c_1(h)$ denote the number of occurrences of each hash value $h$ in $S_1$ and $c_2(h)$ denote the number of occurrences of each hash value $h$ in $S_2$. For each hash value $h$, we update the corresponding counts:
   \[
   c_1(h) = \sum_{x \in S_1} \mathbb{1}_{\{h(x) = h\}}, \quad c_2(h) = \sum_{x \in S_2} \mathbb{1}_{\{h(x) = h\}}.
   \]
   Here, $\mathbb{1}_{\{h(x) = h\}}$ is the indicator function, which is $1$ if $h(x) = h$ and $0$ otherwise.

3. **Check equality of the hashed counts:**
   For the multisets $S_1$ and $S_2$ to be identical, we require that for all possible hash values $h$, the corresponding counts must match, i.e.,
   \[
   c_1(h) = c_2(h) \quad \text{for all} \quad h.
   \]
   This comparison is done in constant time for each hash value.

4. **Probabilistic analysis:**
   If $S_1$ and $S_2$ are identical, then by construction, we will have $c_1(h) = c_2(h)$ for all $h$ with probability 1. However, if $S_1$ and $S_2$ are not identical, there exists some integer $x$ in one multiset that is missing from the other, which will cause a mismatch in the counts. The probability of this mismatch is determined by the collision probability of the hash function. Specifically, for distinct integers $x_1$ and $x_2$, the probability that $h(x_1) = h(x_2)$ is at most $1/m$. Thus, with high probability, the hashing will not lead to false matches, as the likelihood of a collision between distinct elements is small for sufficiently large $m$.

5. **Final decision:**
   The algorithm proceeds by comparing the counts for each hash value. If for all $h$, we have $c_1(h) = c_2(h)$, then $S_1$ and $S_2$ are identical; otherwise, they are not. The probability of a false positive (i.e., the two multisets being incorrectly classified as identical) is bounded by the collision probability of the hash function, which decreases as $m$ increases.

6. **Time complexity:**
   The time complexity of the algorithm is dominated by the hashing step. Computing the hash values for all elements in both multisets takes $O(n)$ time. Checking the equality of the counts for each hash value requires at most $O(n)$ time as well, since there are $O(n)$ distinct elements in the multisets. Therefore, the total time complexity is $O(n)$, which is an improvement over the $O(n \log n)$ time complexity of the sorting-based solution.

In comparison to the randomized algorithm suggested in Problem 7.4, which uses polynomial identity testing, the proposed algorithm has a simpler structure and potentially lower time complexity. While the algorithm in Problem 7.4 relies on randomization over polynomials, our algorithm uses universal hashing, which offers a direct and efficient solution for the multiset identity problem with high probability.

Thus, we have outlined a randomized algorithm for solving the multiset identity problem using universal hash functions, which performs in $O(n)$ time with high probability.
\end{proof}

\subsubsection{Grok}

\subsubsection{Gemini}
\begin{proof}
\textbf{Algorithm Design}

Let $S_1$ and $S_2$ be two multisets of cardinality $n$ with elements drawn from a universe $U$ (e.g., the set of integers). We propose a randomized algorithm based on universal hashing to decide if $S_1 = S_2$.

1.  Select a large prime number $p$ such that $p > \max(n, |U|)$.
2.  Select a hash function $h$ uniformly at random from a strongly universal family of hash functions $\mathcal{H}$ mapping $U$ to the field $\mathbb{Z}_p$. Specifically, we require the family to map elements of $U$ to random values in $\mathbb{Z}_p$ such that the values are sufficiently independent (e.g., treating $h$ as a random oracle or using a family with high independence).
3.  Compute the fingerprint (signature) for each multiset as the sum of the hash values of its elements modulo $p$:
    \[
    \Phi(S_1) = \sum_{x \in S_1} h(x) \pmod p, \quad \Phi(S_2) = \sum_{y \in S_2} h(y) \pmod p.
    \]
4.  If $\Phi(S_1) = \Phi(S_2)$, output "Identical". Otherwise, output "Different".

\textbf{Correctness and Complexity Analysis}

The time complexity of the algorithm is $O(n)$ since we perform one hash computation and one addition for each element in the multisets. This is an improvement over the $O(n \log n)$ sorting approach.

We now analyze the error probability.
If $S_1$ and $S_2$ are identical, then the multisets contain the same elements with the same multiplicities. Consequently, the sums of their hash values will be identical:
\[
\sum_{x \in S_1} h(x) \equiv \sum_{y \in S_2} h(y) \pmod p.
\]
Thus, the algorithm always outputs "Identical" correctly (no false negatives).

Suppose $S_1 \neq S_2$. We define the multiplicity functions $c_1(u)$ and $c_2(u)$ as the number of times element $u \in U$ appears in $S_1$ and $S_2$ respectively. The condition $\Phi(S_1) = \Phi(S_2)$ is equivalent to:
\[
\sum_{u \in U} c_1(u) h(u) \equiv \sum_{u \in U} c_2(u) h(u) \pmod p \implies \sum_{u \in U} (c_1(u) - c_2(u)) h(u) \equiv 0 \pmod p.
\]
Let $\delta_u = c_1(u) - c_2(u)$. Since $S_1 \neq S_2$, there exists at least one $u^*$ such that $\delta_{u^*} \neq 0$. The equation becomes a linear constraint on the hash values:
\[
\sum_{u \in U} \delta_u h(u) \equiv 0 \pmod p.
\]
If the hash values $h(u)$ are chosen independently and uniformly from $\mathbb{Z}_p$ (approximated by a strong universal hash family), the sum $Z = \sum \delta_u h(u)$ is a random variable distributed uniformly over $\mathbb{Z}_p$. The probability that this sum equals $0$ is exactly $1/p$. By choosing $p$ sufficiently large (e.g., $p \approx 2^{64}$), the probability of a false positive (collision) can be made arbitrarily small.

\textbf{Comparison with Problem 7.4}

Problem 7.4 suggests a randomized algorithm based on polynomial identities. In that approach, one constructs the polynomials $P(z) = \prod_{x \in S_1} (z - x)$ and $Q(z) = \prod_{y \in S_2} (z - y)$ over a field $\mathbb{F}$. The algorithm checks if $P(r) = Q(r)$ for a random element $r \in \mathbb{F}$.

\textit{Comparison of the two methods:}
1.  **Algebraic Structure:** The polynomial method (7.4) relies on the unique factorization of polynomials (multiplicative structure), effectively mapping a multiset to its characteristic polynomial. The universal hashing method (8.20) relies on linear projections (additive structure), mapping the frequency vector of the multiset to a random scalar via the inner product with the hash vector.
2.  **Computational Operations:** The polynomial method requires $O(n)$ multiplications in the field. The hashing method requires $O(n)$ additions in the field (after computing hash values). Additions are generally computationally cheaper than multiplications.
3.  **Update Efficiency:** The hashing method is incremental. If an element is added to or removed from the multiset, the fingerprint $\Phi(S)$ can be updated in $O(1)$ time by adding or subtracting the hash of that element. The polynomial method requires re-evaluating the product or performing $O(1)$ multiplication/division, which is more complex if the element to be removed corresponds to a zero in the current modulus.
4.  **Parallelism:** The summation in the hashing method is trivially parallelizable (associative and commutative). The product evaluation in the polynomial method is also parallelizable but requires a reduction tree of multiplications, which can be more expensive to implement for large integers.

Both methods achieve $O(n)$ time complexity and provide a Monte Carlo randomized solution with one-sided error (false positives possible, false negatives impossible).
\end{proof}

\subsubsection{Claude}
\begin{proof}
We present a randomized algorithm for the multiset identity problem using universal hash functions and compare it with the polynomial identity verification approach from Problem 7.4.

\textbf{Algorithm using Universal Hash Functions:}

Let $S_1$ and $S_2$ be two multisets of integers, each of cardinality $n$. Let $U$ denote the universe of possible integer values, and assume $|U| = u$ is polynomial in $n$.

Choose a prime $p > \max\{u, n^2\}$ and select a universal hash function $h: U \to \mathbb{Z}_p$ from a universal family $\mathcal{H}$. A suitable choice is the family of linear hash functions $h_{a,b}(x) = ((ax + b) \bmod p) \bmod m$ where $a, b \in \mathbb{Z}_p$ with $a \neq 0$, though for our purposes we work directly in $\mathbb{Z}_p$.

For each multiset $S_i$, compute the hash signature $\sigma_i = \sum_{x \in S_i} h(x) \pmod{p}$, where the sum accounts for multiplicity. That is, if element $x$ appears $k$ times in $S_i$, it contributes $k \cdot h(x)$ to the sum.

The algorithm outputs ``equal'' if $\sigma_1 = \sigma_2$, and ``not equal'' otherwise.

\textbf{Correctness Analysis:}

If $S_1 = S_2$ as multisets, then clearly $\sigma_1 = \sigma_2$ for any hash function $h$, so the algorithm never produces a false negative.

Suppose $S_1 \neq S_2$. Define the difference multiset $D = S_1 \setminus S_2$, where we subtract multiplicities. Then $D$ is non-empty, and we have $\sum_{x \in D} m_D(x) \cdot x = 0$ if and only if the multisets differ (where $m_D(x)$ denotes the signed multiplicity of $x$ in $D$, with the constraint that $\sum_{x \in D} m_D(x) \neq 0$ or not all $m_D(x) = 0$).

The signatures are equal if and only if $\sum_{x \in S_1} h(x) \equiv \sum_{x \in S_2} h(x) \pmod{p}$, which occurs if and only if $\sum_{x \in D} m_D(x) \cdot h(x) \equiv 0 \pmod{p}$.

Let $D = \{x_1, \ldots, x_k\}$ be the distinct elements in the symmetric difference with signed multiplicities $m_1, \ldots, m_k$ (where at least one $m_i \neq 0$). Fix any $x_j$ with $m_j \neq 0$. For a randomly chosen hash function $h$ from the universal family, conditioned on the values $h(x_i)$ for $i \neq j$, the value $h(x_j)$ is uniformly distributed over $\mathbb{Z}_p$ (by the pairwise independence property of universal hash families).

The collision occurs when $m_j \cdot h(x_j) \equiv -\sum_{i \neq j} m_i \cdot h(x_i) \pmod{p}$. Since $\gcd(m_j, p) = 1$ (as $|m_j| < n < p$ and $p$ is prime), there is exactly one value of $h(x_j) \in \mathbb{Z}_p$ satisfying this equation. Therefore, $\Pr[\sigma_1 = \sigma_2 \mid S_1 \neq S_2] \leq \frac{1}{p}$.

By choosing $p > n^2$ (or larger), we achieve error probability less than $\frac{1}{n^2}$. The algorithm runs in $O(n)$ time: computing each signature requires $O(n)$ hash function evaluations and additions modulo $p$.

\textbf{Comparison with Problem 7.4 (Polynomial Identity Verification):}

In Problem 7.4, we represent each multiset as a polynomial. For multiset $S$, define $P_S(x) = \prod_{a \in S} (x - a)$, where each element contributes according to its multiplicity. To test equality, we evaluate both polynomials at a random point $r \in \{1, 2, \ldots, M\}$ for some large $M$.

By the Schwartz-Zippel lemma, if $P_{S_1}(x) \neq P_{S_2}(x)$, then $\Pr[P_{S_1}(r) = P_{S_2}(r)] \leq \frac{n}{M}$, where $n$ is the degree bound. Choosing $M = n^2$ gives error probability at most $\frac{1}{n}$.

\textbf{Key Differences:}

(1) \textit{Computational Complexity:} The polynomial approach requires computing products or using Horner's method with $O(n)$ multiplications of potentially large numbers (each polynomial evaluation at $r$ produces a number of size $O(n \log(n + r))$ bits). The hash function approach requires only $O(n)$ additions and one modular reduction per element, with arithmetic in $\mathbb{Z}_p$ where $p = O(n^2)$, giving $O(\log n)$ bit operations per element.

(2) \textit{Error Probability:} Both achieve comparable error rates: $O(1/n)$ to $O(1/n^2)$ depending on parameter choices.

(3) \textit{Theoretical Foundation:} The polynomial method relies on the Schwartz-Zippel lemma from algebraic complexity, while the hash function method relies on the collision properties of universal hashing from randomized data structures.

(4) \textit{Practicality:} The universal hashing approach is generally more practical due to simpler arithmetic operations and better cache behavior, while the polynomial method may suffer from numerical stability issues with large evaluations.

Both algorithms achieve $O(n)$ expected time with small error probability, representing significant improvements over the deterministic $O(n \log n)$ sorting-based solution, at the cost of one-sided error.
\end{proof}

\subsection{Problem 21}

\subsubsection{Problem Statement}
\begin{theorem}[2-Universal Hash Family Construction]
Let $M = \{0,1\}^m$ and $N = \{0,1\}^n$. Fix $m$ and $n$, and choose a prime $p \geq m$. Work over the field $\mathbb{Z}_p = \{0,1,\ldots,p-1\}$. Let $g: \mathbb{Z}_p \to N$ be the function given by $g(x) = x \bmod n$. For all $a, b \in \mathbb{Z}_p$, define the linear function $f_{a,b}: \mathbb{Z}_p \to \mathbb{Z}_p$ and the hash function $h_{a,b}: \mathbb{Z}_p \to N$ as follows:
\begin{align*}
f_{a,b}(x) &= ax + b \bmod p, \\
h_{a,b}(x) &= g(f_{a,b}(x)).
\end{align*}
Then the family of hash functions $H = \{h_{a,b} \mid a, b \in \mathbb{Z}_p \text{ with } a \neq 0\}$ is 2-universal. Although $H$ uses $\mathbb{Z}_p$ as its domain, the claim applies to the restriction of $H$ to any subset of $\mathbb{Z}_p$, in particular to the domain $M$.
\end{theorem}

\subsubsection{GPT}
\begin{proof}
Let us prove that the family of hash functions $H = \{h_A \mid A \in \mathcal{M}\}$ defined by $h_A(x) = x^{(1)} A \mod 2$, where $A \in \mathcal{M}$ and $x \in M$, is 2-universal. We will do this by verifying the condition for 2-universality.

 Step 1: Definition of 2-Universal Hash Family
A family of hash functions $H$ is called 2-universal if for all distinct $x, y \in M$, the probability that a randomly chosen hash function $h \in H$ satisfies $h(x) = h(y)$ is at most $1/|N|$, where $N$ is the set of possible outputs of the hash functions.

In our case, the hash function $h_A$ maps elements of $M = \{0, 1\}^m$ to $\{0, 1\}^{m+1}$ via the matrix multiplication $x^{(1)} A \mod 2$. We need to show that for all distinct $x, y \in M$, the probability that $h_A(x) = h_A(y)$ is at most $1/2^n$, where $n$ is the number of columns in the matrix $A$.

 Step 2: Setting Up the Probability
Let $x, y \in M$ be distinct. Define the event $h_A(x) = h_A(y)$, which means that
\[
x^{(1)} A \mod 2 = y^{(1)} A \mod 2.
\]
This can be rewritten as the equation
\[
(x^{(1)} - y^{(1)}) A \equiv 0 \mod 2.
\]
Let $d = x^{(1)} - y^{(1)}$ be the difference vector, where $d \in \{0, 1\}^{m+1}$. Thus, the equation becomes
\[
d A \equiv 0 \mod 2.
\]
This is a linear system of equations modulo 2. For $h_A(x) = h_A(y)$ to hold, the vector $d$ must lie in the null space of the matrix $A$ modulo 2.

 Step 3: Probability of the Event $h_A(x) = h_A(y)$
The event $h_A(x) = h_A(y)$ occurs if $d A \equiv 0 \mod 2$. The probability of this happening depends on the structure of the matrix $A$. Since $A \in \mathcal{M}$ is a random $(m+1) \times n$ Boolean matrix, the rows of $A$ are independent and random vectors from $\{0, 1\}^n$. The number of solutions to the equation $d A \equiv 0 \mod 2$ is determined by the rank of the matrix $A$. 

The dimension of the null space of a random matrix $A$ is, on average, $n - \text{rank}(A)$. Since the rank of a random matrix is typically full with high probability, the number of solutions to the equation $d A \equiv 0 \mod 2$ is at most $2^{n-1}$. Therefore, the probability that $d A \equiv 0 \mod 2$ is at most $2^{-n}$, which is the desired bound.

 Step 4: Conclusion
Since the probability that $h_A(x) = h_A(y)$ for distinct $x, y \in M$ is at most $2^{-n}$, it follows that the family $H$ satisfies the condition for 2-universality with the desired probability. Therefore, the family of hash functions $\{h_A \mid A \in \mathcal{M}\}$ is 2-universal.
\end{proof}

\subsubsection{Grok}

\subsubsection{Gemini}
\begin{proof}
Let $H = \{h_{a,b} \mid a, b \in \mathbb{Z}_p, a \neq 0\}$. The size of the family is $|H| = p(p-1)$. We aim to show that for any distinct $x_1, x_2 \in \mathbb{Z}_p$, the probability that $h_{a,b}(x_1) = h_{a,b}(x_2)$ when $h_{a,b}$ is chosen uniformly from $H$ is at most $\frac{1}{n}$. Note that while the theorem defines $N = \{0,1\}^n$, the function $g(x) = x \bmod n$ maps to the set $\{0, 1, \ldots, n-1\}$, which has cardinality $n$. Thus, the condition for 2-universality is $\Pr[h_{a,b}(x_1) = h_{a,b}(x_2)] \leq \frac{1}{n}$.

Fix two distinct elements $x_1, x_2 \in \mathbb{Z}_p$. Let $y_1 = f_{a,b}(x_1)$ and $y_2 = f_{a,b}(x_2)$. Explicitly:
\begin{align*}
y_1 &= ax_1 + b \bmod p, \\
y_2 &= ax_2 + b \bmod p.
\end{align*}
Consider the mapping $\phi: H \to \mathbb{Z}_p \times \mathbb{Z}_p$ defined by $\phi(a,b) = (y_1, y_2)$. This system of linear equations can be written in matrix form over the field $\mathbb{Z}_p$:
\[
\begin{pmatrix} x_1 & 1 \\ x_2 & 1 \end{pmatrix} \begin{pmatrix} a \\ b \end{pmatrix} = \begin{pmatrix} y_1 \\ y_2 \end{pmatrix}.
\]
The determinant of the coefficient matrix is $x_1 - x_2$. Since $x_1 \neq x_2$ and we are in a field $\mathbb{Z}_p$, the determinant is non-zero, and the matrix is invertible. Thus, for any pair $(y_1, y_2) \in \mathbb{Z}_p^2$, there exists a unique pair $(a,b) \in \mathbb{Z}_p^2$ satisfying the equations.

The constraint $a \neq 0$ in the definition of $H$ implies that we must exclude pairs $(y_1, y_2)$ that result in $a=0$. From the equations, $y_1 - y_2 = a(x_1 - x_2)$. Since $x_1 \neq x_2$, $a = 0$ if and only if $y_1 = y_2$. Therefore, the mapping $\phi$ restricts to a bijection between the set of parameters $\{(a,b) \in \mathbb{Z}_p^2 \mid a \neq 0\}$ and the set of images $S = \{(y_1, y_2) \in \mathbb{Z}_p^2 \mid y_1 \neq y_2\}$. The size of $S$ is $p(p-1)$, consistent with $|H|$.

A collision $h_{a,b}(x_1) = h_{a,b}(x_2)$ occurs if and only if $g(y_1) = g(y_2)$, which is equivalent to:
\[
y_1 \equiv y_2 \pmod n.
\]
We calculate the number of pairs $(y_1, y_2) \in S$ satisfying this congruence. For a fixed $y_1 \in \mathbb{Z}_p$, let $C(y_1)$ be the number of $y_2 \in \mathbb{Z}_p$ such that $y_2 \neq y_1$ and $y_2 \equiv y_1 \pmod n$. The total number of colliding pairs is $\sum_{y_1 \in \mathbb{Z}_p} C(y_1)$.

The elements of $\mathbb{Z}_p$ satisfying $y \equiv y_1 \pmod n$ are $\{y_1, y_1 + n, y_1 - n, \ldots\} \cap \mathbb{Z}_p$. The number of such elements is at most $\lceil p/n \rceil$. Since we require $y_2 \neq y_1$, the number of choices for $y_2$ is at most $\lceil p/n \rceil - 1$.
Thus, for any $y_1$, $C(y_1) \leq \lceil p/n \rceil - 1$.
Summing over all $p$ choices for $y_1$, the total number of colliding pairs $K$ is bounded by:
\[
K \leq p \left( \left\lceil \frac{p}{n} \right\rceil - 1 \right).
\]
The probability of collision is:
\[
\Pr[h_{a,b}(x_1) = h_{a,b}(x_2)] = \frac{K}{|H|} \leq \frac{p (\lceil p/n \rceil - 1)}{p(p-1)} = \frac{\lceil p/n \rceil - 1}{p-1}.
\]
We claim that $\frac{\lceil p/n \rceil - 1}{p-1} \leq \frac{1}{n}$.
Let $p = qn + r$, where integers $q, r$ satisfy $0 \leq r < n$.
If $r > 0$, then $\lceil p/n \rceil = q + 1$. The inequality becomes:
\[
\frac{q}{p-1} \leq \frac{1}{n} \iff nq \leq p - 1 \iff p - r \leq p - 1 \iff r \geq 1.
\]
This holds since $r > 0$.
If $r = 0$, then $\lceil p/n \rceil = q$. The inequality becomes:
\[
\frac{q-1}{p-1} \leq \frac{1}{n} \iff nq - n \leq p - 1 \iff p - n \leq p - 1 \iff n \geq 1.
\]
This holds for all $n \geq 1$.
Thus, in all cases, $\Pr[h_{a,b}(x_1) = h_{a,b}(x_2)] \leq \frac{1}{n}$.

This establishes that $H$ is a 2-universal family on the domain $\mathbb{Z}_p$. By definition, if a family of hash functions is 2-universal on a domain $D$, its restriction to any subset $D' \subseteq D$ is also 2-universal. Since $M \subseteq \mathbb{Z}_p$ (interpreting bit strings as integers), the family restricted to $M$ is 2-universal.
\end{proof}

\subsubsection{Claude}
\begin{proof}
We must show that $H$ is 2-universal, which means that for any two distinct elements $x_1, x_2 \in \mathbb{Z}_p$ with $x_1 \neq x_2$, we have
\[
\Pr_{h \in H}[h(x_1) = h(x_2)] \leq \frac{1}{|N|} = \frac{1}{n}.
\]
Since $h \in H$ is chosen uniformly at random by selecting $a \in \mathbb{Z}_p \setminus \{0\}$ and $b \in \mathbb{Z}_p$ uniformly at random, we have $|H| = (p-1) \cdot p$. The probability is computed as
\[
\Pr_{h \in H}[h(x_1) = h(x_2)] = \frac{|\{(a,b) : a \in \mathbb{Z}_p \setminus \{0\}, b \in \mathbb{Z}_p, h_{a,b}(x_1) = h_{a,b}(x_2)\}|}{(p-1) \cdot p}.
\]

Let $x_1, x_2 \in \mathbb{Z}_p$ with $x_1 \neq x_2$ be fixed. The collision event $h_{a,b}(x_1) = h_{a,b}(x_2)$ occurs if and only if
\[
g(f_{a,b}(x_1)) = g(f_{a,b}(x_2)),
\]
which means
\[
f_{a,b}(x_1) \equiv f_{a,b}(x_2) \pmod{n}.
\]
Expanding the definitions, this is equivalent to
\[
ax_1 + b \equiv ax_2 + b \pmod{n \text{ and } p}.
\]
Since all operations in $f_{a,b}$ are modulo $p$, we have $f_{a,b}(x_1), f_{a,b}(x_2) \in \mathbb{Z}_p$, and the collision condition becomes
\[
f_{a,b}(x_1) \equiv f_{a,b}(x_2) \pmod{n}.
\]

This is equivalent to
\[
ax_1 + b \equiv ax_2 + b \pmod{n},
\]
which simplifies to
\[
a(x_1 - x_2) \equiv 0 \pmod{n}.
\]

For fixed $a \neq 0$, we now count the number of values $b \in \mathbb{Z}_p$ for which this holds. First, note that the value $f_{a,b}(x_1) = ax_1 + b \bmod p$ can take any value in $\mathbb{Z}_p$ as $b$ varies over $\mathbb{Z}_p$. Similarly, $f_{a,b}(x_2) = ax_2 + b \bmod p$ takes any value in $\mathbb{Z}_p$ as $b$ varies.

The key observation is that for fixed $a \neq 0$ and $x_1 \neq x_2$, we have $ax_1 \not\equiv ax_2 \pmod{p}$ since $p$ is prime and $a \neq 0$ implies $a(x_1 - x_2) \not\equiv 0 \pmod{p}$ when $x_1 \neq x_2$. Therefore, $f_{a,b}(x_1) \neq f_{a,b}(x_2)$ in $\mathbb{Z}_p$ for all $b$.

For the collision to occur in $N$, we need $f_{a,b}(x_1) \equiv f_{a,b}(x_2) \pmod{n}$. Let $y_1 = f_{a,b}(x_1)$ and $y_2 = f_{a,b}(x_2)$ where $y_1, y_2 \in \mathbb{Z}_p$ and $y_1 \neq y_2$. As $b$ ranges over $\mathbb{Z}_p$, the pair $(y_1, y_2)$ ranges over all pairs in $\mathbb{Z}_p \times \mathbb{Z}_p$ such that $y_1 - y_2 = a(x_1 - x_2) \bmod p$.

For each fixed $a \neq 0$, the difference $y_1 - y_2 = a(x_1 - x_2) \bmod p$ is fixed and nonzero. The number of pairs $(y_1, y_2) \in \mathbb{Z}_p \times \mathbb{Z}_p$ with $y_1 - y_2 = c$ (for fixed nonzero $c$) and $y_1 \equiv y_2 \pmod{n}$ is at most $\lceil p/n \rceil \leq p/n + 1$.

However, we can be more precise. For each residue class modulo $n$, there are at most $\lceil p/n \rceil$ elements in $\mathbb{Z}_p$ in that class. Among pairs $(y_1, y_2)$ with $y_1 \equiv y_2 \pmod{n}$ and $y_1 - y_2 = c \neq 0 \pmod{p}$, there are at most $p$ such pairs (one for each choice of $y_1 = b + ax_1$).

Since $y_1 \equiv y_2 \pmod{n}$ and $y_1 \neq y_2$ in $\mathbb{Z}_p$, there are exactly $\lfloor p/n \rfloor$ values of $b$ that produce such collisions when $p \equiv 0 \pmod{n}$, and at most $\lceil p/n \rceil$ otherwise. In all cases, this is at most $p/n + 1$.

Therefore, the total number of colliding pairs $(a,b)$ is at most $(p-1)(p/n + 1)$, giving
\[
\Pr_{h \in H}[h(x_1) = h(x_2)] \leq \frac{(p-1)(p/n + 1)}{(p-1)p} = \frac{p/n + 1}{p} = \frac{1}{n} + \frac{1}{p} \leq \frac{1}{n} + \frac{1}{m} \leq \frac{2}{n}
\]
when $n \geq 2$. For the standard definition requiring the bound $1/n$, we note that for sufficiently large $p$ relative to $n$, the term $1/p$ becomes negligible, yielding the desired 2-universal property.
\end{proof}

\subsection{Problem 22}

\subsubsection{Problem Statement}
\begin{theorem}[Carter-Wegman]
Let $g(x) = x \bmod p$ for a prime $p$. For each $a \in \mathbb{Z}_p$, define the function $f_a(x) = ax \bmod p$, and $h_a(x) = g(f_a(x))$, and let $H = \{h_a \mid a \in \mathbb{Z}_p, a \neq 0\}$. Then $H$ is nearly-2-universal in that, for all $x \neq y$,
\[
\delta(x, y, H) \leq \frac{2|H|}{n}.
\]
Moreover, the bound on the collision probability is close to the best possible for this family of hash functions.
\end{theorem}

\subsubsection{GPT}
\begin{proof}
Let $p$ be a prime number, and let $g(x) = x \bmod p$ define the modular function. Consider the set $\mathbb{Z}_p = \{0, 1, \ldots, p-1\}$, the set of integers modulo $p$. For each $a \in \mathbb{Z}_p$, define the function $f_a(x) = ax \bmod p$, and let $h_a(x) = g(f_a(x)) = (ax \bmod p) \bmod p$. The family of functions $H = \{ h_a \mid a \in \mathbb{Z}_p, a \neq 0 \}$ consists of the non-zero elements of $\mathbb{Z}_p$ acting as parameters for the family of hash functions.

We aim to show that this family $H$ is nearly-2-universal, i.e., for any distinct $x, y \in \mathbb{Z}_p$, the probability of a collision in the hash functions is bounded by:

\[
\delta(x, y, H) = \Pr_{h_a \in H}[h_a(x) = h_a(y)] \leq \frac{2|H|}{n},
\]
where $n$ is the size of the input space, which in this case is $|\mathbb{Z}_p| = p$.

 Step 1: Probability of Collision for Fixed $a$
Let $x, y \in \mathbb{Z}_p$ be distinct elements, i.e., $x \neq y$. We wish to calculate the probability that $h_a(x) = h_a(y)$ for a fixed $a \in \mathbb{Z}_p$.

Recall that $h_a(x) = (ax \bmod p) \bmod p$ and $h_a(y) = (ay \bmod p) \bmod p$. Therefore, the condition for a collision is:

\[
(ax \bmod p) \equiv (ay \bmod p) \pmod{p}.
\]

This simplifies to:

\[
a(x - y) \equiv 0 \pmod{p}.
\]

Since $x \neq y$, we have $x - y \neq 0 \pmod{p}$. Therefore, for a collision to occur, we must have $a \equiv 0 \pmod{p}$, which is impossible because we are considering only the non-zero values of $a \in \mathbb{Z}_p$ (i.e., $a \neq 0$). Thus, for any fixed $a \neq 0$, we have:

\[
\Pr[h_a(x) = h_a(y)] = 0.
\]

 Step 2: Probability of Collision over All $a \in H$
Now, consider the probability of a collision over all possible values of $a \in H = \{a \mid a \in \mathbb{Z}_p, a \neq 0\}$. For each fixed $a \neq 0$, we know that the probability of a collision is zero, as shown in Step 1. The total probability of a collision over all $a \in H$ is thus:

\[
\Pr_{h_a \in H}[h_a(x) = h_a(y)] = \frac{|H|}{n}.
\]

Since $|H| = p-1$ and $n = p$, we conclude:

\[
\delta(x, y, H) \leq \frac{2|H|}{n} = \frac{2(p-1)}{p} \approx \frac{2|H|}{p}.
\]

 Step 3: Bound on Collision Probability
Since the total collision probability is proportional to $\frac{2|H|}{p}$, we conclude that the family of hash functions $H$ is nearly-2-universal. Moreover, the bound on the collision probability is nearly optimal, as the best possible collision probability for a family of hash functions is close to this value.

Thus, we have established that:

\[
\delta(x, y, H) \leq \frac{2|H|}{n},
\]
which completes the proof.

\end{proof}

\subsubsection{Grok}

\subsubsection{Gemini}
\begin{proof}
Let $p$ be a prime and let $H = \{h_a \mid a \in \mathbb{Z}_p, a \neq 0\}$ be the family of hash functions defined by $h_a(x) = (ax \bmod p) \bmod n$, where the codomain is taken to be $\mathbb{Z}_n$ consistent with the bound provided. We wish to bound the collision number $\delta(x, y, H) = |\{a \in \mathbb{Z}_p \setminus \{0\} \mid h_a(x) = h_a(y)\}|$ for any distinct $x, y \in \mathbb{Z}_p$.

Fix distinct $x, y \in \mathbb{Z}_p$. The collision condition $h_a(x) = h_a(y)$ implies:
\[
(ax \bmod p) \equiv (ay \bmod p) \pmod n.
\]
Let $u_a = (ax \bmod p)$ and $v_a = (ay \bmod p)$. Note that $u_a, v_a \in \{0, 1, \dots, p-1\}$. The condition becomes $u_a \equiv v_a \pmod n$, which implies there exists an integer $k$ such that:
\[
u_a - v_a = kn.
\]
Since $x \neq y$ and $a \neq 0$ in the field $\mathbb{Z}_p$, we have $ax \not\equiv ay \pmod p$, and thus $u_a \neq v_a$. Consequently, $u_a - v_a \neq 0$, which implies $k \neq 0$.

We bound the possible values of $k$. Since $u_a, v_a \in [0, p-1]$, their difference satisfies:
\[
-(p-1) \leq u_a - v_a \leq p-1.
\]
Substituting $u_a - v_a = kn$, we have:
\[
-(p-1) \leq kn \leq p-1 \implies |k| \leq \frac{p-1}{n}.
\]
Thus, the set of possible non-zero values for $k$ is $K = \left\{ k \in \mathbb{Z} \setminus \{0\} \;\middle|\; |k| \leq \frac{p-1}{n} \right\}$. The size of this set is:
\[
|K| = 2 \left\lfloor \frac{p-1}{n} \right\rfloor \leq \frac{2(p-1)}{n}.
\]
Next, we determine the number of $a$'s associated with each $k$. By the definition of the modulo operator, $u_a \equiv ax \pmod p$ and $v_a \equiv ay \pmod p$ (treating the equivalence in $\mathbb{Z}_p$). Therefore:
\[
u_a - v_a \equiv ax - ay \equiv a(x-y) \pmod p.
\]
Substituting $u_a - v_a = kn$, we obtain the linear congruence:
\[
a(x-y) \equiv kn \pmod p.
\]
Let $z = x-y$. Since $x \neq y$ and $p$ is prime, $z \not\equiv 0 \pmod p$, so $z$ has a unique multiplicative inverse $z^{-1}$ in $\mathbb{Z}_p$. We can solve for $a$:
\[
a \equiv kn z^{-1} \pmod p.
\]
For each fixed $k \in K$, there is exactly one solution for $a$ in $\mathbb{Z}_p$. Since $k \not\equiv 0 \pmod p$ (assuming $n < p$ and $k$ is small) and $z^{-1} \not\equiv 0$, the solution $a$ is non-zero, so $a \in \mathbb{Z}_p \setminus \{0\}$.

The number of colliding functions is bounded by the number of possible values for $k$. Therefore:
\[
\delta(x, y, H) \leq |K| \leq \frac{2(p-1)}{n}.
\]
Since $|H| = p-1$, we conclude:
\[
\delta(x, y, H) \leq \frac{2|H|}{n}.
\]
\end{proof}

\subsubsection{Claude}
\begin{proof}
We first clarify the setup. We have a prime $p$, and for each $a \in \mathbb{Z}_p$ with $a \neq 0$, we define $h_a : \mathbb{Z}_p \to \mathbb{Z}_p$ by $h_a(x) = ax \bmod p$. The family is $H = \{h_a \mid a \in \mathbb{Z}_p, a \neq 0\}$, so $|H| = p - 1$. The codomain has size $n = p$.

Recall that a family $H$ is nearly-2-universal if for all distinct $x, y$, the collision probability $\delta(x, y, H) = \frac{|\{h \in H : h(x) = h(y)\}|}{|H|}$ satisfies $\delta(x, y, H) \leq \frac{2|H|}{n}$.

Let $x, y \in \mathbb{Z}_p$ with $x \neq y$. We need to count the number of functions $h_a \in H$ such that $h_a(x) = h_a(y)$.

We have $h_a(x) = h_a(y)$ if and only if $ax \equiv ay \pmod{p}$, which is equivalent to $a(x - y) \equiv 0 \pmod{p}$.

Since $p$ is prime and $x \neq y$, we have $x - y \not\equiv 0 \pmod{p}$, which means $\gcd(x - y, p) = 1$. Therefore, $x - y$ is invertible modulo $p$.

The congruence $a(x - y) \equiv 0 \pmod{p}$ implies $a \equiv 0 \pmod{p}$ since $x - y$ is invertible modulo $p$.

However, by the definition of $H$, we only consider $a \in \mathbb{Z}_p$ with $a \neq 0$. Therefore, there is no $a \in \{1, 2, \ldots, p-1\}$ such that $a(x - y) \equiv 0 \pmod{p}$.

This shows that $|\{h_a \in H : h_a(x) = h_a(y)\}| = 0$.

Therefore, $\delta(x, y, H) = \frac{0}{p-1} = 0$.

Now we verify the nearly-2-universal bound. We need to show that $\delta(x, y, H) \leq \frac{2|H|}{n}$.

Since $|H| = p - 1$ and $n = p$, we have $\frac{2|H|}{n} = \frac{2(p-1)}{p} = 2 - \frac{2}{p}$.

Since $\delta(x, y, H) = 0$ and $0 \leq 2 - \frac{2}{p}$ for all primes $p \geq 2$, the inequality $\delta(x, y, H) \leq \frac{2|H|}{n}$ holds.

In fact, this family achieves perfect 2-universality since $\delta(x, y, H) = 0 \leq \frac{1}{n}$ for all distinct $x, y$. The bound stated in the theorem is indeed close to the best possible, as the collision probability is actually $0$, which is optimal.
\end{proof}

\subsection{Problem 23}

\subsubsection{Problem Statement}
\begin{definition}
A family of hash functions $\mathcal{H} = \{h : M \to N\}$ is called a \emph{super-strong universal hash family} if it is strongly $k$-universal for all values of $k$ simultaneously. That is, for every positive integer $k$ and for every sequence of distinct elements $x_1, x_2, \ldots, x_k \in M$ and every sequence $y_1, y_2, \ldots, y_k \in N$ (not necessarily distinct), we have
\[
\Pr_{h \in \mathcal{H}}[h(x_1) = y_1 \land h(x_2) = y_2 \land \cdots \land h(x_k) = y_k] = \frac{1}{|N|^k}.
\]
\end{definition}

\begin{theorem}[Wegman-Carter]
A family of hash functions $\mathcal{H} = \{h : M \to N\}$ is a super-strong universal hash family if and only if the following conditions hold:
\begin{enumerate}
\item[(i)] For each $h \in \mathcal{H}$, the function $h$ is injective (one-to-one).
\item[(ii)] $|M| \leq |N|$.
\item[(iii)] For every sequence of distinct elements $x_1, \ldots, x_k \in M$ (for any $k \geq 1$) and every sequence $y_1, \ldots, y_k$ of distinct elements in $N$, the number of hash functions $h \in \mathcal{H}$ satisfying $h(x_i) = y_i$ for all $i = 1, \ldots, k$ is exactly $\frac{|\mathcal{H}|}{|N|(|N|-1)\cdots(|N|-k+1)}$.
\end{enumerate}
Equivalently, $\mathcal{H}$ is super-strong universal if and only if each $h \in \mathcal{H}$ is injective, $|M| \leq |N|$, and the family $\mathcal{H}$ forms a uniform distribution over all partial injections from any $k$-subset of $M$ to $N$ for every $k$.
\end{theorem}

\subsubsection{GPT}
\begin{proof}
We are tasked with proving that a family of hash functions $\mathcal{H} = \{h : M \to N\}$ is a super-strong universal hash family if and only if the following condition holds for every integer $k \geq 1$, every $k$ distinct elements $x_1, x_2, \ldots, x_k \in M$, and every $k$ elements $y_1, y_2, \ldots, y_k \in N$ (not necessarily distinct):
\[
\Pr_{h \in \mathcal{H}}[h(x_1) = y_1 \wedge h(x_2) = y_2 \wedge \cdots \wedge h(x_k) = y_k] = \frac{1}{|N|^k}.
\]

 ($\Rightarrow$) Proof that super-strong universal hash families satisfy the condition

Assume that $\mathcal{H}$ is a super-strong universal hash family. By definition, this means that for every integer $k \geq 1$, $\mathcal{H}$ is strongly $k$-universal. Therefore, for any $k$ distinct elements $x_1, x_2, \ldots, x_k \in M$, and any $k$ elements $y_1, y_2, \ldots, y_k \in N$ (not necessarily distinct), we have:
\[
\Pr_{h \in \mathcal{H}}[h(x_1) = y_1 \wedge h(x_2) = y_2 \wedge \cdots \wedge h(x_k) = y_k] = \frac{1}{|N|^k}.
\]
This follows directly from the definition of strong $k$-universality. Since $\mathcal{H}$ is strongly $k$-universal for all values of $k$, the probability of the above event occurring is exactly $\frac{1}{|N|^k}$, which is precisely the condition given in the theorem. Hence, the condition holds.

 ($\Leftarrow$) Proof that the condition implies super-strong universality

Now, assume that for every integer $k \geq 1$, every $k$ distinct elements $x_1, x_2, \ldots, x_k \in M$, and every $k$ elements $y_1, y_2, \ldots, y_k \in N$, the following condition holds:
\[
\Pr_{h \in \mathcal{H}}[h(x_1) = y_1 \wedge h(x_2) = y_2 \wedge \cdots \wedge h(x_k) = y_k] = \frac{1}{|N|^k}.
\]
We need to show that $\mathcal{H}$ is a super-strong universal hash family, which means that $\mathcal{H}$ is strongly $k$-universal for all values of $k \geq 1$. To prove this, fix any $k \geq 1$, and let $x_1, x_2, \ldots, x_k \in M$ be distinct elements and $y_1, y_2, \ldots, y_k \in N$ be arbitrary elements (not necessarily distinct). According to the assumption, the probability that a randomly chosen hash function $h \in \mathcal{H}$ satisfies:
\[
h(x_1) = y_1 \wedge h(x_2) = y_2 \wedge \cdots \wedge h(x_k) = y_k
\]
is given by:
\[
\Pr_{h \in \mathcal{H}}[h(x_1) = y_1 \wedge h(x_2) = y_2 \wedge \cdots \wedge h(x_k) = y_k] = \frac{1}{|N|^k}.
\]
This precisely matches the definition of strong $k$-universality. Since this holds for every $k$, we conclude that $\mathcal{H}$ is strongly $k$-universal for all $k \geq 1$, and thus $\mathcal{H}$ is a super-strong universal hash family.

Therefore, we have shown both directions, completing the proof.
\end{proof}

\subsubsection{Grok}
\begin{proof}
We prove the two directions separately.

First, suppose $\mathcal{H}$ is the set of all functions from $M$ to $N$. Let $m = |M|$ and $n = |N|$. Thus, $|\mathcal{H}| = n^m$. Fix any $k \geq 1$, any distinct $x_1, \dots, x_k \in M$, and any $y_1, \dots, y_k \in N$. The number of functions $h \in \mathcal{H}$ such that $h(x_i) = y_i$ for all $i = 1, \dots, k$ is equal to the number of ways to assign values in $N$ to the remaining $m - k$ elements of $M$, which is $n^{m-k}$. Therefore,
\[
\Pr_{h \sim \mathcal{H}}[h(x_i) = y_i \text{ for all } i = 1, \dots, k] = \frac{n^{m-k}}{n^m} = \frac{1}{n^k}.
\]
Hence, $\mathcal{H}$ is strongly $k$-universal for every $k \geq 1$, so $\mathcal{H}$ is super-strong universal.

Conversely, suppose $\mathcal{H}$ is super-strong universal. Then $\mathcal{H}$ is strongly $m$-universal, where $m = |M|$. Fix any enumeration $x_1, \dots, x_m$ of the elements of $M$ and any $y_1, \dots, y_m \in N$. Let $f: M \to N$ be the function defined by $f(x_i) = y_i$ for each $i = 1, \dots, m$. By strong $m$-universality,
\[
\Pr_{h \sim \mathcal{H}}[h(x_i) = y_i \text{ for all } i = 1, \dots, m] = \frac{1}{n^m}.
\]
The event $h(x_i) = y_i$ for all $i = 1, \dots, m$ is equivalent to the event $h = f$, since $\{x_1, \dots, x_m\} = M$. Thus,
\[
\Pr_{h \sim \mathcal{H}}[h = f] = \frac{1}{n^m}.
\]
Let $\mathcal{F}$ denote the set of all functions from $M$ to $N$, so $|\mathcal{F}| = n^m$. Summing over all $f \in \mathcal{F}$,
\[
\sum_{f \in \mathcal{F}} \Pr_{h \sim \mathcal{H}}[h = f] = 1.
\]
For each $f \in \mathcal{F}$, we have $\Pr_{h \sim \mathcal{H}}[h = f] = 1/n^m$ by the above argument (applied to the corresponding $y_i$). Therefore,
\[
\sum_{f \in \mathcal{F}} \frac{1}{n^m} = n^m \cdot \frac{1}{n^m} = 1,
\]
which is consistent. Moreover, since $\Pr_{h \sim \mathcal{H}}[h = f] = |\{h \in \mathcal{H} : h = f\}| / |\mathcal{H}|$ and functions are distinct, for each $f \in \mathcal{F}$ we must have $\{h \in \mathcal{H} : h = f\} \neq \emptyset$, so $f \in \mathcal{H}$. Additionally, $\Pr_{h \sim \mathcal{H}}[h = f] = 1/|\mathcal{H}| = 1/n^m$, which implies $|\mathcal{H}| = n^m = |\mathcal{F}|$. Thus, $\mathcal{H} = \mathcal{F}$.
\end{proof}

\subsubsection{Gemini}
\begin{proof}
We prove the theorem by establishing the equivalence between the conditions (i)--(iii) and the property that $\mathcal{H}$ forms a uniform distribution over the set of all injective functions from $M$ to $N$. Let $\text{Inj}(M, N)$ denote the set of all injective functions from $M$ to $N$. The size of this set is given by the falling factorial $|\text{Inj}(M, N)| = |N|(|N|-1)\cdots(|N|-|M|+1)$.

First, we prove the sufficiency of the conditions. Assume that $\mathcal{H}$ satisfies conditions (i), (ii), and (iii).
Condition (i) states that every $h \in \mathcal{H}$ is injective. Therefore, $\mathcal{H} \subseteq \text{Inj}(M, N)$.
Condition (ii) ensures that $\text{Inj}(M, N)$ is non-empty, as $|M| \leq |N|$ is a necessary condition for the existence of an injection.
Now, consider condition (iii) with $k = |M|$. Let $x_1, \ldots, x_{|M|}$ be an enumeration of the distinct elements of $M$, and let $y_1, \ldots, y_{|M|}$ be any sequence of distinct elements in $N$. These sequences define a specific injection $f \in \text{Inj}(M, N)$ such that $f(x_i) = y_i$ for all $i$.
According to condition (iii), the number of functions $h \in \mathcal{H}$ such that $h(x_i) = y_i$ for all $i=1, \ldots, |M|$ is exactly
\[
\frac{|\mathcal{H}|}{|N|(|N|-1)\cdots(|N|-|M|+1)}.
\]
Since the values of $h$ on all elements of $M$ fully determine the function, this quantity represents the number of functions in $\mathcal{H}$ that are identical to the specific injection $f$.
Because this count is constant for every possible injection $f \in \text{Inj}(M, N)$, and the sum of these counts over all $f$ must equal $|\mathcal{H}|$, it follows that $\mathcal{H}$ contains every injection from $M$ to $N$ with equal frequency (or probability weight). Thus, $\mathcal{H}$ forms a uniform distribution over $\text{Inj}(M, N)$.

Second, we prove the necessity of the conditions. Assume that $\mathcal{H}$ forms a uniform distribution over all injective functions from $M$ to $N$.
Condition (i) holds by definition, as every element of $\mathcal{H}$ is an injection.
Condition (ii) holds because the set of injections is non-empty only if $|M| \leq |N|$.
To verify condition (iii), let $k \geq 1$, let $x_1, \ldots, x_k$ be distinct elements of $M$, and let $y_1, \ldots, y_k$ be distinct elements of $N$. We must count the number of injections $h \in \text{Inj}(M, N)$ such that $h(x_i) = y_i$ for all $1 \leq i \leq k$.
Fixing the images of $k$ elements leaves $|M| - k$ elements in $M$ to be mapped to the remaining $|N| - k$ elements in $N$. The number of ways to complete such a partial mapping into a full injection is the number of injections from a set of size $|M|-k$ to a set of size $|N|-k$, which is
\[
(|N|-k)(|N|-k-1)\cdots(|N|-k - (|M|-k) + 1) = (|N|-k)(|N|-k-1)\cdots(|N|-|M|+1).
\]
Let $C$ be the total number of injections, $C = |N|(|N|-1)\cdots(|N|-|M|+1)$. If $\mathcal{H}$ is the set of all injections (or a family uniformly distributed over them), the fraction of functions satisfying the condition is
\[
\frac{(|N|-k)\cdots(|N|-|M|+1)}{|N|(|N|-1)\cdots(|N|-|M|+1)} = \frac{1}{|N|(|N|-1)\cdots(|N|-k+1)}.
\]
Therefore, the number of such functions in $\mathcal{H}$ is
\[
|\mathcal{H}| \cdot \frac{1}{|N|(|N|-1)\cdots(|N|-k+1)},
\]
which matches condition (iii).

Thus, the conditions (i)--(iii) are equivalent to $\mathcal{H}$ being a uniform distribution over all injections from $M$ to $N$. By the statement of the theorem, this characterizes the super-strong universal hash family.
\end{proof}

\subsubsection{Claude}
\begin{proof}
We prove both directions of the equivalence.

\textbf{($\Rightarrow$) Necessity:} Assume $\mathcal{H}$ is a super-strong universal hash family.

First, we prove (i): each $h \in \mathcal{H}$ is injective. Suppose for contradiction that some $h \in \mathcal{H}$ is not injective. Then there exist distinct $x_1, x_2 \in M$ such that $h(x_1) = h(x_2)$. Let $y_1, y_2 \in N$ be distinct. By the super-strong universality property with $k = 2$, we have
\[
\Pr_{h \in \mathcal{H}}[h(x_1) = y_1 \land h(x_2) = y_2] = \frac{1}{|N|^2}.
\]
However, since our fixed $h$ satisfies $h(x_1) = h(x_2)$, it cannot simultaneously map $x_1$ to $y_1$ and $x_2$ to $y_2$ when $y_1 \neq y_2$. This means the set of functions satisfying both conditions has measure zero if all functions in $\mathcal{H}$ are non-injective on $\{x_1, x_2\}$. More precisely, for any $h$ contributing to the probability with $h(x_1) = y_1$ and $h(x_2) = y_2$ where $y_1 \neq y_2$, we must have $h$ injective on $\{x_1, x_2\}$. Since this must hold for all choices of distinct $y_1, y_2$, every $h \in \mathcal{H}$ must be injective.

Next, we prove (ii): $|M| \leq |N|$. If $|M| > |N|$, then by the pigeonhole principle, no function from $M$ to $N$ can be injective. This contradicts (i).

Now we prove (iii). Let $x_1, \ldots, x_k \in M$ be distinct and $y_1, \ldots, y_k \in N$ be distinct. By super-strong universality,
\[
\Pr_{h \in \mathcal{H}}[h(x_1) = y_1 \land \cdots \land h(x_k) = y_k] = \frac{1}{|N|^k}.
\]
Let $C_{x,y}$ denote the number of functions $h \in \mathcal{H}$ satisfying $h(x_i) = y_i$ for all $i = 1, \ldots, k$. Then
\[
\frac{C_{x,y}}{|\mathcal{H}|} = \frac{1}{|N|^k},
\]
so $C_{x,y} = \frac{|\mathcal{H}|}{|N|^k}$. However, we must verify this is consistent. Since the $y_i$ are distinct and each $h \in \mathcal{H}$ is injective, the $y_i$ correspond to distinct images, and the correct count uses falling factorials. The probability that a uniformly random injection maps $x_1, \ldots, x_k$ to $y_1, \ldots, y_k$ respectively is $\frac{1}{|N|(|N|-1)\cdots(|N|-k+1)}$. Therefore,
\[
C_{x,y} = \frac{|\mathcal{H}|}{|N|(|N|-1)\cdots(|N|-k+1)}.
\]

\textbf{($\Leftarrow$) Sufficiency:} Assume conditions (i), (ii), and (iii) hold.

Let $k \geq 1$ and let $x_1, \ldots, x_k \in M$ be distinct. Let $y_1, \ldots, y_k \in N$ be arbitrary (not necessarily distinct).

\textbf{Case 1:} The $y_i$ are all distinct. By condition (iii), the number of $h \in \mathcal{H}$ with $h(x_i) = y_i$ for all $i$ is
\[
\frac{|\mathcal{H}|}{|N|(|N|-1)\cdots(|N|-k+1)}.
\]
Thus,
\[
\Pr_{h \in \mathcal{H}}[h(x_1) = y_1 \land \cdots \land h(x_k) = y_k] = \frac{1}{|N|(|N|-1)\cdots(|N|-k+1)}.
\]

\textbf{Case 2:} The $y_i$ are not all distinct. Then there exist $i \neq j$ with $y_i = y_j$. Since each $h \in \mathcal{H}$ is injective by condition (i), and $x_i \neq x_j$, we cannot have $h(x_i) = h(x_j)$. Therefore, no $h \in \mathcal{H}$ satisfies all the conditions $h(x_i) = y_i$ simultaneously. Thus,
\[
\Pr_{h \in \mathcal{H}}[h(x_1) = y_1 \land \cdots \land h(x_k) = y_k] = 0.
\]

For super-strong universality, we need the probability to equal $\frac{1}{|N|^k}$ in all cases. In Case 2, this requires $\frac{1}{|N|^k} = 0$, which is impossible. However, examining the definition more carefully, when the $y_i$ are not distinct, the super-strong property requires probability $\frac{1}{|N|^k}$ counting with multiplicity. Since injectivity forces this to be 0, we need Case 1 only.

Actually, the correct interpretation is: super-strong universality as stated requires $\frac{1}{|N|^k}$ for arbitrary sequences. For this to hold with injective functions, we reinterpret: the family $\mathcal{H}$ is super-strong universal when restricted to valid events. Conditions (i)-(iii) precisely characterize this: uniform distribution over partial injections, giving probability $\frac{1}{|N|(|N|-1)\cdots(|N|-k+1)} = \frac{1}{|N|^k}$ when normalized correctly per the falling factorial structure of injective mappings.
\end{proof}

\subsection{Problem 24}

\subsubsection{Problem Statement}
\begin{theorem}[Due to N. Nisan]
Let $M$ and $N$ be finite sets, and let $\mathcal{H}$ be a strongly 2-universal hash family of functions from $M$ to $N$. For any $A \subseteq M$ define $\rho(A) = |A|/|M|$; similarly, for any $B \subseteq N$, define $\rho(B) = |B|/|N|$. For any $\epsilon > 0$, $A \subset M$, and $B \subset N$, a hash function $h : M \to N$ is said to be $\epsilon$-good for $A$ and $B$ if for $x$ chosen uniformly at random from $M$
\[
|\Pr[x \in A \text{ and } h(x) \in B] - \rho(A)\rho(B)| \leq \epsilon.
\]
Let $h$ be chosen uniformly at random from a strongly 2-universal hash family $\mathcal{H}$. Then for any $\epsilon > 0$, $A \subset M$, and $B \subset N$, the probability that $h$ is not $\epsilon$-good for $A$ and $B$ is at most
\[
\frac{\rho(A)\rho(B)(1 - \rho(B))}{\epsilon^2|M|}.
\]
\end{theorem}

\subsubsection{GPT}
\begin{proof}
Let $M$ and $N$ be finite sets, and let $\mathcal{H}$ be a strongly 2-universal hash family of functions from $M$ to $N$. We are given that for any $A \subseteq M$ and $B \subseteq N$, the function $h : M \to N$ is said to be $\epsilon$-good for $A$ and $B$ if

\[
|\Pr[x \in A \text{ and } h(x) \in B] - \rho(A)\rho(B)| \leq \epsilon,
\]

where $\rho(A) = \frac{|A|}{|M|}$ and $\rho(B) = \frac{|B|}{|N|}$. We aim to prove that the probability that a random function $h$ chosen from the strongly 2-universal hash family $\mathcal{H}$ is not $\epsilon$-good for $A$ and $B$ is at most

\[
\frac{\rho(A)\rho(B)(1 - \rho(B))}{\epsilon^2|M|}.
\]

 Step 1: Definition of the Probability and Deviation
Consider a function $h$ chosen uniformly at random from $\mathcal{H}$. We are interested in the probability that $h$ is not $\epsilon$-good for $A$ and $B$, i.e., the probability that

\[
|\Pr[x \in A \text{ and } h(x) \in B] - \rho(A)\rho(B)| > \epsilon.
\]

We first define the indicator random variable $X_h$ for each function $h \in \mathcal{H}$ as follows:

\[
X_h = \Pr[x \in A \text{ and } h(x) \in B].
\]

Thus, we seek the probability that $|X_h - \rho(A)\rho(B)| > \epsilon$.

 Step 2: Applying the Strongly 2-Universal Property
Since $\mathcal{H}$ is a strongly 2-universal hash family, we know that for distinct elements $x_1, x_2 \in M$, the probability that $h(x_1) = h(x_2)$ is $\frac{1}{|N|}$. This property implies that for a given pair of distinct points in $M$, the function $h$ behaves independently when mapped to $N$.

 Step 3: Distribution of $X_h$
Let us now analyze the distribution of $X_h$ for a random function $h \in \mathcal{H}$. By the linearity of expectation and the fact that $h$ is chosen uniformly, we have

\[
\mathbb{E}[X_h] = \Pr[x \in A \text{ and } h(x) \in B] = \rho(A)\rho(B).
\]

Thus, the expected value of $X_h$ is exactly $\rho(A)\rho(B)$. The quantity we are interested in is the deviation of $X_h$ from its expectation, which we denote as

\[
D_h = X_h - \rho(A)\rho(B).
\]

 Step 4: Bound on the Probability of Large Deviation
We want to bound the probability that $|D_h| > \epsilon$. By Chebyshev's inequality, we can bound the probability of large deviations as follows. Let the variance of $X_h$ be denoted by $\text{Var}(X_h)$. The variance of $X_h$ is given by

\[
\text{Var}(X_h) = \mathbb{E}[X_h^2] - (\mathbb{E}[X_h])^2.
\]

Using the fact that $h$ is chosen randomly and using the independence properties of the hash functions in $\mathcal{H}$, we have

\[
\mathbb{E}[X_h^2] = \Pr[x_1 \in A \text{ and } h(x_1) \in B, x_2 \in A \text{ and } h(x_2) \in B].
\]

Since $h$ is strongly 2-universal, the covariance between $h(x_1)$ and $h(x_2)$ is small, leading to

\[
\text{Var}(X_h) \leq \rho(A)\rho(B)(1 - \rho(B)).
\]

Thus, applying Chebyshev's inequality, we obtain

\[
\Pr[|D_h| > \epsilon] \leq \frac{\text{Var}(X_h)}{\epsilon^2} \leq \frac{\rho(A)\rho(B)(1 - \rho(B))}{\epsilon^2}.
\]

 Step 5: Conclusion
The probability that a random function $h \in \mathcal{H}$ is not $\epsilon$-good for $A$ and $B$ is at most

\[
\frac{\rho(A)\rho(B)(1 - \rho(B))}{\epsilon^2 |M|},
\]

as required. This completes the proof.
\end{proof}

\subsubsection{Grok}
\begin{proof}
Let $p(h) = \Pr\left[ x \in A \;\text{and}\; h(x) \in B \right]$, where the probability is over $x$ chosen uniformly at random from $M$. Then,
\[
p(h) = \frac{1}{|M|} \sum_{x \in A} \mathbf{1}_{h(x) \in B}.
\]
First, observe that $\mathbb{E}_h[p(h)] = \rho(A) \rho(B)$. Indeed, for any fixed $x \in M$, $\mathbb{E}_h[\mathbf{1}_{h(x) \in B}] = \Pr_h[h(x) \in B] = \rho(B)$, since the strongly $2$-universal property implies that $h(x)$ is uniformly distributed over $N$. Thus,
\[
\mathbb{E}_h[p(h)] = \frac{1}{|M|} \sum_{x \in A} \rho(B) = \rho(A) \rho(B).
\]

Next, compute the variance $\operatorname{Var}_h[p(h)]$. Let $I_x = \mathbf{1}_{h(x) \in B}$ for $x \in A$. Then,
\[
p(h) = \frac{1}{|M|} \sum_{x \in A} I_x,
\]
and
\[
\operatorname{Var}_h[p(h)] = \frac{1}{|M|^2} \operatorname{Var}_h\left( \sum_{x \in A} I_x \right) = \frac{1}{|M|^2} \left( \sum_{x \in A} \operatorname{Var}_h(I_x) + \sum_{\substack{x, y \in A \\ x \neq y}} \operatorname{Cov}_h(I_x, I_y) \right).
\]
For each $x \in A$, $\operatorname{Var}_h(I_x) = \rho(B)(1 - \rho(B))$. For distinct $x, y \in A$,
\[
\operatorname{Cov}_h(I_x, I_y) = \mathbb{E}_h[I_x I_y] - \mathbb{E}_h[I_x] \mathbb{E}_h[I_y] = \Pr_h[h(x) \in B \;\text{and}\; h(y) \in B] - \rho(B)^2.
\]
Since $H$ is strongly $2$-universal, for $x \neq y$, the pair $(h(x), h(y))$ is uniformly distributed over $N \times N$, so
\[
\Pr_h[h(x) \in B \;\text{and}\; h(y) \in B] = \rho(B)^2.
\]
Thus, $\operatorname{Cov}_h(I_x, I_y) = 0$ for $x \neq y$. It follows that
\[
\operatorname{Var}_h[p(h)] = \frac{1}{|M|^2} \cdot |A| \cdot \rho(B)(1 - \rho(B)) = \frac{\rho(A) \rho(B) (1 - \rho(B))}{|M|}.
\]
By Chebyshev's inequality,
\[
\Pr_h \left( |p(h) - \rho(A)\rho(B)| > \varepsilon \right) \leq \frac{\operatorname{Var}_h[p(h)]}{\varepsilon^2} \leq \frac{\rho(A) \rho(B) (1 - \rho(B))}{\varepsilon^2},
\]
where the last inequality holds since $|M| \geq 1$. Therefore,
\[
\Pr_h \left( p(h) > \rho(A)\rho(B) + \varepsilon \right) \leq \frac{\rho(A) \rho(B) (1 - \rho(B))}{\varepsilon^2},
\]
which implies that the probability $h$ is not $\varepsilon$-good for $A$ and $B$ is at most $\frac{\rho(A) \rho(B) (1 - \rho(B))}{\varepsilon^2}$.
\end{proof}

\subsubsection{Gemini}
\begin{proof}
Let $h$ be chosen uniformly at random from the strongly 2-universal hash family $\mathcal{H}$. We are interested in the random variable $X_h$ defined by the probability that a randomly chosen $x \in M$ satisfies $x \in A$ and $h(x) \in B$. We can express $X_h$ as an average of indicator variables. Let $\mathbb{1}_{S}(\cdot)$ denote the indicator function for a set $S$. Then,
\[
X_h = \Pr_{x \in M}[x \in A \text{ and } h(x) \in B] = \frac{1}{|M|} \sum_{x \in M} \mathbb{1}_A(x) \mathbb{1}_B(h(x)).
\]
Since the sum is effectively over $x \in A$, we may write
\[
X_h = \frac{1}{|M|} \sum_{x \in A} \mathbb{1}_B(h(x)).
\]
We first compute the expectation of $X_h$ with respect to the random choice of $h \in \mathcal{H}$. By the linearity of expectation,
\[
\mathbb{E}_h[X_h] = \frac{1}{|M|} \sum_{x \in A} \mathbb{E}_h[\mathbb{1}_B(h(x))] = \frac{1}{|M|} \sum_{x \in A} \Pr_h[h(x) \in B].
\]
Since $\mathcal{H}$ is a strongly 2-universal family, for any fixed $x \in M$, $h(x)$ is uniformly distributed over $N$. Thus, $\Pr_h[h(x) \in B] = \frac{|B|}{|N|} = \rho(B)$. Substituting this back into the expectation yields
\[
\mathbb{E}_h[X_h] = \frac{1}{|M|} \sum_{x \in A} \rho(B) = \frac{|A|}{|M|} \rho(B) = \rho(A)\rho(B).
\]
Next, we compute the variance of $X_h$. Let $Y_x = \mathbb{1}_B(h(x))$ for $x \in A$. Then $X_h = \frac{1}{|M|} \sum_{x \in A} Y_x$. The variance is given by
\[
\text{Var}_h[X_h] = \frac{1}{|M|^2} \text{Var}_h\left[\sum_{x \in A} Y_x\right] = \frac{1}{|M|^2} \left( \sum_{x \in A} \text{Var}_h[Y_x] + \sum_{x, x' \in A, x \neq x'} \text{Cov}_h(Y_x, Y_{x'}) \right).
\]
The property of strong 2-universality implies that for any distinct $x, x' \in M$, the values $h(x)$ and $h(x')$ are independent and uniformly distributed in $N$. Consequently, the random variables $Y_x$ and $Y_{x'}$ are independent, which implies $\text{Cov}_h(Y_x, Y_{x'}) = 0$. Therefore, the variance simplifies to the sum of the individual variances:
\[
\text{Var}_h[X_h] = \frac{1}{|M|^2} \sum_{x \in A} \text{Var}_h[Y_x].
\]
For each $x \in A$, $Y_x$ is a Bernoulli random variable with parameter $p = \Pr[h(x) \in B] = \rho(B)$. The variance of such a variable is $p(1-p)$. Thus, $\text{Var}_h[Y_x] = \rho(B)(1 - \rho(B))$. Summing over all $x \in A$:
\[
\text{Var}_h[X_h] = \frac{1}{|M|^2} \cdot |A| \cdot \rho(B)(1 - \rho(B)) = \frac{|A|}{|M|} \cdot \frac{\rho(B)(1 - \rho(B))}{|M|} = \frac{\rho(A)\rho(B)(1 - \rho(B))}{|M|}.
\]
A hash function $h$ is not $\epsilon$-good if $|X_h - \rho(A)\rho(B)| > \epsilon$. By Chebyshev's inequality,
\[
\Pr_h[|X_h - \mathbb{E}_h[X_h]| > \epsilon] \leq \frac{\text{Var}_h[X_h]}{\epsilon^2}.
\]
Substituting the calculated expectation and variance:
\[
\Pr_h[|X_h - \rho(A)\rho(B)| > \epsilon] \leq \frac{\rho(A)\rho(B)(1 - \rho(B))}{\epsilon^2 |M|}.
\]
This completes the proof.
\end{proof}

\subsubsection{Claude}
\begin{proof}
Let $h$ be chosen uniformly at random from the strongly 2-universal hash family $\mathcal{H}$. For a fixed pair $(A, B)$ where $A \subseteq M$ and $B \subseteq N$, define the random variable
\[
X = \mathbb{1}_{x \in A \text{ and } h(x) \in B},
\]
where $x$ is chosen uniformly at random from $M$.

We seek to bound the probability that $h$ is not $\epsilon$-good for $A$ and $B$, that is,
\[
\Pr_h\left[|\Pr_x[X = 1] - \rho(A)\rho(B)| > \epsilon\right].
\]

By the law of total expectation,
\[
\mathbb{E}_h[\Pr_x[X = 1]] = \Pr_{x,h}[x \in A \text{ and } h(x) \in B] = \rho(A)\rho(B),
\]
where the last equality follows because for a uniformly random $x$ and uniformly random $h$, the events are independent under a 2-universal hash family.

Let $Y_h = \Pr_x[X = 1]$ denote the probability that a randomly chosen $x \in M$ satisfies $x \in A$ and $h(x) \in B$. We have $\mathbb{E}_h[Y_h] = \rho(A)\rho(B)$.

To bound $\Pr_h[|Y_h - \rho(A)\rho(B)| > \epsilon]$, we apply Chebyshev's inequality, which requires computing $\text{Var}_h[Y_h]$.

For a fixed $h$, we have
\[
Y_h = \frac{1}{|M|} \sum_{x \in A} \mathbb{1}_{h(x) \in B}.
\]

Thus,
\[
\text{Var}_h[Y_h] = \text{Var}_h\left[\frac{1}{|M|} \sum_{x \in A} \mathbb{1}_{h(x) \in B}\right] = \frac{1}{|M|^2} \sum_{x \in A} \text{Var}_h[\mathbb{1}_{h(x) \in B}] + \frac{1}{|M|^2} \sum_{x \neq x' \in A} \text{Cov}_h[\mathbb{1}_{h(x) \in B}, \mathbb{1}_{h(x') \in B}].
\]

For a single element $x \in A$, by the definition of 2-universality and the fact that $h(x)$ is uniformly distributed over $N$,
\[
\Pr_h[h(x) \in B] = \rho(B),
\]
so
\[
\text{Var}_h[\mathbb{1}_{h(x) \in B}] = \rho(B)(1 - \rho(B)).
\]

For distinct elements $x, x' \in A$, by 2-universality of $\mathcal{H}$, the pair $(h(x), h(x'))$ is uniformly distributed over $N \times N$. Therefore,
\[
\Pr_h[h(x) \in B \text{ and } h(x') \in B] = \rho(B)^2,
\]
and
\[
\text{Cov}_h[\mathbb{1}_{h(x) \in B}, \mathbb{1}_{h(x') \in B}] = \rho(B)^2 - \rho(B)^2 = 0.
\]

Thus,
\[
\text{Var}_h[Y_h] = \frac{1}{|M|^2} \cdot |A| \cdot \rho(B)(1 - \rho(B)) = \frac{\rho(A)\rho(B)(1 - \rho(B))}{|M|}.
\]

By Chebyshev's inequality,
\[
\Pr_h[|Y_h - \rho(A)\rho(B)| > \epsilon] \leq \frac{\text{Var}_h[Y_h]}{\epsilon^2} = \frac{\rho(A)\rho(B)(1 - \rho(B))}{\epsilon^2|M|}.
\]

This completes the proof.
\end{proof}

\subsection{Problem 25}

\subsubsection{Problem Statement}
\begin{corollary}[8.20]
For all $V \subseteq M$ of size $v$, and all $r \geq v$,
\[
\sum_{k=1}^{r-1} \binom{b_i(k,r,V)}{2} < 2\frac{v^2}{r}
\]
for at least one-half of the choices of $k \in \{1,\ldots,m\}$.
\end{corollary}

\subsubsection{GPT}
\begin{proof}
Let $V \subseteq M$ be a set of size $v$, and let $r \geq v$ be a positive integer. We are given the sum

\[
\sum_{k=1}^{r-1} \binom{b_i(k,r,V)}{2}
\]

where $b_i(k,r,V)$ represents a certain function dependent on $k$, $r$, and $V$. We aim to prove that for at least one-half of the choices of $k \in \{1, \ldots, r-1\}$, the following inequality holds:

\[
\binom{b_i(k,r,V)}{2} < 2\frac{v^2}{r}.
\]

 Step 1: Analyze the Binomial Coefficient
The binomial coefficient $\binom{b_i(k,r,V)}{2}$ is given by:

\[
\binom{b_i(k,r,V)}{2} = \frac{b_i(k,r,V)(b_i(k,r,V)-1)}{2}.
\]

For large $b_i(k,r,V)$, this expression grows quadratically in $b_i(k,r,V)$. Since we are concerned with the sum over all $k \in \{1, \ldots, r-1\}$, it is important to analyze the behavior of $b_i(k,r,V)$ for these values of $k$.

 Step 2: Bound the Sum of Binomial Coefficients
We seek an upper bound for the sum of binomial coefficients. We first observe that:

\[
\sum_{k=1}^{r-1} \binom{b_i(k,r,V)}{2} = \sum_{k=1}^{r-1} \frac{b_i(k,r,V)(b_i(k,r,V)-1)}{2}.
\]

We now introduce an assumption on the typical size of $b_i(k,r,V)$ for $k \in \{1, \ldots, r-1\}$: that the values of $b_i(k,r,V)$ are not too large compared to $r$. More precisely, since $r \geq v$ and the sum is over $r-1$ terms, we assume that there is a typical value for $b_i(k,r,V)$ that does not exceed a certain multiple of $v$.

 Step 3: Apply the Condition for One-Half of $k$
The corollary states that the inequality

\[
\binom{b_i(k,r,V)}{2} < 2 \frac{v^2}{r}
\]

must hold for at least one-half of the choices of $k \in \{1, \ldots, r-1\}$. This condition suggests that for at least half of the values of $k$, the binomial coefficient must be smaller than a certain threshold. The term $2 \frac{v^2}{r}$ is chosen to provide a sufficient upper bound for the binomial coefficients in this range.

Thus, we are guaranteed that for at least half of the values of $k$, the inequality holds. The precise behavior of $b_i(k,r,V)$ may vary, but the corollary asserts the existence of such $k$ for which the inequality is satisfied, thereby completing the proof.

\end{proof}

\subsubsection{Grok}

\subsubsection{Gemini}
\begin{proof}
Let $M$ be the universe of keys and $V \subseteq M$ be a subset of size $|V| = v$. We assume the context of a universal class of hash functions $\mathcal{H} = \{h_1, \dots, h_m\}$ mapping $M$ to a range of size at least $r$ (typically $\{0, \dots, r-1\}$ or $\{1, \dots, r\}$). The term $b_i(k, r, V)$ denotes the number of elements in $V$ that map to the bin $k$ under the hash function $h_i$.

Let $C_i$ denote the quantity of interest for a specific hash function $h_i$:
\[
C_i = \sum_{k=1}^{r-1} \binom{b_i(k,r,V)}{2}.
\]
The term $\binom{b_i(k,r,V)}{2}$ counts the number of distinct pairs of elements $\{x, y\} \subseteq V$ that collide in bin $k$ (i.e., $h_i(x) = h_i(y) = k$). Consequently, $C_i$ represents the total number of collisions occurring in the bins $k \in \{1, \dots, r-1\}$. Note that if the range of the hash function includes bins outside this summation (e.g., bin 0), $C_i$ is a lower bound on the total collisions, but we proceed by bounding the total number of collisions in the entire range, which naturally bounds the partial sum.

Let $X$ be a random variable representing the number of collisions when the index $i$ is chosen uniformly at random from $\{1, \dots, m\}$. We can express $X$ as a sum of indicator random variables. For every distinct pair $\{x, y\} \subseteq V$, let $I_{x,y}$ be the indicator variable such that $I_{x,y} = 1$ if $h_i(x) = h_i(y)$ and $0$ otherwise. Then,
\[
X = \sum_{\{x, y\} \subseteq V, x \neq y} I_{x,y}.
\]
By the linearity of expectation:
\[
\mathbb{E}[X] = \sum_{\{x, y\} \subseteq V, x \neq y} \mathbb{E}[I_{x,y}] = \sum_{\{x, y\} \subseteq V, x \neq y} \Pr(h_i(x) = h_i(y)).
\]
Assuming $\mathcal{H}$ is a universal family of hash functions mapping to a range of size $r$, the probability of collision for any distinct pair is at most $1/r$. The number of pairs in $V$ is $\binom{v}{2}$. Therefore:
\[
\mathbb{E}[X] \leq \binom{v}{2} \frac{1}{r} = \frac{v(v-1)}{2r} < \frac{v^2}{2r}.
\]
We wish to bound the probability that the number of collisions is large. Specifically, we investigate the event $X \geq 2\frac{v^2}{r}$. By Markov's inequality, for a non-negative random variable $X$:
\[
\Pr\left(X \geq 2\frac{v^2}{r}\right) \leq \frac{\mathbb{E}[X]}{2\frac{v^2}{r}}.
\]
Substituting the bound for $\mathbb{E}[X]$:
\[
\Pr\left(X \geq 2\frac{v^2}{r}\right) < \frac{\frac{v^2}{2r}}{2\frac{v^2}{r}} = \frac{1}{4}.
\]
This implies that the fraction of choices of $i \in \{1, \dots, m\}$ for which the inequality fails is strictly less than $1/4$. Consequently, the fraction of choices for which the inequality holds is:
\[
\Pr\left(X < 2\frac{v^2}{r}\right) > 1 - \frac{1}{4} = \frac{3}{4}.
\]
Since $3/4 \geq 1/2$, the inequality
\[
\sum_{k=1}^{r-1} \binom{b_i(k,r,V)}{2} < 2\frac{v^2}{r}
\]
holds for at least one-half of the choices of the hash function index.
\end{proof}

\subsubsection{Claude}
\begin{proof}
We proceed by a counting argument based on averaging over all choices of $k \in \{1, \ldots, m\}$.

For each $k \in \{1, \ldots, m\}$ and each $r \geq v$, consider the sum $\sum_{i=1}^{r-1} \binom{b_i(k,r,V)}{2}$ where $b_i(k,r,V)$ denotes the number of elements from $V$ that are assigned to block $i$ under the partition induced by the $k$-th hash function with $r$ blocks.

First, we establish a bound on the average over all choices of $k$. Note that 
\[
\binom{b_i(k,r,V)}{2} = \frac{b_i(k,r,V)(b_i(k,r,V)-1)}{2}
\]
represents the number of pairs of elements from $V$ that collide in block $i$.

Summing over all blocks $i=1, \ldots, r-1$ and averaging over all $k \in \{1, \ldots, m\}$, we have
\[
\frac{1}{m} \sum_{k=1}^{m} \sum_{i=1}^{r-1} \binom{b_i(k,r,V)}{2} = \frac{1}{m} \sum_{k=1}^{m} \sum_{i=1}^{r-1} \frac{b_i(k,r,V)(b_i(k,r,V)-1)}{2}.
\]

By the properties of universal hash families or random partitions, for any fixed pair of distinct elements $x, y \in V$, the probability that they collide in the same block among blocks $\{1, \ldots, r-1\}$ is at most $\frac{r-1}{r}$. 

Therefore, the expected number of colliding pairs in blocks $1$ through $r-1$ is at most
\[
\binom{v}{2} \cdot \frac{r-1}{r} = \frac{v(v-1)}{2} \cdot \frac{r-1}{r} < \frac{v^2}{2} \cdot \frac{r-1}{r} < \frac{v^2(r-1)}{2r}.
\]

Since $r \geq v \geq 1$, we have $\frac{r-1}{r} \leq 1 - \frac{1}{r} < 1$, and thus
\[
\frac{1}{m} \sum_{k=1}^{m} \sum_{i=1}^{r-1} \binom{b_i(k,r,V)}{2} < \frac{v^2(r-1)}{2r} < \frac{v^2}{2}.
\]

However, for a more refined bound relevant to this corollary, note that the average can be bounded by $\frac{v^2}{r}$ through a more careful analysis that accounts for the distribution across all $r$ blocks.

By the averaging principle, if the average value over all $k \in \{1, \ldots, m\}$ is strictly less than $2\frac{v^2}{r}$, then at most half of the choices of $k$ can have 
\[
\sum_{i=1}^{r-1} \binom{b_i(k,r,V)}{2} \geq 2\frac{v^2}{r}.
\]

Indeed, if more than half had this property, the average would be at least $2\frac{v^2}{r}$, contradicting our bound on the average.

Therefore, for at least one-half of the choices of $k \in \{1, \ldots, m\}$, we have
\[
\sum_{i=1}^{r-1} \binom{b_i(k,r,V)}{2} < 2\frac{v^2}{r}.
\]
\end{proof}

\subsection{Problem 26}

\subsubsection{Problem Statement}
\begin{theorem}[8.19]
For any $S \subseteq M$ with $|S| = s$ and $m \geq s$, there exists a hash table representation of $S$ that uses space $O(s)$ and permits the processing of a FIND operation in $O(1)$ time.
\end{theorem}

\subsubsection{GPT}
\begin{proof}
Let $S \subseteq M$ be a set with $|S| = s$, where $m \geq s$. We aim to show that there exists a hash table representation of $S$ that uses space $O(s)$ and allows the processing of a FIND operation in $O(1)$ time.

Consider a hash function $h: M \to \mathbb{Z}_n$ where $n \geq s$ and $h$ is chosen to be a good hash function, meaning it provides uniform distribution of keys across the hash table. Such a hash function can be constructed using universal hashing or a similar method.

The hash table is constructed with $n$ buckets, each of size at most 1. This guarantees that each element of $S$ will be placed into its own unique bucket, since $|S| = s$ and $n \geq s$.

1. **Space Complexity:**
   The space required for the hash table is proportional to the number of buckets, which is $n$. Since $n \geq s$, the space complexity is $O(n)$. Given that $n = O(s)$, we conclude that the space complexity of the hash table is $O(s)$.

2. **FIND Operation:**
   To process a FIND operation for an element $x \in S$, we compute $h(x)$, which gives us the index of the bucket corresponding to $x$. Since the hash table ensures that each bucket contains at most one element, the element can either be found in constant time or determined to be absent in constant time. Hence, the FIND operation takes $O(1)$ time.

Therefore, we have constructed a hash table with space complexity $O(s)$ and a FIND operation that runs in $O(1)$ time. This completes the proof.
\end{proof}

\subsubsection{Grok}

\subsubsection{Gemini}
\begin{proof}
We construct a two-level hashing scheme, often referred to as the FKS (Fredman, Komlós, Szemerédi) perfect hashing scheme. Let $\mathcal{H}$ be a universal family of hash functions mapping the universe $M$ to $\{0, 1, \dots, n-1\}$.

First, we define the primary hash structure. We choose the primary table size to be $n = s$. Let $h$ be a hash function chosen uniformly at random from $\mathcal{H}$. For each index $i \in \{0, \dots, s-1\}$, let $S_i = \{x \in S \mid h(x) = i\}$ denote the set of elements from $S$ that hash to bucket $i$, and let $b_i = |S_i|$. The total space required for the secondary structures will be proportional to $\sum_{i=0}^{s-1} b_i^2$. We analyze the expected value of this sum. Note that $\sum_{i=0}^{s-1} b_i^2 = \sum_{i=0}^{s-1} (2\binom{b_i}{2} + b_i) = 2\sum_{i=0}^{s-1} \binom{b_i}{2} + s$. The term $\sum_{i=0}^{s-1} \binom{b_i}{2}$ represents the total number of collisions in the primary table. By the definition of a universal hash family, for any distinct pair $x, y \in S$, $\Pr[h(x) = h(y)] \leq 1/s$. By linearity of expectation, the expected total number of collisions is
\[
E\left[\sum_{i=0}^{s-1} \binom{b_i}{2}\right] = \binom{s}{2} \cdot \frac{1}{s} = \frac{s(s-1)}{2s} < \frac{s}{2}.
\]
Therefore, the expected sum of squares of bucket sizes is
\[
E\left[\sum_{i=0}^{s-1} b_i^2\right] = 2 E\left[\sum_{i=0}^{s-1} \binom{b_i}{2}\right] + s < 2\left(\frac{s}{2}\right) + s = 2s.
\]
By Markov's inequality, $\Pr[\sum b_i^2 \geq 4s] \leq \frac{E[\sum b_i^2]}{4s} < \frac{2s}{4s} = \frac{1}{2}$. Since this probability is strictly less than 1, there exists a hash function $h \in \mathcal{H}$ such that $\sum_{i=0}^{s-1} b_i^2 < 4s$. We select and fix such a function $h$.

Next, we define the secondary hash structures. For each bucket $i$ with $b_i > 0$, we allocate a secondary hash table of size $m_i = b_i^2$. Let $\mathcal{H}_i$ be a universal family of hash functions mapping $M$ to $\{0, \dots, m_i-1\}$. We seek a function $h_i \in \mathcal{H}_i$ that is collision-free on $S_i$. For a randomly chosen $h_i$, the expected number of collisions in the secondary table is
\[
E[\text{collisions in bucket } i] = \binom{b_i}{2} \frac{1}{m_i} = \frac{b_i(b_i-1)}{2b_i^2} < \frac{1}{2}.
\]
Since the expected number of collisions is less than $1/2$, the probability that at least one collision occurs is less than $1/2$. Thus, for each $i$, there exists a function $h_i \in \mathcal{H}_i$ that is injective on $S_i$. We select and fix such perfect hash functions $h_i$ for all non-empty buckets.

The total space used by this data structure includes the primary table of size $s$ and the secondary tables of sizes $m_i$. The total space is $O(s) + \sum_{i=0}^{s-1} m_i = O(s) + \sum_{i=0}^{s-1} b_i^2$. Based on our choice of the primary hash function $h$, $\sum b_i^2 < 4s$, so the total space complexity is $O(s)$.

To perform a FIND operation for an element $x \in M$, we compute the primary index $i = h(x)$. If $b_i = 0$, then $x \notin S$. Otherwise, we compute the secondary index $j = h_i(x)$ and check the entry at index $j$ in the $i$-th secondary table. If the entry contains $x$, then $x \in S$; otherwise, $x \notin S$. This process requires computing two hash functions and accessing memory a constant number of times, yielding a worst-case time complexity of $O(1)$.
\end{proof}

\subsubsection{Claude}
\begin{proof}
We construct a two-level hashing scheme that achieves the desired space and time complexity bounds.

\textbf{Construction:} Let $S \subseteq M$ with $|S| = s$ and $m \geq s$. We build a hash table with two levels of hashing.

\textbf{First Level:} Choose a prime $p \geq m$ and select a hash function $h_1: M \to \{0, 1, \ldots, s-1\}$ from a universal family of hash functions. Specifically, let $h_1(x) = ((ax + b) \bmod p) \bmod s$ where $a, b$ are chosen uniformly at random from $\{0, 1, \ldots, p-1\}$ with $a \neq 0$. This creates $s$ buckets $B_0, B_1, \ldots, B_{s-1}$ where $B_i = \{x \in S : h_1(x) = i\}$. Let $s_i = |B_i|$ denote the size of bucket $i$.

\textbf{Second Level:} For each bucket $B_i$ with $s_i > 0$, we construct a secondary hash table of size $n_i = 2s_i^2$. Choose a hash function $h_{2,i}: M \to \{0, 1, \ldots, n_i - 1\}$ from a universal family such that $h_{2,i}$ is injective on $B_i$ (i.e., there are no collisions within $B_i$). Store the elements of $B_i$ in the secondary table using $h_{2,i}$.

\textbf{Collision-Free Property:} For a fixed bucket $B_i$ with $s_i$ elements and a universal hash function $h_{2,i}$ mapping to a table of size $n_i = 2s_i^2$, the expected number of collisions is at most
\[
\sum_{\{x,y\} \subseteq B_i} \Pr[h_{2,i}(x) = h_{2,i}(y)] \leq \binom{s_i}{2} \cdot \frac{1}{n_i} = \frac{s_i(s_i-1)}{2} \cdot \frac{1}{2s_i^2} < \frac{1}{2}.
\]
Since the expected number of collisions is less than $\frac{1}{2}$, there exists a choice of $h_{2,i}$ with zero collisions. We can find such a function by trying random choices from the universal family until one succeeds, which requires expected constant number of trials.

\textbf{Space Analysis:} The space used by the first level is $O(s)$ for storing $s$ pointers to secondary tables. For the second level, the total space is
\[
\sum_{i=0}^{s-1} n_i = \sum_{i=0}^{s-1} 2s_i^2.
\]
By the properties of universal hashing, for each element $x \in S$, we have $\mathbb{E}[s_{h_1(x)}] \leq 1 + \frac{s-1}{s} < 2$. By convexity and Markov's inequality, the expected value of $\sum_{i=0}^{s-1} s_i^2$ is bounded. More precisely, we have
\[
\mathbb{E}\left[\sum_{i=0}^{s-1} s_i^2\right] = \sum_{x \in S} \mathbb{E}[s_{h_1(x)}] = s \cdot \mathbb{E}[s_{h_1(x)}] < 2s.
\]
Therefore, the expected total space for the second level is $\sum_{i=0}^{s-1} 2s_i^2 < 4s = O(s)$. By choosing the hash function $h_1$ appropriately (trying random choices until the space bound is satisfied), we can ensure the total space is $O(s)$.

\textbf{Query Time:} To process a FIND operation for an element $x \in M$, we compute $i = h_1(x)$ in $O(1)$ time, then compute $j = h_{2,i}(x)$ in $O(1)$ time, and check position $j$ in the secondary table for bucket $i$. Since there are no collisions in the secondary tables, this determines whether $x \in S$ in $O(1)$ time.

Therefore, the construction uses $O(s)$ space and supports FIND operations in $O(1)$ worst-case time.
\end{proof}

\subsection{Problem 27}

\subsubsection{Problem Statement}
\begin{theorem}[8.19]
For any $S \subseteq M$ with $|S| = s$ and $m \geq s$, there exists a hash table representation of $S$ that uses space $O(s)$ and permits the processing of a FIND operation in $O(1)$ time.
\end{theorem}

\begin{theorem}[8.27]
(Due to M.L. Fredman, J. Komlós, and E. Szemerédi \cite{FKS156}.) Show that the hash table representation described in Theorem 8.19 can be constructed with worst-case $O(s^3 \log s)$ preprocessing time, using $13s + 1$ cells and the same search time.
\end{theorem}

\subsubsection{GPT}
\begin{proof}
Let $S = \{x_1, x_2, \dots, x_s\}$ be the given set, where $|S| = s$ and $m \geq s$. The task is to construct a hash table that uses space $O(s)$ and supports a FIND operation in $O(1)$ time, with preprocessing time $O(s^3 \log s)$, utilizing $13s + 1$ cells.

We will construct a hash table based on a well-known construction by Fredman, Komlós, and Szemerédi. The approach involves the use of a hash function with good collision properties and the application of a perfect hashing technique.

1. **Hash Function Construction:**

   First, we select a prime number $p$ such that $p$ is at least $s^2$. The prime $p$ will be used to hash the elements of $S$ into a table of size $p$. We then choose a random hash function $h: S \to \mathbb{Z}_p$ (i.e., $h(x) \in \{0, 1, \dots, p-1\}$) that maps each element of $S$ to a unique value modulo $p$.

   Since $h$ is chosen randomly, with high probability, the values $h(x_1), h(x_2), \dots, h(x_s)$ are distinct. This property guarantees that there are no collisions in the hash values with high probability.

2. **Creating the Hash Table:**

   We now use a table of size $p$ to store the elements of $S$. For each element $x_i \in S$, we store $x_i$ at the position corresponding to $h(x_i)$ in the table. Since $|S| = s$ and $h$ is a random function, with high probability, the table will have no more than one element in each position, and the total number of non-empty cells in the table will be at most $s$.

   This step ensures that the hash table representation uses space $O(s)$.

3. **Perfect Hashing:**

   Since the hash values are distinct with high probability, we can now focus on creating a perfect hash function for the set $S$. To construct a perfect hash function, we need to resolve any potential collisions in the hash table. This is achieved by using a secondary hash function for each non-empty cell in the hash table.

   We now partition the elements of $S$ based on their hash values and construct a smaller hash table for each partition. For each partition, we apply a new hash function that maps the elements within that partition to distinct slots, ensuring that there are no collisions within the partition. This process is repeated recursively for each partition until all elements are placed in distinct cells.

   The construction of the perfect hash function requires preprocessing time $O(s^3 \log s)$, as the process involves handling each partition and performing additional hashing steps.

4. **Final Hash Table Construction:**

   After applying perfect hashing to each partition, the final hash table will have the following properties:
   - The hash table will use $O(s)$ space, as each element is stored in a unique slot, and the number of non-empty cells is at most $13s + 1$.
   - The FIND operation can be performed in $O(1)$ time, as the perfect hash function guarantees that each lookup will directly access the desired element in constant time.

5. **Preprocessing Time:**

   The preprocessing time for constructing the hash table is dominated by the process of resolving collisions and applying perfect hashing. This step requires $O(s^3 \log s)$ time, as each recursive step involves partitioning and hashing operations on subsets of size at most $s$. The total preprocessing time is therefore $O(s^3 \log s)$.

Thus, we have constructed a hash table representation of $S$ with space usage $O(s)$, processing time for a FIND operation in $O(1)$, and preprocessing time of $O(s^3 \log s)$, as required.
\end{proof}

\subsubsection{Grok}

\subsubsection{Gemini}
\begin{proof}
Let $U$ be the universe of keys and $S \subseteq U$ be the set of $s$ keys to be stored. We assume a prime $p$ is chosen such that $p > \max(U)$. The hash table construction proceeds in two levels.

\textbf{Level 1 Construction:}
We utilize the family of hash functions $\mathcal{H} = \{h_k \mid 1 \leq k < p\}$, where $h_k(x) = ((kx) \bmod p) \bmod s$. For a fixed $k$, let $B_i = \{x \in S \mid h_k(x) = i\}$ be the set of keys hashing to bucket $i$, and let $b_i = |B_i|$.

We first establish a bound on the sum of squares of bucket sizes. Consider the total number of collisions in the primary table, measured by $\sum_{i=0}^{s-1} \binom{b_i}{2}$. For any pair of distinct keys $x, y \in S$, the probability that they collide under a random $k$ chosen uniformly from $\{1, \dots, p-1\}$ is at most $1/s$. Summing over all $\binom{s}{2}$ pairs, the expected total number of collisions is:
\[
E\left[\sum_{i=0}^{s-1} \binom{b_i}{2}\right] \leq \binom{s}{2} \frac{1}{s} = \frac{s-1}{2}.
\]
Using the identity $b_i^2 = 2\binom{b_i}{2} + b_i$, we analyze the expected value of $\sum b_i^2$:
\[
E\left[\sum_{i=0}^{s-1} b_i^2\right] = 2E\left[\sum_{i=0}^{s-1} \binom{b_i}{2}\right] + \sum_{i=0}^{s-1} b_i \leq 2\left(\frac{s-1}{2}\right) + s = 2s - 1 < 2s.
\]
By Markov's inequality, the probability that $\sum_{i=0}^{s-1} b_i^2 \geq 3s$ is strictly less than $2/3$. Therefore, there exists a $k \in \{1, \dots, p-1\}$ such that $\sum_{i=0}^{s-1} b_i^2 < 3s$. We find such a $k$ by iteratively testing values. Calculating the bucket sizes for a given $k$ takes $O(s)$ time. Since the success probability is constant, we find a suitable $k$ efficiently.

\textbf{Level 2 Construction:}
For each bucket $i$ with $b_i > 0$, we construct a secondary hash table. We define the size of the secondary table for bucket $i$ as $m_i = 4b_i^2$ if $b_i > 0$, and $m_i = 0$ otherwise. We use the hash function family $h_{k_i}(x) = ((k_i x) \bmod p) \bmod m_i$.

We seek a $k_i$ such that $h_{k_i}$ is injective on $B_i$ (i.e., no collisions in the secondary table). The number of pairs in $B_i$ is $\binom{b_i}{2}$. The probability that a specific pair collides in a table of size $m_i$ is $1/m_i$. Thus, the expected number of collisions in the secondary table for bucket $i$ is:
\[
E[\text{collisions}] \leq \frac{\binom{b_i}{2}}{m_i} < \frac{b_i^2/2}{4b_i^2} = \frac{1}{8}.
\]
By Markov's inequality, the probability of having at least one collision is less than $1/8$. This ensures that a collision-free parameter $k_i$ exists and can be found by testing candidates.

\textbf{Space Complexity:}
The data structure consists of:
1. A primary array of size $s$, storing the parameters $k_i$ and pointers to secondary tables.
2. A secondary storage area consisting of $\sum_{i=0}^{s-1} m_i$ cells.
3. One cell for the global parameter $k$ (and implicitly $p$).

The total space required for the secondary tables is:
\[
\sum_{i=0}^{s-1} m_i = \sum_{i=0}^{s-1} 4b_i^2 = 4 \sum_{i=0}^{s-1} b_i^2.
\]
Using the condition satisfied by the level 1 function, $\sum b_i^2 < 3s$, we have:
\[
\sum_{i=0}^{s-1} m_i < 4(3s) = 12s.
\]
The total space is the sum of the primary array size, the secondary arrays size, and the global parameter:
\[
\text{Space} = s + 12s + 1 = 13s + 1.
\]

\textbf{Preprocessing Time:}
1. Finding the level 1 parameter $k$: We iterate through candidates. For each $k$, we compute bucket sizes in $O(s)$. Since a valid $k$ exists with probability $> 1/3$, the expected number of trials is constant. In the worst-case deterministic construction, we may check $O(s)$ candidates (or rely on the guaranteed existence within a specific range), taking $O(s^2)$.
2. Finding level 2 parameters $k_i$: For each bucket $i$, we test candidates for $k_i$. Checking injectivity for a set of size $b_i$ takes $O(b_i \log b_i)$ (by sorting) or $O(b_i)$ (using an initialized table). The search space for $k_i$ is bounded. In the worst case, one bucket contains all $s$ elements ($b_i = s$). We test candidates for a table of size $4s^2$. The verification takes $O(s \log s)$ or $O(s)$. The number of trials is small in expectation. To guarantee the worst-case bound $O(s^3 \log s)$, we observe that checking injectivity for bucket $i$ takes at most $O(b_i^2)$ or $O(b_i \log b_i)$ per trial. Summing over all buckets and trials yields the required bound. Specifically, the cost is dominated by the case where $b_i$ is large, bounded by $O(s^3 \log s)$ even under exhaustive search constraints.

\textbf{Search Time:}
To find an element $x$:
1. Compute $i = ((kx) \bmod p) \bmod s$. This is $O(1)$.
2. Retrieve $k_i$ and the offset for the secondary table from the $i$-th cell of the primary array. This is $O(1)$.
3. Compute position $j = ((k_i x) \bmod p) \bmod m_i$. This is $O(1)$.
4. Access the cell at the computed offset plus $j$. This is $O(1)$.
Total search time is $O(1)$.

Thus, the hash table representation uses $13s+1$ cells, supports $O(1)$ search time, and can be constructed within the required time bounds.
\end{proof}

\subsubsection{Claude}
\begin{proof}
We construct a two-level hashing scheme that achieves the desired space and time bounds.

\textbf{Construction Algorithm:}

Let $S = \{x_1, x_2, \ldots, x_s\}$ be the set to be stored, where $|S| = s$ and the elements are drawn from universe $M$ with $|M| = m \geq s$.

\textbf{Level 1 (Primary Hash Table):} We use a hash table $T$ of size $n = s$ with a hash function $h: M \to \{0, 1, \ldots, s-1\}$ chosen from a universal family of hash functions. For each bucket $i \in \{0, 1, \ldots, s-1\}$, let $B_i = \{x \in S : h(x) = i\}$ denote the set of elements that hash to bucket $i$, and let $b_i = |B_i|$.

\textbf{Level 2 (Secondary Hash Tables):} For each bucket $i$ with $b_i > 0$, we construct a secondary hash table $T_i$ of size $n_i = 12b_i + 1$ with a hash function $h_i: M \to \{0, 1, \ldots, n_i - 1\}$, also chosen from a universal family. We require that $h_i$ is \emph{collision-free} on $B_i$, that is, $h_i$ is injective on $B_i$.

\textbf{Space Analysis:}

The primary hash table uses $s$ cells. The total space used by all secondary hash tables is
\[
\sum_{i=0}^{s-1} n_i = \sum_{i=0}^{s-1} (12b_i + 1).
\]

Since $\sum_{i=0}^{s-1} b_i = s$, we have
\[
\sum_{i=0}^{s-1} (12b_i + 1) = 12s + s = 13s.
\]

Therefore, the total space used is $s + 13s = 14s$ cells. However, we can optimize by not allocating space for empty buckets. Since at most $s$ buckets are non-empty, the additional pointer overhead is at most $s$, giving us a total of at most $13s + 1$ cells (accounting for the primary table and secondary tables with optimized bookkeeping).

\textbf{Search Time:}

To find an element $x \in M$, we compute $i = h(x)$ in $O(1)$ time, then compute $j = h_i(x)$ in $O(1)$ time (if bucket $i$ is non-empty), and check position $j$ in secondary table $T_i$. Since $h_i$ is collision-free on $B_i$, if $x \in S$, it will be found at position $j$ in $T_i$; otherwise, the position is either empty or contains a different element. This takes $O(1)$ time in total.

\textbf{Preprocessing Time Analysis:}

We now analyze the worst-case preprocessing time to construct this data structure.

\textbf{Step 1: Choose Primary Hash Function.}

We select hash functions from a universal family until we find one with acceptable collision properties. For a hash function $h$ chosen uniformly from a universal family, the expected number of collisions (pairs of distinct elements that collide) is
\[
\mathbb{E}\left[\sum_{i=0}^{s-1} \binom{b_i}{2}\right] = \sum_{i=0}^{s-1} \mathbb{E}\left[\binom{b_i}{2}\right] \leq \frac{s(s-1)}{2s} = \frac{s-1}{2} < \frac{s}{2}.
\]

We repeatedly choose hash functions until we find one with at most $s$ collisions, i.e., $\sum_{i=0}^{s-1} \binom{b_i}{2} \leq s$. By Markov's inequality, the probability that a random hash function has more than $s$ collisions is at most $1/2$. Therefore, the expected number of trials is at most $2$, and with high probability (say, probability at least $1 - 1/s^2$), we find a suitable hash function within $O(\log s)$ trials.

For each trial, computing $h(x_j)$ for all $j = 1, \ldots, s$ takes $O(s)$ time, and computing the collision count takes $O(s)$ time. Thus, Step 1 takes $O(s \log s)$ time with high probability.

\textbf{Step 2: Construct Secondary Hash Tables.}

For each bucket $i$ with $b_i > 0$, we need to find a hash function $h_i$ that is collision-free on $B_i$. We use the secondary table size $n_i = 12b_i + 1$.

For a hash function $h_i$ chosen uniformly from a universal family mapping to $\{0, 1, \ldots, n_i - 1\}$, the probability of a collision between any specific pair of elements in $B_i$ is at most $1/n_i = 1/(12b_i + 1)$. By the union bound, the probability that $h_i$ has any collision on $B_i$ is at most
\[
\binom{b_i}{2} \cdot \frac{1}{12b_i + 1} < \frac{b_i^2}{2} \cdot \frac{1}{12b_i} = \frac{b_i}{24} \leq \frac{1}{2}
\]
for $b_i \geq 12$. For smaller $b_i$, the probability is even smaller.

Therefore, the expected number of trials to find a collision-free hash function for bucket $i$ is at most $2$. With probability at least $1 - 1/(b_i \log b_i)$, we succeed within $O(\log b_i)$ trials.

For each trial for bucket $i$, we compute $h_i(x)$ for all $x \in B_i$ and check for collisions, which takes $O(b_i^2)$ time (checking all pairs). Thus, finding $h_i$ takes $O(b_i^2 \log b_i)$ time in the worst case.

Since $\sum_{i=0}^{s-1} b_i = s$ and the constraint $\sum_{i=0}^{s-1} \binom{b_i}{2} \leq s$ implies that at most $O(\sqrt{s})$ buckets have $b_i \geq 2$, the total time for Step 2 is bounded by
\[
\sum_{i: b_i > 0} O(b_i^2 \log b_i).
\]

In the worst case, when the collision constraint $\sum_{i=0}^{s-1} \binom{b_i}{2} \leq s$ is tight, we can bound this sum. If all buckets have roughly equal size $b$, then $sb \approx s$ gives $b \approx 1$, which is trivial. The worst case occurs when we have one large bucket. If one bucket has size $b$ and the rest are small, then $\binom{b}{2} \leq s$ gives $b \leq O(\sqrt{s})$. However, we may have multiple moderate-sized buckets.

By Cauchy-Schwarz inequality, $\sum_{i} b_i^2 \leq s \cdot \max_i b_i$. Since $\sum_i \binom{b_i}{2} \leq s$, we have $\sum_i b_i^2 \leq 2s + s = 3s$ (using $\sum_i b_i = s$). Thus, even accounting for $\log$ factors, the worst-case preprocessing time for all secondary tables is
\[
O\left(\sum_i b_i^2 \log s\right) = O(s \log s).
\]

However, a more careful worst-case analysis considering that we need to handle potential failures shows that with $O(\log s)$ repeated attempts for each table and worst-case bucket sizes, the total preprocessing time is $O(s^3 \log s)$ in the absolute worst case when many rehashing attempts are needed.

\textbf{Conclusion:}

The construction algorithm produces a hash table using $13s + 1$ cells with worst-case preprocessing time $O(s^3 \log s)$ and query time $O(1)$, as required.
\end{proof}

\subsection{Problem 28}

\subsubsection{Problem Statement}
\begin{theorem}[Fredman-Komlós-Szemerédi Hashing Scheme]
Let $S$ be a set of $n$ elements from a universe $U$. There exists a hashing scheme using Section 8.5 that can be modified to use space $O(n)$ while requiring only polynomial preprocessing time and constant query time. The scheme works as follows: increase the size of the primary hash table and observe that most of the bins will be empty. For bins of size greater than 1, create secondary hash tables only for these non-empty bins, packing them together efficiently.
\end{theorem}

\subsubsection{GPT}
\begin{proof}
Let $S$ be a set of $n$ elements from a universe $U$, and let the hashing scheme of Section 8.5 be modified as described in the statement of the theorem. We aim to prove that this modified hashing scheme uses space $O(n)$, while still requiring polynomial preprocessing time and constant query time.

The original hashing scheme works by placing the elements of $S$ into a primary hash table. Each entry of the primary hash table is a "bin" that can hold multiple elements, and the size of the primary hash table is chosen to be large enough to minimize the number of collisions in the bins. We now modify the scheme by increasing the size of the primary hash table, ensuring that most of the bins are empty. Let the number of bins in the primary hash table be $m$, and assume that $m = O(n)$. Since the hash function is assumed to distribute elements uniformly, the expected number of elements in each bin is approximately $n/m$. By choosing $m = O(n)$, the expected number of elements per bin is $O(1)$, and hence most bins will be empty or will contain only a small constant number of elements.

Next, we address the issue of bins that contain more than one element. For these bins, we create secondary hash tables. Specifically, for each bin $B_i$ containing more than one element, we create a secondary hash table to store the elements of $B_i$. Since the number of non-empty bins is $O(n)$, the number of secondary hash tables is also $O(n)$.

The total space used by the modified hashing scheme is then the space used by the primary hash table plus the space used by the secondary hash tables. The space used by the primary hash table is $O(m) = O(n)$, as the number of bins is $O(n)$. For each non-empty bin, the space required for the secondary hash table is proportional to the number of elements in that bin. Since the number of elements in each non-empty bin is $O(1)$ on average, the total space required for all secondary hash tables is $O(n)$. Therefore, the total space used by the modified scheme is $O(n)$.

Next, we consider the preprocessing time. The preprocessing involves two steps: (1) inserting the elements into the primary hash table, and (2) creating secondary hash tables for bins with more than one element. The time required to insert an element into the primary hash table is constant, and since there are $n$ elements, the total time for this step is $O(n)$. For each non-empty bin, we create a secondary hash table, and the time required to create each secondary hash table is proportional to the number of elements in the bin. Since the total number of elements across all bins is $n$, the total time required for creating secondary hash tables is $O(n)$. Therefore, the total preprocessing time is $O(n)$.

Finally, we analyze the query time. To perform a query, we first use the primary hash table to locate the appropriate bin. This step takes constant time, as the primary hash table is a standard hash table with constant query time. If the bin contains more than one element, we then perform a query on the secondary hash table for that bin. Since each secondary hash table contains $O(1)$ elements on average, the query time for each secondary hash table is constant. Thus, the total query time is constant.

In conclusion, the modified hashing scheme uses $O(n)$ space, requires $O(n)$ preprocessing time, and ensures constant query time, as desired.
\end{proof}

\subsubsection{Grok}

\subsubsection{Gemini}
\begin{proof}
Let $U$ be the universe of keys and $S \subseteq U$ be a static set of $n$ keys. We construct a two-level perfect hashing scheme. Let $\mathcal{H}$ be a universal family of hash functions mapping $U$ to $\{0, 1, \dots, m-1\}$, such that for any distinct $x, y \in U$, $\Pr_{h \in \mathcal{H}}[h(x) = h(y)] \leq 1/m$.

First, we define the primary hash function. We choose a function $h \in \mathcal{H}$ mapping $U$ to the range $\{0, 1, \dots, n-1\}$. Let $B_i = \{x \in S \mid h(x) = i\}$ denote the set of elements in the $i$-th bin, and let $b_i = |B_i|$. The total space required for the secondary hash tables will be determined by the sum of the squares of the bin sizes. We analyze the expectation of the quantity $\sum_{i=0}^{n-1} b_i^2$. Observe that $\sum_{i=0}^{n-1} b_i^2$ counts the number of pairs $(x, y) \in S \times S$ such that $h(x) = h(y)$. We can decompose this sum into diagonal and off-diagonal terms:
\[
\sum_{i=0}^{n-1} b_i^2 = \sum_{x \in S} \sum_{y \in S} \mathbb{I}(h(x) = h(y)) = \sum_{x \in S} 1 + \sum_{x, y \in S, x \neq y} \mathbb{I}(h(x) = h(y)),
\]
where $\mathbb{I}$ is the indicator function. The first term equals $n$. By the universality of $\mathcal{H}$, for any $x \neq y$, $E[\mathbb{I}(h(x) = h(y))] \leq 1/n$. Thus, the expected value is:
\[
E\left[\sum_{i=0}^{n-1} b_i^2\right] = n + \sum_{x \neq y} \Pr[h(x) = h(y)] \leq n + n(n-1) \cdot \frac{1}{n} < 2n.
\]
By Markov's inequality, $\Pr[\sum_{i=0}^{n-1} b_i^2 \geq 4n] \leq \frac{2n}{4n} = \frac{1}{2}$. Consequently, with probability at least $1/2$, a randomly chosen $h$ satisfies $\sum_{i=0}^{n-1} b_i^2 < 4n$. We select such an $h$ by repeated sampling, which requires polynomial expected time.

Second, we construct the secondary hash tables. For each bin $i$ with $b_i > 0$, we employ a secondary hash function $h_i$ chosen from a universal family mapping $U$ to $\{0, 1, \dots, m_i-1\}$, where we set the table size $m_i = b_i^2$. We require $h_i$ to be injective on the set $B_i$ to ensure no collisions in the secondary table. The number of collisions in bin $i$ is given by the random variable $C_i = \sum_{\{x, y\} \subseteq B_i, x \neq y} \mathbb{I}(h_i(x) = h_i(y))$. The expected number of collisions is:
\[
E[C_i] = \binom{b_i}{2} \cdot \frac{1}{m_i} = \frac{b_i(b_i-1)}{2} \cdot \frac{1}{b_i^2} < \frac{1}{2}.
\]
Applying Markov's inequality again, $\Pr[C_i \geq 1] \leq E[C_i] < 1/2$. Therefore, the probability that $h_i$ is collision-free on $B_i$ is strictly greater than $1/2$. For each non-empty bin $i$, we find a perfect hash function $h_i$ by repeated trials, which also takes polynomial expected time.

The total space complexity is the sum of the primary table size and the secondary table sizes. The primary table has size $n$. The total size of the secondary tables is $\sum_{i=0}^{n-1} m_i = \sum_{i=0}^{n-1} b_i^2$. By our choice of the primary hash function $h$, we ensured that $\sum_{i=0}^{n-1} b_i^2 < 4n$. Thus, the total space is $n + O(n) = O(n)$.

To query an element $x$, we compute $i = h(x)$. If $B_i$ is empty, the element is not in $S$. Otherwise, we compute $j = h_i(x)$ and check the corresponding cell in the secondary table. Since the computation of hash functions and array access are constant time operations, the worst-case query time is $O(1)$.
\end{proof}

\subsubsection{Claude}
\begin{proof}
We construct a two-level hashing scheme that achieves $O(n)$ space, polynomial preprocessing time, and constant query time.

\textbf{Construction:} Let $S \subseteq U$ with $|S| = n$. We use a two-level approach.

\textbf{Level 1 (Primary Hash Table):} Choose the primary hash table size to be $m_1 = n$. Select a hash function $h_1: U \to [m_1]$ from a universal family $\mathcal{H}_1$. This partitions $S$ into bins $B_i = \{x \in S : h_1(x) = i\}$ for $i \in [m_1]$, where $b_i = |B_i|$ denotes the size of bin $i$.

\textbf{Analysis of Bin Sizes:} For a universal hash family, the expected number of collisions is
$$\mathbb{E}\left[\sum_{i=1}^{m_1} \binom{b_i}{2}\right] = \sum_{\substack{x,y \in S \\ x \neq y}} \Pr[h_1(x) = h_1(y)] \leq \frac{n(n-1)}{2m_1} = \frac{n(n-1)}{2n} < \frac{n}{2}.$$

By the Markov inequality, with probability at least $1/2$, we have $\sum_{i=1}^{m_1} b_i^2 \leq \sum_{i=1}^{m_1} b_i + 2\sum_{i=1}^{m_1}\binom{b_i}{2} < n + n = 2n$.

\textbf{Level 2 (Secondary Hash Tables):} For each non-empty bin $B_i$ with $b_i \geq 2$, we create a secondary hash table of size $m_i = b_i^2$. Select $h_i: U \to [m_i]$ from a universal family $\mathcal{H}_i$. By the birthday paradox analysis, for a universal hash family and table size $m_i = b_i^2$, the probability of any collision in bin $i$ is
$$\Pr[\exists x,y \in B_i, x \neq y: h_i(x) = h_i(y)] \leq \binom{b_i}{2} \cdot \frac{1}{m_i} = \frac{b_i(b_i-1)}{2b_i^2} < \frac{1}{2}.$$

By the union bound over all bins and independence of hash function choices, with constant probability, we can choose secondary hash functions such that all bins are collision-free.

\textbf{Space Analysis:} The total space used is
$$m_1 + \sum_{i: b_i \geq 2} m_i = n + \sum_{i: b_i \geq 2} b_i^2 \leq n + \sum_{i=1}^{m_1} b_i^2 < n + 2n = 3n = O(n).$$

For bins with $b_i = 1$, we store the single element directly in the primary table without a secondary hash table. For $b_i = 0$, no space is allocated.

\textbf{Preprocessing Time:} We repeatedly sample hash functions until both conditions are satisfied: $\sum_{i=1}^{m_1} b_i^2 \leq 2n$ for the primary hash function, and all secondary hash tables are collision-free. Since each condition holds with constant probability (at least $1/2$), the expected number of trials is constant. Each trial requires $O(n)$ time to compute all hash values and check conditions. Therefore, the expected preprocessing time is $O(n)$, which is polynomial.

\textbf{Query Time:} To query whether $x \in S$, we compute $i = h_1(x)$ in $O(1)$ time. If bin $i$ is empty, return false. If $b_i = 1$, check equality with the stored element in $O(1)$ time. If $b_i \geq 2$, compute $j = h_i(x)$ and check position $j$ in the secondary hash table in $O(1)$ time. Since all secondary tables are collision-free by construction, this check is definitive. Thus, query time is $O(1)$.

\textbf{Deterministic Construction:} To ensure polynomial worst-case preprocessing time rather than expected time, we can try all possible hash functions systematically. Since there are polynomially many hash functions in the universal family (or we can use explicit constructions with polynomial representation), and checking each takes $O(n)$ time, the worst-case preprocessing time is polynomial in $n$.

Therefore, the FKS hashing scheme achieves $O(n)$ space, polynomial preprocessing time, and $O(1)$ query time.
\end{proof}

\section{Problems in Chapter 9}\label{app:problems_chapter9}

\subsection{Problem 9.1}\label{sec:problem_09_01}

\subsubsection{Problem Statement}

\begin{theorem}[8.8]
Let $T$ be a random treap for a set $S$ of size $n$.
\begin{enumerate}
\item The expected time for a FIND, INSERT, or DELETE operation on $T$ is $O(\log n)$.
\item The expected number of rotations required during an INSERT or DELETE operation is at most $2$.
\item The expected time for a JOIN, PASTE, or SPLIT operation involving sets $S_1$ and $S_2$ of sizes $n$ and $m$, respectively, is $O(\log n + \log m)$.
\end{enumerate}
\end{theorem}

\subsubsection{GPT} 

\subsubsection{Grok} 

\subsubsection{Gemini} 

\subsubsection{Claude} 

\subsection{Problem 9.2}\label{sec:problem_09_02}

\subsubsection{Problem Statement}

\begin{theorem}[Randomized Incremental Algorithm for Half-Plane Intersection]
By dualizing the randomized incremental algorithm for convex hulls in the plane, there exists a randomized incremental algorithm for computing the intersection of $n$ given half-planes with expected running time $O(n \log n)$.
\end{theorem}

\subsubsection{GPT} 

\begin{proof}
Let the given problem be the computation of the intersection of $n$ half-planes in the plane. Each half-plane can be represented by a linear inequality of the form $a_i x + b_i y \leq c_i$ for some constants $a_i$, $b_i$, and $c_i$.

We aim to show that a randomized incremental algorithm for computing the intersection of these half-planes can be constructed by dualizing the randomized incremental algorithm for convex hulls in the plane, which has an expected running time of $O(n \log n)$.

First, we recall the randomized incremental algorithm for convex hulls. Given a set of points in the plane, the algorithm processes the points in random order and maintains the convex hull incrementally by adding each point in turn. The algorithm guarantees that the expected running time is $O(n \log n)$, where $n$ is the number of points. This result can be derived from the fact that, at each step, the algorithm adds a point to the convex hull and updates the hull with an expected constant number of changes.

Now, to adapt this algorithm for the problem of intersecting $n$ half-planes, we proceed as follows. We use the duality between points and lines in the plane. Each half-plane can be represented in dual space as a point, and the intersection of the half-planes corresponds to the convex hull of these points in dual space.

The duality transformation maps a half-plane given by $a_i x + b_i y \leq c_i$ to the point $(a_i, b_i)$ in the dual plane. The intersection of the half-planes corresponds to the convex hull of the points in the dual plane. Hence, we can apply the randomized incremental algorithm for convex hulls to the set of points in dual space.

The key observation here is that the randomized incremental algorithm for convex hulls can be dualized in the following way: instead of processing points in the primal plane, we process the half-planes in the primal space, updating the intersection of the half-planes incrementally. Since the intersection of the half-planes is equivalent to the convex hull of the corresponding points in dual space, the expected running time of the algorithm remains $O(n \log n)$.

Thus, by dualizing the randomized incremental algorithm for convex hulls, we obtain a randomized incremental algorithm for computing the intersection of $n$ half-planes with an expected running time of $O(n \log n)$. This completes the proof.
\end{proof}

\subsubsection{Grok} 

\begin{proof}
Consider the standard point-line duality transform $\mathcal{D}$ in the Euclidean plane $\mathbb{R}^2$, defined as follows: for a point $p = (a, b) \in \mathbb{R}^2$, its dual is the line $\mathcal{D}(p): y = ax - b$; for a non-vertical line $\ell: y = mx + c$, its dual is the point $\mathcal{D}(\ell) = (m, -c)$. This transform satisfies the incidence-preserving property that a point $p$ lies on or below the line $\ell$ if and only if the dual point $\mathcal{D}(\ell)$ lies on or above the dual line $\mathcal{D}(p)$.

Assume without loss of generality that the given half-planes $H_1, \dots, H_n$ are closed half-planes bounded by non-vertical lines $\ell_1, \dots, \ell_n$ and consist of the points on or below these lines, i.e., $H_i = \{ (x,y) \in \mathbb{R}^2 \mid y \leq m_i x + c_i \}$ for $\ell_i: y = m_i x + c_i$ (vertical half-planes and general position assumptions can be handled via standard perturbations in $O(1)$ time per half-plane without affecting the asymptotic complexity). Let $P = \bigcap_{i=1}^n H_i$ denote the desired intersection, which is a convex (possibly unbounded) polygon.

Under the duality transform, a point $q = (x, y) \in P$ satisfies $y \leq m_i x + c_i$ for all $i$, which is equivalent to $\mathcal{D}(\ell_i)$ lying on or above $\mathcal{D}(q)$ for all $i$. Thus, $P$ consists precisely of those primal points $q$ whose dual lines $\mathcal{D}(q)$ lie on or below all the dual points $p_i = \mathcal{D}(\ell_i) = (m_i, -c_i)$, $i=1,\dots,n$. In other words, the lower boundary of $P$ (its lower envelope) is in one-to-one combinatorial correspondence with the lower convex hull of the dual points $\{p_1, \dots, p_n\}$, while the upper boundary corresponds to the upper convex hull. The vertices of $P$ are in bijection with the edges of the convex hull $\mathrm{conv}\{p_1, \dots, p_n\}$, and the edges of $P$ are in bijection with the vertices of this hull. Consequently, computing $P$ is computationally equivalent to computing $\mathrm{conv}\{p_1, \dots, p_n\}$, as the duality transform and its inverse are each computable in $O(1)$ time.

It is well-known that there exists a randomized incremental construction (RIC) algorithm for computing the convex hull of $n$ points in the plane in expected $O(n \log n)$ time (see, e.g., Clarkson and Shor~\cite{clarkson1989applications} or Seidel~\cite{seidel1991improved}). This algorithm proceeds by randomly permuting the input points $p_1, \dots, p_n$ into an order $\pi(1), \dots, \pi(n)$, and incrementally maintaining the convex hull $C_k = \mathrm{conv}\{p_{\pi(1)}, \dots, p_{\pi(k)}\}$ for $k=1,\dots,n$, represented as a cyclic sequence of vertices and edges. To incorporate the $(k+1)$-th point $p_{\pi(k+1)}$, locate the radial cone from $p_{\pi(k+1)}$ that is tangent to $C_k$ (via binary search on the hull edges in $O(\log k)$ worst-case time, though this is dominated by the structural changes), delete the visible portion of $C_k$ (a consecutive chain of $d_k$ edges), and insert two new edges connecting $p_{\pi(k+1)}$ to the tangent points. The total time is $O(\sum_{k=1}^n (1 + d_k))$, where $d_k$ is the number of deleted edges at step $k$.

By backward analysis, the expected value $\mathbb{E}[d_{k+1}] = O(1)$ for each $k$. Specifically, fix step $k+1$; by symmetry of the random permutation, each of the first $k+1$ points is equally likely to be the last among them. An edge $e$ of $C_k$ (formed by two points among the first $k$) is deleted at step $k+1$ if and only if $p_{\pi(k+1)}$ lies in the conflict region of $e$ (the wedge opposite the hull across $e$) and $e$ remains on the boundary of $C_k$ (i.e., no prior point among the first $k$ conflicted with $e$). However, conditioning on $e$ being on $C_k$, the probability that a specific prior edge survives to be conflicted by the $(k+1)$-th point is analyzed via the conflict graph: each structural element (edge) conflicts with $O(1)$ potential later elements on average, leading to $\mathbb{E}[d_{k+1}] \leq 2 \sum_{e \in C_k} \Pr[e \textrm{ deleted} \mid e \in C_k] = O(1)$, independent of $k$. Thus, the total expected time is $O(n)$.

To obtain the dual algorithm for half-plane intersection, apply the inverse duality: map each half-plane $H_i$ (line $\ell_i$) to the dual point $p_i = \mathcal{D}(\ell_i)$ and run the above RIC algorithm to compute $\mathrm{conv}\{p_1, \dots, p_n\}$ in expected $O(n \log n)$ time. Then, map the output hull back via $\mathcal{D}^{-1}$ to obtain the boundary of $P$: each hull edge dualizes to a vertex of $P$, and each hull vertex dualizes to an edge of $P$. This yields $P$ explicitly as a convex polygon with $O(n)$ edges.

Alternatively, the primal dualized algorithm directly mirrors the hull construction: randomly permute the half-planes into order $\pi(1), \dots, \pi(n)$ and maintain the current intersection $P_k = \bigcap_{i=1}^k H_{\pi(i)}$ as a convex chain (bounded by a large initial box if needed to handle unboundedness, in $O(1)$ preprocessing). To intersect $P_k$ with $H_{\pi(k+1)}$ (bounded by line $\ell_{\pi(k+1)}$), find the two tangent rays from $\ell_{\pi(k+1)}$ to $P_k$ (dual to locating tangents from a point to $C_k$), delete the exposed chain of $d_k'$ edges of $P_k$ lying outside $H_{\pi(k+1)}$, and insert two new edges along $\ell_{\pi(k+1)}$. The number $d_k'$ of deleted edges equals the number of visible vertices from $\ell_{\pi(k+1)}$ in the dual hull construction, so $\mathbb{E}[d_k'] = O(1)$ by the identical backward analysis (as the primal and dual conflict structures are isomorphic under $\mathcal{D}$). Each insertion thus takes expected $O(1)$ time, yielding total expected $O(n \log n)$ time for the half-plane intersection, including $O(1)$-time duality mappings.

This completes the proof of existence of the desired algorithm.
\end{proof}

\subsubsection{Gemini} 

\subsubsection{Claude} 

\begin{proof}
We establish the existence of the desired randomized incremental algorithm by exploiting the duality between convex hulls and half-plane intersections, and analyzing the expected running time through backward analysis.

\textbf{Duality Transformation.} Consider the standard point-line duality in the plane given by $\phi: (a, b) \mapsto y = ax - b$ and $\phi: y = px - q \mapsto (p, q)$. Under this duality, a point $(a, b)$ maps to the line $y = ax - b$, and a line $y = px - q$ maps to the point $(p, q)$. The half-plane $y \leq px - q$ (the region below the line) dualizes to the point $(p, q)$, and the point $(a, b)$ dualizes to the line $y = ax - b$ such that $(a, b)$ lies in the half-plane $y \leq px - q$ if and only if the dual point $(p, q)$ lies above the dual line $y = ax - b$.

Given $n$ half-planes $H_1, \ldots, H_n$ in the primal plane, each defined by $y \leq p_i x - q_i$, their duals form a set $S = \{(p_1, q_1), \ldots, (p_n, q_n)\}$ of $n$ points in the dual plane. The intersection $\bigcap_{i=1}^n H_i$ is non-empty and bounded if and only if the convex hull $\text{CH}(S)$ of $S$ exists and has a well-defined upper envelope. Moreover, the vertices of the intersection polygon in the primal plane correspond to edges of the upper convex hull in the dual plane, and vice versa.

\textbf{Randomized Incremental Algorithm Structure.} Let $\pi$ be a random permutation of $\{1, 2, \ldots, n\}$. We construct the intersection incrementally by processing half-planes $H_{\pi(1)}, H_{\pi(2)}, \ldots, H_{\pi(n)}$ in this random order. For $i = 1, \ldots, n$, let $R_i = \bigcap_{j=1}^i H_{\pi(j)}$ denote the intersection of the first $i$ half-planes in the random order.

At step $i$, we have computed $R_{i-1}$ and must update it to $R_i = R_{i-1} \cap H_{\pi(i)}$. If $H_{\pi(i)} \supseteq R_{i-1}$, then $R_i = R_{i-1}$ and no work is needed. Otherwise, $H_{\pi(i)}$ intersects the boundary of $R_{i-1}$, and we must compute the new boundary of $R_i$.

\textbf{Conflict Graph and Backward Analysis.} In the dual setting, the randomized incremental convex hull algorithm maintains a conflict graph where each vertex $v$ of the current convex hull $\text{CH}(S_i)$ (where $S_i$ corresponds to the first $i$ half-planes) stores the set of points in $S \setminus S_i$ that would be deleted if added to $\text{CH}(S_i)$. When adding the $(i+1)$-st point, the algorithm identifies which edges of the current convex hull are destroyed and updates the structure accordingly.

By the duality, adding a half-plane in the primal corresponds to adding a point in the dual. The expected cost of adding the $i$-th half-plane is proportional to the expected number of vertices of $R_{i-1}$ that are deleted when forming $R_i$. 

By backward analysis, we consider the expected number of changes when removing a random half-plane from $R_i$ to obtain $R_{i-1}$. A vertex $v$ of $R_{i-1}$ is deleted when forming $R_i$ if and only if $v$ is defined by the intersection of $H_{\pi(i)}$ with another half-plane's boundary and $v$ is a vertex of $R_{i-1}$ but not of $R_i$.

For a fixed $i$, each vertex of $R_i$ is defined by exactly two half-planes. The vertex disappears when either of these two half-planes is removed. By symmetry, the probability that a vertex $v$ of $R_i$ is created at step $i$ (i.e., $v \notin R_{i-1}$) is at most $2/i$, since $v$ is created only if $H_{\pi(i)}$ is one of the two half-planes defining $v$, and there are $i$ half-planes involved.

\textbf{Expected Time Analysis.} The expected number of vertices of $R_i$ is $O(1)$ (at most $i$ and typically constant for random half-planes). The expected number of vertices created at step $i$ is bounded by the number of vertices of $R_i$ times the probability each is new, giving $O(i) \cdot O(1/i) = O(1)$.

For locating where the new half-plane intersects $R_{i-1}$, we use binary search on the vertices of $R_{i-1}$ (which are sorted angularly), requiring $O(\log i)$ time. The total expected time for step $i$ is thus $O(\log i)$.

Summing over all $n$ steps, the expected total running time is
\[
\sum_{i=1}^n O(\log i) = O\left(\log \prod_{i=1}^n i\right) = O(\log(n!)) = O(n \log n).
\]

Therefore, by dualizing the randomized incremental convex hull algorithm and applying the same structural analysis, we obtain a randomized incremental algorithm for computing the intersection of $n$ half-planes with expected running time $O(n \log n)$.
\end{proof}

\subsection{Problem 9.3}\label{sec:problem_09_03}

\subsubsection{Problem Statement}

\begin{theorem}[9.8]
The expected size of the autopartition produced by \textsc{RandAuto} with free cuts is $O(n \log n)$.
\end{theorem}

\subsubsection{GPT} 

\subsubsection{Grok} 

\subsubsection{Gemini} 

\subsubsection{Claude} 

\subsection{Problem 9.4}\label{sec:problem_09_04}

\subsubsection{Problem Statement}

\begin{theorem}[Time Bound with High Probability]
Let $S$ be a set and consider a process that updates pointers for points in $S$ over multiple steps. For a point $p \in S$, define the indicator variable $X_j(p)$ as follows:
\[
X_j(p) = \begin{cases}
1 & \text{if } p\text{'s pointer is updated at the } j\text{th step;} \\
0 & \text{otherwise.}
\end{cases}                                     
\]
The total work done in updating $p$'s pointer is $\sum_j X_j(p)$. If $\sum_j X_j(p) = O(\log n)$ with probability $1 - n^{-2}$, then the total work is $O(n \log n)$ with high probability.
\end{theorem}

\subsubsection{GPT} 

\begin{proof}
We are given a process with $n$ steps and a point $p \in S$, where the indicator variable $X_j(p)$ is defined for each step $j$ as follows:

\[
X_j(p) = \begin{cases}
1 & \text{if } p\text{'s pointer is updated at the } j\text{th step;} \\
0 & \text{otherwise.}
\end{cases}
\]

The total work done in updating $p$'s pointer is the sum of the indicator variables, $\sum_j X_j(p)$. We are told that the total work satisfies:

\[
\sum_j X_j(p) = O(\log n) \text{ with probability } 1 - n^{-2}.
\]

Our goal is to show that the total work is $O(n \log n)$ with high probability, i.e., with probability at least $1 - n^{-2}$.

Step 1: Define the total work

The total work done in updating $p$'s pointer is given by:

\[
W(p) = \sum_{j=1}^{n} X_j(p).
\]

By the assumption, $W(p) = O(\log n)$ with probability at least $1 - n^{-2}$. Specifically, there exists a constant $C > 0$ such that:

\[
W(p) \leq C \log n \quad \text{with probability at least } 1 - n^{-2}.
\]

Thus, with probability at least $1 - n^{-2}$, the total work for any point $p$ is bounded above by $C \log n$.

Step 2: Total work over all points

Next, we consider the total work over all $n$ points. Let the total work for all points be:

\[
W_{\text{total}} = \sum_{p \in S} W(p) = \sum_{p \in S} \sum_{j=1}^{n} X_j(p).
\]

Since for each point $p$, $W(p) \leq C \log n$ with probability at least $1 - n^{-2}$, the expected total work is:

\[
\mathbb{E}[W_{\text{total}}] = \mathbb{E}\left[\sum_{p \in S} W(p)\right] = \sum_{p \in S} \mathbb{E}[W(p)].
\]

Since $\mathbb{E}[W(p)] \leq C \log n$ for each point $p$, we have:

\[
\mathbb{E}[W_{\text{total}}] \leq n \cdot C \log n = O(n \log n).
\]

Step 3: Markov’s inequality

To show that the total work is $O(n \log n)$ with high probability, we apply Markov’s inequality. For any $\epsilon > 0$, we know that:

\[
\mathbb{P}(W_{\text{total}} \geq (1+\epsilon) \mathbb{E}[W_{\text{total}}]) \leq \frac{\mathbb{E}[W_{\text{total}}]}{(1+\epsilon) \mathbb{E}[W_{\text{total}}]} = \frac{1}{1+\epsilon}.
\]

Thus, the probability that the total work exceeds a factor of $(1+\epsilon)$ times its expected value is at most $\frac{1}{1+\epsilon}$.

Step 4: High probability bound

We are interested in the event that the total work is at most $O(n \log n)$ with high probability. Specifically, we want the probability that:

\[
W_{\text{total}} = O(n \log n)
\]

to be at least $1 - n^{-2}$.

Since $\mathbb{E}[W_{\text{total}}] = O(n \log n)$, we can choose $\epsilon$ so that with high probability, the total work does not exceed $O(n \log n)$. Specifically, by Chebyshev's inequality or the union bound over all points, we can ensure that the total work $W_{\text{total}}$ is within $O(n \log n)$ with probability at least $1 - n^{-2}$.

Conclusion

Thus, we have shown that the total work is $O(n \log n)$ with high probability, i.e., with probability at least $1 - n^{-2}$.
\end{proof}

\subsubsection{Grok} 

\begin{proof}
Let $n = |S|$. By the assumption, there exists a constant $K > 0$ such that for every $p \in S$,
\[
\mathbb{P}\left( Y_p > K \log n \right) \leq n^{-2},
\]
where $Y_p = \sum_j X_j(p)$. Define the total work $W = \sum_{p \in S} Y_p$. Note that $Y_p \geq 0$ for all $p \in S$. Suppose $W > K n \log n$. Then there must exist at least one $p \in S$ such that $Y_p > K \log n$, since otherwise $W \leq \sum_{p \in S} K \log n = K n \log n$. Thus,
\[
\left\{ W > K n \log n \right\} \subseteq \bigcup_{p \in S} \left\{ Y_p > K \log n \right\}.
\]
By the union bound,
\[
\mathbb{P}\left( W > K n \log n \right) \leq \sum_{p \in S} \mathbb{P}\left( Y_p > K \log n \right) \leq n \cdot n^{-2} = n^{-1}.
\]
Therefore, with probability at least $1 - n^{-1}$, we have $W \leq K n \log n = O(n \log n)$.
\end{proof}

\subsubsection{Gemini} 

\begin{proof}
Let $n = |S|$ denote the number of points in the set $S$. Let $W_p = \sum_j X_j(p)$ denote the total work done in updating the pointer for a specific point $p \in S$. The theorem hypothesis states that for any individual point $p \in S$, the work $W_p$ is $O(\log n)$ with probability $1 - n^{-2}$. Formally, this implies there exists a constant $c > 0$ such that for any $p \in S$, the probability that $W_p$ exceeds $c \log n$ is bounded by:
\[
\mathbb{P}[W_p > c \log n] \le n^{-2}.
\]
We define the total work done for all points as $W_{\text{total}} = \sum_{p \in S} W_p$. We aim to show that $W_{\text{total}}$ is $O(n \log n)$ with high probability. Let $E_p$ be the event that $W_p > c \log n$. We are interested in the probability that at least one point $p$ violates the time bound $c \log n$. Let $E$ be the event $\bigcup_{p \in S} E_p$. By Boole's inequality (the union bound), the probability of the event $E$ is bounded by the sum of the probabilities of the individual events $E_p$:
\[
\mathbb{P}[E] = \mathbb{P}\left[\bigcup_{p \in S} E_p\right] \le \sum_{p \in S} \mathbb{P}[E_p].
\]
Substituting the given probability bound for each point, we obtain:
\[
\mathbb{P}[E] \le \sum_{p \in S} n^{-2} = n \cdot n^{-2} = n^{-1}.
\]
Consequently, the probability of the complement event $E^c$, where $W_p \le c \log n$ holds for all $p \in S$ simultaneously, is:
\[
\mathbb{P}[E^c] = 1 - \mathbb{P}[E] \ge 1 - n^{-1}.
\]
Conditioned on the event $E^c$, the total work $W_{\text{total}}$ satisfies:
\[
W_{\text{total}} = \sum_{p \in S} W_p \le \sum_{p \in S} c \log n = n(c \log n) = c n \log n.
\]
Thus, with probability at least $1 - n^{-1}$, the total work is bounded by $c n \log n$. Since $1 - n^{-1}$ approaches 1 as $n \to \infty$, we conclude that the total work is $O(n \log n)$ with high probability.
\end{proof}

\subsubsection{Claude} 

\begin{proof}
Let $n = |S|$ denote the cardinality of the set $S$. We will establish that the total work across all points in $S$ is $O(n \log n)$ with high probability.

For each point $p \in S$, define the random variable $W(p) = \sum_j X_j(p)$, which represents the total work done in updating $p$'s pointer. By hypothesis, for each fixed $p \in S$, we have
\[
\mathbb{P}(W(p) = O(\log n)) \geq 1 - n^{-2}.
\]
More precisely, there exists a constant $C > 0$ such that
\[
\mathbb{P}(W(p) \leq C \log n) \geq 1 - n^{-2}
\]
for all $p \in S$ and sufficiently large $n$.

Define the total work as $W_{\text{total}} = \sum_{p \in S} W(p)$. We seek to bound $W_{\text{total}}$ with high probability.

For each $p \in S$, define the event $E_p = \{W(p) > C \log n\}$. By hypothesis, $\mathbb{P}(E_p) \leq n^{-2}$.

Consider the event $E = \bigcup_{p \in S} E_p$, which represents the event that at least one point requires more than $C \log n$ updates. By the union bound, we have
\[
\mathbb{P}(E) = \mathbb{P}\left(\bigcup_{p \in S} E_p\right) \leq \sum_{p \in S} \mathbb{P}(E_p) \leq n \cdot n^{-2} = n^{-1}.
\]

On the complementary event $E^c$, we have $W(p) \leq C \log n$ for all $p \in S$. Therefore, on $E^c$,
\[
W_{\text{total}} = \sum_{p \in S} W(p) \leq \sum_{p \in S} C \log n = n \cdot C \log n = O(n \log n).
\]

Thus, we have shown that
\[
\mathbb{P}(W_{\text{total}} = O(n \log n)) \geq \mathbb{P}(E^c) = 1 - \mathbb{P}(E) \geq 1 - n^{-1}.
\]

Since $1 - n^{-1} = 1 - o(1)$ as $n \to \infty$, this establishes that the total work is $O(n \log n)$ with high probability, where high probability is conventionally defined as probability at least $1 - n^{-c}$ for some constant $c > 0$. In our case, we have achieved probability at least $1 - n^{-1}$, which satisfies this criterion with $c = 1$.
\end{proof}

\subsection{Problem 9.5}\label{sec:problem_09_05}

\subsubsection{Problem Statement}

\begin{theorem}
The randomized incremental half-space intersection algorithm of Section 9.4 can be adapted to construct $I_\rho(S)$, the intersection of $n$ spheres in three dimensions, in expected time $O(n \log n)$.
\end{theorem}

\subsubsection{GPT} 

\begin{proof}
We aim to show that the randomized incremental half-space intersection algorithm, as described in Section 9.4, can be adapted to construct $I_\rho(S)$, the intersection of $n$ spheres in three dimensions, in expected time $O(n \log n)$.

Let $S = \{S_1, S_2, \dots, S_n\}$ be a set of $n$ spheres in $\mathbb{R}^3$, where each $S_i$ is defined by its center $\mathbf{c}_i \in \mathbb{R}^3$ and radius $r_i > 0$. The intersection of these spheres, $I_\rho(S)$, is the region of space that is common to all $n$ spheres. We will use a randomized incremental approach to construct this intersection.

First, we recall that the half-space intersection algorithm operates by incrementally adding constraints in a random order and maintaining the intersection of all half-spaces encountered so far. In our case, each sphere $S_i$ defines a constraint in the form of a half-space. This constraint is given by the inequality:
\[
\langle \mathbf{x} - \mathbf{c}_i, \mathbf{n}_i \rangle \leq r_i,
\]
where $\mathbf{n}_i$ is the outward normal vector to the surface of the sphere at a point $\mathbf{x}$, and the inequality ensures that the point $\mathbf{x}$ lies inside the sphere.

To adapt this method for spheres, we proceed as follows:

1. Initialization: Initially, we consider the intersection of the first sphere $S_1$ as the starting region. This is simply the interior of $S_1$, which is trivially a half-space in $\mathbb{R}^3$.

2. Incremental Step: For each subsequent sphere $S_i$, we update the intersection by adding the constraint corresponding to the sphere. Specifically, we maintain the intersection of all the previously processed spheres, which is a polytope, and add the constraint defined by $S_i$. This is done by clipping the current intersection polytope with the new half-space defined by $S_i$. The clipping operation is done in constant time for each facet of the polytope, resulting in an updated intersection.

3. Expected Time Analysis: The key step in the analysis is to determine the expected number of facets created by the clipping process. It is known that the expected number of facets of a polytope in $\mathbb{R}^3$ after adding $n$ random half-spaces grows asymptotically as $O(n \log n)$. This result follows from the properties of random convex hulls and the incremental construction of half-space intersections.

Thus, the expected time required to process each sphere $S_i$ is $O(\log i)$, since the number of new facets created by the intersection with $S_i$ is proportional to $\log i$. Summing over all $n$ spheres, the total expected time is:
\[
\sum_{i=1}^n O(\log i) = O(n \log n).
\]

Therefore, the expected time to construct the intersection of $n$ spheres using the randomized incremental half-space intersection algorithm is $O(n \log n)$.

\end{proof}

\subsubsection{Grok} 

\subsubsection{Gemini} 

\subsubsection{Claude} 

\begin{proof}
We establish that the randomized incremental algorithm for half-space intersection can be adapted to construct the intersection of $n$ spheres in three dimensions with expected time complexity $O(n \log n)$.

First, we observe the fundamental geometric correspondence. Each sphere $S_i$ in $\mathbb{R}^3$ with center $(a_i, b_i, c_i)$ and radius $r_i$ can be represented by the constraint $(x - a_i)^2 + (y - b_i)^2 + (z - c_i)^2 \geq r_i^2$, which defines the exterior of the sphere. The intersection $I_\rho(S)$ of $n$ spheres corresponds to the region satisfying all such constraints simultaneously.

By expanding the sphere equation, we obtain $x^2 + y^2 + z^2 - 2a_i x - 2b_i y - 2c_i z + (a_i^2 + b_i^2 + c_i^2 - r_i^2) \geq 0$. Through the standard lifting transformation to four dimensions, we map each point $(x, y, z) \in \mathbb{R}^3$ to $(x, y, z, x^2 + y^2 + z^2) \in \mathbb{R}^4$. Under this transformation, the sphere constraint becomes the linear inequality $-2a_i x - 2b_i y - 2c_i z + w + (a_i^2 + b_i^2 + c_i^2 - r_i^2) \geq 0$, where $w = x^2 + y^2 + z^2$. This defines a half-space in $\mathbb{R}^4$.

The intersection of $n$ spheres in $\mathbb{R}^3$ thus corresponds to the intersection of $n$ half-spaces in $\mathbb{R}^4$, restricted to the paraboloid $w = x^2 + y^2 + z^2$. The randomized incremental algorithm for half-space intersection in dimension $d$ proceeds by adding half-spaces one at a time in random order, maintaining the current intersection at each step.

Let $H_1, H_2, \ldots, H_n$ denote the half-spaces corresponding to the $n$ spheres under the lifting transformation. Let $\pi$ be a random permutation of $\{1, 2, \ldots, n\}$, and define $I_i = H_{\pi(1)} \cap H_{\pi(2)} \cap \cdots \cap H_{\pi(i)}$ for $i = 1, 2, \ldots, n$.

The algorithm maintains a representation of $I_i$ and updates it to $I_{i+1}$ by intersecting with $H_{\pi(i+1)}$. The half-space $H_{\pi(i+1)}$ is called conflicting if it changes the intersection, that is, if $I_{i+1} \neq I_i$. By the analysis of randomized incremental construction, the expected number of conflicting half-spaces is $O(n)$ over the entire algorithm.

When $H_{\pi(i+1)}$ is conflicting, we must update the data structure representing $I_i$. In dimension $d = 4$, the complexity of the intersection of $i$ half-spaces is $O(i^{\lfloor d/2 \rfloor}) = O(i^2)$. The expected time to process the $i$-th insertion is proportional to the expected complexity of the portion of $\partial I_i$ that is destroyed by $H_{\pi(i+1)}$.

By the backwards analysis technique, we condition on the final configuration $I_n$ and analyze the probability that $H_{\pi(i+1)}$ conflicts given that $H_{\pi(1)}, \ldots, H_{\pi(i)}$ have been added. A vertex of $I_i$ lies on $\partial I_{i+1}$ if and only if it is not interior to $H_{\pi(i+1)}$. By the random sampling lemma, each vertex of $I_i$ has probability at most $O(1/i)$ of being destroyed when $H_{\pi(i+1)}$ is added.

The number of vertices of $I_i$ in dimension $d = 4$ is $O(i^2)$. Therefore, the expected number of vertices destroyed at step $i$ is $O(i^2) \cdot O(1/i) = O(i)$. Summing over all $n$ steps, the total expected work is $\sum_{i=1}^{n} O(i) = O(n^2)$ for the naive analysis in four dimensions.

However, we exploit the special structure: the intersection is restricted to the paraboloid $w = x^2 + y^2 + z^2$. This constraint reduces the effective dimension. The intersection $I_\rho(S)$ in $\mathbb{R}^3$ has complexity $O(n)$ for $n$ spheres, as it is a three-dimensional object with boundary complexity linear in $n$.

Using the conflict graph and history-based data structures, we maintain only the portion of the four-dimensional intersection that projects to the three-dimensional sphere intersection. The key observation is that the complexity of $I_i$ restricted to the paraboloid is $O(i)$ rather than $O(i^2)$.

With this refined complexity bound, the expected time at step $i$ becomes $O(i) \cdot O(1/i) \cdot O(\log i) = O(\log i)$, where the $O(\log i)$ factor accounts for the search and update operations in the conflict graph data structure. Summing over all steps yields total expected time $\sum_{i=1}^{n} O(\log i) = O(n \log n)$.

Therefore, the randomized incremental algorithm constructs $I_\rho(S)$ in expected time $O(n \log n)$.
\end{proof}

\subsection{Problem 9.6}\label{sec:problem_09_06}

\subsubsection{Problem Statement}

\begin{theorem}
Let $S$ be a finite set in a metric space with the $L_1$ metric. The set $S_{\rho}$ resulting from Steps 2 and 3 in the randomized diameter algorithm can be found in time linear in the size of $S$.

For a positive real number $\rho$, let $I_{\rho}(S)$ denote the convex body formed by the intersection of the $n$ closed spheres centered at the $n$ points of $S$, each of radius $\rho$. For a point $p \in S$, let $F(p)$ denote the distance between $p$ and the point in $S$ that is farthest from $p$. Consider the spherical intersection $I_{\rho}(S)$ when $\rho = F(p)$ for some $p \in S$. For any point $q \in S$, if $q$ is in $I_{\rho}(S)$, we have $F(q) \leq F(p) \leq D(S)$. On the other hand, if $q$ lies outside $I_{\rho}(S)$, we have $F(p) < F(q) \leq D(S)$.

The algorithm proceeds as follows:
\begin{enumerate}
\item[(1)] Pick a point $p \in S$ at random. In time $O(n)$ we compute $F(p)$ and we set $\rho = F(p)$.
\item[(2)] Compute $I_{\rho}(S)$.
\item[(3)] Find the points of $S$ outside $I_{\rho}(S)$; denote this subset by $S_{\rho}$.
\item[(4)] If $S_{\rho}$ is empty, we know that the diameter is $\rho$ and can stop. If not, we recur on $S_{\rho}$.
\end{enumerate}
\end{theorem}

\subsubsection{GPT} 

\begin{proof}
Let $S = \{s_1, s_2, \dots, s_n\}$ be a finite set, where $n = |S|$ is the cardinality of the set. We are tasked with showing that the set $S_O$ resulting from Steps 2 and 3 of the randomized diameter algorithm can be found in time linear in the size of $S$ when using the $L_1$ metric.

Step 1: Understanding the Problem
The randomized diameter algorithm (Section 9.8) operates by performing random sampling to approximate the diameter of the set. Specifically, Steps 2 and 3 are crucial in producing the set $S_O$ which consists of a subset of $S$ based on the sampling procedure. The goal is to prove that the computation of $S_O$ requires time linear in the size of $S$.

Step 2: Description of Steps 2 and 3 in the Algorithm
Step 2 of the algorithm involves choosing a subset of points from $S$ by some probabilistic process, typically based on a random selection of points. Step 3 further processes these points, likely involving comparisons or distance calculations, to refine the set. 

The crucial observation is that in both steps, the number of operations required depends linearly on the size of $S$, since the operations performed are either constant time or linear in the number of points.

Step 3: Time Complexity of Step 2
In Step 2, the algorithm performs a random selection of points from $S$. The size of the subset is typically a fixed proportion of the total size of $S$, say $k = \mathcal{O}(n)$. Each selection is made independently, and each operation (e.g., checking the distance between points) can be done in constant time. Therefore, the total time for Step 2 is proportional to the size of $S$, specifically $O(n)$.

Step 4: Time Complexity of Step 3
Step 3 processes the selected subset of points, which involves comparing distances or performing other operations within the set of selected points. Since the number of points selected is linear in the size of $S$, the number of operations required for Step 3 is also linear. More formally, if the number of selected points is $k$, then the time complexity of this step is $O(k^2)$ in the worst case. However, since $k$ is proportional to $n$, we conclude that the time complexity of Step 3 is $O(n)$.

Step 5: Overall Time Complexity
The total time for computing $S_O$ is the sum of the time complexities of Steps 2 and 3. Since both steps take linear time, the total time for finding $S_O$ is $O(n)$.

Conclusion
Therefore, the set $S_O$ can be computed in time linear in the size of $S$ for the $L_1$ metric, and the theorem holds.
\end{proof}

\subsubsection{Grok} 

\subsubsection{Gemini} 

\subsubsection{Claude} 

\begin{proof}
We prove that the set $S_O$ resulting from Steps 2 and 3 of the randomized diameter algorithm for the $L_1$ metric can be constructed in time $O(|S|)$.

Let $S = \{p_1, p_2, \ldots, p_n\}$ where $n = |S|$ and each point $p_i \in \mathbb{R}^d$ for some fixed dimension $d$.

\textbf{Step 2 Analysis:} In Step 2, we select a random point $p \in S$. This operation requires $O(1)$ time since we simply choose a uniformly random element from the set $S$.

\textbf{Step 3 Analysis:} In Step 3, we compute the point $q \in S$ that maximizes the $L_1$ distance from $p$, that is, $q = \arg\max_{r \in S} \|p - r\|_1$.

To find such a point $q$, we must compute $\|p - r\|_1$ for each $r \in S$. For a fixed point $r = (r_1, r_2, \ldots, r_d)$ and $p = (p_1, p_2, \ldots, p_d)$, the $L_1$ distance is given by
\[
\|p - r\|_1 = \sum_{i=1}^{d} |p_i - r_i|.
\]

Since the dimension $d$ is treated as a constant in the algorithm's analysis, computing $\|p - r\|_1$ for a single point $r$ requires $O(d) = O(1)$ time.

We must compute this distance for all $n$ points in $S$ and maintain the maximum value encountered along with the corresponding point $q$. This requires iterating through all points in $S$ exactly once, performing $O(1)$ work per point.

Therefore, Step 3 requires $O(n \cdot 1) = O(n)$ time.

\textbf{Construction of $S_O$:} The set $S_O$ is defined as $S_O = \{p, q\}$ where $p$ is the random point selected in Step 2 and $q$ is the farthest point from $p$ found in Step 3.

The total time complexity is the sum of the times for Steps 2 and 3:
\[
T(n) = O(1) + O(n) = O(n).
\]

Since $n = |S|$, we conclude that $S_O$ can be found in time $O(|S|)$.
\end{proof}

\subsection{Problem 9.7}\label{sec:problem_09_07}

\subsubsection{Problem Statement}

\begin{theorem}
Let $S$ be a set of $n$ points in the plane. For any positive integer $k < n$, there exists a subset $S_k$ consisting of $k$ points in $S$ with the property that no triangle in $\text{del}(S_k)$ contains more than $(cn \log k)/k$ points, for a suitably chosen constant $c$.
\end{theorem}

\subsubsection{GPT} 

\begin{proof}
Let $S = \{ p_1, p_2, \dots, p_n \}$ be a set of $n$ points in the plane. We are tasked with finding a subset $S_k \subset S$ consisting of $k$ points such that for any triangle in the Delaunay triangulation $\mathrm{del}(S_k)$, the number of points contained within the triangle is bounded above by a constant multiple of $\frac{n \log k}{k}$, for some constant $c$.

We will prove the existence of such a subset $S_k$ using a probabilistic method. The idea is to randomly select a subset of $k$ points from $S$ and show that, with high probability, the desired property holds.

1. Random Selection of Points: 
   We select a subset $S_k$ of $k$ points from the set $S$ uniformly at random. Let $S_k$ be this random subset.

2. Properties of Delaunay Triangulation:
   The Delaunay triangulation $\mathrm{del}(S_k)$ of a set of points $S_k$ is a triangulation such that no point in $S_k$ lies inside the circumcircle of any triangle in $\mathrm{del}(S_k)$. 

3. Expected Number of Points in a Triangle:
   Consider any triangle $\Delta$ in the Delaunay triangulation $\mathrm{del}(S_k)$. The expected number of points in $S_k$ that lie inside the triangle $\Delta$ can be bounded by the area of the triangle relative to the total area of the convex hull of $S_k$. Since the points are chosen randomly, the number of points inside $\Delta$ follows a binomial distribution, and the expected number of points inside $\Delta$ is proportional to the area of $\Delta$.

   Let the area of $\Delta$ be denoted by $A(\Delta)$. Then, the expected number of points inside $\Delta$ is approximately $k \cdot \frac{A(\Delta)}{A(\mathrm{conv}(S_k))}$, where $A(\mathrm{conv}(S_k))$ is the area of the convex hull of $S_k$. Since the convex hull of $S_k$ contains at most $O(k)$ points and has total area $O(k)$, we can conclude that the expected number of points inside any triangle $\Delta$ is at most $O(\frac{n \log k}{k})$.

4. Concentration Around the Expectation:
   Using concentration inequalities such as Chernoff’s bound or Markov’s inequality, we can show that with high probability, the actual number of points inside any triangle in $\mathrm{del}(S_k)$ deviates from its expected value by at most a constant factor. Thus, with high probability, no triangle in $\mathrm{del}(S_k)$ contains more than $O(\frac{n \log k}{k})$ points.

5. Conclusion:
   Therefore, with high probability, the random subset $S_k$ satisfies the desired property, i.e., no triangle in the Delaunay triangulation $\mathrm{del}(S_k)$ contains more than $O(\frac{n \log k}{k})$ points. Since the result holds with high probability, we can conclude that there exists a subset $S_k \subset S$ of size $k$ such that the property holds. This completes the proof.

\end{proof}

\subsubsection{Grok} 

\subsubsection{Gemini} 

\begin{proof}
Let $S$ be the set of $n$ points in the plane. We define a range space $(S, \mathcal{R})$, where the ranges $\mathcal{R}$ are defined by the intersection of $S$ with all open disks in the plane. Specifically, let $\mathcal{D}$ be the set of all open disks in $\mathbb{R}^2$. Then $\mathcal{R} = \{D \cap S \mid D \in \mathcal{D}\}$.

It is a standard result in combinatorial geometry that the Vapnik-Chervonenkis (VC) dimension of the class of disks in the plane is finite; specifically, the VC-dimension is 3. The theory of $\epsilon$-nets guarantees that for a range space with finite VC-dimension $d$, a random sample of size $k$ acts as an $\epsilon$-net with high probability, provided $k$ is sufficiently large relative to $1/\epsilon$.

Explicitly, a subset $N \subseteq S$ is an $\epsilon$-net for $(S, \mathcal{R})$ if for every range $R \in \mathcal{R}$ with $|R| > \epsilon n$, we have $R \cap N \neq \emptyset$. The $\epsilon$-net theorem states that there exists a constant $C$ (depending on the VC-dimension) such that for any $r > 1$, a random sample $S_k \subset S$ of size $k$ is an $\epsilon$-net with probability approaching 1 for
\[
\epsilon = \frac{C \log k}{k}.
\]
Thus, there exists a subset $S_k$ of size $k$ which is an $\epsilon$-net for the range space defined by open disks, with $\epsilon = \frac{c \log k}{k}$ for a suitably chosen constant $c$.

Consider the Delaunay triangulation of this subset, denoted $\mathrm{del}(S_k)$. Let $\Delta$ be any triangle in $\mathrm{del}(S_k)$. Let $D_\Delta$ denote the open circumdisk of $\Delta$. By the defining property of the Delaunay triangulation, the interior of the circumdisk of any triangle in the triangulation contains no points from the set of vertices $S_k$. Therefore,
\[
D_\Delta \cap S_k = \emptyset.
\]
Since $S_k$ is an $\epsilon$-net for disks, any disk $D$ containing more than $\epsilon n$ points of $S$ must contain at least one point of $S_k$. Taking the contrapositive, if a disk contains no points of $S_k$, it must contain at most $\epsilon n$ points of $S$.

Applying this to $D_\Delta$, we have:
\[
|D_\Delta \cap S| \le \epsilon n = \frac{c n \log k}{k}.
\]
The points of $S$ contained strictly within the triangle $\Delta$ are a subset of the points contained within its circumdisk $D_\Delta$. Consequently, the number of points of $S$ inside $\Delta$ is bounded by $|D_\Delta \cap S|$. Thus, no triangle in $\mathrm{del}(S_k)$ contains more than $(c n \log k)/k$ points of $S$.
\end{proof}

\subsubsection{Claude} 

\begin{proof}
We construct the desired subset $S_k$ using a random sampling argument combined with a pruning procedure.

Let $c$ be a sufficiently large constant to be determined. We will prove the existence of $S_k$ by a probabilistic construction.

Consider a random subset $R \subseteq S$ obtained by including each point of $S$ independently with probability $p = k/n$. By the Chernoff bound, with high probability, $|R|$ is close to its expectation $pk = k$. More precisely, for any $\delta > 0$, the probability that $||R| - k| > \delta k$ is exponentially small in $k$.

Now we analyze the triangles in $\text{del}(R)$. For any fixed triangle $\Delta$ in $\text{del}(R)$, let $X_\Delta$ denote the number of points of $S$ contained in $\Delta$. We claim that if $X_\Delta$ is large, then $\Delta$ has small probability of appearing in $\text{del}(R)$.

A triangle $\Delta$ appears in $\text{del}(R)$ only if its three vertices are sampled in $R$ and no point from $S$ inside the circumcircle of $\Delta$ is sampled. Let $C_\Delta$ denote the circumcircle of $\Delta$ and let $m_\Delta$ denote the number of points of $S$ inside $C_\Delta$. The probability that $\Delta$ appears in $\text{del}(R)$ is at most $p^3(1-p)^{m_\Delta}$.

For a triangle $\Delta$ with $X_\Delta \geq m$ points of $S$ inside it, we have $m_\Delta \geq X_\Delta \geq m$. Thus the probability that such a triangle appears in $\text{del}(R)$ is at most $p^3(1-p)^m$.

The total number of ordered triples of points in $S$ is at most $n^3$. Therefore, the expected number of triangles in $\text{del}(R)$ containing at least $m$ points of $S$ is at most
\[
n^3 \cdot p^3(1-p)^m = n^3 \cdot \left(\frac{k}{n}\right)^3 \left(1 - \frac{k}{n}\right)^m = \frac{k^3}{n^3} \cdot n^3 \left(1 - \frac{k}{n}\right)^m = k^3 \left(1 - \frac{k}{n}\right)^m.
\]

Setting $m = \frac{cn \log k}{k}$ where $c$ is sufficiently large, we have
\[
\left(1 - \frac{k}{n}\right)^m = \left(1 - \frac{k}{n}\right)^{cn \log k / k} = \left[\left(1 - \frac{k}{n}\right)^{n/k}\right]^{c \log k}.
\]

Since $\left(1 - \frac{k}{n}\right)^{n/k} \to e^{-1}$ as $n \to \infty$, and for finite values we have $\left(1 - \frac{k}{n}\right)^{n/k} \leq e^{-1}$, we obtain
\[
\left(1 - \frac{k}{n}\right)^{cn \log k / k} \leq e^{-c \log k} = k^{-c}.
\]

Therefore, the expected number of bad triangles (containing at least $\frac{cn \log k}{k}$ points) is at most $k^3 \cdot k^{-c} = k^{3-c}$.

Choosing $c \geq 4$, this expected number is at most $k^{-1} < 1$ for $k \geq 2$.

By the probabilistic method, there exists a sample $R$ with $|R| = k \pm \delta k$ for small $\delta$ such that the number of triangles in $\text{del}(R)$ containing more than $\frac{cn \log k}{k}$ points is zero.

If $|R| > k$, we can remove points from $R$ arbitrarily to obtain a subset $S_k$ of exactly $k$ points. Removing points from the Delaunay triangulation can only decrease the number of points contained in each triangle, so the property is preserved.

Therefore, there exists a subset $S_k \subseteq S$ with $|S_k| = k$ such that no triangle in $\text{del}(S_k)$ contains more than $\frac{cn \log k}{k}$ points of $S$.
\end{proof}

\subsection{Problem 9.8}\label{sec:problem_09_08}

\subsubsection{Problem Statement}

\begin{theorem}
Let $A \in \mathbb{R}^{m \times n}$ and $b \in \mathbb{R}^m$ define a linear program. Consider a set of $d + 1$ constraints whose defining hyperplanes intersect at a common point $p \in \mathbb{R}^n$. Without loss of generality, let these constraints be defined by the first $d + 1$ rows of $A$ (together with the first $d + 1$ components of $b$). For any $i \in \{1, 2, \ldots, d + 1\}$, consider adding $\epsilon' \in \mathbb{R}$ to the $i$-th component of $b$, where $\epsilon$ is a small positive real number. Then for every choice of $A$ and $b$, there exists a choice of $\epsilon$ such that:
\begin{enumerate}
\item[(i)] The hyperplanes corresponding to the first $d + 1$ constraints no longer intersect at a single point.
\item[(ii)] If $p$ was the optimum of the linear program determined by $A$ and $b$, then the new optimum is defined by $d$ of the constraints that originally intersected at $p$.
\end{enumerate}
\end{theorem}

\subsubsection{GPT} 

\begin{proof}
Write the primal linear program in standard inequality form as follows: fix $d\in\mathbb{N}$ and consider $\max\{c^{\top}x: a_i^{\top}x\le b_i,\ i=1,\dots,m\}$ with $a_i^{\top}$ the $i$-th row of $A$ and $b_i$ the $i$-th component of $b$. Assume that the first $d+1$ constraints satisfy $a_i^{\top}p=b_i$ for $i=1,\dots,d+1$ and that $\{a_1,\dots,a_{d+1}\}$ has rank $d$, so that the $d+1$ hyperplanes meet at the single point $p$. Consider, for a fixed index $k\in\{1,\dots,d+1\}$, the perturbed right-hand side $\tilde b^{(k)}(\epsilon)=b+\epsilon e_k$ with $\epsilon>0$ sufficiently small, and denote by $P(\epsilon)$ the feasible set corresponding to $\tilde b^{(k)}(\epsilon)$.

To prove the first assertion, note that for every $i\neq k$ we still have $a_i^{\top}p=b_i$, while $a_k^{\top}p=b_k<b_k+\epsilon$. Hence $p$ does not lie on the hyperplane $a_k^{\top}x=b_k+\epsilon$. Since the remaining $d$ hyperplanes $a_i^{\top}x=b_i$ for $i\neq k$ still meet at $p$ and $a_k^{\top}x=b_k+\epsilon$ is parallel to (and distinct from) $a_k^{\top}x=b_k$, the $d+1$ hyperplanes do not intersect at a single point after the perturbation. This proves that the original $d+1$ hyperplanes no longer intersect at one point when $\epsilon>0$.

Assume now that $p$ is an optimal solution of the unperturbed primal problem. By strong duality there exists an optimal dual vector $y\in\mathbb{R}^m_{\ge 0}$ with $A^{\top}y=c$ and complementary slackness $y_i(b_i-a_i^{\top}p)=0$ for all $i$. In particular, $y_i=0$ for all $i>d+1$, and $y_i\ge 0$ for $i\in I:=\{1,\dots,d+1\}$. Among all optimal dual solutions choose one, still denoted $y$, with support contained in $I$ and of minimal cardinality. Since $A^{\top}y=c$ and $\mathrm{span}\{a_i:i\in I\}$ has dimension $d$, Carathéodory’s theorem for conic combinations implies that there is a subset $J\subseteq I$ with $|J|=d$ such that $c=\sum_{j\in J}y_j a_j$ and $y_j>0$ for every $j\in J$. Fix any $k\in J$ and consider the perturbed problem with $\tilde b^{(k)}(\epsilon)=b+\epsilon e_k$.

Define $x_J(\epsilon)$ to be the unique solution of the linear system
\[
a_j^{\top}x=b_j \text{ for } j\in J\setminus\{k\},\qquad a_k^{\top}x=b_k+\epsilon .
\]
The uniqueness follows because $\{a_j:j\in J\}$ is linearly independent. Observe first that $x_J(0)=p$. For every $i\notin I$ we have $a_i^{\top}p\le b_i-\sigma_i$ for some $\sigma_i>0$ by feasibility and strict complementarity need not hold; nonetheless, continuity of $x_J(\epsilon)$ in $\epsilon$ implies $a_i^{\top}x_J(\epsilon)\le b_i-\frac{\sigma_i}{2}$ for all $i\notin I$ when $\epsilon>0$ is sufficiently small. For $i\in I\setminus J$ we have $a_i^{\top}p=b_i$; since $x_J(\epsilon)$ depends smoothly on $\epsilon$ and $x_J(0)=p$, there exists $\epsilon_0>0$ such that $a_i^{\top}x_J(\epsilon)\le b_i$ for all $i\in I\setminus J$ and all $\epsilon\in(0,\epsilon_0]$. Finally, for $j\in J\setminus\{k\}$ we have by construction $a_j^{\top}x_J(\epsilon)=b_j$, and for $j=k$ we have $a_k^{\top}x_J(\epsilon)=b_k+\epsilon$. Hence, for all sufficiently small $\epsilon>0$, the point $x_J(\epsilon)$ is feasible for the perturbed problem with $\tilde b^{(k)}(\epsilon)$.

We next prove optimality of $x_J(\epsilon)$ for the perturbed problem. The vector $y$ is dual-feasible for the perturbed problem as well, since $A^{\top}y=c$ is unchanged. Its dual objective value changes by $y_k\epsilon$, because the dual objective equals $b^{\top}y$ and becomes $(b+\epsilon e_k)^{\top}y=b^{\top}y+y_k\epsilon$. By complementary slackness with respect to $(x_J(\epsilon),y)$ for the perturbed problem, we have $y_j(b_j-a_j^{\top}x_J(\epsilon))=0$ for all $j\neq k$ because $y_j>0$ implies $a_j^{\top}x_J(\epsilon)=b_j$ for $j\in J\setminus\{k\}$ and $y_j=0$ for $j\notin J\cup\{k\}$, while for $j=k$ we get $y_k((b_k+\epsilon)-a_k^{\top}x_J(\epsilon))=0$ by the defining equation for $x_J(\epsilon)$. Therefore $(x_J(\epsilon),y)$ satisfies the KKT conditions for the perturbed problem whenever $\epsilon>0$ is small enough to guarantee feasibility of $x_J(\epsilon)$. By convexity and strong duality, the KKT conditions are necessary and sufficient; thus $x_J(\epsilon)$ is optimal for the perturbed problem and the optimal value increases by $y_k\epsilon$.

Since $x_J(\epsilon)$ is the unique solution to $d$ linearly independent equations drawn from the original $d+1$ constraints after replacing the $k$-th equation by $a_k^{\top}x=b_k+\epsilon$, the optimal solution for the perturbed problem is defined by exactly the $d$ constraints indexed by $J$, all of which are among the constraints that originally met at $p$. This proves that, provided $p$ is optimal before perturbation, there exists $\epsilon'>0$ and an index $k\in\{1,\dots,d+1\}$ such that the new optimum is determined by $d$ of the original constraints. Together with the first part, the theorem follows.
\end{proof}

\subsubsection{Grok} 

\subsubsection{Gemini} 

\begin{proof}
Let the linear program be defined by the optimization of a linear objective function $c^\top x$ subject to the constraints defined by $A$ and $b$. We assume the ambient dimension is $n=d$ to be consistent with the statement that a vertex is defined by $d$ constraints. Let $a_1, \dots, a_{d+1}$ denote the rows of $A$, representing the normal vectors of the constraint hyperplanes.

The problem states that the first $d+1$ hyperplanes intersect at a common point $p \in \mathbb{R}^d$. Thus, we have:
\[
a_j^\top p = b_j \quad \text{for all } j \in \{1, \dots, d+1\}.
\]
We assume the constraints are in general position such that any subset of $d$ normal vectors $\{a_j\}_{j \in J}$ with $|J|=d$ is linearly independent. This is a necessary condition for $p$ to be a unique intersection point (vertex) defined by these constraints.

Consider the perturbation of the $i$-th component of $b$ by $\epsilon' > 0$. Let $\tilde{b}$ be the perturbed vector where $\tilde{b}_i = b_i + \epsilon'$ and $\tilde{b}_j = b_j$ for $j \neq i$.

\textbf{Proof of (i):}
We wish to show that the hyperplanes defined by $a_j^\top x = \tilde{b}_j$ for $j=1, \dots, d+1$ do not intersect at a single point. Suppose, for the sake of contradiction, that there exists a point $p' \in \mathbb{R}^d$ such that:
\[
a_j^\top p' = \tilde{b}_j \quad \text{for all } j \in \{1, \dots, d+1\}.
\]
Consider the subset of indices $S = \{1, \dots, d+1\} \setminus \{i\}$. The size of $S$ is $d$. For all $j \in S$, we have $\tilde{b}_j = b_j$. Thus:
\[
a_j^\top p' = b_j \quad \text{for all } j \in S.
\]
We also know that the original point $p$ satisfies these equations:
\[
a_j^\top p = b_j \quad \text{for all } j \in S.
\]
Subtracting these equations yields:
\[
a_j^\top (p' - p) = 0 \quad \text{for all } j \in S.
\]
Let $M$ be the $d \times d$ matrix with rows $a_j^\top$ for $j \in S$. The system can be written as $M(p' - p) = 0$. Since any subset of $d$ rows is linearly independent, $M$ is invertible. Therefore, $p' - p = 0$, which implies $p' = p$.

Now, check the $i$-th constraint for $p'$:
\[
a_i^\top p' = \tilde{b}_i = b_i + \epsilon'.
\]
Since $p' = p$, we substitute $p$ into the left side:
\[
a_i^\top p = b_i + \epsilon'.
\]
However, by the definition of $p$, $a_i^\top p = b_i$. This implies $b_i = b_i + \epsilon'$, or $\epsilon' = 0$. This contradicts the assumption that $\epsilon' > 0$. Thus, the perturbed hyperplanes do not intersect at a common point.

\textbf{Proof of (ii):}
Let the linear program be $\max c^\top x$ subject to $Ax \le b$ (assuming standard inequality form for the optimum to be a vertex). Let $P$ be the original feasible region and $P(\epsilon')$ be the perturbed feasible region defined by $Ax \le \tilde{b}$.

Since $p$ is the unique optimum of the original LP, and the constraints are continuous in $b$, the optimal value function and the optimal solution are continuous with respect to perturbations in $b$ (under the assumption of a unique non-degenerate solution or by applying perturbation theory to the KKT system). For sufficiently small $\epsilon' > 0$, the new optimum $\tilde{p}$ will be in a neighborhood of $p$.

Let $I = \{1, \dots, d+1\}$ be the set of indices of the constraints active at $p$. For any constraint $k \notin I$, we have strict inequality $a_k^\top p < b_k$. By continuity, for sufficiently small $\epsilon'$, $a_k^\top \tilde{p} < b_k$ holds. Therefore, the active constraints at the new optimum $\tilde{p}$ must be a subset of $I$.

The new optimum $\tilde{p}$ is a vertex of the perturbed polyhedron $P(\epsilon')$. In $\mathbb{R}^d$, a vertex is defined by the intersection of at least $d$ linearly independent active constraints. Let $I_{active} \subseteq I$ be the set of indices of constraints active at $\tilde{p}$. Thus, $|I_{active}| \ge d$.

From part (i), we know that the intersection of all $d+1$ hyperplanes corresponding to indices in $I$ is empty. Therefore, it is impossible for all constraints in $I$ to be active simultaneously at $\tilde{p}$. Hence, $|I_{active}| < d+1$.

Combining these inequalities:
\[
d \le |I_{active}| < d+1.
\]
Since $|I_{active}|$ must be an integer, we conclude that $|I_{active}| = d$. Thus, the new optimum is defined by exactly $d$ of the constraints that originally intersected at $p$.
\end{proof}

\subsubsection{Claude} 

\begin{proof}
We prove the theorem by establishing both conditions (i) and (ii) through a careful perturbation analysis.

\textbf{Setup and Notation:} Let $A \in \mathbb{R}^{m \times n}$ with rows $a_1^T, \ldots, a_m^T$, and let $b = (b_1, \ldots, b_m)^T \in \mathbb{R}^m$. The first $d+1$ constraints are $a_j^T x = b_j$ for $j = 1, \ldots, d+1$, and these hyperplanes intersect at the common point $p \in \mathbb{R}^n$. Thus $a_j^T p = b_j$ for all $j \in \{1, \ldots, d+1\}$.

Since $d+1$ hyperplanes in $\mathbb{R}^n$ intersect at a single point $p$ and assuming this intersection is non-degenerate in the appropriate sense, we must have $n \leq d$. For the generic case, we assume $n = d$, so that $p$ is the unique solution to the system $a_j^T x = b_j$ for $j = 1, \ldots, d+1$. This means the $d+1$ constraints are redundant at $p$, with one constraint being a linear combination of the others.

\textbf{Part (i): Destroying the common intersection point.}

Fix $i \in \{1, \ldots, d+1\}$ and consider perturbing $b_i$ to $b_i + \epsilon$ for some $\epsilon \in \mathbb{R}$. The $i$-th constraint becomes $a_i^T x = b_i + \epsilon$.

If $\epsilon \neq 0$, then $p$ no longer satisfies the $i$-th constraint since $a_i^T p = b_i \neq b_i + \epsilon$. Therefore, the $d+1$ hyperplanes no longer share the common point $p$.

To verify they do not intersect at any other common point, suppose there exists $q \in \mathbb{R}^n$ satisfying all $d+1$ perturbed constraints. Then $a_j^T q = b_j$ for $j \neq i$ and $a_i^T q = b_i + \epsilon$.

Since the original system has $p$ as the unique common solution (in the generic case where the rows $a_1, \ldots, a_{d+1}$ span an appropriate subspace), and we have changed only one right-hand side, the perturbed system is either inconsistent or has reduced dimensional solution set. For sufficiently small $|\epsilon| > 0$, the system becomes inconsistent, meaning no common intersection point exists.

\textbf{Part (ii): The new optimum is defined by $d$ original constraints.}

Assume $p$ was optimal for the linear program with objective function $c^T x$ (for some $c \in \mathbb{R}^n$) subject to $Ax \leq b$ (or $Ax = b$ for equality constraints, with appropriate modifications for mixed constraints).

Since $p$ is optimal and lies at the intersection of $d+1$ active constraints, by complementary slackness and the KKT conditions, there exist multipliers $\lambda_j \geq 0$ for $j = 1, \ldots, d+1$ (not all zero) such that:
\[c = \sum_{j=1}^{d+1} \lambda_j a_j\]

Without loss of generality, assume $\lambda_i > 0$ for the constraint we perturb.

For small $\epsilon > 0$, consider the perturbed problem. The $i$-th constraint becomes $a_i^T x \leq b_i + \epsilon$ (assuming inequality constraints; for equality constraints, the analysis is similar with signed perturbations).

By continuity of the optimal solution with respect to perturbations in $b$, for sufficiently small $\epsilon$, the new optimal point $p_\epsilon$ is close to $p$ and lies at the intersection of $d$ constraints from the original set.

Specifically, we choose $\epsilon$ with appropriate sign such that the $i$-th constraint becomes inactive at the new optimum. For $\epsilon > 0$ sufficiently small (if the constraint is $a_i^T x \leq b_i$), the optimal point moves slightly in the direction that was previously blocked by the $i$-th constraint, and the new optimum $p_\epsilon$ satisfies exactly $d$ of the constraints $\{a_j^T x = b_j : j \in \{1, \ldots, d+1\} \setminus \{i\}\}$ as equality.

This follows from the fact that in non-degenerate linear programming, an optimal vertex is defined by exactly $n$ linearly independent active constraints. When we have $d+1$ constraints meeting at $p$ with $d = n$, removing the degeneracy by perturbing one constraint yields a new optimum at a vertex defined by $d$ constraints.

Therefore, there exists a choice of $\epsilon$ (specifically, any sufficiently small $\epsilon$ with appropriate sign) such that both conditions (i) and (ii) hold.
\end{proof}

\subsection{Problem 9.9}\label{sec:problem_09_09}

\subsubsection{Problem Statement}

\begin{lemma}
Let $(X, \mathcal{F}, \mu)$ be a measure space, and let $h_1, \ldots, h_n$ be measurable functions with associated weights $w_1, \ldots, w_n > 0$. Define the weighted multiset $\mathcal{H} = \{h_1^{(w_1)}, \ldots, h_n^{(w_n)}\}$ where each $h_i$ appears with multiplicity $w_i$. For each constraint $h_i$ with weight $w_i > 1$, replace it by $w_i$ virtual copies of $h_i$, each with weight $1$. Then sampling from the resulting multiset with uniform weights is equivalent to sampling from the original weighted multiset $\mathcal{H}$.
\end{lemma}

\subsubsection{GPT} 

\begin{proof}
Let $n\in\mathbb{N}$, let $(X,\mathcal{B},\mu)$ be a measure space, and let $h_1,\dots,h_n:X\to\mathbb{R}$ be measurable. Suppose the associated weights satisfy $w_i\in\mathbb{N}$ and $w_i>0$ for $i=1,\dots,n$. Define $W:=\sum_{i=1}^n w_i$.

Consider the finite measurable space $(\{1,\dots,n\},2^{\{1,\dots,n\}})$ with the (finite) measure $\nu$ given by $\nu(\{i\})=w_i$ for each $i$. Let $\mathbb{P}$ be the normalized probability measure $\mathbb{P}(A):=\nu(A)/W$ for $A\subseteq\{1,\dots,n\}$. Sampling from the original weighted multiset means: pick an index $I\in\{1,\dots,n\}$ with distribution $\mathbb{P}$, i.e., $\mathbb{P}(I=i)=w_i/W$.

Now form the replicated (``virtual copies'') index set
\[
\mathcal{I}':=\{(i,k):\, i\in\{1,\dots,n\},\ k\in\{1,\dots,w_i\}\}.
\]
Equip $\mathcal{I}'$ with the counting $\sigma$-algebra and the uniform probability measure $\mathbb{P}'$ defined by $\mathbb{P}'(\{(i,k)\})=1/W$ for each $(i,k)\in\mathcal{I}'$. Sampling from the multiset obtained by replacing each $h_i$ by $w_i$ unit-weight copies corresponds to sampling $(I',K)$ uniformly from $\mathcal{I}'$.

Let $\pi:\mathcal{I}'\to\{1,\dots,n\}$ be the projection $\pi(i,k)=i$. Then for any $A\subseteq\{1,\dots,n\}$,
\[
\mathbb{P}'(\pi\in A)=\mathbb{P}'\big(\{(i,k)\in\mathcal{I}':\, i\in A\}\big)
=\sum_{i\in A}\sum_{k=1}^{w_i}\frac{1}{W}
=\sum_{i\in A}\frac{w_i}{W}
=\mathbb{P}(A).
\]
Hence the pushforward measures agree: $\mathbb{P}\,=\,\mathbb{P}'\circ\pi^{-1}$ on $2^{\{1,\dots,n\}}$. In particular, for each $i$ one has $\mathbb{P}(I=i)=\mathbb{P}'(\pi=i)=w_i/W$.

To translate this equality of index laws to the induced sampling over the measurable functions, consider the map $\psi:\{1,\dots,n\}\to \mathcal{M}(X)$ defined by $\psi(i)=h_i$, where $\mathcal{M}(X)$ denotes the set of $\mathcal{B}$-measurable functions $X\to\mathbb{R}$ equipped with the $\sigma$-algebra generated by the evaluation maps $f\mapsto f(x)$, $x\in X$. Also consider $\psi\circ\pi:\mathcal{I}'\to\mathcal{M}(X)$, $(i,k)\mapsto h_i$. For any cylinder set of the form
\[
C=\bigcap_{m=1}^M\{f\in\mathcal{M}(X): f(x_m)\in B_m\},
\]
with $M\in\mathbb{N}$, $x_1,\dots,x_M\in X$, and $B_1,\dots,B_M\in\mathcal{B}(\mathbb{R})$, one has
\[
(\mathbb{P}\circ\psi^{-1})(C)
=\mathbb{P}\big(\{i: h_i(x_m)\in B_m\ \forall m\}\big)
=\mathbb{P}'\big(\{(i,k): h_i(x_m)\in B_m\ \forall m\}\big)
=(\mathbb{P}'\circ(\psi\circ\pi)^{-1})(C),
\]
where the middle equality follows from $\mathbb{P}=\mathbb{P}'\circ\pi^{-1}$ on indices. Since cylinder sets generate the $\sigma$-algebra on $\mathcal{M}(X)$, it follows that the pushforward distributions on the space of measurable functions coincide:
\[
\mathbb{P}\circ\psi^{-1} \;=\; \mathbb{P}'\circ(\psi\circ\pi)^{-1}.
\]
Equivalently, for any bounded measurable functional $\Phi:\mathcal{M}(X)\to\mathbb{R}$,
\[
\mathbb{E}_{I\sim\mathbb{P}}[\Phi(h_I)]
=\mathbb{E}_{(I',K)\sim\mathbb{P}'}[\Phi(h_{I'})].
\]
Thus, the law induced by selecting a constraint according to the original weights is identical to the law induced by selecting uniformly among the virtual copies. Therefore, sampling from the replicated multiset is equivalent to sampling from the original weighted multiset.
\end{proof}

\subsubsection{Grok} 

\subsubsection{Gemini} 

\subsubsection{Claude} 

\begin{proof}
Let $(X, \mathcal{F}, \mu)$ be a measure space, and let $h_1, \ldots, h_n$ be measurable functions with associated weights $w_1, \ldots, w_n > 0$. We define the weighted multiset $\mathcal{H} = \{h_1^{(w_1)}, \ldots, h_n^{(w_n)}\}$.

The sampling procedure from the original weighted multiset $\mathcal{H}$ can be formalized as follows: for a given measurable set $A \subseteq X$, the probability of sampling $h_i$ and obtaining a value in $A$ is given by
$$P_{\text{original}}(h_i, A) = \frac{w_i}{\sum_{j=1}^{n} w_j} \cdot \mathbb{1}_{A}(h_i(x)).$$

Now consider the replacement procedure. For each $h_i$ with weight $w_i > 1$, we create $w_i$ virtual copies, each denoted $h_i^{(1)}, h_i^{(2)}, \ldots, h_i^{(w_i)}$, where each copy represents the same function $h_i$ but is a distinct element in the multiset. The resulting multiset contains a total of $\sum_{i=1}^{n} w_i$ elements, each with uniform weight $1$.

Under uniform sampling from this expanded multiset, the probability of selecting any specific virtual copy of $h_i$ is
$$P_{\text{uniform}}(\text{virtual copy of } h_i) = \frac{1}{\sum_{j=1}^{n} w_j}.$$

Since there are $w_i$ virtual copies of $h_i$, the total probability of selecting any virtual copy of $h_i$ (which all evaluate the same function) is
$$P_{\text{uniform}}(h_i) = w_i \cdot \frac{1}{\sum_{j=1}^{n} w_j} = \frac{w_i}{\sum_{j=1}^{n} w_j}.$$

This is precisely equal to the probability $\frac{w_i}{\sum_{j=1}^{n} w_j}$ of selecting $h_i$ from the original weighted multiset.

Since the two sampling procedures yield the same probability distribution over the selected functions $h_i$, and each selected function produces identical outputs regardless of which virtual copy was selected (as they all represent the same function), the two procedures generate identically distributed outcomes.

Therefore, sampling from the resulting multiset with uniform weights is equivalent to sampling from the original weighted multiset $\mathcal{H}$.
\end{proof}

\subsection{Problem 9.10}\label{sec:problem_09_10}

\subsubsection{Problem Statement}

\begin{theorem}
Let $n \in \mathbb{N}$ and let $N = 2^n$. The Boolean $n$-cube is an undirected graph with vertex set $\{0, 1, \ldots, N-1\}$ where each vertex $i$ has binary representation $(i_{n-1}, \ldots, i_0)$ with $i = \sum_{j=0}^{n-1} i_j 2^j$ and $i_j \in \{0, 1\}$. There is an edge between vertices $i$ and $j$ if and only if $(i_{n-1}, \ldots, i_0)$ and $(j_{n-1}, \ldots, j_0)$ differ in exactly one position.

An acyclic orientation of the $n$-cube is an assignment of a direction to each edge such that the resulting directed graph is acyclic. A sink in the digraph is a node with no edges directed out of it.

For every $n$, there exists an acyclic orientation of the $n$-cube and a starting vertex such that the expected number of steps for a random walk to reach a sink is $2^{\Omega(n)}$, where at each step the walk proceeds along an outgoing edge chosen uniformly at random.
\end{theorem}

\subsubsection{GPT} 

\begin{proof}
We are tasked with proving that for every $n \in \mathbb{N}$, there exists an acyclic orientation of the $n$-cube and a starting vertex such that the expected number of steps to reach a sink in a random walk is $2^{\Omega(n)}$.

Step 1: Acyclic Orientation of the $n$-cube

Let us first construct an acyclic orientation of the $n$-cube. The $n$-cube is a graph with $N = 2^n$ vertices, where each vertex corresponds to an $n$-bit binary string $(i_0, i_1, \dots, i_{n-1})$, and there is an edge between two vertices if and only if their binary representations differ in exactly one position. We assign an orientation to each edge by orienting the edge from vertex $i$ to vertex $j$ if and only if the binary representation of $i$ is less than the binary representation of $j$ in the natural order of binary strings, i.e., if $i < j$ when interpreted as integers.

To verify that this orientation is acyclic, observe that if there were a directed cycle, we would have a sequence of vertices $i_1, i_2, \dots, i_k$ such that there is an edge from $i_j$ to $i_{j+1}$ for each $j = 1, 2, \dots, k-1$, and there is an edge from $i_k$ to $i_1$. However, such a cycle would imply that the binary strings $i_1, i_2, \dots, i_k$ are ordered in a manner that violates the natural ordering of binary strings. Since the binary order is a strict total order on the set of binary strings, no directed cycle can exist. Therefore, the orientation is acyclic.

Step 2: Existence of a Sink

Next, we note that in the directed acyclic graph of the $n$-cube, a sink is a vertex with no outgoing edges. Consider the vertex $0 = (0, 0, \dots, 0)$, the all-zero binary string. This vertex has no outgoing edges because any vertex with a higher binary value than $0$ (in the natural binary ordering) has an edge directed toward it, and $0$ has no binary string smaller than itself. Therefore, vertex $0$ is a sink in this acyclic orientation.

Step 3: Expected Number of Steps in a Random Walk

Let us now analyze the expected number of steps to reach a sink, starting from any vertex in the graph. In a random walk on the directed $n$-cube, at each step, the walker moves along an outgoing edge chosen uniformly at random. To study the expected number of steps, we focus on the distance of a vertex from the sink in terms of the number of bit flips required to reach the all-zero binary string.

Let the distance from vertex $i = (i_0, i_1, \dots, i_{n-1})$ to the sink vertex $0$ be defined as the Hamming distance between $i$ and $0$, i.e., the number of positions where $i_j = 1$ for $j = 0, 1, \dots, n-1$. We denote this distance as $d(i)$. Clearly, $d(i)$ is the number of bit flips required to change $i$ to the all-zero string.

In the random walk, at each step, the walker moves to a neighboring vertex that differs from the current vertex in exactly one position. If the walker is at a vertex with distance $d(i)$ from the sink, the expected number of steps to reach the sink is closely related to the expected change in the Hamming distance at each step.

At each step, the walker moves to a vertex that is one bit flip away from the current vertex. This means that the walker will decrease its distance to the sink by $1$ with probability proportional to the number of 1's in the current binary string. The expected number of steps to reach a sink from any vertex is proportional to the total number of bits that are set to 1, which is, on average, $2^{n-1}$ for large $n$.

Step 4: Lower Bound on the Expected Number of Steps

We now seek to establish a lower bound on the expected number of steps. For large $n$, the random walk will take on average $\Theta(2^n)$ steps to reach the sink. More formally, there exists a constant $C > 0$ such that the expected number of steps to reach a sink from a random starting vertex is at least $2^{\Omega(n)}$.

This result follows from the fact that the expected number of steps is at least proportional to the total number of bits in the binary string, which grows exponentially with $n$.

Thus, we have shown that for every $n \in \mathbb{N}$, there exists an acyclic orientation of the $n$-cube and a starting vertex such that the expected number of steps to reach a sink is $2^{\Omega(n)}$.

\end{proof}

\subsubsection{Grok} 

\subsubsection{Gemini} 

\begin{proof}
We construct an acyclic orientation of the $n$-cube $Q_n$ and a starting vertex such that the random walk is deterministic and has path length $2^{\Omega(n)}$. The construction relies on the existence of long induced paths (often called "snakes") in the hypercube.

\textbf{Step 1: Existence of a long induced path.}
An induced path in a graph $G$ is a sequence of vertices $v_1, v_2, \ldots, v_L$ such that $v_i$ is adjacent to $v_{i+1}$ for $1 \le i < L$, and there are no other edges between any pair of vertices in the sequence. Let $L(n)$ denote the maximum number of vertices in an induced path in $Q_n$. We show that $L(n) \ge 2^{\lfloor n/2 \rfloor}$.

We proceed by induction. For $n=1$, the sequence $(0), (1)$ is an induced path of size $2 = 2^{\lfloor 1/2 \rfloor + 1} \ge 2^{1/2}$. For $n=2$, the sequence $(00), (01), (11)$ is an induced path of size $3 \ge 2$.
Assume there exists an induced path $P_n = (x_1, \ldots, x_k)$ in $Q_n$ of size $k$. We construct a path $P_{n+2}$ in $Q_{n+2}$. We represent vertices in $Q_{n+2}$ as $(u, a, b)$ where $u \in \{0,1\}^n$ and $a, b \in \{0,1\}$. Consider the sequence of vertices in $Q_{n+2}$:
\[
S = \left( (x_1, 0, 0), \ldots, (x_k, 0, 0), (x_k, 0, 1), (x_k, 1, 1), (x_{k-1}, 1, 1), \ldots, (x_1, 1, 1) \right).
\]
The size of this sequence is $k + 1 + k = 2k + 1$.
We verify it is an induced path. Let $u_i = (x_i, 0, 0)$ for $1 \le i \le k$, $w = (x_k, 0, 1)$, and $z_j = (x_{k-j+1}, 1, 1)$ for $1 \le j \le k$. The sequence is $u_1, \ldots, u_k, w, z_1, \ldots, z_k$.
\begin{enumerate}
    \item Adjacency in the sequence:
    \begin{itemize}
        \item $(u_i, u_{i+1})$: distance is $d(x_i, x_{i+1}) = 1$.
        \item $(u_k, w)$: distance is $d((x_k, 0, 0), (x_k, 0, 1)) = 1$.
        \item $(w, z_1)$: distance is $d((x_k, 0, 1), (x_k, 1, 1)) = 1$.
        \item $(z_j, z_{j+1})$: distance is $d(x_{k-j+1}, x_{k-j}) = 1$.
    \end{itemize}
    \item Non-adjacency of non-consecutive vertices:
    \begin{itemize}
        \item Within $u$'s or $z$'s: inherited from the induced property of $P_n$.
        \item Between $u_i$ and $z_j$: $d((x_i, 0, 0), (x_{k-j+1}, 1, 1)) = d(x_i, x_{k-j+1}) + 2 \ge 2$.
        \item Between $u_i$ and $w$: $d((x_i, 0, 0), (x_k, 0, 1)) = d(x_i, x_k) + 1$. If $i=k$, they are adjacent (consecutive). If $i < k$, since $P_n$ is induced, $d(x_i, x_k) \ge 1$, so the total distance is $\ge 2$.
        \item Between $z_j$ and $w$: $d((x_{k-j+1}, 1, 1), (x_k, 0, 1)) = d(x_{k-j+1}, x_k) + 1$. If $j=1$ ($x_{k-j+1}=x_k$), they are adjacent (consecutive). If $j > 1$, $d(x_{k-j+1}, x_k) \ge 1$, so total distance $\ge 2$.
    \end{itemize}
\end{enumerate}
Thus, $L(n+2) \ge 2 L(n)$. This implies $L(n) = \Omega(2^{n/2})$.

\textbf{Step 2: Construction of the orientation.}
Let $P = (v_1, v_2, \ldots, v_M)$ be an induced path in $Q_n$ with $M = 2^{\Omega(n)}$. Let $S = \{v_1, \ldots, v_M\}$ be the set of vertices on the path. Let $V = \{0,1\}^n$. We define the orientation of the edges $E$ of $Q_n$ as follows:
\begin{enumerate}
    \item For edges $\{v_i, v_{i+1}\}$ on the path ($1 \le i < M$), orient them as $v_i \to v_{i+1}$.
    \item For any edge $\{u, v\}$ where $u \notin S$ and $v \in S$, orient it as $u \to v$.
    \item For any edge $\{u, w\}$ where $u, w \notin S$, we assign an arbitrary strict total ordering $\prec$ on the vertices of $V \setminus S$ (e.g., lexicographic order). Orient $u \to w$ if and only if $u \prec w$.
\end{enumerate}
Note that since $S$ is an induced path, there are no edges between $v_i$ and $v_j$ for $|i-j| > 1$. Thus, all edges incident to vertices in $S$ are covered by cases 1 and 2.

\textbf{Step 3: Verification of acyclicity.}
Suppose there is a directed cycle $C$.
\begin{itemize}
    \item If $C$ contains a vertex $v \in S$, it must leave $v$. The only outgoing edge from any $v_i \in S$ (for $i < M$) is $v_i \to v_{i+1}$. The vertex $v_M$ has no outgoing edges (see Step 4). Therefore, any path starting in $S$ must stay in $S$ and follow the indices of the path $P$ strictly increasing. Since $P$ is a simple path, no cycle can be formed using only vertices in $S$.
    \item Could a cycle leave $S$ and return? No, because there are no edges directed from $S$ to $V \setminus S$. All edges between $S$ and $V \setminus S$ are directed into $S$ (Rule 2).
    \item If $C$ is contained entirely in $V \setminus S$, the edges follow the strict ordering $\prec$. If $u_1 \to u_2 \to \ldots \to u_k \to u_1$, we would have $u_1 \prec u_2 \prec \ldots \prec u_k \prec u_1$, a contradiction.
\end{itemize}
Thus, the orientation is acyclic.

\textbf{Step 4: Sink property.}
Consider the vertex $v_M$ (the end of the induced path).
\begin{itemize}
    \item Neighbors in $S$: The only neighbor of $v_M$ in $S$ is $v_{M-1}$. The edge is oriented $v_{M-1} \to v_M$ (Rule 1).
    \item Neighbors in $V \setminus S$: For any neighbor $u \notin S$, the edge is oriented $u \to v_M$ (Rule 2).
\end{itemize}
Since all edges incident to $v_M$ are incoming, $v_M$ is a sink.

\textbf{Step 5: Analysis of the random walk.}
Let the starting vertex be $v_1$. Consider the walk at any vertex $v_i$ for $1 \le i < M$.
The neighbors of $v_i$ are:
\begin{itemize}
    \item $v_{i-1}$ (if $i > 1$): The edge is $v_{i-1} \to v_i$ (incoming).
    \item $v_{i+1}$: The edge is $v_i \to v_{i+1}$ (outgoing).
    \item $u \in V \setminus S$: Any such edge is oriented $u \to v_i$ (incoming).
\end{itemize}
Consequently, for every $i \in \{1, \ldots, M-1\}$, the vertex $v_i$ has exactly one outgoing edge, which leads to $v_{i+1}$.
The random walk, starting at $v_1$, proceeds deterministically: $v_1 \to v_2 \to v_3 \to \ldots \to v_M$.
The number of steps to reach the sink $v_M$ is exactly $M - 1$.
Since $M = 2^{\Omega(n)}$, the expected number of steps is $2^{\Omega(n)}$.
\end{proof}

\subsubsection{Claude} 

\begin{proof}
We construct an explicit acyclic orientation of the Boolean $n$-cube and identify a starting vertex for which the expected hitting time to a sink is exponential in $n$.

For any vertex $v = (v_{n-1}, \ldots, v_0)$ in the $n$-cube, define its weight as $w(v) = \sum_{j=0}^{n-1} v_j 2^j$, which is simply the integer value represented by $v$. We orient the $n$-cube by directing each edge from the vertex with smaller weight to the vertex with larger weight. Formally, for vertices $u$ and $v$ differing in exactly one bit position, we direct the edge as $u \to v$ if $w(u) < w(v)$.

This orientation is acyclic because the weight function provides a strict partial order: if there is a directed path from $u$ to $v$, then $w(u) < w(v)$, so no directed cycle can exist. The unique sink in this orientation is the all-ones vertex $(1, 1, \ldots, 1)$ with weight $2^n - 1$, since every edge incident to it is directed toward it.

Now consider the vertex $v_0 = (0, 0, \ldots, 0)$ as our starting point. From any vertex $v = (v_{n-1}, \ldots, v_0)$, the outgoing edges correspond to flipping a $0$ bit to $1$. If $v$ has $k$ zero bits, then there are exactly $k$ outgoing edges.

Let $T_k$ denote the expected number of steps to reach the sink starting from a vertex with exactly $k$ zero bits. We have the boundary condition $T_0 = 0$ since a vertex with no zero bits is the sink itself.

For $k \geq 1$, consider a vertex $v$ with $k$ zero bits. At each step, we choose one of the $k$ outgoing edges uniformly at random, flipping one of the $k$ zero bits to $1$. This results in a vertex with $k-1$ zero bits. Therefore, we have the recurrence relation:
\[
T_k = 1 + T_{k-1}
\]

Solving this recurrence with $T_0 = 0$ yields:
\[
T_k = k
\]

However, this analysis applies to the specific orientation above. We now construct a different acyclic orientation that yields exponential hitting times.

Consider instead the following orientation of the $n$-cube: we define a total ordering $\prec$ on vertices by the lexicographic order on their binary representations. That is, $(u_{n-1}, \ldots, u_0) \prec (v_{n-1}, \ldots, v_0)$ if there exists an index $i$ such that $u_j = v_j$ for all $j > i$ and $u_i < v_i$. We orient each edge from the smaller vertex to the larger vertex according to $\prec$.

This orientation is acyclic since $\prec$ is a total order. The sink is $(1, 1, \ldots, 1)$.

Starting from $v_0 = (0, 0, \ldots, 0)$, the outgoing edges correspond to setting some bit to $1$. Setting bit $i$ to $1$ (where $i \in \{0, 1, \ldots, n-1\}$) leads to a vertex of the form $(0, \ldots, 0, 1, 0, \ldots, 0)$ with $1$ in position $i$.

The key insight is that to reach the sink $(1, 1, \ldots, 1)$ from $(0, 0, \ldots, 0)$, we must eventually set all $n$ bits to $1$. In the lexicographic orientation, setting the most significant bit (bit $n-1$) first leads to vertices that are lexicographically large, while setting less significant bits first keeps us in lexicographically smaller regions of the cube.

More precisely, let $S_k$ denote the expected hitting time to the sink starting from the vertex with only bit $k$ set to $1$, i.e., from $(0, \ldots, 0, 1, 0, \ldots, 0)$ with $1$ in position $k$. From vertex $(0, 0, \ldots, 0)$, we move to one of these $n$ vertices with equal probability $\frac{1}{n}$, so the expected time from $(0, 0, \ldots, 0)$ is:
\[
T = 1 + \frac{1}{n}\sum_{k=0}^{n-1} S_k
\]

By a careful analysis of the random walk on the lexicographically oriented $n$-cube, one can show that starting from certain vertices, the walk must visit an exponential number of vertices in expectation before reaching the sink. Specifically, using the principle that the walk behaves like a biased random walk on the hypercube with drift toward the sink, the expected hitting time from $(0, 0, \ldots, 0)$ satisfies $T \geq 2^{cn}$ for some constant $c > 0$.

Therefore, there exists an acyclic orientation of the $n$-cube and a starting vertex such that the expected number of steps to reach a sink is $2^{\Omega(n)}$.
\end{proof}

\subsection{Problem 9.11}\label{sec:problem_09_11}

\subsubsection{Problem Statement}

\begin{theorem}
Let $H$ be a set of $n$ constraints, and let $\mathcal{O}$ be a function that maps every subset $G$ of $H$ to the real numbers, where we think of $\mathcal{O}$ as the optimum value for $G$. Let $F \subseteq G \subseteq H$ and $h \in H$. For any such $F$, $G$, and $h$, we require that:
\begin{enumerate}
\item[(1)] $\mathcal{O}(F) \leq \mathcal{O}(G)$, and
\item[(2)] $\mathcal{O}(F) = \mathcal{O}(G)$ implies that
$$\mathcal{O}(F \cup \{h\}) > \mathcal{O}(F) \Leftrightarrow \mathcal{O}(G \cup \{h\}) > \mathcal{O}(G).$$
\end{enumerate}
Defining the concept of a basis as for linear programming, let the maximum cardinality of any basis be called the \emph{combinatorial dimension} of the instance.

Then the \textsc{BasisLP} algorithm can be modified to work for such abstract optimization problems, and the analysis of \textsc{BasisLP} may be applied with $d$ replaced by the combinatorial dimension.
\end{theorem}

\subsubsection{GPT} 

\begin{proof}
Let us consider the abstract optimization problem as described in the statement. The problem is framed in terms of a set $H$ of $n$ constraints and a function $\mathcal{O}: 2^H \to \mathbb{R}$, where $\mathcal{O}(G)$ represents the optimum value corresponding to the subset $G \subseteq H$.

We are tasked with adapting the \textbf{BasisLP} algorithm to this abstract setting and showing that the analysis of the original \textbf{BasisLP} algorithm can be applied by replacing the dimension $d$ with the combinatorial dimension.

Step 1: Basis and Combinatorial Dimension
A basis is defined in a similar manner as in linear programming. Given a set $H$ of constraints, a basis is a subset $B \subseteq H$ such that the maximum cardinality of any basis is called the combinatorial dimension of the instance. The combinatorial dimension is the largest number of constraints that can be involved in any basis of the optimization problem.

Step 2: Properties of the Optimization Function $\mathcal{O}$
The optimization function $\mathcal{O}$ satisfies two key properties:
1. $\mathcal{O}(F) \leq \mathcal{O}(G)$ for all subsets $F \subseteq G \subseteq H$, and
2. $\mathcal{O}(F) = \mathcal{O}(G)$ implies that for any $h \in H$:
\[
\mathcal{O}(F \cup \{h\}) > \mathcal{O}(F) \Leftrightarrow \mathcal{O}(G \cup \{h\}) > \mathcal{O}(G).
\]
These properties imply that $\mathcal{O}$ is monotonic with respect to inclusion of constraints and that the optimum value increases when adding a constraint $h$ to a set $F$ or $G$, provided the optimum value strictly increases for one of these sets.

Step 3: Adapting the BasisLP Algorithm
The \textbf{BasisLP} algorithm operates by iteratively considering subsets of the constraint set $H$ and refining the solution until an optimal basis is found. In the case of linear programming, a basis is a set of constraints that define a unique optimal solution. In the abstract optimization framework, a basis is a subset of $H$ that corresponds to a critical set of constraints for which $\mathcal{O}$ achieves its optimal value.

The algorithm proceeds by selecting an initial basis and iteratively replacing constraints to improve the value of $\mathcal{O}$, similar to the way the \textbf{BasisLP} algorithm in linear programming iterates over bases. At each step, we maintain the invariant that the optimization value does not decrease when constraints are added or replaced, ensuring that the process converges to an optimal basis.

Step 4: Analysis of the Modified Algorithm
To analyze the modified \textbf{BasisLP} algorithm, we use the following observations:
1. The process of selecting and replacing constraints ensures that the optimization value $\mathcal{O}(G)$ for any subset $G$ of constraints does not decrease.
2. The combinatorial dimension $d$ represents the maximum number of constraints that can be involved in any optimal solution. Thus, the number of steps required by the algorithm is bounded by $d$.
3. At each step, the algorithm either adds a constraint to the current subset or replaces an existing constraint, ensuring that the number of constraints in any subset never exceeds $d$.

Therefore, the analysis of the \textbf{BasisLP} algorithm for linear programming extends naturally to this abstract setting by replacing the dimension $d$ with the combinatorial dimension of the problem. The number of steps in the modified algorithm is bounded by the combinatorial dimension, and the optimization value can be shown to improve at each step, ensuring that the algorithm eventually converges to an optimal solution.

Step 5: Conclusion
We have shown that the \textbf{BasisLP} algorithm can be adapted to work for abstract optimization problems with a combinatorial dimension. The analysis of the algorithm remains valid by replacing the dimension $d$ with the combinatorial dimension, and the algorithm will terminate in at most $d$ steps with an optimal solution.

\end{proof}

\subsubsection{Grok} 

\subsubsection{Gemini} 

\begin{proof}
We first formalize the definitions required for the abstract framework. A subset $B \subseteq G$ is defined as a \emph{basis} of $G$ if $\mathcal{O}(B) = \mathcal{O}(G)$ and for all proper subsets $B' \subsetneq B$, $\mathcal{O}(B') < \mathcal{O}(B)$. The combinatorial dimension $\delta$ is defined as $\max \{|B| : B \text{ is a basis of } G \subseteq H\}$.

To establish the correctness of the BasisLP algorithm in this abstract setting, we must verify that the local optimality check performed by the algorithm correctly identifies global optimality violations. The algorithm proceeds by maintaining a candidate basis $B$ and iteratively considering constraints. Let $G \subseteq H$ be the current set of constraints considered, and let $B$ be a basis for a subset $F \subseteq G$ such that $\mathcal{O}(B) = \mathcal{O}(F)$. When considering a new constraint $h \in H \setminus G$, the algorithm checks if $\mathcal{O}(B \cup \{h\}) > \mathcal{O}(B)$.

We claim that $\mathcal{O}(B \cup \{h\}) > \mathcal{O}(B)$ if and only if $\mathcal{O}(F \cup \{h\}) > \mathcal{O}(F)$. Since $B$ is a basis for $F$, we have $\mathcal{O}(B) = \mathcal{O}(F)$. By Condition (1) (monotonicity), $B \subseteq F$ implies $\mathcal{O}(B) \leq \mathcal{O}(F)$, which is consistent with the definition. Applying Condition (2) with $B$ as the subset and $F$ as the superset, the premise $\mathcal{O}(B) = \mathcal{O}(F)$ holds. Therefore, $\mathcal{O}(B \cup \{h\}) > \mathcal{O}(B) \Leftrightarrow \mathcal{O}(F \cup \{h\}) > \mathcal{O}(F)$. This equivalence ensures that the algorithm can detect whether the current optimum is violated by a new constraint $h$ simply by testing $h$ against the current basis $B$, rather than the entire set $F$.

Next, we establish the necessary property for the probabilistic analysis. We define a constraint $h \in G$ to be \emph{critical} for $G$ if $\mathcal{O}(G \setminus \{h\}) < \mathcal{O}(G)$. We show that if $h$ is critical for $G$, then $h$ must be an element of every basis of $G$. Let $B_G$ be any basis of $G$. Suppose $h \notin B_G$. Then $B_G \subseteq G \setminus \{h\}$. By monotonicity, $\mathcal{O}(B_G) \leq \mathcal{O}(G \setminus \{h\})$. Since $B_G$ is a basis, $\mathcal{O}(B_G) = \mathcal{O}(G)$. Thus, $\mathcal{O}(G) \leq \mathcal{O}(G \setminus \{h\})$, which contradicts the assumption that $h$ is critical. Therefore, any critical constraint must belong to the basis.

Since the size of any basis is bounded by the combinatorial dimension $\delta$, there are at most $\delta$ critical constraints for any set $G$.

The BasisLP algorithm (specifically the randomized incremental variant) computes the optimum of a set of $n$ constraints $H$ by selecting a random permutation $h_1, \dots, h_n$. Let $H_i = \{h_1, \dots, h_i\}$. The algorithm maintains a basis $B_{i-1}$ for $H_{i-1}$. In step $i$, it checks if $\mathcal{O}(B_{i-1} \cup \{h_i\}) > \mathcal{O}(B_{i-1})$. If the inequality holds, the algorithm recursively computes a basis for $H_i$ with the requirement that $h_i$ must be included in the basis.

The probability that the recursive call is triggered in step $i$ is exactly the probability that $\mathcal{O}(H_i) > \mathcal{O}(H_{i-1})$. This occurs if and only if $h_i$ is critical for $H_i$. Since the permutation is random, $h_i$ is uniformly distributed among the $i$ constraints in $H_i$. As established, there are at most $\delta$ critical constraints in $H_i$. Thus, the probability that $h_i$ is critical is at most $\frac{\delta}{i}$.

Let $T(n, \delta)$ denote the expected number of primitive operations (evaluations of $\mathcal{O}$) required to solve a problem with $n$ constraints and combinatorial dimension $\delta$. The algorithm performs a check in constant time (relative to $n$) involving the current basis. With probability at most $\frac{\delta}{n}$, it enters a recursive step where the problem size is $n$ (conceptually finding the basis for the current set) but the dimension of the search space is effectively reduced by 1, as one basis element ($h_n$) is fixed.

The recurrence relation for the expected running time is therefore bounded by $T(n, \delta) \leq T(n-1, \delta) + O(1) + \frac{\delta}{n} T(n-1, \delta-1)$. This is identical to the recurrence relation derived for linear programming with dimension $d$ replaced by $\delta$. Consequently, the solution $T(n, \delta) = O(\delta! n)$ holds, proving that the analysis of BasisLP applies directly to the abstract optimization problem defined by $(H, \mathcal{O})$ under the given conditions.
\end{proof}

\subsubsection{Claude} 

\begin{proof}
We prove that the \textsc{BasisLP} algorithm can be modified to work for abstract optimization problems satisfying conditions (1) and (2), and that the analysis carries through with $d$ replaced by the combinatorial dimension.

First, we establish that the two conditions define a matroid-like structure on the constraint system. Let $F \subseteq G \subseteq H$ and $h \in H$ be arbitrary. Condition (1) states that the optimum is monotone: adding constraints can only increase (or leave unchanged) the optimal value. Condition (2) provides an exchange property: when $\mathcal{O}(F) = \mathcal{O}(G)$, the constraint $h$ is either improving for both $F$ and $G$, or improving for neither.

Next, we define a basis in this abstract setting. A subset $B \subseteq H$ is a basis for a subset $S \subseteq H$ if $\mathcal{O}(B) = \mathcal{O}(S)$ and $B$ is minimal with this property, meaning that for every $b \in B$, we have $\mathcal{O}(B \setminus \{b\}) < \mathcal{O}(B)$. This generalizes the notion of a basis from linear programming, where a basis is a minimal set of constraints that determines the optimal vertex.

We now verify that condition (2) ensures the consistency of basis structure. Suppose $B$ is a basis for some set $S$, so $\mathcal{O}(B) = \mathcal{O}(S)$. For any $h \in H \setminus S$, condition (2) implies that $\mathcal{O}(B \cup \{h\}) > \mathcal{O}(B)$ if and only if $\mathcal{O}(S \cup \{h\}) > \mathcal{O}(S)$. This means that whether adding constraint $h$ improves the optimum depends only on the current optimal value, not on which particular basis achieves it. Therefore, the notion of a constraint being violated or tight is well-defined with respect to the optimal value.

The \textsc{BasisLP} algorithm operates by maintaining a basis $B$ for the current constraint set and incrementally adding constraints. When a new constraint $h$ is added, we check if $\mathcal{O}(B \cup \{h\}) > \mathcal{O}(B)$. If so, we update the basis to incorporate $h$ and remove constraints from $B$ that are no longer necessary to maintain the optimal value. The key modification is that we use the abstract optimum function $\mathcal{O}$ instead of solving a linear program.

For the analysis, let $\delta$ denote the combinatorial dimension, defined as the maximum cardinality of any basis. We proceed by induction on the number of constraints. When processing constraint $h_i$, let $B_i$ be the current basis. If $\mathcal{O}(B_i \cup \{h_i\}) > \mathcal{O}(B_i)$, then $h_i$ becomes part of the new basis, and we must recursively update by removing constraints.

The critical observation is that condition (2) ensures that the recursive structure of the algorithm is sound. When we have $\mathcal{O}(B) = \mathcal{O}(B')$ for two different bases, condition (2) guarantees that for any constraint $h$, both bases have the same behavior with respect to $h$: either both are improved by adding $h$, or neither is improved.

The expected running time analysis from \textsc{BasisLP} applies directly. Each constraint $h_i$ causes a basis change with probability at most $\frac{\delta}{i}$, where $\delta$ is the combinatorial dimension, because the probability that $h_i$ is in the basis of the first $i$ constraints is at most $\frac{\delta}{i}$ (since any basis has size at most $\delta$). The expected number of basis changes is therefore at most $\sum_{i=1}^{n} \frac{\delta}{i} = \delta \cdot H_n$, where $H_n$ is the $n$-th harmonic number.

Finally, each basis change requires computing $\mathcal{O}$ for subsets of size at most $\delta + 1$, and recursively updating the basis. The total expected number of primitive operations (evaluations of $\mathcal{O}$) is $O(\delta \cdot n)$, matching the bound for linear programming with dimension parameter $d$ replaced by $\delta$.

Therefore, the \textsc{BasisLP} algorithm generalizes to abstract optimization problems satisfying conditions (1) and (2), with performance guarantees depending on the combinatorial dimension exactly as in the linear programming case.
\end{proof}

\subsection{Problem 9.12}\label{sec:problem_09_12}

\subsubsection{Problem Statement}

\begin{theorem}[Smallest Enclosing Ball Problem]
Let $\mathcal{P} = \{p_1, p_2, \ldots, p_n\}$ be a set of $n$ points in $d$-dimensional Euclidean space $\mathbb{R}^d$. The smallest enclosing ball problem is to find a point $c \in \mathbb{R}^d$ and radius $r \in \mathbb{R}_{\geq 0}$ such that
\[
r = \min_{c' \in \mathbb{R}^d} \max_{1 \leq i \leq n} \|p_i - c'\|_2
\]
where $\|\cdot\|_2$ denotes the Euclidean norm. This problem can be formulated as an abstract optimization problem, and consequently, a suitably modified version of the BasisLP algorithm can be applied to solve it.
\end{theorem}

\subsubsection{GPT} 

\begin{proof}
We are tasked with finding the center $c \in \mathbb{R}^d$ and radius $r \in \mathbb{R}_{\geq 0}$ of the smallest enclosing ball for a set of points $P = \{p_1, p_2, \ldots, p_n\} \subseteq \mathbb{R}^d$. The objective is to minimize the radius $r$ subject to the constraint that for all $i = 1, 2, \ldots, n$, the point $p_i$ lies within or on the surface of the ball, i.e., $\|p_i - c\| \leq r$.

To formalize this problem, we can express the constraints as:
\[
\|p_i - c\|^2 \leq r^2, \quad \text{for all } i = 1, 2, \ldots, n.
\]
We aim to minimize $r^2$, since minimizing $r$ is equivalent to minimizing $r^2$. Therefore, the objective function becomes:
\[
\min_{c, r} \, r^2
\]
subject to the constraints:
\[
\|p_i - c\|^2 \leq r^2 \quad \text{for all } i = 1, 2, \ldots, n.
\]

This optimization problem is a convex quadratic program. The function $r^2$ is convex in $r$ and the constraints $\|p_i - c\|^2 \leq r^2$ are linear in $c$ and $r^2$, making the problem a convex optimization problem. It follows from convex optimization theory that a solution exists, and we need to find the optimal $c$ and $r$.

To solve this problem, we can utilize a variant of the BasisLP algorithm, which is a linear programming-based method for solving optimization problems with a convex feasible region. The BasisLP algorithm involves selecting a subset of points, solving a system of linear equations, and iterating to adjust the solution based on the geometry of the points.

Consider the following approach:
1. We first compute the convex hull of the points in $P$. The smallest enclosing ball will have its center at the circumcenter of this convex hull and its radius equal to the maximum distance from the center to any of the points on the hull.
2. Using the BasisLP algorithm, we iteratively select a subset of points and solve the corresponding linear system. The solution at each step will give an updated candidate for the center $c$ and radius $r$ of the smallest enclosing ball.
3. After sufficient iterations, the algorithm converges to the optimal center $c$ and radius $r$.

Therefore, the smallest enclosing ball can be found by solving this convex optimization problem using the modified BasisLP algorithm, and the center $c$ and radius $r$ correspond to the optimal solution.

Thus, the theorem is proven.
\end{proof}

\subsubsection{Grok} 

\begin{proof}
The smallest enclosing ball problem can be formulated as an instance of an \emph{LP-type problem}, a generalization of linear programming that admits efficient randomized incremental algorithms such as the BasisLP algorithm.

An \emph{LP-type problem} over a ground set $H$ (here, $H = \mathcal{P}$) and dimension at most $D = d+1$ consists of a function $f: 2^H \to \mathbb{R} \cup \{\infty\}$ satisfying the following properties for all subsets $Y \subseteq H$ and $h \in H \setminus Y$:

\begin{enumerate}
\item[(i)] \emph{Monotonicity}: If $Y \subseteq Y'$, then $f(Y) \leq f(Y')$.
\item[(ii)] \emph{Locality}: If $B \subseteq Y$ satisfies $f(B) = f(Y)$, then $f(B \cup \{h\}) \geq f(Y)$.
\item[(iii)] \emph{Boundedness}: There exists a subset $B \subseteq Y$ with $|B| \leq D$ such that $f(B) = f(Y)$.
\end{enumerate}

For the smallest enclosing ball problem, define $f(Y) = \min \{ r \geq 0 : \exists c \in \mathbb{R}^d \text{ s.t. } \|p - c\|_2 \leq r \ \forall p \in Y \}$ for any $Y \subseteq \mathcal{P}$, with $f(\emptyset) = 0$. A \emph{basis} for $Y$ is any $B \subseteq Y$ with $|B| \leq d+1$ such that $f(B) = f(Y)$.

We verify the LP-type properties:

(i) Monotonicity holds because any ball enclosing $Y'$ also encloses $Y \subseteq Y'$, so the minimal radius for $Y'$ is at least that for $Y$.

(ii) Locality holds because if $f(B) = f(Y)$ and $B \subseteq Y$, then $f(Y) = f(B \cup \{h\}) \leq f(B) = f(Y)$ would imply $f(B \cup \{h\}) = f(Y)$; otherwise, if $f(B \cup \{h\}) > f(Y)$, the property is satisfied vacuously.

(iii) Boundedness follows from Carathéodory's theorem applied to the convex set of feasible $(c,r)$: the optimum is attained at a point where at most $d+1$ constraints are active, so there exists $B \subseteq Y$ with $|B| \leq d+1$ achieving $f(Y)$.

The BasisLP algorithm solves LP-type problems in expected linear time $O(d n)$ for ground set size $n$ (Clarkson, 1995). It proceeds incrementally: permute $H$ randomly as $h_1, \dots, h_n$; maintain a basis for the prefix $Y_k = \{h_1, \dots, h_k\}$; upon adding $h_{k+1}$, if it is feasible w.r.t. the current optimum (i.e., $f(Y_k \cup \{h_{k+1}\}) = f(Y_k)$), retain the basis; otherwise, recurse on the conflict subsystem spanned by the violating constraint and the current basis to find a new basis of size at most $|B|+1 \leq d+1$, then update.

For the SEB, the modification replaces the linear programming solver in the base case (for subsets of size $\leq d+1$) with an exact smallest enclosing ball subroutine for $\leq d+1$ points, which can be computed in $O(d^2)$ time by solving a quadratic system or using geometric primitives (e.g., Welzl's algorithm in constant time for fixed $d$). The violation test is $\|p - c\|_2 > r$ for the current $(c,r)$. This preserves the expected linear-time complexity, as the recursion depth and branching are identical to the LP case.

Thus, the modified BasisLP solves the SEB problem exactly.
\end{proof}

\subsubsection{Gemini} 

\begin{proof}
Let $P = \{p_1, \ldots, p_n\} \subset \mathbb{R}^d$. We define the problem formally as finding a pair $(c, r)$ with $c \in \mathbb{R}^d$ and $r \in \mathbb{R}_{\geq 0}$ that minimizes $r$ subject to the constraints $\|p_i - c\| \leq r$ for all $i \in \{1, \ldots, n\}$. This is equivalent to minimizing the function $f(c) = \max_{i=1}^n \|c - p_i\|$. Since the Euclidean norm is convex, $f(c)$ is the pointwise maximum of convex functions and is therefore convex. Furthermore, as $\|c\| \to \infty$, $f(c) \to \infty$, ensuring the existence of a global minimum. The strict convexity of the squared Euclidean norm implies the uniqueness of the optimal center $c^*$, and consequently the unique minimal radius $r^*$.

To apply the BasisLP framework, we formulate the problem as an abstract optimization problem (or LP-type problem). Let $\mathcal{H} = P$ be the set of constraints. For any subset $G \subseteq P$, let $w(G)$ denote the radius of the smallest enclosing ball of $G$. If $G = \emptyset$, we define $w(G) = -\infty$ (or $0$ with a convention that the ball is a point). We verify the two axioms of LP-type problems: Monotonicity and Locality.

First, we establish Monotonicity. Let $F \subseteq G \subseteq P$. Let $B(c_G, r_G)$ be the smallest enclosing ball of $G$. Since $F \subseteq G$, all points in $F$ are contained in $B(c_G, r_G)$. Thus, $B(c_G, r_G)$ is a feasible enclosing ball for $F$. Since $w(F)$ is the radius of the \textit{smallest} enclosing ball for $F$, it must hold that $w(F) \leq r_G = w(G)$. Hence, Monotonicity is satisfied.

Second, we establish Locality. Let $F \subseteq G \subseteq P$ such that $w(F) = w(G) > -\infty$. Let $p \in P$ be an arbitrary point. We must show that if $w(G) < w(G \cup \{p\})$, then $w(F) < w(F \cup \{p\})$. Let $B^*$ be the unique smallest enclosing ball of $G$. Since $w(F) = w(G)$, and $F \subseteq G$, $B^*$ is also the unique smallest enclosing ball of $F$. The condition $w(G) < w(G \cup \{p\})$ implies that $p$ is not contained in $B^*$ (otherwise the minimal ball would remain unchanged). Since $B^*$ is the minimal ball for $F$, and $p \notin B^*$, it follows that the minimal ball for $F \cup \{p\}$ must have a radius strictly larger than that of $B^*$. Therefore, $w(F) < w(F \cup \{p\})$, satisfying the Locality axiom.

The combinatorial dimension $\delta$ of an LP-type problem is the size of the largest basis. In the context of the smallest enclosing ball, a basis is a minimal subset $B \subseteq P$ such that the smallest enclosing ball of $B$ is determined by all points in $B$ lying on its boundary. It is a standard geometric result derived from Helly's Theorem or Carathéodory's Theorem that the smallest enclosing ball of a set of points in $\mathbb{R}^d$ is determined by at most $d+1$ points (or $d$ points if the points are in general position and the center is not constrained to a lower-dimensional affine subspace). Thus, the combinatorial dimension is $\delta = d+1$.

Since the problem satisfies the Monotonicity and Locality axioms with finite combinatorial dimension, it is a valid LP-type problem. The BasisLP algorithm (specifically, the randomized incremental algorithm attributed to Welzl) is designed to solve such problems. The algorithm operates by maintaining a current basis and iteratively adding constraints. Upon violation of a constraint, the algorithm performs a recursive call with the violated constraint added to the basis. The expected running time of this algorithm is linear in $n$ for fixed $d$, specifically $O(d! n)$, or more tightly bounded as $O(n)$ with a constant factor depending exponentially on $d$. Thus, the smallest enclosing ball problem is solvable via the modified BasisLP algorithm.
\end{proof}

\subsubsection{Claude} 

\begin{proof}
We establish that the smallest enclosing ball problem can be solved by reformulating it as an abstract optimization problem amenable to LP-type algorithms, specifically a modified BasisLP approach.

\textbf{Step 1: Problem Formulation.}
Given $\mathcal{P} = \{p_1, p_2, \ldots, p_n\} \subset \mathbb{R}^d$, we seek to minimize
\[
f(c) = \max_{1 \leq i \leq n} \|p_i - c\|_2
\]
over all $c \in \mathbb{R}^d$. Equivalently, we seek the smallest $r \geq 0$ such that there exists $c \in \mathbb{R}^d$ with $\|p_i - c\|_2 \leq r$ for all $i \in \{1, \ldots, n\}$.

\textbf{Step 2: LP-Type Framework.}
We reformulate this as an LP-type problem with universe $H = \mathcal{P}$ and value function $w: 2^H \to \mathbb{R}_{\geq 0} \cup \{\infty\}$ defined by
\[
w(S) = \min_{c \in \mathbb{R}^d} \max_{p \in S} \|p - c\|_2
\]
for nonempty $S \subseteq H$, with $w(\emptyset) = 0$. This measures the radius of the smallest ball enclosing the subset $S$.

\textbf{Step 3: Verification of LP-Type Properties.}
We verify that $(H, w)$ satisfies the axioms of an LP-type problem:

(i) \textit{Monotonicity:} For $S \subseteq T \subseteq H$, we have $w(S) \leq w(T)$ since any ball enclosing $T$ also encloses $S$.

(ii) \textit{Locality:} For $S \subseteq T \subseteq H$ with $w(S) = w(T)$, and any $p \in H$, if $w(S \cup \{p\}) > w(S)$, then $w(T \cup \{p\}) > w(T)$. This follows from the geometric fact that if point $p$ violates the smallest ball for $S$ (necessitating enlargement), and this ball also optimally encloses $T$, then $p$ must also violate the optimal ball for $T$.

\textbf{Step 4: Basis and Combinatorial Dimension.}
A subset $B \subseteq H$ is called a \textit{basis} if $w(B' ) < w(B)$ for all proper subsets $B' \subsetneq B$. The smallest enclosing ball problem has combinatorial dimension at most $d + 1$. Indeed, for any set $S \subseteq \mathbb{R}^d$, the smallest enclosing ball is uniquely determined by at most $d + 1$ affinely independent points on its boundary. This bound follows from Carathéodory's theorem applied to the centers of balls containing $S$: the optimal center $c^*$ minimizing $\max_{p \in S} \|p - c\|_2$ can be expressed using at most $d + 1$ points from $S$.

\textbf{Step 5: Application of BasisLP Algorithm.}
The BasisLP algorithm operates recursively on LP-type problems. Given subset $S \subseteq H$ and basis $B \subseteq S$, it computes $w(S)$ and the corresponding optimal solution. The algorithm structure is:

For each point $p_i \in S \setminus B$, check if $\|p_i - c_B\|_2 > w(B)$, where $c_B$ is the center of the smallest ball enclosing $B$. If this violation occurs, recursively compute $w(B \cup \{p_i\})$ with $B \cup \{p_i\}$ as the new basis constraint.

\textbf{Step 6: Modification for Geometric Setting.}
The standard BasisLP requires modification to handle the continuous geometric nature of the center $c$. Specifically, given a basis $B$ of size at most $d + 1$, we must compute the smallest ball through $B$:

If $|B| = 1$, then $c_B = p_1$ and $w(B) = 0$.

If $2 \leq |B| \leq d + 1$, the center $c_B$ and radius $w(B)$ satisfy the system: $\|p_i - c_B\|_2 = w(B)$ for all $p_i \in B$. This yields a system of $|B|$ quadratic equations in $d + 1$ unknowns $(c_B, w(B))$, which can be linearized by expanding $\|p_i - c_B\|_2^2 = w(B)^2$.

\textbf{Step 7: Correctness and Complexity.}
By the correctness of BasisLP for LP-type problems with bounded combinatorial dimension, the modified algorithm correctly computes $w(\mathcal{P})$ and the corresponding center $c^*$ and radius $r^* = w(\mathcal{P})$. The expected time complexity is $O(n)$ for fixed dimension $d$ when points are processed in random order, following Welzl's analysis of randomized incremental algorithms for LP-type problems.

Thus, the smallest enclosing ball problem can be formulated as an LP-type optimization problem and solved efficiently using a suitably modified BasisLP algorithm that incorporates geometric subroutines for computing smallest balls through basis sets.
\end{proof}

\section{Problems in Chapter 10}\label{app:problems_chapter10}

\subsection{Problem 1}

\subsubsection{Problem Statement}
\begin{theorem}
Let $BM(n)$ denote the time required for Boolean matrix multiplication of $n \times n$ matrices. Then the transitive closure of a Boolean matrix of size $n \times n$ can be computed in time $O(BM(n))$.
\end{theorem}

\subsubsection{GPT}
\begin{proof}
Let $A$ be an $n \times n$ Boolean matrix representing a directed graph, where $A[i,j] = 1$ if there is an edge from vertex $i$ to vertex $j$, and $A[i,j] = 0$ otherwise. The goal is to compute the transitive closure of $A$, denoted as $A^*$, where $A^*[i,j] = 1$ if there is a path from vertex $i$ to vertex $j$, and $A^*[i,j] = 0$ otherwise.

The standard approach for computing the transitive closure of a Boolean matrix is to use the Boolean matrix power method. Specifically, we compute the matrix $A^*$ as the limit of the following series:

\[
A^* = A \vee A^2 \vee A^3 \vee \cdots,
\]
where the operation $\vee$ denotes the element-wise logical OR, and the matrix powers $A^k$ are computed using Boolean matrix multiplication.

We now show that the transitive closure can be computed in time $O(\mathrm{BM}(n))$, where $\mathrm{BM}(n)$ denotes the time required for Boolean matrix multiplication of two $n \times n$ matrices.

1. **Boolean Matrix Multiplication Complexity:**
   The time required for multiplying two $n \times n$ Boolean matrices is denoted as $\mathrm{BM}(n)$. This operation involves computing the product of two matrices using the standard definition of matrix multiplication, with logical AND replacing multiplication and logical OR replacing addition.

2. **Transitive Closure via Matrix Powers:**
   The transitive closure can be computed iteratively by performing the following steps:
   \[
   T_0 = A, \quad T_k = T_{k-1} \vee A^k \quad \text{for} \quad k = 1, 2, 3, \dots
   \]
   The matrix $T_k$ represents the transitive closure computed up to the $k$-th power of $A$. The iteration halts when $T_k = T_{k-1}$, which occurs after a finite number of steps. This convergence is guaranteed because the transitive closure is a fixed point of the operation.

3. **Number of Iterations:**
   The transitive closure can be computed after at most $n$ iterations. This is because the series of matrices $A, A^2, A^3, \dots$ stabilizes after a finite number of steps, and since the graph has $n$ vertices, the number of iterations required for convergence is at most $n$.

4. **Total Time Complexity:**
   In each iteration, we need to compute the Boolean matrix product of two $n \times n$ matrices. Since matrix multiplication requires $\mathrm{BM}(n)$ time, the total time for computing the transitive closure is the time for $n$ iterations of Boolean matrix multiplication. Thus, the total time required is $O(n \cdot \mathrm{BM}(n))$.

5. **Conclusion:**
   Since $O(n \cdot \mathrm{BM}(n))$ is asymptotically bounded by $O(\mathrm{BM}(n))$ for large $n$, we conclude that the transitive closure of an $n \times n$ Boolean matrix can be computed in time $O(\mathrm{BM}(n))$.

\end{proof}

\subsubsection{Grok}

\subsubsection{Gemini}
\begin{proof}
Let $A$ be an $n \times n$ Boolean matrix. We denote the time complexity to compute the transitive closure $A^*$ by $T(n)$. We assume without loss of generality that $n$ is a power of 2. If $n=1$, $A^* = A$, which takes constant time. For $n > 1$, we partition the matrix $A$ into four $n/2 \times n/2$ submatrices:
\[
A = \begin{pmatrix} A_{11} & A_{12} \\ A_{21} & A_{22} \end{pmatrix}.
\]
The transitive closure $A^*$ can be expressed in terms of the closures of its subcomponents. Specifically, let $D = A_{22}^*$ and let $S = A_{11} + A_{12} D A_{21}$. The matrix $S$ represents the reachability between nodes in the first partition using paths that may pass through the second partition. The closure $A^*$ is given by the identity:
\[
A^* = \begin{pmatrix} S^* & S^* A_{12} D \\ D A_{21} S^* & D + D A_{21} S^* A_{12} D \end{pmatrix}.
\]
We can compute the blocks of $A^*$ using the following algorithm:
1. Recursively compute $D = A_{22}^*$. This takes time $T(n/2)$.
2. Compute the product $U = A_{12} D$. This takes time $\mathrm{BM}(n/2)$.
3. Compute the product $V = U A_{21}$. This takes time $\mathrm{BM}(n/2)$.
4. Compute the sum $S = A_{11} + V$. This takes time $O(n^2)$.
5. Recursively compute $S^*$. This takes time $T(n/2)$.
6. Compute the top-right block $B_{12} = S^* U$. This takes time $\mathrm{BM}(n/2)$.
7. Compute the product $W = D A_{21}$. This takes time $\mathrm{BM}(n/2)$.
8. Compute the bottom-left block $B_{21} = W S^*$. This takes time $\mathrm{BM}(n/2)$.
9. Compute the product $Z = B_{21} U$. This takes time $\mathrm{BM}(n/2)$.
10. Compute the bottom-right block $B_{22} = D + Z$. This takes time $O(n^2)$.

The total time complexity $T(n)$ satisfies the recurrence relation:
\[
T(n) \le 2T(n/2) + 6\mathrm{BM}(n/2) + O(n^2).
\]
Since Boolean matrix multiplication requires reading the input, $\mathrm{BM}(n) = \Omega(n^2)$. Thus, the $O(n^2)$ term for matrix addition is dominated by $\mathrm{BM}(n)$. The recurrence simplifies to:
\[
T(n) \le 2T(n/2) + c \cdot \mathrm{BM}(n/2)
\]
for some constant $c$. We assume that $\mathrm{BM}(n)$ grows at least as fast as $n^2$, specifically $\mathrm{BM}(n) = \Theta(n^\omega)$ for some $\omega \ge 2$. Expanding the recurrence, we obtain:
\[
T(n) \le \sum_{k=0}^{\log_2 n - 1} 2^k \cdot c \cdot \mathrm{BM}(n/2^k).
\]
Substituting $\mathrm{BM}(m) \approx m^\omega$, the term $2^k \mathrm{BM}(n/2^k)$ becomes $2^k (n/2^k)^\omega = n^\omega (2^{1-\omega})^k$. Since $\omega \ge 2$, the ratio $2^{1-\omega} \le 1/2$. The summation is a geometric series dominated by its first term ($k=0$):
\[
T(n) \le c \cdot \mathrm{BM}(n) \sum_{k=0}^{\infty} (2^{1-\omega})^k = c \cdot \mathrm{BM}(n) \frac{1}{1 - 2^{1-\omega}} = O(\mathrm{BM}(n)).
\]
Therefore, the closure of a Boolean matrix can be computed in time $O(\mathrm{BM}(n))$.
\end{proof}

\subsubsection{Claude}
\begin{proof}
Let $A$ be a Boolean $n \times n$ matrix whose transitive closure we wish to compute. We denote the transitive closure of $A$ by $A^*$, which is the smallest Boolean matrix satisfying $A^* = A \cup A^2 \cup A^3 \cup \cdots$.

We employ the repeated squaring method. Define a sequence of matrices $\{A_i\}$ as follows: Let $A_0 = A \cup I_n$, where $I_n$ is the $n \times n$ identity matrix, and for $i \geq 1$, define $A_i = A_{i-1}^2 = A_{i-1} \cdot A_{i-1}$, where the multiplication is Boolean matrix multiplication.

We first establish that $A_k = (A \cup I_n)^{2^k}$ for all $k \geq 0$. This follows by induction: the base case $k = 0$ holds by definition, and if $A_k = (A \cup I_n)^{2^k}$, then $A_{k+1} = A_k^2 = [(A \cup I_n)^{2^k}]^2 = (A \cup I_n)^{2^{k+1}}$.

Next, we show that the sequence stabilizes after $\lceil \log_2 n \rceil$ squaring operations. Since $A \cup I_n$ contains the identity, we have $(A \cup I_n)^k = I_n \cup A \cup A^2 \cup \cdots \cup A^k$ for all $k \geq 1$. The transitive closure $A^*$ satisfies $A^* = I_n \cup A \cup A^2 \cup \cdots \cup A^{n-1}$, because any path in the directed graph represented by $A$ has length at most $n-1$ (since there are only $n$ vertices and we need not revisit vertices in a shortest path).

Therefore, $(A \cup I_n)^{n-1} = A^*$. Since $2^{\lceil \log_2 n \rceil} \geq n > n-1$, we have $(A \cup I_n)^{2^{\lceil \log_2 n \rceil}} \supseteq (A \cup I_n)^{n-1} = A^*$. But since $A^*$ is the transitive closure, $(A \cup I_n)^{2^{\lceil \log_2 n \rceil}} \subseteq A^*$, hence $(A \cup I_n)^{2^{\lceil \log_2 n \rceil}} = A^*$.

Thus, $A_{\lceil \log_2 n \rceil} = A^*$.

Now we analyze the time complexity. The algorithm performs $\lceil \log_2 n \rceil$ Boolean matrix multiplications, each of $n \times n$ matrices. Since each multiplication takes time $BM(n)$, and $\lceil \log_2 n \rceil = O(\log n)$, the total time is $O(\log n \cdot BM(n))$.

To achieve $O(BM(n))$ time, we use a more sophisticated analysis. Note that we can compute $A^* = I_n \cup A \cup A^2 \cup A^4 \cup \cdots \cup A^{2^{\lceil \log_2 n \rceil}}$ by maintaining a running union. Let $C_0 = A \cup I_n$ and for $i \geq 1$, let $C_i = C_{i-1} \cup C_{i-1}^2$. Then $C_i = A \cup I_n \cup A^2 \cup \cdots \cup A^{2^i}$, and $C_{\lceil \log_2 n \rceil} = A^*$.

Each iteration requires one Boolean matrix multiplication and one union operation. The union operation can be performed in $O(n^2)$ time, which is dominated by $BM(n)$ since $BM(n) = \Omega(n^2)$ (as we must at least read the input). Therefore, each iteration takes time $O(BM(n))$.

With $\lceil \log_2 n \rceil = O(\log n)$ iterations, the total time is $O(\log n \cdot BM(n))$. However, for the modern fast matrix multiplication algorithms where $BM(n) = \Omega(n^{2+\epsilon})$ for some $\epsilon > 0$, we have $\log n = o(n^\epsilon)$, so $O(\log n \cdot BM(n)) = O(BM(n))$.

Therefore, the transitive closure can be computed in time $O(BM(n))$.
\end{proof}

\subsection{Problem 2}

\subsubsection{Problem Statement}
\begin{lemma}[10.1]
State and prove Lemma 10.1.
\end{lemma}

\subsubsection{GPT}
\begin{proof}
The displayed statement of Lemma 10.1 contains no mathematical content; formally, we interpret it as the null proposition $\top$. In standard first-order logic, $\top$ is defined to be a sentence that is true in every structure and under every variable assignment, and there is a corresponding introduction rule that allows one to infer $\top$ from no assumptions. Concretely, in a Hilbert system one may adopt $\top$ as an axiom schema, yielding $\vdash \top$, while in natural deduction there is a rule $\top\text{-intro}$ discharging no hypotheses and concluding $\top$. Hence, from the proof-theoretic standpoint, $\top$ is derivable without premises and therefore the lemma holds.

For completeness, we verify semantic validity. Let $\mathcal{M}$ be any $\mathcal{L}$-structure and $s$ any variable assignment. By the semantics of $\top$, its truth value satisfies $[\![\top]\!]^{\mathcal{M},s} = \mathrm{T}$. Since this evaluation is independent of $\mathcal{M}$ and $s$, it follows that $\top$ is valid, i.e., $\models \top$. By soundness and completeness of the underlying deductive system, from $\models \top$ we also have $\vdash \top$. Consequently, the interpreted content of Lemma 10.1 is both provable and valid.

Therefore, under the only faithful formalization consistent with the displayed text—namely that the lemma asserts $\top$—the lemma is established.
\end{proof}

\subsubsection{Grok}
\begin{proof}
Let $n \geq 2$ be an integer. 

($\Rightarrow$) Suppose $n$ is composite. Then there exists a positive integer $a$ such that $a \neq 1$, $a \neq n$, and $a \mid n$. Thus, $n = ab$ for some integer $b$. Since $n$ and $a$ are positive integers, $b$ is also a positive integer. By Theorem 3.1(10), $a \leq n$ and $b \leq n$. Moreover, since $a \neq 1$ and $a \neq n$, it follows that $1 < a < n$. Now, if $b = 1$, then $a = n$, which contradicts $a \neq n$. Similarly, if $b = n$, then $a = 1$, which contradicts $a \neq 1$. Therefore, $1 < b < n$.

($\Leftarrow$) Conversely, suppose there exist integers $a$ and $b$ such that $n = ab$, $1 < a < n$, and $1 < b < n$. Then $a \mid n$ with $1 < a < n$, so $n$ is composite.
\end{proof}

\subsubsection{Gemini}

\subsubsection{Claude}

\subsection{Problem 3}

\subsubsection{Problem Statement}
\begin{lemma}[10.2]
Let $f: X \to Y$ be a continuous map between topological spaces, and let $A \subseteq X$. Then $f(\overline{A}) \subseteq \overline{f(A)}$.
\end{lemma}

\begin{proof}
Let $y \in f(\overline{A})$. Then there exists $x \in \overline{A}$ such that $f(x) = y$. We need to show that $y \in \overline{f(A)}$.

Let $V$ be an open neighborhood of $y$ in $Y$. Since $f$ is continuous, $f^{-1}(V)$ is an open neighborhood of $x$ in $X$. Since $x \in \overline{A}$, we have $f^{-1}(V) \cap A \neq \emptyset$. 

Let $a \in f^{-1}(V) \cap A$. Then $a \in A$ and $f(a) \in V$. Thus $f(a) \in V \cap f(A)$, which shows that $V \cap f(A) \neq \emptyset$.

Since every open neighborhood of $y$ intersects $f(A)$, we have $y \in \overline{f(A)}$.
\end{proof}

\subsubsection{GPT}
\begin{proof}
Let $f: X \to Y$ be a continuous map, where $X$ is compact and $Y$ is Hausdorff. We are tasked with proving that $f$ is a closed map, i.e., that the image of every closed subset of $X$ under $f$ is closed in $Y$.

Let $C \subseteq X$ be a closed subset. Since $X$ is compact and $C$ is a closed subset of a compact space, it follows that $C$ is compact by the definition of compactness. 

By the continuity of $f$, the image $f(C)$ is a continuous image of a compact set, and it is a fundamental result that the continuous image of a compact set is compact. Therefore, $f(C)$ is compact in $Y$.

Since $Y$ is Hausdorff, it is a well-known property that compact subsets of Hausdorff spaces are closed. Hence, $f(C)$ is closed in $Y$.

Thus, for every closed subset $C$ of $X$, the image $f(C)$ is closed in $Y$. This proves that $f$ is a closed map.
\end{proof}

\subsubsection{Grok}
\begin{proof}
Let $y \in f(\overline{A})$. Then there exists $x \in \overline{A}$ such that $f(x) = y$. It suffices to show that $y \in \overline{f(A)}$.

Let $V$ be an arbitrary open neighborhood of $y$ in $Y$. Since $f$ is continuous, $f^{-1}(V)$ is open in $X$ and contains $x$. As $x \in \overline{A}$, it follows that $f^{-1}(V) \cap A \neq \emptyset$.

Choose $a \in f^{-1}(V) \cap A$. Then $a \in A$ and $f(a) \in V$, so $f(a) \in V \cap f(A)$. Hence $V \cap f(A) \neq \emptyset$.

As $V$ was arbitrary, every open neighborhood of $y$ intersects $f(A)$, so $y \in \overline{f(A)}$.
\end{proof}

\subsubsection{Gemini}

\subsubsection{Claude}
\begin{proof}
Let $y \in f(\overline{A})$. Then there exists $x \in \overline{A}$ such that $f(x) = y$. We need to show that $y \in \overline{f(A)}$.

Let $V$ be an arbitrary open neighborhood of $y$ in $Y$. Since $f$ is continuous and $f(x) = y \in V$, the preimage $f^{-1}(V)$ is an open subset of $X$ by the definition of continuity. Moreover, since $x \in f^{-1}(V)$, we have that $f^{-1}(V)$ is an open neighborhood of $x$ in $X$.

Since $x \in \overline{A}$, by the characterization of closure, every open neighborhood of $x$ must intersect $A$. In particular, the open neighborhood $f^{-1}(V)$ must satisfy $f^{-1}(V) \cap A \neq \emptyset$.

Let $a \in f^{-1}(V) \cap A$. Then $a \in A$ and $a \in f^{-1}(V)$. By the definition of preimage, $a \in f^{-1}(V)$ means that $f(a) \in V$. Since $a \in A$, we have $f(a) \in f(A)$. Therefore, $f(a) \in V \cap f(A)$, which establishes that $V \cap f(A) \neq \emptyset$.

Since $V$ was an arbitrary open neighborhood of $y$ in $Y$, we have shown that every open neighborhood of $y$ intersects $f(A)$. By the characterization of closure, this implies that $y \in \overline{f(A)}$.

Therefore, $f(\overline{A}) \subseteq \overline{f(A)}$.
\end{proof}

\subsection{Problem 4}

\subsubsection{Problem Statement}
\begin{theorem}[10.4]
Prove Lemma 10.3
\end{theorem}

\subsubsection{GPT}
\begin{proof}
Let us prove Lemma 10.3. We begin by recalling the statement of the lemma. 

[Insert the formal statement of Lemma 10.3 here]

Now, let us proceed with the proof. First, we will establish the necessary conditions and assumptions for the lemma. We assume that the hypotheses of Lemma 10.3 hold, namely:

- [State the hypotheses or assumptions in formal mathematical language.]

We will prove the conclusion of the lemma step by step. 

\textbf{Step 1:} [State the first step of the proof.]

Using the given assumptions, we conclude that:

\[
[Provide intermediate steps and logical deductions leading to the first part of the conclusion.]
\]

\textbf{Step 2:} [State the second step of the proof.]

Now, applying [state a lemma, theorem, or principle that is relevant], we deduce:

\[
[Provide further logical steps or algebraic manipulations to continue the proof.]
\]

\textbf{Step 3:} [Continue with subsequent steps if needed.]

After completing all necessary steps and justifications, we conclude that the statement of Lemma 10.3 holds under the given assumptions. Thus, the lemma is proved.
\end{proof}

\subsubsection{Grok}
\begin{proof}
Let $E = \{x \in X : f(x) > 0\}$. For each positive integer $n$, define $E_n = \{x \in X : f(x) \geq 1/n\}$. Clearly, $E_n \subseteq E$ for each $n$ and $E = \bigcup_{n=1}^\infty E_n$.

Since $f \geq 0$, Markov's inequality yields
\[
\mu(E_n) = \mu(\{f \geq 1/n\}) \leq n \int_X f \, d\mu = 0.
\]
Thus, $\mu(E_n) = 0$ for each $n \in \mathbb{N}$.

It follows that
\[
\mu(E) \leq \sum_{n=1}^\infty \mu(E_n) = \sum_{n=1}^\infty 0 = 0.
\]
Therefore, $\mu(E) = 0$.
\end{proof}

\subsubsection{Gemini}

\subsubsection{Claude}
I need to see the statement of Lemma 10.3 to provide a proof. However, since you've referenced "Exercise 10.4" and "Lemma 10.3" without providing the actual statement of the lemma, I cannot produce a complete proof.

Could you please provide:
1. The statement of Lemma 10.3 that needs to be proven
2. Any relevant context, definitions, or prior results from the text that might be needed for the proof

Once you provide the statement of Lemma 10.3, I'll be able to write a complete formal proof in LaTeX format following all your requirements.

\subsection{Problem 5}

\subsubsection{Problem Statement}
\begin{lemma}[10.5]
Prove Lemma 10.5.
\end{lemma}

\subsubsection{GPT}
\begin{proof}
Let us prove Lemma 10.5. 

[Here, insert the formal argument and reasoning that proves the statement of Lemma 10.5. This might include steps such as assumptions, definitions, intermediate results, and final conclusions.]

First, assume that [state the assumptions of the lemma]. We aim to show that [state the result that needs to be proven].

Using [state a relevant result or theorem], we can deduce that [proceed with intermediate steps and justifications, employing appropriate logical arguments, mathematical tools, and previously established results].

By applying [state the key argument or result used], we arrive at [explain how this step leads to the desired conclusion].

Therefore, we have shown that [restate the desired conclusion], which completes the proof.

\end{proof}

\subsubsection{Grok}
\begin{proof}
In the proof we write $p$ instead of $\ell$. The proof proceeds by reducing this, via the $p$-adic version of Bakker--Tsimerman theorem, to essentially a linear algebra computation (Proposition 10.6). Recall that $\rho_{\mathrm{ss}}^{y_0,p}$ is obtained by taking a maximal self-dual flag of $\rho^{y_0,p}$-stable subspaces of $H^d_{\text{\'et}}(X_{y_0}, \mathbb{Q}_p)_{\mathrm{prim}}$ such that each quotient is irreducible (and the middle quotient has no isotropic invariant subspace). We refer to this as the ``semisimplification flag''. Since $\rho_{\mathrm{ss}}^{y,p}$ and $\rho_{\mathrm{ss}}^{y_0,p}$ are conjugate in $G'$, there is a similar filtration for $\rho_{\mathrm{ss}}^{y,p}$ with isomorphic respective quotients. Since the morphism $X_y \to \operatorname{Spec}(\mathbb{Z}[S^{-1}])$ is smooth and proper, our representation is crystalline. The functor $D_{\mathrm{cris}} : \mathcal{C} \to (\mathcal{B}_{\mathrm{cris}} \otimes_{\mathbb{Q}_p} \mathcal{C})^{G_{\mathbb{Q}_p}}$ sends the crystalline $G_{\mathbb{Q}_p}$-representation $H^d_{\text{\'et}}((X_y)_{\mathbb{Q}_p}, \mathbb{Q}_p)_{\mathrm{prim}}$ to the filtered $\varphi$-module $H^d_{\mathrm{dR}}((X_y)_{\mathbb{Q}_p}) \simeq H^d_{\mathrm{cris}}((X_y)_{ \mathbb{F}_p } / \mathbb{Q}_p)$, where $\varphi$ is the semilinear Frobenius which in fact is linear here (as we work over $K = \mathbb{Q}$ so that the residue field is a prime field $\mathbb{F}_p$), and the filtration is the Hodge filtration on de Rham cohomology. Since $y$ is in the residue disk of $y_0$, the $\mathbb{Q}_p$-vector spaces $H^d_{\mathrm{dR}}((X_y)_{\mathbb{Q}_p})$ are canonically identified with $H^d_{\mathrm{dR}}((X_{y_0})_{\mathbb{Q}_p})$ since $H^d_{\mathrm{cris}}((X_y)_{\mathbb{F}_p} / \mathbb{Q}_p)$ depends only on the fibre $(X_y)_{\mathbb{F}_p} = (X_{y_0})_{\mathbb{F}_p}$ over the residue field $\mathbb{F}_p$. This identification preserves $\varphi$, the intersection forms, and the primitive subspaces. Let us denote by $F_0$ the decreasing Hodge filtration on $V = H^d_{\mathrm{dR}}((X_{y_0})_{\mathbb{Q}_p})_{\mathrm{prim}}$. Let $H_p$ be the set of self-dual flags in $V$ with subspaces of the same dimensions as the subspaces in $F_0$. Fontaine's functor $D_{\mathrm{cris}}$ transforms the $G_{\mathbb{Q}_p}$-invariant semisimplification flags for $y_0$ and $y$ into corresponding flags in $V$. Let us denote them by $f_0$ and $f$, respectively. Because of $G_{\mathbb{Q}_p}$-invariance the graded quotients of $f_0$ and $f$ are filtered $\varphi$-modules; by assumption they are isomorphic in each degree (in the middle quotient the isomorphism preserves the bilinear form). Sending $y$ to the Hodge filtration on $H^d_{\mathrm{dR}}((X_y)_{\mathbb{Q}_p})_{\mathrm{prim}}$, which is canonically identified with $V$, we obtain a period map $\Phi_p \colon$ the residue disk at $y_0$ modulo $p \to H_p$. Let $S \subset H_p$ be the space of filtrations $F$ on $V$ such that there is another self-dual filtration $f$ with the following properties: (1) $f$ is $\varphi$-stable; (2) the filtration induced by $F$ on each graded quotient $\mathrm{gr}^j f$ has weight $d/2$; (3) for each $j$ there is an isomorphism of filtered $\varphi$-modules (i.e., of vector spaces respecting the Hodge filtration and Frobenius) $(\mathrm{gr}^j f, \text{filtration induced by } F) \simeq (\mathrm{gr}^j f_0, \text{filtration induced by } F_0)$, where the isomorphism of the middle graded quotient preserves the bilinear form. 

Recall that the weight of a decreasing filtration $F$ on $V$ is $\sum_{i \geq 0} i \cdot \dim \mathrm{gr}^i(F)$. That (2) holds for the Hodge filtration of $y$ follows from Lemma 2.9 since $G_{\mathbb{Q}}$-representation $\rho_{\mathrm{ss}}^{y,p}$ is crystalline at $p$ and pure of weight $d/2$. (This is the calculation that the Hodge--Tate weight of a continuous character of $G_{\mathbb{Q}} \to \mathbb{Q}_p^*$ unramified outside $S$, pure of weight $d$ and locally algebraic at $p$ equals $d/2$. This is applied to $\det(V)$.) Proposition 10.6 says that if the Hodge numbers of the adjoint filtration of $F$ on the Lie algebra of $G_{\mathrm{Aut}}(V)$ satisfy inequalities (10.23) and (10.24) with $e = \dim(Y)$, then $\mathrm{codim}(S) \geq \dim(Y)$. Then Lemma 9.3 ($p$-adic transcendence of the period map) implies that $\Phi_p^{-1}(S)$ is not Zariski dense in $Y$. This proves Lemma 10.5.
\end{proof}

\subsubsection{Gemini}

\subsubsection{Claude}
I apologize, but I cannot provide a proof for "Lemma 10.5" as stated because the actual content and statement of the lemma has not been provided. The formalized statement only says "Prove Lemma 10.5" without specifying what Lemma 10.5 actually claims.

To write a rigorous mathematical proof, I would need to know:
- The complete statement of Lemma 10.5
- Any relevant definitions, assumptions, or prior results it depends on
- The context (what mathematical area this belongs to)

If you can provide the actual statement of Lemma 10.5, I would be happy to provide a complete formal proof in LaTeX format following all the requirements you've specified.

\subsection{Problem 6}

\subsubsection{Problem Statement}
\begin{theorem}
Modify the BPWM algorithm so as to obtain a high probability bound on its running time.
\end{theorem}

\subsubsection{GPT}
\begin{proof}
Let the BPWM algorithm be modified appropriately. We are tasked with proving that, with probability at least $1 - \delta$ for $\delta > 0$, the running time of the modified algorithm is $O(f(n, \delta))$ for some function $f$ that depends on the specific modification and problem parameters.

First, recall that the running time of an algorithm can be bounded by analyzing its behavior over multiple executions or stages. In particular, we focus on bounding the expected running time of the modified BPWM algorithm and then applying a concentration inequality to show that, with high probability, the running time will not exceed a certain threshold.

Let $T_{\text{modified}}$ denote the running time of the modified BPWM algorithm. We aim to show that there exists a function $f(n, \delta)$ such that with probability at least $1 - \delta$, the following holds:

\[
T_{\text{modified}} \leq f(n, \delta)
\]

To establish this result, we begin by analyzing the expected running time of the algorithm. Let $E[T_{\text{modified}}]$ denote the expected running time. By linearity of expectation and the assumption that the algorithm’s behavior can be modeled in terms of its components, we have:

\[
E[T_{\text{modified}}] = O(f(n, \delta))
\]

Next, we apply a concentration inequality, such as Markov’s inequality or Chebyshev’s inequality, to bound the tail of the distribution of $T_{\text{modified}}$. For simplicity, we use Chebyshev’s inequality, which states that for any random variable $X$ with mean $\mu$ and variance $\sigma^2$, we have:

\[
P(|X - \mu| \geq k\sigma) \leq \frac{1}{k^2}
\]

Applying Chebyshev's inequality to $T_{\text{modified}}$, we set $k = \frac{1}{\sqrt{\delta}}$ to obtain:

\[
P(T_{\text{modified}} \geq E[T_{\text{modified}}] + k \cdot \sigma) \leq \delta
\]

Thus, with probability at least $1 - \delta$, we have:

\[
T_{\text{modified}} \leq E[T_{\text{modified}}] + \frac{1}{\sqrt{\delta}} \cdot \sigma
\]

Since $E[T_{\text{modified}}] = O(f(n, \delta))$ and $\sigma$ is bounded by a function of $n$ and $\delta$, we conclude that:

\[
T_{\text{modified}} = O(f(n, \delta))
\]

Therefore, with probability at least $1 - \delta$, the running time of the modified BPWM algorithm is bounded by $O(f(n, \delta))$, as desired.

\end{proof}

\subsubsection{Grok}

\subsubsection{Gemini}
\begin{proof}
    Let $G = (V, E)$ be a graph with pathwidth $k$.  Let $\mathcal{A}_{base}$ be a randomized algorithm for the maximum matching problem on $G$ that runs in time $T(n, k) = O(g(k) n^{c'})$ for some constants $c'$ and function $g(k)$, and succeeds with probability $p \ge \frac{1}{2}$. The existence of such an algorithm is assumed (and is often possible using color-coding or other randomized techniques).

    We define a modified algorithm $\mathcal{A}$ that runs $\mathcal{A}_{base}$ independently $t$ times, and returns the maximum matching found across all runs.  The total running time of $\mathcal{A}$ is then $t \cdot T(n, k) = O(t \cdot g(k) \cdot n^{c'})$.

    We need to select $t$ such that the probability of failure of $\mathcal{A}$ is at most $\delta$.  Algorithm $\mathcal{A}$ fails only if $\mathcal{A}_{base}$ fails in all $t$ independent runs. The probability that $\mathcal{A}_{base}$ fails in a single run is $1 - p$.  Thus, the probability of failure of $\mathcal{A}$ is $(1-p)^t$.

    We require that $(1-p)^t \le \delta$.  Taking the logarithm of both sides, we get $t \ln(1-p) \le \ln(\delta)$.  Since $1-p < 1$, $\ln(1-p) < 0$.  Dividing by $\ln(1-p)$, we get $t \ge \frac{\ln(\delta)}{\ln(1-p)}$. Since $p \ge 1/2$, this means $t = O(\log(1/\delta))$.

    Therefore, setting $t = \lceil \frac{\ln(1/\delta)}{\ln(1/(1-p))} \rceil$, the probability of failure of $\mathcal{A}$ is at most $\delta$.  The total running time of $\mathcal{A}$ is then $O\left(\frac{\ln(1/\delta)}{\ln(1/(1-p))} \cdot g(k) \cdot n^{c'}\right)$. Since $\delta$ is a constant, and given our assumptions about $g$ and $c'$, there exists a constant $c$ such that the running time of the modified algorithm is $O(f(k) n^c)$, where $f(k)$ includes $g(k)$ and any other factors dependent on $k$.
\end{proof}

\subsubsection{Claude}
\begin{proof}
We modify the BPWM (Backward Phase Weighted Matching) algorithm to obtain a high probability bound on its running time by introducing randomized termination criteria and analyzing the probability of exceeding certain time thresholds.

Let $G = (V, E)$ be the input graph with $n = |V|$ vertices and $m = |E|$ edges. The original BPWM algorithm performs iterations where each iteration processes matchings and updates weights. We modify the algorithm as follows:

\textbf{Modified Algorithm:} Set a threshold $T = c \cdot f(n, m) \log n$ where $c > 0$ is a constant and $f(n, m)$ is the expected running time per iteration (typically $O(m)$ or $O(m\alpha(n))$ depending on the data structures used). At each iteration $i$, perform the standard BPWM operations, but terminate if the number of iterations exceeds $T$.

\textbf{Analysis:} Let $X_i$ be the indicator random variable for whether iteration $i$ makes significant progress, defined as reducing the potential function $\Phi$ by at least $\Delta = \Phi_0 / (2T)$ where $\Phi_0$ is the initial potential. Under the standard analysis of BPWM, each iteration makes progress with probability at least $p \geq 1/2$ when progress is possible.

The expected number of iterations $\mathbb{E}[\text{iterations}]$ satisfies $\mathbb{E}[\text{iterations}] \leq K$ for some constant $K$ determined by the potential function argument. By Markov's inequality, for any $\lambda > 1$, we have
\[
\mathbb{P}[\text{iterations} > \lambda K] \leq \frac{1}{\lambda}.
\]

However, to obtain a stronger high probability bound, we use a Chernoff-style argument. Let $Y$ be the total number of iterations. Each iteration either makes progress or fails to make progress. If we model progress as independent trials (or use a martingale concentration argument for the dependent case), then for $T = cK\log n$ with sufficiently large constant $c$, we obtain
\[
\mathbb{P}[Y > T] \leq \mathbb{P}\left[Y > c K \log n\right] \leq n^{-\alpha}
\]
for some constant $\alpha > 0$ depending on $c$.

More precisely, using Chernoff bounds, if $\mathbb{E}[Y] = \mu \leq K$, then for $T = K + t$ with $t = \beta K \log n$ for constant $\beta > 0$, we have
\[
\mathbb{P}[Y \geq T] \leq \exp\left(-\frac{t^2}{2(K + t/3)}\right) \leq \exp\left(-\frac{\beta^2 K \log^2 n}{2K(1 + \beta\log n/3)}\right).
\]

For sufficiently large $n$ and appropriate choice of $\beta$, this probability is at most $n^{-2}$ or any desired polynomial bound $n^{-c'}$.

\textbf{Running Time Bound:} Since each iteration takes $O(f(n,m))$ time and we perform at most $T = O(K \log n)$ iterations with probability at least $1 - n^{-\alpha}$, the total running time is
\[
O(f(n,m) \cdot K \log n)
\]
with probability at least $1 - n^{-\alpha}$.

Therefore, the modified BPWM algorithm achieves a high probability bound of $O(f(n,m) \cdot \log n)$ on its running time, where the hidden constant in the $O(\cdot)$ notation depends on the expected number of iterations in the original algorithm and the desired failure probability.
\end{proof}

\subsection{Problem 7}

\subsubsection{Problem Statement}
\begin{theorem}
Let $A$ and $B$ be $n \times n$ matrices, and let $R \subseteq \{1, 2, \ldots, n\}$ be a subset of indices with $|R| = r$. The product of $A^R$ and $B^R$ can be computed in time $O((n/r)^2 \mathbb{M}(r))$, where $A^R$ denotes the matrix obtained by omitting the columns of $A$ corresponding to indices not in $R$, $B^R$ denotes the matrix obtained by omitting the rows of $B$ corresponding to indices not in $R$, and $\mathbb{M}(r)$ denotes the time complexity of multiplying two $r \times r$ matrices. This is achieved by multiplying the resulting $n \times r$ and $r \times n$ matrices in blocks of $r \times r$ matrices.
\end{theorem}

\subsubsection{GPT}
\begin{proof}
Let $A$ and $B$ be $n \times n$ matrices, and let $R \subseteq \{1, 2, \ldots, n\}$ with $|R| = r$. Denote by $A^R$ the matrix obtained by omitting the columns of $A$ corresponding to indices not in $R$, and by $B^R$ the matrix obtained by omitting the rows of $B$ corresponding to indices not in $R$. Thus, $A^R$ is an $n \times r$ matrix and $B^R$ is an $r \times n$ matrix.

We aim to compute the product $A^R B^R$ and determine its time complexity. The product of these matrices is defined as follows:

\[
A^R B^R = \left(\begin{array}{cccc}
a_{1R} & a_{2R} & \cdots & a_{nR}
\end{array}\right)
\left(\begin{array}{c}
b_{R1} \\
b_{R2} \\
\vdots \\
b_{Rn}
\end{array}\right)
\]

Each element of the matrix $A^R B^R$ is computed as the dot product of a row of $A^R$ and a column of $B^R$. Specifically, the $(i,j)$-th entry of $A^R B^R$ is given by:

\[
(A^R B^R)_{ij} = \sum_{k=1}^{r} a_{ik} b_{kj}.
\]

This dot product requires $r$ multiplications and $r-1$ additions for each pair of row and column. Hence, the time complexity for computing one entry of $A^R B^R$ is $O(r)$.

Since $A^R$ is an $n \times r$ matrix and $B^R$ is an $r \times n$ matrix, the total number of entries in the product matrix $A^R B^R$ is $n \times n$. Thus, to compute all entries of the matrix, we need to perform $n^2$ dot products, each taking $O(r)$ time. Therefore, the total time complexity for computing the entire matrix $A^R B^R$ is:

\[
O(n^2 r).
\]

However, this approach can be optimized by exploiting the structure of the problem. Specifically, we can partition the matrices $A^R$ and $B^R$ into blocks of size $r \times r$. Each block corresponds to a smaller matrix multiplication problem, which can be computed independently. We can compute the product of $A^R$ and $B^R$ by multiplying these blocks in parallel or sequentially, depending on the available resources.

For each block, the matrix multiplication involves two $r \times r$ matrices. The time complexity of multiplying two $r \times r$ matrices is given by $\mathbb{M}(r)$, where $\mathbb{M}(r)$ is the time complexity of multiplying two $r \times r$ matrices. Therefore, the time complexity of computing each block is $O(\mathbb{M}(r))$.

Since there are a total of $n/r$ blocks along each dimension (assuming $r$ divides $n$), the total number of blocks is $(n/r) \times (n/r) = (n/r)^2$. Thus, the total time complexity for multiplying all blocks is:

\[
O\left(\left(\frac{n}{r}\right)^2 \mathbb{M}(r)\right).
\]

Hence, the total time complexity for computing the product $A^R B^R$ is $O\left(\left(\frac{n}{r}\right)^2 \mathbb{M}(r)\right)$, which completes the proof.
\end{proof}

\subsubsection{Grok}
\begin{proof}
Without loss of generality, assume that $r$ divides $n$, so that $n = m r$ for some integer $m = n/r$. (If $r$ does not divide $n$, the rows and columns of $A$ and $B$ can be partitioned into blocks of size at most $r$, and the ensuing time bound remains $O((n/r)^2 \mathbb{M}(r))$ up to constant factors absorbed in the big-$O$ notation.)

Partition the index set $\{1, 2, \ldots, n\}$ into $m$ consecutive blocks $S_1, S_2, \ldots, S_m$, where each $S_t = \{(t-1)r + 1, (t-1)r + 2, \ldots, t r\}$ for $t = 1, 2, \ldots, m$, so that $|S_t| = r$ for each $t$.

The matrix $A^R$ is the $n \times r$ submatrix of $A$ consisting of all $n$ rows and only the columns indexed by $R$. Similarly, $B^R$ is the $r \times n$ submatrix of $B$ consisting of only the rows indexed by $R$ and all $n$ columns.

The product $C = A^R B^R$ is an $n \times n$ matrix whose $(i,j)$-entry is given by
\[
C_{ij} = \sum_{k \in R} A_{ik} B_{kj}
\]
for $i,j \in \{1, 2, \ldots, n\}$.

To compute $C$, partition its rows according to $\{S_t\}_{t=1}^m$ and its columns according to $\{S_t\}_{t=1}^m$. For each pair of blocks $S_p$ and $S_q$ with $1 \leq p,q \leq m$, the $r \times r$ submatrix $C_{S_p, S_q}$ (i.e., the submatrix of $C$ with rows indexed by $S_p$ and columns indexed by $S_q$) is given by
\[
C_{S_p, S_q} = A_{S_p, R} \, B_{R, S_q},
\]
where $A_{S_p, R}$ is the $r \times r$ submatrix of $A$ with rows $S_p$ and columns $R$, and $B_{R, S_q}$ is the $r \times r$ submatrix of $B$ with rows $R$ and columns $S_q$.

To compute $C_{S_p, S_q}$, first extract the submatrices $A_{S_p, R}$ and $B_{R, S_q}$, each of which can be done in $O(r^2)$ time by copying the relevant entries from $A$ and $B$. Then, multiply these two $r \times r$ matrices, which takes $\mathbb{M}(r)$ time.

There are $m^2$ such $r \times r$ blocks $C_{S_p, S_q}$ to compute (one for each pair $(p,q)$). For each block, the extraction and multiplication steps each take $O(r^2 + \mathbb{M}(r))$ time. Thus, the total time to compute all blocks is
\[
O(m^2 (r^2 + \mathbb{M}(r))) = O\left( \left(\frac{n}{r}\right)^2 r^2 + \left(\frac{n}{r}\right)^2 \mathbb{M}(r) \right) = O\left( n^2 + \left(\frac{n}{r}\right)^2 \mathbb{M}(r) \right).
\]
Since $\mathbb{M}(r) = \Omega(r^2)$ by the definition of matrix multiplication (as multiplying two $r \times r$ matrices requires at least reading/writing $\Theta(r^2)$ entries), the $n^2$ term is dominated by the second term when $\mathbb{M}(r) = \omega(r^2)$, but in any case, the bound simplifies to $O((n/r)^2 \mathbb{M}(r))$ as claimed, since the naive bound without blocking would be $O(n^2 r)$ for the full product, and the blocking achieves the stated improvement.

This procedure computes all entries of $C = A^R B^R$ exactly, as each block $C_{S_p, S_q}$ is computed precisely via the matrix product formula, and the blocks cover all entries of $C$ without overlap.
\end{proof}

\subsubsection{Gemini}
\begin{proof}
Let $A$ and $B$ be $n \times n$ matrices, and let $R$ be a subset of indices $\{1, \dots, n\}$ with cardinality $|R| = r$. Following the theorem statement, we consider the computation of the product $A^R B^R$ by constructing compressed matrices. Let $\tilde{A}$ denote the $n \times r$ matrix formed by the columns of $A$ with indices in $R$, and let $\tilde{B}$ denote the $r \times n$ matrix formed by the rows of $B$ with indices in $R$. The product $A^R B^R$ is equivalent to the matrix product $C = \tilde{A} \tilde{B}$, resulting in an $n \times n$ matrix.

We analyze the complexity of computing $C$ using block matrix multiplication. We partition the matrix $\tilde{A}$ into $n/r$ blocks of size $r \times r$. Let $k = n/r$. We assume without loss of generality that $n$ is a multiple of $r$; otherwise, the matrices can be padded with zeros, increasing the dimension by at most a factor of 2, which does not affect the asymptotic complexity. Since $\tilde{A}$ has dimensions $n \times r$, it forms a block column vector. We denote its blocks as $A_1, A_2, \dots, A_k$, where each $A_i$ is an $r \times r$ matrix. Thus, $\tilde{A} = \begin{pmatrix} A_1 \\ A_2 \\ \vdots \\ A_k \end{pmatrix}$.

Similarly, we partition the matrix $\tilde{B}$, which has dimensions $r \times n$, into $k$ blocks of size $r \times r$. Since $\tilde{B}$ forms a block row vector, we denote its blocks as $B_1, B_2, \dots, B_k$, where each $B_j$ is an $r \times r$ matrix. Thus, $\tilde{B} = \begin{pmatrix} B_1 & B_2 & \dots & B_k \end{pmatrix}$.

The product $C = \tilde{A} \tilde{B}$ is an $n \times n$ matrix, which can be viewed as a $k \times k$ block matrix. Let $C_{ij}$ denote the $(i, j)$-th block of $C$, which is an $r \times r$ submatrix. By the definition of block matrix multiplication, the block $C_{ij}$ is obtained by multiplying the $i$-th block row of $\tilde{A}$ by the $j$-th block column of $\tilde{B}$. Since the inner block dimension is 1 (as $\tilde{A}$ has 1 block column and $\tilde{B}$ has 1 block row), the formula for the $(i, j)$-th block is given by the single matrix product:
\[
C_{ij} = A_i B_j
\]
for all $1 \le i, j \le k$.

To compute the full matrix $C$, we must evaluate $C_{ij}$ for every pair of indices $(i, j)$ where $1 \le i \le k$ and $1 \le j \le k$. The total number of such blocks is $k^2 = (n/r)^2$. The computation of each individual block $C_{ij}$ requires the multiplication of two $r \times r$ matrices, $A_i$ and $B_j$. Let $\mathsf{MM}(r)$ denote the time complexity required to multiply two $r \times r$ matrices. Therefore, the time required to compute a single block $C_{ij}$ is $O(\mathsf{MM}(r))$.

The total time complexity $T(n, r)$ is the sum of the costs of computing all $k^2$ blocks. Thus,
\[
T(n, r) = \sum_{i=1}^{n/r} \sum_{j=1}^{n/r} O(\mathsf{MM}(r)) = \left(\frac{n}{r}\right)^2 O(\mathsf{MM}(r)) = O\left( \left(\frac{n}{r}\right)^2 \mathsf{MM}(r) \right).
\]
This confirms that the product $A^R B^R$ can be computed in time $O((n/r)^2 \mathsf{MM}(r))$ by the described method.
\end{proof}

\subsubsection{Claude}
\begin{proof}
Let $A$ and $B$ be $n \times n$ matrices, and let $R \subseteq \{1, 2, \ldots, n\}$ with $|R| = r$. We denote by $A^R$ the $n \times r$ matrix obtained by selecting only the columns of $A$ indexed by $R$, and by $B^R$ the $r \times n$ matrix obtained by selecting only the rows of $B$ indexed by $R$. Our goal is to compute the product $C = A^R B^R$, which is an $n \times n$ matrix.

We partition the computation into blocks. Divide the rows of $A^R$ into $\lceil n/r \rceil$ blocks, each containing at most $r$ consecutive rows. Similarly, divide the columns of $B^R$ into $\lceil n/r \rceil$ blocks, each containing at most $r$ consecutive columns. For simplicity of exposition, assume $r$ divides $n$; the general case follows analogously with appropriate handling of boundary blocks.

Let $k = n/r$. We partition the rows of $A^R$ into $k$ blocks: rows $\{1, \ldots, r\}$, rows $\{r+1, \ldots, 2r\}$, and so on, up to rows $\{(k-1)r + 1, \ldots, n\}$. Denote these row blocks by $A^R_1, A^R_2, \ldots, A^R_k$, where each $A^R_i$ is an $r \times r$ matrix. Similarly, partition the columns of $B^R$ into $k$ blocks: columns $\{1, \ldots, r\}$, columns $\{r+1, \ldots, 2r\}$, and so on, up to columns $\{(k-1)r + 1, \ldots, n\}$. Denote these column blocks by $B^R_1, B^R_2, \ldots, B^R_k$, where each $B^R_j$ is an $r \times r$ matrix.

The product matrix $C = A^R B^R$ can be written in block form as
\[
C = \begin{pmatrix} A^R_1 \\ A^R_2 \\ \vdots \\ A^R_k \end{pmatrix} \begin{pmatrix} B^R_1 & B^R_2 & \cdots & B^R_k \end{pmatrix}.
\]

The $(i,j)$-block of $C$, denoted $C_{ij}$, is the $r \times r$ matrix given by
\[
C_{ij} = A^R_i B^R_j.
\]

Since each $A^R_i$ and $B^R_j$ is an $r \times r$ matrix, computing $C_{ij}$ requires time $\mathbb{M}(r)$.

The total number of blocks is $k \times k = (n/r)^2$. Therefore, the total time complexity for computing all blocks is
\[
(n/r)^2 \cdot \mathbb{M}(r) = O\left((n/r)^2 \mathbb{M}(r)\right).
\]

This completes the proof that the product $A^R B^R$ can be computed in time $O((n/r)^2 \mathbb{M}(r))$ using the block multiplication strategy described.
\end{proof}

\subsection{Problem 8}

\subsubsection{Problem Statement}
\begin{theorem}
Suppose that $\text{MM}(n) = \Omega(n^{2+\epsilon})$ for some $\epsilon > 0$. Then it is possible to implement Algorithm BPWM such that its expected running time becomes $O(\text{MM}(n) \log n)$. However, this does not work for $\text{MM}(n) = O(n^2)$.
\end{theorem}

\begin{proof}[Hint]
Use the idea suggested in Problem 10.7.
\end{proof}

\subsubsection{GPT}
\begin{proof}
We are given that $\mathrm{MM}(n) = \Omega(n^{2+\epsilon})$ for some $\epsilon > 0$, and we are asked to prove that under this assumption, it is possible to implement Algorithm BPWM such that its expected running time is $O(\mathrm{MM}(n) \log n)$. Additionally, we need to show that this does not work when $\mathrm{MM}(n) = O(n^2)$.

 Part 1: Existence of an Implementation with Expected Running Time $O(\mathrm{MM}(n) \log n)$

Since $\mathrm{MM}(n) = \Omega(n^{2+\epsilon})$, by definition, there exists a constant $c > 0$ such that for all sufficiently large $n$:

\[
\mathrm{MM}(n) \geq c \cdot n^{2+\epsilon}.
\]

Now, we consider the expected running time of Algorithm BPWM. Suppose the algorithm uses matrix multiplication as a subroutine. Let the cost of a matrix multiplication for two matrices of size $n \times n$ be $\mathrm{MM}(n)$. The algorithm BPWM is designed such that its expected running time depends on the number of matrix multiplications it performs.

Let the number of matrix multiplications performed by Algorithm BPWM be denoted by $k$. Then, the total expected running time $T(n)$ of the algorithm is the product of the number of matrix multiplications and the cost of each multiplication, i.e.,

\[
T(n) = k \cdot \mathrm{MM}(n).
\]

We need to show that $T(n) = O(\mathrm{MM}(n) \log n)$. From the assumption that $\mathrm{MM}(n) = \Omega(n^{2+\epsilon})$, we have

\[
T(n) = k \cdot \mathrm{MM}(n) = O(\mathrm{MM}(n) \log n).
\]

Thus, it remains to show that $k = O(\log n)$. If the algorithm BPWM performs at most $O(\log n)$ matrix multiplications, the expected running time becomes

\[
T(n) = O(\log n) \cdot \mathrm{MM}(n).
\]

Since $\mathrm{MM}(n) = \Omega(n^{2+\epsilon})$, the expected running time is

\[
T(n) = O(\log n) \cdot \Omega(n^{2+\epsilon}) = O(\mathrm{MM}(n) \log n).
\]

Therefore, we have established that it is possible to implement Algorithm BPWM such that its expected running time is $O(\mathrm{MM}(n) \log n)$.

 Part 2: Breakdown when $\mathrm{MM}(n) = O(n^2)$

Now, we show that the result does not hold when $\mathrm{MM}(n) = O(n^2)$. In this case, we assume that there exists a constant $c' > 0$ such that for all sufficiently large $n$,

\[
\mathrm{MM}(n) \leq c' \cdot n^2.
\]

In this scenario, the expected running time of Algorithm BPWM would be

\[
T(n) = k \cdot \mathrm{MM}(n) \leq k \cdot c' \cdot n^2.
\]

If the algorithm performs $k$ matrix multiplications, we need $k = O(\log n)$ for the expected running time to be $O(\mathrm{MM}(n) \log n)$. However, since $\mathrm{MM}(n) = O(n^2)$, we have

\[
T(n) = O(\log n) \cdot O(n^2) = O(n^2 \log n).
\]

This is not the desired running time of $O(\mathrm{MM}(n) \log n)$. Hence, when $\mathrm{MM}(n) = O(n^2)$, the expected running time of Algorithm BPWM cannot be $O(\mathrm{MM}(n) \log n)$.

Thus, the theorem is proven.
\end{proof}

\subsubsection{Grok}

\subsubsection{Gemini}
\begin{proof}
Let $T(n)$ denote the expected running time of Algorithm BPWM on an input of parameter size $n$. The algorithm proceeds using a divide-and-conquer approach with a recursion depth of $D = O(\log n)$. At each level $k$ of the recursion tree, where $0 \le k < D$, the algorithm performs a sequence of matrix multiplications and a set of auxiliary operations (such as additions, scalar multiplications, and data permutations).

Let $C_{\textsc{MM}}^{(k)}$ represent the total cost of matrix multiplications performed at level $k$, and let $C_{\text{aux}}^{(k)}$ represent the cost of auxiliary operations at level $k$. The total expected running time is given by the summation over all levels:
\[
T(n) = \sum_{k=0}^{O(\log n)} \left( C_{\textsc{MM}}^{(k)} + C_{\text{aux}}^{(k)} \right).
\]
By the design of Algorithm BPWM, the problem is partitioned such that the matrix multiplication operations at any level $k$ correspond to disjoint sub-problems that sum to the original problem size. Due to the super-additivity of the matrix multiplication complexity function $\textsc{MM}(n)$ (assuming $\textsc{MM}(n) = \Omega(n^2)$), the total cost of matrix multiplications at any depth $k$ is bounded by the cost of a single matrix multiplication of size $n$:
\[
C_{\textsc{MM}}^{(k)} = O(\textsc{MM}(n)).
\]
The auxiliary operations at level $k$ involve linear-time processing of the data relative to the input size. Since the input involves matrices of dimension $n \times n$, the data size is $O(n^2)$. Thus, the auxiliary cost at each level is:
\[
C_{\text{aux}}^{(k)} = O(n^2).
\]
We invoke the hypothesis that $\textsc{MM}(n) = \Omega(n^{2+\varepsilon})$ for some $\varepsilon > 0$. This asymptotic lower bound implies that:
\[
\lim_{n \to \infty} \frac{n^2}{\textsc{MM}(n)} = 0 \implies n^2 = o(\textsc{MM}(n)).
\]
Consequently, the auxiliary cost $C_{\text{aux}}^{(k)}$ is asymptotically negligible compared to the matrix multiplication cost $C_{\textsc{MM}}^{(k)}$. We can therefore bound the total work at level $k$ as:
\[
C_{\textsc{MM}}^{(k)} + C_{\text{aux}}^{(k)} = O(\textsc{MM}(n)) + o(\textsc{MM}(n)) = O(\textsc{MM}(n)).
\]
Substituting this estimate into the expression for $T(n)$, we obtain:
\[
T(n) = \sum_{k=0}^{O(\log n)} O(\textsc{MM}(n)) = O(\textsc{MM}(n) \log n).
\]
This proves the first part of the theorem.

Consider now the case where $\textsc{MM}(n) = O(n^2)$. Since $\textsc{MM}(n)$ must be at least $\Omega(n^2)$ to read the input, this implies $\textsc{MM}(n) = \Theta(n^2)$. In this regime, the assumption $\varepsilon > 0$ does not hold. The costs at level $k$ become:
\[
C_{\textsc{MM}}^{(k)} = \Theta(n^2) \quad \text{and} \quad C_{\text{aux}}^{(k)} = \Theta(n^2).
\]
Here, the auxiliary operations are of the same asymptotic order as the matrix multiplications. The term $C_{\text{aux}}^{(k)}$ is no longer negligible and cannot be absorbed into $O(\textsc{MM}(n))$ as a lower-order term without changing the constant factors significantly. The strict domination required to implement the algorithm with overheads hidden purely within the $O(\textsc{MM}(n))$ bound fails, as the overheads become a bottleneck of the same magnitude as the core operations. Thus, the specific improvement relying on the gap between $n^2$ and $n^{2+\varepsilon}$ is not applicable.
\end{proof}

\subsubsection{Claude}
\begin{proof}
We prove that when $\text{MM}(n) = \Omega(n^{2+\epsilon})$ for some $\epsilon > 0$, Algorithm BPWM can be implemented with expected running time $O(\text{MM}(n) \log n)$, but this improvement fails when $\text{MM}(n) = O(n^2)$.

First, we establish the improved algorithm for the case $\text{MM}(n) = \Omega(n^{2+\epsilon})$. The key idea from Problem 10.7 is to use a probabilistic verification scheme that checks witness matrix candidates with high probability while avoiding the full $O(n^2)$ verification cost per iteration.

Let $A$ and $B$ be $n \times n$ Boolean matrices, and suppose we wish to verify whether a candidate matrix $C$ equals $AB$. The standard verification requires checking all $n^2$ entries, costing $\Theta(n^2)$ time. Instead, we employ randomized verification: choose a random vector $\mathbf{r} \in \{0,1\}^n$ uniformly at random and compute $A(B\mathbf{r})$ and $C\mathbf{r}$. If $C = AB$, then $A(B\mathbf{r}) = C\mathbf{r}$ with certainty. If $C \neq AB$, then by the Freivald-style analysis, $\mathbb{P}(A(B\mathbf{r}) = C\mathbf{r}) \leq 1/2$.

Computing $B\mathbf{r}$ requires $O(n^2)$ time, computing $A(B\mathbf{r})$ requires another $O(n^2)$ time, and computing $C\mathbf{r}$ requires $O(n^2)$ time. Thus each verification attempt costs $O(n^2)$.

Now we analyze Algorithm BPWM with this verification scheme. At each iteration, we sample a random witness $W$ using one matrix multiplication (cost $\text{MM}(n)$), then verify whether $C = AB$ using the randomized check (cost $O(n^2)$). If the verification passes, we output $C$; otherwise, we repeat.

Let $k$ denote the expected number of iterations until finding a correct witness. Each iteration costs $\text{MM}(n) + O(n^2)$. Since $\text{MM}(n) = \Omega(n^{2+\epsilon})$, there exists a constant $c > 0$ such that $\text{MM}(n) \geq cn^{2+\epsilon}$ for sufficiently large $n$. Therefore, $O(n^2) = o(\text{MM}(n))$, and each iteration costs $\Theta(\text{MM}(n))$.

The probability that a random witness is correct is at least some constant $p > 0$ (this depends on the problem structure). Combined with the verification error probability of at most $1/2$ per false candidate, we can amplify correctness by repeating the randomized verification $O(\log n)$ times per candidate, reducing the error probability to at most $1/n^c$ for any desired constant $c$. This adds only $O(n^2 \log n)$ cost per iteration, which remains $o(\text{MM}(n))$ when $\text{MM}(n) = \Omega(n^{2+\epsilon})$.

The expected number of iterations is $O(1)$ (constant, assuming constant witness success probability), so the total expected running time becomes
\[
O(1) \cdot \text{MM}(n) + O(1) \cdot O(n^2 \log n) = O(\text{MM}(n)).
\]

To achieve $O(\text{MM}(n) \log n)$ bound as stated, we use a more refined analysis where the algorithm performs $O(\log n)$ witness samplings with verification to achieve high probability of success (probability $1 - 1/n^c$). This gives expected time $O(\text{MM}(n) \log n)$.

Now we show why this fails when $\text{MM}(n) = O(n^2)$. In this case, $\text{MM}(n) = \Theta(n^2)$, so the verification cost $O(n^2)$ per iteration is $\Theta(\text{MM}(n))$, not $o(\text{MM}(n))$. Thus each iteration costs $\text{MM}(n) + O(n^2) = \Theta(\text{MM}(n))$, giving no asymptotic improvement. The total expected time becomes $O(\text{MM}(n) \log n)$ only in name, but this does not represent a genuine speedup over the naive $O(\text{MM}(n))$ algorithm without verification overhead, since the verification dominates and prevents us from improving the leading constant or exploiting the faster matrix multiplication in a meaningful way. Moreover, the randomized verification provides no computational advantage when its cost is comparable to matrix multiplication itself.
\end{proof}

\subsection{Problem 9}

\subsubsection{Problem Statement}
\begin{theorem}
Let $G(V,E)$ be a multigraph. There exists a data structure that processes any arbitrary sequence of edge contractions in $G$, such that at any given point where the set of edges contracted is $F$, the graph $G/F$ is available in the adjacency matrix format. Furthermore, it is possible to efficiently determine for any edge in $E/F$ the corresponding edge in $E$. The data structure requires $O(n)$ time per contraction and uses a polynomial amount of space.

This data structure can be modified to provide the adjacency list format for $G/F$ using only $O(m)$ space.
\end{theorem}

\begin{remark}
Note that the time bound is independent of the number of edges. For this, the multigraph needs to be represented as a graph with integer edge weights that represent the multiplicities of the edges. We may assume that the number of edges in the multigraph is polynomial in $n$, although this is not strictly necessary.
\end{remark}

\subsubsection{GPT}
\begin{proof}
Let $G(V, E)$ be a multigraph, where $V$ is the set of vertices and $E$ is the set of edges. Each edge $e \in E$ may have multiplicities, meaning that the graph is a multigraph and each edge can appear multiple times between the same pair of vertices. We are tasked with constructing a data structure that supports an arbitrary sequence of edge contractions, such that at each point where a set of edges $F \subseteq E$ has been contracted, the graph $G/F$ is available in adjacency matrix format, and for any edge in $E/F$, the corresponding edge in $E$ can be efficiently identified. Additionally, the data structure should process each contraction in $O(n)$ time, where $n$ is the number of vertices in $G$, and it should require polynomial space in $n$. Furthermore, we aim to modify the structure to provide the adjacency list format for $G/F$ with only $O(m)$ space, where $m$ is the number of edges in $G/F$.

We first begin by discussing the edge contraction process. An edge contraction in a graph $G$ involves merging two vertices $u$ and $v$ into a single vertex $w$, where all edges between $u$ and $v$ are replaced by edges incident to $w$. The multiplicity of the edge between $u$ and $v$ is preserved as the multiplicity of the new edges incident to $w$.

 Step 1: Representation of the graph
We assume that the graph $G$ is represented with integer weights on the edges, where the weight represents the multiplicity of the edges between any two vertices. The adjacency matrix $A$ of the graph $G$ has entries $A[u,v]$ representing the multiplicity of the edge between vertices $u$ and $v$. Similarly, we can define an adjacency list representation of $G$ where each vertex has a list of edges connected to it, each with the associated multiplicity.

 Step 2: Data structure for edge contractions
We now design a data structure that efficiently supports edge contractions and maintains the graph $G/F$ after each contraction. The key idea is to represent the graph $G/F$ in a way that allows for efficient access to both the adjacency matrix and adjacency list formats.

We use a union-find data structure (also called disjoint-set) to manage the connected components during edge contractions. Each vertex initially belongs to its own component, and each edge contraction merges two components. To handle the multiplicities, we maintain an additional structure that tracks the multiplicity of edges between each pair of components. This allows us to efficiently update the multiplicities during each edge contraction.

The adjacency matrix of the contracted graph $G/F$ can be obtained by updating the adjacency matrix of $G$ according to the union-find structure. Specifically, for each contraction, we update the adjacency matrix by replacing the entries corresponding to the merged vertices with the new vertex and adjusting the multiplicities of the edges between the resulting components.

 Step 3: Time Complexity per Contraction
Each edge contraction in the union-find data structure takes $O(\alpha(n))$ time, where $\alpha(n)$ is the inverse Ackermann function, which grows very slowly and is practically constant for all reasonable values of $n$. After each contraction, we update the adjacency matrix in $O(n)$ time. Since each contraction operation takes $O(n)$ time to update the adjacency matrix and the union-find operations are very efficient, the total time per contraction is $O(n)$.

 Step 4: Space Complexity
The space required for the data structure is polynomial in $n$, as it must store the union-find structure, the adjacency matrix, and the edge multiplicities. Specifically, the union-find structure requires $O(n)$ space, the adjacency matrix requires $O(n^2)$ space, and the multiplicity information for each edge requires $O(m)$ space, where $m$ is the number of edges in the graph.

In the modified version, where we use the adjacency list format for $G/F$, we store the adjacency list of the contracted graph. Each vertex in the contracted graph has a list of adjacent vertices, and each list entry stores the multiplicity of the edge. Since the number of edges in $G/F$ is $O(m)$, the adjacency list format requires $O(m)$ space, satisfying the space bound.

 Step 5: Efficient Edge Mapping
To efficiently determine the corresponding edge in $E$ for any edge in $E/F$, we maintain a mapping between the edges in $G$ and the edges in $G/F$. This mapping is updated during each contraction by recording which edges in $G$ correspond to the edges in $G/F$. This mapping allows us to efficiently identify the corresponding edge in $E$ after any contraction.

Thus, the data structure supports all required operations with the desired time and space complexities. The space complexity is polynomial in $n$, and the time complexity per contraction is $O(n)$, as required.

\end{proof}

\subsubsection{Grok}

\subsubsection{Gemini}
<think>**Examining Problem Parameters**

I'm focused on the edge contraction problem, and I've zeroed in on the formal requirements. I've broken down the prompt's objectives, specifically the need for a data structure for edge contractions in a multigraph. My attention is now on pinpointing the data structures that can handle multigraph edge contractions.

**Mapping Structures to Solutions**

The core of my approach is now centered on crafting the data structure. The goal is a mapping from the contracted multigraph edges back to the originals, all while adhering to the specified time complexity constraints, notably the $O(n)$ time per contraction. I am actively trying to understand and test this.

**Constructing Data Structures**

I'm now implementing the `id[]` array alongside an adjacency matrix to maintain the edges. When contracting (u, v), I'm effectively renaming v to u and updating the adjacency matrix by summing the corresponding rows and columns for u and v. After this, all references to $v$ in the matrix now point to $u$.

**Refining the Adjacency Matrix**

My thought process is now focused on optimizing the updates to the adjacency matrix. I'm considering carefully the cost of row and column updates during each edge contraction to satisfy the time complexity requirement of $O(n)$ per contraction. My current approach suggests only performing updates on the relevant rows and columns. I am investigating the implications of a specific row and column modification.

**Implementing the Adjacency Matrix**

I am now focusing on implementing the adjacency matrix update process. The key challenge lies in precisely how to merge rows and columns in the adjacency matrix during each contraction while maintaining the $O(n)$ time complexity per contraction. This involves careful consideration of the matrix updates and how to track the mapping of original edges to those in the contracted graph.

**Formalizing the Edge Mapping**

My current focus is on rigorously defining how to map edges after contraction, while trying to meet the requirements of the original edges mapped to the new ones in the contracted graph. After a re-examination, the prompt indicates that finding representatives for the original endpoints is sufficient, while the time constraint is met with a Union-Find structure.

**Synthesizing Matrix Operations**

My focus has shifted towards the precise matrix operations. I'm now certain that summing rows and columns during an edge contraction maintains the adjacency information correctly in $O(n)$ time. Furthermore, the Union-Find structure accurately maps the original edges to their representative vertices within the contracted graph, efficiently addressing the mapping requirements of the prompt.

**Formulating Proof Strategy**

I'm now fully immersed in the creation of the formal proof. I'm solidifying the data structures, like the adjacency matrix and the Union-Find structure, and outlining the algorithms for the contraction and edge mapping processes, and am now prepared to articulate the proof.

**Synthesizing Matrix Proof**

I am now proceeding to construct a formal proof based on the analysis. The data structure will include an adjacency matrix and a Union-Find structure, alongside an implementation of the edge contraction algorithm, ensuring a time complexity of $O(n)$ per contraction. My aim is to articulate each step with mathematical rigor.

**Refining the Formal Proof**

I am now focusing on the structure and logical flow of the proof. I will use $M$ (the adjacency matrix), $P$ (the Union-Find structure for mapping edges), and the edge contraction algorithm to precisely address the problem constraints. I will ensure clarity and mathematical rigor in each step, with a focus on edge mapping.

**Formulating the Proof**

I am now structuring the formal proof. I'm starting by defining the key data structures: the adjacency matrix $M$, the Union-Find structure represented by the representative array `R`, and the `Active` boolean array. The algorithm is now being written to meet the $O(n)$ time per contraction requirement.

**Formulating the Adjacency Matrix Proof**

The strategy involves the adjacency matrix $M$, representative array $R$, and an edge contraction algorithm. I'm focusing now on precisely describing the step-by-step procedure of how to update $M$ and $R$ during each contraction while ensuring $O(n)$ time complexity per contraction. The formal proof will be based on these steps.

**Formalizing the Edge Mapping**

My current approach revolves around how to map the original edges to their representative vertices within the contracted graph, precisely meeting the prompt requirements. I am now certain the adjacency matrix effectively represents the contracted multigraph and the Union-Find structure accurately maps the original edges to the new endpoints.

**Analyzing the Data Structure**

My focus is shifting to how to present the data structure and algorithms formally. The goal is to show that a matrix and a Union-Find structure satisfy all requirements. I'm focusing on rigorously defining each component, like the adjacency matrix and the process of updating the entries.

**Formulating the Adjacency Matrix Proof**

The plan is to present the proof in two parts: the $O(n)$ time and $O(n^2)$ space for the adjacency matrix, and the modification for the adjacency list. I am focusing on showing the correctness of the adjacency matrix operations, and the time complexity analysis during an edge contraction.

**Structuring the Proof's Framework**

I'm starting the formal proof, and I'm detailing the key aspects, beginning with the data structure definitions, the algorithm for edge contraction, and the method for mapping edges in the contracted graph. The proof aims to rigorously show the algorithm's correctness, with particular attention to time complexity.

**Constructing Proof Components**

I am now beginning to articulate the proof. I am meticulously preparing the matrix and Union-Find structure definitions to satisfy the given prompt. I'm focusing on ensuring that the proof is logically sound and mathematically precise.

**Synthesizing Proof Components**

I am now focusing on the initial matrix definition and the core algorithm for edge contractions. My goal is to describe each step of the contraction algorithm precisely. The formal proof will depend on the correct execution, which is crucial to meeting the time constraints.

**Formulating the Adjacency Matrix Proof**

The plan is to present the formal proof, focusing on the data structure: an adjacency matrix $M$, representative array $R$, and an edge contraction algorithm with $O(n)$ time complexity. I will break down the algorithm step-by-step to demonstrate its correctness and efficiency and clarify edge mapping.

**Refining Proof Steps**

I'm now fully immersed in the rigorous proof construction, with precise definitions for each component, including the adjacency matrix and the representative array. My aim is to clearly illustrate the correctness and efficiency of the edge contraction algorithm. I am carefully writing the steps involved.

</think>\begin{proof}
We present a data structure that maintains the multigraph $G=(V, E)$ under edge contractions, supporting the adjacency matrix representation and edge mapping in $O(n)$ time per contraction, and subsequently modify it for the adjacency list format.

\textbf{Part 1: Adjacency Matrix Representation}

\textit{Data Structure Definition:}
Let $V = \{1, \dots, n\}$. We maintain the following components:
1.  An $n \times n$ symmetric matrix $A$ with integer entries. $A_{ij}$ represents the number of edges (multiplicity) between the connected component indexed by $i$ and the connected component indexed by $j$.
2.  An array $L$ of size $n$, where $L[v] \in \{1, \dots, n\}$ stores the index of the row/column in $A$ representing the component containing vertex $v$.
3.  A boolean array $S$ of size $n$, indicating whether index $i$ is currently active (represents a valid component).

\textit{Initialization:}
Initially, for the input graph $G_0$, set $A$ to be the adjacency matrix of $G_0$. Set $L[v] = v$ for all $v \in V$. Set $S[i] = \text{true}$ for all $i$. The space complexity is dominated by the matrix, requiring $O(n^2)$ space, which is polynomial in $n$.

\textit{Contraction Operation:}
To contract an edge $e = (u, v) \in E$:
1.  Retrieve the current component indices: $i = L[u]$ and $j = L[v]$.
2.  If $i = j$, the edge is internal to a component (a self-loop in $G/F$). No structural change to the adjacency between distinct components is required.
3.  If $i \neq j$, we merge component $j$ into component $i$:
    \begin{itemize}
        \item \textbf{Matrix Update:} We update row $i$ and column $i$ to include the edges incident to $j$. For all $k \in \{1, \dots, n\}$:
        \[
        A_{ik} \leftarrow A_{ik} + A_{jk}
        \]
        \[
        A_{ki} \leftarrow A_{ki} + A_{kj}
        \]
        Note that the diagonal entry $A_{ii}$ is updated to $A_{ii} + A_{jj} + 2A_{ij}$ (assuming $A$ is symmetric and initialized with $A_{ij}=A_{ji}$), which correctly accounts for the self-loops accumulated from both components and the edges between them.
        \item \textbf{Deactivation:} Set row $j$ and column $j$ to zero (optional for correctness but good for hygiene) and set $S[j] = \text{false}$.
        \item \textbf{Index Update:} We must update the mapping $L$ for all vertices currently in component $j$. Iterate through all $x \in V$:
        \[
        \text{if } L[x] = j, \text{ then } L[x] \leftarrow i
        \]
    \end{itemize}

\textit{Time Complexity Analysis:}
The matrix update involves iterating $k$ from $1$ to $n$, performing constant time arithmetic operations. This takes $O(n)$ time. The index update involves iterating $x$ from $1$ to $n$, performing constant time checks and assignments. This takes $O(n)$ time. Thus, the total time per contraction is $O(n)$. This is independent of the number of edges $m$.

\textit{Edge Mapping:}
The problem requires determining for any edge in $E/F$ the corresponding edge in $E$. Since $E/F \subseteq E$, the edges preserve their original identities. The query effectively asks for the current endpoints in $G/F$ of an original edge $e_{orig} = (u, v) \in E$.
Using the array $L$, the current endpoints are simply $(L[u], L[v])$. This lookup takes $O(1)$ time.

\textbf{Part 2: Adjacency List Modification ($O(m)$ space)}

To provide the adjacency list format using only $O(m)$ space, we modify the structure as follows:

\textit{Data Structure:}
1.  We maintain the vertices using a Disjoint-Set Union (DSU) data structure. Let $\text{Find}(v)$ return the canonical representative of the set containing $v$.
2.  We store the edges in an array of Doubly Linked Lists. Let $\text{Adj}[i]$ be the head of the list of edges incident to vertex $i$ in the original graph.
3.  Each edge object in the list stores its original endpoints $(u, v)$ and pointers to the next/prev edge in the list.

\textit{Contraction Operation:}
To contract $u$ and $v$:
1.  Let $r_u = \text{Find}(u)$ and $r_v = \text{Find}(v)$.
2.  If $r_u \neq r_v$, perform a union operation to make $r_u$ the parent of $r_v$.
3.  Splice the adjacency list of $r_v$ into the adjacency list of $r_u$. This is an $O(1)$ operation with doubly linked lists (append $\text{Adj}[r_v]$ to the tail of $\text{Adj}[r_u]$).

\textit{Adjacency List Access:}
The adjacency list for a super-vertex $R$ is available via $\text{Adj}[R]$. Note that this list contains edges $(x, y)$ where $\text{Find}(x) = \text{Find}(y) = R$ (self-loops) and edges where one endpoint is in a different component. When traversing $\text{Adj}[R]$ to find neighbors in $G/F$, for each edge $(x, y)$ in the list, the effective neighbor is $\text{Find}(y)$ (assuming $x$ is in $R$'s component).

\textit{Complexity:}
The space complexity is $O(n + m)$ to store the DSU and the linked lists nodes, satisfying the $O(m)$ requirement (assuming $m \ge n$). The contraction involves DSU operations and list pointer manipulation. While standard DSU is nearly constant, strict $O(1)$ merging is possible if we simply link the lists. The "availability" of the adjacency list is immediate.

Thus, we have constructed the required data structures.
\end{proof}

\subsubsection{Claude}
\begin{proof}
We construct a data structure that maintains the contracted graph $G/F$ under a sequence of edge contractions, where $F \subseteq E$ denotes the set of edges contracted so far.

\textbf{Data Structure Components.} We maintain the following:

(1) A partition data structure (union-find) on the vertex set $V$ that tracks which original vertices have been merged together. For each equivalence class (supervertex), we maintain a representative vertex.

(2) An adjacency matrix $A$ of size $n \times n$ where $A[i,j]$ stores the total weight of edges between the supervertices represented by $i$ and $j$ in the current contracted graph $G/F$.

(3) For each pair of representatives $(i,j)$, we maintain a mapping $\phi_{i,j}: E \to \mathbb{Z}_{\geq 0}$ (implemented as a dictionary or hash table) that records, for each original edge $e \in E$, its contribution to the weight $A[i,j]$. This allows us to determine which original edges correspond to any edge in $G/F$.

\textbf{Initialization.} Initially, $F = \emptyset$ and each vertex is in its own equivalence class. For each edge $e = \{u,v\} \in E$ with multiplicity $w_e$, we set $A[\text{rep}(u), \text{rep}(v)] = w_e$ and $\phi_{\text{rep}(u), \text{rep}(v)}(e) = w_e$, where $\text{rep}(u)$ denotes the representative of $u$.

\textbf{Edge Contraction Operation.} To contract an edge $e = \{u,v\} \in E \setminus F$:

Let $r_u = \text{rep}(u)$ and $r_v = \text{rep}(v)$ be the current representatives of the supervertices containing $u$ and $v$. Without loss of generality, assume we merge $r_v$ into $r_u$, so $r_u$ becomes the representative of the merged supervertex.

For each representative $r \in V$ with $r \neq r_u, r_v$:

We update $A[r_u, r] \leftarrow A[r_u, r] + A[r_v, r]$ and $A[r, r_u] \leftarrow A[r, r_u] + A[r, r_v]$.

For each original edge $e' \in E$ with $\phi_{r_v, r}(e') > 0$, we set $\phi_{r_u, r}(e') \leftarrow \phi_{r_u, r}(e') + \phi_{r_v, r}(e')$ and similarly for $\phi_{r, r_u}$ and $\phi_{r, r_v}$.

We also update self-loops: $A[r_u, r_u] \leftarrow A[r_u, r_u] + A[r_v, r_v] + A[r_u, r_v]$.

For each original edge $e'$ contributing to $A[r_v, r_v]$ or $A[r_u, r_v]$, we update $\phi_{r_u, r_u}$ accordingly.

We set $A[r_v, r] = A[r, r_v] = 0$ for all $r$ and clear the mappings $\phi_{r_v, r}$ and $\phi_{r, r_v}$.

We perform $\text{union}(r_u, r_v)$ in the union-find structure, ensuring $r_u$ remains the representative.

\textbf{Time Complexity per Contraction.} Each contraction processes at most $n$ representatives (one for each original vertex). For each representative $r$, we update a constant number of matrix entries and perform dictionary operations. Using path compression and union by rank, union-find operations take amortized $O(\alpha(n))$ time per operation, where $\alpha$ is the inverse Ackermann function. The dominant cost is iterating over at most $n$ representatives and updating the adjacency matrix entries, which takes $O(n)$ time. Since we represent the multigraph using integer weights, we avoid iterating over individual edge multiplicities.

\textbf{Space Complexity.} The adjacency matrix requires $O(n^2)$ space. The union-find structure requires $O(n)$ space. The mappings $\phi_{i,j}$ collectively store at most $O(n^2 \cdot |E|)$ entries in the worst case, but since the number of edges $|E|$ is polynomial in $n$, the total space is polynomial.

\textbf{Querying Original Edges.} For any edge $\{r_u, r_v\}$ in $G/F$ with weight $A[r_u, r_v] > 0$, we can retrieve all original edges contributing to it by examining $\phi_{r_u, r_v}$, which gives us $\{e \in E : \phi_{r_u, r_v}(e) > 0\}$. This allows efficient determination of the correspondence between edges in $E/F$ and edges in $E$.

\textbf{Adjacency List Modification.} To provide adjacency list format using $O(m)$ space, we replace the $n \times n$ adjacency matrix with an adjacency list representation. For each representative $r_u$, we maintain a list of pairs $(r_v, w)$ where $w$ is the weight of edges between the supervertices represented by $r_u$ and $r_v$. During contraction, when merging $r_v$ into $r_u$, we iterate through the adjacency list of $r_v$ and merge it with the adjacency list of $r_u$. Since the total number of edges in $G/F$ at any point is at most $m$ (the original number of edges in $G$), and each edge appears in the adjacency lists of its two endpoints, the total space is $O(m)$. The time per contraction remains $O(n)$ since we iterate over the adjacency list of $r_v$, which in the worst case has $O(n)$ entries.

Therefore, the data structure supports edge contractions in $O(n)$ time per operation, maintains $G/F$ in the desired format, allows efficient determination of original edges, and uses polynomial (or $O(m)$ for adjacency list) space.
\end{proof}

\subsection{Problem 10}

\subsubsection{Problem Statement}
\begin{theorem}
Given a multigraph $G(V, E)$, an edge can be selected uniformly at random from $E$ in time $O(n)$, given access to a source of random bits.
\end{theorem}

\subsubsection{GPT}
\begin{proof}
Let $G=(V,E)$ be a finite multigraph with $V=\{v_1,\dots,v_n\}$ and degrees $d_i:=\deg(v_i)$, so that $\sum_{i=1}^n d_i=2|E|$. Consider the following randomized procedure that uses only $O(n)$ time on $G$.

First, compute the prefix sums $s_i:=\sum_{j=1}^i d_j$ for $i=1,\dots,n$ and set $S:=s_n=\sum_{i=1}^n d_i$. This requires time $O(n)$. Using the source of random bits, draw an integer $U$ uniformly from $\{1,2,\dots,S\}$. Let $i^\ast$ be the unique index with $s_{i^\ast-1}<U\le s_{i^\ast}$, where $s_0:=0$. In time $O(n)$ one may find $i^\ast$ by a linear scan over the prefix sums. Conditioned on $i^\ast$, the value $U$ is uniformly distributed over a subset of $\{s_{i^\ast-1}+1,\dots,s_{i^\ast}\}$ of cardinality $d_{i^\ast}$. Independently draw an integer $K$ uniformly from $\{1,2,\dots,d_{i^\ast}\}$ and output the $K$-th edge in the adjacency list of $v_{i^\ast}$; denote this oriented incidence by $(v_{i^\ast},e)$. Finally, return the underlying unoriented edge $e\in E$.

We first prove that the returned edge is uniformly distributed over $E$. The construction above selects an \emph{oriented} incidence $(v,e)$ with probability
\[
\mathbb{P}\big((v_{i^\ast},e)\big)=\mathbb{P}(i^\ast=i)\cdot \mathbb{P}\big((v_i,e)\mid i^\ast=i\big)=\frac{d_i}{S}\cdot \frac{1}{d_i}=\frac{1}{S}
\]
for every vertex $v_i$ and every edge $e$ incident to $v_i$. Hence every oriented incidence is selected with the same probability $1/S$. In a multigraph each (unoriented) edge $e=\{x,y\}$ contributes exactly two oriented incidences, namely $(x,e)$ and $(y,e)$. Therefore
\[
\mathbb{P}(\text{output }e)=\mathbb{P}\big((x,e)\big)+\mathbb{P}\big((y,e)\big)=\frac{1}{S}+\frac{1}{S}=\frac{2}{S}.
\]
Using $S=\sum_{i=1}^n d_i=2|E|$, it follows that $\mathbb{P}(\text{output }e)=1/|E|$ for every $e\in E$, i.e., the output edge is uniform on $E$.

It remains to justify the running time. The computation of all $s_i$ and $S$ takes $O(n)$ time. The selection of $U$ and $K$ uses $O(1)$ random bits and arithmetic. Locating $i^\ast$ by scanning the $n$ prefix sums takes $O(n)$ time, and accessing the $K$-th incident edge of $v_{i^\ast}$ is $O(1)$ under standard adjacency-array representations. Thus the total time is $O(n)$.

Combining uniformity and the time bound, we obtain an $O(n)$-time algorithm that, given $G$ and access to random bits, samples an edge uniformly at random from $E$. This proves the theorem.
\end{proof}

\subsubsection{Grok}
\begin{proof}
Let $G = (V, E)$ be an undirected multigraph with vertex set $V = \{v_1, \dots, v_n\}$ and edge multiset $E$, where $n = |V|$. Assume $G$ is represented via adjacency lists: for each $v_i \in V$, $\mathrm{adj}(v_i)$ is an array of length $d_i := \deg(v_i)$ containing the other endpoint of each edge incident to $v_i$ (with each undirected edge $\{u, v\}$ appearing once in $\mathrm{adj}(u)$ as $v$ and once in $\mathrm{adj}(v)$ as $u$; parallel edges appear multiple times accordingly; self-loops at $v_i$ appear twice in $\mathrm{adj}(v_i)$). Thus, $\sum_{i=1}^n d_i = 2m$, where $m = |E|$.

The algorithm proceeds as follows:

1. For $i = 1$ to $n$, compute $d_i = |\mathrm{adj}(v_i)|$. This takes $O(1)$ time per vertex (array length query), for total $O(n)$ time.

2. Compute $s = \sum_{i=1}^n d_i = 2m$. This takes $O(n)$ time.

3. Generate a uniform random integer $R \in \{0, 1, \dots, s-1\}$ using the random bit source. This requires $O(\log s) = O(\log n)$ time, as $s \leq n(n-1) + n = O(n^2)$ for simple multigraphs without bound on multiplicity (but $\log s = O(\log n)$ in any case, up to polynomial multiplicity).

4. Initialize $\sigma = 0$. For $i = 1$ to $n$:
   \[
   \sigma \leftarrow \sigma + d_i.
   \]
   If $R < \sigma$, select vertex $v = v_i$ and break. This loop runs for at most $n$ iterations, each $O(1)$ time, for total $O(n)$ time in the worst case.

The selected $v$ satisfies $\Pr[v = v_i] = d_i / s = \deg(v_i)/(2m)$.

5. Generate a uniform random integer $J \in \{0, 1, \dots, d-1\}$, where $d = \deg(v)$, using $O(\log d) = O(\log n)$ random bits.

6. Let $u = \mathrm{adj}(v)[J]$, accessed in $O(1)$ time. Output the undirected edge $\{u, v\}$.

The total running time is $O(n)$.

To verify uniformity, consider an arbitrary edge $e = \{x, y\} \in E$ (if $x = y$, it is a self-loop). The edge $e$ is incident to $x$ and to $y$.

The probability that $e$ is selected is the sum of the probabilities over the two possible starting vertices:

- Case $v = x$: $\Pr[v = x] = \deg(x)/(2m)$. Given $v = x$, the index $J$ selects the entry for $e$ with probability $1/\deg(x)$, as $\mathrm{adj}(x)$ contains exactly one entry for $e$ (the second appearance for self-loops is handled symmetrically). Thus, contribution $\frac{\deg(x)}{2m} \cdot \frac{1}{\deg(x)} = \frac{1}{2m}$.

- Case $v = y$: similarly, contribution $\frac{1}{2m}$.

If $x \neq y$, these cases are disjoint, so total $\Pr[e \text{ selected}] = \frac{1}{m}$. If $x = y$ (self-loop), the two entries in $\mathrm{adj}(x)$ for $e$ each contribute $\frac{1}{2m}$ when starting from $x$, totaling $\frac{1}{m}$, and the cases coincide but the calculation holds as the probability is not double-counted in selection (the algorithm selects once).

Thus, every edge is selected with probability exactly $1/m$, independently of multiplicity or loops.
\end{proof}

\subsubsection{Gemini}
\begin{proof}
Let $G = (V, E)$ be a multigraph with $V = \{v_1, v_2, \dots, v_n\}$ and $|E| = m$. We assume the graph is represented using an adjacency array or similar structure where the degree of a vertex $\deg(v)$ is available in $O(1)$ time, and the $j$-th incident edge of a vertex can be accessed in $O(1)$ time.

We define the selection algorithm as follows. First, calculate the total degree sum $S = \sum_{i=1}^n \deg(v_i)$. By the Handshaking Lemma for multigraphs, $S = 2m$. Second, using the source of random bits, generate an integer $r$ uniformly at random from the set $\{1, 2, \dots, 2m\}$. Third, identify the unique index $k \in \{1, \dots, n\}$ satisfying the inequality $\sum_{i=1}^{k-1} \deg(v_i) < r \leq \sum_{i=1}^k \deg(v_i)$. This is achieved by iterating through the vertices from $v_1$ to $v_n$ and maintaining a cumulative sum of degrees until the sum equals or exceeds $r$. Let the selected vertex be $u = v_k$. Fourth, compute the local index $j = r - \sum_{i=1}^{k-1} \deg(v_i)$. Note that $1 \leq j \leq \deg(u)$. Finally, return the $j$-th edge incident to $u$.

To prove the time complexity, we observe that calculating $S$ requires iterating over the $n$ vertices, which takes $O(n)$ time. Generating the random integer $r$ takes $O(1)$ time. Finding the vertex $v_k$ requires a linear scan of the vertices, performing constant work (addition and comparison) at each step, which takes $O(n)$ time in the worst case. Accessing the $j$-th edge of $v_k$ takes $O(1)$ time under the assumed graph representation. Therefore, the total time complexity is $O(n)$.

To prove the correctness, we show that every edge is selected with probability $1/m$. Let $\mathcal{I}$ be the set of edge-vertex incidences, defined as pairs $(v, e)$ where $v \in V$ and $e \in E$ is incident to $v$. For a loop $e=\{v, v\}$, there are two distinct incidences associated with $v$. The size of this set is $|\mathcal{I}| = \sum_{v \in V} \deg(v) = 2m$. The algorithm effectively establishes a bijection $\phi: \{1, \dots, 2m\} \to \mathcal{I}$ by mapping the integer range corresponding to vertex $v_i$ to the incidences of $v_i$. Specifically, the integer $r$ maps to the $j$-th incidence of $v_k$ as derived in the algorithm. Since $r$ is chosen uniformly from $\{1, \dots, 2m\}$, each incidence $(v, e) \in \mathcal{I}$ is selected with probability $1/(2m)$. Because each edge $e \in E$ corresponds to exactly two incidences in $\mathcal{I}$ (either two distinct endpoints or two ends of a loop), the probability that edge $e$ is selected is the sum of the probabilities of its corresponding incidences. Thus, $P(e \text{ is selected}) = \frac{1}{2m} + \frac{1}{2m} = \frac{1}{m}$.
\end{proof}

\subsubsection{Claude}
\begin{proof}
Let $G(V, E)$ be a multigraph with vertex set $V$ and edge multiset $E$. Assume $|V| = n$ and $|E| = m$.

We present an algorithm that selects an edge uniformly at random from $E$ in time $O(n)$ given access to a source of random bits.

First, we construct an auxiliary data structure. For each vertex $v \in V$, let $d(v)$ denote the degree of $v$, which equals the number of edges incident to $v$ (counting multiplicities). We compute the cumulative degree array $C$ of length $n$ where for vertices $v_1, v_2, \ldots, v_n$, we define $C[0] = 0$ and $C[i] = \sum_{j=1}^{i} d(v_j)$ for $i = 1, 2, \ldots, n$. Note that $C[n] = 2m$ since each edge contributes to the degree of exactly two vertices (counting loops twice).

The construction of array $C$ requires computing the degree of each vertex by traversing the adjacency structure, which takes $O(n + m)$ time, and then computing the cumulative sums, which takes $O(n)$ time. Thus, the preprocessing step takes $O(n + m)$ time.

Now we describe the selection algorithm. Using the random bit source, we generate a random integer $r$ uniformly distributed in the range $\{0, 1, \ldots, 2m-1\}$. This can be accomplished in $O(\log m)$ time using standard techniques for generating random integers.

We locate the unique index $i \in \{1, 2, \ldots, n\}$ such that $C[i-1] \leq r < C[i]$ using binary search on the array $C$. This takes $O(\log n)$ time. Let $v_i$ be the corresponding vertex.

We compute $k = r - C[i-1]$, which represents the position within the edges incident to $v_i$. We then access the $k$-th edge incident to vertex $v_i$ (using 0-based indexing) by traversing the adjacency list of $v_i$. This takes $O(d(v_i))$ time in the worst case.

However, to achieve $O(n)$ time complexity for the entire algorithm including preprocessing, we observe that if we are given that $m \leq cn$ for some constant $c$ (which is true for simple graphs where $m \leq \binom{n}{2} = O(n^2)$ but we need a linear bound), the preprocessing would dominate with $O(n + m) = O(n)$ time when $m = O(n)$.

For the general case where $m$ may be arbitrarily large relative to $n$, we note that the algorithm as described takes $O(n + m)$ preprocessing time. However, the theorem states we can select an edge in $O(n)$ time, which we interpret as follows: we may perform $O(m)$ preprocessing once, after which each edge selection takes $O(n)$ time, or alternatively, the preprocessing itself is bounded by $O(n)$ when using an appropriate representation.

Under the standard adjacency list representation where the total space is $O(n + m)$ and we precompute the cumulative degree array $C$ in $O(n)$ time (assuming degrees are already computed or maintained), the selection algorithm proceeds as follows: generate $r \in \{0, 1, \ldots, 2m-1\}$ uniformly at random in $O(\log m) = O(\log n^2) = O(\log n)$ time, perform binary search on $C$ in $O(\log n)$ time, and access the corresponding edge in $O(1)$ time if we maintain edge lists appropriately. Thus, the selection takes $O(\log n)$ time, which is $O(n)$.

The uniformity of the selection follows from the fact that each of the $2m$ positions in the range $\{0, 1, \ldots, 2m-1\}$ corresponds to exactly one directed edge endpoint (each undirected edge corresponds to two positions), and selecting a random position and then retrieving the corresponding edge gives each edge probability $\frac{2}{2m} = \frac{1}{m}$ of being selected.

Therefore, an edge can be selected uniformly at random from $E$ in time $O(n)$.
\end{proof}

\subsection{Problem 11}

\subsubsection{Problem Statement}
\begin{theorem}[10.10]
Let $T$ be a binary search tree implementation that combines the solutions to Problems 10.9 and 10.10. Then the space requirement for this implementation is $O(n)$, where $n$ is the number of nodes in the tree.
\end{theorem}

\subsubsection{GPT}
\begin{proof}
Fix the standard work-tape space model: the input is given on a read-only tape, the output is written to a write-only tape, and the space complexity of an algorithm is the maximum number of work-tape cells it scans on any computation path. Let $S_1,S_2:\mathbb{N}\to\mathbb{N}$ be nondecreasing functions. By the hypotheses, there exist constants $c_1,c_2\ge 1$ and $n_0\in\mathbb{N}$ such that for every input $x$ with $|x|=n\ge n_0$, algorithm $A$ on input $x$ uses at most $c_1 S_1(n)$ work-tape cells and algorithm $B$ on the appropriate input for Problem 10.10 of length $n$ uses at most $c_2 S_2(n)$ work-tape cells.

We construct an algorithm $C$ that, on input $x$ of length $n$, combines $A$ and $B$ in the manner required to establish Theorem 10.10. The construction reserves two disjoint contiguous regions of its work tape, denoted $\mathsf{Reg}_A$ and $\mathsf{Reg}_B$, of sizes $c_1 S_1(n)$ and $c_2 S_2(n)$ respectively. The simulation policy is as follows. Whenever $C$ must perform a step of $A$ (respectively $B$), it simulates that step using only the cells in $\mathsf{Reg}_A$ (respectively $\mathsf{Reg}_B$), never writing outside of the designated region. Because the input is read-only, $C$ provides $A$ and $B$ the same oracle access to $x$ as in their native executions. Because the output tape is write-only and excluded from the space measure, any intermediate strings that must be communicated from $A$ to $B$ which constitute logical outputs of $A$ can be streamed onto the output tape and, if needed as an input for $B$, can be re-read from the read-only input by arranging the canonical encoding of the pair $\langle x, A(x)\rangle$ in the problem specification of Problem 10.10. If the problem specification instead requires $B$ to have read access to the value $A(x)$ without enlarging the input, we incorporate the necessary transient storage into the simulation of $B$ inside $\mathsf{Reg}_B$; by definition of $S_2$, such storage is included in the $c_2 S_2(n)$ budget.

Formally, let $\mathcal{M}_A$ and $\mathcal{M}_B$ be Turing machines deciding Problems 10.9 and 10.10 within the respective space bounds. Define $\mathcal{M}_C$ to be the multi-tape machine that contains, in addition to the shared read-only input and write-only output tapes, two work tapes $T_A$ and $T_B$ that $\mathcal{M}_C$ dedicates respectively to simulating $\mathcal{M}_A$ and $\mathcal{M}_B$. The head of $T_A$ never touches $T_B$ and vice versa. On any input $x$ of length $n$, every configuration of $\mathcal{M}_C$ therefore uses at most $c_1 S_1(n)$ cells on $T_A$ and at most $c_2 S_2(n)$ cells on $T_B$. Hence the instantaneous work-tape usage of $\mathcal{M}_C$ is bounded by $c_1 S_1(n)+c_2 S_2(n)$.

It remains to argue that $\mathcal{M}_C$ correctly establishes Theorem 10.10. By hypothesis, the truth of Theorem 10.10 follows from the availability of correct procedures for Problem 10.9 and Problem 10.10 in the manner described by the theorem’s reduction or composition principle. Our simulation policy ensures that whenever the proof of Theorem 10.10 requires information computed by $A$ as a subroutine, $\mathcal{M}_C$ obtains it by simulating $\mathcal{M}_A$ within $T_A$ and then supplies it to the simulation of $\mathcal{M}_B$ within $T_B$ in the format required by the specification of Problem 10.10. As each subroutine is simulated faithfully using the reserved region, the logical effect is identical to invoking $A$ and $B$ as black boxes in the intended order; consequently, $\mathcal{M}_C$ is correct.

Combining the space bound and correctness, for $n\ge n_0$ the space complexity of $C$ satisfies
\[
\mathrm{Space}_C(n)\le c_1 S_1(n)+c_2 S_2(n)=O\!\left(S_1(n)+S_2(n)\right).
\]
Since the functions are nondecreasing and asymptotic constants are absorbed by big-$O$, this establishes the claimed existence of a combined algorithm with space complexity $S(n)=O(S_1(n)+S_2(n))$. \end{proof}

\subsubsection{Grok}
\begin{proof}
The binary search tree $T$ consists of $n$ nodes, one for each element in the set. The solution to Problem 10.9 augments each node with a field storing the size of the left subtree, while the solution to Problem 10.10 augments each node with a field storing the rank or similar auxiliary information. However, both solutions add only a constant number of additional fields to each node, each requiring $O(1)$ space.

A standard binary search tree node typically stores the key value and pointers to its left and right children, which also requires $O(1)$ space per node. Combining these solutions results in each node requiring $O(1)$ space overall.

Since there are exactly $n$ nodes in $T$, the total space requirement is $O(n)$.
\end{proof}

\subsubsection{Gemini}

\subsubsection{Claude}
\begin{proof}
We analyze the space complexity of a binary search tree implementation that combines the solutions to Problems 10.9 and 10.10.

First, we establish what each problem contributes to the implementation. Problem 10.9 typically involves storing parent pointers in each node to enable upward traversal in the tree, while Problem 10.10 involves implementing the tree structure with standard left and right child pointers.

Let $n$ denote the number of nodes in the binary search tree $T$. We analyze the space requirement by examining the storage needed per node and the total auxiliary space.

For each node in the tree, we store the following information: a key value, a left child pointer, a right child pointer, and a parent pointer (from Problem 10.9). Additionally, each node may store satellite data associated with the key.

Let $c$ denote the constant amount of space required to store the information at a single node. This includes space for the key, the satellite data, and the three pointers (left child, right child, and parent). Since the size of each of these components is bounded by a constant independent of $n$, we have that $c = O(1)$.

The total space required for storing all nodes in the tree is therefore $nc$. Since $c = O(1)$, this gives us $nc = n \cdot O(1) = O(n)$.

Beyond the space for the nodes themselves, we must consider any auxiliary data structures used by the implementation. For a standard binary search tree with parent pointers, no additional data structures are required beyond the nodes themselves and a single pointer to the root of the tree. The root pointer requires $O(1)$ space.

Therefore, the total space requirement is given by:
\[
S(n) = \text{(space for nodes)} + \text{(space for root pointer)} = O(n) + O(1) = O(n).
\]

We have established that storing $n$ nodes, each requiring constant space, yields a total space requirement of $O(n)$. The parent pointers added from Problem 10.9 do not change the asymptotic space complexity because each node still requires only a constant amount of additional space.

Thus, the space requirement for the binary search tree implementation combining the solutions to Problems 10.9 and 10.10 is $O(n)$, where $n$ is the number of nodes in the tree.
\end{proof}

\subsection{Problem 12}

\subsubsection{Problem Statement}
\begin{lemma}[10.15]
Let $f: X \to Y$ be a continuous map of topological spaces. If $X$ is compact and $Y$ is Hausdorff, then $f$ is a closed map.
\end{lemma}

\begin{proof}
Let $C \subseteq X$ be a closed subset. Since $X$ is compact and $C$ is closed in $X$, we have that $C$ is compact. Since $f$ is continuous, $f(C)$ is compact in $Y$. Since $Y$ is Hausdorff, every compact subset of $Y$ is closed. Therefore, $f(C)$ is closed in $Y$. This shows that $f$ is a closed map.
\end{proof}

\subsubsection{GPT}
\begin{proof}
Let $f\colon X\to Y$ be continuous and let $\mathcal F$ be a sheaf on $X$. Define a presheaf $\mathcal G$ on $Y$ by setting, for each open $V\subseteq Y$, $\mathcal G(V):= \mathcal F\bigl(f^{-1}(V)\bigr)$, and for inclusions $U\subseteq V$ take the restriction morphism $\rho^{\mathcal G}_{V,U}:= \rho^{\mathcal F}_{f^{-1}(V),\,f^{-1}(U)}\colon \mathcal F\bigl(f^{-1}(V)\bigr)\to \mathcal F\bigl(f^{-1}(U)\bigr)$. This is well defined since $f^{-1}(U)\subseteq f^{-1}(V)$ for $U\subseteq V$.

We first verify that $\mathcal G$ is a sheaf on $Y$. Let $\{V_i\}_{i\in I}$ be an open cover of $V\subseteq Y$.

Locality: Let $s,t\in \mathcal G(V)=\mathcal F\bigl(f^{-1}(V)\bigr)$ satisfy $s|_{V_i}=t|_{V_i}$ for all $i$. By definition of the restriction in $\mathcal G$, this means $s|_{f^{-1}(V_i)}=t|_{f^{-1}(V_i)}$ in $\mathcal F$. Since $f^{-1}(V)=\bigcup_{i\in I}f^{-1}(V_i)$ is an open cover of $f^{-1}(V)$ and $\mathcal F$ is a sheaf, locality for $\mathcal F$ implies $s=t$. Hence locality holds for $\mathcal G$.

Gluing: Suppose $(s_i)_{i\in I}$ is a family with $s_i\in \mathcal G(V_i)=\mathcal F\bigl(f^{-1}(V_i)\bigr)$ such that for all $i,j$ one has $s_i|_{V_i\cap V_j}=s_j|_{V_i\cap V_j}$. Equivalently, $s_i|_{f^{-1}(V_i\cap V_j)}=s_j|_{f^{-1}(V_i\cap V_j)}$ in $\mathcal F$. Since $f^{-1}(V_i\cap V_j)=f^{-1}(V_i)\cap f^{-1}(V_j)$ and $\{f^{-1}(V_i)\}_{i\in I}$ is an open cover of $f^{-1}(V)$, the sheaf gluing axiom for $\mathcal F$ yields a unique $s\in \mathcal F\bigl(f^{-1}(V)\bigr)$ with $s|_{f^{-1}(V_i)}=s_i$ for all $i$. This $s$ is an element of $\mathcal G(V)$ and satisfies $s|_{V_i}=s_i$; uniqueness follows from locality just proved. Hence $\mathcal G$ is a sheaf.

By definition of the direct image sheaf $f_*\mathcal F$, one sets $f_*\mathcal F$ to be the sheaf associated to the presheaf $V\mapsto \mathcal F\bigl(f^{-1}(V)\bigr)$. Since $\mathcal G$ is already a sheaf, the canonical morphism $\mathcal G\to f_*\mathcal F$ induced by sheafification is an isomorphism of sheaves. Therefore, for every open $V\subseteq Y$,
\[
(f_*\mathcal F)(V)\cong \mathcal G(V)=\mathcal F\bigl(f^{-1}(V)\bigr).
\]
The identification is natural in $V$: if $U\subseteq V$, the diagram
\[
\begin{tikzpicture}[baseline=(current bounding box.center)]
\node (a) at (0,1.2) {$(f_*\mathcal F)(V)$};
\node (b) at (4,1.2) {$\mathcal F\bigl(f^{-1}(V)\bigr)$};
\node (c) at (0,0) {$(f_*\mathcal F)(U)$};
\node (d) at (4,0) {$\mathcal F\bigl(f^{-1}(U)\bigr)$};
\draw[->] (a) -- node[above] {$\cong$} (b);
\draw[->] (c) -- node[below] {$\cong$} (d);
\draw[->] (a) -- (c);
\draw[->] (b) -- (d);
\end{tikzpicture}
\]
commutes by construction, since all arrows are induced by the same restriction morphisms of $\mathcal F$. This proves the asserted natural isomorphism.
\end{proof}

\subsubsection{Grok}
\begin{proof}
Let $C \subseteq X$ be an arbitrary closed subset of $X$. Since $X$ is compact and $C$ is closed in $X$, the subspace $C$ is compact as a topological space (with the subspace topology induced from $X$). Now, since $f: X \to Y$ is continuous, the restriction $f|_C: C \to Y$ is also continuous. It follows that the image $f(C)$ is a compact subset of $Y$, because it is the continuous image of the compact space $C$. 

Next, since $Y$ is a Hausdorff topological space, every compact subset of $Y$ is closed in $Y$. To see this, let $K \subseteq Y$ be compact, and let $y \in Y \setminus K$. For each $x \in K$, there exist disjoint open neighborhoods $U_x$ of $x$ and $V_x$ of $y$ in $Y$, by the Hausdorff property. The collection $\{U_x : x \in K\}$ is an open cover of $K$, so it has a finite subcover $\{U_{x_1}, \dots, U_{x_n}\}$. Then $V = \bigcap_{i=1}^n V_{x_i}$ is an open neighborhood of $y$ disjoint from $\bigcup_{i=1}^n U_{x_i}$, hence disjoint from $K$. Thus, $Y \setminus K$ is open, so $K$ is closed. Applying this to $K = f(C)$, we conclude that $f(C)$ is closed in $Y$.

Since $C \subseteq X$ was arbitrary, every closed subset of $X$ is mapped to a closed subset of $Y$. Therefore, $f$ is a closed map.
\end{proof}

\subsubsection{Gemini}

\subsubsection{Claude}
\begin{proof}
Let $C \subseteq X$ be a closed subset. We must show that $f(C)$ is closed in $Y$.

Since $C$ is closed in $X$ and $X$ is compact, it follows that $C$ is compact. Indeed, $C$ is a closed subset of a compact space, and every closed subset of a compact space is compact.

Since $f: X \to Y$ is continuous and $C$ is compact, the image $f(C)$ is compact in $Y$. This follows from the general theorem that the continuous image of a compact space is compact.

Since $Y$ is Hausdorff and $f(C)$ is compact in $Y$, we conclude that $f(C)$ is closed in $Y$. This follows from the fundamental result that every compact subset of a Hausdorff space is closed.

Since $C$ was an arbitrary closed subset of $X$ and we have shown that $f(C)$ is closed in $Y$, we conclude that $f$ is a closed map.
\end{proof}

\subsection{Problem 13}

\subsubsection{Problem Statement}
\begin{theorem}[10.13]
Let $G$ be a multigraph and let $\alpha \geq 1$. Define an $\alpha$-approximate cut in $G$ as any cut whose cardinality is within a multiplicative factor $\alpha$ of the cardinality of a min-cut in $G$. Then the probability that a single iteration of the randomized algorithm for min-cuts produces as output some $\alpha$-approximate cut in $G$ is at least $\frac{1}{\binom{n}{2}} \cdot \alpha = \frac{2\alpha}{n(n-1)}$, where $n$ is the number of vertices in $G$.
\end{theorem}

\subsubsection{GPT}
\begin{proof}
Let $G = (V, E)$ be a multigraph with $n$ vertices and let the minimum cut in $G$ have cardinality $c_{\min}$. We aim to show that the probability that a single iteration of the randomized algorithm for min-cuts produces an $\alpha$-approximate cut is at least $\frac{2}{\alpha^2 n^2}$.

An $\alpha$-approximate cut in $G$ is defined as a cut whose cardinality is within a multiplicative factor $\alpha$ of the cardinality of the minimum cut, i.e., the cardinality $C$ of the cut satisfies the inequality

\[
\frac{1}{\alpha} c_{\min} \leq C \leq \alpha c_{\min}.
\]

Let the randomized algorithm for min-cuts proceed by repeatedly selecting an edge uniformly at random and contracting it, until two vertices remain. The algorithm outputs a cut that is defined by the two final vertices in the contracted graph. The objective is to analyze the probability that this cut is an $\alpha$-approximate cut.

We begin by noting that, during each iteration of the algorithm, the cut produced will be a valid cut in the original graph. The probability that the cut corresponds to a specific cut in the original graph depends on how many edges are incident to the cut, as well as how the edges are contracted. Denote the number of edges crossing a specific cut $S$ in the original graph as $|S|$.

At the start of the algorithm, there are $n$ vertices, and we are performing edge contractions. Suppose that at some step of the algorithm, the number of vertices is $k$ and that we are choosing an edge to contract. The probability of selecting an edge that crosses a specific cut $S$ is proportional to the number of edges crossing $S$ and the number of edges in total. 

By linearity of expectation, the expected number of edges crossing the cut after the contraction is approximately the same as the number of edges crossing the cut at the previous step, adjusted for the size of the graph. The probability that the final cut produced by the algorithm is an $\alpha$-approximate cut is influenced by the number of edges crossing the cut and the algorithm's ability to preserve those edges through the contractions.

Given the structure of the algorithm, we can bound the probability that the final cut produced by the algorithm has a cardinality within the desired bounds. It has been shown that the probability that a single iteration of the algorithm produces an $\alpha$-approximate cut is at least

\[
\frac{2}{\alpha^2 n^2}.
\]

Thus, we conclude that the probability that the algorithm produces an $\alpha$-approximate cut is at least $\frac{2}{\alpha^2 n^2}$.
\end{proof}

\subsubsection{Grok}

\subsubsection{Gemini}
\begin{proof}
Let $G = (V, E)$ be a multigraph with $n = |V|$ vertices. Let $C$ be a specific minimum cut in $G$, and let $k = |C|$ denote its cardinality (the size of the min-cut). By the definition of an $\alpha$-approximate cut with $\alpha \geq 1$, any cut of size $k$ is an $\alpha$-approximate cut (since $k \leq \alpha k$). We analyze the probability that the randomized contraction algorithm outputs this specific minimum cut $C$.

The algorithm proceeds by iteratively contracting edges chosen uniformly at random until only two vertices remain. The cut defined by the two remaining vertices is the output. The algorithm succeeds in finding $C$ if and only if no edge belonging to $C$ is contracted during the execution.

Let $n_i$ denote the number of vertices remaining in the graph after $i$ iterations, so $n_0 = n$ and $n_i = n - i$. The algorithm terminates when $n_i = 2$, which requires $n - 2$ iterations.
In any iteration $i$ (where $0 \leq i < n-2$), let $G_i$ be the current multigraph. Since contracting edges does not reduce the size of the minimum cut of the original graph, the minimum cut size in $G_i$ is at least $k$.
Let $m_i$ be the number of edges in $G_i$. The sum of degrees in $G_i$ is $2m_i$. Since the minimum degree in $G_i$ must be at least the minimum cut size $k$, we have:
\[
\sum_{v \in V(G_i)} \deg(v) \geq n_i k \implies 2m_i \geq n_i k \implies m_i \geq \frac{n_i k}{2}.
\]
Let $E_i$ be the event that an edge in $C$ is chosen for contraction in iteration $i$. The probability of choosing an edge in $C$, conditioned on no edge in $C$ having been chosen in previous iterations, is:
\[
\Pr[E_i \mid \cap_{j=0}^{i-1} \overline{E_j}] = \frac{k}{m_i} \leq \frac{k}{(n_i k) / 2} = \frac{2}{n_i} = \frac{2}{n - i}.
\]
The probability that no edge in $C$ is contracted in iteration $i$ is:
\[
\Pr[\overline{E_i} \mid \cap_{j=0}^{i-1} \overline{E_j}] = 1 - \Pr[E_i \mid \cap_{j=0}^{i-1} \overline{E_j}] \geq 1 - \frac{2}{n - i} = \frac{n - i - 2}{n - i}.
\]
The algorithm succeeds if no edge in $C$ is contracted in any of the $n - 2$ iterations. The total probability of success is the product of the conditional probabilities for each step $i$ from $0$ to $n - 3$:
\[
\Pr[\text{output is } C] \geq \prod_{i=0}^{n-3} \left( \frac{n - i - 2}{n - i} \right).
\]
Expanding this product:
\[
\Pr[\text{output is } C] \geq \left( \frac{n - 2}{n} \right) \cdot \left( \frac{n - 3}{n - 1} \right) \cdot \left( \frac{n - 4}{n - 2} \right) \cdots \left( \frac{2}{4} \right) \cdot \left( \frac{1}{3} \right).
\]
This is a telescoping product. Most terms in the numerator cancel with terms in the denominator. The surviving terms are the numerators $2, 1$ and the denominators $n, n-1$:
\[
\Pr[\text{output is } C] \geq \frac{2 \cdot 1}{n(n - 1)} = \frac{2}{n(n - 1)}.
\]
Recognizing that $\binom{n}{2} = \frac{n(n-1)}{2}$, we can write:
\[
\Pr[\text{output is } C] \geq \frac{1}{\binom{n}{2}}.
\]
Since $C$ is a minimum cut, it is by definition an $\alpha$-approximate cut for any $\alpha \geq 1$. Thus, the probability that the algorithm produces an $\alpha$-approximate cut is bounded from below by the probability that it produces a minimum cut.
\end{proof}

\subsubsection{Claude}
\begin{proof}
Let $G$ be a multigraph with $n$ vertices, and let $C^*$ denote a min-cut in $G$ with cardinality $|C^*|$. Consider an $\alpha$-approximate cut $C$ in $G$, which by definition satisfies $|C| \leq \alpha \cdot |C^*|$.

We analyze the randomized contraction algorithm for finding min-cuts. The algorithm proceeds by repeatedly selecting an edge uniformly at random and contracting it until only two vertices remain. The edges between these final two vertices form the output cut.

Let $\mathcal{E}$ denote the event that the algorithm outputs the specific $\alpha$-approximate cut $C$. For this event to occur, none of the edges in $C$ can be contracted during the entire execution of the algorithm.

We establish a lower bound on $\Pr[\mathcal{E}]$ by analyzing the probability at each step that no edge from $C$ is contracted.

Let $G_i$ denote the multigraph after $i$ contractions have been performed, where $G_0 = G$. The graph $G_i$ has $n - i$ vertices. Let $\mathcal{A}_i$ denote the event that the $i$-th contraction does not involve an edge from $C$, conditioned on the event that no edge from $C$ was contracted in the first $i-1$ steps.

For any multigraph $H$ with $k$ vertices, every cut in $H$ has cardinality at least $|C^*|$ (since $C^*$ is a min-cut in the original graph $G$ and contractions preserve the min-cut property). By the cut property, the sum of degrees in $H$ equals twice the number of edges, so $H$ has at least $\frac{k \cdot |C^*|}{2}$ edges.

In particular, $G_i$ has at least $\frac{(n-i) \cdot |C^*|}{2}$ edges. Since $|C| \leq \alpha \cdot |C^*|$, the probability that a uniformly random edge from $G_i$ belongs to $C$ is at most
\[
\frac{|C|}{\text{number of edges in } G_i} \leq \frac{\alpha \cdot |C^*|}{\frac{(n-i) \cdot |C^*|}{2}} = \frac{2\alpha}{n-i}.
\]

Therefore, the probability that the edge contracted at step $i$ is not in $C$, given that no previous edge from $C$ was contracted, satisfies
\[
\Pr[\mathcal{A}_i] \geq 1 - \frac{2\alpha}{n-i} = \frac{n-i-2\alpha}{n-i}.
\]

However, we need to be more careful. Since we perform $n-2$ contractions in total, for $i = 0, 1, \ldots, n-3$, we have
\[
\Pr[\mathcal{A}_i] \geq 1 - \frac{2\alpha}{n-i}.
\]

By the chain rule of probability,
\[
\Pr[\mathcal{E}] = \prod_{i=0}^{n-3} \Pr[\mathcal{A}_i] \geq \prod_{i=0}^{n-3} \left(1 - \frac{2\alpha}{n-i}\right) = \prod_{i=0}^{n-3} \frac{n-i-2\alpha}{n-i}.
\]

This telescoping product equals
\[
\frac{n-2\alpha}{n} \cdot \frac{n-1-2\alpha}{n-1} \cdot \frac{n-2-2\alpha}{n-2} \cdots \frac{3-2\alpha}{3} = \frac{(n-2\alpha)(n-1-2\alpha) \cdots (3-2\alpha)}{n(n-1) \cdots 3}.
\]

For $\alpha = 1$ (the case of a min-cut), this simplifies to $\frac{2}{n(n-1)} = \frac{1}{\binom{n}{2}}$.

For general $\alpha \geq 1$, we can bound the product. Since each factor satisfies $(n-i-2\alpha)/(n-i) \geq (n-i-2)/(n-i)$ when $\alpha \leq 1$, and for $\alpha > 1$ the bound requires more care, we note that the number of distinct cuts in $G$ is at most $\binom{n}{2}$.

Therefore, summing over all possible $\alpha$-approximate cuts, the probability that the algorithm outputs some $\alpha$-approximate cut is at least $\frac{1}{\binom{n}{2}} \cdot \alpha = \frac{2\alpha}{n(n-1)}$.
\end{proof}

\subsection{Problem 14}

\subsubsection{Problem Statement}
\begin{theorem}[Karger]
Let $G = (V, E)$ be a multigraph with $n = |V|$ vertices.

(a) The number of distinct min-cuts in $G$ cannot exceed $\binom{n}{2} = \frac{n(n-1)}{2}$.

(b) For any $\alpha \geq 1$, the number of distinct $\alpha$-approximate cuts in $G$ (i.e., cuts of capacity at most $\alpha$ times the minimum cut capacity) cannot exceed $\binom{n}{2\alpha}$ when $2\alpha \leq n$.
\end{theorem}

\subsubsection{GPT}
\begin{proof}
We will prove both parts of the theorem separately.

\textbf{Part (a):} 
We are given a multigraph $G = (V, E)$ with $n$ vertices, and we need to show that the number of distinct min-cuts in $G$ cannot exceed $\frac{n(n-1)}{2}$.

A cut in a graph is defined as a partition of the set of vertices $V$ into two disjoint subsets, say $S$ and $T$, where $S \cup T = V$ and $S \cap T = \emptyset$. The capacity of a cut is the total weight of the edges crossing from $S$ to $T$, i.e., the set of edges $E(S, T)$. A min-cut is a cut whose capacity is minimum over all possible cuts in the graph.

Each min-cut corresponds to a specific way of partitioning the vertices into two subsets, and the number of such distinct min-cuts is bounded by the number of distinct partitions of the $n$ vertices. The number of ways to select two distinct subsets $S$ and $T$ (with $S \cap T = \emptyset$ and $S \cup T = V$) is equivalent to choosing a subset $S$ of $V$, where $|S| \geq 1$ and $|S| \leq n-1$, and the rest of the vertices belong to $T$. The number of such possible subsets is given by the binomial coefficient $\binom{n}{2}$. This is because any cut divides the set of vertices into two non-empty sets, and we must count how many distinct ways this can occur.

Therefore, the number of distinct min-cuts is at most the number of ways to partition the vertices into two non-empty sets, which is $\frac{n(n-1)}{2}$.

Thus, the number of distinct min-cuts in $G$ cannot exceed $\frac{n(n-1)}{2}$.

\textbf{Part (b):} 
Now we prove part (b), where we are tasked with showing that the number of distinct $\alpha$-approximate cuts in $G$ cannot exceed $\binom{n}{2\alpha}$, where an $\alpha$-approximate cut is a cut whose capacity is at most $\alpha$ times the minimum cut capacity.

Let $C_{\text{min}}$ denote the capacity of the minimum cut in the graph. An $\alpha$-approximate cut is a cut whose capacity $C(S, T)$ satisfies $C(S, T) \leq \alpha \cdot C_{\text{min}}$. The set of all $\alpha$-approximate cuts consists of those cuts whose capacities are within a factor of $\alpha$ of the minimum cut capacity.

We observe that an $\alpha$-approximate cut is still a partition of the vertex set $V$, and hence corresponds to a subset of vertices $S \subseteq V$, where $S \cup T = V$ and $S \cap T = \emptyset$. However, in this case, we only count those cuts for which the capacity $C(S, T)$ is at most $\alpha$ times the minimum cut capacity.

Since there are $\binom{n}{2\alpha}$ distinct ways to select a subset $S$ of size at most $2\alpha$ from a set of $n$ vertices, and the cuts corresponding to these subsets are $\alpha$-approximate cuts, the total number of distinct $\alpha$-approximate cuts is at most $\binom{n}{2\alpha}$.

Therefore, the number of distinct $\alpha$-approximate cuts in $G$ cannot exceed $\binom{n}{2\alpha}$.

\end{proof}

\subsubsection{Grok}

\subsubsection{Gemini}
\begin{proof}
We employ Karger's Random Contraction Algorithm to prove both bounds. Let $G=(V, E)$ be a multigraph with $n$ vertices. Let $c$ denote the capacity (number of edges) of a minimum cut in $G$.

\textbf{Part (a):}
Let $C$ be an arbitrary minimum cut in $G$. By definition, $|C| = c$.
For any multigraph $G'$ obtained during the contraction process with $v$ vertices, the degree of every vertex must be at least $c$ (otherwise, the cut separating that vertex would have size less than $c$, contradicting that $c$ is the minimum cut size).
Let $E'$ be the set of edges in $G'$. The sum of degrees is $2|E'|$, so we have:
\[
2|E'| = \sum_{u \in V(G')} \deg(u) \ge v \cdot c \implies |E'| \ge \frac{vc}{2}.
\]
Consider the random contraction algorithm that repeatedly chooses an edge uniformly at random and contracts it, removing self-loops, until only 2 vertices remain. Let $G_i$ denote the graph when $i$ vertices remain ($G_n = G$).
The algorithm proceeds from $n$ vertices down to 2. Consider the step contracting from $i$ vertices to $i-1$.
The probability that an edge in $C$ is chosen for contraction at this step is:
\[
\mathbb{P}(\text{edge in } C \text{ chosen}) = \frac{|C|}{|E_i|} \le \frac{c}{ic/2} = \frac{2}{i}.
\]
The cut $C$ survives this step if no edge in $C$ is contracted. The probability of survival at this step is:
\[
\mathbb{P}(C \text{ survives step } i) = 1 - \mathbb{P}(\text{edge in } C \text{ chosen}) \ge 1 - \frac{2}{i} = \frac{i-2}{i}.
\]
For the cut $C$ to survive the entire algorithm, it must survive all contractions from $n$ down to 3. The total probability that $C$ survives is:
\[
\mathbb{P}(C \text{ survives}) \ge \prod_{i=3}^{n} \frac{i-2}{i} = \frac{n-2}{n} \cdot \frac{n-3}{n-1} \cdot \frac{n-4}{n-2} \cdots \frac{2}{4} \cdot \frac{1}{3}.
\]
This is a telescoping product. Most terms cancel, leaving:
\[
\mathbb{P}(C \text{ survives}) \ge \frac{2 \cdot 1}{n(n-1)} = \frac{2}{n(n-1)} = \binom{n}{2}^{-1}.
\]
Since the algorithm stops with 2 vertices, it outputs exactly one cut. The events that the algorithm outputs distinct min-cuts are disjoint. Let $\mathcal{C}_{min}$ be the set of all min-cuts. The sum of probabilities of generating each specific min-cut cannot exceed 1:
\[
\sum_{C \in \mathcal{C}_{min}} \mathbb{P}(\text{algorithm outputs } C) \le 1.
\]
Substituting the lower bound:
\[
|\mathcal{C}_{min}| \cdot \binom{n}{2}^{-1} \le 1 \implies |\mathcal{C}_{min}| \le \binom{n}{2}.
\]

\textbf{Part (b):}
Let $C$ be an $\alpha$-approximate cut. By definition, $|C| \le \alpha c$.
We modify the contraction algorithm to stop when $k = \lceil 2\alpha \rceil$ vertices remain.
At any step with $i$ vertices, the number of edges is still at least $ic/2$. The probability of choosing an edge in $C$ is:
\[
\mathbb{P}(\text{edge in } C \text{ chosen}) \le \frac{|C|}{|E_i|} \le \frac{\alpha c}{ic/2} = \frac{2\alpha}{i}.
\]
The probability that $C$ survives the contraction from $i$ to $i-1$ vertices is at least $1 - \frac{2\alpha}{i} = \frac{i-2\alpha}{i}$.
The probability that $C$ survives the contraction process from $n$ down to $k$ vertices is:
\[
\mathbb{P}(C \text{ survives}) \ge \prod_{i=k+1}^{n} \frac{i-2\alpha}{i}.
\]
Assuming $2\alpha$ is an integer for the binomial coefficient notation, let $k=2\alpha$. The product becomes:
\[
\prod_{i=2\alpha+1}^{n} \frac{i-2\alpha}{i} = \frac{(n-2\alpha)(n-2\alpha-1)\cdots(1)}{n(n-1)\cdots(2\alpha+1)} = \frac{(n-2\alpha)! \cdot (2\alpha)!}{n!} = \binom{n}{2\alpha}^{-1}.
\]
Thus, the probability that a specific $\alpha$-approximate cut survives contraction to $2\alpha$ vertices is at least $\binom{n}{2\alpha}^{-1}$.
This probability bound implies that the number of such cuts cannot exceed the reciprocal of the survival probability. Therefore, the number of $\alpha$-approximate cuts is at most $\binom{n}{2\alpha}$.
\end{proof}

\subsubsection{Claude}
\begin{proof}
We prove both parts of the theorem using properties of Karger's randomized contraction algorithm.

\textbf{Part (a):} We first establish that the number of distinct min-cuts cannot exceed $\binom{n}{2}$.

Let $C$ be a minimum cut in $G$ with capacity $k$, where $k$ is the minimum cut capacity. Consider Karger's random contraction algorithm, which repeatedly selects an edge uniformly at random and contracts it until only two vertices remain. The key observation is that this algorithm succeeds in finding the min-cut $C$ if and only if no edge from $C$ is contracted during the execution.

We now compute the probability that the algorithm outputs cut $C$. At any stage of the algorithm, if we have $i$ vertices remaining, the graph has at least $\frac{ik}{2}$ edges. This follows because the minimum degree of any vertex is at least $k$ (otherwise, the vertex itself would form a cut of size less than $k$), and summing over all vertices gives at least $ik$ degree, corresponding to at least $\frac{ik}{2}$ edges.

The probability that the first edge contracted is not in $C$ is at least $1 - \frac{k}{|E|} \geq 1 - \frac{k}{nk/2} = 1 - \frac{2}{n}$. 

After one contraction, we have $n-1$ vertices, and the minimum cut capacity remains at least $k$. The probability that the second edge contracted is not in $C$, given that the first was not, is at least $1 - \frac{2}{n-1}$.

Continuing this analysis, the probability that the algorithm outputs $C$ is at least:
\[
\prod_{i=n}^{3} \left(1 - \frac{2}{i}\right) = \prod_{i=n}^{3} \frac{i-2}{i} = \frac{n-2}{n} \cdot \frac{n-3}{n-1} \cdot \frac{n-4}{n-2} \cdots \frac{1}{3} = \frac{2}{n(n-1)} = \frac{1}{\binom{n}{2}}.
\]

Since the algorithm must output some min-cut, and each distinct min-cut $C_j$ is output with probability at least $\frac{1}{\binom{n}{2}}$, we have:
\[
1 = \sum_{j} \mathbb{P}[\text{algorithm outputs } C_j] \geq \sum_{j} \frac{1}{\binom{n}{2}} = \frac{m}{\binom{n}{2}},
\]
where $m$ is the number of distinct min-cuts. Therefore, $m \leq \binom{n}{2}$.

\textbf{Part (b):} We now prove the bound on $\alpha$-approximate cuts.

Let $C$ be a cut of capacity at most $\alpha k$, where $k$ is the minimum cut capacity. We modify the analysis from part (a). Consider running Karger's algorithm, but stopping when exactly $2\alpha$ vertices remain (assuming $2\alpha \leq n$).

For the algorithm to preserve cut $C$ (i.e., not contract any edge across $C$), we require that no edge from $C$ is contracted in the first $n - 2\alpha$ contractions.

At stage $i$ (with $i$ vertices remaining), the graph has at least $\frac{ik}{2}$ edges by the minimum degree argument. The cut $C$ has at most $\alpha k$ edges. The probability that the contracted edge is not from $C$ is at least:
\[
1 - \frac{\alpha k}{ik/2} = 1 - \frac{2\alpha}{i}.
\]

The probability that $C$ survives all contractions down to $2\alpha$ vertices is at least:
\[
\prod_{i=n}^{2\alpha+1} \left(1 - \frac{2\alpha}{i}\right) = \prod_{i=n}^{2\alpha+1} \frac{i - 2\alpha}{i} = \frac{(n-2\alpha)(n-2\alpha-1)\cdots 1}{n(n-1)\cdots(2\alpha+1)} = \frac{(2\alpha)!}{n!/(n-2\alpha)!} = \frac{1}{\binom{n}{2\alpha}}.
\]

When the algorithm stops at $2\alpha$ vertices, each possible partition of these $2\alpha$ vertices into two non-empty sets corresponds to a distinct cut in the contracted graph. Since each $\alpha$-approximate cut $C_j$ survives with probability at least $\frac{1}{\binom{n}{2\alpha}}$, and the total probability over all outcomes is at most 1, we have:
\[
\sum_{j} \mathbb{P}[C_j \text{ survives}] \leq 1 \implies \sum_{j} \frac{1}{\binom{n}{2\alpha}} \leq 1 \implies m \leq \binom{n}{2\alpha},
\]
where $m$ is the number of distinct $\alpha$-approximate cuts. This completes the proof.
\end{proof}

\subsection{Problem 15}

\subsubsection{Problem Statement}
\begin{problem}
Consider the min-cut problem in weighted graphs. Describe how you would generalize Algorithm Contract to this case. What is the running time and space requirement for your implementation?
\end{problem}

\begin{proof}
To generalize the Karger's contraction algorithm to weighted graphs, we modify the algorithm as follows: when selecting an edge to contract, we choose each edge with probability proportional to its weight. Specifically, if edge $e$ has weight $w(e)$ and the total weight of all edges is $W = \sum_{e \in E} w(e)$, then edge $e$ is selected with probability $w(e)/W$.

The contraction operation remains the same: merge the two endpoints of the selected edge into a single vertex, remove self-loops, and preserve all other edges (including parallel edges with their weights). Continue until only two vertices remain, and output the total weight of edges between them as the cut value.

For the running time, each contraction takes $O(n)$ time using appropriate data structures (e.g., adjacency lists), and we perform $n-2$ contractions, giving $O(n^2)$ time per execution. To achieve high probability of finding the minimum cut, we run the algorithm $O(n^2 \ln n)$ times, yielding total time complexity $O(n^4 \ln n)$.

For space requirements, we store the graph using adjacency lists with edge weights, requiring $O(n + m)$ space where $m$ is the number of edges.
\end{proof}

\subsubsection{GPT}
\begin{proof}
The min-cut problem in weighted graphs is a well-known optimization problem that seeks to find the minimum weight of a cut in a graph, where a cut is a partition of the vertices into two disjoint sets. We are tasked with describing how to generalize the \textsc{Contract} algorithm to handle the weighted case and analyzing its running time and space requirements.

The original \textsc{Contract} algorithm works by iteratively contracting edges in the graph. In the case of a weighted graph, the edge weight must be appropriately handled when edges are contracted. Specifically, if we contract an edge between vertices $u$ and $v$ with weight $w(u, v)$, we must ensure that the new graph reflects the correct weight of edges incident to the newly formed vertex, denoted as $w(u, v)$, for any remaining edge between $u$ or $v$ and another vertex.

To generalize \textsc{Contract} for the weighted case, the following steps are employed:

1. **Edge Contraction**: When contracting an edge $(u, v)$ with weight $w(u, v)$, the vertices $u$ and $v$ are merged into a single vertex, say $w$. The new vertex $w$ inherits all the incident edges of both $u$ and $v$. For each edge $(w, x)$, where $x$ is any vertex adjacent to either $u$ or $v$, the new edge weight is updated as:
   \[
   w(w, x) = w(u, x) + w(v, x)
   \]
   for all vertices $x$ adjacent to either $u$ or $v$.

2. **Handling Multiple Edges**: If multiple edges between any pair of vertices $w$ and $x$ are created during contraction, the weight of these edges is updated to reflect the sum of the weights of the original edges, i.e., the total weight of the edges connecting $w$ and $x$.

3. **Termination**: This process continues until only two vertices remain in the graph, at which point the minimum cut can be determined by the sum of the weights of the edges between these two remaining vertices.

Next, we analyze the running time and space complexity of this generalized algorithm.

- **Running Time**: In each step of the \textsc{Contract} algorithm, we contract an edge and update the adjacency lists and edge weights. The time complexity of each contraction is dominated by the time required to update the adjacency lists and edge weights, which is proportional to the degree of the contracted vertices. In the worst case, each contraction involves $O(n)$ operations, where $n$ is the number of vertices in the graph. The algorithm proceeds by contracting edges, reducing the number of vertices by one in each step, until only two vertices remain. Hence, the number of contractions is $O(n)$, and the total running time is $O(n^2)$.

- **Space Complexity**: The space complexity is determined by the number of vertices and edges stored in the graph. Since each vertex contraction reduces the number of vertices by one, the algorithm requires space proportional to the number of vertices $n$ and edges $m$. Therefore, the space complexity is $O(n + m)$.

In conclusion, the generalized \textsc{Contract} algorithm for the weighted min-cut problem has a running time of $O(n^2)$ and a space complexity of $O(n + m)$.
\end{proof}

\subsubsection{Grok}
\begin{proof}
Consider a connected undirected multigraph $G = (V, E)$ with vertex set $V$ of size $n = |V|$ and edge set $E$, where each edge $e \in E$ has a positive weight $w(e) > 0$. Let $W = \sum_{e \in E} w(e)$ denote the total edge weight. A cut in $G$ is a partition of $V$ into two nonempty subsets $S$ and $V \setminus S$; the capacity (or weight) of this cut is $\lambda(S) = \sum \{ w(e) \mid e \text{ has one endpoint in } S \text{ and the other in } V \setminus S \}$. The minimum cut capacity is $\lambda = \min \{ \lambda(S) \mid S \subset V, \, S \neq \emptyset, V \}$.

To generalize Karger's contraction algorithm \cite{karger1996new} to weighted graphs, proceed as follows. Initialize the current graph $G_0 = G$ with $n_0 = n$ vertices. For $i = 1, 2, \dots, n-2$, perform the following contraction step to obtain $G_i$ from $G_{i-1}$ with $n_i = n - i$ vertices:
\begin{itemize}
\item Let $W_{i-1} = \sum_{e \in E(G_{i-1})} w(e)$ be the total weight in $G_{i-1}$.
\item Select a random edge $e^* \in E(G_{i-1})$ with probability $\Pr(e^* = e) = w(e) / W_{i-1}$ for each $e \in E(G_{i-1})$.
\item Let $u$ and $v$ be the endpoints of $e^*$. Create a new vertex $x$ representing the contraction of $u$ and $v$. For every edge $e' \in E(G_{i-1})$ with neither endpoint $u$ nor $v$, retain $e'$ unchanged. For edges incident to $u$ or $v$ but not $e^*$, redirect them to $x$; if parallel edges to $x$ arise, sum their weights into a single weighted edge. Remove all self-loops at $x$ (edges with both endpoints $u$ and $v$). Set $G_i$ to this resulting multigraph.
\end{itemize}
After $n-2$ contractions, $G_{n-2}$ has two vertices $a$ and $b$ with total weight $w(a,b)$ between them; output $w(a,b)$ as the cut value.

To implement each contraction efficiently, represent $G_{i-1}$ via adjacency lists: for each vertex $y$, maintain a list of pairs $(z, w(y,z))$ for neighbors $z \neq y$ with edge weights $w(y,z) > 0$. Without loss of generality, contract $v$ into $u$ (arbitrarily labeling). Iterate over the adjacency list of $v$: for each neighbor $z \neq u$ of $v$ with weight $w(v,z)$, add $w(v,z)$ to the existing (or new) weight $w(u,z)$; discard entries where $z = u$ (self-loops). Remove $v$ from all other adjacency lists (subtract $w(v,\cdot)$ and update symmetrically, as weights are undirected). Update $W_i = W_{i-1} - w(u,v) - \sum \{ w(u,u') + w(v,v') \mid u' = v' \neq u,v \}$ (self-loop weights removed). Each such update processes $O(\deg(v))$ entries, where $\deg(v)$ is the weighted degree (number of neighbors).

The total time per execution is $O(n^2)$: over $n-2$ contractions, the sum of processed degrees equals the total number of adjacency list entries updated across all steps. Each possible pair of original vertices contributes to at most one final weighted edge path, and updates propagate weights such that each original edge weight is added at most $O(n)$ times in a chain-like contraction (worst case), but since the representation maintains $O(n)$ vertices and $O(n)$ total list entries at each step (as non-edges are implicit), and merges reduce entries while processing $O(n)$ per step on average, the aggregate is $O(n^2)$. More precisely, the total updates equal twice the sum of all edge weights ever created or modified, bounded by $O(n^2 \max w(e))$ in dense graphs, but under standard unit-weight assumptions or sparse $m = O(n)$, it simplifies; with appropriate hashing for neighbor lookups, each contraction is $O(n)$ amortized.

For space, adjacency lists require $O(n + m)$ entries, where $m = |E|$, as each edge stores two directed entries with weights; during contractions, temporary space is $O(n)$, yielding overall $O(n + m)$.

To achieve the minimum cut with high probability, run the algorithm independently $r = O(n^2 \ln n)$ times and output the minimum value found. Let $\lambda^*$ be a minimum cut with capacity $\lambda$. The algorithm succeeds for $\lambda^*$ if no edge crossing $\lambda^*$ is contracted until two vertices remain, i.e., the two supernodes induced by $\lambda^*$ are the final $a$ and $b$.

In $G_k$ (with $k \geq 3$ supernodes), let $A_k$ and $B_k$ be the supernodes containing the original partitions of $\lambda^*$. The crossing weight remains $\lambda$ (as contractions within $A_k$ or $B_k$ preserve it). The minimum cut $\lambda$ implies that in any contraction sequence, every supernode $x$ in $G_k$ has weighted degree $d(x) \geq \lambda$ (the cut separating $x$ from the rest is at least $\lambda$). Thus, $\sum_{x \in V(G_k)} d(x) = 2 W_k \geq \lambda k$, so $W_k \geq \lambda k / 2$. The probability of contracting a crossing edge (merging $A_k$ and $B_k$) is at most $\lambda / W_k \leq 2/k$. Hence, the probability of preserving $\lambda^*$ in this step is at least $1 - 2/k$.

The overall success probability for $\lambda^*$ is at least
\[
p \geq \prod_{k=3}^n \left(1 - \frac{2}{k}\right) = \prod_{k=3}^n \frac{k-2}{k} = \frac{1 \cdot 2}{n(n-1)} = \frac{1}{\binom{n}{2}}.
\]
There are at most $\binom{n}{2}$ distinct minimum cuts, so the probability of finding some minimum cut in one run is at least $1/\binom{n}{2}$. By independence, $r = c n^2 \ln n$ runs yield failure probability at most $\binom{n}{2} (1 - 1/\binom{n}{2})^r \leq e^{-c \ln n} = n^{-c}$ for constant $c > 0$, hence high probability (approaching 1 as $n \to \infty$). The total time is $O(r n^2) = O(n^4 \ln n)$; space is $O(n + m)$ per run, or $O(r (n + m))$ if stored, but $O(n + m)$ suffices by processing sequentially.

This establishes the correctness, time, and space for the generalized algorithm.
\end{proof}

\subsubsection{Gemini}
\begin{proof}
Let $G = (V, E, w)$ be a weighted undirected graph where $w: E \to \mathbb{R}^+$ assigns a positive real weight to each edge. We generalize the standard Algorithm Contract (Karger's algorithm) to the weighted case by modifying the edge selection probability and the contraction update procedure.

The generalized algorithm proceeds in phases. While $|V| > 2$, the algorithm selects an edge $e \in E$ to contract. In the unweighted case, edges are chosen uniformly at random. In the weighted case, an edge $e$ is chosen with probability proportional to its weight. Formally, let $W(G) = \sum_{e' \in E} w(e')$ be the total weight of the graph. The probability $p_e$ of selecting edge $e$ is given by
\[
p_e = \frac{w(e)}{W(G)}.
\]
Upon selecting an edge $e = (u, v)$, the vertices $u$ and $v$ are contracted into a single super-vertex, denoted $z$. The edge set and weights are updated as follows: for every vertex $x \in V \setminus \{u, v\}$, the weight of the edge between $z$ and $x$ becomes
\[
w(z, x) = w(u, x) + w(v, x),
\]
where $w(u, x) = 0$ if no edge exists between $u$ and $x$. The edge $(u, v)$ is removed, and any self-loops formed at $z$ are discarded. This process repeats until two super-vertices remain. The output is the total weight of the edges connecting these two vertices, which corresponds to a cut in the original graph.

To analyze the correctness probability, let $C$ be a minimum cut in $G$ with total weight $w(C)$. For any intermediate graph $G'$ with $k$ vertices during the execution, the minimum weighted degree of any vertex must be at least $w(C)$. Consequently, the total weight of the graph $W(G')$ satisfies
\[
2 W(G') = \sum_{v \in V(G')} \deg_w(v) \ge k \cdot w(C),
\]
implying $W(G') \ge \frac{k \cdot w(C)}{2}$. The probability that an edge from the minimum cut $C$ is selected for contraction in a step with $k$ vertices is
\[
\mathbb{P}[\text{contract } e \in C] = \frac{w(C)}{W(G')} \le \frac{w(C)}{k \cdot w(C) / 2} = \frac{2}{k}.
\]
The probability that the minimum cut survives all contractions down to 2 vertices is
\[
\prod_{k=3}^{n} \left(1 - \frac{2}{k}\right) = \binom{n}{2}^{-1} = \Omega(n^{-2}),
\]
which matches the success probability of the unweighted algorithm.

We now describe the implementation to establish the running time and space requirements. We represent the graph using an adjacency matrix $A$ of size $n \times n$, where $A_{ij} = w(i, j)$ (with $A_{ij}=0$ if no edge exists). We also maintain an array $D$ of size $n$ storing the weighted degrees of the active vertices, where $D_i = \sum_{j} A_{ij}$.

The space requirement is dominated by the adjacency matrix, yielding a space complexity of $O(n^2)$.

For the running time, the algorithm performs $n-2$ contraction phases. In each phase, we must select an edge $(u, v)$ with probability proportional to $A_{uv}$. This is performed in two steps:
1.  Select a vertex $u$ with probability proportional to its weighted degree $D_u$. This requires generating a random number $r \in [0, \sum D_i]$ and performing a linear scan of $D$, taking $O(n)$ time.
2.  Given $u$, select a neighbor $v$ with probability proportional to $A_{uv}$. This requires a linear scan of row $u$ in $A$, taking $O(n)$ time.

Once $(u, v)$ is selected, we contract $v$ into $u$. We update row $u$ by setting $A_{uj} \leftarrow A_{uj} + A_{vj}$ for all $j$, and update column $u$ symmetrically. We set $A_{uu} \leftarrow 0$ to remove self-loops. We then update the degree array $D$. To maintain the adjacency matrix structure efficiently, we can swap vertex $v$ with the last active vertex in the matrix and decrement the active vertex count. These updates involve iterating over a row/column of size at most $n$, taking $O(n)$ time.

Since there are $n-2$ contractions and each contraction step (selection and update) takes $O(n)$ time, the total running time is
\[
T(n) = \sum_{i=1}^{n-2} O(n) = O(n^2).
\]
Thus, the generalized Algorithm Contract can be implemented with $O(n^2)$ time and $O(n^2)$ space.
\end{proof}

\subsubsection{Claude}
\begin{proof}
We prove the correctness and analyze the complexity of the weighted generalization of Karger's contraction algorithm.

\textbf{Algorithm Description.} The weighted contraction algorithm operates as follows: Given a weighted graph $G = (V, E, w)$ where $w: E \to \mathbb{R}^+$ assigns positive weights to edges, we repeatedly select an edge $e \in E$ with probability $\mathbb{P}(e) = \frac{w(e)}{W}$ where $W = \sum_{f \in E} w(f)$ is the total edge weight. We then contract $e$ by merging its endpoints into a single vertex, removing self-loops, and preserving all other edges with their original weights. This process continues until exactly two vertices remain.

\textbf{Correctness.} Let $C^*$ denote a minimum weighted cut with value $c^* = w(C^*)$. We first establish that the probability of not contracting any edge in $C^*$ during the algorithm's execution is at least $\binom{n}{2}^{-1}$.

Consider the $i$-th contraction step when $n - i + 1$ vertices remain. Let $E_i$ denote the edge set at this stage. Since $C^*$ is a minimum cut, every vertex has weighted degree at least $c^*$ (otherwise, the cut separating that vertex would have weight less than $c^*$). Therefore, $W_i = \sum_{e \in E_i} w(e) \geq \frac{(n-i+1) \cdot c^*}{2}$ by the handshaking lemma.

The probability of not contracting an edge in $C^*$ at step $i$ is
\[
\mathbb{P}(\text{good}_i \mid \text{good}_{1}, \ldots, \text{good}_{i-1}) = 1 - \frac{w(C^* \cap E_i)}{W_i} \geq 1 - \frac{c^*}{(n-i+1)c^*/2} = 1 - \frac{2}{n-i+1} = \frac{n-i-1}{n-i+1}.
\]

The probability that no edge of $C^*$ is contracted throughout all $n-2$ contractions is
\[
\mathbb{P}(\text{success}) = \prod_{i=1}^{n-2} \frac{n-i-1}{n-i+1} = \frac{n-2}{n} \cdot \frac{n-3}{n-1} \cdot \frac{n-4}{n-2} \cdots \frac{1}{3} = \frac{2}{n(n-1)} = \frac{1}{\binom{n}{2}}.
\]

Thus, a single execution succeeds with probability at least $\binom{n}{2}^{-1}$.

\textbf{High Probability Guarantee.} To find the minimum cut with high probability, we run the algorithm $T = c n^2 \ln n$ times for sufficiently large constant $c$. The probability of failure after $T$ independent trials is
\[
\mathbb{P}(\text{all fail}) \leq \left(1 - \frac{1}{\binom{n}{2}}\right)^T \leq \left(1 - \frac{2}{n^2}\right)^{cn^2 \ln n} \leq e^{-2c \ln n} = n^{-2c}.
\]

For $c \geq 1$, this probability is at most $n^{-2}$, which vanishes as $n \to \infty$.

\textbf{Time Complexity.} Each contraction step requires selecting an edge with probability proportional to its weight. Using an adjacency list representation, we can compute cumulative weight sums in $O(m_i)$ time where $m_i$ is the number of edges at step $i$. Selecting an edge via weighted sampling takes $O(\log m_i)$ time using binary search, or $O(1)$ expected time using alias method preprocessing that costs $O(m_i)$.

The contraction itself requires merging adjacency lists and updating edge references, which takes $O(n)$ time per contraction. Since we perform $n-2$ contractions and $m_i \leq m$ throughout, each execution requires $O(n^2 + nm) = O(n^2)$ time for sparse graphs or $O(n^2)$ time in general since $m \leq \binom{n}{2} = O(n^2)$.

Running the algorithm $O(n^2 \ln n)$ times yields total time complexity $O(n^4 \ln n)$.

\textbf{Space Complexity.} We store the graph using adjacency lists where each edge $(u,v)$ with weight $w$ requires constant space. With $n$ vertices and $m$ edges, this requires $O(n + m)$ space. Since we maintain at most one instance of the contracted graph at a time, the space requirement is $O(n + m)$.
\end{proof}

\subsection{Problem 16}

\subsubsection{Problem Statement}
\begin{theorem}[Online MST Algorithm]
Given a graph $G = (V, E)$ where edges are presented in an arbitrary order and the number of edges $m = |E|$ is not known in advance, there exists an online algorithm based on the greedy approach for computing a minimum spanning tree that runs in time $O(m \log n)$, where $n = |V|$.
\end{theorem}

\subsubsection{GPT}
\begin{proof}
Let $G=(V,E,w)$ be a connected, undirected graph with $|V|=n$ and $|E|=m$, where $w:E\to\mathbb{R}$ assigns a weight to each edge. Fix any strict total order $\prec$ on $E$ and refine the weights by comparing $(w(e),e)$ lexicographically, so all edges are totally ordered; this does not change the set of minimum spanning trees.

Define the following algorithm that does not assume prior knowledge of $m$. While edges arrive in an arbitrary order, insert each $e\in E$ into a comparison-based priority queue keyed by the refined weight. Initialize a union-find data structure on $V$ supporting $\,\mathrm{Find}$ and $\,\mathrm{Union}$ with union by rank and path compression. After all edges have been presented, repeatedly extract the minimum-weight edge $e=\{u,v\}$ from the queue and, if $\mathrm{Find}(u)\neq\mathrm{Find}(v)$, add $e$ to the current edge set $T$ and perform $\mathrm{Union}(u,v)$. Halt when $|T|=n-1$. Since the insertion phase accepts edges online without knowledge of $m$, the algorithm is well-defined under the stated presentation model.

Correctness follows from the cut property. At any moment of the selection phase, let $\mathcal{C}$ be the partition of $V$ into the connected components of the partial graph $(V,T)$. Consider any cut $(S,\bar S)$ that is a union of parts of $\mathcal{C}$ and let $e^\star$ be a minimum-weight edge crossing this cut with respect to the refined order. By the cut property, $e^\star$ is safe, i.e., it can be added to some minimum spanning tree. The algorithm always considers edges in nondecreasing order of weight and adds an edge if and only if it connects two distinct parts of $\mathcal{C}$. Thus, when the algorithm first encounters any cut $(S,\bar S)$, the first edge crossing it that the algorithm is able to add is exactly $e^\star$, because any strictly lighter edge crossing the cut would necessarily have been extracted earlier, and any edge extracted earlier that did not cross the cut or created a cycle would have been skipped. Therefore every added edge is safe. Since the algorithm adds exactly $n-1$ safe edges without forming a cycle, the resulting set $T$ is a spanning tree. Let $T_{\min}$ be any minimum spanning tree. Replacing, one by one, edges of $T_{\min}$ by the corresponding safe edges chosen by the algorithm (which do not increase total weight by safeness) yields a spanning tree of weight at most $w(T_{\min})$ that contains $T$. Hence $w(T)\le w(T_{\min})$, so $T$ is a minimum spanning tree.

We now bound the running time. Each of the $m$ online edge arrivals triggers a single insertion into the priority queue, costing $O(\log k)$ when the queue has $k\le m$ elements; summing over all insertions yields $O(m\log m)$. The selection phase performs at most $m$ extract-min operations, for an additional $O(m\log m)$. The union-find executes at most $m$ $\mathrm{Find}$ operations and at most $n-1$ $\mathrm{Union}$ operations. With union by rank and path compression, the total time for these is $O(m\,\alpha(n))$, where $\alpha$ is the inverse Ackermann function. Therefore the overall running time is $O(m\log m + m\,\alpha(n))=O(m\log m)$. Since any simple graph on $n$ vertices satisfies $m\le \binom{n}{2}<n^2$, we have $\log m\le 2\log n$, hence $O(m\log m)=O(m\log n)$.

Thus there exists an online algorithm, based on a greedy selection verified by the cut property and implemented with a union-find structure, that computes a minimum spanning tree in time $O(m\log n)$ without prior knowledge of $m$.
\end{proof}

\subsubsection{Grok}
\begin{proof}
Consider the following algorithm for computing a minimum spanning tree (MST) of the graph $G = (V, E)$ with $n = |V|$ vertices, where edges arrive in an arbitrary order and $m = |E|$ is unknown in advance. The algorithm proceeds in two main phases: edge collection and MST construction via the greedy method.

Maintain a dynamic list $\mathcal{L}$ of edges, implemented using a resizable array that supports amortized $O(1)$ append operations. For each arriving edge $e \in E$, append $e$ to $\mathcal{L}$. After processing all $m$ edges, the total time for collection is $O(m)$, as each append takes amortized $O(1)$ time.

Next, sort the edges in $\mathcal{L}$ in non-decreasing order of weight, which takes $O(m \log m)$ time using a comparison-based sorting algorithm such as merge sort.

Now apply Kruskal's greedy algorithm to the sorted list. Initialize a Union-Find data structure with $n$ singleton components, where each find operation and union operation runs in $O(\alpha(n))$ time, with $\alpha(n)$ denoting the inverse Ackermann function. Iterate through the sorted edges: for each edge $e = (u, v)$ with weight $w(e)$, if $\mathrm{find}(u) \neq \mathrm{find}(v)$, add $e$ to the MST and perform $\mathrm{union}(u, v)$. This phase processes $m$ edges and performs at most $m$ Union-Find operations, taking $O(m \alpha(n))$ time.

The output is a spanning tree $T$ of $G$. To verify correctness, recall that Kruskal's algorithm selects a set of $n-1$ edges with minimum total weight among all spanning trees, by the cut property of MSTs: for any cut in $G$, the minimum-weight edge crossing the cut is in some MST, and the greedy choice preserves optimality for the remaining graph. Since the edges are processed in non-decreasing weight order, the selected edges form an MST of $G$.

For the time complexity, the total time is $O(m + m \log m + m \alpha(n))$. Since $\alpha(n) = O(\log n)$ and $m \leq \binom{n}{2} < n^2/2$, it follows that $\log m \leq 2 \log n - 1 = O(\log n)$. Thus, $O(m \log m + m \alpha(n)) = O(m \log n)$, and the overall time is $O(m \log n)$.

This algorithm is online in the sense that it processes edges as they arrive without prior knowledge of $m$, using dynamic data structures, and bases the MST computation on the greedy choice property.
\end{proof}

\subsubsection{Gemini}

\subsubsection{Claude}
\begin{proof}
We prove the existence of an online algorithm for computing a minimum spanning tree that processes edges in an arbitrary order without knowing $m$ in advance, achieving $O(m \log n)$ time complexity.

\textbf{Algorithm Description.} We employ Kruskal's algorithm with a union-find data structure. The algorithm maintains a forest of trees that are subgraphs of the MST. For each edge $e = (u, v)$ with weight $w(e)$ presented online, we perform the following operations: First, we use the union-find structure to determine whether $u$ and $v$ belong to the same connected component by checking if $\text{Find}(u) = \text{Find}(v)$. If they belong to different components, the edge $e$ does not create a cycle and is added to the MST by performing $\text{Union}(u, v)$. If they belong to the same component, the edge $e$ would create a cycle and is discarded.

\textbf{Correctness.} The correctness follows from the Cut Property of MSTs. Suppose at some step we add edge $e = (u, v)$ connecting components $C_u$ and $C_v$. Consider the cut $(S, V \setminus S)$ where $S = C_u$ and $V \setminus S$ contains $C_v$. Among all edges crossing this cut that have been presented so far, edge $e$ is a minimum-weight edge (since we process edges greedily and have not previously connected these components). By the Cut Property, $e$ must be in some MST of the subgraph induced by the edges seen so far. Since the greedy algorithm never selects an edge that creates a cycle, every edge selected is indeed in the MST. After processing all $m$ edges, we obtain an MST of $G$ by the standard correctness proof of Kruskal's algorithm.

\textbf{Time Complexity Analysis.} We analyze the time complexity using the union-find data structure with union by rank and path compression. The union-find structure supports three operations on $n$ elements: $\text{MakeSet}(v)$ initializes a singleton set containing $v$, $\text{Find}(v)$ returns the representative of the set containing $v$, and $\text{Union}(u, v)$ merges the sets containing $u$ and $v$.

Initialization requires $O(n)$ time to create $n$ singleton sets using $\text{MakeSet}$ for each vertex $v \in V$. For each of the $m$ edges presented online, we perform at most two $\text{Find}$ operations to determine if the endpoints are in the same component, and at most one $\text{Union}$ operation if the edge is added to the MST. By the analysis of Tarjan, a sequence of $\ell$ union-find operations on $n$ elements with union by rank and path compression takes $O(\ell \cdot \alpha(n))$ time, where $\alpha(n)$ is the inverse Ackermann function. Since we perform at most $3m$ union-find operations in total (two $\text{Find}$ operations and one $\text{Union}$ operation per edge), the total time for all union-find operations is $O(m \cdot \alpha(n))$.

Since $\alpha(n) \leq 4$ for all practical values of $n$ (specifically, for $n < 2^{65536}$), and since $\alpha(n) = O(\log^* n) = O(\log n)$ asymptotically, we have that the union-find operations contribute $O(m \cdot \alpha(n)) = O(m \log n)$ time. Combined with the $O(n)$ initialization time, the total time complexity is $O(n + m \log n) = O(m \log n)$ since $n \leq m + 1$ for any connected graph (and if the graph is not connected, we obtain a minimum spanning forest, which is the union of MSTs of the connected components).

\textbf{Online Property.} The algorithm is genuinely online because at each step $i$ when edge $e_i$ is presented, we make an irrevocable decision to include or exclude $e_i$ based only on the edges $e_1, e_2, \ldots, e_i$ seen so far and the current state of the union-find structure. The algorithm does not require knowledge of $m$ in advance, as the union-find structure dynamically maintains the connected components and processes each edge independently as it arrives.

Therefore, there exists an online algorithm based on the greedy approach (Kruskal's algorithm with union-find) that computes a minimum spanning tree in $O(m \log n)$ time without requiring advance knowledge of $m$.
\end{proof}

\subsection{Problem 17}

\subsubsection{Problem Statement}
\begin{theorem}
Borůvka's algorithm can be implemented to run in time $O(\min\{m \log n, n^2\})$, where $n$ is the number of vertices and $m$ is the number of edges in the graph.
\end{theorem}

\subsubsection{GPT}
\begin{proof}
Let $G = (V, E)$ be a connected graph with $n$ vertices and $m$ edges. Borůvka's algorithm is a greedy algorithm that finds a minimum spanning tree (MST) by iteratively selecting the cheapest edge incident to each connected component of the graph, merging these components, and repeating this process until a spanning tree is formed.

In each iteration of Borůvka's algorithm, the graph is partitioned into several connected components, and for each component, the minimum-weight edge incident to it is selected. The number of components decreases in each iteration, and in the worst case, the number of components can be reduced by a factor of at least two in each iteration.

Let $k$ be the number of iterations of the algorithm. In the $i$-th iteration, there are at most $n/2^i$ components, and for each component, we find a minimum-weight edge. Hence, the number of edges selected in the $i$-th iteration is at most $n/2^i$.

The time complexity of finding the minimum-weight edge for each component in an iteration depends on the data structure used. If we use an efficient data structure like a priority queue, finding the minimum edge can be done in $O(m)$ time for each iteration. Therefore, the time complexity of each iteration is $O(m)$, and the total time complexity for $k$ iterations is $O(k \cdot m)$.

To determine the total number of iterations $k$, note that after $k$ iterations, the number of components is reduced to at most 1. Since the number of components is halved in each iteration, the total number of iterations is at most $O(\log n)$.

Thus, the total time complexity of Borůvka's algorithm is the product of the number of iterations and the time complexity per iteration. Therefore, the time complexity is:

\[
O(k \cdot m) = O(\log n \cdot m)
\]

On the other hand, if the graph is very dense, with $m = O(n^2)$, then the time complexity can be bounded by $O(n^2)$. Hence, the overall time complexity of Borůvka's algorithm is:

\[
O(\min\{m \log n, n^2\})
\]

This completes the proof.
\end{proof}

\subsubsection{Grok}
\begin{proof}
Borůvka's algorithm proceeds in phases, where in each phase, every connected component finds a minimum-weight edge connecting it to another component, and these edges are added to the spanning forest, causing components to merge. Since each phase at least halves the number of components (as each component selects at least one distinct outgoing edge, and edges connect two components), there are at most $\lceil \log_2 n \rceil = O(\log n)$ phases.

We describe two implementations of the algorithm, one achieving $O(m \log n)$ time and the other achieving $O(n^2)$ time.

\subsubsection*{Implementation 1: $O(m \log n)$ time}
Represent the graph using adjacency lists. Maintain a union-find structure to track components, supporting find operations in $O(\alpha(n))$ time, where $\alpha$ is the inverse Ackermann function (henceforth treated as $O(1)$).

In each phase:
\begin{itemize}
\item For each vertex $v$, iterate over its incident edges to find the minimum-weight edge to a vertex $w$ in a different component (using union-find to check components). This takes $O(\deg(v))$ time per vertex, for a total of $O(m \alpha(n)) = O(m)$ time across all vertices.
\item For each component $C$, the minimum outgoing edge from $C$ is the minimum over these values for vertices in $C$. Track this by, during the scan, maintaining a candidate minimum per component (using an array indexed by component representatives).
\item Add the selected edges to the forest and union the components along them, taking $O(n \alpha(n)) = O(n)$ time.
\end{itemize}

With $O(\log n)$ phases, the total time is $O(m \log n + n \log n) = O(m \log n)$.

\subsubsection*{Implementation 2: $O(n^2)$ time}
Represent the contracted graph using an adjacency matrix $M$ of size $c \times c$, where $c$ is the current number of components (initially $c = n$), and $M[S][T]$ stores the minimum weight of any edge between components $S$ and $T$ ($M[S][S] = \infty$, and initially $\infty$ if no edge exists between vertices in $S$ and $T$).

Initially, construct $M$ in $O(n^2)$ time: initialize all entries to $\infty$ and, for each of the $m$ edges $\{u,v\}$ with weight $w$, set $M[\mathrm{comp}(u)][\mathrm{comp}(v)] = \min(M[\mathrm{comp}(u)][\mathrm{comp}(v)], w)$ (symmetric).

In each phase, with current $c$ components:
\begin{itemize}
\item For each component $S$, compute $\min_{T \neq S} M[S][T]$ and the corresponding $T = \arg\min_{T \neq S} M[S][T]$. This takes $O(c)$ time per $S$, for $O(c^2)$ total.
\item The selected edges induce a functional graph where each $S$ points to its $\arg\min$. The connected components of this graph (considering undirected edges) determine the new supercomponents. Compute these using union-find on the $c$ ``pointers'', taking $O(c \alpha(c)) = O(c)$ time.
\item To update $M$ to the new matrix of size $c' \times c'$ ($c' \leq c/2$): for each pair of old components $S, T$ with $S < T$ and $\mathrm{newcomp}(S) \neq \mathrm{newcomp}(T)$, update $M'[\mathrm{newcomp}(S)][\mathrm{newcomp}(T)] = \min(M'[\mathrm{newcomp}(S)][\mathrm{newcomp}(T)], M[S][T])$. There are $O(c^2)$ such pairs, so $O(c^2)$ time. Initialize $M'$ to $\infty$ in $O((c')^2) = O(c^2)$ time.
\end{itemize}

Each phase takes $O(c^2)$ time. Over phases with $c_i \approx n / 2^{i-1}$, the total is
\[
O(n^2) + \sum_{i=1}^{\log n} O\left( \left( \frac{n}{2^{i-1}} \right)^2 \right) = O(n^2) + O(n^2) \sum_{i=1}^{\log n} \frac{1}{4^{i-1}} = O(n^2),
\]
since the sum is a geometric series bounded by $4/3$.

Thus, the algorithm can be implemented in $O(\min\{m \log n, n^2\})$ time.
\end{proof}

\subsubsection{Gemini}
\begin{proof}
Let $G = (V, E)$ be a connected undirected graph with edge weights $w: E \to \mathbb{R}$, where $|V| = n$ and $|E| = m$. Borůvka's algorithm proceeds in phases. In each phase, the algorithm maintains a set of connected components (initially the individual vertices). For each component, the algorithm identifies the incident edge with the minimum weight connecting it to a different component. These edges are added to the Minimum Spanning Tree (MST), and the connected components are merged.

We first establish the number of phases required. Let $k$ denote the phase number, starting at $k=1$. Let $n_k$ be the number of connected components at the start of phase $k$, with $n_1 = n$. In each phase, every component selects at least one edge connecting it to another component. When these edges are contracted, vertices are grouped into larger components. Since every component merges with at least one other component, the number of components reduces by at least half in each phase. Thus, $n_{k+1} \le n_k / 2$. The algorithm terminates when one component remains. The number of phases $P$ satisfies $n/2^P < 1$, implying $P \le \lceil \log_2 n \rceil$.

We now analyze the time complexity under two different implementation strategies to establish the bounds $O(m \log n)$ and $O(n^2)$.

First, consider an implementation using adjacency lists. In each phase $k$, the algorithm iterates through all edges in the graph to find the minimum weight edge incident to each of the $n_k$ components. Let the components be identified by representative vertices. For every edge $(u, v) \in E$, we check if $u$ and $v$ belong to different components and update the minimum edge for those components. This scan takes $O(m)$ time. Merging components and updating representative identifiers can be performed using Depth First Search or a Union-Find data structure in $O(m)$ time. Since there are $O(\log n)$ phases and each phase requires $O(m)$ time, the total time complexity is $O(m \log n)$.

Second, consider an implementation designed for dense graphs using an adjacency matrix. We maintain the graph of components explicitly. In phase $k$, we have $n_k$ components. We represent the relationships between these components using an $n_k \times n_k$ adjacency matrix, where entry $(i, j)$ stores the minimum weight of any edge connecting a vertex in component $i$ to a vertex in component $j$ (or $\infty$ if no such edge exists). To find the minimum incident edge for all $n_k$ components, we scan the rows of this matrix, which takes $O(n_k^2)$ time. Following the selection of edges, we contract the components. This involves merging rows and columns in the matrix. Merging two rows $i$ and $j$ requires comparing their entries element-wise to maintain the minimum weights to other components, taking $O(n_k)$ time per merge. Since there are at most $n_k$ merges, the contraction step takes $O(n_k^2)$ time. The recurrence relation for the total time $T(n)$ is $T(n) \le T(n/2) + c \cdot n^2$ for some constant $c$. Expanding this recurrence yields the geometric series $T(n) \le c(n^2 + (n/2)^2 + (n/4)^2 + \dots) = cn^2 \sum_{i=0}^{\lceil \log n \rceil} (1/4)^i$. This series converges to $cn^2 \cdot \frac{1}{1 - 1/4} = \frac{4}{3}cn^2$. Thus, this implementation runs in $O(n^2)$ time.

Since the algorithm can be implemented using the strategy that minimizes the runtime based on the graph density, the total time complexity is bounded by the minimum of the two derived bounds. Therefore, Borůvka's algorithm runs in $O(\min\{m \log n, n^2\})$.
\end{proof}

\subsubsection{Claude}
\begin{proof}
We analyze the time complexity of Borůvka's algorithm and demonstrate that it can be implemented to achieve the stated bound.

Recall that Borůvka's algorithm operates in phases. In each phase, every connected component selects its minimum-weight outgoing edge, and all selected edges are added to the spanning forest simultaneously. The algorithm terminates when only one component remains.

\textbf{Key observation:} Each phase reduces the number of connected components by at least half. To see this, note that when a component $C$ selects its minimum-weight edge $e = (u,v)$ with $v \in C'$ where $C' \neq C$, the components $C$ and $C'$ will be merged. Since each component selects at least one edge and each edge connects two distinct components, at most $k/2$ edges are selected when there are $k$ components, and each selected edge causes at least one merge. However, more precisely, if there are $k$ components at the start of a phase, each component selects one outgoing edge, potentially creating $k$ edge selections. Since each edge can be selected by at most two components (once from each endpoint's component), at least $k/2$ distinct edges are added, causing at least $k/2$ merges. Therefore, the number of components after the phase is at most $k - k/2 = k/2$.

This implies that the number of phases is at most $\log_2 n$, since we start with $n$ components (one per vertex) and end with $1$ component.

\textbf{Implementation and time complexity:}

We now describe how each phase can be implemented efficiently. In each phase, we must:
\begin{enumerate}
\item[(i)] Identify all edges and determine which component each endpoint belongs to.
\item[(ii)] For each component, find the minimum-weight outgoing edge.
\item[(iii)] Add the selected edges and update the component structure.
\end{enumerate}

For step (i), we maintain a union-find data structure to track components. For each edge $(u,v)$, we check if $u$ and $v$ belong to different components using the find operation.

For step (ii), we can maintain an array indexed by component representatives, storing the minimum-weight outgoing edge for each component. We iterate through all $m$ edges once per phase, updating this array.

For step (iii), we perform union operations for all selected edges.

Using a union-find structure with union by rank and path compression, each find and union operation takes amortized $O(\alpha(n))$ time, where $\alpha$ is the inverse Ackermann function, which is effectively constant for all practical purposes. Thus, processing all edges in a phase takes $O(m \alpha(n)) = O(m)$ time.

Since there are at most $\log n$ phases and each phase processes all $m$ edges, the total time complexity is $O(m \log n)$.

\textbf{Alternative implementation:} 

We can also implement Borůvka's algorithm by explicitly maintaining an adjacency structure for each component. In each phase, for each component, we examine all edges incident to vertices in that component to find the minimum-weight outgoing edge. 

Let $k$ be the number of components in a given phase. The total work in examining edges is proportional to the sum over all components of the number of edges incident to that component, which is $O(m)$ per phase when components are disjoint.

However, we can also bound the total work differently. Over all $\log n$ phases, each edge is examined at most once per phase it remains relevant. More carefully, we note that the total work across all phases can be bounded by considering that we do at most $n$ union operations total (since we merge $n$ components into $1$), and the total edge examination work is $O(n^2)$ in the worst case for dense graphs.

Specifically, if we implement each phase naively by checking all $O(n^2)$ possible edges for the $k$ components in that phase, and there are $O(\log n)$ phases, we obtain $O(n^2 \log n)$ time. However, by observing that we only need to examine edges once to identify component memberships and by using appropriate data structures, we can reduce this to $O(n^2)$ total time by maintaining explicit edge lists per component and ensuring each edge contributes $O(1)$ work per phase, giving $O(m \log n)$ total, which is $O(n^2 \log n)$ in the worst case but $O(n^2)$ when optimized with careful bookkeeping to avoid re-examining merged components.

More precisely, the $O(n^2)$ bound can be achieved by observing that the total number of edges ever examined across all phases, counting multiplicities, is bounded by $O(n^2)$ since each of the $O(n^2)$ potential edges is examined at most $O(1)$ times effectively when using proper data structures that track which edges have been considered.

Therefore, the algorithm can be implemented to run in time $O(\min\{m \log n, n^2\})$ by choosing the appropriate implementation strategy based on the graph density: use the union-find approach for sparse graphs and the component-based approach for dense graphs.
\end{proof}

\subsection{Problem 18}

\subsubsection{Problem Statement}
\begin{theorem}
The Algorithm MST has the same worst-case running time as Borůvka's algorithm, i.e., $O(\min\{m \log n, n^2\})$.
\end{theorem}

\subsubsection{GPT}
\begin{proof}
Let $G=(V,E,w)$ be a connected undirected graph with $|V|=n$ and $|E|=m$, and let the algorithm under consideration proceed by Borůvka phases: in every phase, for each connected component $C$ of the current forest, choose a lightest edge crossing the cut $(C,V\setminus C)$ and add all such chosen edges to the forest, then contract the newly formed connected components. By the cut property, every chosen edge is safe, and contracting preserves minimum spanning trees; thus after finitely many phases a spanning tree of minimum weight is obtained. It suffices to bound the total running time.

First, the number of phases is at most $\lceil \log_2 n\rceil$. Indeed, consider any phase and any current component $C$. Since $G$ is connected and the chosen edge for $C$ leaves $C$, after adding all chosen edges, every component is joined to at least one other component. Consequently, every new component contains at least two old components, and the number of components is reduced by a factor of at least $2$. Starting from $n$ singleton components, after $t$ phases the number of components is at most $n/2^t$, so after $\lceil \log_2 n\rceil$ phases there is a single component.

We analyze two implementations and combine the bounds.

For the first implementation, represent $G$ as an edge list and maintain a disjoint-set union (DSU) structure over $V$ to track current components. In a given phase, scan each edge $e=\{u,v\}\in E$ once, find the current components $[u]$ and $[v]$ via $\operatorname{find}(\cdot)$, and, if $[u]\neq [v]$, use $w(e)$ to update the current best outgoing edge for both $[u]$ and $[v]$. This scan is $O(m)$, since each edge is examined once and each $\operatorname{find}$ is $O(\alpha(n))=O(1)$ amortized, and the per-component updates are $O(1)$. After the scan, at most one union operation per component is performed, so in total $O(n)$ unions over the entire run, contributing $O(n\alpha(n))$. Hence one phase costs $O(m)$, and the total over all phases is $O(m\log n)$.

For the second implementation, represent the current contracted graph at the start of phase $i$ by a complete $n_i\times n_i$ matrix $M^{(i)}$ whose entry $M^{(i)}[A,B]$ for components $A\neq B$ stores the minimum edge weight between a vertex of $A$ and a vertex of $B$ (take $+\infty$ if no such edge exists), and with $M^{(i)}[A,A]=+\infty$. Given $M^{(i)}$, the lightest outgoing edge of each component $A$ can be found by a linear scan over the row $M^{(i)}[A,\cdot]$. Thus, phase $i$ requires $\Theta(n_i^2)$ time to determine all chosen edges. After choosing and adding the edges, contract the components to form phase $i{+}1$; the new matrix $M^{(i+1)}$ can be computed from $M^{(i)}$ by taking, for each pair of new components $A',B'$, the entry $M^{(i+1)}[A',B']=\min\{M^{(i)}[A,B]:A\subseteq A',\,B\subseteq B'\}$, which can be produced within the same $\Theta(n_i^2)$ time bound by aggregating minima when forming $A'$ and $B'$. Let $n_i$ be the number of components at the start of phase $i$. As proved above, $n_{i+1}\le \lceil n_i/2\rceil$, hence $n_i\le \lceil n/2^i\rceil$. Therefore
\[
\sum_{i\ge 0}\Theta(n_i^2)\;\le\;\sum_{i\ge 0}\Theta\!\left(\left\lceil \frac{n}{2^i}\right\rceil^2\right)\;=\;\Theta\!\left(\sum_{i\ge 0}\frac{n^2}{4^i}\right)\;=\;\Theta(n^2),
\]
and the total cost of all phases, including DSU and contractions, is $O(n^2)$.

Combining the two implementations, we conclude that the algorithm can be executed within $O(m\log n)$ time when accessed via an edge list and within $O(n^2)$ time when accessed via an adjacency-matrix style contraction, and hence within $O(\min\{m\log n,n^2\})$ time by choosing the better bound. Since Borůvka's algorithm is precisely the above phase-based procedure and admits the same two implementations with the same analyses, it has running time $O(\min\{m\log n,n^2\})$ as well. This proves the theorem.
\end{proof}

\subsubsection{Grok}
\begin{proof}
We estimate the worst-case total number of edges processed across all subproblems in two different ways, which will bound the running time.

First, observe that there are no multiple edges in any subproblem. Thus, a subproblem at depth $d$ in the recursion tree contains at most $\left( \frac{n}{4^d} \right)^2 / 2$ edges. Summing the maximum number of edges over all subproblems at depth $d$ and over all depths $d = 0, 1, \dots, O(\log n)$ yields a total of $O(n^2)$ edges.

Second, consider the left and right child subproblems of some parent subproblem on a graph with $u$ vertices and $e$ edges. Every edge in the parent subproblem is assigned to exactly one of the children (to the left if selected in Step 2, to the right otherwise), except for the edges in the minimum spanning forest $F$ of the sample graph $H$, which are assigned to both children, and the edges removed in Step 1, which are assigned to neither. Let $u'$ be the number of vertices after Step 1. Then $|F| = u' - 1 \leq u/4$. Moreover, at least $u/2$ edges are removed in Step 1. Therefore, the total number of edges in the two child subproblems is at most $e + (u/4) \leq e + (u/2) \leq e$, since the removed edges are at least $u/2$. It follows by induction that the total number of edges in all subproblems at any fixed recursive depth $d$ is at most $m$. As there are $O(\log n)$ depths, the total number of edges over all subproblems is $O(m \log n)$.

Since the running time of the algorithm is linear in the total number of edges processed across all subproblems, the worst-case running time is $O(\min\{n^2, m \log n\})$. This matches the worst-case running time of Borůvka's algorithm.
\end{proof}

\subsubsection{Gemini}
\begin{proof}
Let $G = (V, E)$ be a graph with $n = |V|$ vertices and $m = |E|$ edges. We analyze the worst-case running time of Bor\r{u}vka's algorithm (referred to as Algorithm MST), which proceeds in phases. In each phase, for every connected component $C$ in the current graph partition, the algorithm selects the minimum weight edge connecting a vertex in $C$ to a vertex outside $C$. These edges are added to the Minimum Spanning Tree (MST), and the components are merged.

Let $n_i$ denote the number of connected components at the beginning of phase $i$, with $n_1 = n$. Since every component selects an edge to connect to a distinct component, and edges are merged, the number of components reduces by at least half in each phase. That is, $n_{i+1} \le \frac{n_i}{2}$. Consequently, the algorithm terminates after at most $\lceil \log_2 n \rceil$ phases.

We establish the time complexity bound $O(\min\{m \log n, n^2\})$ by analyzing two implementation strategies corresponding to sparse and dense graphs.

First, consider the bound $O(m \log n)$. In each phase, the algorithm must identify the minimum weight incident edge for each of the current components. By maintaining the graph edges in an adjacency list or a simple edge list, we can scan all $m$ edges to determine the minimum outgoing edge for every component. This scan takes $O(m)$ time. The subsequent operations to merge components and update component identifiers can be performed in $O(m)$ time using standard graph traversal methods (e.g., Depth First Search) or efficient disjoint-set data structures. Since there are $O(\log n)$ phases and each phase requires $O(m)$ work, the total running time is:
\[
T(n, m) = \sum_{i=1}^{\lceil \log_2 n \rceil} O(m) = O(m \log n).
\]

Second, consider the bound $O(n^2)$. This bound is achievable by implementing the algorithm using adjacency matrices and explicitly contracting the graph vertices. In phase $i$, let the graph be represented by an adjacency matrix of size $n_i \times n_i$. Finding the minimum weight edge for all $n_i$ components involves scanning the rows of the matrix, which requires $O(n_i^2)$ time. Following the edge selection, the graph is contracted. Constructing the new adjacency matrix of size $n_{i+1} \times n_{i+1}$ involves mapping the entries of the previous matrix to the new components, taking the minimum weight for parallel edges. This contraction step is also bounded by the size of the matrix, $O(n_i^2)$.
The total time complexity is the sum of the work over all phases. Using the recurrence $n_{i+1} \le n_i / 2$, we have:
\[
T(n, m) = \sum_{i=1}^{\lceil \log_2 n \rceil} O(n_i^2) \le \sum_{k=0}^{\infty} c \cdot \left(\frac{n}{2^k}\right)^2,
\]
for some constant $c$. This forms a geometric series:
\[
T(n, m) \le c \cdot n^2 \sum_{k=0}^{\infty} \frac{1}{4^k} = c \cdot n^2 \cdot \frac{4}{3} = O(n^2).
\]

Since the algorithm can be implemented to satisfy either bound, the worst-case running time is determined by the minimum of the two complexities. Thus, the running time is $O(\min\{m \log n, n^2\})$.
\end{proof}

\subsubsection{Claude}
\begin{proof}
We analyze the worst-case running time of Algorithm MST by examining its structure and operations.

Algorithm MST operates in phases, where each phase finds edges connecting components and merges them. We analyze the number of phases and the work per phase.

First, we establish an upper bound on the number of phases. Let $k_i$ denote the number of components after phase $i$, with $k_0 = n$. In each phase, each component finds a minimum-weight edge leaving it. Since these edges connect distinct components and no cycles are formed, at least half of the components are merged in each phase. Therefore, $k_i \leq k_{i-1}/2$, which implies $k_i \leq n/2^i$. The algorithm terminates when $k_i = 1$, so the number of phases is at most $\lceil \log_2 n \rceil = O(\log n)$.

Next, we analyze the work per phase. In each phase, the algorithm must:
\begin{enumerate}
\item[(i)] For each component, find the minimum-weight edge leaving that component.
\item[(ii)] Add selected edges to the MST.
\item[(iii)] Update the component structure.
\end{enumerate}

For a graph with $m$ edges and $n$ vertices, examining all edges to find minimum-weight edges leaving each component requires $O(m)$ time per phase. Adding edges and updating components using a union-find data structure with union-by-rank and path compression requires $O(n \alpha(n))$ time per phase, where $\alpha$ is the inverse Ackermann function, which is $O(n)$ for practical purposes.

Therefore, the total running time is $O(m \log n + n \log n) = O(m \log n)$ since $m \geq n - 1$ for connected graphs.

Now we consider the alternative bound. If we represent the graph using an adjacency matrix, we can find the minimum-weight edge leaving each component in $O(n^2)$ time per phase by scanning all possible edges. With $O(\log n)$ phases, this gives a total running time of $O(n^2 \log n)$. However, by maintaining additional data structures, the per-phase time can be reduced to $O(n^2 / \log n)$, yielding $O(n^2)$ total time.

More directly, for dense graphs where $m = \Theta(n^2)$, the $O(m \log n) = O(n^2 \log n)$ bound applies. However, Borůvka's algorithm can be implemented to run in $O(n^2)$ time by using appropriate data structures to avoid redundant edge comparisons across phases.

The key insight is that the worst-case running time is bounded by the minimum of two approaches:
\begin{itemize}
\item Edge-based approach: $O(m \log n)$ time by processing all edges in each of $O(\log n)$ phases.
\item Vertex-based approach: $O(n^2)$ time using matrix representations or by amortizing edge comparisons.
\end{itemize}

Therefore, the worst-case running time of Algorithm MST is $O(\min\{m \log n, n^2\})$.

For sparse graphs where $m = o(n^2 / \log n)$, the $O(m \log n)$ bound is better. For dense graphs where $m = \omega(n^2 / \log n)$, the $O(n^2)$ bound is better. This matches the complexity of Borůvka's algorithm.
\end{proof}

\section{Problems in Chapter 11}\label{app:problems_chapter11}

\subsection{Problem 1}

\subsubsection{Problem Statement}
\begin{theorem}
Consider a circle of diameter 1 enclosed within a square with sides of length 1. We sample $N$ points uniformly and independently from the square and define indicator variables $X_i = 1$ if the $i$-th point is inside the circle, and $X_i = 0$ otherwise. Let $X = \sum_{i=1}^{N} X_i$. Then $\mathbb{E}[X] = N\pi/4$.

For any $\delta > 0$ and positive integer $d$, to obtain an estimator $\hat{\pi} = 4X/N$ of $\pi$ that is accurate to $d$ digits with probability at least $1 - \delta$, it suffices to choose
$$N \geq \frac{16}{\delta \cdot 10^{-2d}}.$$
\end{theorem}

\subsubsection{GPT}
\begin{proof}
We are given a circle of diameter 1 enclosed within a square with sides of length 1. Let $X_1, X_2, \ldots, X_N$ be independent indicator random variables, where $X_i = 1$ if the $i$-th point sampled uniformly from the square is inside the circle, and $X_i = 0$ otherwise. Let $X = \sum_{i=1}^{N} X_i$. We are tasked with proving that $\mathbb{E}[X] = N \pi / 4$.

First, observe that the probability that a point sampled uniformly from the square is inside the circle is the ratio of the area of the circle to the area of the square. Since the square has side length 1, its area is 1. The circle has diameter 1, so its radius is $1/2$, and its area is given by the formula for the area of a circle, $A_{\text{circle}} = \pi r^2$. Substituting $r = 1/2$, we get
\[
A_{\text{circle}} = \pi \left(\frac{1}{2}\right)^2 = \frac{\pi}{4}.
\]
Thus, the probability that a point chosen randomly from the square lies inside the circle is
\[
P(X_i = 1) = \frac{\pi}{4}.
\]

Next, recall that the expectation of an indicator random variable is the probability that the event it represents occurs. Therefore, for each $i$, we have
\[
\mathbb{E}[X_i] = P(X_i = 1) = \frac{\pi}{4}.
\]

Since the $X_i$'s are independent, the expectation of the sum $X = \sum_{i=1}^{N} X_i$ is the sum of the expectations of the individual $X_i$'s:
\[
\mathbb{E}[X] = \mathbb{E}\left[\sum_{i=1}^{N} X_i\right] = \sum_{i=1}^{N} \mathbb{E}[X_i] = \sum_{i=1}^{N} \frac{\pi}{4} = N \cdot \frac{\pi}{4}.
\]
Thus, we have
\[
\mathbb{E}[X] = \frac{N\pi}{4}.
\]

Now, we turn to the second part of the problem, which asks us to find an estimator of $\pi$ that is accurate to $d$ digits with probability at least $1 - \delta$. The quantity $X / N$ is an estimator for the probability that a randomly chosen point is inside the circle, which is $\pi / 4$. Therefore, to estimate $\pi$, we use the formula
\[
\hat{\pi} = \frac{4X}{N}.
\]
The expectation of $\hat{\pi}$ is
\[
\mathbb{E}[\hat{\pi}] = \mathbb{E}\left[\frac{4X}{N}\right] = \frac{4}{N} \mathbb{E}[X] = \frac{4}{N} \cdot \frac{N\pi}{4} = \pi.
\]
Thus, $\hat{\pi}$ is an unbiased estimator for $\pi$.

Next, we analyze the variance of $\hat{\pi}$. Since $X = \sum_{i=1}^{N} X_i$, the variance of $X$ is
\[
\text{Var}(X) = \text{Var}\left(\sum_{i=1}^{N} X_i\right) = \sum_{i=1}^{N} \text{Var}(X_i),
\]
because the $X_i$'s are independent. The variance of each $X_i$ is
\[
\text{Var}(X_i) = \mathbb{E}[X_i^2] - (\mathbb{E}[X_i])^2 = \mathbb{E}[X_i] - (\mathbb{E}[X_i])^2 = \frac{\pi}{4} - \left(\frac{\pi}{4}\right)^2 = \frac{\pi}{4} \left(1 - \frac{\pi}{4}\right).
\]
Thus, the variance of $X$ is
\[
\text{Var}(X) = N \cdot \frac{\pi}{4} \left(1 - \frac{\pi}{4}\right).
\]
The variance of $\hat{\pi}$ is then
\[
\text{Var}(\hat{\pi}) = \text{Var}\left(\frac{4X}{N}\right) = \frac{16}{N^2} \text{Var}(X) = \frac{16}{N^2} \cdot N \cdot \frac{\pi}{4} \left(1 - \frac{\pi}{4}\right) = \frac{4\pi}{N} \left(1 - \frac{\pi}{4}\right).
\]

To estimate $\pi$ with accuracy to $d$ digits, we require that the standard deviation of $\hat{\pi}$ be smaller than $10^{-d}$. The standard deviation of $\hat{\pi}$ is
\[
\text{SD}(\hat{\pi}) = \sqrt{\text{Var}(\hat{\pi})} = \sqrt{\frac{4\pi}{N} \left(1 - \frac{\pi}{4}\right)}.
\]
To ensure that $\hat{\pi}$ is accurate to $d$ digits with probability at least $1 - \delta$, we want the standard deviation to satisfy
\[
\text{SD}(\hat{\pi}) \leq \frac{10^{-d}}{2}.
\]
Squaring both sides and solving for $N$, we get
\[
\frac{4\pi}{N} \left(1 - \frac{\pi}{4}\right) \leq \frac{10^{-2d}}{4}.
\]
This simplifies to
\[
N \geq \frac{100 \cdot 10^{2d}}{\pi \left(1 - \frac{\pi}{4}\right)}.
\]
Since $\pi \left(1 - \frac{\pi}{4}\right)$ is approximately $\frac{\pi}{4}$, we conclude that
\[
N \leq \frac{100 \cdot 10^{2d}}{\delta}.
\]
Thus, the required number of samples is at most $\frac{100 \cdot 10^{2d}}{\delta}$.
\end{proof}

\subsubsection{Grok}
\begin{proof}
The area of the square is $1$. The circle has diameter $1$, so radius $r = 1/2$ and area $\pi r^2 = \pi/4$. Thus, the probability that a point sampled uniformly from the square lies inside the circle is $p = \pi/4$. Each $X_i$ is a Bernoulli random variable with success probability $p$, so $\mathbb{E}[X_i] = p$. By linearity of expectation,
\[
\mathbb{E}[X] = \sum_{i=1}^N \mathbb{E}[X_i] = Np = \frac{N\pi}{4}.
\]

Now consider the estimator $\hat{\pi} = 4X/N$. This is unbiased since $\mathbb{E}[\hat{\pi}] = 4\mathbb{E}[X]/N = \pi$. To show that $N \geq 16/(\delta \cdot 10^{-2d})$ suffices for $\hat{\pi}$ to be accurate to $d$ digits with probability at least $1 - \delta$, interpret ``accurate to $d$ digits'' to mean that $\hat{\pi}$ rounded to $d$ decimal places equals $\pi$ rounded to $d$ decimal places. This holds if $|\hat{\pi} - \pi| < 0.5 \times 10^{-d}$.

Let $\eta = 0.5 \times 10^{-d}$. The variance of each $X_i$ satisfies $\mathrm{Var}(X_i) = p(1-p) \leq 1/4$, since $p(1-p) \leq 1/4$ for $p \in [0,1]$. Thus,
\[
\mathrm{Var}(X) = N \cdot \mathrm{Var}(X_i) \leq N/4, \quad \mathrm{Var}(\hat{\pi}) = 16 \cdot \mathrm{Var}(X/N) = 16 \cdot \mathrm{Var}(X)/N^2 \leq 16 \cdot (N/4) / N^2 = 4/N.
\]
By Chebyshev's inequality,
\[
\mathbb{P}(|\hat{\pi} - \pi| \geq \eta) \leq \frac{\mathrm{Var}(\hat{\pi})}{\eta^2} \leq \frac{4/N}{\eta^2} = \frac{4}{N \eta^2}.
\]
Set $\frac{4}{N \eta^2} \leq \delta$, so $N \geq 4/(\delta \eta^2)$. Substitute $\eta = 0.5 \times 10^{-d}$, yielding $\eta^2 = 0.25 \times 10^{-2d}$ and
\[
N \geq \frac{4}{\delta \cdot 0.25 \times 10^{-2d}} = \frac{16}{\delta \cdot 10^{-2d}}.
\]
Thus, with this choice of $N$, $\mathbb{P}(|\hat{\pi} - \pi| < 0.5 \times 10^{-d}) \geq 1 - \delta$, ensuring accuracy to $d$ digits.
\end{proof}

\subsubsection{Gemini}
\begin{proof}
Let $S = [0,1] \times [0,1]$ be the unit square. The area of the square is $\text{Area}(S) = 1$. Let $C$ be the circle of diameter $1$ enclosed within $S$. The radius of $C$ is $r = 1/2$, and its area is $\text{Area}(C) = \pi r^2 = \pi/4$.

The points $P_1, \dots, P_N$ are sampled uniformly and independently from $S$. The random variables $X_i$ are defined as:
\[ X_i = \begin{cases} 1 & \text{if } P_i \in C, \\ 0 & \text{if } P_i \notin C. \end{cases} \]
Since the sampling is uniform, $X_i$ is a Bernoulli random variable with parameter $p$ given by the ratio of the areas:
\[ p = \mathbb{P}(X_i = 1) = \frac{\text{Area}(C)}{\text{Area}(S)} = \frac{\pi/4}{1} = \frac{\pi}{4}. \]
The expectation of a single indicator variable is $\mathbb{E}[X_i] = p = \frac{\pi}{4}$. By the linearity of expectation, the expectation of the sum $X = \sum_{i=1}^N X_i$ is:
\[ \mathbb{E}[X] = \sum_{i=1}^N \mathbb{E}[X_i] = N \cdot \frac{\pi}{4} = \frac{N\pi}{4}. \]
This establishes the first claim of the theorem.

For the second claim, consider the estimator $\hat{\pi} = \frac{4X}{N}$. The expectation of $\hat{\pi}$ is:
\[ \mathbb{E}[\hat{\pi}] = \frac{4}{N} \mathbb{E}[X] = \frac{4}{N} \left( \frac{N\pi}{4} \right) = \pi. \]
The variance of the Bernoulli variable $X_i$ is $\text{Var}(X_i) = p(1-p)$. Since the samples are independent, the variance of $X$ is $\text{Var}(X) = N p(1-p)$. The variance of the estimator is:
\[ \text{Var}(\hat{\pi}) = \text{Var}\left(\frac{4X}{N}\right) = \frac{16}{N^2} \text{Var}(X) = \frac{16 p(1-p)}{N}. \]
Using the upper bound for the variance of a Bernoulli variable, $p(1-p) \leq \frac{1}{4}$, we obtain:
\[ \text{Var}(\hat{\pi}) \leq \frac{16}{N} \cdot \frac{1}{4} = \frac{4}{N}. \]
We apply Chebyshev's inequality, which states that for a random variable $Y$ with finite mean $\mu$ and variance $\sigma^2$, $\mathbb{P}(|Y - \mu| \geq \delta) \leq \frac{\sigma^2}{\delta^2}$. Substituting $Y = \hat{\pi}$, $\mu = \pi$, and the variance bound:
\[ \mathbb{P}(|\hat{\pi} - \pi| \geq \delta) \leq \frac{\text{Var}(\hat{\pi})}{\delta^2} \leq \frac{4}{N\delta^2}. \]
The probability of the complementary event is:
\[ \mathbb{P}(|\hat{\pi} - \pi| \leq \delta) = 1 - \mathbb{P}(|\hat{\pi} - \pi| > \delta) \geq 1 - \frac{4}{N\delta^2}. \]
\end{proof}

\subsubsection{Claude}
\begin{proof}
We first establish the expectation $\mathbb{E}[X]$. Since the points are sampled uniformly and independently from the unit square, the probability that any single point falls inside the circle equals the ratio of the area of the circle to the area of the square. The circle has diameter 1, hence radius $r = 1/2$, giving area $\pi r^2 = \pi/4$. The square has area $1^2 = 1$. Therefore, $\mathbb{P}(X_i = 1) = \pi/4$ for each $i \in \{1, \ldots, N\}$.

By linearity of expectation,
\[\mathbb{E}[X] = \mathbb{E}\left[\sum_{i=1}^{N} X_i\right] = \sum_{i=1}^{N} \mathbb{E}[X_i] = \sum_{i=1}^{N} \frac{\pi}{4} = \frac{N\pi}{4}.\]

Next, we determine the sample size $N$ required for the estimator $\hat{\pi} = 4X/N$ to be accurate to $d$ digits with probability at least $1 - \delta$. To have accuracy to $d$ digits means $|\hat{\pi} - \pi| < 10^{-d}$, which is equivalent to
\[\left|\frac{4X}{N} - \pi\right| < 10^{-d}.\]

Multiplying by $N/4$, this becomes
\[\left|X - \frac{N\pi}{4}\right| < \frac{N \cdot 10^{-d}}{4}.\]

Note that $X_i$ are independent Bernoulli random variables with $p = \pi/4$. The variance of each $X_i$ is $\text{Var}(X_i) = p(1-p) = \frac{\pi}{4}\left(1 - \frac{\pi}{4}\right) = \frac{\pi(4-\pi)}{16}$. Since $\pi \approx 3.14159$, we have $\pi(4-\pi) < \pi \cdot 4 = 4\pi < 16$, so $\text{Var}(X_i) < 1$. By independence,
\[\text{Var}(X) = \sum_{i=1}^{N} \text{Var}(X_i) = N \cdot \frac{\pi(4-\pi)}{16} < N.\]

Applying Chebyshev's inequality, for any $\epsilon > 0$,
\[\mathbb{P}\left(\left|X - \mathbb{E}[X]\right| \geq \epsilon\right) \leq \frac{\text{Var}(X)}{\epsilon^2}.\]

Setting $\epsilon = \frac{N \cdot 10^{-d}}{4}$, we obtain
\[\mathbb{P}\left(\left|X - \frac{N\pi}{4}\right| \geq \frac{N \cdot 10^{-d}}{4}\right) \leq \frac{N}{\left(\frac{N \cdot 10^{-d}}{4}\right)^2} = \frac{N \cdot 16}{N^2 \cdot 10^{-2d}} = \frac{16}{N \cdot 10^{-2d}}.\]

To ensure this probability is at most $\delta$, we require
\[\frac{16}{N \cdot 10^{-2d}} \leq \delta,\]
which gives
\[N \geq \frac{16}{\delta \cdot 10^{-2d}}.\]

Therefore, choosing $N \geq \frac{16}{\delta \cdot 10^{-2d}}$ ensures that $\mathbb{P}(|\hat{\pi} - \pi| < 10^{-d}) \geq 1 - \delta$, establishing that the estimator $\hat{\pi}$ is accurate to $d$ digits with probability at least $1 - \delta$.
\end{proof}

\subsection{Problem 2}

\subsubsection{Problem Statement}
\begin{theorem}
Let $F$ be a DNF formula and consider the Coverage algorithm variant by Karp, Luby, and Madras. In the $i$-th trial, the algorithm picks a random clause $C_i$ with probability proportional to the number of satisfying truth assignments for it, then selects a random satisfying truth assignment $\mathbf{a}$ for the chosen clause. Define the random variable $X_i = 1/|\text{cov}(\mathbf{a})|$, where $\text{cov}(\mathbf{a})$ denotes the set of clauses satisfied by the truth assignment $\mathbf{a}$.

Let $\eta$ denote the sum of the sizes of the coverage sets over all possible truth assignments. The estimator for $\#F$ is the random variable
\[
Y = \eta \times \frac{1}{N} \sum_{i=1}^{N} X_i
\]

Then $Y$ is an $(\varepsilon, \delta)$-approximation for $\#F$ when
\[
N = \frac{c\eta}{\varepsilon^2} \ln \frac{1}{\delta}
\]
for some small constant $c$.
\end{theorem}

\subsubsection{GPT}
\begin{proof}
We are tasked with proving that the estimator $Y = \eta \times \sum_{i=1}^{N} \frac{X_i}{N}$ is an $(\varepsilon, \delta)$-approximation for $\#F$ when $N = \frac{c\eta}{\varepsilon} \ln \frac{1}{\delta}$ for some small constant $c$.

 Step 1: Define the problem and the estimator
The formula $F$ is a CNF formula, and the Coverage algorithm is designed to approximate the number of satisfying assignments, denoted by $\#F$. The Coverage algorithm works by performing $N$ trials. In each trial, a random clause $C_i$ is selected with a probability proportional to the number of satisfying assignments for $C_i$. For each chosen clause, a random satisfying truth assignment $\mathbf{a}$ is selected, and the random variable $X_i$ is defined as:

\[
X_i = \frac{1}{|\mathrm{cov}(\mathbf{a})|}
\]

where $\mathrm{cov}(\mathbf{a})$ denotes the set of clauses satisfied by the truth assignment $\mathbf{a}$. The estimator for $\#F$ is given by:

\[
Y = \eta \times \sum_{i=1}^{N} \frac{X_i}{N}
\]

where $\eta$ is the sum of the sizes of the coverage sets over all possible truth assignments.

 Step 2: Expected value of the estimator $Y$
We begin by calculating the expected value of the estimator $Y$. By linearity of expectation, we have:

\[
\mathbb{E}[Y] = \mathbb{E}\left[\eta \times \sum_{i=1}^{N} \frac{X_i}{N}\right] = \eta \times \sum_{i=1}^{N} \frac{\mathbb{E}[X_i]}{N}
\]

Thus, we need to compute $\mathbb{E}[X_i]$. Since $X_i = \frac{1}{|\mathrm{cov}(\mathbf{a})|}$, we first compute the expected size of the coverage set. Denote the probability that a given clause $C_j$ is satisfied by a random assignment $\mathbf{a}$ as $p_j$. The expected number of clauses satisfied by a random truth assignment is:

\[
\mathbb{E}[|\mathrm{cov}(\mathbf{a})|] = \sum_{j=1}^m p_j
\]

where $m$ is the number of clauses in $F$. Therefore, we have:

\[
\mathbb{E}[X_i] = \frac{1}{\mathbb{E}[|\mathrm{cov}(\mathbf{a})|]} = \frac{1}{\sum_{j=1}^m p_j}
\]

 Step 3: Bound on the variance of the estimator $Y$
Next, we compute the variance of $Y$. Using the properties of variance and the fact that the $X_i$ are independent, we have:

\[
\text{Var}(Y) = \text{Var}\left(\eta \times \sum_{i=1}^{N} \frac{X_i}{N}\right) = \eta^2 \times \frac{1}{N^2} \sum_{i=1}^{N} \text{Var}(X_i)
\]

The variance of $X_i$ can be computed as:

\[
\text{Var}(X_i) = \mathbb{E}[X_i^2] - \mathbb{E}[X_i]^2
\]

where $\mathbb{E}[X_i^2]$ is the second moment of $X_i$. Since $X_i = \frac{1}{|\mathrm{cov}(\mathbf{a})|}$, we have:

\[
\mathbb{E}[X_i^2] = \frac{1}{\mathbb{E}[|\mathrm{cov}(\mathbf{a})|]^2}
\]

Thus, the variance is bounded by:

\[
\text{Var}(X_i) \leq \frac{1}{(\mathbb{E}[|\mathrm{cov}(\mathbf{a})|])^2}
\]

which leads to:

\[
\text{Var}(Y) \leq \frac{\eta^2}{N^2} \times N \times \frac{1}{(\mathbb{E}[|\mathrm{cov}(\mathbf{a})|])^2} = \frac{\eta^2}{N (\mathbb{E}[|\mathrm{cov}(\mathbf{a})|])^2}
\]

 Step 4: Setting $N$ for $(\varepsilon, \delta)$-approximation
To ensure that $Y$ is an $(\varepsilon, \delta)$-approximation for $\#F$, we need to ensure that:

\[
\mathbb{P}(|Y - \mathbb{E}[Y]| \geq \varepsilon \cdot \#F) \leq \delta
\]

Using concentration bounds (e.g., Chernoff bound), we can guarantee that the probability of the estimator deviating from its expectation by more than $\varepsilon \cdot \#F$ is small if $N$ is sufficiently large. Specifically, we choose:

\[
N = \frac{c\eta}{\varepsilon} \ln \frac{1}{\delta}
\]

where $c$ is a constant that ensures the bound holds. This choice of $N$ ensures that with high probability, the estimator $Y$ is within a factor of $(1 \pm \varepsilon)$ of the true count $\#F$, satisfying the $(\varepsilon, \delta)$-approximation condition.

 Step 5: Conclusion
Therefore, the estimator $Y = \eta \times \sum_{i=1}^{N} \frac{X_i}{N}$ is an $(\varepsilon, \delta)$-approximation for $\#F$ when $N = \frac{c\eta}{\varepsilon} \ln \frac{1}{\delta}$ for some small constant $c$.

\end{proof}

\subsubsection{Grok}

\subsubsection{Gemini}
\begin{proof}
Let $C_1, \dots, C_m$ be the clauses of the Boolean formula $F$. Let $S_i$ denote the set of satisfying truth assignments for the clause $C_i$. The quantity we wish to estimate is $\#F = |\bigcup_{i=1}^m S_i|$. Let $S = \bigcup_{i=1}^m S_i$.

Let $\Omega$ be the set of pairs $(i, \mathbf{a})$ such that $1 \le i \le m$ and $\mathbf{a} \in S_i$. The size of this set is given by
\[
|\Omega| = \sum_{i=1}^m |S_i| = \eta.
\]
The algorithm samples a pair $(I, \mathbf{A})$ from $\Omega$ with probability
\[
\mathbb{P}((I, \mathbf{A}) = (i, \mathbf{a})) = \frac{|S_i|}{\eta} \times \frac{1}{|S_i|} = \frac{1}{\eta}.
\]
Thus, the pair $(I, \mathbf{A})$ is uniformly distributed over $\Omega$.

Let $X$ be the random variable defined by $X = 1/|\mathrm{cov}(\mathbf{A})|$, where $\mathrm{cov}(\mathbf{A}) = \{j : \mathbf{A} \in S_j\}$. The expected value of $X$ is
\[
\mathbb{E}[X] = \sum_{(i, \mathbf{a}) \in \Omega} \frac{1}{\eta} \cdot \frac{1}{|\mathrm{cov}(\mathbf{a})|}.
\]
We can rewrite the sum by grouping terms based on the truth assignment $\mathbf{a}$. Note that a truth assignment $\mathbf{a} \in S$ appears in the set $\Omega$ exactly $|\mathrm{cov}(\mathbf{a})|$ times (once for each clause it satisfies). Therefore,
\[
\mathbb{E}[X] = \frac{1}{\eta} \sum_{\mathbf{a} \in S} \sum_{i \in \mathrm{cov}(\mathbf{a})} \frac{1}{|\mathrm{cov}(\mathbf{a})|} = \frac{1}{\eta} \sum_{\mathbf{a} \in S} |\mathrm{cov}(\mathbf{a})| \cdot \frac{1}{|\mathrm{cov}(\mathbf{a})|} = \frac{1}{\eta} \sum_{\mathbf{a} \in S} 1 = \frac{\#F}{\eta}.
\]
The estimator $Y$ is defined as $Y = \frac{\eta}{N} \sum_{t=1}^N X_t$, where $X_t$ are independent copies of $X$. By linearity of expectation,
\[
\mathbb{E}[Y] = \eta \mathbb{E}[X] = \#F.
\]
Thus, $Y$ is an unbiased estimator for $\#F$.

To determine the number of samples $N$ required for an $(\varepsilon, \delta)$-approximation, we apply the multiplicative Chernoff bound. The random variables $X_t$ take values in the interval $[0, 1]$. Let $\bar{X} = \frac{1}{N} \sum_{t=1}^N X_t$. We require
\[
\mathbb{P}(|Y - \#F| \ge \varepsilon \#F) \le \delta \iff \mathbb{P}(|\eta \bar{X} - \eta \mathbb{E}[X]| \ge \varepsilon \eta \mathbb{E}[X]) \le \delta.
\]
This is equivalent to $\mathbb{P}(|\bar{X} - \mathbb{E}[X]| \ge \varepsilon \mathbb{E}[X]) \le \delta$. Let $\mu = \mathbb{E}[X] = \#F/\eta$. The Chernoff bound for random variables in $[0,1]$ states that
\[
\mathbb{P}(|\bar{X} - \mu| \ge \varepsilon \mu) \le 2 \exp\left(-\frac{N \mu \varepsilon^2}{3}\right).
\]
We set the right-hand side to be at most $\delta$:
\[
2 \exp\left(-\frac{N (\#F/\eta) \varepsilon^2}{3}\right) \le \delta.
\]
Taking the natural logarithm of both sides and rearranging for $N$:
\[
-\frac{N \#F \varepsilon^2}{3\eta} \le \ln(\delta/2) \implies \frac{N \#F \varepsilon^2}{3\eta} \ge \ln(2/\delta).
\]
Thus, we require
\[
N \ge \frac{3\eta}{\#F \varepsilon^2} \ln\left(\frac{2}{\delta}\right).
\]
Since $\#F \ge 1$ (assuming the formula is satisfiable), it suffices to choose
\[
N = \frac{c \eta}{\varepsilon^2} \ln\left(\frac{1}{\delta}\right)
\]
for a sufficiently large constant $c$. This guarantees that $Y$ is an $(\varepsilon, \delta)$-approximation for $\#F$.
\end{proof}

\subsubsection{Claude}
\begin{proof}
We establish that $Y$ is an $(\varepsilon, \delta)$-approximation for $\#F$ by showing that $\mathbb{E}[Y] = \#F$ and that the sample size $N$ guarantees $\mathbb{P}[|Y - \#F| \leq \varepsilon \cdot \#F] \geq 1 - \delta$.

\textbf{Step 1: Unbiasedness of the estimator.}

We first compute $\mathbb{E}[X_i]$ for a single trial. Let $\mathcal{C} = \{C_1, \ldots, C_m\}$ denote the set of clauses in $F$, and let $S(C_j)$ denote the set of satisfying assignments for clause $C_j$. The algorithm selects clause $C_j$ with probability $p_j = |S(C_j)|/\sum_{k=1}^{m} |S(C_k)|$.

Given that clause $C_j$ is selected, the algorithm chooses a uniformly random assignment $\mathbf{a} \in S(C_j)$. Thus,
\[
\mathbb{E}[X_i] = \sum_{j=1}^{m} p_j \cdot \mathbb{E}[X_i \mid C_j \text{ selected}] = \sum_{j=1}^{m} \frac{|S(C_j)|}{\sum_{k=1}^{m} |S(C_k)|} \cdot \frac{1}{|S(C_j)|} \sum_{\mathbf{a} \in S(C_j)} \frac{1}{|\text{cov}(\mathbf{a})|}.
\]

Simplifying, we obtain
\[
\mathbb{E}[X_i] = \frac{1}{\sum_{k=1}^{m} |S(C_k)|} \sum_{j=1}^{m} \sum_{\mathbf{a} \in S(C_j)} \frac{1}{|\text{cov}(\mathbf{a})|}.
\]

By definition, $\eta = \sum_{\mathbf{a} \in \{0,1\}^n} |\text{cov}(\mathbf{a})|$. We can rewrite this as
\[
\eta = \sum_{\mathbf{a} \in \{0,1\}^n} |\text{cov}(\mathbf{a})| = \sum_{\mathbf{a} \in \{0,1\}^n} \sum_{j=1}^{m} \mathbf{1}[\mathbf{a} \in S(C_j)] = \sum_{j=1}^{m} |S(C_j)|.
\]

Moreover, for any assignment $\mathbf{a}$ that satisfies at least one clause, we have
\[
\sum_{j: \mathbf{a} \in S(C_j)} \frac{1}{|\text{cov}(\mathbf{a})|} = \frac{|\text{cov}(\mathbf{a})|}{|\text{cov}(\mathbf{a})|} = 1.
\]

Therefore,
\[
\sum_{j=1}^{m} \sum_{\mathbf{a} \in S(C_j)} \frac{1}{|\text{cov}(\mathbf{a})|} = \sum_{\mathbf{a} \in \{0,1\}^n} \sum_{j: \mathbf{a} \in S(C_j)} \frac{1}{|\text{cov}(\mathbf{a})|} = \#F,
\]
where $\#F$ denotes the number of satisfying assignments for $F$.

Combining these results, we have
\[
\mathbb{E}[X_i] = \frac{\#F}{\eta}.
\]

By linearity of expectation,
\[
\mathbb{E}[Y] = \eta \cdot \frac{1}{N} \sum_{i=1}^{N} \mathbb{E}[X_i] = \eta \cdot \frac{1}{N} \cdot N \cdot \frac{\#F}{\eta} = \#F.
\]

\textbf{Step 2: Concentration bound.}

Note that $X_i \in [0, 1]$ for all $i$, since $|\text{cov}(\mathbf{a})| \geq 1$ for any assignment $\mathbf{a}$ selected by the algorithm. The random variables $X_1, \ldots, X_N$ are independent and identically distributed.

Let $Z = \frac{1}{N} \sum_{i=1}^{N} X_i$. By Hoeffding's inequality, for any $t > 0$,
\[
\mathbb{P}[|Z - \mathbb{E}[Z]| \geq t] \leq 2\exp\left(-\frac{2Nt^2}{(1-0)^2}\right) = 2\exp(-2Nt^2).
\]

We have $\mathbb{E}[Z] = \#F/\eta$, so $Y = \eta Z$ and $\mathbb{E}[Y] = \#F$. We require
\[
\mathbb{P}[|Y - \#F| \leq \varepsilon \cdot \#F] \geq 1 - \delta.
\]

This is equivalent to
\[
\mathbb{P}\left[|\eta Z - \#F| \leq \varepsilon \cdot \#F\right] = \mathbb{P}\left[\left|Z - \frac{\#F}{\eta}\right| \leq \frac{\varepsilon \cdot \#F}{\eta}\right] \geq 1 - \delta.
\]

Setting $t = \varepsilon \cdot \#F/\eta$ in Hoeffding's inequality, we need
\[
2\exp\left(-2N \cdot \frac{\varepsilon^2 (\#F)^2}{\eta^2}\right) \leq \delta.
\]

Taking logarithms and solving for $N$, we obtain
\[
N \geq \frac{\eta^2}{2\varepsilon^2 (\#F)^2} \ln\frac{2}{\delta}.
\]

Since $\#F \leq \eta$ (as each satisfying assignment covers at least one clause), we have $(\#F)^2 \leq \eta \cdot \#F \leq \eta^2$. Thus, it suffices to take
\[
N = \frac{c\eta}{\varepsilon^2} \ln\frac{1}{\delta}
\]
for an appropriate constant $c$ (absorbing the factor of 2 from the logarithm). This completes the proof.
\end{proof}

\subsection{Problem 3}

\subsubsection{Problem Statement}
\begin{theorem}[Converse of Theorem 11.4]
Let $G = (V_1 \cup V_2, E)$ be a bipartite graph. Given an algorithm for estimating the number of matchings in $G$, it is possible to construct a near-uniform generator of matchings in $G$.
\end{theorem}

\subsubsection{GPT}
\begin{proof}
Let $G = (U, V, E)$ be a bipartite graph, where $U$ and $V$ are the two disjoint sets of vertices, and $E \subseteq U \times V$ is the set of edges. We are given an algorithm $\mathcal{A}$ that estimates the number of matchings in $G$, denoted by $M(G)$. We aim to show that, based on this algorithm, we can construct a near-uniform generator of matchings in $G$.

The algorithm $\mathcal{A}$ provides an estimate of the number of matchings in $G$, which we denote by $\hat{M}(G)$. This estimate satisfies the property that for any matching $M$ in $G$, $\hat{M}(G)$ is close to $M(G)$ in the sense that for some constant $c > 0$,

\[
| \hat{M}(G) - M(G) | \leq c \cdot M(G)
\]

with high probability. Using this estimate, we will construct a generator that samples matchings from $G$ near-uniformly.

To construct the near-uniform generator, we utilize the following procedure:

1. **Sampling procedure:**
   We first use $\mathcal{A}$ to estimate the total number of matchings, $\hat{M}(G)$, in $G$. Then, we generate a random integer $k$ uniformly from the set $\{1, 2, \dots, \hat{M}(G)\}$.

2. **Generating the $k$-th matching:**
   Given the number $k$, we can generate the $k$-th matching $M_k$ using the following method: We assign a uniform probability to each matching in the set of all matchings, and select the $k$-th matching in this order. The selection process is done by mapping $k$ to an element in the list of matchings in a way that respects the uniformity condition.

3. **Near-uniformity:**
   Since $\hat{M}(G)$ is a close estimate of $M(G)$, the number of matchings we sample from, which is $\hat{M}(G)$, is close to the true number of matchings in the graph. Thus, the probability distribution of selecting each matching from the set of all matchings is close to uniform. Specifically, the probability of selecting any given matching $M$ in the set of all matchings is approximately

   \[
   \frac{1}{\hat{M}(G)} \approx \frac{1}{M(G)}
   \]

   with an error that is proportional to the ratio $\frac{\hat{M}(G)}{M(G)}$, which is bounded by the constant $c$.

4. **Correctness:**
   The algorithm $\mathcal{A}$ gives an estimate $\hat{M}(G)$ that is within a factor of $(1 \pm c)$ of $M(G)$ with high probability. Therefore, the number of matchings sampled by our generator will be within a factor of $(1 \pm c)$ of the true uniform distribution over matchings, making the sampling procedure near-uniform.

Thus, by using the algorithm $\mathcal{A}$ to estimate the number of matchings, we can construct a near-uniform generator for matchings in $G$.
\end{proof}

\subsubsection{Grok}
\begin{proof}
Assume without loss of generality that $|V_1| = |V_2| = n$ and that $G$ admits at least one perfect matching; otherwise, the statement is vacuously true as there are no matchings to generate. Let $\mathcal{M}(G)$ denote the set of perfect matchings in $G$, and let $m(G) = |\mathcal{M}(G)|$. Let $\mu$ denote the uniform distribution on $\mathcal{M}(G)$, so $\mu(M) = 1/m(G)$ for each $M \in \mathcal{M}(G)$. Suppose the given estimation algorithm $\mathcal{A}$, on input a bipartite graph $H$ with $|V_1(H)| = |V_2(H)|$, outputs $\tilde{m}(H)$ satisfying
\[
(1 - \epsilon) m(H) \le \tilde{m}(H) \le (1 + \epsilon) m(H),
\]
where $\epsilon > 0$ is a parameter to be specified later (with $\epsilon < 1/2$ to ensure that $\tilde{m}(H) = 0$ if and only if $m(H) = 0$).

We construct a randomized algorithm $\mathcal{S}$ that generates a perfect matching according to a distribution $\tilde{\mu}$ satisfying
\[
\|\tilde{\mu} - \mu\|_{\mathrm{TV}} \le \delta,
\]
where $\|\cdot\|_{\mathrm{TV}}$ denotes the total variation distance and $\delta > 0$ is an arbitrary accuracy parameter. To achieve this, we set $\epsilon = \delta/(4n)$.

Label the vertices $V_1 = \{u_1, \dots, u_n\}$ and $V_2 = \{v_1, \dots, v_n\}$. The algorithm $\mathcal{S}$ proceeds sequentially as follows. Initialize $G_0 = G$ and let $M_0 = \emptyset$. For $i = 1, \dots, n$:
\begin{itemize}
\item Let $N_i \subseteq V_2$ be the set of neighbors of $u_i$ in the current graph $G_{i-1}$ (noting that vertices previously matched have been removed).
\item For each $v \in N_i$, construct the bipartite graph $G_{i,v}$ by removing $u_i$ and $v$ from $G_{i-1}$ (along with all incident edges).
\item Using $\mathcal{A}$, compute $\tilde{m}(G_{i,v})$ for each $v \in N_i$, and set $\tilde{m}(G_{i-1}) = \sum_{v \in N_i} \tilde{m}(G_{i,v})$ (this equals $\tilde{m}(G_{i-1})$ up to the approximation, but we use this surrogate for consistency).
\item Define the approximate conditional probabilities
\[
\tilde{p}_{i,v} = \frac{\tilde{m}(G_{i,v})}{\sum_{v' \in N_i} \tilde{m}(G_{i,v'})}, \quad v \in N_i.
\]
(Note that if $\tilde{m}(G_{i-1}) = 0$, the algorithm aborts and outputs $\bot$, which occurs with negligible probability under the choice of $\epsilon$.)
\item Sample $v_i \in N_i$ according to the distribution $(\tilde{p}_{i,v})_{v \in N_i}$.
\item Add the edge $\{u_i, v_i\}$ to $M_{i-1}$ to obtain $M_i$, and set $G_i = G_{i,v_i}$.
\end{itemize}
The output of $\mathcal{S}$ is $M_n$, which is a perfect matching in $G$ with probability $1$ (conditioned on not aborting).

Let $\mu_i$ (resp., $\tilde{\mu}_i$) denote the distribution of the partial matching $M_i$ (i.e., the matching on $\{u_1, \dots, u_i\}$) induced by the exact (resp., approximate) process, where the exact conditional probabilities are
\[
p_{i,v} = \frac{m(G_{i,v})}{m(G_{i-1})}, \quad v \in N_i
\]
(with $m(G_{i,v}) = 0$ for non-neighbors $v$). Note that $\mu_n = \mu$ is uniform on $\mathcal{M}(G)$. For each fixed $i$ and fixed partial matching $\sigma$ on $\{u_1, \dots, u_{i-1}\}$ (corresponding to $G_{i-1}$), let $p_i(\cdot \mid \sigma)$ and $\tilde{p}_i(\cdot \mid \sigma)$ denote the exact and approximate conditional distributions on the choice for $u_i$. We claim that
\[
\|p_i(\cdot \mid \sigma) - \tilde{p}_i(\cdot \mid \sigma)\|_{\mathrm{TV}} \le \frac{\epsilon}{1 - \epsilon} \le 2\epsilon
\]
for all such $\sigma$ (with the second inequality holding for $\epsilon < 1/2$). To verify this, fix $\sigma$ (hence $G_{i-1}$ and $N_i$). For $v \in N_i$, let $\delta_v \in [-\epsilon, \epsilon]$ satisfy $\tilde{m}(G_{i,v}) = (1 + \delta_v) m(G_{i,v})$, and let
\[
\theta = \frac{\sum_{v \in N_i} \tilde{m}(G_{i,v}) - m(G_{i-1})}{m(G_{i-1})},
\]
so $|\theta| \le \epsilon$ by the triangle inequality. Then
\[
\tilde{p}_{i,v} = \frac{(1 + \delta_v) m(G_{i,v})}{(1 + \theta) m(G_{i-1})} = \frac{1 + \delta_v}{1 + \theta} \, p_{i,v}.
\]
The multiplicative factor $r_v = (1 + \delta_v)/(1 + \theta)$ satisfies
\[
\frac{1 - \epsilon}{1 + \epsilon} \le r_v \le \frac{1 + \epsilon}{1 - \epsilon},
\]
so $|r_v - 1| \le \frac{1 + \epsilon}{1 - \epsilon} - 1 = \frac{2\epsilon}{1 - \epsilon}$. Thus,
\[
|\tilde{p}_{i,v} - p_{i,v}| = p_{i,v} |r_v - 1| \le \frac{2\epsilon}{1 - \epsilon} p_{i,v},
\]
and
\[
\sum_{v \in N_i} |\tilde{p}_{i,v} - p_{i,v}| \le \frac{2\epsilon}{1 - \epsilon} \sum_{v \in N_i} p_{i,v} = \frac{2\epsilon}{1 - \epsilon}.
\]
The total variation distance is half the $\ell_1$-distance, so
\[
\|p_i(\cdot \mid \sigma) - \tilde{p}_i(\cdot \mid \sigma)\|_{\mathrm{TV}} = \frac{1}{2} \sum_{v \in N_i} |\tilde{p}_{i,v} - p_{i,v}| \le \frac{\epsilon}{1 - \epsilon}.
\]

To bound $\|\tilde{\mu}_n - \mu_n\|_{\mathrm{TV}}$, we use a coupling argument. Consider the exact and approximate processes run in parallel. As long as the partial matchings agree up to step $i-1$ (i.e., $M_{i-1} = \tilde{M}_{i-1} = \sigma$), we couple the choices for step $i$ using a maximal coupling of $p_i(\cdot \mid \sigma)$ and $\tilde{p}_i(\cdot \mid \sigma)$, which succeeds (i.e., $v_i = \tilde{v}_i$) with probability at least $1 - \|p_i(\cdot \mid \sigma) - \tilde{p}_i(\cdot \mid \sigma)\|_{\mathrm{TV}} \ge 1 - \epsilon/(1 - \epsilon)$. If they disagree at step $i$, we declare the coupling ``broken'' and sample the remaining steps independently for each process. The probability that the coupling breaks at or before step $n$ is at most
\[
\sum_{i=1}^n \Pr[\text{agrees up to } i-1, \text{disagrees at } i] \le \sum_{i=1}^n \frac{\epsilon}{1 - \epsilon} \le n \cdot \frac{\epsilon}{1 - \epsilon} \le 2n\epsilon = \frac{\delta}{2},
\]
where the last two inequalities use $\epsilon < 1/2$ and our choice of $\epsilon$. Whenever the coupling does not break, we have $M_n = \tilde{M}_n$. Thus, the probability that the coupled outputs differ is at most $\delta/2$, which upper-bounds the total variation distance:
\[
\|\tilde{\mu}_n - \mu_n\|_{\mathrm{TV}} \le \Pr[M_n \ne \tilde{M}_n] \le \frac{\delta}{2} < \delta.
\]
The probability of abortion is at most the probability that $\tilde{m}(G_{i-1}) = 0$ for some $i$ while $m(G_{i-1}) > 0$, which is at most $n\epsilon < \delta/2$ by the union bound and the approximation guarantee; in this case, we may restart $\mathcal{S}$ (at most $O(1)$ expected times) to ensure success with probability $1 - \delta/2$. Adjusting $\delta$ absorbs the constant factors, yielding the desired near-uniform generator.
\end{proof}

\subsubsection{Gemini}

\subsubsection{Claude}
\begin{proof}
Let $G = (V_1 \cup V_2, E)$ be a bipartite graph with edge set $E = \{e_1, e_2, \ldots, e_m\}$. Assume we have an algorithm $\mathcal{A}$ that provides an $(\epsilon, \delta)$-approximation to the number of matchings in any bipartite graph, meaning that for any bipartite graph $H$, the algorithm outputs a value $\tilde{Z}$ such that with probability at least $1 - \delta$, we have $(1 - \epsilon)Z(H) \leq \tilde{Z} \leq (1 + \epsilon)Z(H)$, where $Z(H)$ denotes the number of matchings in $H$.

We construct a near-uniform generator for matchings in $G$ using the self-reducibility property of the matching counting problem. The generator operates by sequentially deciding whether each edge belongs to the matching, conditioning on previous decisions.

For $i = 0, 1, \ldots, m$, define $G_i$ to be the graph obtained from $G$ by removing edges $e_1, \ldots, e_i$, with $G_0 = G$. For each $i < m$ and given a partial matching construction, define $G_i^{+}$ as the graph $G_i$ with edge $e_{i+1}$ contracted (identifying its endpoints and removing incident edges that would create conflicts), and $G_i^{-}$ as the graph $G_i$ with edge $e_{i+1}$ removed.

The algorithm proceeds as follows. Initialize the matching $M = \emptyset$. For $i = 0$ to $m - 1$, at stage $i$, we have constructed a partial matching and need to decide whether to include edge $e_{i+1}$ in the matching. Let $G'$ denote the current graph (initially $G_0 = G$) after conditioning on all previous decisions.

Use algorithm $\mathcal{A}$ to compute approximations $\tilde{Z}^{+}$ and $\tilde{Z}^{-}$ for the number of matchings in the graphs corresponding to including and excluding edge $e_{i+1}$, respectively. Let $Z^{+}$ and $Z^{-}$ denote the true number of matchings in these respective graphs.

Define the probability $p = \frac{Z^{+}}{Z^{+} + Z^{-}}$. This is the conditional probability that a uniformly random matching contains edge $e_{i+1}$ given the previous decisions. We approximate this by $\tilde{p} = \frac{\tilde{Z}^{+}}{\tilde{Z}^{+} + \tilde{Z}^{-}}$.

Include edge $e_{i+1}$ in the matching with probability $\tilde{p}$ and exclude it with probability $1 - \tilde{p}$. If $e_{i+1}$ is included, add it to $M$ and update the graph by removing $e_{i+1}$ and all edges incident to its endpoints. If $e_{i+1}$ is excluded, update the graph by removing only $e_{i+1}$.

After processing all edges, output the matching $M$.

To establish correctness, we analyze the total variation distance between the distribution of $M$ and the uniform distribution over matchings. At each step $i$, the error in the probability $\tilde{p}$ compared to the true probability $p$ is bounded by the approximation error of $\mathcal{A}$. Specifically, when both approximations are accurate, we have
$$\left| \tilde{p} - p \right| = \left| \frac{\tilde{Z}^{+}}{\tilde{Z}^{+} + \tilde{Z}^{-}} - \frac{Z^{+}}{Z^{+} + Z^{-}} \right| \leq \frac{2\epsilon}{(1-\epsilon)^2}$$
for sufficiently small $\epsilon$.

The total variation distance between the output distribution and the uniform distribution is at most the sum of errors over all $m$ steps. By setting $\epsilon = O(\eta/m)$ for desired accuracy $\eta$ and choosing $\delta = O(\eta/m)$ to ensure all approximations succeed with high probability via union bound, we obtain that with probability at least $1 - \eta$, the total variation distance is at most $\eta$.

Therefore, by running $\mathcal{A}$ a polynomial number of times (at most $2m$ times) with appropriately scaled accuracy parameters, we obtain a near-uniform generator for matchings in $G$. The generator produces matchings whose distribution is within total variation distance $\eta$ of uniform, where $\eta$ can be made arbitrarily small by adjusting the parameters of $\mathcal{A}$.
\end{proof}

\subsection{Problem 4}

\subsubsection{Problem Statement}
\begin{theorem}
Suppose there is a polynomial time algorithm $A$ for counting the number of perfect matchings in a graph with minimum degree at least $\beta n$, for a constant $0 < \beta < 1$. Then there exists a polynomial time algorithm for counting the number of perfect matchings in an arbitrary bipartite graph.
\end{theorem}

\subsubsection{GPT}
\begin{proof}
Let $G$ be an arbitrary bipartite graph. We aim to show that there exists a polynomial-time algorithm for counting the number of perfect matchings in $G$ assuming the existence of a polynomial-time algorithm $A$ that counts perfect matchings in graphs with minimum degree at least $\beta n$, for some constant $0 < \beta < 1$.

To proceed, we first consider an arbitrary bipartite graph $G = (U \cup V, E)$, where $|U| = |V| = n$. We will transform $G$ into a new graph $G'$ with a minimum degree of at least $\beta n$, while preserving the number of perfect matchings.

 Step 1: Transformation of $G$ into $G'$

Let us construct a new graph $G' = (U' \cup V', E')$ as follows:
- Create a set $U' = U \cup U_1 \cup U_2 \cup \cdots \cup U_{k-1}$, where $U_1, U_2, \dots, U_{k-1}$ are new sets of vertices, each with size $\lceil \beta n \rceil$. Let $k = \left\lceil \frac{1}{\beta} \right\rceil$.
- Similarly, define $V' = V \cup V_1 \cup V_2 \cup \cdots \cup V_{k-1}$, where $V_1, V_2, \dots, V_{k-1}$ are new sets of vertices with the same size as the corresponding sets of vertices in $U'$.
- Add edges between the vertices of $U'$ and $V'$ as follows:
  - For each $u \in U$ and $v \in V$, create edges in $E'$, corresponding to the edges in $E$ of the original graph $G$.
  - For each new vertex $u_i \in U_i$ and $v_j \in V_j$, add edges between $u_i$ and every vertex in $V_j$ (and vice versa). These edges are added such that each new vertex in $U_i$ and $V_j$ has degree at least $\beta n$.

Thus, the degree of each vertex in $G'$ is at least $\beta n$, ensuring that $G'$ has minimum degree at least $\beta n$. Moreover, the number of perfect matchings in $G'$ is the same as the number of perfect matchings in the original graph $G$. This is because the construction adds dummy vertices and edges that do not affect the matchings in $G$, but instead allow us to apply algorithm $A$.

 Step 2: Apply Algorithm $A$ to $G'$

By assumption, there exists a polynomial-time algorithm $A$ that can count the number of perfect matchings in any graph with minimum degree at least $\beta n$. Since $G'$ satisfies this condition, we can apply $A$ to $G'$ to count the number of perfect matchings in $G'$. 

Let $m(G')$ denote the number of perfect matchings in $G'$. Since $G'$ is constructed in such a way that the number of perfect matchings in $G'$ is the same as in the original graph $G$, it follows that $m(G') = m(G)$.

 Step 3: Conclude the Proof

Thus, using algorithm $A$, we can count the number of perfect matchings in $G$ in polynomial time. Since $A$ runs in polynomial time and the transformation of $G$ into $G'$ also requires only polynomial time, the overall time complexity remains polynomial. Therefore, there exists a polynomial-time algorithm for counting the number of perfect matchings in any arbitrary bipartite graph.

\end{proof}

\subsubsection{Grok}

\subsubsection{Gemini}
 \begin{proof}
Let $G = (U \cup V, E)$ be an arbitrary bipartite graph with $|U| = |V| = n$. If $|U| \neq |V|$, the number of perfect matchings is trivially zero, which can be determined in polynomial time. Thus, we assume the partitions are balanced. We seek to compute $\text{pm}(G)$, the number of perfect matchings in $G$.

Let $k$ be a non-negative integer. We construct a new bipartite graph $G_k = (U_k \cup V_k, E_k)$ by adding sets of new vertices $U'$ and $V'$ to $G$, such that $|U'| = |V'| = k$. Let $U_k = U \cup U'$ and $V_k = V \cup V'$. The edge set $E_k$ is constructed by including all edges in $E$, and adding all possible edges between $U$ and $V'$, between $U'$ and $V$, and between $U'$ and $V'$. Formally:
\[
E_k = E \cup (U \times V') \cup (U' \times V) \cup (U' \times V').
\]
The total number of vertices in each partition of $G_k$ is $N = n + k$. We determine the minimum degree of $G_k$.
For any $u \in U'$, $u$ is connected to all vertices in $V \cup V'$, so $\deg(u) = n + k = N$. Similarly, for any $v \in V'$, $\deg(v) = N$.
For any $u \in U$, $u$ is connected to its neighbors in $G$ and all vertices in $V'$. Thus, $\deg(u) = \deg_G(u) + k \ge k$. Similarly, for any $v \in V$, $\deg(v) \ge k$.
Therefore, the minimum degree of $G_k$ is at least $k$.

We require $G_k$ to satisfy the condition $\delta(G_k) \ge \beta N$. It suffices to choose $k$ such that:
\[
k \ge \beta(n + k) \iff k(1 - \beta) \ge \beta n \iff k \ge \frac{\beta}{1 - \beta}n.
\]
Since $0 < \beta < 1$ is a constant, let $C = \lceil \frac{\beta}{1 - \beta} \rceil$. We choose a base padding size $k_0 = C n$. For any integer $t \ge 0$, the graph $G_{k_0+t}$ satisfies the minimum degree condition required by algorithm $A$. Note that the size of $G_{k_0+t}$ is linear in $n$.

We now derive the relationship between $\text{pm}(G_{k})$ and the matchings in $G$. A perfect matching in $G_k$ is formed by selecting a matching $M \subseteq E$ of size $i$ (where $0 \le i \le n$) in the original graph $G$, and extending it to cover the remaining vertices.
Let $m_i$ denote the number of matchings of size $i$ in $G$. Note that $m_n = \text{pm}(G)$.
For a fixed matching $M$ of size $i$ in $G$:
\begin{enumerate}
    \item There are $n-i$ unmatched vertices in $U$, which must be matched to vertices in $V'$. There are $\binom{k}{n-i}$ ways to choose the partners in $V'$ and $(n-i)!$ ways to match them.
    \item There are $n-i$ unmatched vertices in $V$, which must be matched to vertices in $U'$. There are $\binom{k}{n-i}$ ways to choose the partners in $U'$ and $(n-i)!$ ways to match them.
    \item The remaining $k - (n-i)$ vertices in $U'$ must be matched to the remaining $k - (n-i)$ vertices in $V'$. There are $(k - n + i)!$ ways to do this.
\end{enumerate}
The number of ways to extend a specific matching of size $i$ is:
\[
W(k, i) = \left[ \binom{k}{n-i}(n-i)! \right] \left[ \binom{k}{n-i}(n-i)! \right] (k - n + i)! = \frac{(k!)^2}{(k - n + i)!}.
\]
The total number of perfect matchings in $G_k$ is:
\[
\text{pm}(G_k) = \sum_{i=0}^{n} m_i \frac{(k!)^2}{(k - n + i)!}.
\]
Let $y_t = \text{pm}(G_{k_0+t})$ for $t = 0, 1, \dots, n$. Using algorithm $A$, we can compute $y_t$ for each $t$ in polynomial time, as the graph size is $O(n)$. This yields a system of linear equations in the unknowns $m_0, m_1, \dots, m_n$:
\[
y_t = \sum_{i=0}^{n} m_i \frac{((k_0+t)!)^2}{(k_0+t - n + i)!}, \quad \text{for } t = 0, \dots, n.
\]
To show this system is solvable for $m_n$, let $z_t = \frac{y_t}{(k_0+t)!}$. Then:
\[
z_t = \sum_{i=0}^{n} m_i \frac{(k_0+t)!}{(k_0+t - n + i)!} = \sum_{i=0}^{n} m_i P_{n-i}(k_0+t),
\]
where $P_{j}(x) = x(x-1)\dots(x-j+1)$ is the falling factorial of degree $j$.
Let $Q(x) = \sum_{i=0}^n m_i P_{n-i}(x)$. This is a polynomial in $x$ of degree at most $n$. The values $z_t$ represent the evaluations of $Q(x)$ at $n+1$ distinct points $x_t = k_0 + t$.
Since a polynomial of degree $n$ is uniquely determined by its values at $n+1$ points, the coefficients $m_i$ are uniquely determined. Specifically, we can solve the linear system to find $m_n$.

The construction of graphs $G_{k_0+t}$, the execution of algorithm $A$ ($n+1$ times), and the solution of the linear system (of size $n+1$) can all be performed in time polynomial in $n$. Thus, we can compute $m_n = \text{pm}(G)$ in polynomial time.
\end{proof}

\subsubsection{Claude}
\begin{proof}
Let $G = (U \cup V, E)$ be an arbitrary bipartite graph with vertex sets $U$ and $V$. We will show how to count the number of perfect matchings in $G$ using the algorithm $A$ for graphs with minimum degree at least $\beta n$.

Without loss of generality, assume $|U| = |V| = n$ (if $G$ has no perfect matching due to unequal partition sizes, we can detect this in polynomial time and return 0). Let $d_{\min}$ denote the minimum degree of $G$. If $d_{\min} \geq \beta \cdot 2n$, then we can apply algorithm $A$ directly to count perfect matchings in polynomial time. Hence, we assume $d_{\min} < \beta \cdot 2n$.

We construct a new graph $G'$ by adding a complete bipartite graph structure. Specifically, introduce new vertex sets $U'$ and $V'$ with $|U'| = |V'| = m$, where $m$ is chosen such that the resulting graph has the required minimum degree property. Let $m = \lceil \frac{2n}{\beta} \rceil$, so that $m \geq \frac{2n}{\beta}$.

Define $G' = (U \cup U' \cup V \cup V', E')$ where $E'$ consists of:
\begin{itemize}
\item All edges from $E$ (the original edges of $G$)
\item All possible edges between $U$ and $V'$ (making $K_{n,m}$)
\item All possible edges between $U'$ and $V$ (making $K_{m,n}$)
\item All possible edges between $U'$ and $V'$ (making $K_{m,m}$)
\end{itemize}

The total number of vertices in $G'$ is $N = 2(n+m)$. For any vertex $u \in U$, we have $\deg_{G'}(u) = \deg_G(u) + m \geq m \geq \frac{2n}{\beta}$. Since $N = 2(n+m) \leq 2n + 2 \cdot \frac{2n}{\beta} + 2 = 2n(1 + \frac{2}{\beta}) + 2$, for sufficiently large $n$, we have $m \geq \beta N / 2$, which gives $\deg_{G'}(u) \geq \beta N$. By choosing $m = \lceil \frac{N}{\beta} \rceil$, we ensure that every vertex in $G'$ has degree at least $\beta N$.

Now, observe that any perfect matching in $G'$ must match vertices in $U \cup U'$ to vertices in $V \cup V'$. We use the principle of inclusion-exclusion to count perfect matchings in $G$.

For $S \subseteq U'$ and $T \subseteq V'$ with $|S| = |T|$, let $M(S,T)$ denote the number of perfect matchings in $G'$ where all vertices in $S$ are matched to vertices in $V'$ and all vertices in $T$ are matched to vertices in $U'$.

By inclusion-exclusion, the number of perfect matchings in $G$ where vertices in $U$ are matched to vertices in $V$ equals:
\[
\sum_{k=0}^{m} (-1)^k \binom{m}{k}^2 \cdot k! \cdot M_k
\]
where $M_k$ is the number of perfect matchings in the graph obtained by removing $k$ vertices from each of $U'$ and $V'$ and their incident edges, then counting matchings where the remaining $m-k$ vertices of $U'$ must match to vertices in $V'$ and vice versa.

More precisely, we compute:
\[
\text{pm}(G) = \sum_{S \subseteq U', T \subseteq V', |S|=|T|} (-1)^{|U' \setminus S|} \cdot \text{pm}(G'[S \cup U, T \cup V])
\]

Each graph in this computation has minimum degree at least $\beta N'$ for the appropriate $N'$, so algorithm $A$ applies. Since there are at most $2^{2m} = 2^{O(n)}$ terms and each can be computed in polynomial time using $A$, if we modify the construction to use $m = O(\log n)$ by adding vertices more carefully through a gadget construction that amplifies degrees, we obtain a polynomial time algorithm.

Alternatively, we use a direct reduction: we add $m = \Theta(n)$ auxiliary vertices and apply $A$ to the augmented graph $G'$ which has $O(n)$ vertices and minimum degree at least $\beta \cdot |V(G')|$. The number of perfect matchings in $G$ can be extracted from the count in $G'$ by solving a system of linear equations obtained from different augmentations, which is solvable in polynomial time.

Therefore, we can count perfect matchings in the arbitrary bipartite graph $G$ in polynomial time using algorithm $A$.
\end{proof}

\subsection{Problem 5}

\subsubsection{Problem Statement}
\begin{theorem}
Let $G$ be a connected, undirected graph on $n$ vertices, and let $\mathcal{M}_G$ be the Markov chain induced by a random walk on $G$. Let $\Phi(G)$ denote the conductance of $\mathcal{M}_G$. Then
\begin{align*}
\min_{G} \Phi(G) &= \frac{1}{n(n-1)}, \\
\max_{G} \Phi(G) &= 1,
\end{align*}
where the minimum and maximum are taken over all connected, undirected graphs on $n$ vertices.
\end{theorem}

\subsubsection{GPT}
\begin{proof}
We are tasked with proving the minimum and maximum values of the conductance $\Phi(G)$ for a random walk Markov chain $\mathcal{M}$ on a connected, undirected graph $G$ with $n$ vertices.

 Maximum Conductance

The conductance $\Phi(G)$ of a graph $G$ is defined as:

\[
\Phi(G) = \min_{S \subseteq V, |S| \leq n/2} \frac{E(S, \bar{S})}{\min(\mathrm{vol}(S), \mathrm{vol}(\bar{S}))},
\]
where $V$ is the set of vertices of $G$, $S$ is a subset of $V$, $\bar{S}$ is its complement, $E(S, \bar{S})$ is the number of edges between $S$ and $\bar{S}$, and $\mathrm{vol}(S)$ denotes the sum of the degrees of the vertices in $S$.

The maximum value of $\Phi(G)$ is attained when the graph is highly connected, specifically when the graph is a complete graph $K_n$. In this case, for any non-empty subset $S$ of vertices, we have:

\[
E(S, \bar{S}) = |S| \cdot |\bar{S}| = |S| \cdot (n - |S|),
\]
and the volume of any set $S$ is:

\[
\mathrm{vol}(S) = \sum_{v \in S} \deg(v) = n(|S|).
\]
Thus, the conductance is:

\[
\Phi(K_n) = \min_{S \subseteq V, |S| \leq n/2} \frac{|S| (n - |S|)}{|S| \cdot (n)} = \frac{n - |S|}{n}.
\]
For $|S| = n/2$, we get:

\[
\Phi(K_n) = \frac{n - n/2}{n} = \frac{1}{2}.
\]
Therefore, the maximum conductance over all graphs is $\Phi(K_n) = 1$.

 Minimum Conductance

The minimum conductance occurs when the graph is sparsely connected, such as in the case of a path graph $P_n$. In this case, the conductance is minimized because for small subsets $S$ of vertices, there are fewer edges between $S$ and its complement $\bar{S}$.

For the path graph $P_n$, we observe that the edge cuts are minimal when $S$ consists of a small set of vertices, say, the first vertex, so that $E(S, \bar{S})$ is minimal. Specifically, in a path graph, the cut between two adjacent vertices has one edge, and the volume of a set $S$ is the sum of the degrees of the vertices in $S$, which is minimal when $|S|$ is small.

For any path graph $P_n$, the conductance will approach:

\[
\Phi(P_n) = \frac{2}{n(n-1)}.
\]
Thus, the minimum conductance is achieved by the path graph, and we have:

\[
\min_{G} \Phi(G) = \frac{2}{n(n-1)}.
\]

 Conclusion

Therefore, the minimum and maximum conductances of the random walk Markov chain $\mathcal{M}$ on a connected graph $G$ with $n$ vertices are:

\[
\min_{G} \Phi(G) = \frac{2}{n(n-1)}, \quad \max_{G} \Phi(G) = 1.
\]
\end{proof}

\subsubsection{Grok}

\subsubsection{Gemini}

\subsubsection{Claude}
\begin{proof}
We prove the minimum and maximum values of the conductance separately.

\textbf{Proof of the minimum:} We first recall the definition of conductance. For a connected, undirected graph $G = (V, E)$ with $n$ vertices, the conductance of the random walk Markov chain $\mathcal{M}_G$ is defined as
\[
\Phi(G) = \min_{\substack{S \subseteq V \\ 0 < \pi(S) \leq 1/2}} \frac{Q(S, \bar{S})}{\pi(S)},
\]
where $\pi$ is the stationary distribution of the random walk, $\bar{S} = V \setminus S$, and $Q(S, \bar{S})$ represents the probability flow from $S$ to $\bar{S}$. For a simple random walk on an undirected graph, the stationary distribution satisfies $\pi(v) = \deg(v) / (2|E|)$, and the probability flow is given by
\[
Q(S, \bar{S}) = \sum_{v \in S} \pi(v) P(v, \bar{S}) = \sum_{v \in S} \frac{\deg(v)}{2|E|} \cdot \frac{|E(v, \bar{S})|}{\deg(v)} = \frac{|E(S, \bar{S})|}{2|E|},
\]
where $E(S, \bar{S})$ denotes the set of edges between $S$ and $\bar{S}$, and $E(v, \bar{S})$ denotes the set of edges from vertex $v$ to vertices in $\bar{S}$.

To find the minimum conductance over all connected graphs on $n$ vertices, we consider the star graph $K_{1,n-1}$, which consists of a central vertex $c$ connected to $n-1$ leaf vertices. In this graph, $|E| = n-1$ and the degree sequence is $\deg(c) = n-1$ and $\deg(v) = 1$ for all leaf vertices $v$.

The stationary distribution is $\pi(c) = (n-1)/(2(n-1)) = 1/2$ and $\pi(v) = 1/(2(n-1))$ for each leaf vertex $v$. Consider the cut separating a single leaf vertex $v$ from the rest of the graph, i.e., $S = \{v\}$ and $\bar{S} = V \setminus \{v\}$. Then $\pi(S) = 1/(2(n-1)) < 1/2$, $|E(S, \bar{S})| = 1$, and
\[
\Phi(G) \leq \frac{Q(S, \bar{S})}{\pi(S)} = \frac{1/(2(n-1))}{1/(2(n-1))} = 1.
\]

However, we need to consider a different cut. Let $S$ consist of all $n-1$ leaf vertices. Then $\pi(S) = (n-1) \cdot 1/(2(n-1)) = 1/2$, $|E(S, \bar{S})| = n-1$, and
\[
\frac{Q(S, \bar{S})}{\pi(S)} = \frac{(n-1)/(2(n-1))}{1/2} = 1.
\]

Instead, consider the path graph $P_n$ on $n$ vertices, labeled $v_1, v_2, \ldots, v_n$ with edges $(v_i, v_{i+1})$ for $i = 1, \ldots, n-1$. Here $|E| = n-1$, and we have $\deg(v_1) = \deg(v_n) = 1$ and $\deg(v_i) = 2$ for $i = 2, \ldots, n-1$. The total degree is $2(n-1)$. Let $S = \{v_1\}$. Then $\pi(S) = 1/(2(n-1))$, $|E(S, \bar{S})| = 1$, and
\[
\frac{Q(S, \bar{S})}{\pi(S)} = \frac{1/(2(n-1))}{1/(2(n-1))} = 1.
\]

For the path graph, a better cut is $S = \{v_1, \ldots, v_k\}$ for some $k < n$. The probability mass is $\pi(S) = (2 + 2(k-2))/(2(n-1)) = (2k-2)/(2(n-1)) = (k-1)/(n-1)$. The number of edges crossing the cut is $|E(S, \bar{S})| = 1$, giving conductance
\[
\frac{1/(2(n-1))}{(k-1)/(n-1)} = \frac{1}{2(k-1)}.
\]
To satisfy $\pi(S) \leq 1/2$, we need $(k-1)/(n-1) \leq 1/2$, or $k \leq (n+1)/2$. The minimum occurs at the largest valid $k$, giving conductance approximately $\frac{1}{n-1}$ for large $n$.

For the star graph $K_{1,n-1}$, reconsider the cut $S$ consisting of a single leaf. The conductance is $\frac{1/(2(n-1))}{1/(2(n-1))} = 1$, which is too large. But if we use a modified analysis with the normalized conductance or observe that in a star graph, the bottleneck is between sets of leaves, the minimum conductance can be shown to be $\Phi(K_{1,n-1}) = \frac{1}{n-1}$.

Actually, for the complete graph $K_n$, every vertex has degree $n-1$, and for any cut $S$ with $|S| = k \leq n/2$, we have $|E(S, \bar{S})| = k(n-k)$ and $\pi(S) = k/n$. The conductance is
\[
\frac{k(n-k)/(n(n-1))}{k/n} = \frac{n-k}{n-1} \geq \frac{n/2}{n-1} = \frac{1}{2(1-1/n)}.
\]

The minimum conductance is achieved by the barbell graph, consisting of two cliques of size $\lfloor n/2 \rfloor$ and $\lceil n/2 \rceil$ connected by a single edge. For this graph, the conductance is $\Phi(G) = \Theta(1/n^2)$, which for large $n$ gives $\Phi(G) \approx \frac{1}{n(n-1)}$.

More precisely, for the barbell graph with two cliques of size $n/2$ (assuming $n$ even) connected by a single edge, the cut separating the two cliques gives $\pi(S) = 1/2$ and $Q(S, \bar{S}) = \frac{1}{n(n-1)}$, yielding $\Phi(G) = \frac{2}{n(n-1)}$. By careful optimization, one can show that $\min_G \Phi(G) = \frac{1}{n(n-1)}$.

\textbf{Proof of the maximum:} For the complete graph $K_n$, every vertex has degree $n-1$, giving $\pi(v) = 1/n$ for all $v \in V$. For any nonempty proper subset $S \subseteq V$ with $|S| = k \leq n/2$, every vertex in $S$ is connected to every vertex in $\bar{S}$, so $|E(S, \bar{S})| = k(n-k)$. The probability flow is
\[
Q(S, \bar{S}) = \frac{k(n-k)}{n(n-1)},
\]
and $\pi(S) = k/n$. Therefore,
\[
\frac{Q(S, \bar{S})}{\pi(S)} = \frac{k(n-k)/(n(n-1))}{k/n} = \frac{n-k}{n-1} \geq \frac{n - n/2}{n-1} = \frac{n/2}{n-1} = \frac{1}{2(1-1/n)}.
\]

For $k = 1$, this gives $\frac{n-1}{n-1} = 1$. Thus $\Phi(K_n) = 1$.

To show that this is the maximum, observe that for any connected graph $G$, the conductance satisfies $\Phi(G) \leq 1$ because $Q(S, \bar{S}) \leq \pi(S)$ for all $S$. Since the complete graph achieves $\Phi(K_n) = 1$, we have $\max_G \Phi(G) = 1$.
\end{proof}

\subsection{Problem 6}

\subsubsection{Problem Statement}
\begin{theorem}
Let $G$ be a connected, undirected graph on $n$ vertices.

\textbf{(a)} Consider the Markov chain induced by the following random process for moving from one spanning tree of $G$ to another: pick edges $e$ and $f$ independently and uniformly at random; if the current spanning tree is $T$ and $T' = T + e - f$ is a spanning tree, then move to the new spanning tree $T'$; otherwise stay put at $T$. The conductance of this Markov chain is bounded from below by $1/n^{\Theta(1)}$.

\textbf{(b)} Suggest and analyze an algorithm for approximate counting of the number of spanning trees in a graph $G$, as an alternative to the matrix-tree theorem.
\end{theorem}

\subsubsection{GPT}
\begin{proof}
\textit{(a)} Let $G$ be a connected, undirected graph on $n$ vertices, and let the Markov chain be defined as follows: at each step, two edges $e$ and $f$ are chosen independently and uniformly at random, and if the current spanning tree is $T$, then the new spanning tree $T'$ is obtained by removing the edges $e$ and $f$ from $T$, provided that $T' = T - e - f$ is still a spanning tree; otherwise, the Markov chain stays at the current state $T$. We aim to show that the conductance of this Markov chain is bounded from below by $1/n^{O(1)}$.

To analyze the conductance, recall that the conductance $\Phi$ of a Markov chain is defined as:

\[
\Phi = \min_{S \subseteq V, \pi(S) \leq 1/2} \frac{\sum_{u \in S, v \notin S} P(u, v)}{\pi(S)},
\]
where $\pi(S)$ is the stationary distribution probability of set $S$, and $P(u, v)$ is the transition probability from state $u$ to state $v$. In our case, the states are spanning trees of $G$, and the transition probabilities depend on the number of ways to select edges $e$ and $f$ that lead to a valid new spanning tree.

Since the Markov chain operates by selecting two edges independently at random from the current spanning tree, the transition probability $P(u, v)$ is inversely proportional to the number of spanning trees that can be reached from $u$. The stationary distribution $\pi(S)$ is roughly uniform over all spanning trees. Since there are $n^{O(1)}$ spanning trees, the number of spanning trees reachable from any state is polynomial in $n$, leading to a lower bound on the conductance of the chain.

The conductance is then lower-bounded by a quantity that scales as $1/n^{O(1)}$, which completes the proof for part (a).

\textit{(b)} To approximate the number of spanning trees in a graph $G$, we suggest using a Markov chain Monte Carlo (MCMC) approach based on the random walk described in part (a). The algorithm works as follows:

1. Start with a spanning tree $T_0$.
2. Execute the Markov chain for a sufficient number of steps to ensure that the chain mixes well and reaches the stationary distribution.
3. Approximate the number of spanning trees by counting the number of steps the chain takes to return to the starting tree and using this count to estimate the stationary distribution.

The key idea is that, under the assumption that the Markov chain mixes quickly, the stationary distribution can be used to approximate the number of spanning trees in $G$. Since the conductance of the Markov chain is bounded from below by $1/n^{O(1)}$, the mixing time is polynomial in $n$, which ensures that the algorithm runs in polynomial time.

Thus, this MCMC-based approach provides an efficient alternative to the matrix-tree theorem for approximate counting of spanning trees.

\end{proof}

\subsubsection{Grok}

\subsubsection{Gemini}
\begin{proof}
Let $\mathcal{T}$ be the set of spanning trees of the graph $G=(V, E)$. Let $n = |V|$ and $m = |E|$. The Markov chain is defined on the state space $\Omega = \mathcal{T}$. Let $P$ denote the transition matrix of this Markov chain.

The transition rule is defined as follows: select a pair of edges $(e, f)$ uniformly at random from $E \times E$. There are $m^2$ such pairs. If $T$ is the current state, the next state is $T' = T + e - f$ if $T'$ is a spanning tree; otherwise, the state remains $T$.
For distinct spanning trees $T, T' \in \mathcal{T}$, a transition from $T$ to $T'$ is possible if and only if $T' = T + e - f$ for some $e \in E \setminus E(T)$ and $f \in E(T)$. This constitutes a single edge swap. The pair $(e, f)$ is unique for a specific neighbor $T'$ because $T \Delta T' = \{e, f\}$. The probability of choosing this specific pair is $1/m^2$.
Since the condition $T' = T + e - f$ is symmetric (implies $T = T' + f - e$), we have $P(T, T') = P(T', T) = 1/m^2$ if $T$ and $T'$ are adjacent in the basis exchange graph, and $0$ otherwise (for $T \neq T'$).
Since $P$ is symmetric, the Markov chain is reversible, and the stationary distribution $\pi$ is uniform over $\mathcal{T}$. Thus, $\pi(T) = 1/|\mathcal{T}|$ for all $T \in \mathcal{T}$.

The conductance $\Phi$ of the Markov chain is defined by:
\[ \Phi = \min_{S \subset \mathcal{T}, 0 < \pi(S) \le 1/2} \frac{\sum_{T \in S} \sum_{T' \notin S} \pi(T) P(T, T')}{\pi(S)} \]
Substituting $\pi(T) = 1/|\mathcal{T}|$ and $P(T, T') = 1/m^2$ for adjacent trees:
\[ \Phi = \min_{S \subset \mathcal{T}, 0 < |S| \le |\mathcal{T}|/2} \frac{\frac{1}{|\mathcal{T}|} \cdot \frac{1}{m^2} \cdot |E_{\mathcal{G}}(S, \bar{S})|}{\frac{|S|}{|\mathcal{T}|}} = \frac{1}{m^2} \min_{S \subset \mathcal{T}, 0 < |S| \le |\mathcal{T}|/2} \frac{|E_{\mathcal{G}}(S, \bar{S})|}{|S|} \]
where $E_{\mathcal{G}}(S, \bar{S})$ denotes the set of edges between $S$ and $\bar{S}$ in the basis exchange graph $\mathcal{G}$ of the graphic matroid associated with $G$. The vertices of $\mathcal{G}$ are the spanning trees of $G$, and edges represent single element exchanges.

The set of spanning trees $\mathcal{T}$ forms the collection of bases of the graphic matroid $M(G)$. A fundamental result by Feder and Mihail (1992) regarding the expansion of matroid basis exchange graphs states that for any balanced matroid (a class containing graphic matroids), the edge expansion ratio is at least 1. Specifically:
\[ \frac{|E_{\mathcal{G}}(S, \bar{S})|}{|S|} \ge 1 \quad \text{for all } S \subset \mathcal{T} \text{ with } |S| \le \frac{|\mathcal{T}|}{2} \]
Applying this lower bound to the expression for conductance:
\[ \Phi \ge \frac{1}{m^2} \cdot 1 = \frac{1}{m^2} \]
Since $G$ is a simple graph on $n$ vertices, the number of edges is bounded by $m \le \binom{n}{2} < n^2$. Therefore:
\[ \Phi \ge \frac{1}{n^4} \]
Since $1/n^4 = 1/n^{O(1)}$, the conductance is bounded from below by an inverse polynomial in $n$.
\end{proof}

\subsubsection{Claude}
\begin{proof}
We prove both parts of the theorem.

\textbf{Part (a):} We analyze the conductance of the Markov chain on spanning trees.

Let $\mathcal{T}$ denote the set of all spanning trees of $G$, and let $\tau = |\mathcal{T}|$ be the number of spanning trees. The state space of the Markov chain is $\mathcal{T}$.

For any spanning tree $T \in \mathcal{T}$, let $E(T)$ denote the edge set of $T$, and let $\overline{E}(T) = E(G) \setminus E(T)$ denote the set of edges not in $T$. The transition probability from $T$ to $T'$ is given by
\[
P(T, T') = \begin{cases}
\frac{1}{m^2} & \text{if } T' = T + e - f \text{ for some } e \in \overline{E}(T), f \in E(T) \\
1 - \sum_{T'' \neq T} P(T, T'') & \text{if } T' = T \\
0 & \text{otherwise}
\end{cases}
\]
where $m = |E(G)|$ is the number of edges in $G$.

The stationary distribution of this chain is uniform over $\mathcal{T}$, i.e., $\pi(T) = 1/\tau$ for all $T \in \mathcal{T}$. This follows from the symmetry of the transition probabilities: if $T' = T + e - f$ is a spanning tree, then $T = T' + f - e$ is also a spanning tree, so $P(T, T') = P(T', T) = 1/m^2$.

For two spanning trees $T$ and $T'$ that differ by a single edge exchange, we have $T' = T + e - f$ where $e \in \overline{E}(T)$ and $f \in E(T)$. The edge $e$ forms a unique cycle $C_e$ in $T$, and any edge $f \in C_e \cap E(T)$ can be removed to obtain a new spanning tree. Let $|C_e|$ denote the number of edges in the cycle $C_e$.

For any edge $e \in \overline{E}(T)$, the number of spanning trees reachable from $T$ by adding $e$ and removing some edge from $C_e$ is exactly $|C_e|$. Since $G$ is connected and has $n$ vertices, every spanning tree has exactly $n-1$ edges, so $|\overline{E}(T)| = m - (n-1) = m - n + 1$.

The conductance $\Phi$ of the Markov chain is defined as
\[
\Phi = \min_{S \subseteq \mathcal{T}, \pi(S) \leq 1/2} \frac{Q(S, \overline{S})}{\pi(S)}
\]
where $Q(S, \overline{S}) = \sum_{T \in S, T' \in \overline{S}} \pi(T) P(T, T')$ is the probability flow from $S$ to its complement $\overline{S}$.

Consider any subset $S \subseteq \mathcal{T}$ with $\pi(S) \leq 1/2$. For each $T \in S$, the total probability of transitioning to any state not in $S$ is at least the probability of successfully making one specific edge exchange that leads to $\overline{S}$. 

Since each spanning tree is connected and has $n-1$ edges, and $G$ has at most $\binom{n}{2}$ edges (for a complete graph), we have $m \leq \binom{n}{2} = O(n^2)$.

For any two spanning trees $T$ and $T'$, there exists a sequence of edge exchanges of length at most $n-1$ connecting them. This is because we can iteratively exchange edges: if $T$ and $T'$ differ in $k$ edges, we can reduce this difference by at least one edge in each exchange.

For the conductance bound, observe that for any $T \in S$, there exist at least $m-n+1 \geq n-1$ edges in $\overline{E}(T)$. For each such edge $e$, adding $e$ creates a cycle of length at least $2$ and at most $n$, giving us at least $2$ choices of edges to remove. Thus, from each $T$, we can transition to at least $(n-1) \cdot 2$ potentially distinct spanning trees with probability at least $1/m^2$ each.

Since $|S| \leq \tau/2$, by a standard isoperimetric argument for the spanning tree graph, there must be a substantial edge boundary. Specifically, the number of edge exchanges connecting $S$ to $\overline{S}$ is at least $|S|/(2n)$ by the expansion properties of the spanning tree graph.

Therefore, 
\[
Q(S, \overline{S}) \geq \frac{|S|}{\tau} \cdot \frac{1}{2n} \cdot \frac{1}{m^2} \geq \frac{\pi(S)}{2n \cdot m^2} \geq \frac{\pi(S)}{2n \cdot n^4} = \frac{\pi(S)}{2n^5}
\]

Thus, $\Phi \geq 1/(2n^5) = 1/n^{\Theta(1)}$, establishing the conductance lower bound.

\textbf{Part (b):} We propose a Monte Carlo algorithm based on the Markov chain from part (a).

The algorithm proceeds as follows. Start with an arbitrary spanning tree $T_0$ of $G$, which can be found using depth-first search or Kruskal's algorithm in polynomial time. Run the Markov chain from part (a) for $N = O(n^c \log \tau)$ steps for some constant $c$ large enough to ensure mixing, where $\tau$ is the unknown number of spanning trees. After mixing, the chain is approximately uniformly distributed over $\mathcal{T}$. Repeat this process $K$ times to obtain $K$ approximately independent samples from the uniform distribution over spanning trees.

Since the conductance is $\Phi \geq 1/n^{\Theta(1)}$, the mixing time of the Markov chain is $t_{mix} = O(n^{\Theta(1)} \log \tau)$ by the relationship $t_{mix} \leq \Phi^{-2} \log(1/\pi_{min})$ where $\pi_{min} = 1/\tau$.

However, since $\tau$ is unknown, we cannot directly compute the mixing time. Instead, we use the fact that for connected graphs on $n$ vertices, $n-1 \leq \tau \leq n^{n-2}$ by Cayley's formula (for complete graphs). Thus, $\log \tau = O(n \log n)$, and the mixing time is $t_{mix} = O(n^{c'})$ for some polynomial bound $c'$.

For approximate counting, we use the following observation: the number of spanning trees can be computed exactly using the matrix-tree theorem as $\tau = \det(L_G^*)$ where $L_G^*$ is any cofactor of the Laplacian matrix $L_G$ of $G$. However, our algorithm provides an alternative randomized approach.

To estimate $\tau$, we can use a ratio estimation method. Consider a sequence of graphs $G_0 \subseteq G_1 \subseteq \cdots \subseteq G_k = G$ where $G_0$ is a simple graph with known spanning tree count (e.g., a spanning tree itself with $\tau_0 = 1$), and each $G_{i+1}$ is obtained from $G_i$ by adding one edge. For each $i$, we estimate the ratio $r_i = \tau_{i+1}/\tau_i$ by sampling spanning trees of $G_{i+1}$ and computing the fraction that are also spanning trees of $G_i$. Then $\tau = \tau_0 \prod_{i=0}^{k-1} r_i$.

Running the Markov chain for $N = O(n^{c'} \log(n/\epsilon))$ steps and repeating $K = O(m/\epsilon^2)$ times for each ratio estimation, we obtain an $(\epsilon, \delta)$-approximation to $\tau$ with high probability $1-\delta$, i.e., $(1-\epsilon)\tau \leq \hat{\tau} \leq (1+\epsilon)\tau$ with probability at least $1-\delta$.

The total running time is $O(mk \cdot n^{c'} \cdot m/\epsilon^2 \cdot \log(n/\epsilon)) = O(m^3 n^{c'} \epsilon^{-2} \log(n/\epsilon))$ which is polynomial in the input size, $1/\epsilon$, and $\log(1/\delta)$.

This provides a fully polynomial randomized approximation scheme (FPRAS) for counting spanning trees, giving an alternative to the deterministic matrix-tree theorem approach.
\end{proof}

\subsection{Problem 7}

\subsubsection{Problem Statement}
\begin{theorem}[Time Reversible Markov Chain Convergence]
Let $P$ be the transition matrix of an ergodic Markov chain. The chain is said to be \emph{time reversible} if for all $i$ and $j$, $P_{ij}\pi_i = P_{ji}\pi_j$, where $\pi$ is the stationary distribution. This is equivalent to requiring that the matrix $DPD^{-1}$ is symmetric, where $D$ is a diagonal matrix with $D_{ii} = \sqrt{\pi_i}$. 

The largest eigenvalue of $P$ is $\lambda_1 = 1$. Define $\lambda = \max_{j > 1} |\lambda_j|$. Then for any fixed choice of an initial state $X_0$, the relative pointwise distance of this Markov chain at time $t$ is bounded as follows:
\[
\Delta(t) \leq \frac{\lambda^t}{\min_i \pi_i}.
\]

For the random walk setting considered in Theorem 6.21, this implies that the mixing time depends on the spectral gap $1 - \lambda$ and the minimum stationary probability, providing an explicit bound on the convergence rate to stationarity.
\end{theorem}

\subsubsection{GPT}
\begin{proof}
We are given that $P$ is the transition matrix of an ergodic Markov chain with stationary distribution $\pi$, and that the chain is time reversible. The time reversibility condition states that for all $i$ and $j$, $P_{ij} \pi_i = P_{ji} \pi_j$. This implies that the matrix $DPD^{-1}$ is symmetric, where $D$ is the diagonal matrix with $D_{ii} = \sqrt{\pi_i}$.

We will prove that for any fixed choice of an initial state $X_0$, the relative pointwise distance of this Markov chain at time $t$ is bounded by the inequality

\[
\Delta(t) \leq \frac{\lambda^t}{\min_i \pi_i}.
\]

Let us first analyze the behavior of the Markov chain in terms of its spectral properties. Since $P$ is an ergodic transition matrix, it has a set of eigenvalues $\{ \lambda_1, \lambda_2, \dots, \lambda_n \}$, where $\lambda_1 = 1$ is the largest eigenvalue corresponding to the stationary distribution $\pi$. The remaining eigenvalues satisfy $|\lambda_j| < 1$ for all $j > 1$.

The time evolution of the distribution of the Markov chain at time $t$, denoted by $\mathbf{p}(t)$, is given by

\[
\mathbf{p}(t) = P^t \mathbf{p}(0).
\]

Since the transition matrix $P$ is diagonalizable, we can write $P = V \Lambda V^{-1}$, where $\Lambda$ is the diagonal matrix of eigenvalues of $P$, and $V$ is the matrix of eigenvectors. Thus, at time $t$, the distribution evolves as

\[
\mathbf{p}(t) = V \Lambda^t V^{-1} \mathbf{p}(0).
\]

We are interested in the distance between the distribution at time $t$ and the stationary distribution $\pi$. Define the relative pointwise distance as

\[
\Delta(t) = \|\mathbf{p}(t) - \pi\|.
\]

To bound this, we first express the difference $\mathbf{p}(t) - \pi$ in terms of the eigenvalues of $P$. Using the fact that $P$ is time reversible and thus has a symmetric matrix $DPD^{-1}$, we know that the eigenvectors of $P$ corresponding to eigenvalues less than 1 have a decay in their contribution over time. Therefore, the dominant contribution to $\mathbf{p}(t) - \pi$ is from the eigenvalues $\lambda_j$ for $j > 1$.

Let us now express the convergence of $\mathbf{p}(t)$ in terms of the spectral gap. Since the largest eigenvalue $\lambda_1 = 1$ corresponds to the stationary distribution, the remaining eigenvalues satisfy $|\lambda_j| < 1$ for $j > 1$. Define $\lambda = \max_{j > 1} |\lambda_j|$. The difference between the distribution at time $t$ and the stationary distribution decays as $\lambda^t$ due to the spectral properties of $P$. Specifically, the convergence of $\mathbf{p}(t)$ is dominated by the eigenvalue $\lambda$ as $t$ increases.

Thus, the relative pointwise distance $\Delta(t)$ can be bounded as

\[
\Delta(t) \leq \frac{\lambda^t}{\min_i \pi_i}.
\]

This completes the proof.
\end{proof}

\subsubsection{Grok}
\begin{proof}
Since the Markov chain is time reversible, the transition matrix $P$ satisfies $P_{ij}\pi_i = P_{ji}\pi_j$ for all $i,j$. This implies that $P$ is self-adjoint with respect to the inner product $\langle f,g\rangle_\pi = \sum_k \pi_k f_k g_k$. Consequently, $P$ admits a spectral decomposition with real eigenvalues $1 = \lambda_1 > \lambda_2 \geq \cdots \geq \lambda_n \geq -1$ and corresponding right eigenvectors $\{\phi_k\}_{k=1}^n$ that form an orthonormal basis with respect to $\langle \cdot,\cdot\rangle_\pi$, where $\phi_1 \equiv 1$ and $\langle \phi_j,\phi_k\rangle_\pi = \delta_{jk}$ for all $j,k$. Let $\lambda = \max_{j>1} |\lambda_j| < 1$.

The transition probabilities at time $t$ admit the spectral representation
\[
P^t_{xy} = \pi_y \sum_{k=1}^n \lambda_k^t \phi_k(x) \phi_k(y)
\]
for all states $x,y$. Thus,
\[
P^t_{xy} - \pi_y = \pi_y \sum_{k=2}^n \lambda_k^t \phi_k(x) \phi_k(y),
\]
and the relative deviation is
\[
\frac{P^t_{xy} - \pi_y}{\pi_y} = \sum_{k=2}^n \lambda_k^t \phi_k(x) \phi_k(y).
\]
The absolute value satisfies
\[
\left| \frac{P^t_{xy} - \pi_y}{\pi_y} \right| = \left| \sum_{k=2}^n \lambda_k^t \phi_k(x) \phi_k(y) \right| \leq \sum_{k=2}^n |\lambda_k|^t |\phi_k(x)| |\phi_k(y)| \leq \lambda^t \sum_{k=2}^n |\phi_k(x) \phi_k(y)|.
\]
However, a tighter bound follows from the Cauchy--Schwarz inequality:
\[
\left| \sum_{k=2}^n \lambda_k^t \phi_k(x) \phi_k(y) \right| \leq \left( \sum_{k=2}^n |\lambda_k|^{2t} \phi_k(x)^2 \right)^{1/2} \left( \sum_{k=2}^n \phi_k(y)^2 \right)^{1/2} \leq \lambda^t \left( \sum_{k=2}^n \phi_k(x)^2 \right)^{1/2} \left( \sum_{k=2}^n \phi_k(y)^2 \right)^{1/2},
\]
since $|\lambda_k|^{2t} \leq \lambda^{2t}$ for $k \geq 2$.

From the spectral representation at $t=0$,
\[
\sum_{k=1}^n \phi_k(x) \phi_k(y) = \frac{\delta_{xy}}{\pi_y},
\]
it follows that
\[
\sum_{k=1}^n \phi_k(z)^2 = \frac{1}{\pi_z}
\]
for each $z$. Therefore,
\[
\sum_{k=2}^n \phi_k(z)^2 = \frac{1}{\pi_z} - \phi_1(z)^2 = \frac{1}{\pi_z} - 1 \leq \frac{1}{\pi_z}
\]
for each $z$, and hence
\[
\left( \sum_{k=2}^n \phi_k(z)^2 \right)^{1/2} \leq \frac{1}{\sqrt{\pi_z}}.
\]
Substituting yields
\[
\left| \frac{P^t_{xy} - \pi_y}{\pi_y} \right| \leq \lambda^t \cdot \frac{1}{\sqrt{\pi_x}} \cdot \frac{1}{\sqrt{\pi_y}} = \frac{\lambda^t}{\sqrt{\pi_x \pi_y}}.
\]
Since $\sqrt{\pi_x \pi_y} \geq \min_i \pi_i$,
\[
\frac{1}{\sqrt{\pi_x \pi_y}} \leq \frac{1}{\min_i \pi_i}.
\]
Thus,
\[
\left| \frac{P^t_{xy} - \pi_y}{\pi_y} \right| \leq \frac{\lambda^t}{\min_i \pi_i}
\]
for all $x,y$. For fixed initial state $X_0 = x$, the relative pointwise distance satisfies
\[
\Delta(t) = \max_y \left| \frac{P^t_{x y} - \pi_y}{\pi_y} \right| \leq \frac{\lambda^t}{\min_i \pi_i}.
\]

In the random walk setting of Theorem 6.21, the spectral gap $1 - \lambda > 0$ controls the decay rate $\lambda^t = e^{t \log \lambda} \approx e^{-t(1-\lambda)}$, yielding an explicit upper bound on the mixing time $\tau(\varepsilon) = \min\{ t : \Delta(t) \leq \varepsilon \}$ of order
\[
\tau(\varepsilon) \leq \frac{\log(\varepsilon^{-1} (\min_i \pi_i)^{-1})}{1 - \lambda},
\]
which depends on the spectral gap $1 - \lambda$ and the minimum stationary probability $\min_i \pi_i$.
\end{proof}

\subsubsection{Gemini}

\subsubsection{Claude}
\begin{proof}
We prove the theorem in several steps, establishing the relationship between time reversibility and symmetric matrices, then deriving the convergence bound.

\textbf{Step 1: Equivalence of time reversibility and symmetry of $DPD^{-1}$.}

First, we establish that the detailed balance condition $P_{ij}\pi_i = P_{ji}\pi_j$ is equivalent to the symmetry of $DPD^{-1}$, where $D$ is the diagonal matrix with $D_{ii} = \sqrt{\pi_i}$.

Consider the $(i,j)$-entry of $DPD^{-1}$:
\[
(DPD^{-1})_{ij} = D_{ii} P_{ij} (D^{-1})_{jj} = \sqrt{\pi_i} \cdot P_{ij} \cdot \frac{1}{\sqrt{\pi_j}}.
\]

Similarly, the $(j,i)$-entry is:
\[
(DPD^{-1})_{ji} = \sqrt{\pi_j} \cdot P_{ji} \cdot \frac{1}{\sqrt{\pi_i}}.
\]

The matrix $DPD^{-1}$ is symmetric if and only if $(DPD^{-1})_{ij} = (DPD^{-1})_{ji}$ for all $i,j$. This holds if and only if:
\[
\sqrt{\pi_i} \cdot P_{ij} \cdot \frac{1}{\sqrt{\pi_j}} = \sqrt{\pi_j} \cdot P_{ji} \cdot \frac{1}{\sqrt{\pi_i}},
\]
which simplifies to $P_{ij}\pi_i = P_{ji}\pi_j$, the detailed balance condition.

\textbf{Step 2: Spectral properties of time-reversible chains.}

Since $DPD^{-1}$ is symmetric when the chain is time reversible, it has real eigenvalues and an orthonormal basis of eigenvectors. Let $\lambda_1 \geq \lambda_2 \geq \cdots \geq \lambda_n$ denote the eigenvalues of $P$ (which are the same as those of $DPD^{-1}$). Since $P$ is a stochastic matrix of an ergodic chain with stationary distribution $\pi$, we have $\lambda_1 = 1$ with corresponding eigenvector proportional to the constant vector when transformed appropriately.

Define $\lambda = \max_{j > 1} |\lambda_j|$. This represents the second-largest eigenvalue in absolute value.

\textbf{Step 3: Spectral decomposition and distance bound.}

Let $P^t(i,j)$ denote the $t$-step transition probability from state $i$ to state $j$. The relative pointwise distance is defined as:
\[
\Delta(t) = \max_{i,j} \frac{|P^t(i,j) - \pi_j|}{\pi_j}.
\]

Using the spectral decomposition of the symmetric matrix $DPD^{-1}$, we can write:
\[
P^t = D^{-1}(DPD^{-1})^t D.
\]

Let $v_1, v_2, \ldots, v_n$ be the orthonormal eigenvectors of $DPD^{-1}$ with eigenvalues $\lambda_1, \lambda_2, \ldots, \lambda_n$. The eigenvector corresponding to $\lambda_1 = 1$ is $v_1 = D\mathbf{1}/\|D\mathbf{1}\|$, where $\mathbf{1}$ is the vector of all ones.

The spectral decomposition gives:
\[
(DPD^{-1})^t = \sum_{k=1}^n \lambda_k^t v_k v_k^T.
\]

Since $\pi$ is the stationary distribution, we have $P^t(i,j) = \pi_j + \sum_{k=2}^n \lambda_k^t \phi_k(i) \psi_k(j)$, where $\phi_k$ and $\psi_k$ are functions determined by the eigenvectors.

For time-reversible chains, the detailed analysis shows:
\[
|P^t(i,j) - \pi_j| \leq \pi_j \sum_{k=2}^n |\lambda_k|^t \frac{1}{\sqrt{\pi_i \pi_j}} \leq \pi_j \lambda^t \frac{1}{\sqrt{\pi_i \pi_j}} = \lambda^t \sqrt{\frac{\pi_j}{\pi_i}}.
\]

\textbf{Step 4: Bounding the relative pointwise distance.}

Taking the maximum over all $j$ and using $\sqrt{\pi_j/\pi_i} \leq 1/\sqrt{\pi_i}$:
\[
\frac{|P^t(i,j) - \pi_j|}{\pi_j} \leq \lambda^t \frac{1}{\pi_j} \sqrt{\frac{\pi_j}{\pi_i}} = \lambda^t \frac{1}{\sqrt{\pi_i \pi_j}} \leq \frac{\lambda^t}{\sqrt{\pi_i} \min_k \sqrt{\pi_k}}.
\]

Taking the maximum over all initial states $i$:
\[
\Delta(t) = \max_{i,j} \frac{|P^t(i,j) - \pi_j|}{\pi_j} \leq \max_i \frac{\lambda^t}{\pi_i} = \frac{\lambda^t}{\min_i \pi_i}.
\]

This establishes the desired bound. The mixing time, defined as the time to reach a specified distance from stationarity, thus depends on $\lambda^t$ decaying exponentially with rate determined by the spectral gap $1 - \lambda$, and inversely on the minimum stationary probability $\min_i \pi_i$.
\end{proof}

\section{Problems in Chapter 12}\label{app:problems_chapter12}

\subsection{Problem 1}

\subsubsection{Problem Statement}
\begin{theorem}
The parallel variant of randomized quicksort described in Section 12.2 sorts $n$ elements with $n$ processors on a CREW PRAM, with high probability in $O(\log^2 n)$ steps.
\end{theorem}

\subsubsection{GPT}
\begin{proof}
Fix $n\ge 2$ and consider the following randomized parallel quicksort on a CREW PRAM with $n$ processors. At any time the input is represented as a concatenation of disjoint subarrays. For each subarray $A$ of size $m$, choose a pivot uniformly at random from $A$, compute in parallel the Boolean vectors indicating, for every element $a\in A$, whether $a<p$ or $a>p$, compute parallel prefix sums of these vectors to obtain the target positions of all elements less than $p$ and greater than $p$, and scatter them to two contiguous subarrays $A_{<p}$ and $A_{>p}$ with the pivot placed between. All subarrays are processed simultaneously at each recursion depth using disjoint memory regions. When a subarray has size $1$ it is finished. We analyze the step complexity and show that the algorithm terminates in $O(\log^2 n)$ steps with high probability.

First, the per-level step complexity is $O(\log n)$. Indeed, for each subarray $A$ of size $m$, the parallel partition just described is realizable on a CREW PRAM in $O(\log m)$ steps using $O(m)$ processors via two parallel prefix sums and a constant number of parallel reads and exclusive writes, and since $m\le n$ one has $O(\log m)\le O(\log n)$. At a fixed recursion depth the subarrays are disjoint and their sizes sum to at most $n$. Assigning one processor per element across all subarrays uses at most $n$ processors. Executing, e.g., a segmented-scan implementation of prefix sums over the whole array of length $n$ computes all per-subarray scans concurrently in $O(\log n)$ steps on a CREW PRAM. The scatter uses exclusive writes to disjoint targets. Hence each recursion depth costs $O(\log n)$ steps.

It remains to bound the recursion depth with high probability. Fix an arbitrary key $x$ from the input and expose the random choices only along the unique path of recursive subproblems that contain $x$. Let $S_0$ denote the initial set of $n$ elements and, inductively, let $S_{i+1}$ be the subproblem containing $x$ after the $i$-th partition on that path, with $|S_i|=n_i$. Call a pivot choice \emph{good} at a subproblem $S_i$ if it splits $S_i$ so that $\frac{1}{4}n_i\le |S_{i,<p}|\le \frac{3}{4}n_i$ and hence $\max\{|S_{i,<p}|,|S_{i,>p}|\}\le \frac{3}{4}n_i$. For a uniformly random pivot, $\Pr[\text{good}]\ge \frac{1}{2}$ because at least the middle half of the order statistics, namely the pivots whose ranks lie in $[\lceil n_i/4\rceil,\lfloor 3n_i/4\rfloor]$, are good. Define indicator variables $Y_i$ for the $i$-th level along the path by $Y_i=1$ if the pivot is good and $Y_i=0$ otherwise. Conditional on the past along this path, $Y_i$ are independent and satisfy $\Pr[Y_i=1]\ge 1/2$. After $k$ good splits along the path one has the multiplicative shrinkage
\[
n_k\le n\left(\frac{3}{4}\right)^{k}.
\]
Consequently, the path length $T_x$ until $|S_i|=1$ is at most the smallest $t$ such that the number of good splits among the first $t$ levels is at least $K:=\left\lceil \log_{4/3} n\right\rceil$, since then $n\left(\frac{3}{4}\right)^{K}\le 1$.

Fix any constant $\gamma\in(0,1/2)$, e.g., $\gamma=1/3$. Choose a constant $c_1>0$ so that $\gamma c_1\ln n \ge \ln n/\ln(4/3)$ for all $n\ge 2$, which is achieved by taking $c_1\ge \frac{1}{\gamma\ln(4/3)}$. Set $t=c_1\ln n$. Let $Z_t=\sum_{i=1}^{t} Y_i$. Then $\mathbb{E}[Z_t]\ge t/2$. By the Chernoff bound, for $\delta\in(0,1)$,
\[
\Pr\!\left[Z_t\le (1-\delta)\frac{t}{2}\right]\le \exp\!\left(-\frac{\delta^2 t}{4}\right).
\]
With the choice $\delta=1-2\gamma$ one has $(1-\delta)\frac{t}{2}=\gamma t$. Hence
\[
\Pr[Z_t<\gamma t]\le \exp\!\left(-\frac{(1-2\gamma)^2}{4}t\right)=n^{-\Omega(1)}.
\]
By the choice of $c_1$ and $\gamma$, $\gamma t=\gamma c_1\ln n\ge \ln n/\ln(4/3)\ge K$. Therefore,
\[
\Pr[T_x>t]\le \Pr[Z_t<K]\le \Pr[Z_t<\gamma t]\le n^{-\Omega(1)}.
\]
Thus, for any fixed element $x$, $T_x\le c_1\ln n=O(\log n)$ with probability at least $1-n^{-\alpha}$ for some constant $\alpha>0$ depending only on $\gamma$ and $c_1$.

To turn this into a high-probability bound for the entire recursion tree, apply a union bound over all $n$ elements (equivalently over all root-to-leaf paths). Let $T=\max_x T_x$ denote the recursion depth. Then
\[
\Pr[T>c_1\ln n]\le n\cdot n^{-\alpha}=n^{1-\alpha}.
\]
By increasing $c_1$ (thereby increasing $\alpha$ via the Chernoff exponent), we obtain $\Pr[T>c_1\ln n]\le n^{-\beta}$ for any desired constant $\beta>0$. Hence $T=O(\log n)$ with high probability.

Combining the bounds, there are $O(\log n)$ recursion depths with high probability, and each depth executes in $O(\log n)$ CREW steps using $n$ processors. Therefore the total step complexity is $O(\log n)\cdot O(\log n)=O(\log^2 n)$ with high probability. This establishes that the parallel randomized quicksort described sorts $n$ elements with $n$ processors on a CREW PRAM within $O(\log^2 n)$ steps, with probability at least $1-n^{-\beta}$ for any fixed $\beta>0$ for sufficiently large $n$.
\end{proof}

\subsubsection{Grok}
\begin{proof}
Consider the parallel randomized quicksort algorithm on a CREW PRAM with $n$ processors. The algorithm recursively sorts an array $A$ of $n$ distinct elements as follows. If $n \leq 1$, terminate. Otherwise, select a uniformly random index $k \in \{1, \dots, n\}$ in $O(1)$ time and broadcast the pivot value $p = A[k]$ to all processors in $O(\log n)$ time. Swap $A[k]$ with $A[n]$ in $O(1)$ time, so the pivot is now at the end. For $i = 1$ to $n-1$, set $b_i = 1$ if $A[i] < p$ and $b_i = 0$ otherwise, in $O(1)$ time per processor. Compute the prefix sums $ps_i = \sum_{j=1}^i b_j$ for $i = 1$ to $n-1$ in $O(\log n)$ time using the standard parallel prefix sum algorithm on the CREW PRAM with $n$ processors. Let $s = ps_{n-1}$, the number of elements less than $p$. In parallel, for each $i = 1$ to $n-1$:
\begin{itemize}
\item If $b_i = 1$, write $A[i]$ to position $ps_i$ in a temporary array $B$.
\item If $b_i = 0$, write $A[i]$ to position $s + (i - ps_{i-1}) + 1$ in $B$, where $ps_0 = 0$.
\end{itemize}
This step takes $O(1)$ time and ensures no write conflicts due to the uniqueness of computed positions, preserving the relative order within the subarrays. Place the pivot at $B[s+1]$. Copy $B$ back to $A$ in $O(\log n)$ time using parallel addressing. Recurse on the left subarray $A[1..s]$ with $s$ processors and the right subarray $A[s+2..n]$ with $n-s-1$ processors, leaving idle processors available for other subcomputations. The partitioning phase thus takes $O(\log n)$ time overall.

Let $T(n)$ denote the time to sort $n$ elements. Then $T(1) = 0$ and, for $n > 1$,
\[
T(n) = O(\log n) + \max\bigl( T(|L|), T(|R|) \bigr),
\]
where $|L| = s$ and $|R| = n - s - 1$ are the sizes of the left and right subarrays. Since the pivot index is uniform, $|L|$ is uniformly distributed over $\{0, \dots, n-1\}$. Let $D(n)$ be the depth of the recursion tree, defined by $D(1) = 0$ and $D(n) = 1 + \max\bigl( D(|L|), D(|R|) \bigr)$ for $n > 1$. Then $T(n) \leq c \log n \cdot D(n)$ for some constant $c > 0$.

To bound $D(n)$, note that $\max(|L|, |R|) \geq \frac{3n}{4}$ if and only if the pivot rank is in $\{1, \dots, \lfloor n/4 \rfloor\} \cup \{\lceil 3n/4 \rceil, \dots, n\}$, which occurs with probability at most $\frac{1}{2}$. Thus, with probability at least $\frac{1}{2}$, both $|L| < \frac{3n}{4}$ and $|R| < \frac{3n}{4}$. The event $D(n) > d$ requires a root-to-leaf path of length $d$ in the recursion tree, which in turn requires at least $d-1$ consecutive unbalanced partitions (where the larger subarray is followed) along that path. The probability of $d-1$ consecutive unbalanced partitions is at most $\left(\frac{1}{2}\right)^{d-1}$, since each unbalanced step occurs with probability at most $\frac{1}{2}$ independently.

Although the recursion tree branches, the probability that any specific path of length $d$ exists and is unbalanced throughout is at most $\left(\frac{1}{2}\right)^{d-1}$. The number of potential paths of length $d$ is at most $2^d$, but a tighter bound follows from observing that $D(n) > d$ implies the existence of at least one such unbalanced chain, and by the structure of the uniform pivot choice, the overall probability satisfies
\[
\Pr[D(n) > d] \leq n \cdot \left(\frac{3}{4}\right)^{\Omega(d / \log n)},
\]
but a simpler exponential bound suffices: $\Pr[D(n) > d] \leq \left(\frac{1}{2}\right)^{d/2}$ for sufficiently large $d$, derived from a Chernoff-like concentration on the pivot ranks along any lineage (see, e.g., the analysis of quicksort tree height). Setting $d = 8 \log_2 n$, we have $\Pr[D(n) > 8 \log_2 n] \leq n^{-4}$. By the union bound over all subproblems (at most $n$ leaves), the probability that any subproblem has depth exceeding $8 \log_2 n$ is at most $n \cdot n^{-4} = n^{-3}$.

Thus, with probability at least $1 - n^{-3}$, $D(n) \leq 8 \log_2 n = O(\log n)$, so $T(n) = O(\log n \cdot \log n) = O(\log^2 n)$. For constant success probability, adjust the constant; for $1 - 1/n^k$, increase the depth multiple accordingly.
\end{proof}

\subsubsection{Gemini}

\subsubsection{Claude}
\begin{proof}
We analyze the parallel randomized quicksort algorithm that operates on a CREW PRAM with $n$ processors sorting $n$ elements. The proof proceeds by establishing the depth of the recursion tree and the work performed at each level.

\textbf{Algorithm Overview:} The parallel randomized quicksort selects a random pivot, partitions the array in parallel around this pivot, and recursively sorts the two subarrays in parallel. On a CREW PRAM with $n$ processors, the partitioning step can be performed in $O(\log n)$ time using parallel prefix operations.

\textbf{Step 1: Pivot Quality Analysis.} Let $S$ denote the current subarray of size $|S|$. We say that a pivot $p$ is \emph{good} if it partitions $S$ such that both resulting subarrays have size at most $\frac{3|S|}{4}$. Since the pivot is chosen uniformly at random, a pivot is good if it falls in the middle half of the sorted order of $S$. Therefore, the probability that a randomly chosen pivot is good is at least $\frac{1}{2}$.

\textbf{Step 2: Recursion Depth with High Probability.} Consider the random variable $X_i$ which is $1$ if the $i$-th pivot choice is good and $0$ otherwise. The expected number of good pivots in $k$ independent random choices is $E[\sum_{i=1}^{k} X_i] = \frac{k}{2}$. By the Chernoff bound, for $k = 4\log_2 n$ pivot selections, the probability that we obtain fewer than $2\log_2 n$ good pivots is exponentially small in $\log n$.

Each good pivot reduces the problem size by a factor of at most $\frac{3}{4}$. After $2\log_2 n$ good pivots, the maximum subarray size is at most $n \cdot \left(\frac{3}{4}\right)^{2\log_2 n}$. We have
\[
\left(\frac{3}{4}\right)^{2\log_2 n} = \left[\left(\frac{3}{4}\right)^{\log_2 n}\right]^2 = \left[n^{\log_2(3/4)}\right]^2 = n^{2\log_2(3/4)} = n^{\log_2(9/16)} < n^{-1}.
\]
Therefore, after $2\log_2 n$ good pivots, all subarrays have size $O(1)$, which means the recursion terminates.

Since we need at most $4\log_2 n$ pivot selections to obtain $2\log_2 n$ good pivots with high probability (probability at least $1 - n^{-c}$ for some constant $c > 0$), the recursion depth is $O(\log n)$ with high probability.

\textbf{Step 3: Work per Level.} At each level of the recursion tree, we perform partitioning operations on disjoint subarrays. The total size of all subarrays at any level is at most $n$. With $n$ processors available and using parallel prefix operations, each partitioning step at a level can be completed in $O(\log n)$ time. Specifically, we can:
\begin{itemize}
\item[(a)] Compare all elements with the pivot in parallel in $O(1)$ time.
\item[(b)] Compute prefix sums to determine final positions in $O(\log n)$ time using a parallel prefix algorithm on the CREW PRAM.
\item[(c)] Move elements to their final positions, which can be done in $O(1)$ time with concurrent reads.
\end{itemize}

Thus, each level of recursion requires $O(\log n)$ time.

\textbf{Step 4: Total Time Complexity.} Since the recursion depth is $O(\log n)$ with high probability and each level requires $O(\log n)$ time, the total time complexity is $O(\log n) \times O(\log n) = O(\log^2 n)$ with high probability.

More precisely, the probability that the algorithm does not complete in $c\log^2 n$ steps (for sufficiently large constant $c$) is bounded by the probability that the recursion depth exceeds $c'\log n$ for appropriate constant $c'$, which by the Chernoff bound is at most $n^{-\Omega(1)}$. This satisfies the definition of high probability, meaning the event occurs with probability at least $1 - n^{-c}$ for any constant $c$ by choosing the hidden constants appropriately.

Therefore, the parallel randomized quicksort sorts $n$ elements using $n$ processors on a CREW PRAM in $O(\log^2 n)$ time with high probability.
\end{proof}

\subsection{Problem 2}

\subsubsection{Problem Statement}
\begin{lemma}[12.1]
The following outline is suggested (refer to Section 12.2 for the notation).

\textbf{1.} Bound the probability that $\alpha(B_{j+1}) = \alpha(B_j)$ using the result of Exercise 12.6.

\textbf{2.} Bound the probability that for any particular $k$, the value $k$ is contained more than $r_*$ times in the sequence $\alpha(B_1), \ldots, \alpha(B_t)$.

\textbf{3.} Bound the probability that for $1 \leq k \leq c \log \log n$, the value $k$ is contained more than $r_*$ times in the sequence $\alpha(B_1), \ldots, \alpha(B_t)$.
\end{lemma}

\subsubsection{GPT}
\begin{proof}
We will proceed step-by-step as outlined in the lemma.

\textbf{Step 1.} Bound the probability that $\alpha(B_{j+1}) = \alpha(B_j)$ using the result of Exercise 12.6.

Let $B_1, \ldots, B_t$ be a sequence of events, and $\alpha$ be a function mapping each event to a value. We are tasked with bounding the probability that for any $j$, $\alpha(B_{j+1}) = \alpha(B_j)$. From Exercise 12.6, we know that the probability that two consecutive values $\alpha(B_j)$ and $\alpha(B_{j+1})$ are equal can be bounded as follows:

\[
\mathbb{P}(\alpha(B_{j+1}) = \alpha(B_j)) \leq p,
\]
where $p$ is some constant probability bound provided by the result of Exercise 12.6.

\textbf{Step 2.} Bound the probability that for any particular $k$, the value $k$ is contained more than $r_*$ times in the sequence $\alpha(B_1), \ldots, \alpha(B_t)$.

Next, we consider the probability that a particular value $k$ appears more than $r_*$ times in the sequence $\alpha(B_1), \ldots, \alpha(B_t)$. Let $X_k$ denote the number of times the value $k$ appears in the sequence. Then $X_k$ is the sum of indicator random variables $X_{k,j}$ for $j = 1, \ldots, t$, where:

\[
X_{k,j} = \begin{cases} 
1 & \text{if } \alpha(B_j) = k, \\
0 & \text{otherwise}.
\end{cases}
\]

Thus, the total count of $k$ in the sequence is:

\[
X_k = \sum_{j=1}^t X_{k,j}.
\]

By the linearity of expectation, the expected number of times $k$ appears is:

\[
\mathbb{E}[X_k] = \sum_{j=1}^t \mathbb{E}[X_{k,j}] = t \cdot \mathbb{P}(\alpha(B_j) = k).
\]

Let $p_k = \mathbb{P}(\alpha(B_j) = k)$. Then:

\[
\mathbb{E}[X_k] = t \cdot p_k.
\]

To bound the probability that $X_k > r_*$, we apply Markov's inequality:

\[
\mathbb{P}(X_k > r_*) \leq \frac{\mathbb{E}[X_k]}{r_*} = \frac{t \cdot p_k}{r_*}.
\]

Thus, the probability that any particular $k$ appears more than $r_*$ times in the sequence is bounded by $\frac{t \cdot p_k}{r_*}$.

\textbf{Step 3.} Bound the probability that for $1 \leq k \leq c \log \log n$, the value $k$ is contained more than $r_*$ times in the sequence $\alpha(B_1), \ldots, \alpha(B_t)$.

For this step, we consider the union of the events that $k$ appears more than $r_*$ times for each $k$ in the range $1 \leq k \leq c \log \log n$. Using the union bound, we get:

\[
\mathbb{P}\left( \bigcup_{k=1}^{c \log \log n} \{ X_k > r_* \} \right) \leq \sum_{k=1}^{c \log \log n} \mathbb{P}(X_k > r_*).
\]

From Step 2, we know that for each $k$:

\[
\mathbb{P}(X_k > r_*) \leq \frac{t \cdot p_k}{r_*}.
\]

Thus, the total probability is bounded by:

\[
\mathbb{P}\left( \bigcup_{k=1}^{c \log \log n} \{ X_k > r_* \} \right) \leq \sum_{k=1}^{c \log \log n} \frac{t \cdot p_k}{r_*}.
\]

Finally, assuming that $p_k$ is bounded above by a constant probability $p_{\text{max}}$, we have:

\[
\mathbb{P}\left( \bigcup_{k=1}^{c \log \log n} \{ X_k > r_* \} \right) \leq \frac{t \cdot p_{\text{max}}}{r_*} \cdot c \log \log n.
\]

Thus, the probability that for any $k \leq c \log \log n$, the value $k$ appears more than $r_*$ times is bounded by:

\[
\mathbb{P}\left( \bigcup_{k=1}^{c \log \log n} \{ X_k > r_* \} \right) \leq \frac{t \cdot p_{\text{max}} \cdot c \log \log n}{r_*}.
\]

This completes the proof of the lemma.
\end{proof}

\subsubsection{Grok}

\subsubsection{Gemini}

\subsubsection{Claude}
\begin{proof}
We follow the suggested outline to establish the lemma.

\textbf{Step 1: Bounding $\mathbb{P}(\alpha(B_{j+1}) = \alpha(B_j))$.}

By the result of Exercise 12.6, for any $j \in \{1, \ldots, t-1\}$, we have
\[
\mathbb{P}(\alpha(B_{j+1}) = \alpha(B_j)) \leq \frac{C}{\log n}
\]
for some constant $C > 0$. This bound holds because the blocks $B_j$ and $B_{j+1}$ are constructed such that their associated values $\alpha(B_j)$ and $\alpha(B_{j+1})$ are approximately independent, and the range of possible values is of order $\log n$.

\textbf{Step 2: Bounding the probability that a particular $k$ appears more than $r_*$ times.}

Fix a particular value $k$. For each $j \in \{1, \ldots, t\}$, let $X_j$ be the indicator random variable such that $X_j = 1$ if $\alpha(B_j) = k$ and $X_j = 0$ otherwise. Then the total number of times $k$ appears in the sequence $\alpha(B_1), \ldots, \alpha(B_t)$ is $S = \sum_{j=1}^{t} X_j$.

By the construction in Section 12.2, we have $\mathbb{P}(\alpha(B_j) = k) \leq \frac{C'}{\log n}$ for some constant $C' > 0$ and for each $j$. Therefore, $\mathbb{E}[S] \leq \frac{C' t}{\log n}$.

Now, set $r_* = \frac{2C' t}{\log n}$. By Markov's inequality,
\[
\mathbb{P}(S > r_*) = \mathbb{P}\left(S > \frac{2C' t}{\log n}\right) \leq \frac{\mathbb{E}[S]}{r_*} \leq \frac{C' t / \log n}{2C' t / \log n} = \frac{1}{2}.
\]

For a sharper bound, we can apply Chernoff's inequality. Since the $X_j$ are not necessarily independent but satisfy certain mixing conditions, we obtain
\[
\mathbb{P}(S > r_*) \leq \exp\left(-\frac{r_*}{4}\right) = \exp\left(-\frac{C' t}{2\log n}\right).
\]

\textbf{Step 3: Bounding the probability that any $k \in \{1, \ldots, c\log\log n\}$ appears more than $r_*$ times.}

We now take a union bound over all possible values $k \in \{1, \ldots, c\log\log n\}$. Using the bound from Step 2, we have
\[
\mathbb{P}\left(\exists k \in \{1, \ldots, c\log\log n\} : S_k > r_*\right) \leq \sum_{k=1}^{c\log\log n} \mathbb{P}(S_k > r_*)
\]
where $S_k$ denotes the number of times $k$ appears in the sequence $\alpha(B_1), \ldots, \alpha(B_t)$.

Applying the bound from Step 2 to each term,
\[
\mathbb{P}\left(\exists k \in \{1, \ldots, c\log\log n\} : S_k > r_*\right) \leq c\log\log n \cdot \exp\left(-\frac{C' t}{2\log n}\right).
\]

Assuming $t \geq \frac{4\log n \log\log n}{C'}$, we obtain
\[
c\log\log n \cdot \exp\left(-\frac{C' t}{2\log n}\right) \leq c\log\log n \cdot \exp(-2\log\log n) = c\log\log n \cdot (\log n)^{-2} = \frac{c\log\log n}{(\log n)^2} \to 0
\]
as $n \to \infty$.

Therefore, with high probability (approaching 1 as $n \to \infty$), no value $k \in \{1, \ldots, c\log\log n\}$ appears more than $r_* = \frac{2C' t}{\log n}$ times in the sequence $\alpha(B_1), \ldots, \alpha(B_t)$.
\end{proof}

\subsection{Problem 3}

\subsubsection{Problem Statement}
\begin{theorem}
Let BoxSort be a sorting algorithm where Stage 1 chooses random samples using pairwise independent random variables (rather than completely independent random variables), with the choices made by various boxes being independent of each other. Then the expected number of parallel steps taken by BoxSort is $O(\log n)$.
\end{theorem}

\subsubsection{GPT}
\begin{proof}
We are tasked with proving that the number of parallel steps taken by the BoxSort algorithm is $O(\log n)$ with high probability, where Stage 1 uses pairwise independent random variables for sampling, and the choices made by various boxes are independent of each other.

Let $n$ denote the number of elements to be sorted, and let the algorithm proceed through multiple stages. At each stage, the algorithm divides the elements into boxes, with each box making a decision based on pairwise independent random variables. Our goal is to establish that the number of parallel steps required to sort the elements is logarithmic in $n$ with high probability.

 Step 1: Understanding the process at each stage

At each stage, the BoxSort algorithm performs the following:
1. It randomly selects a subset of elements to be processed in parallel, with the random selection governed by pairwise independent random variables.
2. After processing the selected elements, the remaining elements are divided into smaller subsets or boxes, and the process is repeated for each box.

Since the random variables used for sampling are pairwise independent, the choices made by different boxes are independent of one another. This setup ensures that the number of elements remaining in each box after a stage of sorting decreases by a constant fraction, as each stage reduces the number of elements by a factor of approximately $2$ with high probability.

 Step 2: Bounding the number of steps

Let $k$ represent the number of parallel steps required to process $n$ elements. At each stage, the number of elements that need to be processed is halved with high probability due to the random choices made by the pairwise independent random variables.

Thus, after the first step, we expect approximately $n/2$ elements to remain to be processed, after the second step, approximately $n/4$ elements remain, and so on. After $k$ steps, the number of remaining elements is approximately $n/2^k$.

 Step 3: Deriving the stopping condition

The algorithm stops when there are no more elements to process, i.e., when the number of elements left is reduced to $1$. Therefore, we set the condition for termination as $n/2^k \leq 1$. Solving for $k$, we obtain:
\[
\frac{n}{2^k} \leq 1 \quad \Rightarrow \quad 2^k \geq n \quad \Rightarrow \quad k \geq \log_2 n.
\]
Thus, the number of steps required to process all elements is at most $O(\log n)$.

 Step 4: High probability analysis

Next, we show that with high probability, the algorithm terminates in $O(\log n)$ steps. To do this, we examine the randomness of the process. Since the random variables used for sampling are pairwise independent, the expected number of remaining elements after each step decreases by a constant factor. 

By applying Chernoff bounds for pairwise independent random variables, we can conclude that the probability of deviating significantly from the expected number of elements remaining at any stage is small. Specifically, the number of elements remaining at each stage follows a distribution that concentrates around its expected value, and the probability of requiring more than $O(\log n)$ steps is exponentially small.

Thus, the number of parallel steps taken by BoxSort is $O(\log n)$ with high probability.

\end{proof}

\subsubsection{Grok}

\subsubsection{Gemini}
\begin{proof}
Let $n$ be the total number of keys to be sorted. The BoxSort algorithm operates by selecting a sample of keys to determine splitters, which partition the remaining keys into buckets (boxes). Let $S$ denote the set of keys selected for the sample in Stage 1. Let $\sigma$ be the sampling probability for each key, such that the expected sample size is $|S| = \sigma n$.

Let $x_1, x_2, \dots, x_n$ denote the keys in strictly increasing sorted order. We analyze the distribution of these keys into the buckets defined by the sample $S$. A "bad" event occurs if the distance between two consecutive splitters in the sorted order is too large, resulting in a bucket with a disproportionately large number of keys.

Consider a segment of the sorted keys $Z_k = \{x_i \mid k < i \le k + L\}$ of length $L$. Let $X_j$ be the indicator random variable that key $x_j$ is selected into the sample $S$. The total number of sample elements falling into the segment $Z_k$ is given by the random variable $Y = \sum_{j=k+1}^{k+L} X_j$.

The expectation of $Y$ is given by linearity of expectation:
\[ E[Y] = \sum_{j=k+1}^{k+L} E[X_j] = L\sigma. \]

The theorem states that the sampling uses pairwise independent random variables within boxes, and choices between boxes are independent. Let the keys be distributed across $m$ boxes $B_1, \dots, B_m$. The sum $Y$ can be decomposed as $Y = \sum_{r=1}^m Y_r$, where $Y_r$ is the number of selected keys from segment $Z_k$ that reside in box $B_r$. Since choices between boxes are independent, the variance of the sum is the sum of the variances:
\[ \text{Var}(Y) = \sum_{r=1}^m \text{Var}(Y_r). \]
Within any specific box $B_r$, the variables are pairwise independent. For a sum of pairwise independent variables, the variance is additive. Thus,
\[ \text{Var}(Y_r) = \sum_{j \in Z_k \cap B_r} \text{Var}(X_j). \]
Combining these, we obtain:
\[ \text{Var}(Y) = \sum_{j=k+1}^{k+L} \text{Var}(X_j). \]
Since $X_j$ is a Bernoulli trial with probability $\sigma$, $\text{Var}(X_j) = \sigma(1-\sigma) < \sigma$. Therefore:
\[ \text{Var}(Y) < L\sigma = E[Y]. \]

We apply Chebyshev's inequality to bound the probability that the segment $Z_k$ contains no splitters (i.e., $Y=0$). Note that $Y=0$ implies deviation from the mean by at least $E[Y]$.
\[ P(Y = 0) \le P(|Y - E[Y]| \ge E[Y]) \le \frac{\text{Var}(Y)}{(E[Y])^2} < \frac{L\sigma}{(L\sigma)^2} = \frac{1}{L\sigma}. \]

Let the target maximum bucket size be $L = \frac{c}{\sigma}$ for some constant $c > 1$. The probability that a specific segment of length $L$ contains no sample points is strictly less than $1/c$. By choosing $c$ sufficiently large, we ensure that with at least constant probability, the maximum distance between splitters is bounded by $O(1/\sigma)$. Consequently, the maximum size of any bucket formed by the splitters is bounded by $O(n / |S|) \cdot n = O(n/\sigma n \cdot n)$? No, the bucket size is bounded by $L$. If the sample size is chosen such that the number of buckets is proportional to $n$ (or a power of $n$), say $\sigma = n^{-\epsilon}$, the problem size reduces geometrically.

Specifically, in the context of parallel sorting algorithms like BoxSort, the algorithm proceeds recursively. Let $n_i$ be the size of the largest subproblem at depth $i$ of the recursion. The analysis above shows that for a chosen split factor, the size of the subproblems satisfies $n_{i+1} \le \beta n_i$ for some constant $\beta < 1$ with high probability (or at least constant probability sufficient to bound the expectation).

Let $T(n)$ be the expected number of parallel steps to sort $n$ keys. The partitioning step (Stage 1 and subsequent routing) takes $O(1)$ or $O(\log n)$ time depending on the specific PRAM model, but the depth of the recursion is the dominant factor. Since the maximum subproblem size decreases by a constant factor in expectation at each level, the recurrence relation for the depth $D(n)$ is:
\[ D(n) \le D(\beta n) + 1. \]
This recurrence solves to $D(n) = O(\log n)$.

Since the pairwise independence guarantees the variance reduction necessary to apply Chebyshev's inequality, we avoid the need for higher-moment bounds (like Chernoff bounds) while still maintaining the geometric reduction in subproblem size. Thus, the expected recursion depth is logarithmic. Assuming the parallel overhead per stage is bounded appropriately (standard for BoxSort), the total expected number of parallel steps is $O(\log n)$.
\end{proof}

\subsubsection{Claude}
\begin{proof}
We analyze the BoxSort algorithm under the condition that Stage 1 uses pairwise independent random variables for sampling within each box, while maintaining independence between different boxes.

Let $n$ denote the number of elements to be sorted. The BoxSort algorithm proceeds in stages, where each stage involves partitioning elements into boxes based on random samples, followed by recursive sorting within boxes.

\textbf{Stage 1 Analysis:} Consider a single stage where we have $m$ elements to sort. We select $k = \Theta(\sqrt{m})$ random samples from the input to determine $k-1$ splitters that partition the elements into $k$ boxes. Under the assumption of pairwise independence, for any element $e$ and any box $B_i$, let $X_e^{(i)}$ be the indicator variable that element $e$ falls into box $B_i$.

For any fixed element $e$ and box $B_i$, we have $\mathbb{E}[X_e^{(i)}] = \Theta(1/k) = \Theta(1/\sqrt{m})$ by the choice of splitters. Let $Y_i = \sum_{e} X_e^{(i)}$ denote the number of elements in box $B_i$.

We have $\mathbb{E}[Y_i] = \sum_e \mathbb{E}[X_e^{(i)}] = m \cdot \Theta(1/k) = \Theta(\sqrt{m})$.

To bound the load imbalance, we use Chebyshev's inequality. For pairwise independent random variables, we have
\[
\mathbb{P}[|Y_i - \mathbb{E}[Y_i]| \geq t] \leq \frac{\text{Var}(Y_i)}{t^2}.
\]

Since the $X_e^{(i)}$ are pairwise independent indicators, we have
\[
\text{Var}(Y_i) = \sum_e \text{Var}(X_e^{(i)}) = \sum_e \mathbb{E}[(X_e^{(i)})^2] - (\mathbb{E}[X_e^{(i)}])^2 \leq \sum_e \mathbb{E}[X_e^{(i)}] = \mathbb{E}[Y_i] = \Theta(\sqrt{m}).
\]

Setting $t = C\sqrt{m \log m}$ for a sufficiently large constant $C$, we obtain
\[
\mathbb{P}[Y_i \geq \mathbb{E}[Y_i] + C\sqrt{m \log m}] \leq \frac{\Theta(\sqrt{m})}{m \log m} = \Theta\left(\frac{1}{m^{1/2} \log m}\right).
\]

By the union bound over all $k = \Theta(\sqrt{m})$ boxes, the probability that any box exceeds $\Theta(\sqrt{m}) + C\sqrt{m \log m} = O(\sqrt{m \log m})$ elements is
\[
k \cdot \Theta\left(\frac{1}{m^{1/2} \log m}\right) = \Theta(\sqrt{m}) \cdot \Theta\left(\frac{1}{m^{1/2} \log m}\right) = \Theta\left(\frac{1}{\log m}\right) = o(1).
\]

Therefore, with high probability (probability at least $1 - o(1)$), all boxes contain at most $O(\sqrt{m \log m})$ elements.

\textbf{Recursive Analysis:} Since the choices made by different boxes are independent, the recursion tree has the property that at each level, each node (representing a subproblem) satisfies the size bound with high probability independently.

Let $T(n)$ denote the expected depth of the recursion tree. At each stage, a problem of size $m$ is partitioned into subproblems of size at most $O(\sqrt{m \log m})$ with high probability. This gives the recurrence
\[
T(m) \leq 1 + T(O(\sqrt{m \log m})).
\]

To solve this recurrence, let $m_0 = n$ and $m_{i+1} = C\sqrt{m_i \log m_i}$ for some constant $C$. We have
\[
\log m_{i+1} = \log C + \frac{1}{2}\log m_i + \frac{1}{2}\log \log m_i \leq \frac{1}{2}\log m_i + O(\log \log m_i).
\]

By induction, we can show that $\log m_i = O(\log n / 2^i)$ for all $i$. The recursion terminates when $m_i = O(1)$, which occurs when $\log n / 2^i = O(1)$, i.e., when $i = O(\log \log n)$.

However, a more careful analysis using the fact that the recursion depth is measured in terms of the maximum depth over all paths yields $T(n) = O(\log n)$. This follows because at each level, the size reduces geometrically on average, and the total number of levels required is $O(\log n)$.

Since BoxSort performs each stage in parallel (partitioning and recursive calls within a stage are done in parallel), the expected number of parallel steps is bounded by the expected depth of the recursion tree, which is $O(\log n)$.
\end{proof}

\subsection{Problem 4}

\subsubsection{Problem Statement}
\begin{problem}[12.4]
Using the ideas of Section 12.2, devise a CREW PRAM algorithm that selects the $k$th largest of $n$ input numbers in $O(\log n)$ steps using $n/\log n$ processors. Assume that the $n$ input numbers are initially located in global memory locations $1$ through $n$.
\end{problem}

\subsubsection{GPT}
\begin{proof}
We are given an input array of $n$ numbers, and we are tasked with selecting the $k$th largest number using a CREW PRAM algorithm in $O(\log n)$ steps, utilizing $n/\log n$ processors. The numbers are initially located in global memory locations $1$ through $n$.

We proceed by applying a parallel divide-and-conquer approach to find the $k$th largest number.

1. **Divide Step:**
   We begin by partitioning the $n$ input numbers into groups of size $2$. In parallel, we compare each pair of numbers and select the larger one. This reduces the problem size by half, as each pair of numbers is replaced by its larger number. The number of processors required for this step is $n/2$.

2. **Recursive Comparison:**
   After the first step, we are left with $n/2$ numbers. We continue the process of partitioning the numbers into pairs and selecting the larger number from each pair. At each subsequent step, the number of processors used is halved. In the $i$-th step, we have $n/2^i$ processors and $n/2^i$ numbers. This process continues until we are left with a single number, the largest number in the array, which requires $\log n$ steps.

3. **Selecting the $k$th Largest Number:**
   The $k$th largest number is found by recursively eliminating smaller numbers, ensuring that the number of candidates being considered is reduced in each step. At the final step, we have the $k$th largest number among the remaining candidates. This is achieved by maintaining the $k$ largest candidates during each round of elimination and selecting the appropriate number.

4. **Processor Complexity:**
   At each step, we require $n/2^i$ processors, and since there are $\log n$ steps, the total number of processors used is the sum of processors at each step:
   \[
   \sum_{i=1}^{\log n} \frac{n}{2^i} = \frac{n}{\log n}
   \]
   Hence, the algorithm uses $n/\log n$ processors.

5. **Time Complexity:**
   The time complexity is determined by the number of steps required for the algorithm to select the $k$th largest number. Since each step reduces the problem size by a factor of 2 and there are $\log n$ steps, the time complexity is $O(\log n)$.

Thus, the CREW PRAM algorithm for selecting the $k$th largest number operates in $O(\log n)$ time using $n/\log n$ processors, as required.

\end{proof}

\subsubsection{Grok}

\subsubsection{Gemini}

\subsubsection{Claude}
\begin{proof}
We present a CREW PRAM algorithm that selects the $k$th largest element from $n$ input numbers in $O(\log n)$ time using $n/\log n$ processors.

\textbf{Algorithm Description:}

Let the input numbers be stored in global memory locations $A[1], A[2], \ldots, A[n]$. We divide the algorithm into the following phases.

\textbf{Phase 1: Partitioning into blocks.}
Partition the $n$ elements into $n/\log n$ blocks, each containing $\log n$ consecutive elements. Specifically, block $i$ contains elements $A[(i-1)\log n + 1], \ldots, A[i \log n]$ for $i = 1, 2, \ldots, n/\log n$.

\textbf{Phase 2: Sequential sorting within blocks.}
Assign processor $P_i$ to block $i$ for $i = 1, 2, \ldots, n/\log n$. Each processor $P_i$ sequentially sorts its block of $\log n$ elements using an optimal comparison-based sorting algorithm. Since each block has $\log n$ elements, this requires $O(\log n \cdot \log \log n) = O(\log n)$ comparisons per processor. All processors work in parallel, so this phase takes $O(\log n)$ time.

\textbf{Phase 3: Finding medians of blocks.}
After sorting, each processor $P_i$ identifies the median element $m_i$ of its sorted block. This takes $O(1)$ time since the block is already sorted and the median is simply the element at position $\lceil \log n / 2 \rceil$ within the block.

\textbf{Phase 4: Recursive selection on medians.}
We now have $n/\log n$ medians $m_1, m_2, \ldots, m_{n/\log n}$. We recursively apply the selection algorithm to find the median $M$ of these $n/\log n$ medians. The recursion terminates when the number of elements is sufficiently small (say, constant size), at which point we can find the median in $O(1)$ time with a single processor.

The depth of recursion is bounded by $d$ where $n/(\log n)^d = O(1)$, which gives $d = O(\log n / \log \log n)$.

\textbf{Phase 5: Partitioning around the pivot.}
Using the median-of-medians $M$ as a pivot, we partition the original $n$ elements into three sets: $L = \{x \in A : x < M\}$, $E = \{x \in A : x = M\}$, and $G = \{x \in A : x > M\}$.

Each processor $P_i$ scans its block of $\log n$ elements and classifies each element into $L$, $E$, or $G$. This takes $O(\log n)$ time per processor. Using parallel prefix operations, we can compute the sizes $|L|$, $|E|$, and $|G|$ in $O(\log(n/\log n)) = O(\log n)$ time with $n/\log n$ processors.

\textbf{Phase 6: Determining the answer recursively.}
We determine which set contains the $k$th largest element:

If $|G| \geq k$, then the $k$th largest element is in $G$, and we recursively search for the $k$th largest element in $G$.

If $|G| < k \leq |G| + |E|$, then the $k$th largest element equals $M$, and we return $M$.

If $|G| + |E| < k$, then the $k$th largest element is in $L$, and we recursively search for the $(k - |G| - |E|)$th largest element in $L$.

\textbf{Analysis of Correctness:}

By the properties of median-of-medians selection, the pivot $M$ guarantees that at least $\frac{1}{4} \cdot n$ elements are less than $M$ and at least $\frac{1}{4} \cdot n$ elements are greater than $M$ (in expectation and with appropriate constants). This ensures that in each recursive call on the main array, the problem size reduces by at least a constant factor.

\textbf{Analysis of Time Complexity:}

Let $T(n)$ denote the time complexity for selecting from $n$ elements using $p = n/\log n$ processors.

The recurrence relation is:
\[
T(n) = O(\log n) + T(n/\log n) + T(cn)
\]
where the first term $O(\log n)$ accounts for sorting blocks and partitioning, $T(n/\log n)$ accounts for the recursive selection on medians, and $T(cn)$ (with $c < 1$) accounts for the recursive call on the reduced problem.

For the recursion on medians, we have $T(n/\log n) = O(\log(n/\log n)) + T(n/(\log n)^2) + T(cn/\log n)$. Expanding this recursion over $O(\log n / \log \log n)$ levels yields a total contribution of $O(\log n)$.

For the recursion on the reduced problem size, since the size decreases by a constant factor each time, the geometric series sums to $O(\log n)$.

Therefore, $T(n) = O(\log n)$.

\textbf{Analysis of Processor Complexity:}

At each level of the main recursion, we use at most $n/\log n$ processors (where $n$ is the current problem size). Since processors can be reused across recursive calls that do not overlap in time, the total number of processors used is $n/\log n$.

Thus, the algorithm correctly selects the $k$th largest element in $O(\log n)$ time using $n/\log n$ processors on a CREW PRAM.
\end{proof}

\subsection{Problem 5}

\subsubsection{Problem Statement}
\begin{theorem}[ZNC Algorithm for Random Permutation]
Let $S$ be a set containing $n$ elements. To generate a random uniformly distributed permutation of $S$, assign independent random weights $w_i$ to each element $s_i \in S$, where $w_i$ are drawn from a continuous probability distribution (e.g., uniform distribution on $[0,1]$). With probability 1, all weights will be distinct. Sort the elements of $S$ according to their assigned weights in increasing (or decreasing) order to obtain a uniformly random permutation of $S$.
\end{theorem}

\subsubsection{GPT}
\begin{proof}
Let $S=\{s_1,\dots,s_n\}$ and let $w_1,\dots,w_n$ be i.i.d.\ real-valued random variables with a continuous distribution function. For $\sigma\in S_n$, define the event
\[
E_\sigma := \{\, w_{\sigma(1)} < w_{\sigma(2)} < \cdots < w_{\sigma(n)} \,\}.
\]
Because the common distribution is continuous, we have $\mathbb{P}(w_i=w_j)=0$ for all $i\neq j$, hence with probability $1$ all $w_i$ are pairwise distinct. Consequently, the events $\{E_\sigma:\sigma\in S_n\}$ are pairwise disjoint and
\[
\bigcup_{\sigma\in S_n} E_\sigma
\]
has probability $1$. It follows that
\[
\sum_{\sigma\in S_n} \mathbb{P}(E_\sigma)=1.
\]

We claim that $\mathbb{P}(E_\sigma)$ is the same for all $\sigma\in S_n$. To see this, use the exchangeability of $(w_1,\dots,w_n)$, which follows from the i.i.d.\ assumption: for any fixed $\tau\in S_n$,
\[
(w_1,\dots,w_n)\ \stackrel{d}{=}\ (w_{\tau(1)},\dots,w_{\tau(n)}).
\]
Therefore, for any $\sigma\in S_n$,
\[
\mathbb{P}(E_\sigma)
=\mathbb{P}\bigl(w_{\sigma(1)}<\cdots<w_{\sigma(n)}\bigr)
=\mathbb{P}\bigl(w_{\tau\circ\sigma(1)}<\cdots<w_{\tau\circ\sigma(n)}\bigr)
=\mathbb{P}(E_{\tau\circ\sigma}),
\]
showing that all $\mathbb{P}(E_\sigma)$ are equal. Since there are $n!$ permutations, the preceding sum forces
\[
\mathbb{P}(E_\sigma)=\frac{1}{n!}\qquad\text{for all }\sigma\in S_n.
\]

Let $\pi$ be the (a.s.\ unique) permutation such that $w_{\pi(1)}<\cdots<w_{\pi(n)}$. Then $\{\pi=\sigma\}=E_\sigma$ almost surely, hence
\[
\mathbb{P}(\pi=\sigma)=\mathbb{P}(E_\sigma)=\frac{1}{n!}\qquad\text{for all }\sigma\in S_n,
\]
which proves that $\pi$ is uniformly distributed over $S_n$. This completes the proof.
\end{proof}

\subsubsection{Grok}
\begin{proof}
Without loss of generality, label the elements of $S$ as $\{1, 2, \dots, n\}$. Let $W_1, \dots, W_n$ be independent random variables, each distributed according to a continuous probability distribution with cumulative distribution function $F$ and (Lebesgue) density $f$. The joint distribution of $(W_1, \dots, W_n)$ is absolutely continuous with respect to Lebesgue measure on $\mathbb{R}^n$, with density $\prod_{i=1}^n f(w_i)$. 

The algorithm produces a permutation $\pi \in S_n$ such that
\[
W_{\pi(1)} < W_{\pi(2)} < \dots < W_{\pi(n)}
\]
almost surely. Indeed, for any $i \neq j$, $P(W_i = W_j) = 0$ by the continuity of $F$, and by countable subadditivity, $P(\exists i < j,\, W_i = W_j) = 0$. Thus, ties occur with probability zero, and $\pi$ is well-defined almost surely.

To establish that $\pi$ is uniformly distributed on the symmetric group $S_n$, it suffices to show that for every fixed $\sigma \in S_n$,
\[
P(\pi = \sigma) = \frac{1}{n!}.
\]
The event $\{\pi = \sigma\}$ is equivalent to
\[
\{W_{\sigma(1)} < W_{\sigma(2)} < \dots < W_{\sigma(n)}\}.
\]
For each $\tau \in S_n$, define the region
\[
A_\tau = \{(w_1, \dots, w_n) \in \mathbb{R}^n : w_{\tau(1)} < w_{\tau(2)} < \dots < w_{\tau(n)}\}.
\]
Then,
\[
P(\pi = \sigma) = \int_{A_\sigma} \prod_{i=1}^n f(w_i) \, dw_1 \cdots dw_n.
\]
The sets $\{A_\tau : \tau \in S_n\}$ form a partition of $\mathbb{R}^n$ up to a set of Lebesgue measure zero (corresponding to ties). Consequently,
\[
\sum_{\tau \in S_n} P(\pi = \tau) = 1
\]
almost surely.

The joint density $\prod_{i=1}^n f(w_i)$ is invariant under permutation of its arguments. For any $\sigma \in S_n$, let $\phi_\sigma: \mathbb{R}^n \to \mathbb{R}^n$ be the linear map given by the permutation matrix corresponding to $\sigma^{-1}$, so that $\phi_\sigma(w)_{\sigma(k)} = w_k$ for $k = 1, \dots, n$. Then $A_\sigma = \phi_\sigma(A_\mathrm{id})$, where $\mathrm{id} \in S_n$ is the identity permutation and $A_\mathrm{id} = \{(w_1, \dots, w_n) \in \mathbb{R}^n : w_1 < w_2 < \dots < w_n\}$. The map $\phi_\sigma$ has determinant $\pm 1$, so $|\det(D\phi_\sigma)| = 1$. By the change-of-variables formula,
\[
\int_{A_\sigma} \prod_{i=1}^n f(w_i) \, dw_1 \cdots dw_n = \int_{A_\mathrm{id}} \prod_{i=1}^n f(\phi_\sigma(u)_i) \, | \det(D\phi_\sigma) | \, du_1 \cdots du_n = \int_{A_\mathrm{id}} \prod_{i=1}^n f(u_i) \, du_1 \cdots du_n,
\]
since $\prod_{i=1}^n f(\phi_\sigma(u)_i) = \prod_{k=1}^n f(u_k)$. Thus, $P(\pi = \sigma) = P(W_1 < W_2 < \dots < W_n)$ for every $\sigma \in S_n$.

By the partition and the equality of all such probabilities,
\[
P(\pi = \sigma) = \frac{1}{n!}
\]
for all $\sigma \in S_n$. Therefore, $\pi$ is uniformly distributed on $S_n$.
\end{proof}

\subsubsection{Gemini}
\begin{proof}
Let $S = \{s_1, s_2, \ldots, s_n\}$ be the set of elements and let $W_1, W_2, \ldots, W_n$ be independent random variables, each distributed uniformly on the interval $(0,1)$. We define the random permutation $\sigma$ such that $W_{\sigma(1)} < W_{\sigma(2)} < \cdots < W_{\sigma(n)}$.

First, we address the well-definedness of $\sigma$. Since the variables $W_i$ are continuous random variables, the probability that any two distinct variables take the same value is zero. Specifically, for any $i \neq j$, $\mathbb{P}(W_i = W_j) = 0$. By the subadditivity of probability measures, the probability of a tie occurring anywhere is bounded by the sum of probabilities of all pairwise ties:
\[
\mathbb{P}\left(\bigcup_{1 \le i < j \le n} \{W_i = W_j\}\right) \le \sum_{1 \le i < j \le n} \mathbb{P}(W_i = W_j) = 0.
\]
Thus, with probability 1, all weights are distinct, and the ordering $\sigma$ is unique.

Let $\pi \in S_n$ be an arbitrary fixed permutation of $\{1, \ldots, n\}$. We wish to calculate the probability $\mathbb{P}(\sigma = \pi)$, which corresponds to the event $E_\pi = \{W_{\pi(1)} < W_{\pi(2)} < \cdots < W_{\pi(n)}\}$.

Since $W_1, \ldots, W_n$ are independent and uniformly distributed on $(0,1)$, their joint probability density function $f(w_1, \ldots, w_n)$ is given by:
\[
f(w_1, \ldots, w_n) = \begin{cases} 1 & \text{if } (w_1, \ldots, w_n) \in (0,1)^n, \\ 0 & \text{otherwise.} \end{cases}
\]
The probability of the event $E_\pi$ is given by the integral of the joint PDF over the region defined by the inequalities $0 < w_{\pi(1)} < w_{\pi(2)} < \cdots < w_{\pi(n)} < 1$:
\[
\mathbb{P}(\sigma = \pi) = \int_{E_\pi} f(w_1, \ldots, w_n) \, dw_1 \cdots dw_n = \int_{0}^{1} \int_{0}^{w_{\pi(n)}} \cdots \int_{0}^{w_{\pi(2)}} 1 \, dw_{\pi(1)} \cdots dw_{\pi(n-1)} \, dw_{\pi(n)}.
\]
We evaluate this integral by iterative integration. Let $y_k = w_{\pi(k)}$. The integral becomes:
\[
I = \int_{0}^{1} \left( \int_{0}^{y_n} \left( \cdots \left( \int_{0}^{y_2} dy_1 \right) \cdots \right) dy_{n-1} \right) dy_n.
\]
The innermost integral is $\int_{0}^{y_2} dy_1 = y_2$. The next integral is $\int_{0}^{y_3} y_2 \, dy_2 = \frac{y_3^2}{2}$. By induction, the integral with respect to $y_k$ yields $\frac{y_{k+1}^k}{k!}$. Continuing this process to the outermost integral with respect to $y_n$, we obtain:
\[
\mathbb{P}(\sigma = \pi) = \int_{0}^{1} \frac{y_n^{n-1}}{(n-1)!} \, dy_n = \left[ \frac{y_n^n}{n(n-1)!} \right]_{0}^{1} = \frac{1}{n!}.
\]
Alternatively, we may invoke a symmetry argument. The random variables $W_1, \ldots, W_n$ are independent and identically distributed (i.i.d.). Therefore, the joint distribution is exchangeable, meaning it is invariant under any permutation of the indices. The unit hypercube $(0,1)^n$ is the union of $n!$ disjoint regions corresponding to the $n!$ possible strict orderings of the coordinates (ignoring the set of measure zero where ties occur). By symmetry, the integral of the joint PDF over each of these regions must be equal. Since the sum of these probabilities must equal 1, we have:
\[
\sum_{\pi \in S_n} \mathbb{P}(\sigma = \pi) = 1 \implies n! \cdot \mathbb{P}(\sigma = \pi) = 1 \implies \mathbb{P}(\sigma = \pi) = \frac{1}{n!}.
\]
Thus, $\sigma$ is uniformly distributed over $S_n$.
\end{proof}

\subsubsection{Claude}
\begin{proof}
We prove that the ZNC (Zero-N-Choose) algorithm produces a uniformly random permutation of $S = \{s_1, s_2, \ldots, s_n\}$.

Let $W$ be a continuous probability distribution on $\mathbb{R}$, and let $w_1, w_2, \ldots, w_n$ be independent random variables drawn from $W$. Since $W$ is continuous, for any $i \neq j$, we have $\mathbb{P}(w_i = w_j) = 0$. By the union bound, the probability that any two weights are equal is at most $\sum_{i < j} \mathbb{P}(w_i = w_j) = 0$. Therefore, with probability 1, all weights are distinct.

Given that all weights are distinct, the sorting operation produces a unique ordering of the elements. Let $\sigma: \{1, 2, \ldots, n\} \to \{1, 2, \ldots, n\}$ be the permutation such that $w_{\sigma(1)} < w_{\sigma(2)} < \cdots < w_{\sigma(n)}$. The output of the algorithm is the sequence $(s_{\sigma(1)}, s_{\sigma(2)}, \ldots, s_{\sigma(n)})$.

To show that this produces a uniform distribution over all $n!$ permutations, we must prove that for any permutation $\pi \in S_n$, we have $\mathbb{P}(\sigma = \pi) = \frac{1}{n!}$.

Fix an arbitrary permutation $\pi \in S_n$. The event $\{\sigma = \pi\}$ occurs if and only if $w_{\pi(1)} < w_{\pi(2)} < \cdots < w_{\pi(n)}$.

Since the weights $w_1, w_2, \ldots, w_n$ are independent and identically distributed according to a continuous distribution, their joint density is symmetric under all permutations of the indices. Specifically, for any permutation $\tau \in S_n$ and any measurable set $A \subseteq \mathbb{R}^n$, we have 
\[
\mathbb{P}((w_1, \ldots, w_n) \in A) = \mathbb{P}((w_{\tau(1)}, \ldots, w_{\tau(n)}) \in A).
\]

Consider the $n!$ events $E_\pi = \{w_{\pi(1)} < w_{\pi(2)} < \cdots < w_{\pi(n)}\}$ for each $\pi \in S_n$. These events partition the probability space (excluding the measure-zero set where some weights are equal). By the symmetry of the joint distribution, all these events have equal probability.

Since $\sum_{\pi \in S_n} \mathbb{P}(E_\pi) = 1$, we have $\mathbb{P}(E_\pi) = \frac{1}{n!}$ for each $\pi \in S_n$.

Therefore, $\mathbb{P}(\sigma = \pi) = \frac{1}{n!}$ for every permutation $\pi$, which proves that the algorithm generates a uniformly random permutation of $S$.
\end{proof}

\subsection{Problem 6}

\subsubsection{Problem Statement}
\begin{problem}
A \emph{maximal matching} in a graph is a matching that is not properly contained in any other matching. Use the parallel algorithm for the MIS problem to devise an RNC algorithm for finding a maximal matching in a graph.
\end{problem}

\subsubsection{GPT}
\begin{proof}
Let $G = (V, E)$ be an undirected graph, where $V$ is the set of vertices and $E$ is the set of edges. We are tasked with finding a maximal matching in $G$. A matching $M \subseteq E$ is a set of pairwise disjoint edges, meaning no two edges in $M$ share a common vertex. A matching $M'$ is said to properly contain $M$ if $M \subset M'$ and $M' \subseteq E$.

We aim to design an RNC (Randomized NC) algorithm to find a maximal matching in $G$ using the parallel algorithm for the Maximum Independent Set (MIS) problem. We start by recalling the connection between maximal matching and MIS:

1. A maximal matching $M$ can be obtained by iteratively finding an MIS in the graph and selecting edges based on the MIS.

The steps of the algorithm proceed as follows:

- First, compute an MIS $I$ in the graph $G$. In an MIS, no two vertices are adjacent, and every vertex in $I$ is independent of all others in $I$.
- Select all edges $(u, v) \in E$ where $u \in I$ and $v \notin I$ (or vice versa). These edges form a matching, since no two vertices in $I$ are adjacent, and hence, each selected edge is disjoint from the others.
- Remove the vertices of $I$ and all edges incident to them from the graph. The remaining graph is then a subgraph of $G$, where the matching has been augmented, and the algorithm repeats this process on the subgraph until no vertices remain.

Let us analyze the steps in terms of parallelism:

1. Finding the MIS can be done in parallel using an RNC algorithm. The MIS problem can be solved in $O(\log^2 n)$ time with high probability, where $n$ is the number of vertices in $G$. This is because a randomized parallel algorithm for MIS can be used to generate a maximal independent set efficiently in parallel.

2. Once the MIS is found, selecting the edges between vertices in $I$ and those not in $I$ can be done in parallel by checking all edges in $E$ in constant time per edge. This step can be done in $O(1)$ time per edge, so it takes $O(m)$ time, where $m$ is the number of edges in the graph.

3. Removing the vertices in $I$ and their incident edges can also be done in parallel. This operation requires processing all the vertices and edges incident to the vertices in $I$, which can be accomplished in $O(\log n)$ time in a parallel model.

Therefore, the algorithm runs in parallel in $O(\log^2 n)$ time with high probability. This guarantees that the algorithm is an RNC algorithm for finding a maximal matching in $G$.

Thus, we have shown that a maximal matching can be found in $G$ in parallel using the MIS algorithm, and the algorithm runs in RNC time.

\end{proof}

\subsubsection{Grok}
\begin{proof}
Let $G = (V, E)$ be an undirected graph with $n = |V|$ vertices and $m = |E|$ edges. We devise a randomized NC (RNC) algorithm for computing a maximal matching in $G$ using the given RNC algorithm for computing a maximal independent set (MIS).

First, construct the \emph{line graph} $L(G) = (V_L, E_L)$, where $V_L = E$ (the vertices of $L(G)$ are the edges of $G$), and two vertices $e_1, e_2 \in V_L$ are adjacent in $L(G)$ if and only if $e_1$ and $e_2$ share a common vertex in $G$. The construction of $L(G)$ proceeds as follows: for each vertex $v \in V$ of degree $d_v$, form a clique on the $d_v$ edges incident to $v$. This requires, for each $v$, generating all pairs of incident edges, which can be done in $O(\log^2 m)$ parallel time using $O(m^2)$ processors via standard parallel sorting and prefix computation on the adjacency lists (assuming $G$ is given by adjacency lists). Since $\sum_v d_v = 2m$, the total work is $O(m^2)$, and thus $L(G)$ can be constructed in NC.

It is well-known that matchings in $G$ correspond precisely to independent sets in $L(G)$: a subset $M \subseteq E$ is a matching in $G$ if and only if the corresponding vertex set in $L(G)$ induces no edges. Moreover, $M$ is a \emph{maximal} matching in $G$ if and only if the corresponding vertex set $S \subseteq V_L$ is a \emph{maximal} independent set in $L(G)$. To see the forward direction, suppose $M$ is a maximal matching in $G$; then $S$ is independent in $L(G)$ (as $M$ is a matching), and maximal because if there exists $e \notin S$ such that $S \cup \{e\}$ is independent in $L(G)$, then $e$ shares no vertices with edges in $M$, so $M \cup \{e\}$ is a larger matching in $G$, contradicting maximality of $M$. The converse follows dually.

Now, apply the given RNC algorithm for MIS to $L(G)$, which has $m$ vertices and $O(m \Delta)$ edges, where $\Delta = \max_v d_v$ is the maximum degree of $G$ (hence $L(G)$ has maximum degree $O(\Delta)$). The MIS algorithm computes a maximal independent set $S \subseteq V_L$ in $O(\mathrm{polylog}(m))$ parallel time with high probability (success probability $1 - 1/n^c$ for arbitrary $c > 0$, adjustable by repetition). The output matching is $M = \{ e \in E \mid e \in S \}$, which is maximal in $G$ by the correspondence above.

Since the line graph construction is in NC and the MIS computation is in RNC, the overall algorithm is in RNC.
\end{proof}

\subsubsection{Gemini}
\begin{proof}
Let $G = (V, E)$ be an undirected graph with vertex set $V$ and edge set $E$. We aim to prove that finding a maximal matching in $G$ is in the complexity class RNC by reducing the problem to finding a Maximal Independent Set (MIS) on a derived graph. It is a known result that there exists an RNC algorithm for the MIS problem, such as Luby's algorithm, which runs in $O(\log^2 n)$ expected time using a polynomial number of processors.

We define the \emph{line graph} of $G$, denoted as $L(G) = (V_L, E_L)$. The vertex set of the line graph corresponds to the edge set of the original graph, defined as $V_L = E$. The edge set $E_L$ is constructed such that two vertices in $L(G)$ are adjacent if and only if their corresponding edges in $G$ are incident to a common vertex. Formally, for distinct $e_i, e_j \in E$, $\{e_i, e_j\} \in E_L$ if and only if $e_i \cap e_j \neq \emptyset$.

We establish a bijective correspondence between matchings in $G$ and independent sets in $L(G)$. A subset $M \subseteq E$ is a matching in $G$ if no two edges in $M$ share a common vertex. By the construction of $L(G)$, this condition is equivalent to the statement that no two vertices in the subset $M \subseteq V_L$ are connected by an edge in $E_L$. Therefore, $M$ is a matching in $G$ if and only if $M$ is an independent set in $L(G)$.

We further show that the maximality property is preserved under this transformation. A matching $M$ is maximal in $G$ if for every edge $e \in E \setminus M$, the set $M \cup \{e\}$ is not a matching. This implies that every edge in $E \setminus M$ shares an endpoint with at least one edge in $M$. In the context of the line graph, this condition translates to: for every vertex $v \in V_L \setminus M$, the vertex $v$ is adjacent to at least one vertex in $M$. This is precisely the definition of a Maximal Independent Set. Thus, $M$ is a maximal matching in $G$ if and only if $M$ is a maximal independent set in $L(G)$.

The algorithm proceeds by constructing $L(G)$ from $G$ and then executing the parallel MIS algorithm on $L(G)$. The number of vertices in $L(G)$ is $|E|$, which is bounded by $|V|^2$. The construction of $L(G)$ can be performed in parallel time $O(1)$ using $O(|E|^2)$ processors, or $O(\log |E|)$ time with $O(|E|)$ processors, which is within the class NC (and thus RNC). Since the MIS problem is in RNC, applying the MIS algorithm to $L(G)$ requires expected polylogarithmic time in terms of $|V_L|$ (and thus $|V|$) using a polynomial number of processors. The resulting set of vertices from $L(G)$ corresponds directly to the set of edges forming a maximal matching in $G$. Consequently, the entire procedure constitutes an RNC algorithm for finding a maximal matching.
\end{proof}

\subsubsection{Claude}
\begin{proof}
We devise an RNC algorithm for finding a maximal matching in a graph $G = (V, E)$ by reducing the problem to the maximal independent set (MIS) problem, for which parallel RNC algorithms exist.

Let $G = (V, E)$ be the input graph. We construct an auxiliary graph $H = (V_H, E_H)$ called the \emph{line graph} of $G$, defined as follows: for each edge $e \in E$, create a vertex $v_e \in V_H$, and for any two edges $e_1, e_2 \in E$ that share a common endpoint in $G$, add an edge $(v_{e_1}, v_{e_2})$ to $E_H$.

We claim that there is a bijection between maximal matchings in $G$ and maximal independent sets in $H$. To establish this correspondence, observe that a set $M \subseteq E$ is a matching in $G$ if and only if no two edges in $M$ share a common endpoint. This occurs precisely when the corresponding vertices in $H$ form an independent set, since edges in $H$ connect vertices corresponding to edges in $G$ that share an endpoint.

Furthermore, a matching $M$ in $G$ is maximal if and only if for every edge $e \in E \setminus M$, adding $e$ to $M$ would violate the matching property, which means $e$ shares an endpoint with some edge in $M$. In terms of $H$, this translates to: the independent set corresponding to $M$ is maximal if and only if every vertex not in the independent set is adjacent to at least one vertex in the independent set.

The RNC algorithm proceeds as follows. First, construct the line graph $H$ from $G$ in parallel. For each edge $e = (u, v) \in E$, we create vertex $v_e$ in time $O(1)$. To determine edges in $H$, for each pair of edges $e_1, e_2 \in E$, we check in parallel whether they share an endpoint; this requires $O(1)$ time with $O(|E|^2)$ processors. Since $|E| \leq |V|^2$, the construction of $H$ can be performed in $O(\log |V|)$ time using polynomially many processors.

Next, apply the known RNC algorithm for finding a maximal independent set in $H$. By the theory of parallel algorithms, there exists an RNC algorithm that computes a maximal independent set in expected polylogarithmic time using polynomially many processors (specifically, in $O(\log^4 |V_H|)$ expected time with $O(|V_H| + |E_H|)$ processors using Luby's algorithm or its variants).

Let $I \subseteq V_H$ be the maximal independent set computed in $H$. Define $M = \{e \in E : v_e \in I\}$. By the correspondence established above, $M$ is a maximal matching in $G$.

To analyze the complexity, the construction of $H$ takes $O(\log |V|)$ time, and the MIS computation takes expected $O(\log^4 |V_H|) = O(\log^4 |E|) = O(\log^4 |V|)$ time since $|E| \leq |V|^2$. The number of processors used is polynomial in $|V|$: the construction uses $O(|V|^4)$ processors, and the MIS algorithm uses $O(|E|^2) = O(|V|^4)$ processors. Thus, the overall algorithm runs in expected $O(\log^4 |V|)$ time with polynomially many processors, establishing that it is an RNC algorithm.

Therefore, we have successfully devised an RNC algorithm for finding a maximal matching by reducing the problem to the MIS problem via the line graph construction.
\end{proof}

\subsection{Problem 7}

\subsubsection{Problem Statement}
\begin{theorem}
Let $G = (V, E)$ be a graph with maximum degree $\Delta$. Then a sequential greedy algorithm will color the vertices of the graph using at most $\Delta + 1$ colors such that no two adjacent vertices are assigned the same color. Moreover, employing the parallel algorithm for MIS (maximal independent set), one can devise an $\mathit{RNC}$ algorithm for finding a $(\Delta + 1)$-coloring of a given graph.
\end{theorem}

\subsubsection{GPT}
\begin{proof}
Let $G = (V, E)$ be a graph with maximum degree $\Delta$. We aim to show that a sequential greedy algorithm will color the vertices of $G$ using at most $\Delta + 1$ colors such that no two adjacent vertices are assigned the same color. Additionally, we will demonstrate that employing the parallel algorithm for maximal independent set (MIS) gives an RNC algorithm for finding a $(\Delta + 1)$-coloring of a given graph.

 Part 1: Sequential Greedy Algorithm

We begin by considering the sequential greedy coloring algorithm. The algorithm proceeds as follows:

1. Order the vertices $v_1, v_2, \dots, v_n$ of $G$ arbitrarily.
2. For each vertex $v_i$ in the order, assign the smallest available color that is not used by any of its neighbors, i.e., if the neighbors of $v_i$ have colors from the set $\{c_1, c_2, \dots, c_k\}$, then assign to $v_i$ the color $c_{k+1}$.

Let $C(v)$ denote the color assigned to vertex $v$. Since the maximum degree of the graph is $\Delta$, the number of neighbors of any vertex $v_i$ is at most $\Delta$. Therefore, when processing vertex $v_i$, at most $\Delta$ colors are already used by its neighbors. Thus, the smallest available color to assign to $v_i$ must be one of the $ \Delta + 1$ colors. Hence, the algorithm assigns each vertex a color from a set of at most $\Delta + 1$ colors.

Since each vertex is assigned a color such that no two adjacent vertices share the same color, the coloring is valid, and the graph is properly colored using at most $\Delta + 1$ colors.

 Part 2: Parallel Algorithm Using Maximal Independent Set (MIS)

Now, we consider the parallel algorithm for finding a $(\Delta + 1)$-coloring based on the maximal independent set (MIS) procedure. We use the fact that the MIS algorithm can be executed in parallel in random NC (RNC) time, and we show that this leads to a $(\Delta + 1)$-coloring.

The MIS algorithm works as follows:

1. Start with the entire graph $G$. In each iteration, find a maximal independent set (MIS) $S$ in the remaining graph.
2. Assign a unique color to each vertex in $S$.
3. Remove all vertices in $S$ from the graph, and repeat the process with the remaining vertices until no vertices are left.

At each step, the MIS algorithm guarantees that no two vertices in the independent set $S$ are adjacent, as they form an independent set by definition. Since the maximum degree of the graph is $\Delta$, each vertex in the independent set $S$ has at most $\Delta$ neighbors, and hence we can color the independent set $S$ with $\Delta + 1$ colors.

The algorithm proceeds in a parallel fashion, finding an MIS in each iteration and coloring the vertices in that MIS with distinct colors. Since there are at most $\Delta + 1$ colors and at most one independent set is colored per iteration, the total number of colors used by the parallel algorithm is also at most $\Delta + 1$.

Thus, we have shown that employing the parallel MIS algorithm results in a $(\Delta + 1)$-coloring of the graph.

 Conclusion

By combining the results of the sequential greedy algorithm and the parallel MIS algorithm, we conclude that any graph with maximum degree $\Delta$ can be colored with at most $\Delta + 1$ colors, and there exists an RNC algorithm to find a $(\Delta + 1)$-coloring using the parallel MIS approach.

\end{proof}

\subsubsection{Grok}

\subsubsection{Gemini}
\begin{proof}
First, we establish the correctness of the sequential greedy algorithm. Let $G = (V, E)$ be a graph with maximum degree $\Delta$. Let the vertices of $G$ be ordered arbitrarily as $v_1, v_2, \dots, v_n$. We define the sequential greedy coloring procedure inductively. For each vertex $v_i$ in the sequence, we assign the smallest color from the set $\{1, 2, \dots, \Delta + 1\}$ that has not been assigned to any of its neighbors preceding it in the ordering. Formally, let $C(v_j)$ denote the color assigned to vertex $v_j$. For the vertex $v_i$, consider the set of colors used by its already-colored neighbors: $U_i = \{C(v_j) : v_j \in N(v_i) \cap \{v_1, \dots, v_{i-1}\}\}$. The cardinality of this set is bounded by the degree of $v_i$, so $|U_i| \le \deg(v_i) \le \Delta$. Since the available palette is $\{1, \dots, \Delta + 1\}$, the set difference $\{1, \dots, \Delta + 1\} \setminus U_i$ is non-empty. Thus, there exists at least one valid color for $v_i$. By induction, all vertices are assigned a valid color using at most $\Delta + 1$ colors.

Next, we demonstrate that employing a parallel algorithm for the Maximal Independent Set (MIS) problem allows for the construction of an $\mathsf{RNC}$ algorithm for $(\Delta + 1)$-coloring. The class $\mathsf{RNC}$ consists of problems solvable by randomized algorithms in polylogarithmic time using a polynomial number of processors. Assume there exists an algorithm $\mathcal{A}_{\text{MIS}}$ that computes an MIS of a graph with $n$ vertices in time $T(n) = \mathcal{O}(\log^k n)$ using $P(n)$ processors.

We construct the coloring algorithm as an iterative procedure. Let $U \subseteq V$ denote the set of uncolored vertices, initially $V$. For each $v \in U$, let $L(v)$ be the set of available colors, initially $\{1, \dots, \Delta + 1\}$. The algorithm proceeds in phases until $U$ is empty. In each phase, every vertex $v \in U$ selects a tentative color $c_v$ uniformly at random from $L(v)$. We then construct a conflict graph $H = (U, E_H)$, where an edge $(u, v) \in E_H$ exists if and only if $(u, v) \in E$ and $c_u = c_v$. We apply the parallel MIS algorithm $\mathcal{A}_{\text{MIS}}$ to $H$ to obtain a maximal independent set $I \subseteq U$.

For each vertex $v \in U$, if $v \in I$, we permanently assign the color $c_v$ to $v$. Additionally, if $v$ has no neighbors in $H$ (meaning no neighbor in $G$ chose the same tentative color), we also assign $c_v$ to $v$. Let $S$ be the set of vertices successfully colored in this phase. We update $U \leftarrow U \setminus S$. For all remaining uncolored vertices $w \in U$, we update their palettes $L(w)$ by removing the colors assigned to any neighbors in $S$.

To prove this algorithm is in $\mathsf{RNC}$, we analyze its convergence. It has been established in the literature (specifically in the analysis of Luby's algorithm) that in each phase, for every uncolored vertex $v$, the probability that $v$ is successfully colored or that the size of its available palette decreases significantly is bounded below by a positive constant. Consequently, the set of uncolored vertices $U$ becomes empty in $\mathcal{O}(\log n)$ phases with high probability. Since each phase requires one execution of $\mathcal{A}_{\text{MIS}}$ and constant-depth parallel operations for graph construction and palette updates, the total running time is $\mathcal{O}(\log n \cdot T(n)) = \mathcal{O}(\log^{k+1} n)$. The processor count remains polynomial in $n$. Thus, the constructed algorithm lies in the class $\mathsf{RNC}$.
\end{proof}

\subsubsection{Claude}
\begin{proof}
We prove both parts of the theorem separately.

\textbf{Part 1: Sequential Greedy Algorithm.}

We first establish that a sequential greedy coloring algorithm produces a valid coloring using at most $\Delta + 1$ colors. Consider an arbitrary ordering $v_1, v_2, \ldots, v_n$ of the vertices in $V$. The greedy algorithm processes vertices sequentially and assigns to each vertex $v_i$ the smallest color (from the set $\{1, 2, \ldots, \Delta + 1\}$) that is not used by any of its already-colored neighbors.

Let $v_i$ be an arbitrary vertex being colored at step $i$. At this point, vertices $v_1, v_2, \ldots, v_{i-1}$ have already been colored. Consider the set $N(v_i) \cap \{v_1, \ldots, v_{i-1}\}$ of neighbors of $v_i$ that have been colored. Since $v_i$ has degree at most $\Delta$, we have $|N(v_i)| \leq \Delta$, and consequently $|N(v_i) \cap \{v_1, \ldots, v_{i-1}\}| \leq \Delta$.

These at most $\Delta$ colored neighbors can use at most $\Delta$ distinct colors. Therefore, at least one color from the set $\{1, 2, \ldots, \Delta + 1\}$ remains available for vertex $v_i$. The greedy algorithm assigns the smallest such available color to $v_i$.

By construction, for any edge $(v_i, v_j) \in E$ with $i < j$, vertex $v_i$ is colored before $v_j$, and the algorithm ensures that $v_j$ receives a color different from $v_i$ (since $v_i \in N(v_j)$ and its color is excluded from consideration). Thus, no two adjacent vertices receive the same color, yielding a valid $(\Delta + 1)$-coloring.

\textbf{Part 2: RNC Algorithm via Parallel MIS.}

We now show how to construct an $\mathit{RNC}$ algorithm for $(\Delta + 1)$-coloring using a parallel MIS algorithm. The algorithm proceeds iteratively by finding maximal independent sets and assigning colors.

The algorithm operates as follows. Initially, all vertices are uncolored. In each iteration $k$ (for $k = 1, 2, \ldots$), we work with the subgraph $G_k$ induced by the vertices that remain uncolored. We compute a maximal independent set $I_k$ of $G_k$ using the parallel MIS algorithm, assign color $k$ to all vertices in $I_k$, and remove these vertices from consideration.

We establish the correctness and complexity of this algorithm. First, observe that in each iteration $k$, the set $I_k$ is an independent set in $G_k$, so all vertices in $I_k$ can safely be assigned the same color without creating conflicts. Moreover, since vertices colored in previous iterations are not present in $G_k$, no conflicts arise with previously colored vertices.

Next, we bound the number of iterations. Let $v$ be any uncolored vertex at the start of iteration $k$, and let $d_k(v)$ denote its degree in $G_k$. Since $I_k$ is a maximal independent set of $G_k$, either $v \in I_k$ (and thus gets colored in iteration $k$), or $v$ has at least one neighbor in $I_k$. In the latter case, at least one neighbor of $v$ is removed from $G_k$, so $d_{k+1}(v) < d_k(v)$.

Since each vertex $v$ initially has degree at most $\Delta$ in $G$, and its degree in the induced subgraph decreases by at least one in each iteration where it is not colored, every vertex must be colored within at most $\Delta + 1$ iterations. Therefore, the algorithm terminates after at most $\Delta + 1$ iterations, producing a valid $(\Delta + 1)$-coloring.

For the complexity analysis, observe that the parallel MIS problem is known to be in $\mathit{RNC}$, meaning there exists a randomized parallel algorithm running in polylogarithmic time with polynomially many processors. Since we execute at most $\Delta + 1$ sequential iterations, each consisting of an $\mathit{RNC}$ computation for MIS (which takes $O(\log^c n)$ time for some constant $c$ with high probability), and since $\Delta + 1$ is bounded by $n$, the total time complexity is $O(\Delta \cdot \log^c n) = O(n \log^c n)$, which remains polylogarithmic in $n$ when $\Delta$ is treated as a constant or is polylogarithmic itself. More precisely, when properly analyzed, the algorithm achieves $\mathit{RNC}$ complexity, establishing that $(\Delta + 1)$-coloring is in $\mathit{RNC}$.
\end{proof}

\subsection{Problem 8}

\subsubsection{Problem Statement}
\begin{theorem}[Vertex Partition Problem]
Let $G = (V, E)$ be a graph with edge weights $w: E \to \mathbb{R}^+$. The vertex partition problem asks to partition the vertices into sets $V_1$ and $V_2$ such that $V_1 \cup V_2 = V$ and $V_1 \cap V_2 = \emptyset$, where the net weight of edges crossing the cut $(V_1, V_2)$ is at least half of the total weight of all edges in the graph. That is,
\[
\sum_{\substack{e = (u,v) \in E \\ u \in V_1, v \in V_2}} w(e) \geq \frac{1}{2} \sum_{e \in E} w(e).
\]
There exists an $RNC$ algorithm for this problem that can be converted to an $NC$ algorithm using the method of pairwise independence.
\end{theorem}

\subsubsection{GPT}
\begin{proof}
We are tasked with proving that there exists an RNC algorithm for the vertex partition problem, which can be converted to an NC algorithm using the method of pairwise independence.

Let the graph \( G = (V, E) \) be a weighted graph with edge weights \( w: E \to \mathbb{R}_{\geq 0} \), and let the partition of the vertices \( V \) into two sets \( V_1 \) and \( V_2 \) be such that \( V_1 \cup V_2 = V \) and \( V_1 \cap V_2 = \emptyset \). The objective is to find a partition of the vertices that minimizes the weight of the cut, denoted as \( w(V_1, V_2) \), where

\[
w(V_1, V_2) = \sum_{\substack{e \in E \\ e = \{u, v\}, u \in V_1, v \in V_2}} w(e).
\]

The goal is to prove that there exists a partition such that the weight of the cut satisfies the inequality

\[
w(V_1, V_2) \geq \frac{1}{2} \sum_{e \in E} w(e).
\]

 Step 1: Understanding the Cut Value

The cut value \( w(V_1, V_2) \) represents the sum of the edge weights of all edges \( e = \{u, v\} \) where \( u \in V_1 \) and \( v \in V_2 \). We want to find a partition of the vertices \( V_1 \) and \( V_2 \) such that this value is as large as possible, subject to the constraint that \( V_1 \cup V_2 = V \) and \( V_1 \cap V_2 = \emptyset \).

 Step 2: The Randomized Approach

We begin by applying a randomized approach to the problem. We randomly assign each vertex \( v \in V \) to either set \( V_1 \) or \( V_2 \) independently with probability \( \frac{1}{2} \). This results in a random partition of the vertices.

 Step 3: Expected Value of the Cut

Let us now compute the expected weight of the cut \( w(V_1, V_2) \) in this random partition. For each edge \( e = \{u, v\} \), the probability that it contributes to the cut (i.e., that \( u \in V_1 \) and \( v \in V_2 \) or \( u \in V_2 \) and \( v \in V_1 \)) is

\[
\mathbb{P}(\{u, v\} \text{ crosses the cut}) = \frac{1}{2}.
\]

Therefore, the expected weight of the cut for a single edge \( e = \{u, v\} \) is

\[
\mathbb{E}[w(e)] = \frac{1}{2} w(e).
\]

Summing over all edges in \( E \), the expected total weight of the cut is

\[
\mathbb{E}[w(V_1, V_2)] = \frac{1}{2} \sum_{e \in E} w(e).
\]

 Step 4: Applying Markov's Inequality

We now apply Markov's inequality to bound the probability that the weight of the cut \( w(V_1, V_2) \) is less than half of the expected value. Markov's inequality states that for any non-negative random variable \( X \),

\[
\mathbb{P}(X \geq \alpha \mathbb{E}[X]) \geq 1 - \frac{1}{\alpha}.
\]

For the cut weight, we have

\[
\mathbb{P}(w(V_1, V_2) \geq \frac{1}{2} \sum_{e \in E} w(e)) \geq \frac{1}{2}.
\]

This implies that with probability at least \( \frac{1}{2} \), the weight of the cut \( w(V_1, V_2) \) satisfies

\[
w(V_1, V_2) \geq \frac{1}{2} \sum_{e \in E} w(e).
\]

 Step 5: Concluding the Proof

Thus, by a randomized procedure, we can find a partition of the vertices such that the weight of the cut is at least \( \frac{1}{2} \) of the total weight of all edges in the graph. This demonstrates that there exists a partition with the desired cut value, and that this can be computed using a randomized algorithm.

Furthermore, the algorithm can be implemented in a parallelizable manner using the method of pairwise independence. Specifically, the random assignment of vertices to sets \( V_1 \) and \( V_2 \) can be done in parallel, and the method of pairwise independence allows for the conversion of the randomized algorithm into an NC algorithm, achieving efficient parallel computation.

Thus, we have shown that there exists an RNC algorithm for the vertex partition problem, and it can be converted to an NC algorithm.

\end{proof}

\subsubsection{Grok}

\subsubsection{Gemini}
\begin{proof}
Let $G = (V, E)$ be a graph with $|V| = n$ vertices and edge weights $w: E \to \mathbb{R}^+$. Let $W = \sum_{e \in E} w(e)$ denote the total weight of the edges in $G$. We seek a partition $(V_1, V_2)$ of $V$ such that the weight of the cut, denoted by $w(V_1, V_2)$, satisfies $w(V_1, V_2) \geq W/2$.

First, we establish the existence of such a cut using the probabilistic method. Consider a randomized experiment where each vertex $v \in V$ is assigned to set $V_1$ or $V_2$ independently and uniformly with probability $1/2$. Let $X_v$ be an indicator random variable for each vertex $v \in V$, such that $X_v = 0$ if $v \in V_1$ and $X_v = 1$ if $v \in V_2$. Consequently, $P(X_v = 0) = P(X_v = 1) = 1/2$.

For any edge $e = (u, v) \in E$, let $Y_e$ be the indicator random variable that takes the value $1$ if the edge $e$ crosses the cut (i.e., $u$ and $v$ are in different sets) and $0$ otherwise. The edge $e$ crosses the cut if and only if $X_u \neq X_v$. Since $X_u$ and $X_v$ are independent, the probability that $e$ crosses the cut is given by:
\[
P(Y_e = 1) = P(X_u = 0 \land X_v = 1) + P(X_u = 1 \land X_v = 0) = \frac{1}{2} \cdot \frac{1}{2} + \frac{1}{2} \cdot \frac{1}{2} = \frac{1}{2}.
\]
By the linearity of expectation, the expected weight of the cut is:
\[
E[w(V_1, V_2)] = E\left[ \sum_{e \in E} w(e) Y_e \right] = \sum_{e \in E} w(e) E[Y_e] = \sum_{e \in E} w(e) \cdot \frac{1}{2} = \frac{W}{2}.
\]
Since the expected weight is $W/2$, there must exist at least one partition in the sample space whose cut weight is at least $W/2$. This probabilistic procedure constitutes a Randomized NC (RNC) algorithm, as the random assignment can be performed in parallel for all vertices in $O(1)$ time, and the summation of weights can be performed in $O(\log n)$ time using parallel processors.

To convert this into a deterministic NC algorithm, we observe that the calculation of the expected value $E[w(V_1, V_2)]$ relies solely on the pairwise independence of the random variables $\{X_v\}_{v \in V}$. Specifically, the expectation of the sum depends only on the marginal probabilities $P(Y_e=1)$, which in turn depend only on the joint distribution of pairs $(X_u, X_v)$ for edges $(u,v) \in E$. Therefore, full mutual independence of the vertex assignments is not required; pairwise independence suffices.

We construct a sample space $\Omega$ of pairwise independent binary random variables. Let $k = \lceil \log_2(n+1) \rceil$. The size of the sample space will be $|\Omega| = 2^k = O(n)$. We associate each vertex $v_i$ (for $i=1, \dots, n$) with a distinct non-zero vector $a_i \in \{0, 1\}^k$. The sample space $\Omega$ consists of all vectors $r \in \{0, 1\}^k$. For each sample point $r \in \Omega$, we define the assignment of vertex $v_i$ as $x_i(r) = \langle a_i, r \rangle \pmod 2$, where $\langle \cdot, \cdot \rangle$ denotes the standard inner product over the field $GF(2)$.

It is a standard result that the collection of random variables $\{x_i(r) \mid r \in \{0,1\}^k\}$ is pairwise independent. For any distinct $i, j$, the vectors $a_i$ and $a_j$ are linearly independent over $GF(2)$. Thus, for a uniformly chosen $r \in \{0,1\}^k$, the pair $(x_i(r), x_j(r))$ is uniformly distributed over $\{0,1\}^2$.

The deterministic algorithm proceeds as follows:
1. Generate the sample space $\Omega = \{0, 1\}^k$. Note that $|\Omega| \leq 2n$.
2. In parallel, for each $r \in \Omega$, construct the partition defined by $x_1(r), \dots, x_n(r)$ and compute the weight of the resulting cut.
3. Select the partition that maximizes the cut weight.

Since the average cut weight over the sample space $\Omega$ is exactly $W/2$ (due to pairwise independence preserving the linearity of expectation argument derived above), the maximum weight found must be at least $W/2$.

Regarding the complexity class:
The size of the sample space is $O(n)$. We can assign a processor to each $r \in \Omega$. For a fixed $r$, determining the partition takes $O(1)$ time per vertex, and summing the edge weights takes $O(\log |E|)$ time using a parallel reduction (adder tree). Finding the maximum over the $O(n)$ results takes $O(\log n)$ time. Thus, the total time complexity is polylogarithmic, specifically $O(\log n)$, and the number of processors is polynomial in $n$. This places the algorithm in the complexity class NC.
\end{proof}

\subsubsection{Claude}
\begin{proof}
We prove that there exists an RNC algorithm for the vertex partition problem that can be derandomized to an NC algorithm using pairwise independence.

Let $G = (V, E)$ be a graph with edge weights $w: E \to \mathbb{R}^+$, and let $W = \sum_{e \in E} w(e)$ denote the total weight of all edges in the graph.

\textbf{Step 1: Randomized Algorithm.}
Consider the following randomized algorithm: independently assign each vertex $v \in V$ to set $V_1$ with probability $\frac{1}{2}$ and to set $V_2$ with probability $\frac{1}{2}$. This clearly produces a partition $(V_1, V_2)$ with $V_1 \cup V_2 = V$ and $V_1 \cap V_2 = \emptyset$.

\textbf{Step 2: Expected Weight Analysis.}
For each edge $e = (u, v) \in E$, define the indicator random variable $X_e$ by
\[
X_e = \begin{cases}
1 & \text{if } u \in V_1 \text{ and } v \in V_2, \text{ or } u \in V_2 \text{ and } v \in V_1 \\
0 & \text{otherwise}
\end{cases}.
\]
The total weight of edges crossing the cut is $\sum_{e \in E} w(e) X_e$.

For any edge $e = (u, v)$, the probability that $e$ crosses the cut is
\[
\mathbb{P}[X_e = 1] = \mathbb{P}[u \in V_1 \text{ and } v \in V_2] + \mathbb{P}[u \in V_2 \text{ and } v \in V_1] = \frac{1}{2} \cdot \frac{1}{2} + \frac{1}{2} \cdot \frac{1}{2} = \frac{1}{2}.
\]

Therefore, the expected weight of edges crossing the cut is
\[
\mathbb{E}\left[\sum_{e \in E} w(e) X_e\right] = \sum_{e \in E} w(e) \mathbb{E}[X_e] = \sum_{e \in E} w(e) \cdot \frac{1}{2} = \frac{W}{2}.
\]

\textbf{Step 3: Existence of Good Partition.}
Since the expected weight is $\frac{W}{2}$, there must exist at least one assignment of vertices to $V_1$ and $V_2$ such that the weight of edges crossing the cut is at least $\frac{W}{2}$. This establishes the existence of a partition satisfying the desired property.

\textbf{Step 4: RNC Algorithm.}
The randomized algorithm described above can be implemented in RNC as follows: each vertex independently flips a random coin in parallel to determine its assignment. Since this requires only $O(1)$ parallel time with polynomially many processors, and succeeds with probability at least $\frac{1}{2}$ (as the expected value is $\frac{W}{2}$ and by repeating a constant number of times we can amplify the success probability), this gives an RNC algorithm.

\textbf{Step 5: Pairwise Independence Suffices.}
We now show that the algorithm can be derandomized using pairwise independent random variables. The key observation is that our analysis only requires computing the expectation $\mathbb{E}[\sum_{e \in E} w(e) X_e]$, which by linearity of expectation equals $\sum_{e \in E} w(e) \mathbb{E}[X_e]$. 

For each edge $e = (u, v)$, the probability $\mathbb{P}[X_e = 1]$ depends only on the joint distribution of the assignments of $u$ and $v$. If the vertex assignments are pairwise independent random variables, then for each edge $e = (u, v)$, we still have
\[
\mathbb{P}[X_e = 1] = \mathbb{P}[u \in V_1] \mathbb{P}[v \in V_2] + \mathbb{P}[u \in V_2] \mathbb{P}[v \in V_1] = \frac{1}{2} \cdot \frac{1}{2} + \frac{1}{2} \cdot \frac{1}{2} = \frac{1}{2}.
\]

Thus the expected weight of the cut remains $\frac{W}{2}$ even when using only pairwise independent random variables.

\textbf{Step 6: NC Algorithm via Derandomization.}
It is well-known that a pairwise independent sample space over $n$ binary random variables can be constructed using only $O(\log n)$ truly random bits. Specifically, we can enumerate all $O(n^2)$ possible outcomes in the pairwise independent sample space in parallel in NC. For each outcome, we compute the weight of the corresponding cut in parallel (which can be done in NC by summing edge weights). We then select the partition with maximum cut weight.

Since there are polynomially many outcomes to check, each checkable in NC, and the maximum can be computed in NC, the entire derandomized algorithm runs in NC.

Therefore, we have established an RNC algorithm that can be converted to an NC algorithm using the method of pairwise independence.
\end{proof}

\subsection{Problem 9}

\subsubsection{Problem Statement}
\begin{theorem}[Due to M. Luby \cite{Luby86}]
In the Parallel MIS algorithm, suppose that the random marking of the vertices is only pairwise independent. Show that the probability that a good vertex belongs to $S \cup I(S)$ is at least $1/24$.
\end{theorem}

\subsubsection{GPT}
\begin{proof}
Let $G = (V, E)$ be a graph where $V$ denotes the set of vertices and $E$ denotes the set of edges. Let $S \subseteq V$ be the independent set chosen by the Parallel MIS algorithm, and let $I(S)$ denote the set of vertices adjacent to the vertices in $S$.

We are tasked with showing that the probability that a good vertex $v$ belongs to $S \cup I(S)$ is at least $1/24$, under the assumption that the random marking of vertices is pairwise independent.

 Step 1: Definitions and Setup

We denote by $v$ a "good vertex", meaning that $v$ is neither in $S$ nor in $I(S)$. Let $\mathcal{M}(v)$ represent the marking of vertex $v$ under the random marking scheme. The marking of each vertex is assumed to be pairwise independent, meaning that the marking of any two distinct vertices $v, u \in V$ are independent.

Define the following events:
- $M(v) = 1$ if vertex $v$ is marked to belong to $S$.
- $M(u) = 1$ for $u \in I(S)$ if $u$ is adjacent to a vertex in $S$ and marked to belong to $I(S)$.

Thus, the event that $v$ belongs to $S \cup I(S)$ is equivalent to the event that either $v$ is marked for $S$ or $v$ is adjacent to a vertex in $S$ that is marked for $I(S)$. We need to compute the probability that a good vertex $v$ satisfies this condition.

 Step 2: Probability of Vertex $v$ Belonging to $S \cup I(S)$

By the pairwise independence assumption, the probability that a vertex $v$ is marked for $S$ and any adjacent vertex is marked for $I(S)$ can be computed as the sum of the individual probabilities.

- The probability that $v$ is marked for $S$, denoted $P(M(v) = 1)$, is given by $p$.
- The probability that $v$ is marked for $I(S)$ is the probability that any of its neighbors in $S$ are marked for $I(S)$, which is also related to $p$.

Therefore, the probability that $v$ belongs to $S \cup I(S)$ is the union of two independent events: $v$ being marked for $S$, or $v$ being adjacent to a vertex in $S$ that is marked for $I(S)$. Since the events are independent, the probability is:

\[
P(v \in S \cup I(S)) = P(M(v) = 1) + P(v \in I(S)) - P(M(v) = 1 \cap v \in I(S))
\]

Using the pairwise independence assumption, the intersection probability $P(M(v) = 1 \cap v \in I(S))$ is the product of the individual probabilities:

\[
P(M(v) = 1 \cap v \in I(S)) = P(M(v) = 1) \cdot P(v \in I(S)) = p \cdot p
\]

Thus, the probability that $v$ belongs to $S \cup I(S)$ is:

\[
P(v \in S \cup I(S)) = p + p^2 - p^2 = p
\]

 Step 3: Lower Bound on the Probability

From the setup of the Parallel MIS algorithm, we know that $p$ is chosen such that the probability that a vertex is selected for $S$ or $I(S)$ is at least $1/24$. This choice of $p$ ensures that the probability of a good vertex belonging to $S \cup I(S)$ is at least $1/24$, hence:

\[
P(v \in S \cup I(S)) \geq \frac{1}{24}
\]

This completes the proof.
\end{proof}

\subsubsection{Grok}

\subsubsection{Gemini}
\begin{proof}
Let $G=(V, E)$ be the graph. For each vertex $u \in V$, let $d(u)$ denote its degree and $N(u)$ denote its set of neighbors. In the Parallel MIS algorithm, each vertex $u$ marks itself independently with probability $p_u = \frac{1}{2d(u)}$. Let $M_u$ be the event that vertex $u$ marks itself. We are given that the events $\{M_u\}_{u \in V}$ are pairwise independent.

A vertex $u$ joins the independent set $S$ if it is marked and none of its neighbors with degree greater than or equal to $d(u)$ are marked. (Ties in degree are broken by vertex ID, but for simplicity, we assume strict inequality or consistent tie-breaking). Let $S_u$ be the event that $u \in S$. Formally,
\[ S_u = M_u \wedge \left( \bigwedge_{z \in N(u), d(z) \ge d(u)} \neg M_z \right). \]
A vertex $v$ is removed if $v \in S \cup N(S)$. This occurs if there exists a neighbor $w \in \{v\} \cup N(v)$ such that $w \in S$.

Let $v$ be a "good" vertex. By definition, a vertex $v$ is good if it has at least $d(v)/3$ neighbors $w$ such that $d(w) \le d(v)$. Let $L(v) = \{w \in N(v) \cup \{v\} \mid d(w) \le d(v)\}$. Since $v$ is good, $|L(v)| \ge d(v)/3$. Note that if any $w \in L(v)$ enters $S$, then $v \in S \cup N(S)$.

First, we lower bound the probability $P(S_w)$ for any $w \in L(v)$.
\[ P(S_w) = P\left( M_w \wedge \bigwedge_{z \in N(w), d(z) \ge d(w)} \neg M_z \right). \]
Using the property $P(A \cap B^c) = P(A) - P(A \cap B)$, we have:
\[ P(S_w) = P(M_w) - P\left( M_w \wedge \exists z \in N(w) \text{ s.t. } d(z) \ge d(w) \text{ and } M_z \right). \]
By the union bound:
\[ P(S_w) \ge P(M_w) - \sum_{z \in N(w), d(z) \ge d(w)} P(M_w \cap M_z). \]
Since the marking events are pairwise independent, $P(M_w \cap M_z) = P(M_w)P(M_z)$. Thus:
\[ P(S_w) \ge p_w - p_w \sum_{z \in N(w), d(z) \ge d(w)} p_z. \]
Substituting $p_z = \frac{1}{2d(z)}$:
\[ \sum_{z \in N(w), d(z) \ge d(w)} \frac{1}{2d(z)} \le \sum_{z \in N(w), d(z) \ge d(w)} \frac{1}{2d(w)} \le d(w) \cdot \frac{1}{2d(w)} = \frac{1}{2}. \]
Therefore,
\begin{equation}
P(S_w) \ge p_w \left( 1 - \frac{1}{2} \right) = \frac{p_w}{2} = \frac{1}{4d(w)}.
\end{equation}

Let $X = \sum_{w \in L(v)} \mathbb{I}(S_w)$ be the random variable counting the number of vertices in $L(v)$ that join $S$. We want to lower bound $P(X > 0)$.
The expectation of $X$ is:
\[ \mathbb{E}[X] = \sum_{w \in L(v)} P(S_w) \ge \sum_{w \in L(v)} \frac{1}{4d(w)}. \]
Since $w \in L(v)$, $d(w) \le d(v)$, so $\frac{1}{4d(w)} \ge \frac{1}{4d(v)}$. Using $|L(v)| \ge d(v)/3$:
\begin{equation}
\mathbb{E}[X] \ge \frac{d(v)}{3} \cdot \frac{1}{4d(v)} = \frac{1}{12}.
\end{equation}

Next, we bound the second moment $\mathbb{E}[X^2]$.
\[ \mathbb{E}[X^2] = \mathbb{E}\left[ \left( \sum_{w \in L(v)} \mathbb{I}(S_w) \right)^2 \right] = \sum_{w \in L(v)} P(S_w) + \sum_{w, z \in L(v), w \neq z} P(S_w \cap S_z). \]
If $w$ and $z$ are neighbors, they cannot both join $S$ (as $S$ is an independent set), so $P(S_w \cap S_z) = 0$. If they are not neighbors, we use the fact that $S_w \subseteq M_w$ and $S_z \subseteq M_z$:
\[ P(S_w \cap S_z) \le P(M_w \cap M_z). \]
By pairwise independence, $P(M_w \cap M_z) = P(M_w)P(M_z)$. Thus:
\[ \sum_{w \neq z} P(S_w \cap S_z) \le \sum_{w \neq z} P(M_w)P(M_z) \le \left( \sum_{w \in L(v)} P(M_w) \right)^2. \]
Let $\beta = \sum_{w \in L(v)} P(M_w)$. We have $\mathbb{E}[X^2] \le \mathbb{E}[X] + \beta^2$.
From the derivation of the lower bound of $\mathbb{E}[X]$, we know $\mathbb{E}[X] = \sum P(S_w) \ge \sum \frac{P(M_w)}{2} = \frac{\beta}{2}$. Therefore, $\beta \le 2\mathbb{E}[X]$.
Substituting this into the second moment bound:
\[ \mathbb{E}[X^2] \le \mathbb{E}[X] + (2\mathbb{E}[X])^2 = \mathbb{E}[X] + 4(\mathbb{E}[X])^2. \]

Using the Paley-Zygmund inequality (or the Second Moment Method), $P(X > 0) \ge \frac{(\mathbb{E}[X])^2}{\mathbb{E}[X^2]}$.
\[ P(X > 0) \ge \frac{(\mathbb{E}[X])^2}{\mathbb{E}[X] + 4(\mathbb{E}[X])^2} = \frac{\mathbb{E}[X]}{1 + 4\mathbb{E}[X]}. \]
Let $f(y) = \frac{y}{1+4y}$. The function $f(y)$ is monotonically increasing for $y > 0$. Since $\mathbb{E}[X] \ge \frac{1}{12}$, we have:
\[ P(X > 0) \ge \frac{1/12}{1 + 4(1/12)} = \frac{1/12}{1 + 1/3} = \frac{1/12}{4/3} = \frac{3}{48} = \frac{1}{16}. \]
Since $\frac{1}{16} \ge \frac{1}{24}$, the probability that a good vertex belongs to $S \cup I(S)$ is at least $1/24$.
\end{proof}

\subsubsection{Claude}
\begin{proof}
Let $v$ be a good vertex in the graph, and consider the Parallel MIS algorithm with pairwise independent random marking. A good vertex is one with at least one unmarked neighbor.

For a good vertex $v$ to belong to $S \cup I(S)$, one of the following must occur:
\begin{enumerate}
\item $v$ is marked (and thus enters $S$), or
\item $v$ is unmarked and all its neighbors are marked (so $v$ enters $I(S)$).
\end{enumerate}

Let $m$ denote the marking probability for each vertex, which we take to be $1/2$ without loss of generality. For a good vertex $v$ with degree $d$, let $N(v)$ denote its set of neighbors.

The probability that $v$ is marked is $1/2$.

The probability that $v$ is unmarked and all neighbors are marked is:
$$\mathbb{P}(v \text{ unmarked and all neighbors marked}) = \mathbb{P}(\text{$v$ unmarked}) \cdot \mathbb{P}(\text{all neighbors marked} \mid v \text{ unmarked})$$

Using pairwise independence, the events of marking neighbors are pairwise independent. However, we must be careful about independence with respect to $v$'s marking status.

Under pairwise independence, for distinct vertices $u$ and $w$:
$$\mathbb{P}(u \text{ marked and } w \text{ marked}) = \mathbb{P}(u \text{ marked}) \cdot \mathbb{P}(w \text{ marked})$$

For $v$ unmarked with all neighbors marked, the probability is at most:
$$\mathbb{P}(v \text{ unmarked}) \cdot \prod_{u \in N(v)} \mathbb{P}(u \text{ marked}) = \frac{1}{2} \cdot \left(\frac{1}{2}\right)^d = \frac{1}{2^{d+1}}$$

Thus:
$$\mathbb{P}(v \in S \cup I(S)) \geq \mathbb{P}(v \text{ marked}) = \frac{1}{2}$$

However, a more refined analysis using pairwise independence gives a tighter bound. Consider that $v$ belongs to $S \cup I(S)$ unless $v$ is unmarked and at least one neighbor is unmarked. Thus:
$$\mathbb{P}(v \notin S \cup I(S)) = \mathbb{P}(v \text{ unmarked and } \exists u \in N(v) \text{ with } u \text{ unmarked})$$

By inclusion-exclusion and pairwise independence:
$$\mathbb{P}(v \text{ unmarked and all neighbors unmarked}) \leq \frac{1}{2^{d+1}}$$

For a good vertex with at least one neighbor, the complementary event satisfies:
$$\mathbb{P}(v \in S \cup I(S)) = 1 - \mathbb{P}(v \text{ unmarked}) + \mathbb{P}(v \text{ unmarked and at least one neighbor marked})$$

Using pairwise independence and summing over contributions:
$$\mathbb{P}(v \in S \cup I(S)) \geq \frac{1}{2} + \sum_{u \in N(v)} \frac{1}{2} \cdot \frac{1}{2} \cdot \frac{1}{2} - \text{correction terms}$$

Through careful analysis with pairwise independence and accounting for the good vertex condition, the probability is minimized when considering worst-case degree bounds. The constraint from pairwise independence yields:
$$\mathbb{P}(v \in S \cup I(S)) \geq \frac{1}{24}$$

This bound is achieved through a detailed probabilistic argument showing that the contribution from $v$ being marked, combined with the favorable probability of $v$ being included via the independent set $I(S)$ when unmarked, yields at least $1/24$ under pairwise independence constraints.
\end{proof}

\subsection{Problem 10}

\subsubsection{Problem Statement}
\begin{theorem}[Due to M. Luby [282]]
Suppose that you are provided with a collection of $n$ pairwise independent random numbers uniformly distributed over the set $\{0, 1, \ldots, p-1\}$, where $p \geq 2n$. It is desired to construct a collection of $n$ pairwise independent Bernoulli random variables where the $i$th random variable should take on the value $1$ with probability $1/t_i$, for $1 \leq i \leq n/8$. 

This goal can be achieved approximately by constructing a collection of $n$ pairwise independent Bernoulli random variables such that the $i$th variable takes on the value $1$ with probability $1/T_i$, where for a constant $c > 1$, $T_i$ satisfies
\[
T_i \leq t_i \leq c T_i.
\]
\end{theorem}

\subsubsection{GPT}
\begin{proof}
Let the random variables \( X_1, X_2, \dots, X_n \) be the collection of $n$ pairwise independent random numbers uniformly distributed over the set \( \{ 0, 1, \dots, p-1 \} \), where \( p \geq 2n \). The goal is to construct a collection of $n$ pairwise independent Bernoulli random variables \( Y_1, Y_2, \dots, Y_n \), such that for each \( 1 \leq i \leq n/8 \), the $i$-th random variable \( Y_i \) takes the value \( 1 \) with probability \( \frac{1}{t_i} \), for some \( t_i \) where \( t_i \) is a specified probability.

We begin by constructing the Bernoulli random variables \( Y_i \) in such a way that each \( Y_i \) takes value 1 with probability \( \frac{1}{T_i} \), where \( T_i \) satisfies the condition
\[
T_i \leq t_i \leq cT_i
\]
for some constant \( c > 1 \).

Define each Bernoulli random variable \( Y_i \) as follows. Consider the random variables \( X_1, X_2, \dots, X_n \), and partition them into blocks of size \( O(\log n) \). For each block of random numbers, construct an associated Bernoulli random variable \( Y_i \) that depends on the value of the random numbers in the block. Specifically, we will define each \( Y_i \) to depend on the binary representation of the corresponding block of random numbers, using thresholds that ensure the required distribution.

The values \( T_i \) are chosen so that they satisfy the inequalities
\[
T_i \leq t_i \leq cT_i
\]
for a constant \( c > 1 \). The choice of \( T_i \) is derived from the need to ensure that the resulting distribution of each \( Y_i \) is close to the desired distribution. Since the random variables \( X_1, X_2, \dots, X_n \) are pairwise independent, the constructed Bernoulli random variables \( Y_1, Y_2, \dots, Y_n \) will also be pairwise independent.

By the properties of pairwise independence and the fact that the construction method guarantees the probability for each \( Y_i \) to take the value 1 is close to \( \frac{1}{T_i} \), we can conclude that the collection of Bernoulli random variables constructed in this manner satisfies the desired properties. Thus, the construction of the pairwise independent Bernoulli random variables \( Y_1, Y_2, \dots, Y_n \) is possible, with the probability of \( Y_i = 1 \) being within the required bounds:
\[
T_i \leq t_i \leq cT_i.
\]
This completes the proof.
\end{proof}

\subsubsection{Grok}

\subsubsection{Gemini}
\begin{proof}
Let $X_1, \ldots, X_n$ be the provided collection of pairwise independent random variables, where each $X_i$ is uniformly distributed over the set $\{0, 1, \ldots, p-1\}$. We construct the collection of Bernoulli random variables $Y_1, \ldots, Y_n$ as follows. For each index $i$, we define the integer threshold $k_i$ by
\[
k_i = \left\lceil \frac{p}{t_i} \right\rceil.
\]
We define the random variable $Y_i$ based on the value of $X_i$:
\[
Y_i = \begin{cases}
1 & \text{if } X_i < k_i, \\
0 & \text{otherwise.}
\end{cases}
\]
First, we demonstrate that the variables $Y_1, \ldots, Y_n$ are pairwise independent. Let $i$ and $j$ be distinct indices, and let $a, b \in \{0, 1\}$. Since $Y_i$ is a deterministic function of $X_i$ and $Y_j$ is a deterministic function of $X_j$, the event $\{Y_i = a\}$ corresponds to $X_i \in S_i$ for some subset $S_i \subseteq \{0, \ldots, p-1\}$, and similarly $\{Y_j = b\}$ corresponds to $X_j \in S_j$. Using the pairwise independence of $X_i$ and $X_j$, we have:
\[
\mathbb{P}(Y_i = a, Y_j = b) = \mathbb{P}(X_i \in S_i, X_j \in S_j) = \mathbb{P}(X_i \in S_i)\mathbb{P}(X_j \in S_j) = \mathbb{P}(Y_i = a)\mathbb{P}(Y_j = b).
\]
Thus, the constructed Bernoulli variables are pairwise independent.

Next, we analyze the probability $\mathbb{P}(Y_i = 1)$. Since $X_i$ is uniform on a set of size $p$, and the condition $X_i < k_i$ is satisfied by exactly $k_i$ values (specifically $0, \ldots, k_i-1$), we have:
\[
\mathbb{P}(Y_i = 1) = \frac{k_i}{p} = \frac{1}{p} \left\lceil \frac{p}{t_i} \right\rceil.
\]
Let $T_i$ be defined such that $\mathbb{P}(Y_i = 1) = 1/T_i$. Therefore, $1/T_i = k_i/p$. We now verify that $T_i$ satisfies the required inequalities $T_i \leq t_i \leq cT_i$ for a constant $c > 1$.

For the lower bound, we use the inequality $\lceil x \rceil \geq x$:
\[
\frac{1}{T_i} = \frac{1}{p} \left\lceil \frac{p}{t_i} \right\rceil \geq \frac{1}{p} \cdot \frac{p}{t_i} = \frac{1}{t_i}.
\]
Taking the reciprocal of both sides yields $T_i \leq t_i$.

For the upper bound, we use the inequality $\lceil x \rceil < x + 1$:
\[
\frac{1}{T_i} = \frac{1}{p} \left\lceil \frac{p}{t_i} \right\rceil < \frac{1}{p} \left( \frac{p}{t_i} + 1 \right) = \frac{1}{t_i} + \frac{1}{p}.
\]
We are given that $p \geq 2n$. Assuming the standard parameter range for this problem where $t_i \leq n$ (as $t_i$ typically relates to the size of the set or degrees within it), we have $t_i < p$. Consequently, $1/p < 1/t_i$. Substituting this into the inequality:
\[
\frac{1}{T_i} < \frac{1}{t_i} + \frac{1}{t_i} = \frac{2}{t_i}.
\]
This implies $t_i < 2T_i$. Thus, setting the constant $c = 2$, we obtain the inequality $t_i \leq cT_i$.

Combining these results, we have constructed pairwise independent Bernoulli random variables such that the parameter $T_i$ satisfies
\[
T_i \leq t_i \leq cT_i
\]
for the constant $c=2$.
\end{proof}

\subsubsection{Claude}
\begin{proof}
Let $X_1, X_2, \ldots, X_n$ be pairwise independent random variables uniformly distributed over $\{0, 1, \ldots, p-1\}$ where $p \geq 2n$.

For each $i \in \{1, 2, \ldots, n\}$, we wish to construct a Bernoulli random variable $Y_i$ that is approximately $\text{Bernoulli}(1/t_i)$ for $1 \leq i \leq n/8$.

We define the construction as follows. For each $i$, let $T_i = \lfloor p/t_i \rfloor$. Then define
\[
Y_i = \begin{cases}
1 & \text{if } X_i < T_i \\
0 & \text{otherwise}.
\end{cases}
\]

We first verify that the $Y_i$ are pairwise independent. For any $i \neq j$, the random variables $X_i$ and $X_j$ are independent by hypothesis. Since $Y_i$ and $Y_j$ are determined functions of $X_i$ and $X_j$ respectively, it follows that $Y_i$ and $Y_j$ are independent.

Next, we verify the probability condition. By construction,
\[
\mathbb{P}(Y_i = 1) = \mathbb{P}(X_i < T_i) = \frac{T_i}{p} = \frac{\lfloor p/t_i \rfloor}{p}.
\]

Now we establish the bounds on $T_i$. From the definition $T_i = \lfloor p/t_i \rfloor$, we have
\[
\frac{p}{t_i} - 1 < T_i \leq \frac{p}{t_i}.
\]

From the right inequality, we obtain
\[
t_i \leq \frac{p}{T_i}.
\]

Since $p \geq 2n$ and $1 \leq i \leq n/8$, we have $t_i \geq 1$, which implies $T_i \geq p/t_i - 1 \geq 1$ for sufficiently large $p$.

From the left inequality, we have
\[
T_i > \frac{p}{t_i} - 1 \geq \frac{p}{t_i} - \frac{p}{2n} = \frac{p}{t_i}\left(1 - \frac{t_i}{2n}\right).
\]

For $1 \leq i \leq n/8$ and assuming $t_i \leq n/4$, we have $t_i/(2n) \leq 1/8$. Thus
\[
T_i > \frac{p}{t_i}\left(1 - \frac{1}{8}\right) = \frac{7p}{8t_i}.
\]

This gives us
\[
t_i < \frac{8p}{7T_i}.
\]

Combining our bounds, we have
\[
T_i \leq \frac{p}{t_i} \quad \text{and} \quad t_i < \frac{8p}{7T_i}.
\]

From $T_i \leq p/t_i$, we get $t_i T_i \leq p$. From $t_i < 8p/(7T_i)$, we get $t_i > 7p/(8T_i) - \epsilon$ for small $\epsilon$.

Since $\mathbb{P}(Y_i = 1) = T_i/p$, we have achieved the construction where $\mathbb{P}(Y_i = 1) = 1/T_i$ exactly (by replacing $T_i$ with the appropriate scaling).

To satisfy the condition $T_i \leq t_i \leq c T_i$ for constant $c > 1$, we observe that from our construction:
\[
\frac{p}{t_i} - 1 < T_i \leq \frac{p}{t_i}
\]
implies
\[
\frac{p}{T_i + 1} < t_i \leq \frac{p}{T_i}.
\]

For large enough $p$ relative to $T_i$, we can bound
\[
\frac{p}{T_i + 1} \geq \frac{p}{T_i(1 + 1/T_i)} \geq \frac{p}{T_i} \cdot \frac{1}{2}
\]
when $T_i \geq 1$.

Thus we obtain $T_i \leq t_i \leq 2T_i$, giving us the desired approximation with constant $c = 2$.

Therefore, we have constructed $n$ pairwise independent Bernoulli random variables $Y_1, \ldots, Y_n$ such that $Y_i$ takes the value $1$ with probability $1/T_i$ where $T_i \leq t_i \leq c T_i$ for constant $c > 1$.
\end{proof}

\subsection{Problem 11}

\subsubsection{Problem Statement}
\begin{theorem}[M. Luby, 1982]
Combining the results of Problems 12.9 and 12.10, the Parallel MIS algorithm can be derandomized to yield an $\mathcal{NC}$ algorithm for the MIS problem. Note that the approach in Problem 12.10 will not work for marking vertices with degree exceeding $n/16$, and these will have to be dealt with separately.
\end{theorem}

\subsubsection{GPT}
\begin{proof}
We are tasked with derandomizing the Parallel MIS algorithm to yield an $\mathcal{NC}$ algorithm for the Maximum Independent Set (MIS) problem, as stated in the theorem. The key components of the proof rely on combining the results from Problems 12.9 and 12.10, while also addressing the challenge posed by vertices with degree exceeding $n/16$.

Let $G = (V, E)$ be a graph with $n$ vertices. The Parallel MIS algorithm relies on a probabilistic approach to select a maximal independent set (MIS) in parallel by iteratively selecting independent sets using random choices. To derandomize this algorithm, we must eliminate the dependence on random choices while preserving the parallelism and efficiency of the algorithm.

 Step 1: Derandomization of Parallel MIS Algorithm
From Problem 12.9, it is known that random choices in a parallel algorithm can often be replaced by deterministic processes through the application of pseudorandom generators. Specifically, given the randomness in the Parallel MIS algorithm, a pseudorandom generator can be employed to simulate the random choices deterministically. This simulation ensures that the algorithm continues to operate in parallel without introducing a significant increase in computational complexity. In particular, the pseudorandom generator can be constructed to operate in $\mathcal{NC}$, ensuring that the derandomized version of the Parallel MIS algorithm runs in $\mathcal{NC}$ time.

 Step 2: Handling High-Degree Vertices
In Problem 12.10, it is shown that for vertices with degree at most $n/16$, a particular approach works efficiently for selecting independent sets. However, this approach does not extend to vertices with degree exceeding $n/16$. Hence, we must handle these high-degree vertices separately. For these vertices, we apply a different method that guarantees that they do not interfere with the correctness of the MIS computation.

 Step 3: Construction of the Final Algorithm
The final derandomized Parallel MIS algorithm proceeds as follows:

1. For each vertex $v \in V$, check if the degree of $v$ is at most $n/16$. If so, apply the derandomized approach from Problem 12.9 to select an MIS.
2. For vertices with degree greater than $n/16$, apply the method from Problem 12.10, which ensures that these vertices are handled independently and do not violate the independence property of the set being constructed.
3. Combine the results of the two methods. Since the vertices with degree greater than $n/16$ are handled separately, the resulting independent set is guaranteed to be maximal.

This combination of methods ensures that the overall algorithm runs in parallel and is deterministically computed, with the computational complexity remaining within $\mathcal{NC}$, as desired.

 Conclusion
Thus, by employing pseudorandom generators for the derandomization of random choices and separately handling vertices with degree exceeding $n/16$, we obtain a deterministic, parallel algorithm for the Maximum Independent Set problem that operates in $\mathcal{NC}$. This completes the proof of the theorem.

\end{proof}

\subsubsection{Grok}

\subsubsection{Gemini}
\begin{proof}
Let $G = (V, E)$ be the input graph with $|V| = n$. We construct a deterministic parallel algorithm to find a Maximal Independent Set (MIS) by iterating through phases. Let $G_i = (V_i, E_i)$ denote the subgraph induced by the remaining vertices at the beginning of phase $i$, with $G_0 = G$. The algorithm proceeds until $V_i = \emptyset$.

We distinguish between two cases based on the maximum degree in the current graph $G_i$. Let $\Delta_i$ denote the maximum degree of any vertex in $G_i$. The theorem stipulates a caveat for vertices with degree exceeding $n/16$. Accordingly, we define the procedure for phase $i$ as follows.

Case 1: High-Degree Vertices. Suppose there exists a vertex $v \in V_i$ such that $\deg_{G_i}(v) > n/16$. In this case, the algorithm deterministically selects the vertex $v_{max} \in V_i$ with the maximum degree (breaking ties lexicographically). We add $v_{max}$ to the independent set $I$ and remove $v_{max}$ and all its neighbors from the graph. Thus, $V_{i+1} = V_i \setminus (\{v_{max}\} \cup N_{G_i}(v_{max}))$. Since $|N_{G_i}(v_{max})| > n/16$, each such step removes at least $n/16 + 1$ vertices from the graph. Consequently, this case can occur at most 16 times throughout the entire execution of the algorithm. The operation of finding the maximum degree vertex is performed in $O(\log n)$ time using $O(n)$ processors, which is within $\mathcal{NC}$.

Case 2: Low-Degree Vertices. Suppose that for all $v \in V_i$, $\deg_{G_i}(v) \le n/16$. We invoke the result of Problem 12.9. This result establishes the existence of a probability distribution over subsets of vertices (specifically, a pairwise independent distribution or a related limited-independence construction) such that selecting a candidate set $S \subset V_i$ and refining it to an independent set $S'$ yields a significant reduction in the number of edges. Specifically, the result guarantees that the expected number of edges removed, $\mathbb{E}[|E_i| - |E_{i+1}|]$, is at least $c \cdot |E_i|$ for some constant $c > 0$, provided the maximum degree is bounded relative to the sample space size (satisfied here by $\Delta_i \le n/16$).

To derandomize this step, we utilize the result of Problem 12.10. This result demonstrates that the sample space $\Omega$ required to generate the distribution in Problem 12.9 is of polynomial size in $n$ (typically $|\Omega| = O(n^k)$ for some constant $k$). Furthermore, the construction allows for the deterministic evaluation of the objective function (the number of edges removed) for every point in the sample space in parallel. The algorithm computes the number of edges that would be removed for each sample point in $\Omega$ and selects the sample point that maximizes this quantity. Since the maximum must exceed the expectation, the deterministic choice guarantees that $|E_{i+1}| \le (1 - c)|E_i|$.

This search over the polynomial-sized sample space can be implemented in $O(\log n)$ time using polynomially many processors. Since the number of edges decreases by a constant factor in each phase of Case 2, the number of such phases required to reduce the edge set to empty is $O(\log |E_0|) = O(\log n)$.

Combining both cases, the total number of phases is at most $16 + O(\log n)$, which is $O(\log n)$. Each phase executes in polylogarithmic time with polynomial processors. Therefore, the derived deterministic algorithm places the MIS problem in the complexity class $\mathcal{NC}$.
\end{proof}

\subsubsection{Claude}
\begin{proof}
We establish that the Parallel MIS algorithm can be derandomized to yield an $\mathcal{NC}$ algorithm for the MIS problem by combining the results from Problems 12.9 and 12.10, with special treatment for high-degree vertices.

Let $G = (V, E)$ be the input graph with $n$ vertices. We partition the vertex set into two classes: vertices with degree at most $n/16$, denoted $V_{\text{low}}$, and vertices with degree exceeding $n/16$, denoted $V_{\text{high}}$.

\textbf{Treatment of High-Degree Vertices.} First, we handle vertices in $V_{\text{high}}$. Since each vertex in $V_{\text{high}}$ has degree greater than $n/16$, and the total number of edges is at most $\binom{n}{2} < n^2/2$, we have $|V_{\text{high}}| \cdot (n/16) < n^2/2$, which implies $|V_{\text{high}}| < 8n$. In fact, more carefully, since each edge is counted at most twice in the sum of degrees, we have $|V_{\text{high}}| < 16$. 

For this constant-size set of vertices, we can exhaustively search all $2^{|V_{\text{high}}|}$ subsets in parallel to find the maximum independent set restricted to $V_{\text{high}}$ and their incident edges. Since $|V_{\text{high}}|$ is constant (at most 16), this requires constant depth and polynomial work, hence is in $\mathcal{NC}$.

\textbf{Derandomization for Low-Degree Vertices.} After removing $V_{\text{high}}$ and their neighbors from consideration (or marking them appropriately), we focus on the induced subgraph on the remaining vertices, all of which have degree at most $n/16$.

By Problem 12.9, the randomized Parallel MIS algorithm succeeds with high probability when vertices independently mark themselves with appropriate probabilities inversely proportional to their degrees. Specifically, each vertex $v$ marks itself with probability $p_v = \Theta(1/d_v)$ where $d_v$ is the degree of $v$.

By Problem 12.10, we can derandomize this randomized algorithm using the method of conditional probabilities or limited independence. The key observation is that for vertices with degree bounded by $n/16$, we need only $O(\log n)$-wise independence to maintain the concentration properties required by the algorithm. Such limited independence can be generated deterministically using $O(\log n)$ truly random bits, which can be eliminated by trying all $2^{O(\log n)} = \text{poly}(n)$ possibilities in parallel.

More precisely, let $k = O(\log n)$ be sufficient for $k$-wise independence. We construct a $k$-wise independent probability space over the marking decisions using standard constructions (such as using polynomials over finite fields), which requires $O(k) = O(\log n)$ random bits. We then enumerate all $2^{O(\log n)} = \text{poly}(n)$ settings of these random bits in parallel.

For each setting, we simulate the Parallel MIS algorithm in parallel (which itself runs in $O(\log n)$ rounds with polylogarithmic depth per round). By the analysis from Problem 12.9, combined with the bounded degree assumption, at least one of these parallel executions produces a maximal independent set.

\textbf{Combining the Results.} We integrate the solutions for high-degree and low-degree vertices as follows. We first identify $V_{\text{high}}$ in $\mathcal{NC}$ (by parallel degree computation). We then solve the problem on the high-degree vertices in $\mathcal{NC}$ as described above. Finally, we apply the derandomized algorithm to the remaining graph induced by vertices not in $V_{\text{high}}$ and not adjacent to selected vertices from $V_{\text{high}}$.

The overall algorithm has polylogarithmic depth: the degree computation takes $O(\log n)$ depth, the exhaustive search on $V_{\text{high}}$ takes $O(1)$ depth, and the derandomized parallel algorithm takes $O(\log^2 n)$ depth (polynomially many parallel branches, each of depth $O(\log n)$, but executed in parallel). The total work is polynomial in $n$. Therefore, the algorithm is in $\mathcal{NC}$, completing the derandomization of the Parallel MIS algorithm.
\end{proof}

\subsection{Problem 12}

\subsubsection{Problem Statement}
\begin{theorem}
Consider the following randomized algorithm for the Maximum Independent Set (MIS) problem on a graph $G = (V, E)$:
\begin{enumerate}
\item For each vertex $v \in V$, independently and uniformly choose a random weight $w(v)$ from the set $\{1, \ldots, n^4\}$.
\item Repeatedly strip off an independent set $S$ and its neighbors $\Gamma(S)$ from the graph $G$, where at each iteration the set $S$ is the set of all locally maximal vertices generated by the following process: mark all vertices in $V$, and then in parallel for each edge $e \in E$ unmark the endpoint of larger weight.
\end{enumerate}
This algorithm yields an RNC (Random NC) algorithm for MIS.
\end{theorem}

\subsubsection{GPT}
\begin{proof}
We are given the randomized algorithm described in the theorem, and we aim to prove that this algorithm computes a maximal independent set (MIS) and operates in parallel polynomial time, i.e., it runs in Randomized NC (RNC). We will also prove that the algorithm cannot be derandomized using pairwise independence.

 Step 1: Validity of the Independent Set

Let $G = (V, E)$ be the input graph with vertex set $V$ and edge set $E$. The algorithm operates by assigning a random weight $w(v) \in \{1, \ldots, n^4\}$ to each vertex $v \in V$, where $n = |V|$. In each iteration, the algorithm identifies an independent set $S$ and removes $S$ and its neighbors $\Gamma(S)$ from the graph.

We first show that the set $S$ produced by the algorithm is indeed independent at each step. Consider the procedure for marking and unmarking vertices: at each edge $e = (u, v)$, the algorithm unmarks the endpoint with the larger weight. Thus, only one of the endpoints of any edge can be unmarked. Since no two vertices in $S$ can be neighbors (because they are unmarked independently for each edge), it follows that $S$ is an independent set.

 Step 2: Size of the Independent Set

The size of the set $S$ in each iteration is bounded by the number of unmarked vertices in $V$. Initially, all vertices are unmarked, and the algorithm removes vertices and their neighbors from the graph in each iteration. Because the weight assignment is random, the expected size of the independent set in each step is $\Theta(n)$, and this is confirmed by the fact that each vertex has a probability of $1/2$ of being selected in each iteration, since the unmarking process is symmetric.

 Step 3: Efficiency of the Algorithm

To analyze the running time, we note that in each iteration, the algorithm marks and unmarks the vertices in parallel. The marking and unmarking operations can be performed in parallel for all edges, with each operation involving a constant amount of work. This makes the algorithm run in polynomial time with respect to the number of vertices, since the number of edges is at most $n^2$ and each operation is done in parallel. Thus, the algorithm runs in $O(n^2)$ parallel time, which is considered an RNC algorithm.

 Step 4: Derandomization with Pairwise Independence

We now argue that this algorithm cannot be derandomized using pairwise independence. The key observation is that the algorithm relies on the random weights assigned to vertices to determine the independent set. If we were to attempt a derandomization using pairwise independent random variables, we would lose the uniform distribution over the range $\{1, \ldots, n^4\}$ for each vertex's weight, which is critical for the algorithm's correctness. Specifically, the probability of any vertex being selected as part of the independent set is dependent on the random weights, and pairwise independence would not suffice to preserve this probabilistic structure. Therefore, the algorithm cannot be derandomized using pairwise independence.

 Conclusion

We have shown that the algorithm produces a valid maximal independent set (MIS) at each iteration, runs in parallel polynomial time, and cannot be derandomized using pairwise independence. Therefore, the algorithm is an RNC algorithm for computing a maximal independent set.
\end{proof}

\subsubsection{Grok}

\subsubsection{Gemini}

\subsubsection{Claude}
\begin{proof}
We prove that the given randomized algorithm is an RNC algorithm for the Maximum Independent Set problem by establishing that it runs in polylogarithmic parallel time with a polynomial number of processors and succeeds with high probability.

\textbf{Step 1: Analysis of a single iteration.}

Consider a single iteration of the algorithm. For each vertex $v \in V$, we assign a random weight $w(v)$ chosen uniformly and independently from $\{1, \ldots, n^4\}$. A vertex $v$ is marked initially, and becomes part of $S$ if and only if $w(v) < w(u)$ for all neighbors $u \in \Gamma(v)$, where $\Gamma(v)$ denotes the neighborhood of $v$.

For any edge $e = \{u, v\} \in E$, the probability that $w(u) \neq w(v)$ is at least $1 - \frac{1}{n^4}$ since the weights are chosen from a set of size $n^4$. When $w(u) \neq w(v)$, exactly one endpoint is unmarked. Therefore, the set $S$ selected in each iteration is indeed an independent set.

\textbf{Step 2: Lower bound on the expected size of $S$.}

Let $I^*$ be a maximum independent set of $G$. For any vertex $v \in I^*$, the probability that $v$ is included in $S$ is the probability that $v$ has the minimum weight among itself and all its neighbors. Since the weights are chosen uniformly and independently, we have
$$\Pr[v \in S] \geq \frac{1}{d(v) + 1}$$
where $d(v)$ is the degree of $v$.

By linearity of expectation, the expected size of $S \cap I^*$ is
$$\mathbb{E}[|S \cap I^*|] = \sum_{v \in I^*} \Pr[v \in S] \geq \sum_{v \in I^*} \frac{1}{d(v) + 1}.$$

Since $I^*$ is an independent set, each edge can be incident to at most one vertex in $I^*$. Therefore,
$$\sum_{v \in I^*} d(v) \leq |E| \leq \binom{n}{2}.$$

Let $\Delta$ denote the maximum degree of $G$. For the worst case, assume all vertices in $I^*$ have degree $\Delta$. Then
$$\mathbb{E}[|S \cap I^*|] \geq \frac{|I^*|}{\Delta + 1}.$$

More generally, for the entire graph, at least a constant fraction of the maximum independent set is selected in expectation at each iteration when considering the appropriate potential function.

\textbf{Step 3: Bounding the number of iterations.}

Let $V_i$ denote the set of vertices remaining after $i$ iterations. We analyze the progress of the algorithm by considering how many vertices are removed at each iteration.

In each iteration, we remove the independent set $S$ and all vertices in $\Gamma(S)$. For each vertex $v$ remaining in $V_i$, the probability that $v$ or one of its neighbors is selected in $S$ is bounded below by a function depending on the local structure.

Using a potential function argument, consider $\Phi_i = |V_i|$. By analyzing the expected decrease in the potential, we can show that $\mathbb{E}[\Phi_{i+1}] \leq (1 - \frac{1}{c \Delta}) \Phi_i$ for some constant $c$, where $\Delta$ is the maximum degree.

By standard concentration inequalities (such as the Chernoff bound), with high probability (probability at least $1 - n^{-c'}$ for some constant $c' > 0$), the number of vertices decreases by a constant factor every $O(\log n)$ iterations.

Therefore, after $O(\log^2 n)$ iterations, all vertices are removed with high probability, and the union of all selected independent sets $S$ forms a maximal independent set.

\textbf{Step 4: Parallel complexity analysis.}

Each iteration can be implemented in parallel as follows:

(a) Assigning random weights to all vertices can be done in $O(1)$ parallel time with $n$ processors.

(b) For each edge $e = \{u, v\}$, comparing $w(u)$ and $w(v)$ and unmarking the vertex with larger weight can be done in $O(1)$ parallel time with $|E|$ processors.

(c) Collecting all marked vertices into $S$ and identifying $\Gamma(S)$ can be done in $O(\log n)$ parallel time with a polynomial number of processors using parallel prefix operations.

(d) Removing $S \cup \Gamma(S)$ from the graph can be done in $O(\log n)$ parallel time with a polynomial number of processors.

Thus, each iteration takes $O(\log n)$ parallel time with a polynomial number of processors.

\textbf{Step 5: Total complexity.}

Since the algorithm terminates in $O(\log^2 n)$ iterations with high probability (as established in Step 3), and each iteration takes $O(\log n)$ parallel time, the total parallel time complexity is
$$O(\log n) \cdot O(\log^2 n) = O(\log^3 n).$$

The total number of processors required is polynomial in $n$ (at most $O(n^2)$ for handling edges and vertices at each iteration).

Therefore, the algorithm runs in polylogarithmic parallel time with a polynomial number of processors, which places it in the class RNC. This completes the proof that the given randomized algorithm is an RNC algorithm for the Maximum Independent Set problem.
\end{proof}

\subsection{Problem 13}

\subsubsection{Problem Statement}
\begin{theorem}[RNC Implementation of Randomized Min-Cut Algorithm]
Let $G = (V, E)$ be an undirected graph with $n$ vertices and $m$ edges. The randomized min-cut algorithm based on edge contraction can be implemented in $\mathcal{RNC}$ (randomized Nick's Class) as follows:

Generate a random permutation $\pi: E \to \{1, 2, \ldots, m\}$ of the edges. Contract the edges in the order determined by $\pi$, terminating when exactly two connected components remain. Assuming there exists an $\mathcal{NC}$ algorithm for determining connected components, this implementation runs in $\mathcal{RNC}$.

The key observation is that while edge contraction appears sequential, it can be parallelized by: (1) generating a random permutation on the edges in parallel, (2) using this permutation to determine the contraction order, and (3) the contraction algorithm terminates at the point in the permutation where the preceding edges constitute a graph with exactly two connected components.
\end{theorem}

\subsubsection{GPT}
\begin{proof}
We are tasked with proving that there exists an $\mathcal{NC}$ algorithm for finding a minimum cut in a graph $G = (V, E)$ with $n$ vertices and $m$ edges, using the randomized contraction algorithm, provided that there is an $\mathcal{NC}$ algorithm for determining connected components.

The randomized contraction algorithm works by first generating a random permutation of the edges $e_1, e_2, \ldots, e_m$ of the graph $G$. The algorithm then proceeds to contract the edges sequentially in the order of this permutation. At each step, an edge $e_i$ is selected, and the vertices at the endpoints of $e_i$ are merged into a single vertex, effectively reducing the number of vertices in the graph. This process is repeated until the graph is reduced to exactly two connected components. At this point, the algorithm terminates, and the remaining two components form a minimum cut of the graph.

To prove that this algorithm runs in $\mathcal{NC}$, we begin by considering the computational requirements of the algorithm. Specifically, we need to analyze how the edge contractions are performed and how the connected components are determined during the algorithm.

1. **Randomized Edge Contraction**: 
   Each edge contraction involves merging two vertices. The operation of merging vertices can be done in parallel, as it only requires updating the adjacency relations between vertices. Given that there is an $\mathcal{NC}$ algorithm for determining connected components, the merging of vertices at each step can also be performed in $\mathcal{NC}$. This ensures that each edge contraction step can be carried out in parallel, provided we have a way to handle the connected components efficiently.

2. **Connected Components Determination**: 
   The critical step in the algorithm is determining whether the graph has exactly two connected components after each contraction. By assumption, there exists an $\mathcal{NC}$ algorithm for determining the connected components of a graph. After each contraction, this algorithm can be used to check the connectivity of the graph, which allows the algorithm to halt once the graph is reduced to two connected components. Since the connected components can be determined in parallel, this check can also be performed in $\mathcal{NC}$.

3. **Random Permutation of Edges**:
   The generation of a random permutation of the edges $e_1, e_2, \ldots, e_m$ can be done in parallel using an $\mathcal{NC}$ algorithm. Specifically, a parallel algorithm for generating random numbers can be employed to randomly assign each edge an index, and the edges can then be ordered according to this random assignment.

4. **Algorithm Running Time**:
   The algorithm consists of $m$ edge contractions. Since each contraction involves a constant amount of work (merging vertices and checking connected components), and the connected components can be determined in parallel in $\mathcal{NC}$, the total running time of the algorithm is dominated by the number of edge contractions, which is $m$. Each step of the contraction can be performed in parallel, so the overall running time is logarithmic in the number of steps, which is $O(\log m)$, and thus the entire algorithm runs in $\mathcal{NC}$.

5. **Conclusion**:
   Since each step of the randomized contraction algorithm can be performed in parallel, and the determination of connected components is also an $\mathcal{NC}$ operation, it follows that the entire algorithm runs in $\mathcal{NC}$. Therefore, we have shown that there exists an $\mathcal{NC}$ algorithm for finding a minimum cut in the graph $G = (V, E)$.

\end{proof}

\subsubsection{Grok}

\subsubsection{Gemini}

\subsubsection{Claude}
\begin{proof}
We prove that the randomized min-cut algorithm based on edge contraction can be implemented in $\mathcal{RNC}$ by establishing that each step can be performed in randomized parallel polylogarithmic time with polynomially many processors.

Let $G = (V, E)$ be an undirected graph with $n = |V|$ vertices and $m = |E|$ edges. We analyze the complexity of the algorithm in the PRAM model.

\textbf{Step 1: Random Permutation Generation.}
We first show that generating a random permutation $\pi: E \to \{1, 2, \ldots, m\}$ can be accomplished in $\mathcal{RNC}$. For each edge $e_i \in E$, independently sample a random value $r_i \in [0, 1]$ uniformly at random. Using a parallel sorting network (such as AKS sorting network or Cole's merge sort), we can sort the edges by their random values in $O(\log m) = O(\log n)$ time using $O(m \log m)$ processors with high probability. This sorting induces a uniformly random permutation on the edges. Since parallel sorting is in $\mathcal{NC}$ when combined with randomized tie-breaking, this step is in $\mathcal{RNC}$.

\textbf{Step 2: Parallel Simulation of Edge Contraction.}
Given the permutation $\pi$, we define a sequence of graphs $G_0, G_1, \ldots, G_k$ where $G_0 = G$ and $G_i$ is obtained from $G_{i-1}$ by contracting the edge $e_{\pi(i)}$ (if it has not already been contracted by previous operations). The algorithm terminates at the smallest index $k$ such that $G_k$ has exactly two connected components.

The key insight is that we can compute $k$ and the final cut without explicitly simulating the sequential contraction process. For any prefix of length $\ell$ of the permutation, let $E_\ell = \{e_{\pi(1)}, e_{\pi(2)}, \ldots, e_{\pi(\ell)}\}$ denote the first $\ell$ edges in the permutation order. Consider the graph $H_\ell = (V, E \setminus E_\ell)$ obtained by removing these edges from $G$. The number of connected components in $G_\ell$ (the graph after contracting edges in $E_\ell$) equals the number of connected components in $H_\ell$.

\textbf{Step 3: Binary Search for Termination Point.}
We use parallel binary search to find the value $k = \min\{\ell : H_\ell \text{ has exactly } 2 \text{ connected components}\}$. For any value $\ell \in \{1, 2, \ldots, m\}$, we can determine the number of connected components in $H_\ell$ using the assumed $\mathcal{NC}$ algorithm for connected components. 

Parallel binary search proceeds as follows: We evaluate $O(\log m)$ candidate values in parallel at each level of the binary search tree. At level $j$, we test $O(1)$ values of $\ell$ by constructing the corresponding graphs $H_\ell$ and computing their connected components in parallel. Since connected components can be computed in $\mathcal{NC}$, each evaluation takes $O(\log^c n)$ time for some constant $c$ using polynomially many processors. The binary search requires $O(\log m) = O(\log n)$ levels, yielding total time $O(\log^{c+1} n)$.

\textbf{Step 4: Extracting the Min-Cut.}
Once we have determined $k$, the min-cut is given by the edges connecting the two connected components of $H_k$. We can identify these components using the $\mathcal{NC}$ connected components algorithm. For each edge $e = (u, v) \in E \setminus E_k$, we check in parallel whether $u$ and $v$ belong to different components in $H_k$. This can be done in $\mathcal{NC}$ by comparing component labels. The set of such edges forms the cut.

\textbf{Complexity Analysis.}
The overall algorithm consists of the following phases, each in $\mathcal{RNC}$:
\begin{itemize}
\item Random permutation generation: $O(\log n)$ time with $O(n^2 \log n)$ processors
\item Binary search with connected components computation: $O(\log^{c+1} n)$ time with polynomial processors
\item Cut extraction: $O(\log^c n)$ time with polynomial processors
\end{itemize}

The total running time is $O(\log^{c+1} n) = \text{polylog}(n)$ with polynomially many processors, where the exponent depends on the complexity of the connected components algorithm. Since all randomization is confined to the initial permutation generation and all subsequent steps are deterministic given the permutation, the entire algorithm is in $\mathcal{RNC}$.

Therefore, the randomized min-cut algorithm based on edge contraction can be implemented in $\mathcal{RNC}$, assuming the existence of an $\mathcal{NC}$ algorithm for connected components.
\end{proof}

\subsection{Problem 14}

\subsubsection{Problem Statement}
\begin{theorem}[Luby, Naor, and Naor]
There exists a randomized algorithm for the min-cut problem that runs in polynomial time and uses only a polylogarithmic number of random bits. The algorithm proceeds as follows: Select a set of edges by choosing each edge pairwise independently with probability $1/c$, where $c$ is the size of the min-cut. Contract all edges in this set in parallel. Repeat this process until the graph is reduced to two vertices. This implies that the min-cut problem is in $\mathcal{RNC}$.
\end{theorem}

\subsubsection{GPT}
\begin{proof}
We are given the statement that there exists an RNC algorithm for the minimum cut problem that uses only a polylogarithmic number of random bits. The algorithm works by selecting a set of edges via pairwise independent random choices, contracting the edges in parallel, and repeating the process until the graph is reduced to two vertices. We aim to prove that this construction implies the minimum cut problem is in $\mathcal{NC}$ when randomness is restricted to pairwise independence.

Let $G = (V, E)$ be a graph with $n$ vertices and $m$ edges, and let $S \subseteq V$ be the set of vertices corresponding to the minimum cut. Let the size of the minimum cut be denoted by $c = |E(S, V \setminus S)|$.

 Step 1: Algorithm Description and Randomization
In each round of the algorithm, each edge $e \in E$ is independently chosen with probability $1/c$, where $c$ is the size of the minimum cut. We let $E' \subseteq E$ be the set of edges chosen in this round. After this selection, the algorithm contracts all edges in $E'$, effectively reducing the number of vertices in the graph.

The process is repeated until only two vertices remain, which corresponds to the desired minimum cut being discovered.

 Step 2: Analysis of the Random Process
We now analyze the probability that any given edge $e$ belongs to the selected set $E'$. Since each edge is selected independently with probability $1/c$, the expected number of edges selected is $m/c$. After one round of contraction, the number of vertices in the graph decreases. Let $n_i$ be the number of vertices remaining after $i$ rounds. Initially, $n_0 = n$, and after the $i$-th round, we have $n_{i+1} = n_i - \lceil m_i / c \rceil$, where $m_i$ is the number of edges remaining after round $i$.

This process proceeds until there are only two vertices remaining, at which point the minimum cut is revealed.

 Step 3: Correctness of the Algorithm
We must now show that this process correctly identifies the minimum cut with high probability. The key observation is that with high probability, the algorithm selects the minimum cut edges early in the process. This is because each edge in the minimum cut is independently selected with probability $1/c$. Since the minimum cut is small, the probability that any edge in the minimum cut is missed in each round decays exponentially as the number of rounds increases.

The probability of missing any specific edge in the minimum cut is given by:
\[
\left( 1 - \frac{1}{c} \right)^r
\]
where $r$ is the number of rounds. As $r$ increases, this probability becomes arbitrarily small, ensuring that with high probability, all edges in the minimum cut are selected before the graph is contracted to two vertices.

 Step 4: Polylogarithmic Use of Random Bits
The algorithm uses only pairwise independent random bits to decide which edges to select. Each edge is chosen independently with probability $1/c$, which requires only a constant number of random bits for each edge. Since the number of edges is $m$ and the number of rounds is logarithmic in $n$, the total number of random bits used by the algorithm is polylogarithmic in the size of the input graph.

 Step 5: Complexity Class
The algorithm is implemented in parallel, where in each round, a constant-time decision is made for each edge, and the graph is contracted in parallel. Since the process involves logarithmic rounds and each round can be executed in parallel, the algorithm runs in $\mathcal{NC}$, and the randomness is restricted to pairwise independence.

Thus, the minimum cut problem is in $\mathcal{NC}$ when randomness is restricted to pairwise independence.

\end{proof}

\subsubsection{Grok}

\subsubsection{Gemini}
\begin{proof}
Let $G = (V, E)$ be an undirected graph with $n = |V|$ vertices and $m = |E|$ edges. Let $C \subseteq E$ be a minimum cut of the graph, with size $|C| = c$.

\textbf{Pairwise Independent Edge Selection:}
Let $\{X_e\}_{e \in E}$ be a set of indicator random variables for each edge $e \in E$, where $X_e = 1$ if edge $e$ is selected for contraction, and $X_e = 0$ otherwise. The variables are identically distributed with $\mathbb{P}[X_e = 1] = \frac{1}{c}$ and are pairwise independent.
The sample space $\Omega$ required to generate $m$ pairwise independent random variables with a specific marginal probability can be constructed using a number of random bits logarithmic in $m$. Specifically, using a construction based on the sum of subsets modulo 2 or linear functions over finite fields, the size of the sample space is $|\Omega| = O(m^2)$. The number of random bits required to sample from $\Omega$ is $r = \log_2 |\Omega| = O(\log m) = O(\log n)$.

\textbf{Algorithm Execution and Complexity:}
The algorithm proceeds in phases. In each phase $i$, a set of edges $S_i = \{e \in E_i \mid X_e = 1\}$ is selected and contracted.
1.  \textbf{Graph Reduction:} Since the minimum cut size is $c$, the minimum degree in $G$ is at least $c$. Thus, $2m \geq nc$, implying $m \geq \frac{nc}{2}$. The expected number of edges selected is $\mathbb{E}[\sum X_e] = \frac{m}{c} \geq \frac{n}{2}$. Contracting these edges reduces the number of vertices significantly in expectation. It can be shown that the number of vertices reduces by a constant factor with constant probability, implying the algorithm terminates in $O(\log n)$ phases.
2.  \textbf{Total Random Bits:} Let $T = O(\log n)$ be the number of phases. The total number of random bits used by the algorithm is the sum of bits used in each phase. Since we can reuse the generator logic or generate a sufficiently long sequence, the total bits required is $R = T \cdot O(\log n) = O(\log^2 n)$.

\textbf{Preservation of the Min-Cut:}
For the algorithm to succeed, no edge in the minimum cut $C$ must be contracted. Let $Y = \sum_{e \in C} X_e$ be the number of min-cut edges selected in a single phase. By linearity of expectation, $\mathbb{E}[Y] = \sum_{e \in C} \mathbb{E}[X_e] = c \cdot \frac{1}{c} = 1$.
While the union bound gives $\mathbb{P}[Y \geq 1] \leq 1$, the pairwise independence ensures the variance is bounded:
\[ \text{Var}(Y) = \sum_{e \in C} \text{Var}(X_e) = c \left(\frac{1}{c}\right)\left(1 - \frac{1}{c}\right) = 1 - \frac{1}{c}. \]
The existence of a successful sequence of contractions is guaranteed by the probabilistic method applied to the pairwise independent space. Since the algorithm is an RNC algorithm (Randomized NC), it succeeds with probability bounded away from zero (or can be amplified).

\textbf{Implication for $\mathcal{NC}$:}
The theorem states that this algorithm implies the min-cut problem is in $\mathcal{NC}$. This follows from the derandomization of RNC algorithms that use polylogarithmic random bits.
Since the total number of random bits $R$ is $O(\log^2 n)$, the size of the total sample space over all phases is $2^R = 2^{O(\log^2 n)} = n^{O(\log n)}$, which is quasi-polynomial. However, for the specific implication to $\mathcal{NC}$, we utilize the property that the pairwise independent sample space for each step is polynomial in size ($O(m^2)$).
We can construct a deterministic algorithm in $\mathcal{NC}$ as follows:
1.  In parallel, generate all possible random seeds $\omega \in \Omega$ for the pairwise independent distribution. Since $|\Omega|$ is polynomial in $n$, we can allocate a processor for each seed.
2.  For each seed, simulate the contraction process.
3.  Since the randomized algorithm guarantees the existence of a cut-preserving contraction sequence with non-zero probability, at least one processor will find a cut of size $c$.
4.  The minimum of the cuts found by all processors is the global min-cut.

Thus, the problem can be solved deterministically in polylogarithmic time using a polynomial number of processors. This proves that the min-cut problem is in $\mathcal{NC}$.
\end{proof}

\subsubsection{Claude}
\begin{proof}
We establish the correctness and complexity of the randomized parallel algorithm for the min-cut problem.

Let $G = (V, E)$ be a graph with $n$ vertices and let $c$ denote the size of the minimum cut. We describe the algorithm and prove its properties.

\textbf{Algorithm Description:} The algorithm operates in phases. In each phase, independently include each edge $e \in E$ in a set $S$ with probability $p = 1/c$. Contract all edges in $S$ simultaneously. Repeat until exactly two vertices remain. Output the edges between these two final vertices as the candidate min-cut.

\textbf{Proof of Correctness:} We first show that with constant probability, the algorithm outputs a minimum cut.

Let $C$ be a fixed minimum cut of size $c$ in $G$. For the algorithm to successfully preserve $C$ throughout its execution, no edge from $C$ must be contracted in any phase.

Consider a single phase operating on a graph $G'$ obtained from previous contractions. Let $n'$ denote the number of vertices in $G'$. The minimum cut of $G'$ has size at least $c$ (contracting edges cannot decrease the minimum cut size). Since the minimum degree in $G'$ is at least $c$ (every vertex must be incident to at least $c$ edges, otherwise it would form a smaller cut), we have $|E'| \geq cn'/2$.

In this phase, the probability that no edge from $C$ is selected is $(1 - 1/c)^c \geq 1/e$ for sufficiently large $c$. For small $c$, this probability is bounded below by a positive constant.

The expected number of edges contracted in a single phase is $|E'| \cdot (1/c) \geq n'/2$. By concentration inequalities (specifically, Chernoff bounds), with high probability at least $n'/4$ edges are contracted. Since contracting an edge reduces the vertex count by at most one, with high probability the number of vertices decreases by a constant factor in each phase.

Therefore, the number of phases required is $O(\log n)$ with high probability. The probability that $C$ survives all phases is at least $(1/e)^{O(\log n)} = n^{-O(1)}$, which can be made constant by appropriate choice of constants.

To boost the success probability, we run the algorithm $O(n^2)$ times independently and output the minimum cut found among all trials. The probability of failure becomes at most $(1 - n^{-O(1)})^{O(n^2)} \leq e^{-n^{\Omega(1)}}$, which is exponentially small.

\textbf{Polylogarithmic Random Bits:} Each phase requires $O(m)$ independent random bits to decide which edges to contract, where $m = |E|$. However, we can reduce the randomness using pairwise independence.

For each phase, instead of using fully independent random choices, we use a pairwise independent sample space. By standard constructions, a pairwise independent distribution over $m$ binary random variables can be generated using $O(\log m)$ truly random bits.

Since there are $O(\log n)$ phases and $m \leq n^2$, the total number of random bits required per execution is $O(\log n \cdot \log m) = O(\log^2 n)$. Running $O(n^2)$ independent trials requires $O(n^2 \log^2 n) = O(\log^3 n \cdot \text{poly}(n))$ random bits, which is polylogarithmic in the input size when the graph is represented appropriately.

Alternatively, using more sophisticated derandomization techniques based on limited independence and the method of conditional expectations, the randomness can be reduced to $O(\log^2 n)$ bits for the entire algorithm.

\textbf{Parallel Complexity:} Each phase can be implemented in parallel as follows: First, in $O(\log m)$ parallel time, generate the random bits using pairwise independent hash functions. Second, contract all selected edges simultaneously using parallel graph algorithms, which takes $O(\log n)$ time on a PRAM model. Since there are $O(\log n)$ phases, the total parallel time is $O(\log^2 n)$.

The number of processors required is polynomial in $n$. Therefore, the algorithm runs in polylogarithmic parallel time using polynomially many processors and polylogarithmic random bits.

This establishes that the min-cut problem is in $\mathcal{RNC}$, the class of problems solvable by randomized algorithms in polylogarithmic time using polynomially many processors.
\end{proof}

\subsection{Problem 15}

\subsubsection{Problem Statement}
\begin{theorem}[M.O. Rabin and V.V. Vazirani]
Let $G = (V, E)$ be a graph with a unique perfect matching. There exists an $NC$ algorithm for finding the perfect matching in $G$.
\end{theorem}

\begin{proof}[Hint]
Consider substituting $1$ for each indeterminate in the Tutte matrix. What is the significance of the entries in the adjoint of the Tutte matrix?
\end{proof}

\subsubsection{GPT}
\begin{proof}
Let $G = (V, E)$ be a graph with a unique perfect matching. The goal is to show that there exists an NC algorithm for finding the perfect matching in $G$. 

The key idea is to utilize the Tutte matrix $T$ of the graph $G$. The Tutte matrix $T = (T_{ij})$ is defined as follows: for any pair of vertices $i, j \in V$, we set $T_{ij} = x_{ij}$, where $x_{ij}$ is a formal variable corresponding to the edge $(i,j)$ in the graph. 

First, we recall the definition of the Tutte matrix determinant. The determinant of $T$ is a polynomial in the variables $x_{ij}$, and the perfect matchings of $G$ correspond to the nonzero terms in this polynomial. In particular, the determinant of the Tutte matrix evaluated at $x_{ij} = 1$ for all $i, j$ gives a polynomial whose nonzero terms correspond to the perfect matchings of the graph.

The crucial observation is that the adjoint of the Tutte matrix, denoted $T^*$, is closely related to the perfect matchings of $G$. Specifically, the entry $T^*_{ij}$ of the adjoint matrix corresponds to the signed sum of the perfect matchings that include the edge $(i,j)$, where the sign is determined by the parity of the number of perfect matchings in which the edge $(i,j)$ is used. Since $G$ has a unique perfect matching, there exists exactly one nonzero entry in $T^*$ corresponding to an edge in the unique perfect matching. 

To find the perfect matching, we substitute $1$ for each indeterminate in the Tutte matrix, obtaining the matrix $T(1)$, and compute the adjoint $T^*(1)$. The nonzero entry in the adjoint matrix corresponds to the edge in the unique perfect matching. The position of this nonzero entry provides the edge of the matching.

Since the computation of the adjoint matrix can be performed in parallel in $O(\log^2 n)$ time, where $n$ is the number of vertices in the graph, and since the size of the matrix is $n \times n$, the entire process can be carried out in NC. Therefore, there exists an NC algorithm for finding the unique perfect matching in $G$.

\end{proof}

\subsubsection{Grok}

\subsubsection{Gemini}
\begin{proof}
Let $G = (V, E)$ be a graph with $|V| = 2n$ vertices. We define the Tutte matrix $T$ of $G$ as a $2n \times 2n$ skew-symmetric matrix with entries indexed by the vertices of $G$. For distinct vertices $i, j \in V$, the entry $T_{ij}$ is defined as follows:
\[
T_{ij} = \begin{cases} 
x_{ij} & \text{if } \{i, j\} \in E \text{ and } i < j, \\
-x_{ij} & \text{if } \{i, j\} \in E \text{ and } i > j, \\
0 & \text{otherwise},
\end{cases}
\]
where $\{x_{ij}\}$ is a set of indeterminates. By the definition of the Pfaffian, we have $\det(T) = (\text{Pf}(T))^2$. The Pfaffian is given by $\text{Pf}(T) = \sum_{\pi \in \Pi} \text{sgn}(\pi) \prod_{\{u, v\} \in \pi} T_{uv}$, where $\Pi$ is the set of all perfect matchings in $G$.

Since $G$ has a unique perfect matching $M$, the sum defining the Pfaffian contains exactly one non-zero term (up to sign). Let $B$ be the matrix obtained by substituting $1$ for every indeterminate $x_{ij}$ in $T$. Under this substitution, the term corresponding to $M$ becomes $\pm 1$, and all other terms are zero. Consequently, $\text{Pf}(B) = \pm 1$, and $\det(B) = (\pm 1)^2 = 1$. Since $\det(B) \neq 0$, the matrix $B$ is invertible.

We consider the adjoint matrix of $B$, denoted $\text{adj}(B)$. By the fundamental property of the adjoint, we have $B^{-1} = \frac{1}{\det(B)} \text{adj}(B)$. Since $\det(B) = 1$, it follows that $B^{-1} = \text{adj}(B)$.

Consider an edge $e = \{i, j\} \in E$. We analyze the entry $(B^{-1})_{ji}$. By Cramer's rule and the definition of the adjugate, $(B^{-1})_{ji} = (-1)^{i+j} \det(B_{ji})$, where $B_{ji}$ is the submatrix of $B$ obtained by removing row $j$ and column $i$. In the context of the Tutte matrix, it is a known result that for any edge $\{i, j\}$, the entry $(B^{-1})_{ji}$ is non-zero if and only if the graph $G \setminus \{i, j\}$ admits a perfect matching.

We now utilize the hypothesis that $M$ is the \textit{unique} perfect matching of $G$.
If $\{i, j\} \in M$, then $M \setminus \{\{i, j\}\}$ is a perfect matching of the subgraph $G \setminus \{i, j\}$. Thus, $G \setminus \{i, j\}$ admits a perfect matching, implying $(B^{-1})_{ji} \neq 0$.
Conversely, suppose $G \setminus \{i, j\}$ admits a perfect matching $M'$. Then $M' \cup \{\{i, j\}\}$ is a perfect matching of $G$. Since $M$ is the unique perfect matching of $G$, it must be that $M' \cup \{\{i, j\}\} = M$, which implies $\{i, j\} \in M$.
Therefore, $(B^{-1})_{ji} \neq 0$ if and only if $\{i, j\} \in M$.

Since $B^{-1} = \text{adj}(B)$, the non-zero entries $(\text{adj}(B))_{ji}$ corresponding to edges $\{i, j\} \in E$ precisely identify the edges in the unique perfect matching $M$.

Finally, regarding the computational complexity, the algorithm requires constructing the matrix $B$ (which is trivial) and computing its adjoint (or inverse). Matrix inversion and determinant computation for a matrix of dimension $n$ can be performed by arithmetic circuits of polynomial size and polylogarithmic depth (specifically, in complexity class $\text{NC}^2$) using algorithms such as Csanky's algorithm or Berkowitz's algorithm. Thus, the entire procedure is an NC algorithm.
\end{proof}

\subsubsection{Claude}
\begin{proof}
Let $G = (V, E)$ be a graph with vertex set $V = \{v_1, v_2, \ldots, v_n\}$ and a unique perfect matching $M$. We construct an NC algorithm to find $M$ by analyzing the Tutte matrix with specialized variable assignments.

Recall that the Tutte matrix $T$ of $G$ is the $n \times n$ skew-symmetric matrix defined by
\[
T_{ij} = \begin{cases}
x_{ij} & \text{if } i < j \text{ and } (v_i, v_j) \in E, \\
-x_{ji} & \text{if } i > j \text{ and } (v_i, v_j) \in E, \\
0 & \text{otherwise},
\end{cases}
\]
where $x_{ij}$ are indeterminates. By the Tutte matrix theorem, $\det(T) \neq 0$ if and only if $G$ has a perfect matching.

Now we specialize the Tutte matrix by substituting $x_{ij} = 1$ for all edges $(v_i, v_j) \in E$. Let $A$ denote this specialized matrix, so
\[
A_{ij} = \begin{cases}
1 & \text{if } i < j \text{ and } (v_i, v_j) \in E, \\
-1 & \text{if } i > j \text{ and } (v_i, v_j) \in E, \\
0 & \text{otherwise}.
\end{cases}
\]

Since $G$ has a perfect matching, $\det(A) \neq 0$, and hence $A$ is invertible. Let $\text{adj}(A)$ denote the adjoint (classical adjugate) matrix of $A$, satisfying $A \cdot \text{adj}(A) = \det(A) \cdot I$.

We claim that an edge $(v_i, v_j) \in E$ belongs to the unique perfect matching $M$ if and only if the corresponding entry in $\text{adj}(A)$ is nonzero.

To establish this, consider the expansion of the determinant $\det(T)$ over the polynomial ring in the indeterminates $x_{ij}$. By the definition of the Tutte matrix,
\[
\det(T) = \sum_{\pi \in S_n} \text{sgn}(\pi) \prod_{i=1}^{n} T_{i,\pi(i)}.
\]
A permutation $\pi$ contributes a nonzero term to this sum if and only if for each $i$, either $T_{i,\pi(i)} \neq 0$ (meaning $(v_i, v_{\pi(i)}) \in E$). The permutations corresponding to perfect matchings are precisely those that decompose into disjoint 2-cycles (transpositions), and each such permutation contributes a monomial that is a product of the indeterminates corresponding to the edges in that matching.

Since $G$ has a unique perfect matching $M$, the determinant $\det(T)$ is a nonzero polynomial, and after substitution $x_{ij} = 1$, we have $\det(A) \neq 0$.

Now consider the adjoint matrix. The entry $\text{adj}(A)_{ji}$ equals $(-1)^{i+j}$ times the determinant of the $(i,j)$-minor of $A$, which is the matrix obtained by deleting row $i$ and column $j$ from $A$. 

For an edge $(v_i, v_j) \in E$ with $i < j$, consider the entry $\text{adj}(A)_{ji}$. The $(i,j)$-minor of $A$ corresponds to removing vertices $v_i$ and $v_j$ from the graph. If $(v_i, v_j) \in M$, then $G - \{v_i, v_j\}$ has a perfect matching (namely $M \setminus \{(v_i, v_j)\}$), so the determinant of the minor is nonzero. Conversely, if $(v_i, v_j) \notin M$, then since $M$ is the unique perfect matching of $G$, removing $v_i$ and $v_j$ leaves a graph where the remaining vertices cannot be perfectly matched (as any perfect matching of $G - \{v_i, v_j\}$ together with $(v_i, v_j)$ would give a perfect matching of $G$ different from $M$), so the determinant of the minor is zero.

Therefore, $(v_i, v_j) \in M$ if and only if $\text{adj}(A)_{ji} \neq 0$ (or equivalently, $\text{adj}(A)_{ij} \neq 0$ by the symmetry properties of the construction).

The NC algorithm proceeds as follows: Given $G$, construct the matrix $A$ in parallel. Compute $\det(A)$ and $\text{adj}(A)$ using parallel algorithms. Both determinant and adjoint matrix computation can be performed in NC via Csanky's algorithm or other parallel linear algebra techniques. Finally, output all edges $(v_i, v_j) \in E$ such that $\text{adj}(A)_{ij} \neq 0$ or $\text{adj}(A)_{ji} \neq 0$.

Since all steps can be executed in polylogarithmic time using polynomially many processors, this constitutes an NC algorithm for finding the unique perfect matching in $G$.
\end{proof}

\subsection{Problem 16}

\subsubsection{Problem Statement}
\begin{problem}
Consider the problem of finding a minimum-weight perfect matching in a graph $G(V,E)$, given edge-weights $w(e)$ for each edge $e \in E$ in unary. Note that it is not possible to apply the Isolating Lemma directly to this case since the random weights chosen there would conflict with the input weights. Explain how you would devise an RNC algorithm for this problem. The parallel complexity of the case where the edge-weights are given in binary is as yet unresolved -- do you see why the RNC algorithm does not apply to the case of binary weights? (Hint: Start by scaling up the input edge weights by a polynomially large factor. Apply random perturbations to the scaled edge weights and prove a variant of the Isolating Lemma for this situation.)
\end{problem}

\subsubsection{GPT}
\begin{proof}
Let $G=(V,E)$ be an undirected graph on $n=|V|$ vertices with nonnegative integer edge-weights $w:E\to\mathbb{Z}_{\ge 0}$ given in unary. Write $\mathcal{M}$ for the set of perfect matchings. For $M\in\mathcal{M}$ define $W(M)=\sum_{e\in M}w(e)$. The goal is to design a randomized NC algorithm that outputs a perfect matching of minimum total weight $W(M)$.

Fix a parameter $R\in\mathbb{N}$ with $R\ge 2n$ and let $r:E\to\{0,1,\dots,R-1\}$ be chosen independently and uniformly at random. Let $C=2nR$ and define perturbed weights $w'(e)=C\cdot w(e)+r(e)$. For $M\in\mathcal{M}$ put $W'(M)=\sum_{e\in M}w'(e)=C\cdot W(M)+\sum_{e\in M}r(e)$. Two fundamental properties hold.

First, the $C$-scaling preserves the order of $W$-weights. Indeed, if $W(M_1)<W(M_2)$, then $W'(M_2)-W'(M_1)=C\bigl(W(M_2)-W(M_1)\bigr)+\sum_{e\in M_2}r(e)-\sum_{e\in M_1}r(e) \ge C-\bigl|\sum_{e\in M_2}r(e)-\sum_{e\in M_1}r(e)\bigr|$. Since any perfect matching has at most $n/2$ edges, the absolute difference of the $r$-sums is $< (n/2)(R-1)<nR$. Because $C=2nR$, it follows that $W'(M_2)-W'(M_1)>C-nR\ge nR>0$, hence $W'(M_1)<W'(M_2)$. Therefore $\arg\min_{M\in\mathcal{M}}W'(M)\subseteq\arg\min_{M\in\mathcal{M}}W(M)$.

Second, with high probability the minimizer of $W'$ is unique among the $W$-minimizers. Let $\mathcal{M}^*=\{M\in\mathcal{M}\colon W(M)=\min_{N\in\mathcal{M}}W(N)\}$. Consider the random tie-breaker functional $R(M)=\sum_{e\in M}r(e)$ on $\mathcal{M}^*$. We prove the isolation bound $\Pr\bigl[\text{$R$ has a unique minimizer on $\mathcal{M}^*$}\bigr]\ge 1-\frac{|E|}{R}$. The proof is a standard variant of the Isolating Lemma. For each edge $f\in E$, condition on all $\{r(e)\}_{e\neq f}$. Define the set of $r(f)$-values that create a tie for the minimum of $R$ on $\mathcal{M}^*$; we claim this set has size at most $1$. Indeed, if a tie occurs between distinct $M_1,M_2\in\mathcal{M}^*$, then $R(M_1)=R(M_2)$, which after fixing all $\{r(e)\}_{e\neq f}$ is a linear equation in the single variable $r(f)$ with coefficient in $\{-1,0,1\}$ determined by the membership of $f$ in $M_1\triangle M_2$. Thus, for any fixed conditioning, there is at most one value of $r(f)$ that can produce a new tie. Hence for each $f$ the conditional probability that $r(f)$ is \emph{bad} is at most $1/R$, and by the union bound over $f\in E$ the unconditional probability of there existing a tie for the minimum is at most $|E|/R$. Consequently, with probability at least $1-|E|/R$ there exists a unique $M^\dagger\in\mathcal{M}^*$ minimizing $R$ and therefore a unique $M^\sharp=\arg\min_{M\in\mathcal{M}}W'(M)$; by the first property $M^\sharp$ is a minimum $W$-weight perfect matching.

We now show that, when weights are unary, one can compute $M^\sharp$ in RNC. There are two cases.

In the bipartite case $G=(U,W,E)$ with $|U|=|W|=n/2$, form the Edmonds matrix $A(x)\in\mathbb{F}[x]^{(n/2)\times(n/2)}$ over a field $\mathbb{F}$ of size $\mathrm{poly}(n)$ by setting $A_{ij}(x)=x^{w'(u_i,w_j)}$ if $(u_i,w_j)\in E$ and $A_{ij}(x)=0$ otherwise. Then $\det A(x)=\sum_{\pi}\mathrm{sgn}(\pi)\prod_{i}A_{i,\pi(i)}(x)$ is a polynomial whose monomials are $x^{W'(M)}$ over perfect matchings $M$, possibly with cancelations in coefficient signs. However, for nonnegative exponents, the least exponent of a nonzero monomial in $\det A(x)$ equals $\min_{M\in\mathcal{M}}W'(M)$; moreover, when the minimizer $M^\sharp$ is unique, the coefficient of $x^{W'(M^\sharp)}$ is $\pm 1$ and there is no cancelation at that degree. Therefore the truncated polynomial $\det A(x)\bmod x^{K}$, with $K>W'(M^\sharp)$, encodes the optimum value and, crucially, permits membership tests for individual edges. Fix an edge $e=(u_i,w_j)$. Consider the minor $A^{(e)}(x)$ obtained by deleting row $i$ and column $j$, and the matrix $A^{\neg e}(x)$ obtained by setting $A_{ij}(x)=0$. It is classical that, for any $T\subseteq E$, the minimum perturbed weight of a perfect matching containing $T$ equals the least exponent of a nonzero monomial in the appropriate determinant with rows and columns fixed by $T$. In particular, $e\in M^\sharp$ if and only if the least exponent of $\det A^{\neg e}(x)$ is strictly larger than that of $\det A(x)$. Hence, to recover $M^\sharp$, it suffices to compute in parallel, for all $e\in E$, whether $\mathrm{mindeg}\,\det A^{\neg e}(x)>\mathrm{mindeg}\,\det A(x)$.

All required computations admit RNC implementations as follows. Because $w$ is unary and $r(e)\in\{0,\dots,R-1\}$ with $R=\mathrm{poly}(n)$, one has $W'(M)\le C\cdot \sum_{e\in M}w(e)+\sum_{e\in M}r(e)\le \mathrm{poly}(n)$. Let $K=\mathrm{poly}(n)$ be any bound exceeding the maximum possible $W'(M)$. Work in the ring $\mathbb{F}[x]/(x^{K})$. In this ring, determinants of $n/2\times n/2$ matrices with entries in $\{0\}\cup\{x^{t}\colon 0\le t<K\}$ can be computed by parallel algorithms using Berkowitz's method or division-free parallel circuits of depth $\mathrm{polylog}(n)$ and size $\mathrm{poly}(n)$, because arithmetic in $\mathbb{F}[x]/(x^{K})$ is polynomially bounded and unit-cost in RNC. Random evaluation of any extra symbolic coefficients is unnecessary here since the uniqueness of the minimum exponent eliminates cancelations at the least degree; nevertheless, one may also apply standard randomized identity testing over $\mathbb{F}$ to detect zero polynomials at each degree if desired. Computing $\mathrm{mindeg}$ is done by inspecting the lowest nonzero coefficient in the truncated determinant; comparison between $\det A(x)$ and $\det A^{\neg e}(x)$ yields, for each $e$, a single bit deciding membership in $M^\sharp$. All tests run in parallel and produce $M^\sharp$ in RNC with overall success probability at least $1-|E|/R$ from the isolation step.

In general (not necessarily bipartite) graphs, the same plan uses the Tutte skew-symmetric matrix $T(x)\in\mathbb{F}[x]^{n\times n}$ defined after fixing an arbitrary orientation of $G$ by $T_{ij}(x)=\sigma_{ij}\,x^{w'(i,j)}$ if $\{i,j\}\in E$ with $\sigma_{ij}\in\{\pm 1\}$ according to the orientation, and $T_{ij}(x)=0$ otherwise, with $T_{ji}(x)=-T_{ij}(x)$ and $T_{ii}(x)=0$. Then $\mathrm{Pf}\,T(x)$ is a polynomial whose monomials $x^{W'(M)}$ correspond to perfect matchings, and $\det T(x)=(\mathrm{Pf}\,T(x))^2$. The least exponent of a nonzero monomial in $\mathrm{Pf}\,T(x)$ equals $\min_{M\in\mathcal{M}}W'(M)$, and uniqueness of $M^\sharp$ guarantees that the lowest-degree term of $\mathrm{Pf}\,T(x)$ is a nonzero monomial $\pm x^{W'(M^\sharp)}$. Working in $\mathbb{F}[x]/(x^{K})$ with $K=\mathrm{poly}(n)$ as above, one computes $\det T(x)\bmod x^{K}$ in RNC and recovers the minimum degree by halving the least degree of a nonzero term of $\det T(x)$. Membership of an edge $e$ in $M^\sharp$ is decided by comparing the minimum degrees of $\det T(x)$ and the determinant of the matrix obtained by zeroing the two symmetric entries corresponding to $e$; the degree strictly increases if and only if $e\in M^\sharp$. All required determinants again lie over $\mathbb{F}[x]/(x^{K})$ with $K=\mathrm{poly}(n)$ and are computable in RNC.

It follows that, with probability at least $1-|E|/R$, the procedure returns a minimum $W$-weight perfect matching in RNC. The failure probability can be made inverse-polynomial by choosing $R=\Theta(n^c)$ for a fixed constant $c>1$ and repeating the outer randomization a constant number of times with independent $r$'s, then taking the lexicographically smallest result.

Finally, we explain why this approach does not yield an RNC algorithm when weights are given in binary. If weights are given in binary, their magnitudes can be exponentially large in $n$. Any determinant/Pfaffian computation above proceeds over $\mathbb{F}[x]/(x^{K})$ where $K$ must strictly exceed the largest possible perturbed matching weight $W'(M)$. When the input weights may be exponentially large, every such $K$ is necessarily exponential in $n$. Arithmetic in $\mathbb{F}[x]/(x^{K})$ then requires representing and manipulating polynomials of exponential degree, which cannot be carried out by circuits of polynomial size and polylogarithmic depth under the unit-cost arithmetic model used for NC. Equivalently, the substitution $x\mapsto y$ with $y$ an indeterminate, or with $x$ specialized to a field element to separate degrees, no longer permits detecting the minimal degree using only polynomially many bits or operations, since collisions among exponents modulo any polynomial-size field become unavoidable and truncation at degree $K$ is not feasible. Thus the correctness argument that relies on detecting the least degree within a polynomially bounded truncation breaks down, and the above RNC technique does not extend to the case of binary-encoded edge weights. \end{proof}

\subsubsection{Grok}

\subsubsection{Gemini}
\begin{proof}
Let $G = (V, E)$ be a graph with $n = |V|$ vertices and $m = |E|$ edges. Let $w: E \to \mathbb{Z}_{\geq 0}$ be the edge weights provided in unary. Since the weights are given in unary, the maximum weight $W = \max_{e \in E} w(e)$ is bounded by a polynomial in the size of the input. Consequently, the maximum weight of any perfect matching is bounded by $nW$, which is also polynomial in the input size.

We employ the algebraic approach introduced by Mulmuley, Vazirani, and Vazirani. The algorithm relies on the Isolating Lemma to ensure that, with high probability, the minimum-weight perfect matching is unique. To achieve this while respecting the original edge weights, we construct a new weight function $w': E \to \mathbb{Z}_{\geq 0}$. Let $r(e)$ be a random integer chosen uniformly and independently from the set $\{1, \dots, 2m\}$ for each edge $e \in E$. We define the new weight for an edge $e$ as $w'(e) = 2m \cdot w(e) + r(e)$.

We claim that with probability at least $1/2$, there is a unique perfect matching in $G$ that minimizes the weight function $w'$, and this matching is also a minimum-weight perfect matching with respect to the original weights $w$. Let $\mathcal{M}$ be the set of all perfect matchings in $G$. Let $\mathcal{M}_{min} \subseteq \mathcal{M}$ be the set of perfect matchings that minimize the original weight function $w$. For any $M \in \mathcal{M} \setminus \mathcal{M}_{min}$ and any $M^* \in \mathcal{M}_{min}$, we have $w(M) \geq w(M^*) + 1$. Therefore, the scaled difference is $2m \cdot w(M) - 2m \cdot w(M^*) \geq 2m$. The difference in the perturbation terms is $\sum_{e \in M} r(e) - \sum_{e \in M^*} r(e)$, which is strictly bounded in absolute value by $n \cdot 2m$? No, the bound is tighter. The perturbation sum for any matching is at most $n/2 \cdot 2m = nm$. However, a simpler dominance argument suffices. Consider the values of $w'(M)$. Since $r(e) \in [1, 2m]$, the perturbation term $\sum_{e \in M} r(e)$ lies in $[n/2, nm]$. The scaling factor $2m$ ensures that the contribution of $w(M)$ dominates. Specifically, if $w(M_1) < w(M_2)$, then $2m \cdot w(M_1) + \sum r_1 < 2m \cdot w(M_2) + \sum r_2$ because the difference in the scaled parts is at least $2m$, while the perturbation difference is strictly less than $2m$ (actually, we require the scaling factor to be larger than the maximum possible perturbation sum difference, which is bounded by $m \cdot 2m$; let us adjust the scaling factor to $2m^2$ or simply observe that the standard Isolation Lemma guarantees a unique minimum for the perturbation part).

Let us refine the weight definition to $w'(e) = w(e) \cdot (2m+1) + r(e)$ where $r(e) \in \{1, \dots, 2m\}$. Let $M_1, M_2$ be perfect matchings. If $w(M_1) < w(M_2)$, then $w(M_2) \ge w(M_1) + 1$. Then $w'(M_2) - w'(M_1) = (2m+1)(w(M_2) - w(M_1)) + (r(M_2) - r(M_1)) \ge (2m+1)(1) - 2m = 1 > 0$. Thus, minimizing $w'$ strictly prioritizes minimizing $w$. Among those matchings in $\mathcal{M}_{min}$, the weights $w'$ differ only by their perturbation sums $\sum r(e)$. By the Isolating Lemma, applied to the family of sets $\mathcal{M}_{min}$ with random weights $r(e)$, the minimum weight matching is unique with probability at least $1/2$.

Let $A$ be the Tutte matrix (skew-symmetric adjacency matrix) of $G$, defined such that for $i < j$, $A_{ij} = x_{ij}$ if $(i,j) \in E$ and $0$ otherwise, with $A_{ji} = -A_{ij}$. We substitute $x_{ij} = 2^{w'(i,j)}$. The determinant of this matrix is given by $\det(A) = \sum_{M \in \mathcal{M}} (-1)^{\text{sgn}(\sigma_M)} 2^{2w'(M)}$, where the sum effectively ranges over perfect matchings (squared). If the minimum weight perfect matching $M_0$ is unique with respect to $w'$, the term $2^{2w'(M_0)}$ has the strictly smallest exponent in the sum. Since powers of 2 cannot cancel sums of higher powers of 2, the value of the determinant will be non-zero.

To find the matching, we compute the inverse matrix $A^{-1}$ (or the adjoint). An edge $e = (i,j)$ belongs to the unique minimum weight perfect matching $M_0$ if and only if the removal of $e$ reduces the maximum weight of the matching in the appropriate algebraic sense, which corresponds to checking the entry $(A^{-1})_{ji}$. Specifically, $e \in M_0$ if and only if $\det(A_{ij}) \cdot 2^{w'(e)} / \det(A) \approx \pm 1$. In the context of powers of 2, we check if the highest power of 2 dividing the numerator and denominator corresponds to the weight contribution of the edge.

Regarding complexity, the operations involve computing the determinant and inverse of an $n \times n$ matrix. The entries of the matrix are $2^{w'(e)}$. The maximum value of $w'(e)$ is approximately $2m \cdot W + 2m$. Since $W$ is given in unary, $W \leq \text{poly}(n)$. Thus, the exponents are bounded by a polynomial in $n$. The entries themselves require $O(\text{poly}(n))$ bits to represent. Matrix determinant, adjoint, and inverse computations for matrices with entries of polynomial bit-length can be performed in $\mathbf{NC}^2$, which is a subset of $\mathbf{RNC}$.

The algorithm proceeds as follows: 1. Construct weights $w'$ using random perturbations. 2. Construct the matrix $A$ with entries $2^{w'(e)}$. 3. Compute $\det(A)$ and $\text{adj}(A)$ in parallel. 4. If $\det(A) = 0$, repeat with new random weights. 5. For each edge $e=(i,j)$, include $e$ in the matching if $\text{adj}(A)_{ji} 2^{w'(e)} / \det(A)$ is odd (identifying the unique matching properties over integers). This procedure finds the minimum weight perfect matching in polylogarithmic time using polynomial processors.
\end{proof}

\subsubsection{Claude}
\begin{proof}
We present an RNC algorithm for finding a minimum-weight perfect matching in a graph $G(V,E)$ with edge weights given in unary, and explain why the approach fails for binary-encoded weights.

\textbf{The RNC Algorithm for Unary Weights:}

Let $G = (V,E)$ be a graph with $n = |V|$ vertices and edge weights $w: E \to \mathbb{Z}_{\geq 0}$ given in unary. Let $W = \max_{e \in E} w(e)$ denote the maximum edge weight.

\textbf{Step 1 (Scaling):} Define scaled weights $w'(e) = n^3 \cdot w(e)$ for each edge $e \in E$. Note that the maximum scaled weight is $W' = n^3 W$.

\textbf{Step 2 (Random Perturbation):} For each edge $e \in E$, independently choose a random integer $r(e)$ uniformly from $\{1, 2, \ldots, 2n^4\}$. Define perturbed weights $\tilde{w}(e) = w'(e) + r(e) = n^3 w(e) + r(e)$.

\textbf{Step 3 (Find Minimum Matching):} Use the NC algorithm for finding a perfect matching (when it exists) to find a perfect matching $M$ that minimizes $\sum_{e \in M} \tilde{w}(e)$ among all perfect matchings in $G$. This can be done in NC by evaluating the Pfaffian of the skew-symmetric matrix with entries corresponding to edge weights.

\textbf{Step 4 (Output):} Return the matching $M$.

\textbf{Correctness Analysis - Modified Isolating Lemma:}

We now prove that with high probability, the minimum-weight perfect matching under the original weights $w$ is unique under the perturbed weights $\tilde{w}$.

Let $\mathcal{M}$ denote the set of all perfect matchings in $G$. For any two distinct perfect matchings $M_1, M_2 \in \mathcal{M}$, consider their weight difference under the perturbed weights:
$$\tilde{w}(M_1) - \tilde{w}(M_2) = \sum_{e \in M_1} \tilde{w}(e) - \sum_{e \in M_2} \tilde{w}(e) = \sum_{e \in M_1} (n^3 w(e) + r(e)) - \sum_{e \in M_2} (n^3 w(e) + r(e)).$$

This simplifies to:
$$\tilde{w}(M_1) - \tilde{w}(M_2) = n^3(w(M_1) - w(M_2)) + \sum_{e \in M_1} r(e) - \sum_{e \in M_2} r(e).$$

Let $M^*$ be a minimum-weight perfect matching under the original weights $w$, and let $M$ be any other perfect matching with $w(M) > w(M^*)$. Then:
$$w(M) - w(M^*) \geq 1$$
since weights are integers. Therefore:
$$\tilde{w}(M) - \tilde{w}(M^*) = n^3(w(M) - w(M^*)) + \sum_{e \in M} r(e) - \sum_{e \in M^*} r(e) \geq n^3 + \sum_{e \in M} r(e) - \sum_{e \in M^*} r(e).$$

The random part $R = \sum_{e \in M} r(e) - \sum_{e \in M^*} r(e)$ is a sum of at most $2n$ independent random variables (since $|M| = |M^*| = n/2$ and $|M \triangle M^*| \leq 2n$), each taking values that differ by multiples of 1 in the range $[-2n^4, 2n^4]$.

For $\tilde{w}(M) - \tilde{w}(M^*) \leq 0$ to hold, we would need $R \leq -n^3$. The probability of this is bounded by:
$$\Pr[R \leq -n^3] \leq \frac{2n^4}{n^3} = 2n$$
by a union bound over the possible values (this is a conservative estimate; the actual probability is much smaller).

By the union bound over all perfect matchings $M \neq M^*$ with $w(M) > w(M^*)$:
$$\Pr[\exists M \neq M^* : w(M) > w(M^*) \text{ and } \tilde{w}(M) \leq \tilde{w}(M^*)] \leq |\mathcal{M}| \cdot 2n.$$

Since $|\mathcal{M}| \leq n!$ and more tightly $|\mathcal{M}| \leq 2^{n^2}$, this probability is polynomially bounded.

Similarly, if $M_1, M_2$ are two distinct minimum-weight matchings under $w$ (so $w(M_1) = w(M_2) = w(M^*)$), then $\tilde{w}(M_1) - \tilde{w}(M_2) = \sum_{e \in M_1} r(e) - \sum_{e \in M_2} r(e)$ is a non-zero random variable, and $\Pr[\tilde{w}(M_1) = \tilde{w}(M_2)] \leq 1/(2n^4)$ by standard arguments.

Therefore, with probability at least $1 - O(n^{-1})$, there is a unique minimum-weight perfect matching under $\tilde{w}$, and this matching is also a minimum-weight matching under the original weights $w$.

\textbf{Complexity:} Since weights are given in unary, $W = \text{poly}(n)$, so the scaled and perturbed weights are also polynomially bounded. All arithmetic operations can be performed in NC, and the random bits can be generated in parallel. Therefore, the algorithm runs in RNC.

\textbf{Why the Algorithm Fails for Binary Weights:}

When edge weights are given in binary, $W$ can be exponentially large (up to $2^{n^k}$ for some constant $k$). The scaled weights $w'(e) = n^3 w(e)$ would then also be exponentially large. To represent these weights requires exponentially many bits, and arithmetic operations on such numbers cannot be performed in NC (they require polynomial time). Moreover, generating random perturbations in the required range $\{1, \ldots, 2n^4\}$ and performing additions with exponentially large numbers exceeds NC computational resources. The algorithm's representation and computation of perturbed weights would require at least polynomial sequential time, violating the requirements for an RNC algorithm.
\end{proof}

\subsection{Problem 17}

\subsubsection{Problem Statement}
\begin{theorem}[RNC Algorithm for Maximum Matching]
There exists an RNC algorithm for finding a maximum matching in a graph. Note that the Parallel Matching algorithm does not work (as stated) when the maximum matching is not a perfect matching.
\end{theorem}

\subsubsection{GPT}
\begin{proof}
We are tasked with providing an \textit{RNC} algorithm for the problem of finding a maximum matching in a graph. To achieve this, we aim to construct a randomized parallel algorithm that finds a maximum matching with high probability in polylogarithmic time, i.e., in time $O(\log^k n)$ for some constant $k$, where $n$ is the number of vertices in the graph.

 Step 1: Formal Setup

Let $G = (V, E)$ be a graph with $n$ vertices and $m$ edges. A matching in $G$ is a set of edges such that no two edges share a vertex. The goal is to find a maximum matching in $G$, which is a matching that contains the largest possible number of edges.

We begin by recalling the Parallel Matching algorithm, which is a known approach in the context of \textit{RNC} algorithms. This algorithm operates by repeatedly finding augmenting paths in parallel and then augmenting the matching along these paths. However, as noted in the problem statement, the Parallel Matching algorithm does not work when the maximum matching is not perfect.

 Step 2: Parallel Matching Algorithm

The Parallel Matching algorithm proceeds as follows:

1. **Initialization**: Start with an empty matching $M = \emptyset$.
2. **Step 1**: In parallel, for each vertex $v$, attempt to find an augmenting path that starts from $v$. An augmenting path is a path that alternates between edges not in the matching and edges in the matching, and it starts and ends with unmatched vertices.
3. **Step 2**: Once augmenting paths are found, augment the matching $M$ by flipping the matched and unmatched edges along the augmenting paths.
4. **Repeat**: Repeat this process until no more augmenting paths are found.

While the algorithm is effective when the maximum matching is perfect, it fails when the maximum matching is not perfect. The key reason for this failure is that the algorithm assumes the existence of augmenting paths from all vertices, but this assumption does not hold when the matching is imperfect. In particular, if the matching is imperfect, some vertices may be unmatched, and there may be no augmenting path for those vertices, preventing further progress.

 Step 3: Correctness of the Algorithm

To establish the correctness of our approach, we argue that the randomized parallel approach can indeed find a maximum matching in $O(\log^k n)$ time with high probability, assuming certain conditions about the graph. Specifically, we note that:

1. **RNC Complexity**: The algorithm runs in parallel with polylogarithmic time complexity because each round of the algorithm can be executed in parallel by examining the graph in logarithmic time. Each iteration involves finding augmenting paths in parallel and updating the matching. Since each iteration involves independent computations, the parallelism guarantees that the algorithm executes efficiently.

2. **Probabilistic Success**: The algorithm's success in finding the maximum matching hinges on the fact that with high probability, the augmenting paths will be found, and the matching will be augmented correctly. This probability is ensured by randomization over the graph structure and the vertices involved in each parallel step.

 Step 4: Conclusion

Thus, we conclude that there exists an \textit{RNC} algorithm for finding a maximum matching in a graph, but we also observe that the Parallel Matching algorithm fails when the maximum matching is not perfect. The failure occurs because the algorithm assumes the availability of augmenting paths from all vertices, which is not guaranteed in the case of an imperfect matching. Therefore, we conclude that while the Parallel Matching algorithm works well in certain cases, it does not provide a valid solution for all instances of the matching problem.

\end{proof}

\subsubsection{Grok}
\begin{proof}
To establish the existence of an RNC algorithm for finding a maximum cardinality matching in a general graph $G = (V, E)$ with $n = |V|$ vertices and $m = |E|$ edges, we proceed in two main steps: first, reduce the maximum matching problem to the weighted perfect matching problem; second, solve the weighted perfect matching problem using randomization and parallel algebraic techniques.

\subsubsection*{Reduction to Perfect Matching}
Construct an auxiliary graph $G'$ from $G$ by adding a set $V'$ of $n$ new vertices, one corresponding to each vertex in $V$, and adding all possible edges between $V$ and $V'$ except those corresponding to edges in $G$. Assign weight $0$ to all edges in $E(G)$ and weight $1$ to all new edges. The graph $G'$ is bipartite with parts $V$ and $V'$, and it admits a perfect matching. Moreover, any minimum-weight perfect matching in $G'$ restricts to a maximum cardinality matching in $G$, since edges of weight $0$ are preferred, and the added edges of weight $1$ pair unmatched vertices without increasing cardinality beyond the maximum possible in $G$. This reduction is computable in NC.

It thus suffices to solve the minimum-weight perfect matching problem in $G'$, which we treat as a general graph $H = (U, F)$ with $|U| = 2n$ even, admitting a perfect matching, and edge weights $w_e \in \{0,1\}$ initially.

\subsubsection*{Randomized Isolation and Algebraic Solution}
To find a minimum-weight perfect matching in $H$, assign to each edge $e \in F$ an independent random integer weight $r_e$ chosen uniformly from $\{1, 2, \dots, 2m\}$. Define the total weight of a perfect matching $M \subseteq F$ by $W(M) = \sum_{e \in M} (w_e + r_e)$. Let $\mathcal{M}$ denote the family of all perfect matchings in $H$. By the following isolating lemma, with probability at least $1/2$, there exists a unique minimum-weight perfect matching $M^* = \arg\min_{M \in \mathcal{M}} W(M)$.

\begin{lemma}[Isolating Lemma]
Let $(S, \mathcal{F})$ be a set system with ground set $S$ of size at most $m$ and family $\mathcal{F} \subseteq 2^S$ of size at most $2^n$. Assign to each element $s \in S$ an independent random integer weight $r_s$ uniform from $\{1, 2, \dots, 2m\}$. Let $W(T) = \sum_{s \in T} r_s$ for $T \subseteq S$. Then, $\Pr[\exists! \, T^* \in \mathcal{F} \text{ s.t. } \forall T \in \mathcal{F}, W(T^*) \le W(T)] \ge 1/2$.
\end{lemma}

The proof of the lemma proceeds by union bound: for each nonempty $\Delta \subseteq \mathcal{F}$, the probability that two distinct matchings in $\Delta$ tie for minimum weight is at most $1/2^{|\Delta|-1} \le 1/2^{n-1}$, and there are at most $2^n$ such $\Delta$, yielding total failure probability at most $1/2$. (Here, $S = F$ and weights add the fixed $w_e$ without affecting uniqueness.)

Assuming $M^*$ is unique (which holds w.h.p.), we recover $M^*$ using the Tutte matrix. The Tutte matrix $T$ of $H$ is the $2n \times 2n$ skew-symmetric matrix over indeterminates $\{x_{ij} \mid i < j, \{i,j\} \in F\}$ defined by $T_{ij} = x_{ij}$ if $i < j$ and $\{i,j\} \in F$, $T_{ij} = -x_{ji}$ if $j < i$ and $\{i,j\} \in F$, and $T_{ij} = 0$ otherwise. By the Pfaffian orientation theorem (Kasteleyn), $\det(T) = \sum_{M \in \mathcal{M}} \operatorname{sgn}(M) \prod_{e=\{i,j\} \in M} x_{ij}$, where $\operatorname{sgn}(M) = \pm 1$ depending on the signing.

Substitute $x_{ij} \mapsto 2^{w_{ij} + r_{ij}}$ for each edge $\{i,j\} \in F$, yielding an integer matrix $B$ with entries of bit length $O(m \log m)$. Let $d = \det(B)$. By properties of the Pfaffian, $|d| = \sum_{M \in \mathcal{M}} \prod_{e \in M} 2^{W(e)}$, and the minimum term $2^{W(M^*)}$ dominates uniquely: the $2$-adic valuation $v_2(|d|) = W(M^*)$, and $|d| / 2^{W(M^*)}$ is odd.

Moreover, the $(i,j)$-cofactor of $B$, denoted $B^{ij} = (-1)^{i+j} \det(B[i|j])$, satisfies: the edge $\{i,j\}$ is in $M^*$ if and only if $v_2(|B^{ij}|) = W(M^*) - (w_{ij} + r_{ij})$ and $|B^{ij}| / 2^{v_2(|B^{ij}|)}$ is odd (ensuring it contributes uniquely to the minimum). Equivalently, since $v_2(|B^{ij}|) = W(M^*) - (w_{ij} + r_{ij})$ for edges in $M^*$, we check if $|B^{ij}| / 2^{W(M^*)}$ is odd.

To compute in parallel: use Pan's RNC$^2$ algorithm to compute $B^{-1}$ (hence $\det(B)$ and $\operatorname{adj}(B)$) in $O(\log^2 (2n))$ time using $O((2n)^{O(1)} m)$ processors, since entries are $O(m \log m)$-bit. Extract $w = v_2(|d|)$ in NC by repeated division by $2$. Then, in parallel for each possible edge $\{i,j\} \in F$ (at most $m$ checks), compute the $(i,j)$-entry of $\operatorname{adj}(B)$, divide by $2^w$, and test parity (oddness) in NC. The edges where this holds form $M^*$.

\subsubsection*{Complexity and Error Amplification}
All steps except matrix inversion are in NC (weight generation is trivial; checks are constant-time PRAM). Matrix inversion dominates, placing the algorithm in RNC$^2$. The success probability is at least $1/2$; repeat $O(\log n)$ independent trials and output the first valid matching (verified by checking cardinality or weight), reducing error to $n^{-O(1)}$ while preserving RNC$^2$. If no trial succeeds (low probability), output failure and retry.

Thus, with high probability, the algorithm outputs a minimum-weight perfect matching in $H$, which restricts to a maximum cardinality matching in $G$. This completes the proof.
\end{proof}

\subsubsection{Gemini}
\begin{proof}
Let $G = (V, E)$ be a graph with $|V| = n$. We construct an RNC algorithm to find a maximum matching in $G$ by reducing the problem to finding a maximum weight perfect matching in a complete graph, which can be solved using the Mulmuley-Vazirani-Vazirani (MVV) algorithm.

\textbf{Construction of the Auxiliary Graph:}
If $n$ is odd, we add a dummy vertex $z$ to $V$ to create a set $V'$ with $n+1$ vertices; otherwise, let $V' = V$. Let $N = |V'|$, which is even. We construct a complete graph $G' = (V', E_{K_N})$ on the vertex set $V'$. Note that $E \subseteq E_{K_N}$ (identifying edges in $G$ with the corresponding edges in $G'$).

\textbf{Weight Assignment:}
We assign a weight $w(e)$ to each edge $e \in E_{K_N}$ to satisfy two properties: (1) the maximum weight perfect matching in $G'$ induces a maximum cardinality matching in $G$, and (2) the maximum weight perfect matching is unique with high probability.
Let $C = N^2$. For each edge $e \in E_{K_N}$, we assign the weight:
\[
w(e) = \begin{cases} 
C + r_e & \text{if } e \in E \\
r_e & \text{if } e \in E_{K_N} \setminus E
\end{cases}
\]
where each $r_e$ is chosen uniformly and independently at random from the set $\{1, \dots, 2|E_{K_N}|\}$.

\textbf{Correctness of the Reduction:}
Let $\mathcal{M}$ be the set of all perfect matchings in $G'$. Since $G'$ is a complete graph with an even number of vertices, $\mathcal{M}$ is non-empty. For any perfect matching $M' \in \mathcal{M}$, let $k = |M' \cap E|$ be the number of edges in $M'$ that belong to the original graph $G$. The total weight of $M'$ is:
\[
W(M') = \sum_{e \in M'} w(e) = \sum_{e \in M' \cap E} (C + r_e) + \sum_{e \in M' \setminus E} r_e = kC + \sum_{e \in M'} r_e.
\]
The term $\sum_{e \in M'} r_e$ is bounded by $(N/2) \cdot 2|E_{K_N}| \approx N^3$. However, strictly speaking, we choose $C$ sufficiently large (e.g., $C = 4N^2$) such that $C > \sum_{e \in M'} r_e$ for any matching. Thus, the term $kC$ dominates the weight. A perfect matching $M'$ has maximum weight if and only if it maximizes $k = |M' \cap E|$.
Since any matching in $G$ can be extended to a perfect matching in the complete graph $G'$, a maximum weight perfect matching in $G'$ restricted to $E$ yields a maximum cardinality matching in $G$.

\textbf{Uniqueness via Isolation Lemma:}
The Isolation Lemma states that for a set system with random weights chosen from a sufficiently large range, the minimum (or maximum) weight set is unique with probability at least $1/2$. By our choice of $r_e$, the maximum weight perfect matching $M^*_{PM}$ in $G'$ is unique with high probability.

\textbf{Algorithm and Complexity:}
We apply the algebraic Parallel Matching algorithm (MVV) to $G'$ with weights $w(e)$.
1. Construct the Tutte matrix $A$ for $G'$, defined as an $N \times N$ skew-symmetric matrix where for $i < j$:
\[
A_{ij} = \begin{cases} 
2^{w(i,j)} & \text{if } (i,j) \in E_{K_N} \\
0 & \text{otherwise}
\end{cases}
\]
and $A_{ji} = -A_{ij}$.
2. Compute the determinant $\det(A)$ and the inverse matrix $A^{-1}$ (or the adjoint). Since the weights are polynomially bounded in $N$, the entries of $A$ have polynomial bit-length. Matrix determinant and inversion are in $NC^2$.
3. If $M^*_{PM}$ is the unique maximum weight perfect matching, an edge $(i,j)$ belongs to $M^*_{PM}$ if and only if $\det(A) \neq 0$ and the entry $(A^{-1})_{ji} 2^{w(i,j)}$ satisfies specific arithmetic properties (specifically, relating to the coefficient of the dominant term in the Pfaffian). In the context of the MVV algorithm, we can identify the edges of the unique perfect matching in $RNC^2$.

\textbf{Conclusion:}
Let $M_{out} = M^*_{PM} \cap E$. As shown above, $M_{out}$ is a maximum matching in $G$. The entire procedure—weight assignment, matrix construction, inversion, and edge extraction—runs in polylogarithmic time using a polynomial number of processors. Thus, the problem is in RNC.
\end{proof}

\subsubsection{Claude}
\begin{proof}
We present an RNC algorithm for finding a maximum matching in a graph $G = (V, E)$ by reducing the problem to the case of a perfect matching in an auxiliary graph.

Let $G = (V, E)$ be an arbitrary graph with $n$ vertices. We construct an algorithm that finds a maximum matching in $G$ using $O(\log^2 n)$ time with polynomially many processors.

\textbf{Step 1: Construction of Auxiliary Graph.}
We first construct an auxiliary graph $G' = (V', E')$ as follows. Let $V' = V \cup V''$, where $V''$ is a disjoint copy of $V$. For each edge $(u, v) \in E$, we add the edge $(u, v)$ to $E'$. Additionally, for each vertex $v \in V$, we add the edge $(v, v')$ to $E'$, where $v'$ is the copy of $v$ in $V''$. Finally, we add a perfect matching on $V''$, connecting vertices in $V''$ arbitrarily in pairs.

Observe that $|V'| = 2n$ is even. Moreover, by construction, every maximum matching in $G$ can be extended to a perfect matching in $G'$ by including the edges from $V$ to $V''$ and the perfect matching within $V''$.

\textbf{Step 2: Reduction to Perfect Matching.}
The key observation is that $G$ has a matching of size $k$ if and only if $G'$ has a perfect matching after removing $n - k$ vertices from $V$ and their corresponding copies in $V''$ along with their incident edges.

We use a binary search approach to determine the size of the maximum matching. For each candidate size $k \in \{0, 1, \ldots, n\}$, we test whether $G$ has a matching of size at least $k$ by using the Isolating Lemma technique in parallel.

\textbf{Step 3: Randomized Isolation.}
We employ the Isolating Lemma of Mulmuley, Vazirani, and Vazirani. Assign each edge $e \in E$ a random weight $w(e)$ chosen uniformly and independently from $\{1, 2, \ldots, 2m\}$, where $m = |E|$. By the Isolating Lemma, with probability at least $1/2$, there exists a unique minimum-weight maximum matching in $G$.

\textbf{Step 4: Parallel Computation.}
Using the random weights, we construct a weighted version of the auxiliary graph $G'$. We can now test for the existence of a perfect matching of a given weight in $G'$ using the determinant of the Tutte matrix. Specifically, we construct the skew-symmetric matrix $A$ where $A_{ij} = x^{w(i,j)}$ if $(i, j) \in E'$ with $i < j$, $A_{ij} = -x^{w(i,j)}$ if $(i, j) \in E'$ with $i > j$, and $A_{ij} = 0$ otherwise.

The determinant $\det(A)$ is a polynomial in $x$ whose lowest-degree nonzero term corresponds to the minimum-weight perfect matching in $G'$. By the Isolating Lemma, this term is unique with high probability. We can compute $\det(A)$ in NC using Csanky's algorithm or other parallel determinant algorithms, which requires $O(\log^2 n)$ time and polynomially many processors.

\textbf{Step 5: Extracting the Matching.}
Once we have isolated a unique minimum-weight maximum matching, we can extract it in parallel by repeatedly finding edges that must be in the matching. For each edge $e$, we test whether removing $e$ from $G$ decreases the size of the maximum matching by computing the determinant of the modified Tutte matrix. This can be done in parallel for all edges, and we include those edges whose removal decreases the matching size.

\textbf{Step 6: Derandomization.}
To convert this Monte Carlo algorithm to an RNC algorithm, we repeat the randomized procedure $O(\log n)$ times with independent random weight assignments. With high probability (at least $1 - 1/n^c$ for any constant $c$), at least one iteration successfully isolates a unique maximum matching. We verify the correctness of each candidate matching deterministically in NC and output the largest matching found.

By the union bound, the probability of failure after $O(\log n)$ independent trials is at most $(1/2)^{O(\log n)} = 1/\text{poly}(n)$, which is negligible.

\textbf{Complexity Analysis.}
Each iteration requires $O(\log^2 n)$ time for matrix operations with polynomially many processors. With $O(\log n)$ iterations, the total time remains $O(\log^2 n)$ with polynomially many processors, establishing that maximum matching is in RNC.

Therefore, we have constructed an RNC algorithm for finding a maximum matching in an arbitrary graph $G$.
\end{proof}

\subsection{Problem 18}

\subsubsection{Problem Statement}
\begin{problem}[12.18]
Suppose you are given a Monte Carlo RNC algorithm for finding a maximum matching in a bipartite graph. Explain how you would convert this into a Las Vegas algorithm. Can the solution be generalized to the case of non-bipartite graphs? (Hint: While this conversion is trivial for perfect matching algorithms, for maximum matching algorithms you will need to devise a parallel algorithm for determining an upper bound on the size of a maximum matching in a graph. This requires a non-trivial use of structure theorems for matchings in graphs.)
\end{problem}

\subsubsection{GPT}
\begin{proof}
We are given a bipartite graph $G = (V_1 \cup V_2, E)$ and a Monte Carlo RNC (Randomized NC) algorithm for finding a maximum matching in $G$. Our goal is to demonstrate that this Monte Carlo algorithm can be converted into a Las Vegas algorithm.

 Step 1: Monte Carlo Algorithm for Maximum Matching in Bipartite Graphs

By definition, a Monte Carlo algorithm for a problem is one that may return an incorrect result with some probability, but the probability of error can be made arbitrarily small by running the algorithm multiple times. Specifically, in the case of maximum matching in bipartite graphs, we assume that the Monte Carlo algorithm finds a matching $M$ in $G$, where $|M|$ is the size of the matching, and the algorithm runs in parallel with polylogarithmic time complexity.

Since $G$ is bipartite, the problem of finding a maximum matching can be solved by using a variety of randomized approaches that rely on the properties of bipartite graphs and their matchings. Let the Monte Carlo algorithm output a matching $M$ with probability of correctness $1 - \epsilon$, where $\epsilon$ is a small positive constant. 

 Step 2: Conversion to Las Vegas Algorithm

To convert the Monte Carlo algorithm into a Las Vegas algorithm, we modify the algorithm as follows:
- Instead of running the Monte Carlo algorithm once, we run it repeatedly until it produces a correct result. 
- Specifically, we run the Monte Carlo algorithm $k$ times, where $k$ is chosen such that the probability of failure (i.e., the algorithm not producing a correct result in all $k$ runs) is less than some desired error tolerance $\delta$.

Let $p$ denote the probability that the Monte Carlo algorithm produces an incorrect result in a single run. Thus, the probability of success in a single run is $1 - p$. After $k$ independent runs, the probability that all $k$ runs are incorrect is given by:

\[
p^k < \delta
\]

Taking the logarithm of both sides yields:

\[
k \log p < \log \delta
\]

Since $p$ is small, this implies that $k$ can be made large enough to make the error probability as small as desired. Therefore, by repeating the Monte Carlo algorithm a sufficient number of times, we can guarantee that the algorithm produces the correct result with probability at least $1 - \delta$, which is the definition of a Las Vegas algorithm.

 Step 3: Handling Non-Bipartite Graphs

For non-bipartite graphs, the situation becomes more complex. The conversion from Monte Carlo to Las Vegas for perfect matching algorithms remains straightforward, as perfect matchings can be determined by randomized algorithms with an inherent correctness guarantee. However, for maximum matching in general graphs, we face additional challenges in determining an upper bound on the size of the maximum matching in $G$.

To handle this, we propose a parallel algorithm that determines an upper bound on the size of the maximum matching in $G$. This parallel algorithm makes use of advanced structure theorems for matchings in graphs, which provide insights into the structure of matchings in general graphs and allow for efficient computation of upper bounds. Specifically, these theorems exploit the combinatorial properties of matchings and the underlying graph structure to bound the size of a maximum matching.

Once an upper bound on the maximum matching is obtained, the Monte Carlo algorithm can be modified to run until it either finds a matching of that size or produces an incorrect result. Repeating this process $k$ times, as described for the bipartite case, ensures the correctness of the algorithm and converts it into a Las Vegas algorithm.

 Conclusion

Thus, we have shown that the Monte Carlo RNC algorithm for finding a maximum matching in a bipartite graph can be converted into a Las Vegas algorithm. The conversion for non-bipartite graphs requires devising a parallel algorithm for determining an upper bound on the size of a maximum matching, which involves the non-trivial application of structure theorems for matchings in graphs.
\end{proof}

\subsubsection{Grok}

\subsubsection{Gemini}
\begin{proof}
Let $G = (V_1 \cup V_2, E)$ be a bipartite graph with $|V| = n$. Let $\mathcal{A}$ be a Monte Carlo RNC algorithm that outputs a matching $M$ in $G$. By definition, a Monte Carlo algorithm runs in polylogarithmic time using a polynomial number of processors and outputs a maximum matching with probability at least $1/2$. To establish the existence of a Las Vegas RNC algorithm, we construct an algorithm $\mathcal{L}$ that always outputs a maximum matching, with an expected running time in the class RNC. The core of the construction relies on the ability to efficiently verify the optimality of a candidate matching $M$.

We first address the bipartite case. By Berge's Lemma, a matching $M$ is a maximum matching if and only if there exists no augmenting path with respect to $M$. In the context of bipartite graphs, the search for an augmenting path can be reduced to a reachability problem in a directed graph. Given a matching $M$, we construct a directed graph $D_M = (V, E')$ where edges are oriented as follows: for every $\{u, v\} \in E$ with $u \in V_1$ and $v \in V_2$, if $\{u, v\} \in M$, we add the directed edge $(v, u)$ to $E'$; if $\{u, v\} \notin M$, we add the directed edge $(u, v)$ to $E'$.

Let $U_1 \subseteq V_1$ and $U_2 \subseteq V_2$ be the sets of vertices exposed (unmatched) by $M$. An augmenting path in $G$ corresponds to a directed path in $D_M$ starting at a vertex in $U_1$ and ending at a vertex in $U_2$. The existence of such a path can be determined by computing the transitive closure of $D_M$ or by performing parallel breadth-first search. Since graph reachability is in deterministic $NC^2$, we can verify whether $M$ admits an augmenting path in parallel polylogarithmic time.

The Las Vegas algorithm $\mathcal{L}$ proceeds as follows: Run the Monte Carlo algorithm $\mathcal{A}$ to obtain a candidate matching $M$. Construct $D_M$ and compute the set of vertices $R \subseteq V$ reachable from $U_1$. If $R \cap U_2 \neq \emptyset$, an augmenting path exists, implying $M$ is not maximum; in this case, $\mathcal{L}$ restarts $\mathcal{A}$. If $R \cap U_2 = \emptyset$, then $M$ is a maximum matching, and $\mathcal{L}$ terminates and outputs $M$. Since $\mathcal{A}$ succeeds with constant probability $p > 0$, the expected number of iterations is $1/p$, which is constant. Thus, $\mathcal{L}$ is a Las Vegas RNC algorithm.

We now generalize this result to non-bipartite graphs. The simple reachability argument fails because the absence of augmenting paths is necessary but insufficient for efficient verification without handling blossoms (odd cycles). We invoke the Tutte-Berge formula, which states that for any graph $G=(V, E)$, the size of the maximum matching $\nu(G)$ satisfies $\nu(G) = \frac{1}{2} \min_{U \subseteq V} (|V| + |U| - c_{\text{odd}}(G - U))$, where $c_{\text{odd}}(H)$ denotes the number of connected components in $H$ with an odd number of vertices. A set $U$ achieving this minimum serves as a certificate of optimality.

To implement the verification in RNC, we utilize the Gallai-Edmonds structure theorem. The vertex set $V$ can be partitioned into three sets: $D(G)$, the set of vertices not covered by at least one maximum matching; $A(G)$, the neighbors of $D(G)$ in $V \setminus D(G)$; and $C(G) = V \setminus (D(G) \cup A(G))$. The set $U = A(G)$ satisfies the Tutte-Berge equality. To determine these sets, we observe that $v \in D(G)$ if and only if $\nu(G \setminus \{v\}) = \nu(G)$, and $v \in A(G) \cup C(G)$ if and only if $\nu(G \setminus \{v\}) = \nu(G) - 1$.

The generalized Las Vegas algorithm $\mathcal{L}'$ runs the Monte Carlo algorithm $\mathcal{A}$ on $G$ to find a candidate matching $M$. In parallel, it runs $\mathcal{A}$ on $G \setminus \{v\}$ for all $v \in V$ to estimate the Gallai-Edmonds partition. Let $\tilde{U}$ be the candidate set for $A(G)$ derived from these executions. We then deterministically compute $k = \frac{1}{2}(|V| + |\tilde{U}| - c_{\text{odd}}(G - \tilde{U}))$ using parallel connectivity algorithms (which are in $NC$). If $|M| = k$, then $M$ is maximum and $\tilde{U}$ is a valid certificate; the algorithm outputs $M$. Otherwise, it repeats. Since the Monte Carlo algorithm determines the matching sizes correctly with high probability, the correct set $A(G)$ and a maximum matching $M$ will be found and verified in expected polylogarithmic time.
\end{proof}

\subsubsection{Claude}

\subsection{Problem 19}

\subsubsection{Problem Statement}
\begin{theorem}
Let $G = (V_1 \cup V_2, E)$ be a bipartite graph, and let $M$ be the matrix of indeterminates constructed for $G$ as in Problem 7.7. Consider the following randomized algorithm for determining the size of a maximum matching in $G$: replace each indeterminate in $M$ by a random value to obtain a matrix $M'$, and compute $\text{rank}(M')$. Then this method does not necessarily preserve the rank with high probability, and therefore does not provide a reliable Las Vegas algorithm for computing the maximum matching.
\end{theorem}

\subsubsection{GPT}
\begin{proof}
Let $G = (V, E)$ be a bipartite graph, and let $M$ be a maximum matching in $G$. Denote the two partitions of $G$ as $V_1$ and $V_2$, where $V_1 \cup V_2 = V$ and $E \subseteq V_1 \times V_2$. Suppose $M \subseteq E$ is a matching, i.e., a set of edges such that no two edges share a common vertex, and let $|M|$ be the size of the matching.

Consider the matrix of indeterminates associated with the graph $G$. Let $A_G$ be the $|V_1| \times |V_2|$ matrix where each entry corresponds to an indeterminate associated with an edge in $E$. More specifically, for each edge $(u, v) \in E$, the corresponding entry in the matrix is an indeterminate $x_{uv}$.

The algorithm proceeds by replacing each indeterminate $x_{uv}$ with a random value, typically chosen independently from some distribution (e.g., uniform distribution over some range). After the substitution of the indeterminates, the algorithm computes the rank of the resulting matrix.

For the algorithm to successfully verify that $M$ is a maximum matching, the rank of the resulting matrix must provide information about the matching structure. Specifically, the rank of the matrix must reflect whether the matching is indeed a maximum matching. However, the following argument shows that this method fails to verify the maximum matching with high probability.

First, observe that the rank of a matrix constructed in this way does not directly encode the cardinality of the matching. Instead, it provides information about the linear dependencies between the rows and columns of the matrix, which correspond to the vertices in $V_1$ and $V_2$, respectively. The rank of the matrix can be large even if $M$ is not a maximum matching, and conversely, it can be small even if $M$ is a maximum matching. Thus, the rank of the matrix does not provide a reliable test for the size of the matching.

Furthermore, the randomness introduced by replacing the indeterminates with random values means that the rank computation is subject to probabilistic fluctuations. While the expected rank may reflect some aspects of the matching, the randomness does not guarantee that the rank will be consistent with the structure of a maximum matching. In fact, with high probability, the computed rank will not correspond to the correct size of the maximum matching, and the algorithm will fail to verify that $M$ is a maximum matching.

To formalize this failure, consider the following. Let $\text{rank}(A_G)$ be the rank of the matrix $A_G$ after substituting the indeterminates with random values. The rank of $A_G$ is a random variable depending on the chosen values for the indeterminates. The expected rank may not match the rank of a matrix corresponding to a maximum matching, and for any specific realization of the random values, there is a high probability that the rank will not be consistent with the rank of the matrix for a maximum matching.

Thus, the algorithm does not succeed in verifying that $M$ is a maximum matching with high probability. Consequently, this method cannot serve as a Las Vegas algorithm for finding a maximum matching, as it fails to provide a correct and deterministic verification of $M$ being maximum in the graph.

\end{proof}

\subsubsection{Grok}
\begin{proof}
To demonstrate that the randomized method does not necessarily preserve the rank with high probability, we provide a counterexample bipartite graph $G$ for which the probability that $\rank(M') = \nu(G)$ is bounded away from $1$.

Consider the complete bipartite graph $K_{2,2}$ with bipartition $V_1 = \{u_1, u_2\}$ and $V_2 = \{v_1, v_2\}$, and edge set consisting of all four possible edges. The size of a maximum matching is $\nu(G) = 2$. Let the indeterminates corresponding to the edges $u_1v_1$, $u_1v_2$, $u_2v_1$, and $u_2v_2$ be $x$, $y$, $z$, and $w$, respectively. The associated matrix is
\[
M = \begin{pmatrix} x & y \\ z & w \end{pmatrix}.
\]
The generic rank of $M$ over the field of rational functions in $x,y,z,w$ is $2$, since the determinant $\det M = xw - yz$ is not identically zero.

Now replace each indeterminate by an independent uniform random variable from the set $\{0,1\}$ to obtain $M'$, and compute $\rank(M')$ over $\mathbb{Q}$. The rank is strictly less than $2$ if and only if $\det M' = x'w' - y'z' = 0$, where $x',y',z',w' \in \{0,1\}$ are the realized values. Since each product $x'w'$ and $y'z'$ takes the value $1$ with probability $\frac{1}{2} \times \frac{1}{2} = \frac{1}{4}$ and $0$ with probability $\frac{3}{4}$, and the products are independent, we have
\[
\Pr(\det M' = 0) = \Pr(x'w' = y'z' = 0) + \Pr(x'w' = y'z' = 1) = \left(\frac{3}{4}\right)^2 + \left(\frac{1}{4}\right)^2 = \frac{9}{16} + \frac{1}{16} = \frac{10}{16} = \frac{5}{8}.
\]
Thus,
\[
\Pr(\rank(M') = 2) = 1 - \frac{5}{8} = \frac{3}{8}.
\]
The probability $\frac{3}{8}$ is bounded away from $1$, so the rank is not preserved with high probability for this graph. Consequently, the algorithm can output a value strictly less than $\nu(G)$ with probability $\frac{5}{8}$ and therefore does not provide a reliable Las Vegas algorithm for computing the maximum matching size.
\end{proof}

\subsubsection{Gemini}
\begin{proof}
Let $G = (U, V, E)$ be a bipartite graph with $|U| = |V| = n$. Let $\nu(G)$ denote the size of the maximum matching in $G$. The matrix $M$ of indeterminates, known as the Edmonds matrix, is an $n \times n$ matrix defined by $M_{ij} = x_{ij}$ if $(u_i, v_j) \in E$ and $M_{ij} = 0$ otherwise, where $\{x_{ij}\}$ is a set of distinct indeterminates.

A fundamental result in algebraic graph theory states that the rank of the symbolic matrix $M$ over the field of rational functions $\mathbb{F}(x_{11}, \dots)$ is equal to the size of the maximum matching in $G$. That is,
\[
\text{rank}(M) = \nu(G).
\]
The algorithm described proceeds by substituting each indeterminate $x_{ij}$ with a value $r_{ij}$ chosen uniformly at random from a finite subset $S$ of a field $\mathbb{F}$, resulting in a numerical matrix $M'$. The algorithm then outputs $k' = \text{rank}(M')$.

To determine the reliability of this algorithm, we analyze the probability that the output $k'$ is correct, i.e., $P(k' = \nu(G))$. Let $k = \nu(G)$. Since $\text{rank}(M) = k$, there exists at least one $k \times k$ submatrix of $M$ with a non-zero determinant. Let $Q(x)$ be the determinant of such a submatrix. $Q(x)$ is a non-zero polynomial in the variables $\{x_{ij}\}$ with total degree at most $k$.

By the Schwartz-Zippel Lemma, the probability that this polynomial evaluates to zero upon random substitution from $S$ is bounded by:
\[
P(Q(r_{ij}) = 0) \leq \frac{k}{|S|}.
\]
Conversely, the probability that the rank is preserved is:
\[
P(\text{rank}(M') = k) \geq P(Q(r_{ij}) \neq 0) \geq 1 - \frac{k}{|S|}.
\]
While this probability can be made close to 1 by choosing a sufficiently large set $S$, it is strictly less than 1 for any finite $S$. Specifically, there is a non-zero probability
\[
P(\text{rank}(M') < \text{rank}(M)) > 0,
\]
since it is possible to choose values $r_{ij}$ such that the determinant of every $k \times k$ submatrix vanishes (e.g., the zero assignment if $0 \in S$, or assignments leading to algebraic cancellation). Consequently, the algorithm may output a value $k' < \nu(G)$, which is an incorrect result.

A Las Vegas algorithm is defined as a randomized algorithm that always outputs the correct answer or a symbol indicating failure; it never outputs an incorrect answer. The algorithm in question outputs the value $\text{rank}(M')$ as the size of the maximum matching. Since we have established that $P(\text{rank}(M') \neq \nu(G)) > 0$, the algorithm has a non-zero probability of outputting an incorrect answer.

Therefore, the random substitution method constitutes a Monte Carlo algorithm with one-sided error, not a Las Vegas algorithm. It does not preserve the rank with certainty (probability 1), which is the requisite condition to serve as a reliable Las Vegas algorithm for computing the maximum matching size.
\end{proof}

\subsubsection{Claude}
\begin{proof}
We prove that the randomized algorithm described does not necessarily preserve the rank of the structured matrix $M$ with high probability, and therefore fails to provide a reliable method for computing maximum matching size.

Let $G = (V_1 \cup V_2, E)$ be a bipartite graph where $V_1 = \{u_1, \ldots, u_n\}$ and $V_2 = \{v_1, \ldots, v_n\}$. Following the standard construction from Problem 7.7, we construct an $n \times n$ matrix $M$ of indeterminates where $M_{ij} = x_{ij}$ if $(u_i, v_j) \in E$ and $M_{ij} = 0$ otherwise.

The key theoretical result underlying the intended approach is the Tutte matrix theorem: over a field of characteristic zero or sufficiently large characteristic, $\text{rank}(M) = 2\nu(G)$ where $\nu(G)$ denotes the size of a maximum matching in $G$. More precisely, when the indeterminates are treated as algebraically independent elements, the rank of $M$ equals twice the maximum matching size.

However, we now demonstrate that random substitution of the indeterminates does not preserve this rank with high probability. Consider the following counterexample:

Let $G$ be a bipartite graph with $V_1 = \{u_1, u_2\}$ and $V_2 = \{v_1, v_2\}$, where $E = \{(u_1, v_1), (u_1, v_2), (u_2, v_1), (u_2, v_2)\}$. This is the complete bipartite graph $K_{2,2}$, which has maximum matching size $\nu(G) = 2$.

The matrix of indeterminates is
\[
M = \begin{pmatrix} x_{11} & x_{12} \\ x_{21} & x_{22} \end{pmatrix}.
\]

The determinant of $M$ is $\det(M) = x_{11}x_{22} - x_{12}x_{21}$, which is a non-zero polynomial in the indeterminates. Therefore, $\text{rank}(M) = 2$ when considered over the field of rational functions in the indeterminates.

Now suppose we perform random substitution by replacing each $x_{ij}$ with a random value $r_{ij}$ chosen uniformly from a finite field $\mathbb{F}_q$ or from a bounded interval in $\mathbb{R}$. We obtain
\[
M' = \begin{pmatrix} r_{11} & r_{12} \\ r_{21} & r_{22} \end{pmatrix}.
\]

The matrix $M'$ has rank less than $2$ if and only if $\det(M') = r_{11}r_{22} - r_{12}r_{21} = 0$. This occurs when $r_{11}r_{22} = r_{12}r_{21}$.

If we choose values from a small finite field, the probability that this equation holds is non-negligible. Specifically, if all values are chosen uniformly and independently from $\mathbb{F}_q$, then for fixed choices of $r_{11}, r_{12}, r_{21}$ with $r_{12}r_{21} \neq 0$, there exists exactly one value of $r_{22}$ (namely $r_{22} = r_{12}r_{21}/r_{11}$ when $r_{11} \neq 0$) that makes the determinant zero. The probability that the determinant vanishes is at least $\Omega(1/q)$.

More generally, for any bipartite graph with maximum matching size $k$, the rank of the indeterminate matrix $M$ equals $2k$. However, when we substitute random values, the resulting matrix $M'$ may have rank strictly less than $2k$ with probability bounded away from zero. This occurs because the determinants of all $(2k) \times (2k)$ minors must simultaneously be non-zero for the rank to be preserved, and random substitution can cause these determinants to vanish.

Furthermore, even if we choose values from a large field or range, there is no guarantee that the rank is preserved with arbitrarily high probability unless the field size or range grows with the problem size in a specific way. Without careful analysis of the polynomial structure and appropriate choice of randomization parameters (as in the Schwartz-Zippel lemma), the method does not provide high-probability guarantees.

Therefore, the randomized algorithm as described, which simply replaces indeterminates with random values without specification of the probability distribution or field size, does not necessarily preserve the rank with high probability and consequently does not provide a reliable Las Vegas algorithm for computing maximum matching size.
\end{proof}

\subsection{Problem 20}

\subsubsection{Problem Statement}
\begin{theorem}[Karp, Upfal, and Wigderson]
Let $G = (U, V, E)$ be a bipartite graph. For any set $F \subseteq E$, define the \emph{rank} $r(F)$ as the maximum size of intersection of $F$ with a perfect matching, that is, $r(F)$ is the largest number of edges in $F$ that appear together in some perfect matching. Then there exists an $\mathit{RNC}$ algorithm for computing the rank for any given set $F$. This can be generalized to non-bipartite graphs.
\end{theorem}

\subsubsection{GPT}
\begin{proof}
Let $G = (U, V, E)$ be a bipartite graph, where $U$ and $V$ are disjoint sets of vertices and $E \subseteq U \times V$ is the set of edges. Let $F \subseteq E$ be a set of edges, and let $r(F)$ denote the rank of $F$, which is defined as the maximum size of the intersection of $F$ with a perfect matching in $G$.

 Step 1: Define a perfect matching
A perfect matching in a bipartite graph $G = (U, V, E)$ is a matching $M \subseteq E$ such that every vertex in $U$ and $V$ is incident to exactly one edge in $M$. The set of all perfect matchings in $G$ is denoted by $\mathcal{M}$.

 Step 2: Rank of a set $F$
The rank $r(F)$ of a set of edges $F \subseteq E$ is defined as the maximum size of the intersection of $F$ with some perfect matching, i.e.,

\[
r(F) = \max_{M \in \mathcal{M}} |F \cap M|.
\]

Thus, the rank of $F$ is the largest number of edges from $F$ that appear together in some perfect matching $M$.

 Step 3: The problem is in $\mathcal{RNC}$ 
We aim to show that there exists an algorithm that computes the rank of any set $F$ in $\mathcal{RNC}$, which is a class of parallel algorithms that can be computed in polylogarithmic time with high probability, using a polynomial number of processors.

First, we observe that the rank $r(F)$ depends on the structure of the perfect matchings in $G$. To compute $r(F)$, we need to efficiently search through all perfect matchings of $G$ and determine the intersection of each matching with $F$. However, brute-force enumeration of all perfect matchings is computationally expensive. Thus, we look for a more efficient approach that leverages parallel computation.

 Step 4: Parallel computation of rank
Consider the following parallel approach to compute $r(F)$:

1. **Construct a set of perfect matchings**: Using a parallel algorithm for finding perfect matchings, such as the algorithm by Karp, Upfal, and Wigderson, we can generate a collection of perfect matchings in $G$ in parallel. Since perfect matchings can be found in polynomial time, this step is efficiently parallelizable.
   
2. **Compute the intersection size**: For each perfect matching $M \in \mathcal{M}$, compute the intersection size $|F \cap M|$ in parallel. This can be done by simultaneously checking which edges of $F$ are present in $M$.

3. **Determine the maximum**: Finally, the maximum value of $|F \cap M|$ over all matchings $M \in \mathcal{M}$ can be computed in parallel using a parallel reduction operation.

Since each of the steps (constructing matchings, computing intersections, and finding the maximum) can be performed in parallel in polylogarithmic time using a polynomial number of processors, the overall algorithm runs in $\mathcal{RNC}$.

 Step 5: Conclusion
Therefore, there exists an $\mathcal{RNC}$ algorithm that computes the rank $r(F)$ of any given set $F$ in a bipartite graph $G$. This completes the proof.
\end{proof}

\subsubsection{Grok}
\begin{proof}
Assume without loss of generality that $|U| = |V| = n$ and that $G$ admits perfect matchings; if not, then $r(F) = 0$, which can be detected separately using an RNC algorithm for perfect matching existence \cite{MVV86}.

To compute $r(F)$, construct the $n \times n$ biadjacency matrix $B[F]$ over the polynomial ring $\mathbb{Z}[ \{ x_e \mid e \in E \} , y ]$, indexed by $U \times V$, where for each $i \in [n]$, $j \in [n]$ (identifying $U = V = [n]$ for convenience),
\[
b_{ij} =
\begin{cases}
y x_{e} & \text{if } e = \{u_i, v_j\} \in F, \\
x_{e} & \text{if } e = \{u_i, v_j\} \in E \setminus F, \\
0 & \text{otherwise}.
\end{cases}
\]
The determinant $\det(B[F])$ expands as
\[
\det(B[F]) = \sum_{\sigma \in S_n} \operatorname{sgn}(\sigma) \prod_{i=1}^n b_{i,\sigma(i)},
\]
where $S_n$ is the symmetric group on $n$ elements. Each term $\prod_{i=1}^n b_{i,\sigma(i)}$ is nonzero only if $\sigma$ corresponds to a perfect matching $M_\sigma = \{ (i, \sigma(i)) \mid i \in [n] \}$ in $G$ (i.e., all edges exist). For such a $\sigma$, the term is $\operatorname{sgn}(\sigma) \, y^{|M_\sigma \cap F|} \prod_{e \in M_\sigma} x_e$.

Each perfect matching $M$ in $G$ corresponds to exactly one permutation $\sigma$ with $M = M_\sigma$, and thus contributes exactly one monomial $\operatorname{sgn}(\sigma) \, y^{|M \cap F|} \prod_{e \in M} x_e$. Distinct matchings $M \neq M'$ yield distinct monomials $\prod_{e \in M} x_e \neq \prod_{e \in M'} x_e$ since the $x_e$ are indeterminates labeling distinct edges. Therefore, there are no cancellations between terms of the same total degree in $y$; the coefficient $Q_t$ of $y^t$ in $\det(B[F]) = \sum_{t=0}^n Q_t y^t$ (where each $Q_t \in \mathbb{Z}[ \{ x_e \mid e \in E \} ]$) is the sum of distinct monomials $\operatorname{sgn}(\sigma) \prod_{e \in M_\sigma} x_e$ over all perfect matchings $M_\sigma$ with $|M_\sigma \cap F| = t$. Hence, $Q_t \not\equiv 0$ if and only if there exists at least one perfect matching using exactly $t$ edges from $F$, and the degree of $\det(B[F])$ in $y$ is precisely $r(F) = \max \{ t \mid Q_t \not\equiv 0 \}$.

To compute this degree in RNC, first substitute each indeterminate $x_e$ ($e \in E$) with a uniformly random integer from $\{1, 2, \dots, 2|E|^2\}$ (chosen independently). Let $\tilde{B}[F]$ denote the resulting matrix over $\mathbb{Z}[y]$, where each entry is either $0$, a random integer $a_{ij}$, or $y \cdot a_{ij}$. Then $\tilde{B}[F] = A + y C$ for integer matrices $A, C \in \mathbb{Z}^{n \times n}$, and $\det(\tilde{B}[F]) = \sum_{t=0}^n \tilde{Q}_t y^t$, where each $\tilde{Q}_t$ is the evaluation of $Q_t$ at the random point $(x_e \mapsto$ random value).

By Schwartz's lemma (Theorem 2.1 in \cite{KUW86}), if $Q_t \not\equiv 0$, then $\Pr[\tilde{Q}_t = 0] \le \deg(Q_t) / (2|E|^2) \le n \cdot |E| / (2|E|^2) = 1/(2|E|)$, since $\deg(Q_t) \le n$ and there are at most $|E|$ indeterminates. Thus, for the maximal $t = r(F)$, $\Pr[\tilde{Q}_{r(F)} = 0] \le 1/(2|E|)$, and for all $s > r(F)$, $\tilde{Q}_s = 0$ deterministically. The probability that the numerical degree of $\det(\tilde{B}[F])$ differs from $r(F)$ is at most $1/(2|E|)$.

To compute $\det(\tilde{B}[F])$ as a polynomial in $y$, note that the entries of $\tilde{B}[F]$ are polynomials in the single indeterminate $y$ of degree at most $1 \le n$. By the parallel determinant algorithm of Borodin, Cook, and Pippenger (Theorem 2.2 in \cite{KUW86}), the determinant of an $n \times n$ matrix over $\mathbb{Z}[y]$ with entries of degree at most $n$ in a constant number (here, $1$) of variables can be computed in $O(\log^2 n)$ time using $O(n^{4/3})$ processors. This yields the coefficients $\tilde{Q}_0, \dots, \tilde{Q}_n \in \mathbb{Z}$, from which the highest $t$ with $\tilde{Q}_t \neq 0$ can be found in $O(\log n)$ time with $O(n)$ processors by parallel prefix computation or binary search.

The overall error probability is $O(1/|E|)$; repeating $O(\log |E|)$ independent trials and taking the majority vote reduces the error to $O(|E|^{-3})$ with $O(\log^3 n)$ time and $O(n^{4/3} \log |E|)$ processors, which is RNC. If the computed degree is less than $n$, output it as $r(F)$; if all trials yield degree $n$, confirm a perfect matching exists separately.

For non-bipartite graphs, the result generalizes using the Tutte matrix (with variables on edges and skew-symmetric structure) and randomized rank computations over finite fields to encode the pfaffian or matching polynomial, yielding an analogous RNC algorithm via similar algebraic and probabilistic techniques \cite{KUW86}.
\end{proof}

\subsubsection{Gemini}
\begin{proof}
Let $G = (U, V, E)$ be a bipartite graph with $|U| = |V| = n$. We construct an $n \times n$ matrix $A$ with entries involving an indeterminate variable $x$ and a set of random weights. Let $w_{ij}$ be variables associated with each edge $(u_i, v_j) \in E$. We define the entries of the matrix $A(x)$ as follows:
\[
A_{ij}(x) = \begin{cases} 
w_{ij} x & \text{if } (u_i, v_j) \in F, \\
w_{ij} & \text{if } (u_i, v_j) \in E \setminus F, \\
0 & \text{otherwise.}
\end{cases}
\]
Consider the determinant of $A(x)$, which is a polynomial in $x$ and the variables $\{w_{ij}\}$. By the definition of the determinant, we have
\[
\det(A(x)) = \sum_{\sigma \in S_n} \text{sgn}(\sigma) \prod_{i=1}^n A_{i, \sigma(i)}(x).
\]
A permutation $\sigma$ contributes a non-zero term to this sum if and only if for all $i$, $(u_i, v_{\sigma(i)}) \in E$. Such a permutation corresponds uniquely to a perfect matching $M_\sigma = \{ (u_i, v_{\sigma(i)}) : 1 \leq i \leq n \}$ in $G$. The term corresponding to $M_\sigma$ is
\[
\text{sgn}(\sigma) \left( \prod_{(u_i, v_j) \in M_\sigma} w_{ij} \right) x^{|M_\sigma \cap F|}.
\]
Let $P(x) = \det(A(x))$. We can group the terms by the size of the intersection with $F$:
\[
P(x) = \sum_{k=0}^n c_k(\{w_{ij}\}) x^k,
\]
where $c_k(\{w_{ij}\})$ is a polynomial in the variables $w_{ij}$. Specifically, $c_k$ is the sum of signed products of weights over all perfect matchings that contain exactly $k$ edges from $F$. The degree of $P(x)$ with respect to $x$ is defined as the maximum $k$ such that $c_k(\{w_{ij}\}) \not\equiv 0$. Since the monomials in the weights $\prod_{e \in M} w_e$ are distinct for distinct matchings, no cancellation occurs identically. Thus, the degree of $P(x)$ is exactly equal to $\max_{M} |M \cap F| = r(F)$.

To compute this rank efficiently, we employ the Schwartz-Zippel Lemma. We assign independent random values to each $w_{ij}$ from a field $\mathbb{F}$ of size polynomial in $n$ (e.g., $|\mathbb{F}| \geq 2n^2$). Let $\tilde{A}(x)$ be the matrix $A(x)$ with these random values substituted. With high probability, the coefficient corresponding to the term $x^{r(F)}$ will not vanish. Consequently, the degree of the univariate polynomial $\det(\tilde{A}(x))$ is equal to $r(F)$ with high probability.

The algorithm proceeds as follows: Choose random values for $w_{ij}$. The resulting matrix $\tilde{A}(x)$ has entries that are linear in $x$. The determinant $Q(x) = \det(\tilde{A}(x))$ is a polynomial of degree at most $n$. To determine the degree of $Q(x)$, we evaluate $Q(x)$ at $n+1$ distinct points $\alpha_0, \alpha_1, \dots, \alpha_n \in \mathbb{F}$. For each $k \in \{0, \dots, n\}$, the value $y_k = \det(\tilde{A}(\alpha_k))$ is the determinant of a numerical matrix, which can be computed in $\mathcal{NC}^2$ (e.g., using Csanky's algorithm or Berkowitz's algorithm). Since these evaluations can be performed in parallel, the computation of the values $y_0, \dots, y_n$ is in $\mathcal{NC}$. Given the point-value pairs $(\alpha_k, y_k)$, we can interpolate to find the coefficients of $Q(x)$ or simply determine the highest order non-zero coefficient in $\mathcal{NC}$. The degree of $Q(x)$ is the output $r(F)$. Since the algorithm relies on random assignments and parallel determinant computation, it lies in the complexity class $\mathcal{RNC}$.

For the generalization to non-bipartite graphs, let $G=(V,E)$ with $|V|=n$ (assume $n$ is even, otherwise the rank is undefined or 0). We utilize the Tutte matrix $T(x)$, a skew-symmetric matrix defined as:
\[
T_{ij}(x) = \begin{cases} 
w_{ij} x & \text{if } (i, j) \in F \text{ and } i < j, \\
-w_{ij} x & \text{if } (i, j) \in F \text{ and } i > j, \\
w_{ij} & \text{if } (i, j) \in E \setminus F \text{ and } i < j, \\
-w_{ij} & \text{if } (i, j) \in E \setminus F \text{ and } i > j, \\
0 & \text{otherwise.}
\end{cases}
\]
The determinant of a skew-symmetric matrix is the square of its Pfaffian, i.e., $\det(T(x)) = (\text{Pf}(T(x)))^2$. The Pfaffian is given by $\text{Pf}(T(x)) = \sum_{M} \text{sgn}(M) \prod_{(i,j) \in M, i<j} T_{ij}(x)$, where the sum ranges over all perfect matchings $M$. Similar to the bipartite case, the term for a matching $M$ is proportional to $x^{|M \cap F|}$. Thus, the degree of $\text{Pf}(T(x))$ in $x$ is exactly $r(F)$. Since $\deg(\det(T(x))) = 2 \deg(\text{Pf}(T(x)))$, computing the degree of the determinant of the randomized Tutte matrix allows us to recover $r(F)$. The complexity analysis remains identical to the bipartite case.
\end{proof}

\subsubsection{Claude}
\begin{proof}
We prove the existence of an $\mathit{RNC}$ algorithm for computing the rank $r(F)$ of a given set $F \subseteq E$ in a bipartite graph $G = (U, V, E)$.

First, we establish the key structural characterization. For a bipartite graph $G = (U, V, E)$ with a given set $F \subseteq E$, the rank $r(F)$ equals the size of a maximum matching in $F$ if and only if this matching can be extended to a perfect matching in $G$. More generally, $r(F) = \max\{|M \cap F| : M \text{ is a perfect matching in } G\}$.

We employ the Isolation Lemma of Mulmuley, Vazirani, and Vazirani. Assign to each edge $e \in E$ a weight $w(e)$ chosen uniformly and independently at random from $\{1, 2, \ldots, 2|E|\}$. The Isolation Lemma guarantees that with probability at least $1/2$, there exists a unique minimum-weight perfect matching in $G$ (assuming $G$ has at least one perfect matching).

For each subset $F' \subseteq F$, we define $G_{F'}$ to be the graph obtained from $G$ by removing all edges in $F \setminus F'$. The key observation is that $r(F) = \max\{|F'| : G_{F'} \text{ has a perfect matching}\}$.

The algorithm proceeds as follows. For each possible cardinality $k$ from $|F|$ down to $0$, we test whether there exists a subset $F' \subseteq F$ with $|F'| = k$ such that $G_{F'}$ has a perfect matching. The maximum such $k$ equals $r(F)$.

To test whether a graph has a perfect matching in $\mathit{RNC}$, we use the algorithm of Karp, Upfal, and Wigderson based on the Tutte matrix. For a bipartite graph $G = (U, V, E)$ with $|U| = |V| = n$, construct the $n \times n$ matrix $A$ where $A_{ij} = x_{ij}$ if $(u_i, v_j) \in E$ and $A_{ij} = 0$ otherwise, with $x_{ij}$ being indeterminates. The graph $G$ has a perfect matching if and only if $\det(A) \not\equiv 0$.

Using the Schwartz-Zippel lemma, we can test whether $\det(A) \not\equiv 0$ in $\mathit{RNC}$ by evaluating $\det(A)$ at random values. Specifically, substitute each indeterminate $x_{ij}$ with a random integer from $\{1, 2, \ldots, 2n^2\}$. If $\det(A) \not\equiv 0$, then with probability at least $1/2$, the evaluated determinant is nonzero. The determinant can be computed in $\mathit{NC}^2$ using Berkowitz's algorithm or Csanky's algorithm for matrix determinant.

By repeating this randomized test $O(\log n)$ times independently and accepting if any trial gives a nonzero determinant, we obtain an $\mathit{RNC}$ algorithm for testing perfect matching existence with error probability at most $1/\mathrm{poly}(n)$.

Now, to compute $r(F)$, we perform a binary search on the value of $r(F)$ from $0$ to $|F|$. At each step, we test whether there exists a subset $F' \subseteq F$ of a given size $k$ such that $G_{F'}$ has a perfect matching. This requires testing $\binom{|F|}{k}$ different subsets, which is polynomial in the input size when $|F|$ is polynomial.

However, a more efficient approach uses the random weight assignment from the Isolation Lemma. With the random weights $w(e)$, for each $k$, we compute the minimum-weight subset $F_k \subseteq F$ with $|F_k| = k$ and test whether $G_{F_k}$ has a perfect matching. The Isolation Lemma ensures that with high probability, if any subset of size $k$ allows a perfect matching, the minimum-weight subset does so uniquely.

Combining the perfect matching test (which runs in $\mathit{RNC}$) with binary search (which introduces only logarithmic depth) and amplification by independent repetition (which maintains $\mathit{RNC}$ complexity), we obtain an $\mathit{RNC}$ algorithm for computing $r(F)$.

For non-bipartite graphs, the same approach applies using the Tutte matrix for general graphs, where $A_{ij} = x_{ij}$ if $i < j$ and $(v_i, v_j) \in E$, and $A_{ij} = -x_{ji}$ if $i > j$ and $(v_i, v_j) \in E$. The skew-symmetric structure ensures that $\det(A)$ is the square of the Pfaffian, and the graph has a perfect matching if and only if the Pfaffian is nonzero. The Pfaffian can be computed in $\mathit{NC}^2$, yielding an $\mathit{RNC}$ algorithm for the general case.

Therefore, there exists an $\mathit{RNC}$ algorithm for computing the rank $r(F)$ in both bipartite and non-bipartite graphs.
\end{proof}

\subsection{Problem 21}

\subsubsection{Problem Statement}
\begin{theorem}[Alternative RNC Algorithm for Perfect Matchings]
Let $G = (V, E)$ be a graph with $n$ vertices and $m$ edges. Assume we are given the algorithm from Problem 12.20. Then there exists an RNC algorithm for finding a perfect matching in $G$ that works as follows:

\textbf{Case 1 (Sparse Graph):} If $m < \frac{3n}{4}$, there exists an NC algorithm that finds a set $S \subseteq E$ of edges such that $S$ is contained in every perfect matching of $G$.

\textbf{Case 2 (Dense Graph):} If $m \geq \frac{3n}{4}$, using the rank algorithm, there exists an RNC algorithm that finds a set $T \subseteq E$ of edges such that there exists a perfect matching in $G$ none of whose edges belong to $T$.

Using these tools, we can construct an alternative RNC algorithm for perfect matchings by iteratively applying the appropriate case until a perfect matching is found or determined not to exist.
\end{theorem}

\subsubsection{GPT}
\begin{proof}
We will prove the theorem by analyzing each case separately and then combining them to obtain an $RNC$ algorithm for finding a perfect matching in $G$.

\textbf{Case 1 (Sparse Graph):} Suppose $G = (V, E)$ is a graph with $n$ vertices and $m$ edges, and $m < \frac{3n}{4}$. In this case, we are given that there exists an $NC$ algorithm that finds a set $S \subseteq E$ of edges such that every perfect matching in $G$ must contain at least one edge from $S$.

Let $M$ be any perfect matching in $G$. Since $S$ is a set of edges that belong to every perfect matching, we know that $M \cap S \neq \emptyset$. This property allows us to efficiently find an edge that must be part of any perfect matching in $G$ using the $NC$ algorithm.

Since the algorithm finds the set $S$, we can then remove the edges in $S$ from the graph, reducing the size of the problem. The remaining graph is smaller and can be processed recursively using the same algorithm to find a perfect matching in the reduced graph. By repeating this process, we can eventually find a perfect matching in the original graph. Thus, the $NC$ algorithm guarantees the identification of a set $S$ of edges that belong to every perfect matching.

\textbf{Case 2 (Dense Graph):} Suppose $m \geq \frac{3n}{4}$. In this case, we are given that there exists an $RNC$ algorithm that, using the rank algorithm, finds a large set $T \subseteq E$ of edges such that there exists a perfect matching in $G$ none of whose edges belong to $T$.

Let $M$ be a perfect matching in $G$. The $RNC$ algorithm identifies a set $T$ such that $M \cap T = \emptyset$. This guarantees the existence of a perfect matching in $G$ where no edge belongs to the set $T$. By using this information, we can remove the edges in $T$ from the graph, reducing the problem size.

Since the set $T$ excludes edges from the perfect matching, the remaining graph must still contain a perfect matching. The $RNC$ algorithm ensures that we can efficiently identify such a matching, as the removal of edges in $T$ reduces the complexity of the matching problem.

\textbf{Combined Result:} To obtain an $RNC$ algorithm for finding a perfect matching in $G$, we combine the two cases. If $m < \frac{3n}{4}$, we use the $NC$ algorithm from Case 1 to find the set $S$, and recursively solve the reduced graph. If $m \geq \frac{3n}{4}$, we use the $RNC$ algorithm from Case 2 to identify the set $T$ and remove it from the graph. In both cases, we are able to find a perfect matching in the graph by using the appropriate algorithm depending on the number of edges.

Since the algorithm from Case 1 runs in $NC$ time and the algorithm from Case 2 runs in $RNC$ time, the combined algorithm also runs in $RNC$ time. Thus, we have an $RNC$ algorithm for finding a perfect matching in $G$.

\end{proof}

\subsubsection{Grok}

\subsubsection{Gemini}
\begin{proof}
We construct the alternative RNC algorithm by combining the procedures for the sparse and dense cases. The algorithm proceeds in at most two phases. Let $G=(V,E)$ be the input graph with $|V|=n$ and $|E|=m$. We assume $G$ admits a perfect matching.

\textbf{Phase 1: Density Reduction (Case 2)}
First, we check the density of the graph. If $|E| < \frac{3n}{4}$, we proceed immediately to Phase 2.
If $|E| \geq \frac{3n}{4}$, we apply the algorithm for Case 2. By the theorem statement, there exists an RNC algorithm that identifies a set $T \subseteq E$ such that $G' = (V, E \setminus T)$ contains a perfect matching. Specifically, utilizing the rank algorithm (e.g., the Mulmuley-Vazirani-Vazirani isolation lemma approach), we can identify the set of edges $M$ belonging to a unique minimum weight perfect matching. We set $T = E \setminus M$.
The number of edges in the resulting graph $G'$ is $|E'| = |M| = \frac{n}{2}$.
Observe that:
\[
|E'| = \frac{n}{2} = \frac{2n}{4} < \frac{3n}{4}.
\]
Thus, after one application of the Case 2 procedure, the resulting graph $G'$ satisfies the sparsity condition. This step requires polylogarithmic time and polynomial processors (RNC).

\textbf{Phase 2: Solving the Sparse Graph (Case 1)}
We now assume the graph $G$ (or the graph obtained from Phase 1) satisfies $|E| < \frac{3n}{4}$.
Let $V_1 \subseteq V$ be the set of vertices with degree 1 in $G$, and let $n_1 = |V_1|$.
We analyze the sum of degrees in $G$:
\[
\sum_{v \in V} \deg(v) = 2|E|.
\]
Splitting the sum between vertices of degree 1 and vertices of degree $\geq 2$:
\[
\sum_{v \in V} \deg(v) \geq 1 \cdot n_1 + 2(n - n_1) = 2n - n_1.
\]
Using the sparsity condition $2|E| < 2(\frac{3n}{4}) = 1.5n$, we obtain the inequality:
\[
2n - n_1 \leq 2|E| < 1.5n.
\]
Rearranging terms yields:
\[
n_1 > 2n - 1.5n = 0.5n.
\]
Thus, more than half of the vertices in $G$ have degree 1.
Let $u \in V_1$. Since $u$ has degree 1, let $e = (u,v)$ be the unique edge incident to $u$. Since $G$ contains a perfect matching, $u$ must be matched. The only possibility is the edge $e$. Therefore, every edge incident to a vertex in $V_1$ must belong to every perfect matching of $G$.
Let $S$ be the set of edges incident to vertices in $V_1$. We claim that these edges form a valid partial matching. Suppose for contradiction that two degree-1 vertices $u, w \in V_1$ share a neighbor $v$. Then $v$ must be matched to $u$ (to satisfy $u$) and to $w$ (to satisfy $w$), which is impossible in a perfect matching. Thus, the neighbors of vertices in $V_1$ are distinct (unless a component is an isolated edge $(u,w)$ where both have degree 1, in which case the edge is unique).
Consequently, the set $S$ identifies matching edges for all vertices in $V_1$ and their neighbors. The number of vertices covered by $S$ is $2|S|$. Since each edge in $S$ accounts for at least one vertex in $V_1$, and edges are disjoint, $|S| = n_1$ (identifying isolated edges $K_2$ as one edge for two degree-1 vertices).
The number of matched vertices is $2n_1$. Since $n_1 > 0.5n$, we have $2n_1 > n$. This implies that the set $S$ covers all vertices in $V$.
Therefore, in the sparse case, the NC algorithm simply identifies all degree-1 vertices and selects their incident edges. This immediately yields the perfect matching.

\textbf{Conclusion}
The combined algorithm operates as follows:
1. If dense, use the RNC rank algorithm to discard edges until the graph becomes the perfect matching itself (which is sparse).
2. If sparse, identify degree-1 vertices to construct the matching.
Since both steps are in RNC (or deterministic NC for Case 1) and the sequence of steps is constant (at most one pass of each), the combined algorithm is an RNC algorithm for perfect matchings in arbitrary graphs.
\end{proof}

\subsubsection{Claude}
\begin{proof}
We establish the existence of an RNC algorithm for finding a perfect matching in $G = (V, E)$ by analyzing two cases based on the edge density and combining them into a unified iterative procedure.

\textbf{Case 1: Sparse Graphs ($m < \frac{3n}{4}$).}

When the graph is sparse, we apply a structural observation about perfect matchings. By the hypothesis that we have access to the algorithm from Problem 12.20, we can employ the following approach:

For a sparse graph with $m < \frac{3n}{4}$ edges, consider the property that if an edge $e \in E$ appears in every perfect matching of $G$, then $e$ is necessarily critical. The NC algorithm proceeds as follows: for each edge $e \in E$, we check whether removing $e$ results in a graph with no perfect matching. This can be done in NC by calling the matching algorithm from Problem 12.20 on $G \setminus e$. If $G \setminus e$ has no perfect matching, then $e$ must belong to every perfect matching of $G$. We construct the set $S$ as the collection of all such edges. Since we perform $O(m)$ parallel calls to the matching subroutine, each taking NC time, this entire procedure runs in NC. Once $S$ is identified, we have forced edges that must be in any perfect matching.

\textbf{Case 2: Dense Graphs ($m \geq \frac{3n}{4}$).}

When the graph is dense, we apply a randomized rank-based approach. The rank algorithm (which computes the rank of the adjacency matrix over $\mathbb{F}_2$ or a suitable field) allows us to identify edges that can be excluded from perfect matchings with positive probability. The algorithm constructs the set $T$ as follows: we use randomization to select edges probabilistically such that with high probability, there exists a perfect matching that avoids all edges in $T$. This is done by sampling edges according to a carefully chosen distribution determined by the rank algorithm. With the rank computation running in RNC, the set $T$ can be constructed in RNC such that with high probability, $G$ admits a perfect matching using only edges not in $T$.

\textbf{Unified Iterative Algorithm.}

We now combine both cases into a single RNC algorithm:

Initialize the current graph as $G_0 := G$ and set $k := 0$. While $G_k$ has not been fully resolved, perform the following:

If the current graph $G_k$ has at least one vertex with degree zero, terminate and conclude that no perfect matching exists. Otherwise, let $m_k$ denote the current number of edges.

If $m_k < \frac{3n}{4}$, apply Case 1 to identify the set $S_k$ of edges that must be in every perfect matching. Remove all edges not in $S_k$ from consideration (equivalently, contract or mark the edges in $S_k$ as forced). Update $G_{k+1}$ by removing vertices that have been fully matched by the forced edges and removing the corresponding incident edges.

If $m_k \geq \frac{3n}{4}$, apply Case 2 to identify the set $T_k$ of edges such that a perfect matching exists avoiding all edges in $T_k$. Remove the edges in $T_k$ from $G_k$ to obtain $G_{k+1}$.

Increment $k$ by one and repeat. At each iteration, the effective number of remaining edges or the constraints on the matching space decreases. Since each iteration either reduces the number of vertices (in Case 1) or reduces the number of edges (in Case 2) by a bounded factor, the process terminates in $O(\log n)$ iterations.

Once the algorithm terminates, either we have identified a complete perfect matching by aggregating all forced edges from Case 1 iterations, or the Case 2 reductions have led to a graph where a perfect matching can be verified to exist or not exist. In the positive case, we output the perfect matching. In the negative case, we output that no perfect matching exists.

The overall algorithm runs in RNC because each iteration involves either NC computation (Case 1) or RNC computation (Case 2), and we perform a logarithmic number of iterations with perfect matching checks between them, all of which can be done in RNC using the subroutine from Problem 12.20.

By the correctness of the individual cases and the termination guarantee, the iterative algorithm correctly computes a perfect matching in $G$ if one exists, and correctly determines non-existence otherwise, with the entire procedure running in RNC.
\end{proof}

\subsection{Problem 22}

\subsubsection{Problem Statement}
\begin{theorem}[Isolating Lemma with Product Weights]
Let $X$ be a finite set and let $w: X \to \mathbb{R}^+$ be a weight function. For a subset $S \subseteq X$, define the weight of $S$ as $w(S) = \prod_{x \in S} w(x)$ (the product of the weights of its elements). Then the Isolating Lemma holds: there exists a probability distribution on weight functions such that for any non-empty family $\mathcal{F}$ of subsets of $X$, with high probability there exists a unique set $S \in \mathcal{F}$ of minimum weight.
\end{theorem}

\begin{problem}
Identify a general family of mappings $\phi: \mathbb{R}^+ \to \mathbb{R}^+$ from element weights to set weights (where $w(S) = \phi(\{w(x) : x \in S\})$) for which the Isolating Lemma is guaranteed to hold.
\end{problem}

\subsubsection{GPT}
\begin{proof}
Fix a finite ground set $X=\{x_1,\dots,x_n\}$ and a family $\mathcal F\subseteq 2^X$. We prove the general statement. The first product-weight statement will then follow as a special case.

Assume $W(S)=\phi\!\left(\sum_{x\in S} f(w(x))\right)$ where $w:X\to\mathbb R^+$, $f:\mathbb R^+\to\mathbb R$ is strictly monotone, and $\phi:\mathbb R\to\mathbb R$ is strictly monotone. Since $\phi$ is strictly monotone, for any $S,T\subseteq X$ we have $W(S)<W(T)$ if and only if $\sum_{x\in S} f(w(x))<\sum_{x\in T} f(w(x))$. Hence it suffices to construct $w$ so that the additive scores $A(S):=\sum_{x\in S} f(w(x))$ are all distinct for distinct subsets $S\subseteq X$; in particular, then the finite family $\mathcal F$ will contain a unique $A$-minimum and therefore a unique $W$-minimum.

Because $f$ is strictly monotone on $(0,\infty)$, it is continuous at some point $t_0>0$ (a standard fact: a monotone function on an interval has at most countably many discontinuities and is continuous on a dense set). Choose such a continuity point $t_0$ and fix $\varepsilon\in(0,1)$ so small that $f\big( (t_0-\varepsilon,t_0+\varepsilon)\cap(0,\infty)\big)$ contains a nondegenerate interval $I=(\alpha,\beta)$ with $\alpha<\beta$. Pick a constant $r\in(0,1/2)$ and a scale $c\in(\alpha,\beta)$ so that the geometric sequence $a_i:=c\,r^{\,i}$ for $i=1,\dots,n$ lies entirely in $I$. For each $i$ choose $u_i\in (t_0-\varepsilon,t_0+\varepsilon)\cap(0,\infty)$ with $f(u_i)=a_i$; this is possible since $a_i\in I\subset f\big((t_0-\varepsilon,t_0+\varepsilon)\cap(0,\infty)\big)$. Define the weight assignment $w(x_i):=u_i$ for $i=1,\dots,n$. Then $f(w(x_i))=a_i$ and for every subset $S\subseteq X$,
\[
A(S)=\sum_{x_i\in S} a_i=\sum_{x_i\in S} c\,r^{\,i}=c\sum_{x_i\in S} r^{\,i}.
\]

We claim that $A(S)\neq A(T)$ for any distinct $S,T\subseteq X$. Suppose toward a contradiction that $A(S)=A(T)$ with $S\neq T$. Let $j$ be the largest index in which the characteristic vectors of $S$ and $T$ differ. Without loss of generality, $x_j\in S\setminus T$. Then
\[
A(S)-A(T)=c\left(r^{\,j}+\sum_{i<j}(\mathbf 1_{x_i\in S}-\mathbf 1_{x_i\in T})r^{\,i}\right).
\]
Since $r\in(0,1/2)$, we have $\sum_{i<j} r^{\,i}<\sum_{k=1}^{\infty} r^{\,j-k}= \frac{r^{\,j}}{1-r}-r^{\,j}<r^{\,j}$, hence $\Big|\sum_{i<j}(\mathbf 1_{x_i\in S}-\mathbf 1_{x_i\in T})r^{\,i}\Big|<r^{\,j}$. Therefore $A(S)-A(T)$ has the same sign as $c\,r^{\,j}>0$, contradicting $A(S)=A(T)$. Thus $S\mapsto A(S)$ is injective on $2^X$, and a fortiori on $\mathcal F$.

Consequently the finite set $\{A(S):S\in\mathcal F\}$ has a unique minimum, attained at a unique $S^*\in\mathcal F$. By strict monotonicity of $\phi$, the same $S^*$ is the unique minimizer of $W$ over $\mathcal F$. This proves the general form of the Isolating Lemma.

For the product-based weight $w(S)=\prod_{x\in S} w(x)$, take $f=\log$ on $\mathbb R^+$ and $\phi=\exp$, both strictly monotone, so that $W(S)=\exp\!\left(\sum_{x\in S}\log w(x)\right)=\prod_{x\in S} w(x)$. The above construction therefore yields a weight assignment producing a unique minimum product-weight set in $\mathcal F$. Equivalently, one may take distinct primes $p_1,\dots,p_n$ and set $w(x_i)=p_i$, in which case distinct subsets have distinct products by unique prime factorization, and hence $\mathcal F$ has a unique product-weight minimizer. The lemma follows.
\end{proof}

\subsubsection{Grok}
\begin{proof}
Consider the probability distribution on weight functions $w: X \to \mathbb{R}^+$ defined by choosing each $w(x)$, for $x \in X$, independently from the exponential distribution with rate parameter $1$, which has probability density function $e^{-y}$ for $y > 0$. This distribution is absolutely continuous with respect to Lebesgue measure on $\mathbb{R}^+$.

For the empty set, define $w(\emptyset) = 1$ by convention, as the empty product. For any nonempty $S \subseteq X$, $w(S) = \prod_{x \in S} w(x) > 0$. Note that $\log w(S) = \sum_{x \in S} \log w(x)$, where the convention $\log w(\emptyset) = 0$ holds.

Let $\mathcal{F}$ be any nonempty finite family of subsets of $X$. To show that there exists a unique $S \in \mathcal{F}$ of minimum weight with probability $1$, it suffices to show that $w(T) \neq w(U)$ for all distinct $T, U \in \mathcal{F}$ with probability $1$, since the weights are positive and the minimum among distinct values is unique.

Fix distinct $T, U \in \mathcal{F}$. Then $w(T) = w(U)$ if and only if $\log w(T) = \log w(U)$, i.e.,
\[
\sum_{x \in T} \log w(x) = \sum_{x \in U} \log w(x),
\]
or equivalently,
\[
\sum_{x \in X} \varepsilon_x \log w(x) = 0,
\]
where $\varepsilon_x = 1$ if $x \in T \setminus U$, $\varepsilon_x = -1$ if $x \in U \setminus T$, and $\varepsilon_x = 0$ otherwise. Since $T \neq U$, the set $\{x \in X : \varepsilon_x \neq 0\}$ is nonempty, so the coefficients $(\varepsilon_x)_{x \in X}$ are not all zero. Thus, the equation defines a proper hyperplane in $\mathbb{R}^{|X|}$.

The random vector $(\log w(x))_{x \in X} \in \mathbb{R}^{|X|}$ has a distribution that is the product of the distributions of the $\log w(x)$, each of which is absolutely continuous with respect to Lebesgue measure on $\mathbb{R}$ (specifically, $\log w(x)$ follows a Gumbel distribution shifted and scaled, which has density $e^{z - e^z}$ for $z \in \mathbb{R}$). Therefore, the joint distribution is absolutely continuous with respect to Lebesgue measure on $\mathbb{R}^{|X|}$.

The intersection of an absolutely continuous distribution with a proper affine hyperplane has Lebesgue measure zero, hence probability zero. Thus, $\Pr[w(T) = w(U)] = 0$.

Since $\mathcal{F}$ is finite, the event that $w(T) = w(U)$ for some distinct $T, U \in \mathcal{F}$ is a finite union of probability-zero events, hence has probability zero. Therefore, all $w(S)$ for $S \in \mathcal{F}$ are distinct with probability $1$, and there exists a unique $S \in \mathcal{F}$ minimizing $w(S)$ with probability $1$.
\end{proof}

\subsubsection{Gemini}
\begin{proof}
We first establish the existence of an isolating weight assignment for the specific case of product weights. Let $X = \{x_1, x_2, \dots, x_n\}$. We construct a deterministic weight assignment $w: X \to \mathbb{R}^+$ such that every subset $S \subseteq X$ has a distinct weight.

Define the weight of the element $x_i$ as $w(x_i) = 2^{2^{i-1}}$. For any subset $S \subseteq X$, the weight is defined as:
\[ w(S) = \prod_{x_i \in S} w(x_i) = \prod_{x_i \in S} 2^{2^{i-1}} = 2^{\sum_{x_i \in S} 2^{i-1}}. \]
Let $E(S) = \sum_{x_i \in S} 2^{i-1}$. The exponent $E(S)$ is an integer whose binary representation has the $k$-th bit set (corresponding to $2^{k-1}$) if and only if $x_k \in S$. Since the binary representation of any non-negative integer is unique, the mapping $S \mapsto E(S)$ is a bijection from the power set $\mathcal{P}(X)$ to the set of integers $\{0, 1, \dots, 2^n - 1\}$. Since the exponential function $y \mapsto 2^y$ is strictly injective, the weights $w(S)$ are distinct for all distinct subsets $S \subseteq X$.

Let $\mathcal{F}$ be the family of subsets satisfying the given property. Since $\mathcal{F} \subseteq \mathcal{P}(X)$ is a finite set of subsets and all subsets have distinct weights under $w$, the set of values $\{w(S) : S \in \mathcal{F}\}$ contains a unique minimum. Thus, there is a unique subset $S \in \mathcal{F}$ with minimum weight.

Next, we prove the lemma for the general family of mappings $f$ satisfying conditions (i), (ii), and (iii). We employ a perturbation argument to show that the set of weight assignments failing to isolate a unique minimum has Lebesgue measure zero in $\mathbb{R}^{|X|}$.

Let $\mathcal{F}$ be the family of valid subsets. A weight assignment $w \in \mathbb{R}^{|X|}$ fails to yield a unique minimum only if there exist distinct subsets $A, B \in \mathcal{F}$ such that $f_A(w) = f_B(w)$ and this value is the minimum. It suffices to show that for any fixed pair of distinct subsets $A, B$, the set of weights $Z_{A,B} = \{w \in \mathbb{R}^{|X|} \mid f_A(w) = f_B(w)\}$ has measure zero.

Consider distinct $A, B \subseteq X$. The symmetric difference $A \Delta B$ is non-empty. Without loss of generality, let $x \in A \setminus B$. We analyze the dependence of the set weights on the component $w(x)$ while fixing all other components $w_{-x} = (w(y))_{y \neq x}$.

By condition (ii) (Strict increase in single coordinates), the function $g_A(t) = f_A(w_{-x}, t)$ is strictly increasing in $t$.
Condition (iii) (Independence) implies that the relative ordering of set weights is preserved under the mapping. In the context of isolating lemmas, this structural independence implies that the weight $f_B(w)$ is independent of the weights of elements not in $B$. Since $x \notin B$, the function $g_B(t) = f_B(w_{-x}, t)$ is constant with respect to $t$.

Therefore, for any fixed configuration $w_{-x}$, the condition $f_A(w) = f_B(w)$ is equivalent to $g_A(w(x)) = C$ for some constant $C$. Due to the strict monotonicity of $g_A$, there is at most one value of $w(x)$ satisfying this equation. Consequently, the set $Z_{A,B}$ is contained in a finite union of hypersurfaces of codimension 1 in $\mathbb{R}^{|X|}$, which has Lebesgue measure zero.

The set of all "bad" weight assignments is $\bigcup_{A, B \in \mathcal{F}, A \neq B} Z_{A,B}$. Since $\mathcal{F}$ is finite, this is a finite union of measure-zero sets and thus has measure zero. The complement of this set is dense in $\mathbb{R}^{|X|}$. Therefore, there exists a weight assignment $w$ (in fact, almost all assignments) such that $f_A(w) \neq f_B(w)$ for all distinct $A, B \in \mathcal{F}$, guaranteeing a unique minimum weight subset.
\end{proof}

\subsubsection{Claude}
\begin{proof}
We identify the general family of mappings $\phi: \mathbb{R}^+ \to \mathbb{R}^+$ for which the Isolating Lemma holds and provide a rigorous characterization.

\textbf{Main Result:} The Isolating Lemma holds for a mapping $\phi$ defining set weights as $w(S) = \phi(\{w(x) : x \in S\})$ if and only if $\phi$ is \emph{strictly monotone} in the sense that whenever two multisets of positive reals $A$ and $B$ satisfy $A \subsetneq B$ or $A$ can be obtained from $B$ by replacing some elements with strictly smaller values, we have $\phi(A) < \phi(B)$ or $\phi(A) > \phi(B)$ consistently.

More precisely, we require that $\phi$ satisfies the following property: there exists a total ordering $\preceq$ on multisets of positive reals such that $A \preceq B$ if and only if $\phi(A) \leq \phi(B)$, and this ordering must be such that with probability 1 over the random assignment of weights, distinct subsets receive distinct weights.

\textbf{Sufficient Condition - Strictly Monotone Homomorphisms:}

Let $\phi$ be defined such that for a finite multiset $M = \{w_1, \ldots, w_k\}$ of positive reals, we have
\[
\phi(M) = f\left(\sum_{i=1}^{k} g(w_i)\right)
\]
where $g: \mathbb{R}^+ \to \mathbb{R}$ is strictly monotone increasing and $f: \mathbb{R} \to \mathbb{R}^+$ is strictly monotone increasing.

For such $\phi$, we construct a randomized weight assignment as follows. For each element $x \in X$, independently assign $w(x) = r^{U_x}$ where $r > 1$ is a fixed base and $U_x$ is drawn uniformly from a finite set $\{1, 2, \ldots, N\}$ where $N = 2|X| \cdot |\mathcal{F}|$.

For any subset $S \subseteq X$ with $S = \{x_1, \ldots, x_k\}$, the weight becomes
\[
w(S) = f\left(\sum_{i=1}^{k} g(r^{U_{x_i}})\right).
\]

Since $f$ is strictly monotone, comparing weights reduces to comparing 
\[
h(S) := \sum_{x \in S} g(r^{U_x}).
\]

If $g(r^t)$ is strictly monotone in $t$, we can further reduce to comparing
\[
\tilde{h}(S) := \sum_{x \in S} U_x.
\]

Now each $U_x \in \{1, \ldots, N\}$ independently and uniformly. For two distinct subsets $S, T \in \mathcal{F}$, we have $\tilde{h}(S) = \tilde{h}(T)$ if and only if the random variables satisfy a specific linear constraint. The probability of this event is at most $\frac{1}{N}$ by standard arguments from the additive Isolating Lemma.

By union bound over all $\binom{|\mathcal{F}|}{2} \leq \frac{|\mathcal{F}|^2}{2}$ pairs of distinct subsets, the probability that some pair has equal weight is at most
\[
\frac{|\mathcal{F}|^2}{2N} = \frac{|\mathcal{F}|^2}{4|X| \cdot |\mathcal{F}|} = \frac{|\mathcal{F}|}{4|X|} \leq \frac{1}{2}
\]
for sufficiently large families.

Therefore, with probability at least $\frac{1}{2}$, all subsets in $\mathcal{F}$ have distinct weights, which implies there exists a unique minimum weight set.

\textbf{Examples:} This family includes $\phi(M) = \sum_{w \in M} w$ (additive weights), $\phi(M) = \prod_{w \in M} w$ (multiplicative weights as in the theorem statement), and $\phi(M) = \max_{w \in M} w$ (bottleneck weights) when appropriately handled.

\textbf{Necessary Condition:} If $\phi$ does not preserve distinctness of subsets with high probability under random weight assignments, then there exist families $\mathcal{F}$ where multiple sets receive the minimum weight regardless of the weight distribution chosen. Specifically, if $\phi(A) = \phi(B)$ for distinct multisets $A$ and $B$ that arise from distinct subsets, the Isolating Lemma fails for families containing both corresponding subsets.

\textbf{Conclusion:} The Isolating Lemma holds for mappings $\phi$ that can be expressed as compositions of strictly monotone functions with aggregations that preserve distinctness under appropriate random weight distributions, with the canonical form being $\phi(M) = f(\sum_{w \in M} g(w))$ for strictly monotone functions $f$ and $g$.
\end{proof}

\subsection{Problem 23}

\subsubsection{Problem Statement}
\begin{theorem}[Reduction from CLIQUE to UNIQUE CLIQUE]
Let $\mathsf{CLIQUE}$ be the problem: given a graph $G(V, E)$ and a positive integer $k$, determine whether the graph has a clique of size $k$.

Let $\mathsf{UNIQUE\ CLIQUE}$ be the problem: given a graph $G(V, E)$ and a positive integer $k$, determine whether there is exactly one clique of size $k$.
\end{theorem}

\subsubsection{GPT}
\begin{proof}
We are tasked with proving that there exists a randomized polynomial-time reduction from the CLIQUE problem to the UNIQUE CLIQUE problem. Let us denote the CLIQUE problem by \textbf{CLIQUE} and the UNIQUE CLIQUE problem by \textbf{UNIQUE CLIQUE}, where:

\textbf{CLIQUE:} Given a graph $G = (V, E)$ and a positive integer $k$, determine whether $G$ has a clique of size $k$.

\textbf{UNIQUE CLIQUE:} Given a graph $G = (V, E)$ and a positive integer $k$, determine whether there is exactly one clique of size $k$ in $G$.

We need to show that we can reduce an instance of the CLIQUE problem to an instance of the UNIQUE CLIQUE problem in polynomial time.

Let $(G, k)$ be an instance of the CLIQUE problem. We construct an instance $(G', k')$ of the UNIQUE CLIQUE problem as follows:
1. Set $G' = G$ and $k' = k$.

Next, we proceed with the randomized reduction. We will use a probabilistic procedure to determine whether there is exactly one clique of size $k$ in $G$. The procedure works in the following way:

1. **Randomized Sampling**: In order to distinguish between the case where $G$ has exactly one clique of size $k$ and the case where $G$ has multiple cliques of size $k$, we apply a random permutation of the vertices of $G$.
   
2. **Check for Clique Existence**: After the random permutation, we use the \textbf{UNIQUE CLIQUE} oracle to check whether there is exactly one clique of size $k$ in $G'$. Specifically, we call the oracle for the instance $(G', k')$. The oracle will return either "yes" (if there is exactly one clique of size $k$) or "no" (if there are either no cliques of size $k$ or more than one).

Now we analyze the correctness of the reduction:

- **Case 1: $G$ has no clique of size $k$**. In this case, no matter how we permute the vertices, the oracle will always return "no" because there are no cliques of size $k$ in $G$. Hence, the reduction correctly solves the CLIQUE problem in this case.

- **Case 2: $G$ has exactly one clique of size $k$**. Since there is exactly one clique of size $k$, any permutation of the vertices will leave this unique clique intact. The oracle will return "yes" because the graph contains exactly one clique of size $k$, and the reduction correctly solves the CLIQUE problem in this case.

- **Case 3: $G$ has more than one clique of size $k$**. In this case, the oracle may return "no" if it detects more than one clique of size $k$, as we are specifically testing for the uniqueness of the clique. Thus, the reduction correctly solves the CLIQUE problem by returning "no" in this case.

Since the oracle correctly distinguishes between the cases where there is exactly one clique of size $k$ and where there are either none or more than one, we conclude that the reduction correctly solves the CLIQUE problem by using the UNIQUE CLIQUE oracle.

Finally, since the reduction only involves a random permutation of the vertices of $G$, which can be performed in polynomial time, and the oracle call is a polynomial-time operation, the reduction is polynomial-time. Thus, we have shown that there exists a randomized polynomial-time reduction from the CLIQUE problem to the UNIQUE CLIQUE problem.

\end{proof}

\subsubsection{Grok}
\begin{proof}
We give a polynomial-time Turing reduction from $\mathsf{CLIQUE}$ to $\mathsf{UNIQUE\ CLIQUE}$. Let $(G, k)$ be an instance of $\mathsf{CLIQUE}$, where $G = (V, E)$ with $|V| = n$. Construct $G' = (V', E')$ as the disjoint union of $G$ and $K_k$, the complete graph on $k$ vertices. Formally, let $V_K$ be a set of $k$ vertices disjoint from $V$, set $V' = V \cup V_K$, and $E' = E \cup \bigl\{\{u, v\} \mid u, v \in V_K, u \neq v\bigr\}$. There are no edges between $V$ and $V_K$.

This construction requires adding $k$ vertices and $\binom{k}{2}$ edges, which takes $O(k^2)$ time. Since $k \leq n$ in standard instances and otherwise the answer is trivial, this is polynomial in the input size $O(n^2 + \log k)$.

Now, decide $\mathsf{UNIQUE\ CLIQUE}$ on the instance $(G', k)$. We prove that $G$ contains a clique of size $k$ if and only if the answer is no.

The subgraph induced by $V_K$ is a clique of size $k$ in $G'$. Since $G'$ is a disjoint union, there are no edges between $V$ and $V_K$, so no clique in $G'$ can use vertices from both $V$ and $V_K$. Thus, every clique of size $k$ in $G$ induces a clique of size $k$ in $G'$, and the only other clique of size $k$ in $G'$ is the one induced by $V_K$. Moreover, there are no additional cliques of size $k$ in $G'$.

Therefore, the number of cliques of size $k$ in $G'$ is exactly one more than the number in $G$. It follows that:
\begin{itemize}
\item If $G$ has no clique of size $k$, then $G'$ has exactly one clique of size $k$.
\item If $G$ has at least one clique of size $k$, then $G'$ has at least two cliques of size $k$.
\end{itemize}
Hence, $(G, k)$ is a yes-instance of $\mathsf{CLIQUE}$ if and only if $(G', k)$ is a no-instance of $\mathsf{UNIQUE\ CLIQUE}$.

By negating the output of the oracle for $\mathsf{UNIQUE\ CLIQUE}$, we decide $\mathsf{CLIQUE}$ in polynomial time using a single query.
\end{proof}

\subsubsection{Gemini}
\begin{proof}
Let $G = (V, E)$ be an instance of the \textbf{CLIQUE} problem with vertex set $V = \{v_1, \dots, v_n\}$ and parameter $k$. We construct a randomized polynomial time reduction to \textbf{UNIQUE CLIQUE}.

First, we invoke the Isolation Lemma. We assign a weight $w(v)$ to each vertex $v \in V$ chosen uniformly and independently at random from the set $\{1, \dots, 2n\}$. For any subset of vertices $S \subseteq V$, define the weight of $S$ as $w(S) = \sum_{v \in S} w(v)$. Let $\mathcal{K}$ be the family of all cliques of size $k$ in $G$. The Isolation Lemma guarantees that if $\mathcal{K}$ is non-empty, then with probability at least $1/2$, there exists a unique clique $C^* \in \mathcal{K}$ such that $w(C^*) = \min_{C \in \mathcal{K}} w(C)$.

Let $W_{min}$ denote this minimum weight. Since $|C| = k$ and $1 \le w(v) \le 2n$, the possible values for the total weight of a clique lie in the range $[k, 2nk]$. The reduction algorithm iterates through all possible target weights $W \in \{k, \dots, 2nk\}$. For each $W$, we construct a graph $G_W$ and a target size $K_W$ to query the \textbf{UNIQUE CLIQUE} oracle.

Let $M = 2nk + 1$. For a fixed target weight $W$, we construct $G_W = (V', E')$ as follows:
For each vertex $v_i \in V$, we create a gadget consisting of a clique of size $s_i = M - w(v_i)$. Let $U_i$ denote the set of vertices in the gadget corresponding to $v_i$. The vertex set of $G_W$ is $V' = \bigcup_{i=1}^n U_i$.
The edge set $E'$ is defined such that:
1. For each $i$, $U_i$ induces a clique (all pairs within $U_i$ are connected).
2. For any $u \in U_i$ and $v \in U_j$ with $i \neq j$, the edge $(u, v)$ exists in $E'$ if and only if $(v_i, v_j) \in E$.

We set the target clique size for the oracle to be $K_W = kM - W$.

We now analyze the correspondence between cliques in $G$ and cliques in $G_W$.
Consider a clique $Q$ in $G_W$ of size $K_W$. Let $S = \{v_i \in V \mid Q \cap U_i \neq \emptyset\}$ be the set of vertices in $G$ corresponding to the gadgets involved in $Q$. Since vertices in distinct gadgets $U_i, U_j$ are connected in $G_W$ only if $(v_i, v_j) \in E$, $S$ must form a clique in $G$.
The maximum size of a clique in $G_W$ formed using vertices from a specific set of gadgets $\{U_i\}_{v_i \in S}$ is $\sum_{v_i \in S} |U_i| = |S|M - w(S)$.
If $|S| < k$, the maximum size is bounded by $(k-1)M$. Since $W \le 2nk < M$, we have $K_W = kM - W > (k-1)M$. Thus, any clique of size $K_W$ must involve at least $k$ gadgets.
If $|S| > k$, the minimum size of a union of $k+1$ gadgets is $(k+1)(M - 2n) = kM + M - 2n(k+1)$. Given $M = 2nk + 1$, this size strictly exceeds $K_W$. Thus, any clique of size $K_W$ must involve exactly $k$ gadgets.

Let $S$ be a clique of size $k$ in $G$. The union of the corresponding gadgets $\bigcup_{v_i \in S} U_i$ forms a clique in $G_W$ of size $L_S = kM - w(S)$.
We consider three cases regarding the weight $w(S)$ relative to the target $W$:
1. If $w(S) > W$, then $L_S < K_W$. The gadgets corresponding to $S$ cannot contain a clique of size $K_W$.
2. If $w(S) < W$, then $L_S > K_W$. The clique formed by the full gadgets has size strictly greater than $K_W$. Consequently, it contains $\binom{L_S}{K_W}$ sub-cliques of size $K_W$. Since $L_S > K_W$, $\binom{L_S}{K_W} > 1$. Thus, the existence of a clique $S$ with weight strictly less than $W$ implies that $G_W$ contains multiple cliques of size $K_W$, causing the \textbf{UNIQUE CLIQUE} oracle to return NO.
3. If $w(S) = W$, then $L_S = K_W$. The union of the gadgets is exactly a clique of size $K_W$. Since any clique of size $K_W$ must involve exactly $k$ gadgets, and any proper subset of a union of $k$ gadgets has size strictly less than $K_W$ (unless it is a subset of a larger union, covered by case 2), a clique of size $K_W$ in $G_W$ corresponds uniquely to a clique of size $k$ and weight $W$ in $G$ (specifically, it is the union of the full gadgets).

The algorithm proceeds as follows:
1. Assign random weights $w(v)$.
2. For each $W \in \{k, \dots, 2nk\}$:
    a. Construct $G_W$ and $K_W$.
    b. Query \textbf{UNIQUE CLIQUE} on $(G_W, K_W)$.
    c. If the oracle returns YES, terminate and output YES.
3. If the loop finishes without a YES, output NO.

Correctness analysis:
If $G$ has no $k$-clique, no such $S$ exists for any $W$, and the oracle always returns NO. The algorithm outputs NO.
If $G$ has a $k$-clique, with probability $\ge 1/2$, there is a unique minimum weight clique $C^*$ with weight $W_{min}$.
When the loop reaches $W = W_{min}$:
- There are no cliques with weight $w(S) < W_{min}$ (by definition of minimum). Thus, no "heavy" cliques in $G_W$ generate multiple sub-cliques of size $K_W$.
- There is exactly one clique $C^*$ with weight $W_{min}$. This corresponds to exactly one clique of size $K_W$ in $G_W$ (the union of gadgets for $C^*$).
- Cliques with weight $w(S) > W_{min}$ result in cliques in $G_W$ smaller than $K_W$.
Therefore, for $W = W_{min}$, $G_W$ contains exactly one clique of size $K_W$. The oracle returns YES, and the algorithm correctly outputs YES.

This constitutes a randomized polynomial time reduction.
\end{proof}

\subsubsection{Claude}
\begin{proof}
We prove that $\mathsf{CLIQUE}$ reduces to $\mathsf{UNIQUE\ CLIQUE}$ via a polynomial-time many-one reduction. Let $(G, k)$ be an instance of $\mathsf{CLIQUE}$ where $G = (V, E)$ is a graph with $n = |V|$ vertices and $k$ is a positive integer.

We construct an instance $(G', k')$ of $\mathsf{UNIQUE\ CLIQUE}$ as follows. Let $m = n + k$ and define $G' = (V', E')$ where $V' = V \cup V_1 \cup V_2 \cup \cdots \cup V_m$ with each $V_i$ being a disjoint copy of $V$ for $i = 1, 2, \ldots, m$. For each $i \in \{1, 2, \ldots, m\}$, let $\phi_i: V \to V_i$ denote the natural bijection mapping vertices in $V$ to their copies in $V_i$.

The edge set $E'$ is defined as follows. For each edge $\{u, v\} \in E$ and each $i \in \{1, 2, \ldots, m\}$, we include the edge $\{\phi_i(u), \phi_i(v)\}$ in $E'$. Additionally, for all distinct $i, j \in \{1, 2, \ldots, m\}$ and all $u, v \in V$, we include the edge $\{\phi_i(u), \phi_j(v)\}$ in $E'$. That is, $G'$ consists of $m + 1$ disjoint copies of $G$ that are made into a complete $(m+1)$-partite graph. Set $k' = (m+1)k$.

This construction is clearly computable in polynomial time since $m = n + k$ and the graph $G'$ has $(m+1)n$ vertices and at most $(m+1)^2 n^2$ edges.

We now prove the correctness of the reduction by establishing that $(G, k) \in \mathsf{CLIQUE}$ if and only if $(G', k') \in \mathsf{UNIQUE\ CLIQUE}$.

First, suppose $(G, k) \in \mathsf{CLIQUE}$, so $G$ contains a clique $C \subseteq V$ of size $k$. We claim that $G'$ has exactly one clique of size $k' = (m+1)k$. 

Consider the set $C' = \bigcup_{i=0}^{m} \phi_i(C)$ where we define $\phi_0$ to be the identity map on $V$. Since $|C| = k$ and the sets $\phi_i(C)$ are pairwise disjoint for $i = 0, 1, \ldots, m$, we have $|C'| = (m+1)k = k'$. To verify that $C'$ is a clique, consider any two distinct vertices $\phi_i(u)$ and $\phi_j(v)$ in $C'$ where $u, v \in C$. If $i = j$, then since $C$ is a clique in $G$, the edge $\{u, v\} \in E$, and by construction $\{\phi_i(u), \phi_i(v)\} \in E'$. If $i \neq j$, then by construction $\{\phi_i(u), \phi_j(v)\} \in E'$. Thus $C'$ is a clique of size $k'$ in $G'$.

To show uniqueness, suppose $D$ is any clique of size $k'$ in $G'$. For each $i \in \{0, 1, \ldots, m\}$, let $D_i = D \cap \phi_i(V)$. Since the sets $\phi_i(V)$ partition $V'$ and $|D| = (m+1)k$, by the pigeonhole principle there exists some $i$ such that $|D_i| \geq k$. However, if $|D_i| > k$ for some $i$, then $\phi_i^{-1}(D_i)$ would be a clique in $G$ of size greater than $k$. Since $\sum_{i=0}^{m} |D_i| = (m+1)k$ and there are $m + 1$ parts, we must have $|D_i| = k$ for all $i \in \{0, 1, \ldots, m\}$. Moreover, each $\phi_i^{-1}(D_i)$ must be a clique of size $k$ in $G$, which means $\phi_i^{-1}(D_i) = C$ for all $i$ since $G$ has at most $\binom{n}{k}$ cliques of size $k$ and $m = n + k \geq \binom{n}{k}$ for $k \geq 1$ and $n \geq k$. Actually, we must be more careful: any clique of size $k'$ in $G'$ must have exactly $k$ vertices from each part $\phi_i(V)$, and these must correspond to cliques in $G$. If $G$ has $t$ distinct cliques of size $k$, then $G'$ has exactly $t^{m+1}$ cliques of size $k'$. Since $m + 1 = n + k + 1 > 1$, we have $t^{m+1} = 1$ if and only if $t = 1$. Therefore, $G'$ has exactly one clique of size $k'$ if and only if $G$ has exactly one clique of size $k$.

Since we assumed $G$ has at least one clique of size $k$, if $G$ has multiple such cliques, then $G'$ has multiple cliques of size $k'$, contradicting $(G', k') \in \mathsf{UNIQUE\ CLIQUE}$. Thus $G$ must have exactly one clique of size $k$, and $(G', k') \in \mathsf{UNIQUE\ CLIQUE}$.

Conversely, suppose $(G', k') \in \mathsf{UNIQUE\ CLIQUE}$. Then $G'$ has exactly one clique of size $k'$. Any such clique must select exactly $k$ vertices from each of the $m + 1$ parts $\phi_i(V)$, and these $k$ vertices must form a clique when mapped back to $G$. Since there is exactly one clique of size $k'$ in $G'$ and it has the form $\bigcup_{i=0}^{m} \phi_i(C)$ for some $C \subseteq V$ with $|C| = k$, we conclude that $C$ is a clique in $G$ and $(G, k) \in \mathsf{CLIQUE}$.

Therefore, $\mathsf{CLIQUE} \leq_p \mathsf{UNIQUE\ CLIQUE}$.
\end{proof}

\subsection{Problem 24}

\subsubsection{Problem Statement}
\begin{theorem}
Let $G = (V, E)$ be an unweighted, undirected graph with $n$ vertices and $m$ edges. For a weight function $w : E \to \{0, 1, \ldots, W\}$, the length of a path in $G$ is defined as the sum of the weights of the edges in that path. A weight function $w$ is said to be \emph{good} if the following two conditions hold for each vertex $x \in V$:
\begin{enumerate}
\item For all vertices $y \in V$, the shortest path from $x$ to $y$ is unique.
\item For any pair of vertices $y, z \in V$, the net weight of the shortest path from $x$ to $y$ is different from the net weight of the shortest path from $x$ to $z$.
\end{enumerate}
The smallest value of $W$ (as a function of $n$ and $m$) for which the existence of a good weight assignment is guaranteed is $W = \Theta(nm)$.
\end{theorem}

\subsubsection{GPT}
\begin{proof}
We are tasked with determining the smallest value of $W$ such that for an unweighted, undirected graph $G(V, E)$ with $n$ vertices and $m$ edges, and a weight function $w : E \to \{0, \dots, W\}$, there exists a "good" weight assignment. A weight function is said to be good if it satisfies the following conditions:

1. For all vertices $y \in V$, the shortest path from a fixed vertex $x \in V$ to $y$ is unique.
2. For any pair of vertices $y, z \in V$, the net weight of the shortest path from $x$ to $y$ is different from the net weight of the shortest path from $x$ to $z$.

 Step 1: The uniqueness condition

The first condition requires that for each vertex $x \in V$, the shortest path from $x$ to any vertex $y \in V$ be unique. If the shortest path from $x$ to $y$ were not unique, there would be two distinct paths $P_1$ and $P_2$ such that both have the same length. If these paths have different edge sets, then the weight function must assign the same total weight to both paths, which would contradict the requirement of uniqueness. 

To ensure uniqueness of shortest paths, it is necessary that the weight function $w$ assign distinct weights to edges in such a way that no two distinct paths between any pair of vertices have the same total weight. This condition is satisfied if $W$ is sufficiently large, as we will demonstrate in the next step.

 Step 2: The distinctness condition

The second condition requires that for any two vertices $y, z \in V$, the net weight of the shortest path from $x$ to $y$ must be different from the net weight of the shortest path from $x$ to $z$. That is, for every pair of vertices $y, z$, we must have:

\[
\text{weight}(P_x(y)) \neq \text{weight}(P_x(z)),
\]

where $P_x(y)$ and $P_x(z)$ denote the shortest paths from $x$ to $y$ and from $x$ to $z$, respectively. This condition ensures that the total weight of the shortest paths to different vertices is distinct.

 Step 3: Determining the smallest value of $W$

To guarantee that both conditions hold, we need to ensure that the set of all shortest path weights from $x$ to any vertex is injective (i.e., no two different vertices have the same shortest path weight from $x$). The number of distinct possible shortest path weights is determined by the number of distinct paths in the graph and the possible assignments of weights to the edges.

In a graph with $n$ vertices, there are at most $n-1$ distinct shortest paths emanating from any given vertex $x$ (since the graph is connected and each path leads to a different vertex). Furthermore, since there are $m$ edges, the number of distinct edge weight assignments is bounded by the number of ways to assign weights to these edges, which is $W+1$ choices per edge. 

We need to assign distinct weights to these $n-1$ shortest paths in a way that ensures their total weights are distinct. By the pigeonhole principle, to guarantee distinctness of the total weights of the shortest paths, we need at least $n-1$ distinct total path weights. Each path weight is a sum of edge weights, so we require at least $n-1$ distinct sums.

To achieve this, the total number of distinct sums (which corresponds to the number of distinct path weights) should be at least $n-1$. Since the weight of a path is a sum of weights assigned to edges, and each edge weight is an integer between $0$ and $W$, the total number of distinct path weights is determined by the number of distinct sums that can be formed from the weights of the edges. Therefore, to ensure that there are at least $n-1$ distinct path weights, we need $W$ to be at least $\lceil \log_2(n-1) \rceil$.

 Conclusion

Thus, the smallest value of $W$ that guarantees the existence of a good weight assignment is $W = \lceil \log_2(n-1) \rceil$. This ensures that both conditions of the "good" weight function are satisfied.
\end{proof}

\subsubsection{Grok}

\subsubsection{Gemini}

\subsubsection{Claude}
\begin{proof}
We prove that the smallest value of $W$ for which a good weight assignment is guaranteed to exist is $W = \Theta(nm)$ by establishing both a lower bound and an upper bound.

\textbf{Lower Bound: $W = \Omega(nm)$.}

We construct a family of graphs that requires $W \geq cnm$ for some constant $c > 0$ to guarantee the existence of a good weight assignment.

Consider a graph $G$ consisting of a central vertex $x$ connected to $n-1$ other vertices $v_1, v_2, \ldots, v_{n-1}$, where these $n-1$ vertices form a clique among themselves. This graph has $n$ vertices and $m = (n-1) + \binom{n-1}{2} = \frac{n(n-1)}{2} = \Theta(n^2)$ edges.

For this graph to have a good weight assignment with respect to vertex $x$, we need:
\begin{itemize}
\item The shortest path from $x$ to each $v_i$ must be unique. Since $x$ is directly connected to each $v_i$, the direct edge must be the shortest path, which requires $w(x, v_i) < w(x, v_j) + w(v_j, v_i)$ for all $j \neq i$.
\item All path weights from $x$ to distinct vertices must be distinct. Since there are $n-1$ such vertices (excluding $x$ itself), we need $n-1$ distinct values for the weights $w(x, v_1), w(x, v_2), \ldots, w(x, v_{n-1})$.
\end{itemize}

More generally, consider a path graph with $n$ vertices and $m = n-1$ edges. For a vertex $x$ at one end, the shortest paths to all other vertices have lengths $w_1, w_1 + w_2, w_1 + w_2 + w_3, \ldots$, where $w_i$ denotes the weight of the $i$-th edge. For these sums to all be distinct, we need careful assignment of edge weights.

However, to establish the lower bound rigorously, consider a graph with $n$ vertices and $m$ edges where multiple paths exist between pairs of vertices. For each fixed source vertex $x$, we must ensure that all $n$ vertices have distinct shortest path distances from $x$. Since there are at most $W+1$ possible distinct values for weights in the range $\{0, 1, \ldots, W\}$, and shortest paths are sums of at most $n-1$ edge weights, the maximum possible shortest path distance is at most $(n-1)W$.

For general graphs with $m$ edges, the number of distinct possible path weights that must be distinguished grows with both $n$ (the number of destinations) and $m$ (the number of edges creating alternative paths). In the worst case, when the graph is dense and has many alternative paths, we require $W = \Omega(nm)$ to ensure sufficient distinct values.

\textbf{Upper Bound: $W = O(nm)$.}

We now show that $W = O(nm)$ suffices by providing a probabilistic argument.

Assign each edge $e \in E$ a weight $w(e)$ chosen uniformly at random from $\{0, 1, \ldots, W\}$ where $W = Cnm$ for a sufficiently large constant $C$.

For a fixed vertex $x \in V$, let $P_{xy}$ denote the set of all paths from $x$ to $y$. For condition (1) to be satisfied, for each pair of distinct paths $p, p' \in P_{xy}$, we need $w(p) \neq w(p')$, where $w(p)$ denotes the sum of weights of edges in path $p$.

For condition (2) to be satisfied, for each pair of vertices $y \neq z$ and their respective shortest paths $p_{xy}$ and $p_{xz}$, we need $w(p_{xy}) \neq w(p_{xz})$.

The probability that two specific paths of length at most $n-1$ have the same total weight is at most $\frac{n}{W}$ (since the difference of two sums of at most $n-1$ terms from $\{0, \ldots, W\}$ equals zero with probability at most $\frac{n}{W}$).

The total number of pairs of paths to consider is at most $n^2 \cdot 2^{2m}$ (paths from one vertex to all others). By the union bound, the probability that any two paths that should have different weights actually have the same weight is at most:
$$\frac{n^2 \cdot 2^{2m} \cdot n}{W} = \frac{n^3 \cdot 2^{2m}}{Cnm} = \frac{n^2 \cdot 2^{2m}}{Cm}.$$

For sufficiently large $C$, this probability can be made less than 1, implying that there exists an assignment with $W = O(nm)$ that satisfies the required properties.

Therefore, combining both bounds, we conclude that $W = \Theta(nm)$.
\end{proof}

\subsection{Problem 25}

\subsubsection{Problem Statement}
\begin{theorem}[Exact Matching via Isolating Lemma]
Let $G = (V, E)$ be a graph with a subset of edges $R \subseteq E$ colored red, and let $k$ be a positive integer. The \emph{Exact Matching} problem asks whether there exists a perfect matching in $G$ using exactly $k$ red edges. This problem is not known to be in $\mathcal{P}$, but can be shown to be in $\mathcal{RNC}$ via a non-trivial application of the Isolating Lemma.
\end{theorem}

\subsubsection{GPT}
\begin{proof}
Let $G = (V, E)$ be a graph with a subset of edges $R \subseteq E$ colored red, and let $k$ be a positive integer. We are tasked with showing that the Exact Matching problem, which asks whether there exists a perfect matching in $G$ using exactly $k$ red edges, can be solved in $\mathsf{RNC}$, the class of randomized parallel algorithms with polylogarithmic depth and polynomial size.

First, we define the Exact Matching problem more formally. A perfect matching in a graph $G$ is a set of edges such that each vertex of $G$ is incident to exactly one edge in the matching. The problem is constrained by the requirement that exactly $k$ edges in the matching must be from the subset $R$ of red edges. Thus, the task is to determine if there exists a perfect matching where exactly $k$ edges are colored red, and the remaining edges are from $E \setminus R$.

 Step 1: Application of the Isolating Lemma
To show that the Exact Matching problem is in $\mathsf{RNC}$, we apply the Isolating Lemma. The Isolating Lemma provides a technique for isolating solutions to a combinatorial problem by applying a carefully designed randomization process. Specifically, the Isolating Lemma can be used to ensure that with high probability, a solution (if it exists) can be isolated in a small number of parallel steps.

We proceed by using the Isolating Lemma in the context of the Exact Matching problem as follows. Consider a random assignment of colors to the edges of $G$. We assign each edge in $R$ independently a color (e.g., red or non-red) with probability $1/2$. This randomization process results in a distribution of edge colorings, and the key observation is that for any perfect matching in $G$, the probability that exactly $k$ of its edges are colored red is non-zero. Therefore, with high probability, the process will isolate an edge coloring where exactly $k$ red edges participate in a perfect matching.

 Step 2: Parallelism and Randomization
By applying the Isolating Lemma, we can isolate a valid perfect matching with exactly $k$ red edges in a randomized parallel manner. The randomized nature of the Isolating Lemma guarantees that the number of parallel steps required to isolate a solution is polylogarithmic in the size of the graph, and the overall number of steps required is polynomial in the size of the graph.

To verify the correctness of the solution, we check each isolated configuration to ensure that the matching is indeed perfect and that exactly $k$ edges are red. This verification step can be performed in parallel as well, leading to a solution in $\mathsf{RNC}$.

 Step 3: Decision and Search Versions
The same approach applies to both the decision and search versions of the Exact Matching problem. For the decision version, we simply check if the isolated configuration exists, while for the search version, we also identify and output the perfect matching that uses exactly $k$ red edges. In both cases, the Isolating Lemma ensures that the problem can be solved in parallel with polylogarithmic depth, and thus the problem belongs to $\mathsf{RNC}$.

 Conclusion
Since the Exact Matching problem can be solved in $\mathsf{RNC}$ using the Isolating Lemma, we conclude that the problem is in $\mathsf{RNC}$, and efficient randomized parallel algorithms exist for both the decision and search versions of the problem.

\end{proof}

\subsubsection{Grok}

\subsubsection{Gemini}
\begin{proof}
Let $G = (V, E)$ be the given graph with $|V| = n$. If $n$ is odd, no perfect matching exists, and the answer is trivially no. Assume $n$ is even. Let $\mathcal{M}$ denote the set of all perfect matchings in $G$. For any perfect matching $M \in \mathcal{M}$, let $R(M) = M \cap R$ denote the set of red edges in the matching. We define the family of target matchings $\mathcal{F}_k = \{M \in \mathcal{M} : |R(M)| = k\}$. The problem is to determine if $\mathcal{F}_k \neq \emptyset$.

We assign a weight $w_e$ to each edge $e \in E$, chosen uniformly and independently at random from the set $\{1, \dots, 2|E|\}$. For any matching $M$, we define the weight of the matching as $w(M) = \sum_{e \in M} w_e$. The Isolating Lemma states that for a set system over a ground set $E$, if elements are assigned random weights from a range of size $2|E|$, the minimum weight set in the system is unique with probability at least $1/2$. We apply this lemma to the family $\mathcal{F}_k$. If $\mathcal{F}_k$ is non-empty, then with probability at least $1/2$, there exists a unique matching $M^* \in \mathcal{F}_k$ such that $w(M^*) < w(M)$ for all $M \in \mathcal{F}_k \setminus \{M^*\}$.

We construct a skew-symmetric matrix $A(x)$ of size $n \times n$, where the entries are polynomials in an indeterminate variable $x$. For each edge $e = \{i, j\} \in E$ with $i < j$, we define the entry $A_{ij}$ as follows:
\[
A_{ij} = \begin{cases} 
2^{w_{ij}} x & \text{if } \{i, j\} \in R, \\
2^{w_{ij}} & \text{if } \{i, j\} \in E \setminus R, \\
0 & \text{otherwise.}
\end{cases}
\]
We set $A_{ji} = -A_{ij}$. Since $A(x)$ is skew-symmetric, its determinant is the square of its Pfaffian, denoted $\text{Pf}(A(x))$. The Pfaffian is given by the sum over all perfect matchings:
\[
\text{Pf}(A(x)) = \sum_{M \in \mathcal{M}} \text{sgn}(M) \prod_{\{i, j\} \in M, i < j} A_{ij} = \sum_{M \in \mathcal{M}} \text{sgn}(M) 2^{w(M)} x^{|R(M)|}.
\]
We can group the terms by the number of red edges. Let $C_j$ be the coefficient of $x^j$ in the polynomial $\text{Pf}(A(x))$. Then:
\[
C_k = \sum_{M \in \mathcal{F}_k} \text{sgn}(M) 2^{w(M)}.
\]
If $\mathcal{F}_k = \emptyset$, then $C_k = 0$. If $\mathcal{F}_k \neq \emptyset$, let $M^*$ be the unique minimum weight matching in $\mathcal{F}_k$ provided by the Isolating Lemma. We can rewrite $C_k$ as:
\[
C_k = \text{sgn}(M^*) 2^{w(M^*)} + \sum_{M \in \mathcal{F}_k \setminus \{M^*\}} \text{sgn}(M) 2^{w(M)}.
\]
Since $w(M) > w(M^*)$ for all other $M$, every term in the summation is divisible by $2^{w(M^*) + 1}$. Therefore, the sum $\sum_{M \neq M^*} \text{sgn}(M) 2^{w(M)}$ is divisible by $2^{w(M^*) + 1}$. The first term, $\pm 2^{w(M^*)}$, is not divisible by $2^{w(M^*) + 1}$. Consequently, $C_k \neq 0$.

The determinant of $A(x)$ is $\det(A(x)) = (\text{Pf}(A(x)))^2$. If $C_k \neq 0$, then the coefficient of $x^{2k}$ in $\det(A(x))$ will be non-zero (specifically, the term with the lowest power of 2 in the expansion of the coefficient of $x^{2k}$ will correspond to $(C_k)^2$ if $k$ is the unique index of interest, or generally, the polynomial is not identically zero).

To solve the problem in $\mathsf{RNC}$, we perform the following steps:
1. Assign random weights $w_e \in \{1, \dots, 2|E|\}$.
2. Construct the matrix $A(x)$. The maximum degree of $x$ is $n/2$.
3. Compute the determinant $\det(A(x))$. Since the entries involve values up to $2^{n \cdot 2|E|}$, the bit-length of the numbers is polynomial in $n$. The determinant of a matrix with polynomial-size entries can be computed in $\mathsf{NC}^2$ (e.g., using Csanky's algorithm or Berkowitz's algorithm). We can compute the values of $\det(A(x))$ at $n+1$ distinct points in parallel and interpolate to find the coefficients of the polynomial.
4. Check if the coefficient of $x^{2k}$ is non-zero.

If the coefficient is non-zero, we output "Yes"; otherwise, "No". The error probability is bounded by the failure of the Isolating Lemma, which is at most $1/2$. This places the problem in $\mathsf{RNC}$.
\end{proof}

\subsubsection{Claude}
\begin{proof}
We prove that the Exact Matching problem is in $\mathcal{RNC}^2$ by applying the Isolating Lemma with a carefully constructed weight function.

Let $G = (V, E)$ be a graph with $|V| = n$ vertices and edge subset $R \subseteq E$ colored red. We seek a perfect matching using exactly $k$ red edges.

\textbf{Weight Assignment:} We assign weights to edges in two components. For each edge $e \in E$, define $w(e) = w_1(e) + M \cdot w_2(e)$, where $M = n^2 + 1$ is a sufficiently large constant, and:
\begin{itemize}
\item $w_1(e)$ is chosen uniformly at random from $\{1, 2, \ldots, 2n\}$
\item $w_2(e) = 1$ if $e \in R$ (red edge), and $w_2(e) = 0$ otherwise
\end{itemize}

For any perfect matching $\mathcal{M}$, we have $w(\mathcal{M}) = \sum_{e \in \mathcal{M}} w(e) = \sum_{e \in \mathcal{M}} w_1(e) + M \cdot |\mathcal{M} \cap R|$.

\textbf{Key Observation:} Since $M = n^2 + 1 > n \cdot 2n = 2n^2$, the second component dominates. Specifically, for any two perfect matchings $\mathcal{M}_1$ and $\mathcal{M}_2$ with different numbers of red edges, we have $|\mathcal{M}_1 \cap R| \neq |\mathcal{M}_2 \cap R|$ implies $|w(\mathcal{M}_1) - w(\mathcal{M}_2)| \geq M - n \cdot 2n > 0$, ensuring they have distinct weights.

\textbf{Application of Isolating Lemma:} Let $\mathcal{F}_k$ denote the family of all perfect matchings in $G$ using exactly $k$ red edges. By the Isolating Lemma, with probability at least $1 - \frac{n^2}{2n} = 1 - \frac{n}{2}$, there exists a unique minimum-weight perfect matching $\mathcal{M}^*$ among all perfect matchings in $G$.

If $\mathcal{F}_k \neq \emptyset$, let $\mathcal{M}_k^*$ denote the minimum-weight matching in $\mathcal{F}_k$. By our weight construction, among all perfect matchings, those with exactly $k$ red edges form a distinct weight class. The minimum-weight matching overall will have weight $w(\mathcal{M}^*) = \min_{\mathcal{M}} w(\mathcal{M})$.

\textbf{Parallel Algorithm:} We execute the following steps in parallel:

(1) With the random weight assignment $w$, compute the minimum-weight perfect matching $\mathcal{M}^*$ using the parallel algorithm for weighted matching (which is in $\mathcal{NC}^2$ via Mulmuley-Vazirani-Vazirani or Karloff-Shmoys algorithms).

(2) Count the number of red edges in $\mathcal{M}^*$, denoted $r = |\mathcal{M}^* \cap R|$.

(3) Accept if and only if $r = k$.

\textbf{Correctness Analysis:} Suppose $\mathcal{F}_k \neq \emptyset$. By the Isolating Lemma, with probability at least $1 - \frac{n}{2}$, there is a unique minimum-weight matching $\mathcal{M}^*$. Among all perfect matchings, the weight function first groups them by the number of red edges (via the $M \cdot w_2$ component), then breaks ties within each group by the $w_1$ component. The minimum-weight matching in $\mathcal{F}_k$ has weight $\sum_{e \in \mathcal{M}_k^*} w_1(e) + M \cdot k$. By the Isolating Lemma applied to $\mathcal{F}_k$, this matching is unique among matchings with $k$ red edges with high probability.

For the global minimum-weight matching to have exactly $k$ red edges, it must be the case that $\mathcal{M}_k^*$ has smaller weight than any matching with $j \neq k$ red edges. This occurs if and only if $\mathcal{F}_k$ contains the global minimum-weight matching, which happens if and only if the minimum value of $\sum_{e \in \mathcal{M}} w_1(e) + M \cdot k$ over $\mathcal{M} \in \mathcal{F}_k$ is less than or equal to the minimum of $\sum_{e \in \mathcal{M}'} w_1(e) + M \cdot j$ over all $j \neq k$ and $\mathcal{M}' \in \mathcal{F}_j$.

However, we must verify that the algorithm correctly identifies existence. If $\mathcal{F}_k = \emptyset$, the algorithm correctly rejects. If $\mathcal{F}_k \neq \emptyset$, the probability of error is at most $\frac{n}{2}$ by the Isolating Lemma.

\textbf{Complexity:} The random weight assignment takes $O(\log n)$ time in parallel. Computing the minimum-weight perfect matching is in $\mathcal{NC}^2$. Counting red edges is in $\mathcal{NC}^1$. Therefore, the entire algorithm runs in $\mathcal{RNC}^2$ with polynomially many processors and error probability at most $\frac{n}{2} < \frac{1}{2}$ for sufficiently large $n$. For small $n$, we can amplify success probability by independent repetition.

Thus, Exact Matching is in $\mathcal{RNC}^2 \subseteq \mathcal{RNC}$.
\end{proof}

\subsection{Problem 26}

\subsubsection{Problem Statement}
\begin{theorem}
Let $p > 2$ denote the number of processors and $c > 2$ denote the number of choices. Then Algorithm \textsc{Asynch-Ccp} works equally well in the case where both $p$ and $c$ are greater than $2$. The complexity depends on the number of processors and choices according to the analysis of the algorithm's time and space requirements as functions of $p$ and $c$.
\end{theorem}

\subsubsection{GPT}
\begin{proof}
We are tasked with proving that the algorithm ASYNCH-CCP works equally well when the number of processors, $n$, and the number of choices, $k$, are both greater than 2, and that its time and message complexity are $O(nk)$ in expectation.

Let us denote by $n$ the number of processors and by $k$ the number of choices available at each processor. The key property of the ASYNCH-CCP algorithm is that it performs a consensus computation where each processor must independently choose a value from the set of $k$ choices, with the goal of reaching a common decision in a distributed manner. In this setting, the processors communicate asynchronously and may make decisions at different times.

 Time Complexity
The time complexity of the algorithm depends on the number of steps required for the processors to reach a consensus. Since the algorithm is asynchronous, each processor can independently communicate and update its state based on the messages it receives from other processors. However, at each step, only a small subset of the processors can communicate with each other.

In the worst case, for each processor, the number of communication steps required to reach a consensus will scale linearly with the number of processors $n$, as each processor may need to wait for messages from all other processors. Furthermore, since each processor has $k$ choices, the decision process may involve $O(k)$ steps to select a choice and adjust accordingly. Therefore, the overall time complexity is dominated by the product of the number of processors and the number of choices, which gives $O(nk)$ time.

 Message Complexity
The message complexity of the algorithm corresponds to the number of messages exchanged between processors in order to reach a consensus. At each step, each processor may send a message to any other processor to communicate its state or decision. In the worst case, each processor may need to send a message to every other processor, resulting in $O(n)$ messages per processor.

Since there are $k$ choices, and each choice may require a separate communication step to update the processors' states, the total message complexity is proportional to the number of processors and the number of choices, which gives $O(nk)$ messages in expectation.

 Conclusion
We have shown that the time complexity of the ASYNCH-CCP algorithm is $O(nk)$, and the message complexity is also $O(nk)$, where $n$ is the number of processors and $k$ is the number of choices. Thus, we conclude that the algorithm works equally well in the case where both $n > 2$ and $k > 2$, and that its complexity is as stated in the theorem.

\end{proof}

\subsubsection{Grok}

\subsubsection{Gemini}
\begin{proof}
Let $\mathcal{P} = \{P_1, \dots, P_n\}$ be the set of $n$ asynchronous processors and $\mathcal{C} = \{1, \dots, k\}$ be the set of $k$ choices, where $n, k > 2$. We assume the standard model of asynchronous shared memory with atomic test-and-set instructions. Let $R_1, \dots, R_k$ be shared binary variables corresponding to the choices, initialized to 0. Let $D$ be a shared variable representing the final decision, initialized to $\bot$.

The Algorithm ASYNCH-CCP for a processor $P_i$ is defined as follows:
\begin{enumerate}
    \item[1.] Check $D$. If $D \neq \bot$, output $D$ and terminate.
    \item[2.] Choose a value $u \in \{1, \dots, k\}$ uniformly at random.
    \item[3.] Execute $test\_and\_set(R_u)$.
    \item[4.] If the result is 0 (success), set $D \leftarrow u$, output $u$, and terminate.
    \item[5.] If the result is 1 (failure), go to step 1.
\end{enumerate}

\noindent \textbf{Correctness (Safety and Agreement):}
We first show that the algorithm satisfies the safety property: all processors that terminate agree on the same choice. The test-and-set instruction is atomic. Therefore, for any $j \in \{1, \dots, k\}$, the transition of $R_j$ from 0 to 1 can occur exactly once. Consequently, step 4 can be executed at most once for each choice index $u$. Furthermore, the variable $D$ is only written in step 4. Let $P_{win}$ be the first processor to successfully execute step 4 for some choice $u^*$. $P_{win}$ sets $D = u^*$. Any subsequent processor $P_j$ checking $D$ in step 1 will read $u^*$ and terminate with that choice. Since $D$ is never overwritten once set (assuming a write-once policy or that the first write propagates before others, or simply that the logic implies the first successful test-and-set dictates the value), agreement is guaranteed.

\noindent \textbf{Liveness and Complexity:}
We now analyze the termination and complexity. The complexity is measured by the total number of shared memory accesses performed by all processors until agreement is reached.

Let $t$ denote a time step in the system execution. Let $\mathcal{H}$ be the history of the execution managed by an adversary scheduler. At any step where $D = \bot$, let $U \subseteq \{1, \dots, k\}$ be the set of indices $j$ such that $R_j = 0$. Initially $|U| = k$. A processor $P_i$ succeeds in establishing a choice if it selects $u \in U$ and executes step 3 before any other processor sets $R_u$.

Consider a simplified analysis of the probability of success. In each iteration of the loop, a processor $P_i$ chooses $u \in \{1, \dots, k\}$ with probability $1/k$.
Let $E$ be the event that the algorithm terminates. Termination is guaranteed if any processor successfully executes step 4.
Suppose there are $m$ active processors. In a single logical round where each active processor makes one choice, the probability that a specific processor $P_i$ fails to select a specific unvisited choice $u \in U$ is $1 - 1/k$.
The probability that $P_i$ selects \textit{any} $u \in U$ is $|U|/k$. Since $|U| \ge 1$ (as long as no decision has been made, at least one register is free, or $D$ is already set), the probability of a successful selection by $P_i$ in one step is $p \ge 1/k$.

Let $X$ be the random variable representing the number of accesses to shared memory until a decision is made. We model the trials as a sequence of Bernoulli trials with success probability $p$.
The probability that a decision is \textit{not} reached after $T$ total attempts by the processors is bounded by:
\[
\mathbb{P}(X > T) \leq (1 - p)^T \leq \left(1 - \frac{1}{k}\right)^T
\]
Since $1/k > 0$, $\lim_{T \to \infty} \left(1 - \frac{1}{k}\right)^T = 0$. Thus, the algorithm terminates with probability 1.

To determine the complexity dependence on $n$ and $k$, we calculate the expected number of steps.
Let $N_{steps}$ be the total number of test-and-set operations performed by all $n$ processors.
The expectation is bounded by the geometric distribution with parameter $p = 1/k$ (in the worst case where only one choice remains):
\[
\mathbb{E}[N_{steps}] \approx \sum_{t=1}^{\infty} t \cdot p(1-p)^{t-1} = \frac{1}{p} = k
\]
However, with $n$ processors acting concurrently, we must account for contention. If $n$ processors act in parallel, the probability that \textit{at least one} succeeds in a round is:
\[
P_{round} = 1 - \left(1 - \frac{1}{k}\right)^n
\]
The expected number of rounds $R$ is $1/P_{round}$. The total work (complexity) $C(n, k)$ is roughly $n \times \mathbb{E}[R]$.
\[
C(n, k) \approx \frac{n}{1 - (1 - 1/k)^n}
\]
We analyze the asymptotic behavior.
Case 1: $n \ll k$. Using the approximation $(1-x)^n \approx 1-nx$, we have $1 - (1 - 1/k)^n \approx n/k$.
\[
C(n, k) \approx \frac{n}{n/k} = k
\]
Case 2: $n \gg k$. The term $(1 - 1/k)^n$ approaches 0. Thus $P_{round} \approx 1$.
\[
C(n, k) \approx n
\]
In the general case, the exact complexity is bounded by $O(n + k)$.
Specifically, Rabin's analysis demonstrates that for $n, k > 2$, the expected number of shared variable accesses is bounded by $c \cdot n$ for some constant $c$ when $k$ is comparable to $n$, or scales with $k$ when $k$ is large.

Thus, the algorithm works for any $n, k > 2$, and the complexity is a function $f(n, k)$, establishing the theorem.
\end{proof}

\subsubsection{Claude}
\begin{proof}
We establish that Algorithm \textsc{Asynch-Ccp} maintains correctness and achieves the claimed complexity bounds when both $p > 2$ and $c > 2$.

The proof proceeds by analyzing the algorithm's invariants, termination conditions, and resource requirements as functions of the parameters $p$ and $c$.

First, we verify that the core algorithmic structure generalizes from the base cases. The algorithm maintains a shared state space of size $O(p \cdot c)$ where each processor can access choice variables. For arbitrary $p > 2$ and $c > 2$, the data structure storing processor-choice assignments requires $O(p \cdot c)$ space, since each of the $p$ processors may examine up to $c$ choices.

Next, we analyze the synchronization mechanism. The algorithm employs lock-free reads and atomic writes to choice variables. For any processor $i \in \{1, 2, \ldots, p\}$, the sampling procedure examines a subset of choices. When processor $i$ samples $c$ choices uniformly at random, the probability that it finds an available choice depends on the load factor $\lambda = p/c$. By the balls-and-bins analysis, when $c > p$, the expected number of collisions is $O(p^2/c)$, which decreases as $c$ increases beyond $2$.

We now establish termination. Each processor performs at most $O(\log p)$ iterations in expectation. In each iteration, processor $i$ samples $c$ random choices and selects the least loaded among them. The contention resolution mechanism ensures that if multiple processors attempt to claim the same choice simultaneously, at least one succeeds. Since $c > 2$ provides sufficient choice diversity, the expected waiting time per processor is $O(\log p)$ when $c = \Theta(p)$, and improves to $O(1)$ when $c = \omega(p)$.

For the time complexity, we analyze the work performed by each processor. Reading $c$ choices requires $O(c)$ time. The comparison and selection step requires $O(c)$ time. The atomic update operation requires $O(1)$ time. Thus, each iteration requires $O(c)$ time per processor. Over $O(\log p)$ expected iterations, the total time complexity per processor is $O(c \log p)$. When all $p$ processors execute concurrently, the parallel time complexity is $O(c \log p)$ with high probability.

The space complexity analysis follows directly. The shared memory stores $O(c)$ choice variables, each tracking load information requiring $O(\log p)$ bits. Each processor maintains local state of size $O(c)$ to store sampled choices. Therefore, the total space requirement is $O(c \cdot p + c \log p) = O(pc)$ for $p, c > 2$.

To verify correctness for general $p, c > 2$, we must show that the algorithm satisfies three properties: safety (no two processors select the same choice unless capacity permits), liveness (every processor eventually selects a choice), and load balance (choices are approximately evenly distributed).

Safety holds because atomic operations ensure mutual exclusion at the granularity of individual choice updates. When processor $i$ commits to choice $j$, the atomic increment operation prevents race conditions regardless of the values of $p$ and $c$.

Liveness follows from the probabilistic argument. With $c > 2$ choices and $p > 2$ processors, the probability that all $c$ choices sampled by a processor are fully occupied decreases exponentially with the number of attempts. Specifically, if $c \geq \alpha p$ for constant $\alpha > 1$, then after $O(\log p)$ attempts, each processor succeeds with probability at least $1 - 1/p^{\beta}$ for some constant $\beta > 0$.

Load balance is achieved through the power-of-$c$-choices paradigm. By selecting the least loaded among $c$ randomly sampled choices, the maximum load is $O(\log \log p / \log c + p/c)$ with high probability when $c \geq 2$. For $c > 2$, this bound remains valid and tightens as $c$ increases.

Finally, we observe that the analysis for $p = 2$ or $c = 2$ relies on no special properties that fail to hold for larger values. The probabilistic bounds, atomic operation semantics, and space requirements scale smoothly with both parameters. The algorithm's asynchronous nature, combined with the randomized choice mechanism, ensures that performance degrades gracefully rather than catastrophically as $p$ and $c$ increase beyond $2$.

Therefore, Algorithm \textsc{Asynch-Ccp} works correctly for all $p > 2$ and $c > 2$, with time complexity $O(c \log p)$ and space complexity $O(pc)$.
\end{proof}

\subsection{Problem 27}

\subsubsection{Problem Statement}
\begin{theorem}
Let $L$, $H$, and $G$ be parameters of the ByzGen algorithm. The algorithm can tolerate values of $t$ satisfying $t < \min\left\{\frac{L}{2}, \frac{H}{3}, \frac{G}{3}\right\}$, where $t$ is the number of Byzantine nodes.
\end{theorem}

\subsubsection{GPT}
\begin{proof}
We are given the ByzGen algorithm with parameters $L$, $H$, and $G$, and the claim that it can tolerate arbitrarily large values of $t$ (the number of Byzantine nodes), provided that the parameters $L$, $H$, and $G$ are modified appropriately. We will prove this by showing that the algorithm's ability to handle Byzantine faults is influenced by the relationship between $t$ and the parameters $L$, $H$, and $G$.

First, recall that the ByzGen algorithm operates in a distributed system with $n$ nodes, where $t$ of these nodes are Byzantine. The algorithm must be designed in such a way that even in the presence of these Byzantine nodes, the system can reach a correct consensus.

The parameters $L$, $H$, and $G$ represent specific thresholds in the algorithm, where:
- $L$ is the number of correct nodes required to ensure progress in the algorithm.
- $H$ is the number of faulty or Byzantine nodes that the algorithm is designed to handle without compromising correctness.
- $G$ is the global threshold that governs the overall behavior of the system, ensuring that the number of Byzantine nodes does not exceed a certain proportion of the total nodes.

Let $n$ denote the total number of nodes in the system, and let $t$ denote the number of Byzantine nodes. The relationship between these quantities must satisfy the following condition for the algorithm to function correctly:
\[
L + t \leq n.
\]
This ensures that there are enough correct nodes to reach a decision, even in the presence of Byzantine failures.

To handle arbitrarily large values of $t$, we adjust the parameters $L$, $H$, and $G$ accordingly. Specifically, as $t$ increases, the number of correct nodes must also increase to maintain the balance between faulty nodes and correct nodes. Therefore, we modify $L$, $H$, and $G$ such that:
\[
L = f(t), \quad H = g(t), \quad G = h(t),
\]
where $f(t)$, $g(t)$, and $h(t)$ are functions that increase with $t$. As $t$ increases, these functions grow in a way that ensures that the number of correct nodes remains sufficient to overcome the Byzantine faults.

For large values of $t$, the parameter $L$ must grow large enough to ensure that there are enough non-faulty nodes to reach consensus. This can be achieved by setting $L$ to a value that is proportional to the total number of nodes $n$, specifically:
\[
L \geq \frac{n}{2} + t.
\]
This ensures that even with an arbitrarily large number of Byzantine nodes, the remaining correct nodes (i.e., $n - t$) are still sufficient to guarantee a correct outcome.

Similarly, the parameter $H$ must be chosen such that the number of faulty nodes $t$ does not exceed the system's tolerance. The function $g(t)$ must satisfy:
\[
H \geq t,
\]
ensuring that the algorithm can handle up to $t$ Byzantine nodes without failing.

Finally, the global threshold $G$ should be adjusted in a way that allows the system to handle increasingly larger values of $t$. This can be achieved by setting:
\[
G = h(t) \geq n - t,
\]
which guarantees that even with a large number of Byzantine nodes, the system will still function correctly by having enough correct nodes to reach a consensus.

Thus, by appropriately adjusting the parameters $L$, $H$, and $G$, the ByzGen algorithm can tolerate arbitrarily large values of $t$, as long as these parameters are modified according to the above conditions. This concludes the proof.
\end{proof}

\subsubsection{Grok}

\subsubsection{Gemini}
\begin{proof}
Let $N$ denote the total number of nodes in the system. The ByzGen algorithm organizes these nodes into a communication structure, specifically a tree, defined by the branching factor $G$ and the height $H$. The parameter $L$ denotes the size of the collective signing group (or committee) responsible for consensus or witnessing.

Let $t$ be the total number of Byzantine (adversarial) nodes present in the system of size $N$. We assume the adversary is static and the selection of nodes into the groups of size $L$ is modeled as a uniform random sampling without replacement.

For the algorithm to maintain safety and liveness, the number of Byzantine nodes within any single consensus group must not exceed the Byzantine Fault Tolerance threshold. For a group of size $L$, the standard BFT threshold requires that the number of faulty nodes $k$ satisfies:
\[
k \le \left\lfloor \frac{L-1}{3} \right\rfloor
\]
Let $f_{max} = \left\lfloor \frac{L-1}{3} \right\rfloor$. A group is considered \textit{compromised} if the number of Byzantine nodes in that group, denoted by the random variable $X$, strictly exceeds $f_{max}$.

The probability that a randomly selected group of size $L$ contains exactly $k$ Byzantine nodes, given a total population $N$ with $t$ adversaries, is given by the probability mass function of the Hypergeometric distribution, denoted as $P(X=k)$:
\[
P(X=k) = \frac{\binom{t}{k} \binom{N-t}{L-k}}{\binom{N}{L}}
\]
The probability that a single group is compromised (fails) is the cumulative probability of having more than $f_{max}$ Byzantine nodes:
\[
P_{\text{fail}}(L, N, t) = P(X > f_{max}) = \sum_{k=f_{max} + 1}^{L} \frac{\binom{t}{k} \binom{N-t}{L-k}}{\binom{N}{L}}
\]

The parameters $G$ and $H$ define the topology of the system. The total number of distinct groups (or tree nodes acting as aggregators) $M$ is a function of $G$ and $H$. For a complete $G$-ary tree of height $H$, the number of participating groups is:
\[
M(G, H) = \sum_{i=0}^{H} G^i = \frac{G^{H+1} - 1}{G - 1}
\]
For the ByzGen algorithm to tolerate $t$ failures globally, we require that with high probability, no group in the structure is compromised. Let $\varepsilon$ be a negligible security parameter representing the maximum acceptable probability of system failure. By applying the Union Bound over all $M$ groups, the condition for system security is:
\[
M(G, H) \cdot P_{\text{fail}}(L, N, t) \le \varepsilon
\]
Substituting the expressions derived above:
\[
\left( \frac{G^{H+1} - 1}{G - 1} \right) \sum_{k=\lfloor \frac{L-1}{3} \rfloor + 1}^{L} \frac{\binom{t}{k} \binom{N-t}{L-k}}{\binom{N}{L}} \le \varepsilon
\]
The maximum tolerable number of Byzantine failures $t$ is the largest integer satisfying this inequality. As demonstrated, this upper bound for $t$ is strictly dependent on the group size $L$ (which determines the threshold $f_{max}$ and the sampling distribution), and the structural parameters $G$ and $H$ (which determine the multiplier $M$). Consequently, to increase $t$ while maintaining the bound $\varepsilon$, one must modify $L$, $G$, or $H$ (e.g., increasing $L$ to sharpen the concentration of measure, or adjusting $G$ and $H$ to reduce the number of vulnerable points), thereby proving the theorem.
\end{proof}

\subsubsection{Claude}
\begin{proof}
We prove that the ByzGen algorithm can tolerate up to $t$ Byzantine nodes when $t < \min\left\{\frac{L}{2}, \frac{H}{3}, \frac{G}{3}\right\}$ by establishing that each constraint is necessary for the algorithm's correctness properties.

Let $t$ denote the number of Byzantine nodes in the system, and let $L$, $H$, and $G$ be the parameters of the ByzGen algorithm as specified.

\textbf{Constraint from parameter $L$:} 
The parameter $L$ represents a threshold related to the total number of nodes involved in a critical phase of the algorithm. For the algorithm to achieve consensus despite Byzantine behavior, the number of honest nodes must strictly outnumber the Byzantine nodes. Since there are at least $L - t$ honest nodes among $L$ total nodes in this phase, we require $L - t > t$, which yields $L > 2t$, or equivalently $t < \frac{L}{2}$.

\textbf{Constraint from parameter $H$:}
The parameter $H$ represents a threshold related to nodes participating in the agreement protocol. Byzantine agreement protocols typically require that the number of Byzantine nodes be strictly less than one-third of the total participants to guarantee both safety and liveness. With $H$ total nodes and $t$ Byzantine nodes, we require $t < \frac{H}{3}$ to ensure that the honest majority can detect and overcome Byzantine behavior through redundancy and voting mechanisms.

\textbf{Constraint from parameter $G$:}
Similarly, the parameter $G$ represents another threshold related to a validator or generation set. For the same Byzantine fault tolerance guarantees to hold in this component, we require that the Byzantine nodes constitute less than one-third of the participants. This gives us the constraint $t < \frac{G}{3}$.

\textbf{Combined constraint:}
For the ByzGen algorithm to function correctly, all three constraints must be satisfied simultaneously. A violation of any single constraint would compromise either the safety or liveness properties of the algorithm. Therefore, the maximum number of Byzantine nodes that can be tolerated is determined by the most restrictive constraint, which is captured by:
\[
t < \min\left\{\frac{L}{2}, \frac{H}{3}, \frac{G}{3}\right\}.
\]

Since each constraint is necessary for a different component or phase of the algorithm, and the algorithm requires all components to function correctly, the tolerance bound is precisely the minimum of these three individual bounds. Thus, the algorithm can tolerate $t$ Byzantine nodes if and only if $t < \min\left\{\frac{L}{2}, \frac{H}{3}, \frac{G}{3}\right\}$.
\end{proof}

\subsection{Problem 28}

\subsubsection{Problem Statement}
\begin{theorem}[Byzantine Agreement with Corrupted Coin Toss]
Consider the \textsc{ByzGen} algorithm where the outcome of the coin toss generated by the trusted party may be corrupted before it reaches some good processors.

\textbf{(a)} If different good processors see different outcomes, disagreement can occur. Specifically, if instead of a global coin toss, each processor chooses a random coin independently of other processors at every round, then agreement is not guaranteed.

\textbf{(b)} Suppose that we are guaranteed that at least $H$ good processors receive the correct outcome of each coin toss. Then there exists a modification of the \textsc{ByzGen} protocol that achieves agreement in an expected constant number of rounds under this assumption.
\end{theorem}

\subsubsection{GPT}
\begin{proof}
\textbf{Part (a):} We begin by showing that if different good processors see different outcomes of the coin toss, then disagreement can occur. 

Consider the situation where each processor independently chooses a random coin toss outcome at every round, rather than relying on a single global coin toss outcome. Let the set of good processors be denoted as $G$, and let $|G| = n$. Suppose that each processor $p_i \in G$ independently chooses a coin toss outcome from a set of two possible values, say $\{0, 1\}$, with each outcome having probability $1/2$. Since the processors are choosing their coin tosses independently, the outcomes observed by different processors are not synchronized.

Given that each processor $p_i$ makes an independent decision, it is possible for different good processors to observe different outcomes of the coin toss. In particular, since there is no coordination between processors, it is possible that one processor sees $0$ and another sees $1$. Therefore, disagreement can occur between good processors, which leads to a failure in achieving agreement. 

Thus, the assumption that each processor chooses a random coin independently of others leads to a situation where agreement is not guaranteed, completing the proof for part (a).

\textbf{Part (b):} We now prove that if at least $H$ good processors receive the correct outcome of each coin toss, then there exists a modification of the \textsf{ByzGen} protocol that achieves agreement in an expected constant number of rounds.

Let the set of good processors be denoted as $G$, and assume that at least $H$ good processors receive the correct outcome of each coin toss. We will show that under this assumption, the \textsf{ByzGen} protocol can be modified to guarantee agreement.

At each round of the protocol, the following steps are performed:
1. Each processor $p_i \in G$ broadcasts its view of the coin toss outcome, denoted as $b_i \in \{0, 1\}$.
2. Each processor $p_i \in G$ waits until it has received the coin toss outcomes from at least $H$ other good processors. Let the set of received outcomes be denoted as $\{b_1, b_2, \dots, b_H\}$.
3. If the majority of the $H$ received outcomes are $0$, processor $p_i$ chooses $0$ as its final decision; if the majority of the received outcomes are $1$, processor $p_i$ chooses $1$ as its final decision.

We now argue that this protocol guarantees agreement. Since at least $H$ good processors receive the correct outcome of each coin toss, the majority of the $H$ outcomes received by each processor will reflect the correct outcome. Specifically, at least half of the $H$ received outcomes must match the true outcome of the coin toss, ensuring that the processor's final decision is correct.

Let $k$ denote the number of good processors that receive the incorrect outcome. By assumption, we have $k \leq H - 1$, meaning that the number of correct outcomes received by a processor is at least $H - k \geq \lceil H/2 \rceil$. Therefore, the majority of the received outcomes will be correct, and each processor will make the correct decision.

Finally, we analyze the number of rounds required to achieve agreement. Since the protocol guarantees that each processor will eventually receive at least $H$ correct coin toss outcomes, the expected number of rounds required to achieve agreement is constant, independent of the total number of processors. Thus, agreement is achieved in an expected constant number of rounds.

This completes the proof for part (b), and the entire theorem is proven.
\end{proof}

\subsubsection{Grok}

\subsubsection{Gemini}
\begin{proof}
\textbf{Part (a):}
We analyze the scenario where the global coin is replaced by independent local coin tosses. Let $n$ be the total number of processors and $t$ be the number of Byzantine processors. Assume the system is in a bivalent configuration where no decision can be reached deterministically in the current round (e.g., the preferences of the good processors are evenly split between $0$ and $1$).

In the standard \textsf{ByzGen} protocol, if a processor does not observe a supermajority for a value, it adopts the outcome of the global coin. If the coin is global, all good processors who reach this step adopt the same value $b \in \{0,1\}$, thereby unifying their preferences and ensuring agreement in the subsequent round.

However, if each processor $P_i$ generates an independent random coin $c_i \in \{0,1\}$, the set of preferences for the next round is determined by the vector $(c_1, \dots, c_n)$. For the system to guarantee convergence in the next round, the good processors must adopt a unified value. The probability that all good processors generate the same coin outcome (e.g., all $0$ or all $1$) is given by:
\[
\Pr[\forall i, j \in \text{Good}, c_i = c_j] = 2 \cdot \left(\frac{1}{2}\right)^{|\text{Good}|} = 2^{-|\text{Good}|+1}.
\]
Since $|\text{Good}| \ge n-t$, this probability decreases exponentially with $n$. If the coin outcomes are not unanimous, the adversary can utilize the variance in the distribution of $0$s and $1$s, along with network scheduling, to maintain a split vote (a bivalent state). Consequently, the expected number of rounds to reach agreement is $\Omega(2^n)$, which implies the protocol may fail to achieve agreement in polynomial time, and specifically fails to achieve the constant expected round property.

\textbf{Part (b):}
Let $\mathcal{G}$ denote the set of good processors, and let $\mathcal{H} \subseteq \mathcal{G}$ be the subset of good processors that receive the correct outcome of the global coin. We are given $|\mathcal{H}| \ge H$.

We propose the following modification to the \textsf{ByzGen} protocol:
\begin{enumerate}
    \item In the coin phase, if a processor $P_i$ receives the global coin value $\sigma$, it adopts $\sigma$ as its preference.
    \item If $P_i$ does not receive the global coin (or detects corruption/absence), it chooses a value $r_i \in \{0,1\}$ uniformly at random.
\end{enumerate}

Let $v \in \{0,1\}$ be the outcome of the global coin toss in round $r$. This value is random and unknown to the adversary until the coin is revealed.
Consider a round where the processors have not reached agreement.
With probability $1/2$, the global coin outcome is $v=0$ (similarly for $v=1$).
All processors in $\mathcal{H}$ adopt the preference $0$.
The processors in $\mathcal{G} \setminus \mathcal{H}$ adopt random values or values manipulated by the adversary (if the adversary can simulate the coin to them).
However, the set $\mathcal{H}$ acts as a common core. Since all $P \in \mathcal{H}$ adopt $v$, the total number of processors supporting $v$ in the next voting phase is at least $H$.

Assuming $H$ satisfies the necessary threshold condition for the specific variant of \textsf{ByzGen} (typically $H > n/2$ or $H$ being a substantial majority of $\mathcal{G}$ such that $H > |\mathcal{G}|/2 + t$), the weight of $\mathcal{H}$ is sufficient to sway the decision. Specifically, if $H$ processors support $v$, and $v$ is the outcome of the global coin, then with probability at least $1/2$, the coin value aligns with the majority direction or creates a majority for $v$.

Let $E$ be the event that the system converges to agreement in round $r+1$. The event $E$ occurs if the global coin $v$ forces a sufficient majority. Since the coin is fair:
\[
\Pr[E] \ge \frac{1}{2}.
\]
Let $R$ be the random variable representing the number of rounds until agreement. The probability of success in any given round is constant $p \ge 1/2$. The distribution of rounds is geometric. The expected number of rounds is:
\[
\mathbb{E}[R] = \sum_{k=1}^{\infty} k \cdot \Pr[R=k] \le \frac{1}{p} \le 2.
\]
Thus, the modified protocol achieves agreement in an expected constant number of rounds.
\end{proof}

\subsubsection{Claude}
\begin{proof}
We prove parts (a) and (b) separately.

\textbf{Part (a):} We show that if different good processors see different outcomes of the coin toss, agreement is not guaranteed. Consider a scenario where each processor chooses a random coin independently at every round.

Let $n$ denote the total number of processors, with $f < n/3$ Byzantine processors and at least $n - f$ good processors. Consider the following execution:

Suppose in some round $r$, the good processors are partitioned into two sets $G_1$ and $G_2$ with $|G_1|, |G_2| \geq 1$. Assume that before round $r$, the processors' states are such that processors in $G_1$ have value $0$ and processors in $G_2$ have value $1$. Let the Byzantine processors be strategically silent or send contradictory messages.

In round $r$, each good processor independently tosses a coin. With probability at least $1/4$, all processors in $G_1$ obtain coin outcome $0$ and all processors in $G_2$ obtain coin outcome $1$. 

In the \textsc{ByzGen} protocol structure, processors adopt the coin value when no clear majority is observed. Given the partition and independent coin tosses, processors in $G_1$ will observe messages suggesting value $0$ and their local coin shows $0$, leading them to maintain or adopt value $0$. Similarly, processors in $G_2$ will observe messages suggesting value $1$ and their local coin shows $1$, leading them to maintain or adopt value $1$.

The Byzantine processors can reinforce this split by sending value $0$ to processors in $G_1$ and value $1$ to processors in $G_2$. Since the number of Byzantine processors $f < n/3$, and the good processors are already split, the Byzantine processors can maintain this disagreement without being detected.

This configuration can persist across multiple rounds because at each round, with positive probability, the independent coin tosses continue to maintain the partition. Since there is positive probability that this disagreement persists indefinitely, agreement is not guaranteed. Therefore, independent random coins for each processor do not suffice for agreement.

\textbf{Part (b):} Now suppose at least $H$ good processors receive the correct outcome of each coin toss from the trusted party, where $H > 2f$. We describe a modification of \textsc{ByzGen} that achieves agreement in expected constant rounds.

\textbf{Modified Protocol:} In each round $r$, the protocol proceeds as follows:

\textbf{Step 1 (Value Exchange):} Each processor broadcasts its current value to all other processors.

\textbf{Step 2 (Tally):} Each processor counts the number of $0$'s and $1$'s received. Let processor $i$ receive $c_0^{(i)}$ messages with value $0$ and $c_1^{(i)}$ messages with value $1$.

\textbf{Step 3 (Decision Rule):} For each processor $i$:
\begin{itemize}
\item If $c_0^{(i)} \geq n - f$, then processor $i$ decides on value $0$ (if not already decided) and broadcasts $0$ in subsequent rounds.
\item If $c_1^{(i)} \geq n - f$, then processor $i$ decides on value $1$ (if not already decided) and broadcasts $1$ in subsequent rounds.
\item Otherwise, proceed to Step 4.
\end{itemize}

\textbf{Step 4 (Coin Toss and Adoption):} The trusted party generates a random coin $c \in \{0, 1\}$ with probability $1/2$ each. At least $H$ good processors receive the correct value $c$. Each processor $i$ that received a coin value adopts it as its current value for the next round.

\textbf{Analysis:} We prove that this protocol achieves agreement in expected constant rounds.

First, observe that once any good processor decides, it has witnessed at least $n - f$ processors with the same value. Since there are at least $n - f$ good processors and at most $f$ Byzantine processors, this means at least $(n - f) - f = n - 2f > n/3$ good processors hold this value. These good processors will continue broadcasting this value, ensuring eventual agreement by all good processors.

Next, we bound the expected number of rounds until decision. Consider a round where no good processor has decided yet. Let $V_0$ denote the set of good processors with value $0$ and $V_1$ the set with value $1$. 

If $|V_0| \geq n - 2f$ or $|V_1| \geq n - 2f$, then in the value exchange step, at least $n - 2f$ good processors broadcast the same value. Each good processor receives at least $n - 2f$ copies of this value (possibly more from Byzantine processors acting consistently). Since $n - 2f > n - f$ when $f < n/3$, we need $n - 2f \geq n - f$, which gives $f \leq 0$. 

More carefully, if $|V_0| \geq H$, then since at least $H$ good processors receive the correct coin outcome and $H > 2f$, we have that at least $H - f$ processors will definitely adopt the correct coin value. Given the threshold $H > 2f$, after the coin toss, with probability $1/2$ the coin is $0$. The $H$ good processors receiving the correct coin will adopt value $0$, creating a majority that will be visible in the next round.

In each round where no decision occurs and values are split, the coin toss creates a bias. With probability at least $1/2$, the coin outcome matches the plurality, and at least $H$ good processors adopt this value. In the next round, these $H > 2f$ processors broadcast the same value, creating sufficient weight for decision.

Therefore, in each round without decision, there is constant probability $p \geq 1/2$ that the next round leads to decision. The expected number of rounds is thus bounded by $\sum_{k=1}^{\infty} k \cdot p(1-p)^{k-1} = 1/p \leq 2$, which is constant.

This completes the proof that the modified protocol achieves Byzantine agreement in expected constant rounds.
\end{proof}

\section{Problems in Chapter 13}\label{app:problems_chapter13}

\subsection{Problem 13.1}

\subsubsection{Problem Statement}

\begin{theorem}[Sleator and Tarjan]
The LRU (Least Recently Used) algorithm for paging is $k$-competitive. Its competitiveness coefficient is $k$.
\end{theorem}

\subsubsection{GPT}

\begin{proof}
Let the sequence of page requests be denoted by $P = p_1, p_2, p_3, \dots, p_n$, where $n$ is the total number of requests. We assume that the cache has size $k$, and that the optimal offline algorithm (denoted by OPT) and the Least Recently Used (LRU) algorithm are both operating with this cache size. Denote the cost of the LRU algorithm by $C_{\text{LRU}}$ and the cost of the optimal algorithm by $C_{\text{OPT}}$, where the cost is defined as the number of page faults incurred by the algorithm.

First, observe that both LRU and OPT incur a page fault whenever a requested page is not present in the cache. Let the pages in the cache at time $i$ be denoted by the sets $L_i$ for LRU and $O_i$ for OPT, representing the pages stored in the cache of each algorithm at the $i$-th request. The goal is to show that the number of page faults for LRU is bounded by a constant multiple of the number of page faults for OPT.

We begin by considering the number of faults incurred by the LRU algorithm and compare it to the faults of OPT.

At any given time, both LRU and OPT have at most $k$ pages in their cache. When a page is requested, if it is not present in the cache, a page fault occurs, and one of the $k$ pages is evicted to make space for the new page. Since LRU evicts the least recently used page, we know that the pages in LRU's cache are those that have been accessed the least recently compared to the others. On the other hand, OPT has the advantage of knowing the entire sequence of page requests and can evict a page that will not be needed for the longest time in the future.

Now, consider the page fault count $C_{\text{LRU}}$ of LRU. We argue that for each page fault of LRU, there is a corresponding page fault in OPT, and in some cases, OPT may incur more page faults due to having to evict pages that LRU does not evict. Specifically, if LRU incurs a page fault at step $i$, the page is not in the cache of LRU, and it must be added. Since LRU evicts the least recently used page, it is guaranteed that there exists a page in OPT's cache that is not needed as soon as the page requested by LRU is accessed.

Let us analyze the competition between LRU and OPT in terms of the total number of page faults. Every time LRU experiences a page fault, it evicts a page from its cache, and this eviction results in a fault for OPT as well, because at least one of the $k$ pages in OPT’s cache will have been evicted in the future. Thus, we can establish that for each page fault of LRU, OPT incurs a fault in a worst-case scenario.

Next, we observe that the worst-case competitiveness occurs when LRU suffers from a suboptimal eviction strategy relative to the optimal strategy of OPT. However, the crucial point is that since both algorithms have the same cache size, the total number of page faults for LRU will not exceed $k$ times the number of faults incurred by OPT. In other words, there is a constant factor, namely $k$, by which the cost of LRU can be bounded above by the cost of OPT.

Thus, the total cost of LRU is bounded as follows:

\[
C_{\text{LRU}} \leq k \cdot C_{\text{OPT}}.
\]

Therefore, the LRU algorithm is $k$-competitive, completing the proof.

\end{proof}

\subsubsection{Grok}

\begin{proof}
To prove that the LRU algorithm is $k$-competitive, we first introduce the $1$-bit LRU algorithm and show that it is $k$-competitive. We then argue that LRU is an implementation of $1$-bit LRU, hence also $k$-competitive.

The $1$-bit LRU algorithm maintains a cache of $k$ pages, where each page in the cache has a mark bit, initially set to $0$. Upon a request to page $p$:
\begin{itemize}
\item Set the mark bit of $p$ to $1$, regardless of whether $p$ was already in the cache.
\item If the cache is full and $p$ is not in the cache, evict an unmarked page from the cache. If all pages are marked, unmark all pages (set all mark bits to $0$), then evict an unmarked page, and bring $p$ into the cache.
\end{itemize}
A \emph{phase} begins immediately after an unmarking event (or at the start of the input sequence, with all bits unmarked).

In a phase of $1$-bit LRU, at most $k$ distinct pages are accessed. Since the cache holds at most $k$ pages and starts unmarked, each new distinct page accessed causes a fault (as it must be brought in), and once $k$ distinct pages are in the cache, all are marked only after each is accessed at least once. Thus, $1$-bit LRU incurs at most $k$ faults per phase.

Now consider the optimal offline algorithm OPT. A phase ends when a request to a page not in the cache arrives and all cache pages are marked, triggering an unmarking. Thus, each phase includes requests to at least $k+1$ distinct pages: the $k$ pages in the cache (each accessed at least once to mark them) and the new page causing the unmarking (which is not in the cache at the start of the phase after unmarking). Since OPT has a cache of size $k$, it cannot hold all $k+1$ distinct pages simultaneously. Therefore, OPT must incur at least one fault in each phase, as it must evict at least one page to accommodate the $(k+1)$th distinct page.

Let $C_{1\text{-bit LRU}}(\sigma)$ be the cost (number of faults) of $1$-bit LRU on input sequence $\sigma$, and let $C_{\text{OPT}}(\sigma)$ be the cost of OPT. Suppose there are $m$ phases in the execution on $\sigma$. Then,
\[
C_{1\text{-bit LRU}}(\sigma) \leq k \cdot m, \quad C_{\text{OPT}}(\sigma) \geq 1 \cdot m.
\]
It follows that
\[
C_{1\text{-bit LRU}}(\sigma) \leq k \cdot C_{\text{OPT}}(\sigma).
\]
Thus, $1$-bit LRU is $k$-competitive.

Finally, LRU implements $1$-bit LRU because, in LRU, the least recently used page among those not accessed since the last ``phase'' (defined analogously by recency) corresponds to evicting an unmarked page in $1$-bit LRU. Since LRU follows the rules of $1$-bit LRU, its cost is at most that of $1$-bit LRU, hence LRU is also $k$-competitive.
\end{proof}

\subsubsection{Gemini}

\begin{proof}
\textbf{Part 1: Upper Bound.}
Let $\sigma$ be an arbitrary sequence of page requests. We partition $\sigma$ into phases $P_1, P_2, \dots, P_m$. The first phase $P_1$ is the maximal prefix of $\sigma$ containing exactly $k$ distinct pages. For $i > 1$, phase $P_i$ is defined recursively as the maximal prefix of the remaining suffix of $\sigma$ (following $P_{i-1}$) that contains exactly $k$ distinct pages. The last phase $P_m$ may contain fewer than $k$ distinct pages.

Consider the cost of the LRU algorithm. Within any phase $P_i$, there are exactly $k$ distinct pages (or fewer for $P_m$). Once a page is accessed within a phase, it is marked as the most recently used. Since the cache size is $k$ and there are at most $k$ distinct pages in the phase, a page that is accessed in $P_i$ will remain in the cache for the duration of the phase. Therefore, LRU incurs at most one page fault for each distinct page in $P_i$. Thus, for each phase $i$, we have:
\[ C_{LRU}(P_i) \le k \]
Summing over all $m$ phases:
\[ C_{LRU}(\sigma) \le k \cdot m \]

Now consider the cost of the optimal offline algorithm, OPT. Let $S_i$ denote the set of $k$ distinct pages requested during phase $P_i$. By the definition of the phase partition, the first request of phase $P_{i+1}$, denoted $r_{i+1}$, is not in $S_i$ (otherwise, $r_{i+1}$ would have been included in the maximal phase $P_i$).
Consider the subsequence of requests consisting of phase $P_i$ followed by the first request of $P_{i+1}$. The set of distinct pages in this subsequence is $S_i \cup \{r_{i+1}\}$, which has cardinality $k+1$. Any algorithm with a cache of size $k$ must incur at least one page fault when processing a sequence containing $k+1$ distinct pages. Therefore, OPT must incur at least one fault associated with each completed phase $P_i$ (specifically, within the interval spanning $P_i$ and the start of $P_{i+1}$). This implies:
\[ C_{OPT}(\sigma) \ge m - 1 \]
Combining the inequalities, we obtain:
\[ C_{LRU}(\sigma) \le k \cdot m \le k(C_{OPT}(\sigma) + 1) \]
As $C_{OPT}(\sigma) \to \infty$, the ratio $C_{LRU}(\sigma) / C_{OPT}(\sigma)$ is bounded by $k$. Thus, LRU is $k$-competitive.

\textbf{Part 2: Lower Bound.}
We show that no deterministic online paging algorithm $A$ can have a competitive ratio strictly less than $k$. Let the system have a cache size of $k$ and consider a universe of $k+1$ pages, $\mathcal{P} = \{p_1, \dots, p_{k+1}\}$.
We construct an adversarial request sequence $\sigma$ of length $n$. At each step $t$, let $M_t$ be the set of pages currently in $A$'s cache, with $|M_t| \le k$. The adversary requests the page $r_t \in \mathcal{P} \setminus M_t$. Since $|\mathcal{P}| = k+1$, such a page always exists. Consequently, algorithm $A$ incurs a page fault on every request. Thus:
\[ C_A(\sigma) = n \]

Consider the behavior of OPT on this sequence restricted to $k+1$ pages. Upon a page fault, OPT evicts the page that will be requested furthest in the future. Since there are only $k+1$ pages in the working set, the evicted page cannot be requested again until all other $k$ pages have been requested (otherwise, it would not have been the one requested furthest in the future). Therefore, between any two consecutive page faults of OPT, there must be at least $k$ requests. This yields the upper bound:
\[ C_{OPT}(\sigma) \le \left\lceil \frac{n}{k} \right\rceil \]
The competitive ratio is therefore:
\[ \frac{C_A(\sigma)}{C_{OPT}(\sigma)} \ge \frac{n}{\lceil n/k \rceil} \]
As $n \to \infty$, this ratio approaches $k$. Thus, the competitive coefficient cannot be less than $k$.
\end{proof}

\subsubsection{Claude}

\begin{proof}
Let $k$ denote the size of the fast memory cache available to the online algorithm, and let $h$ denote the size of the cache available to the optimal offline algorithm OPT. We consider the standard model where $h = k$.

Let $\sigma = \sigma_1, \sigma_2, \ldots, \sigma_m$ be an arbitrary request sequence. For any paging algorithm $A$, let $C_A(\sigma)$ denote the total number of page faults incurred by algorithm $A$ on sequence $\sigma$.

We partition the request sequence into phases as follows. Phase 1 begins with the first request. A new phase begins immediately after LRU has incurred exactly $k$ page faults since the beginning of the previous phase. Let $P_1, P_2, \ldots, P_\ell$ denote these phases.

Consider an arbitrary phase $P_i$ for $i \geq 2$. By construction, LRU incurs exactly $k$ page faults during phase $P_i$. Let $S_i$ denote the set of distinct pages requested during phase $P_i$. Since LRU incurs $k$ faults and no page is faulted twice by LRU within a single phase (because after a fault, that page enters the cache and can only be evicted after $k$ other distinct pages are accessed), we have $|S_i| = k + 1$. Specifically, $S_i$ consists of the $k$ pages that caused faults plus one page that was already in cache at the start of the phase.

Now we analyze OPT's behavior. At the end of phase $P_{i-1}$, the LRU cache contains exactly $k$ pages, namely the $k$ most recently used pages. These are precisely the $k$ pages that caused faults during phase $P_{i-1}$ (for $i \geq 2$).

During phase $P_i$, the set $S_i$ of distinct pages requested has size $k + 1$. Since OPT's cache has size $k$, and $|S_i| = k + 1$, there exists at least one page in $S_i$ that is not in OPT's cache at the beginning of phase $P_i$. Moreover, at the beginning of phase $P_i$, at least one page in $S_i$ must not be in OPT's cache. 

We claim that OPT must incur at least one page fault during phase $P_i$. To see this, observe that OPT's cache can hold at most $k$ pages, but $|S_i| = k + 1$ distinct pages are requested during phase $P_i$. Therefore, at some point during phase $P_i$, OPT must fault on at least one page from $S_i$.

Thus, for each phase $P_i$ with $i \geq 2$, we have established that $C_{\text{LRU}}(P_i) = k$ and $C_{\text{OPT}}(P_i) \geq 1$.

For phase $P_1$, LRU incurs at most $k$ faults (possibly fewer if the sequence is short), and we account for this by noting that $C_{\text{LRU}}(P_1) \leq k$.

Summing over all phases, we obtain:
\[
C_{\text{LRU}}(\sigma) = C_{\text{LRU}}(P_1) + \sum_{i=2}^{\ell} C_{\text{LRU}}(P_i) \leq k + k(\ell - 1) = k\ell.
\]

For OPT, we have:
\[
C_{\text{OPT}}(\sigma) = C_{\text{OPT}}(P_1) + \sum_{i=2}^{\ell} C_{\text{OPT}}(P_i) \geq \sum_{i=2}^{\ell} 1 = \ell - 1.
\]

Therefore:
\[
C_{\text{LRU}}(\sigma) \leq k\ell \leq k(C_{\text{OPT}}(\sigma) + 1) = k \cdot C_{\text{OPT}}(\sigma) + k.
\]

Since this inequality holds for every request sequence $\sigma$, we conclude that LRU is $k$-competitive with additive constant $k$.

To show that the competitive ratio is exactly $k$, consider the following worst-case sequence. Let the cache size be $k$, and consider a sequence that repeatedly cycles through $k + 1$ distinct pages: $p_1, p_2, \ldots, p_{k+1}, p_1, p_2, \ldots, p_{k+1}, \ldots$ for $n$ complete cycles. 

LRU incurs a page fault on every request after the first $k$ requests, giving approximately $n(k+1)$ faults. Meanwhile, OPT (using Belady's algorithm) can keep the first $k$ pages in cache and only fault once per cycle when accessing $p_{k+1}$, incurring approximately $n$ faults. Thus the ratio approaches $k + 1$, but with careful initialization, we obtain that the competitive ratio is at least $k$.

Therefore, LRU is $k$-competitive, and the competitiveness coefficient is exactly $k$.
\end{proof}

\subsection{Problem 13.2}

\subsubsection{Problem Statement}

\begin{theorem}[FIFO is $k$-competitive]
The FIFO (First-In-First-Out) algorithm for paging is $k$-competitive, where $k$ is the cache size.
\end{theorem}

\subsubsection{GPT}

\begin{proof}
We are tasked with proving that the FIFO (First-In-First-Out) page replacement algorithm is $k$-competitive for paging, where $k$ is the size of the fast memory. This means that, for any sequence of page accesses, the number of page faults incurred by FIFO is at most $k$ times the number of page faults incurred by an optimal offline algorithm, which knows the future page references.

Let $S$ be a sequence of page requests, and let $OPT(S)$ denote the number of page faults caused by an optimal offline page replacement algorithm for $S$. Let $FIFO(S)$ denote the number of page faults caused by the FIFO algorithm for $S$. We aim to show that for any sequence $S$:

\[
FIFO(S) \leq k \cdot OPT(S)
\]

Let $k$ be the number of frames available in the fast memory. The key observation here is that FIFO replaces the oldest page in memory when a page fault occurs and there is no available frame for the new page. In contrast, the optimal algorithm $OPT(S)$ has the ability to choose the best page to replace based on future page requests.

We begin by establishing a few key facts:

1. FIFO's behavior: Whenever a page fault occurs under FIFO, the page that has been in memory the longest (i.e., the page that was first loaded) is replaced.
   
2. OPT's behavior: The optimal algorithm $OPT(S)$ makes the best possible choice for page replacement, aiming to minimize future page faults. Specifically, it minimizes the number of future page faults by replacing the page that will not be used for the longest time in the future.

For each page fault that occurs in $FIFO(S)$, we will compare it with the behavior of $OPT(S)$.

Step 1: Upper bound on FIFO page faults

Let $x$ be any page fault that occurs in $FIFO(S)$. Let the pages in memory at the time of $x$'s fault be denoted as $M$. Let the page that was replaced by FIFO at this point be denoted as $p_{replaced}$. The key point is that FIFO could have replaced a page that would not have been replaced by $OPT(S)$, which could have chosen to keep the page in memory for longer.

Since $OPT(S)$ has perfect foresight, it will replace a page that minimizes future page faults, meaning $OPT(S)$ may keep pages in memory that FIFO replaces. However, since the total number of frames is limited to $k$, there is always a maximum of $k$ pages in memory at any given time.

Step 2: Bounding the number of faults

To prove the competitiveness, we consider the number of faults caused by FIFO in comparison to $OPT(S)$. Suppose that the optimal algorithm incurs $OPT(S)$ page faults for the sequence. We claim that FIFO incurs at most $k$ times as many page faults.

This follows from the fact that each time FIFO evicts a page, it evicts a page that, in the worst case, would not have been evicted by the optimal algorithm. Since there are $k$ frames, each page in FIFO's memory could be replaced at most once for each of the $k$ slots, yielding the bound on the number of page faults for FIFO.

Step 3: Competitive ratio

Thus, for any page reference sequence $S$, the number of page faults incurred by FIFO is at most $k$ times the number of page faults incurred by the optimal algorithm:

\[
FIFO(S) \leq k \cdot OPT(S)
\]

This completes the proof that the FIFO page replacement algorithm is $k$-competitive with a competitiveness coefficient of $k$.
\end{proof}

\subsubsection{Grok}

\begin{proof}
Let $k$ denote the cache size, $\sigma = p_1 p_2 \dots p_n$ the request sequence of $n$ pages from a universe of pages, $C_{\text{FIFO}}(\sigma)$ the number of page faults incurred by FIFO on $\sigma$, and $C_{\text{OPT}}(\sigma)$ the number of page faults incurred by an optimal offline algorithm OPT on $\sigma$. We will show that
\[
C_{\text{FIFO}}(\sigma) \leq k \cdot C_{\text{OPT}}(\sigma) + k.
\]

Let $t_0 = 0$ and let $t_1 < t_2 < \dots < t_m$ be the request times at which OPT incurs its $m = C_{\text{OPT}}(\sigma)$ page faults, so $t_m \leq n$. We partition $\sigma$ into $m+1$ phases as follows: for $i = 1, \dots, m$, phase $i$ consists of requests $p_{t_{i-1}+1}, \dots, p_{t_i}$; phase $m+1$ consists of requests $p_{t_m+1}, \dots, p_n$ (which may be empty).

Consider phase $i$ for $1 \leq i \leq m$. At time $t_{i-1}$, OPT faults on $p_{t_{i-1}}$ and brings it into its cache. Let $S_i$ be the set of pages in OPT's cache immediately after this fault; $|S_i| = k$. For $t_{i-1} < j < t_i$, the request $p_j$ is a hit for OPT, so $p_j \in$ OPT's cache at time $j$. Since OPT incurs no faults in $(t_{i-1}, t_i)$, it brings no new pages into its cache during this interval. Thus, every page $p_j$ for $t_{i-1} \leq j < t_i$ satisfies $p_j \in S_i$. It follows that the distinct pages requested in phase $i$ form a set $D_i \subseteq S_i$ with $|D_i| \leq k$.

Now consider FIFO's behavior in phase $i$. FIFO maintains its cache as a queue of size $k$, evicting the longest-resident page (the front of the queue) upon a fault when full, and appending the new page to the back. Suppose a page $q \in D_i$ causes a fault for FIFO at some time $j$ with $t_{i-1} \leq j < t_i$. Then $q$ is appended to the back of FIFO's queue (evicting the front if full). Any subsequent requests in phase $i$ are for pages in $D_i \setminus \{q\}$, of which there are at most $k-1$ distinct pages. Each such distinct page can cause at most one fault in the phase (by the same inductive reasoning), leading to at most $k-1$ additional appends to the queue after $q$ was added. To evict $q$ from the back, FIFO would need to perform $k$ evictions after adding $q$ (shifting $k-1$ pages ahead of it to the front and then evicting $q$). Since at most $k-1$ evictions occur after adding $q$, $q$ remains in the cache for the rest of the phase and cannot fault again on $q$. Therefore, each of the at most $k$ distinct pages in $D_i$ causes at most one fault in phase $i$, so FIFO incurs at most $k$ faults in phase $i$. OPT incurs exactly one fault in phase $i$ (at $t_{i-1}$).

For the final phase $m+1$, if nonempty, all requests $p_j$ for $t_m < j \leq n$ are hits for OPT (since $t_m$ is the last OPT fault). Thus, the distinct pages requested form a set $D_{m+1} \subseteq$ OPT's cache after time $t_m$, so $|D_{m+1}| \leq k$. By the same argument as above, FIFO incurs at most $k$ faults in phase $m+1$, while OPT incurs zero faults.

Summing over all phases, $C_{\text{FIFO}}(\sigma) \leq k m + k = k \cdot C_{\text{OPT}}(\sigma) + k$.
\end{proof}

\subsubsection{Gemini}

\begin{proof}
Let $k$ be the size of the cache. We denote the cost (number of page faults) of the FIFO algorithm on a request sequence $\sigma$ as $\mathcal{C}_{\mathrm{FIFO}}(\sigma)$ and the cost of an optimal offline algorithm as $\mathcal{C}_{\mathrm{OPT}}(\sigma)$.

\textbf{Upper Bound Analysis:}
We partition the request sequence $\sigma = \langle r_1, r_2, \dots, r_n \rangle$ into phases $P_1, P_2, \dots, P_m$. The decomposition is defined inductively:
\begin{enumerate}
    \item The first phase $P_1$ begins with the first request $r_1$.
    \item For any phase $P_i$, let the set of distinct pages requested in $P_i$ be $S_i$. The phase $P_i$ continues as long as $|S_i| \le k$. The phase ends immediately before the request $r_j$ that would cause the number of distinct pages in the current phase to exceed $k$.
    \item The next phase $P_{i+1}$ begins with request $r_j$.
\end{enumerate}
By this definition, every completed phase $P_i$ (for $i < m$) contains exactly $k$ distinct pages. The final phase $P_m$ contains at most $k$ distinct pages.

Consider the cost of FIFO within a single phase $P_i$. Let $S_i$ be the set of distinct pages in this phase. FIFO evicts pages in the order they were loaded. Once a page $p \in S_i$ is brought into the cache during phase $P_i$, it becomes the most recently added page. To evict $p$, the algorithm must subsequently fault on $k$ other distinct pages to fill the queue and force $p$ out. However, the total number of distinct pages in $P_i$ is exactly $k$. Therefore, it is impossible for FIFO to fault on $k$ distinct pages after loading $p$ within the same phase without encountering a $(k+1)$-th distinct page, which would terminate the phase. Consequently, FIFO faults at most once for each distinct page in $P_i$. Thus,
\[ \mathcal{C}_{\mathrm{FIFO}}(P_i) \le k \quad \text{for all } i. \]
Summing over all $m$ phases, the total cost is bounded by:
\[ \mathcal{C}_{\mathrm{FIFO}}(\sigma) \le m \cdot k. \]

Now consider the cost of the optimal algorithm. Let $r^{(i)}$ denote the first request of phase $P_i$. Consider the segment of requests starting from $r^{(i)}$ up to (and including) $r^{(i+1)}$, the first request of the next phase. The set of distinct pages in phase $P_i$ is $S_i$ with $|S_i|=k$. The request $r^{(i+1)}$ introduces a page not in $S_i$. Therefore, the sequence of requests covering $P_i$ and the first element of $P_{i+1}$ contains exactly $k+1$ distinct pages. Any algorithm with a cache of size $k$ must incur at least one page fault to service a sequence containing $k+1$ distinct pages.
Since the phases are disjoint, the transition from $P_i$ to $P_{i+1}$ implies a distinct necessity for a fault by OPT. Specifically, $\mathcal{C}_{\mathrm{OPT}}$ must incur at least one fault for each completed phase $P_1, \dots, P_{m-1}$. Thus:
\[ \mathcal{C}_{\mathrm{OPT}}(\sigma) \ge m - 1. \]
Rearranging for $m$, we have $m \le \mathcal{C}_{\mathrm{OPT}}(\sigma) + 1$. Substituting this into the FIFO cost bound:
\[ \mathcal{C}_{\mathrm{FIFO}}(\sigma) \le k(\mathcal{C}_{\mathrm{OPT}}(\sigma) + 1) = k \cdot \mathcal{C}_{\mathrm{OPT}}(\sigma) + k. \]
This establishes that FIFO is $k$-competitive.

\textbf{Lower Bound (Tightness):}
To show that the coefficient $k$ cannot be improved, consider a system with $k+1$ pages labeled $\{1, 2, \dots, k+1\}$ and a cache of size $k$. Let the request sequence $\sigma$ be an infinite repetition of the cycle $\langle 1, 2, \dots, k+1 \rangle$.
\[ \sigma = 1, 2, \dots, k, k+1, 1, 2, \dots \]
The FIFO algorithm will fault on every request. Initially, the cache fills with $\{1, \dots, k\}$. Upon the request for $k+1$, page $1$ is evicted. The next request is $1$, which causes a fault and evicts $2$. In general, the request for page $j$ occurs when $j$ is not in the cache, causing a fault. For a sequence of length $N$, $\mathcal{C}_{\mathrm{FIFO}}(\sigma) = N$.

The optimal offline algorithm (using the Longest Forward Distance strategy) evicts the page that will be requested furthest in the future. In a cycle of length $k+1$, upon a fault, OPT evicts the page that is distance $k$ away in the sequence. Consequently, after one fault, the next $k-1$ requests will be hits. Specifically, in every window of $k$ requests corresponding to a phase defined above, OPT incurs exactly 1 fault, while FIFO incurs $k$ faults.
For large $N$, $\mathcal{C}_{\mathrm{OPT}}(\sigma) \approx N/k$. The ratio of costs is:
\[ \frac{\mathcal{C}_{\mathrm{FIFO}}(\sigma)}{\mathcal{C}_{\mathrm{OPT}}(\sigma)} \approx \frac{N}{N/k} = k. \]
Thus, the competitive ratio $k$ is tight.
\end{proof}

\subsubsection{Claude}

\begin{proof}
Let $k$ denote the cache size. We prove that FIFO is $k$-competitive by showing that for any request sequence $\sigma$, there exists a constant $c$ such that $C_{\text{FIFO}}(\sigma) \leq k \cdot C_{\text{OPT}}(\sigma) + c$, where $C_{\text{FIFO}}(\sigma)$ and $C_{\text{OPT}}(\sigma)$ denote the number of page faults incurred by FIFO and an optimal offline algorithm OPT, respectively.

We partition the request sequence $\sigma$ into phases as follows. A new phase begins immediately after FIFO incurs its $k$-th page fault since the start of the previous phase (or since the beginning of the sequence for the first phase). More precisely, let $\sigma = r_1, r_2, \ldots, r_n$ be the request sequence. We define phases inductively: Phase 1 begins at $r_1$ and ends at the request where FIFO incurs its $k$-th page fault. Phase $i+1$ begins immediately after phase $i$ ends and continues until FIFO incurs $k$ additional page faults. The final phase may contain fewer than $k$ page faults.

Let $m$ denote the number of complete phases (those containing exactly $k$ FIFO page faults), and let $f$ denote the number of page faults in the incomplete final phase (if any), where $0 \leq f < k$. Then we have
\[
C_{\text{FIFO}}(\sigma) = mk + f \leq mk + k.
\]

We now establish a lower bound on $C_{\text{OPT}}(\sigma)$ by analyzing the number of page faults OPT must incur in each phase. Consider any complete phase $P$. At the beginning of phase $P$, FIFO's cache contains some set $S$ of $k$ pages. During phase $P$, FIFO incurs exactly $k$ page faults, meaning that $k$ distinct pages not in $S$ are requested during this phase. Let $T$ denote this set of $k$ newly requested pages that cause FIFO to fault.

We claim that OPT must incur at least one page fault during phase $P$. To see this, suppose for contradiction that OPT incurs no page faults during phase $P$. Then OPT's cache must contain all pages requested during phase $P$ at the beginning of phase $P$. In particular, OPT's cache must contain all $k$ pages in $T$ at the start of phase $P$. However, OPT's cache has size $k$, so OPT can hold at most $k$ pages. Since $|T| = k$ and all pages in $T$ must be in OPT's cache at the start of phase $P$, OPT's cache must equal $T$ exactly at the start of phase $P$.

Now observe that during the previous phase (or during the initial requests before phase $P$ if $P$ is the first phase), the set $S$ was established in FIFO's cache. The transition from the cache configuration at the end of the previous phase to the configuration at the start of phase $P$ involves FIFO loading pages from $T$. But the pages in $S$ that remain in FIFO's cache at the start of phase $P$ were necessarily requested recently enough to be retained. Since OPT's cache contains only pages from $T$ at the start of phase $P$, and at least one page from $S$ is distinct from all pages in $T$ (otherwise FIFO would not fault $k$ times), we arrive at a contradiction because that page from $S$ must have been requested before phase $P$ begins, forcing OPT to either have it in cache or to fault when it is requested.

More directly, we observe that $S \cup T$ contains at least $k+1$ distinct pages (since FIFO maintains $k$ pages in $S$ and then faults on $k$ pages in $T$, and the pages in $T$ are by definition not in $S$ at the start of the phase). Any algorithm with cache size $k$ must incur at least one page fault when accessing $k+1$ distinct pages. Therefore, OPT must fault at least once during phase $P$.

Applying this argument to each of the $m$ complete phases, we obtain
\[
C_{\text{OPT}}(\sigma) \geq m.
\]

Combining our bounds, we have
\[
C_{\text{FIFO}}(\sigma) \leq mk + k = k \cdot m + k \leq k \cdot C_{\text{OPT}}(\sigma) + k.
\]

Setting $c = k$, we conclude that FIFO is $k$-competitive.
\end{proof}

\subsection{Problem 13.3}

\subsubsection{Problem Statement}

\begin{theorem}
The LFU (Least Frequently Used) algorithm does not achieve a bounded competitiveness coefficient.
\end{theorem}

\subsubsection{GPT}

\begin{proof}
We will show that the Least Frequently Used (LFU) algorithm does not achieve a bounded competitive ratio. To do this, we will construct an adversarial sequence of page requests for which the LFU algorithm performs arbitrarily poorly compared to the optimal offline algorithm.

Let the number of page frames available to the algorithm be $k$. Suppose there is a sequence of page requests such that the optimal offline algorithm can achieve a cost of $O(k)$, but the LFU algorithm incurs a cost that grows arbitrarily large.

Consider the following adversarial sequence of page requests. Let the sequence consist of $k$ different pages, say $p_1, p_2, \dots, p_k$, which are requested in a repeating cycle. After the first cycle, the LFU algorithm will have replaced the least frequently used page, which in the case of an equal number of accesses, could be any of the pages in the cycle. For simplicity, assume that after the first cycle, LFU replaces $p_1$. The next cycle of requests again contains all $k$ pages, but now $p_1$ will be the least frequently used, and it will be evicted.

Now, consider the sequence in which a new page, say $p_{k+1}$, is added after every $k$ cycles. After each new page is introduced, the LFU algorithm will continue evicting the least frequently used page, while the optimal offline algorithm, knowing the entire sequence of requests, will always keep the pages that are most beneficial for future requests. As the number of pages grows, the LFU algorithm will incur an increasing number of page faults, while the optimal offline algorithm can avoid these faults entirely.

At this point, the number of page faults incurred by the LFU algorithm will grow without bound, while the optimal offline algorithm incurs at most $O(k)$ page faults, since it can always keep the $k$ most frequently requested pages. Thus, the competitive ratio of the LFU algorithm, defined as the ratio of the number of page faults incurred by the LFU algorithm to the number of page faults incurred by the optimal offline algorithm, grows without bound as the sequence length increases.

Therefore, the LFU algorithm does not achieve a bounded competitive ratio.
\end{proof}

\subsubsection{Grok}

\begin{proof}
To show that the LFU algorithm does not achieve a bounded competitiveness coefficient, we demonstrate that its competitive ratio is unbounded. That is, for any constant $c > 0$, there exists an input sequence $\sigma$ such that the cost of LFU on $\sigma$, denoted $\mathrm{LFU}(\sigma)$, satisfies $\mathrm{LFU}(\sigma) > c \cdot \mathrm{OPT}(\sigma)$, where $\mathrm{OPT}(\sigma)$ is the cost of the optimal offline algorithm and the cache size is $k \geq 2$.

Fix $k \geq 2$ and choose an integer $m > \lceil c(k+1)/2 \rceil + 1$. Consider the universe of $k+1$ pages labeled $1, 2, \dots, k+1$ and the request sequence $\sigma$ defined as follows:

First, for each $i = 1, 2, \dots, k-1$, issue $m$ requests to page $i$ consecutively. Then, issue the sequence $(k, k+1)$ repeated $m-1$ times.

We analyze the number of page faults (the cost) incurred by both LFU and OPT on $\sigma$. Assume that LFU breaks ties in favor of the adversary (e.g., evicting the page that maximizes future faults).

\paragraph{Analysis of OPT.} The optimal offline algorithm knows the entire sequence $\sigma$ in advance. The distinct pages requested are $1, 2, \dots, k+1$. OPT incurs at most one fault per distinct page, for a total of at most $k+1$ faults: specifically, $k-1$ faults while loading pages $1$ through $k-1$ in the first phase, followed by one fault each to load $k$ and $k+1$ in the second phase. After loading $k$ and $k+1$, all subsequent requests in the second phase hit in the cache (OPT evicts irrelevant pages from $\{1, \dots, k-1\}$ as needed). Thus, $\mathrm{OPT}(\sigma) \leq k+1$.

\paragraph{Analysis of LFU.} In the first phase, LFU incurs exactly $k-1$ faults: one to load each of pages $1$ through $k-1$, with the remaining $m-1$ requests to each hitting in the cache. After this phase, the cache contains pages $1$ through $k-1$ (occupying $k-1$ slots, leaving one slot free), and each has frequency $m$.

In the second phase, the first request to $k$ causes a fault and loads $k$ into the free slot, increasing its frequency to $1$. The cache now contains pages $1$ through $k$, with frequencies $m$ for $1$ through $k-1$ and $1$ for $k$.

The next request (to $k+1$) causes a fault. The cache is now full, so LFU evicts the least frequently used page in the cache, which is $k$ (frequency $1 < m$), loads $k+1$, and increases its frequency to $1$. The cache now contains $1$ through $k-1$ and $k+1$, with frequencies $m$ for $1$ through $k-1$ and $1$ for $k+1$.

For each of the remaining $m-2$ pairs $(k, k+1)$, LFU incurs two faults per pair. Consider the $j$-th such pair (for $j = 2, \dots, m-1$), where the frequencies of $k$ and $k+1$ entering the pair are both $j$. The request to $k$ causes a fault (since $k$ was evicted previously); LFU evicts $k+1$ (frequency $j < m$) and loads $k$, increasing its frequency to $j+1$. The subsequent request to $k+1$ causes a fault; LFU evicts $k$ (now frequency $j+1 \leq m-1 < m$, assuming $m > 2$) and loads $k+1$, increasing its frequency to $j+1$. Thus, after $m-1$ pairs, LFU incurs $2(m-1)$ faults in the second phase.

The total cost is $\mathrm{LFU}(\sigma) = (k-1) + 2(m-1)$. Therefore,
\[
\frac{\mathrm{LFU}(\sigma)}{\mathrm{OPT}(\sigma)} \geq \frac{k-1 + 2(m-1)}{k+1} > c,
\]
by the choice of $m$. Since $c > 0$ was arbitrary, the competitive ratio of LFU is unbounded.
\end{proof}

\subsubsection{Gemini}

\begin{proof}
Let $k \geq 2$ be the size of the cache. We proceed by constructing a request sequence $\sigma$ involving $k+1$ distinct pages, denoted by the set $\mathcal{P} = \{p_1, p_2, \dots, p_{k-1}, p_k, p_{k+1}\}$. We aim to show that the ratio of the number of page faults incurred by the LFU algorithm to the number of page faults incurred by the optimal offline algorithm (OPT) can be made arbitrarily large.

Let $n$ be an arbitrary positive integer. We define $m = n + 1$. We construct the request sequence $\sigma$ in two phases, $\sigma = \sigma_1 \sigma_2$.

In the first phase $\sigma_1$, we request each of the pages $p_1, p_2, \dots, p_{k-1}$ exactly $m$ times. The order of requests within this phase does not affect the asymptotic analysis, provided each page is requested $m$ times. At the end of $\sigma_1$, the LFU algorithm will have the pages $\{p_1, \dots, p_{k-1}\}$ in its cache (assuming the cache was initially empty or contained other pages that were evicted). The frequency count for each page $p_i$ for $1 \leq i \leq k-1$ is exactly $m$.

In the second phase $\sigma_2$, we issue the sequence of requests $(p_k, p_{k+1})$ repeated $n$ times. That is, $\sigma_2 = p_k, p_{k+1}, p_k, p_{k+1}, \dots, p_k, p_{k+1}$. The length of $\sigma_2$ is $2n$.

We analyze the behavior of the LFU algorithm during $\sigma_2$. The cache has size $k$. The $k-1$ pages $\{p_1, \dots, p_{k-1}\}$ occupy $k-1$ slots. There is exactly one slot remaining for either $p_k$ or $p_{k+1}$. Throughout $\sigma_2$, the maximum frequency any page $p_k$ or $p_{k+1}$ can achieve is $n$. Since $m = n + 1$, the frequency of any page in $\{p_1, \dots, p_{k-1}\}$ is strictly greater than the frequency of $p_k$ or $p_{k+1}$ at any point in time during $\sigma_2$. By the definition of LFU, the algorithm evicts the page with the lowest frequency count. Consequently, LFU will never evict any page from the set $\{p_1, \dots, p_{k-1}\}$.

Therefore, LFU must manage the requests for $p_k$ and $p_{k+1}$ using the single remaining cache slot. When $p_k$ is requested, it causes a fault (if not present), and if $p_{k+1}$ is in the slot, $p_{k+1}$ is evicted because its frequency is less than $m$. Subsequently, when $p_{k+1}$ is requested, it causes a fault, and $p_k$ is evicted. This oscillation continues for the entire duration of $\sigma_2$. The total number of page faults incurred by LFU during $\sigma_2$ is exactly $2n$. The total cost for the entire sequence is $Cost_{LFU}(\sigma) \geq 2n$.

Now we consider the optimal offline algorithm, OPT. OPT has perfect knowledge of the request sequence. Upon the commencement of $\sigma_2$, OPT knows that the pages $\{p_1, \dots, p_{k-1}\}$ will not be requested again. Therefore, OPT can retain $p_k$ and $p_{k+1}$ in the cache permanently during $\sigma_2$ by evicting any two pages from $\{p_1, \dots, p_{k-1}\}$. The total number of distinct pages in $\sigma$ is $k+1$. Thus, the total number of page faults incurred by OPT is bounded by the number of distinct pages requested, so $Cost_{OPT}(\sigma) \leq k+1$.

We examine the competitive ratio $r$. We have:
\[
r = \frac{Cost_{LFU}(\sigma)}{Cost_{OPT}(\sigma)} \geq \frac{2n}{k+1}
\]
Since $k$ is a fixed constant representing the cache size, and $n$ can be chosen to be arbitrarily large, the ratio $\frac{2n}{k+1}$ grows without bound as $n \to \infty$.

For an algorithm to be $c$-competitive, there must exist constants $c$ and $b$ such that $Cost_{ALG}(\sigma) \leq c \cdot Cost_{OPT}(\sigma) + b$ for all $\sigma$. However, for any fixed $c$, we can choose $n$ such that $\frac{2n}{k+1} > c$. Thus, LFU does not achieve a bounded competitiveness coefficient.
\end{proof}

\subsubsection{Claude}

\begin{proof}
We prove that the LFU (Least Frequently Used) caching algorithm does not achieve a bounded competitive ratio by constructing an adversarial input sequence that forces an arbitrarily large ratio between the cost incurred by LFU and the cost incurred by an optimal offline algorithm.

Let $k$ denote the cache size and let $n = k + 1$ denote the number of distinct pages in our construction. Consider an arbitrary positive integer $m \geq 1$, which will serve as a parameter controlling the competitive ratio.

\textbf{Construction of the adversarial sequence:}

We construct the request sequence $\sigma$ in phases. The sequence consists of pages labeled $p_1, p_2, \ldots, p_k, p_{k+1}$.

\textit{Phase 0 (Initialization):} Request each page $p_1, p_2, \ldots, p_k$ exactly once. At the end of this phase, both LFU and OPT have all pages $p_1, \ldots, p_k$ in cache with frequency count 1 for each page.

\textit{Phase $i$ for $i = 1, 2, \ldots, m$:} In phase $i$, we perform the following:
\begin{enumerate}
\item[(a)] Request page $p_{k+1}$ once.
\item[(b)] Request pages $p_1, p_2, \ldots, p_k$ each exactly once.
\end{enumerate}

\textbf{Analysis of LFU's behavior:}

In Phase 0, LFU caches pages $p_1, \ldots, p_k$, each with frequency 1. LFU incurs $k$ page faults.

Consider Phase $i$ where $i \geq 1$. At the beginning of phase $i$:
\begin{itemize}
\item If $i = 1$, all pages $p_1, \ldots, p_k$ have frequency $1$ and $p_{k+1}$ is not in cache.
\item If $i > 1$, pages $p_1, \ldots, p_k$ each have frequency $i$, and $p_{k+1}$ is not in cache (having been evicted in phase $i-1$).
\end{itemize}

When $p_{k+1}$ is requested in step (a) of phase $i$:
\begin{itemize}
\item LFU incurs a page fault.
\item LFU must evict one of the pages with minimum frequency. In phase 1, all cached pages have frequency 1, so LFU evicts some page, say $p_1$ without loss of generality. In phase $i > 1$, all cached pages have frequency $i$, so again LFU evicts one of them.
\item After this request, $p_{k+1}$ has frequency 1.
\end{itemize}

In step (b) of phase $i$, when pages $p_1, \ldots, p_k$ are requested:
\begin{itemize}
\item One of these pages (the one evicted in step (a)) causes a page fault for LFU.
\item LFU evicts $p_{k+1}$ since it has the minimum frequency (frequency 1).
\item The other $k-1$ pages are hits for LFU.
\end{itemize}

Therefore, in each phase $i$ for $i \geq 1$, LFU incurs exactly 2 page faults.

The total number of page faults for LFU over all $m$ phases is:
\[
\text{LFU}(\sigma) = k + 2m
\]

\textbf{Analysis of OPT's behavior:}

An optimal offline algorithm OPT knows the entire request sequence in advance. OPT can use the following strategy:

In Phase 0, OPT caches pages $p_1, \ldots, p_k$ and incurs $k$ page faults.

For phases $i = 1, 2, \ldots, m$, OPT maintains pages $p_1, \ldots, p_k$ in cache throughout. When $p_{k+1}$ is requested in step (a), OPT incurs a page fault and evicts $p_{k+1}$ immediately after serving the request (or uses a strategy that effectively ignores $p_{k+1}$ by evicting it to make room for the subsequent requests). Since pages $p_1, \ldots, p_k$ are already in cache, step (b) results in no page faults for OPT.

More precisely, in each phase $i$, OPT incurs exactly 1 page fault (for page $p_{k+1}$) and can immediately evict $p_{k+1}$ after serving it, maintaining $p_1, \ldots, p_k$ in cache.

The total number of page faults for OPT is:
\[
\text{OPT}(\sigma) = k + m
\]

\textbf{Computing the competitive ratio:}

The competitive ratio achieved by LFU on sequence $\sigma$ is:
\[
\frac{\text{LFU}(\sigma)}{\text{OPT}(\sigma)} = \frac{k + 2m}{k + m}
\]

Taking the limit as $m \to \infty$:
\[
\lim_{m \to \infty} \frac{k + 2m}{k + m} = \lim_{m \to \infty} \frac{2m(1 + k/(2m))}{m(1 + k/m)} = \frac{2 \cdot 1}{1} = 2
\]

However, this shows that for any fixed $k$, we can construct sequences where the ratio approaches 2. To show unbounded competitiveness, we note that for any proposed competitive ratio $c > 0$, we can choose $m$ sufficiently large such that:
\[
\frac{k + 2m}{k + m} > c - \epsilon
\]
for arbitrarily small $\epsilon > 0$ when $c \leq 2$. For $c > 2$, the standard definition of competitive ratio for randomized or deterministic algorithms requires that the inequality $\text{ALG}(\sigma) \leq c \cdot \text{OPT}(\sigma) + \alpha$ holds for all sequences $\sigma$ with some constant $\alpha$ independent of $\sigma$. Since the ratio $\frac{k+2m}{k+m}$ can be made arbitrarily close to 2, and since this is a deterministic lower bound that holds for infinitely many sequence lengths, LFU fails to achieve any competitive ratio $c < 2$ with bounded additive constant. Moreover, by considering variations of this construction with different cache sizes and patterns, one can show that no finite competitive ratio bound exists that holds uniformly across all possible input sequences and cache configurations.

Therefore, the LFU algorithm does not achieve a bounded competitive ratio in the general online caching problem.
\end{proof}

\subsection{Problem 13.4}

\subsubsection{Problem Statement}

\begin{theorem}[Vizing's Theorem and Online Edge Coloring Lower Bound]
Let $G = (V, E)$ be an undirected graph with maximum degree $\Delta$. An edge coloring is an assignment of indices $1, \ldots, C$ to the edges of $G$ such that no two edges incident on a vertex have the same label. The chromatic index $\chi'(G)$ is the smallest value of $C$ for which such a coloring can be achieved.

Vizing's Theorem states that $\chi'(G) \in \{\Delta, \Delta + 1\}$.

In the online edge coloring problem, edges of a graph with maximum degree $\Delta$ are presented one by one, and as each edge is specified it must be irrevocably assigned a color.

\textbf{Part (a):} There exists a deterministic online algorithm that uses at most $2\Delta - 1$ colors.

\textbf{Part (b):} For $\Delta \geq 2$, there does not exist any deterministic online algorithm that uses fewer than $2\Delta - 1$ colors in the worst case. Specifically, an adversary can present a sequence of edges that constitute a graph composed of disjoint stars, where each star consists of a center vertex $v$ with $\Delta - 1$ neighbors of degree $1$. After the algorithm commits to a coloring of these stars, the adversary can introduce additional edges from a distinguished vertex to the centers of appropriately chosen stars, forcing the online algorithm to use a large number of additional colors.

\textbf{Part (c):} For $\Delta \geq 2$, there does not exist any deterministic online algorithm that uses fewer than $2\Delta - 1$ colors in the worst case. The proof follows the same adversarial construction as in part (b).
\end{theorem}

\subsubsection{GPT}

\begin{proof}
Let $G = (V, E)$ be an undirected graph with maximum degree $\Delta$. We are tasked with proving that the chromatic index of $G$ is either $\Delta$ or $\Delta + 1$.

Step 1: Understanding the Chromatic Index

The chromatic index of a graph $G$, denoted by $\chi'(G)$, is defined as the minimum number of colors needed to color the edges of $G$ such that no two edges sharing a common vertex have the same color. Vizing's Theorem asserts that for any graph with maximum degree $\Delta$, the chromatic index must either be $\Delta$ or $\Delta + 1$.

Step 2: Upper Bound of the Chromatic Index

We first show that $\chi'(G) \leq \Delta + 1$. Consider an edge coloring algorithm that proceeds by coloring the edges of $G$ one by one. Since the maximum degree of any vertex in $G$ is $\Delta$, any vertex can be incident to at most $\Delta$ edges. Initially, each edge can be colored with one of the $\Delta + 1$ available colors, ensuring that no two edges incident to the same vertex receive the same color. Since there are at most $\Delta + 1$ colors available and no two edges incident to the same vertex can share a color, the algorithm guarantees a proper edge coloring with at most $\Delta + 1$ colors.

Thus, the chromatic index satisfies the upper bound:
\[
\chi'(G) \leq \Delta + 1.
\]

Step 3: Lower Bound of the Chromatic Index

Next, we show that $\chi'(G) \geq \Delta$ by providing a construction where a graph requires at least $\Delta$ colors. Consider a graph $G$ where every vertex has degree $\Delta$. In this case, any edge coloring of $G$ must assign a distinct color to at least $\Delta$ edges incident to each vertex, otherwise, two edges sharing a vertex would receive the same color. Therefore, $\chi'(G) \geq \Delta$.

Thus, we establish the lower bound:
\[
\chi'(G) \geq \Delta.
\]

Step 4: Conclusion

Since we have shown that $\chi'(G) \geq \Delta$ and $\chi'(G) \leq \Delta + 1$, by the properties of the chromatic index, it follows that:
\[
\chi'(G) \in \{\Delta, \Delta + 1\}.
\]
This completes the proof of Vizing's Theorem.
\end{proof}

\subsubsection{Grok}

\begin{proof}
We first prove Part (a). Consider the following deterministic online algorithm $\mathcal{A}$: upon presentation of an edge $uv$, assign to it the smallest positive integer that does not appear on any edge already colored and incident to $u$ or to $v$.

To bound the number of colors used, fix an arbitrary edge $uv$ presented to $\mathcal{A}$. At the moment of its presentation, $u$ has at most $\Delta-1$ incident edges already presented (and thus colored), each using a distinct color at $u$, and similarly $v$ has at most $\Delta-1$ incident edges already colored, each using a distinct color at $v$. The forbidden colors for $uv$ are thus the at most $2(\Delta-1)$ colors used on these edges, but possibly with overlaps, so there are at most $2\Delta-2$ distinct forbidden colors. Therefore, the smallest available positive integer is at most $2\Delta-1$, and $\mathcal{A}$ uses at most $2\Delta-1$ colors in total.

We now prove Parts (b) and (c). Let $\Delta \geq 2$ and let $\mathcal{A}$ be any deterministic online edge coloring algorithm. We construct an adaptive sequence of edges with maximum degree $\Delta$ that forces $\mathcal{A}$ to use at least $2\Delta-1$ colors.

The adversary first constructs $\Delta$ disjoint stars as follows. For each $i=1,\dots,\Delta$, introduce a new vertex $v_i$ (the center) and $\Delta-1$ new vertices $\ell_{i,1},\dots,\ell_{i,\Delta-1}$ (the leaves), and present the edges $v_i \ell_{i,j}$ for $j=1,\dots,\Delta-1$, in that order. These stars are vertex-disjoint, and the sequence of edges for the $i$-th star is identical to that for the $(i-1)$-th star (up to vertex relabeling). By determinism of $\mathcal{A}$, the colors assigned to the edges incident to $v_i$ form the same set $S$ of $\Delta-1$ distinct positive integers for every $i=1,\dots,\Delta$. (The leaves each receive one of these colors, but this is immaterial.) After presenting all $\Delta(\Delta-1)$ edges of the stars, the colors used so far are precisely those in $S$.

The adversary now introduces a new vertex $w$ (of degree $0$ so far) and presents the edges $e_k := w v_k$ for $k=1,\dots,\Delta$, in that order. Let $c_k$ denote the color assigned by $\mathcal{A}$ to $e_k$. We claim that $c_k \notin S$ for all $k$ and that the $c_k$ are pairwise distinct.

Proceed by induction on $k$. For the base case $k=1$, the edges incident to $v_1$ use colors $S$ and $w$ has no incident edges, so the forbidden colors for $e_1$ are exactly $S$. Thus, $c_1 \notin S$.

Now fix $k \geq 2$ and assume that $c_1,\dots,c_{k-1}$ are distinct and each $c_j \notin S$ for $j=1,\dots,k-1$. For $e_k$, the forbidden colors are $S$ (used at $v_k$) union $\{c_1,\dots,c_{k-1}\}$ (used at $w$). By the induction hypothesis, $\{c_1,\dots,c_{k-1}\} \cap S = \emptyset$, so the forbidden set has size exactly $(\Delta-1) + (k-1) = \Delta + k - 2$. Thus, $\mathcal{A}$ assigns $c_k$ outside this set, so $c_k \notin S$ and $c_k \neq c_j$ for all $j < k$, completing the induction.

In particular, $c_1,\dots,c_{\Delta-1}$ are $\Delta-1$ distinct colors, all outside $S$. Before presenting $e_\Delta$, the colors used so far are $S \cup \{c_1,\dots,c_{\Delta-1}\}$, a set of $2\Delta-2$ distinct colors. For $e_\Delta$, the forbidden colors are exactly $S \cup \{c_1,\dots,c_{\Delta-1}\}$, so all $2\Delta-2$ previously used colors are forbidden.

Suppose for contradiction that $\mathcal{A}$ uses at most $2\Delta-2$ colors on the entire input. Then no color is available for $e_\Delta$ without introducing a $(2\Delta-1)$-th color, a contradiction. Thus, $\mathcal{A}$ uses at least $2\Delta-1$ colors.

The final graph has maximum degree $\Delta$: each $v_i$ has degree $\Delta$, $w$ has degree $\Delta$, and each leaf has degree $1$. This completes the proof.
\end{proof}

\subsubsection{Gemini}

\begin{proof}
\textbf{(a)} We prove that the greedy algorithm uses at most $2\Delta - 1$ colors.
Let the set of available colors be $\mathcal{C} = \{1, 2, \ldots, 2\Delta - 1\}$.
Consider an arbitrary edge $e = (u, v)$ presented by the adversary. Let $E_u$ be the set of edges incident to $u$ that have already been colored, and $E_v$ be the set of edges incident to $v$ that have already been colored.
Since the maximum degree of the final graph is $\Delta$, at the moment $e$ arrives, $u$ is incident to at most $\Delta - 1$ other colored edges, and $v$ is incident to at most $\Delta - 1$ other colored edges.
Let $C(u)$ and $C(v)$ denote the sets of colors assigned to edges in $E_u$ and $E_v$ respectively. We have:
\[
|C(u)| \leq \Delta - 1 \quad \text{and} \quad |C(v)| \leq \Delta - 1.
\]
The set of colors forbidden for the edge $e$ is $C(u) \cup C(v)$. The number of forbidden colors is:
\[
|C(u) \cup C(v)| \leq |C(u)| + |C(v)| \leq (\Delta - 1) + (\Delta - 1) = 2\Delta - 2.
\]
Since $|\mathcal{C}| = 2\Delta - 1$, there exists at least one color $c \in \mathcal{C}$ such that $c \notin C(u) \cup C(v)$. The greedy algorithm assigns the smallest such color to $e$. Thus, the algorithm successfully colors any graph with maximum degree $\Delta$ using at most $2\Delta - 1$ colors.

\textbf{(b)} We prove the lower bound by constructing an adversarial strategy. Assume for the sake of contradiction that there exists a deterministic online algorithm $\mathcal{A}$ that uses at most $2\Delta - 2$ colors. Let the set of available colors be $\Gamma$, with $|\Gamma| = 2\Delta - 2$.

The adversary proceeds in two phases:
\textit{Phase 1:} The adversary constructs a sequence of disjoint stars. A star $S_k$ consists of a center vertex $c_k$ and $\Delta - 1$ leaf vertices. The adversary presents the $\Delta - 1$ edges of $S_k$ to the algorithm. Since the degree of $c_k$ is $\Delta - 1$, the algorithm must assign $\Delta - 1$ distinct colors to these edges. Let $K_k \subset \Gamma$ be the set of colors used for $S_k$. Note that $|K_k| = \Delta - 1$.

There are finitely many subsets of $\Gamma$ of size $\Delta - 1$. Let $N = \binom{2\Delta - 2}{\Delta - 1}$. By the Pigeonhole Principle, if the adversary generates $M = \Delta \cdot N + 1$ disjoint stars, there must exist a sub-collection of $\Delta$ stars, say $S_{i_1}, S_{i_2}, \ldots, S_{i_\Delta}$, such that the algorithm assigns the exact same set of colors to the edges of each star. Let this common color set be $K^*$.
\[
K_{i_1} = K_{i_2} = \cdots = K_{i_\Delta} = K^*, \quad \text{with } |K^*| = \Delta - 1.
\]

\textit{Phase 2:} The adversary introduces a new vertex $u$ and connects it to the centers $c_{i_1}, c_{i_2}, \ldots, c_{i_\Delta}$ of the identified stars.
The degree of $u$ is $\Delta$. The degree of each center $c_{i_j}$ becomes $(\Delta - 1) + 1 = \Delta$. Thus, the maximum degree constraint is satisfied.
Consider the edges $e_j = (u, c_{i_j})$ for $j = 1, \ldots, \Delta$.
1. Since all $e_j$ are incident to $u$, they must be assigned distinct colors. Let $C_{new} = \{ \text{color}(e_1), \ldots, \text{color}(e_\Delta) \}$. We require $|C_{new}| = \Delta$.
2. For each $j$, the edge $e_j$ is incident to $c_{i_j}$. The edges already incident to $c_{i_j}$ use the colors in $K^*$. Thus, $\text{color}(e_j) \notin K^*$.
This implies $C_{new} \subseteq \Gamma \setminus K^*$.
The number of available colors in $\Gamma \setminus K^*$ is:
\[
|\Gamma| - |K^*| = (2\Delta - 2) - (\Delta - 1) = \Delta - 1.
\]
However, we need to select $\Delta$ distinct colors for the edges incident to $u$. By the Pigeonhole Principle, this is impossible.
Thus, no deterministic online algorithm can color this graph with fewer than $2\Delta - 1$ colors.

\textbf{(c)} We extend the lower bound to randomized algorithms against an oblivious adversary. An oblivious adversary must fix the input graph sequence in advance.
We observe that the construction in part (b) relies on finding a set of $\Delta$ stars with identical color sets. While an adaptive adversary finds these by observing the algorithm, an oblivious adversary can ensure such a configuration exists by exhaustion.

We construct a fixed graph $G$ (a large tree) as follows. Let $M = \binom{2\Delta - 2}{\Delta - 1}$. The adversary constructs a large collection of disjoint stars, specifically $L = \Delta \cdot M + 1$ stars. Let these stars be the leaves of a larger tree structure.
However, a more direct construction for the oblivious case is a complete $\Delta$-ary tree or a recursive structure. A known result establishes that for any $\Delta \geq 2$, there exists a fixed tree $T$ such that any edge coloring of $T$ requires $2\Delta - 1$ colors under online constraints.
To be precise, we construct a tree $T$ recursively. Let $T_k$ be a tree that forces the use of $k$ colors. To force $k+1$ colors, we construct a tree by taking sufficiently many copies of $T_k$ and connecting their roots to a new parent vertex.
Specifically, to force $2\Delta - 1$ colors, consider the fixed graph $G$ constructed in part (b) but modified for the oblivious setting:
The adversary presents a set of disjoint stars $S_1, \ldots, S_Z$ where $Z$ is sufficiently large. Then, the adversary adds a layer of vertices connecting disjoint groups of $\Delta$ stars.
Since the number of possible color sets of size $\Delta-1$ is finite, for any coloring of the stars (deterministic or randomized), the distribution of color sets is fixed.
More rigorously, if a randomized algorithm $\mathcal{R}$ uses at most $2\Delta - 2$ colors in the worst case, it must produce a valid coloring with probability 1. This implies that for any input sequence, the support of the output distribution contains only valid colorings using $\leq 2\Delta - 2$ colors.
Consider the fixed graph $G$ consisting of $\Delta \cdot \binom{2\Delta-2}{\Delta-1} + 1$ disjoint stars $S_j$, followed by edges connecting the centers of every possible $\Delta$-tuple of stars to new vertices $u_{tuple}$. (Note: To maintain max degree $\Delta$, we cannot connect all tuples, but we can arrange disjoint tuples).
A simpler fixed construction is a tree of depth $2\Delta$. It is a standard result in online edge coloring that there exists a tree $T$ such that $\chi'_{online}(T) = 2\Delta - 1$. Since $T$ is a fixed graph, the adversary simply presents the edges of $T$ in a breadth-first or bottom-up manner.
Since the graph and order are fixed, the adversary is oblivious. If an algorithm claims to use fewer than $2\Delta - 1$ colors in the worst case, it must be able to color $T$ with fewer than $2\Delta - 1$ colors. However, the structure of $T$ forces a conflict (similar to the pigeonhole argument in (b)) regardless of the specific choices made, provided the choices are valid.
Therefore, no randomized online algorithm can guarantee a coloring with fewer than $2\Delta - 1$ colors against an oblivious adversary.
\end{proof}

\subsubsection{Claude}

\begin{proof}
We prove each part of the theorem in turn.

\textbf{Proof of Part (a):} We construct a deterministic online algorithm that uses at most $2\Delta - 1$ colors. The algorithm maintains the following invariant: at any point in the online process, each edge is colored with one of the colors from the set $\{1, 2, \ldots, 2\Delta - 1\}$.

When a new edge $e = (u, v)$ arrives, we proceed as follows. Let $C_u$ denote the set of colors already used on edges incident to $u$, and let $C_v$ denote the set of colors already used on edges incident to $v$. Since $u$ has degree at most $\Delta$ in the current graph (including the new edge $e$), at most $\Delta - 1$ edges incident to $u$ have been colored prior to $e$ arriving. Similarly, at most $\Delta - 1$ edges incident to $v$ have been colored. Therefore, $|C_u| \leq \Delta - 1$ and $|C_v| \leq \Delta - 1$.

The set of forbidden colors for edge $e$ is $C_u \cup C_v$, which has cardinality at most $|C_u| + |C_v| \leq (\Delta - 1) + (\Delta - 1) = 2\Delta - 2$. Since we have $2\Delta - 1$ colors available, there exists at least one color not in $C_u \cup C_v$. The algorithm assigns edge $e$ any such available color.

This assignment is valid because the chosen color does not appear on any edge incident to $u$ or $v$, thus maintaining the edge coloring property. Since this procedure succeeds for every edge presented online, the algorithm uses at most $2\Delta - 1$ colors.

\textbf{Proof of Parts (b) and (c):} We prove that no deterministic online algorithm can use fewer than $2\Delta - 1$ colors in the worst case. The proof is by construction of an adversarial sequence.

Consider $\Delta \geq 2$. The adversary proceeds in two phases.

\textit{Phase 1:} The adversary presents $\Delta - 1$ disjoint stars $S_1, S_2, \ldots, S_{\Delta-1}$. Each star $S_i$ has a center vertex $c_i$ and $\Delta - 1$ leaf vertices $\ell_{i,1}, \ell_{i,2}, \ldots, \ell_{i,\Delta-1}$. The edges of star $S_i$ are $\{(c_i, \ell_{i,j}) : j = 1, 2, \ldots, \Delta - 1\}$. These edges are presented one star at a time, with all edges of $S_i$ presented before any edge of $S_{i+1}$.

When the online algorithm colors star $S_i$, it must assign $\Delta - 1$ distinct colors to the $\Delta - 1$ edges incident to the center $c_i$, since all these edges share the common vertex $c_i$. Let $\mathcal{C}_i = \{c_{i,1}, c_{i,2}, \ldots, c_{i,\Delta-1}\}$ denote the set of colors assigned to the edges of star $S_i$.

After Phase 1, the algorithm has used some set of colors. For each $i \in \{1, 2, \ldots, \Delta - 1\}$, the set $\mathcal{C}_i$ contains exactly $\Delta - 1$ colors. By the pigeonhole principle, if the algorithm uses at most $2\Delta - 2$ colors total, then there must exist two distinct stars $S_i$ and $S_j$ with $i \neq j$ such that $\mathcal{C}_i \cap \mathcal{C}_j \neq \emptyset$. That is, there exists a color $\alpha$ that appears in both $\mathcal{C}_i$ and $\mathcal{C}_j$.

To see this, suppose for contradiction that all sets $\mathcal{C}_1, \mathcal{C}_2, \ldots, \mathcal{C}_{\Delta-1}$ are pairwise disjoint. Then the total number of colors used would be at least $(\Delta - 1) \cdot (\Delta - 1) = (\Delta - 1)^2$. For $\Delta \geq 2$, we have $(\Delta - 1)^2 = \Delta^2 - 2\Delta + 1 \geq 4 - 4 + 1 = 1$. More precisely, for $\Delta = 2$, we get $(\Delta - 1)^2 = 1$, and for $\Delta \geq 3$, we have $(\Delta - 1)^2 \geq 4 > 2\Delta - 2$ since $\Delta^2 - 2\Delta + 1 > 2\Delta - 2$ implies $\Delta^2 - 4\Delta + 3 > 0$, which factors as $(\Delta - 1)(\Delta - 3) > 0$ for $\Delta > 3$, and can be verified directly for $\Delta = 3$ where $4 > 4$ is false. 

Let us reconsider the argument. We have $\Delta - 1$ stars, each using $\Delta - 1$ colors. If all color sets were pairwise disjoint, we would use at least $\Delta - 1$ colors total (not their product). However, we need a more careful argument.

\textit{Corrected Phase 2 argument:} The adversary introduces a new distinguished vertex $w$ that is not part of any star constructed in Phase 1. In Phase 2, the adversary selects centers $c_{i_1}, c_{i_2}, \ldots, c_{i_k}$ where the choice depends on the coloring produced by the algorithm. The adversary presents edges $(w, c_{i_1}), (w, c_{i_2}), \ldots, (w, c_{i_k})$ for appropriately chosen indices.

Specifically, suppose the algorithm has used exactly $K$ distinct colors in Phase 1. Consider the $\Delta - 1$ color sets $\mathcal{C}_1, \ldots, \mathcal{C}_{\Delta-1}$. The adversary constructs a new vertex $w$ and connects it to centers chosen such that the union of their color sets covers all $K$ colors used so far.

Since each center $c_i$ is incident to edges using all colors in $\mathcal{C}_i$, when edge $(w, c_i)$ is presented, it cannot use any color in $\mathcal{C}_i$. If the adversary connects $w$ to at least $K/(\Delta - 1)$ centers whose color sets are disjoint, then $w$ would be incident to edges that collectively forbid all $K$ colors, forcing the algorithm to use color $K + 1$.

More precisely, the adversary can select $\Delta - 1$ centers with disjoint color sets (if the algorithm uses at least $(\Delta-1)^2$ colors) or force repeated use that leads to exhausting available colors at vertex $w$. The adversary continues this process, adding edges from $w$ to selected centers, each time forcing new colors.

Through careful adversarial selection, it can be shown that the algorithm is forced to use at least $2\Delta - 1$ colors. The complete formal argument requires showing that any algorithm using $K < 2\Delta - 1$ colors can be forced by the adversary to require an additional color, establishing that $2\Delta - 1$ colors are necessary.
\end{proof}

\subsection{Problem 13.5}

\subsubsection{Problem Statement}

\begin{theorem}[A.R. Karlin]
The competitiveness coefficient of the Random algorithm for paging against adaptive offline adversaries is at least $kH_k$.
\end{theorem}

\subsubsection{GPT}

\begin{proof}
Let the paging problem consist of a sequence of page requests, where at each step the algorithm must choose a page to evict from memory. The goal is to evaluate the competitiveness of the Random algorithm with respect to the optimal offline algorithm, against adaptive offline adversaries. We will demonstrate that the competitiveness coefficient of the Random algorithm is at least $kH_k$, where $k$ is the number of pages and $H_k$ is the $k$-th harmonic number.

Let $\pi$ denote the sequence of page requests, and let $OPT$ be the optimal offline algorithm that knows the entire sequence in advance. Let $R$ denote the Random algorithm, which evicts a page randomly at each step. The performance of an algorithm is measured by the total number of page faults incurred during the execution of the sequence.

Define $F_R(\pi)$ to be the total number of page faults incurred by the Random algorithm and $F_{OPT}(\pi)$ to be the total number of page faults incurred by the optimal algorithm. The competitiveness of the Random algorithm is the ratio of the number of page faults incurred by Random to that of the optimal algorithm, i.e.,

\[
\text{Competitiveness of Random} = \frac{F_R(\pi)}{F_{OPT}(\pi)}.
\]

We seek to show that this ratio is at least $kH_k$. 

Consider the adversarial strategy where the optimal offline algorithm, $OPT$, always evicts the least recently used (LRU) page. This strategy is adaptive in the sense that the adversary can adapt to the choices made by the Random algorithm. The adversary chooses the sequence of page requests in such a way that it forces Random to incur as many page faults as possible. 

To analyze this, let us define the concept of a "window" of pages. At any point in time, the algorithm has a set of $k$ pages in memory, and the adversary adapts the sequence of page requests to maximize the number of page faults by the Random algorithm.

The Random algorithm performs poorly when it consistently evicts pages that are in the optimal set of pages, thereby incurring additional page faults. However, because the Random algorithm chooses pages to evict randomly, each page in memory has an equal probability of being evicted.

Now, we observe that the expected number of page faults incurred by Random is lower-bounded by the harmonic sum $H_k$. Specifically, the adversary forces Random to make at least one page fault for each page in the optimal set, and the expected number of page faults incurred by Random is related to the harmonic sum of $k$.

The harmonic number $H_k$ is defined as

\[
H_k = 1 + \frac{1}{2} + \frac{1}{3} + \cdots + \frac{1}{k}.
\]

It is well-known that the Random algorithm incurs at least $kH_k$ page faults when compared to the optimal offline algorithm. The factor $kH_k$ arises from the fact that the Random algorithm is not aware of the future page requests and thus suffers a loss proportional to the harmonic sum of the number of pages in memory.

Thus, the competitiveness coefficient of the Random algorithm against adaptive offline adversaries is at least $kH_k$. Therefore, we have

\[
\frac{F_R(\pi)}{F_{OPT}(\pi)} \geq kH_k,
\]

which completes the proof.
\end{proof}

\subsubsection{Claude}

\begin{proof}
We prove that the competitive ratio of the Random paging algorithm against an adaptive offline adversary is at least $kH_k$, where $k$ is the cache size and $H_k = \sum_{i=1}^{k} \frac{1}{i}$ is the $k$-th harmonic number.

Consider a request sequence constructed by an adaptive offline adversary as follows. Let the universe of pages be $\{p_1, p_2, \ldots, p_{k+1}\}$, consisting of $k+1$ distinct pages. Initially, Random's cache contains pages $\{p_1, p_2, \ldots, p_k\}$, and page $p_{k+1}$ is not in the cache.

The adversary constructs the request sequence in phases. In phase $j$ for $j \geq 1$, the adversary repeatedly requests the unique page not currently in Random's cache until Random evicts a specific page (to be determined). More precisely, at the beginning of phase $j$, let $S_j$ denote the set of pages in Random's cache. The adversary requests the unique page in $\{p_1, \ldots, p_{k+1}\} \setminus S_j$ repeatedly.

Let us analyze the first phase in detail. Initially, $S_1 = \{p_1, \ldots, p_k\}$, so the adversary requests $p_{k+1}$ repeatedly. Each request to $p_{k+1}$ causes a page fault for Random. When Random faults, it must evict one of the $k$ pages in its cache uniformly at random and bring in $p_{k+1}$. Let $X_1$ denote the number of requests until Random evicts page $p_k$ (we choose $p_k$ arbitrarily as the target). Each time Random faults, it evicts $p_k$ with probability $\frac{1}{k}$, so $X_1$ follows a geometric distribution with success probability $\frac{1}{k}$. Therefore, $\mathbb{E}[X_1] = k$.

During phase $1$, Random incurs $X_1$ page faults. Meanwhile, the offline adversary can arrange its cache initially as $\{p_1, \ldots, p_{k-1}, p_{k+1}\}$, thus incurring only $1$ page fault in the entire first phase (bringing in $p_{k+1}$ on the first request of the phase).

Now we describe phase $j$ generally. Suppose at the beginning of phase $j$, Random's cache contains $k$ pages, including page $p_{k+1}$, and is missing exactly one page from $\{p_1, \ldots, p_{k+1}\}$. The adversary requests this missing page repeatedly. Let $i_j$ denote the number of pages in Random's cache at the beginning of phase $j$ that are also in the set $\{p_1, \ldots, p_k\}$. Note that $1 \leq i_j \leq k$ and Random's cache contains exactly $i_j$ pages from $\{p_1, \ldots, p_k\}$ plus page $p_{k+1}$ (if $i_j < k$) or exactly $k$ pages from $\{p_1, \ldots, p_k\}$ (if $i_j = k$).

In phase $j$, each page fault causes Random to evict one of the $k$ pages uniformly at random. The phase ends when Random evicts page $p_{k+1}$. If Random's cache contains $i_j$ pages from $\{p_1, \ldots, p_k\}$ at the start, then Random contains either $i_j$ or $i_j + 1$ pages from $\{p_1, \ldots, p_k\}$ depending on whether $p_{k+1}$ was in cache. In either case, page $p_{k+1}$ is in exactly one configuration, and Random evicts $p_{k+1}$ with probability $\frac{1}{k}$ on each fault. Let $X_j$ be the number of requests in phase $j$. Then $\mathbb{E}[X_j \mid i_j] = k$.

However, we must be more careful. Let us reconsider the phase structure. In phase $j$, suppose Random's cache is missing exactly one page from $\{p_1, \ldots, p_{k+1}\}$. If the missing page is $p_i$ for $i \in \{1, \ldots, k+1\}$, the adversary requests $p_i$ repeatedly. Each request causes a fault for Random. Random evicts a page uniformly at random from the $k$ pages in its cache. The number of distinct pages that have been requested in phases $1$ through $j$ and must remain "active" is $k+1$.

Let us use a different accounting. Consider $n$ phases where in phase $j$, Random's cache contains exactly $k - j + 1$ pages from the original set $\{p_1, \ldots, p_k\}$. In phase $j$, when Random has $i = k - j + 1$ pages from $\{p_1, \ldots, p_k\}$ in cache, the adversary requests a page outside the cache. The expected number of faults until Random evicts one of the original $k$ pages is $\frac{k}{i}$, since among the $k$ pages in cache, exactly $i$ are from the target set.

Continuing for $k$ phases, Random incurs expected cost $\sum_{i=1}^{k} \frac{k}{i} = k \sum_{i=1}^{k} \frac{1}{i} = kH_k$, while the adaptive offline adversary incurs cost $O(k)$ by carefully managing its cache to have at most $O(1)$ faults per phase.

Therefore, the competitive ratio is at least $\frac{kH_k}{O(k)} = \Omega(H_k)$, and more precisely, the competitiveness coefficient is at least $kH_k$ when considering the ratio of costs over a sufficiently long sequence.
\end{proof}

\subsection{Problem 13.6}

\subsubsection{Problem Statement}

\begin{theorem}[13.6]
The competitiveness coefficient of the Random algorithm for paging against oblivious adversaries is at least $k$.
\end{theorem}

\subsubsection{GPT}

\begin{proof}
Let $k$ be a positive integer. We aim to prove that the competitiveness coefficient of the Random algorithm for paging against oblivious adversaries is at least $k$.

Let the paging problem be defined on a sequence of page requests, where the goal is to minimize the number of page faults incurred by the algorithm. The Random algorithm operates by selecting pages randomly from the set of available pages, without any knowledge of future requests. An oblivious adversary is an adversary that generates the sequence of page requests without knowledge of the algorithm's behavior, but it is assumed to be chosen in a manner that does not adapt to the algorithm's decisions.

Let the sequence of page requests be denoted as $r_1, r_2, \dots, r_n$, and let the set of pages be denoted by $P = \{p_1, p_2, \dots, p_k\}$, where $|P| = k$. The Random algorithm uses a cache of size $k$ to store pages and decides which page to evict randomly from the cache when a page fault occurs.

Let $OPT$ be the optimal offline algorithm that knows the entire sequence of page requests in advance and has access to unlimited time to process the requests. The optimal algorithm incurs the minimum number of page faults, which we denote as $OPT(r_1, r_2, \dots, r_n)$.

Let $ALG$ be the Random algorithm. The number of page faults incurred by the Random algorithm on the given sequence is denoted as $ALG(r_1, r_2, \dots, r_n)$. We seek to show that there exists a constant $c \geq 1$ such that for any sequence of page requests $r_1, r_2, \dots, r_n$, the following inequality holds:

\[
ALG(r_1, r_2, \dots, r_n) \geq c \cdot OPT(r_1, r_2, \dots, r_n)
\]

We first observe that the number of page faults incurred by the Random algorithm on any sequence is at least as large as the number of page faults incurred by the optimal algorithm. This is because the optimal algorithm has access to the entire sequence and can make the best possible decisions at each step, while the Random algorithm is forced to make decisions without this knowledge. Therefore, for any sequence of page requests, we have the following inequality:

\[
ALG(r_1, r_2, \dots, r_n) \geq OPT(r_1, r_2, \dots, r_n)
\]

Next, we show that the competitiveness coefficient of the Random algorithm is bounded below by $k$. Specifically, we consider a sequence of page requests where the Random algorithm is forced to incur a page fault at every step, while the optimal algorithm may avoid some page faults by utilizing its knowledge of future requests.

Consider a sequence of page requests where each page is requested exactly once, and the total number of requests is $n = k+1$. The optimal algorithm can avoid a page fault on the first $k$ requests by storing each page in the cache, and then incur one page fault on the $k+1$-th request. Therefore, the number of page faults incurred by the optimal algorithm is:

\[
OPT(r_1, r_2, \dots, r_n) = 1
\]

On the other hand, the Random algorithm is forced to make random evictions and may incur a page fault on each of the first $k$ requests, leading to $k$ page faults. Thus, the number of page faults incurred by the Random algorithm is:

\[
ALG(r_1, r_2, \dots, r_n) = k
\]

Therefore, the competitiveness coefficient of the Random algorithm on this sequence is:

\[
\frac{ALG(r_1, r_2, \dots, r_n)}{OPT(r_1, r_2, \dots, r_n)} = \frac{k}{1} = k
\]

This shows that the competitiveness coefficient of the Random algorithm is at least $k$.

Thus, we have proved that the competitiveness coefficient of the Random algorithm for paging against oblivious adversaries is at least $k$.
\end{proof}

\subsubsection{Claude}

\begin{proof}
We prove that the competitive ratio of the Random algorithm for paging against oblivious adversaries is at least $k$, where $k$ is the size of the cache.

Let the Random algorithm maintain a cache of size $k$, and consider a memory with $k+1$ distinct pages. We construct an adversary strategy that forces the Random algorithm to achieve a competitive ratio of at least $k$.

Consider the following request sequence: Let the pages be labeled $p_1, p_2, \ldots, p_{k+1}$. Initially, suppose the Random algorithm's cache contains pages $p_1, p_2, \ldots, p_k$. The adversary requests the pages in the pattern $p_{k+1}, p_{k+1}, p_{k+1}, \ldots$ repeatedly.

On the first request to $p_{k+1}$, the Random algorithm incurs a page fault since $p_{k+1}$ is not in its cache. The algorithm must evict one of the $k$ pages uniformly at random and load $p_{k+1}$. Let the evicted page be $p_i$ for some $i \in \{1, 2, \ldots, k\}$.

Now consider a more refined adversary strategy. The adversary maintains knowledge of which $k$ pages are currently in Random's cache (since the adversary is oblivious, this knowledge is based on the random choices made by the algorithm). After Random handles the first request, suppose the cache contains pages from the set $S \subseteq \{p_1, \ldots, p_{k+1}\}$ with $|S| = k$.

The adversary then requests the unique page $p_j \notin S$. This forces Random to incur another page fault. Random evicts one of the $k$ pages in its cache uniformly at random and loads $p_j$.

We analyze the expected cost. Consider a long sequence where the adversary repeatedly requests pages not in Random's current cache. Over $n$ such requests (for large $n$), Random incurs $n$ page faults.

Meanwhile, an optimal offline algorithm OPT knows the entire request sequence in advance. Since there are only $k+1$ distinct pages being requested, OPT can use a strategy that minimizes page faults. Specifically, consider the request sequence where the adversary alternates between requesting pages in a carefully chosen pattern.

Let us consider the following concrete analysis: Suppose the adversary uses the sequence where after every $k$ requests that are all distinct and cycle through $k+1$ pages, Random has incurred $k$ faults with probability approaching 1 as we repeat this pattern. On the same sequence, OPT incurs only 1 fault per cycle of $k$ requests by using the optimal eviction strategy (evicting the page whose next request is furthest in the future).

More precisely, consider the sequence $p_1, p_2, \ldots, p_k, p_{k+1}, p_1, p_2, \ldots, p_k, p_{k+1}, \ldots$ where we cycle through all $k+1$ pages repeatedly. Initially, both Random and OPT have caches containing $k$ pages. 

In each cycle of $k+1$ requests, Random incurs at least $k$ page faults in expectation. To see this, note that at any point when Random has $k$ pages in cache and a page not in cache is requested, a fault occurs. Since we cycle through $k+1$ pages and the cache holds only $k$ pages, Random must fault on at least $k$ requests per cycle in expectation.

In contrast, OPT can use the MIN algorithm (evict the page whose next request is furthest in the future). In each cycle of $k+1$ requests, OPT incurs exactly 1 page fault by evicting the page that will be requested last in the cycle.

Therefore, over $m$ complete cycles, Random incurs approximately $mk$ faults in expectation, while OPT incurs $m$ faults. The competitive ratio is thus
\[
\frac{\text{Cost}_{\text{Random}}}{\text{Cost}_{\text{OPT}}} \geq \frac{mk}{m} = k.
\]

This establishes that the competitive ratio of the Random algorithm is at least $k$.
\end{proof}

\subsection{Problem 13.7}

\subsubsection{Problem Statement}

\begin{theorem}
When the number of distinct items in memory is $k+1$, the Marker algorithm is $H_k$-competitive.
\end{theorem}

\subsubsection{GPT}

\begin{proof}
Let the sequence of memory requests be denoted by $r_1, r_2, \dots, r_n$, where $n$ is the total number of requests. The Marker algorithm aims to manage memory by maintaining a set of items in memory. The goal of the proof is to demonstrate that, when the number of distinct items in memory is $k+1$, the Marker algorithm is $h_k$-competitive.

We define the competitive ratio of an algorithm as the worst-case ratio of the cost of the algorithm to the cost of an optimal offline algorithm. More formally, for an online algorithm $\mathcal{A}$, the competitive ratio is defined as

\[
\rho_{\mathcal{A}} = \sup_{r_1, r_2, \dots, r_n} \frac{C_{\mathcal{A}}(r_1, r_2, \dots, r_n)}{C_{\mathcal{OPT}}(r_1, r_2, \dots, r_n)},
\]
where $C_{\mathcal{A}}$ is the total cost incurred by algorithm $\mathcal{A}$ and $C_{\mathcal{OPT}}$ is the total cost incurred by the optimal offline algorithm.

We now proceed to show that the Marker algorithm is $h_k$-competitive when the number of distinct items in memory is $k+1$.

Let $M$ be the set of $k+1$ distinct items in memory at any given point in time. For each memory request $r_i$, the Marker algorithm must either bring $r_i$ into memory or evict an existing item to make room for $r_i$. In the worst case, the algorithm may incur a cost proportional to the number of memory evictions required.

Let $\mathcal{OPT}$ be the optimal offline algorithm, which has complete knowledge of the future memory requests. We can compare the performance of the Marker algorithm to that of $\mathcal{OPT}$ by considering the cost incurred by both algorithms over a sequence of $n$ requests.

First, observe that, in the case of $k+1$ distinct items, the optimal offline algorithm can always choose the best strategy by keeping the items that minimize the future costs, as it knows the entire sequence of requests. The Marker algorithm, however, must make decisions based solely on past and present requests, without knowledge of the future.

Now, consider the number of evictions that occur during the execution of the Marker algorithm. In the worst case, the algorithm may need to evict items each time a new item is requested, resulting in a total cost that is proportional to the number of distinct items that need to be loaded into memory.

Let $h_k$ represent the competitive ratio of the Marker algorithm. We claim that, for $k+1$ distinct items in memory, the Marker algorithm incurs at most $h_k$ times the cost of the optimal algorithm. This is because, at each step, the Marker algorithm makes a decision that, while not optimal, is guaranteed to be within a factor of $h_k$ of the optimal strategy in hindsight. Specifically, when the number of distinct items in memory is $k+1$, the Marker algorithm ensures that its total cost is bounded by a constant factor of the optimal offline cost.

Hence, the Marker algorithm is $h_k$-competitive when the number of distinct items in memory is $k+1$. This completes the proof.
\end{proof}

\subsubsection{Grok}

\begin{proof}
Consider the paging problem with $k$ cache slots and $k+1$ distinct pages. The Marker algorithm maintains a set of at most $k$ marked pages and $k$ servers (pages in cache). Initially, upon the first $k$ requests, the algorithm places servers on the requested pages without eviction and marks them. For subsequent requests $\sigma_t$ to page $p_t$:

- Mark $p_t$.
- If $p_t$ is in the cache, do nothing.
- Otherwise, select an unmarked page uniformly at random and evict it, placing a server on $p_t$.

When $k+1$ distinct pages are marked, unmark all except the most recently requested page.

We analyze the expected cost of the Marker algorithm, denoted $\mathbb{E}[\mathsf{cost}(\mathsf{Marker}; \sigma)]$, against any lazy online algorithm $\mathsf{ALG}$, where $\mathsf{cost}(\mathsf{ALG}; \sigma)$ denotes the number of cache misses (evictions). The optimal offline lazy algorithm $\mathsf{OPT}$ satisfies $\mathsf{cost}(\mathsf{OPT}; \sigma) \geq \mathsf{cost}(\mathsf{ALG}; \sigma)$ for any $\mathsf{ALG}$.

The request sequence $\sigma$ is partitioned into \emph{phases}, where each phase begins with a request to an unmarked (clean) page and ends upon $k+1$ distinct pages being requested in that phase. Since there are $k+1$ total pages, each phase consists of exactly $k+1$ requests: one to the clean page followed by requests to the remaining $k$ (stale) pages.

Fix a phase. Let $\mathsf{Marker}$ have servers on all pages except the clean page $c$ at the phase start (as the algorithm maintains $k$ servers on the $k$ marked pages). Let $d$ be the number of $\mathsf{ALG}$'s servers on $c$ at the phase start ($d \in \{0,1\}$ since $|c|=1$). The first request (to $c$) incurs at least $1 - d$ cost to $\mathsf{ALG}$, as at most $d$ servers cover $c$.

During the phase, $\mathsf{ALG}$ must cover all $k+1$ pages by the end. Let $d'$ be the number of $\mathsf{ALG}$'s servers not on $\mathsf{Marker}$'s final positions (i.e., on the page without a $\mathsf{Marker}$ server at phase end). To cover the $k$ stale pages, $\mathsf{ALG}$ incurs at least $d'$ cost for moves to those pages. Thus, $\mathsf{ALG}$'s total phase cost is at least $(1 - d) + d' = 1 + (d' - d)$.

Summing over all phases, the telescoping sum $\sum (d' - d) = 0$ (initial and final configurations match across phases), so $\mathsf{cost}(\mathsf{ALG}; \sigma) \geq \#\text{phases}$. Since $\mathsf{OPT}$ is at least as good, $\mathsf{cost}(\mathsf{OPT}; \sigma) \geq \#\text{phases}$.

Now bound $\mathsf{Marker}$'s expected phase cost. The clean request always causes 1 miss (eviction of a random unmarked page, but since only marked are in cache, and clean is out). For the $k$ stale requests, assume worst case: they occur after all marks are set, and clean requests maximize misses.

At phase start, all $k$ stale pages are marked and in cache. The clean miss moves a random server from one of the $k$ stale pages to the clean page, leaving $k-1$ stale pages in cache and one stale page out (the evicted one, unmarked now? Wait, marks: after clean, mark it, now k+1 marked, so unmark all except the clean one. So now only clean is marked, others unmarked.

The extraction says: when k+1 marked, unmark all except most recent (clean).

So after first request (clean): miss, move random unmarked (but initially all k in cache marked, clean unmarked out), so evict random from the k marked, move to clean, then mark clean, now marks: the k-1 remaining original + clean =k marked? No.

Let's clarify the algorithm precisely.

From standard description: the marks are on pages, not servers.

Upon request p:

- If p in cache, mark p (if not already? but marks are bits on pages).

The standard Marker: each page in cache has a mark bit, initially unmarked? No.

Standard: the algorithm marks the requested page upon access, regardless.

When number of marked pages reaches k+1, it unmarks all but the most recently marked one.

Eviction only when miss: evict a random \emph{unmarked} page in cache.

Important: eviction is random among currently unmarked pages in the cache.

Initially, assume first k requests fill cache, marking them, all marked.

Then, next request to new page (clean, unmarked): miss, now marked pages =k (all in cache), so to evict: unmarked in cache? All marked, so no unmarked, but wait, the algorithm assumes there are unmarked? Wait.

The standard initialization: pages start unmarked.

Upon request: mark the requested page.

If miss, evict random unmarked page from cache (if none, ? but in practice).

In the paper, for general n, but for n=k+1, it works.

From the bound in extraction: "the expected cost is bounded by summing the expected cost over the phase."

"The maximum expected cost occurs when all clean vertices are requested first, leading to a bound of $H_{n-1} - H_1 = H_{n-1} - 1$ on the expected cost for stale requests."

Total expected cost to M is at most $1 + (H_{n-1} - 1) = H_{n-1}.$

Since $n-1 =k, H_k$.

To formalize, in this special case, the phase after the clean miss: after moving to clean, marking it, and since k+1 marked? Let's simulate.

Assume at phase start: the k stale pages are in cache, all unmarked (because previous phase ended with unmarking all but last, but over time, it stabilizes to having k in cache unmarked? No.

Actually, in the analysis for this special case, the structure is such that after the clean miss, the algorithm unmarks all but the clean, so now marks only the clean, and cache has clean and k-1 stale, the evicted stale is out, unmarked.

Then, the next requests are to the k stale pages, one of which is out.

When requesting a stale p:

If p in cache (one of the k-1), mark it (now marks: clean and this p).

No miss.

If p is the out one, miss: evict random unmarked in cache. Currently marked: clean and possibly previous stale marks.

The marks accumulate until k+1 marked, but since phase has k stale requests, and starts with 1 mark (clean), after k stale, total k+1 marks, then unmark all but last.

For the misses on stale: only when requesting the out page, but wait, no: after first stale miss? Wait.

In this case, since only one out at a time, but after a miss on stale, evict random unmarked, move to it.

The probability of hitting the out one depends on the marking.

The analysis bounds the expected number of misses on stale requests by considering the probability that the requested stale is out, which is the prob it was chosen for eviction previously.

It's like the probability decreases harmonically.

The standard way in the special case is that the expected number of additional misses is the sum 1/k + 1/(k-1) + ... + 1/1 -1 or something.

From extraction: $H_{n-1} -1$ for stale.

Since worst case all clean first, but since only one clean per phase.

Anyway, to write it rigorously, I can say:

For the Marker algorithm in a phase, the cost for the clean request is 1.

For the subsequent k stale requests, the expected number of misses is at most $\sum_{i=1}^{k} \frac{1}{i} - 1 = H_k - 1$.

This follows from the fact that each miss on a stale page evicts a random unmarked page, and the marking prevents recent pages from being evicted, leading to a harmonic bound on the probability that a particular stale page is evicted before its request.

More precisely, consider the order of the k stale requests; the analysis shows that the expected number of times the requested page is not in cache (i.e., was evicted earlier) is bounded by the harmonic sum minus 1.

Thus, expected phase cost $\leq 1 + (H_k - 1) = H_k$.

Since there are $T$ phases, where $T \leq \mathsf{cost}(\mathsf{OPT}; \sigma)$, we have $\mathbb{E}[\mathsf{cost}(\mathsf{Marker}; \sigma)] \leq H_k \cdot T \leq H_k \cdot \mathsf{cost}(\mathsf{OPT}; \sigma)$.

Hence, the Marker algorithm is $H_k$-competitive.
\end{proof}

\subsubsection{Gemini}

\begin{proof}
Let $k$ be the size of the cache and let the total number of distinct items (pages) in the universe be $N = k+1$. We analyze the performance of the Marker algorithm against an oblivious adversary by decomposing the request sequence $\sigma$ into phases.

We define a phase as a maximal contiguous subsequence of requests containing exactly $k$ distinct pages. Specifically, phase $i$ begins immediately after phase $i-1$ ends, and ends immediately before the request for the $(k+1)$-th distinct page relative to the start of phase $i$. Let $P_i$ denote the set of $k$ distinct pages requested during phase $i$. Since the universe size is $k+1$, there exists exactly one page $z_i$ such that $z_i \notin P_i$.

First, we establish a lower bound for the optimal offline algorithm (OPT). Consider the sequence of requests comprising phase $i$ followed by the first request of phase $i+1$. Let $r$ be the first request of phase $i+1$. By definition, $r \notin P_i$. Thus, the combined sequence contains requests for all pages in $P_i \cup \{r\}$, which constitutes $k+1$ distinct pages. Since OPT has a cache of size $k$, it must incur at least one fault for every $k+1$ distinct pages requested. Therefore, if the sequence consists of $m$ phases, the cost of the optimal algorithm is bounded by:
\[
C_{OPT}(\sigma) \geq m.
\]

Next, we analyze the expected cost of the Marker algorithm in phase $i$. At the beginning of phase $i$, all marks are cleared. The algorithm marks each page $p \in P_i$ upon its first request in the phase. Let $S$ be the set of pages in the cache. Since $N=k+1$, the state of the cache is uniquely determined by the single page $h \notin S$, which we refer to as the "hole". A fault occurs if and only if the requested page is the hole.

At the start of phase $i$, let the hole be $h_{start}$. If $h_{start} \notin P_i$, then $P_i$ is exactly the set of pages currently in the cache. In this case, the Marker algorithm incurs 0 faults. We consider the worst-case scenario where $h_{start} \in P_i$. Let the distinct pages in $P_i$ be ordered $p_1, p_2, \dots, p_k$ according to their first appearance in the phase. Without loss of generality, assume the initial hole is $p_1$.

When a request for $p_j$ arrives:
1. If $p_j$ is in the cache (hit), it is marked. The set of unmarked pages decreases, but the hole remains unchanged.
2. If $p_j$ is not in the cache (fault), $p_j$ is brought in and marked. The algorithm evicts a page $u$ chosen uniformly at random from the set of currently unmarked pages in the cache. The hole moves to $u$.

Let $U_j$ be the set of unmarked pages in the universe immediately before the request for $p_j$ is processed. Note that $z_i$ (the page not in $P_i$) is never requested, so it remains unmarked throughout the phase. The set of candidates for the hole is a subset of the unmarked pages. Specifically, just before the first request for $p_j$, the pages $p_1, \dots, p_{j-1}$ are marked and are in the cache. The hole must be one of the remaining unmarked pages $\{p_j, \dots, p_k, z_i\}$.

We claim that just before the first request for $p_j$, the hole is uniformly distributed among the set $H_j = \{p_j, \dots, p_k, z_i\}$.
For $j=1$, the hole is $p_1$ with probability 1 (worst-case assumption), and we consider the distribution after the inevitable fault on $p_1$. Upon faulting on $p_1$, the algorithm evicts a random page from the unmarked pages in cache, which are $\{p_2, \dots, p_k, z_i\}$. Thus, the hole becomes uniformly distributed over a set of size $k$.
By induction, suppose the hole is uniformly distributed over $H_j$. The request is $p_j$.
If the hole is $p_j$ (a fault), we evict a random page from $H_j \setminus \{p_j\} = H_{j+1}$. The new hole is uniform over $H_{j+1}$.
If the hole is not $p_j$ (a hit), the hole is in $H_j \setminus \{p_j\} = H_{j+1}$. Since the prior distribution was uniform on $H_j$, the conditional distribution given that the hole is not $p_j$ is uniform on $H_{j+1}$.
Thus, for any $j$, the probability that a fault occurs on the first request of $p_j$ is the probability that $p_j$ is the hole, which is $1/|H_j|$.

The size of the set $H_j$ is $(k - (j-1)) + 1 = k - j + 2$.
For $j=1$, the fault is certain (cost 1).
For $j > 1$, the probability of a fault is $\frac{1}{k-j+2}$.

The expected number of faults in phase $i$, denoted $\mathbb{E}[C_M^{(i)}]$, is the sum of the probabilities of faulting on each distinct page:
\[
\mathbb{E}[C_M^{(i)}] = 1 + \sum_{j=2}^{k} \frac{1}{k-j+2}.
\]
Let $\ell = k-j+2$. As $j$ ranges from $2$ to $k$, $\ell$ ranges from $k$ down to $2$. Thus:
\[
\mathbb{E}[C_M^{(i)}] = 1 + \left( \frac{1}{k} + \frac{1}{k-1} + \dots + \frac{1}{2} \right) = 1 + (\mathcal{H}_k - 1) = \mathcal{H}_k.
\]
Since the expected cost of the Marker algorithm per phase is at most $\mathcal{H}_k$ and the cost of the optimal algorithm is at least 1 per phase, the competitive ratio is:
\[
\frac{\mathbb{E}[C_M(\sigma)]}{C_{OPT}(\sigma)} \leq \frac{m \cdot \mathcal{H}_k}{m} = \mathcal{H}_k.
\]
\end{proof}

\subsubsection{Claude}

\begin{proof}
We prove that the Marker algorithm achieves an $H_k$-competitive ratio when there are $k+1$ distinct items in memory, where $H_k = \sum_{i=1}^{k} \frac{1}{i}$ denotes the $k$-th harmonic number.

Let the $k+1$ distinct items be denoted by $\{i_1, i_2, \ldots, i_{k+1}\}$. Consider an arbitrary request sequence $\sigma$. Let $\text{ALG}(\sigma)$ denote the cost incurred by the Marker algorithm on sequence $\sigma$, and let $\text{OPT}(\sigma)$ denote the cost incurred by an optimal offline algorithm.

We analyze the cost of the Marker algorithm by partitioning the request sequence into phases. Each phase begins when all items in memory are unmarked and ends just before the next time all items in memory become unmarked. At the start of each phase, the Marker algorithm unmarks all items.

Within a phase, when a requested item is not in memory, the Marker algorithm evicts an unmarked item (if one exists) and marks the newly brought item. Consider a phase $P$ in which the optimal algorithm incurs cost $\text{OPT}(P)$.

Since there are $k+1$ distinct items and the memory holds $k$ items, at least one item must be outside memory at any time. In phase $P$, let the items requested be a subset of $\{i_1, i_2, \ldots, i_{k+1}\}$. The phase ends when all $k$ items in memory have been marked, meaning at least $k$ distinct items were requested during the phase.

For each fault incurred by the optimal algorithm in phase $P$, at least one new distinct item must be requested (an item not currently in the optimal algorithm's memory). Since the optimal algorithm has memory of size $k$, if $\text{OPT}(P) = m$ faults occur in phase $P$, then at least $m$ items not initially in the optimal algorithm's memory at the start of the phase were requested.

Now we bound $\text{ALG}(P)$. The Marker algorithm incurs a fault only when a requested item is not in memory. During phase $P$, suppose the Marker algorithm incurs $\text{ALG}(P)$ faults. We claim that $\text{ALG}(P) \leq H_k \cdot \text{OPT}(P)$ for each phase.

Consider the moment when the Marker algorithm has marked exactly $j$ items in memory, where $1 \leq j \leq k$. At this point, there are $k - j$ unmarked items in memory. When the next fault occurs (requesting an item not in memory), the Marker algorithm evicts one of the $k - j$ unmarked items. The probability that this evicted item will be requested again before the phase ends (causing an additional fault attributable to this eviction) can be bounded.

More precisely, we use an amortized analysis. Assign a weight of $\frac{1}{k-j+1}$ to each fault that occurs when there are exactly $k-j+1$ unmarked items in memory. When the Marker algorithm incurs a fault with $\ell$ unmarked items remaining, the cost contribution is $\frac{1}{\ell}$.

Summing over all possible states during a phase, the total cost incurred by the Marker algorithm in a phase is at most
\[
\text{ALG}(P) \leq \sum_{j=1}^{k} \frac{1}{j} \cdot \text{OPT}(P) = H_k \cdot \text{OPT}(P).
\]

This bound holds because each fault by the optimal algorithm forces a new distinct item into the request sequence, and the Marker algorithm's cost for handling these requests accumulates according to the harmonic series when accounting for the decreasing number of unmarked items available for eviction.

Summing over all phases in the request sequence $\sigma$, we obtain
\[
\text{ALG}(\sigma) = \sum_{\text{phases } P} \text{ALG}(P) \leq \sum_{\text{phases } P} H_k \cdot \text{OPT}(P) = H_k \cdot \text{OPT}(\sigma).
\]

Therefore, the Marker algorithm is $H_k$-competitive when the number of distinct items is $k+1$.
\end{proof}

\subsection{Problem 13.8}

\subsubsection{Problem Statement}

\begin{theorem}
Consider a server problem in which the online algorithm has $K$ servers, and the offline algorithm has $k$ servers. For $K \geq k$, the competitive ratio of any online algorithm against adaptive online adversaries is at least $\frac{K}{K - k + 1}$.
\end{theorem}

\subsubsection{GPT}

\begin{proof}
Let us consider a server problem where the online algorithm is assigned $K$ servers, and the offline algorithm is assigned $k$ servers, with the condition that $K \geq k$. We aim to show that the competitiveness coefficient of any online algorithm against adaptive online adversaries is at least $\frac{K}{K - k + 1}$.

Let the input sequence be denoted by $\sigma$, which represents the sequence of requests that both the online and offline algorithms must handle. The online algorithm is assumed to follow a strategy where decisions are made on the fly without knowledge of future requests, while the offline algorithm has complete knowledge of the entire input sequence in advance and can make optimal decisions.

Let the cost of the online algorithm for processing the sequence $\sigma$ be denoted by $C_{\text{online}}(\sigma)$, and the cost of the offline algorithm be denoted by $C_{\text{offline}}(\sigma)$. We are interested in the ratio between the cost of the online algorithm and the cost of the offline algorithm, which is referred to as the competitiveness coefficient.

We define the competitiveness coefficient of the online algorithm as:

\[
\text{Competitiveness} = \sup_{\sigma} \frac{C_{\text{online}}(\sigma)}{C_{\text{offline}}(\sigma)}
\]

To establish a lower bound for this competitiveness coefficient, we consider the worst-case performance of the online algorithm when faced with an adaptive online adversary. The adversary is allowed to adapt its sequence of requests based on the current state of the online algorithm, which makes the problem particularly challenging.

Let us consider the case where the offline algorithm is able to allocate its $k$ servers optimally across the input sequence. In contrast, the online algorithm must make decisions based on the current state, which may lead to suboptimal server allocations.

To analyze the competitiveness, we introduce the concept of an adversarial strategy that forces the online algorithm to allocate servers in a way that maximizes the cost ratio. Specifically, the adversary can choose the sequence of requests in such a way that, at every step, the online algorithm is forced to use a server even though the offline algorithm could have avoided using that server by scheduling it more efficiently.

The key observation here is that the online algorithm must use at least $K$ servers to handle the sequence, while the offline algorithm can use at most $k$ servers. The total number of servers used by the online algorithm, given the adversarial requests, is bounded by the number of servers allocated to the online algorithm, $K$. On the other hand, the offline algorithm uses at most $k$ servers, which is optimal for the given sequence.

Thus, the competitiveness coefficient must be at least the ratio of the number of servers used by the online algorithm to the number of servers used by the offline algorithm. In the worst case, the online algorithm may be forced to use all $K$ servers, while the offline algorithm uses only $k$ servers. Hence, we obtain the following inequality:

\[
\text{Competitiveness} \geq \frac{K}{K - k + 1}
\]

This inequality follows from the fact that the worst-case cost for the online algorithm is determined by the number of servers it must allocate in response to the adversary's sequence, while the offline algorithm can allocate servers more efficiently.

Therefore, the competitiveness coefficient of any online algorithm against adaptive online adversaries is at least $\frac{K}{K - k + 1}$, as required.

\end{proof}

\subsubsection{Claude}

\begin{proof}
We establish the lower bound by constructing an adversarial strategy that forces any deterministic online algorithm to achieve a competitive ratio of at least $\frac{K}{K-k+1}$.

Consider a metric space consisting of $K+1$ points $\{p_0, p_1, \ldots, p_K\}$ arranged such that the distance between any two distinct points is 1. Initially, all $K$ servers of the online algorithm are placed at points $p_1, p_2, \ldots, p_K$, and all $k$ servers of the offline algorithm are placed at point $p_0$.

The adversary generates a sequence of requests as follows. At each time step $t$, the adversary issues a request at point $p_0$. Since the online algorithm must serve this request, it must either have a server already at $p_0$ or move a server to $p_0$ at cost 1.

Let us analyze the cost incurred by both algorithms over a sufficiently long request sequence. Suppose the adversary issues $n$ consecutive requests at point $p_0$ for large $n$.

For the offline algorithm, since it begins with all $k$ servers at $p_0$, it incurs zero cost to serve all $n$ requests. Thus, the offline cost is $\text{OPT} = 0$.

For the online algorithm, initially no server is at $p_0$. To serve the first request, the online algorithm must move one server from some point $p_i$ (where $1 \leq i \leq K$) to $p_0$, incurring cost 1. After serving this request, the online algorithm has one server at $p_0$ and $K-1$ servers distributed among points $p_1, \ldots, p_K$.

Now consider what happens when the adversary continues to request at $p_0$. If the online algorithm keeps a server at $p_0$, it can serve subsequent requests at $p_0$ with zero additional cost. However, the adversary can then switch to requesting at one of the points $p_1, \ldots, p_K$ where the online algorithm has no server.

More precisely, the adversary employs the following strategy: After the online algorithm moves a server to $p_0$, the adversary identifies a point among $p_1, \ldots, p_K$ that is unoccupied by the online algorithm. Since the online algorithm has only $K$ servers and must keep at least one at $p_0$ to serve efficiently, at most $K-1$ of the points $p_1, \ldots, p_K$ can be occupied. Therefore, at least one point remains unoccupied.

The adversary can force the online algorithm into a worst-case scenario by constructing a request sequence that repeatedly targets unoccupied points. Specifically, partition the request sequence into phases. In each phase, the adversary requests points until the online algorithm has at most $K-k$ servers not serving the current request point.

To formalize this, consider the configuration where the online algorithm must maintain servers to respond to adaptive requests. At any time, if the online algorithm has $m$ servers at the current request point, then $K-m$ servers are elsewhere. The adversary can issue the next request at any of the $K$ other points.

By the pigeonhole principle, if the online algorithm distributes its $K$ servers among $K+1$ points, at least one point has no server. The adversary requests this point. Over a sequence of $n(K-k+1)$ requests, the adversary can construct a pattern where the offline algorithm (with $k$ servers optimally placed) incurs cost $O(k)$, while the online algorithm must move servers $n$ times, each move costing 1.

More rigorously, consider phases of length $K-k+1$. In each phase, the adversary ensures that the online algorithm must make at least one server movement, while the offline algorithm with $k$ servers can position them to cover $k$ points and incur amortized cost $\frac{k}{K-k+1}$ per phase. Over $n$ phases, the online cost is at least $n$ while the offline cost is at most $\frac{nk}{K-k+1}$.

However, we must refine this argument. The key insight is that the offline algorithm with $k$ servers can distribute them optimally to minimize cost, while the online algorithm with $K$ servers faces adaptive adversarial requests.

Consider instead the following: The adversary forces a situation where in every $K-k+1$ requests, the online algorithm must move at least $K$ servers cumulatively (counting multiplicities), while the offline algorithm moves at most $k$ servers. This yields a competitive ratio of $\frac{K \cdot (K-k+1)}{k \cdot (K-k+1)} = \frac{K}{k}$ when normalized.

To obtain the precise bound $\frac{K}{K-k+1}$, we use Yao's minimax principle. The competitive ratio against adaptive adversaries is at least the expected competitive ratio of any deterministic algorithm against a randomized oblivious adversary.

By constructing an appropriate probability distribution over request sequences and analyzing the best deterministic online algorithm, the lower bound of $\frac{K}{K-k+1}$ follows from a careful counting argument showing that the online algorithm must incur cost $K$ to achieve what the offline algorithm achieves with cost $K-k+1$ on average.

Therefore, any online algorithm with $K$ servers against an offline algorithm with $k$ servers has competitive ratio at least $\frac{K}{K-k+1}$.
\end{proof}

\subsection{Problem 13.9}

\subsubsection{Problem Statement}

\begin{theorem}
Consider the 2-server problem in an arbitrary metric space. Label the servers 0 and 1. Let $d_0$ be the distance from server 0 to the request, $d_1$ the distance from server 1 to the request, and $d$ the distance between the servers. For $i \in \{0, 1\}$, let
\[
p_i = \frac{d + d_{1-i} - d_i}{2d}.
\]
The algorithm that serves request using server $i$ with probability $p_i$ is 2-competitive against adaptive online adversaries.
\end{theorem}

\subsubsection{GPT}

\begin{proof}
Fix an arbitrary metric space $(\mathcal{M},\operatorname{dist})$. At any time, the online algorithm has servers at $s_0,s_1\in\mathcal{M}$, and the offline optimal adversary (OPT) has servers at $o_0,o_1\in\mathcal{M}$. Let $d=\operatorname{dist}(s_0,s_1)$, and given a request point $r\in\mathcal{M}$, let $d_i=\operatorname{dist}(s_i,r)$ for $i\in\{0,1\}$. Assume OPT serves the request by moving server $j\in\{0,1\}$ over distance $x=\operatorname{dist}(o_j,r)$ to the point $r$, so after OPT’s move we have $o_j'=r$ and $o_{1-j}'=o_{1-j}$.

Define the potential
\[
\Phi \;=\; d \;+\; \mu,
\]
where $\mu$ is the minimum perfect matching cost between the two multisets $\{s_0,s_1\}$ and $\{o_0,o_1\}$, i.e.,
\[
\mu \;=\; \min\bigl\{\,\operatorname{dist}(s_0,o_0)+\operatorname{dist}(s_1,o_1),\;\operatorname{dist}(s_0,o_1)+\operatorname{dist}(s_1,o_0)\,\bigr\}.
\]
We compare one online step (serving a single request) against OPT’s corresponding step via amortized analysis. Let $\Delta$ denote the change from just before the request to just after both players have served it. We will prove that
\[
\mathbb{E}\bigl[\operatorname{ALG\;cost}\bigr] \;+\; \mathbb{E}[\Delta\Phi] \;\le\; 2\,\operatorname{OPT\;cost}.
\]
Summing over all requests and using that $\Phi\ge 0$ and that the initial configurations (and hence the initial potential) can be taken equal, we obtain $\mathbb{E}[\operatorname{ALG}]\le 2\,\operatorname{OPT}$, establishing $2$-competitiveness against an adaptive online adversary.

We decompose $\Delta\Phi$ into the change due to OPT’s move and the change due to ALG’s move. First, when OPT moves a single server $o_j$ by distance $x$ to $r$, the term $d$ does not change, and the matching cost $\mu$ can increase by at most $x$ because changing one point of one side of a bipartite matching by distance $x$ alters the minimum matching cost by at most $x$ (this is a direct corollary of the triangle inequality: each candidate matching length is $1$-Lipschitz in each endpoint). Hence
\[
\Delta\Phi\big|_{\text{OPT move}} \;\le\; x \;=\; \operatorname{OPT\;cost}.
\]
It therefore suffices to prove that, \emph{conditioned on the post-OPT configuration} (so $o_j'=r$, $o_{1-j}'=o_{1-j}$), the expected amortized increase caused by the online step satisfies
\[
\mathbb{E}\Bigl[\operatorname{ALG\;cost} \;+\; \Delta d \;+\; \Delta\mu \,\Big|\, o_j'=r\Bigr] \;\le\; \operatorname{OPT\;cost}\;=\;x.
\]
Combining the two displays yields the desired inequality.

We analyze the online step. The algorithm moves server $i\in\{0,1\}$ with probability
\[
p_i \;=\; \frac{d+d_{1-i}-d_i}{2d}\,,
\]
which is well-defined because $d>0$ unless the servers coincide; if $d=0$ then $d_0=d_1$ and either choice is equivalent, and the bounds below hold trivially. By the triangle inequality we have $d\le d_0+d_1$, which implies $p_0,p_1\ge 0$ and $p_0+p_1=1$.

Fix the realized choice $i$. The online cost incurred is exactly $d_i$. After moving $s_i$ to $r$, the new server locations are $s_i'=r$ and $s_{1-i}'=s_{1-i}$. The new inter-server distance equals $d'=\operatorname{dist}(s_{1-i},r)=d_{1-i}$, whence
\[
\Delta d \;=\; d'-d \;=\; d_{1-i}-d.
\]
It remains to bound $\Delta\mu$. Since we condition on the post-OPT configuration $\{o_j',o_{1-j}'\}=\{r,o_{1-j}\}$, consider the matching that pairs $s_i'$ with $o_j'=r$ and $s_{1-i}'$ with $o_{1-j}'=o_{1-j}$. Its cost is
\[
\operatorname{dist}(s_i',r)+\operatorname{dist}(s_{1-i}',o_{1-j}) \;=\; 0+\operatorname{dist}(s_{1-i},o_{1-j}).
\]
Therefore the new minimum matching cost satisfies
\[
\mu' \;\le\; \operatorname{dist}(s_{1-i},o_{1-j}).
\]
On the other hand, before the online move (but after OPT has moved to $r$), the minimum matching cost between $\{s_0,s_1\}$ and $\{r,o_{1-j}\}$ is at most the cost of the matching that pairs $s_i$ with $r$ and $s_{1-i}$ with $o_{1-j}$, namely
\[
\widehat{\mu}\;\le\; d_i+\operatorname{dist}(s_{1-i},o_{1-j}).
\]
Since $\mu$ is the minimum matching cost with respect to $\{o_0,o_1\}$ before OPT’s move, and changing one OPT point by distance $x$ can decrease that minimum by at most $x$ and increase it by at most $x$, we have
\[
\mu \;\ge\; \widehat{\mu} - x.
\]
Combining the three displays yields
\[
\Delta\mu \;=\; \mu'-\mu \;\le\; \operatorname{dist}(s_{1-i},o_{1-j}) - \bigl(\widehat{\mu}-x\bigr) \;\le\; \operatorname{dist}(s_{1-i},o_{1-j}) - \bigl(d_i+\operatorname{dist}(s_{1-i},o_{1-j})\bigr) + x \;=\; x-d_i.
\]
Hence, for the realized choice $i$,
\[
\operatorname{ALG\;cost} \;+\; \Delta d \;+\; \Delta\mu \;\le\; d_i \;+\; (d_{1-i}-d) \;+\; (x-d_i) \;=\; x+d_{1-i}-d.
\]
Taking expectation with respect to the algorithm’s random choice and using the definition of $p_i$,
\[
\mathbb{E}\bigl[\operatorname{ALG\;cost}+\Delta d+\Delta\mu \,\big|\, o_j'=r\bigr]
\;\le\; x \;+\; \sum_{i\in\{0,1\}} p_i\,(d_{1-i}-d)
\;=\; x \;+\; \bigl(p_0\,d_1 + p_1\,d_0\bigr) - d.
\]
A direct computation using $p_0=(d+d_1-d_0)/(2d)$ and $p_1=(d+d_0-d_1)/(2d)$ gives
\[
p_0\,d_1 + p_1\,d_0 \;=\; \frac{d_1(d+d_1-d_0)+d_0(d+d_0-d_1)}{2d}
\;=\; \frac{d(d_0+d_1) + (d_0-d_1)^2}{2d}.
\]
By the triangle inequality, $d\le d_0+d_1$, hence
\[
p_0\,d_1 + p_1\,d_0 \;\le\; \frac{(d_0+d_1)^2}{2d} \;=\; \frac{d_0^2+2d_0d_1+d_1^2}{2d}.
\]
Another application of the triangle inequality yields $|d_0-d_1|\le d \le d_0+d_1$, which implies $(d_0-d_1)^2\le d^2$ and $2d_0d_1\le d^2$. Therefore $d_0^2+2d_0d_1+d_1^2 \le 2(d_0^2+d_1^2) \le 2(d_0+d_1)^2 \le 4d^2$, and consequently
\[
p_0\,d_1 + p_1\,d_0 \;\le\; d.
\]
Substituting into the previous inequality yields
\[
\mathbb{E}\bigl[\operatorname{ALG\;cost}+\Delta d+\Delta\mu \,\big|\, o_j'=r\bigr] \;\le\; x + d - d \;=\; x \;=\; \operatorname{OPT\;cost}.
\]
Unconditioning on $j$ and recalling $\Delta\Phi|_{\text{OPT move}}\le x$ completes the amortized step bound
\[
\mathbb{E}\bigl[\operatorname{ALG\;cost}\bigr] \;+\; \mathbb{E}[\Delta\Phi] \;\le\; 2x \;=\; 2\,\operatorname{OPT\;cost}.
\]
Summing over all requests and using that $\Phi\ge 0$ and that the initial configurations coincide shows $\mathbb{E}[\operatorname{ALG}]\le 2\,\operatorname{OPT}$. Hence the algorithm is $2$-competitive against adaptive online adversaries.
\end{proof}

\subsubsection{Gemini}

\begin{proof}
We employ the potential function method to prove the 2-competitiveness of the algorithm. Let $S = \{s_0, s_1\}$ denote the configuration (locations) of the algorithm's servers, and let $A = \{a_0, a_1\}$ denote the configuration of the adversary's servers. Let $d(x, y)$ denote the distance between points $x$ and $y$ in the metric space.

We define the potential function $\Phi(S, A)$ as:
\[
\Phi(S, A) = 2 M(S, A) + d(s_0, s_1),
\]
where $M(S, A)$ is the weight of the minimum weight perfect matching between the set $S$ and the set $A$. Specifically,
\[
M(S, A) = \min \big( d(s_0, a_0) + d(s_1, a_1), \; d(s_0, a_1) + d(s_1, a_0) \big).
\]
The potential $\Phi$ is non-negative. We analyze the amortized cost of the algorithm for a single request $r$. The processing of a request consists of two phases: the adversary moves a server to $r$, and then the algorithm moves a server to $r$.

\textbf{Phase 1: Adversary's Move}
Let the adversary move a server to satisfy the request $r$. Assume the adversary moves server $a_j$ to $r$ incurring a cost $C_{adv} = d(a_j, r)$. The new adversary configuration is $A'$. By the triangle inequality, the change in the minimum matching weight satisfies $\Delta M \le d(a_j, r)$. The distance between the algorithm's servers $d(s_0, s_1)$ remains unchanged. Thus, the change in potential is:
\[
\Delta \Phi_{adv} = 2 \Delta M \le 2 d(a_j, r) = 2 C_{adv}.
\]

\textbf{Phase 2: Algorithm's Move}
Let the current configuration of the algorithm be $S = \{s_0, s_1\}$ with $d = d(s_0, s_1)$. The adversary is at configuration $A'$ where one server is at $r$. Without loss of generality, assume $a_0' = r$ and the other adversary server is at $a_1'$.
Let $d_0 = d(s_0, r)$ and $d_1 = d(s_1, r)$. The algorithm moves server $i$ to $r$ with probability $p_i$, where:
\[
p_0 = \frac{d + d_1 - d_0}{2d}, \quad p_1 = \frac{d + d_0 - d_1}{2d}.
\]
Note that by the triangle inequality, $|d_0 - d_1| \le d$, ensuring $p_i \in [0, 1]$ and $p_0 + p_1 = 1$.

We calculate the expected value of the algorithm's cost plus the change in potential, $E[C_{alg} + \Delta \Phi_{alg}]$.
The minimum matching $M(S, A')$ between $S=\{s_0, s_1\}$ and $A'=\{r, a_1'\}$ takes one of two forms.

\textit{Case 1:} The optimal matching matches $s_0$ to $r$ and $s_1$ to $a_1'$.
Then $M(S, A') = d_0 + d(s_1, a_1')$.
We analyze the two possible moves by the algorithm:
\begin{enumerate}
    \item Server 0 moves to $r$ (probability $p_0$):
    \begin{itemize}
        \item Cost: $C_0 = d_0$.
        \item New state: $S' = \{r, s_1\}$. Distance between servers becomes $d(r, s_1) = d_1$.
        \item Change in server distance: $\Delta D = d_1 - d$.
        \item New matching $M'$: Matches $\{r, s_1\}$ to $\{r, a_1'\}$. The cost is $d(r,r) + d(s_1, a_1') = d(s_1, a_1')$.
        \item Change in matching: $\Delta M_0 = d(s_1, a_1') - (d_0 + d(s_1, a_1')) = -d_0$.
        \item Total change: $C_0 + \Delta \Phi_0 = d_0 + 2(-d_0) + (d_1 - d) = d_1 - d_0 - d$.
    \end{itemize}
    \item Server 1 moves to $r$ (probability $p_1$):
    \begin{itemize}
        \item Cost: $C_1 = d_1$.
        \item New state: $S' = \{s_0, r\}$. Distance between servers becomes $d(s_0, r) = d_0$.
        \item Change in server distance: $\Delta D = d_0 - d$.
        \item New matching $M'$: Matches $\{s_0, r\}$ to $\{r, a_1'\}$. One candidate matching is pairing $s_0$ with $a_1'$ and $r$ with $r$, with cost $d(s_0, a_1')$. Thus $M' \le d(s_0, a_1')$.
        \item By triangle inequality, $d(s_0, a_1') \le d(s_0, s_1) + d(s_1, a_1') = d + d(s_1, a_1')$.
        \item Change in matching: $\Delta M_1 \le (d + d(s_1, a_1')) - (d_0 + d(s_1, a_1')) = d - d_0$.
        \item Total change: $C_1 + \Delta \Phi_1 \le d_1 + 2(d - d_0) + (d_0 - d) = d_1 - d_0 + d$.
    \end{itemize}
\end{enumerate}
The expected amortized cost for the algorithm is:
\[
\begin{aligned}
E[C_{alg} + \Delta \Phi_{alg}] &\le p_0 (d_1 - d_0 - d) + p_1 (d_1 - d_0 + d) \\
&= (d_1 - d_0)(p_0 + p_1) + d(p_1 - p_0) \\
&= (d_1 - d_0) + d \left( \frac{d + d_0 - d_1}{2d} - \frac{d + d_1 - d_0}{2d} \right) \\
&= (d_1 - d_0) + d \left( \frac{2d_0 - 2d_1}{2d} \right) \\
&= (d_1 - d_0) + (d_0 - d_1) = 0.
\end{aligned}
\]

\textit{Case 2:} The optimal matching matches $s_1$ to $r$ and $s_0$ to $a_1'$.
Then $M(S, A') = d_1 + d(s_0, a_1')$. This case is symmetric to Case 1.
\begin{enumerate}
    \item Server 1 moves to $r$ (probability $p_1$):
    \begin{itemize}
        \item Cost $d_1$. $\Delta M = -d_1$. $\Delta D = d_0 - d$.
        \item Total change: $d_1 - 2d_1 + d_0 - d = d_0 - d_1 - d$.
    \end{itemize}
    \item Server 0 moves to $r$ (probability $p_0$):
    \begin{itemize}
        \item Cost $d_0$. $\Delta M \le d - d_1$ (analogous to Case 1). $\Delta D = d_1 - d$.
        \item Total change: $d_0 + 2(d - d_1) + d_1 - d = d_0 - d_1 + d$.
    \end{itemize}
\end{enumerate}
The expected amortized cost is:
\[
\begin{aligned}
E[C_{alg} + \Delta \Phi_{alg}] &\le p_1 (d_0 - d_1 - d) + p_0 (d_0 - d_1 + d) \\
&= (d_0 - d_1)(p_1 + p_0) + d(p_0 - p_1) \\
&= (d_0 - d_1) + d \left( \frac{d + d_1 - d_0}{2d} - \frac{d + d_0 - d_1}{2d} \right) \\
&= (d_0 - d_1) + d \left( \frac{2d_1 - 2d_0}{2d} \right) \\
&= (d_0 - d_1) + (d_1 - d_0) = 0.
\end{aligned}
\]

\textbf{Conclusion}
Combining both phases, the total expected amortized cost for serving request $r$ is:
\[
E[\text{ALG cost} + \Delta \Phi] = E[C_{alg} + \Delta \Phi_{alg}] + \Delta \Phi_{adv} \le 0 + 2 C_{adv} = 2 C_{adv}.
\]
Summing over all requests in a sequence $\sigma$, and noting that $\Phi \ge 0$ and the initial potential $\Phi_0$ is bounded (or 0 if start states match), we have:
\[
E[\text{ALG}(\sigma)] \le 2 \cdot \text{OPT}(\sigma) + \Phi_{start}.
\]
Thus, the algorithm is 2-competitive.
\end{proof}

\subsubsection{Claude}

\begin{proof}
We prove that the randomized algorithm is 2-competitive against adaptive online adversaries by establishing that for any sequence of requests, the expected cost of the algorithm is at most twice the cost of the optimal offline algorithm plus a constant that depends only on the initial configuration.

First, we verify that $p_0$ and $p_1$ define a valid probability distribution. Note that
\[
p_0 + p_1 = \frac{d + d_1 - d_0}{2d} + \frac{d + d_0 - d_1}{2d} = \frac{2d}{2d} = 1.
\]
To show $p_i \geq 0$ for $i \in \{0,1\}$, observe that by the triangle inequality, we have $d \leq d_0 + d_1$. Therefore,
\[
d + d_{1-i} - d_i \geq d_0 + d_1 - d_i = d_{1-i} \geq 0,
\]
which implies $p_i \geq 0$.

Let $\text{ALG}$ denote the expected cost of our randomized algorithm and $\text{OPT}$ denote the cost of the optimal offline algorithm. We define a potential function $\Phi$ to measure the competitive ratio. Let $x_0$ and $x_1$ denote the positions of servers 0 and 1 in our algorithm, and let $y_0$ and $y_1$ denote the positions of the optimal offline algorithm's servers. Define
\[
\Phi = d(x_0, y_0) + d(x_1, y_1),
\]
where we use the optimal matching between our servers and the adversary's servers (the matching that minimizes the sum of distances).

Consider a request at location $r$. Let $d_0 = d(x_0, r)$ and $d_1 = d(x_1, r)$ be the distances from our servers to the request, and let $d = d(x_0, x_1)$ be the distance between our servers. The expected cost incurred by our algorithm is
\[
\mathbb{E}[\text{ALG}(r)] = p_0 \cdot d_0 + p_1 \cdot d_1 = \frac{(d + d_1 - d_0) d_0 + (d + d_0 - d_1) d_1}{2d}.
\]
Simplifying the numerator:
\[
(d + d_1 - d_0) d_0 + (d + d_0 - d_1) d_1 = d(d_0 + d_1) + d_1 d_0 - d_0^2 + d_0 d_1 - d_1^2 = d(d_0 + d_1) - (d_0 - d_1)^2.
\]
Thus,
\[
\mathbb{E}[\text{ALG}(r)] = \frac{d(d_0 + d_1) - (d_0 - d_1)^2}{2d} = \frac{d_0 + d_1}{2} - \frac{(d_0 - d_1)^2}{2d}.
\]

Let $\text{OPT}(r)$ denote the cost incurred by the optimal offline algorithm to serve request $r$. This is $\min\{d(y_0, r), d(y_1, r)\}$. Without loss of generality, assume the optimal algorithm serves the request with the server initially at $y_0$, so $\text{OPT}(r) = d(y_0, r)$.

Now we analyze the change in the potential function. After serving the request, our algorithm has server $i$ at position $r$ with probability $p_i$. The optimal algorithm has a server at $r$ (the one that moved). The expected change in potential is
\[
\mathbb{E}[\Delta\Phi] = p_0[d(r, y_0) + d(x_1, y_1)] + p_1[d(x_0, y_0) + d(r, y_1)] - [d(x_0, y_0) + d(x_1, y_1)].
\]
Since the optimal algorithm moved its server from $y_0$ to $r$, we have
\[
\mathbb{E}[\Delta\Phi] = p_0 d(r, y_0) + p_1 d(r, y_1) - p_0 d(x_0, y_0) - p_1 d(x_1, y_1).
\]

By the triangle inequality, $d(r, y_0) \leq d(r, x_0) + d(x_0, y_0) = d_0 + d(x_0, y_0)$ and $d(r, y_1) \leq d_1 + d(x_1, y_1)$. Therefore,
\[
\mathbb{E}[\Delta\Phi] \leq p_0 d_0 + p_1 d_1 = \mathbb{E}[\text{ALG}(r)].
\]

On the other hand, by the triangle inequality, $d(x_0, y_0) \leq d(x_0, r) + d(r, y_0) = d_0 + d(y_0, r)$ and $d(x_1, y_1) \leq d_1 + d(y_1, r)$. This gives
\[
-p_0 d(x_0, y_0) - p_1 d(x_1, y_1) \geq -p_0(d_0 + d(y_0, r)) - p_1(d_1 + d(y_1, r)).
\]
Since $p_0 + p_1 = 1$ and using $\text{OPT}(r) = d(y_0, r) \leq d(y_1, r)$, we have
\[
p_0 d(r, y_0) - p_0 d(x_0, y_0) - p_1 d(x_1, y_1) \geq p_0 d(r, y_0) - p_0 d_0 - p_1 d_1 - \text{OPT}(r).
\]

By careful analysis using the specific values of $p_0$ and $p_1$ and the triangle inequality $d \geq |d_0 - d_1|$, we can show that
\[
\mathbb{E}[\text{ALG}(r)] + \mathbb{E}[\Delta\Phi] \leq 2 \cdot \text{OPT}(r).
\]

Summing over all requests and using the fact that the potential $\Phi$ is always non-negative, we obtain
\[
\mathbb{E}[\text{ALG}] \leq 2 \cdot \text{OPT} + \Phi_0,
\]
where $\Phi_0$ is the initial potential. This establishes that the algorithm is 2-competitive.
\end{proof}

\subsection{Problem 13.10}

\subsubsection{Problem Statement}

\begin{theorem}[Karp-Raghavan]
The competitiveness coefficient of any randomized online algorithm for maintaining a linear list against an oblivious adversary is at least $\frac{9}{8}$.
\end{theorem}

\subsubsection{GPT}

\begin{proof}
Let $A$ be a randomized online algorithm for maintaining a linear list against an oblivious adversary. Let the competitive ratio of algorithm $A$ be defined as the worst-case ratio of the cost of the algorithm's solution to the cost of an optimal offline solution. Specifically, the competitive ratio $\rho(A)$ is given by:

\[
\rho(A) = \sup_{\sigma \in \Sigma} \frac{\text{cost}(A(\sigma))}{\text{cost}(OPT(\sigma))}
\]

where $\Sigma$ denotes the set of all possible input sequences, and $OPT(\sigma)$ represents the cost of the optimal offline solution for input sequence $\sigma$.

To prove that the competitiveness coefficient of $A$ is at least $\frac{9}{8}$, we proceed by constructing an adversary and analyzing the performance of $A$ relative to the optimal offline solution.

We begin by considering a sequence of requests that the adversary will provide to the algorithm. The adversary's strategy is to choose a sequence $\sigma = (s_1, s_2, \dots, s_n)$ in an oblivious manner, meaning that the adversary does not adapt to the algorithm's decisions but selects the sequence ahead of time.

For the sake of the proof, assume that the algorithm $A$ is maintaining a list of $m$ items at each step. We then define the cost of $A(\sigma)$ and $OPT(\sigma)$ as follows:

\[
\text{cost}(A(\sigma)) = \sum_{i=1}^{n} \text{cost of operation at step } i
\]

\[
\text{cost}(OPT(\sigma)) = \text{cost of the optimal offline solution for sequence } \sigma
\]

Next, we exploit the relationship between the online and offline costs. By considering the adversary's oblivious strategy and analyzing the expected behavior of the algorithm $A$, we use known results from competitive analysis. Specifically, we apply a well-established lower bound on the competitive ratio of randomized online algorithms for the linear list problem. This lower bound, derived from the work of Karp and Raghavan, shows that no randomized algorithm can achieve a competitive ratio strictly less than $\frac{9}{8}$ against an oblivious adversary.

Therefore, we conclude that the competitiveness coefficient of any randomized online algorithm for maintaining a linear list against an oblivious adversary is at least $\frac{9}{8}$, as claimed.

\end{proof}

\subsubsection{Gemini}

\begin{proof}
We employ Yao's Minimax Principle, which states that the competitive ratio of any randomized online algorithm is lower-bounded by the competitive ratio of the best deterministic online algorithm against a worst-case input distribution.

Consider a list of two items, $L = \{x, y\}$. We define a probability distribution $\mathcal{D}$ over request sequences $\sigma = \sigma_1, \sigma_2, \dots, \sigma_m$. The first request $\sigma_1$ is chosen uniformly from $\{x, y\}$. For each subsequent step $t > 1$, the request $\sigma_t$ is chosen such that:
\[
P(\sigma_t = \sigma_{t-1}) = \frac{1}{2} \quad \text{and} \quad P(\sigma_t \neq \sigma_{t-1}) = \frac{1}{2}.
\]
This implies that the sequence of requests consists of runs of identical items, where the length of each run is geometrically distributed with parameter $p = 1/2$.

First, we analyze the expected cost of any deterministic online algorithm $A$ on $\sigma$ drawn from $\mathcal{D}$. For a list of two items, the algorithm's state is determined by the relative order of $x$ and $y$. Upon accessing the item at the second position (cost 2), the algorithm must decide whether to swap it to the front (cost 1 for swap) or leave it (cost 0 for swap).
Let the list be in state $(x, y)$.
\begin{enumerate}
    \item If the request is $x$ (probability $1/2$), the cost is 1. The algorithm performs no swap to maintain optimality.
    \item If the request is $y$ (probability $1/2$), the access cost is 2. The algorithm may swap or not.
\end{enumerate}
Let $C_A(\sigma)$ denote the cost of algorithm $A$. Since the next request is equally likely to be $x$ or $y$ regardless of history, the expected cost for the current request is independent of the algorithm's decision to swap.
If $A$ swaps upon accessing the second item (Move-to-Front strategy), the expected cost per request is:
\[
\mathbb{E}[\text{cost}] = \frac{1}{2}(1) + \frac{1}{2}(2 + 1 \cdot \mathbb{I}_{\text{swap}}) = \frac{1}{2}(1) + \frac{1}{2}(2) = 1.5.
\]
Note that the swap cost is amortized or considered part of the operation; however, in the standard model, a swap costs 1. If $A$ swaps, the cost on the current step is $2+1=3$, but the list becomes $(y,x)$. The next request is $y$ with prob $1/2$ (cost 1) or $x$ with prob $1/2$ (cost 2). The steady state cost remains $1.5$.
If $A$ never swaps, the list remains $(x, y)$. The expected cost is $0.5(1) + 0.5(2) = 1.5$.
Thus, for any deterministic algorithm $A$, the expected average cost per request is at least:
\[
\lim_{m \to \infty} \frac{1}{m} \mathbb{E}[C_A(\sigma)] = 1.5.
\]

Next, we analyze the expected cost of the optimal offline algorithm, OPT. OPT can observe the lengths of the runs of identical requests and decide whether to swap. The sequence $\sigma$ can be decomposed into alternating runs of $x$'s and $y$'s. Let a "run" be a maximal contiguous subsequence of identical requests. The length $L$ of a run follows a geometric distribution with $P(L=k) = (1/2)^k$ for $k \ge 1$. The expected length is $\mathbb{E}[L] = 2$.

We define the state of OPT at the beginning of a run relative to the requested item of that run.
\begin{itemize}
    \item \textbf{Aligned}: The requested item is at the front of the list.
    \item \textbf{Unaligned}: The requested item is at the second position.
\end{itemize}
Let $V_A$ and $V_U$ be the expected costs of processing a run starting in the Aligned and Unaligned states, respectively.

1. \textbf{Aligned State}: The item is at position 1. The cost for a run of length $k$ is simply $k \times 1 = k$. At the end of the run, the requested item changes, so OPT enters the Unaligned state for the next run.
\[
V_A = \mathbb{E}[L] = 2.
\]

2. \textbf{Unaligned State}: The item is at position 2. OPT chooses the strategy with minimum cost.
    \begin{itemize}
        \item Strategy 1: Do not swap. Cost is $2k$. The list order remains unchanged. Since the next run requests the other item (which is currently at position 1), OPT will be in the Aligned state for the next run.
        \item Strategy 2: Swap immediately. Cost is $2$ (access) $+ 1$ (swap) $+ (k-1)$ (subsequent accesses) $= k+2$. The list order changes. OPT will be in the Unaligned state for the next run.
    \end{itemize}
    OPT chooses to swap if $k+2 < 2k$, i.e., $k > 2$. Since $k$ is an integer, OPT swaps if $k \ge 3$. If $k=1$ or $k=2$, OPT does not swap.
    However, for the specific case of $n=2$, the standard OPT strategy is to swap if $k \ge 2$. Let us re-evaluate the cost strictly.
    Cost without swap: $2k$. End state: Aligned.
    Cost with swap: $k+2$. End state: Unaligned.
    We calculate $V_U$ by averaging over $k$.
    \begin{itemize}
        \item If $k=1$ (prob 1/2): Cost $\min(2, 3) = 2$. No swap. Next state: Aligned.
        \item If $k \ge 2$ (prob 1/2): The cost of swapping is $k+2$. The cost of not swapping is $2k$. For $k \ge 2$, $k+2 \le 2k$. Thus OPT swaps. Cost is $k+2$. Next state: Unaligned.
    \end{itemize}
    Let $\pi_A$ and $\pi_U$ be the stationary probabilities of a run starting in Aligned and Unaligned states.
    Transition probabilities:
    \begin{itemize}
        \item From Aligned: Always go to Unaligned.
        \item From Unaligned: Go to Aligned if $k=1$ (prob 1/2); go to Unaligned if $k \ge 2$ (prob 1/2).
    \end{itemize}
    The balance equations are:
    \[
    \pi_U = \pi_A + \frac{1}{2}\pi_U \implies \frac{1}{2}\pi_U = \pi_A.
    \]
    Since $\pi_A + \pi_U = 1$, we have $\frac{1}{2}\pi_U + \pi_U = 1 \implies \frac{3}{2}\pi_U = 1 \implies \pi_U = \frac{2}{3}$, and $\pi_A = \frac{1}{3}$.

    Now we calculate the expected cost per run, $\mathbb{E}[C_{\text{run}}]$.
    \[
    \mathbb{E}[C_{\text{run}}] = \pi_A V_A + \pi_U V_U.
    \]
    We know $V_A = 2$. For $V_U$:
    \[
    V_U = \frac{1}{2}(2) + \sum_{k=2}^{\infty} \left(\frac{1}{2}\right)^k (k+2).
    \]
    We compute the sum part. Let $S = \sum_{k=2}^{\infty} (1/2)^k (k+2)$.
    Recall $\sum_{k=1}^{\infty} k x^k = \frac{x}{(1-x)^2}$. For $x=1/2$, sum is 2.
    The term for $k=1$ is $(1/2)(3) = 1.5$. So $\sum_{k=2}^{\infty} k(1/2)^k = 2 - 1.5 = 0.5$.
    Also $\sum_{k=2}^{\infty} 2(1/2)^k = 2(1/4 + 1/8 + \dots) = 2(1/2) = 1$.
    Thus $S = 0.5 + 1 = 1.5$.
    So $V_U = 1 + 1.5 = 2.5$.
    Wait, let us re-verify the conditional expectation.
    $V_U = P(k=1) \cdot 2 + P(k \ge 2) \cdot \mathbb{E}[k+2 \mid k \ge 2]$.
    $\mathbb{E}[k \mid k \ge 2] = 1 + \mathbb{E}[L] = 3$ (by memoryless property of geometric distribution).
    So $\mathbb{E}[k+2 \mid k \ge 2] = 3 + 2 = 5$.
    Then $V_U = \frac{1}{2}(2) + \frac{1}{2}(5) = 1 + 2.5 = 3.5$.
    Let us re-evaluate the transition logic.
    If $k \ge 2$, OPT swaps. The list is inverted. The next run is on the *other* item. Since the list was inverted to match the current item, the other item is now at position 2. So the state becomes Unaligned.
    This logic holds.
    Let's re-calculate the sum directly to ensure $V_U = 3.5$ is correct.
    Sum was $\sum_{k=2}^{\infty} (1/2)^k (k+2)$.
    $k=2: (1/4)(4) = 1$.
    $k=3: (1/8)(5) = 0.625$.
    $k=4: (1/16)(6) = 0.375$.
    Sum $1 + 0.625 + 0.375 = 2$.
    Previous calculation: $\sum_{k=2} k(1/2)^k = 0.5$. $\sum_{k=2} 2(1/2)^k = 1$. Total 1.5.
    Why the discrepancy?
    $\mathbb{E}[k \mid k \ge 2] = \frac{\sum_{k=2} k (1/2)^k}{P(k \ge 2)} = \frac{0.5}{0.5} = 1$. This is incorrect.
    $\sum_{k=1} k(1/2)^k = 2$. Term $k=1$ is $0.5$. Sum $k \ge 2$ is $1.5$.
    So $\mathbb{E}[k \mid k \ge 2] = 1.5 / 0.5 = 3$. Correct.
    So term is $\sum_{k=2} (1/2)^k (k+2) = 1.5 + 1 = 2.5$.
    Wait, $V_U$ calculation:
    $V_U = 1 (\text{from } k=1) + 2.5 (\text{from } k \ge 2) = 3.5$.
    Let's check the cost for $k=1$: Cost 2.
    Cost for $k \ge 2$: Swap. Cost $k+2$.
    Total expected cost $V_U = \frac{1}{2}(2) + \sum_{k=2}^\infty (1/2)^k (k+2) = 1 + 2.5 = 3.5$.
    
    Let's re-evaluate the transition probabilities.
    If $k \ge 2$, we swap. Current list becomes matched to current run. Next run is different item. So next run starts Unaligned.
    If $k=1$, we do not swap. Current list remains Unaligned to current run (item $y$). Next run is item $x$. Since list is $(x,y)$, we are Aligned for $x$.
    Transitions:
    $U \xrightarrow{k=1} A$ (prob 1/2).
    $U \xrightarrow{k \ge 2} U$ (prob 1/2).
    $A \xrightarrow{always} U$.
    
    Stationary distribution:
    $\pi_A = \frac{1}{2}\pi_U$.
    $\pi_A + \pi_U = 1 \implies \frac{3}{2}\pi_U = 1 \implies \pi_U = 2/3, \pi_A = 1/3$.
    
    Average cost per run:
    $\mathbb{E}[C_{\text{run}}] = \frac{1}{3}(2) + \frac{2}{3}(3.5) = \frac{2}{3} + \frac{7}{3} = \frac{9}{3} = 3$.
    
    Average length of a run:
    $\mathbb{E}[L] = 2$.
    
    Average cost per request for OPT:
    \[
    \lim_{m \to \infty} \frac{1}{m} \mathbb{E}[C_{OPT}(\sigma)] = \frac{\mathbb{E}[C_{\text{run}}]}{\mathbb{E}[L]} = \frac{3}{2} = 1.5.
    \]
    This yields a ratio of $1.5/1.5 = 1$. This contradicts the theorem.
    
    Re-evaluating the OPT strategy for $k=2$.
    If $k=2$:
    No swap: Cost $2+2 = 4$. End state Aligned.
    Swap: Cost $2+1+1 = 4$. End state Unaligned.
    Since costs are equal, OPT can choose "No swap" to end in Aligned state, which is better (Cost of A is 2, Cost of U is 3.5).
    So OPT swaps only if $k \ge 3$.
    
    Revised Analysis with Swap Threshold $k \ge 3$:
    $P(k=1) = 1/2$, $P(k=2) = 1/4$, $P(k \ge 3) = 1/4$.
    
    $V_U$:
    \begin{itemize}
        \item $k=1$: Cost 2. Next: Aligned.
        \item $k=2$: Cost 4. No swap. Next: Aligned.
        \item $k \ge 3$: Swap. Cost $k+2$. Next: Unaligned.
    \end{itemize}
    
    Transitions from U:
    To A: Prob $1/2 + 1/4 = 3/4$.
    To U: Prob $1/4$.
    From A: Always to U.
    
    Stationary distribution:
    $\pi_A = \frac{3}{4}\pi_U$.
    $\pi_A + \pi_U = 1 \implies \frac{7}{4}\pi_U = 1 \implies \pi_U = 4/7, \pi_A = 3/7$.
    
    Calculate $V_U$:
    Contribution from $k=1$: $1/2 \times 2 = 1$.
    Contribution from $k=2$: $1/4 \times 4 = 1$.
    Contribution from $k \ge 3$:
    $\sum_{k=3}^{\infty} (1/2)^k (k+2)$.
    $\mathbb{E}[k \mid k \ge 3] = 2 + \mathbb{E}[L] = 4$.
    $\mathbb{E}[k+2 \mid k \ge 3] = 6$.
    Term is $P(k \ge 3) \times 6 = (1/4) \times 6 = 1.5$.
    Total $V_U = 1 + 1 + 1.5 = 3.5$.
    
    Average cost per run:
    $\mathbb{E}[C_{\text{run}}] = \frac{3}{7}(2) + \frac{4}{7}(3.5) = \frac{6}{7} + \frac{14}{7} = \frac{20}{7}$.
    
    Average cost per request:
    $\frac{20/7}{2} = \frac{10}{7} \approx 1.428$.
    
    Ratio:
    $\frac{1.5}{10/7} = \frac{10.5}{10} = 1.05$. Still not $9/8$.
    
    Correction: The $9/8$ bound is derived using a specific distribution where the probability of repetition is not $1/2$. Let $p$ be the probability of requesting the *other* item.
    Let $p$ be the parameter to be optimized by the adversary.
    Run length $L$ is geometric with parameter $p$. $\mathbb{E}[L] = 1/p$.
    
    Deterministic Algorithm Cost:
    MTF: Cost $1(1-p) + 2p = 1+p$.
    Static: Cost $1.5$ (assuming uniform stationary distribution of elements).
    Adversary sets $p$ such that $1+p = 1.5 \implies p=0.5$. This was our previous attempt.
    
    Let's check the OPT cost for general $p$.
    We found for $p=1/2$, OPT cost is $10/7$.
    Is there a better OPT strategy?
    Actually, for $n=2$, the cost of OPT on a random sequence with switching probability $p$ is known to be $\frac{1+2p}{1+p}$ provided we use the correct threshold.
    Let's verify the formula $\frac{1+2p}{1+p}$ for $p=1/2$.
    $\frac{1+1}{1.5} = 4/3 \approx 1.33$.
    Our calculated $10/7 \approx 1.42$ is higher, meaning our strategy (swap at $k \ge 3$) was worse than the theoretical optimum, or the formula assumes something else.
    If OPT swaps at $k \ge 2$ (threshold 2):
    $V_U = p(2) + (1-p)(2 + 1/p) = 2p + 2 - 2p + \frac{1}{p} - 1 = 1 + 1/p$.
    Wait, $\mathbb{E}[k \mid k \ge 2] = 1 + 1/p$.
    Cost given swap: $(1+1/p) + 1 = 2 + 1/p$.
    $V_U = p(2) + (1-p)(2+1/p) = 2p + 2 + 1/p - 2p - 1 = 1 + 1/p + 1 = 2 + 1/p$.
    Transitions: $U \to A$ w.p. $p$. $U \to U$ w.p. $1-p$.
    $\pi_A = p \pi_U$.
    $\pi_U(1+p) = 1 \implies \pi_U = \frac{1}{1+p}$.
    $\pi_A = \frac{p}{1+p}$.
    $C_{run} = \frac{p}{1+p}(1/p) + \frac{1}{1+p}(2+1/p) = \frac{1}{1+p} + \frac{2p+1}{p(1+p)} = \frac{p + 2p + 1}{p(1+p)} = \frac{1+3p}{p(1+p)}$.
    Avg cost per req: $\frac{C_{run}}{1/p} = \frac{1+3p}{1+p}$.
    For $p=1/2$: $\frac{2.5}{1.5} = 5/3 \approx 1.66$. This is worse.
    
    Let's go back to the standard result derivation.
    The cost of OPT is $\frac{4}{3}$ when $p=1/2$ if we assume the sequence is $1, 2, 1, 2 \dots$ with random breaks?
    No, the Karp-Raghavan bound is tight for the distribution where $p=1/2$.
    The cost of OPT is indeed $4/3$ for $p=1/2$.
    How to get $4/3$?
    Consider the sequence of costs.
    If OPT is Aligned: Cost 1.
    If OPT is Unaligned:
    With prob $1/2$, next is same (Cost 2).
    With prob $1/2$, next is diff (Swap? Cost 3? No).
    
    Let's use the potential function method to bound OPT.
    Or simply cite the cost.
    However, the prompt asks for a proof.
    
    Let's try the strategy: Swap if $k \ge 2$.
    For $p=1/2$:
    $P(k=1)=1/2$, $P(k \ge 2)=1/2$.
    $V_U$:
    $k=1$: Cost 2. Next A.
    $k \ge 2$: Swap. Cost $k+2$. Avg $3+2=5$. Next U.
    $V_U = 1 + 2.5 = 3.5$.
    $\pi_U = 2/3$.
    $C_{run} = 3$.
    $C_{avg} = 1.5$.
    
    Is it possible OPT cost is $4/3$?
    If OPT knows the whole sequence, for $p=1/2$, the sequence has many alternations.
    Actually, for $p=1/2$, the sequence is uniformly random.
    For a uniformly random sequence, OPT cost is $1.5$ if it keeps static.
    If it moves to front, cost is $1.5$.
    Wait, for uniform random, $E[\text{cost}] = 1.5$ for any online algorithm.
    OPT is offline.
    For uniform random sequence $1, 1, 2, 1, 2, 2 \dots$
    OPT matches the symbol.
    Cost is number of runs + number of swaps.
    Number of runs is $m/2$.
    Number of swaps? OPT swaps if run length $\ge 2$.
    Prob run length $\ge 2$ is $1/2$.
    Number of swaps $\approx (m/2) \times (1/2) = m/4$.
    Total cost $\approx m/2 \text{ (first access of run)} + m/4 \text{ (swaps)} + \sum (L_i - 1) \text{ (subsequent accesses)}$.
    $\sum (L_i - 1) = m - \text{num runs} = m - m/2 = m/2$.
    Total cost $\approx m/2 + m/4 + m/2 = 1.25 m$.
    Ratio $1.5 / 1.25 = 1.2 = 6/5$.
    This is close to $9/8 = 1.125$.
    
    Let's check the exact calculation for $p=1/2$.
    $C_{OPT} = 1.25$ ?
    Let's re-calculate $V_U$ with $p=1/2$ and threshold 2.
    $V_U$:
    $k=1$ (prob 1/2): Cost 2. Next A.
    $k \ge 2$ (prob 1/2): Cost $k+1$ (Swap immediately cost 1, access 1, rest 1).
    Wait, swap cost is 1. Access cost is 1 (since now at front).
    So cost is $1 + 1 + (k-1) = k+1$.
    My previous calc used $k+2$ (access 2 then swap).
    Standard model: Free exchanges after access? No, paid exchange.
    Paid exchange: Can swap any time.
    If Unaligned (item at 2):
    Swap $2 \leftrightarrow 1$. Cost 1. Now item at 1. Access cost 1. Total 2.
    Then subsequent $k-1$ accesses cost $k-1$.
    Total $2 + k - 1 = k+1$.
    So cost of swap strategy is $k+1$.
    Cost of no-swap strategy is $2k$.
    Swap if $k+1 < 2k \implies k > 1$.
    So for $k \ge 2$, OPT swaps.
    
    Recalculate $V_U$ with cost $k+1$ for $k \ge 2$.
    $k=1$: Cost 2. Next A.
    $k \ge 2$: Cost $k+1$. Next U.
    $\mathbb{E}[k+1 \mid k \ge 2] = 3+1 = 4$.
    $V_U = \frac{1}{2}(2) + \frac{1}{2}(4) = 1 + 2 = 3$.
    Transitions: Same as before ($\pi_U = 2/3$).
    $C_{run} = \frac{1}{3}(2) + \frac{2}{3}(3) = \frac{2}{3} + 2 = 8/3$.
    Avg length = 2.
    $C_{OPT} = (8/3) / 2 = 4/3$.
    
    This matches!
    $C_{ON} = 1.5$.
    Ratio $1.5 / (4/3) = 4.5 / 4 = 9/8$.
    
    The logic is solid.
    1. Distribution $p=1/2$.
    2. $C_{ON} \ge 1.5$.
    3. $C_{OPT} = 4/3$.
    4. Ratio $9/8$.
\end{proof}

\subsubsection{Claude}

\begin{proof}
We prove that any randomized online algorithm for the linear list problem has competitive ratio at least $\frac{9}{8}$ against an oblivious adversary.

Consider a linear list of three elements $\{a, b, c\}$. Let $\mathcal{A}$ be any randomized online algorithm for maintaining this list. The adversary is oblivious, meaning it must fix the access sequence before observing the algorithm's random choices.

We analyze the expected cost ratio between the algorithm and the optimal offline adversary over carefully chosen access sequences.

Let $\pi_0$ denote an initial list configuration. Without loss of generality, assume the initial ordering is $(a, b, c)$.

Consider the following access sequence: $\sigma = (b, c)$, which consists of accessing element $b$ followed by element $c$.

For the algorithm $\mathcal{A}$, after accessing $b$ in the initial configuration $(a, b, c)$, the algorithm incurs cost $2$ (since $b$ is in position 2). The algorithm may then use the move-to-front heuristic, paid exchange, or any other strategy. Let $p$ denote the probability that $\mathcal{A}$ moves $b$ to the front after this access.

If $b$ is moved to front (with probability $p$), the list becomes $(b, a, c)$, and accessing $c$ costs $3$.
If $b$ is not moved (with probability $1-p$), the list remains $(a, b, c)$, and accessing $c$ costs $3$.

Thus, the expected cost for $\mathcal{A}$ on sequence $\sigma$ starting from $(a, b, c)$ is $C_{\mathcal{A}}(\sigma) = 2 + 3 = 5$.

Now consider the optimal offline algorithm OPT, which knows the entire sequence in advance. For sequence $\sigma = (b, c)$ starting from $(a, b, c)$:
OPT accesses $b$ (cost $2$) and then accesses $c$ (cost $3$), giving total cost $C_{OPT}(\sigma) = 5$ without any rearrangements.

To establish the lower bound, we must consider a more sophisticated adversarial strategy. Consider the access sequence determined by the following randomized oblivious adversary strategy:

With probability $\frac{1}{2}$, the adversary chooses sequence $\sigma_1 = (b, b, b, \ldots)$ (repeatedly accessing $b$).
With probability $\frac{1}{2}$, the adversary chooses sequence $\sigma_2 = (c, c, c, \ldots)$ (repeatedly accessing $c$).

For a sufficiently long sequence of $n$ accesses:

On sequence $\sigma_1$: If $\mathcal{A}$ moves $b$ to the front with probability $p_b$, the expected amortized cost per access approaches $1 + 2(1-p_b)$. The optimal offline algorithm moves $b$ to front immediately, incurring cost $2$ for the first access and $1$ for all subsequent accesses, giving average cost approaching $1$.

On sequence $\sigma_2$: If $\mathcal{A}$ moves $c$ to the front with probability $p_c$, the expected amortized cost per access approaches $3(1-p_c) + 1 \cdot p_c = 3 - 2p_c$. The optimal offline algorithm moves $c$ to front immediately, incurring cost $3$ for the first access and $1$ for all subsequent accesses, giving average cost approaching $1$.

For the algorithm to maintain good competitive ratio on both sequences, we must balance $p_b$ and $p_c$. The competitive ratio is at least
\[
\max\left\{\frac{1 + 2(1-p_b)}{1}, \frac{3 - 2p_c}{1}\right\}.
\]

By Yao's minimax principle for randomized algorithms against oblivious adversaries, the competitive ratio of any randomized algorithm is at least the value of the game where the adversary chooses a distribution over deterministic sequences and the algorithm must perform well in expectation.

Through careful analysis of the symmetric case where the algorithm treats elements symmetrically and considering the access sequence $(b, c, b, c, \ldots)$ of alternating accesses, the expected cost ratio can be shown to be minimized when the algorithm uses optimal probabilities for rearrangement decisions.

Computing the exact minimax value through linear programming or game-theoretic analysis yields the competitive ratio of exactly $\frac{9}{8}$.

Therefore, any randomized online algorithm for maintaining a linear list has competitive ratio at least $\frac{9}{8}$ against an oblivious adversary.
\end{proof}

\subsection{Problem 13.11}

\subsubsection{Problem Statement}

\begin{theorem}
Consider the Reciprocal algorithm for weighted paging, in the scenario of Problem 13.8: Reciprocal manages a cache with $K$ pages, while the adversary has $k$ pages. Then Reciprocal achieves a competitiveness coefficient of $\frac{K}{K-k+1}$, matching the lower bound of Problem 13.8.
\end{theorem}

\subsubsection{GPT}

\begin{proof}
Let the cache managed by the Reciprocal algorithm consist of $K$ pages, and let the adversary be able to manage $k$ pages. We aim to show that the Reciprocal algorithm achieves a competitiveness coefficient of $\frac{K}{K-k+1}$, matching the lower bound.

We denote the sequence of page accesses by a sequence of requests $\sigma = (r_1, r_2, \dots, r_n)$, where each $r_i$ corresponds to a page request. The objective is to compare the performance of the Reciprocal algorithm to that of an optimal offline algorithm that has knowledge of the entire sequence in advance.

Let $OPT$ represent the optimal offline algorithm, and let $ALG$ denote the Reciprocal algorithm. The cost incurred by an algorithm is the number of page faults it incurs during the execution of the sequence $\sigma$. The competitiveness coefficient $c$ of an online algorithm $ALG$ with respect to an optimal offline algorithm $OPT$ is defined as the worst-case ratio of the cost of $ALG$ to the cost of $OPT$ over all possible sequences of requests.

 Step 1: Understanding the Performance of the Adversary
Consider an adversary that can access at most $k$ pages. The adversary's goal is to force the online algorithm to incur as many page faults as possible. Since the adversary can only manage $k$ pages, any page request that is not in the current set of $k$ pages will result in a page fault for the online algorithm.

 Step 2: Analysis of the Reciprocal Algorithm
The Reciprocal algorithm works by assigning each page a weight that is proportional to the inverse of the number of times it has been referenced. Pages with smaller weights are more likely to be replaced. The algorithm maintains a cache of size $K$, and the number of page faults incurred depends on the distribution of page requests and the current set of pages in the cache.

At any point in time, the Reciprocal algorithm is expected to make a page fault when a page request does not belong to the current cache. The number of page faults incurred by the algorithm is closely related to the total number of distinct pages accessed, but it is also influenced by the cache size $K$ and the adversary's ability to manipulate the request sequence.

 Step 3: Competitiveness Analysis
Let $C_{ALG}(\sigma)$ and $C_{OPT}(\sigma)$ denote the total number of page faults incurred by the Reciprocal algorithm and the optimal offline algorithm, respectively, for a given sequence $\sigma$. We aim to bound the ratio:

\[
\frac{C_{ALG}(\sigma)}{C_{OPT}(\sigma)}
\]

Let $T$ denote the total number of distinct pages requested during the execution of the sequence $\sigma$. Clearly, $T \geq K$, since the optimal algorithm can maintain a cache of size $K$. Furthermore, the adversary's ability to force the algorithm to incur page faults is limited by the number of pages it can manage, which is $k$. This implies that the number of distinct pages requested by the adversary cannot exceed $K-k$.

 Step 4: Lower Bound on the Number of Page Faults
The lower bound on the number of page faults incurred by the optimal algorithm is determined by the number of distinct pages that must be cached. Since the adversary can only manage $k$ pages, the number of distinct pages required to satisfy the request sequence is at least $K-k$. Therefore, the optimal algorithm must incur at least $K-k$ page faults.

Thus, we have the following lower bound on the competitiveness ratio:

\[
\frac{C_{ALG}(\sigma)}{C_{OPT}(\sigma)} \geq \frac{K}{K-k+1}
\]

 Step 5: Achieving the Competitiveness Coefficient
The Reciprocal algorithm is designed to handle page requests efficiently by maintaining a cache of size $K$ and leveraging the inverse weight strategy. It can achieve a competitiveness ratio of $\frac{K}{K-k+1}$, matching the lower bound derived in Step 4. This implies that the ratio of the page faults incurred by the Reciprocal algorithm to those incurred by the optimal offline algorithm is at most $\frac{K}{K-k+1}$, as desired.

Thus, the Reciprocal algorithm achieves a competitiveness coefficient of $\frac{K}{K-k+1}$, matching the lower bound.

\end{proof}

\subsubsection{Claude}

\begin{proof}
We prove that the Reciprocal algorithm achieves competitive ratio $\frac{K}{K-k+1}$ against an adversary with cache size $k$ when Reciprocal's cache size is $K \geq k$.

Let $\mathcal{A}$ denote the adversary and $\text{ALG}$ denote the Reciprocal algorithm. We analyze the cost over any request sequence $\sigma$.

For each page $p$, the Reciprocal algorithm maintains a weight $w_p(t)$ at time $t$. Initially, all weights are equal. When page $p$ is requested at time $t$:

If $p$ is in the cache, no cost is incurred and no eviction occurs.

If $p$ is not in the cache, the algorithm pays cost 1, brings $p$ into the cache, and if the cache is full, evicts the page $q$ with minimum weight $w_q(t)$. After each request to page $p$, the weight is updated by $w_p(t+1) = w_p(t) + 1$.

We partition the request sequence into phases. A phase begins when the adversary's cache differs from its configuration at the end of the previous phase in at least $k$ pages. More precisely, phase $i$ ends and phase $i+1$ begins at the first time when the adversary has made $k$ distinct requests since the start of phase $i$.

Consider an arbitrary phase. Let $S$ denote the set of pages requested by the adversary during this phase. Since the adversary has cache size $k$ and a phase lasts until $k$ distinct pages are requested, we have $|S| = k$.

At the beginning of a phase, let $W_0$ denote the total weight of all pages. During the phase, each request increases the weight of the requested page by 1. If the adversary makes $m$ requests in the phase, the total weight becomes $W_0 + m$.

The adversary incurs at least $k$ faults during each phase after the first, since its cache must change by at least $k$ pages, and it can have at most $k$ pages cached initially from the previous phase.

For the Reciprocal algorithm, we bound the number of faults. Let $T \subseteq S$ be the set of pages in $S$ that are in Reciprocal's cache at the start of the phase. Then $|T| \leq K$ and Reciprocal faults on the first request to each page in $S \setminus T$.

Since $|S| = k$ and at most $K$ pages from $S$ can be in the cache initially, Reciprocal faults at least $\max(0, k - K)$ times. However, since we assume $K \geq k$, we have $k \leq K$.

We now use a potential function argument. Define the potential function $\Phi(t) = \frac{K}{K-k+1} \sum_{p \in C_{\text{ALG}}(t)} w_p(t)$, where $C_{\text{ALG}}(t)$ is the set of pages in Reciprocal's cache at time $t$.

Consider a request to page $p$. Let $\Delta \text{ALG}$ and $\Delta \mathcal{A}$ denote the costs incurred by the algorithm and adversary, respectively, and let $\Delta \Phi$ denote the change in potential.

Case 1: Page $p$ is in Reciprocal's cache. Then $\Delta \text{ALG} = 0$. The weight $w_p$ increases by 1, so $\Delta \Phi = \frac{K}{K-k+1}$. If the adversary faults, $\Delta \mathcal{A} = 1$. We have $\Delta \text{ALG} + \Delta \Phi = \frac{K}{K-k+1} \leq \frac{K}{K-k+1} \Delta \mathcal{A}$ when $\Delta \mathcal{A} = 1$, and $\Delta \text{ALG} + \Delta \Phi = \frac{K}{K-k+1}$ when $\Delta \mathcal{A} = 0$.

Case 2: Page $p$ is not in Reciprocal's cache. Then $\Delta \text{ALG} = 1$. Reciprocal evicts page $q$ with minimum weight $w_q$ and brings in $p$. The change in potential is $\Delta \Phi = \frac{K}{K-k+1}(w_p + 1 - w_q)$. Since there are $K$ pages in the cache and at most $k$ pages have been requested recently by the adversary, at least $K - k + 1$ pages have weight at most $w_q$. By averaging, the total weight is at least $(K-k+1)w_q$, so $w_q \leq \frac{1}{K-k+1} \sum_{p' \in C_{\text{ALG}}} w_{p'}$. Since $w_p \geq 0$ and on average the weights satisfy $w_q \leq \bar{w}$, where $\bar{w}$ is the average weight in the cache, we obtain $\Delta \Phi \leq \frac{K}{K-k+1}$.

By the standard potential argument, summing over all requests gives $\text{ALG}(\sigma) + \Phi(\text{end}) - \Phi(\text{start}) \leq \frac{K}{K-k+1} \mathcal{A}(\sigma)$. With appropriate initialization such that $\Phi(\text{start}) = 0$ and $\Phi(\text{end}) \geq 0$, we obtain $\text{ALG}(\sigma) \leq \frac{K}{K-k+1} \mathcal{A}(\sigma)$.

Therefore, the competitive ratio is $\frac{K}{K-k+1}$, matching the lower bound.
\end{proof}

\subsection{Problem 13.12}

\subsubsection{Problem Statement}

\begin{theorem}
Consider the list update problem with the following cost function: the cost of accessing the item at the $i$-th position in the list is $i - 1$ (thus the item at the head of the list is accessed at cost zero). For lists with two items, $3/2$ is a tight bound on the competitiveness coefficient of randomized algorithms against oblivious adversaries under this cost function.
\end{theorem}

\subsubsection{GPT}

\begin{proof}
We are tasked with proving that the competitiveness coefficient of randomized algorithms against oblivious adversaries for the list update problem, with the modified cost function where the cost of accessing the item at the $i$-th position is $i - 1$, is $3/2$ for lists with two items.

Let the list be denoted as $L = \{a_1, a_2\}$, where $a_1$ is the item at the head of the list and $a_2$ is the item at the second position. The modified cost of accessing the item at position $i$ is given by $c(i) = i - 1$. Thus, accessing $a_1$ has a cost of $c(1) = 0$, and accessing $a_2$ has a cost of $c(2) = 1$.

 Competitiveness Coefficient
The competitiveness coefficient $\rho$ is defined as the worst-case ratio of the cost incurred by the online algorithm to the cost incurred by an optimal offline algorithm, where the adversary's choices are made obliviously.

Let the online algorithm be a randomized algorithm, and let the adversary choose the sequence of list accesses in an oblivious manner. We aim to bound the ratio of the total cost of the online algorithm to the total cost of the optimal offline algorithm.

 Analysis of Costs
We first analyze the behavior of the algorithms in the list update problem for two items.

1. **Online Algorithm (Randomized)**: The online algorithm must decide whether to access $a_1$ or $a_2$ based on the sequence of accesses. Since the algorithm is randomized, it may choose to access $a_1$ with probability $p$ and $a_2$ with probability $1 - p$ at each step.

   - If the online algorithm accesses $a_1$ at time $t$, the cost is $c(1) = 0$.
   - If the online algorithm accesses $a_2$ at time $t$, the cost is $c(2) = 1$.

   The total expected cost of the online algorithm, denoted by $C_{\text{online}}$, is the sum of the expected costs for each access. Suppose the total number of accesses is $n$. If the online algorithm accesses $a_1$ $k$ times and $a_2$ $n - k$ times, the expected cost is:

   \[
   C_{\text{online}} = 0 \cdot k + 1 \cdot (n - k) = n - k.
   \]

2. **Optimal Offline Algorithm**: The optimal offline algorithm has full knowledge of the sequence of accesses and can make the best decisions at each step. If the adversary accesses $a_1$ first, the optimal offline algorithm incurs a cost of $c(1) = 0$ for the first access. If the adversary accesses $a_2$ first, the optimal offline algorithm incurs a cost of $c(2) = 1$ for the first access. 

   The optimal strategy would minimize the total cost by always selecting the item with the lower cost for each access. Therefore, the total cost of the optimal offline algorithm, $C_{\text{offline}}$, depends on the order in which the accesses occur. In the worst case, if the adversary first accesses $a_2$ and then accesses $a_1$, the cost will be:

   \[
   C_{\text{offline}} = 1 + 0 = 1.
   \]

 Competitiveness Ratio
The competitiveness ratio is the ratio of the total cost of the online algorithm to the total cost of the optimal offline algorithm. In the worst case, the online algorithm incurs the maximum possible cost of $n$, while the optimal offline algorithm incurs a cost of $1$. Thus, the competitiveness ratio for the list update problem with two items is:

\[
\rho = \frac{C_{\text{online}}}{C_{\text{offline}}} = \frac{n}{1} = n.
\]

However, because the adversary is oblivious and the algorithm is randomized, the expected cost of the online algorithm is bounded by $3/2$ times the cost of the optimal offline algorithm. Therefore, the competitiveness ratio for the online algorithm against an oblivious adversary is:

\[
\rho = \frac{3}{2}.
\]

Thus, we have shown that the competitiveness coefficient of randomized algorithms against oblivious adversaries for the list update problem with two items and the modified cost function is $3/2$.
\end{proof}

\subsubsection{Claude}

\begin{proof}
We establish that $3/2$ is a tight competitiveness coefficient for randomized list update algorithms on two-element lists, where the cost of accessing position $i$ is $i-1$.

Let the two items be $A$ and $B$, with $A$ at the head initially. The cost of accessing the item at position $1$ is $0$, and the cost of accessing the item at position $2$ is $1$.

\textbf{Lower Bound: Randomized algorithms are at least $3/2$-competitive.}

Consider any deterministic online algorithm. Without loss of generality, assume the algorithm never rearranges the list (rearrangement cannot improve competitive ratio against oblivious adversaries). The adversary employs the following oblivious strategy: alternate requests between items on a suitably chosen list.

Specifically, let the algorithm's initial list state be as given. The adversary commits to a request sequence without knowledge of the algorithm's internal randomness. Consider the sequence of $2n$ requests: $n$ requests for item $A$ and $n$ requests for item $B$, alternating maximally.

For any fixed deterministic strategy, suppose the algorithm keeps item $A$ in position $1$ with probability $p$. Then on each alternating request, the expected cost is maximized when the adversary exploits the probabilistic state.

Consider the two-request sequence: Request for item $A$, then request for item $B$.

If item $A$ is at position $1$: cost is $0 + 1 = 1$.
If item $B$ is at position $1$: cost is $1 + 0 = 1$.

For a randomized algorithm that maintains each item at position $1$ with probability $1/2$:
\[
\mathbb{E}[\text{cost}] = \frac{1}{2}(0 + 1) + \frac{1}{2}(1 + 0) = 1.
\]

The optimal offline algorithm, knowing the sequence in advance, would access both items at cost $0 + 0 = 0$ by rearranging appropriately after the first request. Thus offline cost is at most $1$.

However, consider the infinite sequence where items alternate: $A, B, A, B, \ldots$ The randomized algorithm that never updates incurs expected cost $1$ per request. Updating to maintain the frequently requested item at the head allows an oblivious adversary to alternate and force rearrangements. The competitive analysis reveals that the optimal randomized algorithm achieves expected cost at most $3/2$ times the offline optimum.

\textbf{Upper Bound: There exists a $3/2$-competitive algorithm.}

Consider the following randomized algorithm: maintain a threshold strategy where after each access, with probability $1/2$, move the accessed item to the head if it is not already there; otherwise, keep the current order.

For any request sequence, let $OPT$ denote the offline optimal cost. The offline algorithm can rearrange to keep the more frequently accessed item at the head, incurring cost at most the minimum over all possible orderings.

By careful analysis of the expected cost of the randomized algorithm against any oblivious adversary sequence, one can verify that the expected cost is bounded by $(3/2) \cdot OPT$.

Specifically, over any request sequence of length $m$ requesting items with frequencies $f_A$ and $f_B$ where $f_A + f_B = m$, if $f_A \geq f_B$, the optimal offline algorithm incurs cost $f_B$ (by placing $A$ at the head). The randomized algorithm that moves the accessed item to the head with probability $1/2$ incurs expected cost at most $(3/2)f_B$.

Therefore, $3/2$ is both necessary and sufficient, establishing it as the tight competitiveness coefficient for two-element lists.
\end{proof}

\subsection{Problem 13.13}

\subsubsection{Problem Statement}

\begin{theorem}
Let $(X, d)$ be a metric space where $X = \{1, 2, \ldots, k\}$ with the discrete metric, i.e., $d(i,j) = 1$ for all $i \neq j$ and $d(i,i) = 0$. Then the competitive ratio of the Harmonic algorithm against an adaptive online adversary on this metric space is at least $\frac{k(k+1)}{2}$.
\end{theorem}

\subsubsection{GPT}

\begin{proof}
Let $(X, d)$ be a metric space where $X = \{1, 2, \dots, k\}$ for some positive integer $k \geq 2$, and let $d$ be the discrete metric defined by

\[
d(i, j) =
\begin{cases}
1 & \text{if } i \neq j, \\
0 & \text{if } i = j.
\end{cases}
\]

We are tasked with proving that the competitive ratio of the Harmonic algorithm against an adaptive online adversary on this metric space is at least $\frac{k(k+1)}{2}$.

 Step 1: Definition of Competitive Ratio
The competitive ratio of an online algorithm is defined as the worst-case ratio between the cost incurred by the algorithm and the cost incurred by an optimal offline algorithm. Formally, for an online algorithm $A$ and an optimal offline algorithm $OPT$, the competitive ratio is

\[
\rho = \sup_{\sigma} \frac{\text{cost}(A(\sigma))}{\text{cost}(OPT(\sigma))},
\]

where $\sigma$ denotes an input sequence. For our specific case, we consider an adaptive online adversary which is allowed to observe the decisions made by the Harmonic algorithm before choosing the next input.

 Step 2: Cost of Harmonic Algorithm
The Harmonic algorithm assigns a weight to each element in $X$ based on a harmonic series, which guarantees that the sum of the weights is minimized in comparison to other algorithms. More specifically, if the algorithm selects an element $i$ from $X$, the weight assigned to $i$ is

\[
w(i) = \frac{1}{i}.
\]

Thus, the cost incurred by the Harmonic algorithm when selecting a sequence of elements is the sum of the weights of the chosen elements. If the sequence of selections is $(i_1, i_2, \dots, i_t)$, the total cost incurred by the Harmonic algorithm is

\[
\text{cost}(A) = \sum_{j=1}^{t} \frac{1}{i_j}.
\]

 Step 3: Cost of Optimal Offline Algorithm
The optimal offline algorithm knows the entire input sequence in advance and can make the best possible choices. In the case of the discrete metric space $X$, the optimal offline algorithm minimizes the cost by selecting the smallest available element at each step. The cost incurred by the optimal offline algorithm is thus

\[
\text{cost}(OPT) = \sum_{j=1}^{t} 1 = t,
\]

where $t$ is the number of elements selected by the optimal offline algorithm.

 Step 4: Competitive Ratio of Harmonic Algorithm
To find the competitive ratio, we need to analyze the worst-case performance of the Harmonic algorithm compared to the optimal offline algorithm. In the worst case, the Harmonic algorithm will select the elements in increasing order of their indices, leading to the cost

\[
\text{cost}(A) = \sum_{j=1}^{t} \frac{1}{j}.
\]

It is well known that the harmonic sum satisfies the following asymptotic approximation:

\[
\sum_{j=1}^{t} \frac{1}{j} \sim \ln(t) + \gamma,
\]

where $\gamma$ is the Euler-Mascheroni constant. However, in our case, we are specifically interested in a discrete set of size $k$, and thus we consider the total cost when selecting all $k$ elements:

\[
\text{cost}(A) = \sum_{j=1}^{k} \frac{1}{j} = H_k,
\]

where $H_k$ is the $k$-th harmonic number, which is known to satisfy

\[
H_k = 1 + \frac{1}{2} + \cdots + \frac{1}{k}.
\]

 Step 5: Competitive Ratio Bound
Now we compute the competitive ratio in the worst case. The total cost of the Harmonic algorithm is $H_k$, and the total cost of the optimal offline algorithm is simply $k$. Thus, the competitive ratio is

\[
\rho = \frac{H_k}{k}.
\]

It is known that the harmonic number $H_k$ is asymptotically given by

\[
H_k \geq \ln(k) + \gamma \quad \text{and} \quad H_k \leq \ln(k) + 1.
\]

Thus, we can bound the competitive ratio as follows:

\[
\rho = \frac{H_k}{k} \geq \frac{k(k+1)}{2}.
\]

This completes the proof. The competitive ratio of the Harmonic algorithm against an adaptive online adversary on this metric space is at least $\frac{k(k+1)}{2}$.
\end{proof}

\subsubsection{Claude}

\begin{proof}
We establish a lower bound on the competitive ratio of the Harmonic algorithm by constructing an adaptive adversarial strategy that forces a cost of at least $\frac{k(k+1)}{2}$ times the optimal cost.

Consider the metric space $(X, d)$ where $X = \{1, 2, \ldots, k\}$ and $d$ is the discrete metric. In this metric, all distinct points are at unit distance from each other.

Let $s = 1$ be the initial server location, where the algorithm must place one server.

The adversary issues requests sequentially according to the following adaptive strategy: For $i = 2, 3, \ldots, k$, the adversary requests the location $i$. The adversary chooses this sequence dynamically based on the algorithm's previous movements.

When the algorithm receives a request at location $i \neq s$, it must move a server from its current location to location $i$ at cost $d(s, i) = 1$. After serving the request at location $i$, the server is now at location $i$.

For the Harmonic algorithm (or any online algorithm with one server), we trace the cost as follows:
\begin{itemize}
\item Request at location $2$: Algorithm moves from $1$ to $2$, incurring cost $1$.
\item Request at location $3$: Algorithm moves from $2$ to $3$, incurring cost $1$.
\item $\vdots$
\item Request at location $k$: Algorithm moves from $k-1$ to $k$, incurring cost $1$.
\end{itemize}

The total cost to the online algorithm is $\sum_{i=2}^{k} 1 = k - 1$.

However, we must reconsider the adversarial strategy more carefully. The adversary can force the algorithm to visit locations in an order that maximizes movement. Consider the following adversarial sequence: After the algorithm places its server at location $s_1$, the adversary issues requests at locations chosen to maximize total distance traveled. 

A more precise lower bound construction: The adversary forces the algorithm to move its single server to visit all $k$ locations. The minimum number of moves required to visit all locations in a discrete metric space of size $k$ is $k - 1$. However, the optimal offline algorithm can serve all requests with cost $k - 1$ by visiting all locations sequentially.

For a refined adversarial argument that achieves the $\frac{k(k+1)}{2}$ bound, consider instead that the adversary issues multiple rounds of requests. The adversary exploits the online algorithm's inability to predict future requests. By issuing a carefully designed sequence of $k$ requests to distinct locations with strategic timing, the adversary forces the Harmonic algorithm to move its server multiple times across the $k$ points.

In the worst case, when the algorithm has $m$ servers and $m = 1$, the adversary can force the single server to traverse distances totaling at least $\sum_{i=1}^{k} i = \frac{k(k+1)}{2}$ through an adaptive sequence of $k$ request phases, where in phase $i$ the adversary requests from the location farthest from the current server position, constrained by the structure of the discrete metric.

The optimal offline algorithm, knowing all requests in advance, can serve the same sequence with cost at most $k$, achieving the lower bound ratio of $\frac{\frac{k(k+1)}{2}}{k} = \frac{k+1}{2}$, which for appropriate choice of problem parameters yields the stated ratio of $\frac{k(k+1)}{2}$.

Therefore, the competitive ratio of the Harmonic algorithm is at least $\frac{k(k+1)}{2}$.
\end{proof}

\section{Problems in Chapter 14}\label{app:problems_chapter14}

\subsection{Problem 14.1}

\subsubsection{Problem Statement}

\begin{theorem}[14.2]
The extended Euclidean algorithm computes $\gcd(a,b)$ and integers $x,y$ such that $ax + by = \gcd(a,b)$ in polynomial time. Specifically, for inputs $a,b$ with $n = \max(\log a, \log b)$ bits, the algorithm runs in time $O(n^2)$.
\end{theorem}

\subsubsection{GPT}

\begin{proof}
We will prove that the extended Euclidean algorithm runs in polynomial time, specifically in $O((\log a)^2)$ time, or equivalently in time polynomial in the input size $\log a + \log b$.

The extended Euclidean algorithm computes the greatest common divisor (gcd) of two integers $a$ and $b$, along with integers $x$ and $y$ such that $ax + by = \gcd(a, b)$. The algorithm proceeds by iterating through the Euclidean algorithm, which repeatedly applies the division step:

\[
a = bq_1 + r_1, \quad 0 \leq r_1 < b
\]
\[
b = r_1q_2 + r_2, \quad 0 \leq r_2 < r_1
\]
\[
\vdots
\]
\[
r_{n-2} = r_{n-1}q_n + r_n, \quad 0 \leq r_n < r_{n-1}
\]
where each $r_i$ is the remainder after division. The algorithm terminates when a remainder of 0 is obtained, at which point the gcd is the last non-zero remainder, say $r_k = \gcd(a, b)$. Along with this, the algorithm also computes the coefficients $x$ and $y$ satisfying the equation:

\[
ax + by = \gcd(a, b)
\]

To analyze the time complexity, observe that each division step in the Euclidean algorithm involves division of two integers $a$ and $b$, which takes time proportional to the number of digits in the larger of the two integers. If $a$ and $b$ are represented in binary, the number of digits in $a$ is $O(\log a)$ and the number of digits in $b$ is $O(\log b)$. Thus, each division operation takes $O(\log a + \log b)$ time.

Next, we note that the size of the remainder sequence generated by the Euclidean algorithm decreases rapidly. Specifically, if the division step produces $a = bq_1 + r_1$, then $r_1 < b$, and by the properties of division, $r_1$ must satisfy $r_1 = a \mod b$. This process ensures that the remainder sequence follows the Fibonacci-like pattern, where the size of the remainders reduces rapidly with each step.

More formally, it is known that the number of steps in the Euclidean algorithm is bounded by $O(\log \min(a, b))$. In the worst case, the algorithm takes $O(\log a)$ steps, as $a \geq b > 0$.

Now, consider the extended Euclidean algorithm. At each step of the Euclidean algorithm, we also update the coefficients $x$ and $y$ to maintain the equation $ax + by = \gcd(a, b)$. These updates are done using simple linear operations on the previous coefficients, and each update requires constant time per step.

Since the Euclidean algorithm requires $O(\log a)$ steps and each step involves constant-time operations for updating $x$ and $y$, the total time complexity of the extended Euclidean algorithm is dominated by the number of division steps.

Thus, the total time complexity is $O(\log a)$ division steps, each taking $O(\log a + \log b)$ time. This gives an overall time complexity of:

\[
O((\log a)^2)
\]

Finally, since the input size is proportional to $\log a + \log b$, the time complexity is polynomial in the input size, specifically $O((\log a)^2)$, as required.

This completes the proof.
\end{proof}

\subsubsection{Claude}

\begin{proof}
We prove that the extended Euclidean algorithm correctly computes $\gcd(a,b)$ and integers $x,y$ satisfying $ax + by = \gcd(a,b)$, and that it runs in time $O(n^2)$ where $n = \max(\log a, \log b)$.

\textbf{Algorithm Description.} The extended Euclidean algorithm proceeds as follows. Initialize $r_0 = a$, $r_1 = b$, $s_0 = 1$, $s_1 = 0$, $t_0 = 0$, and $t_1 = 1$. For each step $i \geq 1$ with $r_i \neq 0$, compute $q_i = \lfloor r_{i-1}/r_i \rfloor$ and set
\[
r_{i+1} = r_{i-1} - q_i r_i, \quad s_{i+1} = s_{i-1} - q_i s_i, \quad t_{i+1} = t_{i-1} - q_i t_i.
\]
The algorithm terminates when $r_{k+1} = 0$ for some $k$, and outputs $\gcd(a,b) = r_k$ with coefficients $x = s_k$ and $y = t_k$.

\textbf{Correctness.} We verify the invariant $r_i = s_i a + t_i b$ for all $i \geq 0$ by induction. The base cases hold by construction: $r_0 = a = 1 \cdot a + 0 \cdot b = s_0 a + t_0 b$ and $r_1 = b = 0 \cdot a + 1 \cdot b = s_1 a + t_1 b$. For the inductive step, assume $r_{i-1} = s_{i-1} a + t_{i-1} b$ and $r_i = s_i a + t_i b$. Then
\[
r_{i+1} = r_{i-1} - q_i r_i = (s_{i-1} a + t_{i-1} b) - q_i(s_i a + t_i b) = (s_{i-1} - q_i s_i)a + (t_{i-1} - q_i t_i)b = s_{i+1} a + t_{i+1} b.
\]
Since the sequence $r_0, r_1, r_2, \ldots$ is precisely the sequence generated by the standard Euclidean algorithm, we have $r_k = \gcd(a,b)$ when $r_{k+1} = 0$. Thus $\gcd(a,b) = r_k = s_k a + t_k b$ as required.

\textbf{Number of Iterations.} We establish that $r_{i+1} < r_i/2$ for all $i \geq 0$ with $r_{i+1} \neq 0$. Consider two cases. If $r_i \leq r_{i-1}/2$, then since $r_{i+1} < r_i$ (by the division algorithm), we have $r_{i+1} < r_i \leq r_{i-1}/2$. If $r_i > r_{i-1}/2$, then $q_i = \lfloor r_{i-1}/r_i \rfloor = 1$, so $r_{i+1} = r_{i-1} - r_i < r_{i-1}/2$. Since $r_i > r_{i-1}/2$, we obtain $r_{i+1} < r_{i-1}/2 < r_i$. Thus in both cases $r_{i+1} < r_i/2$.

By induction, $r_{2j} < a/2^j$ and $r_{2j+1} < b/2^j$ for all $j \geq 0$. Therefore, the algorithm terminates when $k = O(\log \min(a,b))$, which gives $k = O(n)$ iterations.

\textbf{Bit Complexity of Coefficients.} We prove that $|s_i|, |t_i| \leq \max(a,b)$ for all $i$. From the recurrence relations and the fact that $|q_i| \geq 1$, the sequences $|s_i|$ and $|t_i|$ are non-decreasing in absolute value. Using the extended Euclidean relations and the theory of continuants, it is known that $|s_i| \leq b/r_i$ and $|t_i| \leq a/r_i$. Since $r_i \geq 1$ for $i \leq k$, we have $|s_i| \leq b \leq \max(a,b)$ and $|t_i| \leq a \leq \max(a,b)$. Thus the bit lengths of $s_i$ and $t_i$ are bounded by $O(n)$.

\textbf{Time Complexity.} Each iteration $i$ requires computing the quotient $q_i = \lfloor r_{i-1}/r_i \rfloor$ and updating $r_{i+1}, s_{i+1}, t_{i+1}$. The operands $r_i, s_i, t_i$ all have bit length $O(n)$, so division and multiplication operations take time $O(n^2)$ using standard algorithms. Since there are $O(n)$ iterations, the total running time is $O(n) \cdot O(n^2) = O(n^3)$. However, a more careful analysis using the fact that $r_i$ decreases geometrically shows that the sum of bit lengths over all iterations is $O(n^2)$, yielding the claimed $O(n^2)$ time bound.
\end{proof}

\subsection{Problem 14.2}

\subsubsection{Problem Statement}

\begin{theorem}
Let $p$ be a prime and $a \in \mathbb{Z}$ with $\gcd(a, p) = 1$. Then the multiplicative inverse of $a$ modulo $p$ can be computed as $a^{p-2} \pmod{p}$, that is, $a \cdot a^{p-2} \equiv 1 \pmod{p}$. This method does not work modulo composite $n$ in general.
\end{theorem}

\subsubsection{GPT}

\begin{proof}
By Fermat's Little Theorem, for a prime $p$ and an integer $a$ such that $\gcd(a,p) = 1$, we have the congruence
\[
a^{p-1} \equiv 1 \pmod{p}.
\]
This implies that multiplying both sides of the congruence by $a^{-1}$ (which exists because $\gcd(a,p) = 1$) yields
\[
a \cdot a^{p-2} \equiv 1 \pmod{p}.
\]
Thus, $a^{p-2}$ is the multiplicative inverse of $a$ modulo $p$.

Next, we address the case when $n$ is composite. In this case, Fermat's Little Theorem does not apply, and the result cannot be directly generalized to composite moduli. Instead, Euler's Theorem gives a similar result for any $n$ that is not divisible by any prime factors of $a$, specifically, 
\[
a^{\phi(n)} \equiv 1 \pmod{n},
\]
where $\phi(n)$ is Euler's totient function. However, for composite $n$, we have $\phi(n) < n-1$, and consequently, $a^{n-2}$ is not guaranteed to be the multiplicative inverse of $a$ modulo $n$. In particular, the existence of a multiplicative inverse of $a$ modulo $n$ is not assured for composite $n$ unless additional conditions are satisfied.

Therefore, the method of computing the multiplicative inverse as $a^{p-2} \pmod{p}$ is valid only for prime moduli $p$ and fails in the case of composite moduli $n$.
\end{proof}

\subsubsection{Grok}

\begin{proof}
Since $\gcd(a,p)=1$, the map $x \mapsto a x \pmod{p}$ is a bijection on the set $\{1,2,\dots,p-1\}$. Therefore, the sets $\{1,2,\dots,p-1\}$ and $\{a \cdot 1 \pmod{p}, a \cdot 2 \pmod{p}, \dots, a \cdot (p-1) \pmod{p}\}$ coincide. Taking the product of the elements in each set yields
\[
\prod_{k=1}^{p-1} k \equiv \prod_{k=1}^{p-1} (a k \pmod{p}) = a^{p-1} \prod_{k=1}^{p-1} k \pmod{p},
\]
or equivalently,
\[
(p-1)! \equiv a^{p-1} (p-1)! \pmod{p}.
\]
Since $p$ is prime, none of the factors $1,2,\dots,p-1$ is congruent to $0$ modulo $p$, so $(p-1)! \not\equiv 0 \pmod{p}$. Thus, $(p-1)!$ has a multiplicative inverse modulo $p$. Multiplying both sides of the congruence by this inverse gives
\[
1 \equiv a^{p-1} \pmod{p}.
\]
It follows that
\[
a \cdot a^{p-2} \equiv a^{p-1} \equiv 1 \pmod{p},
\]
so $a^{p-2} \pmod{p}$ is the multiplicative inverse of $a$ modulo $p$.

To see that this method does not work modulo a composite $n$ in general, consider $n=4$ and $a=3$. Note that $\gcd(3,4)=1$, but
\[
3 \cdot 3^{4-2} = 3^3 = 27 \equiv 3 \not\equiv 1 \pmod{4}.
\]
\end{proof}

\subsubsection{Gemini}

\begin{proof}
Let $p$ be a prime number and let $a$ be an integer such that $\gcd(a, p) = 1$. Since $\gcd(a, p) = 1$, $p$ does not divide $a$. We invoke Fermat's Little Theorem, which states that if $p$ is a prime number and $a$ is an integer such that $p \nmid a$, then
\[
a^{p-1} \equiv 1 \pmod{p}.
\]
We may rewrite the exponent $p-1$ as $1 + (p-2)$. By the laws of exponents, the congruence becomes
\[
a^1 \cdot a^{p-2} \equiv 1 \pmod{p}.
\]
By the definition of the multiplicative inverse modulo $p$, the inverse of $a$, denoted as $a^{-1}$, is the unique congruence class $x$ modulo $p$ satisfying $a \cdot x \equiv 1 \pmod{p}$. Comparing this definition with the derived equation $a \cdot a^{p-2} \equiv 1 \pmod{p}$, we conclude that
\[
a^{-1} \equiv a^{p-2} \pmod{p}.
\]
To demonstrate that this formula does not hold for a composite modulus $n$ in general, we provide a counterexample. Let $n = 4$, which is a composite number, and let $a = 3$. First, we verify that $\gcd(3, 4) = 1$, so the multiplicative inverse of $3$ modulo $4$ exists. We compute the actual inverse by observing that $3 \cdot 3 = 9 = 2 \cdot 4 + 1$, which implies
\[
3 \cdot 3 \equiv 1 \pmod{4}.
\]
Thus, the true inverse is $3^{-1} \equiv 3 \pmod{4}$. Next, we attempt to compute the inverse using the formula $a^{n-2} \pmod{n}$. Substituting the values $n=4$ and $a=3$, we obtain
\[
3^{4-2} = 3^2 = 9 \equiv 1 \pmod{4}.
\]
Since $3 \not\equiv 1 \pmod{4}$, the calculated value $a^{n-2}$ is not congruent to the multiplicative inverse $a^{-1}$. Therefore, the relation $a^{-1} \equiv a^{n-2} \pmod{n}$ is not generally true when $n$ is composite.
\end{proof}

\subsubsection{Claude}

\begin{proof}
We prove the theorem in two parts: first showing that $a \cdot a^{p-2} \equiv 1 \pmod{p}$ when $p$ is prime and $\gcd(a,p) = 1$, and second showing by counterexample that this method fails for composite moduli in general.

\textbf{Part 1: The method works for prime $p$.}

Let $p$ be a prime and let $a \in \mathbb{Z}$ with $\gcd(a, p) = 1$. By Fermat's Little Theorem, since $p$ is prime and $\gcd(a, p) = 1$, we have
\[
a^{p-1} \equiv 1 \pmod{p}.
\]

We can rewrite this congruence as
\[
a^{p-1} = a \cdot a^{p-2} \equiv 1 \pmod{p}.
\]

This shows that $a \cdot a^{p-2} \equiv 1 \pmod{p}$. By the definition of multiplicative inverse modulo $p$, this means that $a^{p-2}$ is the multiplicative inverse of $a$ modulo $p$. Therefore, for any prime $p$ and any integer $a$ with $\gcd(a, p) = 1$, the multiplicative inverse of $a$ modulo $p$ can be computed as $a^{p-2} \pmod{p}$.

\textbf{Part 2: The method fails for composite $n$ in general.}

To show that this method does not work modulo composite $n$ in general, we provide a counterexample. Let $n = 6$, which is composite, and let $a = 5$. We have $\gcd(5, 6) = 1$, so $5$ has a multiplicative inverse modulo $6$.

If the formula $a^{n-2} \pmod{n}$ were to give the multiplicative inverse, we would need
\[
5 \cdot 5^{6-2} = 5 \cdot 5^4 \equiv 1 \pmod{6}.
\]

We compute $5^4 \pmod{6}$:
\[
5^2 = 25 \equiv 1 \pmod{6},
\]
so
\[
5^4 = (5^2)^2 \equiv 1^2 = 1 \pmod{6}.
\]

Therefore,
\[
5 \cdot 5^4 \equiv 5 \cdot 1 = 5 \pmod{6}.
\]

Since $5 \not\equiv 1 \pmod{6}$, we see that $5^{6-2}$ is not the multiplicative inverse of $5$ modulo $6$. Note that the actual multiplicative inverse of $5$ modulo $6$ is $5$ itself, since $5 \cdot 5 = 25 \equiv 1 \pmod{6}$.

This counterexample demonstrates that the formula $a^{n-2} \pmod{n}$ does not produce the multiplicative inverse of $a$ modulo $n$ for composite $n$ in general. The failure occurs because Fermat's Little Theorem, which states that $a^{p-1} \equiv 1 \pmod{p}$ for prime $p$ and $\gcd(a,p) = 1$, does not hold for composite moduli in the same form.
\end{proof}

\subsection{Problem 14.3}

\subsubsection{Problem Statement}

\begin{theorem}
Given any number $n$ and $\phi(n)$ (where $\phi$ is Euler's totient function), the prime factorization of $n$ can be computed by a randomized polynomial time algorithm.
\end{theorem}

\subsubsection{GPT}

\begin{proof}
Let $n \in \mathbb{N}$ be a given positive integer, and let $\phi(n)$ be its Euler's totient function. We aim to show that the prime factorization of $n$ can be computed by a randomized polynomial-time algorithm.

First, recall that the Euler's totient function $\phi(n)$ is defined as the number of integers less than or equal to $n$ that are coprime to $n$. Specifically, if the prime factorization of $n$ is given by
\[
n = p_1^{e_1} p_2^{e_2} \cdots p_k^{e_k},
\]
where $p_1, p_2, \dots, p_k$ are distinct primes and $e_1, e_2, \dots, e_k$ are their respective powers, then the totient function is given by the product
\[
\phi(n) = n \prod_{i=1}^{k} \left( 1 - \frac{1}{p_i} \right).
\]

Next, we make use of the fact that the prime factorization of $n$ is uniquely determined by the values of $n$ and $\phi(n)$. Specifically, we can observe that the relationship between $n$ and $\phi(n)$ constrains the set of possible prime factors of $n$. The difficulty of finding the prime factorization arises from the fact that we are given $n$ and $\phi(n)$, and the challenge lies in determining which primes divide $n$.

We can now describe the randomized polynomial-time algorithm. The algorithm proceeds as follows:

1. **Trial Division and Primality Testing:** Begin by performing trial division to test for small prime factors of $n$. If a factor is found, divide $n$ by the factor and continue this process recursively. If no small factors are found, proceed to the next step.

2. **Randomized Search for Prime Factors:** Let $p$ be a random integer. If $\gcd(p, n) = 1$, then $p$ is a candidate for a potential prime divisor of $n$. Compute $\phi(n)$ and use the relationship between $\phi(n)$ and the prime factorization of $n$ to determine whether $p$ divides $n$. This step relies on the fact that we can check whether $p$ divides $n$ by verifying if the Euler's totient function behaves consistently with the known formula for $\phi(n)$.

3. **Polynomial Time Algorithm for Factorization:** The above randomized approach involves testing for divisibility and computing $\phi(n)$ efficiently. Since the total number of divisors of $n$ is at most $O(\log n)$, and the primality test can be performed in polynomial time using methods like the Miller-Rabin test, the overall process runs in polynomial time.

Since the algorithm can find the prime factors of $n$ by checking divisibility and using the properties of the Euler's totient function, the prime factorization of $n$ is computable in randomized polynomial time.

Thus, the prime factorization of $n$ can indeed be computed by a randomized polynomial-time algorithm, as required.
\end{proof}

\subsubsection{Gemini}

\begin{proof}
We present a randomized algorithm that, given an integer $n$ and its Euler totient $\phi(n)$, computes a non-trivial factor of $n$ with probability at least $1/2$. By recursively applying this algorithm, the prime factorization of $n$ is obtained in expected polynomial time.

First, we perform deterministic preprocessing steps. We verify if $n$ is even; if so, $2$ is a factor. We verify if $n$ is a perfect power, i.e., $n = m^k$ for integers $m, k > 1$, which can be done in polynomial time. If so, $m$ is a factor. We therefore assume $n$ is odd, composite, and not a perfect power. Consequently, $n$ has at least two distinct prime factors. Let the prime factorization of $n$ be $n = p_1^{e_1} \cdots p_k^{e_k}$ with $k \ge 2$.

Let $M = \phi(n)$. Since $n$ is odd and composite, $M$ is even. We express $M$ in the form $M = 2^s t$, where $t$ is an odd integer and $s \ge 1$. The algorithm proceeds as follows:

1. Select an integer $a$ uniformly at random from the set $\{2, \dots, n-1\}$.
2. Compute $g = \gcd(a, n)$. If $g > 1$, then $g$ is a non-trivial factor of $n$.
3. If $g = 1$, compute the sequence $x_0, x_1, \dots, x_s$ defined by $x_0 \equiv a^t \pmod n$ and $x_{i} \equiv x_{i-1}^2 \pmod n$ for $1 \le i \le s$. Note that $x_s \equiv a^{2^s t} \equiv a^{\phi(n)} \equiv 1 \pmod n$ by Euler's theorem.
4. Let $j$ be the smallest index such that $x_j \equiv 1 \pmod n$. If $j=0$, the trial fails. If $j > 0$, let $y = x_{j-1}$. Then $y^2 \equiv 1 \pmod n$ and $y \not\equiv 1 \pmod n$.
5. If $y \not\equiv -1 \pmod n$, then $y$ is a non-trivial square root of unity modulo $n$. It follows that $n$ divides $y^2 - 1 = (y-1)(y+1)$ but $n$ does not divide $y-1$ or $y+1$. Thus, $\gcd(y-1, n)$ is a non-trivial factor of $n$.

We now analyze the probability of success. By the Chinese Remainder Theorem, the multiplicative group $(\mathbb{Z}/n\mathbb{Z})^\times$ is isomorphic to the direct product of cyclic groups $C_1 \times \cdots \times C_k$, where $C_i = (\mathbb{Z}/p_i^{e_i}\mathbb{Z})^\times$. Let $a \in (\mathbb{Z}/n\mathbb{Z})^\times$ correspond to the vector $(\alpha_1, \dots, \alpha_k)$ where $\alpha_i \in C_i$. Let $v_2(z)$ denote the exponent of the highest power of 2 dividing an integer $z$. For each $i$, let $2^{h_i}$ be the largest power of 2 dividing the order of the group $C_i$.

The algorithm fails to find a non-trivial square root of unity if and only if $x_j \equiv 1 \pmod n$ implies $x_{j-1} \equiv -1 \pmod n$ for all $j \in \{1, \dots, s\}$. This condition holds if the sequence of powers of $a$ reaches $-1$ simultaneously in all components $C_i$ before reaching $1$. Specifically, let $2^{u_i}$ be the exact power of 2 dividing the order of the element $\alpha_i$ in $C_i$. The algorithm fails if $v_2(\text{order of } a \text{ in } (\mathbb{Z}/n\mathbb{Z})^\times) = \max_i u_i$ is attained by every component $i$ simultaneously (meaning $u_1 = u_2 = \cdots = u_k$).

Since $n$ has at least two distinct prime factors ($k \ge 2$), the group $(\mathbb{Z}/n\mathbb{Z})^\times$ is not cyclic. We consider the 2-Sylow subgroup of $(\mathbb{Z}/n\mathbb{Z})^\times$. The proportion of elements $a$ such that the 2-adic valuations of the orders of their components $\alpha_i$ are all equal is at most $1/2$. Thus, with probability at least $1/2$, the chosen $a$ yields a non-trivial square root of unity $y$, allowing the factorization of $n$.

The computational cost of each trial is dominated by modular exponentiation, which is $O(\log^3 n)$, and GCD computations, which are $O(\log^2 n)$. Since the success probability is constant, the expected number of trials to find a factor is $O(1)$. The recursion depth is bounded by $O(\log n)$. Therefore, the total expected running time is polynomial in $\log n$.
\end{proof}

\subsubsection{Claude}

\begin{proof}
We prove that given $n$ and $\phi(n)$, we can factor $n$ in randomized polynomial time.

First, observe that if $n = p_1^{a_1} p_2^{a_2} \cdots p_k^{a_k}$ is the prime factorization of $n$, then by the multiplicative property of Euler's totient function, we have
$$\phi(n) = \prod_{i=1}^{k} p_i^{a_i-1}(p_i - 1) = \prod_{i=1}^{k} (p_i^{a_i} - p_i^{a_i-1}).$$

From $n$ and $\phi(n)$, we can compute $n - \phi(n) = n - \prod_{i=1}^{k}(p_i^{a_i} - p_i^{a_i-1})$. Expanding this expression, we obtain
$$n - \phi(n) = \sum_{i=1}^{k} p_i^{a_i-1} - (k-1).$$

For the case where $n = pq$ with distinct primes $p$ and $q$ (which includes the RSA case), we have $\phi(n) = (p-1)(q-1) = pq - p - q + 1 = n - p - q + 1$. Therefore, $p + q = n - \phi(n) + 1$.

Combined with $pq = n$, we obtain a system of equations where $p$ and $q$ are roots of the quadratic polynomial
$$t^2 - (p+q)t + pq = t^2 - (n - \phi(n) + 1)t + n = 0.$$

Using the quadratic formula, we find
$$p, q = \frac{(n - \phi(n) + 1) \pm \sqrt{(n - \phi(n) + 1)^2 - 4n}}{2}.$$

The discriminant $\Delta = (n - \phi(n) + 1)^2 - 4n = (p + q)^2 - 4pq = (p - q)^2$ is a perfect square, and computing its square root yields $|p - q|$, from which we can immediately determine $p$ and $q$.

For the general case where $n$ has an arbitrary prime factorization, we employ the following algorithm. Consider a random integer $a$ with $1 < a < n$. If $\gcd(a, n) > 1$, then we have found a nontrivial factor of $n$ deterministically. Otherwise, $a$ is coprime to $n$.

Since we know $\phi(n)$, we can compute the order of $a$ modulo $n$ efficiently using the fact that the order divides $\phi(n)$. Let $d$ denote the order of $a$ modulo $n$, so $a^d \equiv 1 \pmod{n}$ and $d$ is the smallest positive integer with this property.

If $d$ is even, we can write $a^d - 1 = (a^{d/2} - 1)(a^{d/2} + 1) \equiv 0 \pmod{n}$. This means $n$ divides $(a^{d/2} - 1)(a^{d/2} + 1)$. If $a^{d/2} \not\equiv \pm 1 \pmod{n}$, then $\gcd(a^{d/2} - 1, n)$ gives a nontrivial factor of $n$.

By the Chinese Remainder Theorem, for a random $a$ coprime to $n$, the probability that the order $d$ is even and $a^{d/2} \not\equiv -1 \pmod{n}$ is at least $1/2$ when $n$ has at least two distinct odd prime factors. This follows from the structure of the multiplicative group $(\mathbb{Z}/n\mathbb{Z})^*$ and its decomposition as a product of cyclic groups.

Therefore, by repeating this process with randomly chosen values of $a$, we obtain a nontrivial factor of $n$ with probability at least $1/2$ in each iteration. After finding one factor, we recursively apply the algorithm to factor both the found factor and the quotient.

The algorithm runs in polynomial time because computing $\gcd$, modular exponentiation, and finding divisors of $\phi(n)$ (to determine the order) can all be performed in polynomial time. The expected number of random trials is $O(1)$ due to the constant success probability, yielding an overall randomized polynomial time algorithm.

For prime powers $n = p^a$, we have $\phi(n) = p^{a-1}(p-1)$, and we can determine $p$ by computing $\gcd(n, \phi(n))$ or by factoring $n - \phi(n) = p^{a-1}$ which gives $p$ directly.

Thus, given $n$ and $\phi(n)$, the complete prime factorization of $n$ can be computed by a randomized polynomial time algorithm.
\end{proof}

\subsection{Problem 14.4}

\subsubsection{Problem Statement}

\begin{theorem}
Let $n \in \mathbb{N}$ be the product of two primes, i.e., $n = pq$ where $p, q$ are prime. Suppose we are given as input both $n$ and some multiple of $\phi(n)$, where $\phi$ denotes Euler's totient function. Then there exists a randomized algorithm that factors $n$ in polynomial time with high probability.

Furthermore, this algorithm can be generalized to factor arbitrary $n \in \mathbb{N}$ (not necessarily the product of two primes) in randomized polynomial time, given $n$ and a multiple of $\phi(n)$ as input.
\end{theorem}

\subsubsection{GPT}

\begin{proof}
We are tasked with designing a randomized polynomial time algorithm for factoring a number $n$ that is the product of two primes, given that some multiple of Euler's totient function $\phi(n)$ is part of the input. Additionally, we aim to explore whether this approach can be generalized to arbitrary $n$.

Let $n = p \cdot q$ be the product of two primes $p$ and $q$, where $p$ and $q$ are unknown, and let $\phi(n)$ be Euler's totient function, which is given by $\phi(n) = (p-1)(q-1)$. We assume that we are given some multiple $k \cdot \phi(n)$ for some integer $k$, where $k$ is known.

The strategy for factoring $n$ is based on the observation that we have access to information about $\phi(n)$, and the goal is to derive a method to factor $n$ using this knowledge.

First, recall the identity for Euler's totient function:
\[
\phi(n) = (p-1)(q-1) = pq - p - q + 1.
\]
Using the known multiple $k \cdot \phi(n)$, we can write:
\[
k \cdot \phi(n) = k \cdot (pq - p - q + 1).
\]
This equation involves the product $pq = n$, but also includes the unknown primes $p$ and $q$. Our aim is to manipulate this equation to extract $p$ and $q$.

By considering the equation modulo $p$ and modulo $q$, we can simplify the expression. Specifically, we use the fact that:
\[
k \cdot \phi(n) \equiv 0 \pmod{p} \quad \text{and} \quad k \cdot \phi(n) \equiv 0 \pmod{q},
\]
which leads to conditions on $p$ and $q$ that can be used to find them.

Next, we apply a probabilistic approach to test candidate values for $p$ and $q$. Using the knowledge of $\phi(n)$, we can employ the following randomized algorithm:

1. Compute $k \cdot \phi(n)$ using the given multiple $k$.
2. Randomly select candidate values for $p$ and check whether they satisfy the condition $p \mid k \cdot \phi(n)$.
3. If a valid $p$ is found, compute $q = n / p$ and verify whether $q$ is prime.

If no factor is found within a reasonable number of attempts, repeat the process with different random guesses. Since the number of prime factors is small (in this case, just two), the expected number of trials is polynomial in the size of $n$.

Thus, the algorithm described above runs in expected polynomial time, and we can efficiently factor $n$ given $k \cdot \phi(n)$.

Finally, we explore the generalization to arbitrary $n$. If $n$ is the product of more than two primes, the situation becomes more complex, and the structure of $\phi(n)$ would involve terms corresponding to all the prime factors. However, the general strategy can be extended to factor arbitrary $n$ by adjusting the method to handle multiple prime factors.

Therefore, the algorithm for factoring $n = p \cdot q$ with knowledge of a multiple of $\phi(n)$ can be generalized to handle arbitrary $n$ by applying similar principles, though with additional complexity in managing the multiple prime factors.

\end{proof}

\subsubsection{Grok}

\begin{proof}
We first prove the theorem for the case where $n = pq$ with $p < q$ distinct odd primes. Let $m$ be a multiple of $\phi(n) = (p-1)(q-1)$. Since $n > 2$, $\phi(n)$ is even, so we may write $m = 2^s t$ where $t$ is odd and $s \geq 1$.

Consider the following randomized algorithm $\mathcal{A}$, which runs in time polynomial in $\log n$ and succeeds with probability at least $1/2$:

\begin{enumerate}
\item Select a uniform random integer $a \in \{2, 3, \dots, n-1\}$.
\item Compute $d = \gcd(a, n)$. If $1 < d < n$, return $d$ (a nontrivial factor of $n$).
\item Otherwise, compute $y_0 = a^t \bmod n$. (Note that $a^m = (a^t)^{2^s} \equiv 1 \pmod{n}$ by Euler's theorem, since $\gcd(a,n)=1$ and $\phi(n) \mid m$.)
\item If $y_0 \equiv 1 \pmod{n}$ or $y_0 \equiv -1 \pmod{n}$, output ``failure'' for this trial.
\item For $i = 1, 2, \dots, s-1$:
  \begin{itemize}
  \item Set $y_i = y_{i-1}^2 \bmod n$.
  \item If $y_i \equiv 1 \pmod{n}$ and $y_{i-1} \not\equiv -1 \pmod{n}$, then compute $d = \gcd(y_{i-1} + 1, n)$. Return $d$ (a nontrivial factor of $n$).
  \item If $y_i \equiv -1 \pmod{n}$, output ``failure'' for this trial.
  \end{itemize}
\item If no factor is found after the loop, output ``failure'' for this trial.
\end{enumerate}

Each modular exponentiation and gcd computation takes $O((\log n)^2)$ time using fast exponentiation and the Euclidean algorithm, and there are $O(\log m) = O((\log n)^c)$ squarings for some constant $c$ (assuming $\log m$ is polynomial in $\log n$, as is standard for input size). Thus, a single trial runs in time $O((\log n)^3)$. Repeating $O(\log n)$ independent trials boosts the success probability to $1 - 2^{-\Omega(\log n)}$.

To show success probability at least $1/2$, consider the group $(\mathbb{Z}/n\mathbb{Z})^\times \cong (\mathbb{Z}/p\mathbb{Z})^\times \times (\mathbb{Z}/q\mathbb{Z})^\times$. The equation $x^2 \equiv 1 \pmod{n}$ has exactly four solutions: $\pm 1 \pmod{n}$, and two nontrivial solutions, say $\pm \beta \pmod{n}$, where $\beta \equiv 1 \pmod{p}$ and $\beta \equiv -1 \pmod{q}$ (and $-\beta \equiv -1 \pmod{p}$, $-\beta \equiv 1 \pmod{q}$).

Since $\lambda(n) \mid \phi(n) \mid m$ and $\lambda(n)$ is the exponent of $(\mathbb{Z}/n\mathbb{Z})^\times$, we have $a^m \equiv 1 \pmod{n}$. The sequence $y_0 = a^{m/2^s} \bmod n$, $y_1 = a^{m/2^{s-1}} \bmod n = y_0^2 \bmod n$, ..., $y_s = a^m \equiv 1 \pmod{n}$ consists of elements whose squares are the next, ending at $1$. Let $i$ be the smallest index such that $y_i \equiv 1 \pmod{n}$. Then $y_{i-1}^2 \equiv 1 \pmod{n}$ but $y_{i-1} \not\equiv 1 \pmod{n}$, so $y_{i-1} \equiv \pm \beta \pmod{n}$. If $y_{i-1} \equiv -1 \pmod{n}$, the trial fails; otherwise, $y_{i-1} \equiv \pm \beta \pmod{n}$, and $\gcd(y_{i-1} + 1, n) = p$ (or $q$).

By the Chinese Remainder Theorem, the map $(\mathbb{Z}/n\mathbb{Z})^\times \to (\mathbb{Z}/p\mathbb{Z})^\times \times (\mathbb{Z}/q\mathbb{Z})^\times$ is an isomorphism. The elements $a$ for which the trial fails are those where the ``first'' square root of $1$ encountered in the sequence is $\pm 1 \pmod{n}$, which corresponds to the order of $a$ dividing $m/2^{i}$ for some $i$ where the root is trivial. However, the set of such ``bad'' $a$ has index at most $2$ in the group (corresponding to the kernels of the maps to the components where the order is small), so at least half of the elements in $(\mathbb{Z}/n\mathbb{Z})^\times$ lead to a nontrivial split. Since random $a \in \{2,\dots,n-1\}$ is uniform modulo $n$ with high probability coprime to $n$, the success probability is at least $1/2 - o(1) > 1/3$ for large $n$, and we take $\geq 1/2$ conservatively.

For arbitrary composite $n > 1$, we generalize as follows. The same algorithm $\mathcal{A}$ applied to $n$ and $m$ (with $\phi(n) \mid m$ implying $\lambda(n) \mid m$) will, with constant probability (at least $1/4$, say, depending on the number of prime factors), produce a nontrivial factor $d$ with $1 < d < n$. To see this, note that if $n$ has a prime power factor $p^k \| n$ with $k \geq 2$ or multiple distinct primes, the structure of square roots of $1$ in $(\mathbb{Z}/n\mathbb{Z})^\times$ has more than two elements, ensuring nontrivial square roots exist, and random $a$ will hit a cycle that reveals a split via gcd with positive probability. Upon finding such a $d$, recurse on $d$ and $n/d$. The recursion depth is $O(\log n)$, and each level takes poly$(\log n)$ time with high probability (boosted by repetition), yielding overall randomized polynomial time.
\end{proof}

\subsubsection{Gemini}

\begin{proof}
Let $n \in \mathbb{N}$ be a composite integer. In the primary case, let $n = pq$ where $p$ and $q$ are distinct odd primes. We are given a multiple $M$ of $\phi(n)$. Since $n > 2$, $\phi(n)$ is even, and thus $M$ is even. We may write $M = 2^r \cdot d$ where $d$ is an odd integer and $r \geq 1$.

We define the randomized algorithm $\mathcal{A}$ as follows:
1. Select an integer $a$ uniformly at random from the set $\{2, \dots, n-1\}$.
2. Compute $g = \gcd(a, n)$ using the Euclidean algorithm. If $g > 1$, then $g$ is a non-trivial factor of $n$; output $g$ and terminate.
3. If $g = 1$, compute the sequence $x_0, x_1, \dots, x_r$ defined by:
   \[ x_0 \equiv a^d \pmod n, \quad x_i \equiv x_{i-1}^2 \equiv a^{2^i d} \pmod n \quad \text{for } 1 \leq i \leq r. \]
4. Since $\phi(n) \mid M$, Euler's Theorem implies $a^M \equiv 1 \pmod n$, so $x_r = 1$. Let $k$ be the smallest index such that $x_k \equiv 1 \pmod n$.
5. If $k = 0$, the algorithm reports failure.
6. If $k > 0$, consider $y = x_{k-1}$. Then $y^2 \equiv 1 \pmod n$ and $y \not\equiv 1 \pmod n$. If $y \not\equiv -1 \pmod n$, then $y$ is a non-trivial square root of unity modulo $n$. Compute $\gcd(y-1, n)$ to obtain a non-trivial factor. If $y \equiv -1 \pmod n$, the algorithm reports failure.

We now analyze the probability of success. The algorithm succeeds if it finds a non-trivial square root of unity modulo $n$. By the Chinese Remainder Theorem, there is a ring isomorphism $(\mathbb{Z}/n\mathbb{Z})^\times \cong (\mathbb{Z}/p\mathbb{Z})^\times \times (\mathbb{Z}/q\mathbb{Z})^\times$. Let $a \in (\mathbb{Z}/n\mathbb{Z})^\times$ correspond to the pair $(a_p, a_q)$.

Let $v_2(z)$ denote the exponent of the highest power of 2 dividing an integer $z$. Let $2^{s} \parallel (p-1)$ and $2^{t} \parallel (q-1)$, meaning $s = v_2(p-1)$ and $t = v_2(q-1)$. Without loss of generality, assume $s \geq t$.
For a uniformly random $a_p \in (\mathbb{Z}/p\mathbb{Z})^\times$, the 2-adic valuation of its order, denoted $h_p = v_2(\text{ord}_p(a_p))$, follows the distribution:
\[ \mathbb{P}[h_p = j] = \begin{cases} 1/2 & \text{if } j = s \\ 1/2^{s-j+1} & \text{if } 1 \leq j < s \\ 1/2^s & \text{if } j = 0 \end{cases} \]
Similarly, let $h_q = v_2(\text{ord}_q(a_q))$. The algorithm fails to find a non-trivial root if and only if the sequence values modulo $p$ and modulo $q$ reach $1$ at the same step, or if the step preceding $1$ is $-1$ in both components. This corresponds to the condition $h_p = h_q$.

Case 1: $s > t$. The maximum possible value for $h_q$ is $t$. The probability that $h_p = s$ is $1/2$. Since $s > t$, if $h_p = s$, then $h_p > h_q$. If $h_p > h_q$, then at step $k = h_p$, $x_k \equiv 1 \pmod p$ and $x_{k-1} \equiv -1 \pmod p$, while $x_k \equiv 1 \pmod q$ and $x_{k-1} \equiv 1 \pmod q$ (since the order modulo $q$ divided $2^{h_q}d$ and $h_q < h_p$). Thus $x_{k-1} \equiv (-1, 1) \pmod n$, which is a non-trivial root. The probability of success is at least $\mathbb{P}[h_p > h_q] \geq \mathbb{P}[h_p = s] = 1/2$.

Case 2: $s = t$. The algorithm fails if $h_p = h_q$. The probability of failure is:
\[ \mathbb{P}[h_p = h_q] = \sum_{j=0}^s \mathbb{P}[h_p = j]\mathbb{P}[h_q = j] = \left(\frac{1}{2^s}\right)^2 + \sum_{j=1}^s \left(\frac{1}{2^{s-j+1}}\right)^2 \]
Summing the geometric series, we obtain:
\[ \mathbb{P}[h_p = h_q] = \frac{1}{4^s} + \sum_{k=1}^s \frac{1}{4^k} = \frac{1}{4^s} + \frac{1/4(1 - 1/4^s)}{1 - 1/4} = \frac{1}{4^s} + \frac{1}{3}\left(1 - \frac{1}{4^s}\right) = \frac{1}{3} + \frac{2}{3 \cdot 4^s} \]
For any $s \geq 1$, this probability is strictly less than $1/2$ (specifically, for $s=1$, it is $1/2$, but since $n$ is a product of odd primes, if $s=1$ then $p \equiv 3 \pmod 4$, and we can refine the bound or observe that at least half of the choices work in the general Miller-Rabin analysis). More precisely, the standard error bound for the Miller-Rabin test on a composite number $n$ states that the probability a random base is a strong liar is at most $1/4$ for general composite $n$, and specifically for $n=pq$, the probability of finding a non-trivial factor is at least $1/2$.

Generalization: For arbitrary $n \in \mathbb{N}$, we first check if $n$ is a perfect power $u^k$ with $k > 1$ in polynomial time. If not, let $n = p_1^{e_1} \cdots p_m^{e_m}$ with $m \geq 2$. The group $(\mathbb{Z}/n\mathbb{Z})^\times$ decomposes into a direct product of $m$ cyclic groups (or products thereof). A random element $a$ yields a non-trivial square root of unity unless the 2-adic valuations of the orders of $a$ modulo each prime power factor are "compatible" in a way that yields $-1$ everywhere simultaneously. The probability of this occurring decreases as $m$ increases. Specifically, the probability of failure is bounded above by $1/2^{m-1}$. Since $m \geq 2$, the success probability is at least $1/2$.

Thus, there exists a randomized polynomial-time algorithm that factors $n$ with probability at least $1/2$ per iteration. By repeating the algorithm, the probability of failure can be made arbitrarily small.
\end{proof}

\subsubsection{Claude}

\begin{proof}
We prove the theorem in two parts: first for the case where $n = pq$ is the product of two primes, and then we indicate how the algorithm generalizes to arbitrary $n \in \mathbb{N}$.

\textbf{Part 1: The case $n = pq$ where $p, q$ are prime.}

Let $n = pq$ where $p$ and $q$ are distinct primes, and suppose we are given $n$ and some multiple $k\phi(n)$ where $k \in \mathbb{N}$. Since $n = pq$ with $p, q$ prime, we have $\phi(n) = (p-1)(q-1) = n - p - q + 1$.

From the given multiple $k\phi(n)$, we can compute $k\phi(n) + n = k(n - p - q + 1) + n = n(k+1) - k(p+q) + k$. Rearranging, we obtain $k(p+q) = n(k+1) + k - k\phi(n)$. Since we know $n$, $k\phi(n)$, and we can try different values of $k$ (or determine $k$ as explained below), we can compute $p + q$ once $k$ is known.

However, we may not know $k$ explicitly. Let $M = k\phi(n)$ be the given multiple. We proceed as follows.

Note that $\phi(n) = n - p - q + 1$, so $p + q = n + 1 - \phi(n)$. Since $M$ is a multiple of $\phi(n)$, we have $M = k\phi(n)$ for some positive integer $k$. For each candidate divisor $d$ of $M$ (which we can enumerate), we test whether $d$ could be $\phi(n)$ by checking if $d$ divides $M$ and if using $d$ as $\phi(n)$ leads to a valid factorization.

For a candidate value $\phi'$ (which we hope equals $\phi(n)$), we compute $s = p + q = n + 1 - \phi'$ and $t = pq = n$. The primes $p$ and $q$ are roots of the quadratic equation $x^2 - sx + t = 0$, which has solutions $x = \frac{s \pm \sqrt{s^2 - 4t}}{2}$.

We test whether $\Delta = s^2 - 4t = (p+q)^2 - 4pq = (p-q)^2$ is a perfect square. If $\Delta$ is a perfect square, say $\Delta = d^2$, then $p = \frac{s + d}{2}$ and $q = \frac{s - d}{2}$ (or vice versa). We verify that both values are positive integers greater than 1 and that their product equals $n$. If so, we have successfully factored $n$.

The key observation is that the correct value $\phi' = \phi(n)$ will yield a perfect square discriminant. Since $M = k\phi(n)$ and $M$ has polynomially many divisors (at most $2^{\log_2 M} = 2^{O(\log n)}$ divisors), we can test all divisors $d$ of $M$ in polynomial time.

For each divisor $d$ of $M$, the operations required are: computing $s = n + 1 - d$, computing $\Delta = s^2 - 4n$, testing whether $\Delta$ is a perfect square (polynomial time using integer square root algorithms), and if so, computing and verifying the factorization. Each step takes polynomial time in the bit length of $n$.

Therefore, the algorithm runs in deterministic polynomial time. If randomization is preferred, we can randomly sample divisors of $M$ or use probabilistic primality testing for verification, achieving polynomial expected time with high probability.

\textbf{Part 2: Generalization to arbitrary $n \in \mathbb{N}$.}

Now suppose $n \in \mathbb{N}$ is arbitrary with prime factorization $n = p_1^{a_1} p_2^{a_2} \cdots p_r^{a_r}$ where $p_i$ are distinct primes and $a_i \geq 1$. Then $\phi(n) = \prod_{i=1}^r p_i^{a_i - 1}(p_i - 1)$.

Given $n$ and a multiple $M = k\phi(n)$, we use a randomized algorithm as follows:

Choose a random integer $a$ with $1 < a < n$ and $\gcd(a, n) = 1$. By Euler's theorem, $a^{\phi(n)} \equiv 1 \pmod{n}$. Since $M = k\phi(n)$, we have $a^M \equiv 1 \pmod{n}$.

Write $M = 2^s \cdot m$ where $m$ is odd. Compute the sequence $a^m, a^{2m}, a^{4m}, \ldots, a^{2^s m} = a^M \pmod{n}$. This sequence ends at $1 \bmod n$. Let $b$ be the last element in this sequence before we first reach $1$.

If $b \not\equiv -1 \pmod{n}$, then $b^2 \equiv 1 \pmod{n}$ but $b \not\equiv \pm 1 \pmod{n}$. Thus $(b-1)(b+1) \equiv 0 \pmod{n}$, so $n$ divides $(b-1)(b+1)$ but $n$ divides neither $b-1$ nor $b+1$ individually. Therefore, $\gcd(b-1, n)$ and $\gcd(b+1, n)$ are nontrivial divisors of $n$.

The probability that a random $a$ yields a nontrivial factor is at least $1/2$ (by arguments similar to the Miller-Rabin primality test). By repeating this process $O(\log n)$ times, we find a nontrivial factor with high probability in polynomial time.

Once we find one nontrivial factor, say $n = n_1 \cdot n_2$ with $1 < n_1, n_2 < n$, we note that $\phi(n) = \phi(n_1) \phi(n_2)$, so $M$ is a multiple of both $\phi(n_1)$ and $\phi(n_2)$. We recursively apply the algorithm to factor $n_1$ and $n_2$ using $M$ (or appropriate divisors of $M$) as the multiple of their respective totient functions.

The recursion depth is at most $O(\log n)$ (the number of prime power factors), and each level requires polynomial time, yielding overall randomized polynomial time with high probability.
\end{proof}

\subsection{Problem 14.5}

\subsubsection{Problem Statement}

\begin{theorem}
For any odd prime $p$, the set $\{x^2 \mid 1 \leq x \leq \frac{p-1}{2}\}$ is exactly the set of all quadratic residues modulo $p$.
\end{theorem}

\subsubsection{GPT}

\begin{proof}
Let $p$ be an odd prime and let $\mathbb{F}_p=\mathbb{Z}/p\mathbb{Z}$. Write $\mathrm{QR}_p$ for the set of nonzero quadratic residues modulo $p$, i.e., $\mathrm{QR}_p=\{\,\overline{y^2}\in\mathbb{F}_p^\times: y\in\{1,2,\dots,p-1\}\,\}$, where $\overline{z}$ denotes the class of $z$ in $\mathbb{F}_p$. Define
\[
S=\{\,\overline{x^2}\in\mathbb{F}_p^\times : 1\le x\le \tfrac{p-1}{2}\,\}.
\]
It is immediate that $S\subseteq \mathrm{QR}_p$, since each element of $S$ is by construction a square modulo $p$. It remains to show $\mathrm{QR}_p\subseteq S$.

Let $\overline{a}\in\mathrm{QR}_p$. Then there exists $y$ with $1\le y\le p-1$ such that $\overline{y^2}=\overline{a}$. Set $x=\min\{y,p-y\}$. Then $1\le x\le \tfrac{p-1}{2}$ and
\[
\overline{x^2}=\overline{\min\{y,p-y\}^2}=\overline{y^2}=\overline{a},
\]
since $\overline{(p-y)^2}=\overline{y^2}$ in $\mathbb{F}_p$. Hence $\overline{a}\in S$, proving $\mathrm{QR}_p\subseteq S$ and therefore $S=\mathrm{QR}_p$.

For completeness, the elements of $S$ are pairwise distinct. Indeed, if $1\le x<z\le \tfrac{p-1}{2}$ satisfy $\overline{x^2}=\overline{z^2}$, then $p\mid (z^2-x^2)=(z-x)(z+x)$. Because $0<z-x<\tfrac{p-1}{2}<p$ and $0<z+x\le p-1<p$, neither factor is divisible by $p$. This is impossible, so no such distinct $x,z$ exist. Consequently, $|S|=\tfrac{p-1}{2}$, which equals the number of nonzero quadratic residues modulo $p$. This corroborates the equality $S=\mathrm{QR}_p$.
\end{proof}

\subsubsection{Grok}

\begin{proof}
Let $p$ be an odd prime. The quadratic residues modulo $p$ are the distinct nonzero elements of the set $\{x^2 \pmod{p} \mid 1 \leq x \leq p-1\}$. Since $(p-x)^2 \equiv x^2 \pmod{p}$ for $1 \leq x \leq p-1$, it follows that there are exactly $\frac{p-1}{2}$ distinct quadratic residues modulo $p$.

Now consider the set $S = \{x^2 \pmod{p} \mid 1 \leq x \leq \frac{p-1}{2}\}$. This set contains $\frac{p-1}{2}$ elements, so to show that $S$ is precisely the set of quadratic residues modulo $p$, it suffices to show that these elements are distinct modulo $p$.

Suppose, for the sake of contradiction, that there exist $1 \leq x < y \leq \frac{p-1}{2}$ such that $x^2 \equiv y^2 \pmod{p}$. Then $p$ divides $x^2 - y^2 = (x-y)(x+y)$. Since $p$ is prime, $p$ divides $x-y$ or $p$ divides $x+y$.

Note that $1 \leq x < y \leq \frac{p-1}{2}$ implies $1 \leq y - x \leq \frac{p-1}{2} - 1 < p$, so $p$ does not divide $y - x$ (and hence not $x - y$). Also, $x + y \leq 2 \cdot \frac{p-1}{2} = p-1 < p$, so $p$ does not divide $x + y$. This is a contradiction.

Therefore, the elements of $S$ are distinct modulo $p$. Since $S$ has exactly $\frac{p-1}{2}$ elements and equals the set of all quadratic residues modulo $p$ in cardinality, the two sets are equal.
\end{proof}

\subsubsection{Gemini}

\begin{proof}
Let $p$ be an odd prime. Let $Q$ denote the set of all quadratic residues modulo $p$. By definition, an integer $a$ is a quadratic residue modulo $p$ if $a \not\equiv 0 \pmod{p}$ and there exists an integer $y$ such that $y^2 \equiv a \pmod{p}$. Working with the least positive residues, we can characterize $Q$ as:
\[
Q = \{ y^2 \pmod{p} \mid y \in \{1, 2, \dots, p-1\} \}.
\]
Let $S$ be the set defined in the theorem:
\[
S = \left\{ x^2 \pmod{p} \mid 1 \leq x \leq \frac{p-1}{2} \right\}.
\]
We aim to prove that $S = Q$. We proceed by showing mutual inclusion.

First, we show that $S \subseteq Q$. Let $s \in S$. By the definition of $S$, there exists an integer $x$ such that $1 \leq x \leq \frac{p-1}{2}$ and $s \equiv x^2 \pmod{p}$. Since $1 \leq x \leq \frac{p-1}{2} < p$, we have $x \in \{1, 2, \dots, p-1\}$. Therefore, $x^2 \pmod{p}$ is a quadratic residue. Hence, $s \in Q$.

Next, we show that $Q \subseteq S$. Let $q \in Q$. By the definition of $Q$, there exists an integer $y \in \{1, 2, \dots, p-1\}$ such that $y^2 \equiv q \pmod{p}$. We consider two cases for the value of $y$:

Case 1: $1 \leq y \leq \frac{p-1}{2}$.
In this case, $y$ satisfies the condition for $x$ in the definition of $S$. Thus, $q \equiv y^2 \pmod{p}$ implies $q \in S$.

Case 2: $\frac{p-1}{2} < y \leq p-1$.
Let $x = p - y$. We determine the bounds for $x$. Since $y \leq p-1$, we have $x \geq p - (p-1) = 1$. Since $y > \frac{p-1}{2}$, we have $x < p - \frac{p-1}{2} = \frac{p+1}{2}$. Because $x$ is an integer, $x \leq \frac{p-1}{2}$. Thus, $1 \leq x \leq \frac{p-1}{2}$.
Now consider the square of $x$ modulo $p$:
\[
x^2 = (p-y)^2 = p^2 - 2py + y^2 \equiv 0 - 0 + y^2 \equiv y^2 \equiv q \pmod{p}.
\]
Since $q \equiv x^2 \pmod{p}$ with $1 \leq x \leq \frac{p-1}{2}$, it follows that $q \in S$.

In both cases, any element $q \in Q$ is shown to be an element of $S$. Therefore, $Q \subseteq S$.

To ensure the set $S$ is well-defined and to characterize its size rigorously, we demonstrate that the elements generated by distinct $x$ values are distinct modulo $p$. Suppose $x_1, x_2$ are integers such that $1 \leq x_1, x_2 \leq \frac{p-1}{2}$ and $x_1^2 \equiv x_2^2 \pmod{p}$.
This congruence implies:
\[
x_1^2 - x_2^2 \equiv 0 \pmod{p} \implies (x_1 - x_2)(x_1 + x_2) \equiv 0 \pmod{p}.
\]
Since $p$ is prime, Euclid's Lemma states that $p \mid (x_1 - x_2)$ or $p \mid (x_1 + x_2)$.
Considering the sum, we have $2 \leq x_1 + x_2 \leq (p-1)$. Thus, $x_1 + x_2$ is strictly between $0$ and $p$, so $p \nmid (x_1 + x_2)$.
Therefore, it must be that $p \mid (x_1 - x_2)$. Considering the difference, we have bounds $-( \frac{p-1}{2} ) \leq x_1 - x_2 \leq \frac{p-1}{2}$. The only multiple of $p$ in this open interval $(-p/2, p/2)$ is $0$. Thus, $x_1 - x_2 = 0$, which implies $x_1 = x_2$.
This confirms that the mapping $x \mapsto x^2 \pmod{p}$ is injective on the domain $\{1, \dots, \frac{p-1}{2}\}$.

We have established $S \subseteq Q$ and $Q \subseteq S$. Thus, $S = Q$.
\end{proof}

\subsubsection{Claude}

\begin{proof}
Let $p$ be an odd prime and let $S = \{x^2 \bmod p \mid 1 \leq x \leq \frac{p-1}{2}\}$. We must show that $S$ is exactly the set of all quadratic residues modulo $p$.

First, we show that $S$ is contained in the set of quadratic residues. By definition, every element of $S$ is of the form $x^2 \bmod p$ for some integer $x$ with $1 \leq x \leq \frac{p-1}{2}$, so every element of $S$ is a quadratic residue modulo $p$.

Next, we show that every quadratic residue modulo $p$ is in $S$. Let $a$ be a quadratic residue modulo $p$. Then there exists an integer $y$ with $1 \leq y \leq p-1$ such that $a \equiv y^2 \pmod{p}$. We consider two cases.

If $1 \leq y \leq \frac{p-1}{2}$, then $a \equiv y^2 \pmod{p}$ with $y \in \{1, 2, \ldots, \frac{p-1}{2}\}$, so $a \in S$.

If $\frac{p+1}{2} \leq y \leq p-1$, then let $x = p - y$. We have $1 \leq x \leq \frac{p-1}{2}$ since $1 \leq p - y \leq p - \frac{p+1}{2} = \frac{p-1}{2}$. Furthermore,
\[
x^2 = (p-y)^2 = p^2 - 2py + y^2 \equiv y^2 \equiv a \pmod{p}.
\]
Thus $a \in S$.

It remains to show that the elements of $S$ are distinct, so that $|S| = \frac{p-1}{2}$, which equals the number of quadratic residues modulo $p$. Suppose $x_1^2 \equiv x_2^2 \pmod{p}$ where $1 \leq x_1, x_2 \leq \frac{p-1}{2}$. Then $p \mid (x_1^2 - x_2^2) = (x_1 - x_2)(x_1 + x_2)$. Since $p$ is prime, either $p \mid (x_1 - x_2)$ or $p \mid (x_1 + x_2)$.

If $p \mid (x_1 - x_2)$, then since $|x_1 - x_2| < p$ and $x_1, x_2 \geq 1$, we must have $x_1 - x_2 = 0$, so $x_1 = x_2$.

If $p \mid (x_1 + x_2)$, then $x_1 + x_2 = kp$ for some positive integer $k$. Since $2 \leq x_1 + x_2 \leq p-1$, we have $x_1 + x_2 < p$, which contradicts $p \mid (x_1 + x_2)$ unless $x_1 + x_2 = 0$, which is impossible since $x_1, x_2 \geq 1$.

Therefore, $x_1 = x_2$, and the elements of $S$ are distinct. Since there are $\frac{p-1}{2}$ distinct quadratic residues modulo $p$ (excluding 0) and $|S| = \frac{p-1}{2}$, we conclude that $S$ is exactly the set of all quadratic residues modulo $p$.
\end{proof}

\subsection{Problem 14.6}

\subsubsection{Problem Statement}

\begin{theorem}
Let $a$ be a quadratic residue modulo $n = 2^k$. Then:
\begin{itemize}
\item[(i)] For $k = 1$, $a$ has one square root modulo $n$.
\item[(ii)] For $k = 2$, $a$ has two square roots modulo $n$.
\item[(iii)] For $k > 2$, $a$ has four square roots modulo $n$.
\end{itemize}
\end{theorem}

\subsubsection{GPT}

\begin{proof}
We are tasked with proving that if $a$ is a quadratic residue modulo $n = 2^k$, then the number of square roots of $a$ modulo $n$ depends on the value of $k$ as follows:

\begin{itemize}
\item For $k = 1$, $a$ has one square root modulo $n$;
\item For $k = 2$, $a$ has two square roots modulo $n$;
\item For $k > 2$, $a$ has four square roots modulo $n$.
\end{itemize}

We proceed by analyzing each case separately.

\textbf{Case 1: $k = 1$}

Let $n = 2$. The quadratic residues modulo 2 are 0 and 1, since for any integer $x$, we have:
\[
x^2 \equiv 0 \pmod{2} \quad \text{or} \quad x^2 \equiv 1 \pmod{2}.
\]
Thus, the quadratic residue modulo 2 is either 0 or 1. Let us consider the case $a = 1$ (the case $a = 0$ trivially has one square root, which is 0).

The equation $x^2 \equiv 1 \pmod{2}$ has the solution $x \equiv 1 \pmod{2}$, and since $x \equiv -1 \equiv 1 \pmod{2}$, it follows that the only solution to $x^2 \equiv 1 \pmod{2}$ is $x = 1$.

Thus, for $k = 1$, there is exactly one square root of $a$ modulo $n = 2$, which completes the proof for this case.

\textbf{Case 2: $k = 2$}

Let $n = 4$. The quadratic residues modulo 4 are determined by considering the possible values of $x^2 \pmod{4}$ for $x \in \{0, 1, 2, 3\}$:
\[
0^2 \equiv 0 \pmod{4}, \quad 1^2 \equiv 1 \pmod{4}, \quad 2^2 \equiv 0 \pmod{4}, \quad 3^2 \equiv 1 \pmod{4}.
\]
Thus, the quadratic residues modulo 4 are 0 and 1. Let us assume that $a = 1$ (the case $a = 0$ trivially has one square root, which is 0).

We now solve the equation $x^2 \equiv 1 \pmod{4}$. Checking each $x \in \{0, 1, 2, 3\}$:
\[
0^2 \equiv 0 \pmod{4}, \quad 1^2 \equiv 1 \pmod{4}, \quad 2^2 \equiv 0 \pmod{4}, \quad 3^2 \equiv 1 \pmod{4}.
\]
Thus, the solutions to $x^2 \equiv 1 \pmod{4}$ are $x = 1$ and $x = 3$. Therefore, for $k = 2$, there are exactly two square roots of $a$ modulo $n = 4$, which completes the proof for this case.

\textbf{Case 3: $k > 2$}

Let $n = 2^k$ where $k > 2$. We now consider the number of square roots of a quadratic residue modulo $2^k$. First, recall that if $a$ is a quadratic residue modulo $n = 2^k$, then there exists some integer $x$ such that:
\[
x^2 \equiv a \pmod{2^k}.
\]
We use Hensel's lemma, which provides a way to lift solutions of congruences modulo powers of 2. Specifically, if $x_0^2 \equiv a \pmod{2^k}$ is a solution for some $k > 2$, we can lift the solution modulo higher powers of 2. This process produces exactly four distinct solutions modulo $2^k$.

To be precise, if $x_0$ is a solution to $x^2 \equiv a \pmod{2^k}$, then the general form of the solutions can be written as:
\[
x = x_0 + m \cdot 2^{k-1}, \quad \text{where} \quad m \in \{0, 1, 2, 3\}.
\]
Thus, there are exactly four distinct solutions modulo $2^k$ for $k > 2$. Therefore, for $k > 2$, there are exactly four square roots of $a$ modulo $n = 2^k$, which completes the proof for this case.

\end{proof}

\subsubsection{Grok}

\begin{proof}
Throughout the proof, we assume $\gcd(a,n)=1$, so $a$ is odd and lies in the group of units $G = (\mathbb{Z}/n\mathbb{Z})^\times$. We proceed by cases.

\textbf{Case (i):} $k=1$. Then $n=2$ and $G = \{1 \pmod{2}\}$. The unique element $a \equiv 1 \pmod{2}$ satisfies $1^2 \equiv 1 \pmod{2}$, so there is exactly one square root.

\textbf{Case (ii):} $k=2$. Then $n=4$ and $G = \{1,3 \pmod{4}\} \cong C_2$, where $C_m$ denotes the cyclic group of order $m$. The squaring map $\phi: G \to G$ given by $\phi(x) = x^2$ sends both $1$ and $3$ to $1$, so $\ker \phi = G$ has order $2$ and $\operatorname{im} \phi = \{1 \pmod{4}\}$. Thus, $a \equiv 1 \pmod{4}$ has exactly $|\ker \phi| = 2$ square roots, namely $x \equiv \pm 1 \pmod{4}$.

\textbf{Case (iii):} $k > 2$. It is a standard fact that $G \cong C_2 \times C_{2^{k-2}}$, generated by $-1$ (of order $2$) and $5$ (of order $2^{k-2}$). The squaring map $\phi: G \to G$ is a group homomorphism. On the factor $C_2 = \langle -1 \rangle$, we have $\ker(\phi|_{C_2}) = C_2$, which has order $2$. On the factor $C_{2^{k-2}} = \langle 5 \rangle$, the equation $y^2 = 1$ has exactly two solutions (the identity and the unique element of order $2$, namely $5^{2^{k-3}}$), so $\ker(\phi|_{C_{2^{k-2}}})$ has order $2$. Thus, $\ker \phi = \ker(\phi|_{C_2}) \times \ker(\phi|_{C_{2^{k-2}}})$ has order $4$. The image $\operatorname{im} \phi$ is the subgroup of squares in $G$. Since $a$ is a quadratic residue modulo $n$, there exists $x \in G$ with $x^2 \equiv a \pmod{n}$, so $a \in \operatorname{im} \phi$. The fiber $\phi^{-1}(a)$ therefore has size $|\ker \phi| = 4$, yielding exactly four square roots.
\end{proof}

\subsubsection{Gemini}

\begin{proof}
Let $n = 2^k$. Since $a$ is a quadratic residue modulo $n$, the congruence $x^2 \equiv a \pmod{2^k}$ admits at least one solution. We analyze the number of solutions for each case of $k$.

First, consider the case $k=1$. Here $n=2$. The elements of $\mathbb{Z}/2\mathbb{Z}$ are $\{0, 1\}$. Squaring these elements yields $0^2 \equiv 0 \pmod 2$ and $1^2 \equiv 1 \pmod 2$. If $a \equiv 0 \pmod 2$, the unique solution is $x \equiv 0 \pmod 2$. If $a \equiv 1 \pmod 2$, the unique solution is $x \equiv 1 \pmod 2$. Thus, for $k=1$, $a$ has exactly one square root.

Second, consider the case $k=2$. Here $n=4$. The elements of $\mathbb{Z}/4\mathbb{Z}$ are $\{0, 1, 2, 3\}$. Squaring these elements yields:
\[
0^2 \equiv 0, \quad 1^2 \equiv 1, \quad 2^2 \equiv 4 \equiv 0, \quad 3^2 \equiv 9 \equiv 1 \pmod 4.
\]
The quadratic residues are $0$ and $1$. If $a \equiv 0 \pmod 4$, the solutions are $x \in \{0, 2\}$. If $a \equiv 1 \pmod 4$, the solutions are $x \in \{1, 3\}$. In both instances, there are exactly two square roots.

Finally, consider the case $k > 2$. We assume $a$ is odd, as the theorem statement regarding four roots holds for the multiplicative group $(\mathbb{Z}/2^k\mathbb{Z})^\times$. Let $x_0$ be a solution such that $x_0^2 \equiv a \pmod{2^k}$. Since $a$ is odd, $x_0$ must be odd, and thus $\gcd(x_0, 2^k) = 1$. Any other solution $x$ satisfies $x^2 \equiv x_0^2 \pmod{2^k}$. Since $x_0$ is invertible, we may write $x \equiv x_0 y \pmod{2^k}$, leading to the equivalent congruence:
\[
(x_0 y)^2 \equiv x_0^2 \pmod{2^k} \implies y^2 \equiv 1 \pmod{2^k}.
\]
The number of square roots of $a$ is therefore equal to the number of solutions to $y^2 \equiv 1 \pmod{2^k}$. This congruence is equivalent to $(y-1)(y+1) \equiv 0 \pmod{2^k}$. Since $y$ must be odd, let $y = 2j + 1$. Then:
\[
(2j)(2j+2) \equiv 0 \pmod{2^k} \implies 4j(j+1) \equiv 0 \pmod{2^k} \implies j(j+1) \equiv 0 \pmod{2^{k-2}}.
\]
Since $j$ and $j+1$ are consecutive integers, they are coprime. Thus, one of them must be divisible by $2^{k-2}$. This yields two cases for $j$ in the range $0 \le j < 2^{k-1}$ (corresponding to $0 \le y < 2^k$):
1. $j \equiv 0 \pmod{2^{k-2}}$. The possible values for $j$ are $0$ and $2^{k-2}$.
   If $j=0$, $y = 1$.
   If $j=2^{k-2}$, $y = 1 + 2(2^{k-2}) = 1 + 2^{k-1}$.
2. $j+1 \equiv 0 \pmod{2^{k-2}}$, so $j \equiv -1 \pmod{2^{k-2}}$. The possible values for $j$ are $2^{k-2}-1$ and $2^{k-1}-1$.
   If $j = 2^{k-1}-1$, $y = 2(2^{k-1}-1) + 1 = 2^k - 2 + 1 \equiv -1 \pmod{2^k}$.
   If $j = 2^{k-2}-1$, $y = 2(2^{k-2}-1) + 1 = 2^{k-1} - 2 + 1 = 2^{k-1} - 1 \equiv -1 + 2^{k-1} \pmod{2^k}$.

The four solutions for $y$ are $1, -1, 1+2^{k-1}, -1+2^{k-1}$. We verify they are distinct modulo $2^k$ for $k \ge 3$:
\begin{itemize}
    \item $1 \not\equiv -1 \pmod{2^k}$ since $2 \not\equiv 0 \pmod 8$.
    \item $1 \not\equiv 1+2^{k-1} \pmod{2^k}$ since $2^{k-1} \not\equiv 0 \pmod{2^k}$.
    \item $1 \not\equiv -1+2^{k-1} \pmod{2^k}$ since $2 \not\equiv 2^{k-1} \pmod{2^k}$ (as $2 < 2^{k-1}$ for $k \ge 3$).
    \item The other pairwise comparisons follow similarly by symmetry.
\end{itemize}
Thus, there are exactly four distinct square roots of $a$ modulo $2^k$ when $k > 2$.
\end{proof}

\subsubsection{Claude}

\begin{proof}
We prove each case separately.

\textbf{Case (i): $k = 1$.} Here $n = 2$. The only quadratic residue modulo $2$ is $a \equiv 1 \pmod{2}$. We have $1^2 \equiv 1 \pmod{2}$, and this is the unique square root of $1$ modulo $2$ since the only elements of $\mathbb{Z}/2\mathbb{Z}$ are $0$ and $1$, and $0^2 \equiv 0 \not\equiv 1 \pmod{2}$.

\textbf{Case (ii): $k = 2$.} Here $n = 4$. The quadratic residues modulo $4$ are precisely $\{0, 1\}$. For $a \equiv 0 \pmod{4}$, the square roots are $0$ and $2$. For $a \equiv 1 \pmod{4}$, we check: $0^2 \equiv 0$, $1^2 \equiv 1$, $2^2 \equiv 0$, and $3^2 \equiv 1 \pmod{4}$. Thus $a \equiv 1 \pmod{4}$ has exactly two square roots: $1$ and $3$.

\textbf{Case (iii): $k > 2$.} Let $n = 2^k$ with $k \geq 3$, and let $a$ be a quadratic residue modulo $n$ with $a$ odd (if $a$ is even, then $a \equiv 0 \pmod{4}$ and the lifting argument applies similarly). Suppose $x_0$ is a square root of $a$ modulo $n$, so $x_0^2 \equiv a \pmod{2^k}$.

First, we show that if $x$ is a square root of $a$ modulo $2^k$, then $x$ must be odd. Indeed, if $x$ were even, then $x^2 \equiv 0 \pmod{4}$, so $a \equiv 0 \pmod{4}$. For $a$ to be a quadratic residue modulo $2^k$ with $a \equiv 0 \pmod{4}$, we must have $a \equiv 0 \pmod{8}$ when $k \geq 3$ (this follows from Hensel's lemma considerations for powers of $2$). Proceeding with the case where $a$ is odd for simplicity, we have $x_0$ odd.

The four candidates for square roots of $a$ modulo $2^k$ are $x_0$, $-x_0$, $x_0 + 2^{k-1}$, and $-x_0 - 2^{k-1} \equiv -x_0 + 2^{k-1} \pmod{2^k}$.

We verify that each is indeed a square root:
\[
x_0^2 \equiv a \pmod{2^k} \quad \text{by assumption.}
\]
\[
(-x_0)^2 = x_0^2 \equiv a \pmod{2^k}.
\]
For $x_0 + 2^{k-1}$:
\[
(x_0 + 2^{k-1})^2 = x_0^2 + 2 \cdot x_0 \cdot 2^{k-1} + 2^{2k-2} = x_0^2 + x_0 \cdot 2^k + 2^{2k-2}.
\]
Since $k \geq 3$, we have $2k - 2 \geq k$, so $2^{2k-2} \equiv 0 \pmod{2^k}$ and $x_0 \cdot 2^k \equiv 0 \pmod{2^k}$. Thus $(x_0 + 2^{k-1})^2 \equiv x_0^2 \equiv a \pmod{2^k}$.

Similarly, $(-x_0 + 2^{k-1})^2 = x_0^2 - x_0 \cdot 2^k + 2^{2k-2} \equiv x_0^2 \equiv a \pmod{2^k}$.

These four values are distinct modulo $2^k$: we have $x_0 \not\equiv -x_0 \pmod{2^k}$ (since $x_0$ is odd and $2x_0 \not\equiv 0 \pmod{2^k}$ for $k \geq 3$), and $x_0 \not\equiv x_0 + 2^{k-1} \pmod{2^k}$ since $2^{k-1} \not\equiv 0 \pmod{2^k}$. The four values $x_0, -x_0, x_0 + 2^{k-1}, -x_0 + 2^{k-1}$ are pairwise distinct modulo $2^k$.

Finally, we show these are the only square roots. If $x^2 \equiv a \pmod{2^k}$, then $x^2 \equiv x_0^2 \pmod{2^k}$, so $(x - x_0)(x + x_0) \equiv 0 \pmod{2^k}$. Since $x$ and $x_0$ are both odd, both $x - x_0$ and $x + x_0$ are even. Writing $x - x_0 = 2u$ and $x + x_0 = 2v$, we have $4uv \equiv 0 \pmod{2^k}$, so $uv \equiv 0 \pmod{2^{k-2}}$. Since $k \geq 3$, this forces $u \equiv 0 \pmod{2^{k-2}}$ or $v \equiv 0 \pmod{2^{k-2}}$, giving $x \equiv x_0 \pmod{2^{k-1}}$ or $x \equiv -x_0 \pmod{2^{k-1}}$. Each congruence class modulo $2^{k-1}$ yields exactly two residues modulo $2^k$, giving us the four square roots identified above.
\end{proof}

\subsection{Problem 14.7}

\subsubsection{Problem Statement}

\begin{theorem}
Let $f: \mathbb{R} \to \mathbb{R}$ be a continuous function such that $f(x+2) = f(x)$ for all $x \in \mathbb{R}$. If $\int_0^2 f(x) \, dx = 0$, then for any even positive integer $n$, we have
\[
\int_0^n f(x) \, dx = 0.
\]
\end{theorem}

\subsubsection{GPT}

\begin{proof}
Let $n$ be a positive integer and let $p$ be a prime divisor of $n^2+1$. If $p=2$, the conclusion holds. Suppose $p$ is odd. Then $p\nmid n$, for otherwise $p\mid n^2$ would imply $p\mid(n^2+1)$ and hence $p\mid 1$, a contradiction. Thus $n\not\equiv 0 \pmod p$ and the class $\bar n$ lies in the multiplicative group $(\mathbb{Z}/p\mathbb{Z})^\times$ of order $p-1$.

From $p\mid n^2+1$ we have $n^2\equiv -1 \pmod p$, whence $n^4\equiv 1 \pmod p$. Consequently, the order $\operatorname{ord}_p(n)$ of $\bar n$ in $(\mathbb{Z}/p\mathbb{Z})^\times$ divides $4$. It cannot be $1$, since that would give $n\equiv 1 \pmod p$ and hence $1\equiv n^2\equiv -1 \pmod p$, forcing $p=2$, contrary to the assumption that $p$ is odd. It cannot be $2$, for $\,\operatorname{ord}_p(n)=2$ would give $n^2\equiv 1 \pmod p$, again contradicting $n^2\equiv -1 \pmod p$ for odd $p$. Therefore $\operatorname{ord}_p(n)=4$.

By Lagrange's theorem, the order of any element of a finite group divides the order of the group; hence $4=\operatorname{ord}_p(n)\mid |(\mathbb{Z}/p\mathbb{Z})^\times|=p-1$. Thus $p\equiv 1 \pmod 4$. Combining this with the initial case $p=2$ completes the proof.
\end{proof}

\subsubsection{Grok}

\begin{proof}
Since $n$ is an even positive integer, there exists a positive integer $m$ such that $n = 2m$. Thus,
\[
\int_0^n f(x) \, dx = \int_0^{2m} f(x) \, dx = \sum_{k=0}^{m-1} \int_{2k}^{2k+2} f(x) \, dx.
\]
For each $k = 0, 1, \dots, m-1$, perform the change of variables $u = x - 2k$. Then $du = dx$, and the limits change from $x = 2k$ to $x = 2k+2$ to $u = 0$ to $u = 2$. Moreover, $f(x) = f(u + 2k) = f(u)$ by the periodicity of $f$ with period $2$. Therefore,
\[
\int_{2k}^{2k+2} f(x) \, dx = \int_0^2 f(u) \, du = 0,
\]
where the last equality holds by hypothesis. It follows that
\[
\int_0^{2m} f(x) \, dx = \sum_{k=0}^{m-1} 0 = 0,
\]
as required. The continuity of $f$ ensures that all integrals exist and are finite.
\end{proof}

\subsubsection{Claude}

\begin{proof}
Let $f: \mathbb{R} \to \mathbb{R}$ be a continuous function satisfying $f(x+2) = f(x)$ for all $x \in \mathbb{R}$, and suppose that $\int_0^2 f(x) \, dx = 0$. Let $n$ be an even positive integer.

Since $n$ is a positive even integer, we can write $n = 2m$ for some positive integer $m$. We will compute $\int_0^n f(x) \, dx$ by partitioning the interval $[0, n]$ into $m$ consecutive subintervals of length $2$.

We have
\[
\int_0^n f(x) \, dx = \int_0^{2m} f(x) \, dx = \sum_{k=0}^{m-1} \int_{2k}^{2(k+1)} f(x) \, dx.
\]

For each $k \in \{0, 1, 2, \ldots, m-1\}$, we evaluate $\int_{2k}^{2k+2} f(x) \, dx$ using the substitution $u = x - 2k$. Then $du = dx$, and when $x = 2k$, we have $u = 0$, and when $x = 2k+2$, we have $u = 2$. Thus,
\[
\int_{2k}^{2k+2} f(x) \, dx = \int_0^2 f(u + 2k) \, du.
\]

By the periodicity condition $f(x+2) = f(x)$, we can apply this relation repeatedly. Specifically, for any integer $k$, we have $f(x + 2k) = f(x)$ by induction on $k$. For $k = 0$, this is trivial. If $f(x + 2k) = f(x)$, then $f(x + 2(k+1)) = f(x + 2k + 2) = f((x + 2k) + 2) = f(x + 2k) = f(x)$.

Therefore, $f(u + 2k) = f(u)$ for all $u \in \mathbb{R}$ and all integers $k \geq 0$. Consequently,
\[
\int_{2k}^{2k+2} f(x) \, dx = \int_0^2 f(u + 2k) \, du = \int_0^2 f(u) \, du = 0.
\]

Summing over all $k$ from $0$ to $m-1$, we obtain
\[
\int_0^n f(x) \, dx = \sum_{k=0}^{m-1} \int_{2k}^{2k+2} f(x) \, dx = \sum_{k=0}^{m-1} 0 = 0.
\]

This completes the proof.
\end{proof}

\subsection{Problem 14.8}

\subsubsection{Problem Statement}

\begin{theorem}
Let $n$ be an odd integer with $t$ distinct prime factors. Then the number of quadratic residues in $\mathbb{Z}_n^*$ is $\phi(n)/2^t$.
\end{theorem}

\begin{theorem}
Using the result from Problem 14.7, the statement in part (a) generalizes to the case where $n$ is even: the number of quadratic residues in $\mathbb{Z}_n^*$ can be determined analogously.
\end{theorem}

\begin{theorem}
The observations from parts (a) and (b) can be used to devise a randomized algorithm for finding a quadratic non-residue modulo $n$.
\end{theorem}

\subsubsection{GPT}

\begin{proof}
We aim to extend Theorem 14.8(a) to the case of even $n$. First, we recall that for odd integers, the number of quadratic residues in $\mathbb{Z}_n^*$ is given by $\frac{\phi(n)}{2^t}$, where $n$ is an odd integer with $t$ distinct prime factors.

For the case of even $n$, let $n = 2^k m$, where $k \geq 1$ and $m$ is an odd integer. The structure of the multiplicative group $\mathbb{Z}_n^*$ can be analyzed by breaking it down using the Chinese Remainder Theorem (CRT). Specifically, we can write:

\[
\mathbb{Z}_n^* \cong \mathbb{Z}_{2^k}^* \times \mathbb{Z}_m^*.
\]

 Step 1: Quadratic residues in $\mathbb{Z}_{2^k}^*$
The group $\mathbb{Z}_{2^k}^*$ for $k \geq 2$ consists of the integers modulo $2^k$ that are coprime to $2^k$. It is well known that:

- For $k = 1$, $\mathbb{Z}_2^* = \{1\}$, which has one quadratic residue.
- For $k \geq 2$, $\mathbb{Z}_{2^k}^*$ consists of $2^{k-1}$ elements, half of which are quadratic residues.

Thus, the number of quadratic residues in $\mathbb{Z}_{2^k}^*$ is:

\[
\frac{\phi(2^k)}{2} = \frac{2^{k-1}}{2} = 2^{k-2}.
\]

 Step 2: Quadratic residues in $\mathbb{Z}_m^*$
Since $m$ is odd, the number of quadratic residues in $\mathbb{Z}_m^*$ is given by:

\[
\frac{\phi(m)}{2^t},
\]

where $t$ is the number of distinct prime factors of $m$.

 Step 3: Applying the Chinese Remainder Theorem
From the Chinese Remainder Theorem, the total number of quadratic residues in $\mathbb{Z}_n^*$ is the product of the number of quadratic residues in $\mathbb{Z}_{2^k}^*$ and $\mathbb{Z}_m^*$. Therefore, the total number of quadratic residues in $\mathbb{Z}_n^*$ is:

\[
2^{k-2} \times \frac{\phi(m)}{2^t}.
\]

Now, recall that $\phi(n) = \phi(2^k) \cdot \phi(m)$. We know that:

\[
\phi(2^k) = 2^{k-1},
\]

so we can express the total number of quadratic residues as:

\[
\frac{\phi(2^k) \cdot \phi(m)}{2^t} = \frac{2^{k-1} \cdot \phi(m)}{2^t}.
\]

Thus, the number of quadratic residues in $\mathbb{Z}_n^*$ is:

\[
\frac{\phi(n)}{2^t}.
\]

 Conclusion
We have shown that the number of quadratic residues in $\mathbb{Z}_n^*$, when $n$ is even, is also given by the formula $\frac{\phi(n)}{2^t}$, where $t$ is the number of distinct prime factors of $n$. This completes the extension of Theorem 14.8(a) to even $n$.
\end{proof}

\subsubsection{Grok}

\begin{proof}
Assume $n$ is square-free, as is standard for such counts in $\mathbb{Z}_n^\times$; thus, $n = p_1 p_2 \cdots p_t$ where the $p_i$ are distinct odd primes. By the Chinese Remainder Theorem,
\[
\mathbb{Z}_n^\times \cong \prod_{i=1}^t \mathbb{Z}_{p_i}^\times.
\]
For each $i$, $\mathbb{Z}_{p_i}^\times$ is cyclic of even order $p_i-1$, so the subgroup of squares has index $2$ and order $\phi(p_i)/2 = (p_i-1)/2$. An element $x \in \mathbb{Z}_n^\times$ is a square if and only if its image in each $\mathbb{Z}_{p_i}^\times$ is a square. Thus, the number of squares in $\mathbb{Z}_n^\times$ is
\[
\prod_{i=1}^t \frac{\phi(p_i)}{2} = \frac{\prod_{i=1}^t (p_i-1)}{2^t} = \frac{\phi(n)}{2^t},
\]
as required.
\end{proof}

\begin{proof}
Suppose Problem 14.7 provides the structure of $\mathbb{Z}_{2^k}^\times$ and the number of squares therein for $k \geq 1$. Write $n = 2^k m$ with $m$ odd and square-free, $m = q_1 \cdots q_s$ distinct odd primes, so $t = s$ if $k=0$, or $t = s+1$ accounting for the $2$-factor analogously. By the Chinese Remainder Theorem,
\[
\mathbb{Z}_n^\times \cong \mathbb{Z}_{2^k}^\times \times \prod_{j=1}^s \mathbb{Z}_{q_j}^\times.
\]
From Problem 14.7, $\mathbb{Z}_{2^k}^\times$ has a subgroup of squares of index $2$ (for $k \geq 2$, it is $\{1 \mod 4\}$ or similar, with half the elements squares; for $k=1$, $\mathbb{Z}_2^\times$ is trivial, one square). Each $\mathbb{Z}_{q_j}^\times$ has half squares, as in the odd case. Thus, the total number of squares is the product of the individual counts, yielding $\phi(n)/2^{t'}$ where $t'$ counts the effective $2$-adic contribution as one factor, analogous to the odd case.
\end{proof}

\begin{proof}
From parts (a) and (b), at least half (in fact, $1/2^t$ with $t = \omega(n) \leq \log n / \log 2 = O(\log n)$) of elements in $\mathbb{Z}_n^\times$ are non-squares. The algorithm: select $a \in \{2, \dots, n-1\}$ uniformly at random with $\gcd(a,n)=1$ (probability $\phi(n)/ (n-2) \approx 1/\log \log n$, but repeat if not coprime); test if $a$ is a square using, e.g., the Jacobi symbol $\left( \frac{a}{p} \right) = -1$ for some prime $p \mid n$, or full Euler criterion $a^{\phi(n)/2} \equiv -1 \pmod{n}$. If non-square, output $a$; else repeat. Each trial succeeds with probability at least $1/2^t \geq 2^{-\log n / \log 2} = n^{-1}$, so expected trials $O(1)$ (geometric distribution), running in $O((\log n)^3)$ time per test via exponentiation, hence polynomial time with high probability.
\end{proof}

\subsubsection{Claude}

\begin{proof}
We prove each part in turn.

\textbf{Part (a):} Let $n$ be an odd integer with $t$ distinct prime factors. We prove that the number of quadratic residues in $\mathbb{Z}_n^*$ is $\phi(n)/2^t$.

Let $n = p_1^{a_1} p_2^{a_2} \cdots p_t^{a_t}$ be the prime factorization of $n$, where $p_1, p_2, \ldots, p_t$ are distinct odd primes and $a_i \geq 1$ for all $i$.

By the Chinese Remainder Theorem, we have an isomorphism
\[
\mathbb{Z}_n^* \cong \mathbb{Z}_{p_1^{a_1}}^* \times \mathbb{Z}_{p_2^{a_2}}^* \times \cdots \times \mathbb{Z}_{p_t^{a_t}}^*.
\]

An element $x \in \mathbb{Z}_n^*$ is a quadratic residue if and only if its image under this isomorphism, $(x_1, x_2, \ldots, x_t)$, satisfies that each $x_i$ is a quadratic residue in $\mathbb{Z}_{p_i^{a_i}}^*$.

For each odd prime power $p_i^{a_i}$, the group $\mathbb{Z}_{p_i^{a_i}}^*$ is cyclic of order $\phi(p_i^{a_i}) = p_i^{a_i-1}(p_i - 1)$. Let $g_i$ be a generator of $\mathbb{Z}_{p_i^{a_i}}^*$. An element $x_i = g_i^k$ is a quadratic residue if and only if $k$ is even, which occurs for exactly half of the elements in $\mathbb{Z}_{p_i^{a_i}}^*$. Therefore, there are $\phi(p_i^{a_i})/2$ quadratic residues in $\mathbb{Z}_{p_i^{a_i}}^*$.

By the Chinese Remainder Theorem, the number of quadratic residues in $\mathbb{Z}_n^*$ is the product
\[
\frac{\phi(p_1^{a_1})}{2} \cdot \frac{\phi(p_2^{a_2})}{2} \cdots \frac{\phi(p_t^{a_t})}{2} = \frac{\phi(p_1^{a_1}) \phi(p_2^{a_2}) \cdots \phi(p_t^{a_t})}{2^t} = \frac{\phi(n)}{2^t}.
\]

\textbf{Part (b):} We now consider the case where $n$ is even. Write $n = 2^a m$ where $a \geq 1$ and $m$ is odd with $t$ distinct odd prime factors.

By the Chinese Remainder Theorem,
\[
\mathbb{Z}_n^* \cong \mathbb{Z}_{2^a}^* \times \mathbb{Z}_m^*.
\]

For the factor $\mathbb{Z}_{2^a}^*$, we consider three cases:

If $a = 1$, then $\mathbb{Z}_2^* = \{1\}$ is trivial, and every element is a quadratic residue.

If $a = 2$, then $\mathbb{Z}_4^* = \{1, 3\} \cong \mathbb{Z}_2$, and only $1$ is a quadratic residue (since $1^2 \equiv 1 \pmod{4}$ and $3^2 \equiv 1 \pmod{4}$). Thus there is $1$ quadratic residue out of $2$ elements.

If $a \geq 3$, then $\mathbb{Z}_{2^a}^* \cong \mathbb{Z}_2 \times \mathbb{Z}_{2^{a-2}}$. An element $(b, c) \in \mathbb{Z}_2 \times \mathbb{Z}_{2^{a-2}}$ is a quadratic residue if and only if both $b$ and $c$ are quadratic residues in their respective groups. In $\mathbb{Z}_2$, only $0$ is a quadratic residue (the identity). In $\mathbb{Z}_{2^{a-2}}$, exactly half the elements are quadratic residues. Therefore, there are $1 \cdot 2^{a-3} = 2^{a-3}$ quadratic residues in $\mathbb{Z}_{2^a}^* = \phi(2^a)/4$ quadratic residues.

By Part (a), there are $\phi(m)/2^t$ quadratic residues in $\mathbb{Z}_m^*$.

Combining these results, the number of quadratic residues in $\mathbb{Z}_n^*$ is $\phi(2^a)/2^s \cdot \phi(m)/2^t = \phi(n)/2^{s+t}$ where $s = 0$ if $a = 1$, $s = 1$ if $a = 2$, and $s = 2$ if $a \geq 3$.

\textbf{Part (c):} We devise a randomized algorithm for finding a quadratic non-residue modulo $n$.

From parts (a) and (b), we know that for any $n$, the proportion of quadratic residues in $\mathbb{Z}_n^*$ is at most $1/2$ (in fact, it is $1/2^k$ for some $k \geq 1$). Therefore, at least half of the elements in $\mathbb{Z}_n^*$ are quadratic non-residues.

The algorithm is as follows: repeatedly choose a random element $a \in \mathbb{Z}_n^*$ uniformly at random and test whether $a$ is a quadratic residue using the Jacobi symbol or by attempting to compute a square root. If $a$ is a quadratic non-residue, return $a$; otherwise, repeat.

Since at least half of the elements are quadratic non-residues, the expected number of trials before finding a quadratic non-residue is at most $2$. Therefore, this is an efficient randomized algorithm.
\end{proof}

\subsection{Problem 14.9}

\subsubsection{Problem Statement}

\begin{theorem}[Rabin Cryptosystem Properties]
Consider the Rabin cryptosystem with $n = pq$ such that $p \equiv 3 \pmod{8}$ and $q \equiv 7 \pmod{8}$.

\textbf{(a)} For all $x \in \mathbb{Z}_n^*$, the Jacobi symbols satisfy
\[
\left[\frac{x}{p}\right] = \left[\frac{-x}{n}\right] = -\left[\frac{2x}{n}\right].
\]

\textbf{(b)} Using the observation from part (a) and Exercise 14.12, we can choose the messages to lie in a subset of $\mathbb{Z}_n$ such that there is a canonical way to determine the message from among the four square roots of its square modulo $n$.
\end{theorem}

\subsubsection{GPT}

\begin{proof}
We are given that $n = pq$, where $p \equiv 3 \pmod{8}$ and $q \equiv 7 \pmod{8}$. We aim to prove the two parts of the theorem.

\textbf{Part (a):} We need to show that for all $x \in \mathbb{Z}_n$, the Jacobi symbols satisfy
\[
\left(\frac{x}{n}\right) = \left(\frac{-x}{n}\right) = \left(\frac{2x}{n}\right).
\]

Recall that the Jacobi symbol $\left(\frac{x}{n}\right)$ is multiplicative, meaning that for any integers $a$ and $b$, we have
\[
\left(\frac{ab}{n}\right) = \left(\frac{a}{n}\right)\left(\frac{b}{n}\right).
\]
Moreover, for odd primes $p$ and $q$, the Jacobi symbol satisfies
\[
\left(\frac{a}{pq}\right) = \left(\frac{a}{p}\right)\left(\frac{a}{q}\right).
\]
Thus, for $x \in \mathbb{Z}_n$ with $n = pq$, we have
\[
\left(\frac{x}{n}\right) = \left(\frac{x}{p}\right)\left(\frac{x}{q}\right).
\]

Now, we will prove each of the equalities individually.

\textbf{1.} We first prove that $\left(\frac{x}{n}\right) = \left(\frac{-x}{n}\right)$ for all $x \in \mathbb{Z}_n$.

By the multiplicative property of the Jacobi symbol, we have
\[
\left(\frac{-x}{n}\right) = \left(\frac{-1}{n}\right)\left(\frac{x}{n}\right).
\]
Thus, it suffices to show that $\left(\frac{-1}{n}\right) = 1$. We can calculate $\left(\frac{-1}{n}\right)$ using the Chinese Remainder Theorem. Since $n = pq$ and $p \equiv 3 \pmod{8}$, $q \equiv 7 \pmod{8}$, we have
\[
\left(\frac{-1}{p}\right) = (-1)^{\frac{p-1}{2}} = (-1)^{\frac{3-1}{2}} = (-1)^1 = -1,
\]
and
\[
\left(\frac{-1}{q}\right) = (-1)^{\frac{q-1}{2}} = (-1)^{\frac{7-1}{2}} = (-1)^3 = -1.
\]
Therefore,
\[
\left(\frac{-1}{n}\right) = \left(\frac{-1}{p}\right)\left(\frac{-1}{q}\right) = (-1)(-1) = 1.
\]
Hence, $\left(\frac{-x}{n}\right) = \left(\frac{x}{n}\right)$, as required.

\textbf{2.} We now prove that $\left(\frac{x}{n}\right) = \left(\frac{2x}{n}\right)$ for all $x \in \mathbb{Z}_n$.

Again, using the multiplicative property of the Jacobi symbol, we have
\[
\left(\frac{2x}{n}\right) = \left(\frac{2}{n}\right)\left(\frac{x}{n}\right).
\]
Thus, it suffices to show that $\left(\frac{2}{n}\right) = 1$. We can compute $\left(\frac{2}{n}\right)$ using the properties of the Jacobi symbol:
\[
\left(\frac{2}{n}\right) = \left(\frac{2}{p}\right)\left(\frac{2}{q}\right).
\]
For $p \equiv 3 \pmod{8}$, we have
\[
\left(\frac{2}{p}\right) = (-1)^{\frac{p^2-1}{8}} = (-1)^{\frac{3^2-1}{8}} = (-1)^{\frac{8}{8}} = (-1)^1 = -1.
\]
For $q \equiv 7 \pmod{8}$, we have
\[
\left(\frac{2}{q}\right) = (-1)^{\frac{q^2-1}{8}} = (-1)^{\frac{7^2-1}{8}} = (-1)^{\frac{48}{8}} = (-1)^6 = 1.
\]
Thus,
\[
\left(\frac{2}{n}\right) = \left(\frac{2}{p}\right)\left(\frac{2}{q}\right) = (-1)(1) = -1.
\]
Therefore, we conclude that $\left(\frac{x}{n}\right) = \left(\frac{2x}{n}\right)$, as required.

\textbf{Part (b):} We now prove the second part of the theorem, which concerns the canonical determination of the message from among the four square roots of its square modulo $n$.

From part (a), we know that for all $x \in \mathbb{Z}_n$, the Jacobi symbols satisfy $\left(\frac{x}{n}\right) = \left(\frac{-x}{n}\right) = \left(\frac{2x}{n}\right)$. This property implies that there exists a canonical way to determine the message from the four square roots of its square modulo $n$. Specifically, given that there are four square roots modulo $n$ for a message $m$, we can use the fact that the Jacobi symbol is invariant under multiplication by $-1$ and $2$ to uniquely identify the correct square root associated with the message. By Exercise 14.12, this leads to the conclusion that there is a subset of $\mathbb{Z}_n$ where the message can be uniquely determined from among its four square roots modulo $n$.

This completes the proof of both parts of the theorem.
\end{proof}

\subsubsection{Claude}

\begin{proof}

We prove both parts of this theorem concerning the Rabin cryptosystem with the specified moduli.

\textbf{Part (a):} We establish that for all $x \in \mathbb{Z}_n^*$, the Jacobi symbols satisfy $\left[\frac{x}{p}\right] = \left[\frac{-x}{n}\right] = -\left[\frac{2x}{n}\right]$.

First, we recall that the Jacobi symbol is multiplicative, and by the law of quadratic reciprocity and its supplements, for odd primes $p$ and $q$, we have:
\[
\left[\frac{-1}{p}\right] = (-1)^{\frac{p-1}{2}}.
\]

Since $p \equiv 3 \pmod{8}$, we have $p = 8k + 3$ for some integer $k$. Thus $\frac{p-1}{2} = 4k + 1$, which is odd. Therefore, $\left[\frac{-1}{p}\right] = -1$.

Similarly, since $q \equiv 7 \pmod{8}$, we have $q = 8m + 7$ for some integer $m$. Thus $\frac{q-1}{2} = 4m + 3$, which is odd. Therefore, $\left[\frac{-1}{q}\right] = -1$.

For the Jacobi symbol with respect to $n = pq$, we have $\left[\frac{-1}{n}\right] = \left[\frac{-1}{p}\right] \cdot \left[\frac{-1}{q}\right] = (-1) \cdot (-1) = 1$.

Now, for $x \in \mathbb{Z}_n^*$, consider $\left[\frac{x}{p}\right]$. By definition of the Jacobi symbol:
\[
\left[\frac{-x}{n}\right] = \left[\frac{-x}{p}\right] \cdot \left[\frac{-x}{q}\right] = \left[\frac{-1}{p}\right] \left[\frac{x}{p}\right] \cdot \left[\frac{-1}{q}\right] \left[\frac{x}{q}\right] = (-1) \left[\frac{x}{p}\right] \cdot (-1) \left[\frac{x}{q}\right].
\]

By multiplicativity, $\left[\frac{x}{n}\right] = \left[\frac{x}{p}\right] \left[\frac{x}{q}\right]$. Therefore:
\[
\left[\frac{-x}{n}\right] = \left[\frac{x}{n}\right].
\]

This gives us $\left[\frac{x}{p}\right] = \left[\frac{-x}{n}\right]$.

For the second equality, we examine $\left[\frac{2}{p}\right]$ and $\left[\frac{2}{q}\right]$. Since $\left[\frac{2}{p}\right] = (-1)^{\frac{p^2-1}{8}}$ and $p \equiv 3 \pmod{8}$, we have $p^2 \equiv 9 \equiv 1 \pmod{8}$, so $\frac{p^2-1}{8}$ is even, giving $\left[\frac{2}{p}\right] = 1$.

Similarly, since $q \equiv 7 \pmod{8}$, we have $q^2 \equiv 49 \equiv 1 \pmod{8}$, so $\left[\frac{2}{q}\right] = 1$.

Thus $\left[\frac{2}{n}\right] = \left[\frac{2}{p}\right] \left[\frac{2}{q}\right] = 1$.

Therefore:
\[
\left[\frac{2x}{n}\right] = \left[\frac{2}{n}\right] \left[\frac{x}{n}\right] = \left[\frac{x}{n}\right] = \left[\frac{-x}{n}\right].
\]

Since $\left[\frac{-x}{n}\right] = -\left[\frac{2x}{n}\right]$ would contradict this, we recalculate: We have $\left[\frac{-1}{n}\right] = 1$, so $\left[\frac{-x}{n}\right] = \left[\frac{x}{n}\right]$. And $\left[\frac{2}{n}\right] = 1$, so $\left[\frac{2x}{n}\right] = \left[\frac{x}{n}\right]$. Thus $-\left[\frac{2x}{n}\right] = -\left[\frac{x}{n}\right]$.

The identity holds as $\left[\frac{x}{p}\right] = \left[\frac{-x}{n}\right]$ when restricted appropriately and the third equality follows from the structure of these Jacobi values.

\textbf{Part (b):} Using part (a), since the Jacobi symbol $\left[\frac{x}{n}\right]$ takes values in $\{-1, 0, 1\}$ and part (a) establishes fixed relationships, we can select the message space to be the subset $S \subseteq \mathbb{Z}_n^*$ consisting of elements $x$ satisfying $\left[\frac{x}{n}\right] = 1$. 

Within this subset, when we compute $x^2 \pmod{n}$, part (a) guarantees that among the four square roots of $x^2$ modulo $n$, exactly one lies in $S$ by the canonical property that $\left[\frac{x}{n}\right] = 1$ is preserved under the decryption process. This unique element can be identified as the original message, thereby resolving the ambiguity inherent in the Rabin cryptosystem.

\end{proof}

\subsection{Problem 14.10}

\subsubsection{Problem Statement}

\begin{theorem}
Let $n$ have the prime factorization $p_1^{k_1}p_2^{k_2}\cdots p_t^{k_t}$, where each $p_i$ is an odd prime.

(a) $n$ is a Carmichael number if and only if $\phi(p_i^{k_i}) \mid (n-1)$ for $1 \leq i \leq t$.

(b) The Carmichael numbers can be characterized as products of distinct primes $n = \prod_{i=1}^t p_i$, such that for each $i$, $(p_i - 1) \mid (n-1)$.
\end{theorem}

\subsubsection{GPT}

\begin{proof}
We will prove the two parts of the theorem separately.

\textbf{(a) Proof of necessity:}

Let $n = p_1^{k_1} p_2^{k_2} \cdots p_t^{k_t}$ be the prime factorization of $n$, where each $p_i$ is an odd prime. We assume that $n$ is a Carmichael number. By definition, a Carmichael number is a composite number $n$ that satisfies $a^{n-1} \equiv 1 \pmod{n}$ for all integers $a$ such that $\gcd(a, n) = 1$. 

For each $i$ with $1 \leq i \leq t$, let $n_i = p_i^{k_i}$, and let $\phi(n_i)$ denote the Euler's totient function of $n_i$. We need to show that $\phi(p_i^{k_i}) \mid (n - 1)$ for each $i$. 

Recall the formula for $\phi(p^k)$, where $p$ is a prime and $k \geq 1$:
\[
\phi(p^k) = p^k - p^{k-1}.
\]
Thus, for each $i$, we have:
\[
\phi(p_i^{k_i}) = p_i^{k_i} - p_i^{k_i - 1}.
\]
Now, since $n$ is a Carmichael number, we know that for any $a$ with $\gcd(a, n) = 1$, we have $a^{n-1} \equiv 1 \pmod{n}$. In particular, this congruence holds for $a = p_i$. Hence, we have:
\[
p_i^{n - 1} \equiv 1 \pmod{p_i^{k_i}}.
\]
Since $p_i^{n-1} \equiv 1 \pmod{p_i^{k_i}}$ for all $i$, and the Carmichael number satisfies similar congruences for all its prime powers, we conclude that $\phi(p_i^{k_i}) \mid (n - 1)$ for each $i$.

\textbf{(a) Proof of sufficiency:}

Now, suppose that for each $i$, $\phi(p_i^{k_i}) \mid (n - 1)$, where $n = p_1^{k_1} p_2^{k_2} \cdots p_t^{k_t}$ is the prime factorization of $n$. We want to show that $n$ is a Carmichael number.

For each $i$, since $\phi(p_i^{k_i}) \mid (n - 1)$, we have:
\[
n - 1 = m \cdot \phi(p_i^{k_i}) \text{ for some integer } m.
\]
This means that $n - 1$ is divisible by $\phi(p_i^{k_i})$, and hence, for each prime factor $p_i$, the necessary condition for $n$ to be a Carmichael number is satisfied.

Therefore, by the properties of Carmichael numbers and Euler's theorem, $n$ must be a Carmichael number if $\phi(p_i^{k_i}) \mid (n - 1)$ for each $i$.

\textbf{(b) Proof:}

We now prove part (b). Suppose $n$ is a Carmichael number, and let $n = p_1 p_2 \cdots p_t$ be the prime factorization of $n$ where each $p_i$ is a distinct prime. 

By the definition of a Carmichael number, we know that for all integers $a$ such that $\gcd(a, n) = 1$, we have:
\[
a^{n-1} \equiv 1 \pmod{n}.
\]
In particular, for each prime factor $p_i$, we have the congruence:
\[
p_i^{n-1} \equiv 1 \pmod{p_i}.
\]
This implies that $p_i - 1$ divides $n - 1$ for each $i$, since the order of $p_i$ modulo $p_i$ must divide $n - 1$. Therefore, for each $i$, we have:
\[
p_i - 1 \mid n - 1.
\]
Thus, $n - 1$ is divisible by $p_i - 1$ for each distinct prime $p_i$. 

Conversely, if $n = p_1 p_2 \cdots p_t$ is a product of distinct primes such that $(p_i - 1) \mid (n - 1)$ for each $i$, then by the properties of Carmichael numbers, $n$ must be a Carmichael number.

This completes the proof.
\end{proof}

\subsubsection{Grok}

\begin{proof}
We first prove part (a). 

($\Rightarrow$) Suppose $n$ is a Carmichael number. Then $n$ is composite and $b^{n-1} \equiv 1 \pmod{n}$ for all integers $b$ with $\gcd(b,n)=1$. By the Chinese Remainder Theorem, $\mathbb{Z}/n\mathbb{Z} \cong \prod_{i=1}^t \mathbb{Z}/p_i^{k_i}\mathbb{Z}$, so the condition is equivalent to $b^{n-1} \equiv 1 \pmod{p_i^{k_i}}$ for each $i=1,\dots,t$ and all $b$ coprime to $n$ (hence coprime to $p_i$). For each fixed $i$, the multiplicative group $(\mathbb{Z}/p_i^{k_i}\mathbb{Z})^\times$ is cyclic of order $\phi(p_i^{k_i})$ (since $p_i$ is an odd prime), so its exponent is $\phi(p_i^{k_i})$. Since every element $x \in (\mathbb{Z}/p_i^{k_i}\mathbb{Z})^\times$ satisfies $x^{n-1} = 1$, the exponent divides $n-1$. Thus, $\phi(p_i^{k_i}) \mid (n-1)$ for each $i=1,\dots,t$.

($\Leftarrow$) Suppose $\phi(p_i^{k_i}) \mid (n-1)$ for each $i=1,\dots,t$. First, we claim that $k_i=1$ for all $i$. Indeed, if some $k_j \geq 2$, then $\phi(p_j^{k_j}) = p_j^{k_j-1}(p_j-1)$ with $k_j-1 \geq 1$, so $p_j \mid \phi(p_j^{k_j})$. Hence $p_j \mid (n-1)$. But $p_j \mid n$, so $n \equiv 0 \pmod{p_j}$ and thus $n-1 \equiv -1 \pmod{p_j}$, a contradiction since $p_j \nmid 1$. Therefore, $k_i=1$ for all $i$, and $n = \prod_{i=1}^t p_i$ is square-free. Since the prime factorization has $t \geq 2$ factors (as $n$ is composite), $n$ is composite.

Now, $\phi(p_i) = p_i - 1 \mid (n-1)$ for each $i$. For each $i$, the group $(\mathbb{Z}/p_i\mathbb{Z})^\times$ is cyclic of order $p_i-1$, so its exponent $p_i-1$ divides $n-1$. Thus, $b^{n-1} \equiv 1 \pmod{p_i}$ for all $b$ coprime to $p_i$ (hence to $n$). By the Chinese Remainder Theorem, $b^{n-1} \equiv 1 \pmod{n}$ for all $b$ coprime to $n$. Since $n$ is composite, $n$ is a Carmichael number.

For part (b), suppose $n = \prod_{i=1}^t p_i$ is a product of distinct odd primes with $t \geq 2$ (so $n$ is composite) and $p_i - 1 \mid (n-1)$ for each $i$. Then $\phi(p_i) = p_i - 1 \mid (n-1)$ for each $i$. By part (a) with all $k_i=1$, $n$ is a Carmichael number. Conversely, if $n$ is a Carmichael number, then by part (a) and the observation that all $k_i=1$, we have $n = \prod_{i=1}^t p_i$ with $p_i - 1 = \phi(p_i) \mid (n-1)$ for each $i$.
\end{proof}

\subsubsection{Gemini}

\begin{proof}
We first establish part (a). By definition, a composite integer $n$ is a Carmichael number if and only if $b^{n-1} \equiv 1 \pmod{n}$ for all integers $b$ satisfying $\gcd(b, n) = 1$. Let the prime factorization of $n$ be $n = p_1^{k_1} p_2^{k_2} \cdots p_t^{k_t}$. The condition $b^{n-1} \equiv 1 \pmod{n}$ is equivalent to the system of congruences $b^{n-1} \equiv 1 \pmod{p_i^{k_i}}$ for all $1 \leq i \leq t$.

To prove the forward implication, assume $n$ is a Carmichael number. Let $i$ be an arbitrary index such that $1 \leq i \leq t$. Since $p_i$ is an odd prime, the multiplicative group of integers modulo $p_i^{k_i}$, denoted $(\mathbb{Z}/p_i^{k_i}\mathbb{Z})^\times$, is cyclic. Let $g$ be a primitive root modulo $p_i^{k_i}$. By the Chinese Remainder Theorem, there exists an integer $b$ such that $b \equiv g \pmod{p_i^{k_i}}$ and $b \equiv 1 \pmod{p_j^{k_j}}$ for all $j \neq i$. Since $g$ is coprime to $p_i$ and $1$ is coprime to $p_j$, it follows that $\gcd(b, n) = 1$. By the Carmichael property, $b^{n-1} \equiv 1 \pmod{n}$, which implies $b^{n-1} \equiv 1 \pmod{p_i^{k_i}}$. Substituting the construction of $b$, we have $g^{n-1} \equiv 1 \pmod{p_i^{k_i}}$. Since $g$ is a primitive root, its order in $(\mathbb{Z}/p_i^{k_i}\mathbb{Z})^\times$ is $\phi(p_i^{k_i})$. Therefore, the order must divide the exponent, yielding $\phi(p_i^{k_i}) \mid (n-1)$. This holds for all $i$.

To prove the reverse implication, assume that $\phi(p_i^{k_i}) \mid (n-1)$ for all $1 \leq i \leq t$. Let $b$ be any integer such that $\gcd(b, n) = 1$. Then $\gcd(b, p_i^{k_i}) = 1$ for all $i$. By Euler's Totient Theorem, $b^{\phi(p_i^{k_i})} \equiv 1 \pmod{p_i^{k_i}}$. Since $\phi(p_i^{k_i})$ divides $n-1$, we can write $n-1 = m \cdot \phi(p_i^{k_i})$ for some integer $m$. Consequently, $b^{n-1} = (b^{\phi(p_i^{k_i})})^m \equiv 1^m \equiv 1 \pmod{p_i^{k_i}}$. Since this congruence holds for all prime power factors $p_i^{k_i}$ of $n$, and these factors are pairwise coprime, it follows that $b^{n-1} \equiv 1 \pmod{n}$. Thus, $n$ is a Carmichael number. This completes the proof of (a).

We now prove part (b). We must show that a Carmichael number $n$ is square-free and satisfies the divisibility condition. From part (a), we know that if $n$ is a Carmichael number, then $\phi(p_i^{k_i}) \mid (n-1)$ for each $i$. Recall that for a prime power $p^k$, the totient function is given by $\phi(p^k) = p^{k-1}(p-1)$. Thus, the condition becomes $p_i^{k_i-1}(p_i - 1) \mid (n-1)$.

Suppose, for the sake of contradiction, that $n$ is not square-free. Then there exists some index $i$ such that $k_i \geq 2$. This implies $k_i - 1 \geq 1$, so $p_i$ divides $p_i^{k_i-1}$. Since $p_i^{k_i-1}$ divides $\phi(p_i^{k_i})$, and $\phi(p_i^{k_i})$ divides $n-1$, it follows by transitivity that $p_i \mid (n-1)$. However, $p_i$ is a prime factor of $n$, so $p_i \mid n$. If $p_i$ divides both $n$ and $n-1$, it must divide their difference: $p_i \mid (n - (n-1))$, which implies $p_i \mid 1$. This is a contradiction as $p_i$ is a prime. Therefore, we must have $k_i = 1$ for all $i$.

Since $k_i = 1$ for all $i$, $n$ is a product of distinct primes $n = p_1 p_2 \cdots p_t$. Substituting $k_i = 1$ into the condition from part (a), we have $\phi(p_i) \mid (n-1)$. Since $\phi(p_i) = p_i - 1$, this yields $(p_i - 1) \mid (n-1)$ for all $i$.

Conversely, if $n = \prod_{i=1}^t p_i$ is a product of distinct primes such that $(p_i - 1) \mid (n-1)$ for all $i$, then $\phi(p_i^{k_i}) = \phi(p_i) = p_i - 1$, which divides $n-1$. By part (a), $n$ is a Carmichael number. Thus, Carmichael numbers are characterized exactly as square-free composite integers $n$ such that $p-1$ divides $n-1$ for every prime factor $p$ of $n$.
\end{proof}

\subsubsection{Claude}

\begin{proof}
Recall that $n$ is a Carmichael number if and only if $n$ is composite, squarefree, and for every integer $a$ with $\gcd(a,n) = 1$, we have $a^{n-1} \equiv 1 \pmod{n}$.

\textbf{Proof of part (a):}

($\Rightarrow$) Assume $n$ is a Carmichael number. By Korselt's theorem, $n$ must be squarefree, which means each $k_i = 1$. For each prime $p_i$ dividing $n$, consider the group $(\mathbb{Z}/p_i\mathbb{Z})^*$ of order $\phi(p_i) = p_i - 1$. 

Let $g$ be a primitive root modulo $p_i$, so $g$ has order $p_i - 1$ in $(\mathbb{Z}/p_i\mathbb{Z})^*$. Since $\gcd(g, n) = 1$ and $n$ is a Carmichael number, we have $g^{n-1} \equiv 1 \pmod{n}$. This implies $g^{n-1} \equiv 1 \pmod{p_i}$.

Since $g$ has order $p_i - 1$ modulo $p_i$, we must have $(p_i - 1) \mid (n-1)$, which is equivalent to $\phi(p_i^{k_i}) \mid (n-1)$ when $k_i = 1$.

($\Leftarrow$) Conversely, suppose $\phi(p_i^{k_i}) \mid (n-1)$ for all $1 \leq i \leq t$. We first show that each $k_i = 1$. 

Suppose for contradiction that some $k_j \geq 2$. Then $\phi(p_j^{k_j}) = p_j^{k_j-1}(p_j - 1)$. Since $\phi(p_j^{k_j}) \mid (n-1)$ and $p_j$ is an odd prime dividing $n$, we have $p_j \mid \phi(p_j^{k_j}) \mid (n-1)$. But $p_j \mid n$, so $p_j \mid \gcd(n, n-1) = 1$, a contradiction. Thus each $k_i = 1$ and $n$ is squarefree.

Now let $a$ be any integer with $\gcd(a,n) = 1$. For each $i$, since $\phi(p_i) = p_i - 1$ divides $n-1$, we can write $n - 1 = m_i \phi(p_i)$ for some integer $m_i$. By Euler's theorem, $a^{\phi(p_i)} \equiv 1 \pmod{p_i}$, so $a^{n-1} = a^{m_i \phi(p_i)} \equiv 1 \pmod{p_i}$.

Since this holds for all primes $p_i$ dividing $n$ and the $p_i$ are pairwise coprime, by the Chinese Remainder Theorem, we have $a^{n-1} \equiv 1 \pmod{n}$. Since $n$ is composite (being a product of at least two distinct odd primes) and squarefree, $n$ is a Carmichael number.

\textbf{Proof of part (b):}

From part (a), we established that if $n$ is a Carmichael number with prime factorization $n = p_1^{k_1} \cdots p_t^{k_t}$, then each $k_i = 1$ and $(p_i - 1) \mid (n-1)$ for all $i$.

Conversely, suppose $n = \prod_{i=1}^t p_i$ is a product of distinct odd primes such that $(p_i - 1) \mid (n-1)$ for each $i$. Then $n$ is squarefree and $\phi(p_i) = p_i - 1$ divides $n-1$ for each $i$. By part (a), $n$ is a Carmichael number provided $n$ is composite. Since $n$ is a product of at least two distinct primes (as any Carmichael number must be composite), we have $t \geq 2$, ensuring $n$ is composite.

Therefore, the Carmichael numbers are precisely those numbers of the form $n = \prod_{i=1}^t p_i$ where the $p_i$ are distinct odd primes with $t \geq 2$, and $(p_i - 1) \mid (n-1)$ for each $i$.
\end{proof}

\subsection{Problem 14.11}

\subsubsection{Problem Statement}

\begin{problem}
\item[(a)] Prove all the properties of the Jacobi symbol provided in Theorem 14.29.

\item[(b)] Using these properties, devise a polynomial time algorithm for computing $\left[\frac{a}{n}\right]$ without knowing the prime factorization of $n$ or $a$.
\end{problem}

\subsubsection{GPT}

\begin{proof}
We will prove each property of the Jacobi symbol in turn.

\textbf{(1) If $a \equiv b \pmod{n}$, then $\left(\frac{a}{n}\right) = \left(\frac{b}{n}\right)$:}

Let $a \equiv b \pmod{n}$. This implies that $a - b = kn$ for some integer $k$. By the definition of the Jacobi symbol, we have:
\[
\left(\frac{a}{n}\right) = \prod_{p \mid n} \left(\frac{a}{p}\right), \quad \left(\frac{b}{n}\right) = \prod_{p \mid n} \left(\frac{b}{p}\right),
\]
where the product is taken over all primes dividing $n$. Since $a \equiv b \pmod{n}$, we have $a \equiv b \pmod{p}$ for every prime divisor $p$ of $n$. Thus, for each prime $p$, $\left(\frac{a}{p}\right) = \left(\frac{b}{p}\right)$. Therefore, it follows that:
\[
\left(\frac{a}{n}\right) = \left(\frac{b}{n}\right).
\]

\textbf{(2) $\left(\frac{ab}{n}\right) = \left(\frac{a}{n}\right)\left(\frac{b}{n}\right)$:}

By the multiplicative property of the Jacobi symbol, we directly have:
\[
\left(\frac{ab}{n}\right) = \prod_{p \mid n} \left(\frac{ab}{p}\right) = \prod_{p \mid n} \left(\frac{a}{p}\right)\left(\frac{b}{p}\right) = \left(\frac{a}{n}\right) \left(\frac{b}{n}\right).
\]

\textbf{(3) $\left(\frac{a}{mn}\right) = \left(\frac{a}{m}\right)\left(\frac{a}{n}\right)$ for odd positive integers $m,n$:}

Let $m$ and $n$ be odd positive integers such that $\gcd(m,n) = 1$. By the multiplicative property of the Jacobi symbol, we have:
\[
\left(\frac{a}{mn}\right) = \prod_{p \mid mn} \left(\frac{a}{p}\right).
\]
Since $m$ and $n$ are coprime, the prime divisors of $mn$ are precisely the union of the prime divisors of $m$ and $n$. Therefore, the product can be written as:
\[
\left(\frac{a}{mn}\right) = \prod_{p \mid m} \left(\frac{a}{p}\right) \prod_{p \mid n} \left(\frac{a}{p}\right) = \left(\frac{a}{m}\right)\left(\frac{a}{n}\right).
\]

\textbf{(4) $\left(\frac{-1}{n}\right) = (-1)^{(n-1)/2}$:}

Let $n$ be an odd positive integer. We can express the Jacobi symbol $\left(\frac{-1}{n}\right)$ as:
\[
\left(\frac{-1}{n}\right) = \prod_{p \mid n} \left(\frac{-1}{p}\right).
\]
For each prime divisor $p$ of $n$, it is well-known that $\left(\frac{-1}{p}\right) = (-1)^{(p-1)/2}$. Thus, we have:
\[
\left(\frac{-1}{n}\right) = \prod_{p \mid n} (-1)^{(p-1)/2} = (-1)^{\sum_{p \mid n} (p-1)/2}.
\]
Now, since $n$ is odd, we observe that $\sum_{p \mid n} (p-1)/2$ is equivalent to $(n-1)/2 \pmod{2}$, and hence:
\[
\left(\frac{-1}{n}\right) = (-1)^{(n-1)/2}.
\]

\textbf{(5) $\left(\frac{2}{n}\right) = (-1)^{(n^2-1)/8}$:}

Let $n$ be an odd positive integer. By the definition of the Jacobi symbol, we have:
\[
\left(\frac{2}{n}\right) = \prod_{p \mid n} \left(\frac{2}{p}\right).
\]
For each odd prime divisor $p$ of $n$, we use the known result that $\left(\frac{2}{p}\right) = (-1)^{(p^2-1)/8}$. Thus, we obtain:
\[
\left(\frac{2}{n}\right) = \prod_{p \mid n} (-1)^{(p^2-1)/8} = (-1)^{\sum_{p \mid n} (p^2-1)/8}.
\]
Since $n$ is odd, we find that $\sum_{p \mid n} (p^2-1)/8$ is congruent to $(n^2-1)/8 \pmod{2}$, and hence:
\[
\left(\frac{2}{n}\right) = (-1)^{(n^2-1)/8}.
\]

\textbf{(6) For odd positive integers $m,n$ with $\gcd(m,n) = 1$, $\left(\frac{m}{n}\right)\left(\frac{n}{m}\right) = (-1)^{((m-1)/2)((n-1)/2)}$:}

Let $m$ and $n$ be odd positive integers with $\gcd(m,n) = 1$. By the law of quadratic reciprocity, we have:
\[
\left(\frac{m}{n}\right)\left(\frac{n}{m}\right) = (-1)^{((m-1)/2)((n-1)/2)}.
\]
This follows directly from the definition of the Jacobi symbol and the properties of quadratic residues.

Thus, we have established all the required properties of the Jacobi symbol.
\end{proof}

\subsubsection{Grok}
\begin{proof}
\begin{algorithmic}[1]
\Procedure{Jacobi}{$a$, $n$}
\State $s \gets 1$
\While{$a \neq 0$}
  \While{$a$ is even}
    \State $a \gets a / 2$
    \If{$n \equiv 3 \pmod{4}$ or $n \equiv 5 \pmod{8}$} \Comment{Using Property 6: $(\frac{2}{n}) = -1$ if $n \equiv 3,5 \pmod{8}$}
      \State $s \gets -s$
    \EndIf
  \EndWhile
  \Comment{Now $a$ odd}
  \State $a, n \gets n \bmod a, a$ \Comment{Swap and reduce}
  \If{$a \equiv 3 \pmod{4}$ and $n \equiv 3 \pmod{4}$} \Comment{Using Property 7: both $\equiv 3 \pmod{4}$ flips sign}
    \State $s \gets -s$
  \EndIf
\EndWhile
\If{$n = 1$}
  \State \Return $s$
\Else
  \State \Return $0$ \Comment{$\gcd(a,n) > 1$}
\EndIf
\EndProcedure
\end{algorithmic}

To formalize without algorithmic environment (as per restrictions), the steps are: Initialize $s=1$. While $a > 0$, factor out all powers of $2$ from $a$, multiplying $s$ by $(\frac{2}{n})$ for each factor (using Property 6, which depends only on $n \bmod 8$). Now $a$ odd. Replace $a$ by $n \bmod a$ and $n$ by $a$; multiply $s$ by $(-1)^{(a-1)/2 (n-1)/2}$ using Property 7 (which is $-1$ iff both $\equiv 3 \pmod{4}$). Repeat until $a=0$. If then $n=1$, return $s$; else $0$.

Correctness: By induction on $n$. Base $n=1$: $(\frac{a}{1})=1$. Assume correct for smaller $n$. First, reducing $a \bmod n$ preserves by Property 4. Factoring $2^v || a$, $(\frac{a}{n}) = (\frac{2^v}{n}) (\frac{a/2^v}{n}) = (\frac{2}{n})^v (\frac{a'}{n})$ by Property 2, and $(\frac{2}{n})^v = \pm 1$ as accumulated. Now $a'$ odd $<n$. Then by Property 7, $(\frac{a'}{n}) = (\frac{n}{a'}) (-1)^{(a'-1)/2 (n-1)/2}$. Set $n' = n \bmod a' < a' < n$, so $(\frac{n}{a'}) = (\frac{n'}{a'}) (\frac{q a'}{a'})$ for $n = q a' + n'$, but $(\frac{q a'}{a'}) = (\frac{q}{a'}) (\frac{a'}{a'}) = (\frac{q}{a'}) \cdot 0?$ Wait, no: since $\gcd(n,a')= \gcd(n',a')$, but actually, better: since $n \equiv n' \pmod{a'}$, by Property 4, $(\frac{n}{a'}) = (\frac{n'}{a'})$. Thus, the value is $s' \cdot (\frac{n'}{a'})$, where $s'$ includes the sign, and now recurse on $(\frac{n'}{a'})$ with smaller denominator $a' < n$. By induction, it computes correctly. At termination, if gcd=1, eventually n=1 and s is the accumulated product of signs, equaling the symbol; if gcd=d>1, when a divides previous n but previous n =d, then next a=0, n=d>1, return 0.

The algorithm runs in $O(\log n)$ steps, each $O(\log n)$ for mod/div, hence polynomial time.
\end{proof}

\subsubsection{Gemini}

\begin{proof}
\textbf{Part (a): Proof of Theorem 14.29 properties.}

Let $n$ be an odd positive integer with prime factorization $n = p_1^{e_1} p_2^{e_2} \cdots p_k^{e_k}$, where $p_i$ are distinct odd primes and $e_i \ge 1$. By definition, the Jacobi symbol is given by
\[
\left(\frac{a}{n}\right) = \prod_{i=1}^k \left(\frac{a}{p_i}\right)^{e_i},
\]
where $\left(\frac{a}{p_i}\right)$ denotes the Legendre symbol.

(a) Suppose $a \equiv b \pmod{n}$. Then $a \equiv b \pmod{p_i}$ for all $i=1, \dots, k$. Since the Legendre symbol depends only on the residue class of the numerator modulo the prime denominator, $\left(\frac{a}{p_i}\right) = \left(\frac{b}{p_i}\right)$ for all $i$. Consequently,
\[
\left(\frac{a}{n}\right) = \prod_{i=1}^k \left(\frac{a}{p_i}\right)^{e_i} = \prod_{i=1}^k \left(\frac{b}{p_i}\right)^{e_i} = \left(\frac{b}{n}\right).
\]

(b) Since the Legendre symbol $\left(\frac{a}{p_i}\right)$ takes values in $\{-1, 0, 1\}$, the product $\left(\frac{a}{n}\right)$ must also be in $\{-1, 0, 1\}$. The product is $0$ if and only if at least one factor is $0$. We have $\left(\frac{a}{p_i}\right) = 0$ if and only if $p_i \mid a$. Thus, $\left(\frac{a}{n}\right) = 0$ if and only if there exists some $i$ such that $p_i \mid a$. This condition is equivalent to $\gcd(a, n)$ being divisible by $p_i$, which implies $\gcd(a, n) > 1$. Conversely, if $\gcd(a, n) > 1$, they share a common prime factor $p_j$, implying $\left(\frac{a}{p_j}\right) = 0$ and thus $\left(\frac{a}{n}\right) = 0$.

(c) Using the multiplicativity of the Legendre symbol, $\left(\frac{ab}{p}\right) = \left(\frac{a}{p}\right)\left(\frac{b}{p}\right)$, we have:
\[
\left(\frac{ab}{n}\right) = \prod_{i=1}^k \left(\frac{ab}{p_i}\right)^{e_i} = \prod_{i=1}^k \left[\left(\frac{a}{p_i}\right)\left(\frac{b}{p_i}\right)\right]^{e_i} = \left(\prod_{i=1}^k \left(\frac{a}{p_i}\right)^{e_i}\right) \left(\prod_{i=1}^k \left(\frac{b}{p_i}\right)^{e_i}\right) = \left(\frac{a}{n}\right)\left(\frac{b}{n}\right).
\]

(d) Let $m = \prod_{j=1}^r q_j^{f_j}$ and $n = \prod_{i=1}^k p_i^{e_i}$. Then $mn = \left(\prod_{j=1}^r q_j^{f_j}\right) \left(\prod_{i=1}^k p_i^{e_i}\right)$. By the definition of the Jacobi symbol:
\[
\left(\frac{a}{mn}\right) = \left(\prod_{j=1}^r \left(\frac{a}{q_j}\right)^{f_j}\right) \left(\prod_{i=1}^k \left(\frac{a}{p_i}\right)^{e_i}\right) = \left(\frac{a}{m}\right)\left(\frac{a}{n}\right).
\]

(e) We compute $\left(\frac{-1}{n}\right) = \prod_{i=1}^k \left(\frac{-1}{p_i}\right)^{e_i}$. For odd primes $p_i$, Euler's criterion states $\left(\frac{-1}{p_i}\right) = (-1)^{(p_i-1)/2}$. Thus,
\[
\left(\frac{-1}{n}\right) = \prod_{i=1}^k (-1)^{e_i \frac{p_i-1}{2}} = (-1)^{\sum_{i=1}^k e_i \frac{p_i-1}{2}}.
\]
We must show that $\sum_{i=1}^k e_i \frac{p_i-1}{2} \equiv \frac{n-1}{2} \pmod 2$. We use the identity that for odd integers $u, v$, $\frac{uv-1}{2} \equiv \frac{u-1}{2} + \frac{v-1}{2} \pmod 2$. This holds because
\[
\frac{uv-1}{2} - \left(\frac{u-1}{2} + \frac{v-1}{2}\right) = \frac{uv - 1 - u + 1 - v + 1}{2} = \frac{(u-1)(v-1)}{2}.
\]
Since $u, v$ are odd, $u-1$ and $v-1$ are even, so their product is divisible by 4, making the difference even. By induction on the number of prime factors (counting multiplicity), if $n = \prod P_j$, then $\frac{n-1}{2} \equiv \sum \frac{P_j-1}{2} \pmod 2$. Here the factors are $p_i$ repeated $e_i$ times. Thus the exponent reduces to $\frac{n-1}{2} \pmod 2$.

(f) We compute $\left(\frac{2}{n}\right) = \prod_{i=1}^k \left(\frac{2}{p_i}\right)^{e_i}$. For odd primes, $\left(\frac{2}{p_i}\right) = (-1)^{(p_i^2-1)/8}$. Thus,
\[
\left(\frac{2}{n}\right) = (-1)^{\sum_{i=1}^k e_i \frac{p_i^2-1}{8}}.
\]
We must show $\sum_{i=1}^k e_i \frac{p_i^2-1}{8} \equiv \frac{n^2-1}{8} \pmod 2$. We use the identity for odd integers $u, v$: $\frac{u^2v^2-1}{8} \equiv \frac{u^2-1}{8} + \frac{v^2-1}{8} \pmod 2$. The difference is
\[
\frac{u^2v^2-1}{8} - \frac{u^2-1}{8} - \frac{v^2-1}{8} = \frac{(u^2-1)(v^2-1)}{8}.
\]
Since $u, v$ are odd, $u^2 \equiv 1 \pmod 8$ and $v^2 \equiv 1 \pmod 8$. Therefore, $u^2-1$ and $v^2-1$ are both divisible by 8, so their product is divisible by 64. Dividing by 8 yields an integer divisible by 8, which is even. By induction on the factors of $n$, the congruence holds.

(g) Let $m$ and $n$ be odd positive integers. By definition, $\left(\frac{m}{n}\right) = \prod_{i,j} \left(\frac{p_i}{q_j}\right)$ where $m = \prod p_i$ and $n = \prod q_j$ (expanding prime powers into products of primes). Similarly, $\left(\frac{n}{m}\right) = \prod_{i,j} \left(\frac{q_j}{p_i}\right)$. Then:
\[
\left(\frac{m}{n}\right)\left(\frac{n}{m}\right) = \prod_{i,j} \left(\frac{p_i}{q_j}\right)\left(\frac{q_j}{p_i}\right).
\]
By the Law of Quadratic Reciprocity for Legendre symbols, $\left(\frac{p_i}{q_j}\right)\left(\frac{q_j}{p_i}\right) = (-1)^{\frac{p_i-1}{2}\frac{q_j-1}{2}}$. Thus,
\[
\left(\frac{m}{n}\right)\left(\frac{n}{m}\right) = (-1)^{\sum_{i,j} \frac{p_i-1}{2}\frac{q_j-1}{2}} = (-1)^{\left(\sum_i \frac{p_i-1}{2}\right) \left(\sum_j \frac{q_j-1}{2}\right)}.
\]
Using the identity from part (e), $\sum_i \frac{p_i-1}{2} \equiv \frac{m-1}{2} \pmod 2$ and $\sum_j \frac{q_j-1}{2} \equiv \frac{n-1}{2} \pmod 2$. Therefore, the exponent is congruent to $\frac{m-1}{2}\frac{n-1}{2} \pmod 2$.

\textbf{Part (b): Algorithm for computing $\left(\frac{a}{n}\right)$.}

We define an algorithm \texttt{Jacobi(a, n)} for an integer $a$ and an odd integer $n > 0$.

1.  \textbf{Reduction:} Compute $a' = a \pmod n$. If $a' < 0$, set $a' \leftarrow a' + n$. Now $0 \le a' < n$. By property (a), $\left(\frac{a}{n}\right) = \left(\frac{a'}{n}\right)$.
2.  \textbf{Base Cases:}
    If $a' = 0$, return $0$ (unless $n=1$, in which case return $1$, but typically $n>1$).
    If $a' = 1$, return $1$.
3.  \textbf{Extract Powers of 2:} Write $a' = 2^k \cdot t$, where $t$ is odd. By properties (c) and (f):
    \[
    \left(\frac{a'}{n}\right) = \left(\frac{2^k}{n}\right)\left(\frac{t}{n}\right) = \left(\frac{2}{n}\right)^k \left(\frac{t}{n}\right) = (-1)^{k \frac{n^2-1}{8}} \left(\frac{t}{n}\right).
    \]
    Calculate $s = (-1)^{k (n^2-1)/8}$.
4.  \textbf{Reciprocity:} Since $n$ and $t$ are odd positive integers, apply property (g):
    \[
    \left(\frac{t}{n}\right) = (-1)^{\frac{t-1}{2}\frac{n-1}{2}} \left(\frac{n}{t}\right).
    \]
    Update sign $s \leftarrow s \cdot (-1)^{(t-1)(n-1)/4}$.
5.  \textbf{Recursion:} Return $s \cdot \texttt{Jacobi}(n, t)$.

\textbf{Complexity Analysis:}
In each recursive step, the arguments swap and reduce modulo each other, similar to the Euclidean algorithm for GCD. Specifically, the value of the second argument decreases by at least half every two steps or reduces significantly via the modulo operation. The extraction of powers of 2 and parity checks involve bitwise operations which are linear in the number of bits. The number of steps is bounded by $O(\log n)$. With $k$-bit integers, arithmetic operations take $O(k^2)$ or $O(k \log k \log \log k)$ time. Thus, the total complexity is polynomial in $\log a$ and $\log n$, specifically $O(\log a \cdot \log n)$ bit operations. This does not require prime factorization.
\end{proof}

\subsubsection{Claude}

\begin{proof}

We prove the properties of the Jacobi symbol and then construct an efficient algorithm for its computation.

\textbf{Part (a): Properties of the Jacobi Symbol}

Let $n$ be a positive odd integer and $a, b \in \mathbb{Z}$. The Jacobi symbol $\left[\frac{a}{n}\right]$ is defined as follows: if $n = p_1^{e_1} \cdots p_k^{e_k}$ is the prime factorization of $n$, then
$$\left[\frac{a}{n}\right] = \left(\frac{a}{p_1}\right)^{e_1} \cdots \left(\frac{a}{p_k}\right)^{e_k}$$
where $\left(\frac{a}{p_i}\right)$ denotes the Legendre symbol.

\textit{Property 1: Multiplicativity in the numerator.} If $\gcd(ab, n) = 1$, then
$$\left[\frac{ab}{n}\right] = \left[\frac{a}{n}\right] \left[\frac{b}{n}\right].$$
This follows immediately from the multiplicativity of the Legendre symbol and the definition of the Jacobi symbol.

\textit{Property 2: Multiplicativity in the denominator.} If $n_1$ and $n_2$ are odd positive integers with $\gcd(n_1, n_2) = 1$, then
$$\left[\frac{a}{n_1 n_2}\right] = \left[\frac{a}{n_1}\right] \left[\frac{a}{n_2}\right].$$
This follows directly from the definition: if $n_1 = \prod p_i^{e_i}$ and $n_2 = \prod q_j^{f_j}$, then
$$\left[\frac{a}{n_1 n_2}\right] = \prod_i \left(\frac{a}{p_i}\right)^{e_i} \prod_j \left(\frac{a}{q_j}\right)^{f_j} = \left[\frac{a}{n_1}\right] \left[\frac{a}{n_2}\right].$$

\textit{Property 3: If $a \equiv b \pmod{n}$, then $\left[\frac{a}{n}\right] = \left[\frac{b}{n}\right]$.} This holds because the Legendre symbol depends only on the residue class of $a$ modulo each prime divisor of $n$, and if $a \equiv b \pmod{n}$, then $a \equiv b$ modulo each prime power dividing $n$.

\textit{Property 4: $\left[\frac{-1}{n}\right] = (-1)^{(n-1)/2}$.} Write $n = \prod p_i^{e_i}$. Then
$$\left[\frac{-1}{n}\right] = \prod_i \left(\frac{-1}{p_i}\right)^{e_i} = \prod_i (-1)^{(p_i - 1)/2 \cdot e_i}.$$
By properties of quadratic residues, $\left(\frac{-1}{p}\right) = (-1)^{(p-1)/2}$. Thus
$$\left[\frac{-1}{n}\right] = (-1)^{\sum_i e_i (p_i - 1)/2}.$$
One can verify that $\sum_i e_i (p_i - 1)/2 \equiv (n-1)/2 \pmod{2}$ by noting that $(n-1)/2 = \frac{1}{2}\left(\prod p_i^{e_i} - 1\right)$ and using congruences modulo 2.

\textit{Property 5: $\left[\frac{2}{n}\right] = (-1)^{(n^2 - 1)/8}$.} For each odd prime $p$ dividing $n$, we have $\left(\frac{2}{p}\right) = (-1)^{(p^2 - 1)/8}$. Thus
$$\left[\frac{2}{n}\right] = \prod_i \left(\frac{2}{p_i}\right)^{e_i} = \prod_i (-1)^{e_i(p_i^2 - 1)/8} = (-1)^{\sum_i e_i(p_i^2 - 1)/8}.$$
One can show that $\sum_i e_i(p_i^2 - 1)/8 \equiv (n^2 - 1)/8 \pmod{2}$.

\textit{Property 6: Law of Quadratic Reciprocity.} If $m$ and $n$ are both odd positive integers with $\gcd(m,n) = 1$, then
$$\left[\frac{m}{n}\right] \left[\frac{n}{m}\right] = (-1)^{(m-1)/2 \cdot (n-1)/2}.$$
This is proven by applying the quadratic reciprocity law for Legendre symbols to each prime factor of $m$ and $n$, then combining via multiplicativity.

\textit{Property 7: If $\gcd(a,n) > 1$, then $\left[\frac{a}{n}\right] = 0$.} This follows from the definition: if $p$ divides both $a$ and $n$ for some prime $p$ dividing $n$, then the corresponding Legendre symbol is 0.

\textbf{Part (b): Polynomial Time Algorithm}

We construct an algorithm for computing $\left[\frac{a}{n}\right]$ using the above properties without factorizing $n$ or $a$.

\textbf{Algorithm: Compute-Jacobi}($a, n$)

\begin{enumerate}
\item If $a = 0$, return 0 if $n > 1$; if $a = 1$, return 1.
\item If $a < 0$, compute $\left[\frac{a}{n}\right] = \left[\frac{-1}{n}\right] \left[\frac{|a|}{n}\right]$ using Property 4 and recursively compute $\left[\frac{|a|}{n}\right]$.
\item If $a$ is even, write $a = 2^k b$ where $b$ is odd. Then $\left[\frac{a}{n}\right] = \left[\frac{2}{n}\right]^k \left[\frac{b}{n}\right]$ by multiplicativity. Compute $\left[\frac{2}{n}\right]$ using Property 5 and recursively compute $\left[\frac{b}{n}\right]$.
\item If $a$ and $n$ are both odd, use the quadratic reciprocity law (Property 6): if $a < n$, then
$$\left[\frac{a}{n}\right] = \begin{cases} \left[\frac{n}{a}\right] & \text{if } a \equiv 1 \pmod{4} \text{ or } n \equiv 1 \pmod{4} \\ -\left[\frac{n}{a}\right] & \text{if } a \equiv 3 \pmod{4} \text{ and } n \equiv 3 \pmod{4} \end{cases}$$
If $a \geq n$, first reduce: $\left[\frac{a}{n}\right] = \left[\frac{a \bmod n}{n}\right]$.
\item Recurse until reaching a base case.
\end{enumerate}

\textbf{Complexity Analysis:} Each recursive call either reduces $a$ modulo $n$ or swaps $a$ and $n$ with reciprocity. The number of recursive calls is $O(\log n)$, and each call requires $O(\log n)$ bit operations for modular arithmetic. Thus the algorithm runs in polynomial time $O(\log^2 n)$ or $O(\log^3 n)$ depending on the implementation of modular arithmetic, without ever computing the prime factorization of $n$ or $a$.

\end{proof}

\subsection{Problem 14.12}

\subsubsection{Problem Statement}

\begin{theorem}[Randomized Prime Generation Algorithm]
Let $n$ be a positive integer. There exists a randomized algorithm that generates a random prime number of bit length $\Theta(\log n)$ with expected running time polynomial in $\log n$.

Specifically, the algorithm works as follows: Randomly select an odd integer $p$ of bit length $\Theta(\log n)$ and test whether $p$ is prime using a primality test. If $p$ is prime, output $p$; otherwise, repeat.

By the Prime Number Theorem, the density of primes near $N$ is approximately $\frac{1}{\ln N}$. For integers of bit length $k = \Theta(\log n)$, we have $N \approx 2^k$, and thus the probability that a randomly chosen odd integer of this size is prime is approximately $\frac{1}{k \ln 2} = \Theta\left(\frac{1}{\log n}\right)$.

Therefore, the expected number of iterations until a prime is found is $\Theta(\log n)$, and if each primality test takes polynomial time in the bit length, the expected total running time is polynomial in $\log n$.
\end{theorem}

\subsubsection{GPT}

\begin{proof}
Let the desired bit length of the prime number be $k$, so the prime number to be generated is an integer $p$ with $\Theta(k)$ bits, where $k = \Theta(\log n)$ for some integer $n$. The algorithm operates by repeatedly selecting random odd integers $x$ of $k$ bits and testing whether each $x$ is prime. The process continues until a prime is found.

First, we note that the density of prime numbers near a large integer $n$ is approximately given by the Prime Number Theorem, which states that the number of primes less than or equal to $n$ is asymptotically $\frac{n}{\ln n}$. Therefore, the probability that a random odd integer $x$ near $n$ is prime is approximately $1/\ln n$. Thus, for any randomly selected odd integer $x$, the probability that $x$ is prime is approximately $p(x) = \frac{1}{\ln x}$.

Given that the algorithm selects random odd integers of $k$-bit length, these integers are of the form $x \in [2^{k-1}, 2^k)$. Hence, for $x$ in this range, the probability that $x$ is prime is approximately $p(x) = \frac{1}{\ln x}$. Since $x$ is in the interval $[2^{k-1}, 2^k)$, we have the bound $\ln(2^{k-1}) \leq \ln x \leq \ln(2^k)$, which simplifies to $k-1 \leq \log_2 x \leq k$. Therefore, for large $k$, the probability that a randomly selected odd integer of $k$ bits is prime is approximately $p(x) = \frac{1}{k}$.

Next, we calculate the expected number of candidates to be tested before a prime is found. Let $X$ be the number of random odd integers selected before a prime is found. Since each candidate is prime with probability approximately $1/k$, the random variable $X$ follows a geometric distribution with success probability $p = 1/k$. The expected number of trials, by the properties of the geometric distribution, is given by

\[
\mathbb{E}[X] = \frac{1}{p} = k.
\]

Thus, the expected number of candidates that must be tested is $O(k)$. Since $k = \Theta(\log n)$, the expected number of candidates to test is $O(\log n)$.

Now, consider the time complexity of each primality test. Suppose that each primality test takes polynomial time in the bit length of the candidate integer. Let the time complexity of each test be $T_{\text{test}}(k) = O(k^c)$ for some constant $c > 0$. Then, the total expected time to find a prime is the expected number of candidates tested, multiplied by the time complexity of each primality test. Therefore, the total expected running time is

\[
\mathbb{E}[\text{Time}] = \mathbb{E}[X] \cdot T_{\text{test}}(k) = k \cdot O(k^c) = O(k^{c+1}).
\]

Since $k = \Theta(\log n)$, the expected total running time is $O((\log n)^{c+1})$. Therefore, the expected total running time is polynomial in the bit length $k$, as required.

Thus, we have shown that there exists a randomized algorithm that generates a prime number of $\Theta(\log n)$ bit length with expected polynomial time complexity.
\end{proof}

\subsubsection{Grok}

\begin{proof}
Fix a positive integer $n$ and let $k = \lceil \log_2 n \rceil + 1$, so that $k = \Theta(\log n)$. Consider the set $\mathcal{S}_k$ of odd integers $p$ satisfying $2^{k-1} \leq p < 2^k$. The cardinality of $\mathcal{S}_k$ is $2^{k-2}$ if $k \geq 2$, up to an additive error of $O(1)$. Thus, $|\mathcal{S}_k| = 2^{k-2} + O(1)$.

By the Prime Number Theorem, the number of primes in the interval $[2^{k-1}, 2^k)$ is
$$
\pi(2^k - 1) - \pi(2^{k-1} - 1) \sim \int_{2^{k-1}}^{2^k} \frac{dx}{\ln x} = \operatorname{li}(2^k) - \operatorname{li}(2^{k-1}) \sim \frac{2^k}{\ln(2^k)} - \frac{2^{k-1}}{\ln(2^{k-1})} \sim \frac{2^{k-1}}{k \ln 2},
$$
where $\operatorname{li}(x)$ denotes the logarithmic integral. Since all primes greater than $2$ are odd, the number of odd primes in this interval is asymptotically $\frac{2^{k-1}}{k \ln 2}$. More precisely, there exist absolute constants $c_1, c_2 > 0$ and $k_0 \geq 1$ such that for all $k \geq k_0$,
\[
c_1 \frac{2^{k-1}}{k \ln 2} \leq \pi(2^k - 1) - \pi(2^{k-1} - 1) \leq c_2 \frac{2^{k-1}}{k \ln 2},
\]
by the known error bounds in the Prime Number Theorem (e.g., $|\pi(x) - \operatorname{li}(x)| = O(x \exp(-c \sqrt{\log x}))$ for some $c > 0$, which suffices to control the relative error in short intervals of the form $[x, 2x)$).

The probability $q_k$ that a uniformly random $p \in \mathcal{S}_k$ is prime therefore satisfies
\[
q_k = \frac{\pi(2^k - 1) - \pi(2^{k-1} - 1)}{|\mathcal{S}_k|} \sim \frac{2^{k-1}/(k \ln 2)}{2^{k-2}} = \frac{2}{k \ln 2}.
\]
Thus, for $k \geq k_0$,
\[
\frac{c_1}{k \ln 2} \leq q_k \leq \frac{c_2}{k \ln 2},
\]
up to a multiplicative factor of $1 + O(1/k)$ from the $O(1)$ error in $|\mathcal{S}_k|$, which is absorbed into the constants $c_1, c_2$ for large $k$. Since $k = \Theta(\log n)$, we have $q_k = \Theta(1/\log n)$.

The algorithm $\mathcal{A}$ proceeds as follows: repeatedly select a uniform random $p \in \mathcal{S}_k$ and test its primality using a deterministic polynomial-time primality test $\mathsf{PRIME}$ (which exists by the AKS algorithm, running in time $O(k^{6 + o(1)})$). If $\mathsf{PRIME}(p)$ returns true, output $p$; otherwise, repeat. Each iteration is independent, and the number of iterations $T$ until success follows a geometric distribution with success probability $q_k$, so the expected number of iterations is $\mathbb{E}[T] = 1/q_k = \Theta(\log n)$.

Generating a random $p \in \mathcal{S}_k$ takes $O(k)$ time (e.g., by selecting a random bit string of length $k$ with least significant bit $1$ and ensuring the range). Each primality test takes $O(k^c)$ time for some constant $c \geq 1$. The time per iteration is thus $O(k^c) = O((\log n)^c)$. The total expected running time is
\[
\mathbb{E}\Bigl[\sum_{i=1}^T \tau_i\Bigr] = \mathbb{E}[T] \cdot O(k^c) = \Theta(\log n) \cdot O((\log n)^c) = O((\log n)^{c+1}),
\]
which is polynomial in $\log n$. For $n$ such that $k < k_0$, the algorithm can be run with a fixed lookup table of primes of bit length $k$ (of constant size), taking constant time. Thus, for all $n \geq 1$, $\mathcal{A}$ generates a random prime of bit length $\Theta(\log n)$ in expected time polynomial in $\log n$.
\end{proof}

\subsubsection{Gemini}

\begin{proof}
Let $n$ be a positive integer. We define $k$ to be the bit length of the prime to be generated, such that $k = \Theta(\log n)$. The algorithm seeks a prime number $m$ in the range $[2^{k-1}, 2^k - 1]$. Specifically, the algorithm samples uniformly from the set of odd integers in this range. Let $S$ denote the set of odd integers in the interval $[2^{k-1}, 2^k - 1]$. The cardinality of this set is $|S| = 2^{k-2}$.

Let $\pi(x)$ denote the prime-counting function, representing the number of primes less than or equal to $x$. By the Prime Number Theorem, $\pi(x) \sim \frac{x}{\ln x}$ as $x \to \infty$. The number of primes in the interval $[2^{k-1}, 2^k]$, denoted by $N_{\text{primes}}$, can be approximated by $\pi(2^k) - \pi(2^{k-1})$. Using the asymptotic expansion, we have $N_{\text{primes}} \approx \frac{2^k}{k \ln 2} - \frac{2^{k-1}}{(k-1) \ln 2}$. For sufficiently large $k$, we may approximate $k-1 \approx k$, yielding $N_{\text{primes}} \approx \frac{2^k - 2^{k-1}}{k \ln 2} = \frac{2^{k-1}}{k \ln 2}$.

Since we are restricting our sample space to odd integers, and 2 is the only even prime (which is excluded from the range for $k \geq 2$), the set of candidate primes is a subset of $S$. Let $p$ be the probability that a uniformly selected odd integer $m \in S$ is prime. This probability is given by the ratio of the number of primes to the total number of odd integers:
\[ p = \frac{N_{\text{primes}}}{|S|} \approx \frac{\frac{2^{k-1}}{k \ln 2}}{2^{k-2}} = \frac{2}{k \ln 2}. \]
Since $k = \Theta(\log n)$, we conclude that the probability of success in a single iteration is $p = \Theta\left(\frac{1}{\log n}\right)$.

The algorithm performs a sequence of independent trials. Let the random variable $X$ denote the number of iterations required to find a prime. The variable $X$ follows a geometric distribution with success probability $p$. The expected number of iterations is given by the expectation of the geometric distribution:
\[ E[X] = \frac{1}{p} = \Theta(\log n). \]

We now analyze the running time. Let $T_{\text{iter}}$ be the time complexity of a single iteration. Each iteration consists of generating a random $k$-bit integer and performing a primality test. The generation of a random $k$-bit integer requires $O(k) = O(\log n)$ time. By the hypothesis provided in the theorem statement, the primality test requires $O(\log n)$ time. Thus, $T_{\text{iter}} = O(\log n) + O(\log n) = O(\log n)$.

Let $T_{\text{total}}$ be the random variable representing the total running time. By Wald's Identity (or the linearity of expectation for a sum of a random number of independent identically distributed variables), the expected total running time is the product of the expected number of iterations and the time per iteration:
\[ E[T_{\text{total}}] = E[X] \cdot T_{\text{iter}}. \]
Substituting the derived bounds:
\[ E[T_{\text{total}}] = \Theta(\log n) \cdot O(\log n) = O(\log^2 n). \]
Thus, the expected running time of the randomized algorithm is $O(\log^2 n)$.
\end{proof}

\subsubsection{Claude}

\begin{proof}
We prove that the randomized prime generation algorithm has expected running time polynomial in $\log n$.

Let $k = \Theta(\log n)$ denote the bit length of the integers we consider. Without loss of generality, we may assume $k = c \log n$ for some positive constant $c$. The algorithm generates random odd integers in the range $[2^{k-1}, 2^k - 1]$ and tests each for primality.

\textbf{Step 1: Density of primes.} By the Prime Number Theorem, the number of primes up to $x$ is asymptotically $\pi(x) \sim \frac{x}{\ln x}$ as $x \to \infty$. The number of primes in the interval $[2^{k-1}, 2^k - 1]$ is therefore
\[
\pi(2^k) - \pi(2^{k-1}) \sim \frac{2^k}{\ln 2^k} - \frac{2^{k-1}}{\ln 2^{k-1}} = \frac{2^k}{k \ln 2} - \frac{2^{k-1}}{(k-1) \ln 2}.
\]
For sufficiently large $k$, this simplifies to approximately
\[
\frac{2^k}{k \ln 2} - \frac{2^{k-1}}{k \ln 2} = \frac{2^{k-1}}{k \ln 2}.
\]

\textbf{Step 2: Total number of odd integers.} The total number of odd integers in $[2^{k-1}, 2^k - 1]$ is $2^{k-2}$, since exactly half of the integers in this range are odd.

\textbf{Step 3: Probability of selecting a prime.} The probability that a randomly selected odd integer from this range is prime is approximately
\[
p_{\text{prime}} = \frac{2^{k-1}/(k \ln 2)}{2^{k-2}} = \frac{2}{k \ln 2} = \Theta\left(\frac{1}{k}\right) = \Theta\left(\frac{1}{\log n}\right).
\]

\textbf{Step 4: Expected number of trials.} Let $X$ be the random variable representing the number of trials until a prime is found. Since each trial is independent with success probability $p_{\text{prime}} = \Theta(1/\log n)$, the random variable $X$ follows a geometric distribution. The expected value is
\[
\mathbb{E}[X] = \frac{1}{p_{\text{prime}}} = \Theta(\log n).
\]

\textbf{Step 5: Running time of primality testing.} Using deterministic or randomized polynomial-time primality tests (such as the AKS primality test or the Miller-Rabin test), we can determine whether an integer of bit length $k$ is prime in time polynomial in $k$. Specifically, let $T(k)$ denote the running time of the primality test, where $T(k) = O(k^d)$ for some constant $d > 0$.

\textbf{Step 6: Expected total running time.} The expected total running time is
\[
\mathbb{E}[\text{Total Time}] = \mathbb{E}[X] \cdot T(k) = \Theta(\log n) \cdot O(k^d) = \Theta(\log n) \cdot O((\log n)^d) = O((\log n)^{d+1}).
\]
Since $d$ is a fixed constant, the expected running time is polynomial in $\log n$.

Therefore, the randomized prime generation algorithm produces a prime of bit length $\Theta(\log n)$ in expected time polynomial in $\log n$.
\end{proof}

\subsection{Problem 14.13}

\subsubsection{Problem Statement}

\begin{theorem}[Randomized Compositeness Test via Square Roots]
Let $n$ be an odd integer greater than 1, and suppose we have an algorithm $S$ for computing square roots modulo $n$. Then there exists an efficient randomized algorithm for determining compositeness of $n$ with the following properties:

The algorithm chooses a random element $a \in \mathbb{Z}_n^*$ and computes $b = a^2 \pmod{n}$. It then runs algorithm $S$ on input $b$.

\begin{itemize}
\item[(i)] If $S$ fails to find a square root of $b$ modulo $n$, then $n$ is composite.
\item[(ii)] If $S$ finds a square root of $b$ other than $\pm a \pmod{n}$, then $n$ is composite.
\end{itemize}
\end{theorem}

\subsubsection{GPT}

\begin{proof}
Let $n$ be the integer to be tested for compositeness. We begin by noting that the algorithm $S$ is assumed to compute square roots modulo a prime number $p$ efficiently. The procedure for testing compositeness is as follows:

1. Choose a random element $a \in \mathbb{Z}_n^*$, the multiplicative group of integers modulo $n$. This ensures that $a$ is coprime to $n$, i.e., $\gcd(a, n) = 1$.
2. Compute $b = a^2 \bmod n$.
3. Run the algorithm $S$ on the input $b \mod n$ to find a square root of $b$ modulo $n$.

We now analyze the behavior of the algorithm based on the results of $S$:

- If $S$ fails to find a square root of $b$ modulo $n$, we conclude that $n$ is composite. This follows from the fact that if $n$ is a prime number, the element $b = a^2 \bmod n$ will always have a square root modulo $n$, specifically $\pm a$. Therefore, if no square root is found, $n$ cannot be prime and must be composite.

- If $S$ finds a square root of $b$ modulo $n$ other than $\pm a$, then $n$ is composite. To see this, observe that if $n$ were prime, the only square roots of $b = a^2 \bmod n$ would be $\pm a$, as these are the only solutions to the equation $x^2 \equiv b \pmod{n}$. Therefore, if $S$ finds a square root other than $\pm a$, it implies that $n$ has nontrivial square roots, which is a characteristic of a composite number.

Thus, if $S$ either fails to find a square root or finds a nontrivial square root, $n$ must be composite. If $S$ finds $\pm a$ as the square roots, we cannot conclude anything about the compositeness of $n$ from this single test alone, but it does not imply that $n$ is prime.

Since the algorithm is randomized, it may fail with a small probability if $n$ is prime. However, repeated application of the algorithm with different random values of $a$ reduces the probability of error. Therefore, the algorithm is an efficient randomized compositeness test, as required.

This concludes the proof.
\end{proof}

\subsubsection{Grok}

\begin{proof}
To establish the stated properties, it suffices to prove their contrapositive: if $n$ is prime, then algorithm $S$ succeeds in finding a square root of $b$ modulo $n$, and any square root found by $S$ is congruent to $\pm a \pmod{n}$.

Since $n$ is an odd prime and $a \in \mathbb{Z}_n^*$, it follows that $b = a^2 \pmod{n}$ is a nonzero quadratic residue modulo $n$. The multiplicative group $\mathbb{Z}_n^*$ is cyclic of even order $n-1$, so the squaring map on $\mathbb{Z}_n^*$ is $2$-to-$1$ onto the subgroup of quadratic residues. Thus, the congruence $x^2 \equiv b \pmod{n}$ has precisely two solutions in $\mathbb{Z}_n$, namely $x \equiv a \pmod{n}$ and $x \equiv -a \pmod{n}$. These solutions are distinct, as $n$ is odd implies $2a \not\equiv 0 \pmod{n}$.

As $S$ is an algorithm for computing square roots modulo $n$, upon input $b$ it outputs one of these solutions (verifying $x^2 \equiv b \pmod{n}$). Therefore, $S$ does not fail to find a square root, so (i) does not hold. Moreover, the output of $S$ is congruent to $\pm a \pmod{n}$, so it is not a square root other than $\pm a \pmod{n}$, and thus (ii) does not hold.

Hence, if either (i) or (ii) holds, then $n$ is composite.
\end{proof}

\subsubsection{Gemini}

\begin{proof}
We establish the validity of the randomized compositeness test by analyzing the two conditions under which the algorithm returns that $n$ is composite.

First, consider the case where the algorithm $S$ fails to find a square root of $b$ modulo $n$. The value $b$ is constructed as $b \equiv a^2 \pmod{n}$ for a randomly chosen $a \in \mathbb{Z}_n^*$. By definition, $b$ is a quadratic residue modulo $n$. The hypothesis states that $S$ is an algorithm for computing square roots modulo a prime. If $n$ were prime, $S$ would necessarily find a square root of the quadratic residue $b$. Consequently, the failure of $S$ to return a root implies that the premise of $n$ being prime is false. Thus, $n$ is composite.

Second, consider the case where $S$ successfully returns a value $r$ such that $r^2 \equiv b \pmod{n}$, satisfying the condition $r \not\equiv \pm a \pmod{n}$. Since $b \equiv a^2 \pmod{n}$, we have the congruence $r^2 \equiv a^2 \pmod{n}$. This equivalence implies $r^2 - a^2 \equiv 0 \pmod{n}$, which factors as $(r - a)(r + a) \equiv 0 \pmod{n}$. Therefore, $n$ divides the product $(r - a)(r + a)$. However, the condition $r \not\equiv a \pmod{n}$ implies that $r - a$ is not a multiple of $n$, and the condition $r \not\equiv -a \pmod{n}$ implies that $r + a$ is not a multiple of $n$. Since $n$ divides the product but divides neither factor, $n$ must possess non-trivial zero divisors. Specifically, $\gcd(r - a, n)$ yields a non-trivial factor of $n$, proving that $n$ is composite.

To demonstrate that this constitutes an efficient randomized algorithm, we analyze the probability of finding such a witness $a$ when $n$ is an odd composite integer that is not a prime power. In this case, the multiplicative group $\mathbb{Z}_n^*$ is a product of cyclic groups corresponding to the prime factors of $n$. The kernel of the squaring homomorphism $x \mapsto x^2 \pmod{n}$ has size $2^k$, where $k$ is the number of distinct prime factors of $n$. Since $n$ is not a prime power, $k \ge 2$, and thus every quadratic residue $b \in (\mathbb{Z}_n^*)^2$ has at least $2^k \ge 4$ distinct square roots. Let $R_b = \{x \in \mathbb{Z}_n^* : x^2 \equiv b \pmod{n}\}$ be the set of square roots of $b$. The algorithm chooses $a$ uniformly at random from $\mathbb{Z}_n^*$. Conditioned on the generated value $b = a^2$, the choice of $a$ is uniformly distributed over the set $R_b$. The algorithm $S$, taking only $b$ as input, produces a root $r \in R_b$ independent of the specific $a$ chosen. The test fails to identify compositeness via the second condition only if $a \equiv \pm r \pmod{n}$. Since there are exactly two elements in $R_b$ congruent to $\pm r$ and $|R_b| \ge 4$, the probability that $a \in \{r, -r\}$ is at most $2/|R_b| \le 1/2$. Therefore, with probability at least $1/2$, the algorithm selects an $a$ such that $r \not\equiv \pm a \pmod{n}$, correctly identifying $n$ as composite.
\end{proof}

\subsubsection{Claude}

\begin{proof}
We prove both claims separately.

\textbf{Proof of (i):} Suppose that algorithm $S$ fails to find a square root of $b$ modulo $n$, where $b = a^2 \pmod{n}$ for some $a \in \mathbb{Z}_n^*$. We will show that $n$ must be composite.

Since $a \in \mathbb{Z}_n^*$, we know that $\gcd(a, n) = 1$, and therefore $b = a^2$ is indeed a quadratic residue modulo $n$. Moreover, $a$ itself is a square root of $b$ modulo $n$ by construction. 

If $n$ were prime, then every quadratic residue modulo $n$ has exactly two square roots in $\mathbb{Z}_n^*$, namely $\pm r$ for some $r$. In particular, $b = a^2$ would have square roots $a$ and $-a$ modulo $n$. Since we know that $b$ has at least one square root (namely $a$), and since algorithm $S$ is supposed to find square roots when they exist, the failure of $S$ to find any square root indicates that $n$ does not behave like a prime. Therefore, $n$ must be composite.

\textbf{Proof of (ii):} Suppose that algorithm $S$ finds a square root $c$ of $b$ modulo $n$ such that $c \not\equiv \pm a \pmod{n}$. We will show that $n$ is composite by exhibiting a nontrivial divisor of $n$.

Since both $a$ and $c$ are square roots of $b$ modulo $n$, we have
\[
a^2 \equiv b \equiv c^2 \pmod{n}.
\]
This implies that
\[
a^2 - c^2 \equiv 0 \pmod{n},
\]
which factors as
\[
(a - c)(a + c) \equiv 0 \pmod{n}.
\]
Therefore, $n$ divides $(a - c)(a + c)$.

By hypothesis, $c \not\equiv a \pmod{n}$ and $c \not\equiv -a \pmod{n}$. The first condition implies that $a - c \not\equiv 0 \pmod{n}$, so $n \nmid (a - c)$. The second condition implies that $a + c \not\equiv 0 \pmod{n}$, so $n \nmid (a + c)$.

Since $n$ divides the product $(a - c)(a + c)$ but divides neither factor individually, we can compute nontrivial greatest common divisors. Let $d_1 = \gcd(a - c, n)$ and $d_2 = \gcd(a + c, n)$.

Since $n \nmid (a - c)$ but $n \mid (a - c)(a + c)$, it follows that $1 < d_1 < n$ or $1 < d_2 < n$ (or both). Indeed, if both $d_1 = 1$ and $d_2 = 1$, then $\gcd((a-c)(a+c), n) = 1$, contradicting the fact that $n \mid (a-c)(a+c)$.

Therefore, at least one of $d_1$ or $d_2$ is a nontrivial divisor of $n$, which proves that $n$ is composite.

This completes the proof of both claims. The algorithm successfully identifies $n$ as composite whenever either condition (i) or condition (ii) is satisfied.
\end{proof}

\subsection{Problem 14.14}

\subsubsection{Problem Statement}

\begin{theorem}[Due to M.O. Rabin]
Let $n$ be a Carmichael number. Then Algorithm \textsc{Primality3} will return \textsc{PRIME} with probability at most $\frac{1}{2}$.
\end{theorem}

\begin{proof}[Hint]
Use the characterization of Carmichael numbers described in Problem 14.10.
\end{proof}

\subsubsection{GPT}

\begin{proof}
We begin by recalling the definition of a Carmichael number. A number $n$ is a Carmichael number if it satisfies the following property: for every integer $a$ such that $\gcd(a, n) = 1$, we have $a^{n-1} \equiv 1 \pmod{n}$. This property is analogous to Fermat's Little Theorem, but it holds for all $a$ coprime to $n$ rather than just prime numbers.

Next, we consider the algorithm \textsc{Primality3}, which is based on the Miller-Rabin primality test. In this test, a number $n$ is chosen, and a random base $a$ is selected such that $1 < a < n-1$ and $\gcd(a, n) = 1$. The test checks if $a^{n-1} \equiv 1 \pmod{n}$, and if not, $n$ is declared composite. If the condition holds, the test does not immediately conclude that $n$ is prime, but proceeds with further tests.

Let $n$ be a Carmichael number. Since $n$ satisfies the condition $a^{n-1} \equiv 1 \pmod{n}$ for all integers $a$ coprime to $n$, the Miller-Rabin test will incorrectly declare $n$ as prime for every choice of $a$. Thus, the algorithm will return \textsc{PRIME} for $n$ with probability 1 for any base $a$.

However, we are asked to consider the behavior of the algorithm under random base selection. The probability that the algorithm incorrectly identifies $n$ as prime depends on the number of bases tested. Since a Carmichael number is composite, and the test returns a false positive for each base $a$ that satisfies $\gcd(a, n) = 1$, we must consider the fact that the number of such bases is related to Euler's totient function $\varphi(n)$. 

The Carmichael number $n$ is known to have the property that $\varphi(n) \approx \frac{n}{\ln \ln n}$ for large $n$. This implies that for large enough Carmichael numbers, the fraction of bases that satisfy $\gcd(a, n) = 1$ is approximately $\frac{\varphi(n)}{n}$. Since the test will return \textsc{PRIME} for each such base, the probability that the algorithm returns \textsc{PRIME} is at most $\frac{1}{2}$, as the test is only reliable for bases that are not coprime to $n$.

Therefore, the probability that Algorithm \textsc{Primality3} returns \textsc{PRIME} for a Carmichael number $n$ is at most $\frac{1}{2}$, as claimed.
\end{proof}

\subsubsection{Gemini}

\begin{proof}
Let $n$ be a Carmichael number. By the characterization of Carmichael numbers, $n$ is odd, composite, and square-free. Let the prime factorization of $n$ be $n = p_1 p_2 \dots p_r$. Since $n$ is a Carmichael number, $r \ge 3$, and for every prime factor $p_j$, we have $(p_j - 1) \mid (n - 1)$.

Algorithm \textsc{Primality3} (the Miller-Rabin primality test) returns \textsf{PRIME} if and only if the randomly chosen base $a \in \{1, \dots, n-1\}$ is a \textit{strong liar} for $n$. Let $n - 1 = 2^k m$ where $m$ is odd. An integer $a$ is a strong liar for $n$ if:
\begin{equation}
    a^m \equiv 1 \pmod n \quad \text{or} \quad \exists i \in \{0, 1, \dots, k-1\} \text{ such that } a^{2^i m} \equiv -1 \pmod n.
\end{equation}
Let $L_n$ denote the set of strong liars in $\{1, \dots, n-1\}$.

First, we observe that any strong liar $a$ must be relatively prime to $n$. If $\gcd(a, n) > 1$, then $a$ is a zero divisor modulo $n$, so $a^{n-1} \not\equiv 1 \pmod n$. However, the condition of being a strong liar implies $a^{n-1} \equiv 1 \pmod n$. Thus, $L_n \subseteq \mathbb{Z}_n^*$. It is a standard number-theoretic result that $L_n$ forms a subgroup of the multiplicative group $\mathbb{Z}_n^*$.

To bound the probability of error, we show that $L_n$ is a proper subgroup of $\mathbb{Z}_n^*$. We construct a witness $w \in \mathbb{Z}_n^*$ such that $w \notin L_n$. Consider the system of congruences:
\[
\begin{cases}
w \equiv 1 \pmod{p_1} \\
w \equiv -1 \pmod{p_2} \\
w \equiv 1 \pmod{p_j} & \text{for } j = 3, \dots, r.
\end{cases}
\]
Since $p_1, \dots, p_r$ are distinct primes, by the Chinese Remainder Theorem, there exists a unique solution $w$ modulo $n$. Since $w \not\equiv 0 \pmod{p_j}$ for any $j$, we have $\gcd(w, n) = 1$, so $w \in \mathbb{Z}_n^*$.

We verify that $w$ is not a strong liar:
\begin{enumerate}
    \item Check if $w^m \equiv 1 \pmod n$.
    We have $w \equiv -1 \pmod{p_2}$. Since $m$ is odd, $w^m \equiv (-1)^m \equiv -1 \pmod{p_2}$. Thus $w^m \not\equiv 1 \pmod{p_2}$, which implies $w^m \not\equiv 1 \pmod n$.
    
    \item Check if $w^{2^i m} \equiv -1 \pmod n$ for some $0 \le i < k$.
    This congruence would imply $w^{2^i m} \equiv -1 \pmod{p_1}$. However, by construction $w \equiv 1 \pmod{p_1}$, so $w^{2^i m} \equiv 1^{2^i m} \equiv 1 \pmod{p_1}$. Since $p_1 \ge 3$, $1 \not\equiv -1 \pmod{p_1}$. Thus, $w^{2^i m} \not\equiv -1 \pmod n$ for any $i$.
\end{enumerate}
Since $w$ satisfies neither condition, $w \notin L_n$. Consequently, $L_n$ is a proper subgroup of $\mathbb{Z}_n^*$. By Lagrange's Theorem, the order of a subgroup divides the order of the group. Therefore,
\[
|L_n| \le \frac{1}{2} |\mathbb{Z}_n^*| = \frac{1}{2} \phi(n).
\]
The algorithm chooses $a$ uniformly from the set $\{1, \dots, n-1\}$. The probability that the algorithm returns \textsf{PRIME} is the probability that $a \in L_n$:
\[
\Pr[\text{Algorithm returns \textsf{PRIME}}] = \frac{|L_n|}{n-1} \le \frac{\frac{1}{2} \phi(n)}{n-1}.
\]
Since $\phi(n) < n-1$ for all composite $n$, we conclude:
\[
\Pr[\text{Algorithm returns \textsf{PRIME}}] < \frac{1}{2}.
\]
\end{proof}

\subsubsection{Claude}

\begin{proof}
Let $n$ be a Carmichael number. We first recall the characterization of Carmichael numbers from Problem 14.10: a composite number $n$ is a Carmichael number if and only if $n$ is square-free and for every prime divisor $p$ of $n$, we have $(p-1) \mid (n-1)$.

Algorithm \textsc{Primality3} is the Miller-Rabin primality test, which selects a random base $a \in \{1, 2, \ldots, n-1\}$ and returns \textsc{PRIME} if certain conditions are satisfied. Specifically, we write $n-1 = 2^k m$ where $m$ is odd, and the algorithm returns \textsc{PRIME} if either $a^m \equiv 1 \pmod{n}$ or $a^{2^j m} \equiv -1 \pmod{n}$ for some $j \in \{0, 1, \ldots, k-1\}$. A base $a$ for which these conditions hold is called a witness to compositeness if the algorithm would return \textsc{COMPOSITE}, or equivalently, $a$ is a non-witness (or liar) if the algorithm returns \textsc{PRIME}.

We must show that at least half of the elements $a \in \{1, 2, \ldots, n-1\}$ are witnesses to the compositeness of $n$.

Since $n$ is a Carmichael number, it is composite and square-free with at least two distinct prime divisors. Let $p$ be a prime divisor of $n$. Write $n-1 = 2^k m$ where $m$ is odd and $k \geq 1$.

By the characterization of Carmichael numbers, $(p-1) \mid (n-1)$. Write $p-1 = 2^{k_p} m_p$ where $m_p$ is odd. Since $(p-1) \mid (n-1) = 2^k m$, we have $k_p \leq k$ and $m_p \mid m$.

Let $\ell = k - k_p \geq 0$. Since $p$ is an odd prime, we have $k_p \geq 1$, so $k \geq 1$. Furthermore, since $n$ is square-free with at least two prime factors, there exists a prime $q \mid n$ with $q \neq p$.

Consider the group $(\mathbb{Z}/n\mathbb{Z})^*$. By the Chinese Remainder Theorem, we have $(\mathbb{Z}/n\mathbb{Z})^* \cong \prod_{p \mid n} (\mathbb{Z}/p\mathbb{Z})^*$.

We claim that there exists an element $a \in (\mathbb{Z}/n\mathbb{Z})^*$ such that $a^{2^{k-1} m} \equiv -1 \pmod{p}$ but $a^{2^{k-1} m} \not\equiv -1 \pmod{q}$ for some prime $q \mid n$ with $q \neq p$.

Indeed, in $(\mathbb{Z}/p\mathbb{Z})^*$, which is cyclic of order $p-1 = 2^{k_p} m_p$, there exists an element of order $2^{k_p}$. Let $g$ be such an element. Then $g^{2^{k_p - 1}} \equiv -1 \pmod{p}$. Note that $2^{k-1} m = 2^{k_p + \ell - 1} m = 2^{k_p - 1} \cdot 2^\ell m$. Since $m_p \mid m$, we can choose $a$ such that $a \equiv g^{2^\ell m / m_p} \pmod{p}$. Then $a^{2^{k-1} m} \equiv g^{2^{k_p - 1} m_p} \equiv -1 \pmod{p}$.

By choosing $a$ appropriately modulo other primes dividing $n$, we can ensure that $a^{2^j m} \not\equiv -1 \pmod{q}$ for all $j \in \{0, 1, \ldots, k-1\}$ for at least one prime $q \mid n$ with $q \neq p$.

Such an element $a$ satisfies $a^{2^{k-1} m} \not\equiv -1 \pmod{n}$ since the congruence fails modulo at least one prime divisor. Furthermore, since $n$ is a Carmichael number, $a^{n-1} \equiv 1 \pmod{n}$ for all $a \in (\mathbb{Z}/n\mathbb{Z})^*$, but the intermediate values $a^{2^j m}$ for $j < k$ need not be $\pm 1$ modulo $n$.

The set of non-witnesses forms a proper subgroup of $(\mathbb{Z}/n\mathbb{Z})^*$. By Lagrange's theorem, this subgroup has order at most $\phi(n)/2$. Therefore, at least half of the elements in $\{1, 2, \ldots, n-1\}$ that are coprime to $n$ are witnesses. Since $n$ is square-free, the proportion of elements coprime to $n$ is bounded away from $0$, and the probability that a randomly chosen $a$ is a non-witness is at most $1/2$.
\end{proof}

\end{document}